\documentclass[10pt,a4paper,twoside,onecolumn]{book}

\usepackage{etex}
\reserveinserts{100}

  \usepackage[margin=10pt,font=small,labelfont=bf]{caption}
\usepackage{chngcntr}
\counterwithout{footnote}{chapter}

\usepackage[all]{xy}

\usepackage{makeidx}  % allows for indexgeneration
\usepackage{times}
\usepackage{graphicx}
\usepackage{latexsym,footnote}
\usepackage{amsmath}
\DeclareFontFamily{U}{MnSymbolC}{}
\DeclareSymbolFont{MnSyC}{U}{MnSymbolC}{m}{n}
\DeclareFontShape{U}{MnSymbolC}{m}{n}{
    <-6>  MnSymbolC5
   <6-7>  MnSymbolC6
   <7-8>  MnSymbolC7
   <8-9>  MnSymbolC8
   <9-10> MnSymbolC9
  <10-12> MnSymbolC10
  <12->   MnSymbolC12%
}{}
\DeclareMathSymbol{\powerset}{\mathord}{MnSyC}{180}\usepackage{amsthm}
\usepackage{algorithm}
\usepackage[noend]{algpseudocode}
\algrenewcommand\algorithmicrequire{\textbf{Input:}}
\algrenewcommand\algorithmicensure{\textbf{Output:}}
\usepackage{mathnotation}
\usepackage{array}
\usepackage{dashrule}

\usepackage[T1]{fontenc} 

\usepackage[draft]{fixme}

\usepackage{subfigure}
\usepackage{multirow}
\usepackage{tabularx}
\usepackage{cuted}

\usepackage{amssymb}

\usepackage{changepage}
\usepackage{enumerate}

\usepackage{color}															 % für Farben im allgemeinen
\usepackage[table]{xcolor}
\usepackage{booktabs}
\usepackage{arydshln}
\usepackage{cancel}
\usepackage{rotating}
\usepackage{framed}
\usepackage{float}
\usepackage{ marvosym }
\usepackage[a4paper]{geometry}
\usepackage{changepage}

\usepackage[nottoc,notlot,notlof]{tocbibind}

\usepackage{url}
\usepackage{tikz}
\newcommand*\circled[1]{\tikz[baseline=(char.base)]{
            \node[shape=circle,draw,inner sep=0.8pt] (char) {#1};}}
\usepackage{amsfonts}
\usepackage{wasysym}
\usepackage{pdfpages}
\usepackage[nottoc]{tocbibind}
\usepackage{fancyhdr}
\usepackage{emptypage}
\definecolor{light-gray1}{gray}{0.95}

\fxsetup{
nomargin,inline,index,
theme=color
}

\newcounter{examplecounter}
\newenvironment{example}{%\begin{quote}%
    \refstepcounter{examplecounter}%
  
	\vspace{7pt}
	\noindent\textbf{Example \arabic{chapter}.\arabic{examplecounter}}%
  \quad
}{

\vspace{7pt}
}
\numberwithin{examplecounter}{chapter}

\newcounter{remarkcounter}
\newenvironment{remark}{%\begin{quote}%
    \refstepcounter{remarkcounter}%
  
	\vspace{7pt}
	\noindent\textbf{Remark \arabic{chapter}.\arabic{remarkcounter}}%
  \quad
}{

\vspace{7pt}
}
\numberwithin{remarkcounter}{chapter}

\definecolor{darkgray}{rgb}{0.8,0.8,0.8}
\definecolor{lightgray}{rgb}{0.95,0.95,0.95}

\newcommand{\mC}{{\bf{C}}}

\newcommand{\Q}{{\mathit{Q}}}

\newcommand{\riotax}{\mathit{ax}}
\newcommand{\riomo}{\mathcal{O}}
\newcommand{\riomt}{\mathcal{T}}
\newcommand{\rioma}{\mathcal{A}}
\newcommand{\riomb}{\mathcal{B}}
\newcommand{\riomd}{\mathcal{D}}
\newcommand{\rioTp}{\mathit{P}}
\newcommand{\rioTn}{\mathit{N}}
\newcommand{\riotp}{\mathit{p}}
\newcommand{\riotn}{\mathit{n}}
\newcommand{\riodt}{\mathcal{D}^{*}}
\newcommand{\rioot}{\mathcal{O}^{*}}
\newcommand{\riomX}{{\bf{X}}}
\newcommand{\riomD}{{\bf{D}}}
\newcommand{\riominD}{{\bf{mD}}}
\newcommand{\riodx}[1]{{\bf D}_{#1}^+}
\newcommand{\riodnx}[1]{{\bf D}_{#1}^{-}}
\newcommand{\riodz}[1]{{\bf D}_{#1}^0}
\newcommand{\rioqc}{\mathit{c}_q}
\newcommand{\riouc}{c}
\newcommand{\rioRQ}{{\mathit{R}}}
\newcommand{\rioXsc}{\mathit{X_{sc}}}
\newcommand{\rioXalt}{\mathit{X_{alt}}}
\newcommand{\rioAl}{{\mathcal{M}}}
\newcommand{\rioSig}{{\mathbf{S}}}
\newcommand{\rioNHR}{{\mathit{NHR}}}
\newcommand{\rioN}{\phi}
\newcommand{\jwstax}{\mathit{ax}}
\newcommand{\jwsmo}{\mathcal{O}}
\newcommand{\jwsmt}{\mathcal{T}}
\newcommand{\jwsma}{\mathcal{A}}
\newcommand{\jwsmd}{\mathcal{D}}
\newcommand{\jwsmb}{\mathcal{B}}
\newcommand{\jwsTe}{P}
\newcommand{\jwsTne}{N}
\newcommand{\jwste}{p}
\newcommand{\jwstne}{n}
\newcommand{\jwsqry}{Q}
\newcommand{\jwsent}{E}
\newcommand{\jwsmD}{{\bf{D}}}
\newcommand{\jwsdx}{{\bf{D^{P}}}}
\newcommand{\jwsdnx}{{\bf{D^{N}}}}
\newcommand{\jwsdz}{{\bf{D^\emptyset}}}
\newcommand{\jwsdxi}[1]{{\bf{D^{P}_{#1}}}}
\newcommand{\jwsdnxi}[1]{{\bf{D^{N}_{#1}}}}
\newcommand{\jwsdzi}[1]{{\bf{D^\emptyset_{#1}}}}
\newcommand{\jwsmT}{\mathcal{T}}
\newcommand{\jwsmA}{\mathcal{A}}
\newcommand{\jwsmCN}{\mathcal{CN}}
\newcommand{\jwsmRN}{\mathcal{RN}}
\newcommand{\jwsmIN}{\mathcal{IN}}
\newcommand{\jwsmI}{\mathcal{I}}
\newcommand{\dxtax}{\mathit{ax}}
\newcommand{\dxmo}{\mathcal{K}}
\newcommand{\dxoo}{\mathcal{K}}
\newcommand{\dxmd}{\mathcal{D}}
\newcommand{\dxmD}{{\bf{D}}}
\newcommand{\dxmb}{\mathcal{B}}
\newcommand{\dxTe}{P}
\newcommand{\dxTne}{N}
\newcommand{\dxte}{p}
\newcommand{\dxtne}{n}
\newcommand{\dxqry}{Q}
\newcommand{\dxot}{\mathcal{K}^{*}}
\newcommand{\dxFP}{\mathcal{F}}
\newcommand{\dxqss}{s_Q}
\newcommand{\orig}{\mathsf{orig}}

\newcommand{\tax}{\mathit{ax}}

\newcommand{\mo}{\mathcal{K}}
\newcommand{\mt}{\mathcal{T}}
\newcommand{\ma}{\mathcal{A}}
\newcommand{\mi}{\mathcal{I}}
\newcommand{\mb}{\mathcal{B}}
\newcommand{\md}{\mathcal{D}}
\newcommand{\me}{\mathcal{E}}
\newcommand{\mc}{\mathcal{C}}

\newcommand{\Tp}{\mathit{P}}
\newcommand{\Tn}{\mathit{N}}
\newcommand{\tp}{\mathit{p}}
\newcommand{\tn}{\mathit{n}}
\newcommand{\dt}{\mathcal{D}_{t}}

\newcommand{\ot}{\mathcal{K}^{*}}
\newcommand{\mS}{{\bf{S}}}

\newcommand{\mD}{{\bf{D}}}
\newcommand{\allC}{{\bf{aC}}}
\newcommand{\minC}{{\bf{mC}}}
\newcommand{\minD}{{\bf{mD}}}
\newcommand{\allD}{{\bf{aD}}}
\newcommand{\dx}[1]{{\bf D}_{#1}^+}
\newcommand{\dnx}[1]{{\bf D}_{#1}^{-}}
\newcommand{\dz}[1]{{\bf D}_{#1}^0}

\newcommand{\scHS}{{\textsc{HS}}}

\newcommand{\RQ}{{\mathit{R}}}

\newcommand{\EX}[2]{{\mathbf{EX}(#1)_{#2}}}
\newcommand{\SO}{{\mathbf{Sol}}}

\newcommand{\Queue}{{\mathbf{Q}}}
\newcommand{\scQX}{{\textsc{QX}}}

\newcommand{\mQ}{{\bf{Q}}}

\newcommand{\FP}{\mathbf{FP}}
\newcommand{\QP}{\mathbf{QP}}
\newcommand{\Pt}{\mathfrak{P}}
\newcommand{\Just}{\mathsf{Just}}

\newtheorem{definition}{Definition}[chapter]{}
\newtheorem{prob_def}{Problem Definition}[chapter]{}
\newtheorem{proposition}{Proposition}[chapter]{}
\newtheorem{lemma}{Lemma}[chapter]{}
\newtheorem{corollary}{Corollary}[chapter]{}
\newtheorem{property}{Property}
\newcommand{\RM}[1]{\MakeUppercase{\romannumeral #1}}

\renewcommand{\listoffigures}{\begingroup
\tocsection
\tocfile{\listfigurename}{lof}
\endgroup}

\renewcommand{\listoftables}{\begingroup
\tocsection
\tocfile{\listtablename}{lot}
\endgroup}

\begin{document}
\includepdf{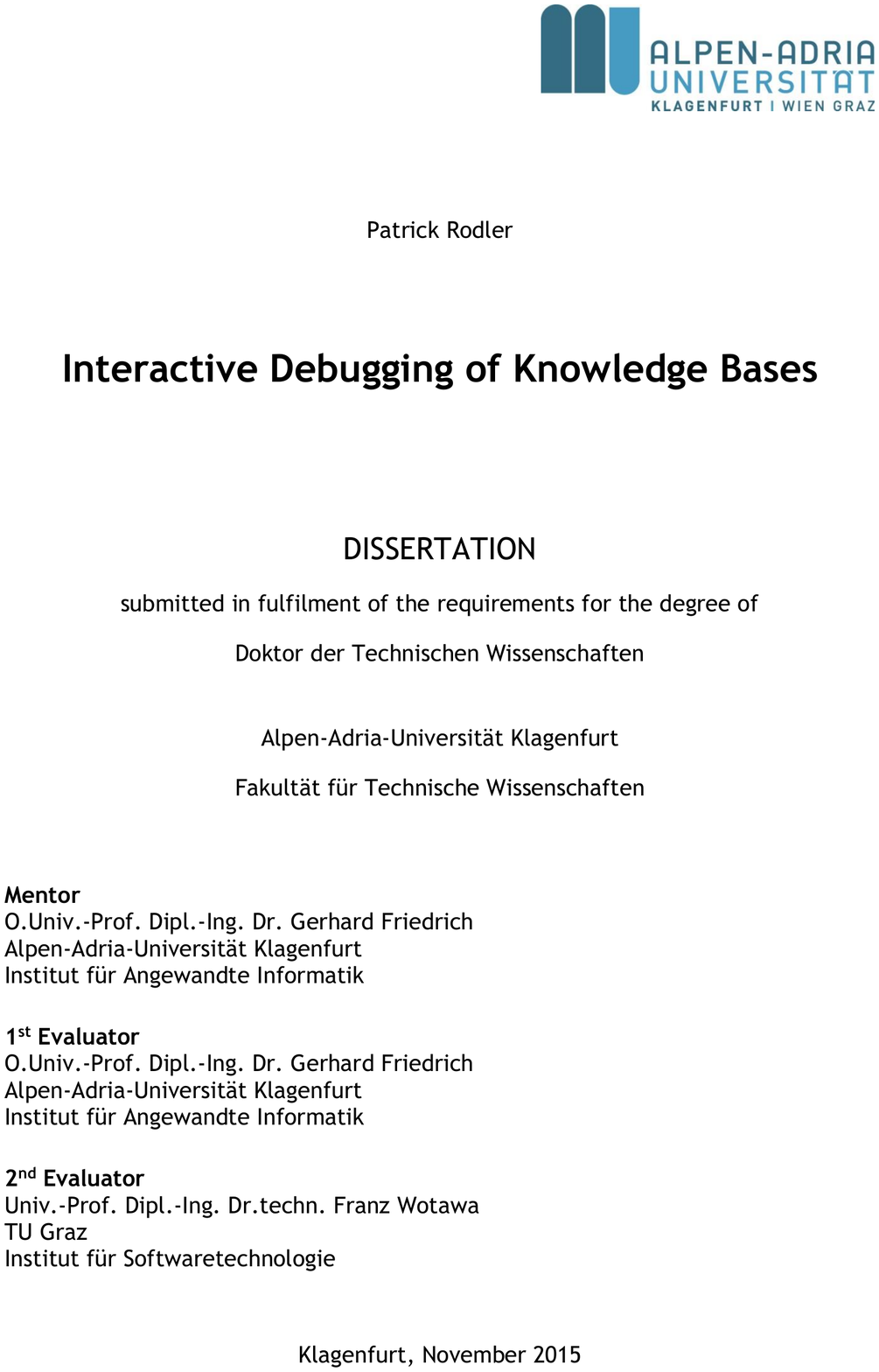}
\includepdf{empty_page.pdf}
\includepdf{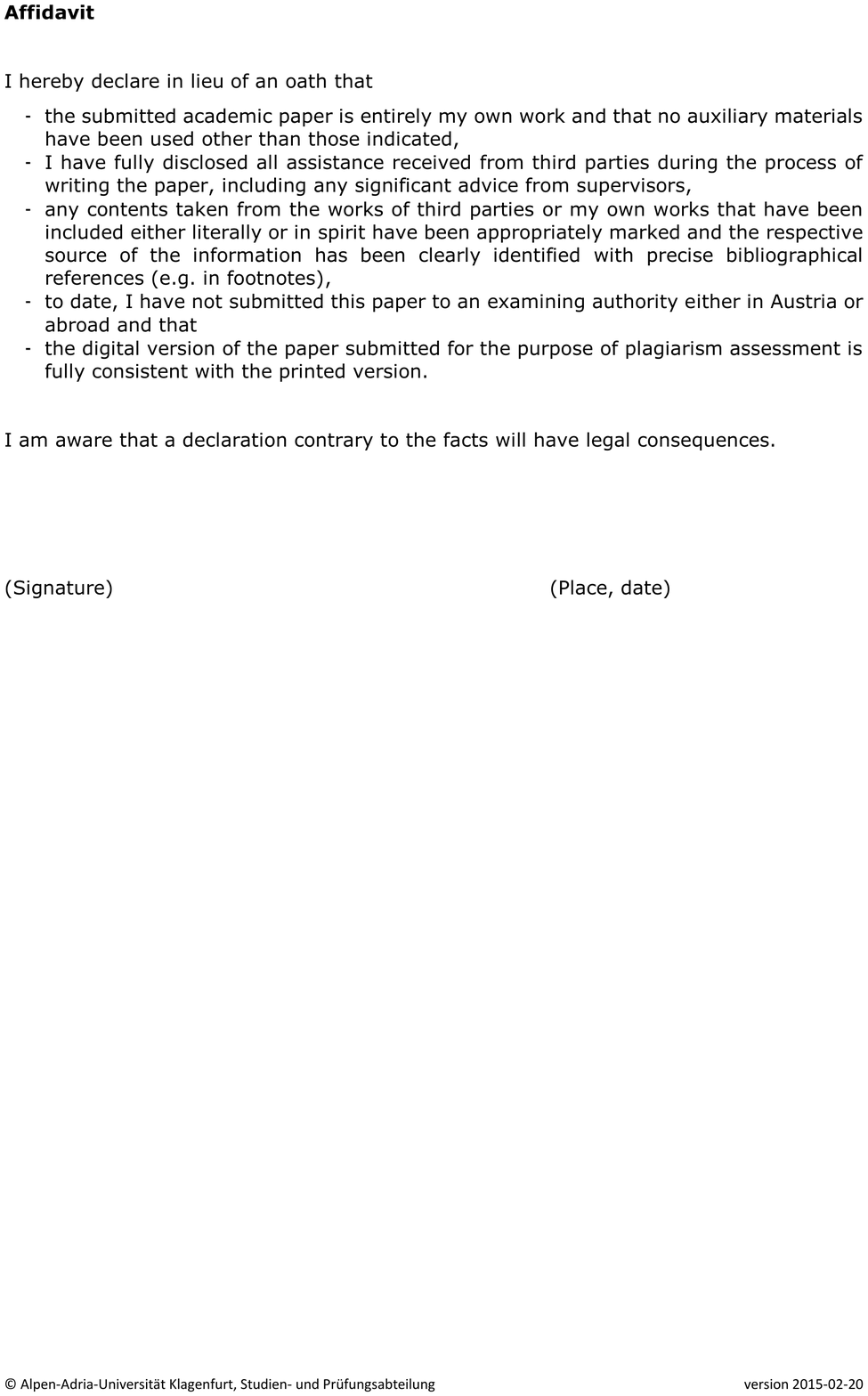}
\thispagestyle{empty}
{\pagestyle{empty}
\tableofcontents
\cleardoublepage}
\thispagestyle{empty}
%%%%%%%%%%%%%%%%%%%%%%%%%% new end%%%%%%%%%%

\frontmatter
%%%%%%%%%%%%%%%%%%%%%%%%%% new start %%%%%%%%%%
\listoffigures
\clearpage
\thispagestyle{empty}

\listoftables
\clearpage
\thispagestyle{empty}

\addtocontents{toc}{\protect\thispagestyle{empty}}
\thispagestyle{empty}
%%%%%%%%%%%%%%%%%%%%%%%%%% new end %%%%%%%%%%
\chapter{Abstract} 
%\addcontentsline{toc}{chapter}{Abstract}
Most artificial intelligence applications rely on knowledge about a relevant real-world domain that is encoded in a knowledge base (KB) by means of some logical knowledge representation language. The most essential benefit of such logical KBs is the opportunity to perform automatic reasoning to derive implicit knowledge or to answer complex queries about the modeled domain. The feasibility of meaningful reasoning requires a KB to meet some minimal quality criteria such as consistency; that is, there must not be any contradictions in the KB. Without adequate tool assistance, the task of resolving such violated quality criteria in a KB can be extremely hard even for domain experts, especially when the problematic KB includes a large number of logical formulas, comprises complicated formalisms, was developed by multiple people or in a distributed fashion or was (partially) generated by means of some automatic systems. 

Non-interactive debugging systems published in research literature often cannot localize all possible faults (\emph{incompleteness}), suggest the deletion or modification of unnecessarily large parts of the KB (\emph{non-minimality}), return incorrect solutions which lead to a repaired KB not satisfying the imposed quality requirements (\emph{unsoundness}) or suffer from \emph{poor scalability} due to the inherent complexity of the KB debugging problem. Even if a system is complete and sound and considers only minimal solutions, there are generally exponentially many solution candidates to select one from. However, any two repaired KBs obtained from these candidates differ in their semantics in terms of entailments and non-entailments. Selection of just any of these repaired KBs might result in unexpected entailments, the loss of desired entailments or unwanted changes to the KB which in turn might cause unexpected new faults during the further development or application of the repaired KB. Also, manual inspection of a large set of solution candidates can be time-consuming (if not practically infeasible), tedious and error-prone since human beings are normally not capable of fully realizing the semantic consequences of deleting a set of formulas from a KB. Hence there is a need for adequate tools that support a user when facing a faulty KB.

In this work, we account for these issues and propose methods for the interactive debugging of KBs which are complete and sound and compute only minimally invasive solutions, i.e.\ suggest the deletion or modification of just a set-minimal subset of the formulas in the problematic KB. User interaction takes place in the form of queries asked to a person, e.g.\ a domain expert, about intended and non-intended entailments of the correct KB. To construct a query, only a minimal set of two solution candidates must be available. After the answer to a query is known, the search space for solutions is pruned. Iteration of this process until 
%a solution candidate has overwhelming probability or until 
there is only a single solution candidate left yields a repaired KB which features exactly the semantics desired and expected by the user. 

The novel contributions of this work are:
\begin{itemize}
	\item \emph{Thorough Theoretical Workup of the Topic of Interactive Debugging of Monotonic KBs:} We evolve the theory of the topic by first elaborating on the theory of non-interactive KB debugging, revealing crucial shortcomings in the application of non-interactive methods and thereby motivating the development and deployment of interactive approaches in KB debugging. Then, we give some important results that guarantee the feasibility of interactive KB debugging, give some precise definitions of the problems interactive KB debugging aims to solve and present algorithms that provably solve these problems.
	\item \emph{A Complete Picture of an Interactive Debugging System is Drawn:} This is the first work that deals with an entire system of algorithms that are required for the interactive debugging of monotonic KBs, considers and details all algorithms separately, proves their correctness and demonstrates how all these algorithms are orchestrated to make up a full-fledged and provably correct interactive KB debugging system. 
	\item \emph{Two New Algorithms for the Iterative Computation of Candidate Solutions} in the scope of interactive KB debugging are proposed. The first one guarantees constant convergence towards the exact solution of the interactive KB problem by the ascertained reduction of the number of remaining solutions after any query is answered. The second one features powerful search tree pruning techniques and might thus be expected to exhibit a more time- and space-saving behavior than existing algorithms, in particular for growing problem instances.
%	(an evaluation that compares the overall efficiency of these new algorithms with the ones proposed in literature must still be conducted and is part of our future research). 
	\item \emph{Suggestion and Extensive Analysis of Different Methods for Selection of the ``Best'' Query} to ask the user next. We compare a greedy ``split-in-half'' strategy that proposes queries which eliminate half of the known candidate solutions with a strategy relying on information entropy that chooses the query with highest information gain based on (a user's) beliefs about faults in the KB. Comprehensive experiments manifest that an average guess of the fault information suffices to reduce the query answering effort for the interacting user, often to a significant extent, by means of the latter strategy compared to the former. Moreover, we demonstrate that both methods clearly outperform a random way of selecting queries.
	\item \emph{Presentation of a Reinforcement Learning Query Selection Strategy.} Minimal effort for the interacting user can be achieved if both the query selection method is chosen carefully and the provided fault information satisfies some minimum quality requirements. In particular, for deficient fault information and unfavorable strategy for query selection, we observe cases where the overhead in terms of user effort exceeds 2000\% (!) in comparison to employing a more favorable query selection strategy. Since, unfortunately, assessment of the fault information is only possible a-posteriori (after the debugging session is finished and the correct solution is known), we devise a learning strategy (RIO) that continuously adapts its behavior depending on the performance achieved and in this vein minimizes the risk of using low-quality fault information. This approach makes interactive debugging practical even in scenarios where reliable fault estimates are difficult to obtain. Evaluations provide evidence that for 100\% of the cases in the hardest (from the debugging point of view) class of faulty test KBs, RIO performed at least as good as the \emph{best} other strategy and in more than 70\% of these cases it even manifested superior behavior to the \emph{best} other strategy. Choosing RIO over other approaches can involve an improvement by the factor of up to 23, meaning that more than 95\% of user time and effort might be saved per debugging session.
	\item \emph{Provisioning of Mechanisms for Efficiently Dealing with KB Debugging Problems Involving High Cardinality Faults.} In the standard interactive debugging approach described in this work, the computation of queries is based on the generation of the set of \emph{most probable} solution candidates. By this postulation, certain quality guarantees about the output solution can be given. However, we learn that dropping this requirement can bring about substantial savings in terms of time and especially space complexity of interactive debugging, in particular in debugging scenarios where faulty KBs are (partly) generated as a result of the application of automatic systems. 
In such situations, we propose to base query computation on \emph{any} set of solution candidates using a ``direct'' method for candidate generation.
We study the application of this direct method to high cardinality faults in KBs and find out that the number of required queries per debugging session is scarcely affected for cases when the standard approach is also applicable. However, the direct method proves applicable in situations when the standard approach is not (due to time or memory issues) and is still able to locate the correct solution.
\end{itemize}

\mainmatter
%%%%%%%%%%%%%%%
\thispagestyle{empty}
\part{Prolog}
\label{part:Prolog}
%This part is organized as follows:\\

In this part, we first give an introduction in Chapter~\ref{chap:intro}. This includes a motivation why knowledge base debugging is a ``hot topic'' (and even getting hotter as intelligent applications and devices become more and more ubiquitous), an introduction to the non-interactive debugging of knowledge bases and the revealment of decisive shortcomings of this paradigm, e.g.\ poor scalability and the risk of obtaining solutions of inferior quality. As a solution to the identified issues we then explain how a (group of) user(s) might collaborate with an interactive debugging system to determine high-quality solutions even in scenarios where non-interactive systems fail. Further, we discuss the design and the components of a generic interactive debugger, provide an illustrating example and outline the powerful feature of our system to be able to incorporate background knowledge into the debugging process which can drastically reduce the search space for solutions and disclose faults in the knowledge base that could be missed otherwise. Finally, we provide an enumeration of the contributions of this work and discuss the further organization of this part and of the rest of this work.\footnote{Parts of Part~\ref{part:Prolog} already appeared in \cite{Rodler2015}.}

\chapter{Introduction} 
\label{chap:intro}

\paragraph{Motivation.}
Most artificial intelligence applications rely on knowledge that is encoded in a knowledge base (KB) by means of some logical knowledge representation language such as propositional logic (PL)~\cite{chang1973}, Datalog~\cite{Ceri1989a}, first-order logic (FOL)~\cite{chang1973}, The Web Ontology Language (OWL~\cite{patel2004owl}, OWL~2~\cite{Grau2008a,Motik2009a}) or Description Logic (DL)~\cite{Baader2007}.
% bechhoferOWLref2004
Experts in a variety of application domains keep developing KBs of constantly growing size. A concrete example of a repository containing biomedical KBs is the Bioportal\footnote{\url{http://bioportal.bioontology.org}}, which comprises vast ontologies with tens or even hundreds of thousands of terms each (e.g.~the SNOMED-CT ontology with currently over 395.000 terms). Such KBs however pose a significant challenge for people as well as tools involved in their evolution, maintenance and application. 

All these activities are based on the most essential benefit of logical KBs, namely the opportunity to perform automatic reasoning to derive implicit knowledge or to answer complex queries about the modeled domain. 
%In case this is not demanded, one could simply formulate the knowledge in some non-logical language, for example natural language, and store it in some kind of database which, equipped with an adequate search feature, could serve the user to retrieve the specified knowledge at some later point in time.
The feasibility of meaningful reasoning requires a KB to meet the minimum quality criterion \emph{consistency}, i.e.\ there must not be any contradictions in the KB. Because \emph{any} logical formula can be derived from an inconsistent KB. Further on, one might postulate further requirements to be met by a KB. For instance, one might consider faulty a FOL KB entailing $\forall X\, \lnot p(X)$ for some predicate symbol $p$ occurring in the KB. Such a KB would be incoherent, i.e.\ it would violate the requirement \emph{coherency} (which was originally defined for DL KBs~\cite{Schlobach2007,Parsia2005}). Additionally, test cases can be specified giving information about desired (\emph{positive test cases}) and non-desired (\emph{negative test cases}) entailments a correct KB should feature. This characterization of a KB's intended semantics is a direct analogon to the field of software debugging, where test cases are exploited as a means to verify the correct semantics of the program code.

As KBs are growing in size and complexity, their likeliness of violating one of these criteria increases. Faults in KBs may, for instance, arise because human reasoning is simply overstrained~\cite{Horridge2011b, Horridge2009}. That is, generally a person will not be capable of completely grasping or mentally processing the entire knowledge contained in a (large or complex) KB at once. In fact, a person might fully comprehend some isolated part of a the KB, but might
not be able to determine or understand all implications or non-implications of 
% be stretched to their limits 
this isolated part combined with other parts of a KB, i.e.\ when new logical formulas are added.

Another reason for the non-compliance with the mentioned quality criteria imposed on KBs might be that multiple (independently working) editors contribute to the development of the KB~\cite{Noy2006a} which may lead to contradictory formulas. The OBO Project\footnote{\url{http://obo.sourceforge.net}} and the NCI Thesaurus\footnote{\url{http://nciterms.nci.nih.gov/ncitbrowser}} are examples of collaborative KB development projects. Employing automatic tools, e.g.\ \cite{Jimenez-Ruiz2011,Ngo2012,Jean-Mary2009}, to generate (parts of) KBs can further exacerbate the task of KB quality assurance~\cite{meilicke2011,Ferrara2011}. 

Moreover, as studies in cognitive psychology~\cite{Ceraso71,Johnson1999} attest, humans make systematic errors while formulating or interpreting logical formulas. These observations are confirmed by \cite{Rector2004,Roussey2009} which present common faults people make when developing a KB (ontology). Hence, it is essential to devise methods that can efficiently identify and correct faults in a KB. 

\begin{figure*}[htbp]
	\centering
		\includegraphics[width=0.6\textwidth]{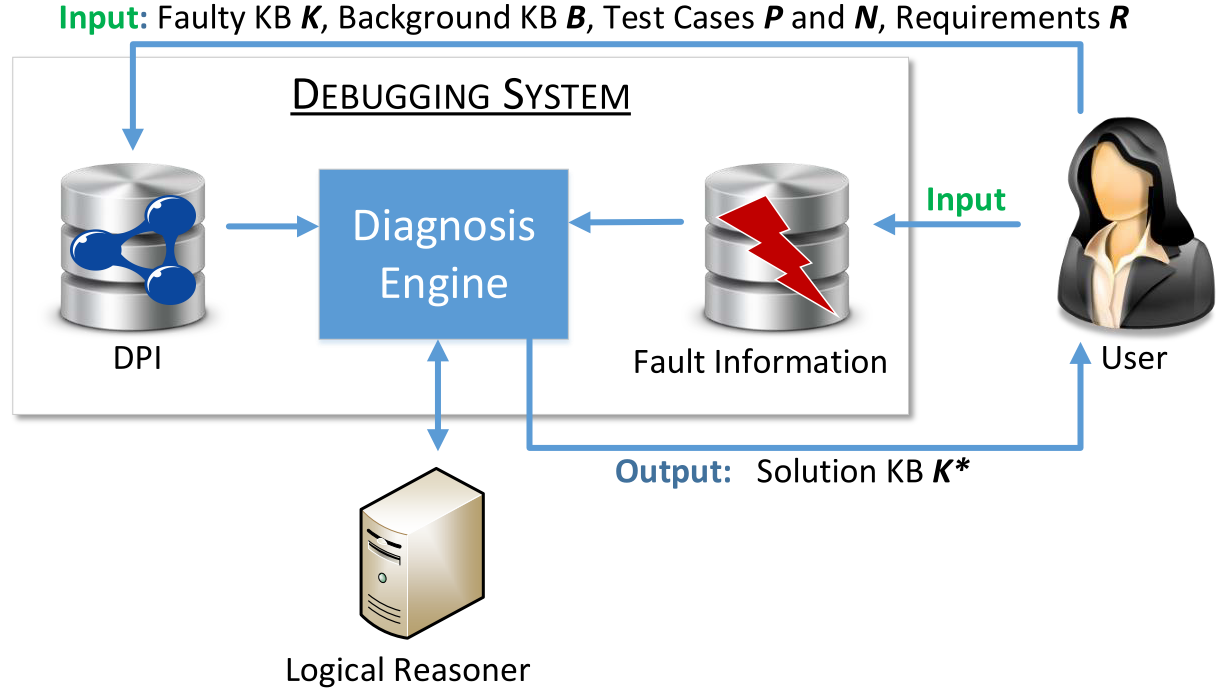}
	\caption[The Principle of Non-Interactive KB Debugging]{The principle of non-interactive KB debugging.}
	\label{fig:non-interactive_debugging_workflow}
\end{figure*}

\paragraph{Non-Interactive KB Debugging.}
Given a set of requirements to the KB and sets of test cases, KB debugging methods~\cite{Schlobach2007,Kalyanpur.Just.ISWC07,friedrich2005gdm,Horridge2008} can localize a (potential) fault by computing a subset $\md$ of the formulas in the KB $\mo$ called a \emph{diagnosis}. At least all formulas in a diagnosis must be (adequately) modified or deleted in order to obtain a KB $\ot$ that satisfies all postulated requirements and test cases. Such a KB $\ot$ constitutes the solution to the \emph{KB debugging problem}. Figure~\ref{fig:non-interactive_debugging_workflow}\footnote{Thanks to Kostyantyn Shchekotykhin for making available to me parts of this diagram.} outlines such a KB debugging system. The input to the system is a \emph{diagnosis problem instance (DPI)} defined by 
\begin{itemize}
	\item some KB $\mo$ formulated using some (monotonic) logical language $\mathcal{L}$ (every formula in $\mo$ might be correct or faulty),
	\item (optionally) some KB $\mb$ (over $\mathcal{L}$) formalizing some background knowledge relevant for the domain modeled by $\mo$ (such that $\mb$ and $\mo$ do not share any formulas; all formulas in $\mb$ are considered correct)
	\item a set of requirements $\RQ$ to the correct KB,
	\item sets of positive ($\Tp$) and negative ($\Tn$) test cases (over $\mathcal{L}$) asserting desired semantic properties of the correct KB and
	\item (optionally) some fault information $\FP$, e.g.\ in terms of fault probabilities of logical formulas in $\mo$.
\end{itemize}
Moreover, the system requires a sound and complete logical reasoner for deciding consistency (coherency) and calculating logical entailments of a KB formulated over the language $\mathcal{L}$. Some approaches (including the ones presented in this work) use the reasoner as a black-box (e.g.\ \cite{Shchekotykhin2012,Horridge2011a}) within the debugging system. That is, the reasoner is called as is and serves as an oracle independent from other computations during the debugging process; that is, the internals of the reasoner are irrelevant for the debugging task. On the other hand, glass-box approaches (e.g.\ \cite{Schlobach2007,Horridge2011a,kalyanpur2005}) attempt to exploit internal modifications of the reasoner for debugging purposes; in other words, the sources of problems (e.g.\ contradictory formulas) in the KB are computed as a direct consequence of reasoning~\cite{Horridge2011a}. The advantages of a black-box approach over a glass-box approach are the lower memory consumption and better performance~\cite{kalyanpur2005} of the reasoner and the reasoner independence of the debugging method. The latter benefit is essential for the generality of our approaches and their applicability to various knowledge representation formalisms. 

Given these inputs, the debugging system focuses on (a subset of) all possible fault candidates (usually the set of minimal, i.e.\ irreducible, diagnoses) and usually outputs the most probable one amongst these if some fault information is provided or the minimum cardinality one, otherwise. Alternatively, a debugging system might also be employed to calculate a predefined number of (most probable or minimum cardinality) minimal diagnoses or to determine all minimal diagnoses computable within a predefined time limit. 
%The output set of diagnoses might then be manually inspected by the user.

%\begin{figure}[htbp]
	%\centering
		%\includegraphics[width=0.7\columnwidth]{debugging_workflow.pdf}
	%\caption{The principle of non-interactive KB debugging.}
	%\label{fig:non-interactive_debugging_workflow}
%\end{figure}  

\paragraph{Issues with Non-Interactive KB Debugging Systems.}
In real-world scenarios, debugging tools often have to cope with large numbers of minimal diagnoses where the trivial application, i.e.\ deletion, of any minimal diagnosis leads to a (repaired) KB with different semantics in terms of entailed and non-entailed formulas. For example, in~\cite{ksgf2010} a sample study of real-world KBs revealed that the number of different minimal diagnoses might exceed thousand by far (1782 minimal diagnoses for a KB with only 1300 formulas). In such situations simple visualization of all these alternative modifications of the ontology is clearly ineffective. Selecting a wrong diagnosis (in terms of its semantics, \emph{not} in terms of fulfillment of test cases and requirements) can lead to unexpected entailments or non-entailments, lost desired entailments and surprising future faults when the KB is further developed. Manual inspection of a large set of (minimal) diagnoses is time-consuming (if not practically infeasible), error-prone and often computationally infeasible due to the complexity of diagnosis computation. 

Moreover, \cite{Stuckenschmidt2008} has put several (non-interactive) debugging systems to the test using a test set of faulty (incoherent OWL) real-world KBs which were partly designed by humans and partly by the application of automatic systems. The result was that most of the investigated systems had serious performance problems, ran out of memory, were not able to locate all the existing faults in the KB (incompleteness), reported parts of a KB as faulty which actually were not faulty (unsoundness), produced only trivial solutions or suggested non-minimal faults (non-minimality). Often, performance problems and incompleteness of non-interactive debugging methods can be traced back to an explosion of the search tree for minimal diagnoses.

\paragraph{The Solution: Interactive KB Debugging.}
In this work we present algorithms for interactive KB debugging. These aim at the gradual reduction of compliant minimal diagnoses by means of user interaction, thereby seeking to prevent the search tree for minimal diagnoses from exploding in size by performing regular pruning operations. ``User'' in this case might refer to a single person or multiple persons, usually experts of the particular domain the faulty KB is dealing with such as biology, medicine or chemistry. 
%An interactive KB debugging problem is defined
Throughout an interactive debugging session, the user is asked a set of automatically chosen queries about the domain that should be modeled by a given faulty KB. A query can be created by the system after a set $\mD$ of a minimum of two minimal diagnoses has been precomputed (we call $\mD$ the \emph{leading diagnoses}). Each query is a conjunction (i.e.\ a set) of logical formulas that are entailed by some correct subset of the formulas in the KB. With regard to one particular query $Q$, any set of minimal diagnoses for the KB, in particular the set $\mD$ which has been utilized to generate $Q$, can be partitioned into three sets, the first one ($\dx{}$) including all diagnoses in $\mD$ compliant only with a positive answer to $Q$, the second ($\dnx{}$) including all diagnoses in $\mD$ compliant only with a negative answer to $Q$, and the third ($\dz{}$) including all diagnoses in $\mD$ compliant with both answers. A positive answer to $Q$ signalizes that the conjunction of formulas in $Q$ must be entailed by the correct KB wherefore $Q$ is added to the set of positive test cases. Likewise, if the user negates $Q$, this is an indication that at least one formula in $Q$ must not be entailed by the correct KB. As a consequence, $Q$ is added to the set of negative test cases.

Assignment of a query $Q$ to either set of test cases results in a new debugging scenario. In this new scenario, all elements of $\dnx{}$ are no longer minimal diagnoses given that $Q$ has been classified as a positive test case. Otherwise, all diagnoses in $\dx{}$ are invalidated. In this vein, the successive reply to queries generated by the system will lead the user to the single minimal solution diagnosis that perfectly reflects their intended semantics. In other words, after deletion of all formulas in the solution diagnosis from the KB and the addition of the conjunction of all formulas in the specified positive test cases to the KB, the resulting KB meets all requirements and positive as well as negative test cases. In that, the added formulas contained in the positive test cases serve to replace the desired entailments that are broken due to the deletion of the solution diagnosis from the KB. 

Thence, in the interactive KB debugging scenario the user is not required to cope with the understanding of which faults (e.g.\ sources of inconsistency or implications of negative test cases) occur in the faulty initial KB, why they are faults (i.e.\ why particular entailments are given and others not) and how to repair them. All these tasks are undertaken by the interactive debugging system. 

The proposed approaches to interactive KB debugging in this work follow the standard \emph{model-based diagnosis (MBD)} technique~\cite{Reiter87,dekleer1987}. MBD has been successfully applied to a great variety of problems in various fields such as robotics~\cite{Steinbauer2005}, planning~\cite{Steinbauer2009}, debugging of software programs~\cite{wotawa2002}, configuration problems~\cite{Felfernig2004213}, hardware designs~\cite{Friedrich1999}, constraint satisfaction problems and spreadsheets~\cite{Abreu2012}. Given a description (model) of a system, together with an observation of the system's behavior which conflicts with the intended behavior of the system, the task of MBD is to find those components of the system (a diagnosis) which, when assumed to be functioning abnormally, provide an explanation of the discrepancy between the intended and the observed system behavior. 
%Translated to the setting of KB debugging, the ``model of a system'' is the description of a (real world) domain in terms of logical formulas, i.e.\ the faulty KB along with the background KB and the positive test cases, formally $\mo \cup \mb \cup \bigcup_{\tp \in \Tp} \tp$. The ``observation of the system's behavior which conflicts with the intended behavior of the system'' corresponds to the recognition that $\mo \cup \mb \cup \bigcup_{\tp \in \Tp} \tp$ violates some predefined requirements in $\RQ$ (e.g.\ consistency, coherency) or logically entails some of the negative test cases $\tn \in \Tn$.
Translated to the setting of KB debugging, the set of ``system components'' comprises the formulas $\tax_i$ in the given faulty KB $\mo$. The ``system description'' refers to the statement that the KB $\mo$ along with the background KB $\mb$ and the positive test cases $\tp\in\Tp$
%, formally $\mo \cup \mb \cup \bigcup_{\tp \in \Tp} \tp$
%the union of the KB $\mo$ with the background KB $\mb$ and with the union of the positive test cases $\tp_j \in \Tp$ 
must meet all predefined requirements (e.g.\ consistency, coherency) and must not logically entail any of the negative test cases $\tn \in \Tn$, i.e.\ 
\begin{enumerate}[(i)]
	\item $\mo \cup \mb \cup \bigcup_{\tp\in\Tp} \tp$ satisfies requirement $r$ for all $r\in\RQ$ and 
	\item $\mo \cup \mb \cup \bigcup_{\tp\in\Tp} \tp \not\models \tn$ for all $\tn \in \Tn$.
\end{enumerate}
%(i)~$\mo \cup \mb \cup \bigcup_{\tp\in\Tp} \tp$ satisfies requirement $r$ for all $r\in\RQ$ and (ii)~$\mo \cup \mb \cup \bigcup_{\tp\in\Tp} \tp \not\models \tn$ for all $\tn \in \Tn$.
The ``observation which conflicts with the intended behavior of the system'' corresponds to the finding that (i) or (ii) or both are violated. That is, the ``system description'' along with the ``observation'' and the assumption that all components are sound yields an inconsistency. An ``explanation for the discrepancy between observed and intended system behavior'' (i.e.\ a diagnosis) is the assumption $\md$ that all formulas in a subset $\md$ of $\mo$ are faulty (``behave abnormally'') and all formulas in $\mo \setminus \md$ are correct (``do not behave abnormally'') such that the ``system description'' along with the ``observation'' and the assumption $\md$ is consistent.
Computation of (minimal) diagnoses is accomplished 
% That means minimal diagnoses are calculated 
with the aid of \emph{minimal conflict sets}, i.e.\ irreducible sets of formulas in the KB $\mo$ that preserve the violation of (i) or (ii) or both.
%at least one requirement or test case. 

\label{etc:MBD_problem_is_abduction_problem} An MBD problem can be modeled as an abduction problem~\cite{Bylander1991}, i.e.\ finding an explanation for a set of data. It was proven in~\cite{Bylander1991} that the computation of the first explanation (minimal diagnosis) is in $\textsc{P}$. However, given a set of explanations (minimal diagnoses) it is $\textsc{NP}$-complete to decide whether there is an additional explanation (minimal diagnosis). Stated differently, the detection of the first explanation can be efficiently accomplished whereas the finding of any further one is intractable (unless $\textsc{P} = \textsc{NP}$). When seeing the (interactive) KB debugging problem as an abduction problem, one must additionally take into account the costs for reasoning. Because, a call to a logical reasoner is required in order to decide whether or not a set of hypotheses (a subset of the KB) is an explanation (minimal diagnosis). Incorporating the necessary reasoning costs and assuming consistency a minimal requirement to the correct KB, the finding of the first explanation (minimal diagnosis) is already $\textsc{NP}$-hard even for propositional KBs~\cite{Selman1989} (since propositional satisfiability checking is $\textsc{NP}$-complete). The worst case complexity for the debugging of KBs formulated over more expressive logics such as OWL 2 (reasoning is 2-$\textsc{NExpTime}$-complete~\cite{Grau2008a,Kazakov2008}) will be of course even worse. This seems quite discouraging. However, we have shown in our previous works~\cite{Rodler2013, Shchekotykhin2012, Shchekotykhin2014} that for many real-world KBs interactive KB debugging is feasible in reasonable time, despite high (or intractable) worst case reasoning costs and the intractable complexity of the abduction (i.e.\ minimal diagnosis finding) problem as such. Hence, the goal of this work is amongst others to present algorithms that work well in many \emph{practical} scenarios.

\paragraph{Assumptions about the Interacting User.}
About a user $u$ consulting an (interactive) debugging system, we 
%Let us suppose a user specifying a KB arriving at a point where the KB is faulty, i.e.\
%\begin{itemize}
	%\item the KB is inconsistent, i.e.\ includes at least one contradiction, or violates certain other desired requirements, e.g.\ coherence is violated in case an $1$-place predicate symbol must be false in all interpretations of the KB, or features undesired entailments or 
	%\item the addition of some desired entailments to the KB yields a new KB that features undesired entailments or violates consistency or conherency (where the latter two are called requirements for short).
%\end{itemize}
%About such a user $u$, we 
make the following plausible assumptions:
\begin{enumerate}[U1]
	\item $u$ is not 
	%(efficiently) 
	able to explicitly enumerate a set of logical formulas that express the intended domain that should be modeled in a satisfactory way, i.e.\ without unwanted entailments or non-fulfilled requirements, %inconsistencies or incoherencies (provided that coherence is defined for the logical laneguage used in the KB).  
	\item $u$ is able to answer concrete queries about the intended domain that should be modeled, i.e.\ $u$ can classify a given logical formula (or a conjunction of logical formulas) as a wanted or unwanted proposition in the intended domain (i.e.\ an entailment or non-entailment of the correct domain model). 
\end{enumerate}
The first assumption is obviously justified since otherwise $u$ could have never obtained a faulty KB, i.e.\ a KB that violates at least one requirement or test case, and there would be no need for $u$ to employ a debugging system. 
%Also if $u$ is not the same (group of) person(s) who developed the KB (which must not necessarily be the case, e.g.\ a knowledge engineer could be the author of a faulty KB and $u$ a domain expert), this assumption is still reasonable. For, otherwise there would be no need for $u$ to consult  
%or a KB which violates at least one requirement or test case

Regarding the second assumption, the first thing to be noted is that any KB (i.e.\ any model of the intended domain) either does entail a certain logical formula $\tax$ or it does not entail $\tax$. Second, if $u$ is assumed to bring along enough expertise in that domain, $u$ should be able to gauge the truth of (at least) some formulas about that domain, especially if these formulas constitute logical entailments of parts of the specified knowledge in KB so far. We want to emphasize that $u$ is not required to be capable of answering all possible queries (or formulas) about the respective domain since $u$ might always skip a particular query in our system without any noticeable disadvantages. In such a case, the system keeps generating further queries, one at a time (usually the next-best one according to some quality measure for queries), until $u$ is ready to answer it. As the number of possible queries is usually exponential in the number of minimal diagnoses exploited to compute it, there will be plenty of different ``surrogate queries'' in most scenarios.

\paragraph{A Motivating Example.} 
To get a more concrete idea of these assumptions, the reader is invited to think about whether the following first-order KB $\mo$
%$\mo := \setof{\forall X (researcher(X) \leftrightarrow (\forall Y (writes(X,Y) \rightarrow paper(Y)))), \forall X ((\exists Y writes(X,Y)) \rightarrow researcher(X)), \forall X (secretary(X) \rightarrow \lnot researcher(X)), secretary(pam)}$ 
is consistent (a similar example is discussed in \cite{Horridge2009}):
\begin{align}
\label{ex0:s1}&\forall X (res(X) \leftrightarrow \forall Y (writes(X,Y) \rightarrow paper(Y))) \\
\label{ex0:s2}&\forall X ((\exists Y writes(X,Y)) \rightarrow res(X)) \\
\label{ex0:s3}&\forall X (secr(X) \rightarrow gen(X)) \\
\label{ex0:s4}&\forall X (gen(X) \rightarrow \lnot res(X)) \\
\label{ex0:s5}&secr(pam)
\end{align}
If we assume that the predicate symbols $res$, $secr$ and $gen$ stand for 'researcher', 'secretary' and 'general employee', respectively, and the constant $pam$ stands for the person Pam, the KB says the following:
\begin{itemize}
	\item Formula~\ref{ex0:s1}: ``Somebody is a researcher if and only if everything they write is a paper.''
	\item Formula~\ref{ex0:s2}: ``Everybody who writes something is a researcher.''
	\item Formula~\ref{ex0:s3}: ``Each secretary is a general employee.''
	\item Formula~\ref{ex0:s4}: ``No general employee is a researcher.''
	\item Formula~\ref{ex0:s5}: ``Pam is a secretary.''
\end{itemize}
This KB is indeed inconsistent. The reader might agree that it is not very easy to understand why this is the case. The observations made in \cite{Horridge2009} concerning a slight modification $\mo'$ of the KB $\mo$ extracted from a real-world KB confirm this assumption. Compared to $\mo$, the KB $\mo'$ included only Formulas~\ref{ex0:s1}-\ref{ex0:s3} of $\mo$, was formulated in DL (cf.\ Section~\ref{sec:DL}), and used the terms $A,C,\dots$ instead of $res, paper, \dots$. 
Amongst others, this KB $\mo'$ was used as a sample KB in a study where participants had to find out whether a concrete given formula is or is not entailed by a concrete given KB. In the case of  the KB $\mo'$, the assignment (translated to the terminology in our KB $\mo$) was to find out whether $\forall X (secr(X) \rightarrow res(X))$ is an entailment of formulas~\ref{ex0:s1}-\ref{ex0:s3}. Although $\mo'$ contains only three formulas, the result was that even participants with many years of experience in DL, among them also DL reasoner developers, did not realize that this is in fact the case (the reason for this entailment to hold is that formulas~\ref{ex0:s1}-\ref{ex0:s3} imply that $\forall X\, res(X)$ holds).      

Since $\forall X\, res(X)$ is also necessary for the inconsistency of $\mo$, this suggests that people might also have severe difficulties in comprehending why $\mo$ is inconsistent. Once the validity of this entailment is clear, it is relatively straightforward to see that $\mo$ cannot have any models. For, $res(pam)$ (due to $\forall X\, res(X)$) and $\lnot res(pam)$ (due to formulas~\ref{ex0:s3}-\ref{ex0:s5}) are implications of $\mo$.

Consequently, we might also assume that even experienced knowledge engineers (not to mention pure domain experts) could end up with a contradictory KB like $\mo$, which substantiates our first assumption (U1) about $u$. Probably, the intention of those people who specified formulas~\ref{ex0:s1}-\ref{ex0:s3} was not that $\forall X\, res(X)$ should be entailed. That is, it might be already a too complex task for many people to (mentally) reason even with such a small KB like this and manually derive implicit knowledge from it. 

However, on the other hand, we might well assume $u$ to be able to answer a concrete query about the intended domain they tried to model by $\mo$. For instance, one such query could be whether $Q_1 := \setof{\forall X\, res(X)}$ is a desired entailment of their model (i.e.\ ``should everybody be a researcher in your intended model of the domain?''). If we assume the (seemingly obvious) case that $u$ negates this query, i.e.\ asserts that this is an unwanted entailment, then an interactive debugging system (employing a logical reasoner) can derive that at least one of the formulas~\ref{ex0:s1} and \ref{ex0:s2} must be faulty. This holds because the only set-minimal explanation in terms of formulas in $\mo$ for the entailment $\forall X\, res(X)$ is given by these two formulas. In other words, the set of formulas $\setof{\ref{ex0:s1},\ref{ex0:s2}}$ is the only minimal conflict set in $\mo$ given that $Q_1$ is a negative test case.
Hence, the deletion (or suitable modification) of any of these formulas will break this unwanted entailment.
%We would call the negatively answered query $Q_1 := \setof{\forall X\, res(X)}$ a negative test case and the set of formulas~\ref{ex0:s1} and \ref{ex0:s2} a minimal conflict set (since this set entails a negative test case, namely $Q_1$). 

Before it is known that $Q_1$ must not be entailed by the correct KB, given consistency is the only requirement to the KB postulated by $u$, the complete KB $\mo$ is a minimal conflict set. That is, after the assignment of a (strategically well-chosen) query to the set of positive or, in this case, negative test cases can already shift the focus of potential modifications or deletions to a subset of only two candidate formulas. We would call these two formulas the remaining minimal diagnoses after an answer to the query $Q_1$ has been submitted. 

Initially, there are five minimal diagnoses, each formula in $\mo$ is one. The meaning of a diagnosis is that its deletion from $\mo$ leads to the fulfillment of all requirements and (so-far-)specified positive and negative test cases. As the reader should be easily able to see, the deletion of any formula from $\mo$ yields a consistent KB; e.g.\ removing formula~\ref{ex0:s5} prohibits the entailment $\lnot res(pam)$ whereas discarding formula~\ref{ex0:s2} prohibits the entailment $res(pam)$. The reader should notice that, as soon as the negative test case $Q_1$ is known, removing (only) formula~\ref{ex0:s5} does not yield a correct KB since $\setof{\ref{ex0:s1},\ref{ex0:s2},\ref{ex0:s3},\ref{ex0:s4}}$ still entails $Q_1$ which must not be entailed. 

A second query to $u$ could be, for example, 
%$Q_2 : \setof{(\exists Y writes(pam,Y)) \rightarrow res(pam))}$ (i.e.\ ``should it be sufficient for Pam to be a researcher that she just writes anything?'').
$Q_2 : \setof{\exists X ((\exists Y writes(X,Y)) \land \lnot res(X))}$ (i.e.\ ``is there somebody who writes something, but is no researcher?''). 
Again, it is reasonable to suppose that $u$ might know whether or not this should hold in their intended domain model. The (seemingly obvious) answer in this case would be positive, e.g.\ because $u$ intends to model students who write homework, exams, etc., but are no researchers. 
%not every student who writes an exam should be automatically deduced to be a researcher, 
This positive answer leads to the new positive test case $Q_2$. Adding this positive test case, like a set of new formulas, to the KB $\mo$ would result in $\mo_{new} := \mo \cup Q_2$. The debugging system would then figure out that formula~\ref{ex0:s2} is the only minimal conflict set in the KB $\mo_{new}$. The reason for this is that the elimination of formula~\ref{ex0:s2} breaks the entailment $Q_1$ (negative test case) and enables the addition of a new desired entailment $Q_2$ (positive test case) without involving the violation of any requirements (consistency). Therefore, formula~\ref{ex0:s2} is the only minimal diagnosis that is still compliant with the new knowledge in terms of $\Q_1 = \false$ and $Q_2 = \true$ obtained. 

It is important to notice that the solution KB $\mo_{new}$ that is returned to the user as a result of the interactive debugging session includes a new logical formula $Q_2$ that can be seen as a repair of the deleted formula~\ref{ex0:s2}. Since the knowledge after the debugging session is that $\lnot \ref{ex0:s2} \equiv Q_2$ must be true, this new knowledge is incorporated into the KB $\mo_{new}$. This indicates that the fault in KB was simply that the $\lnot$ in front of formula~\ref{ex0:s2} had been forgotten. 

Notice however that the positive test case $Q_2$ is not added to $\mo$ as a usual KB formula, but rather as an \emph{extension} of $\mo$ that has already been approved by the user. 
%In this vein, as an element of the set of positive test cases, $Q_2$ can be regarded as a background  
Should the user at some later point in time commit the same fault again (and explicitly specify some formula $x$ equivalent to formula~\ref{ex0:s2}), then the interactive debugging system, owing to the positive test case $Q_2$, would immediately detect a singleton conflict comprising only formula $x$. As a consequence, each diagnosis considered during this later debugging session would suggest to delete or modify (at least) $x$. 
%compensates for the lost \emph{correct} entailments due to the deletion of formula~\ref{ex0:s2}. Overall, the transition from the faulty initial KB $\mo$ to the correct solution KB $\mo_{new}$ can be 

This scenario should illustrate that, in spite of not being able to specify their domain knowledge in a logically consistent way, the user $u$ might still be able to answer questions about the intended domain, which supports our second assumption made about the user $u$ (the reader might agree that answering $Q_1$ and $Q_2$ is much easier than recognizing the entailment $\forall X\, res(X)$ of the KB). In other words, the availability of an (efficient) debugging system could help $u$ debug their KB, without needing to analyze \emph{which} entailments hold or do not hold, \emph{why} certain entailments hold or do not hold or \emph{why} exactly the KB does not meet certain imposed requirements or test cases, by simply answering queries \emph{whether} a certain entailment \emph{should} or \emph{should not} hold. These queries are automatically generated by the system in a way that they focus on the problematic parts of the KB, i.e.\ the minimal conflict sets, and discriminate between the possible solution candidates, i.e.\ the minimal diagnoses. 

\paragraph{Benefits of the Usage of Conflict Sets.}
We want to remark that the usage of minimal conflict sets ``naturally'' forces the system to take into consideration only the smallest relevant (faulty) parts of the problematic KB. This is owed to the property of minimal conflict sets to abstract from what \emph{all} the reasons for a certain entailment or requirements violation are. Instead, only the ``root'' (subset-minimal) causes for such violations are examined and no computation time is wasted to extract ``purely derived'' causes (those which are resolved as a byproduct of fixing all root causes from which it is derived, cf.\ \cite{Horridge2011a,Kalyanpur2006a}). For example, assuming the debugging scenario involving our example KB consisting only of formulas \ref{ex0:s1}-\ref{ex0:s4} which is incoherent and a requirements set including coherency. Then, there are two entailments reflecting the incoherency of this KB, first $\forall X\, \lnot secr(X)$ and second $\forall X\, \lnot gen(X)$ (these entailments hold due to $\forall X\, res(X)$ which follows from formulas~\ref{ex0:s1} and \ref{ex0:s2}). Of these two, only the second one is a ``root'' problem; the first one is a ``purely derived'' problem. That means, the entailment $\forall X\, \lnot secr(X)$ only holds due to the presence of the entailment $\forall X\, \lnot gen(X)$. So, the cause for $\forall X\, \lnot gen(X)$ is given by the set of formulas $\setof{\ref{ex0:s1},\ref{ex0:s2},\ref{ex0:s4}}$ whereas the proper superset $\setof{\ref{ex0:s1},\ref{ex0:s2},\ref{ex0:s3},\ref{ex0:s4}}$ of this set accounts for the entailment $\forall X\, \lnot secr(X)$. The exploitation of minimal conflict sets (the only minimal conflict set for this KB is $\setof{\ref{ex0:s1},\ref{ex0:s2},\ref{ex0:s4}}$) ascertains that such ``purely derived'' causes of requirements or test case violations will not be considered at all.  

\paragraph{The Ability to Incorporate Background Knowledge.}
Another feature of the approaches described in this work is their ability to incorporate relevant additional information in terms of a background knowledge KB $\mb$ (which is regarded to be correct). $\mb$ is a (consistent) KB which is usually semantically related with the faulty KB, e.g.\ $\mb$ represents knowledge about the domain modeled by $\mo$ that has already been sufficiently endorsed by domain experts. For instance, a doctor who wants to express their knowledge of dermatology in terms of a KB might resort to an approved background KB that specifies the human anatomy. Taking this background information into account puts the problematic KB into some context with existing knowledge and can thereby help a great deal to restrict the search space for solutions of the (interactive) KB debugging problem. This has also been found in~\cite{Stuckenschmidt2008}. This useful strategy of prior search space restriction is also exploited in the field of ontology matching\footnote{\url{http://www.ontologymatching.org/}} where automatic systems are employed to generate an alignment, i.e.\ a set of correspondences between semantically related entities of two different ontologies (KBs). Here, both ontologies are considered correct and diagnoses are only allowed to include elements of the alignment~\cite{Meilicke2007}. 

Applying a strategy like that to our example KB given above, supposing that we know that Pam is not a researcher in the world the KB should model, we might specify the background KB $\mb := \setof{\lnot res(pam)}$ prior to starting the interactive debugging session. This would immediately reduce the initial set of possible minimal diagnoses from five (i.e.\ the entire KB) to two (i.e.\ the first two formulas \ref{ex0:s1} and \ref{ex0:s2}). Reason for this is that the entailment $\forall X\, res(X)$ of formulas \ref{ex0:s1} and \ref{ex0:s2} already conflicts with the background knowledge $\lnot res(pam)$.    

%\begin{figure*}[htbp]
	%\centering
		%\includegraphics[width=0.7\textwidth]{debugging_workflow_inter.pdf}
	%\caption{The principle of interactive KB debugging.}
	%\label{fig:interactive_debugging_workflow}
%\end{figure*}

\begin{figure*}[t]
	\centering
		\includegraphics[width=0.85\textwidth]{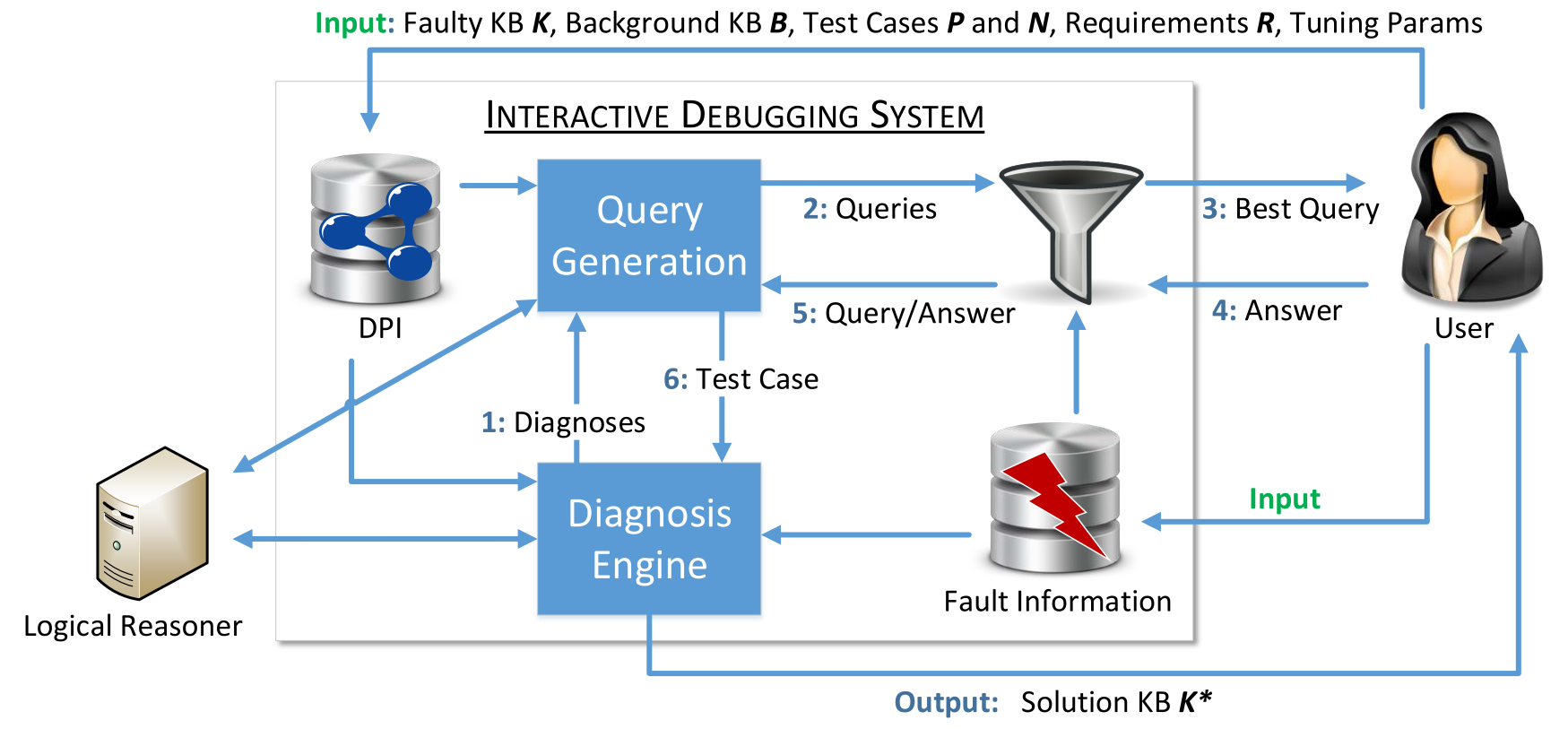}
	\caption[The Principle of Interactive KB Debugging]{The principle of interactive KB debugging.}
	\label{fig:interactive_debugging_workflow}
\end{figure*}

\paragraph{Outline of an Interactive KB Debugging System.}
The schema of an interactive debugging system is pictured by Figure~\ref{fig:interactive_debugging_workflow}.\footnote{Thanks to Kostyantyn Shchekotykhin for making available to me parts of this diagram.} As in the case of a non-interactive debugging system (see above), the system receives as input a \emph{diagnosis problem instance (DPI)}.
 %specified by 
%\begin{itemize}
	%\item some KB $\mo$ formulated using some (monotonic) logical language $\mathcal{L}$ (every formula in $\mo$ might be correct or faulty),
	%\item (optionally) some KB $\mb$ formalizing some background knowledge relevant for the domain modeled by $\mo$ (such that $\mb$ and $\mo$ do not share any formulas; all formulas in $\mb$ are considered correct)
	%\item a set of requirements $\RQ$ to the correct KB,
	%\item sets of positive ($\Tp$) and negative ($\Tn$) test cases (over $\mathcal{L}$) asserting desired semantic properties of the correct KB and
	%\item (optionally) some fault information $\FP$, e.g.\ in terms of fault probabilities of logical formulas in $\mo$.
%\end{itemize}
Further on, a range of additional parameters might be provided to the system. These serve as a means to fine-tune the system's behavior in various aspects. 
%Whereas the inputs given above describe the diagnosis problem instance at hand, 
Hence, we call these inputs \emph{tuning parameters}. These are (roughly) explained next.

First, some parameters might be specified that take influence on the number of leading diagnoses used for query generation and the necessary computation time invested for leading diagnoses computation. Moreover, some parameter determining the quantity of (pre-)generated queries (of which one is selected to be asked to the user) versus the reaction time (the time it takes the system to compute the next query after the current one has been answered) of the system can be chosen. A further input argument is a query selection measure constituting a notion of query ``goodness'' that is employed to filter out the ``best'' query among the set of generated queries. To give the system a criterion specifying when a solution of the interactive KB debugging problem is ``good enough'', the user is allowed to define a fault tolerance parameter $\sigma$. The lower this parameter is chosen, the better the (possibly ``approximate'') solution that is guaranteed to be found. In case of specifying this parameter to zero, the system will (if feasible) return the ``exact'' solution of the interactive KB debugging problem. Roughly, the exact solution is given in terms of a solution KB obtained by means of a \emph{single} solution candidate (minimal diagnosis) that is left after a sufficient number of queries have been answered (and added to the test cases). On the contrary, an approximate solution is represented by a solution KB obtained by means of a solution candidate with sufficiently high probability (where ``sufficiently high'' is determined by $\sigma$) at some point where there are still multiple solution candidates available. 

Finally, the user may choose between two different modes ($static$ or $dynamic$) of determining the leading diagnoses. The $static$ diagnosis computation strategy guarantees a constant ``convergence'' towards the exact solution by ``freezing'' the set of solution candidates at the very beginning and exploiting answered queries only for the deletion of minimal diagnoses. A possible disadvantage of this approach is the lack of efficient pruning of the used search tree. On the other hand, the $dynamic$ method of calculating leading diagnoses has a primary focus on the preservation of a search tree of small size, thereby aiming at being able to solve diagnosis problem instances which are not solvable by the $static$ approach due to high time and (more critically) space complexity. To this end, more powerful pruning rules are applied in this case which do not permit the algorithm to consider only a fixed set of solution candidates. Rather, the set of minimal diagnoses and minimal conflict sets are generally variable in this case which means that they are subject to change after assignment of an answered query to the test cases.    

Like in the case of a non-interactive debugger, an interactive debugging system requires a sound and complete logical reasoner for deciding consistency (coherency) and calculating logical entailments of a KB formulated over the language $\mathcal{L}$. 

The workflow in interactive KB debugging illustrated by Figure~\ref{fig:interactive_debugging_workflow} is the following:
\begin{enumerate}
	\item A set of leading diagnoses is computed by the diagnosis engine (by means of the fault information, if available) using the logical reasoner and passes it to the query generation module.
	\item The query generation module computes a pool of queries exploiting the set of leading diagnoses and delivers it to the query selection module.
	\item The query selection module filters out the ``best query'' (often by means of the fault information, if available) and shows it to the interacting user.
	\item The user submits an answer to the query.
	\item The query along with the given answer is used to formulate a new test case.
	\item This new test case is transferred back to the diagnosis engine and taken into account in prospective iterations. If the stop criterion (as per $\sigma$, see above) is not met, another iteration starts at step 1. Otherwise, the solution KB $\ot$ constructed from the currently most probable minimal diagnosis is output.
\end{enumerate}

\paragraph{Contributions of this Work.}
The contributions of this work are the following:
\begin{itemize}
	\item This work provides a thorough account of the subject and evolves the theory of interactive KB debugging (for monotonic KBs) by presupposing a reader to have only some basic knowledge of logic. Hence, this work addresses newbies as well as people already familiar with related topics. Whereas the comprehensive theoretical considerations might appeal to the more theoretically oriented readers such as researchers, the precise and exhaustive description of all discussed algorithms might be interesting from the implementation point of view and might serve more practically oriented people such as programmers or engineers as an algorithmic cookbook. Further on, the extensive illustration of the way algorithms work by examples might also serve a merely superficially interested reader to just receive a rough impression of how KBs might be interactively debugged.  
	\item Except for basics in FOL and PL, this work is self-contained and provides all necessary definitions and proofs to make the topic of interactive KB debugging accessible to the reader.
	\item To the best of our knowledge, this work provides the most comprehensive and detailed introduction to the field of interactive debugging of (monotonic) KBs. Our previous works on the topic~\cite{Shchekotykhin2012,ksgf2010,Rodler2013,friedrich2005gdm,Shchekotykhin2014} are more application-oriented and thus abstract from some details and omit some of the proofs in favor of comprehensive evaluations of the presented strategies.
	\item This is the first work that gives formal and precise definitions of problems dealt with in interactive KB debugging and introduces methods that provably solve these problems. We believe that precise problem statements are the very basis for all further scientific investigations in a field. Hence, we hope that this work can ``open'' the important subject of interactive KB debugging to a broader audience of interested researchers. This can lead to further progress and improvements in debugging techniques which we deem essential in the light of the growing number of intelligent applications incorporating KBs of growing size and complexity (keyword: The Semantic Web \cite{berners2001semantic}).
	\item An in-depth discussion of query computation including computational complexity considerations together with an accentuation of potential ways of improving these methods is given. The investigated methods for query computation have been used also in~\cite{Shchekotykhin2012,Rodler2013,ksgf2010,Shchekotykhin2014}, but have not been addressed in depth in these works.
	\item We are concerned with the discussion of different ways of exploiting diverse sources of meta information in the KB debugging process from which diagnosis probabilities can be extracted. Our previous works on this topic~\cite{Shchekotykhin2012,Rodler2013,ksgf2010,Shchekotykhin2014} do not address this matter in a comparable depth.
	\item We give a formal proof of the soundness of an algorithm $\scQX$ (based on~\cite{junker04}) for the detection of a minimal conflict set in a KB and we show the correctness (completeness, soundness, optimality) of a hitting set tree algorithm \textsc{HS} (based on~\cite{Reiter87}) for finding minimal diagnoses in a KB in best-first order (i.e.\ most probable diagnoses first) which uses $\scQX$ for conflict set computation only on-demand. We are not aware of any other work that comprises such proofs. 
	\item We establish the theoretical relationship between the widely-used notions of a conflict set and a justification. The former is i.a.\ used in~\cite{dekleer1987,Reiter87,Shchekotykhin2012,Rodler2013} and the latter i.a.\ in~\cite{Horridge2008,Horridge2009,Horridge2010,Horridge2011a,Horridge2011b,Horridge2012,Suntisrivaraporn2008,Kalyanpur2006a,MeilickeStuck2009,SattlerSZ09,Nikitina2011}. As a consequence, empirical results concerning the one might be translated to the other. For instance, since each minimal conflict set is an subset of a justification and there is an efficient (polynomial) method for computing a minimal conflict set given a superset of a minimal conflict set, a result manifesting the efficiency of justification computation for a set of KBs (e.g.\ \cite{Horridge2012b}) implies the efficiency of conflict set computation for the same set of KBs. Moreover, we argue that minimal conflict sets are the better choice for our system since these put the focus of the debugger only on the smallest faulty subsets of the KB whereas justifications are better suited in scenarios where exact explanations for the presence of certain entailments are sought.
	\item Two new algorithms for iterative (leading) diagnosis computation in interactive KB debugging are proposed. One that is guaranteed to reduce the number of remaining solutions after a query is answered and one that features more powerful pruning techniques than our previously published algorithms~\cite{Shchekotykhin2012,Rodler2013} (an evaluation that compares the overall efficiency of our previous algorithms with the ones proposed in this work must still be conducted and is part of our future research).
	\item We suggest and extensively analyze different methods for the selection of an ``optimal'' query to ask the user out of a pool of possible queries. We compare a greedy ``split-in-half'' strategy that proposes queries which eliminate half of the leading diagnoses with a strategy relying on information entropy \cite{shannon1948} that chooses the query with highest information gain based on some statistic or (a user's) beliefs about faults in the KB. Comprehensive experiments manifest that only an average guess of the fault information suffices to reduce the query answering effort for the interacting user, often to a significant extent, by means of the latter strategy compared to the former. Moreover, we demonstrate that both methods clearly outperform a random query selection strategy. The latter result witnesses that incorporation of meta (fault) information into the debugging process is in fact reasonable and might relieve the interacting user of a significant proportion of the effort required without taking into account any meta information.
	\item Addressing the issue of choosing the suitable query selection method for some given fault information, we present a reinforcement learning query selection strategy. For, reliance upon a strategy (e.g.\ information entropy) that fully exploits and gains from the given fault information can speed up the debugging procedure in the normal case, but can also have a negative impact on the performance in the bad case where the actual solution diagnosis is rated as highly improbable. As an alternative, one might prefer to rely on a tool (e.g.\ ``split-in-half'') which does not consider any fault information at all. In this case, however, possibly well-chosen information cannot be exploited, resulting again in inefficient debugging actions. 
	
Minimal effort for the interacting user can be achieved if both the query selection method is chosen carefully and the provided fault information satisfies some minimum quality requirements. In particular, for deficient fault information and unfavorable strategy for query selection, we observe cases where the overhead in terms of user effort exceeds 2000\% (!) in comparison to employing a more favorable query selection strategy. Since, unfortunately, assessment of the fault information is only possible a-poteriori (after the debugging session is finished and the correct solution is known), we devise a learning strategy (RIO) that continuously adapts its behavior depending on the performance achieved and in this vein minimizes the risk of using low-quality fault information. 

This approach makes interactive debugging practical even in scenarios where reliable fault estimates are difficult to obtain. Evaluations provide evidence that for 100\% of the cases in the hardest (from the debugging point of view) class of faulty test KBs, RIO performed at least as good as the \emph{best} other strategy and in more than 70\% of these cases it even manifested superior behavior to the \emph{best} other strategy. Choosing RIO over other approaches can involve an improvement by the factor of up to 23, meaning that more than 95\% of user time and effort might be saved per debugging session.
	\item We come up with mechanisms for efficiently dealing with KB debugging problems involving high cardinality (minimal) diagnoses. In the standard interactive debugging approach described in the first parts of this work, the computation of queries is based on the generation of the set of \emph{most probable (or minimum cardinality)} leading diagnoses. By this postulation, certain quality guarantees about the output solution can be given. However, we learn that dropping this requirement can bring about substantial savings in terms of time and especially space complexity of interactive debugging, in particular in debugging scenarios where faulty KBs are (partly) generated as a result of the application of automatic systems, e.g.\ KB (ontology) learning or matching systems~\cite{Huber2011,Ngo2012,Jean-Mary2009,Reul2010,Jimenez-Ruiz2012a,meilicke2011}.
	
To cope with such situations, we propose to base query computation on \emph{any} set of leading diagnoses using a ``direct'' method for diagnosis generation. Contrary to the standard method that exploits minimal conflict sets, this approach takes advantage of the duality between minimal diagnoses and minimal conflict sets and employs ``inverse'' algorithms to those used in the standard approach in order to determine minimal diagnoses directly from the DPI without the indirection via conflict sets.

We study the application of this direct method to high cardinality faults in KBs and find out that the number of required queries per debugging session is hardly affected for cases when the standard approach is also applicable. However, the direct method proves applicable and able to locate the correct solution diagnosis in situations when the standard approach (albeit one that not yet incorporates the powerful search tree pruning techniques introduced in this work) is not due to time or memory issues.
\end{itemize}

\begin{figure*}[tp]
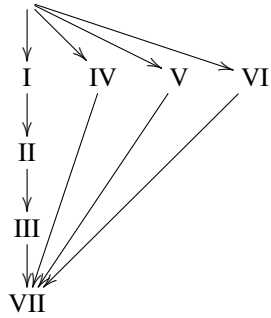

\centering
\begin{minipage}[c]{0.99\textwidth}
\xygraph{
!{<0cm,0cm>;<1cm,0cm>:<0cm,1cm>::}
%!~-{@[|(4)]}
%d=0
!{(1,4)}*+{}="start"
!{(1,3)}*+{\text{\RM 1}}="1"
!{(1,2)}*+{\text{\RM 2}}="2"
!{(1,1)}*+{\text{\RM 3}}="3"
!{(2,3)}*+{\text{\RM 4}}="4"
!{(3,3)}*+{\text{\RM 5}}="5"
!{(4,3)}*+{\text{\RM 6}}="6"
!{(1,0)}*+{\text{\RM 7}}="7"
%d=1
%%d=3
%!{(1,1) }*+{\times}="inv_d2_q2"
%!{(2,1) }*+{\times}="inv_d3_q2"
%!{(3,1) }*+{\checkmark}="d5"
%!{(4,1) }*+{\checkmark}="val_d3_q2"
%%d=4
%!{(1,0) }*+{\times}="inv_d2_q3"
%!{(3,0) }*+{\times}="inv_d5_q4"
%!{(4,0) }*+{\checkmark}="val_d4_q4"
%d0->d1
"start":"1"
"1":"2"
"2":"3"
"start":"4"
"start":"5"
"start":"6"
"3":"7"
"4":"7"
"5":"7"
"6":"7"
%%d2->d3
%"val_d2_q1":@2{->}"inv_d2_q2"^{Q_2}
%"d3":@2{->}"inv_d3_q2"^{Q_2}
%"nonmin1":@2{->}"d5"^{Q_3}
%"d3":@2{->}"val_d3_q2"^{Q_2}
%%d3->d4
%"d5":@2{->}"inv_d5_q4"^{Q_4}
%"val_d4_q3":@2{->}"val_d4_q4"^{Q_4}   
%"val_d2_q2":@2{->}"inv_d2_q3"^{Q_3}
}
\end{minipage}
\caption[Precedence Constraints among the Parts]{Precedence constraints among the parts of this work.}
\label{fig:precedence_constraints_among_parts}
\end{figure*}

\paragraph{Organization of this Work.}
%The rest of this work is organized as follows. 
This work is subdivided into seven parts. Figure~\ref{fig:precedence_constraints_among_parts} illustrates the precedence constraints among the parts. We want to point out that Parts~\ref{part:JWS}-\ref{part:DX} correspond to works that have already been published and are thus self-contained, both from the notation and the content point of view. Parts~\ref{part:Prolog}-\ref{part:IterativeDiagnosisComputation}, on the contrary, are constructive and should thence be read in order.\\

\noindent\emph{(Rest of) Part~\ref{part:Prolog}.} \quad
In Chapter~\ref{chap:basics}, besides introducing the notation used in this work, we describe the requirements imposed on logical knowledge representation languages $\mathcal{L}$ 
%(or better: on the entailment relation in $\mathcal{L}$) 
that might be used with our approaches.
%, namely monotonicity, idempotency as well as extensiveness. 
It should be noted that the postulated properties do not restrict the applications of our approaches very much. For instance, these might be employed to resolve over-constrained constraint satisfaction problems (CSPs) or repair faulty KBs in PL, FOL, DL, Datalog or OWL. Since DL provides the logical underpinning of OWL which has recently received increasing attention due to the extensive research in the field of The Semantic Web \cite{berners2001semantic}, we will also give a short introduction to DL. For, to underline the flexibility of the presented debugging systems in this work, we will illustrate how they work by means of examples involving PL, FOL as well as DL KBs. 

In Chapter~\ref{chap:KBDebugging}, we first give a formal definition of the \emph{KB debugging} problem and define a diagnosis problem instance (DPI), the input of a KB debugger, and a solution KB, the output of a KB debugger. Further on, we formally characterize a diagnosis and give the notion of KB validity and what it means for a KB to be faulty. We discuss and prove relationships between these notions and specify properties a DPI must satisfy in order to be solvable by a KB debugger.  

We motivate why it makes sense to focus on set-minimal diagnoses instead of all diagnoses, i.e.\ to stick to ``The Principle of Parsimony''~\cite{Reiter87,Bylander1991}. This results in the definition of the problem of \emph{parsimonious KB debugging}. Then, we prove that solving this problem is equivalent to the computation of a minimal diagnosis.  
Finally, we explain the benefits of using some background KB in (parsimonious) KB debugging.% (Section~\ref{sec:BackgroundKnowledge}), 

In Chapter~\ref{chap:DiagnosisComputation} we describe methods for diagnosis computation.
%(Chapter~\ref{chap:DiagnosisComputation}). 
To this end, we first introduce the notion of a (minimal) conflict set, discuss some properties of conflict sets related to the notion of KB validity and give sufficient and necessary criteria for the existence of non-trivial conflict sets w.r.t.\ a DPI. Subsequently, we derive the relationship between a conflict set and the notion of a justification (a minimal set of formulas necessary for a particular entailment to hold) which is well-known and frequently used, especially in the fields of DL, OWL and The Semantic Web~\cite{Horridge2008, Horridge2009, Horridge2010, Horridge2011a, Horridge2011b, Horridge2012b}. Concretely, we will demonstrate that a minimal conflict set is a subset of a justification for some negative test case or for some inconsistency (entailment $\false$) or incoherency (entailment $\forall X_1,\dots,X_k\, \lnot p(X_1,\dots,X_k)$ for some predicate symbol $p$ of arity $k$) of the given KB. Moreover, we will learn that, for the debugging tasks we consider, conflict sets are better suited than justifications.

Having deduced all relevant characteristics of (minimal) conflict sets, we proceed to give a description of a method ($\scQX$, Algorithm~\ref{algo:qx}) due to~\cite{junker04} which was originally presented as a method for finding preferred explanations (conflicts) in over-constrained CSPs, but can also be employed for an efficient computation of a minimal conflict set w.r.t.\ a DPI in KB debugging. We discuss and exemplify this algorithm in detail, prove its correctness as a routine for minimal conflict set computation and give complexity results.

Having at our disposal a proven sound method for generation of a minimal conflict set, we continue with the delineation of a hitting set tree algorithm similar to the one originally presented in~\cite{Reiter87} which enables the computation of different minimal conflict sets by means of successive calls to $\scQX$, each time given an (adequately) modified DPI. In this manner, a hitting set tree can be constructed (breadth-first) which facilitates the computation of minimal diagnoses (minimum cardinality diagnoses first). We prove the correctness (termination, soundness, completeness, minimum-cardinality-first property) of this hitting set tree algorithm coupled with the $\scQX$ method which serves to solve the problem of parsimonious KB debugging.

In order to be able to incorporate fault information into the diagnoses finding process, we deal with the induction of a probability space over diagnoses in Section~\ref{sec:DiagnosisProbabilitySpace}. We discuss several ways of constructing a probability space including different sources of fault information. 
%in Section~\ref{sec:prob_space_construction}. 
Hereinafter, we detail how diagnosis probabilities can be determined on the basis of some available fault information and how these can be appropriately updated after new observations (in terms of answered queries) have been made. 
%In Section~\ref{sec:probs_diag_comp} 
Furthermore, we outline how fault probabilities can be appropriately incorporated into the hitting set search tree in order to guarantee the discovery of minimal diagnoses in best-first order, i.e.\ most probable ones first. Then, 
%in Section~\ref{sec:CorrectnessOfAlgorithmHs}, 
we prove the correctness (termination, soundness, completeness, best-first property) of this best-first diagnosis finding algorithm for parsimonious KB debugging.

%Section~\ref{sec:non_int_debug_procedure} 
Finally, we describe a non-interactive KB debugging procedure (Algorithm~\ref{algo:non_int_debug}) that relies on this best-first diagnosis finding algorithm. Some illustrating examples are provided which at the same time reveal significant shortcomings present in non-interactive KB debugging. This motivates the development of interactive KB debugging algorithms.

Readers not theoretically inclined or non-interested in the technical details might well skip Sections~\ref{sec:ConflictSetsVersusJustifications}, \ref{sec:cs_comp_correctness}, \ref{sec:CorrectnessOfBreadthFirstDiagnosisComputation} and \ref{sec:DiagnosisProbabilitySpace} in Part~\ref{part:Prolog}.\\

\noindent\emph{Part~\ref{part:InteractiveKBDebugging}.} \quad
%In Part~\ref{part:InteractiveKBDebugging} 
In Chapter~\ref{chap:MotivationAndProblemDefinitions}, we first discuss how disadvantages of non-interactive KB debugging procedures can be overcome by allowing a user to take part in the debugging process. Then, we define the problem of \emph{interactive static KB debugging} as well as the problem of \emph{interactive dynamic KB debugging} which ``naturally'' arise from the fact that the DPI in interactive KB debugging is always renewed after a new test case has been specified (a new query has been answered). The former problem searches for a solution KB \emph{w.r.t.\ the DPI given as input} such that this solution KB satisfies all test cases added during the debugging session and there is no other such solution KB. The latter problem searches for a solution KB \emph{w.r.t.\ the current DPI} (i.e.\ the input DPI including all new test cases added throughout the debugging session so far) such that there is no other solution KB w.r.t.\ the current DPI.

Next, in Chapter~\ref{chap:UserInteraction}, the central term of a \emph{query} is specified which constitutes the medium for user interaction. Queries are generated from a set of \emph{leading diagnoses} which is characterized thereafter. The set of leading diagnoses is uniquely partitioned into three subsets by each query. The tuple including these subsets is called \emph{q-partition}. Subsequently, the reader is given some explanations how the q-partition can be interpreted, and how it relates to a query. In fact, we will prove that the notion of a q-partition can serve as a criterion for checking whether a set of logical formulas is a query or not. After that, we will learn that a query exists for any set of (at least two) leading diagnoses which grants that the presented algorithms will definitely be able to come up with a query without the need to impose any restrictions on which (minimal) diagnoses are computed by the diagnosis engine in each iteration. 

Chapter~\ref{chap:QueryGeneration} shows a method for the generation of (a pool of) set-minimal queries (Algorithm~\ref{algo:query_gen}) aiming at stressing the interacting user as sparsely as possible, features in-depth discussions of this method's properties, proves its correctness, provides complexity results and gives some illustrating examples. Further on, drawbacks of this method are pointed out and possible solutions are discussed. 

Subsequently, Chapter~\ref{chap:WorkflowInInteractiveKBDebugging} deals with the presentation of the central algorithm of this work which implements an interactive KB debugging system (Algorithm~\ref{algo:inter_onto_debug}). First, an overview of the workflow of interactive KB debugging %(Section~\ref{sec:AlgorithmOverview}) 
is given, followed by a more comprehensive detailed specification of the algorithm. 
%(Section~\ref{sec:DetailedAlgorithmDescription}). 
Some query selection measures are discussed~\cite{Rodler2013,Shchekotykhin2012} 
%(Section~\ref{sec:query_selection_measures}) 
and optimization versions of the problems of interactive dynamic and static KB debugging are defined where the goal is to obtain the solution to these problems by asking the user a minimal number of queries. Finally, %Section~\ref{sec:CorrectnessOfAlgorithmInterOntoDebug}
we prove the correctness of the interactive KB debugging algorithm and provide a discussion of its complexity.

Non-theoretically-oriented readers might well skip Sections~\ref{sec:remarks_query_gen}, \ref{sec:SoundnessOfQueryMinimization}, \ref{sec:query_gen_complexity}, \ref{sec:query_gen_correctness} and \ref{sec:CorrectnessOfAlgorithmInterOntoDebug} in Part~\ref{part:InteractiveKBDebugging}. Moreover, for the superficially interested reader, it may suffice to concentrate only on Chapter~\ref{chap:MotivationAndProblemDefinitions} and Sections~\ref{sec:Queries}, \ref{sec:LeadingDiagnoses} and \ref{sec:AlgorithmOverview} in Part~\ref{part:InteractiveKBDebugging}.\\

\noindent\emph{Part~\ref{part:IterativeDiagnosisComputation}.} \quad
Here, we go into detail w.r.t.\ the two strategies for iterative diagnoses computation introduced in Part~\ref{part:InteractiveKBDebugging} that might be plugged into Algorithm~\ref{algo:inter_onto_debug} to solve either the interactive static or dynamic KB debugging problem. 

Chapter~\ref{chap:StaticHSTree} describes the $static$ method and proves its soundness and completeness w.r.t.\ the computation of minimal diagnoses w.r.t.\ the DPI given as an input to the interactive KB debugging algorithm and its optimality w.r.t.\ the discovery of minimal diagnoses in best-first order (most-probable or minimum cardinality diagnoses first). Incorporation of the $static$ method as a routine for leading diagnosis computation into Algorithm~\ref{algo:inter_onto_debug} provably solves the problem of interactive static KB debugging.
%Used as a routine for leading diagnosis computation in Algorithm~\ref{algo:inter_onto_debug}, the $static$ method solves the problem of interactive static KB debugging.

Chapter~\ref{chap:DynamicHSTree} details the $dynamic$ method and proves its soundness and completeness w.r.t.\ the computation of minimal diagnoses w.r.t.\ the current DPI and its optimality w.r.t.\ the discovery of minimal diagnoses in best-first order (most-probable or minimum cardinality diagnoses first). Employing the $dynamic$ method as a routine for leading diagnosis computation in Algorithm~\ref{algo:inter_onto_debug} provably solves the problem of interactive dynamic KB debugging.

The practically oriented reader or the one that is willing to believe that the presented iterative diagnosis computation techniques in fact work as claimed might skip Sections~\ref{sec:CorrectnessOfTextscStaticHS} as well as \ref{sec:TextscDynamicHSDetailsAndCorrectness} in Part~\ref{part:IterativeDiagnosisComputation}.\\

\noindent\emph{Part~\ref{part:JWS}.} \quad
In this part, we suggest and extensively analyze different methods for the selection of an ``optimal'' query (see above). The material dealt with in Part~\ref{part:JWS} is based on the publications \cite{Shchekotykhin2012, ksgf2010} where the former was published in the journal \emph{Web
Semantics: Science, Services and Agents on the World Wide Web} and the latter in the \emph{Proceedings of the 9th International Semantic Web Conference (ISWC 2010)}.\\

\noindent\emph{Part~\ref{part:RIO}.} \quad
The reinforcement learning query selection strategy (RIO) that makes the presented debugging system robust against the usage of low-quality fault information is presented and thoroughly analyzed in this part which is based on the works 
\cite{Rodler2013, Rodler2012OM, Rodler2011, Shchekotykhin2011} published in \emph{Web Reasoning and Rule Systems (RR-2013)}, in the \emph{Proceedings of the 7th International Workshop on Ontology Matching (OM-2012)}, in the \emph{Proceedings of the Joint Workshop on Knowledge Evolution and Ontology Dynamics 2011 (EvoDyn2011)} and in \emph{DX 2011 - 22nd International Workshop on Principles of Diagnosis}, respectively.\\
%\cite{Rodler2013, Rodler2012OM, Rodler2011, Shchekotykhin2011} published in \emph{Web Reasoning and Rule Systems}, in the \emph{Proceedings of the 7th International Workshop on Ontology Matching (OM-2012)} and in the \emph{Proceedings of the Joint Workshop on Knowledge Evolution and Ontology Dynamics 2011 (EvoDyn2011)}, respectively.\\

\noindent\emph{Part~\ref{part:DX}.} \quad
This part covers the topic of efficiently dealing with KB debugging problems involving high cardinality faults (see above) and relies on material presented in \cite{Shchekotykhin2014, Shchekotykhin2014a, Shchekotykhin2014b} and published in the \emph{Proceedings of the 21st European Conference on Artificial Intelligence (ECAI 2014)}, in \emph{DX 2014 - 25th International Workshop on Principles of Diagnosis} and in the \emph{Proceedings of the Third International Workshop on
Debugging Ontologies and Ontology Mappings (WoDOOM14)}, respectively.\footnote{We are glad to report that the publication \cite{Shchekotykhin2014a} was awarded the Best Paper Award at the DX Workshop that took place in Graz, Austria, in September 2014 (see \url{http://dx-2014.ist.tugraz.at}).} \\

\noindent\emph{Part~\ref{part:Epilog}.} \quad
To round this work off, we provide a discussion of related work in Chapter~\ref{chap:RelatedWork},\footnote{Note that related work specific to topics addressed in Parts~\ref{part:JWS}-\ref{part:DX} is separately treated in these parts.}
%we talk about related work. 
summarize the contributions of this work in Chapter~\ref{chap:summary} and deal with our future work topics in Chapter~\ref{chap:future_work}.\\

\chapter{Preliminaries} \label{chap:basics}

\section{Assumptions}
\label{sec:assumptions}
The techniques described in this work are applicable for any logical knowledge representation formalism $\mathcal{L}$ for which 
%the following conditions are satisfied:
%entailment relation satisfies the following conditions:
the entailment relation is
\begin{enumerate}
	\item \label{logic:cond1} \emph{monotonic:} is given when adding a new logical formula to a KB $\mo_{\mathcal{L}}$ cannot invalidate any entailments of the KB, i.e. $\mo_{\mathcal{L}} \models \alpha_{\mathcal{L}}$ implies that $\mo_{\mathcal{L}} \cup \setof{\beta_{\mathcal{L}}} \models \alpha_{\mathcal{L}}$,
	\item \label{logic:cond2} \emph{idempotent:} is given when adding implicit knowledge explicitly to a KB $\mo_{\mathcal{L}}$ does not yield new entailments of the KB, i.e. $\mo_{\mathcal{L}} \models \alpha_{\mathcal{L}}$ and $\mo_{\mathcal{L}} \cup \setof{\alpha_{\mathcal{L}}} \models \beta_{\mathcal{L}}$ implies $\mo_{\mathcal{L}} \models \beta_{\mathcal{L}}$ and
	\item \label{logic:cond3} \emph{extensive:} is given when each logical formula entails itself, i.e. $\{\alpha_{\mathcal{L}}\} \models \alpha_{\mathcal{L}}$ for all $\alpha_{\mathcal{L}}$,
\end{enumerate}
and for which 
\begin{enumerate}
\setcounter{enumi}{3}
	\item \label{logic:cond4} reasoning procedures for \emph{deciding consistency} and \emph{calculating logical entailments} of a KB are available, 
	%\item availability of reasoning services for \emph{deciding consistency} of a KB,
\end{enumerate}
where $\alpha_{\mathcal{L}}, \beta_{\mathcal{L}}$ are logical formulas and $\mo_{\mathcal{L}}$ is a set $\setof{\tax_\mathcal{L}^{(1)},\dots,\tax_\mathcal{L}^{(n)}}$ of logical formulas formulated over the language $\mathcal{L}$. $\mo_{\mathcal{L}}$ is to be understood as the conjunction $\bigwedge_{i=1}^{n} \tax_\mathcal{L}^{(i)}$. Notice that the elements of a KB are called quite differently in literature. Possible denotations are logical formula (e.g.\ \cite{kreuzer2006}), well-formed formula (e.g.\ \cite{chang1973}), (logical) sentence or axiom (e.g.\ \cite{russellnorvig2003}) and axiom (in most of the description logic literature, e.g.\ \cite{Baader2007}). We will mainly stick to the term \emph{formula} (sometimes \emph{axiom}) to refer to the elements of a KB.
As the logic will be clear from the context in the sequel, we will omit the index $\mathcal{L}$ when referring to formulas or KBs over $\mathcal{L}$ throughout the rest of this work.

\section{Considered Logics}
\label{sec:ConsideredLogics}
To underline the general character of this work, we will illustrate our approaches using example diagnosis problem instances expressed in different logical languages. In this section we give notational remarks concerning these different logics used, namely propositional logic (PL), first-order logic (FOL) as well as description logic (DL). Whereas we assume the reader to be familiar with FOL and PL (a good introduction to PL and FOL can be found in~\cite{chang1973}), we will give a short introduction to DL.

\begin{remark}\label{rem:used_logics}
It is important to notice that the usage of DL as well as FOL examples throughout this work should \emph{not} suggest that the Properties \ref{logic:cond1} -- \ref{logic:cond4} stated above are satisfied for any DL or FOL language $\mathcal{L}$. In fact, it is well-known by the theorems of Church and Turing (cf.\ \cite{mendelson2009}; the original works are \cite{church1936,turing1937}) that FOL is not decidable in general, i.e.\ Property~\ref{logic:cond4} above is not met. Also in the case of DL, which subsumes a range of different logical languages featuring different expressivity and thus different computational complexity of reasoning procedures, there are languages which are undecidable. For instance, a DL language allowing the formalism of equality role-value-maps which facilitates the expression of concepts like ``persons whose co-workers coincide with their relatives'' can be proven undecidable \cite{Baader2007,schmidt1989}.

Property~\ref{logic:cond4} is satisfied, for example, for the DL language $\mathcal{SROIQ}$ which is the logical underpinning of OWL~2 \cite{Grau2008a}. However, the complexity (2-$\textsc{NExpTime}$-complete~\cite{Kazakov2008}) of logical reasoning is intractable in the worst case for this language which implies the intractability of our methods in the worst case.
Nevertheless, other DL languages applied with similar systems as those described in this paper have been showing reasonable performance~\cite{Shchekotykhin2014, Rodler2013, Shchekotykhin2012}. Also from the theoretical point of view, there are DL languages that allow for efficient reasoning. One example is the OWL 2 EL profile which enables polynomial time reasoning~\cite{baader2005}. For this language, the efficient reasoning service ELK has been presented by~\cite{kazakov2014}.
For FOL, datalog is an example of a decidable sublanguage where reasoning is efficient \cite{russellnorvig2003}. Further, restricted sublanguages of FOL can often be translated to some DL language wherefore DL positive results concerning the decidability of reasoning as well as complexity results can be adopted for these restricted FOL languages~\cite[chapter~4]{Baader2007} \cite{borgida1996}.
%proven decidable DL languages have been demonstrated to be less expressive than FOL~\cite[chapter~4]{Baader2007} \cite{borgida1996}. Therefore, the decidability 

Moreover, we want to point out that the practical efficiency of our systems depends strongly on the practical performance (which might be by far better than suggested by the worst case reasoning complexities) of the reasoning services called by our algorithms since the reasoning services are used as a black-box (as mentioned in Chapter~\ref{chap:intro}). Possible strategies for improving the reasoning efficiency in the black-box setting are briefly discussed in Chapter~\ref{chap:future_work}.\qed 
%
%Despite that for this language the complexity (2-NExpTime-complete~\cite{Kazakov2008}) of logical reasoning is intractable in the worst case, it   
%
%Hence, the message should rather be that the methods of this work
%
%Hence, special care must be taken
%
 %it does not hold that all     
\end{remark}

\subsection*{Ontologies and The Semantic Web}
\label{sec:OntologiesAndDescriptionLogic}
%Our framework aims at providing support for a user when formulating KBs. In particular, it helps a user to assure that desired entailments hold in the formulated KB, non-desired entailments do not hold and that meaningful reasoning can be conducted with KB. 
Ontologies are KBs that formally and explicitly represent common knowledge about a domain in the form of individuals, concepts (set of individuals) and roles (binary relationships between individuals). As, in the last decade, extensive research has been done in the area of The Semantic Web~\cite{berners2001semantic} making (automatic) ontology development tools and reasoning services more efficient, ontology engineering for the Semantic Web is on the upswing. 
The Semantic Web aims at the enrichment of unstructured information on the web by semantic meta data which should facilitate the usage of the web as structured database of knowledge of all kinds where computers are able to ``understand'' this structured data, establish relationships between different data sources, combine information from different data sources and (most essentially) derive new (implicit) knowledge from the structured data. At this, ontologies are the key to a common vocabulary used for the semantic meta data. Ontologies are employed to precisely define the meaning of different terms, state relationships between different terms and to introduce new terms by means of already specified ones.

The constantly increasing number of people creating ontologies of increasing size (examples were given in Chapter~\ref{chap:intro}) results in more and more (faulty) ontologies which constitute useful application scenarios and test cases for our approaches. For that reason, we also want to use ontology engineering for The Semantic Web as a concrete use case for the presented work. The standard knowledge representation formalism for ontologies is OWL 2~\cite{Motik2009a,Grau2008a} which relies on DL. A short introduction to DL is given next.
%Without loss of generality, we will thus evolve our theory and methods with a focus on OWL~\cite{}, the standard knowledge representation formalism for ontologies, which relies on description logics. %underlying OWL and sublanguages thereof.

\subsubsection*{Description Logic}
\label{sec:DL}
\emph{Description Logic (DL)}~\cite{Baader2007} is a family of knowledge representation languages with a formal logic-based semantics that are designed to represent knowledge about a domain in form of concept descriptions. 
The \emph{syntax} of a description language $\mathcal{L}$ is defined by its signature and a set of constructors. The \emph{signature} of $\mathcal{L}$ corresponds to the union of possibly disjoint sets $N_C$, $N_R$ and $N_I$, where $N_C$ contains all concept names (unary predicates), $N_R$ comprises all role names (binary predicates) and $N_I$ is the set of all individuals (constants) in $\mathcal{L}$. Each concept and role description can be either atomic or complex. The latter ones are composed using constructors defined in the particular language $\mathcal{L}$.
A typical set of DL \emph{constructors} for complex concepts includes conjunction $A \sqcap B$, disjunction $A \sqcup B$, negation $\lnot A$, existential $\exists r.A$ and value $\forall r.A$ restrictions, where $A,B$ are concept descriptions and $r \in N_R$.

\emph{Axioms} are statements of knowledge that must be true in a domain. An \emph{ontology} $\mo$ %, called ontology henceforth, 
is defined as a tuple $(\mt, \ma)$, where $\mt$ (\emph{TBox}) is a set of terminological axioms and $\ma$ (\emph{ABox}) a set of assertional axioms. Each TBox axiom is expressed by a general concept inclusion $A \sqsubseteq B$, a form of logical implication, or by a definition $A \equiv B$, a kind of logical equivalence, where $A$ and $B$ are concept descriptions or role descriptions. 
ABox axioms are used to assert properties of individuals in terms of the vocabulary defined in the TBox, e.g.\ concept $A(x)$ or role $r(x,y)$ assertions, where $A$ is a concept description, $r$ a role description, and $x,y \in N_I$.
%Two kinds of ABox axioms are, e.g.,
%%ABox axioms can be of two kinds: 
%concept assertions $A(x)$ and role assertions $r(x,y)$, where $x,y \in N_I$. 

%The semantics of DLs is given in terms of interpretations $\mi =(\Delta^\mi, \cdot^\mi)$ consisting of a non-empty domain $\Delta^\mi$ and a function $\cdot^\mi$ that maps each concept to a subset of $\Delta^\mi$, each role to a subset of $\Delta^\mi \times \Delta^\mi$ and each individual to some value in $\Delta^\mi$. That is, $A^\mi \subseteq \Delta^\mi$, $r^\mi \subseteq \Delta^\mi \times \Delta^\mi$ and $x^\mi \in \Delta^\mi$ for all $A \in N_C$, $r \in N_R$ and $x \in N_I$. Complex concept description are mapped as follows: $(A \sqcap B)^\mi = A^\mi \cap B^\mi$, $(A \sqcup B)^\mi = A^\mi \cup B^\mi$, $(\lnot A)^\mi = \Delta^\mi \setminus A^\mi$, $(\exists r.A)^\mi = \setof{x\in\Delta\mi\,|\, \exists y. (x,y) \in r^\mi \land y \in A^\mi}$, and $\forall r.A = \setof{x\in\Delta^\mi\,|\, \forall y. (x,y) \in r^\mi \Rightarrow y^\mi \in A^\mi}$.
The semantics of a description language is given in terms of \emph{interpretations} $\mi =(\Delta^\mi, \cdot^\mi)$ consisting of a non-empty domain $\Delta^\mi$ and a function $\cdot^\mi$ that assigns to every atomic concept $A \in N_C$ a set $A^\mi \subseteq\Delta^\mi$, to every atomic role $r \in N_R$ a set $r^\mi \subseteq \Delta^\mi \times \Delta^\mi$ and to every individual $x\in N_I$ some value $x^\mi \in \Delta^\mi$. 
%That is, $A^\mi \subseteq \Delta^\mi$, $r^\mi \subseteq \Delta^\mi \times \Delta^\mi$ and $x^\mi \in \Delta^\mi$ for all $A \in N_C$, $r \in N_R$ and $x \in N_I$. Complex concept description are mapped 
The interpretation function is extended to complex concept descriptions by the following inductive definitions: 
\begin{align*}
\top^\mi &= \Delta^\mi\\ 
\bot^\mi &= \emptyset\\ 
(A \sqcap B)^\mi &= A^\mi \cap B^\mi\\ 
(A \sqcup B)^\mi &= A^\mi \cup B^\mi\\ 
(\lnot A)^\mi &= \Delta^\mi \setminus A^\mi\\ 
(\exists r.A)^\mi &= \setof{x\in\Delta^\mi\,|\, \exists y.\, (x,y) \in r^\mi \land y \in A^\mi} \\  
(\forall r.A)^\mi &= \setof{x\in\Delta^\mi\,|\, \forall y.\, (x,y) \in r^\mi \rightarrow y \in A^\mi}
\end{align*}
%$\top^\mi = \Delta^\mi$, $\bot^\mi = \emptyset$, $(A \sqcap B)^\mi = A^\mi \cap B^\mi$, $(A \sqcup B)^\mi = A^\mi \cup B^\mi$, $(\lnot A)^\mi = \Delta^\mi \setminus A^\mi$, $(\exists r.A)^\mi = \setof{x\in\Delta^\mi\,|\, \exists y. (x,y) \in r^\mi \land y \in A^\mi}$, and $\forall r.A = \setof{x\in\Delta^\mi\,|\, \forall y. (x,y) \in r^\mi \rightarrow y \in A^\mi}$ 
where $\top$
%~($\mathsf{Thing}$) 
and $\bot$
%~($\mathsf{Nothing}$) 
are predefined concepts; the former is the universal concept and the latter the bottom concept.
% referring to the set of all individuals and the empty set of individuals, respectively.

The semantics of axioms is defined as follows for (1) TBox and (2) ABox axioms: 
(1) Interpretation $\mi$ satisfies $A\sqsubseteq B$ iff $A^\mi \subseteq B^\mi$ and it satisfies $A\equiv B$ iff $A^\mi = B^\mi$. (2) $A(x)$ is satisfied by $\mi$ iff $x^\mi \in A^\mi$ and $r(x,y)$ is satisfied iff $(x^\mi,y^\mi) \in r^\mi$. 
%TBox axioms (1)~$A\sqsubseteq B$ and (2)~$A\equiv B$, respectively, are satisfied by interpretation $\mi$ iff (1)~$A^\mi \subseteq B^\mi$ and (2)~$A^\mi = B^\mi$. ABox axioms (1)~$A(x)$ and (2)~$r(x,y)$, respectively, are satisfied by $\mi$ iff (1)~$x^\mi \in A^\mi$ and (2)~$(x^\mi,y^\mi) \in r^\mi$. 
An interpretation $\mi$ is a \emph{model} of $\mo = (\mt,\ma)$ iff it satisfies all TBox axioms in $\mt$ and all ABox axioms in $\ma$. An \emph{ontology} $\mo$ \emph{is consistent} iff it has a model. A \emph{concept} $A$ (\emph{role} $r$) \emph{is satisfiable} w.r.t\ $\mo$ iff there is a model $\mi$ of $\mo$ with $A^\mi \neq \emptyset$ ($r^\mi \neq \emptyset$). An \emph{ontology} $\mo$ \emph{is coherent} iff all concepts and roles occurring in $\mo$ are satisfiable. An \emph{axiom $\alpha$ is entailed by $\mo$} iff $\alpha$ is true in all models $\mi$ of $\mo$. For a set of axioms $X$ we write $\mo \models X$ as a shorthand for $\mo \models \alpha$ for all $\alpha\in X$. 

Usually description logic systems provide sound and complete reasoning services to their users. Besides \emph{verification of coherency and consistency of $\mo$} and \emph{satisfiability checking of concepts}, reasoner tasks include classification and realization. \emph{Classification} 
%is a subsumption algorithm that 
determines, for each concept name $A$ occurring in $\mo$, most specific (general) concepts that subsume (are subsumed by) $A$. A concept $A$ subsumes (is subsumed by) a concept $B$ iff $\mo \models B \sqsubseteq A$ ($\mo \models A \sqsubseteq B$).
%$B^\mi \subseteq A^\mi$ ($A^\mi \subseteq B^\mi$) in every model $\mi$ of $\mo$. 
Classification is employed to build a taxonomy of concepts in $\mo$.
%hierarchy of concepts according to their generality.
\emph{Realization}, given an individual name $x$ occurring in $\mo$ and a given set of concepts in $\mo$ (usually all concepts in $\mo$), computes the most specific concepts $A_1,\dots,A_n$ from the set such that $\mo\models A_i(x)$ for all $i=1,\dots,n$. The most specific concepts are those that are minimal w.r.t.\ the subsumption ordering $\sqsubseteq$. 
%Note, when we speak of entailments below, we address (only) the output computed by the classification and realization services of a DL-reasoner.
%DLs are fragments of first-order logic.....
%Note, wherever we will speak of \emph{entailment computation} we address the classification and realization reasoning tasks, i.e.\ the extraction of direct sub-concept ($A \sqsubseteq B$) and direct concept assertion ($A(a)$) entailments.

\begin{example}\label{example:FOL_to_DL}
The example KB given in the Introduction (Chapter~\ref{chap:intro}) can be equivalently represented in DL (cf.\ Remark~\ref{rem:used_logics}) as follows:
\begin{align}
\label{ex0:dl:s1}Res &\equiv \forall writes.Paper \\
\label{ex0:dl:s2}\exists writes.\top &\sqsubseteq Res \\
\label{ex0:dl:s3}Secr &\sqsubseteq Gen \\
\label{ex0:dl:s4}Gen &\sqsubseteq \lnot Res \\
\label{ex0:dl:s5}Secr&(pam)
\end{align}
where $Res$ is the concept symbol with equivalent meaning as the predicate symbol $res$, the role symbol $writes$ corresponds to the equally named binary predicate, $Paper$ to $paper$, and so on. Notice that axiom~\ref{ex0:dl:s2} states that the domain of $writes$ is $Res$. \qed
\end{example}

\section[Notational Remarks]{Notational Remarks\footnote{\label{foot:notational_conventions}These conventions apply to Parts~\ref{part:Prolog}-\ref{part:IterativeDiagnosisComputation} and Part~\ref{part:Epilog}. Each of the Parts~\ref{part:JWS}-\ref{part:DX} is self-contained w.r.t.\ the used notation.}}
\label{sec:NotationalRemarks}

\noindent\emph{General Notational Conventions.} Throughout this work, the nomenclature given by Table~\ref{tab:abbreviations} is used (many of the designators in the table will be explained later in this work). We will mainly refer to an ontology by the term \emph{KB}. 
%
% refer to axioms (because an axiom w.r.t.\ an ontology is analogue to a formula w.r.t.\ a KB).
 %
%A \emph{KB} $\mo$ is regarded as a set $\setof{\tax_1,\dots,\tax_n}$ which is to be understood as the conjunction $\bigwedge_{i=1}^{n} \tax_i$ where $\tax_i, i\in\setof{1,\dots,n}$ is called a logical formula (e.g.\ \cite{kreuzer2006}), a well-formed formula (e.g.\ \cite{chang1973}), a (logical) sentence or axiom (e.g.\ \cite{russellnorvig2003}) and axiom in most of the description logic literature (e.g.\ \cite{Baader2007}). 
%In the sequel we mainly refer to an ontology by the term \emph{KB} and mainly use simply the term \emph{formula} to refer to the elements of a KB (ontology).

In order to make a clear distinction between scalars and functions, we denote all scalars $g$ by $g$ and all functions $g$ by $g()$. If an ordered list occurs in a set operation, then this list is interpreted as a (non-ordered) set. For example, let $L := [1,3,4,2]$ be an ordered list; then $L \cap \setof{1,2,3}$ yields the \emph{set} $\setof{1,2,3}$.\\

\noindent\emph{Notational Convention for PL (cf.\ \cite{russellnorvig2003}).}
%The following notational convention will be used for PL: 
We use uppercase letters $A,B,\dots$ to denote atoms and the standard logical connectives to build PL formulas from atoms. The operator precedence we use is $\lnot$, $\land$, $\lor$, $\rightarrow$, $\leftrightarrow$, from highest to lowest. Given a PL KB $\mo$ and a PL formula $\tax$, we call $\widetilde{\mo}$ and $\widetilde{\tax}$ the \emph{signature of $\mo$} and the \emph{signature of $\tax$}, respectively. The former comprises all atoms occurring in $\mo$ and the latter all atoms occurring in $\tax$.\\

\noindent\emph{Notational Convention for FOL (cf.\ \cite{Ceri1989a}).}
%The following notational convention will be used for FOL (cf.\ \cite{Ceri1989a}): 
Variables are denoted by uppercase letters; constants and predicate symbols are denoted by strings beginning with a lowercase letter\footnote{We do not use any function symbols throughout this work.}. Recalling the example KB given in Chapter~\ref{chap:intro}, $X, Y$ are variables, $pam$ is a constant and $res$, $writes$, $paper$, $secr$ and $gen$ are predicate symbols. FOL formulas are built from the standard logical connectives described for PL above. The operator precedence we use for FOL formulas is the same as stated above\footnote{We do not use equality $=$ in FOL formulas throughout this work.}. The precedence of quantifiers $\forall$, $\exists$ is such that a quantifier outside of any parenthesized expression holds over everything to the right of it; if occurring in a parenthesized expression, a quantifier holds over everything to the right of it within this expression. For example, 
$\forall X \mathit{prof}(X) \rightarrow \exists Y secr(Y)$ is equivalent to $(\forall X (\mathit{prof}(X) \rightarrow (\exists Y (secr(Y)))))$ (i.e.\ ``for each professor there is at least one secretary'') and not to $(\forall X \mathit{prof}(X)) \rightarrow \exists Y secr(Y)$ (i.e.\ ``if everybody is a professor, then there is at least one secretary''). 

Given a FOL KB $\mo$ and a FOL formula $\tax$, we call $\widetilde{\mo}$ and $\widetilde{\tax}$ the \emph{signature of $\mo$} and the \emph{signature of $\tax$}, respectively. The former comprises all predicate, function and constant symbols occurring in $\mo$ and the latter all predicate, function and constant symbols occurring in $\tax$. The signature of the example KB given in Chapter~\ref{chap:intro} is $\setof{res, writes, paper, secr, gen, pam}$ and the signature of formula~\ref{ex0:s2} of this KB is $\setof{writes,res}$.\vspace{7pt}

%\noindent\emph{Remark Regarding FOL.} 
\begin{remark}\label{rem:def_incoherency_FOL}
By analogy with the definition of coherency in DL (see Section~\ref{sec:DL}), we call a FOL KB $\mo$ \emph{incoherent} iff $\mo \models \forall X_1,\dots,X_k\, \lnot p(X_1,\dots,X_k)$ for some $k$-place predicate symbol $p$ in the signature of $\mo$ where $k \geq 1$.\qed
\end{remark}
%
%$\forall X prof(X) \rightarrow \exists Y course(Y) \land holds(X,Y)$ is equivalent to $(\forall X (prof(X) \rightarrow (\exists Y (course(Y) \land holds(X,Y)))))$ (i.e.\ ``for each professor there is at least one course he/she holds'') and not to $(\forall X p(X)) \rightarrow \exists Y q(Y) \land r(X,Y)$ (i.e.\ ``'') 
%$\forall X p(X) \rightarrow \exists Y q(Y)$ is equivalent to $\forall X (p(X) \rightarrow \exists Y q(Y))$ () 
%and is not the same formula as $(\forall X p(X)) \rightarrow \exists Y q(Y)$.
%% , e.g.\ $x$, $pam$ are constants, functions by lowercase letters $f,g$

\begin{remark}\label{rem:entailment_computation_finite_types_of_entailments}
We want to point out that whenever we will speak of \emph{entailment computation} we address the invocation of a sound reasoning service that is guaranteed to terminate after \emph{finite} execution time and returns a \emph{finite} number of entailments for any KB given as input (cf.\ Remark~\ref{rem:used_logics}). Similarly, when we say that \emph{all entailments} of a KB are computed, we always refer to a \emph{finite} set of entailments of certain types output by such a reasoning service. Examples of such entailment types regarding DL are the (a)~classification and (b)~realization entailments, by which we mean (a)~all the subsumption relationships between concept names appearing in the KB, i.e.\ entailments of the form $C_1 \sqsubseteq C_2$ for concept names $C_1,C_2 \in \widetilde{\mo}$ and (b)~all the concept names instantiated by a given individual for all individuals appearing in the KB, i.e.\ entailments of the form $C(a)$ for concepts names $C \in \widetilde{\mo}$ and individual names $a \in \widetilde{\mo}$.\qed
% where $$ where both (a) and (b) can be extracted from the KB's classification.\qed
% the extraction of direct sub-concept ($A \sqsubseteq B$) and direct concept assertion ($A(a)$) entailments.
\end{remark}

%\begin{table}[h]
%\small
%\centering
%\rowcolors[]{2}{gray!8}{gray!16}
%\begin{tabular}{p{0.23\textwidth} p{0.33\textwidth} p{0.33\textwidth}}
%%input $X$ to \textsc{prune}/\textsc{pruneQdup} & $\mathsf{nd.cs}$ after execution of \textsc{prune}/\textsc{pruneQdup} \\
%\rowcolor{gray!40}
%\toprule\addlinespace[0pt] 
%&\textsc{staticHS} & \textsc{dynamicHS} \\
%\addlinespace[0pt]\bottomrule 
%\end{tabular}
%\caption{Comparison: \textsc{staticHS} versus \textsc{dynamicHS}.}
%\label{tab:comparison_static_vs_dynamic}
%\vspace{-10pt}
%\end{table}

\newgeometry{margin=2cm}

\renewcommand{\arraystretch}{1.4}
\begin{table*}[!htbp]
\normalsize
	\centering
	\rowcolors[]{2}{gray!8}{gray!16}
		\begin{tabular}{ll}
		\rowcolor{gray!40}
		\toprule\addlinespace[0pt] 
			Symbol & Meaning \\ \hline
			$2^X$ & the powerset of $X$ where $X$ is a set \\
			$U_X$ & the union of all elements in $X$ where $X$ is a set of sets \\
			%$[a_1,\dots,a_n]$ & an ordered list of the elements $a_1,\dots,a_n$ \\
			$\mathcal{L}$ & a (monotonic, idempotent, extensive) logical knowledge representation language \\
			%$\mathsf{A, B},\dots$ & a concept in DL \\
			$\mo_{(i)}$ & a (faulty) KB (optionally with an index)\\
			$\tax_{(i)}$ & a formula in a KB (an axiom in an ontology)\\
			$\mb_{(i)}$ & a (correct) background KB (optionally with an index)\\
			$\Tp$ & the set of positive test cases (each test case is a set of logical formulas) \\
			$\tp_{(i)}$ & a positive test case (optionally with an index) \\
			$\Tn$ & the set of negative test cases (each test case is a set of logical formulas) \\
			$\tn_{(i)}$ & a negative test case (optionally with an index) \\
			$\RQ$ & the set of requirements to the correct KB \\
			$\langle\mo,\mb,\Tp,\Tn\rangle_\RQ$ & a diagnosis problem instance (DPI) \\
			$\allD_{DPI}$ & the set of all diagnoses w.r.t.\ the DPI $DPI$ \\
			$\minD_{DPI}$ & the set of minimal diagnoses w.r.t.\ the DPI $DPI$ \\
			$\md_{(i)}$ & a (minimal) diagnosis (optionally with an index) \\
			$\dt$ & the true diagnosis \\
			$\allC_{DPI}$ & the set of all conflict sets w.r.t.\ the DPI $DPI$ \\
			$\minC_{DPI}$ & the set of minimal conflict sets w.r.t.\ the DPI $DPI$ \\
			$\mc_{(i)}$ & a (minimal) conflict set (optionally with an index) \\
			$\Queue$ & an ordered queue of open nodes in a hitting set tree algorithm \\
			$\mathsf{n}_{(i)},\mathsf{nd}_{(i)},\mathsf{node}_{(i)}$ & nodes in a hitting set tree algorithm (optionally with an index) \\
			%\multirow{3}{*}{$[a_1,\dots,a_n]$} %& \multirow{3}{*}{$[a_1,\dots,a_n]$} \\
			%$[a_1,\dots,a_n]$	
														& context-dependent (will be clear from the context): \\ 
			\cellcolor{gray!8}		&\cellcolor{gray!8}(1)~an ordered list of the elements $a_1,\dots,a_n$ or\\
			\multirow{-3}{*}{$[a_1,\dots,a_n]$}	& (2)~a (non-ordered) minimal diagnosis comprising formulas $a_1,\dots,a_n$ \\
			%$\tuple{a_1,\dots,a_n}$ 
												& context-dependent (will be clear from the context): \\ 
			\cellcolor{gray!16}	&\cellcolor{gray!16} (1)~a tuple of elements $a_1,\dots,a_n$ or\\
			\multirow{-3}{*}{$\tuple{a_1,\dots,a_n}$}	& (2)~a (non-ordered) minimal conflict set comprising formulas $a_1,\dots,a_n$ \\
			$u$ & the user interacting with the debugging system \\	
			$u()$ & the (user) function that maps queries to answers \\	
			$Q_{(i)}$ & a query (optionally with an index) \\
			$\mQ_{\mD,DPI}$ & the set of all queries w.r.t.\ the leading diagnoses $\mD$ and the DPI $DPI$ \\
			$\Pt(Q)$ & the q-partition of the query $Q$ (abbreviated form) \\
			$\tuple{\dx{}(Q),\dnx{}(Q),\dz{}(Q)}$ & the q-partition of the query $Q$ (written-out form) \\
			$\EX{\md}{DPI}$ & the set of all extensions w.r.t.\ a diagnosis $\md$ and a DPI $DPI$ \\
			$\SO_{DPI}$ & the set of all solution KBs w.r.t.\ the DPI $DPI$ \\
			$\SO^{\max}_{DPI}$ & the set of all maximal solution KBs w.r.t.\ the DPI $DPI$ \\
		\addlinespace[0pt]\bottomrule 	 
		\end{tabular}
		\caption[Symbols and Abbreviations]{Symbols and abbreviations used throughout this work (cf.\ footnote~\ref{foot:notational_conventions}).}
		\label{tab:abbreviations}
\end{table*}

\restoregeometry
%Throughout this work we will adhere to the following notation: If $X$ be a set of sets, the $U_X$ denotes the union and $I_X$ the intersection of all sets in $X$. 

\chapter[Knowledge Base Debugging]{Knowledge Base Debugging%
\chaptermark{KB Debugging}}
\chaptermark{KB Debugging}
\label{chap:KBDebugging}

%%%%%
%\chapter{Knowledge Base Debugging}
%%\label{sec:OntologyDebugging}
%\label{chap:KBDebugging}
%%%%%
%First we provide an informal introduction to ontology debugging, particularly addressing readers unfamiliar with the topic. Later we introduce precise formalizations. We assume the reader to be familiar with description logics~\cite{DLHandbook}.
%
KB debugging can be seen as a test-driven procedure comparable to test-driven software development and debugging, where test cases are specified to restrict the possible faults until the user detects the actual fault manually or there is only one (highly probable) fault remaining which is in line with the specified test cases. In this chapter, we want to study the theory of (non-interactive) KB debugging, present and discuss mechanisms that can be employed for the debugging of KBs and reveal drawbacks of such systems. In (non-interactive) KB debugging we assume \emph{test cases fixed during the debugging procedure}. That is, a user might specify a set of test cases offline, run a debugging system and investigate the output solution(s). In case no satisfactory solution has been returned, some additional test cases might be defined offline before the debugger might be invoked again.
 
The inputs to a KB debugging problem can be characterized as follows:
%is specified by the following inputs:
%The problem addressed by ontology debugging is the following: 
Given is a KB $\mo$ and a KB $\mb$ (background knowledge), both formulated over some logic $\mathcal{L}$ complying with the conditions~\ref{logic:cond1} -- \ref{logic:cond4} given in Chapter~\ref{chap:basics}.  
%$\mb \cap \mo = \emptyset$, 
All formulas in $\mb$ are considered to be correct and all formulas in $\mo$ are considered potentially faulty. $\mo \cup \mb$ does not meet postulated requirements 
$\RQ$ where $\setof{\text{consistency}} \subseteq \RQ \subseteq \setof{\text{coherency, consistency}}$
%$\RQ \supseteq \setof{\text{consistency}}$,\footnote{We assume consistency a minimal requirement to a solution ontology provided by a debugging system, as inconsistency makes an ontology completely useless from the semantic point of view.} e.g. $\RQ=\{\text{coherence},\text{consistency}\}$
or does not feature desired semantic properties, called test cases.\footnote{We assume consistency a minimal requirement to a solution KB provided by a debugging system, as inconsistency makes a KB completely useless from the semantic point of view.} Positive test cases (aggregated in the set $\Tp$) correspond to desired entailments and negative test cases ($\Tn$) represent undesired entailments of the correct (repaired) KB (along with the background KB $\mb$). \label{etc:test_cases_are_sets_or_conjuntions_of_formulas} 
%Each test case $\tp \in \Tp$ and $\tn \in \Tn$ is \emph{a set of} logical sentences over $\mathcal{L}$. 
%Each positive ($\tp \in \Tp$) or negative ($\tn\in\Tn$) test case is a set of formulas over $\mathcal{L}$. 
Each test case $\tp \in \Tp$ and $\tn \in \Tn$ is \emph{a set of} logical formulas over $\mathcal{L}$. The meaning of a positive test case $\tp \in \Tp$ is that the correct KB integrated with $\mb$ must entail each formula (or the conjunction of formulas) in $\tp$, whereas a negative test case $\tn \in \Tn$ signalizes that some formula (or the conjunction of formulas) in $\tn$ must not be entailed by the correct KB integrated with $\mb$. 
\begin{remark}\label{rem:entailments_as_sets_of_formulas}
In the sequel, we will write $\mo \models X$ for some set of formulas $X$ to denote that $\mo \models \tax$ for all $\tax \in X$ and $\mo \not\models X$ to state that $\mo \not\models \tax$ for some $\tax \in X$.\qed
\end{remark} 
%
%$\Tp$ (positive test cases) and as a set of undesired entailments $\Tn$ (negative test cases). Thereby, each test case $\tp \in \Tp$ and $\tn \in \Tn$ is \emph{a set of} logical sentences over $\mathcal{L}$. 
%%A test case $t = \setof{s_1,\dots,s_k}$ is associated with the logical conjunction of the comprised sentences $s_{i(i=1,\dots,k)}$. 
%A positive (negative) test case $t = \setof{s_1,\dots,s_k}$ can be understood as the conjunction $s_1 \land \dots \land s_k$ and means that each (at least one) of the comprised sentences $s_{i(i=1,\dots,k)}$ must (not) be entailed by the desired ontology. 
%So, $\Tp$ and $\Tn$ are sets of sets of logical sentences. 
%We call the tuple $\langle\mo,\mb,\Tp,\Tn\rangle_\RQ$ a \emph{diagnosis problem instance (DPI) over $\mathcal{L}$} or simply a \emph{diagnosis problem instance (DPI)} if the used DL is clear from the context or not explicitly relevant.
%The task is then to find a solution ontology w.r.t. the given DPI.
The described inputs to the KB debugging problem are captured by the notion of a diagnosis problem instance:
\begin{definition}[Diagnosis Problem Instance]\label{def:dpi}
Let 
\begin{itemize}
	%\item $\mo, \mb$ be KBs over $\mathcal{L}$ with $\mo \cap \mb = \emptyset$,
	\item $\mo$ be a KB over $\mathcal{L}$,
	\item $\Tp, \Tn$ sets including sets of formulas over $\mathcal{L}$,
	\item $\setof{\text{consistency}}\subseteq \RQ \subseteq \setof{\text{coherency, consistency}}$, 
	\item $\mb$ be a KB over $\mathcal{L}$ such that $\mo \cap \mb = \emptyset$ and $\mb$ satisfies all requirements $r \in \RQ$,
	\item the cardinality of all sets $\mo$, $\mb$, $\Tp$, $\Tn$ be finite.
\end{itemize}
%(1)~$\mo, \mb$ be KBs over $\mathcal{L}$ with $\mo \cap \mb = \emptyset$, (2)~$\Tp, \Tn$ sets including sets of formulas over $\mathcal{L}$ and (3)~$\setof{\text{coherence, consistency}} \supseteq \RQ \supseteq \setof{\text{consistency}}$. Further, let the cardinality of the sets $\mo, \mb,\Tp,\Tn$ be finite. 
Then we call the tuple $\langle\mo,\mb,\Tp,\Tn\rangle_\RQ$ a \emph{diagnosis problem instance (DPI) over $\mathcal{L}$}.\footnote{
%We will omit the ``over $\mathcal{L}$'' in the following for brevity and since $\mathcal{L}$ always stands for any logic compliant with the conditions in Chapter~\ref{chap:basics}.
In the following we will often call a DPI over $\mathcal{L}$ simply a DPI for brevity and since the concrete logic will not be relevant in our theoretical analyses as long as it is compliant with the conditions \ref{logic:cond1} -- \ref{logic:cond4} given in Chapter~\ref{chap:basics}. Nevertheless we will mean exactly the logic over which a particular DPI is defined when we use the designator $\mathcal{L}$.
}
\end{definition}
%used DL is clear or not explicitly relevant in a certain context. 
Note that, for now, we do not make any assumptions about the contents of the sets $\mo$, $ \mb$, $\Tp$ and $\Tn$ that go beyond Definition~\ref{def:dpi}.
%, as long as all sentences comprised by these sets are formulated over one and the same language $\mathcal{L}$. 
So, it might be well the case, for example, to specify a DPI according to Definition~\ref{def:dpi} for which there are no solutions or for which only trivial solutions exist. Later on, we will discuss properties a DPI must fulfill to guarantee existence of solutions for it.

We define a solution KB for a DPI as follows:
%A solution ontology w.r.t.\ a DPI is defined as follows:
\begin{definition}[Solution KB]\label{def:target_ont} Let $\langle\mo,\mb,\Tp,\Tn\rangle_\RQ$ be a DPI. Then a KB $\ot$ is called \emph{solution KB w.r.t. $\langle\mo,\mb,\Tp,\Tn\rangle_\RQ$}, written as $\ot \in \SO_{\langle\mo,\mb,\Tp,\Tn\rangle_\RQ}$, iff all the following conditions hold:
%\vspace{-6pt}
\begin{eqnarray}
		 \forall \, r  \in \RQ&:& \;\ot \cup \mb \,\text{ fulfills }\, r  \label{e:1} \\ 
		 \forall \,\tp \in \Tp&:& \;\ot \cup \mb \,\models\, \tp					\label{e:2} \\ 
		 \forall \,\tn \in \Tn&:& \;\ot \cup \mb \,\not\models\, \tn .		\label{e:3}  
		 %\not\exists\,\mo'&:& \;\mo'\cap\mo \supset \ot \cap \mo \,\text{ and } \mo' \nonumber \\
		 %&&\text{ fulfills conditions (\ref{e:1}), (\ref{e:2}) and (\ref{e:3})}.				
\end{eqnarray}
A solution KB $\ot$ w.r.t. a DPI is called \emph{maximal}, written as $\ot \in \SO^{\max}_{\langle\mo,\mb,\Tp,\Tn\rangle_\RQ}$, iff there is no solution KB $\mo'$ such that $\mo' \cap \mo \supset \ot\cap\mo$. 
%there is no $\mo'$ such that $\mo'\cap\mo \supset \ot \cap \mo$ and $\mo'$ fulfills conditions (\ref{e:1}), (\ref{e:2}) and (\ref{e:3}).
%\vspace{-14pt}
\end{definition}

Now, the problem of KB debugging can be formalized:\vspace{3pt}

\noindent\fcolorbox{black}{light-gray1}{\parbox[c][2em][c]{0.975\linewidth}{\vspace{-4pt}
\begin{prob_def}[KB Debugging] \label{prob_def:onto_debug}
Given a DPI $\langle\mo,\mb,\Tp,\Tn\rangle_\RQ$, 
%the task of ontology debugging is to 
find a solution KB w.r.t.\ $\langle\mo,\mb,\Tp,\Tn\rangle_\RQ$.
\end{prob_def}\vspace{-4pt}
}}

\vspace{3pt}
%A solution ontology w.r.t.\ a DPI is defined as follows:
%\begin{definition}[Solution Ontology]\label{def:target_ont} Let $\langle\mo,\mb,\Tp,\Tn\rangle_\RQ$ be a DPI. Then an ontology $\ot$ is called \emph{solution ontology w.r.t. $\langle\mo,\mb,\Tp,\Tn\rangle_\RQ$}, written as $\ot \in \SO_{\langle\mo,\mb,\Tp,\Tn\rangle_\RQ}$, iff all the following conditions hold:
%%\vspace{-6pt}
%\begin{eqnarray}
		 %\forall \, r  \in \RQ&:& \;\ot \cup \mb \,\text{ fulfills }\, r  \label{e:1} \\ 
		 %\forall \,\tp \in \Tp&:& \;\ot \cup \mb \,\models\, \tp					\label{e:2} \\ 
		 %\forall \,\tn \in \Tn&:& \;\ot \cup \mb \,\not\models\, \tn .		\label{e:3}  
		 %%\not\exists\,\mo'&:& \;\mo'\cap\mo \supset \ot \cap \mo \,\text{ and } \mo' \nonumber \\
		 %%&&\text{ fulfills conditions (\ref{e:1}), (\ref{e:2}) and (\ref{e:3})}.				
%\end{eqnarray}
%A solution ontology $\ot$ w.r.t. a DPI is called \emph{maximal}, written as $\ot \in \SO^{\max}_{\langle\mo,\mb,\Tp,\Tn\rangle_\RQ}$, iff there is no solution ontology $\mo'$ such that $\mo' \cap \mo \supset \ot\cap\mo$. 
%%there is no $\mo'$ such that $\mo'\cap\mo \supset \ot \cap \mo$ and $\mo'$ fulfills conditions (\ref{e:1}), (\ref{e:2}) and (\ref{e:3}).
%%\vspace{-14pt}
%\end{definition}
Note that basically any KB $\ot$ that meets conditions~(\ref{e:1}) - (\ref{e:3}) is a solution KB in the sense of Definition~\ref{def:target_ont}. Hence, $\ot$ does not even need to have a non-empty intersection with $\mo$. Only the postulation of maximality of a solution KB (as detailed later in Section~\ref{sec:MinimallyInvasiveOntologyDebugging}) establishes a relationship to the given KB $\mo$.

\begin{remark}\label{rem:reduce_conditions_of_dpi_def}
Let $\mo' := \mo\cup\mb\cup U_{\Tp}$. Then, conditions~(\ref{e:1}) - (\ref{e:3}) can be reduced to conditions~(\ref{e:2}) and (\ref{e:3}) if 
\begin{itemize}
	%\item $\Tn := \Tn \cup \setof{\setof{\top \sqsubseteq \bot}}$ 
%%is added to $\Tn$ 
%given $\RQ = \setof{\mbox{consistency}}$, or, respectively,
	\item $\Tn := \Tn \cup \setof{\setof{\false}}$ 
%is added to $\Tn$ 
given $\RQ = \setof{\mbox{consistency}}$ or
	%\item $\Tn := \Tn \cup \setof{\setof{A_i \sqsubseteq \bot}\,|\,A_i\in N_C} \cup \setof{\setof{\top \sqsubseteq \bot}}$ 
%%is added to $\Tn$ 
%in case $\RQ = \setof{\mbox{consistency, coherency}}$.\footnote{Remember that $N_C$ is the subset of the signature of $\mathcal{L}$ that contains all concept names (cf.\ Section~\ref{sec:DL}).}
	\item $\Tn := \Tn \cup \{\{\forall X_1,\dots,X_k\, p(X_1,\dots,X_k) \rightarrow\false\}\,|\, p \mbox{ is $k$-place predicate}$ \\
	$ \mbox{symbol in } \widetilde{\mo'}, k \geq 1\} \cup \setof{\setof{\false}}$ 
%is added to $\Tn$ 
in case $\RQ = \setof{\mbox{consistency, coherency}}$.
\end{itemize}
This holds because a KB $\mo$ is inconsistent iff $\mo \models \setof{\false}$ and $\mo$ is incoherent iff some predicate symbol in $\mo'$ must be $\false$ for any instantiation. Notice that the latter must hold for all predicate symbols in $\mo'$ and not only in $\mo$ (see Example~\ref{example:illustration_of_rem:reduce_conditions_of_dpi_def}). For PL and DL, the definitions of $\Tn$ are analogous (cf.\ Chapter~\ref{chap:basics}), but for PL coherency is not defined wherefore only the first bullet is relevant for PL. In what follows we will stick to the more explicit characterization of a solution KB given by Definition~\ref{def:target_ont}.\qed
\end{remark}
\begin{example}\label{example:illustration_of_rem:reduce_conditions_of_dpi_def}
Let a DL DPI be defined as 
\begin{align*}
\mo &:= \setof{B \sqsubseteq C} \\
\mb &:= \setof{A \sqsubseteq B, C \sqsubseteq \lnot A} \\
\Tp &:= \emptyset \\
\Tn &:= \emptyset \\
\RQ &:= \setof{\mbox{coherency, consistency}}
\end{align*}
Then, $\widetilde{\mo} = \setof{B,C}$, but there is some concept $A \notin \widetilde{\mo}$, but $A \in \widetilde{\mo'}$, which is unsatisfiable w.r.t.\ $\mo \cup \mb$. Since we want a solution KB \emph{integrated with $\mb$} to meet the conditions~(\ref{e:1}) - (\ref{e:3}), $\mo$ is not a solution KB w.r.t.\ $\langle\mo,\mb,\Tp,\Tn\rangle_\RQ$ despite the fact that it is perfectly consistent and coherent as an isolated KB.\qed 
%$\mo = \setof{B \sqsubseteq \lnot C}$ and $\mb := \setof{A \sqsubseteq \lnot B, C \sqsubseteq \lnot A}$
\end{example}

Whereas the definition of a solution KB refers to the desired properties of the output of a KB debugging system, the following definition can be seen as a characterization of KBs provided as an input to a KB debugger. 
If a KB is valid w.r.t.\ the background knowledge, the requirements and the test cases, then finding a solution KB w.r.t.\ the DPI is trivial. 
Otherwise, obtaining a solution KB from it involves modification of the input KB and subsequent addition of suitable formulas. Usually, the KB $\mo$ part of the DPI given as an input to a debugger is assumed to be invalid w.r.t.\ this DPI.
%
%So, if an (input) KB is valid, then a solution KB can be directly built from it. That is, the input KB only needs to be extended by the postulated entailments $U_\Tp$. In case an (input) KB is invalid, obtaining a solution KB from it involves modification of the KB and subsequent addition of formulas $U_\Tp$. Usually, the KB $\mo$ part of the DPI given as an input to a debugger is assumed to be invalid w.r.t.\ this DPI.
%We say that an ontology $\mo$ \emph{contradicts the positive test cases $\Tp$ w.r.t.\ $\langle\mo,\mb,\Tp,\Tn\rangle_\RQ$} iff some $r\in\RQ$ is violated by $\mo \cup \mb \cup U_\Tp$ or $\mo \cup \mb \cup U_\Tp \models \tn$ for some $\tn \in \Tn$, where $U_\Tp := \bigcup_{\tp\in\Tp} \tp$. 
\begin{definition}[Valid KB]\label{def:valid_onto}
Let $\langle\mo,\mb,\Tp,\Tn\rangle_\RQ$ be a DPI.
% and $U_\Tp := \bigcup_{\tp\in\Tp} \tp$. 
Then, we say that a KB $\mo'$ is \emph{valid w.r.t.\ $\langle\cdot, \mb, \Tp, \Tn \rangle_{\RQ}$} iff 
$\mo' \cup \mb \cup U_\Tp$ does not violate any $r\in\RQ$ and does not entail any $\tn \in \Tn$. A KB is said to be \emph{invalid (or faulty) w.r.t.\ $\langle\cdot,\mb,\Tp,\Tn\rangle_\RQ$} iff it is not valid w.r.t.\ $\langle\cdot,\mb,\Tp,\Tn\rangle_\RQ$.\footnote{
%We use the dot $\cdot$ in $\langle\cdot,\mb,\Tp,\Tn\rangle_\RQ$ to signalize that only the elements $\mb$, $\Tp$, $\Tn$ as well as $\RQ$ are relevant to determine the validity of a KB. 
It would be more precise to call a KB valid \emph{w.r.t.\ the elements $\mb$, $\Tp$, $\Tn$, $\RQ$ of a DPI}. Though, for brevity, we stick to the presented notation where the dot $\cdot$ in $\langle\cdot,\mb,\Tp,\Tn\rangle_\RQ$ signalizes the irrelevance of the first element $\mo$ of a DPI $\langle\mo,\mb,\Tp,\Tn\rangle_\RQ$ for determining validity of a KB $\mo'$ w.r.t.\ this DPI.} 
%ontology which is not valid w.r.t.\ $\langle\cdot,\mb,\Tp,\Tn\rangle_\RQ$ is said to be \emph{faulty (or non-valid) w.r.t. $\langle\cdot,\mb,\Tp,\Tn\rangle_\RQ$}.
\end{definition} 
%and $\mo$ \emph{contradicts a negative test case} $n\in\Tn$ iff $\mo \cup \mb \models n$. 
Intuitively, if a KB $\mo$ is faulty w.r.t.\ $\langle\cdot,\mb,\Tp,\Tn\rangle_\RQ$, then there is at least one incorrect formula in $\mo$ that needs to be corrected or deleted; 
if a KB $\mo$ is valid w.r.t.\ $\langle\cdot,\mb,\Tp,\Tn\rangle_\RQ$, a solution KB can be \emph{directly} obtained by simply extending $\mo$ by the set $U_\Tp$ of all sentences comprised in positive test cases. 
%can be added to $\mo$ without causing $\mo \cup U_\Tp$ to violate any conditions for a solution ontology. 
Note, however, that $\mo$ being valid w.r.t.\ $\langle\cdot,\mb,\Tp,\Tn\rangle_\RQ$ does not necessarily mean that $\mo \cup \mb$ entails any $\tp \in \Tp$. 
\begin{proposition}\label{prop:validonto_targetonto}
Let $\langle\mo,\mb,\Tp,\Tn\rangle_\RQ$ be a DPI. Then, $\mo' \cup U_\Tp \in \SO_{\langle\mo,\mb,\Tp,\Tn\rangle_\RQ}$ 
%is a solution ontology w.r.t.\ $\langle\mo,\mb,\Tp,\Tn\rangle_\RQ$ 
iff $\mo'$ is valid w.r.t.\ $\langle\cdot,\mb,\Tp,\Tn\rangle_\RQ$.
\end{proposition}
\begin{proof}
``$\Rightarrow$'': If $\mo' \cup U_\Tp$ is a solution KB, then $\mo' \cup U_\Tp \cup \mb$ meets all $r\in\RQ$ as per condition~(\ref{e:1}) and does not entail any $\tn\in\Tn$ as per condition~(\ref{e:3}). Hence, $\mo'$ is valid w.r.t.\ $\langle\cdot,\mb,\Tp,\Tn\rangle_\RQ$.

``$\Leftarrow$'': If $\mo'$ is valid w.r.t.\ $\langle\cdot,\mb,\Tp,\Tn\rangle_\RQ$, then $(\mo' \cup U_\Tp) \cup \mb$ meets all $r\in\RQ$, i.e.\ meets condition~(\ref{e:1}). Moreover, $(\mo' \cup U_\Tp) \cup \mb \not\models \tn$ for all $\tn\in\Tn$, i.e.\ $(\mo' \cup U_\Tp) \cup \mb$ meets condition~(\ref{e:3}). By extensiveness of the used language $\mathcal{L}$, $(\mo' \cup U_\Tp) \cup \mb \models \tp$ for all $\tp\in\Tp$, i.e.\ condition~(\ref{e:2}) is fulfilled by $(\mo' \cup U_\Tp) \cup \mb$. Thus, $\mo' \cup U_\Tp$ is a solution KB.
\end{proof}
\begin{definition}[Extension]\label{def:extension}
Let $\langle\mo,\mb,\Tp,\Tn\rangle_\RQ$ be a DPI over $\mathcal{L}$ and $\mo' \subseteq \mo$. A set of formulas $\me$ over $\mathcal{L}$ is called an \emph{extension w.r.t.\ $\mo'$ and $\langle\mo,\mb,\Tp,\Tn\rangle_\RQ$}, written as $\me \in \EX{\mo'}{\langle\mo,\mb,\Tp,\Tn\rangle_\RQ}$, iff $(\mo\setminus\mo') \cup \me$ is a solution KB w.r.t.\ $\langle\mo,\mb,\Tp,\Tn\rangle_\RQ$.
\end{definition}
\begin{definition}[Diagnosis]\label{def:diagnosis}
Let $\langle\mo,\mb,\Tp,\Tn\rangle_\RQ$ be a DPI. A set of formulas $\md \subseteq \mo$ is called a \emph{diagnosis w.r.t.\ $\langle\mo,\mb,\Tp,\Tn\rangle_\RQ$}, written as $\md \in \allD_{\langle\mo,\mb,\Tp,\Tn\rangle_\RQ}$, iff there exists some $\me \in \EX{\md}{\langle\mo,\mb,\Tp,\Tn\rangle_\RQ}$, i.e.\ $(\mo\setminus\md)\cup \me$ is a solution KB w.r.t.\ $\langle\mo,\mb,\Tp,\Tn\rangle_\RQ$.
%$\EX{\md} \cap \md = \emptyset$ and 
%$(\mo\setminus\md)\cup \EX{\md}$ is a solution ontology w.r.t.\ $\langle\mo,\mb,\Tp,\Tn\rangle_\RQ$.

A diagnosis $\md$ w.r.t.\ $\langle\mo,\mb,\Tp,\Tn\rangle_\RQ$ is \emph{minimal}, written as $\md \in \minD_{\langle\mo,\mb,\Tp,\Tn\rangle_\RQ}$, iff there is no $\md' \subset \md$ such that 
$\md'$ is a diagnosis w.r.t.\ $\langle\mo,\mb,\Tp,\Tn\rangle_\RQ$. 
A diagnosis $\md$ w.r.t.\ $\langle\mo,\mb,\Tp,\Tn\rangle_\RQ$ is a \emph{minimum cardinality diagnosis w.r.t.\ $\langle\mo,\mb,\Tp,\Tn\rangle_\RQ$} iff there is no diagnosis $\md'$ w.r.t.\ $\langle\mo,\mb,\Tp,\Tn\rangle_\RQ$ such that $|\md'| < |\md|$.
%$\md' \in \allD_{\langle\mo,\mb,\Tp,\Tn\rangle_\RQ}$.
%The set of all (minimal) diagnoses w.r.t. a DPI is denoted by ($\minD_{\langle\mo,\mb,\Tp,\Tn\rangle_\RQ}$) $\allD_{\langle\mo,\mb,\Tp,\Tn\rangle_\RQ}$.
\end{definition}
\begin{proposition}\label{prop:validonto_diag}
Let $\langle\mo,\mb,\Tp,\Tn\rangle_\RQ$ be a DPI. 
%over some DL $\mathcal{L}$. 
Then, $\md \in \allD_{\langle\mo,\mb,\Tp,\Tn\rangle_\RQ}$ 
%is a diagnosis w.r.t.\ $\langle\mo,\mb,\Tp,\Tn\rangle_\RQ$} 
iff $\mo \setminus \md$ is valid w.r.t.\ $\langle\cdot,\mb,\Tp,\Tn\rangle_\RQ$. 
\end{proposition}
\begin{proof}
``$\Rightarrow$'': If $\md$ is a diagnosis w.r.t.\ $\langle\mo,\mb,\Tp$, $\Tn\rangle_\RQ$, there is some extension $\me$ w.r.t.\ $\md$ and $\langle\mo,\mb,\Tp$, $\Tn\rangle_\RQ$ which implies that $(\mo \setminus \md) \cup \me$ is a solution KB w.r.t.\ $\langle\mo,\mb,\Tp,\Tn\rangle_\RQ$. Now, assume that $\mo \setminus \md$ is not valid w.r.t.\ $\langle\cdot,\mb,\Tp,\Tn\rangle_\RQ$. By Proposition~\ref{prop:validonto_targetonto}, this means that $(\mo \setminus \md) \cup U_\Tp$ is not a solution KB. Hence, $(\mo \setminus \md) \cup U_\Tp \cup \mb$ violates some $r\in\RQ$ or entails some $\tn\in\Tn$. As $(\mo \setminus \md) \cup \me$ is a solution KB, we have that $(\mo \setminus \md) \cup \me \cup \mb \models \tp$ for all $\tp\in\Tp$. So, by idempotency of $\mathcal{L}$, $(\mo \setminus \md) \cup \me \cup \mb \equiv (\mo \setminus \md) \cup \me \cup \mb \cup U_\Tp \supseteq (\mo \setminus \md) \cup U_\Tp \cup \mb$ which violates some $r\in\RQ$ or entails some $\tn\in\Tn$. By monotonicity of $\mathcal{L}$, $(\mo \setminus \md) \cup \me \cup \mb$ also violates some $r\in\RQ$ or entails some $\tn\in\Tn$ whereby $(\mo \setminus \md) \cup \me$ is not a solution KB which is a contradiction.

``$\Leftarrow$'': If $\mo \setminus \md$ is valid w.r.t.\ $\langle\cdot,\mb,\Tp,\Tn\rangle_\RQ$, then $(\mo \setminus \md) \cup \mb \cup U_\Tp$ does not violate any $r\in\RQ$ and does not entail any $\tn \in \Tn$. Since $(\mo \setminus \md) \cup \mb \cup U_\Tp$ also entails each positive test case $\tp \in \Tp$ by extensiveness of $\mathcal{L}$, we can conclude that $(\mo \setminus \md) \cup U_\Tp$ is a solution KB. By Definition~\ref{def:extension}, $U_\Tp \in \EX{\md}{\langle\mo,\mb,\Tp,\Tn\rangle_\RQ}$ and thus $\md$ is a diagnosis w.r.t.\ $\langle\mo,\mb,\Tp,\Tn\rangle_\RQ$.
\end{proof}
In other words, $\md$ is a diagnosis w.r.t.\ $\langle\mo,\mb,\Tp,\Tn\rangle_\RQ$ iff $(\mo \setminus \md) \cup \mb$ meets all requirements, i.e.\ consistency and/or coherency, as per condition~(\ref{e:1}),
% and does not contradict any positive or negative test cases as per conditions~(\ref{e:2}) and (\ref{e:3}). 
does not entail any negative test cases as per condition~(\ref{e:3}), and the positive test cases $\tp \in \Tp$ can be added to $(\mo \setminus \md) \cup \mb$ without violating any of the conditions~(\ref{e:1}) or (\ref{e:3}).

From a given DPI $\langle\mo,\mb,\Tp,\Tn\rangle_\RQ$, a solution KB $\ot$ can be obtained by a deletion and an expansion step. 
The deletion step involves the elimination of a diagnosis $\md \subseteq \mo$ from $\mo$. Note that, due to monotonicity of $\mathcal{L}$, only deletion (and not expansion) of the KB can effectuate a repair of inconsistencies, incoherencies and unwanted entailments. Note, if $\mo$ is already valid w.r.t.\ $\langle\cdot,\mb,\Tp,\Tn\rangle_\RQ$, then $\md$ can be set to $\emptyset$ and the deletion step can be omitted.
%there is a single minimal diagnosis $\md = \emptyset$ w.r.t.\ $\langle\mo,\mb,\Tp,\Tn\rangle_\RQ$ and the deletion step is obsolete.
%
The expansion step aims at the fulfillment of positive test cases $\Tp$, i.e.\ condition~(\ref{e:2}), which is not necessarily 
%given by only deleting formulas from a faulty KB $\mo$.
the case after the deletion step. In fact, some new logical sentences $\me \in \EX{\md}{\langle\mo,\mb,\Tp,\Tn\rangle_\RQ}$ may need to be added to $(\mo\setminus\md) \cup \mb$ to grant entailment of all positive test cases. 
\begin{corollary}\label{cor:diag_properties}
Let $\md$ be a diagnosis w.r.t.\ $\langle\mo,\mb,\Tp,\Tn\rangle_\RQ$. Then there is a set of logical sentences $\me\in\EX{\md}{\langle\mo,\mb,\Tp,\Tn\rangle_\RQ}$ over $\mathcal{L}$ such that:
\begin{eqnarray*}
		 \forall \, r  \in \RQ&:& \;(\mo\setminus\md) \cup \me \cup \mb \,\text{ fulfills }\, r   \\ 
		 \forall \,\tp \in \Tp&:& \;(\mo\setminus\md) \cup \me \cup \mb \,\models\, \tp					  \\ 
		 \forall \,\tn \in \Tn&:& \;(\mo\setminus\md) \cup \me \cup \mb \,\not\models\, \tn .		  
\end{eqnarray*} 
\end{corollary}
\begin{proof}
The proposition of the corollary is a direct consequence of Definition~\ref{def:target_ont} and Definition~\ref{def:diagnosis}.
\end{proof}
%%%%%%%%%%%%%%%%%%%%%%%%%%%%%%START CUT
%We call two DPIs $\langle\mo,\mb,\Tp,\Tn\rangle_\RQ$ and $\langle\mo',\mb',\Tp',\Tn'\rangle_\RQ$ equivalent iff each diagnosis w.r.t.\ $\langle\mo,\mb,\Tp,\Tn\rangle_\RQ$ is a diagnosis w.r.t.\ $\langle\mo',\mb',\Tp',\Tn'\rangle_\RQ$ and vice versa.
%\begin{proposition}\label{dpi_equiv}
%The DPI $\langle\mo,\mb,\Tp,\Tn\rangle_\RQ$ is equivalent to the DPI $\langle\mo \cup \Tp,\mb,\emptyset,\Tn\rangle_\RQ$.
%\end{proposition}
%\begin{proof}
%%
%\end{proof}
%%%%%%%%%%%%%%%%%%%%%%%%%%%%%%END CUT
From the point of view of a solution KB $\ot$ w.r.t.\ $\langle\mo,\mb,\Tp,\Tn\rangle_\RQ$, $\mo \setminus \ot$ is a diagnosis w.r.t.\ $\langle\mo,\mb,\Tp,\Tn\rangle_\RQ$ and $\ot \setminus \mo$ is one possible extension w.r.t.\ $\md$ and $\langle\mo,\mb,\Tp,\Tn\rangle_\RQ$.
\begin{proposition}\label{prop:targetonto_diag}
For each solution KB $\ot$ w.r.t.\ $\langle\mo,\mb,\Tp,\Tn\rangle_\RQ$ there is a diagnosis w.r.t.\ $\langle\mo,\mb,\Tp,\Tn\rangle_\RQ$ and an extension $\me$ w.r.t.\ $\md$ and $\langle\mo,\mb,\Tp,\Tn\rangle_\RQ$ such that $\ot = (\mo\setminus\md) \cup \me$ and $\me \cap \md = \emptyset$.
\end{proposition}
\begin{proof}
Let $\ot$ be a solution KB w.r.t. $\langle\mo,\mb,\Tp,\Tn\rangle_\RQ$. Then $\ot$ can be written as $\ot = (\mo \cap \ot) \cup (\ot \setminus \mo) = (\mo \setminus (\mo \setminus \ot)) \cup (\ot \setminus \mo)$. Let $\mo \setminus \ot =: \md$ and $\ot \setminus \mo =: \me$, then $\me \cap \md = \emptyset$. Further on, $\md \subseteq \mo$ holds and $\me$ is a set of logical sentences such that $\ot = (\mo\setminus\md)\cup \me \in \SO_{\langle\mo,\mb,\Tp,\Tn\rangle_\RQ}$. Therefore, $\md \in \allD_{\langle\mo,\mb,\Tp,\Tn\rangle_\RQ}$ and $\me \in \EX{\md}{\langle\mo,\mb,\Tp,\Tn\rangle_\RQ}$.
%Since $\ot$ is a solution KB, $\md$ is a diagnosis (for $\ot$) w.r.t. $\langle\mo,\mb,\Tp,\Tn\rangle_\RQ$ per Definition~\ref{def:diagnosis}. The vice versa direction is a direct consequence of Definition~\ref{def:diagnosis}. 
\end{proof}
\begin{corollary}
%Let $\langle\mo,\mb,\Tp,\Tn\rangle_\RQ$ be a DPI. Then 
The \mbox{(non-)}existence of a diagnosis w.r.t. $\langle\mo,\mb,\Tp,\Tn\rangle_\RQ$ is equivalent to the \mbox{(non-)}existence of a solution KB w.r.t. $\langle\mo,\mb,\Tp,\Tn\rangle_\RQ$.
\end{corollary}
\begin{proof}
Proposition~\ref{prop:targetonto_diag} shows that there is a diagnosis for each solution KB. By Definition~\ref{def:diagnosis}, there is also a solution KB for each diagnosis.
\end{proof}
%A diagnosis w.r.t. a DPI $\langle\mo,\mb,\Tp,\Tn\rangle_\RQ$ exists iff the background knowledge $\mb$ meets all requirements in $\RQ$ and does neither contradict any positive test case in $\Tp$ nor any negative test case in $\Tn$.
The next Proposition gives sufficient and necessary criteria for the existence of a solution, i.e.\ a diagnosis or a solution KB, respectively, for a given DPI.
%
%and only if the background knowledge $\mb$ together with the desired entailments $\tp \in \Tp$ does neither violate a requirement nor contradicts a negative test case.
%\begin{lemma}
%Let $\langle\mo,\mb,\Tp,\Tn\rangle_\RQ$ be a DPI over some DL $\mathcal{L}$ and $\md \in \allD$ some diagnosis w.r.t. $\langle\mo,\mb,\Tp,\Tn\rangle_\RQ$. Then $\EX{\md} \models U_\Tp$.
%\end{lemma}
%\begin{proof}
%Assume that $\EX{\md} \not\models U_\Tp$ holds. 
%\end{proof}
\begin{proposition}\label{prop:exist_diag}
Let $\langle\mo,\mb,\Tp,\Tn\rangle_\RQ$ be a DPI. Then, a diagnosis $\md$ w.r.t. $\langle\mo,\mb,\Tp,\Tn\rangle_\RQ$ exists iff 
%$\mb \cup U_\Tp \not\models \tn$ for each $\tn \in \Tn$ and $\mb \cup U_P$ satisfies each $r\in\RQ$.
\begin{itemize}
	\item $\forall \,r\in\RQ \,\;: \; \mb \cup U_P$ fulfills $r$ and
	\item $\forall \,\tn\in\Tn\,: \;\mb \cup U_\Tp \not\models \tn$.
\end{itemize}
\end{proposition}
\begin{proof}

``$\Leftarrow$'': Let us define $\md := \mo$. Then $X := (\mo \setminus \md) \cup \mb \cup U_P = \mb \cup U_P$. Consequently, $X$ satisfies each $r\in\RQ$ as per condition (\ref{e:1}), $X \not\models \tn$ for each $\tn \in \Tn$ as per condition (\ref{e:3}), and finally $X \models \tp$ for each $\tp \in \Tp$ by extensiveness of $\mathcal{L}$ and thus meets condition (\ref{e:2}). So, $X$ is a solution KB w.r.t. $\langle\mo,\mb,\Tp,\Tn\rangle_\RQ$ wherefore $\md$ must be a diagnosis. 

``$\Rightarrow$'': Let $\md \subseteq \mo$ be some diagnosis w.r.t. $\langle\mo,\mb,\Tp,\Tn\rangle_\RQ$. Then, by definition of a diagnosis, there is some solution KB $\ot$ w.r.t. $\langle\mo,\mb,\Tp,\Tn\rangle_\RQ$. Then $\ot \cup \mb \models \tp$ for all $\tp \in \Tp$ by condition~(\ref{e:2}), which implies that $\ot \cup \mb \cup U_\Tp$ does not feature any new entailments compared to $\ot \cup \mb$ by idempotency of $\mathcal{L}$. So, $\ot \cup \mb \equiv \ot \cup \mb \cup U_\Tp$ holds. Now, for arbitrary $\tn \in \Tn$, since $\ot \cup \mb \not\models \tn$ we have that $\ot \cup \mb \cup U_\Tp \not\models \tn$, and, by monotonicity of $\mathcal{L}$, that $\mb \cup U_\Tp \not\models \tn$. Analogously, for any $r\in\RQ$, because $\ot \cup \mb$ satisfies $r$, it must be true that $\ot \cup \mb \cup U_\Tp$ satisfies $r$ and, by monotonicity of $\mathcal{L}$, that $\mb \cup U_\Tp$ satisfies $r$.
%$\mb \cup U_\Tp \not\models \tn$, then, by monotonicity of $\mathcal{L}$,
%- then there is a solution KB
%- if B u UP models n for some n in N, then, by mono, there is no solution onto
%- if B u UP violates r for some r in R, then, by mono, there is no solution onto
%Then there must exist some extension $\EX{\md}$ such that $\ot:=(\mo \setminus \md) \cup \EX{\md} \cup \mb$ is a solution KB. Since $\ot$ meets all requirements $r\in\RQ$ and does not entail any $\tn \in \Tn$, by monotonicity of $\mathcal{L}$, also $\EX{\md} \cup \mb \subseteq \ot$ must meet all $r\in\RQ$ and cannot entail any $\tn \in \Tn$. However, this holds also for $U_\Tp \cup \mb$ since $\EX{\md} \cup \mb \models U_\Tp \cup \mb$ by the fact that $\EX{\md}\models U_\Tp$ and by extensiveness of $\mathcal{L}$.
%by the fact that $\EX{\md}\models U_\Tp$ and by extensiveness of $\mathcal{L}$ we have that $\EX{\md} \cup \mb \models U_\Tp \cup \mb$.
\end{proof}
\begin{definition}[Admissible DPI]\label{def:admissible}
We call a DPI $\langle\mo,\mb,\Tp,\Tn\rangle_\RQ$ \emph{admissible} iff there is at least one diagnosis $\md \in \allD_{\langle\mo,\mb,\Tp,\Tn\rangle_\RQ}$.
\end{definition}
A non-admissible DPI may arise in a situation where a user specifies test cases manually. For this procedure a similar error-proneness as for the user's formulation of KB formulas can be assumed. And there are lots of pitfalls to escape, as Proposition~\ref{prop:exist_diag} shows. In particular, 
%Proposition~\ref{prop:exist_diag} gives sufficient and necessary conditions that must hold for a DPI in order for a solution, i.e. a solution KB or diagnosis, respectively, to exist. 
%In particular, 
%Proposition~\ref{prop:exist_diag} states that 
%According to Proposition~\ref{prop:exist_diag}, 
the specified test cases in $\Tp$ and $\Tn$ must be ``compatible'' with each other, i.e. positive test cases must not contradict negative ones.
%, in order for a diagnosis to exist. 
For example, adding $\tp_1 := \setof{A \sqsubseteq C, E \equiv B}$ and $\tp_2 := \setof{C \sqsubseteq E}$ to $\Tp$ and $\tn_1 := \setof{A \sqsubseteq B}$ to $\Tn$ leads to a contradiction between $\Tp$ and $\Tn$ and consequently to the non-admissibility of a DPI comprising $\Tp$ and $\Tn$. Furthermore, the background KB $\mb$ which is considered as correct, must indeed be correct, at least in terms of $\RQ$; and negative test cases must be specified in a way not to postulate non-entailment of knowledge specified in $\mb$. A counterexample is $\mb := \setof{\exists r.\top \sqsubseteq A, r(x,y), A \sqsubseteq C}$ and $\Tn := \setof{\setof{C(x)}}$.
% it must not entail a negative test case. 
%It is straightforward that $\ot \cup \mb \supseteq \mb$ can never meet some requirement or non-entailment which is not even met by $\mb$. 
And third, the union of positive test cases together with $\mb$ must be in compliance with $\RQ$, particularly the formulas in $\Tp$ must not be inconsistent or incoherent. Because the union of positive test cases $U_\Tp$ can be viewed as an own KB since all logical sentences occurring in some $\tp\in\Tp$ must be true in the solution KB. So, in a setting where test cases are specified manually, faults occur as likely in $U_\Tp$ as they do in $\mo$.

The debugging system presented in this work, however, guarantees by automatic test case generation that 
%all the conditions for a diagnosis to exist are satisfied 
admissibility of a DPI is satisfied at any time, provided that 
%these conditions are true 
an admissible DPI is given as an initial input to the debugging system. 
%at the beginning of the debugging session. 

\begin{remark}\label{rem:from_non_admissible_DPI_to_admissible_DPI}
In case of a present DPI $\langle\mo,\mb,\Tp,\Tn\rangle_\RQ$ which is non-admissible, the DPI must be properly modified before it can be used with our debugging system. More concretely, the sets $\mb$, $\Tp$ as well as $\Tn$ must be prepared in a way that the two conditions in Proposition~\ref{prop:exist_diag} are satisfied. When supposing that $\mb$ is an already approved and correct KB (which is a reasonable assumption for a KB used as background knowledge during a debugging session), then there are (at least) the following ways to obtain an admissible DPI from a given non-admissible DPI without modifying $\mb$. 

(a)~One straightforward way to achieve that is the deletion of all manually specified test cases from $\Tp$ and $\Tn$. After that, both sets are either the empty set (if no automatic test cases, e.g.\ from former debugging sessions were included in these sets) or comprise only automatically generated test cases. The former case yields an admissible DPI independently of $\mo$ by the property of $\mb$ to not violate any requirements in $\RQ$ (see Definition~\ref{def:dpi}). That the latter case implies the admissibility of the DPI is a property of the debugging system described in this work (as we will show later by Corollary~\ref{cor:query_leaves_valid_diag}).

(b)~Another way to resolve the non-admissibility of a DPI $\langle\mo,\mb,\Tp,\Tn\rangle_\RQ$ is to first check whether $\langle U_{\Tp},\mb,\emptyset,\Tn\rangle_\RQ$ is admissible (verification of Proposition~\ref{prop:exist_diag} by means of a reasoning service). If so, it is clear that $\mb$ does not conflict with $\Tn$. Then, a debugger (like the one presented in this work) can be exploited to find an as small as possible subset of the set of all formulas occurring in the positive test cases, the removal of which causes the DPI to become admissible. This would be accomplished by the computation of a minimal diagnosis $\md_{\Tp}$ w.r.t.\ $\langle U_{\Tp},\mb,\emptyset,\Tn\rangle_\RQ$ and the usage of the modified admissible DPI $\langle\mo,\mb,\setof{U_{\Tp}\setminus\md_{\Tp}},\Tn\rangle_\RQ$ instead of the original one. In this case, only a set-minimal set $\md_{\Tp}$ of formulas that were desired entailments of the user are lost. This modification is possible in polynomial time apart from the reasoning costs, i.e.\ by means of a polynomial number of calls to a reasoner (cf.\ Chapter~\ref{chap:intro}).

(c)~Otherwise, i.e.\ if $\mb$ already conflicts with the negative test cases $\Tn$, then an algorithm similar to Algorithm~\ref{algo:qx} (that will be presented in Section~\ref{sec:cs_comp}) can be employed to determine a maximal subset $\Tn'$ of $\Tn$ w.r.t.\ set inclusion such that $\mb$ will not be in conflict with $\Tn'$. This approach also requires only a polynomial number of calls to a reasoner (cf.\ Proposition~\ref{prop:qx_complexity}). If the resulting modified DPI $\langle\mo,\mb,\Tp,\Tn'\rangle_\RQ$ is not yet admissible, i.e.\ after adding the positive test cases $U_{\Tp}$ to $\mb$ there \emph{are} again conflicts with $\Tn'$, method~(b) must be executed in order to finally obtain an admissible DPI. 

That is, given a non-admissible DPI, there is a transformation achievable in polynomial time which enables the establishment of admissibility involving a set-minimal number of modifications to the given test cases. Thence, in the rest of this work, we will assume that a DPI given as an input to our algorithms is admissible.\qed
\end{remark}
%%%%%%%%%%%%%%% MAKE AN EXTRA SUBSECTION FOR THIS
%In case a user specifies test cases manually, we must assume a similar error-proneness as for the user's formulation of KB formulas. Consequently, the situation may arise where the debugging system says there is no diagnosis for some DPI. In other words, some above-mentioned condition is not met by the DPI at hand. The debugging system presented in this work, however, guarantees by automatic test case generation that all the conditions for a diagnosis to exist are satisfied at any time, provided that these conditions are true at the beginning of the debugging session. If not the case, the initial validity of these conditions can be enforced by first considering the DPI $\langle\mb\cup U_\Tp,\emptyset,\emptyset,\emptyset\rangle_\RQ$ and interactively debugging it. Thereby $\mb' \subseteq \mb$ and $U_\Tp' \subseteq U_\Tp$ is obtained such that $\mb' \cup U_\Tp'$ meets the requirements in $\RQ$, which implies that $\mb'$ as well as $\Tp'$ comply with $\RQ$. Second, $\mb' \cup U_\Tp' \not\models \tn$ for each $\tn \in \Tn$ is checked. If the result is positive, then we set $\mb := \mb'$ and $\Tp := \Tp'$. Otherwise, we know that either some positive and negative test cases are in contradiction with each other or $\mb$ entails some negative test case. So, we consider $\langle\mb',\emptyset,\Tp,\Tn\rangle_\RQ$ is first.....
%%%%%%%%%%%%%%%%%%%%%%%%%%%%%%%%%%%%%%%%%%%%%%%%%%

In general, there are multiple (minimal) diagnoses for a DPI, i.e.\ $|\allD_{\langle\mo,\mb,\Tp,\Tn\rangle_\RQ}| \geq$ $|\minD_{\langle\mo,\mb,\Tp,\Tn\rangle_\RQ}|$ $> 1$, and there are multiple, in fact infinitely many, extensions $\me \in \EX{\md}{\langle\mo,\mb,\Tp,\Tn\rangle_\RQ}$ for a fixed diagnosis $\md \in \allD_{\langle\mo,\mb,\Tp,\Tn\rangle_\RQ}$. The task addressed in this work is finding an optimal diagnosis 
%(or, \emph{equivalently}, an optimal solution KB, as we will show) 
for a given DPI, whereas the identification of an \emph{optimal} extension w.r.t.\ that diagnosis and the DPI is not the aim. What we understand by ``optimality'' of a diagnosis will be addressed in more detail in Part~\ref{part:InteractiveKBDebugging}. Instead, we will content ourselves with finding any extension that enables to formulate a solution KB given a DPI and a diagnosis for that DPI. In fact, the problem of finding a solution KB for a DPI can be reduced to finding a diagnosis for that DPI since a suitable extension 
%exists and 
can be easily formulated for any diagnosis, as the next proposition shows:
%
%Note that for one and the same diagnosis $\md$, there can be many extensions $\EX{\md}$. However, as $\ot \cup \mb$ must entail each $\tp \in \Tp$, the set $U_\Tp := \bigcup_{\tp \in \Tp} \tp$ is necessarily a valid extension independently of $\md$.
\begin{proposition}\label{prop:ex_exist}
%Finding a solution KB is equivalent to finding a diagnosis. I.e. only the deletion step is necessary. 
Let $\langle\mo,\mb,\Tp,\Tn\rangle_\RQ$ be a DPI and $\md\in\allD_{\langle\mo,\mb,\Tp,\Tn\rangle_\RQ}$. Then $U_\Tp$ is an extension w.r.t.\ $\md$ and $\langle\mo,\mb,\Tp,\Tn\rangle_\RQ$.
\end{proposition}
\begin{proof}
Let us assume that there is some $\md \in \allD_{\langle\mo,\mb,\Tp,\Tn\rangle_\RQ}$ and 
%$U_\Tp \notin \EX{\md}{\langle\mo,\mb,\Tp,\Tn\rangle_\RQ}$. 
$U_\Tp$ is not an extension w.r.t.\ $\md$ and $\langle\mo,\mb,\Tp,\Tn\rangle_\RQ$.
By the definition of a diagnosis, this is equivalent to stating that $(\mo \setminus \md) \cup U_\Tp$ is not a solution KB which in turn means that at least one condition (\ref{e:1}), (\ref{e:2}) or (\ref{e:3}) of Definition~\ref{def:target_ont} is violated by $(\mo \setminus \md) \cup U_\Tp$. However, the fact that $\md$ is a diagnosis implies the existence of some extension $\me \in \EX{\md}{\langle\mo,\mb,\Tp,\Tn\rangle_\RQ}$ that can be added to $(\mo \setminus \md)$ to obtain a solution KB. This means that conditions (\ref{e:1}) and (\ref{e:3}) must be already valid 
%before $\EX{\md}$ is appended
for $(\mo \setminus \md)$, since, by monotonicity of $\mathcal{L}$, addition of logical sentences $\me$ can neither solve inconsistencies or incoherencies necessary for fulfillment of condition (\ref{e:1}) nor invalidate non-desired entailments as per condition (\ref{e:3}). As a consequence, condition (\ref{e:2}) must be violated by $(\mo \setminus \md) \cup U_\Tp$. By extensiveness of $\mathcal{L}$ it holds that $(\mo \setminus \md) \cup U_\Tp \models \tp$ for all $\tp \in \Tp$ whereby we obtain that condition (\ref{e:2}) is fulfilled which yields a contradiction.
\end{proof}
%
%The set of formulas $\md = \mo \setminus \ot$ is called diagnosis and the set of logical sentences $\EX{\md} = \ot \setminus \mo$ is called extension. Note that for one and the same diagnosis $\md$, there can be many extensions $\EX{\md}$. However, as $\ot \cup \mb$ must entail each $\tp \in \Tp$, the set $U_\Tp := \bigcup_{\tp \in \Tp} \tp$ is necessarily a valid extension independently of $\md$.
%Equivalent to searching for a solution KB is searching for a set of formulas $\md = \mo \setminus \ot \subseteq \mo$ to be deleted from the given KB in order to specify a solution KB. We call such a set of formulas diagnosis.
%The above proposition leads to a more precise definition of what exactly a diagnosis is.
%\begin{proposition}
%For each solution KB w.r.t. $\langle\mo,\mb,\Tp,\Tn\rangle_\RQ$ there is a solution KB w.r.t. $\langle\mo,\mb,\Tp,\Tn\rangle_\RQ$.
%\end{proposition}
%\begin{proof}
%\end{proof}
Proposition~\ref{prop:ex_exist} claims that the expansion operation, i.e.\ identifying a concrete extension for a diagnosis, is trivial, at least for our purposes, namely formulating an extension reflecting only evident entailments given by the set of positive test cases $\Tp$. Consequently, in order to find a solution KB for some DPI, it is sufficient to concentrate on the deletion step, i.e.\ on the search for diagnoses. 

Note that using $U_\Tp$ as a canonical extension when computing diagnoses does not affect the set of identified diagnoses. In other words, exchanging $\me \in \EX{\md}{\langle\mo,\mb,\Tp,\Tn\rangle_\RQ}$ for $U_\Tp$ in Definition~\ref{def:diagnosis} yields an equivalent definition.
%each diagnosis w.r.t.\ a DPI $\langle\mo,\mb,\Tp,\Tn\rangle_\RQ$ is a diagnosis if a particular extension instead of an arbitrary one
The following corollary proves this statement and summarizes the relationship between the notions \emph{diagnosis}, \emph{solution KB} and \emph{valid KB}.
\begin{corollary}\label{cor:notions_equiv}
The following statements are equivalent:
\begin{enumerate}
\item $\md$ is a diagnosis w.r.t.\ $\langle\mo,\mb,\Tp,\Tn\rangle_\RQ$
\item $(\mo \setminus \md) \cup U_\Tp$ is a solution KB w.r.t.\ $\langle\mo,\mb,\Tp,\Tn\rangle_\RQ$
\item $(\mo \setminus \md)$ is valid w.r.t.\ $\langle\cdot,\mb,\Tp,\Tn\rangle_\RQ$.
\end{enumerate}
\end{corollary}
\begin{proof}
That (1) is equivalent to (2) follows from Definition~\ref{def:diagnosis} which states that $\md$ is a diagnosis w.r.t.\ $\langle\mo,\mb,\Tp,\Tn\rangle_\RQ$ iff there is some set of sentences $\me\in\EX{\md}{\langle\mo,\mb,\Tp,\Tn\rangle_\RQ}$ such that $(\mo \setminus \md) \cup \me$ is a solution KB, and from Proposition~\ref{prop:ex_exist} which proves that $U_\Tp$ is an extension w.r.t.\ \emph{any} diagnosis $\md$ and $\langle\mo,\mb,\Tp,\Tn\rangle_\RQ$. 

That (1) is equivalent to (3) follows directly from Proposition~\ref{prop:validonto_diag} and the equivalence of (2) and (3) has been shown in Proposition~\ref{prop:validonto_targetonto}.
\end{proof}

\section[Parsimonious Knowledge Base Debugging]{Parsimonious Knowledge Base Debugging%
\sectionmark{Parsimonious KB Debugging}}
\sectionmark{Parsimonious KB Debugging}
\label{sec:MinimallyInvasiveOntologyDebugging} 
%%%%%
%\section{Parsimonious Knowledge Base Debugging}
%\label{sec:MinimallyInvasiveOntologyDebugging}
%%%%%
Why are \emph{minimal} diagnoses interesting? 
%The answer is twofold: 
First, the set of minimal diagnoses w.r.t.\ a DPI captures all the information that explains the unwanted properties, i.e.\ violation of requirements or test cases, of the DPI. In other words, the minimal diagnoses represent all subset-minimal possibilities to modify a KB in a way it becomes a valid KB w.r.t.\ the given DPI (e.g.\ by simply deleting a minimal diagnosis from the KB in the trivial case). By monotonicity of the logic $\mathcal{L}$, each superset of a minimal diagnosis w.r.t.\ a DPI is a diagnosis w.r.t.\ this DPI. That is, $\allD_{\langle\mo,\mb,\Tp,\Tn\rangle_\RQ}$ can be easily reconstructed given $\minD_{\langle\mo,\mb,\Tp,\Tn\rangle_\RQ}$.
% by computing $2^\mo \setminus \minD^{\subset}_{\langle\mo,\mb,\Tp,\Tn\rangle_\RQ}$ where $\minD^{\subset}_{\langle\mo,\mb,\Tp,\Tn\rangle_\RQ}$ is the set containing all proper subsets of sets in $\minD_{\langle\mo,\mb,\Tp,\Tn\rangle_\RQ}$. 
There is however no evidence (in terms of specified requirements and test cases) in a DPI that would justify the selection of a non-minimal diagnosis. That is, if $\mo$ is a KB and $\md \subseteq\mo$ a minimal diagnosis w.r.t.\ a DPI including $\mo$, $\mo\setminus\md$ does not violate any of the postulated properties that must hold for a KB to be valid w.r.t.\ this DPI. For that reason, there is no \emph{evident} need to delete or modify any other sentences in $\mo$ except for the ones in some \emph{minimal} diagnosis $\md$.
%
%Therefore, the focus can be completely on the minimal diagnoses when debugging ontologies.

Second, usually a setting can be assumed where the author of a KB specifies formulas to the best of their knowledge. Hence, the assumption that a formula is rather correct than faulty, or in other words, that the KB author wants to keep as many formulated sentences as possible in a solution KB obtained from a debugger, is practical. 

This also motivates the importance of a certain subset of minimal diagnoses, namely minimum cardinality diagnoses, which are the solutions of choice in scenarios where no probabilistic information about the KB authors' faults is available, e.g.\ in terms of statistics retrieved from log data of the used IDE (see Section~\ref{sec:DiagnosisProbabilitySpace} for details). In an application where such information is given, minimum cardinality diagnoses might not always be the appropriate choice (for details see Part~\ref{part:InteractiveKBDebugging}). In this case the aim is to find a minimal diagnosis with a maximal probability of including only sentences that are actually faulty (which might not necessarily be a minimum cardinality diagnosis).
%the minimal diagnoses are exactly these repairs of a DPI that

Third, minimality of diagnoses will be a necessary condition to \emph{guarantee} the possibility of discrimination between different (candidate) diagnoses to formulate a solution KB, as will be seen later in Chapter~\ref{chap:UserInteraction}.

Fourth, focusing only on minimal diagnoses rather than all diagnoses can greatly reduce the search space for diagnoses and therefore greatly speed up the debugging procedure (cf.~\cite{dekleer1987}).

Projected to the task of KB debugging, namely finding a solution KB w.r.t.\ a given DPI, this means we are interested in minimal invasiveness, that is making as few formula-deletion-modifications to the input KB $\mo$ as possible in the course of the performed debugging actions. That is, the actual goal is to find some \emph{maximal} solution KB $\ot$ for a DPI. Compare with ``The Principle of Parsimony'' in \cite[p.~7]{Reiter87} \cite{Bylander1991}.\vspace{3pt}

\noindent\fcolorbox{black}{light-gray1}{\parbox[c][2em][c]{0.975\linewidth}{\vspace{-4pt}
\begin{prob_def}[Parsimonious KB Debugging] \label{prob_def:evidence_just}
Given a DPI $\langle\mo,\mb,\Tp,\Tn\rangle_\RQ$, the task is to find a maximal solution KB w.r.t.\ $\langle\mo,\mb,\Tp,\Tn\rangle_\RQ$.
\end{prob_def}\vspace{-4pt}
}}

\vspace{3pt}
The next proposition shows that this problem can be reduced to finding a minimal diagnosis. 

\begin{proposition}\label{prop:min_max}
%\begin{enumerate}
%\item 

\textbf{(i)} 
$\mo \setminus \ot$ is a minimal diagnosis w.r.t.\ $\langle\mo,\mb,\Tp,\Tn\rangle_\RQ$ for each maximal solution KB $\ot$ w.r.t.\ $\langle\mo,\mb,\Tp,\Tn\rangle_\RQ$. 
%For each maximal solution KB $\ot$ w.r.t.\ $\langle\mo,\mb,\Tp,\Tn\rangle_\RQ$, $\mo \setminus \ot$ is a minimal diagnosis w.r.t.\ $\langle\mo,\mb,\Tp,\Tn\rangle_\RQ$.  
%\item
 
\textbf{(ii)} If $\md$ is a minimal diagnosis w.r.t.\ $\langle\mo,\mb,\Tp,\Tn\rangle_\RQ$, then $(\mo\setminus\md) \cup \me$ is a maximal solution KB w.r.t.\ $\langle\mo,\mb,\Tp,\Tn\rangle_\RQ$ for all extensions $\me \in \EX{\md}{\langle\mo,\mb,\Tp,\Tn\rangle_\RQ}$.
%\end{enumerate}
\end{proposition}
\begin{proof}
\textbf{Ad (i):} Let $\ot$ be an arbitrary maximal solution KB w.r.t.\ $\langle\mo,\mb,\Tp,\Tn\rangle_\RQ$. The first observation is that $\md := \mo \setminus \ot$ is a diagnosis w.r.t.\ $\langle\mo,\mb,\Tp,\Tn\rangle_\RQ$ since 
%there is an extension $\EX{\md} := \ot \setminus \mo$ 
$\ot \setminus \mo \in \EX{\md}{\langle\mo,\mb,\Tp,\Tn\rangle_\RQ}$ by the fact that $\ot = (\mo \setminus \md) \cup (\ot \setminus \mo)$ is a solution KB by assumption. Let us assume that there is a diagnosis $\md_k \in \allD_{\langle\mo,\mb,\Tp,\Tn\rangle_\RQ}$ such that $\md \supset \md_k$. Since $\md_k$ is a diagnosis, it holds per Definition~\ref{def:diagnosis} that there is an extension $\me \in \EX{\md_k}{\langle\mo,\mb,\Tp,\Tn\rangle_\RQ}$ such that $\ot_k := (\mo \setminus \md_k) \cup \me$ is a solution KB. Further on, $\mo \cap \ot_k = \mo \cap ((\mo \setminus \md_k) \cup \me) = (\mo \setminus \md_k) \cup (\mo \cap \me)$. Since $\mo \cap \ot$ can be written as $\mo \setminus (\mo \setminus \ot) = \mo \setminus \md$ which is a strict subset of $\mo \setminus \md_k$ which in turn is a subset of $(\mo \setminus \md_k) \cup (\mo \cap \me) = \mo \cap \ot_k$. Consequently, $\mo \cap \ot \subset \mo \cap \ot_k$ holds, which is by Definition~\ref{def:target_ont} a contradiction to the maximality of the solution KB $\ot$. Thus, $\md = \mo \setminus \ot$ is a minimal diagnosis w.r.t.\ $\langle\mo,\mb,\Tp,\Tn\rangle_\RQ$.

\textbf{Ad (ii):} Let $\md$ be a minimal diagnosis w.r.t.\ $\langle\mo,\mb,\Tp,\Tn\rangle_\RQ$. Then, by Definition~\ref{def:diagnosis}, there is an extension $\me \in \EX{\md}{\langle\mo,\mb,\Tp,\Tn\rangle_\RQ}$ such that $\ot := (\mo \setminus \md) \cup \me$ is a solution KB. Let us assume that $\me \cap \md \neq \emptyset$. We can rewrite $\ot$ as $\ot = (\mo \setminus \md) \cup (\me \cap \md) \cup (\me \setminus \md)$. Since $\emptyset \subset \me \cap \md \subseteq \md$, we have that $(\mo \setminus \md) \cup (\me \cap \md) \supset \mo \setminus \md$. Thus, there is a $\md' := \md \setminus (\me \cap \md) \subset \md$ 
%such that $\mo \setminus \md' \subseteq \ot$; in other words, there is 
and an extension $\me' \in \EX{\md'}{\langle\mo,\mb,\Tp,\Tn\rangle_\RQ}$ such that $\me' := \me \setminus \md$ such that $\ot = (\mo \setminus \md') \cup \me'$. As $\ot$ is a solution KB, this is a contradiction to the minimality of $\md$. Therefore, (*) $\me \cap \md = \emptyset$ for all $\me \in \EX{\md}{\langle\mo,\mb,\Tp,\Tn\rangle_\RQ}$ must hold.

Let $\me$ be any extension w.r.t.\ $\md$ and $\langle\mo,\mb,\Tp,\Tn\rangle_\RQ$. Then we can write $\mo \cap \ot = \mo \cap ((\mo \setminus \md) \cup \me) = (\mo \setminus \md) \cup (\mo \cap \me)$ and by (*) also $\mo \cap \me = ((\mo \setminus \md) \cup \md) \cap \me = ((\mo \setminus \md)\cap \me) \cup (\md \cap \me) = (\mo \setminus \md)\cap \me \subseteq \mo \setminus \md$. Consequently, (**) $\mo \cap \ot = \mo\setminus\md$. Now, assume that there is a solution KB $\ot_k$ with the property $\mo \cap \ot_k \supset \mo \cap \ot$. By (**), this implies that $\mo \cap \ot_k \supset \mo \setminus \md$ which means that there is a $\md_k \subset \md \subseteq \mo$ such that $\mo \cap \ot_k = \mo \setminus \md_k \subseteq \ot_k$. Now $\ot_k$ is a solution KB w.r.t.\ $\langle\mo,\mb,\Tp,\Tn\rangle_\RQ$ and can be written as $\ot_k = (\ot_k \cap \mo) \cup (\ot_k \setminus \mo) = (\mo \setminus \md_k) \cup (\ot_k \setminus \mo)$. By $\md_k \subseteq \mo$ and since there is a set of formulas $\me := \ot_k \setminus \mo$ such that $(\mo\setminus\md_k) \cup \me \in \SO_{\langle\mo,\mb,\Tp,\Tn\rangle_\RQ}$ we have that $\me \in \EX{\md_k}{\langle\mo,\mb,\Tp,\Tn\rangle_\RQ}$ must hold wherefore $\md_k$ is a diagnosis by Definition~\ref{def:diagnosis}.
%there is an extension $\EX{\md_k}$, namely $\ot_k \setminus (\mo \setminus \md_k)$, such that $\ot_k = (\mo \setminus \md_k) \cup \EX{\md_k}$. Given that $\ot_k$ is a solution KB, we can conclude by Definition~\ref{def:diagnosis} that $\md_k$ must be a diagnosis which 
This, however, is a contradiction to the minimality of $\md$. Therefore, $\ot = (\mo \setminus \md) \cup \me$ must be a maximal solution KB for any $\me \in \EX{\md}{\langle\mo,\mb,\Tp,\Tn\rangle_\RQ}$.
%%%%%%%%%%%%%%%%%%%%%%%%%%%%%%%%% (i) OLD --- START 
%Ad (ii): Let $\md_m$ be a minimal diagnosis w.r.t.\ $\langle\mo,\mb,\Tp,\Tn\rangle_\RQ$. Then, for all diagnoses $\md_k$ w.r.t.\ $\langle\mo,\mb,\Tp,\Tn\rangle_\RQ$ it holds that $\md_k \supseteq \md_m$. So, we have (*) $\mo \setminus \md_m \supseteq \mo \setminus \md_k$. As $\md_m$ and $\md_k$ are diagnoses, there are extensions $\EX{\md_m}$ and $\EX{\md_k}$ such that $\ot_m = (\mo \setminus \md_m) \cup \EX{\md_m}$ and $\ot_k = (\mo \setminus \md_k) \cup \EX{\md_k}$ are solution ontologies for $\md_m$ and $\md_k$, respectively. For an arbitrary solution KB $\ot = (\mo \setminus \md) \cup \EX{\md}$ it is true that $\mo \cap \ot = \mo \cap ((\mo \setminus \md) \cup \EX{\md}) = (\mo \setminus \md) \cup (\mo \cap \EX{\md})$. But, by $\md \cap \EX{\md} = \emptyset$, we obtain that $\mo \cap \EX{\md} = ((\mo \setminus \md) \cup \md) \cap \EX{\md} = ((\mo \setminus \md)\cap \EX{\md}) \cup (\md \cap \EX{\md}) = (\mo \setminus \md)\cap \EX{\md} \subseteq \mo \setminus \md$. Consequently, $\mo \cap \ot = \mo \setminus \md$. Replacing $\ot$ and $\md$, respectively, by $\setof{\ot_m,\ot_k}$ and $\setof{\md_m,\md_k}$ together with (*) completes the proof.
%%%%%%%%%%%%%%%%%%%%%%%%%%%%%%%%% (ii) OLD --- END
\end{proof}
By claim~(i), Proposition~\ref{prop:min_max} assures that each maximal solution KB can be found by investigating all minimal diagnoses w.r.t. a DPI. Claim~(ii) shows that any solution KB built from a minimal diagnosis is indeed maximal. Thus, finding a suitable minimal diagnosis solves the problem of parsimonious KB debugging completely.
%But this is equivalent to searching for some minimal diagnosis $\md$. 
%due to the fact 
%The reason for this is that a maximal solution KB $\ot$ means set-maximal $\mo \cap \ot$, which is equivalent to set-minimal $\mo \setminus \ot = \md$.
%, i.e. $\md$ is a minimal diagnosis such that $\ot = (\mo \setminus \md) \cup \EX{\md}$.
%superset-maximal $\mo \cap \ot$, i.e. maximal solution KB $\ot$, is equivalent to subset-minimal $\mo \setminus \ot = \md$, i.e. minimal diagnosis $\md$ such that $\ot = (\mo \setminus \md) \cup \EX{\md}$.

%\noindent\emph{Background Knowledge.} 
\section{Background Knowledge}
\label{sec:BackgroundKnowledge}
The general debugging setting considered in this work envisions the opportunity for the user to specify some background knowledge $\mb$, i.e.~a set of formulas that are known (or strongly assumed) to be correct in advance.
Note that, in order for the debugging procedure to work soundly, before some background knowledge is incorporated into the DPI, it is necessary to verify its conformance with the postulated requirements $\RQ$ (cf.\ Definition~\ref{def:dpi}).%, i.e.\ at least consistency of $\mb$ must be ensured.
We can distinguish between two basic scenarios how background knowledge can be leveraged: (1)~We have an initial KB $\mo_{\mathsf{init}}$ and we know or want to assume that a subset of formulas in $\mo_{\mathsf{init}}$ is correct, i.e.\ $\mb \cap \mo_{\mathsf{init}} \neq \emptyset$, and (2)~we have an initial KB $\mo_{\mathsf{init}}$ and some background knowledge disjoint from $\mo_{\mathsf{init}}$, i.e.\ $\mb \cap \mo_{\mathsf{init}} = \emptyset$.

Example use cases for scenario~(1) are situations where a user knows that a subset of formulas $\mb$ in $\mo$ is definitely sound or wants to restrict the scope of debugging to a particular part of the KB. Concretely, this may occur, for instance, when $\mb$ is the result, i.e.\ the finally output solution KB $\ot$, of a former successful debugging session and $\mo$ is a further development of $\ot$, or in a collaborative setting where many users are involved in the development of $\mo$ and one of them may want to debug only formulas authored by herself and not touch foreign formulas, which are thus assumed as correct and assigned to $\mb$. 
%; for instance, because $\mb$ is the result, i.e.\ the solution KB $\ot$, of a former successful debugging session and $\mo$ is a further development of $\ot$, or when many users are involved in the development of $\mo$, one of these users may want to debug only formulas authored by herself and assume foreign formulas as correct and assign them to $\mb$. 
In (1), $\mo_{\mathsf{init}}\cap\mb$ and $\mo_{\mathsf{init}}~\setminus~\mb$ partition the original KB $\mo_{\mathsf{init}}$ into a set of correct and a set of possibly incorrect formulas, respectively. The corresponding DPI would thus be $\langle\mo_{\mathsf{init}}~\setminus~\mb,\mb,\Tp,\Tn\rangle_\RQ$ for some sets of test cases $\Tp$ and $\Tn$. Note that this DPI \emph{does} meet the necessary condition (cf. Definition~\ref{def:dpi}) $\mo \cap \mb = \emptyset$ as $(\mo_{\mathsf{init}}~\setminus~\mb) \cap \mb = \emptyset$.
So, in the debugging session, only $\mo := \mo_{\mathsf{init}}~\setminus~\mb$ is used to search for diagnoses, which can reduce the search space substantially. Though, $\mb$ is incorporated in the calculations throughout the KB debugging procedure, but no formula in $\mb$ may take part in a diagnosis. The advantage of this over simply not considering the formulas in $\mb$ at all is, that the semantics of formulas in $\mb$ is not lost and can be exploited, e.g., to grant the desired semantic properties also in the context of existing approved knowledge or to facilitate a greater choice of queries to interact with a user, which can be exploited to ask queries with lower cardinality or involving less complex formulas (see Chapter~\ref{chap:UserInteraction} for details on queries).
%where better is meant in the sense of easier or with less adherent effort for the respective user.

In scenario~(2), the corresponding DPI looks like $\langle\mo_{\mathsf{init}},\mb,\Tp,\Tn\rangle_\RQ$ for some sets of test cases $\Tp$ and $\Tn$. An application of this scenario could be the reuse of an existing KB to support an increase of the fault detection rate and thus more sustainable debugging. For example, when formulating a KB $\mo_{\mathsf{init}}$ about a domain, a reference KB $\mb$ in that domain that is thoroughly curated by experts could be leveraged.  
%both if $\mo$ meets $\RQ$ and not. 
The use of such a KB $\mb$ is possible both if $\mo_{\mathsf{init}}$ is correct as a standalone KB, i.e.\ $\mo_{\mathsf{init}}$ is already a solution KB for $\langle\mo_{\mathsf{init}},\emptyset,\Tp,\Tn\rangle_\RQ$, or not. In the first case, $\mo_{\mathsf{init}}$ might still contain formulations conflicting with $\mb$. In this vein, in both cases, faults may be detected that would have been missed otherwise.

%\fixme{insert motivating example for background KB}

%Say $\RQ$ hold for $\mo$, i.e. coherence and consistency of $\mo$ is given, then $\mb$ can be employed to find formulations in $\mo$ contradicting the correct ones in $\mb$, that would not be found without integration of $\mb$ into the debugging procedure.
%
%Of course, mixtures of scenarios (1) and (2) are possible as well.

%\noindent\emph{Diagnosis Computation.} 
\chapter{Diagnosis Computation}
\label{chap:DiagnosisComputation}
In this chapter we describe methods for computing minimal diagnoses w.r.t.\ a given admissible DPI, provide an in-depth theoretical analysis of these methods including correctness proofs and illustrate the presented algorithms by various examples.

\section{Conflict Sets}
\label{sec:ConflictSets}
The search space for minimal diagnoses w.r.t.\ $\langle\mo,\mb,\Tp,\Tn\rangle_\RQ$ the size of which is in general $O(2^{|\mo|})$ (if all subsets of the KB $\mo$ are investigated) can be reduced to a great extent by exploiting the notion of a conflict set~\cite{Reiter87,dekleer1987,Shchekotykhin2012}. 
\begin{definition}[Conflict Set]\label{def:cs} Let $\langle\mo,\mb,\Tp,\Tn\rangle_\RQ$ be a DPI. A set of formulas $\mc \subseteq \mo$ is called a \emph{conflict set w.r.t.\ $\langle\mo,\mb,\Tp,\Tn\rangle_\RQ$}, written as $\mc \in \allC_{\langle\mo,\mb,\Tp,\Tn\rangle_\RQ}$, iff $\mc \cup U_\Tp$ is not a solution KB w.r.t.\ $\langle\mo,\mb,\Tp,\Tn\rangle_\RQ$. A conflict set $\mc$ is minimal, written as $\mc \in \minC_{\langle\mo,\mb,\Tp,\Tn\rangle_\RQ}$, iff there is no $\mc' \subset \mc$ such that $\mc'$ is a conflict set.
\end{definition}
Simply put, a (minimal) conflict set is a (minimal) faulty KB that is a subset of $\mo$. That is, a conflict set is one source causing the faultiness of $\mo$ in the context of $\mb \cup U_\Tp$.
%a conflict set $\mc$ is a faulty subset of the KB (w.r.t.\ the DPI $\langle\mo,\mb,\Tp,\Tn\rangle_\RQ$), where faulty refers to a violation of requirements $r\in\RQ$ or test cases $\tp\in\Tp$ or $\tn\in\Tn$. 
In other words, a valid KB may not include all the formulas of any conflict set.
\begin{corollary}\label{cor:validonto_cs}
$\mc \subseteq \mo$ is a conflict set w.r.t.\ $\langle\mo,\mb,\Tp,\Tn\rangle_\RQ$ iff $\mc$ is invalid w.r.t.\ $\langle\cdot,\mb,\Tp,\Tn\rangle_\RQ$.
\end{corollary}
\begin{proof}
If $\mc$ is a conflict set w.r.t.\ $\langle\mo,\mb,\Tp,\Tn\rangle_\RQ$, then $\mc \cup U_\Tp$ is not a solution KB, i.e.\ $\mc \cup \mb \cup U_\Tp$ violates some $r\in\RQ$, some $\tp \in \Tp$ or some $\tn\in\Tn$. By extensiveness of $\mathcal{L}$, $\mc \cup \mb \cup U_\Tp \models \tp$ for all $\tp \in \Tp$, so $\mc \cup \mb \cup U_\Tp$ must violate some $r\in\RQ$ or entail some $\tn\in\Tn$. Thus, by Definition~\ref{def:valid_onto}, $\mc$ is invalid w.r.t.\ $\langle\cdot,\mb,\Tp,\Tn\rangle_\RQ$.

If $\mc \subseteq \mo$ is not valid w.r.t.\ $\langle\cdot,\mb,\Tp,\Tn\rangle_\RQ$, then $\mc \cup \mb \cup U_\Tp$ violates some $r\in\RQ$ or entails some $\tn\in\Tn$, wherefore $\mc \cup U_\Tp \notin \SO_{\langle\mo,\mb,\Tp,\Tn\rangle_\RQ}$. Hence, by Definition~\ref{def:cs}, $\mc$ is a conflict set w.r.t.\ $\langle\mo,\mb,\Tp,\Tn\rangle_\RQ$.
\end{proof}
%This means that $\mc$ does not support the positive test cases $\Tp$ w.r.t.\ $\langle\mo,\mb,\Tp,\Tn\rangle_\RQ$ (cf. Proposition~\ref{prop:contradict_P_targetonto}), i.e.\ 
Consequently, a conflict set $\mc$ along with the background knowledge $\mb$ either violates some $r\in\RQ$, entails some $\tn \in \Tn$, or yields to a violation of some $r\in\RQ$ or entailment of some $\tn \in \Tn$ if all formulas $U_\Tp$ comprised by the positive test cases are added to $\mc$. Any KB $\mo$
that is not valid w.r.t.\ $\langle\cdot,\mb,\Tp,\Tn\rangle_\RQ$ 
%violates any of the conditions~(\ref{e:1}) or (\ref{e:3}) in Definition~\ref{def:target_ont}, i.e.\ $\mo \cup \mb$ violates some $r\in\RQ$ or $\mo \cup \mb \models \tn$ for some $\tn \in \Tn$, 
is itself a conflict set and includes at least one minimal conflict set. 
%If only condition~(\ref{e:2}) of Definition~\ref{def:target_ont} is violated by 
%\begin{proposition}\label{prop:non_target_includes_cs}
%Let $\langle\mo,\mb,\Tp,\Tn\rangle_\RQ$ be a DPI. Then, $\mo \cup U_\Tp$ is not a solution KB w.r.t.\ $\langle\mo,\mb,\Tp,\Tn\rangle_\RQ$ iff $\mo$ includes at least one minimal conflict set w.r.t.\ $\langle\mo,\mb,\Tp,\Tn\rangle_\RQ$.
%\end{proposition}
%\begin{proof}
%``$\Rightarrow$'': $\mo \cup U_\Tp$ being no solution KB w.r.t.\ $\langle\mo,\mb,\Tp,\Tn\rangle_\RQ$ means that $\mo$ is a conflict set w.r.t.\ $\langle\mo,\mb,\Tp,\Tn\rangle_\RQ$ by definition~\ref{def:cs}. So, either $\mo$ is a already a minimal conflict set or there must be some subset $\mc \subset \mo$ which is a minimal conflict set.
%%$\mo \cup U_\Tp \cup \mb \models \tn$ for some $\tn\in\Tn$ or $\mo \cup U_\Tp \cup \mb$ violates some $r\in\RQ$.
%
%``$\Leftarrow$'': Let $\mo$ include at least one minimal conflict set w.r.t.\ $\langle\mo,\mb,\Tp,\Tn\rangle_\RQ$. Then, by Definition~\ref{def:cs}, there is some $\mc \subseteq \mo$ such that $\mc \cup U_\Tp$ is not a solution KB. Hence, by monotonicity of $\mathcal{L}$, $\mo \cup U_\Tp$ cannot be a solution KB either. 
%\end{proof}
\begin{proposition}\label{prop:non_validonto_includes_cs}
Let $\langle\mo,\mb,\Tp,\Tn\rangle_\RQ$ be a DPI. Then, $\mo$ is not valid w.r.t.\ $\langle\cdot,\mb,\Tp,\Tn\rangle_\RQ$ iff $\mo$ includes at least one minimal conflict set w.r.t.\ $\langle\mo,\mb,\Tp,\Tn\rangle_\RQ$.
\end{proposition}
\begin{proof}
``$\Rightarrow$'': Let $\mo$ be not valid w.r.t.\ $\langle\cdot,\mb,\Tp,\Tn\rangle_\RQ$. Then $\mo \cup U_\Tp$ is not a solution KB w.r.t.\ $\langle\mo,\mb,\Tp,\Tn\rangle_\RQ$, which means that $\mo$ is a conflict set w.r.t.\ $\langle\mo,\mb,\Tp,\Tn\rangle_\RQ$ by definition~\ref{def:cs}. So, either $\mo$ is a already a minimal conflict set or there must be some subset $\mc \subset \mo$ which is a minimal conflict set w.r.t. $\langle\mo,\mb,\Tp,\Tn\rangle_\RQ$.
%$\mo \cup U_\Tp \cup \mb \models \tn$ for some $\tn\in\Tn$ or $\mo \cup U_\Tp \cup \mb$ violates some $r\in\RQ$.

``$\Leftarrow$'': Let $\mo$ include at least one minimal conflict set w.r.t.\ $\langle\mo,\mb,\Tp,\Tn\rangle_\RQ$. Then, by Definition~\ref{def:cs}, there is some $\mc \subseteq \mo$ such that $\mc \cup U_\Tp$ is not a solution KB. Hence, by the monotonicity of $\mathcal{L}$, $\mo \cup U_\Tp$ cannot be a solution KB either. So, by Proposition~\ref{prop:validonto_targetonto}, $\mo$ is not valid w.r.t.\ $\langle\cdot,\mb,\Tp,\Tn\rangle_\RQ$.
\end{proof}
As a consequence, a complete and sound method for computing minimal conflict sets w.r.t.\ a DPI $\langle\mo,\mb,\Tp,\Tn\rangle_\RQ$ can be used to decide validity of $\mo$ w.r.t.\ $\langle\cdot,\mb,\Tp,\Tn\rangle_\RQ$. Moreover, such a method can be used to decide whether a given DPI is admissible, i.e.\ has solutions.
For, if a DPI is admissible and the given KB is invalid w.r.t.\ this DPI, then there cannot be an empty conflict set. In other words, if the empty KB is a conflict set -- or, equivalently, an empty conflict set exists w.r.t.\ a DPI --, then the DPI is not admissible.
\begin{proposition}\label{prop:cs_admissible}
Let $\langle\mo,\mb,\Tp,\Tn\rangle_\RQ$ be a DPI and $\mo$ be invalid w.r.t.\ $\langle\cdot,\mb,\Tp,\Tn\rangle_\RQ$. Then, there exists a minimal conflict set $\mc \neq \emptyset$ w.r.t.\ $\langle\mo,\mb,\Tp,\Tn\rangle_\RQ$ iff $\langle\mo,\mb,\Tp,\Tn\rangle_\RQ$ is admissible.
\end{proposition}
\begin{proof}
Since $\mo$ is not valid w.r.t.\ $\langle\cdot,\mb,\Tp,\Tn\rangle_\RQ$, there must be at least one conflict set w.r.t.\ $\langle\mo,\mb,\Tp,\Tn\rangle_\RQ$ by Proposition~\ref{prop:non_validonto_includes_cs}. Assume that there exists a minimal conflict set $\mc \neq \emptyset$ w.r.t.\ $\langle\mo,\mb,\Tp,\Tn\rangle_\RQ$. This can be true iff $\emptyset$ is not a (minimal) conflict set w.r.t.\ $\langle\mo,\mb,\Tp,\Tn\rangle_\RQ$. By Corollary~\ref{cor:validonto_cs} and Definition~\ref{def:valid_onto}, this is equivalent to the fact that $\emptyset \cup \mb \cup U_\Tp \equiv \mb \cup U_\Tp$ does not violate any $r\in\RQ$ and does not entail any $\tn \in \Tn$. By Proposition~\ref{prop:exist_diag}, this holds iff there exists a diagnosis w.r.t.\ $\langle\mo,\mb,\Tp,\Tn\rangle_\RQ$. By Definition~\ref{def:admissible}, this is equivalent to $\langle\mo,\mb,\Tp,\Tn\rangle_\RQ$ being admissible.
\end{proof}
The following proposition provides information about the relationship between (minimal) conflict sets and the background knowledge as well as the positive test cases.
\begin{proposition}\label{prop:cs_properties}
Let $\langle\mo,\mb,\Tp,\Tn\rangle_\RQ$ be a DPI and $\mc$ a conflict set w.r.t.\ $\langle\mo,\mb,\Tp,\Tn\rangle_\RQ$. Then the following holds:
\begin{enumerate}
	\item $\mc \cap \mb = \emptyset$.
	\item If $\mc$ is a minimal conflict set w.r.t.\ $\langle\mo,\mb,\Tp,\Tn\rangle_\RQ$, then $\mc \cap U_\Tp = \emptyset$.
\end{enumerate}
\end{proposition}
\begin{proof}
1): $\mc \cap \mb = \emptyset$ holds since $\mc \subseteq \mo$ (Definition~\ref{def:cs}) and $\mo \cap \mb = \emptyset$ (Definition~\ref{def:dpi}).

2):	Assume that $\mc$ is a minimal conflict set w.r.t.\ $\langle\mo,\mb,\Tp,\Tn\rangle_\RQ$ and $\mc \cap U_\Tp \neq \emptyset$. Since $\mc$ is a conflict set, we have that $\mc \cup \mb \cup U_\Tp$ violates some $r\in\RQ$ or entails some $\tn\in\Tn$ by Corollary~\ref{cor:validonto_cs} and Definition~\ref{def:valid_onto}. Since $(\mc\setminus U_\Tp) \cup \mb \cup U_\Tp = \mc \cup \mb \cup U_\Tp$ and $(\mc\setminus U_\Tp) \subset \mc$, this implies that $(\mc\setminus U_\Tp)$ is a conflict set w.r.t.\ $\langle\mo,\mb,\Tp,\Tn\rangle_\RQ$ which in turn implies that $\mc \notin \minC_{\langle\mo,\mb,\Tp,\Tn\rangle_\RQ}$ which is a contradiction.
\end{proof}

\section{Conflict Sets versus Justifications}
\label{sec:ConflictSetsVersusJustifications}
The notion of a conflict set is closely related to the notion of a justification~\cite{Horridge2008, Horridge2009, Horridge2010, Horridge2011a, Horridge2011b, Horridge2012b} which is frequently adopted in the field of the Semantic Web (cf.\ Section~\ref{sec:OntologiesAndDescriptionLogic}) in order to find minimal explanations for particular entailments in DL ontologies. Thus, the paradigm of a justification can be a useful aid in the debugging of faulty ontologies~\cite{Kalyanpur2006a}. Note that sometimes justifications are referred to as \mbox{MinAs} (Minimal Axiom Sets) \cite{Baader2008} or MUPS (Minimal Unsatisfiability Preserving Sub-TBoxes) \cite{Schlobach2007} where the latter term is mostly used in the context of ontology debugging.
The notion of a (minimal) conflict set, on the other hand, has been mainly adopted in the Diagnosis community~\cite{Reiter87,dekleer1987,Peischl2003,wotawa2002,Felfernig2004213}. In this section we want to establish a relationship between these two widely used instruments used for debugging. It will turn out that both terms are strongly related, but in debugging systems like the ones proposed in our work conflict sets are better suited as they automatically focus \emph{only} on the minimal explanations for \emph{faults} in a KB.

For example, the author of \cite{Kalyanpur2006a} i.a.\ discusses the use of justifications to aid the debugging of incoherent ontologies, i.e.\ ontologies that include unsatisfiable concepts (cf.\ Section~\ref{sec:DL}). If there are multiple unsatisfiable concepts, then some of these might be only unsatisfiable due to the unsatisfiability of another concept. 
%In other words, resolving the unsatisfiability 
Assume, for instance, an incoherent DL KB $\mo := \setof{A \sqsubset B, B \sqsubseteq E \sqcap \lnot E}$. In $\mo$ there are two unsatisfiable concepts $A$ and $B$ where $A$'s unsatisfiability is dependent on $B$'s unsatisfiability. Using the terminology of \cite{Kalyanpur2006a,Horridge2011a}, $A$ would be called a \emph{purely derived} unsatisfiable concept whereas $B$ would be called a \emph{root} unsatisfiable concept. Because the (only) justification for the unsatisfiability of $A$ is $J_A := \mo$ whereas the (only) justification for the unsatisfiability of $B$ is $J_B = \setof{B \sqsubseteq E \sqcap \lnot E} \subset J_A$. Therefore, \cite{Kalyanpur2006a} proposes to resolve root unsatisfiable concepts first since this might resolve some (purely) derived concepts as well, as in this example. However, finding out whether a concept is root or derived involves the computation of justifications for all unsatisfiable concepts in a KB. On the other hand, reliance on minimal conflict sets would implicate a direct focus on the faultiness (in this example: the incoherency) \emph{of the KB} and not necessarily on the exact explanations of all unsatisfiable concepts that cause the incoherency. In this vein, no justification for a purely derived concept can be a minimal conflict set. So, the computation of minimal conflict sets involves only the determination of those justifications for faults that \emph{must necessarily} be resolved. Therefore, for the given example, the only minimal conflict set is $J_B$.

A justification for a given formula (axiom) relative to a KB is a (subset-)minimal subset of the KB that entails the given formula.
\begin{definition}[Justification for a Formula]\label{def:just_sentence}\cite{Kalyanpur.Just.ISWC07}
Let $\mo$ be a KB and $\alpha$ a formula, both over $\mathcal{L}$. Then $J \subseteq \mo$ is called a \emph{justification for $\alpha$ w.r.t.\ $\mo$}, written as $J \in \Just(\alpha,\mo)$, iff $J \models \alpha$ and for all $J' \subset J$ it holds that $J' \not\models \alpha$.
\end{definition}
Since we consider test cases which are \emph{sets of formulas} over $\mathcal{L}$, we generalize the definition of a justification as follows:
\begin{definition}[Justification for a Set of Formulas]\label{def:just_set}
Let $\mo$, $\mo'$ be KBs over $\mathcal{L}$. Then $J \subseteq \mo$ is called a \emph{justification for $\mo'$ w.r.t.\ $\mo$}, written as $J \in \Just(\mo',\mo)$, iff $J \models \mo'$ and for all $J' \subset J$ it holds that $J' \not\models \mo'$.\footnote{Remember that $J \models \mo'$ means that $J \models \tax$ for each $\tax\in\mo'$ (cf.\ Remark~\ref{rem:entailments_as_sets_of_formulas}).}
\end{definition}
In order to express the connection between justifications and conflict sets, we require yet another generalization of this definition. To this end, the following definition characterizes a justification for a set $X$ of KBs relative to a KB $\mo$ as a (subset-)minimal subset of $\mo$ such that this subset entails \emph{some} KB in $X$.
\begin{definition}[Justification for a Set of Sets of Formulas]\label{def:just_set_of_set}
Let $\mo$ be a KB over $\mathcal{L}$ and $X$ a set of KBs over $\mathcal{L}$. Then $J \subseteq \mo$ is called \emph{justification for $X$ w.r.t.\ $\mo$}, written as $J \in \Just(X,\mo)$, iff $J \models \mo'$ for some $\mo' \in X$ and for all $J' \subset J$ it holds that $J' \not\models \mo''$ for all $\mo'' \in X$.
\end{definition}
Based on Definition~\ref{def:just_set_of_set}, the relation between conflict sets and justifications is captured by the following Proposition~\ref{prop:cs_just}. Intuitively, any conflict set w.r.t.\ $\langle\mo,\mb,\Tp,\Tn\rangle_\RQ$ is the part of a justification for a fault that is relevant for the debugging task, where fault refers to an inconsistency (and/or incoherency) and/or a negative test case entailed by $\mo \cup \mb \cup U_\Tp$. Since debugging focuses on the deletion of KB formulas only, ``relevant'' in this context refers to the subset of the justification that does not contain any sentences in $\mb$ and $U_\Tp$, but solely sentences from $\mo$. Importantly, there may be justifications, in general, the relevant subset of which is not a \emph{minimal} conflict set. The reason why this case can arise in spite of the set-minimality of justifications is that the \emph{relevant} part of a justification (for some set of sentences $\mo_1$, e.g.\ a negative test case $\tn_1 \in \Tn$) may be a superset of the \emph{relevant} part of another justification (for some other set of sentences $\mo_2$, e.g.\ another negative test case $\tn_2 \in \Tn$) whereas both justifications are not in a subset-relationship (i.e.\ contain different sentences from $\mb$ and/or $U_\Tp$). This circumstance is illustrated by the following example:
\begin{example}\label{example:min_conflict_set_focuses_on_relevant_part_of_just}
Let a DPI $\langle\mo,\mb,\Tp,\Tn\rangle_\RQ$ be defined as 
\begin{align*}
\mo &:= \setof{B \sqsubseteq E, E \sqsubseteq \exists r.G} \\
\mb &:= \setof{A \sqsubseteq B} \\
\Tn &:= \setof{\setof{A \sqsubseteq E},\setof{B \sqsubseteq \exists r.G}} \\
\Tp &:= \emptyset \\
\RQ &:= \setof{\mbox{consistency}}
\end{align*}
%$\mo := \setof{B \sqsubseteq E, E \sqsubseteq \exists r.G}$ and $\mb := \setof{A \sqsubseteq B}$ and $\Tn := \setof{\setof{A \sqsubseteq E},\setof{B \sqsubseteq \exists r.G}}$ and $\Tp = \emptyset$ and $\RQ = \setof{\mbox{consistency}}$. 
We have that $\mo \cup \mb \cup U_{\Tp}$ is consistent and thus no requirement in $\RQ$ is violated. But, the two negative test cases are both entailed by $\mo \cup \mb \cup U_{\Tp}$ wherefore $\mo$ is invalid w.r.t.\ $\langle\cdot,\mb,\Tp,\Tn\rangle_\RQ$. The set of justifications for the violation of the first negative test case is $J_{\tn_1} = \setof{\setof{A \sqsubseteq B, B \sqsubseteq E}}$; for the second one it is $J_{\tn_2} = \setof{\setof{B \sqsubseteq E, E \sqsubseteq \exists r.G}}$. The relevant subset of the justification $J_1$ in $J_{\tn_1}$ is $J_{1,rel} = \setof{B \sqsubseteq E}$ (since $\setof{A \sqsubseteq B}$ is in $\mb$) whereas the relevant subset of the justification $J_2$ in $J_{\tn_2}$ is $J_{2,rel} = \{B \sqsubseteq E$, $E \sqsubseteq \exists r.G\}$, i.e.\ $J_{1,rel} \subset J_{2,rel}$ despite that there is no set subset-relationship between $J_1$ and $J_2$. Hence, there are two justifications that explain the invalidity of $\mo$ w.r.t.\ $\langle\cdot,\mb,\Tp,\Tn\rangle_\RQ$, but there is only one minimal conflict set $\mc = J_{1,rel}$ w.r.t.\ $\langle\mo,\mb,\Tp,\Tn\rangle_\RQ$.\qed
\end{example}

So, generally, the set of minimal conflict sets w.r.t.\ a DPI is a subset of the set of justifications for faults in $\mo \cup \mb \cup U_\Tp$, which is due to the focus on just the parts of justifications that are relevant for the KB debugging task.
%\fixme{method automatically concentrates on root errors and not derived ones by considering only set-minimal repairs}
\begin{proposition}\label{prop:cs_just}
Let $\langle\mo,\mb,\Tp,\Tn\rangle_\RQ$ be a DPI. Additionally, let 
\begin{enumerate}[(a)]
	\item $X := \setof{\setof{A_i \sqsubseteq \bot}\,|\,A_i \in N_C} \cup \setof{\setof{r_i \sqsubseteq \bot}\,|\,r_i \in N_R} \cup \setof{\setof{\top \sqsubseteq \bot}} \cup \Tn$ \\ if $\RQ = \setof{\mbox{consistency, coherency}}$ and 
	\item \label{prop:cs_just:bullet_consistency} $X := \setof{\setof{\top \sqsubseteq \bot}} \cup \Tn$ if $\RQ = \setof{\mbox{consistency}}$.\footnote{We use DL notation in this proposition since justifications, as argued, are mostly applied to DL KBs. An equivalent formulation of the proposition for FOL or PL is straightforward (cf.\ Example~\ref{example:FOL_to_DL} and Remark~\ref{rem:reduce_conditions_of_dpi_def}). Note that for PL only (\ref{prop:cs_just:bullet_consistency}) is relevant since coherency is not defined for PL. Further, recall that $N_C$ and $N_R$ are defined in Section~\ref{sec:DL}.} 
\end{enumerate}
%(1)~if $\RQ = \setof{\mbox{consistence, coherence}}$, then let $X := \setof{\setof{A_i \sqsubseteq \bot}\,|\,A_i \in N_C} \cup \setof{\setof{\top \sqsubseteq \bot}} \cup \Tn$, and (2)~if $\RQ = \setof{\mbox{consistence}}$, then let $X := \setof{\setof{\top \sqsubseteq \bot}} \cup \Tn$. 
Then the following holds: 
\begin{enumerate}
		\item If $\mc$ is a minimal conflict set w.r.t.\ $\langle\mo,\mb,\Tp,\Tn\rangle_\RQ$, then there is some $J \in \Just(X,\mo\cup\mb\cup U_\Tp)$ such that $(J \cap \mo)\setminus U_\Tp = \mc$.
		\item For all $J \in \Just(X,\mo\cup\mb\cup U_\Tp)$ it is true that $\mc := (J \cap \mo)\setminus U_\Tp$ is a conflict set w.r.t.\ $\langle\mo,\mb,\Tp,\Tn\rangle_\RQ$, but not necessarily a minimal one.
		%\item $\mc$ is a conflict set w.r.t.\ $\langle\mo,\mb,\Tp,\Tn\rangle_\RQ$ iff there is some $J \in \Just(X,\mo\cup\mb\cup U_\Tp)$ such that $J \cap \mo = \mc$.
\end{enumerate}
\end{proposition}
\begin{proof}
\textbf{1):} Assume that $\mc\in\minC_{\langle\mo,\mb,\Tp,\Tn\rangle_\RQ}$ and for all $J \in \Just(X,\mo\cup\mb\cup U_\Tp)$ it holds that $(J \cap \mo) \setminus U_\Tp \neq \mc$. 
%One observation is then that $J$ can be partitioned into $S_1 := [(J \cap \mo)\setminus U_\Tp]$, $S_2:=[J \cap \mb]$ and $S_3:=[J \cap U_\Tp]$. $S_1 \cap S_2 = \emptyset$ since $S_1 \subset\mo$, $S_2 \subset \mb$ and $\mo \cap \mb = \emptyset$. $S_1 \cap S_3 = \emptyset$ since $S_1 \cap U_\Tp = \emptyset$ and $S_3 \subseteq U_\Tp$. $S_2 \cap S_3 = \emptyset$ since
There are two cases to distinguish between: (a)~there is some sentence in $(J \cap \mo) \setminus U_\Tp$ that is not in $\mc$ and (b)~there is some sentence in $\mc$ that is not in $(J \cap \mo) \setminus U_\Tp$.

Let us first assume (a), i.e.\ for all $J\in \Just(X,\mo\cup\mb\cup U_\Tp)$ it holds that there is some sentence $\tax$ in $(J \cap \mo) \setminus U_\Tp$ that is not in $\mc$. 
Additionally, assume there is a $J\in \Just(X,\mo\cup\mb\cup U_\Tp)$ such that $J \subseteq \mc \cup \mb \cup U_\Tp$. We can write $J$ as $J = S_1 \cup S_2 \cup S_3$ for $S_1 := [(J \cap \mo)\setminus U_\Tp]$, $S_2:=[J \cap \mb]$ and $S_3:=[J \cap U_\Tp]$. Since $J = S_1 \cup S_2 \cup S_3 \subseteq \mc \cup \mb \cup U_\Tp$ it must hold in particular that $S_1 \subseteq \mc \cup \mb \cup U_\Tp$ and therefore $\tax \in \mc \cup \mb \cup U_\Tp$. However, $\tax \notin \mc$ by assumption, $\tax \notin \mb$ since $\tax \in \mo$ and $\mb \cap \mo = \emptyset$, and $\tax \notin U_\Tp$ since $\tax \in S_1$ and $S_1 \cap U_\Tp = \emptyset$. This is a contradiction. Hence, for all $J\in \Just(X,\mo\cup\mb\cup U_\Tp)$ it holds that $J \not\subseteq \mc \cup \mb \cup U_\Tp$. Since $X$ captures all $r\in\RQ$ and $\tn\in\Tn$, we can conclude that $\mc$ is not a conflict set w.r.t.\ $\langle\mo,\mb,\Tp,\Tn\rangle_\RQ$ which is a contradiction to $\mc\in\minC_{\langle\mo,\mb,\Tp,\Tn\rangle_\RQ}$.

Let us now assume (b), i.e.\ for all $J\in \Just(X,\mo\cup\mb\cup U_\Tp)$ it holds that there is some sentence $\tax$ in $\mc$ that is not in $(J \cap \mo) \setminus U_\Tp$. Since 
$\mc$ is a conflict set and since $X$ captures all $r\in\RQ$ and $\tn\in\Tn$, we have that $\mc \cup \mb \cup U_\Tp \models \mo'$ for some $\mo' \in X$. So, there must be some $J_0 \in \Just(X,\mo\cup\mb\cup U_\Tp)$ such that $J_0 \subseteq \mc \cup \mb \cup U_\Tp$. As $\mc \in \minC_{\langle\mo,\mb,\Tp,\Tn\rangle_\RQ}$, there cannot be any $J \in \Just(X,\mo\cup\mb\cup U_\Tp)$ with $J \subseteq \mc' \cup \mb \cup U_\Tp$ for arbitrary $\mc' \subset \mc$. This must hold in particular for $J_0$ which implies that $J_0 \cap \mc = \mc$ which is equivalent to $\mc \subseteq J_0$. As (1)~$\mc \subseteq \mo$ (Definition~\ref{def:cs}) and, by Proposition~\ref{prop:cs_properties} and by the fact that $\mc \in \minC_{\langle\mo,\mb,\Tp,\Tn\rangle_\RQ}$, (2)~$\mc \cap U_\Tp = \emptyset$, we can conclude that $\mc \subseteq (J_0 \cap \mo) \setminus U_\Tp$ which is a contradiction since there cannot be a $\tax$ in $\mc$ that is not in $(J_0 \cap \mo) \setminus U_\Tp$.
%
%``$\Leftarrow$'' of (i): Assume there is some $X \in \setof{\setof{A_i \sqsubseteq \bot}\,|\,A_i \in N_C} \cup \setof{\setof{\top \sqsubseteq \bot}} \cup \Tn$ and some $J \in \Just(X,\mo\cup\mb\cup U_\Tp)$. Then $J \models X$ and $J \subseteq \mo\cup\mb\cup U_\Tp$. So, $(J \cap \mo) \cup \mb \cup U_\Tp \supseteq J$ wherefore $(J \cap \mo) \cup \mb \cup U_\Tp \models X$ by monotonicity of $\mathcal{L}$. Consequently (cf. discussion in Section~\ref{chap:KBDebugging}), $(J \cap \mo) \cup \mb \cup U_\Tp$ violates some $r\in\RQ$ or entails some $\tn \in \Tn$ which implies that $(J \cap \mo) \cup U_\Tp \notin \SO_{\langle\mo,\mb,\Tp,\Tn\rangle_\RQ}$. Since $J \cap \mo \subseteq \mo$ is also true, $J\cap\mo \in \allC_{\langle\mo,\mb,\Tp,\Tn\rangle_\RQ}$. Now assume that $J\cap\mo$ is not a minimal conflict set w.r.t.\ $\langle\mo,\mb,\Tp,\Tn\rangle_\RQ$. This means that there is a $\mc' \subset J \cap\mo$ such that $\mc'$ is a conflict set w.r.t.\ $\langle\mo,\mb,\Tp,\Tn\rangle_\RQ$. So, by Corollary~\ref{cor:validonto_cs} and Definition~\ref{def:valid_onto}, $\mc' \cup \mb \cup U_\Tp$ violates some $r\in\RQ$ or entails some $\tn \in\Tn$ which implies that there is some $J' \subset J$ with the property $J' \models$

\textbf{2):} If $J \in \Just(X,\mo\cup\mb\cup U_\Tp)$, then, by Definition~\ref{def:just_set_of_set}, $J \models \mo'$ for some $\mo' \in X$ and $J \subseteq \mo\cup\mb\cup U_\Tp$. So, $[(J \cap \mo)\setminus U_\Tp] \cup \mb \cup U_\Tp = (J \cap \mo) \cup \mb \cup U_\Tp \supseteq J$ wherefore $[(J \cap \mo)\setminus U_\Tp] \cup \mb \cup U_\Tp \models \mo'$ by monotonicity of $\mathcal{L}$. As $\mo' \in X$ and $X$ captures all the reasons why some $r\in\RQ$ or some $\tn\in\Tn$ may not be fulfilled (cf.\ the discussion in Chapter~\ref{chap:KBDebugging}), we have that $[(J \cap \mo)\setminus U_\Tp] \cup \mb \cup U_\Tp$ violates some $r\in\RQ$ or entails some $\tn \in \Tn$. This implies that $[(J \cap \mo)\setminus U_\Tp] \cup U_\Tp \notin \SO_{\langle\mo,\mb,\Tp,\Tn\rangle_\RQ}$. Since $(J \cap \mo)\setminus U_\Tp \subseteq \mo$ is also true, $(J \cap \mo)\setminus U_\Tp \in \allC_{\langle\mo,\mb,\Tp,\Tn\rangle_\RQ}$ by Definition~\ref{def:cs}.

To see that $(J \cap \mo)\setminus U_\Tp \notin \minC_{\langle\mo,\mb,\Tp,\Tn\rangle_\RQ}$ holds in general, reconsider Example~\ref{example:min_conflict_set_focuses_on_relevant_part_of_just} where $(J_2 \cap \mo) \setminus U_{\Tp} = J_2 \supset \mc$ holds for the justification $J_2$ and the minimal conflict set $\mc$.
\end{proof}

\section[The Relation between Conflict Sets and Diagnoses]{The Relation between Conflict Sets and Diagnoses%
\sectionmark{Relation between Conflict Sets and Diagnoses}}
\sectionmark{Relation between Conflict Sets and Diagnoses}
\label{sec:TheRelationBetweenConflictSetsAndDiagnoses}
%%%%%
%\section{The Relation between Conflict Sets and Diagnoses}
%\label{sec:TheRelationBetweenConflictSetsAndDiagnoses}
%%%%%
A minimal conflict set has the property that deletion of any formula in it yields a set of formulas which is correct in the context of $\mb$, $\Tp$, $\Tn$ and $\RQ$.
\begin{proposition}
If $\mc$ is a minimal conflict set w.r.t.\ $\langle\mo,\mb,\Tp,\Tn\rangle_\RQ$, then $\mc'$ is valid w.r.t.\ $\langle\cdot,\mb,\Tp,\Tn\rangle_\RQ$ for each $\mc'\subset\mc$.
\end{proposition}
\begin{proof}
Since $\mc \in \minC_{\langle\mo,\mb,\Tp,\Tn\rangle_\RQ}$, it must hold that $\mc' \notin \allC_{\langle\mo,\mb,\Tp,\Tn\rangle_\RQ}$. Then, by Corollary~\ref{cor:validonto_cs}, $\mc'$ is valid w.r.t.\ $\langle\cdot,\mb,\Tp,\Tn\rangle_\RQ$.
\end{proof}
Hence, by deletion of at least one formula from \emph{each} minimal conflict set w.r.t.\ $\langle\mo,\mb,\Tp,\Tn\rangle_\RQ$, a valid KB can be obtained from $\mo$. Thus, a solution KB $(\mo\setminus\md) \cup U_\Tp$ can be obtained by calculation of a hitting set $\md$ of all minimal conflict sets in $\minC_{\langle\mo,\mb,\Tp,\Tn\rangle_\RQ}$. The Hitting Set problem is defined as follows:
\begin{definition}[Hitting Set]\label{def:hs}
Let $S=\setof{S_1,\dots,S_n}$ be a set of sets. Then, $H$ is called a \emph{hitting set of $S$} 
%or a \emph{hitting set of all $S_i$} 
iff $H \subseteq U_S$ and $H \cap S_i \neq \emptyset$ for all $i=1,\dots,n$. 

A hitting set $H$ of $S$ is \emph{minimal} iff there is no hitting set $H'$ of $S$ such that $H' \subset H$.
\end{definition}
\begin{proposition}\cite{friedrich2005gdm}\label{prop:mindiag_mincs}
A (minimal) diagnosis w.r.t.\ the DPI $\langle\mo,\mb,\Tp,\Tn\rangle_\RQ$ is a (minimal) hitting set of all minimal conflict sets w.r.t.\ $\langle\mo,\mb,\Tp,\Tn\rangle_\RQ$.
\end{proposition}

Now, we want to contemplate two example DPIs and analyze them regarding the their minimal conflict sets and minimal diagnoses:
\begin{example}\label{example:analysis_TabExDpi2}
In this example, we analyze the PL DPI $\tuple{\mo,\mb,\Tp,\Tn}_\RQ$ given by Table~\ref{tab:example2}. There are two minimal conflict sets w.r.t.\ $\tuple{\mo,\mb,\Tp,\Tn}_\RQ$, i.e.\ $\minC_{\tuple{\mo,\mb,\Tp,\Tn}_\RQ} =\setof{\mc_1, \mc_2} = \setof{\tuple{1,2,5},\tuple{1,2,7}}$.\footnote{Please notice that we sometimes write $i$ instead of $\tax_i$ for brevity when it is clear what is meant. We will do so in many other examples as well.}

Why is $\mc_1$ a conflict set w.r.t.\ $\tuple{\mo,\mb,\Tp,\Tn}_\RQ$? We recall Definition~\ref{def:cs} and argue as follows to deduce the entailment $\mc_1 \models \tn_1$ where $\tn_1 \in \Tn$ (left of the colon: the formulas used in the deduction are underlined; right of the colon: the relevant implications are underlined):
\begin{align*}
\underline{\tax_1}:&\quad  \underline{A \;\rightarrow\; E} \\
\underline{\tax_2}:&\quad  X \lor \underline{E \;\rightarrow} \;F \land \underline{Y} \land Z \\
\underline{\tax_5}:&\quad  \underline{Y \;\rightarrow\; \lnot A} \\
\underline{\tax_1, \tax_2, \tax_5}:&\quad  \underline{A \;\rightarrow\; \lnot A} \equiv \underline{\lnot A \lor \lnot A} \equiv \underline{\lnot A} \\
\tn_1 \in \Tn:&\quad  \underline{\lnot A}    \quad\qed
\end{align*}
Minimality of $\mc_2$ is obvious from this argumentation. i.e.\ we cannot deduce $\tn_1$ if any one of the formulas 1, 2 or 5 is omitted, and there is no other fault except for the violation of $\tn_1$.  

Why is $\mc_2$ a conflict set w.r.t.\ $\tuple{\mo,\mb,\Tp,\Tn}_\RQ$? We recall Definition~\ref{def:cs} and argue as follows to deduce the entailment $\mc_2 \cup \mb \models \tn_1$ where $\tn_1 \in \Tn$ (left of the colon: the formulas used in the deduction are underlined; right of the colon: the relevant implications are underlined):
\begin{align*}
\underline{\tax_1}:&\quad  \underline{A \;\rightarrow\; E} \\
\underline{\tax_2}:&\quad  X \lor \underline{E \;\rightarrow} \;F \land Y \land \underline{Z} \\
\underline{\tax_7}:&\quad  \underline{Z \;\rightarrow\; G} \\
(G \;\rightarrow\; \lnot A) \in \mb:&\quad \underline{G \;\rightarrow\; \lnot A} \\ 
\underline{\tax_1, \tax_2, \tax_7},\mb:&\quad  \underline{A \;\rightarrow\; \lnot A} \equiv \underline{\lnot A \lor \lnot A} \equiv \underline{\lnot A} \\
\tn_1 \in \Tn:&\quad  \underline{\lnot A}    \quad\qed
\end{align*}
Minimality of $\mc_2$ is obvious from this argumentation. i.e.\ we cannot deduce $\tn_1$ if any one of the formulas 1, 2 or 7 is omitted, and there is no other fault except for the violation of $\tn_1$.

There are no further minimal conflict sets w.r.t.\ $\tuple{\mo,\mb,\Tp,\Tn}_\RQ$. This is fairly easy to see since
\begin{itemize}
	\item $\mo \cup \mb \cup U_{\Tp} = \mo \cup \mb$ cannot be inconsistent due to the fact that the only negative literal occurring on the righthand side of an implication is $\lnot A$ and $A$ does not occur at the righthand side of any implication in $\mo \cup \mb$,
	\item there is no other way to deduce $\tn_1$ than using a superset of the formulas in $\mc_1$ or $\mc_2$ and
	\item $\tn_1$ is the only negative test case in $\Tn$.
\end{itemize}
Hence, the set of all minimal diagnoses $\minD_{\tuple{\mo,\mb,\Tp,\Tn}_\RQ} =\setof{\md_1, \md_2, \md_3} = \{[1]$, $[2]$, $[5,7]\}$ is obtained by computing all minimal hitting sets of $\minC_{\tuple{\mo,\mb,\Tp,\Tn}_\RQ} =\setof{\mc_1, \mc_2}$ (cf.\ Proposition~\ref{prop:mindiag_mincs}).\qed  
\end{example}

\begin{example}\label{example:analysis_TabExDpi3}
In this example, we analyze the DL DPI $\tuple{\mo,\mb,\Tp,\Tn}_\RQ$ given by Table~\ref{tab:example3}. There are four minimal conflict sets w.r.t.\ $\tuple{\mo,\mb,\Tp,\Tn}_\RQ$, i.e.\ 
\begin{align*}
\minC_{\tuple{\mo,\mb,\Tp,\Tn}_\RQ} =\setof{\mc_1, \mc_2, \mc_3, \mc_4} = \setof{\tuple{1,2,5},\tuple{2,4,6},\tuple{1,3,4},\tuple{1,5,6,8}}
\end{align*}
%$\minC_{\tuple{\mo,\mb,\Tp,\Tn}_\RQ} =\setof{\mc_1, \mc_2, \mc_3, \mc_4} = \setof{\tuple{1,2,5},\tuple{2,4,6},\tuple{1,3,4},\tuple{1,5,6,8}}$.
Why is $\mc_1$ a conflict set w.r.t.\ $\tuple{\mo,\mb,\Tp,\Tn}_\RQ$? We recall Definition~\ref{def:cs} and argue as follows to deduce the entailment $\mc_1 \models \tn_1$ where $\tn_1 \in \Tn$ (left of the colon: the formulas used in the deduction are underlined; right of the colon: the relevant implications are underlined):
\begin{align*}
\underline{\tax_1}:&\quad  \underline{A \;\sqsubseteq\; B} \\
\underline{\tax_2}:&\quad  \underline{B \;\sqsubseteq\; G} \\
\underline{\tax_5}:&\quad  \underline{G \;\sqsubseteq\; K} \\
\underline{\tax_1, \tax_2, \tax_5}:&\quad  \underline{A \;\sqsubseteq\; K} \\
\tn_1  \in \Tn:&\quad  \underline{A \;\sqsubseteq\; K}    \quad\qed
\end{align*}
Minimality of $\mc_1$ is follows from this argumentation. i.e.\ we cannot deduce $\tn_1$ if any one of the formulas 1, 2 or 5 is omitted, and from the fact that we cannot deduce an incoherency ($r_2$), inconsistency ($r_1$) or the entailment of any other negative test case $\tn\in\Tn$ for any KB $\mc'_1 \cup \mb \cup U_{\Tp}$ for any $\mc'_1 \subset \mc_1$.  

Why is $\mc_2$ a conflict set w.r.t.\ $\tuple{\mo,\mb,\Tp,\Tn}_\RQ$? We recall Definition~\ref{def:cs} and argue as follows to deduce 
that $\mc_2 \cup \mb$ is incoherent and thus violates the requirement $r_2 \in \RQ$ (left of the colon: the formulas used in the deduction are underlined; right of the colon: the relevant implications are underlined):
\begin{align*}
\underline{\tax_2}:&\quad  \underline{B \;\sqsubseteq\; G} \\
\underline{\tax_6}:&\quad  \underline{G \;\sqsubseteq\; \exists r.F} \\
(1):\, \underline{\tax_2,\tax_6}:&\quad  \underline{B \;\sqsubseteq\; \exists r.F} \\
\underline{\tax_4}:&\quad  \underline{B \;\sqsubseteq\; \forall r.H} \\
(H \;\sqsubseteq\; \lnot F) \in \mb:&\quad \underline{H \;\sqsubseteq\; \lnot F} \\ 
(2):\, \underline{\tax_4},\mb:&\quad  \underline{B \;\sqsubseteq\; \forall r.\lnot F} \\
(1) \mbox{ and } (2):&\quad  \underline{B \;\sqsubseteq\; \bot} \\
r_1 \in \RQ:&\quad  \underline{B \;\not\sqsubseteq\; \bot}    \quad\qed
\end{align*}
Since we cannot deduce an incoherency ($r_2$), inconsistency ($r_1$) or the entailment of any negative test case $\tn\in\Tn$ for any KB $\mc'_2 \cup \mb \cup U_{\Tp}$ for any $\mc'_2 \subset \mc_2$, the minimality of $\mc_2$ follows.

Why is $\mc_3$ a conflict set w.r.t.\ $\tuple{\mo,\mb,\Tp,\Tn}_\RQ$? We recall Definition~\ref{def:cs} and argue as follows to deduce 
that $\mc_3 \cup \mb \cup U_{\Tp}$ is inconsistent and thus violates the requirement $r_1 \in \RQ$ (left of the colon: the formulas used in the deduction are underlined; right of the colon: the relevant implications are underlined):
\begin{align*}
A(x) \in \mb:&\quad  \underline{A(x)} \\
\underline{\tax_1}:&\quad  \underline{A \;\sqsubseteq\; B} \\
(1):\,\underline{\tax_1},\mb:&\quad  \underline{B(x)} \\
(2):\, \tp_1 \in \Tp:&\quad  \underline{r(x,y)} \\
\underline{\tax_4}:&\quad  \underline{B \;\sqsubseteq\; \forall r.H} \\
(3):\,(1) \mbox{ and } \underline{\tax_4}:&\quad \underline{H(y)} \\
(4):\,\underline{\tax_3}:&\quad  \underline{\lnot H(y)} \\
(3) \mbox{ and } (4):&\quad \mbox{\Lightning} \quad\qed
\end{align*}
No inconsistency ($r_1$) or incoherency ($r_2$) can be derived and no negative test case $\tn \in \Tn$ is entailed from any $\mc'_3 \cup \mb \cup U_{\Tp}$ for $\mc'_3 \subset \mc_3$. Hence, $\mc_3$ is a minimal conflict set w.r.t.\ $\tuple{\mo,\mb,\Tp,\Tn}_\RQ$.

Why is $\mc_4$ a conflict set w.r.t.\ $\tuple{\mo,\mb,\Tp,\Tn}_\RQ$? We recall Definition~\ref{def:cs} and argue as follows to deduce the entailment $\mc_4 \cup \mb \models \tn_2$ where $\tn_2 \in \Tn$ (left of the colon: the formulas used in the deduction are underlined; right of the colon: the relevant implications are underlined):
\begin{align*}
\underline{\tax_8}:&\quad  \underline{L \;\sqsubseteq\; G} \\
\underline{\tax_6}:&\quad  \underline{G \;\sqsubseteq\; \exists r.F} \\
(1):\,\underline{\tax_6,\tax_8}:&\quad  \underline{L \;\sqsubseteq\; \exists r.F} \\
A(x) \in \mb:&\quad  \underline{A(x)} \\
(2):\,\underline{\tax_1},\mb:&\quad  \underline{B(x)} \\
(3):\,\underline{\tax_5}:&\quad  \underline{G \;\sqsubseteq\; K} \\
(1) \mbox{ and } (2) \mbox{ and } (3):&\quad \underline{L \;\sqsubseteq\; \exists r.F,\, B(x),\, G \;\sqsubseteq\; K} \\
\tn_1 \in \Tn:&\quad  \underline{L \;\sqsubseteq\; \exists r.F,\, B(x),\, G \;\sqsubseteq\; K}    \quad\qed
\end{align*}
No inconsistency ($r_1$) or incoherency ($r_2$) can be derived and no negative test case $\tn \in \Tn$ is entailed from any $\mc'_4 \cup \mb \cup U_{\Tp}$ for $\mc'_4 \subset \mc_4$. Thus, $\mc_4$ is a minimal conflict set w.r.t.\ $\tuple{\mo,\mb,\Tp,\Tn}_\RQ$.

Hence, the set of all minimal diagnoses $\minD_{\tuple{\mo,\mb,\Tp,\Tn}_\RQ}$, obtained by computing all minimal hitting sets of $\minC_{\tuple{\mo,\mb,\Tp,\Tn}_\RQ} =\setof{\mc_1, \mc_2, \mc_3, \mc_4}$ (cf.\ Proposition~\ref{prop:mindiag_mincs}), comprises ten minimal diagnoses $\md_i$ for $i = 1,\dots,10$: 
%=\setof{\md_1, \md_2, \md_3, \md_4, \md_5, \md_6, \md_7, \md_8, \md_9, \md_{10}}$ where 
\begin{align*}
\md_1 &= [1,2]   &\md_2 &= [1,4] \\
\md_3 &= [1,6]   &\md_4 &= [2,3,5] \\
\md_5 &= [2,3,6] &\md_6 &= [2,3,8] \\
\md_7 &= [2,4,6] &\md_8 &= [2,4,8] \\
\md_9 &= [3,5,6] &\md_{10} &= [4,5] 
\end{align*}

Although the DPI $\tuple{\mo,\mb,\Tp,\Tn}_\RQ$ is very small in size, i.e.\ number of formulas occurring in it is very small, the reader might agree that it is not trivial on the one hand (1)~to realize which subsets of this KB $\mo$ are (minimal) conflict sets, (2)~to see \emph{that} or \emph{why} a subset of this KB $\mo$ along with the background knowledge $\mb$ and the union of the positive test cases $U_{\Tp}$ is a (minimal) conflict set (cf.\ \cite{Horridge2011b}), and (3)~to assess that there are no further minimal conflict sets w.r.t.\ $\tuple{\mo,\mb,\Tp,\Tn}_\RQ$. This example gives a little bit of an impression that tool assistance in the debugging of KBs is inevitable especially for real-world KBs that are huge in size and/or complex in terms of the expressivity of the used logic or in terms of their ``debugging properties'', i.e.\ large number and/or size of minimal conflict sets and/or minimal diagnoses. 

A means to handle problems~(1) and (3) is provided by some method for the computation of a minimal conflict set (e.g.\ $\scQX$ given by Algorithm~\ref{algo:qx} below, see Section~\ref{sec:cs_comp}) coupled with a hitting set tree algorithm (e.g.\ \textsc{HS} described by Algorithm~\ref{algo:hs} below, see Section~\ref{sec:hs_comp}) for the systematic computation of \emph{different} minimal conflict sets, or other mechanisms such as the ALL\_JUST\_ALG presented in~\cite{Kalyanpur.Just.ISWC07} which computes all justifications for some particular entailment (but, some post-processing of the justifications is necessary to obtain minimal conflict sets, cf.\ Section~\ref{sec:ConflictSetsVersusJustifications}). 
%A technique for dealing with problem~(3) is the construction of a pruned HS-tree $T$ w.r.t.\ $\tuple{\mo,\mb,\Tp,\Tn}_\RQ$ (cf.\ Definition~\ref{def:pruned_hs_tree} and Section~\ref{sec:hs_comp}) to find all hitting sets of $\minC_{\tuple{\mo,\mb,\Tp,\Tn}_\RQ}$. The set of all the labels of non-leaf nodes in the \emph{complete} pruned HS-tree $T$ coincide with $\minC_{\tuple{\mo,\mb,\Tp,\Tn}_\RQ}$.

Problem (2) and its complexity for humans has been studied in \cite{Horridge2011b} with a focus on justifications in DL or OWL KBs. Since a minimal conflict set can be regarded as the relevant (i.e.\ potentially faulty) part of a justification for some undesired entailment (i.e.\ a violated requirement or test case) as we analyzed in Section~\ref{sec:ConflictSetsVersusJustifications},
%minimal conflict sets are equal to or subsets of justifications for undesired entailments as we analyzed in Section~\ref{sec:ConflictSetsVersusJustifications} 
the cognitive complexity model proposed by \cite{Horridge2011b} applies also to minimal conflict sets. Ways to facilitate the understanding of justifications for humans (that might be successfully applied also to conflict sets) have been addressed in \cite{Horridge2010, Horridge2009, Horridge2008}. Moreover, there is an ontology editing browser SWOOP~\cite{Kalyanpur2006b} equipped with a strikeout feature~\cite{Kalyanpur2006a} that highlights parts of justifications that are relevant for the entailment by striking out all irrelevant parts. This is more or less the automation of our analyses of the conflict sets by underlining the relevant parts of the formulas in this example and Example~\ref{example:analysis_TabExDpi2}.\qed   
\end{example}

\renewcommand{\arraystretch}{1.4} 
\begin{table}[h]
	\footnotesize
	\centering
		\rowcolors[]{2}{gray!8}{gray!16} %\arrayrulecolor{black}
		%\extrarowheight10pt
		\begin{tabular}{ c c c c } 
		%\hline\hline
			\rowcolor{gray!40}
			\toprule\addlinespace[0pt]
			$i$ & $\tax_i$ & $\mo$ & $\mb$  \\ \addlinespace[0pt]\midrule\addlinespace[0pt]
			%\hline
			1 & $A \rightarrow E$ & $\bullet$ & 	\\
			%\hline
			2 & $X \lor E \rightarrow F \land Y \land Z$ & $\bullet$ &  	\\
			%\hline
			3 & $F \rightarrow B$ & $\bullet$ &  	\\
			%\hline
			4 & $B \rightarrow X$ & $\bullet$ &  	\\
			%\hline
			5 & $Y \rightarrow \lnot A$ & $\bullet$ &  	\\
			%\hline
			6 & $B \rightarrow Z$ & $\bullet$ &  	\\
			%\hline
			7 & $Z \rightarrow G$ & $\bullet$ & 	\\
			%\hline
			8 & $G \rightarrow \lnot A$ &  & $\bullet$  	\\ 
			%\hline\hline
			\addlinespace[0pt]\bottomrule 
			\rowcolor{gray!40}
			$i$ & \multicolumn{3}{c}{$\tp_i\in\Tp$} \\ \addlinespace[0pt]\midrule\addlinespace[0pt]
			$\times$ & \multicolumn{3}{c}{$\times$} 	\\ \addlinespace[0pt]\toprule\addlinespace[0pt]
			\rowcolor{gray!40}
			$i$ & \multicolumn{3}{c}{$\tn_i\in\Tn$} \\ \addlinespace[0pt]\midrule\addlinespace[0pt]
			$1$ & \multicolumn{3}{c}{$\lnot A$} 	\\ \addlinespace[0pt]\toprule\addlinespace[0pt]
			\rowcolor{gray!40}
			$i$ & \multicolumn{3}{c}{$r_i\in\RQ$} \\ \addlinespace[0pt]\midrule\addlinespace[0pt]
			$1$ & \multicolumn{3}{c}{consistency} \\ \addlinespace[0pt]\bottomrule
			\end{tabular}
		%\begin{tabular}{ c c } 
		%%\hline\hline
			%\rowcolor{gray!40}
			%\toprule\addlinespace[0pt]
			%$i$ & $\tp_i\in\Tp$  \\ \addlinespace[0pt]\midrule\addlinespace[0pt]
			%%\hline
			%$\times$ & $\times$ 	\\ \addlinespace[0pt]\toprule\addlinespace[0pt]
			%%\hline
			%\rowcolor{gray!40}
			%$i$ & $\tn_i\in\Tn$  \\ \addlinespace[0pt]\midrule\addlinespace[0pt]
			%%\hline
			%$\times$ & $\times$  \\
			%\addlinespace[0pt]\bottomrule
		%\end{tabular}	
	\caption{Propositional Logic Example DPI}
	\label{tab:example2}
\end{table}

%%%%%%%%%%%%%%%%%%%%%%% WORKING TABLE start
%\renewcommand{\arraystretch}{1.4} 
%\begin{table}[h]
	%\centering
		%\rowcolors[]{2}{gray!8}{gray!16} %\arrayrulecolor{black}
		%%\extrarowheight10pt
		%\begin{tabular}{ c c c c } 
		%%\hline\hline
			%\rowcolor{gray!40}
			%\toprule\addlinespace[0pt]
			%$\tax$ no. & $\tax$ & $\mo$ & $\mb$ \\ \addlinespace[0pt]\midrule\addlinespace[0pt]
			%%\hline
			%1 & $A \rightarrow E$ & $\bullet$ & 	\\
			%%\hline
			%2 & $X \lor E \rightarrow F \land Y \land Z$ & $\bullet$ & 	\\
			%%\hline
			%3 & $F \rightarrow B$ & $\bullet$ & 	\\
			%%\hline
			%4 & $B \rightarrow X$ & $\bullet$ & 	\\
			%%\hline
			%5 & $Y \rightarrow \lnot A$ & $\bullet$ & 	\\
			%%\hline
			%6 & $B \rightarrow Z$ & $\bullet$ & 	\\
			%%\hline
			%7 & $Z \rightarrow G$ & $\bullet$ & 	\\
			%%\hline
			%8 & $G \rightarrow \lnot A$ &  & $\bullet$	\\ 
			%%\hline\hline
			%\addlinespace[0pt]\bottomrule
		%\end{tabular}
	%\caption{Example DPI}
	%\label{tab:ExampleDPI}
%\end{table}
%%%%%%%%%%%%%%%%%%%%%%%% WORKING TABLE end

\renewcommand{\arraystretch}{1.4} 
\begin{table}[h]
\footnotesize
	\centering
		\rowcolors[]{2}{gray!8}{gray!16} %\arrayrulecolor{black}
		%\extrarowheight10pt
		\begin{tabular}{ c c c c } 
		%\hline\hline
			\rowcolor{gray!40}
			\toprule\addlinespace[0pt]
			$i$ & $\tax_i$ & $\mo$ & $\mb$  \\ \addlinespace[0pt]\midrule\addlinespace[0pt]
			%\hline
			1 & $A \sqsubseteq B$ & $\bullet$ & 	\\
			%\hline
			2 & $B \sqsubseteq G$ & $\bullet$ &  	\\
			%\hline
			3 & $\lnot H(y)$ & $\bullet$ &  	\\
			%\hline
			4 & $B \sqsubseteq \forall r.H$ & $\bullet$ &  	\\
			%\hline
			5 & $G \sqsubseteq K$ & $\bullet$ &  	\\
			%\hline
			6 & $G \sqsubseteq \exists r.F$ & $\bullet$ &  	\\
			%\hline
			7 & $A(x)$ &  & $\bullet$	\\
			%\hline
			8 & $L \sqsubseteq G$ & $\bullet$ &   	\\ 
			%\hline
			9 & $H \sqsubseteq \lnot F$ &  &  $\bullet$ 	\\ 
			%\hline\hline
			\addlinespace[0pt]\bottomrule 
			\rowcolor{gray!40}
			$i$ & \multicolumn{3}{c}{$\tp_i\in\Tp$} \\ \addlinespace[0pt]\midrule\addlinespace[0pt]
			1 & \multicolumn{3}{c}{$r(x,y)$} 	\\ \addlinespace[0pt]\toprule\addlinespace[0pt]
			\rowcolor{gray!40}
			$i$ & \multicolumn{3}{c}{$\tn_i\in\Tn$} \\ \addlinespace[0pt]\midrule\addlinespace[0pt]
			1 & \multicolumn{3}{c}{$A \sqsubseteq K$} 	\\ %\addlinespace[0pt]\toprule\addlinespace[0pt]
			2 & \multicolumn{3}{c}{$L \sqsubseteq \exists r.F, B(x), G \sqsubseteq K$} 	\\ \addlinespace[0pt]\toprule\addlinespace[0pt]
			\rowcolor{gray!40}
			$i$ & \multicolumn{3}{c}{$r_i\in\RQ$} \\ \addlinespace[0pt]\midrule\addlinespace[0pt]
			$1$ &\multicolumn{3}{c}{consistency} \\ 
			$2$ &\multicolumn{3}{c}{coherency} \\ \addlinespace[0pt]\bottomrule
			\end{tabular}
		%\begin{tabular}{ c c } 
		%%\hline\hline
			%\rowcolor{gray!40}
			%\toprule\addlinespace[0pt]
			%$i$ & $\tp_i\in\Tp$  \\ \addlinespace[0pt]\midrule\addlinespace[0pt]
			%%\hline
			%$\times$ & $\times$ 	\\ \addlinespace[0pt]\toprule\addlinespace[0pt]
			%%\hline
			%\rowcolor{gray!40}
			%$i$ & $\tn_i\in\Tn$  \\ \addlinespace[0pt]\midrule\addlinespace[0pt]
			%%\hline
			%$\times$ & $\times$  \\
			%\addlinespace[0pt]\bottomrule
		%\end{tabular}	
	\caption{Description Logic Example DPI}
	\label{tab:example3}
\end{table}

%\paragraph{Methods for Diagnosis Computation}
%\label{sec:MethodsForDiagnosisComputation}

%\noindent\emph{Methods for Diagnosis Computation.} 
\section{Methods for Diagnosis Computation}
\label{sec:MethodsForDiagnosisComputation}
Two common methods employed for the computation of (minimal) diagnoses~\cite{Shchekotykhin2012,Rodler2013} are the
%in this work is accomplished as described in~\cite{Shchekotykhin2012} by means of 
QuickXPlain algorithm~\cite{junker04} (in short $\scQX$) and a hitting set search tree~\cite{Reiter87,greiner1989correction} (in short $\scHS$). Thereby, $\scQX$ serves as a deterministic method for computing one minimal conflict set w.r.t.\ a given DPI $\langle\mo,\mb,\Tp,\Tn\rangle_\RQ$ per call. Since a diagnosis is a hitting set of \emph{all} minimal conflict sets, more than one minimal conflict set is generally required to compute a diagnosis. Due to its determinism, however, $\scQX$ always computes the same minimal conflict set for the same input DPI. Thus, in order to compute different (or all) minimal conflict sets, the input to $\scQX$ needs to be varied accordingly. This can be done by means of $\scHS$ which serves as a search tree to systematically and successively explore all minimal conflict sets w.r.t.\ an initially given DPI. Note that often not all minimal conflict sets w.r.t.\ a DPI are necessary to obtain a minimal diagnosis w.r.t.\ this DPI. This is the case when different minimal conflict sets overlap, i.e.\ have a non-empty intersection. In the extreme case, when all minimal conflict sets w.r.t.\ a DPI share some formulas, then the computation of any single minimal conflict set can suffice to obtain a minimal diagnosis, which is actually even a minimum cardinality diagnosis.

Another approach for computing a minimal conflict set (or justification) is the ``expand-and-shrink'' algorithm presented in \cite{Kalyanpur.Just.ISWC07}. However, empirical evaluations and a theoretical analysis of the best and worst case complexity of the ``expand-and-shrink'' method compared to $\scQX$ performed in \cite{Shchekotykhin2008} revealed that the latter is preferable over the former. 

Also, alternative strategies for the computation of minimal diagnoses have been suggested. One common method is to avoid the indirection of diagnosis computation via minimal conflict sets and use algorithms that determine diagnoses \emph{directly}~\cite{satoh2006}, i.e.\ without the necessity to compute conflict sets. This approach has been applied for the non-interactive debugging of ontologies~\cite{Du2011} and constraints~\cite{Felfernig2011}. In our previous work, we adopted such a direct technique for the interactive debugging of KBs~\cite{Shchekotykhin2014}. The reason why we stick to the conflict-based approach in this work is that we want to present best-first algorithms that figure out minimal diagnoses in descending order of their probability. This is not (systematically) realizable with a direct approach. 
%\fixme{mention the sliding window technique of Kalyanpur and cite Kosyta showing that performance of QX is better.}
%To this end, $\HST$ records the according modifications made to the DPI in order to obtain 

\subsection{Computation of a Minimal Conflict Set}
\label{sec:cs_comp}
The $\scQX$ algorithm takes a DPI $\langle\mo_\orig,\mb_\orig,\Tp,\Tn\rangle_\RQ$ over some monotonic logic $\mathcal{L}$ as input and returns a minimal conflict set $\mc\subseteq\mo_\orig$ w.r.t.\ $\langle\mo_\orig,\mb_\orig,\Tp,\Tn\rangle_\RQ$ as output, if some conflict set exists for the DPI, and 'no conflict' otherwise.

%\noindent\textbf{Monotonic Properties.} 
\paragraph{Monotonic Properties.}
Basically, $\scQX$ can be employed to find for an input set $X$ a set-minimal subset $X_{\min} \subseteq X$ that has a certain property $prop$ 
%of a set $X$ 
for problems of completely different nature such as propositional unsatisfiability or over-constrainedness of constraint satisfaction problems. The only postulated prerequisite for $\scQX$ to work correctly is that $prop$ is a monotonic property. A property is monotonic if and only if the binary function that returns 1 if the property holds for the input set and 0 otherwise is a monotonic function.
\begin{definition}[Binary Monotonic Function]\label{def:monotonic}
Let $X$ be a set and $f:2^X \rightarrow \setof{0,1}$ be a binary function defined for all subsets of $X$. Then, $f$ is monotonic iff 
\begin{align*}
\forall X', X'' \subseteq X:\;\, X' \subset X'' \land f(X') = 1 \implies f(X'')=1
\end{align*}
\end{definition}
So, $prop$ is monotonic iff, given that $prop$ holds for some set $X'$, it follows that $prop$ also holds for any superset $X''$ of $X'$. Note that, by simple logical transformation, an equivalent statement can be derived from Definition~\ref{def:monotonic}; namely that, given that $prop$ does not hold for some set $X''$, it follows that $prop$ does not hold for any subset $X'$ of $X''$ either. 

As inconsistency and incoherency as well as the entailment of some $\tn\in\Tn$ over some monotonic language $\mathcal{L}$ are clearly monotonic properties, the following proposition holds.
\begin{proposition}
Let $\langle\mo,\mb,\Tp,\Tn\rangle_\RQ$ be a DPI. Then, the invalidity of $\mo' \subseteq \mo$ w.r.t.\ $\langle\cdot,\mb,\Tp,\Tn\rangle_\RQ$ (as per Definition~\ref{def:valid_onto}) is a monotonic property.
\end{proposition}
By Corollary~\ref{cor:validonto_cs}, a (minimal) conflict set w.r.t.\ $\langle\mo,\mb,\Tp,\Tn\rangle_\RQ$ is a (minimal) invalid sub-KB of $\mo$ w.r.t.\ $\langle\cdot,\mb,\Tp,\Tn\rangle_\RQ$. Therefore:
\begin{corollary}
Let $\langle\mo,\mb,\Tp,\Tn\rangle_\RQ$ be a DPI. Then, being a conflict set w.r.t.\ $\langle\mo,\mb,\Tp,\Tn\rangle_\RQ$ is a monotonic property.
\end{corollary}
Thus, $\scQX$ is applicable for the problem of finding a minimal conflict set w.r.t.\ a DPI.  
%Other examples of monotonic properties are propositional unsatisfiability, incoherence of description logic KBs or over-constrainedness of constraint satisfaction problems. 
As we shall see later in Chapter~\ref{chap:QueryGeneration}, another monotonic property will enable us to apply $\scQX$ also for the minimization of queries asked to an interacting user in the interactive debugging of KBs.

%\noindent\textbf{How $\scQX$ (Algorithm~\ref{algo:qx}) Works.}
\paragraph{How $\scQX$ (Algorithm~\ref{algo:qx}) Works.}
After verifying that the trivial cases, i.e.\ $\mo_\orig$ is already a valid KB w.r.t.\ $\langle\cdot,\mb_\orig,\Tp,\Tn\rangle_\RQ$ or $\mo_\orig = \emptyset$, are not met, a non-empty minimal conflict set w.r.t.\ $\langle\mo_\orig,\mb_\orig$, $\Tp,\Tn\rangle_\RQ$ must exist. So, the algorithm enters the recursive procedure $\scQX'(\emptyset,\langle\mo_\orig,\mb_\orig,\Tp,\Tn\rangle_\RQ)$. Note that the parameters $\Tp, \Tn, \RQ$ of $\scQX'$ are used for validity tests (\textsc{isKBValid}, line~\ref{algoline:validitytest2}) only and are maintained invariant during the entire recursive execution. 
%At its first call (line~\ref{algoline:call_QX'}), the arguments $\emptyset$ and $\langle\mo_\orig,\mb_\orig,\Tp,\Tn\rangle_\RQ$ are passed to $\scQX'$. So, the first condition tested in line~\ref{algoline:validitytest2} is always false wherefore $\emptyset$ is never returned as a minimal conflict set w.r.t.\ $\langle\mo_\orig,\mb,\Tp,\Tn\rangle_\RQ$. This is correct as $\scQX'$ is called only if there is a non-trivial minimal conflict set w.r.t.\ the original DPI (see above). Next, a test whether $\mo_\orig$ is a singleton set is performed in line~\ref{algoline:test_singleton}. If so, then $\scQX'$ returns $\mo_\orig$ since $\mo_\orig$ must be a minimal conflict set w.r.t.\ $\langle\mo_\orig,\mb_\orig,\Tp,\Tn\rangle_\RQ$. This holds since $\mo_\orig$ is the only non-trivial conflict set. 
In case $\mo_\orig$ is not a singleton, i.e.\ it does not hold for sure that $\mo_\orig$ is an element of a minimal conflict set w.r.t.\ $\langle\mo_\orig,\mb_\orig,\Tp,\Tn\rangle_\RQ$, the idea is to apply a divide-and-conquer strategy to reduce $\mo_\orig$ into two subproblems and solve one subproblem first, i.e.\ find a minimal conflict set for this subproblem, and then the second subproblem. The union of the minimal conflict sets found for the subproblems is then a minimal conflict set for the original problem. This division into smaller problems is recursively executed for each subproblem until the trivial case, i.e.\ the KB of the subproblem that is analyzed includes only one element, occurs. Then this element is an element of a minimal conflict set w.r.t.\ the original problem.

Simply put, one can imagine that $\scQX$ takes $\mo_\orig$, partitions it into $\mo_1$ and $\mo_2$ and first considers the DPI with KB $\mo_2$ and background knowledge $\mb\cup\mo_1$ (line~\ref{algoline:recursive_call1}). If the latter already includes a conflict set (second condition in line~\ref{algoline:validitytest2}), then $\mo_2$ can be safely discarded and does not need to be further considered. 
Instead, $\mo_1$ is further investigated, i.e.\ the DPI with KB $\mo_{1,2}$ and background knowledge $\mb \cup \mo_{1,1}$ where $\mo_{1,1}$ and $\mo_{2,2}$ partition $\mo_1$. Notice that, in this way, $|\mo_2|$ sentences can be dismissed by a single call to \textsc{isKBValid} which is the only function in Algorithm~\ref{algo:qx} that calls a reasoner.

If, on the other hand, $\mb\cup\mo_1$ includes no conflict set, $\mo_2$ is partitioned into $\mo_{2,1}$ and $\mo_{2,2}$ and the two DPIs, the first with KB $\mo_{2,2}$ and background knowledge $\mb\cup\mo_1\cup\mo_{2,1}$ and the second with KB $\mo_{2,1}$ and background knowledge $\mb \cup \mo_1 \cup \mc_{2,2}$, are recursively analyzed where $\mc_{2,2}$ is the result computed for the first DPI.

This recursion is executed until encountering a trivial case, i.e.\ a leaf node of the recursion tree, along each path. Then, the recursion unwinds by building the union of all leaf nodes, i.e.\ the union of all returned sets for subproblems where a trivial case occurred.

\begin{algorithm}
\small
\caption{$\scQX$: Computation of a Minimal Conflict Set} \label{algo:qx}
\begin{algorithmic}[1]
\Require a DPI $\langle\mo_\orig,\mb_\orig,\Tp,\Tn\rangle_\RQ$
%, (optionally) a function $\tax \mapsto w(\tax)$ that assigns a weight $w(\tax)>0$ to each $\tax\in\mo$
\Ensure a minimal conflict set w.r.t.\ $\langle\mo_\orig,\mb_\orig,\Tp,\Tn\rangle_\RQ$ %\fixme{TODO}
%\SetKwFunction{vr}{\normalfont \textsc{verifyRequirements}}
%\SetKwFunction{generate}{\normalfont \textsc{findDiagnosis}}
%\SetKwFunction{split}{\normalfont \textsc{split}}
%\SetKwFunction{get}{\normalfont \textsc{getElements}}
%\SetKwFunction{iscons}{\normalfont \textsc{isConsistent}}
%\SetKwFunction{entails}{\normalfont \textsc{entails}}
%\SetKwBlock{part}{function {\normalfont \textsc{findDiagnosis} ($\mb, \md, \Delta, \mo_\Delta, \mo, \Tne$)} returns {\normalfont a minimal diagnosis $\md$}}{end}
%\SetKwBlock{ver}{function {\normalfont \textsc{verifyRequirements} ($\mb, \md, \mo, \Tne$)} returns {\normalfont \emph{true} or \emph{false}}}{end}

\Procedure{$\scQX$}{$\langle\mo_\orig,\mb_\orig,\Tp,\Tn\rangle_\RQ$}
%\State $\mo' \gets \mo \setminus\mb$
%\State $\mo' \gets \Call{sortAscendingByWeight}{\mo,w}$
%\State $\mb' \gets \mb \cup \bigcup_{\tp\in\Tp} \tp$
\If{$\Call{isKBValid}{\mo_\orig, (\mb_\orig, \Tp, \Tn, \RQ)}$} \label{algoline:validitytest1}
    \State \Return~`no conflict'
\ElsIf{$\mo_\orig = \emptyset$}\label{algoline:O=0}
		\State \Return~$\emptyset$\label{algoline:emptyset}
\Else \State \Return \Call{$\scQX'$}{$\emptyset, \langle\mo_\orig,\mb_\orig,\Tp,\Tn\rangle_\RQ$} \label{algoline:call_QX'}
\EndIf
\EndProcedure

\vspace{10pt}

\Procedure{$\scQX'$}{$\mc,\langle\mo,\mb,\Tp,\Tn\rangle_\RQ$}
\If{$\mc \neq \emptyset \land \neg \Call{isKBValid}{\mb,\langle\cdot,\emptyset,\Tp,\Tn\rangle_\RQ}$}\label{algoline:validitytest2} 
% \fixme{\mb, (\emptyset, \Tp, \Tn, \RQ)} 
	\State \Return $\emptyset$ \label{algoline:return_emptyset}
\EndIf
\If{$|\mo| = 1$}  \label{algoline:test_singleton}              
  \State \Return $\mo$ \label{algoline:return_O}
\EndIf
\State $k \gets \Call{split}{|\mo|}$\label{algoline:split}
\State $\mo_1 \gets \Call{get}{\mo, 1, k}$\label{algoline:get1} 
\State $\mo_2 \gets \Call{get}{\mo, k + 1, |\mo|}$\label{algoline:get2}
\State $\mc_2 \gets \Call{$\scQX'$}{\mo_1, \langle\mo_2,\mb\cup\mo_1,\Tp,\Tn\rangle_\RQ}$ \label{algoline:recursive_call1}
\State $\mc_1 \gets \Call{$\scQX'$}{\mc_2, \langle\mo_1,\mb\cup\mc_2,\Tp,\Tn\rangle_\RQ}$ \label{algoline:recursive_call2}
\State \Return $\mc_1 \cup \mc_2$ \label{algoline:return_upwards}
\EndProcedure

\vspace{10pt}

\Procedure{\textsc{isKBValid}}{$\mo, \langle\cdot,\mb,\Tp,\Tn\rangle_\RQ$}
\State $\mo' \gets \mo \cup \mb \cup \bigcup_{\tp\in\Tp} \tp$
\If{$\neg \Call{verifyReq}{\mo',\RQ}$}
	\State \Return \false
\EndIf
\For{$\tn \in \Tn$}
\If{$\Call{entails}{\mo', \tn}$}
\State \Return \false
\EndIf
\EndFor
\State \Return \true
\EndProcedure

\end{algorithmic}
\normalsize
\end{algorithm}

%
%\tax_1 : {A \sqsubseteq B},$ $\tax_2:B \sqsubseteq E, \tax_3 : B \sqsubseteq D \sqcap \lnot \exists s.C, \tax_4 : C \sqsubseteq \lnot(D \sqcup E), \tax_5 : D \sqsubseteq \lnot B\}

\renewcommand{\arraystretch}{1.4} 
\begin{table}[t]
\footnotesize
	\centering
		\rowcolors[]{2}{gray!8}{gray!16} %\arrayrulecolor{black}
		%\extrarowheight10pt
		\begin{tabular}{ c c c c } 
		%\hline\hline
			\rowcolor{gray!40}
			\toprule\addlinespace[0pt]
			$i$ & $\tax_i$ & $\mo$ & $\mb$  \\ \addlinespace[0pt]\midrule\addlinespace[0pt]
			%\hline
			1 & $A \sqsubseteq B$ & $\bullet$ & 	\\
			%\hline
			2 & $B \sqsubseteq E$ & $\bullet$ &  	\\
			%\hline
			3 & $B \sqsubseteq D \sqcap \lnot \exists s.C$ & $\bullet$ &  	\\
			%\hline
			4 & $C \sqsubseteq \lnot(D \sqcup E)$ & $\bullet$ &  	\\
			%\hline
			5 & $D \sqsubseteq \lnot B$ & $\bullet$ &  	\\
			%\hline
			6 & $A(w)$ &  & $\bullet$   	\\
			%\hline
			7 & $A(v)$ &  & $\bullet$	\\
			%\hline
			8 & $s(v,w)$ & & $\bullet$   	\\ 
			%\hline
			\addlinespace[0pt]\bottomrule 
			\rowcolor{gray!40}
			$i$ & \multicolumn{3}{c}{$\tp_i\in\Tp$} \\ \addlinespace[0pt]\midrule\addlinespace[0pt]
			1 & \multicolumn{3}{c}{$B(w)$} 	\\ \addlinespace[0pt]\toprule\addlinespace[0pt]
			\rowcolor{gray!40}
			$i$ & \multicolumn{3}{c}{$\tn_i\in\Tn$} \\ \addlinespace[0pt]\midrule\addlinespace[0pt]
			1 & \multicolumn{3}{c}{$\lnot C(w)$} 	%\addlinespace[0pt]\toprule\addlinespace[0pt]
			\\ \addlinespace[0pt]\toprule\addlinespace[0pt]
			\rowcolor{gray!40}
			$i$ & \multicolumn{3}{c}{$r_i\in\RQ$} \\ \addlinespace[0pt]\midrule\addlinespace[0pt]
			$1$ &\multicolumn{3}{c}{consistency} \\ 
			$2$ &\multicolumn{3}{c}{coherency} \\ \addlinespace[0pt]\bottomrule
			\end{tabular}
	\caption{Description Logic Example DPI 2}
	\label{tab:example4}
\end{table}

%\noindent\textbf{Example 1.} Consider an ontology $\mo = \mt \cup \ma$ with terminology $\mt:\{
%\tax_1 : {A \sqsubseteq B},$ $\tax_2:B \sqsubseteq E, \tax_3 : B \sqsubseteq D \sqcap \lnot \exists s.C, \tax_4 : C \sqsubseteq \lnot(D \sqcup E), \tax_5 : D \sqsubseteq \lnot B\}$
%\begin{center}
%\begin{tabular}{p{4cm}p{4cm}l}
%$\tax_1 : A \sqsubseteq B$ & $\tax_2 : B \sqsubseteq E$ &  $\tax_3 : B \sqsubseteq D \sqcap \lnot \exists s.C$\\
%$\tax_4 : C \sqsubseteq \lnot(D \sqcup E)$ & $\tax_5 : D \sqsubseteq \lnot B$& \\
%\end{tabular}
%\end{center}
%and assertions $\ma :\{A(w), A(v), s(v,w)\}$.
The next example illustrates one execution of $\scQX$ which computes one minimal conflict set:
\begin{example}\label{example:qx} Let us consider the DL example DPI depicted by Table~\ref{tab:example4}. We will now demonstrate how a minimal conflict set is computed by Algorithm~\ref{algo:qx} (see Fig.~\ref{fig:qx_example}).
Since $\mo$ is not the empty set and not a valid KB w.r.t.\ the DPI (conditions in lines~\ref{algoline:O=0} and \ref{algoline:validitytest1} are false), $\scQX'(\emptyset,\langle\mo,\mb,\Tp,\Tn\rangle_{\RQ})$ is called in line~\ref{algoline:call_QX'}. This call is illustrated by the root node (node \textcircled{\scriptsize 1}) of the recursion tree given in Fig.~\ref{fig:qx_example} (whereas the evaluations made by $\scQX$ prior to this call are not depicted in the figure). Notice that each node in the tree shows only the values of $\mc$, $\mo$ and $\mb$ since all other parameters $\Tp$, $\Tn$ and $\RQ$ are invariant throughout the entire execution of Algorithm~\ref{algo:qx}. 

Due to the fact that $\mc = \emptyset$ and $\mo$ includes five formulas and is thus not a singleton, $\mo=\{\tax_1,\dots$, $\tax_5\}$ is partitioned into $\mo_1=\setof{\tax_1,\tax_2,\tax_3}$ and $\mo_2=\setof{\tax_4,\tax_5}$ and $\scQX'$ is recursively called in line~\ref{algoline:recursive_call1} with parameters $\mc = \mo_1$, $\mo = \mo_2$ and $\mb = \mb\cup\setof{\tax_1,\tax_2,\tax_3}$ which is expressed in the figure by a left branch to node \textcircled{\scriptsize 2}. 
This call, however, returns $\emptyset$ directly since $\mb\cup\setof{\tax_1,\tax_2,\tax_3}$ is already invalid w.r.t.\ $\langle\cdot,\emptyset,\Tp,\Tn\rangle_\RQ$ because $\mb\cup\setof{\tax_1,\tax_2,\tax_3}\cup U_\Tp = \setof{A(w), \underline{A(v)}, \underline{s(v,w)}} \cup \setof{\underline{A \sqsubseteq B}, B \sqsubseteq E, \underline{B \sqsubseteq} D \sqcap \underline{\lnot \exists s.C}} \cup \setof{\setof{B(w)}} \models \setof{\lnot C(w)}$ which is a negative test case, i.e.\ must not be entailed by a solution KB w.r.t.\ the input DPI (the parts of the formulas relevant for the entailment to hold are underlined). Returning $\emptyset$ in this case means discarding $\mo_2 = \setof{\tax_4,\tax_5}$. 

So, the algorithm opens a right branch from the root to node \textcircled{\scriptsize 3} by calling $\scQX'$ (line~\ref{algoline:recursive_call2}) with parameters $\mc = \emptyset$ (result of left branch), $\mo=\mo_1=\setof{\tax_1,\tax_2,\tax_3}$ and $\mb = \mb$. During the execution of this call $\mo_1$ is partitioned into $\setof{\tax_1,\tax_2}$ (left branch to node \textcircled{\scriptsize 4}) and $\setof{\tax_3}$ (right branch to node \textcircled{\scriptsize 5}). In node \textcircled{\scriptsize 4}, it holds that $\mb \cup \setof{\tax_1,\tax_2}$ can be extended to a solution KB by adding $U_\Tp$, i.e.\ $\mb \cup \setof{\tax_1,\tax_2}$ is valid. As it is already an established fact since the execution of node \textcircled{\scriptsize 2} that $\mb \cup \setof{\tax_1,\tax_2,\tax_3}$ is invalid, it must be the case that $\tax_3$ is an element of a minimal conflict set w.r.t.\ the input DPI (as there is a conflict set w.r.t.\ the input DPI in $\setof{\tax_1,\tax_2,\tax_3}$, but there is none in $\setof{\tax_1,\tax_2}$). The algorithm accounts for that by checking whether $\mo$ is a singleton (line~\ref{algoline:test_singleton}) in which case it is guaranteed that $\mo$ is a subset of a minimal conflict set w.r.t.\ the input DPI. So, node \textcircled{\scriptsize 4} returns $\setof{\tax_3}$. This procedure is continued until each path from the root node reaches a node where a trivial case is met. Then the recursion unwinds and, when arrived at the root node, the minimal conflict set $\tuple{\tax_1,\tax_3}$ is returned.

That $\mc := \tuple{\tax_1,\tax_3}$ is indeed a conflict set can be recognized easily by the underlinings in the formulas given before. Minimality is given since $\mb\cup\mc\cup U_\Tp$ is neither inconsistent nor incoherent and the deletion of any formula from $\mc$ breaks the entailment of $\tn_1$. Hence, $\scQX$ has returned a sound output.
\qed

\end{example}

\begin{figure*}[tb]
\setlength{\fboxsep}{2pt}
\centering
\begin{minipage}[c]{0.99\linewidth} 
\small
\begin{displaymath}
\xymatrix{
         \boxed{\emptyset , \setof{\tax_1,\tax_2,\tax_3,\tax_4,\tax_5} , \mb}^{\textcircled{\scriptsize 1}} \ar[dd] \ar[ddr] \ar[r]^{\qquad\qquad\qquad\mbox{output}} &  \tuple{\tax_1,\tax_3}     \\ \\
%depth 1
\boxed{\setof{\tax_1,\tax_2,\tax_3}, \setof{\tax_4,\tax_5}, 
				\mb \cup\setof{\tax_1,\tax_2,\tax_3}}^{\textcircled{\scriptsize 2}} \ar@/^2pc/@{-->}[uu]^-{\setof{}}   & 	
\boxed{\emptyset, \setof{\tax_1,\tax_2,\tax_3}, \mb}^{\textcircled{\scriptsize 3}} \ar[ddl] \ar[dd] \ar@/_2pc/@{-->}[uul]_-{\setof{\tax_1,\tax_3}}  \\ \\
%depth 2
\boxed{\setof{\tax_1,\tax_2}, \setof{\tax_3}, \mb\cup\setof{\tax_1,\tax_2}}^{\textcircled{\scriptsize 4}} \ar@/_2pc/@{-->}[uur]_-{\setof{\tax_3}}  & 
\boxed{\setof{\tax_3}, \setof{\tax_1,\tax_2}, \mb \cup \setof{\tax_3}}^{\textcircled{\scriptsize 5}} \ar[ddl] \ar[dd] \ar@/_2pc/@{-->}[uu]_-{\setof{\tax_1}}  \\ \\
%depth 3
\boxed{\setof{\tax_1}, \setof{\tax_2}, \mb\cup\setof{\tax_3,\tax_1}}^{\textcircled{\scriptsize 6}} \ar@/_2pc/@{-->}[uur]_-{\setof{}} & 
\boxed{\emptyset, \setof{\tax_1}, \mb}^{\textcircled{\scriptsize 7}}  \ar@/_2pc/@{-->}[uu]_-{\setof{\tax_1}}  \\ 
%\ar@/_2pc/@{.>}[uu]|-{\setof{\tax_1}}
}
\end{displaymath}
\end{minipage}
\caption[Recursion Tree for the Computation of a Minimal Conflict Set]{Recursion tree produced during the computation of the minimal conflict set $\tuple{\tax_1,\tax_3}$ w.r.t.\ the DPI shown by Table~\ref{tab:example4} using Algorithm~\ref{algo:qx}. Nodes in the depicted tree represent calls $\scQX'(\mc,\langle\mo,\mb,\Tp,\Tn\rangle_\RQ)$ and are written in format 
%\setlength{\fboxrule}{1pt}
%\framebox[1.1\width]{$\mc,\mo,\mb$}
$\boxed{\mc,\mo,\mb}^{\textcircled{\scriptsize $k$}}$ where $k$ is a counter starting from 1 that indicates when the respective call is made. 
%Notice that $\Tp, \Tn$ and $\RQ$ are not given as these sets are invariants and correspond to $\Tp_{\mathsf{ex}},\Tn_{\mathsf{ex}}$ and $\setof{\mbox{cons,coh}}$, respectively, throughout all computations. 
%The root of the tree corresponds to $\scQX'(\mc,\langle\mo_{\mathsf{ex}},\mb_{\mathsf{ex}},\Tp_{\mathsf{ex}},\Tn_{\mathsf{ex}}\rangle_{\setof{\mbox{cons,coh}}})$. 
A recursive call to $\scQX'$ (left branch = call in line~\ref{algoline:recursive_call1}; right branch = call in line~\ref{algoline:recursive_call2}) is denoted by a normal arrow whereas the return of a set is visualized by a dashed arrow. 
} 
\label{fig:qx_example}
%\vspace{-15pt}
\end{figure*}   

The complexity of Algorithm~\ref{algo:qx} in terms of the number of calls to the function \textsc{isKBValid}, which is the only place in the algorithm where a reasoning service is consulted, is captured by the following proposition.
\begin{proposition}[Complexity of $\scQX$]\label{prop:qx_complexity}\cite{junker04}
Let $\langle\mo,\mb,\Tp,\Tn\rangle_\RQ$ be a DPI and the function \textsc{split} (line~\ref{algoline:split} of Algorithm~\ref{algo:qx}) be defined as $\textsc{split}(n) = \lfloor \frac{n}{2}\rfloor$ where $n$ is a natural number. Then, the worst case number of calls to \textsc{isKBValid} during one call to $\scQX(\langle\mo,\mb,\Tp,\Tn\rangle_\RQ)$ is in $O(|\mc|\log \frac{|\mo|}{|\mc|})$ where $\mc$ is the output of $\scQX(\langle\mo,\mb,\Tp,\Tn\rangle_\RQ)$.

For any other definition of the function \textsc{split}, the worst case number of {\textsc{isKBValid}} invocations gets larger.
\end{proposition}

%\subsection{Correctness of Algorithm~\ref{algo:qx}}
\subsection{Correctness of Conflict Set Computation}
\label{sec:cs_comp_correctness}
This section is dedicated to the proof of correctness of Algorithm~\ref{algo:qx}. First, we show some essential properties of $\scQX$ by various Lemmata which will finally be exploited to demonstrate the overall soundness of $\scQX$. 

%\noindent\emph{Computation of a Minimal Conflict Set.} 
The $\scQX$ algorithm accepts a DPI $\langle\mo_\orig,\mb_\orig,\Tp,\Tn\rangle_\RQ$ over some monotonic language $\mathcal{L}$ 
%and optionally a relation $\prec$ which is a strict partial order between sentences $\tax\in\mo$
%function $w: \mo \rightarrow \mathbb{R}^+$ that assigns a positive real-valued weight $w(\tax)$ to each sentence $\tax\in\mo$ 
%as input parameters. Use cases for the usage of $\prec$ are, for example, if there is a known preference relation between sentences or there are known fault probabilities for sentences.
as input and returns a minimal conflict set $\mc\subseteq\mo_\orig$ w.r.t.\ $\langle\mo_\orig,\mb_\orig,\Tp,\Tn\rangle_\RQ$ as output. 
%In line~\ref{algoline:B+UP}, the algorithm adds $U_\Tp$ to $\mb$ and uses the result $\mb'$ to perform validity checks (lines~\ref{algoline:validitytest1} and \ref{algoline:validitytest2})  
First, the algorithm checks whether $\mo_\orig$ is a valid KB w.r.t.\ the input DPI $\langle\cdot,\mb_\orig,\Tp,\Tn\rangle_\RQ$ (line~\ref{algoline:validitytest1}). If so, there is no conflict set for the DPI by Proposition~\ref{prop:non_validonto_includes_cs} and the algorithm returns 'no conflict'. Otherwise, the test $\mo_\orig = \emptyset$ is performed (line~\ref{algoline:O=0}). If so, then the negative outcome of the validity test executed in line~\ref{algoline:validitytest1} actually means that one of the two criteria of Proposition~\ref{prop:exist_diag} is violated which, by Definition~\ref{def:admissible}, implies that the DPI is not admissible. Invalidity of $\mo_\orig$ w.r.t.\ $\langle\cdot,\mb_\orig,\Tp,\Tn\rangle_\RQ$ and non-admissiblity of $\langle\mo_\orig,\mb_\orig,\Tp,\Tn\rangle_\RQ$ mean that there is only one minimal conflict set $\mc = \emptyset$ by Proposition~\ref{prop:cs_admissible}. Thus, $\emptyset$ is returned in line~\ref{algoline:emptyset}.

\begin{lemma}\label{lem:qx_recursion_finds_cs}
Let $\langle\mo,\mb,\Tp,\Tn\rangle_\RQ$ be an admissible DPI and $\mo$ be invalid w.r.t.\ $\langle\cdot,\mb,\Tp,\Tn\rangle_\RQ$. Then, there is a minimal conflict set $\mc \supset \emptyset$ w.r.t.\ $\langle\mo,\mb,\Tp,\Tn\rangle_\RQ$. 
%such that $\emptyset \subset \mc \subseteq \mo$.
\end{lemma}
\begin{proof}
The proposition is a direct consequence of Proposition~\ref{prop:cs_admissible}.
\end{proof}

So, if both initial tests (lines~\ref{algoline:validitytest1} and \ref{algoline:O=0}) are negative, then, by Lemma~\ref{lem:qx_recursion_finds_cs}, there is a non-trivial minimal conflict set w.r.t.\ $\langle\mo_\orig,\mb_\orig,\Tp,\Tn\rangle_\RQ$ 
%by Proposition~\ref{prop:cs_admissible} 
wherefore the algorithm enters the recursion by a call to the procedure $\scQX'$. 

The argumentation so far proves the following lemma.
\begin{lemma}\label{lem:qx_start_conditions}
\leavevmode
\begin{itemize}
\item \label{lem_enum:no_conflict} $\scQX(\langle\mo,\mb,\Tp,\Tn\rangle_\RQ)$ returns 'no conflict' iff there is no (minimal) conflict w.r.t.\ $\langle\mo,\mb,\Tp,\Tn\rangle_\RQ$.
\item \label{lem_enum:emptyset} $\scQX(\langle\mo,\mb,\Tp,\Tn\rangle_\RQ)$ returns $\emptyset$ iff $\emptyset$ is the only (minimal) conflict w.r.t.\ $\langle\mo,\mb,\Tp,\Tn\rangle_\RQ$.
\item \label{lem_enum:qx'} $\scQX(\langle\mo,\mb,\Tp,\Tn\rangle_\RQ)$ returns $\scQX'(\emptyset,\langle\mo,\mb,\Tp,\Tn\rangle_\RQ)$ iff there is some minimal conflict $\mc \supset \emptyset$ w.r.t.\ $\langle\mo,\mb,\Tp,\Tn\rangle_\RQ$.
\end{itemize}
\end{lemma}

\begin{corollary}\label{cor:call_QX'_admissible}
$\scQX(\langle\mo,\mb,\Tp,\Tn\rangle_\RQ)$ returns $\scQX'(\emptyset,\langle\mo,\mb,\Tp,\Tn\rangle_\RQ)$ iff $\langle\mo,\mb,\Tp,\Tn\rangle_\RQ$ is an admissible DPI.
\end{corollary}
\begin{proof}
By the third proposition of Lemma~\ref{lem:qx_start_conditions} and Proposition~\ref{prop:non_validonto_includes_cs} we have that $\scQX(\langle\mo,\mb,\Tp,\Tn\rangle_\RQ)$ returns $\scQX'(\emptyset,\langle\mo,\mb,\Tp,\Tn\rangle_\RQ)$ iff $\mo$ is invalid w.r.t.\ $\langle\cdot,\mb,\Tp,\Tn\rangle_\RQ$. By Proposition~\ref{prop:cs_admissible}, we can then conclude that $\scQX(\langle\mo,\mb,\Tp,\Tn\rangle_\RQ)$ returns $\scQX'(\emptyset,\langle\mo,\mb,\Tp,\Tn\rangle_\RQ)$ iff $\langle\mo,\mb,\Tp,\Tn\rangle_\RQ$ is an admissible DPI.
\end{proof}

The input arguments (at any call) to $\scQX'$ are (a)~some subset $\mc$ of the original input KB $\mo_\orig$ to $\scQX$ and (b)~a DPI $\langle\mo,\mb,\Tp,\Tn\rangle_\RQ$ where $\mo \subseteq \mo_\orig$ and $\mb \supseteq \mb_\orig$. 

The principle of $\scQX'$ relies on the following fact.
\begin{lemma}\label{lem:qx_recursion_principle}\cite{junker04}
Let $\mo_1, \mo_2$ be a partition of $\mo$. If $\mc_2$ is a minimal conflict set w.r.t.\ $\langle\mo_2, \mb \cup \mo_1,\Tp,\Tn\rangle_\RQ$ and $\mc_1$ is a minimal conflict set w.r.t.\ $\langle\mo_1, \mb \cup \mc_2,\Tp,\Tn\rangle_\RQ$, then $\mc_1 \cup \mc_2$ is a minimal conflict set w.r.t.\ $\langle\mo_1 \cup \mo_2, \mb,\Tp,\Tn\rangle_\RQ = \langle\mo, \mb,\Tp,\Tn\rangle_\RQ$.
\end{lemma}
\begin{proof}
Since $\mc_1$ is a minimal conflict set w.r.t.\ $\langle\mo_1, \mb \cup \mc_2,\Tp,\Tn\rangle_\RQ$, we have that $\mc_1$ is invalid w.r.t.\ $\langle\cdot, \mb \cup \mc_2,\Tp,\Tn\rangle_\RQ$. From that we obtain that $\mc_1 \cup \mc_2$ must be invalid w.r.t.\ $\langle\cdot, \mb,\Tp,\Tn\rangle_\RQ$. Further on, by the fact that $\mo_1,\mo_2$ partition $\mo$ we have that $\mc_1 \subseteq \mo_1 \subseteq \mo$ since $\mc_1$ is a minimal conflict set w.r.t.\ $\langle\mo_1, \mb \cup \mc_2,\Tp,\Tn\rangle_\RQ$ and $\mc_2 \subseteq \mo_2 \subseteq \mo$ since $\mc_2$ is a minimal conflict set w.r.t.\ $\langle\mo_2, \mb \cup \mo_1,\Tp,\Tn\rangle_\RQ$. Consequently, $\mc_1\cup\mc_2 \subseteq\mo$ must be true. So, by Corollary~\ref{cor:validonto_cs}, $\mc_1 \cup \mc_2$ is a conflict set w.r.t.\ $\langle\mo, \mb,\Tp,\Tn\rangle_\RQ$. 

To show the minimality of $\mc_1 \cup \mc_2$, assume that $\mc \subset \mc_1 \cup \mc_2$ is a minimal conflict set w.r.t.\ $\langle\mo$, $\mb,\Tp,\Tn\rangle_\RQ$. Due to $\mo_1 \cap \mo_2 = \emptyset$ and $\mc_1 \subseteq \mo_1$ and $\mc_2 \subseteq \mo_2$, it must hold that $\mc_1\cap\mc_2 = \emptyset$. Thus, (1)~$\mc \cap \mc_1 \subset \mc_1$ or (2)~$\mc \cap \mc_2 \subset \mc_2$. 

Let us assume (1) holds. Then, $\mc$ is invalid w.r.t.\ $\langle\cdot, \mb,\Tp,\Tn\rangle_\RQ$, i.e.\ $\mc \cup \mb \cup U_\Tp = (\mc'_1 \cup \mc_2) \cup \mb \cup U_\Tp = \mc'_1 \cup (\mb \cup \mc_2) \cup U_\Tp$ violates some $r\in\RQ$ or some $\tn\in\Tn$ where $\mc'_1 \subset \mc_1$. This, however, is a contradiction to the minimality of the conflict set $\mc_1$ w.r.t.\ $\langle\mo_1, \mb \cup \mc_2,\Tp,\Tn\rangle_\RQ$.
%by minimality of $\mc_1$, we obtain that $\mc$ cannot be a conflict set w.r.t.\ $\langle\mo_1, \mb \cup \mc_2,\Tp,\Tn\rangle_\RQ$, which is a contradiction. 

Now, let us assume (2) holds. Then, $\mc$ is invalid w.r.t.\ $\langle\cdot, \mb,\Tp,\Tn\rangle_\RQ$, i.e.\ $\mc \cup \mb \cup U_\Tp = (\mc_1 \cup \mc'_2) \cup \mb \cup U_\Tp$ violates some $r\in\RQ$ or some $\tn\in\Tn$ where $\mc'_2 \subset \mc_2$. By monotonicity of $\mathcal{L}$ and $\mc_1 \subseteq \mo_1$, this implies $\mc'_2 \cup (\mo_1 \cup \mb) \cup U_\Tp$ violates some $r\in\RQ$ or some $\tn\in\Tn$, i.e.\ $\mc'_2 \subset \mo_2$ is a conflict set w.r.t.\ 
$\langle\mo_2, \mb \cup \mo_1,\Tp,\Tn\rangle_\RQ$ which is a contradiction due to $\mc'_2 \subset \mc_2$ and the minimality of the conflict set $\mc_2$ w.r.t.\ 
$\langle\mo_2, \mb \cup \mo_1,\Tp,\Tn\rangle_\RQ$.
\end{proof}

$\scQX'(\mc,\langle\mo,\mb,\Tp,\Tn\rangle_\RQ)$ computes a minimal conflict set w.r.t.\ $\langle\mo,\mb,\Tp,\Tn\rangle_\RQ$ in a divide-and-conquer fashion whereby the argument $\mc$ is the set of sentences of $\mo_\orig$ that has been added to $\mb$ in the current iteration. That is, in this iteration $\scQX'$ will output either (1)~$\emptyset$ if the current $\mb$ (which includes $\mc$) already contains a minimal conflict set w.r.t.\ the original DPI $\langle\mo_\orig,\mb_\orig,\Tp,\Tn\rangle_\RQ$ or (2)~a minimal conflict set w.r.t.\ the current DPI $\langle\mo,\mb,\Tp,\Tn\rangle_\RQ$ (i.e.\ a subset of a minimal conflict set w.r.t.\ the original DPI) which does not include any sentence from $\mc$.

\begin{lemma}\label{lem:qx'_termination_etc}
\leavevmode
\begin{enumerate}
	\item \label{lem_enum:c_sub_b} For each call $\scQX'(\mc,\langle\mo,\mb,\Tp,\Tn\rangle_\RQ)$ within Algorithm~\ref{algo:qx} it holds that $\mc \subseteq \mb$.
	\item \label{lem_enum:c_neq_emptyset} If $\scQX'(\mc,\langle\mo,\mb,\Tp,\Tn\rangle_\RQ)$ is called in line~\ref{algoline:recursive_call1} of Algorithm~\ref{algo:qx}, $\mc \neq \emptyset$ holds.
	\item \label{lem_enum:return_emptyset} If $\scQX'(\mc,\langle\mo,\mb,\Tp,\Tn\rangle_\RQ)$ returns $\emptyset$, then there is some non-empty minimal conflict set w.r.t.\ $\langle\mc,\mb\setminus\mc,\Tp,\Tn\rangle_\RQ$.
	\item \label{lem_enum:emptyset_only_cs} If $\scQX'(\mc,\langle\mo,\mb,\Tp,\Tn\rangle_\RQ)$ returns $\emptyset$, then $\emptyset$ is the only minimal conflict set w.r.t.\ $\langle\mo,\mb,\Tp,\Tn\rangle_\RQ$.
	%\item If $\scQX'(\mc,\langle\mo,\mb,\Tp,\Tn\rangle_\RQ)$ is called, then there is some non-empty minimal conflict set w.r.t.\ $\langle\mo\cup\mc,\mb\setminus\mc,\Tp,\Tn\rangle_\RQ$.
	%\item If $\scQX'(\mc,\langle\mo,\mb,\Tp,\Tn\rangle_\RQ)$ returns $\mo$ with $|\mo|=1$, then $\mo$ is a minimal conflict set w.r.t.\ $\langle\mo,\mb,\Tp,\Tn\rangle_\RQ$.
	\item \label{lem_enum:terminates} $\scQX'(\mc,\langle\mo,\mb,\Tp,\Tn\rangle_\RQ)$ terminates.
	
	%\item If $\scQX'(\mc,\langle\mo,\mb,\Tp,\Tn\rangle_\RQ)$ is in line~\ref{algoline:split}, then there is a minimal conflict set w.r.t.\ $\langle\mo,\mb,\Tp,\Tn\rangle_\RQ$.
	%\item After dismissal of some 
	%$\mo \subset \mo_\orig$
	%$\mo$, i.e.\ the call to $\scQX'(\mc,\langle\mo,\mb,\Tp,\Tn\rangle_\RQ)$ in line~\ref{algoline:recursive_call1} returns $\emptyset$, $\mo_\orig\setminus\mo$ is invalid w.r.t.\  $\langle\mo_\orig\setminus\mo,\mb_\orig,\Tp,\Tn\rangle_\RQ$.
\end{enumerate}
\end{lemma}
\begin{proof}
\leavevmode\\
\indent 1): There are three situations when $\scQX'(\mc,\langle\mo,\mb,\Tp,\Tn\rangle_\RQ)$ is called within Algorithm~\ref{algo:qx}, namely in lines~\ref{algoline:call_QX'}, \ref{algoline:recursive_call1} and \ref{algoline:recursive_call2}. In line~\ref{algoline:call_QX'}, $\mc := \emptyset \subseteq \mb$ holds. In line~\ref{algoline:recursive_call1}, $\mc:=\mo_1 \subseteq \mb \cup \mo_1 =: \mb$ holds. In line~\ref{algoline:recursive_call2}, $\mc:=\mc_2 \subseteq \mb \cup \mc_2 =: \mb$ holds.

2): In line~\ref{algoline:recursive_call1}, $\scQX'$ is called with $\mc:=\mo_1$, which is always not the empty set due to the definition of the \textsc{split} function in line~\ref{algoline:split} that is used to extract $\mo_1$ from $\mo$.

3): The first observation is that $\scQX'(\mc,\langle\mo,\mb,\Tp,\Tn\rangle_\RQ)$ cannot return $\emptyset$ if $\mc = \emptyset$ as in this case the first condition in line~\ref{algoline:validitytest2} is not met. Thus, in particular, $\scQX'$ cannot return $\emptyset$ if called in line~\ref{algoline:call_QX'}.

So, $\emptyset$ can be returned by $\scQX'(\mc,\langle\mo,\mb,\Tp,\Tn\rangle_\RQ)$ only if it is called (1)~in line~\ref{algoline:recursive_call1} or (2)~in line~\ref{algoline:recursive_call2}. 

If $\scQX'(\mc,\langle\mo,\mb,\Tp,\Tn\rangle_\RQ)$ returns $\emptyset$, then $\mc \neq \emptyset$ and $\mb$ is invalid w.r.t.\ $\langle\cdot,\emptyset,\Tp,\Tn\rangle_\RQ$ (line~\ref{algoline:validitytest2}), i.e.\ $\mb$ contains a minimal conflict set w.r.t.\ $\tuple{\mb,\emptyset, \Tp, \Tn}_\RQ$ which is non-empty by Proposition~\ref{prop:cs_admissible} since $\tuple{\mb,\emptyset, \Tp, \Tn}_\RQ$ is an admissible DPI by admissibility of the input DPI and the invariance of $\Tp, \Tn, \RQ$ throughout $\scQX'$. Additionally, $\mc \subseteq \mb$ holds by the first proposition of this lemma. Now, assume that there is no non-empty (minimal) conflict set w.r.t.\ $\langle\mc,\mb\setminus\mc,\Tp,\Tn\rangle_\RQ$. Then, for each minimal conflict set $\mc'$ (which we know is non-empty) w.r.t.\ $\langle\mb,\emptyset,\Tp,\Tn\rangle_\RQ$ it must hold that $\mc \cap \mc' = \emptyset$, i.e.\ there is already a non-empty minimal conflict set w.r.t.\ $\langle\mb\setminus\mc,\emptyset,\Tp,\Tn\rangle_\RQ$. 

Case~(1): Let us assume first that the call to $\scQX'$ was made in line~\ref{algoline:recursive_call1}. Then, before this call to $\scQX'$, $\mb$ was exactly $\mb\setminus\mc$. By the second proposition of this lemma, $\mc\neq\emptyset$ as $\scQX'$ was called in line~\ref{algoline:recursive_call1}. Thus, before the current call to $\scQX'$, the algorithm must have already returned $\emptyset$ (both conditions in line~\ref{algoline:validitytest2} are met) in line~\ref{algoline:return_emptyset} which is a contradiction to the assumption that $\scQX'(\mc,\langle\mo,\mb,\Tp,\Tn\rangle_\RQ)$ was called in line~\ref{algoline:recursive_call1}.

Case~(2): Now, assume that the call to 
%$\scQX'(\mc,\langle\mo,\mb,\Tp,\Tn\rangle_\RQ)$
 %= 
$\scQX'(\mc_2,\langle\mo_1,\mb\cup\mc_2,\Tp,\Tn\rangle_\RQ)$ 
was made in line~\ref{algoline:recursive_call2}. Then $\mc_2$ is the result of the call to $\scQX'(\mo_1,\langle\mo_2,\mb\cup\mo_1,\Tp,\Tn\rangle_\RQ)$ in line~\ref{algoline:recursive_call1}. By the argumentation above, we have that $\mc_2 \neq \emptyset$ and there is a non-empty minimal conflict set w.r.t.\ $\langle\mb\cup\mc_2,\emptyset,\Tp,\Tn\rangle_\RQ$. Moreover, we have that there is a non-empty minimal conflict set w.r.t.\ $\langle\mb,\emptyset,\Tp,\Tn\rangle_\RQ$. However, as $\scQX'(\mo_1,\langle\mo_2,\mb\cup\mo_1,\Tp,\Tn\rangle_\RQ)$ in line~\ref{algoline:recursive_call1} did not return $\emptyset$ and $\mo_1 \neq \emptyset$ by the second proposition of this lemma, it must hold that $\mb\cup\mo_1$ is valid w.r.t.\ $\langle\cdot,\emptyset,\Tp,\Tn\rangle_\RQ$, i.e.\ there is no (minimal) conflict set w.r.t.\ $\langle\mb\cup\mo_1,\emptyset,\Tp,\Tn\rangle_\RQ$. By monotonicity of $\mathcal{L}$, this is a contradiction to the fact that there is a non-empty minimal conflict set w.r.t.\ $\langle\mb,\emptyset,\Tp,\Tn\rangle_\RQ$.

4): Assume $\scQX'(\mc,\langle\mo,\mb,\Tp,\Tn\rangle_\RQ)$ returns $\emptyset$ and there is some non-empty minimal conflict set w.r.t.\ $\langle\mo,\mb,\Tp,\Tn\rangle_\RQ$. Since $\emptyset$ is returned, both conditions in line~\ref{algoline:validitytest1} must be met, i.e.\ 
%$\mc \neq \emptyset$ and 
in particular $\mb$ must be invalid w.r.t.\ $\langle\cdot,\emptyset,\Tp,\Tn\rangle_\RQ$ which means that $\langle\mo,\mb,\Tp,\Tn\rangle_\RQ$ is not admissible. By Proposition~\ref{prop:cs_admissible}, there cannot be a non-empty (minimal) conflict set w.r.t.\ $\langle\mo,\mb,\Tp,\Tn\rangle_\RQ$. This yields a contradiction.

5): $\scQX'(\mc,\langle\mo,\mb,\Tp,\Tn\rangle_\RQ)$ either returns $\emptyset$ in line~\ref{algoline:return_emptyset} iff the conditions in line~\ref{algoline:validitytest2} are met or otherwise returns $\mo$ in line~\ref{algoline:return_O} iff $|\mo|=1$ or otherwise calls itself recursively in lines~\ref{algoline:recursive_call1} and \ref{algoline:recursive_call2}. However, for each recursive call $\scQX'(\mc',\langle\mo',\mb',\Tp,\Tn\rangle_\RQ)$ within $\scQX'(\mc,\langle\mo,\mb,\Tp,\Tn\rangle_\RQ)$ it holds that $\mo' \subset \mo$ as $\mo' \in \setof{\mo_1,\mo_2}$ and $\mo_1, \mo_2 \subset \mo$ due to the definition of the \textsc{split} function in line~\ref{algoline:split} that is used to compute $\mo_1$ and $\mo_2$ from $\mo$ in lines~\ref{algoline:get1} and \ref{algoline:get2}. Hence, each recursive call must finally reach the stopping criterion $|\mo|=1$ and return $\mo$ if it does not reach the stopping criterion in line~\ref{algoline:validitytest2} before.
\end{proof}
\begin{lemma}\label{lem:one_rec_call_adm}
Let $\langle\mo,\mb,\Tp,\Tn\rangle_\RQ$ be an admissible DPI. If $\scQX'(\mc,\langle\mo,\mb,\Tp,\Tn\rangle_\RQ)$ is called, then at least one of the immediate recursive calls of $\scQX'$ in line~\ref{algoline:recursive_call1} or line~\ref{algoline:recursive_call2} is given an admissible DPI as argument. 
\end{lemma}
\begin{proof}
Let us assume that $\langle\mo,\mb,\Tp,\Tn\rangle_\RQ$ is an admissible DPI. Within $\scQX'(\mc$, $\langle\mo,\mb$, $\Tp,\Tn\rangle_\RQ)$, the immediate recursive call is $\scQX'(\mo_1,\langle\mo_2,\mb\cup\mo_1,\Tp,\Tn\rangle_\RQ)$ in line~\ref{algoline:recursive_call1} and $\scQX'(\mc_2,\langle\mo_1,\mb\cup\mc_2,\Tp,\Tn\rangle_\RQ)$ in line~\ref{algoline:recursive_call2} where $\mo_1,\mo_2$ is a partition of $\mo$ and $\mc_2$ is the result of $\scQX'(\mo_1,\langle\mo_2,\mb\cup\mo_1,\Tp,\Tn\rangle_\RQ)$. If $\langle\mo_2,\mb\cup\mo_1,\Tp,\Tn\rangle_\RQ$ is admissible, then the proposition of the lemma is fulfilled. So, assume that that $\langle\mo_2,\mb\cup\mo_1,\Tp,\Tn\rangle_\RQ$ is not admissible. Due to this non-admissibility, it must hold that $\mb\cup\mo_1$ is invalid w.r.t.\ $\langle\cdot,\emptyset,\Tp,\Tn\rangle_\RQ$, so the second condition in line~\ref{algoline:validitytest1} is met. As the call to $\scQX'(\mo_1,\langle\mo_2,\mb\cup\mo_1,\Tp,\Tn\rangle_\RQ)$ was made in line~\ref{algoline:recursive_call1}, it must be true by Lemma~\ref{lem:qx'_termination_etc}, prop.~\ref{lem_enum:c_neq_emptyset} that $\mo_1 \neq \emptyset$ wherefore the first condition in line~\ref{algoline:validitytest1} is met as well. Thus, the result of the call of $\scQX'$ in line~\ref{algoline:recursive_call1} must be $\emptyset$. So, the call of $\scQX'$ in line~\ref{algoline:recursive_call2} looks like $\scQX'(\emptyset,\langle\mo_1,\mb,\Tp,\Tn\rangle_\RQ)$. However, the DPIs $\langle\mo_1,\mb,\Tp,\Tn\rangle_\RQ$ and $\langle\mo,\mb,\Tp,\Tn\rangle_\RQ$ are identical except for the first entries, i.e.\ $\mo_1$ and $\mo$. We know that the latter DPI is admissible. Due to the fact that admissibility of a DPI is defined independently of the KB (the first entry of the DPI tuple), we have that $\langle\mo_1,\mb,\Tp,\Tn\rangle_\RQ$ must be admissible. This completes the proof.
\end{proof}
As long as the algorithm goes downwards in the recursion tree (and has never gone upwards), (1)~the invariant that a minimal conflict set exists for each recursive call to $\scQX'$ holds, (2)~each call to $\scQX'$ that returns, returns a singleton or empty set and (3)~the two calls to $\scQX'$ immediately before going upwards in the recursion tree for the first time must both return either a singleton or an empty set.

\begin{lemma}[QX: Downwards Correctness]\label{lem:qx_downwards}
Let $\langle\mo,\mb,\Tp,\Tn\rangle_\RQ$ be an admissible DPI and let
	%$\scQX'(\mc,\langle\mo,\mb,\Tp,\Tn\rangle_\RQ)$ where 
	there be a non-empty minimal conflict set w.r.t.\ $\langle\mo,\mb,\Tp,\Tn\rangle_\RQ$. Then, the following propositions hold: 
\begin{enumerate}	
	\item \label{lem_enum:downwards_1} Before line~\ref{algoline:return_upwards} has ever been reached during the execution of $\scQX'(\mc,\langle\mo$, $\mb,\Tp$, $\Tn\rangle_\RQ)$, the following holds: If some call to $\scQX'(\mc',\langle\mo',\mb',\Tp,\Tn\rangle_\RQ)$ returns a set $S$, then $S = \emptyset$ or $|S| = 1$.
	\item \label{lem_enum:downwards_2} Before line~\ref{algoline:return_upwards} has ever been reached during the execution of $\scQX'(\mc,\langle\mo$, $\mb,\Tp$, $\Tn\rangle_\RQ)$, the following holds: If $\scQX'(\mc',\langle\mo',\mb',\Tp,\Tn\rangle_\RQ)$ is recursively called, then there is some non-empty minimal conflict set w.r.t.\ $\langle\mo'\cup\mc',\mb'\setminus\mc',\Tp,\Tn\rangle_\RQ$.
%Before line~\ref{algoline:return_upwards} has ever been reached during the execution of Algorithm~\ref{algo:qx}, the following holds: If $\scQX'(\mc,\langle\mo,\mb,\Tp,\Tn\rangle_\RQ)$ is called, then there is some minimal conflict set w.r.t.\ $\langle\mo\cup\mc,\mb\setminus\mc,\Tp,\Tn\rangle_\RQ$.
	\item \label{lem_enum:downwards_3} Before line~\ref{algoline:return_upwards} has ever been reached during the execution of $\scQX'(\mc,\langle\mo,\mb,\Tp$, $\Tn\rangle_\RQ)$, the following holds: If some call to $\scQX'(\mc',\langle\mo',\mb',\Tp,\Tn\rangle_\RQ)$ returns a set $S$, then 
	%:
	%\begin{itemize}
		%\item $S = \emptyset$ or $|S| = 1$ \quad and
		%\item 
		$S$ is a minimal conflict set w.r.t.\ $\langle\mo,\mb,\Tp,\Tn\rangle_\RQ$.
	%\end{itemize}
	%it returns $\emptyset$ or some $\mo$ with $|\mo|=1$.
	\item \label{lem_enum:downwards_4} When line~\ref{algoline:return_upwards} is reached for the first time, each of the calls to $\scQX'$ immediately before in lines~\ref{algoline:recursive_call1} and \ref{algoline:recursive_call2} must have returned $\emptyset$ or some $\mo$ with $|\mo|=1$.
\end{enumerate}
\end{lemma}
\begin{proof}
\leavevmode\\
\indent 1): Assume the opposite, i.e.\ some call to $\scQX'(\mc',\langle\mo',\mb',\Tp,\Tn\rangle_\RQ)$ returns a set $S$ with $|S| > 1$ before line~\ref{algoline:return_upwards} has ever been reached. There are three places where $\scQX'$ can return, namely in line~\ref{algoline:return_emptyset}, in line~\ref{algoline:return_O} or in line~\ref{algoline:return_upwards}. However, in line~\ref{algoline:return_emptyset}, only $\emptyset$ and in line~\ref{algoline:return_O} only a singleton set can be returned. That is, $S$ must be returned in line~\ref{algoline:return_upwards} which is a contradiction to the assumption that line~\ref{algoline:return_upwards} has not yet been reached.

2): \emph{Induction Base:} The first recursive call $\scQX'(\mc',\langle\mo',\mb',\Tp,\Tn\rangle_\RQ)$ can only occur at line~\ref{algoline:recursive_call1} where $\mc' = \mo_1$, $\mo'=\mo_2$ and $\mb' = \mb \cup \mo_1$ and $\mo_1,\mo_2$ is a partition of $\mo$ as per the definition of the \textsc{split} and \textsc{get} functions in lines~\ref{algoline:split}-\ref{algoline:get2}. So, $\mo'\cup\mc' = \mo$ and $\mb'\setminus\mc' = \mb$. The latter holds since $\mc' \subseteq \mo$ and for each DPI $\mo \cap \mb =\emptyset$ holds by Definition~\ref{def:dpi}. As there is a non-empty minimal conflict set w.r.t.\ $\langle\mo,\mb,\Tp,\Tn\rangle_\RQ$ we have that there is a non-empty minimal conflict set w.r.t.\ $\langle\mo'\cup\mc',\mb'\setminus\mc',\Tp,\Tn\rangle_\RQ$ by the fact that $\langle\mo,\mb,\Tp,\Tn\rangle_\RQ=\langle\mo'\cup\mc',\mb'\setminus\mc',\Tp,\Tn\rangle_\RQ$. Thus, the existence of a non-empty minimal conflict set w.r.t.\ $\langle\mo'\cup\mc',\mb'\setminus\mc',\Tp,\Tn\rangle_\RQ$ is given 
%for the first recursive call to $\scQX'$ and consequently also before any further 
during the execution of the first
recursive call to $\scQX'$.

\emph{Induction Assumption:} Now, let us assume that the existence of a non-empty minimal conflict set w.r.t.\ $\langle\mo\cup\mc,\mb\setminus\mc,\Tp,\Tn\rangle_\RQ$ is given during some call $\scQX'(\mc,\langle\mo,\mb,\Tp$, $\Tn\rangle_\RQ)$. The goal is now to show that the existence of a non-empty minimal conflict set w.r.t.\ $\langle\mo'\cup\mc',\mb'\setminus\mc',\Tp,\Tn\rangle_\RQ$ is given during any recursive call $\scQX'(\mc',\langle\mo',\mb',\Tp,\Tn\rangle_\RQ)$ that is invoked during execution of $\scQX'(\mc,\langle\mo,\mb,\Tp,\Tn\rangle_\RQ)$.

\emph{Induction Step:} Now, there are three cases where this recursive call to $\scQX'$ can take place, namely (1)~in line~\ref{algoline:recursive_call1}, (2)~in line~\ref{algoline:recursive_call2} where the result of $\scQX'$ in line~\ref{algoline:recursive_call1} is $\mc_2 = \emptyset$ and (3)~in line~\ref{algoline:recursive_call2} where the result of $\scQX'$ in line~\ref{algoline:recursive_call1} is some $\mc_2$ with $|\mc_2|=1$. The case where some $\mc_2$ with $|\mc_2|>1$ is returned by $\scQX'$ in line~\ref{algoline:recursive_call1}, is impossible due to the assumption that line~\ref{algoline:return_upwards} has not yet been reached and the first proposition of this lemma.

Case (1): Let us assume that the call $\scQX'(\mc',\langle\mo',\mb',\Tp,\Tn\rangle_\RQ)$ is made in line~\ref{algoline:recursive_call1}. Since that call is made within $\scQX'(\mc,\langle\mo,\mb,\Tp,\Tn\rangle_\RQ)$, it must hold that some condition in line~\ref{algoline:validitytest1} during $\scQX'(\mc,\langle\mo,\mb,\Tp$, $\Tn\rangle_\RQ)$ is violated, as otherwise a return would have taken place in line~\ref{algoline:return_emptyset} which is a contradiction to the assumption that $\scQX'(\mc',\langle\mo',\mb',\Tp,\Tn\rangle_\RQ)$ is called in line~\ref{algoline:recursive_call1}. 

Let us first assume that $\mc = \emptyset$ holds. In this case, the first condition in line~\ref{algoline:validitytest1} is violated and, by the \emph{Induction Assumption}, it is true that there is a non-empty minimal conflict set w.r.t.\ the DPI $\langle\mo\cup\mc,\mb\setminus\mc,\Tp,\Tn\rangle_\RQ$ which is equal to the DPI $\langle\mo,\mb,\Tp,\Tn\rangle_\RQ$ by $\mc = \emptyset$. So, an equal argumentation to the one of the \emph{Induction Base} can be applied to derive that there is a non-empty minimal conflict set w.r.t.\ $\langle\mo'\cup\mc',\mb'\setminus\mc',\Tp,\Tn\rangle_\RQ$.

If $\mc \neq \emptyset$ holds, on the other hand, then the first condition in line~\ref{algoline:validitytest1} is satisfied wherefore the second condition in line~\ref{algoline:validitytest1} must be violated. That is, there is no conflict set w.r.t.\ $\langle\mb,\emptyset,\Tp,\Tn\rangle_\RQ$. As there is a non-empty minimal conflict set w.r.t.\ $\langle\mo\cup\mc,\mb\setminus\mc,\Tp,\Tn\rangle_\RQ$ by the \emph{Induction Assumption}, $\mc \subseteq \mb$ by Lemma~\ref{lem:qx'_termination_etc}, prop.~\ref{lem_enum:c_sub_b} and $|\mo|\geq 2$ by the fact that there was no return in line~\ref{algoline:return_O}, there must be a non-empty minimal conflict set w.r.t.\ $\langle\mo,\mb,\Tp,\Tn\rangle_\RQ$. Again, an equal argumentation to the one of the \emph{Induction Base} can be applied to derive that there is a non-empty minimal conflict set w.r.t.\ $\langle\mo'\cup\mc',\mb'\setminus\mc',\Tp,\Tn\rangle_\RQ$.
%However, in $\scQX'(\mc',\langle\mo',\mb',\Tp,\Tn\rangle_\RQ)$ that is called in line~\ref{algoline:recursive_call1}, it holds that $\mc' = \mo_1$,
%
%is $\scQX'(\mo_1,\langle\mo_2,\mb\cup\mo_1,\Tp,\Tn\rangle_\RQ)$. Then, either $$ 
%
%it holds that $\langle\mo,\mb,\Tp,\Tn\rangle_\RQ$ is admissible

Case (2): Here, we assume that the recursive call $\scQX'(\mc',\langle\mo',\mb',\Tp,\Tn\rangle_\RQ)$ is made in line~\ref{algoline:recursive_call2} and the result of $\scQX'$ in line~\ref{algoline:recursive_call1} is $\mc_2 = \emptyset$. So, it holds that $\mc' = \mc_2 = \emptyset$, $\mo' = \mo_1$ and $\mb' = \mb$, i.e.\ the recursive call can be written as $\scQX'(\emptyset,\langle\mo_1,\mb,\Tp,\Tn\rangle_\RQ)$. By the fact that $\scQX'(\mo_1,\langle\mo_2,\mb\cup\mo_1,\Tp,\Tn\rangle_\RQ)$ called in line~\ref{algoline:recursive_call1} returned $\emptyset$, both conditions in line~\ref{algoline:validitytest1} during $\scQX'(\mo_1,\langle\mo_2,\mb\cup\mo_1,\Tp,\Tn\rangle_\RQ)$ must have been met. Thus, in particular the existence of a non-empty minimal conflict set w.r.t.\ $\langle\mb\cup\mo_1,\emptyset,\Tp,\Tn\rangle_\RQ$ must be given. Further on, by the \emph{Induction Assumption} there is a non-empty minimal conflict set w.r.t.\ $\langle\mc\cup\mo,\mb\setminus\mc,\Tp,\Tn\rangle_\RQ$.
%, i.e.\ in particular there is no non-empty conflict set w.r.t.\ $\langle\mb\setminus\mc,\emptyset,\Tp,\Tn\rangle_\RQ$.

Let us first assume $\mc = \emptyset$. In this case $\langle\mc\cup\mo,\mb\setminus\mc,\Tp,\Tn\rangle_\RQ$ can be written as $\langle\mo,\mb,\Tp,\Tn\rangle_\RQ$ and it holds that there is a non-empty minimal conflict set w.r.t.\ $\langle\mo,\mb,\Tp,\Tn\rangle_\RQ$, i.e.\ $\mo$ is invalid w.r.t.\ $\langle\cdot,\mb,\Tp,\Tn\rangle_\RQ$. By Proposition~\ref{prop:cs_admissible}, this implies that $\langle\mo,\mb,\Tp,\Tn\rangle_\RQ$ is admissible. In other words, there is no conflict set w.r.t.\ $\langle\mb,\emptyset,\Tp,\Tn\rangle_\RQ$. Consequently, there must be a non-empty minimal conflict set w.r.t.\ $\langle\mo_1,\mb,\Tp,\Tn\rangle_\RQ$.

If $\mc \neq \emptyset$, on the other hand, then the second condition in line~\ref{algoline:validitytest1} during $\scQX'(\mc,\langle\mo,\mb,\Tp,\Tn\rangle_\RQ)$ must be invalid, i.e.\ there is no conflict set w.r.t.\ $\langle\mb,\emptyset,\Tp,\Tn\rangle_\RQ$. Consequently, there must be a non-empty minimal conflict set w.r.t.\ $\langle\mo_1,\mb,\Tp,\Tn\rangle_\RQ$.

Case (3): Here, we assume that the recursive call $\scQX'(\mc',\langle\mo',\mb',\Tp,\Tn\rangle_\RQ)$ is made in line~\ref{algoline:recursive_call2} and the result of $\scQX'$ in line~\ref{algoline:recursive_call1} is $\mc_2 \neq \emptyset$. As $\mc_2 \neq \emptyset$ and line~\ref{algoline:return_upwards} has never been reached by assumption, $\mc_2$ must have been returned in line~\ref{algoline:return_O} of $\scQX'(\mo_1,\langle\mo_2,\mb\cup\mo_1,\Tp,\Tn\rangle_\RQ)$ (which was called in line~\ref{algoline:recursive_call1}) wherefore $\mc_2 = \mo_2$ must hold.
So, it holds that $\mc' = \mo_2$, $\mo' = \mo_1$ and $\mb' = \mb \cup \mo_2$, i.e.\ the recursive call can be written as $\scQX'(\mo_2,\langle\mo_1,\mb\cup\mo_2,\Tp,\Tn\rangle_\RQ)$. 
%
%By the fact that $\scQX'(\mo_1,\langle\mo_2,\mb\cup\mo_1,\Tp,\Tn\rangle_\RQ)$ called in line~\ref{algoline:recursive_call1} returned $\mc_2\neq\emptyset$ and $\mo_1 \neq \emptyset$ by Lemma~\ref{lem:qx'_termination_etc}.\ref{lem_enum:c_neq_emptyset}, the second condition in line~\ref{algoline:validitytest1} during $\scQX'(\mo_1,\langle\mo_2,\mb\cup\mo_1,\Tp,\Tn\rangle_\RQ)$ must have been violated. That is, 
%there is no conflict set w.r.t.\ $\langle\mb\cup\mo_1,\emptyset,\Tp,\Tn\rangle_\RQ$. 
%%Further on, $|\mo_2|=1$ must hold due to the first proposition of this lemma.
 %
By the \emph{Induction Assumption}, there is a non-empty minimal conflict set w.r.t.\ $\langle\mc\cup\mo,\mb\setminus\mc,\Tp,\Tn\rangle_\RQ$. Moreover, $\mc\subseteq\mb$ by Lemma~\ref{lem:qx'_termination_etc}, prop.~\ref{lem_enum:c_sub_b} and (*) there is a non-empty minimal conflict set w.r.t.\ the DPI $\langle\mo,\mb,\Tp,\Tn\rangle_\RQ$ which is equal to the DPI $\langle\mo_1\cup\mo_2,\mb,\Tp,\Tn\rangle_\RQ$ by the fact that $\mo_1,\mo_2$ partition $\mo$ as per the definition of the \textsc{split} and \textsc{get} functions in lines~\ref{algoline:split}-\ref{algoline:get2}. 
%Due to the fact that $\mo_1,\mo_2$ is a partition of $\mo$, it must be true that there is a non-empty minimal conflict set w.r.t.\ $\langle\mo_1\cup\mo_2,\rangle$

What must still be proven, is (*):
Let us first assume that $\mc = \emptyset$ holds. In this case, $\langle\mc\cup\mo,\mb\setminus\mc,\Tp,\Tn\rangle_\RQ = \langle\mo,\mb,\Tp,\Tn\rangle_\RQ$ and thus there is a non-empty minimal conflict set w.r.t.\ $\langle\mo,\mb,\Tp,\Tn\rangle_\RQ$.

If $\mc \neq \emptyset$, on the other hand, then the second condition in line~\ref{algoline:validitytest1} during $\scQX'(\mc,\langle\mo,\mb,\Tp,\Tn\rangle_\RQ)$ must be invalid as otherwise $\emptyset$ would have been returned which is a contradiction to the assumption that the recursive call $\scQX'(\mc',\langle\mo',\mb',\Tp,\Tn\rangle_\RQ)$ was invoked in line~\ref{algoline:recursive_call2}. So, there is no conflict set w.r.t.\ $\langle\mb,\emptyset,\Tp,\Tn\rangle_\RQ$. Consequently, there must be a non-empty minimal conflict set w.r.t.\ $\langle\mo,\mb,\Tp,\Tn\rangle_\RQ$ due to $\mc\subseteq\mb$ by Lemma~\ref{lem:qx'_termination_etc}, prop.~\ref{lem_enum:c_sub_b}.

3): Case $S\neq\emptyset$: By $S \neq \emptyset$ and the fact that line~\ref{algoline:return_upwards} has not yet been reached, we obtain by the first proposition of this lemma that $|S|=1$ must hold.

There are two cases that can trigger $\scQX'(\mc,\langle\mo,\mb,\Tp,\Tn\rangle_\RQ)$ to return $\mo$ with $|\mo|=1$, i.e.\ case~1 involving $\mc \neq \emptyset$ and case~2 involving $\mc =\emptyset$. 

In case~1, $\mb$ must be valid w.r.t.\ $\langle\cdot,\emptyset,\Tp,\Tn,\rangle_\RQ$ as otherwise $\emptyset$ would be returned in line~\ref{algoline:return_emptyset}. So, there is no (minimal) conflict set w.r.t.\ $\langle\mb,\emptyset,\Tp,\Tn\rangle_\RQ$. 

As $|\mo|=1$ by assumption and by the fact that $\mc \subseteq\mb$ (holds by Lemma~\ref{lem:qx'_termination_etc}, prop.~\ref{lem_enum:c_sub_b}) and there is some non-empty minimal conflict set w.r.t.\ $\langle\mo\cup\mc,\mb\setminus\mc,\Tp,\Tn\rangle_\RQ$ (holds by the second proposition of this lemma), $\mo$ must include a non-empty minimal conflict set w.r.t.\ $\langle\mo,\mb,\Tp,\Tn\rangle_\RQ$. Since the only proper subset of $\mo$ is the empty set, $\mo$ must be a minimal conflict set w.r.t.\ $\langle\mo,\mb,\Tp,\Tn\rangle_\RQ$.

Case~2 can arise only when $\scQX'(\mc,\langle\mo,\mb,\Tp,\Tn\rangle_\RQ)$ is called in line~\ref{algoline:call_QX'} or line \ref{algoline:recursive_call2}. In line~\ref{algoline:recursive_call1} $\scQX'$ is called with $\mc\neq\emptyset$ by Lemma~\ref{lem:qx'_termination_etc}, prop.~\ref{lem_enum:c_neq_emptyset}.

In line~\ref{algoline:call_QX'} $\scQX'$ is called with $\mc=\emptyset$ and, by Corollary~\ref{cor:call_QX'_admissible}, with an admissible DPI $\langle\mo,\mb,\Tp,\Tn\rangle_\RQ$ for which a non-empty minimal conflict set exists as arguments. By the second proposition of this lemma, there is some non-empty minimal conflict set w.r.t.\ $\langle\mo\cup\emptyset,\mb\setminus\emptyset,\Tp,\Tn\rangle_\RQ = \langle\mo,\mb,\Tp,\Tn\rangle_\RQ$, and, by admissibility of $\langle\mo,\mb,\Tp,\Tn\rangle_\RQ$, there is no (minimal) conflict set w.r.t.\ $\langle\mb,\emptyset,\Tp,\Tn\rangle_\RQ$. By $|\mo|=1$, $\mo$ must be a minimal conflict set w.r.t.\ $\langle\mo,\mb,\Tp,\Tn\rangle_\RQ$.
% and, as $|\mo|=1$, it must be minimal as well.

A necessary condition for $\scQX'$ to be called with $\mc =\emptyset$ in line~\ref{algoline:recursive_call2} is obviously that $\scQX'(\mo_1,\langle\mo_2,\mb\cup\mo_1,\Tp,\Tn\rangle_\RQ)$ called in line~\ref{algoline:recursive_call1} returns $\emptyset$. By the Lemma~\ref{lem:qx'_termination_etc}, prop.~\ref{lem_enum:return_emptyset}, there is some non-empty minimal conflict set w.r.t.\ $\langle\mo_1,\mb,\Tp,\Tn\rangle_\RQ$. In line~\ref{algoline:recursive_call2}, the call $\scQX'(\emptyset,\langle\mo_1,\mb,\Tp,\Tn\rangle_\RQ)$ is made which, by assumption, returns $\mo_1$ with $|\mo_1|=1$. That means
%, there is some minimal conflict set w.r.t.\ $\langle\mo_1,\mb,\Tp,\Tn\rangle_\RQ$ and $|\mo_1|=1$ wherefore we can derive that 
$\mo_1$ is a minimal conflict set w.r.t.\ $\langle\mo_1,\mb,\Tp,\Tn\rangle_\RQ$.

Case $S=\emptyset$: Here, both conditions in line~\ref{algoline:validitytest1} must be met, i.e.\ in particular $\mb$ is invalid w.r.t.\ $\langle\cdot,\emptyset,\Tp,\Tn\rangle_\RQ$ which implies that $\mo$ is invalid w.r.t.\ $\langle\cdot,\mb,\Tp,\Tn\rangle_\RQ$ and $\langle\mo,\mb,\Tp,\Tn\rangle_\RQ$ is admissible. Therefore, by Proposition~\ref{prop:cs_admissible}, there is no non-empty minimal conflict set w.r.t.\ $\langle\mo,\mb,\Tp,\Tn\rangle_\RQ$. However, since $\mo$ is invalid w.r.t.\ $\langle\cdot,\mb,\Tp,\Tn\rangle_\RQ$, there must be a conflict set w.r.t.\ $\langle\mo,\mb,\Tp,\Tn\rangle_\RQ$. So, there is only the empty minimal conflict set w.r.t.\ $\langle\mo,\mb,\Tp,\Tn\rangle_\RQ$.

4): This proposition is an immediate consequence of the first proposition of this lemma.
\end{proof}

\begin{lemma}\label{lem:non_adm_return_emptyset}
Let $\langle\mo,\mb,\Tp,\Tn\rangle_\RQ$ be a non-admissible DPI. Then, $\emptyset$ is the only minimal conflict set w.r.t.\ $\langle\mo,\mb,\Tp,\Tn\rangle_\RQ$ and $\scQX'(\mc,\langle\mo,\mb,\Tp,\Tn\rangle_\RQ)$ with $\mc \neq \emptyset$ returns $\emptyset$ immediately in line~\ref{algoline:return_emptyset}.
%
%-- then $\emptyset$ is the only conflict set w.r.t. DPI
%-- $\scQX'(\mc,\langle\mo,\mb,\Tp,\Tn\rangle_\RQ) = \emptyset$ for $\mc \neq \emptyset$.
\end{lemma}
\begin{proof}
Since $\langle\mo,\mb,\Tp,\Tn\rangle_\RQ$ is non-admissible, $\mb\cup U_\Tp$ violates some $r\in\RQ$ or $\mb\cup U_\Tp \models \tn$ for some $\tn\in\Tn$. Therefore, $\emptyset$ is invalid w.r.t.\ $\langle\cdot,\mb,\Tp,\Tn\rangle_\RQ$, which, by Corollary~\ref{cor:validonto_cs}, implies that $\emptyset$ is a (minimal) conflict set w.r.t.\ $\langle\mo,\mb,\Tp,\Tn\rangle_\RQ$.

$\scQX'(\mc,\langle\mo,\mb,\Tp,\Tn\rangle_\RQ)$ returns $\emptyset$ in line~\ref{algoline:return_emptyset} as both conditions in line~\ref{algoline:validitytest2} are satisfied due to $\mc \neq \emptyset$ and the non-admissibility of $\langle\mo,\mb,\Tp,\Tn\rangle_\RQ$.
\end{proof}

\begin{lemma}\label{lem:adm_not_return_emptyset}
Let $\langle\mo,\mb,\Tp,\Tn\rangle_\RQ$ be an admissible DPI. Then $\scQX'(\mc,\langle\mo,\mb,\Tp,\Tn\rangle_\RQ)$ does not return in line~\ref{algoline:return_emptyset}.
\end{lemma}
\begin{proof}
By Definition~\ref{def:admissible}, $\mb$ must be valid w.r.t.\ $\langle\cdot,\emptyset,\Tp,\Tn\rangle_\RQ$. Hence, the second condition in line~\ref{algoline:validitytest2} is not satisfied wherefore a return cannot take place in line~\ref{algoline:return_emptyset}.
\end{proof}
\begin{lemma}\label{lem:qx_result_non-empty}
Let $\langle\mo,\mb,\Tp,\Tn\rangle_\RQ$ be an admissible DPI and let there be a non-empty minimal conflict set w.r.t.\ $\langle\mo,\mb,\Tp,\Tn\rangle_\RQ$. Then the following holds:
When $\scQX'(\mc,\langle\mo,\mb,\Tp,\Tn\rangle_\RQ)$
%Algorithm~\ref{algo:qx} 
reaches line~\ref{algoline:return_upwards} for the first time, $\mc_1\cup\mc_2$ is a non-empty minimal conflict set w.r.t.\ $\langle\mo,\mb,\Tp,\Tn\rangle_\RQ$.
\end{lemma}
\begin{proof}
%In order for Algorithm~\ref{algo:qx} to reach line~\ref{algoline:return_upwards}, $\scQX'$ must be called in line~\ref{algoline:call_QX'}. Due to Lemma~\ref{lem:qx_start_conditions}.\ref{lem_enum:qx'} and Corollary~\ref{cor:call_QX'_admissible}, it must hold that $\scQX'$ is called with an admissible DPI as argument for which a non-empty minimal conflict set exists. As a consequence, the premises of Lemma~\ref{lem:qx_downwards} are met.
The premises of this lemma are the same as those of Lemma~\ref{lem:qx_downwards}.
By Lemma~\ref{lem:qx_downwards}, prop.~\ref{lem_enum:downwards_4} we know that for $\mc_2$ and $\mc_1$ that are returned by the the calls to $\scQX'$ in lines~\ref{algoline:recursive_call1} and \ref{algoline:recursive_call2} $|\mc_1|\leq 1$ and $|\mc_2|\leq 1$ holds. Moreover, we know by Lemma~\ref{lem:qx_recursion_principle} that $\mc_1 \cup \mc_2$ is a minimal conflict set w.r.t.\ $\langle\mo,\mb,\Tp,\Tn\rangle_\RQ$.

What remains open is to show that $\mc_1 \cup \mc_2 \neq \emptyset$. To this end, we first assume that $\mc \neq \emptyset$. Then, by Lemma~\ref{lem:non_adm_return_emptyset}, $\langle\mo,\mb,\Tp,\Tn\rangle_\RQ$ must be an admissible DPI since it does not return in line~\ref{algoline:return_emptyset}, but only in line~\ref{algoline:return_upwards}.

If, on the other hand, $\mc = \emptyset$ holds, we can apply Lemma~\ref{lem:qx_downwards}, prop.~\ref{lem_enum:downwards_2} to obtain that there is a non-empty minimal conflict set w.r.t.\ $\langle\mo,\mb,\Tp,\Tn\rangle_\RQ$. This implies that $\mo$ is invalid w.r.t.\ $\langle\cdot,\mb,\Tp,\Tn\rangle_\RQ$. Therefore, we can conclude by means of Proposition~\ref{prop:cs_admissible} that $\langle\mo,\mb,\Tp,\Tn\rangle_\RQ$ is an admissible DPI.

Thus, in both cases we have that $\langle\mo,\mb,\Tp,\Tn\rangle_\RQ$ is an admissible DPI. Applying Lemma~\ref{lem:one_rec_call_adm} yields that at least one recursive call to $\scQX'$ in lines~\ref{algoline:recursive_call1} and \ref{algoline:recursive_call2} is given an admissible DPI as argument. By Lemma~\ref{lem:adm_not_return_emptyset}, this call cannot return in line~\ref{algoline:return_emptyset}. So, it must return in line~\ref{algoline:return_O} by the assumption that line~\ref{algoline:return_upwards} has not yet been reached before, wherefore it must return a set of cardinality 1. This completes the proof.
%either line 16 or line 17 must return a non-empty set due to the invariant an lemma(3).
\end{proof}
As long as the algorithm goes upwards after going upwards for the first time, a non-empty minimal conflict set is propagated upwards.
\begin{lemma}[QX: Upwards Correctness]\label{lem:qx_upwards}
Let $\langle\mo,\mb,\Tp,\Tn\rangle_\RQ$ be an admissible DPI and let there be a non-empty minimal conflict set w.r.t.\ $\langle\mo,\mb,\Tp,\Tn\rangle_\RQ$. Then:
%
%After Algorithm~\ref{algo:qx} 
After $\scQX'(\mc,\langle\mo,\mb,\Tp,\Tn\rangle_\RQ)$ has reached line~\ref{algoline:return_upwards} for the first time, the following holds: As long as line~\ref{algoline:recursive_call1} is not reached, each return in line~\ref{algoline:return_upwards} returns a minimal conflict set w.r.t.\ $\langle\mo,\mb,\Tp,\Tn\rangle_\RQ$.
\end{lemma}
\begin{proof}
The premises of this lemma are the same as those of Lemma~\ref{lem:qx_downwards}.
By Lemma~\ref{lem:qx_result_non-empty} we know that a non-empty minimal conflict $\mc$ set is returned at the first return that is made in line~\ref{algoline:return_upwards}. As, by assumption, $\mc$ is not the result $\mc_2$ of a prior call to $\scQX'$ in line~\ref{algoline:recursive_call1}, it must be the result $\mc_1$ of a prior call to $\scQX'$ in line~\ref{algoline:recursive_call2}. 
%
%As the first call to $\scQX'$ in line~\ref{algoline:call_QX'}, by Lemma~\ref{lem:qx_start_conditions}.\ref{lem_enum:qx'} and Corollary~\ref{cor:call_QX'_admissible}, got an admissible DPI as argument for which a non-empty minimal conflict set exists, the premises of Lemma~\ref{lem:qx_downwards} are fulfilled. 
%
Since the premises of Lemma~\ref{lem:qx_downwards} are fulfilled, Lemma~\ref{lem:qx_downwards} can be applied.
Since the call $\scQX'(\mo_1,\langle\mo_2,\mb\cup\mo_1,\Tp,\Tn\rangle)$ (that returned $\mc_2$) in line~\ref{algoline:recursive_call1} took place before line~\ref{algoline:return_upwards} was first reached, we have that $\mc_2$ is a minimal conflict set w.r.t.\ $\langle\mo_2,\mb\cup\mo_1,\Tp,\Tn\rangle$ by Lemma~\ref{lem:qx_downwards}, prop.~\ref{lem_enum:downwards_3}. By Lemma~\ref{lem:qx_recursion_principle}, we have that $\mc_2 \cup \mc$ is a minimal conflict set w.r.t.\ $\langle\mo,\mb,\Tp,\Tn\rangle$. As long as line~\ref{algoline:recursive_call1} is not reached, the same argumentation can be used to show that a minimal conflict set is returned in line~\ref{algoline:return_upwards}.
%When line~\ref{algoline:return_upwards} is first reached, we
%follows by junker theorem and corollary above.
\end{proof}
When the algorithm goes downwards again after going upwards for the first time, the invariant that that a minimal conflict set exists for each recursive downwards call to $\scQX'$ holds.
\begin{lemma}[QX: Downwards-after-upwards Correctness]\label{lem:qx_downwards_after_upwards}
Let $\langle\mo,\mb,\Tp,\Tn\rangle_\RQ$ be an admissible DPI and let there be a non-empty minimal conflict set w.r.t.\ $\langle\mo,\mb,\Tp,\Tn\rangle_\RQ$. Then:
%After Algorithm~\ref{algo:qx} 
After $\scQX'(\mc,\langle\mo,\mb,\Tp$, $\Tn\rangle_\RQ)$ has reached line~\ref{algoline:return_upwards} for the first time, the following holds: If line~\ref{algoline:recursive_call1} is reached for the first time, then, if the DPI $\langle\mo_1,\mb\cup\mc_2,\Tp,\Tn\rangle_\RQ$ which is the argument to the immediate call $\scQX'(\mc_2,\langle\mo_1,\mb\cup\mc_2,\Tp,\Tn\rangle_\RQ)$ 
in line~\ref{algoline:recursive_call2} is admissible, then there is a non-empty minimal conflict set w.r.t.\ $\langle\mo_1,\mb\cup\mc_2,\Tp,\Tn\rangle_\RQ$.
%for all calls $\scQX'(\mc',\langle\mo',\mb',\Tp,\Tn\rangle_\RQ)$ that take place in the time from the immediate call to 
%%$\scQX'(\mc,\langle\mo,\mb,\Tp,\Tn\rangle_\RQ)$ 
%$\scQX'(\mc_2,\langle\mo_1,\mb\cup\mc_2,\Tp,\Tn\rangle_\RQ)$ 
%in line~\ref{algoline:recursive_call2} until its return 
%%the immediate call to $\scQX'(\mc,\langle\mo,\mb,\Tp,\Tn\rangle_\RQ)$ in line~\ref{algoline:recursive_call2} and all further recursive calls 
%it holds that there is some minimal conflict set w.r.t.\ $\langle\mo'\cup\mc',\mb'\setminus\mc',\Tp,\Tn\rangle_\RQ$.
\end{lemma}
\begin{proof}
The premises of this lemma are the same as those of Lemma~\ref{lem:qx_downwards}.
Since line~\ref{algoline:recursive_call1} is first reached after line~\ref{algoline:return_upwards} has been reached for the first time, it must hold that $\scQX'(\mo_1,\langle\mo_2,\mb\cup\mo_1,\Tp,\Tn\rangle_\RQ)$ in line~\ref{algoline:recursive_call1} was called before line~\ref{algoline:return_upwards} has been reached. The reason for this to hold is the fact that only returns and no new calls to $\scQX'$ can have been made between the first occurrence of line~\ref{algoline:return_upwards} and the next occurrence of line~\ref{algoline:recursive_call1}.  

Therefore, the result $\mc_2$ of the call $\scQX'(\mo_1,\langle\mo_2,\mb\cup\mo_1,\Tp,\Tn\rangle_\RQ)$ in line~\ref{algoline:recursive_call1} is a minimal conflict set w.r.t.\ $\langle\mo_2,\mb\cup\mo_1,\Tp,\Tn\rangle_\RQ$ due to Lemma~\ref{lem:qx_downwards}, prop.~\ref{lem_enum:downwards_3}. As a consequence, $\mc_2 \cup \mb \cup \mo_1 \cup U_\Tp$ violates some $r\in\RQ$ or some $\Tn\in\Tn$. As the DPI $\langle\mo_1,\mb\cup\mc_2,\Tp,\Tn\rangle_\RQ$ is admissible by assumption, it holds that $\mc_2 \cup \mb \cup U_\Tp$ does not violate any $r\in\RQ$ or $\Tn\in\Tn$. Hence, $\mo_1$ must be invalid w.r.t.\ $\langle\cdot,\mb\cup\mc_2,\Tp,\Tn\rangle_\RQ$ which implies that there must be a non-empty minimal conflict set $S$ w.r.t.\ $\langle\mo_1,\mb\cup\mc_2,\Tp,\Tn\rangle_\RQ$.
\end{proof}
By applying the argumentation of Lemmas~\ref{lem:qx_downwards}, \ref{lem:qx_upwards} and \ref{lem:qx_downwards_after_upwards} 
%and corollary~\ref{cor:qx_downwards_return_cs} 
recursively on the entire recursion tree, we can prove the correctness of $\scQX'$.
\begin{lemma}\label{lem:qx'_correctness}
If $\scQX'(\mc,\langle\mo_{\mathsf{orig}},\mb_{\mathsf{orig}},\Tp,\Tn\rangle_\RQ)$ is called in line~\ref{algoline:call_QX'} by Algorithm~\ref{algo:qx}, it returns a non-empty minimal conflict set w.r.t.\ $\langle\mo_{\mathsf{orig}},\mb_{\mathsf{orig}},\Tp,\Tn\rangle_\RQ$.
% if $\langle\mo,\mb,\Tp,\Tn\rangle_\RQ$ is an admissible DPI.
\end{lemma}
\begin{proof}
If $\scQX'(\mc,\langle\mo_{\mathsf{orig}},\mb_{\mathsf{orig}},\Tp,\Tn\rangle_\RQ)$ is called in line~\ref{algoline:call_QX'} of Algorithm~\ref{algo:qx}, it must be true, by Lemma~\ref{lem:qx_start_conditions}, prop.~\ref{lem_enum:qx'} and Corollary~\ref{cor:call_QX'_admissible}, that $\langle\mo_{\mathsf{orig}},\mb_{\mathsf{orig}},\Tp,\Tn\rangle_\RQ$ is an admissible DPI for which a non-empty minimal conflict set exists.
As a consequence, the premises of Lemma~\ref{lem:qx_downwards} are met for $\langle\mo_{\mathsf{orig}},\mb_{\mathsf{orig}},\Tp,\Tn\rangle_\RQ$.

There are two cases to consider: Either (a)~$|\mo_{\mathsf{orig}}| \leq 1$ or (b)~$|\mo_{\mathsf{orig}}| > 1$ for the initial call to $\scQX'(\mc,\langle\mo_{\mathsf{orig}},\mb_{\mathsf{orig}},\Tp,\Tn\rangle_\RQ)$ in line~\ref{algoline:call_QX'}. In case (a), $0=|\mo_{\mathsf{orig}}| < 1$ cannot hold as there must be a non-empty minimal conflict set $\mc$ w.r.t.\ $\langle\mo_{\mathsf{orig}},\mb_{\mathsf{orig}},\Tp,\Tn\rangle_\RQ$ due to Lemma~\ref{lem:qx_start_conditions}, prop.~\ref{lem_enum:qx'}. Since $\emptyset \subset \mc \subseteq \mo_{\mathsf{orig}}$ must hold for $\mc$, this would be a contradiction to $|\mo_{\mathsf{orig}}|=0$.

So, $|\mo_{\mathsf{orig}}| = 1$ holds in case (a). In this case, $\scQX'$ returns $\mo_{\mathsf{orig}}$ immediately in line~\ref{algoline:return_O}, since $\mc=\emptyset$ and thus the conditions checked in line~\ref{algoline:validitytest2} cannot be met. In this case, $\mo_{\mathsf{orig}}$ is indeed a non-empty minimal conflict set since
%, by Corollary~\ref{cor:call_QX'_admissible}, $\scQX'$ is called in line~\ref{algoline:call_QX'} only if 
for the DPI $\langle\mo_{\mathsf{orig}},\mb_{\mathsf{orig}},\Tp,\Tn\rangle_\RQ$ given as argument 
%is admissible and for which 
there is a non-empty minimal conflict set by Lemma~\ref{lem:qx_start_conditions}, prop.~\ref{lem_enum:qx'}. Therefore $\emptyset$ cannot be a conflict set w.r.t.\ this DPI whereby $\mo_{\mathsf{orig}}$ is the only possible minimal conflict set due to $|\mo_{\mathsf{orig}}| = 1$.

Case (b): In this case, a direct return can neither take place in line~\ref{algoline:return_emptyset} by $\mc = \emptyset$ nor in line~\ref{algoline:return_O} by $|\mo_{\mathsf{orig}}| > 1$. So, $\scQX'$ is called recursively in lines~\ref{algoline:recursive_call1} and \ref{algoline:recursive_call2}. Since $\scQX'$ terminates due to Lemma~\ref{lem:qx_start_conditions}, prop.~\ref{lem_enum:terminates}, $\scQX'$ must reach line~\ref{algoline:return_upwards}. The first time some recursive call $\scQX'(\mc,\langle\mo,\mb,\Tp,\Tn\rangle_\RQ)$ reaches line~\ref{algoline:return_upwards}, it returns a non-empty minimal conflict set w.r.t.\ $\langle\mo,\mb,\Tp,\Tn\rangle_\RQ$ due to Lemma~\ref{lem:qx_result_non-empty}.
% where $\langle\mo,\mb,\Tp,\Tn\rangle_\RQ$ is the DPI which is the argument of the call to $\scQX'$ that returns in line~\ref{algoline:return_upwards}. 

By Lemma~\ref{lem:qx_upwards}, as long as line~\ref{algoline:recursive_call1} is not reached, i.e.\ no ``left branch'' (call to $\scQX'$ in line~\ref{algoline:recursive_call1}) but only ``right branches'' (calls to $\scQX'$ in line~\ref{algoline:recursive_call2}) return, a minimal conflict set $S$ is returned for each call to $\scQX'$ that ``wraps'' (is higher in the recursion tree than) the call that was the first to reach line~\ref{algoline:return_upwards}. It holds that $S \neq \emptyset$ since $S$ is a union of sets including the non-empty set returned when line~\ref{algoline:return_upwards} was first reached.

When it comes to an execution of line~\ref{algoline:recursive_call1}, i.e.\ the left branch returns, then the algorithm will take the right branch by executing line~\ref{algoline:recursive_call2}, i.e.\ calling $\scQX'(\mc_2,\langle\mo_1,\mb\cup\mc_2,\Tp,\Tn\rangle_\RQ)$, and go downwards in the recursion tree. 

Now, there are two cases. First, $\langle\mo_1,\mb\cup\mc_2,\Tp,\Tn\rangle_\RQ$ is non-admissible. Then, by Lemma~\ref{lem:non_adm_return_emptyset}, there is only one minimal conflict set w.r.t.\ $\langle\mo_1,\mb\cup\mc_2,\Tp,\Tn\rangle_\RQ$, namely $\emptyset$, and $\scQX'(\mc_2,\langle\mo_1,\mb\cup\mc_2,\Tp,\Tn\rangle_\RQ)$ directly returns $\emptyset$. As also the result $\mc_2$ of the call to $\scQX'(\mo_1,\langle\mo_2,\mb\cup\mo_1,\Tp,\Tn\rangle_\RQ)$ immediately before in line~\ref{algoline:recursive_call1} is a minimal conflict set w.r.t.\ $\langle\mo_2,\mb\cup\mo_1,\Tp,\Tn\rangle_\RQ$, as established above, we can apply Lemma~\ref{lem:qx_recursion_principle} to derive that indeed a minimal conflict set w.r.t.\ $\langle\mo,\mb,\Tp,\Tn\rangle_\RQ$ is returned in line~\ref{algoline:return_upwards}. Thus, Lemma~\ref{lem:qx_upwards} can be further applied to move upwards in the recursion tree until line~\ref{algoline:recursive_call1} occurs again.

Second, $\langle\mo_1,\mb\cup\mc_2,\Tp,\Tn\rangle_\RQ$ is admissible. Then, by Lemma~\ref{lem:qx_downwards_after_upwards}, there is a non-empty minimal conflict set w.r.t.\ $\langle\mo_1,\mb\cup\mc_2,\Tp,\Tn\rangle_\RQ$. Hence, Lemma~\ref{lem:qx_downwards} can be used again for the subtree of the recursion tree rooted at the call $\scQX'(\mc_2,\langle\mo_1,\mb\cup\mc_2,\Tp,\Tn\rangle_\RQ)$. That is, it can be used to show that each call to $\scQX'$ within this subtree returns a minimal conflict set w.r.t.\ the DPI given as argument as long as the algorithm moves downwards in the tree. Having reached line~\ref{algoline:return_upwards} for the first time, Lemma~\ref{lem:qx_result_non-empty} lets us conclude again that a non-empty conflict set w.r.t.\ the respective argument DPI is actually returned at this place. Subsequently, Lemma~\ref{lem:qx_upwards} 
%and \ref{lem:qx_downwards_after_upwards} 
can be applied to show that each return gives back a minimal conflict set w.r.t.\ the argument DPI of the respective call, as long as the algorithm moves upwards in the recursion tree.

What is still open is to show that the call $\scQX'(\mc_2,\langle\mo_1,\mb\cup\mc_2,\Tp,\Tn\rangle_\RQ)$ in line~\ref{algoline:recursive_call2} that is made immediately after the algorithm first reached line~\ref{algoline:recursive_call1} after moving upwards after reaching line~\ref{algoline:return_upwards} for the first time returns a minimal conflict set w.r.t.\ $\langle\mo_1,\mb\cup\mc_2,\Tp,\Tn\rangle_\RQ$, indeed. This holds by the fact that Lemmas~\ref{lem:qx_downwards} and ~\ref{lem:qx_upwards} guarantee that a left branch always returns a minimal conflict set, Lemma~\ref{lem:qx_downwards_after_upwards} guarantees that Lemmas~\ref{lem:qx_downwards} and ~\ref{lem:qx_upwards} can be applied after making a single right branch. However, as $\scQX'$ terminates the recursion tree is finite and thus the case must arise where the right branch directly returns. In case the DPI $\langle\mo,\mb,\Tp,\Tn\rangle_\RQ$ given as argument for this right branch is non-admissible, the only minimal conflict set $\emptyset$ is returned, as established above. If the DPI $\langle\mo,\mb,\Tp,\Tn\rangle_\RQ$ given as argument for this right branch is admissible, on the other hand, then we have already shown above that there is a non-empty minimal conflict set w.r.t.\ this DPI. Moreover, $|\mo|=1$ must hold due to the fact that this right branch directly returns (without entering a further recursion). Therefore, $\mo$ is returned which is actually a minimal conflict set w.r.t.\ $\langle\mo,\mb,\Tp,\Tn\rangle_\RQ$ as $\mo$ is the only non-empty subset of $\mo$.  
%According to Lemma~\ref{lem:qx_downwards_after_upwards}, it holds for the call $\scQX'(\mc,\langle\mo,\mb,\Tp,\Tn\rangle_\RQ)$ made in line~\ref{algoline:recursive_call2} that there is a minimal conflict set w.r.t.\ $\langle\mo\cup\mc,\mb\setminus\mc,\Tp,\Tn\rangle_\RQ$. 
%where $\mo \subset \mo_{\mathsf{orig}}$ and $\mb \supseteq \mb_{\mathsf{orig}}$ which holds since $\scQX'$ was first called with the DPI $\langle\mo_{\mathsf{orig}},\mb_{\mathsf{orig}},\Tp,\Tn\rangle_\RQ$ and only recursive calls were made until line~\ref{algoline:return_upwards}. For the argument DPI $\langle\mo',\mb',\Tp,\Tn\rangle_\RQ$ in each recursive call within $\scQX'(\mc,\langle\mo,\mb,\Tp,\Tn\rangle_\RQ)$ it holds that $\mo' \subset \mo$ and $\mb' \supseteq \mb$ by the way the \textsc{split} function in line~\ref{algoline:split} works. 
\end{proof}

\begin{proposition}\label{prop:qx_correctness}
Let $\langle\mo,\mb,\Tp,\Tn\rangle_\RQ$ be a DPI. Then, $\scQX(\langle\mo,\mb,\Tp,\Tn\rangle_\RQ)$ terminates and returns 
\begin{itemize}
	\item 'no conflict' iff there is no conflict w.r.t.\ $\langle\mo,\mb,\Tp,\Tn\rangle_\RQ$ $\quad$ \\ ($\mo$ is valid w.r.t.\ $\langle\cdot,\mb,\Tp,\Tn\rangle_\RQ$)
	\item $\emptyset$ iff $\emptyset$ is the only minimal conflict set w.r.t.\ $\langle\mo,\mb,\Tp,\Tn\rangle_\RQ$ $\quad$ \\(DPI is non-admissible)
	\item a non-empty minimal conflict set w.r.t.\ $\langle\mo,\mb,\Tp,\Tn\rangle_\RQ$ iff there is a non-empty minimal conflict set w.r.t.\ $\langle\mo,\mb,\Tp,\Tn\rangle_\RQ$ $\quad$ \\(DPI is admissible and $\mo$ is invalid w.r.t.\ $\langle\cdot,\mb,\Tp,\Tn\rangle_\RQ$).
\end{itemize}
\end{proposition}
\begin{proof}
The proposition is a direct consequence of Lemma~\ref{lem:qx_start_conditions} and Lemma~\ref{lem:qx'_correctness}.
\end{proof}
%\begin{proof}
%During the recursion $\scQX'$ successively discards sentences $\mo \subseteq \mo_\orig$ iff $\mo_\orig \setminus \mo$ includes at least one minimal conflict set w.r.t.\ the original DPI $\langle\mo_\orig,\mb_\orig,\Tp,\Tn\rangle_\RQ$. That is, for the call $\scQX'(\mc,\langle\mo,\mb,\Tp,\Tn\rangle_\RQ)$ $\mo$ is dismissed iff $\mc \supset \emptyset$, i.e.\ $$
%- show that it returns a non-minimal set
%- assume that this set is 
%\end{proof}

%%%%%%%%%%%%%%%%%%%%%%%%%%%%%%%%%%%%%%%%%%%%%%%%%%%%%%%%%%%%%%%%%%%%%%

\section{Hitting Set Tree Based Diagnosis Computation}
\label{sec:hs_comp}
%\noindent\emph{Computation of Hitting Sets of all Minimal Conflict Sets.} 
One way to compute minimal diagnoses from minimal conflict sets is to use a hitting set tree algorithm which was originally proposed by Reiter~\cite{Reiter87}. In this work we describe methods for non-interactive and interactive diagnosis computation based on the ones used in~\cite{friedrich2005gdm, ksgf2010, Shchekotykhin2012} which are closely related to the original hitting set tree algorithm. Differences of the described non-interactive algorithm to the original one of Reiter are 
%(1)~on-demand computation of conflict sets, 
%(2)~a conflict set reuse strategy for node labeling,
\begin{enumerate} 
\item the usage of different edge weights (probabilities) inducing an order of node generation (uniform-cost) different to breadth-first and 
\item the opportunity to specify an execution time threshold $t$ as well as a minimal ($n_{\min}$) and maximal ($n_{\max}$) desired number of minimal diagnoses to be computed by the algorithm. 
\end{enumerate}
In this vein, the algorithm computes at least the $n_{\min}$ most-probable minimal diagnoses w.r.t.\ the given probabilities and goes on computing further next most-probable minimal diagnoses until either overall computation time reaches the time limit $t$ or $n_{\max}$ diagnoses have been computed.
 
Such a time threshold and an interval of minimal and maximal number of diagnoses is particularly relevant in settings where not all potential minimal faulty sets need to be computed, such as iterative, interactive settings where reaction time is crucial (since a user is waiting to interact with the system). Instead, in such settings only a ``representative'' set of minimal diagnoses is exploited to decide which question to ask a user such that the answer to that question allows the constructed partial tree to be pruned. After pruning, the tree is expanded again to compute another ``representative'' set of minimal diagnoses.
% in order to execute further computations that exploit the leading diagnoses and additional information obtained by asking the user to prune down the search space in advance to prevent the system from delivering diagnoses that are not in line with the currently collected information. The described interactive search strategies have not been discussed and analyzed in such a detailed way in literature as is done in this paper (see Section~\ref{sec:InteractiveOntologyDebugging}).
%
%\noindent\textbf{Inputs.} 
%
Such an interactive KB debugging algorithm will be presented in Part~\ref{part:InteractiveKBDebugging}.
The non-interactive version of the KB debugging algorithm is delineated by Algorithm~\ref{algo:hs} and described next. 

\paragraph{Inputs.} The algorithm takes as input an admissible DPI $\tuple{\mo,\mb,\Tp,\Tn}_\RQ$, some computation timeout $t$, a
desired minimal ($n_{\min}$) and maximal ($n_{\max}$) number of minimal diagnoses to be returned, and a function $p:\mo \rightarrow (0,0.5)$ that assigns to each formula $\tax \in \mo$ a weight that represents the (estimated) likeliness of $\tax$ to be faulty and thereby determines the search strategy, e.g.\ breadth-first or uniform-cost. 
Within the algorithm, $p()$ is used to impose an order on open nodes that tells the algorithm which node to expand next. Details concerning the function $p()$ will be discussed in Section~\ref{sec:DiagnosisProbabilitySpace} after demonstrating various ways of obtaining information relevant to $p()$ and detailing how $p()$ can be defined by means of such information. Throughout the rest of the current Section~\ref{sec:hs_comp} we assume that $p()$ implies a first-in-first-out sorting of open nodes, i.e.\ a breadth-first search strategy as described in~\cite{Reiter87}.

\subsection{Breadth-First Diagnosis Computation}
\label{sec:BreadthFirstDiagnosisComputation}

%$p()$ is specified in a way that its definition can be extended to sets of sentences such that $p(S_1) \geq p(S_2)$ iff $S_1 \subseteq S_2$ for $S_1,S_2 \subseteq \mo$. For example, $p(S) = \frac{1}{|S|}$ would satisfy this property.
%Usually the weight of an formula will represent the (estimated) likeliness of the formula to be faulty (for more details on the weight function see Section~\ref{sec:DiagnosisProbabilitySpace}). 

%\noindent\textbf{Algorithm Overview and Implementation Remarks.} 
\paragraph{Algorithm Overview and Implementation Remarks.}
To compute minimal diagnoses w.r.t.\ $\langle\mo,\mb,\Tp$, $\Tn\rangle_\RQ$ from minimal conflict sets w.r.t.\ $\langle\mo,\mb,\Tp,\Tn\rangle_\RQ$, the algorithm produces a labeled tree where a non-closed node is labeled by a minimal conflict set and a closed node is labeled by either $valid$ or $closed$. From a non-closed node labeled by a minimal conflict set $\mc=\setof{\tax_p, \dots, \tax_q}$ there are $|\mc|$ outgoing edges, each labeled by one $\tax\in\mc$ and each leading to a new node that needs to be labeled. Closed nodes are leaf nodes of the produced tree, i.e.\ they have no successor nodes, and correspond to non-minimal or duplicate hitting sets (label $closed$) or to minimal hitting sets (label $valid$) of all minimal conflict sets w.r.t.\ the input DPI $\langle\mo,\mb,\Tp,\Tn\rangle_\RQ$. Conflict sets to label nodes are computed only on-demand for time efficiency after the attempt to reuse an already computed one fails. In case an appropriate order of node labeling (e.g.\ breadth-first tree construction) is used, the complete tree given when all nodes in the tree are closed contains all minimal diagnoses w.r.t.\ the DPI $\langle\mo,\mb,\Tp,\Tn\rangle_\RQ$ provided as input. In this complete tree, the set of edge labels on each path from the root node to a node labeled by $valid$ is a minimal diagnosis. 

What Algorithm~\ref{algo:hs} actually does is building up a \emph{pruned HS-tree} for a given DPI. So, we next provide formal definitions of a \emph{(partial) HS-tree} and a \emph{(partial) pruned HS-tree} based on the definitions given in~\cite{Reiter87}.
%provide a formal definition of a (partial) pruned HS-tree based on the definition given in~\cite{Reiter87}.
\begin{definition}[HS-Tree]\label{def:hs_tree}
Let $\langle\mo,\mb,\Tp,\Tn\rangle_\RQ$ be an admissible DPI. An edge-labeled and node-labeled tree $T$ is called an \emph{HS-tree w.r.t.\ $\langle\mo,\mb,\Tp,\Tn\rangle_\RQ$} iff it is a smallest tree with the following properties: 
\begin{enumerate}
\item The root of $T$ is labeled by $valid$ if $\mo$ is valid w.r.t.\ $\langle\cdot,\mb,\Tp,\Tn\rangle_\RQ$. Otherwise, the root is labeled by a conflict set w.r.t.\ $\langle\mo,\mb,\Tp,\Tn\rangle_\RQ$.
\item If $\mathsf{n}$ is a node of $T$, define $H(\mathsf{n})$ to be the set of edge labels on the path in
$T$ from the root node to $\mathsf{n}$. 
If $\mathsf{n}$ is labeled by $valid$, it has no successor nodes in $T$. If $\mathsf{n}$ is labeled by a conflict set $\mc$ w.r.t.\ $\langle\mo,\mb,\Tp,\Tn\rangle_\RQ$, then for each $\tax \in \mc$, $\mathsf{n}$ has a successor node $\mathsf{n}_{\tax}$ joined to $\mathsf{n}$ by an edge labeled by $\tax$. The label for $\mathsf{n}_{\tax}$ is a conflict set $\mc'$ w.r.t.\ $\langle\mo,\mb,\Tp,\Tn\rangle_\RQ$ such that $\mc' \cap H(\mathsf{n}_{\tax}) = \emptyset$ if such a set $\mc'$ exists. Otherwise, $\mathsf{n}_{\tax}$ is labeled by $valid$.
\end{enumerate}
$T$ is called a \emph{partial HS-tree w.r.t.\ $\langle\mo,\mb,\Tp,\Tn\rangle_\RQ$} iff $T$ is a HS-tree w.r.t.\ $\langle\mo,\mb,\Tp,\Tn\rangle_\RQ$ where not all nodes in $T$ are labeled and non-labeled nodes have no successors. 
\end{definition}
\begin{definition}[Pruned HS-Tree]\label{def:pruned_hs_tree}
Let $\langle\mo,\mb,\Tp,\Tn\rangle_\RQ$ be an admissible DPI. An edge-labeled and node-labeled tree $T$ is called a \emph{pruned HS-tree (pHS-tree) w.r.t.\ $\langle\mo,\mb,\Tp,\Tn\rangle_\RQ$} iff $T$ is the result of constructing an HS-tree w.r.t.\ $\langle\mo,\mb,\Tp,\Tn\rangle_\RQ$ with due regard to the following rules: 
\begin{enumerate}
\item Label nodes in the HS-tree in \emph{breadth-first order}.
%Label nodes in the HS-tree in a way that $\mathsf{n}_1$ is labeled before $\mathsf{n}_2$ if $\mathsf{n}_1, \mathsf{n}_2$ are nodes with $H(\mathsf{n}_1) \subset H(\mathsf{n}_2)$.
\item Use only \emph{minimal} conflict sets w.r.t.\ $\langle\mo,\mb,\Tp,\Tn\rangle_\RQ$ to label nodes in $T$.\label{def:pruned_hs_tree:use_only_min_cs}
\item \emph{Reusing node labels:} If node $\mathsf{n}$ is labeled by $\mc$ and $\mathsf{n}'$ is a node such that $H(\mathsf{n}') \cap \mc = \emptyset$, label $\mathsf{n}'$ by $\mc$. 
\item\label{def:pruned_hs_tree:non_min_pruning_rule} \emph{Non-minimality pruning rule:} If node $\mathsf{n}$ is labeled by $valid$ and node $\mathsf{n}'$ is such that $H(\mathsf{n}) \subseteq H(\mathsf{n}')$, label $\mathsf{n}'$ by $closed$.
\item If node $\mathsf{n}$ is labeled by $closed$, it has no successors.
\item\label{def:pruned_hs_tree:rule6} \emph{Duplicate pruning rule:} If node $\mathsf{n}$ is next to be labeled and there is some node $\mathsf{n}'$ such that $H(\mathsf{n}') = H(\mathsf{n})$, then label
$\mathsf{n}$ by $closed$. 
\end{enumerate}
$T$ is called a \emph{partial pruned HS-tree} iff $T$ is a pruned HS-tree where not all nodes in $T$ have been labeled yet and non-labeled nodes have no successors. 
\end{definition}

\begin{remark}\label{rem:pruned_hs_tree}
Notice that we use a definition of a pruned HS-tree that slightly differs from the definition given in~\cite{Reiter87} in that we inherently assume that only \emph{minimal} conflict sets w.r.t.\ the given DPI are used to label nodes in the tree. Therefore we could omit the last rule in the definition of \cite{Reiter87}. Namely, such a situation where some node has been labeled by a subset of the label of another node cannot arise in our definition since no minimal conflict set can be a subset of another different minimal conflict set w.r.t.\ the same DPI.

In general, there are multiple different pHS-trees w.r.t.\ one and the same DPI~\cite{greiner1989correction}. Reason for this is that 
\begin{itemize}
\item the order of adding successor nodes (on the same tree level) to the queue $\Queue$ and
\item which of generally multiple minimal conflict sets to (re)use to label a node 
\end{itemize} 
is not determined by Definition~\ref{def:pruned_hs_tree}.\qed
\end{remark} 

By~\cite[Theorem~4.8]{Reiter87} and Proposition~\ref{prop:mindiag_mincs}, the following holds:
\begin{proposition}\label{prop:pruned_hs_tree_finds_all_min_diags}
Let $\langle\mo,\mb,\Tp,\Tn\rangle_\RQ$ be an admissible DPI and $T$ a pHS-tree w.r.t.\ $\langle\mo,\mb,\Tp,\Tn\rangle_\RQ$. Then, $\setof{H(\mathsf{n})\,|\, \mathsf{n} \mbox{ is a node of } T \mbox{ labeled by } valid} = \minD_{\langle\mo,\mb,\Tp,\Tn\rangle_\RQ}$, i.e.\ the set of all minimal diagnoses w.r.t.\ $\langle\mo,\mb,\Tp,\Tn\rangle_\RQ$.
\end{proposition}

\begin{remark}\label{rem:algo_hs:internal_representation} A \emph{node} $\mathsf{nd}$ in Algorithm~\ref{algo:hs} is defined as the set of formulas that label the edges on the path from the root node to $\mathsf{nd}$. In other words, we associate a node $\mathsf{n}$ with $H(\mathsf{n})$.
In this vein, Algorithm~\ref{algo:hs} \emph{internally} does not store a labeled tree, but only ``relevant'' sets of nodes 
%and a set of minimal 
and conflict sets. That is, it does not store any
\begin{itemize}
	\item non-leaf nodes,
	\item labels of non-leaf nodes, i.e.\ it does not store which minimal conflict set labels which node,
	\item edges between nodes, 
	\item labels of edges and
	\item leaf nodes labeled by $closed$.
\end{itemize}
%Labels of nodes are not explicitly stored. 
%
Let $T$ denote the (partial) pHS-tree produced by Algorithm~\ref{algo:hs} at some point during its execution (Corollary~\ref{cor:algo_hs_returns_relevant_data_of_(partial)_pruned_hs_tree} will show that Algorithm~\ref{algo:hs} using breadth-first search	in fact produces a (partial) pHS-tree). Then, Algorithm~\ref{algo:hs} only stores 
\begin{itemize}
	\item a set of nodes $\mD_{calc}$ where each node corresponds to the edge labels along a path in $T$ leading to a leaf node that has been labeled by $valid$ (minimal diagnoses w.r.t.\ $\langle\mo,\mb,\Tp,\Tn\rangle_\RQ$), 
	%i.e.\ with the definition of a node given above $\mD_{calc} = \setof{\mathsf{nd}\,|\,label(\mathsf{nd}) = valid}$
	\item a list of open (non-closed) nodes $\Queue$ where each node in $\Queue$ corresponds to the edge labels along a path in $T$ leading from the root node to a leaf node that has been generated, but has not yet been labeled and
	\item the set $\mC_{calc}$ of already computed minimal conflict sets w.r.t.\ $\langle\mo,\mb,\Tp,\Tn\rangle_\RQ$ that have been used to label non-leaf nodes in $T$.
	%\item a queue of open nodes $\Queue$, i.e.\ nodes that have been generated, but not yet labeled,
	%\item a set of nodes $\mD_{calc}$ that are minimal diagnoses w.r.t.\ $\langle\mo,\mb,\Tp,\Tn\rangle_\RQ$ and 
	%\item a set $\mC_{calc}$ of already computed minimal conflict sets w.r.t.\ $\langle\mo,\mb,\Tp,\Tn\rangle_\RQ$.
\end{itemize}
We call $\tuple{\mD_{calc}, \Queue, \mC_{calc}}$ the \emph{relevant data of $T$}. If $T$ is a pHS-tree, then $\Queue$ is the empty list.

This internal representation of the constructed (partial) pHS-tree by its relevant data does not constrain the functionality of the algorithm. 
This holds as diagnoses are paths from the root, i.e.\ nodes in the internal representation, and the goal of a (partial) pHS-tree is to determine minimal diagnoses w.r.t.\ the given DPI. 
The node labels or edge labels along a certain path and their order along this path is completely irrelevant when it comes to finding a label for the leaf node of this path. Instead, only the set of edge labels is required for the computation of the label for a leaf node. Also, to rule out nodes corresponding to non-minimal diagnoses, it is sufficient to know the set of already found diagnoses $\mD_{calc}$. No already closed nodes are needed for the correct functionality of Algorithm~\ref{algo:hs}.\qed
%Paths from the root in a (partial) pHS-tree are (partial) hitting sets of all minimal conflict sets w.r.t.\ $\langle\mo,\mb,\Tp,\Tn\rangle_\RQ$, i.e.\ (partial) diagnoses w.r.t.\ $\langle\mo,\mb,\Tp,\Tn\rangle_\RQ$ (cf.\ Proposition~\ref{prop:mindiag_mincs}). 
%
%Algorithm~\ref{algo:hs} does not store nodes corresponding to paths in $T$ leading to leaf nodes that have been labeled by $closed$.
\end{remark}

%\noindent\textbf{Initialization.} 
\paragraph{Initialization.}
First, Algorithm~\ref{algo:hs} initializes the variable $t_{start}$ with the current system time (\textsc{getTime}), the set of calculated minimal diagnoses $\mD_{calc}$ to the empty set and the ordered queue of open nodes $\Queue$ to a list including the empty set only (i.e.\ only the unlabeled root node). 
%Note that in this algorithm nodes are represented as sets of formulas. Thereby, each node $\mathsf{nd}$ represents the set of edge labels on the path from the root node to $\mathsf{nd}$. 

%\noindent\textbf{The Main Loop.} 
\paragraph{The Main Loop.}
Within the loop (line~\ref{algoline:hs:repeat}) the algorithm gets the node to be processed next, namely the first node $\mathsf{node}$ (\textsc{getFirst}, line~\ref{algoline:hs:getfirst}) in the list of open nodes $\Queue$ ordered by the function $p_{nodes}()$ and removes $\mathsf{node}$ from $\Queue$ (\textsc{deleteFirst}, line~\ref{algoline:hs:remove_from_queue}). Note that $p_{nodes}()$ can be directly obtained from $p()$. As mentioned before, for the moment the reader should simply suppose that $p_{nodes}()$ imposes an order on $\Queue$ which effectuates a breadth-first labeling of open nodes in the tree. A definition of $p_{nodes}()$ will be given by Definition~\ref{def:p_node()} after a motivation and detailed explanation of $p_{nodes}()$ will have been given in Section~\ref{sec:DiagnosisProbabilitySpace}. 
%Details regarding the ordering of $\Queue$ are given in Section~\ref{sec:DiagnosisProbabilitySpace}. For the moment, the reader should simply assume that $\Queue$ is ordered FIFO (breadth-first search). 

%\noindent\textbf{Computation of Node Labels.} 
\paragraph{Computation of Node Labels.}
Then, a label is computed for $\mathsf{node}$ in line~\ref{algoline:hs:label}. Nodes are labeled by $valid$, $closed$ or a minimal conflict set w.r.t.\ $\langle\mo,\mb,\Tp,\Tn\rangle_\RQ$ by the procedure \textsc{label} (line~\ref{algoline:hs:procedure_label} ff.). This procedure gets as inputs the DPI $\tuple{\mo,\mb,\Tp,\Tn}_\RQ$, the current node $\mathsf{node}$, the set of already computed minimal conflicts ($\mC_{calc}$) and minimal diagnoses ($\mD_{calc}$) and the queue $\Queue$ of open nodes, and it returns an updated set of computed minimal conflicts $\mC_{calc}$ and a label for $\mathsf{node}$. It works as follows:
%Within the loop (line~\ref{algoline:hs:repeat}) the algorithm gets the first element $\mathsf{node}$ of the ordered set of open nodes $\Queue$, i.e.\ the best node (with \emph{minimal} weight) according to $w()$. The weight $w(\mathsf{nd})$ of a node $\mathsf{nd} = \setof{\tax_s,\dots, \tax_t}$ is defined through $w(\tax), \tax\in\mo$ as follows: 
%\begin{align}
%w(\mathsf{nd}) = \prod_{\tax_i\in \mathsf{nd}} \tax_i \prod_{\tax_j \in \mo \setminus \mathsf{nd}} \tax_j
%\end{align}
%Then, $\mathsf{node}$ is removed from $\Queue$ (line~\ref{algoline:hs:remove_from_queue}) and a label is computed for $\mathsf{node}$ in line\ref{algoline:hs:label}. Nodes are labeled by $valid$, $closed$ or a minimal conflict set by the procedure \textsc{label} (line~\ref{algoline:hs:procedure_label} ff.). This procedure gets as inputs the DPI $\tuple{\dot,\mb,\Tp,\Tn}_\RQ$, the current node $\mathsf{node}$, the set of already computed minimal conflicts ($\mC_{calc}$) and minimal diagnoses ($\mD_{calc}$) and the queue $\Queue$ of open nodes, and it returns an updated set of computed minimal conflicts $\mC_{calc}$ and a label for $\mathsf{node}$. It works as follows:

A node $\mathsf{node}$ is labeled by $closed$ iff (a)~there is an already computed minimal diagnosis $\md$ in $\mD_{calc}$ that is a subset of this node, i.e.\ $\md \subseteq \mathsf{node}$, which means that $\mathsf{node}$ cannot be a minimal diagnosis (non-minimality criterion, lines~\ref{algoline:hs:non_min_crit_start}-\ref{algoline:hs:non_min_crit_end}) or (b)~there is some node $\mathsf{nd}$ in the queue of open nodes $\Queue$ such that $\mathsf{node} = \mathsf{nd}$ which means that one of the two tree branches with an equal set of edge labels can be closed, i.e.\ removed from $\Queue$ (duplicate criterion, lines~\ref{algoline:hs:duplicate_crit_start}-\ref{algoline:hs:duplicate_crit_end}). 

If none of these $closed$-criteria is met, the algorithm searches for some $\mc$ in $\mC_{calc}$, the set of already computed minimal conflict sets, such that $\mc \cap \mathsf{node} = \emptyset$ and returns the label $\mc$ for $\mathsf{node}$ (reuse criterion, lines~\ref{algoline:hs:reuse_crit_start}-\ref{algoline:hs:reuse_crit_end}). This means that the path represented by $\mathsf{node}$ cannot be a diagnosis as there is (at least) one minimal conflict set, namely $\mc$, that is not hit by $\mathsf{node}$. 

If the reuse criterion does not apply, a call to $\scQX(\tuple{\mo \setminus \mathsf{node}, \mb, \Tp, \Tn}_\RQ)$ is made (line~\ref{algoline:hs:qx}) in order to check whether there is a not-yet-computed minimal conflict set that is not hit by $\mathsf{node}$. Note that the KB $\mo \setminus \mathsf{node}$ that is given to $\scQX$ as part of the argument DPI ensures that only minimal conflict sets $\mc \subseteq \mo \setminus \mathsf{node}$ can be computed, i.e.\ ones that do not share any single formula with $\mathsf{node}$ (cf. Section~\ref{sec:cs_comp}).
 
\begin{remark}\label{rem:qx_with_O_setminus_node_yields_cs_wrt_O}
A minimal conflict set computed by $\scQX(\tuple{\mo \setminus \mathsf{node}, \mb, \Tp, \Tn}_\RQ)$ is a minimal conflict set w.r.t.\ $\tuple{\mo, \mb, \Tp, \Tn}_\RQ$ indeed since (i)~$\scQX(\tuple{\mo \setminus \mathsf{node}, \mb, \Tp, \Tn}_\RQ)$ returning a set $\mc$ means that $\mc$ is a minimal conflict set w.r.t.\ $\tuple{\mo \setminus \mathsf{node}, \mb, \Tp, \Tn}_\RQ$ by Proposition~\ref{prop:qx_correctness} and (ii)~the ``$\Rightarrow$'' direction of Corollary~\ref{cor:validonto_cs} implies that $\mc$ is not valid w.r.t.\ $\tuple{\cdot, \mb, \Tp, \Tn}_\RQ$ and (iii)~the ``$\Leftarrow$'' direction of Corollary~\ref{cor:validonto_cs} lets us conclude that $\mc$ is a minimal conflict w.r.t.\ $\tuple{X, \mb, \Tp, \Tn}_\RQ$ where $X$ is any superset of $\mc$, in particular $X := \mo$. 
%\supseteq(\mo\setminus\mathsf{node})$.
\qed
%\begin{lemma}
 %$\mc$ is not valid w.r.t.\ $\tuple{\cdot, \mb, \Tp, \Tn}_\RQ$ iff $\mc$ is a conflict set w.r.t.\ $\tuple{X, \mb, \Tp, \Tn}_\RQ$ for any $X \supseteq \mc$.
%\end{lemma}
\end{remark}

$\scQX$ may then return %(a)~$valid$ iff $\scQX$ returns 
(a)~'no conflict', i.e.\ $\mo \setminus \mathsf{node}$ is already valid w.r.t.\ $\tuple{\cdot,\mb,\Tp,\Tn}_\RQ$, 
%(d)~$closed$ iff $\scQX$ returns $\emptyset$, i.e.\ the input DPI is non-admissible which means that there is no diagnosis for it, 
or (b)~a new conflict set $L \neq \emptyset$ such that $L \notin \mC_{calc}$. 
Note that the case of the output $L = \emptyset$ of $\scQX$ cannot arise since (i)~the DPI provided as input to the algorithm is assumed to be admissible, (ii)~no other DPI for which $\scQX$ is called can be non-admissible since admissibility is defined only by the sets $\mb,\Tp,\Tn,\RQ$ which remain unmodified throughout the execution of Algorithm~\ref{algo:hs}, and (iii)~as per Proposition~\ref{prop:qx_correctness}, $\scQX$ returns $\emptyset$ only if the DPI given to it as an argument is non-admissible.
%can only arise for when labeling the root node $\emptyset$ as $\tuple{\mo,\mb,\Tp,\Tn}_\RQ$ being admissible means that all DPIs $\tuple{\mo\setminus\mathsf{node},\mb,\Tp,\Tn}_\RQ$ must be admissible as well since admissibility is defined only by the sets $\mb,\Tp,\Tn,\RQ$ which remain invariant throughout the execution of Algorithm~\ref{algo:hs}. 
%Note that case (d) can only arise for when labeling the root node $\emptyset$ as $\tuple{\mo,\mb,\Tp,\Tn}_\RQ$ being admissible means that all DPIs $\tuple{\mo\setminus\mathsf{node},\mb,\Tp,\Tn}_\RQ$ must be admissible as well since admissibility is defined only by the sets $\mb,\Tp,\Tn,\RQ$ which remain invariant throughout the execution of Algorithm~\ref{algo:hs}. 
Further on, we point out that the conflict set $L$ in case (b) must be a \emph{new} conflict set since the reuse criterion is always checked \emph{before} the call to $\scQX$ and thus must be negative. That is, each $\mc\in\mC_{calc}$ is hit by $\mathsf{node}$ and $L$ is not hit by $\mathsf{node}$ wherefore $L \neq \mc$ must hold for all $\mc\in\mC_{calc}$. 

In each of the described cases, the \textsc{label} procedure returns a tuple including the respective label as explained and the set $\mC_{calc}$ where $\mC_{calc}$ is equal to the input argument $\mC_{calc}$ in all cases except for the case where a new minimal conflict set is computed by $\scQX$. In this case, the newly computed conflict set is added to $\mC_{calc}$ (line~\ref{algoline:hs:add_conflict}) before the procedure returns.

%\noindent\textbf{Processing of a Node Label.} 
\paragraph{Processing of a Node Label.}
Back in the main procedure, $\mC_{calc}$ is updated (line~\ref{algoline:hs:update_Ccalc}) and then the label $L$ returned by procedure \textsc{label} is processed as follows: 

If $L = valid$, then there is no minimal conflict set w.r.t.\ $\langle\mo,\mb,\Tp,\Tn\rangle_\RQ$ that is not hit by (i.e.\ has an empty intersection with) the current node $\mathsf{node}$. 
%(which is a set of formulas corresponding to the edge labels along the path from the root node to $\mathsf{node}$). 
Thus, $\mathsf{node}$ is added to the set of calculated minimal diagnoses $\mD_{calc}$. Minimality of diagnoses added to $\mD_{calc}$ is guaranteed by the pruning rule (lines~\ref{algoline:hs:non_min_crit_start}-\ref{algoline:hs:non_min_crit_end}) which eliminates non-minimal nodes (paths) and the way the tree is built level by level by the used breadth-first strategy. In case a uniform-cost variant of tree construction is used, certain properties of the function $p()$ need to be postulated to preserve this minimality guarantee. We discuss these properties in Section~\ref{sec:DiagnosisProbabilitySpace}.

If, on the other hand, $L=closed$ is the returned label of the procedure $\textsc{label}$, then there is either a minimal diagnosis in $\mD_{calc}$ that is a subset of the current node $\mathsf{node}$ or a duplicate of $\mathsf{node}$ is already included in $\Queue$. Consequently, $\mathsf{node}$ must simply be removed from $\Queue$ which has already been executed in line~\ref{algoline:hs:remove_from_queue}.

In the third case, if a minimal conflict set $L$ is returned in line~\ref{algoline:hs:label}, then $L$ is a label for $\mathsf{node}$ meaning that $|L|$ successor nodes of $\mathsf{node}$ need to be added to $\Queue$ in sorted order using the function $p_{nodes}()$ (\textsc{insertSorted}, line~\ref{algoline:hs:generate_nodes}), as will be explained in more detail in Section~\ref{sec:DiagnosisProbabilitySpace}.

%\noindent\textbf{Recap.} 
\paragraph{Recap.}
To summarize, in each iteration, the node $\mathsf{node}$ that is the first element of the queue $\Queue$ is deleted from $\Queue$ and,
\begin{enumerate}
	\item if $\mathsf{node}$ is a diagnosis, it is added to the set $\mD_{calc}$
	\item if there is some diagnosis in $\mD_{calc}$ that is a proper subset of $\mathsf{node}$ or $\mathsf{node}$ is equal to some other node in $\Queue$, no action is performed, i.e.\ the algorithm deletes $\mathsf{node}$ without substitution
	\item if there is some minimal conflict set that $\mathsf{node}$ does not hit, then such a conflict set $\mc$ is computed and for each $\tax \in \mc$ a new node $\mathsf{node} \cup \setof{\tax}$ is added to $\Queue$.
\end{enumerate}
We call each node $\mathsf{nd}$ that is added to $\Queue$ in the latter case a \emph{successor of the node $\mathsf{node}$}. 
%Alternatively, we write $\mathsf{nd} \in succ(\mathsf{node})$.\label{etc:def_successor_of_node}

\subsection{Correctness of Breadth-First Diagnosis Computation}
\label{sec:CorrectnessOfBreadthFirstDiagnosisComputation}
For the discussion of the output of Algorithm~\ref{algo:hs} we will exploit the following result saying that Algorithm~\ref{algo:hs} computes all and only minimal diagnoses, if it executes until the queue of open nodes becomes the empty set. 
%which is a direct consequence of Proposition~\ref{prop:pruned_hs_tree_finds_all_min_diags} and the fact that Algorithm~\ref{algo:hs} using a breadth-first tree construction obviously creates a pHS-tree as per Definition~\ref{def:pruned_hs_tree}.
%established by Theorem~4.8 in~\cite{Reiter87}, applied to our Algorithm~\ref{algo:hs}:
\begin{proposition}[Soundness and Completeness of Algorithm~\ref{algo:hs} using Breadth-First Search]\label{prop:hs-tree_bfs_soundness_completeness}
Let $\langle\mo$, $\mb,\Tp,\Tn\rangle_\RQ$ be an admissible DPI given as input to Algorithm~\ref{algo:hs}. If Algorithm~\ref{algo:hs} using a breadth-first tree construction strategy terminates due to $\Queue = []$, 
%i.e.\ the pHS-tree w.r.t.\ $\langle\mo,\mb,\Tp,\Tn\rangle_\RQ$ has been constructed, 
then the algorithm returns exactly the set of all minimal diagnoses w.r.t.\ $\langle\mo,\mb,\Tp,\Tn\rangle_\RQ$. 
\end{proposition}
\begin{proof}
%\fixme{check for correctness regarding the corollary below}
This proposition is a consequence of Proposition~\ref{prop:pruned_hs_tree_finds_all_min_diags} and the following Lemma~\ref{lem:algo_hs_produces_pruned_hs_tree} which witnesses that Algorithm~\ref{algo:hs} using a breadth-first tree construction strategy produces a pHS-tree as per Definition~\ref{def:pruned_hs_tree}.
\end{proof}
%\begin{proof}
%It is easy to see that Algorithm~\ref{algo:hs} with parameter setting $n_{\min} = \infty$ can terminate only due to $\Queue = \emptyset$, i.e.\ . This however applies only if all nodes (paths) in the HS-tree produced by the algorithm are closed. For some function $p()$ with $p(S_1) \geq p(S_2)$ iff $S_1 \subseteq S_2$ for $S_1,S_2 \subseteq \mo$ this holds directly due to Theorem~4.8 in \cite{Reiter87} since such a definition of $p()$ leads to a breadth-first exploration of the tree, as in \cite{Reiter87}. That is, the proposition holds for a function $p()$ that yields to the finding of minimal diagnoses first.
%
%We now show that the proposition still holds if $p()$ is defined in any other way. In this case, non-minimal diagnoses might be identified before minimal diagnoses that are subsets thereof. So, some non-minimal diagnoses w.r.t.\ $\langle\mo,\mb,\Tp,\Tn\rangle_\RQ$ might be in $\mD_{calc}$. But, nevertheless all minimal diagnoses w.r.t.\ $\langle\mo,\mb,\Tp,\Tn\rangle_\RQ$ will be in $\mD_{calc}$. This holds since a (subset of a) minimal diagnosis $\md$ can never be closed due to a diagnosis $\md' \supset \md$ in $\mD_{calc}$ by the fact that \textsc{label} can only return $closed$ for some node $\mathsf{node} \supseteq \md'$. That is, $\mathsf{node} \neq \md_{sub}$ for all $\md_{sub} \subseteq \md$.   
%\end{proof}
\begin{lemma}\label{lem:algo_hs_produces_pruned_hs_tree} Algorithm~\ref{algo:hs} with the admissible input DPI $\langle\mo,\mb,\Tp,\Tn\rangle_\RQ$ using a breadth-first tree construction strategy is a procedure for producing a pHS-tree $T$ w.r.t.\ $\langle\mo,\mb,\Tp,\Tn\rangle_\RQ$.
% if Algorithm~\ref{algo:hs} returns due to $\Queue = \emptyset$.
\end{lemma}
\begin{proof}
We verify whether all rules given by Definitions~\ref{def:hs_tree} and \ref{def:pruned_hs_tree} are satisfied by Algorithm~\ref{algo:hs}.
\begin{itemize}
\item Definition~\ref{def:hs_tree}, rule 1: The root node $\emptyset$ which is the only element of the initial list $\Queue$ is labeled by the first call to \textsc{label} for $\mathsf{node} := \emptyset$ in line~\ref{algoline:hs:label}. If $valid$ is returned, then $\scQX(\langle\mo,\mb,\Tp,\Tn\rangle_\RQ)$ must have returned 'no conflict' which is the case if $\mo$ is valid w.r.t.\ $\langle\cdot,\mb,\Tp,\Tn\rangle_\RQ$. 

Otherwise, if $valid$ is not returned by \textsc{label}, then some minimal conflict set $L$ w.r.t.\ $\langle\mo,\mb,\Tp,\Tn\rangle_\RQ$ must have been returned in line~\ref{algoline:hs:return_computed_cs}. $L$ is a minimal conflict set w.r.t.\ $\langle\mo,\mb,\Tp,\Tn\rangle_\RQ$ by Proposition~\ref{prop:qx_correctness} and since $\scQX(\langle\mo,\mb,\Tp,\Tn\rangle_\RQ)$ has not returned 'no conflict' as otherwise $valid$ would have been returned contradicting our assumption and since $\langle\mo,\mb,\Tp,\Tn\rangle_\RQ$ is an admissible DPI by assumption. \textsc{label} cannot have returned earlier in line~\ref{algoline:hs:non_min_crit_end} or line~\ref{algoline:hs:duplicate_crit_end}, since $\mD_{calc}$ is the empty set and $\Queue$ the empty list at this time. The former holds since $\mD_{calc}$ is only extended in line~\ref{algoline:hs:update_Dcalc} which cannot ever have been reached before the first call to \textsc{label} has returned. The latter holds as $\Queue$ initially contained only $\emptyset$ and as $\emptyset$ was deleted from $\Queue$ in line~\ref{algoline:hs:remove_from_queue} before the call to \textsc{label} was made in line~\ref{algoline:hs:label}.

\item Definition~\ref{def:hs_tree}, rule 2: Suppose a node $\mathsf{node}$ is labeled by $valid$, then it is added to $\mD_{calc}$ in line~\ref{algoline:hs:update_Dcalc}. Since $\mathsf{node}$ can only get a label different from $closed$ if it is the only exemplar of this node in $\Queue$ due to the duplicate criterion (lines~\ref{algoline:hs:duplicate_crit_start}-\ref{algoline:hs:duplicate_crit_end}), it must be the case that $\mathsf{node}\notin\Queue$ (line~\ref{algoline:hs:remove_from_queue}) after $\mathsf{node}$ has been labeled by $valid$. Only nodes that get labeled by a conflict set can have successor nodes added to $\Queue$ in line~\ref{algoline:hs:generate_nodes}. Only nodes in $\Queue$ can get a label (cf.\ lines~\ref{algoline:hs:getfirst} and \ref{algoline:hs:label}). For $\mathsf{node}$ to be added to $\Queue$ at some later point in time there must be a proper subset of $\mathsf{node}$ that is still in $\Queue$ as each node newly added to $\Queue$ is a proper superset of some node in $\Queue$ (cf.\ line~\ref{algoline:hs:generate_nodes} which is the only position in the algorithm where nodes are added to $\Queue$). This is impossible due to the breadth-first tree construction strategy which implies that all nodes of cardinality $|\mathsf{node}|-1$ have already been labeled (and thus deleted from $\Queue$ in line~\ref{algoline:hs:remove_from_queue}) when $\mathsf{node}$ is being labeled. Hence, if $\mathsf{node}$ is labeled by $valid$, then it has no successors.

If $\mathsf{node}$ is labeled by some conflict set $L$, then Algorithm must come to line~\ref{algoline:hs:generate_nodes}, where a successor $\mathsf{node} \cup \setof{e}$ is added to $\Queue$ for all $e \in L$.

How node $\mathsf{node}_e := \mathsf{node} \cup \setof{e}$ must be labeled is overridden by the rules 3, 4 and 6 of Definition~\ref{def:pruned_hs_tree} (see below). 

\item Definition~\ref{def:pruned_hs_tree}, rule 1: This is true by our assumption about $p()$ and $p_{nodes}()$.
%That this is true for a breadth-first tree construction was already argued in Definition~\ref{def:hs_tree}, rule 2. Notice that a node $\mathsf{node}$ in Algorithm~\ref{algo:hs} is defined as the set $H(\mathsf{node})$ of Definitions~\ref{def:hs_tree} and \ref{def:pruned_hs_tree} (cf.\ Remark~\ref{rem:algo_hs:internal_representation}).

\item Definition~\ref{def:pruned_hs_tree}, rule 2: This holds since $\scQX(\langle\mo\setminus\mathsf{node},\mb,\Tp,\Tn\rangle_\RQ)$ computes only minimal conflict sets w.r.t.\ $\langle\mo,\mb,\Tp,\Tn\rangle_\RQ$ (cf.\ Remark~\ref{rem:qx_with_O_setminus_node_yields_cs_wrt_O}).

\item Definition~\ref{def:pruned_hs_tree}, rule 3: All minimal conflict sets that have been used to label nodes so far are stored in $\mC_{calc}$. Before a minimal conflict to label $\mathsf{node}$ might be computed by a call to $\scQX$ in line~\ref{algoline:hs:qx}, the reuse criterion in lines~\ref{algoline:hs:reuse_crit_start}-\ref{algoline:hs:reuse_crit_end} checks whether there is a set $\mc$ in $\mC_{calc}$ with $\mc \cap \mathsf{node}$. If positive, $\mc$ is returned as a label for $\mathsf{node}$.

\item Definition~\ref{def:pruned_hs_tree}, rule 4: This is accomplished by the non-minimality criterion in lines~\ref{algoline:hs:non_min_crit_start}-\ref{algoline:hs:non_min_crit_end} which checks for existence of a node already labeled by $valid$ which is a subset of the node to be labeled right now. All nodes labeled by $valid$ are stored in $\mD_{calc}$ (cf. lines~\ref{algoline:hs:L=valid} and \ref{algoline:hs:update_Dcalc}).

\item Definition~\ref{def:pruned_hs_tree}, rule 5: If some node $\mathsf{node}$ is labeled by $closed$, then no action is performed (cf.\ line~\ref{algoline:hs:do_nothing}). Before each node is labeled in line~\ref{algoline:hs:label}, it is deleted from $\Queue$ in line~\ref{algoline:hs:remove_from_queue}. That $\mathsf{node}$ cannot be inserted into $\Queue$ at some later point in time follows from the argumentation used above to demonstrate that Definition~\ref{def:hs_tree}, rule 2 is met.

\item Definition~\ref{def:pruned_hs_tree}, rule 6: This is achieved by the duplicate criterion in lines~\ref{algoline:hs:duplicate_crit_start}-\ref{algoline:hs:duplicate_crit_end} where $\Queue$ is browsed for some node equal to the one that is to be labeled right now. 
When some node $\mathsf{node}$ is next to be labeled, then all duplicates of $\mathsf{node}$ must already be in $\Queue$ as reasoned above in the argumentation to show that Definition~\ref{def:hs_tree}, rule 2 is satisfied. Thus, the criterion must search for duplicates in no other collections than $\Queue$. Indeed, only one (i.e.\ the last non-deleted) exemplar of these duplicates of $\mathsf{node}$ in $\Queue$ can get a label other than $closed$ due to the duplicate criterion which closes duplicates as long as there are any.  
%Note that the duplicate of some node to be labeled right now must be in $\Queue$ and not in some other set. This holds because only the last exemplar of some node in $\Queue$ can get a label 
\end{itemize}
We conclude that Algorithm~\ref{algo:hs} is a procedure for constructing a pHS-tree.
\end{proof}
%By the correctness of Lemma~\ref{lem:algo_hs_produces_pruned_hs_tree}, 
By Proposition~\ref{prop:hs-tree_bfs_soundness_completeness} and the fact that there is no place in Algorithm~\ref{algo:hs} where nodes are removed from $\mD_{calc}$ (which
%, by Proposition~\ref{prop:hs-tree_bfs_soundness_completeness}, 
implies that only \emph{minimal} diagnoses can be added to $\mD_{calc}$), the following corollary is obvious.
\begin{corollary}\label{cor:algo_hs_returns_relevant_data_of_(partial)_pruned_hs_tree}
Algorithm~\ref{algo:hs} with the admissible input DPI $\langle\mo,\mb,\Tp,\Tn\rangle_\RQ$ using a breadth-first tree construction strategy stores by $\tuple{\mD_{calc},\Queue,\mC_{calc}}$ the relevant data of 
\begin{itemize}
\item a pHS-tree w.r.t.\ $\langle\mo,\mb,\Tp,\Tn\rangle_\RQ$ if Algorithm~\ref{algo:hs} stops due to $\Queue = []$,
\item a partial pHS-tree w.r.t.\ $\langle\mo,\mb,\Tp,\Tn\rangle_\RQ$ otherwise.
\end{itemize}
\end{corollary}
If a pHS-tree is computed in breath-first order, minimal diagnoses are generated with increasing cardinality, as the following Corollary~\ref{cor:algo_hs_finds_min_card_diags_first} attests. Consequently, for the generation of all minimum cardinality diagnoses, only the first level of the tree has to be generated, where a node is labeled.
\begin{corollary}\label{cor:algo_hs_finds_min_card_diags_first}
The following holds for the set $\mD$ returned by Algorithm~\ref{algo:hs} using breadth-first search: %returns a set $\mD$, then the following holds: 
%$\mD$ is the set of diagnoses of minimum cardinality w.r.t.\ the DPI $\langle\mo,\mb,\Tp,\Tn\rangle_\RQ$ given as input to the algorithm. That is, 
If $\mD$ contains some diagnosis of cardinality $k$, then it includes all diagnoses w.r.t.\ $\langle\mo,\mb,\Tp,\Tn\rangle_\RQ$ of cardinality lower than $k$.
\end{corollary}
\begin{proof}
By Proposition~\ref{prop:hs-tree_bfs_soundness_completeness}, it is a fact that Algorithm~\ref{algo:hs} computes all and only minimal diagnoses w.r.t.\ $\langle\mo,\mb,\Tp,\Tn\rangle_\RQ$. As these are computed in breadth-first order, the first computed diagnoses must be the minimum cardinality ones. To see this, assume that Algorithm~\ref{algo:hs} returns $\mD$ 
%with $|\mD| = n$ is returned 
which includes one non-minimum cardinality diagnosis $\md$ and does not comprise a minimum cardinality diagnosis $\md'$, i.e.\ $|\md| > |\md'|$. By breadth-first search, 
nodes are labeled in ascending order of their cardinality. And, if the first node of cardinality $k$ is labeled, no more nodes of cardinality $k-1$ can be in $\Queue$ (cf.\ proof of Lemma~\ref{lem:algo_hs_produces_pruned_hs_tree}).
%diagnoses are explored by ascending cardinality, 
So, we have that the pHS-tree obtained by further execution of the algorithm until $\Queue = []$ can never label $\md'$ since $|\md| > |\md'|$ and $\md$ has already been labeled. Hence, the algorithm would not return $\md'$ in its final output $\mD$. 
Since each minimum cardinality diagnosis is a minimal diagnosis, $\md'$ is a minimal diagnosis. Thus, we have a contradiction to the fact that the algorithm computes all minimal diagnoses.
%By soundness of the algorithm established by Proposition~\ref{prop:hs-tree_bfs_soundness_completeness}, $\md$ 
\end{proof}

%\noindent\textbf{Output.} 
\paragraph{Output.}
The repeat-loop is iterated until the stop criterion (line~\ref{algoline:hs:until}) applies. 
%The algorithm outputs a set of minimal diagnoses $\mD$ w.r.t.\ $\tuple{\mo,\mb,\Tp,\Tn}_\RQ$ which is (a)~a set of best (as per $p()$) 
%, see Section~\ref{sec:DiagnosisProbabilitySpace}) 
% minimal diagnoses w.r.t.\ $\langle\mo,\mb,\Tp,\Tn\rangle_\RQ$ such that $n_{\min} \leq |\mD| \leq n_{\max}$, if at least $n_{\min}$ minimal diagnoses exist w.r.t.\ $\langle\mo,\mb,\Tp,\Tn\rangle_\RQ$, or (b)~the set of all minimal diagnoses w.r.t.\ $\langle\mo,\mb,\Tp,\Tn\rangle_\RQ$ otherwise. 
In case at least $n_{\min}$ minimal diagnoses w.r.t.\ $\langle\mo,\mb,\Tp,\Tn\rangle_\RQ$ exist, there are two cases:
\begin{itemize}
\item If the finding of the $n_{\min}$-th minimal diagnosis happens after $t' < t$ time has passed since the start of Algorithm~\ref{algo:hs}, then the algorithm will continue iterating and terminate only if execution time amounts to at least $t$ time or $|\mD|=n_{\max}$ at the time line~\ref{algoline:hs:until} is processed.
\item Otherwise, if the detection of the $n_{\min}$-th minimal diagnosis takes place after processing longer than $t$ time, then the algorithm will terminate immediately after having determined the $n_{\min}$-th minimal diagnosis.
\end{itemize}
In both cases, the output is a set $\mD$ of minimal diagnoses w.r.t.\ $\langle\mo,\mb,\Tp,\Tn\rangle_\RQ$ such that $n_{\min} \leq |\mD| \leq n_{\max}$ and $\mD$ is the set of best minimal diagnoses as per $p()$, in this case the set of minimal diagnoses with minimum cardinality since $p()$ is assumed to be specified as to cause a breadth-first tree construction. 
%The latter holds since the queue $\Queue$ is processed in descending order prescribed by $p()$. 

If fewer than $n_{\min}$ minimal diagnoses exist w.r.t.\ $\langle\mo,\mb,\Tp,\Tn\rangle_\RQ$, then $\Queue = []$ will be the cause for the algorithm to terminate. In this case, the pHS-tree w.r.t.\ $\langle\mo,\mb,\Tp,\Tn\rangle_\RQ$ has been built up and all minimal diagnoses w.r.t.\ $\langle\mo,\mb,\Tp,\Tn\rangle_\RQ$ are stored in $\mD_{calc}$. Thus, the output is the set $\minD_{\langle\mo,\mb,\Tp,\Tn\rangle_\RQ}$ of all minimal diagnoses w.r.t.\ $\langle\mo,\mb,\Tp,\Tn\rangle_\RQ$.

%Additionally, if $n_{\min}$ minimal diagnoses can be found within the time prescribed by $t$, then Algorithm~\ref{algo:hs} will terminate and return the current set $\mD_{calc}$ after having processed for a time period $t$. If it takes the algorithm longer than $t$ time to compute $n_{\min}$ minimal diagnoses, then it will terminate immediately after the $n_{\min}$-th diagnosis is found.
%Notice that case (b) arises if $\Queue = \emptyset$ applies which means that the complete (pruned) hitting set tree for the given DPI $\langle\mo,\mb,\Tp,\Tn\rangle_\RQ$ has been built up and all minimal diagnoses w.r.t.\ $\langle\mo,\mb,\Tp,\Tn\rangle_\RQ$ are stored in $\mD_{calc}$. 

%\noindent\textbf{Termination.} 
\paragraph{Termination.}
The next proposition shows that Algorithm~\ref{algo:hs} must yield a set of minimal diagnoses after finite time.
\begin{proposition}\label{prop:hs:termination}
Algorithm~\ref{algo:hs} always terminates.
\end{proposition}
\begin{proof}
This is due to the fact that minimal conflict sets used to label non-leaf nodes are subsets of $\mo$ and that nodes in $\Queue$ are subsets of $\mo$, which is a finite set by Definition~\ref{def:dpi}. Moreover, 
%during the execution of the algorithm, 
a node in $\Queue$ is either deleted without substitution from $\Queue$ if $valid$ or $closed$ (line~\ref{algoline:hs:remove_from_queue}) or deleted (line~\ref{algoline:hs:remove_from_queue}) and replaced by proper supersets of it (\textsc{insertSorted} in line~\ref{algoline:hs:generate_nodes}). This means that the cardinality of all nodes in $\Queue$ is strictly monotonically increasing. Thus each node (path) $\mathsf{node}$ is guaranteed to be closed ($valid$ or $closed$) when $\mathsf{node} = \mo$ as in this case $\mathsf{node}$ must hit all possible (minimal) conflict sets $\mc_i$ w.r.t.\ $\langle\mo,\mb,\Tp,\Tn\rangle_\RQ$ since $\mc_i \subseteq\mo$ holds by Definition~\ref{def:cs}. So, after finite time the queue $\Queue$ definitely becomes the empty list which is a stop criterion (line~\ref{algoline:hs:until}).
\end{proof}

The argumentation so far proves the following 
\begin{proposition}\label{prop:hs_bfs_correct}
Let $\langle\mo,\mb,\Tp,\Tn\rangle_\RQ$ be an admissible DPI, $t,n_{\min},n_{\max}\in\mathbb{N}$ 
%$t$ some computation timeout, $n_{\min}$ a desired minimal and $n_{\max}$ a desired maximal number of diagnoses to be returned 
%and $p()$ a function $p(\tax) \in 	(0,0.5), \tax\in\mo$ 
and $p:\mo \rightarrow (0,0.5)$ defined in a way that $\Queue$ is always ordered first-in-first-out.
% if node probabilities are computed as per $p$ and Formula~\ref{eq:path_prob_calc}. 
For these inputs, Algorithm~\ref{algo:hs} always terminates and returns a set $\mD$ of minimal diagnoses w.r.t.\ $\langle\mo,\mb,\Tp,\Tn\rangle_\RQ$ which is 
\begin{itemize}
\item the set of the $|\mD|$ minimal diagnoses of minimum cardinality w.r.t.\ $\langle\mo,\mb,\Tp,\Tn\rangle_\RQ$ (i.e.\ the first $|\mD|$ elements in $\minD_{\langle\mo,\mb,\Tp,\Tn\rangle_\RQ}$ if $\minD_{\langle\mo,\mb,\Tp,\Tn\rangle_\RQ}$ is assumed to be sorted in ascending order by cardinality) such that $n_{\min} \leq |\mD| \leq n_{\max}$, if at least $n_{\min}$ minimal diagnoses exist w.r.t.\ $\langle\mo,\mb,\Tp,\Tn\rangle_\RQ$, or
%a set of minimal diagnoses of minimum cardinality w.r.t.\ $\langle\mo,\mb,\Tp,\Tn\rangle_\RQ$ such that $n_{\min} \leq |\mD| \leq n_{\max}$, if at least $n_{\min}$ minimal diagnoses exist w.r.t.\ $\langle\mo,\mb,\Tp,\Tn\rangle_\RQ$, or 
\item the set of all minimal diagnoses w.r.t.\ $\langle\mo,\mb,\Tp,\Tn\rangle_\RQ$, 
%including all minimum-cardinality diagnoses, 
otherwise.
\end{itemize}
\end{proposition}

\begin{algorithm}[!htp]
\small
\caption{$\scHS$: Computation of Minimal Diagnoses} \label{algo:hs}
\begin{algorithmic}[1]
\Require an admissible DPI $\langle\mo,\mb,\Tp,\Tn\rangle_\RQ$, 
a desired computation timeout $t$, a desired minimal ($n_{\min}$) and maximal ($n_{\max}$) number of diagnoses to be returned, a function $p:\mo \rightarrow (0,0.5)$ 
% that assigns to each $\tax \in \mo$ a probability $$ (of $\tax$ being faulty) and thereby determines the search strategy (breadth-first or uniform-cost)
\Ensure a set $\mD$ which is \newline
(a)~a set of most probable (according to $p()$) minimal diagnoses w.r.t.\ $\langle\mo,\mb,\Tp,\Tn\rangle_\RQ$ such that $n_{\min} \leq |\mD| \leq n_{\max}$, if at least $n_{\min}$ minimal diagnoses exist w.r.t.\ $\langle\mo,\mb,\Tp,\Tn\rangle_\RQ$, or \newline
(b)~the set of all minimal diagnoses w.r.t.\ $\langle\mo,\mb,\Tp,\Tn\rangle_\RQ$ otherwise
\vspace{10pt}
\Procedure{$\scHS$}{$\langle\mo,\mb,\Tp,\Tn\rangle_\RQ, t, n_{\min}, n_{\max}, p()$}
\State $t_{start} \gets \Call{getTime}{ }$
\State $\mD_{calc}, \mC_{calc} \gets \emptyset$
%\State $\Queue \gets \setof{\emptyset}$	
\State $\Queue \gets [\emptyset]$							
\Repeat 																							\label{algoline:hs:repeat}
\State $\mathsf{node} \gets \Call{getFirst}{\Queue}$	\label{algoline:hs:getfirst}
%\State $\Queue \gets \Queue \setminus \setof{\mathsf{node}}$ 	\label{algoline:hs:remove_from_queue}
\State $\Queue \gets \Call{deleteFirst}{\Queue}$ 	\label{algoline:hs:remove_from_queue}
\State $\tuple{L,\mathbf{C}} \gets \Call{label}{\langle\mo,\mb,\Tp,\Tn\rangle_\RQ, \mathsf{node}, \mathbf{C}_{calc}, \mD_{calc}, \Queue}$\label{algoline:hs:label}
\State $\mathbf{C}_{calc} \gets \mathbf{C}$    				\label{algoline:hs:update_Ccalc}
\If{$L = valid$}\label{algoline:hs:L=valid}
	\State $\mD_{calc} \gets \mD_{calc} \cup \setof{\mathsf{node}}$		\label{algoline:hs:update_Dcalc}
\ElsIf{$L = closed$}\label{algoline:hs:do_nothing}  \Comment{do nothing}
	%\State $\\Queue_{closed} \gets \\Queue_{closed} \cup \setof{\mathsf{node}}$
\Else 	\Comment{$L$ must be a minimal conflict set}
%\ElsIf{$L = \mc$}
	\For{$e \in L$}						
		\State $\Queue \gets \Call{insertSorted}{ \mathsf{node} \cup \setof{e}, \Queue, p_{nodes}()}$\label{algoline:hs:generate_nodes}
\EndFor
\EndIf
\Until{$\Queue = [] \lor [|\mD_{calc}| \geq n_{\min} \land ( |\mD_{calc}| = n_{\max} \lor \Call{getTime}{ } - t_{start} > t)]$}  \label{algoline:hs:until}
\State \Return $\mD_{calc}$
\EndProcedure
\vspace{10pt}
\Procedure{\textsc{label}}{$\langle\mo,\mb,\Tp,\Tn\rangle_\RQ,\mathsf{node},\mathbf{C}_{calc},\mD_{calc}, \Queue$}    \label{algoline:hs:procedure_label}
\For{$\mathsf{nd} \in \mD_{calc}$}																								\label{algoline:hs:non_min_crit_start}
	\If{$\mathsf{node} \supseteq \mathsf{nd}$}  \Comment{non-minimality}
			\State \Return $\tuple{closed,\mathbf{C}_{calc}}$														\label{algoline:hs:non_min_crit_end}
	\EndIf
\EndFor
\For{$\mathsf{nd} \in \Queue$}																										\label{algoline:hs:duplicate_crit_start}
	\If{$\mathsf{node} = \mathsf{nd}$}  \Comment{remove duplicates}
			\State \Return $\tuple{closed,\mathbf{C}_{calc}}$														\label{algoline:hs:duplicate_crit_end}
	\EndIf
\EndFor
\For{$\mc \in \mathbf{C}_{calc}$}																									\label{algoline:hs:reuse_crit_start}
	\If{$\mc \cap \mathsf{node} = \emptyset$}\label{algoline:hs:test_cs_not_hit}  \Comment{reuse $\mc$}
		\State \Return $\tuple{\mc,\mathbf{C}_{calc}}$																\label{algoline:hs:reuse_crit_end}
	\EndIf
\EndFor
\State $L\gets \Call{QX}{\langle\mo\setminus\mathsf{node},\mb,\Tp,\Tn\rangle_\RQ}$	\label{algoline:hs:qx}
\If{$L$ = \text{'no conflict'}}   \Comment{$\mathsf{node}$ is a diagnosis}
	\State \Return $\tuple{valid,\mathbf{C}_{calc}}$														\label{algoline:hs:return_valid}
%\ElsIf{$L = \emptyset$}		\Comment{non-admissible DPI}
	%\State \Return $\tuple{closed,\mathbf{C}_{calc}}$
\Else										\Comment{$L$ is \emph{new} minimal conflict set ($\notin \mathbf{C}_{calc}$)}
	\State $\mathbf{C}_{calc} \gets \mathbf{C}_{calc} \cup \setof{L}$							\label{algoline:hs:add_conflict}  
	\State \Return $\tuple{L,\mathbf{C}_{calc}}$\label{algoline:hs:return_computed_cs}
\EndIf
\EndProcedure
\end{algorithmic}
\normalsize
\end{algorithm}

%\fixme{define true diagnosis in a concise way}
\section{Diagnosis Probability Space}
\label{sec:DiagnosisProbabilitySpace}
The induction of a probability space~\cite{durrett2010} over diagnoses facilitates incorporation of well-established probability theoretic methods into the process of KB debugging; for example, a Bayesian approach \cite{Shchekotykhin2012,Rodler2013,dekleer1987} for identifying the \emph{true diagnosis}, i.e.\ the one which leads to a solution KB with the desired semantics, by repeated measurements (see Part~\ref{part:InteractiveKBDebugging}). Let the true diagnosis be denoted as $\dt$ in the sequel. %Intuitively, KB debugging can be interpreted as a random experiment where the debugging system draws (outputs) one diagnosis w.r.t.\ the given DPI. 

%\noindent\textbf{The Probability Space of All Diagnoses.} 
\paragraph{The Probability Space of All Diagnoses.}
From the point of view of probability theory, a diagnosis can be viewed as an atomic event in a probability space $\langle\Omega,\mathcal{E},p\rangle$ defined as follows:
\begin{itemize}
	\item $\Omega$ is the sample space consisting of all possible diagnoses w.r.t. a DPI $\tuple{\mo,\mb,\Tp,\Tn}_\RQ$, i.e. $\Omega = \allD_{\tuple{\mo,\mb,\Tp,\Tn}_\RQ}$,
	\item $\mathcal{E}$ is a sigma-algebra on $\Omega$, in our case 
	%i.e.\ is a collection of all events, where each event is a subset of $\Omega$, i.e. $\mathcal{E}$ is 
	the powerset $2^{\Omega}$ of $\Omega$, and
	\item $p$ is a probability measure assigning a probability to each event in $\mathcal{E}$, i.e. $p:\mathcal{E} \rightarrow [0,1]$ such that $\sum_{\omega\in\Omega} p(\setof{\omega}) = 1$ which means $\sum_{\md\in\allD_{\tuple{\mo,\mb,\Tp,\Tn}_\RQ}} p(\setof{\md}) = 1$.
\end{itemize}
So, $p(\setof{\md})$ for $\md\in\allD_{\tuple{\mo,\mb,\Tp,\Tn}_\RQ}$ can be seen as the probability that $\md$ is the true diagnosis, i.e.\ the probability of the event $\dt = \md$ (or $\dt \in \setof{\md}$). Consequently, $p(\setof{\md})$ for $\md\in\allD_{\tuple{\mo,\mb,\Tp,\Tn}_\RQ}$ is the probability distribution of the random variable $\dt$, i.e.\ the probability distribution of the true diagnosis.
In this vein, the probability of a set $\setof{\md_i,\dots,\md_j} \in \mathcal{E}$ is interpreted as the likeliness of this set to comprise the true diagnosis $\dt$. That is, $p(\{\md_i,\dots,\md_j\}) = p(\dt\in\{\md_i,\dots,\md_j\}) = p(\dt=\md_i \vee \dots \vee \dt=\md_j) = 0.3$ means that $\dt$ is an element of $\setof{\md_i,\dots,\md_j}$ with 30\% probability. Note that singletons are often written without curly braces, i.e.\ $p(\setof{\md_i})$ is usually written as $p(\md_i)$; we will also do so in the rest of this work. 
%\footnote{For brevity, we will write $p(\md_i)$ and $p(\setof{\md_i,\dots,\md_j})$ instead of $p(\dt \in\setof{\md_i})$ and $p(\dt\in\setof{\md_i,\dots,\md_j})$, respectively, below.} 

The elements of the sample space $\Omega$ of a probability space are often called atomic events because they must be \emph{mutually exclusive} (i.e.\ two atomic events cannot ``happen'' at the same time as an outcome of the fictive experiment a probability space describes) and \emph{exhaustive} (i.e.\ for each ``execution'' of the experiment the probability space describes one atomic event must ``happen''). Since the true diagnosis $\dt$ must be a diagnosis w.r.t.\ $\tuple{\mo,\mb,\Tp,\Tn}_\RQ$ and $\Omega$ by definition comprises all such diagnoses, exhaustiveness is clearly fulfilled.
Mutual exclusiveness is a consequence of the fact that each diagnosis $\md$ 
%A diagnosis $\md$ is indeed an atomic event since it 
gives complete information about the correctness of each formula $\tax_k \in \mo$. In other words, $\dt \in \setof{\md}$ is a shorthand for the statement that all $\tax_i \in \md$ are faulty and all $\tax_j \in \mo\setminus\md$ are correct. Thus, any two different diagnoses are mutually exclusive events, i.e. $\dt = \md_i$ implies $\dt \neq \md_j$ for all $\md_j \in \allD$ such that $\md_i \neq \md_j$. 

The probability measure $p$ is completely defined if a probability $p(\md)$ for each diagnosis $\md \in \Omega$ is given. Then, by the mutual exclusiveness of events $\dt \in \setof{\md_i}$ and $\dt \in \setof{\md_j}$ for $\md_i \neq \md_j$, the probability 
\begin{align}
p(E) = \sum_{\md\in E} p(\md)
\end{align}
for each event $E \in \mathcal{E}$.
% contains the true diagnosis $\dt$.

%\noindent\textbf{Restricted Probability Spaces of Diagnoses.}
\paragraph{Restricted Probability Spaces of Diagnoses.} 
In many cases, only a restricted set of diagnoses w.r.t.\ a DPI is considered relevant for the debugging task. That is, the focus is on locating the true diagnosis among a predefined subset of all diagnoses $\allD_{\tuple{\mo,\mb,\Tp,\Tn}_\RQ}$.
%Note that a restriction of the set of all diagnoses w.r.t.\ a DPI to a set of diagnoses that is considered relevant also involves an adaptation of the probability space, in particular of the set $\Omega$. 
This involves an adaptation of the probability space, in particular of the set $\Omega$. For instance, if not the set of all, but only the set of minimal diagnoses $\minD_{\tuple{\mo,\mb,\Tp,\Tn}_\RQ}$ w.r.t.\ $\tuple{\mo,\mb,\Tp,\Tn}_\RQ$ should be considered by a debugging system -- as motivated in Section~\ref{sec:MinimallyInvasiveOntologyDebugging} --  then $\Omega := \minD_{\tuple{\mo,\mb,\Tp,\Tn}_\RQ}$. The other properties $\mathcal{E} = 2^{\Omega}$ and $\sum_{\omega\in\Omega} p(\setof{\omega}) = 1$ remain the same for each restricted probability space, but depend on $\Omega$. Thus, for example, a probability $p(\md)$ for $\md \in \minD_{\tuple{\mo,\mb,\Tp,\Tn}_\RQ} \subseteq \allD_{\tuple{\mo,\mb,\Tp,\Tn}_\RQ}$ must be generally defined differently, i.e.\ assigned a higher value, when $\Omega = \minD_{\tuple{\mo,\mb,\Tp,\Tn}_\RQ}$ instead of $\Omega = \allD_{\tuple{\mo,\mb,\Tp,\Tn}_\RQ}$. This is due to the condition that all probabilities of atomic events in $\Omega$ must sum up to 1. In practice, because of the computational complexity of diagnosis computation, the used probability space will usually need to be restricted even further in that $\Omega$ comprises only a set of ``leading diagnoses'' which is a subset of all minimal diagnoses w.r.t.\ a DPI (see Chapter~\ref{chap:UserInteraction}).

%\noindent\textbf{Estimation/computation of diagnosis probability space.}
\subsection{Construction of a Probability Space} 
\label{sec:prob_space_construction}
%\fixme{ cite \cite{Kalyanpur2006} in this section at the place of ``common mistakes'' and ``provenance''}
Since a diagnosis constitutes an assumption about the correctness of each formula in the KB, the probability of a diagnosis $\md$ (to be the true diagnosis $\dt$) can be computed by means of fault probabilities of formulas. In other words, computing the probability of the event $\md = \dt$ corresponds to computing the probability of the event that exactly all formulas in $\md$ are faulty and all other formulas in the KB are correct.

\subsubsection*{Estimating Fault Probabilities of Formulas in the KB}
%\noindent\textbf{Estimating Formula Probabilities.} 
Next we discuss various possibilities of how the probability of an $\tax\in\mo$ might be assessed. To this end, we first make a distinction between situations where some useful empirical data is available or not and then we differentiate between different sorts of such available data and how to take advantage of it. 

%\noindent\textbf{Empirical Data is Accessible.} 
\paragraph{Empirical Data is Accessible.}
Let us first reflect on how to utilize different empirical data sources in order to compute formula probabilities. Data can be of the following kinds (enumeration may not be complete): 
\begin{enumerate}[(a)]
	\item Regarding formulas: Change logs of formulas in the KB
	\item Regarding the user: Data about common mistakes of the user who has formulated the KB
			%-- past identified diagnoses
			%-- change logs of formulas in KBs the user has created in the past
\end{enumerate}
%in collaborative setting: provenance information and level of expertise w.r.t.\ certain terms or logical operators of people involved in evolution of the KB

\noindent\emph{Ad (a):} Prerequisite for the availability of change logs of formulas in the KB is the usage of some KB engineering software with integrated
%which provides the opportunity of 
logging or change management. Examples of such KB (ontology) developing environments are Prot\'{e}g\'{e}~\cite{Noy2000}, Web Prot\'{e}g\'{e}~\cite{Tudorache2013}, SWOOP~\cite{Kalyanpur2006b}, OntoEdit~\cite{Sure2002} or KAON2.\footnote{\url{http://kaon2.semanticweb.org/}} %, with integrated logging or change management. 
Given a formula $\tax\in\mo$ and its change log, the fault probability $p(\tax)$ of this formula can be estimated by counting the number of modifications accomplished for $\tax$ in the change log. The intuition is, the more often $\tax$ has been altered, the more uncertain the (set of) author(s) might be about its correctness. This method of probability computation however suffers from a cold-start problem. If a KB is completely newly created, then such information is not available at all. On the other hand, for KBs that are being developed over a long period of time, this method can be assumed to be a rather reliable way of assessing the likeliness of formulas to be faulty.\\

\noindent\emph{Ad (b):} Clearly, data about common mistakes of a user has to be related to some type of entity that is recurrent and not dependent on a particular KB. Formulas are therefore not suitable and too coarse-grained since one and the same formula will rarely occur in many KBs. More adequate entities to relate a user fault to are predicates (terms) and logical connectives -- these usually (re-)appear in many different KBs. In this way, the extrapolation and reusability of collected personal fault information of a user within one KB and between different KBs is granted.    

One way of obtaining data about common mistakes of user $u$ on this syntactical level is, for instance, the examination of diagnoses got as a result of past debugging sessions performed on KBs authored by $u$. Another way is, again, to use the change logs (if available) of formulas in KBs user $u$ has created in the past. 

Given such a past diagnosis $\md$, we know that all formulas $\tax\in\md$ \emph{that had been written by $u$} have been confirmed to be faulty by a user. So, these formulas could be analyzed for contained predicates (terms) and logical connectives and the probability of being faulty of those syntactical constructs could be raised relative to those constructs that do not occur in formulas in $\md$. At this, the following assumptions could be made:
\begin{itemize}
\item If a formula has been confirmed to be faulty by the user, then the meaning of all predicates (terms) appearing in this formula is not correct (because in the domain that should be modeled the relationship between the predicates (terms) occurring in the formula stated by the formula must not hold). So, all predicates (terms) in $\tax$ get more suspicious of being faulty \emph{in general} if $\tax\in\md$ for some past solution diagnosis $\md$.
\item If a formula including some logical connective is part of some past solution diagnosis, then this type of logical connective gets more suspicious of being faulty \emph{in general}.
\end{itemize}

When exploiting change logs of formulas \emph{authored by $u$}, the following assumptions could be made: 
%At this, the following assumptions could be made: 
\begin{itemize}
\item If a formula has been modified, then a user has changed the meaning of all predicates (terms) appearing in this formula. So, all predicates (terms) in $\tax$ get more suspicious of being faulty \emph{in general} if $\tax$ has been edited at least once. The more often it has been altered, the more suspicious the predicates (terms) get.
%the more frequently the 
\item If some logical connective in a formula is modified, i.e.\ deleted or added, then this type of logical connective gets more suspicious of being faulty \emph{in general}.
\end{itemize}

The following example should give an intuition of these assumptions:
\begin{example}
Imagine the situation where the author of formula $\tax := \forall X\, pet(X) \leftrightarrow animal(X) \land (\exists Y hasOwner(X,Y) \land person(Y))$ is known to have only vague knowledge about the predicate $pet$ and to frequently interchange $\land$ and $\lor$ when formulating logical formulas. This could be reflected by the assignment of higher fault probability to the predicate $pet$ than to the predicates $animal, hasChild$ and $person$ and by raising the fault probability of $\land$ as well as $\lor$ compared to other logical connectives available in the used logic $\mathcal{L}$. Then, formula $\tax$ should intuitively have a higher probability of being faulty than, e.g., formula $\tax' = \forall X\, animal(X) \rightarrow \lnot person(X)$ since $\tax'$ does not include any of the ``suspicious'' terms or connectives as $\tax$ does.\qed
\end{example}

A probability of $0.25$ of some predicate (term) $a$ occurring in $\mo$ could then account for the observation made in the logs that, in past debugging sessions (not necessarily related to the current KB $\mo$), every fourth formula formulated by user $u$ which includes the term $a$ was modified at least once. Similarly, another term $b$ could be assigned fault probability $0.5$ which could reflect that formulas formulated by $u$ including $b$ have been altered twice as often as formulas formulated by $u$ comprising $a$. Given additionally that $a$ occurred in two formulas formulated by $u$ of past diagnoses whereas $b$ did not occur in any, the probability of $a$ could be increased by some addend or factor to take account of this. 

Concerning some logical connective, say $\exists$, the observation that all past diagnosis formulas contained $\exists$ and in $80\%$ of formulas formulated by this user including $\exists$ the $\exists$ connective has been modified at least once, the fault probability of $\exists$ might be assigned rather high. In comparison, the probability of some other connective, say $\lnot$, occurring in no diagnosis and having been altered only in $10\%$ of the formulas comprising $\lnot$, the probability of the $\lnot$ connective might be estimated rather low.
%by the same user who originally formulated the formula. 
%Such modifications could be interpreted as uncertainty of the author(s) about the correctness of a formula, i.e.\ about the correct meaning of term $A$ (or any other term in the formula).

A shortcoming of this approach is again a cold-start problem. If a user is new to conceptualizing knowledge in a structured logical manner or at least in the given logical language $\mathcal{L}$, then no such (personalized) past diagnoses or change logs will be available. So, this issue especially concerns beginners who are usually anyhow more prone to errors than expert-users. On the positive side, utilization of such empirical data can yield to fault information that is very well tailored for the user and that can imply a significant reduction of computation time and user effort necessary for debugging of the KB at hand~\cite{Shchekotykhin2012}. 

%\noindent\textbf{No Empirical Data is Available.} 
\paragraph{No Empirical Data is Available.}
If no data of the kinds (a) and (b) discussed above is available to a debugging system, then we have the following possibilities:
\begin{enumerate}[(a)]
	\setcounter{enumi}{+2}
	\item Common fault patterns
	\item Subjective self-assessment of a user
	\item Examination of structural complexity of logical formulas
	\item Using no probabilities
\end{enumerate}

\noindent\emph{Ad (c):} A common fault pattern~\cite{Rector2004,Corcho09,Kalyanpur2006}, also called anti-pattern, refers to a set of formulas that either leads to an inconsistency (logical anti-pattern) or corresponds to a potential modeling error that -- alone -- does not lead to a inconsistency or incoherency (non-logical anti-pattern), but still might become a source of inconsistency if merged with other formulas (cf.\ Section~\ref{sec:BackgroundKnowledge}).
Although most of these patterns incorporate more than one formula which makes the individual consideration of a formula in terms of fault probability calculation difficult, an idea to incorporate knowledge about anti-patterns to probability estimation of formulas could be to count for each $\tax\in\mo$ in how many different (logical or non-logical) anti-patterns it occurs. The higher this count, the more likely a formula might be involved in a conflict set and thus in the true diagnosis. 

A drawback of this method could be that most of the formulas involved in a KB might not correspond to any formula occurring in an anti-pattern. Thus, one might end up with no probability estimate for most of the formulas in a KB $\mo$. Besides that, the information provided by these anti-patterns is not personalized at all 
and therefore might significantly diverge from the true fault probabilities for a user and lead to a false bias in the used fault data. This justifies to basically rely on another approach to get a first estimate of a formula's likeliness of being faulty and use this method only to make adaptations to already established probabilities.\\

\noindent\emph{Ad (d):} The method of a user's self-assessment of own fault probabilities supposes a user to be able to specify fault probabilities of predicates (terms), logical connectives or complete formulas by themselves. Since users not always have a clear picture of own strengths and weaknesses, this variant must be regarded with suspicion. Furthermore, in settings where several persons are involved in the engineering of the KB, a reasonable rating of fault probabilities of terms, connectives or formulas authored by other persons might be difficult or impossible for a user.\\

\noindent\emph{Ad (e):} Here the idea is to examine ``grammatical'' (i.e.\ syntactical) aspects of formulas such as the ``nesting depth'' of subordinate clauses or the mere ``length'' of a formula. The underlying assumption can be that higher length and/or deeper nesting  means higher complexity and cognitive difficulty in understanding of the formula's semantics -- as it does in natural language. For instance, it is reasonable to expect formulas like $\tax_1 := \forall X\, a(X) \rightarrow (\exists Y\, r_1(X,Y) \land (\forall Z\, r_2(Y,Z) \rightarrow b(Z)))$ 
%with a maximum ``nesting depth'' of two 
to tend to be more error-prone and more likely to be faulty than $\tax_2 := \forall X\, g(X) \rightarrow b(X)$. 
%with a maximum ``nesting depth'' of zero. 
This intuition is modeled by the maximum nesting depth as well as by the length of $\tax_1$ in comparison to $\tax_2$. Using the analogy to natural language, the maximum nesting depth of a formula could roughly be defined as the maximum number of encapsulated subordinate clauses that cannot be ``flattened'' occurring in the natural language translation of the formula. For formula $\tax_1$, this would imply a maximum nesting depth of two; for $\tax_2$ it would amount to zero. The reason is that $\tax_1$ stated in natural language would sound ``if somebody $X$ is $a$, then there is somebody $Y$, who satisfies property $r_1$ with $X$ and for whom anybody, who satisfies property $r_2$ with $Y$ is $b$''. In this natural language formulation, there are two subordinate clauses, i.e.\ the clauses beginning with the word ``who''; the first is at nesting depth one and the second at depth two. These subordinate clauses cannot be flattened, i.e.\ be brought to some lower depth, because the $Z$ is related to the $Y$ which in turn is related to the $X$.
%by property $r_1$ and $b$ holds for anybody $Z$, who relates to $Y$ by property $r_2$''. 
%Roughly, the maximum nesting depth of a formula could be defined as the maximum number of complex formulas that are part of a complex formula where every formula that is not a predicate $P(t_1,\dots,t_k)$ for terms $t_1,\dots,t_k$ and $k \geq 1$ is a complex formula. As per this definition, the maximum nesting depth of $\tax_1$ would amount to two; for $\tax_2$ it would be zero. 
%could has a maximum nesting depth of two since on the right side we have two nested ``complex'' formulas, i.e.\ one ``complex'' formula that is part of another ``complex'' formula. The explanation of a nesting depth of zero of $\tax_2$ is that left and right sides contain only a ``simple'' formula. 
The length of formulas could be defined similarly as in~\cite{Horridge2008} which provides such a definition for DL languages. % which are fragments of first-order logic~\cite{Baader2007}. 
In this case the length of $\tax_1$ and $\tax_2$ would be four (roughly: four predicates in $\tax_1$) and two (two predicates in $\tax_2$), respectively. 

A disadvantage of such a ``grammatical'' approach gets evident when most of the formulas in a KB are rather ``simple'', i.e.\ have a low nesting depth and a short length. In such case this method will give little differentiation between different formulas and should thus be combined with another method of probability estimation in general.\\

\noindent\emph{Ad (f):} In a situation where all the aforementioned ways of gauging probabilities do not apply or are believed to have a too high risk of introducing a false bias into the debugging system, the solution is to define all formulas to be equally probably faulty. The obvious pro of this is that the system cannot get misled by unreasonable fault probabilities whereas the con is that possibly well-suited probabilistic information cannot be exploited. Moreover, experiments in our previous work \cite{Shchekotykhin2012} have manifested that fault information of only ``average'' quality most often leads to a better performance than no fault information. Apart from that, we have suggested a reinforcement learning ``plug-in'' to a debugger which could successfully mitigate the negative effect of low-quality fault information and in many cases, in spite of the low-quality fault information, even led to lower resource consumption (user, time) than a debugger without this plug-in using good fault information \cite{Rodler2013}.

%\noindent\textbf{Collaborative KB Development.} 
\paragraph{Collaborative KB Development.}
In a collaborative development scenario involving several authors, provenance information could be additionally leveraged to refine probability estimates (cf.\ \cite{Kalyanpur2006}). At this point, user skills could come into play; that is, formulas authored by more experienced authors get a lower overall fault probability as opposed to beginners concerning KB engineering or logic skills or expertise in the modeled domain. This probability adaptation can also affect syntactical elements in that one and the same predicate (term) or logical connective can get a different probability depending on in which formula it occurs and who authored that formula.

\begin{remark}\label{rem:problems_with_probs_of_syntactical_elements}
Of course, these assumptions and methods of obtaining fault probabilities of syntactical elements and formulas are only \emph{some} possible ways of doing so. For example, one might argue that the ``authorship'' of a formula is somewhat not clearly defined. What if user $u_1$ has originally written formula $\tax$ and then user $u_2$ alters the formula to become $\tax'$? Who is the author of $\tax'$? $u_1$, $u_2$ or both? For whose fault probability computation should the renewed modification of $\tax'$ to $\tax''$ count? Questions like this one need to be discussed and maybe evaluations using real data need to be accomplished in order to find a practical answer; or perhaps to find out that completely different approaches turn out to be reasonable. This is a topic of our future work.\qed
\end{remark}

\begin{remark}\label{rem:ax_prob_not_zero}
By the definition of a DPI (Definition~\ref{def:dpi}) stating that the KB $\mo$ must be disjoint with the background knowledge $\mb$ and the role $\mb$ has within a DPI, namely to comprise all formulas that are definitely correct, we postulate that no formula $\tax \in \mo$ must have a probability of zero. 
%Similarly, we want to avoid a probability of 1 for any formula $\tax \in \mo$ since this might imply that  
In a situation when this is not the case, a modified DPI must be used where such formulas have been moved from $\mo$ to $\mb$.\qed 
\end{remark}

%\noindent\textbf{Computation of Diagnosis Probabilities.} 
\paragraph{Computation of Diagnosis Probabilities.}
In the following, we denote by $\overline{\tax}$ ($\overline{\mo}$) the set of logical connectives and quantifiers occurring in a formula $\tax$ (in the KB $\mo$) and by $\widetilde{\tax}$ ($\widetilde{\mo}$) the signature of $\tax$ (of $\mo$). 
\begin{example}
Considering the DL formula $\tax := Pet \equiv Animal \sqcap \exists hasOwner.Person$, we have that $\overline{\tax} = \setof{\equiv,\sqcap,\exists}$ and $\widetilde{\tax}=\setof{Pet, Animal, hasOwner, Person}$.\qed
\end{example}
We now suppose that either a fault probability $p(e) := p(``e \mbox{ is faulty''})$ of each element $e\in\overline{\mo}\cup\widetilde{\mo}$ or the fault probability $p(\tax) := p(``\tax \mbox{ is faulty''})$ of each formula $\tax\in\mo$ is given. For estimation of these probabilities any (combination) of the methods mentioned above might be employed. In case formula probabilities are given, diagnosis probabilities can be directly computed by Formula~\ref{eq:diag_prob_calc}. Otherwise, the following pre-computations must be performed.
%From the quantitative perspective, given fault probabilities $p(e) = p(``e \mbox{ is faulty''})$ for the author to make a mistake when using a syntactical element $e \in \overline{\tax} \cup \widetilde{\tax}$, 

%events ``$e_i$ is faulty'' and `$e_j$ is faulty'' for $p(e)$ for $e\in\overline{\mo}\cup\widetilde{\mo}$, 
The fault probability $p(\tax)$ of $\tax$ can be calculated as the probability that at least one (occurrence of a) syntactical element in $\tax$ is faulty. So, $p(\tax)$ is equal to 1 minus the probability that none of the syntactical elements occurring in $\tax$ is faulty. Hence, under the assumption of mutual independence of syntactical faults concerning elements $e\in\overline{\tax}\cup\widetilde{\tax}$,
\begin{align}
p(\tax) = 1 - \prod_{e \in \overline{\tax} \cup \widetilde{\tax}} (1-p(e))^{n(e)} \label{eq:ax_prob_calc}
\end{align}
where $n(e)$ is the number of occurrences of syntactical element $e$ in $\tax$. %Thereby, $p(\tax)$ is the probability that at least one syntactical element in $\tax$ is faulty. 

If $p(\tax)$ for all $\tax \in \mo$ is known, the fault probability $p(\md)$ of any diagnosis $\md \in \Omega \subseteq \allD_{\langle\mo,\mb,\Tp,\Tn\rangle_\RQ}$ can be determined as the probability that each formula in $\md$ is faulty whereas each formula in $\mo \setminus \md$ is correct, i.e.\ not faulty. Thence, 
\begin{align}
p(\md) = \prod_{\tax_r \in \md} p(\tax_r) \prod_{\tax_s \in \mo\setminus\md} (1-p(\tax_s))  \label{eq:diag_prob_calc}
\end{align}
Recall that probabilities of all atomic events in a well-defined probability space must sum up to $1$. As not every subset of $\mo$ is a diagnosis, this is in general not the case. Therefore, diagnosis probabilities need to be normalized, i.e.\ each diagnosis probability $p(\md)$ must be divided by the sum of all diagnosis probabilities for diagnoses in $\Omega$. That is, the following adjustment is necessary:
\begin{align}
p(\md) \quad\gets\quad \frac{p(\md)}{\sum_{\md_k\in\Omega} p(\md_k)}  \label{eq:diag_prob_norm}
\end{align}
We want to emphasize that the probability measures $p(e)$ of syntactical elements $e$ and $p(\tax)$ of formulas $\tax$ are not required to satisfy any conditions except for $p(e) \in (0,1]$ and $p(\tax) \in (0,1]$ for all $e \in \overline{\tax} \cup \widetilde{\tax}$ and all $\tax \in \mo$ (see Remark~\ref{rem:ax_prob_not_zero} why the intervals $(0,1]$ are open). In particular, no normalization is needed. The reason for this is that ``$e$ is faulty'' and ``$\tax$ is faulty'' are assumptions about a \emph{single} logical connective and a \emph{single} logical formula, respectively. ``$\md$ is the true diagnosis'', to the contrary, is an assumption about \emph{each} formula in the KB $\mo$. So, the probabilities of two different syntactical elements $e_i\neq e_j$ 
%(as well as of each formula $\tax$) 
are computed on the basis of two different probability spaces, namely 
%with $\Omega = \setof{``e \mbox{ is faulty''}, ``e \mbox{ is not faulty''}}$. Clearly, the probability spaces with 
$\Omega_{e_i} = \setof{``e_i \mbox{ is faulty''}, ``e_i \mbox{ is not faulty''}}$ and $\Omega_{e_j} = \setof{``e_j \mbox{ is faulty''}, ``e_j \mbox{ is not faulty''}}$ 
%for $e_i\neq e_j$ 
which clearly do not depend on each other at all. The same argumentation holds for probabilities of formulas.
\paragraph{More Reliable Probabilities through Observations.}
%All these approaches have in common that 
As we argued before, the basic fault information from which diagnosis probabilities are deduced might be rather vague. A usual way of dealing with scenarios of that kind, is to regard the initial probabilities as a first (a-priori) estimation and to gather additional information, e.g.\ by making measurements or observations, and exploit this information to adapt the a-priori estimation in order to obtain a more reliable a-posteriori estimation. The more additional information has been accumulated and incorporated, the more realistic is the resulting updated estimation of probabilities. 

A well-known technique enabling computation of a-posteriori probabilities from a-priori probabilities is Bayes' Theorem. Let $p(\md)$ be the a-priori probability of some $\md\in\Omega \subseteq \allD_{\tuple{\mo,\mb,\Tp,\Tn}_\RQ}$ and $Obs$ be a new observation. Then, the a-posteriori probability $p(\md\,|\,Obs)$ of $\md$, i.e.\ the probability that the true diagnosis $\dt = \md$ taking into account the new information $Obs$, is computed according to Bayes' Theorem as
\begin{align}
p(\md\,|\,Obs) = \frac{p(Obs\,|\,\md)\;\,p(\md)}{p(Obs)} \label{eq:bayes}
\end{align}
where $p(Obs)$ is the (a-priori) probability that observation $Obs$ is made and $p(Obs\,|\,\md)$ is the (a-priori) probability that the observation $Obs$ is made under the assumption that $\md$ is the true diagnosis, i.e.\ $\dt = \md$. That is, the a-priori probability $p(\md)$, i.e.\ the probability that $\dt = \md$ without any additional knowledge, must be multiplied by $p(Obs\,|\,\md) / p(Obs)$ which is often referred to as the \emph{support $Obs$ provides for $\md$}. If the support is greater than 1, then the a-posteriori probability of $\md$ is greater than its a-priori probability, otherwise the a-posteriori probability gets smaller after incorporating the new information $Obs$. Note that Bayes' Theorem is only applicable to KB debugging if a suitable class of observations can be defined such that $p(Obs)$ and $p(Obs\,|\,\md)$ can be computed for observations $Obs$ of this class. As we shall see in Chapter~\ref{chap:UserInteraction}, the assignment of test cases to either $\Tp$ or $\Tn$ is one such class of observations. For instance, $t_i \in \Tp$ and $t_j \in \Tn$ for sets of formulas $t_i,t_j$ over $\mathcal{L}$ are two such observations.  
%%%%%%%%%%%%%%% Only commented out for the journal paper, not for the thesis!!! %%%%%%%%%%%%%%%%%%%%%%%
%Section~\ref{} will show that careful selection of fault probabilities is absolutely crucial in ontology debugging. How this problem can be handled will be explained in detail in Section~\ref{}.
%%%%%%%%%%%%%%% Only commented out for the journal paper, not for the thesis!!! %%%%%%%%%%%%%%%%%%%%%%%

\subsection{Using Probabilities for Diagnosis Computation}
\label{sec:probs_diag_comp}

If available, formula fault probabilities can be exploited during construction of the pHS-tree (Algorithm~\ref{algo:hs}, Chapter~\ref{chap:DiagnosisComputation}) in that most probable instead of minimum cardinality diagnoses are calculated first. To achieve that, breadth-first construction of the tree must be replaced by uniform-cost order of node expansion by means of the function $p()$ that assigns a fault probability to each formula $\tax\in\mo$. Thereby, the ``probability'' $p(\mathsf{nd})$ of a node 
%(path) 
$\mathsf{nd}= \setof{\tax_s,\dots, \tax_t}$ in Algorithm~\ref{algo:hs} is defined through $p(\tax), \tax\in\mo$ as 
\begin{align}\label{eq:path_prob_calc}
p(\mathsf{nd}) = \prod_{\tax_i\in\mathsf{nd}} p(\tax_i) \prod_{\tax_j \in \mo\setminus\mathsf{nd}} (1-p(\tax_j))
\end{align}
%. Thereby, $p(\mathsf{nd})$ is computed 
Notice that this formula extends the definition of Formula~\ref{eq:diag_prob_calc} to arbitrary subsets of $\mo$, not only diagnoses. Thus, Formula~\ref{eq:diag_prob_calc} is a special case of Formula~\ref{eq:path_prob_calc}.
 
First, note that we put ``probability'' of a node in quotation marks as, to be concise, each node (path) which is not yet a diagnosis, i.e.\ needs to be further expanded to become one, has probability zero (of being the true diagnosis $\dt$). For, a probability space is defined on a set of diagnoses and not on a set of arbitrary subsets $\mathsf{nd}$ of the KB. However, we misuse the diagnosis probability space in this case to determine the probability of ``pseudo-diagnoses'' in order to impose an order on the queue of open nodes in the tree. This will guarantee the finding of the most probable diagnoses first, as we shall see below (Proposition~\ref{prop:hs_prob_correct}).

Second, note that no normalization, i.e.\ application of Formula~(\ref{eq:diag_prob_norm}), is necessary within the scope of the non-interactive Algorithm~\ref{algo:hs} since the aim here is only the expansion of nodes $\mathsf{nd}$ in the order of $p(\mathsf{nd})$ and the return of the most probable identified diagnoses at a certain point in time. For this, the comparison of the probability of one node $\mathsf{nd}$ with the probability of another node $\mathsf{nd}'$ suffices.
%(for now, i.e.\ for the scope of the non-interactive Algorithm~\ref{algo:hs}) 
Thus, no other calculations using the properties of a probability space are performed by Algorithm~\ref{algo:hs}. We shall recognize in Chapter~\ref{chap:WorkflowInInteractiveKBDebugging} that this will not hold for the interactive Algorithm~\ref{algo:inter_onto_debug} where Formula~(\ref{eq:diag_prob_norm}) is essential.

So, nodes $\mathsf{nd}$ are inserted into $\Queue$ in a way descending order of node probabilities in $\Queue$ is always maintained. Consequently, nodes with highest fault probability are processed first. This is practical since a user will usually be most interested in seeing those possible faults first that have the highest (estimated) probability to be the actual fault they seek.
%\begin{align}
%w(\mathsf{nd}) = \prod_{\tax_i\in \mathsf{nd}} p(\tax_i) \prod_{\tax_j \in \mo \setminus \mathsf{nd}} (1-p(\tax_j))
%\end{align}

However, one needs to be careful when using probabilities as weights in order not to lose the property of Algorithm~\ref{algo:hs} to compute \emph{minimal} diagnoses only. To this end,
%amounts to the probability $p(\HS(\nd^l))$ that is given by $\prod_{\tax_r \in \HS(\nd^l)}p(\tax_r) \prod_{\tax_s \in \mo\setminus\HS(\nd^l)}(1-p(\tax_s))$. Nodes with highest cost are expanded first. To prevent the so-constructed tree from yielding non-minimal diagnoses, which are undesired since we want to repair an ontology through minimal changes, a technical detail needs to be paid attention to. Namely, 
the formula probabilities $p(\tax)$ for all $\tax\in\mo$ must be adapted as
\begin{align}\label{eq:adapt_ax_prob_to_get_min_diags}
p(\tax) \quad\gets\quad c\,\;\;p(\tax)
\end{align}
where the factor $c$ is an arbitrary positive real number smaller than $0.5$, e.g.\ $c := 0.49 / \max_{\setof{\tax\in\mo}}(p(\tax))$. This transformation effects that all probabilities $p(\tax)$ become smaller than $50\%$. In other words, each formula must be more likely to be correct than faulty which in turn means that a minimal diagnosis is more likely than any of its supersets.
%This property guarantees that minimal diagnoses are discovered first, as the following lemma and proposition show.
\begin{definition}\label{def:p_node()}
Let $p:\mo \rightarrow [0,1]$ be some function that assigns to each $\tax\in\mo$ some $p(\tax) \in [0,1]$. Then, we denote by $p_{nodes}: 2^{\mo} \rightarrow [0,1]$ the function that assigns to each node $\mathsf{nd} \subseteq \mo$ some $p_{nodes}(\mathsf{nd})\in [0,1]$ which is obtained by means of Formula~\ref{eq:path_prob_calc} and $p()$. 
\end{definition}
\begin{lemma}\label{lem:superset_lower_prob}
Let $\mathsf{nd},\mathsf{nd}'\subseteq\mo$ where $\mathsf{nd}\subset \mathsf{nd}'$ and $p: \mo \rightarrow (0,0.5)$ a function which assigns to each $\tax\in\mo$ some probability $p(\tax) \in (0,0.5)$. %Further, let $p(X)$ for a set of formulas $X\subseteq \mo$ be defined by means of $p(\tax), \tax\in\mo$ as per Formula~\ref{eq:path_prob_calc}. 
Then $p_{nodes}(\mathsf{nd}) > p_{nodes}(\mathsf{nd}')$ holds. 
\end{lemma}
\begin{proof}
According to Formula~\ref{eq:path_prob_calc} and Definition~\ref{def:p_node()} we have that 
\begin{align*}
p_{nodes}(\mathsf{nd}) = \prod_{\tax_i\in \mathsf{nd}} p(\tax_i) \prod_{\tax_j \in \mo \setminus \mathsf{nd}} (1-p(\tax_j))
\end{align*}
%$p_{nodes}(\mathsf{nd}) = \prod_{\tax_i\in \mathsf{nd}} p(\tax_i) \prod_{\tax_j \in \mo \setminus \mathsf{nd}} (1-p(\tax_j))$. 
Then the probability $p_{nodes}(\mathsf{nd}')$ can be computed from $p_{nodes}(\mathsf{nd})$ in that, for each formula $\tax$ in $\mathsf{nd}'\setminus \mathsf{nd} \subseteq \mo \setminus \mathsf{nd}$, we multiply $p_{nodes}(\mathsf{nd})$ by a factor $f_\tax := p(\tax) / (1-p(\tax))$ because $\tax$ ``moves'' from $\mo\setminus \mathsf{nd}$ to $\mathsf{nd}$. However, $f_\tax < 1$ holds due to $p(\tax) < 0.5$ and thus $1-p(\tax) > 0.5$.
\end{proof}
This result will be a key to proving the completeness, soundness and correctness of Algorithm~\ref{algo:hs} in the next Section.

The next definition characterizes a (partial) weighted pHS-tree, the type of hitting set tree constructed by Algorithm~\ref{algo:hs} given any function $p(\tax) \in (0,0.5)$ for all $\tax\in\mo$ as input which is not necessarily specified in a way a breadth-first tree construction is forced.
\begin{definition}[Weighted Pruned HS-Tree]\label{def:weighted_pruned_hs_tree}
Let $\langle\mo,\mb,\Tp,\Tn\rangle_\RQ$ be an admissible DPI and let $w:\mo\rightarrow [0,1]$ be a weight function which assigns a weight to each node $\mathsf{n} \subseteq \mo$ with the property that $w(\mathsf{n}_1) > w(\mathsf{n}_2)$ if $\mathsf{n}_1 \subset \mathsf{n}_2$. An edge-labeled and node-labeled tree $T$ is called a \emph{weighted pruned HS-tree (wpHS-tree) w.r.t.\ $\langle\mo,\mb,\Tp,\Tn\rangle_\RQ$ and $w()$} iff $T$ is the result of constructing an HS-tree w.r.t.\ $\langle\mo,\mb,\Tp,\Tn\rangle_\RQ$ with due regard to the following rule  
\begin{enumerate}
\item Label open nodes in the HS-tree in order of descending $w()$,
\end{enumerate}
and the rules 2 to 6 as per Definition~\ref{def:pruned_hs_tree}. 

$T$ is called a \emph{partial weighted pruned HS-tree w.r.t.\ $\langle\mo,\mb,\Tp,\Tn\rangle_\RQ$ and $w()$} iff $T$ is a weighted pruned HS-tree w.r.t.\ $\langle\mo,\mb,\Tp,\Tn\rangle_\RQ$ and $w()$ where not all nodes in $T$ have been labeled yet and non-labeled nodes have no successors. 
\end{definition}
Then, we have the following relationship between a (partial) pHS-tree
and a (partial) wpHS-tree. An explanation why this holds will be given in Section~\ref{sec:UsingProbabilitiesToComputeMinimumCardinalityDiagnoses}.
\begin{proposition}
A (partial) pHS-tree w.r.t.\ $\langle\mo,\mb,\Tp,\Tn\rangle_\RQ$ is a (partial) wpHS-tree w.r.t.\ $\langle\mo,\mb,\Tp,\Tn\rangle_\RQ$ and $w()$ where $w()$ is a weight function which, additionally to the property postulated in Definition~\ref{def:weighted_pruned_hs_tree}, satisfies $w(\mathsf{n}_1) = w(\mathsf{n}_2)$ if $|\mathsf{n}_1| = |\mathsf{n}_2|$.

In general, a (partial) wpHS-tree w.r.t.\ $\langle\mo,\mb,\Tp,\Tn\rangle_\RQ$ and $w()$ is not a (partial) pHS-tree w.r.t.\ $\langle\mo,\mb,\Tp,\Tn\rangle_\RQ$.
\end{proposition}

\begin{lemma}\label{lem:algo_hs_produces_weighted_pruned_hs_tree} Algorithm~\ref{algo:hs} 
%given as input an admissible DPI $\langle\mo,\mb,\Tp,\Tn\rangle_\RQ$ and a function $p()$ with $p(\tax)\in(0,0.5)$ for all $\tax\in\mo$ 
is a procedure for producing a wpHS-tree $T$ w.r.t.\ $\langle\mo,\mb,\Tp,\Tn\rangle_\RQ$ and $p_{nodes}()$. 
%where $p_{nodes}()$ 
%is the function obtained by means of Formula~\ref{eq:path_prob_calc} and $p(\tax),\tax\in\mo$.
%
% if Algorithm~\ref{algo:hs} returns due to $\Queue = \emptyset$.
\end{lemma}
\begin{proof}
First, the property $p_{nodes}(\mathsf{n}_1) > p_{nodes}(\mathsf{n}_2)$ if $\mathsf{n}_1 \subset \mathsf{n}_2$ postulated by Definition~\ref{def:weighted_pruned_hs_tree} holds by Lemma~\ref{lem:superset_lower_prob} and the fact that the function $p$ given as input to Algorithm~\ref{algo:hs} satisfies $p(\tax)\in(0,0.5)$ for all $\tax\in\mo$. Moreover, the DPI $\langle\mo,\mb,\Tp,\Tn\rangle_\RQ$ provided as an input to Algorithm~\ref{algo:hs} is admissible, as postulated by Definition~\ref{def:weighted_pruned_hs_tree}.

The compliance with rule 1 of Definition~\ref{def:hs_tree} as well as with rules 2 to 6 of Definition~\ref{def:pruned_hs_tree} is a simple consequence of Lemma~\ref{lem:algo_hs_produces_pruned_hs_tree}. In the following we prove that rule 2 of Definition~\ref{def:hs_tree} and rule 1 of Definition~\ref{def:weighted_pruned_hs_tree} are satisfied.
\begin{itemize}
	\item Definition~\ref{def:hs_tree}, rule 2: Suppose a node $\mathsf{nd}$ is labeled by $valid$. Then it is added to $\mD_{calc}$ in line~\ref{algoline:hs:update_Dcalc}. Since $\mathsf{nd}$ can only get a label different from $closed$ if it is the only exemplar of this node in $\Queue$ due to the duplicate criterion (lines~\ref{algoline:hs:duplicate_crit_start}-\ref{algoline:hs:duplicate_crit_end}), it must be the case that $\mathsf{nd}\notin\Queue$ (line~\ref{algoline:hs:remove_from_queue}) after $\mathsf{nd}$ has been labeled by $valid$. Only nodes that get labeled by a conflict set can have successor nodes added to $\Queue$ in line~\ref{algoline:hs:generate_nodes}. Only nodes in $\Queue$ can get a label (cf.\ lines~\ref{algoline:hs:getfirst} and \ref{algoline:hs:label}). For $\mathsf{nd}$ to be added to $\Queue$ at some later point in time there must be a proper subset of $\mathsf{nd}$ that is still in $\Queue$ as each node newly added to $\Queue$ is a proper superset of some node in $\Queue$ (cf.\ line~\ref{algoline:hs:generate_nodes} which is the only position in the algorithm where nodes are added to $\Queue$). This is impossible since $\Queue$ is ordered descending by $p_{nodes}()$. %and $p()$ meets Lemma~\ref{lem:superset_lower_prob}. 
Hence, each proper subset of $\mathsf{nd}$ must have been ranked before $\mathsf{nd}$ in $\Queue$ and thus must have already been labeled because
$\mathsf{nd}$ is already labeled by assumption.
Hence, if $\mathsf{nd}$ is labeled by $valid$, then it has no successors.
	\item Definition~\ref{def:weighted_pruned_hs_tree}, rule 1: 
%First, $p()$ with $p(\tax) < 0.5$ for all $\tax \in \mo$ and $p(\mathsf{node})$ being defined using $p(\tax),\tax\in\mo$ as per Formula~\ref{eq:path_prob_calc} implies by Lemma~\ref{lem:superset_lower_prob} that $p(\mathsf{n}_1) > p(\mathsf{n}_2)$ if $\mathsf{n}_1 \subset \mathsf{n}_2$. 
That nodes are processed and labeled in order of descending $p_{nodes}()$ follows from the fact that new nodes are inserted into $\Queue$ only in a way that the order of $\Queue$ by descending $p_{nodes}()$ is maintained (\textsc{insertSorted} in line~\ref{algoline:hs:generate_nodes}) and by the fact that always the first element of $\Queue$ is selected to be labeled next (\textsc{getFirst} in line~\ref{algoline:hs:getfirst}).
\end{itemize}
This completes the proof.
\end{proof}
Let the relevant data of a wpHS-tree be defined as for a pHS-tree (cf.\ Remark~\ref{rem:algo_hs:internal_representation}). By the correctness of Lemma~\ref{lem:algo_hs_produces_weighted_pruned_hs_tree}, we have:
\begin{corollary}\label{cor:algo_hs_returns_relevant_data_of_weighted_(partial)_pruned_hs_tree}
Algorithm~\ref{algo:hs} 
%given as input an admissible DPI $\langle\mo,\mb,\Tp,\Tn\rangle_\RQ$ and a function $p()$ with $p(\tax)\in(0,0.5)$ for all $\tax\in\mo$ 
stores by $\tuple{\mD_{calc},\Queue,\mC_{calc}}$ the relevant data of 
\begin{itemize}
\item a wpHS-tree w.r.t.\ $\langle\mo,\mb,\Tp,\Tn\rangle_\RQ$ and $p_{nodes}()$ if Algorithm~\ref{algo:hs} stops due to $\Queue = []$, and
\item a partial wpHS-tree w.r.t.\ $\langle\mo,\mb,\Tp,\Tn\rangle_\RQ$ and $p_{nodes}()$ otherwise.
\end{itemize}
%where $p_{nodes}()$ is the function obtained by means of Formula~\ref{eq:path_prob_calc} and $p(\tax),\tax\in\mo$.
\end{corollary}

\subsection{Correctness of Weighted Diagnosis Computation}
\label{sec:CorrectnessOfAlgorithmHs}
First, we show the completeness of Algorithm~\ref{algo:hs} regarding minimal diagnoses, i.e.\ that it computes \emph{all} minimal diagnoses w.r.t.\ the DPI it is given as input.
\begin{lemma}\label{lem:algo_hs_only_diags_in_Dcalc}
Only diagnoses w.r.t.\ $\tuple{\mo,\mb,\Tp,\Tn}_\RQ$ can be added to $\mD_{calc}$ by Algorithm~\ref{algo:hs}.
% with $\tuple{\mo,\mb,\Tp,\Tn}_\RQ$ given as an input DPI.
\end{lemma}
\begin{proof}
A node $\mathsf{nd}$ can be added to $\mD_{calc}$ only in line~\ref{algoline:hs:update_Dcalc}. To reach this line, \textsc{label} must have returned $valid$ for $\mathsf{nd}$. For this to hold, $\scQX(\tuple{\mo\setminus\mathsf{nd},\mb,\Tp,\Tn}_\RQ)$ must have returned 'no conflict' which implies that $\mathsf{nd}$ is a diagnosis w.r.t.\ $\tuple{\mo,\mb,\Tp,\Tn}_\RQ$ by Propositions~\ref{prop:qx_correctness} and \ref{prop:validonto_diag}.
\end{proof}
\begin{lemma}\label{lem:algo_hs_path+cs}
Let $T$ denote a (partial) wpHS-tree produced by Algorithm~\ref{algo:hs}. Further, let $\Queue$ be the queue of open nodes in $T$ maintained by Algorithm~\ref{algo:hs} and let $\mathsf{nd}$ be some node which occurs only once in $\Queue$ and which is a proper subset of some minimal diagnosis w.r.t.\ $\tuple{\mo,\mb,\Tp,\Tn}_\RQ$. Then:
\begin{enumerate}[(1)]
%the pruned HS-tree produced by Algorithm~\ref{algo:hs} when it stops due to $\Queue = \emptyset$. 
\item The nodes $\emptyset=\mathsf{nd}_1,\dots,\mathsf{nd}_k$ along any path from the root node $\emptyset$ to $\mathsf{nd}_k$ in $T$ satisfy $\mathsf{nd}_i \subset \mathsf{nd}_{i+1}$ and $|\mathsf{nd}_i|+1 = |\mathsf{nd}_{i+1}|$ and $\mathsf{nd}_i \subseteq \mo$ for $1 \leq i \leq k$. 
\item  If the \textsc{label} function is called for $\mathsf{nd}$, then it yields some minimal conflict set $\mc$ w.r.t.\ $\tuple{\mo,\mb,\Tp,\Tn}_\RQ$ with $\mathsf{nd} \cap \mc = \emptyset$.
%\item Each node $\mathsf{nd}$ in $T$ which has a successor node is labeled by some minimal conflict set $\mc$ w.r.t.\ $\tuple{\mo,\mb,\Tp\cup\Tp',\Tn\cup\Tn'}_\RQ$ with $\mathsf{nd} \cap \mc = \emptyset$.
%The \textsc{label} function called for (the last exemplar of) some node $\mathsf{nd}$ in $\Queue$ which is a proper subset of some minimal diagnosis w.r.t.\ $\tuple{\mo,\mb,\Tp,\Tn}_\RQ$ labels $\mathsf{nd}$ by some minimal conflict set $\mc$ w.r.t.\ $\tuple{\mo,\mb,\Tp,\Tn}_\RQ$ with $\mathsf{nd} \cap \mc = \emptyset$.
\end{enumerate}
\end{lemma}
\begin{proof}
%Given the path $\emptyset=\mathsf{nd}_1,\dots,\mathsf{nd}_k$, each node $\mathsf{nd}_{i+1}$ is a successor node of $\mathsf{nd}_i$. 
(1): In the representation used by Algorithm~\ref{algo:hs}, a node $\mathsf{nd}$ in the (partial) wpHS-tree $T$ produced by Algorithm~\ref{algo:hs} is defined as the set of all edge labels on the path from the root node 
%$\emptyset$ 
to $\mathsf{nd}$ (see Remark~\ref{rem:algo_hs:internal_representation}) and the successor of a node is defined as a node added to $\Queue$ after $\mathsf{nd}$ has been labeled by a minimal conflict set.% (see page~\pageref{etc:def_successor_of_node}). 
%$\mathsf{nd} \cup \setof{\tax}$ for some $\tax$.
%A node $\mathsf{nd}$ in Algorithm~\ref{algo:hs} is defined as the set of all edge labels on the path from the root node $\emptyset$ to $\mathsf{nd}$. The label of the edge going from some node $\mathsf{nd}$ to a successor node consists of exactly one sentence $\tax\in\mc$ where $\mc$ is the minimal conflict set that labels $\mathsf{nd}$. 
%Reflecting this, 
After the \textsc{label} function for node $\mathsf{nd}$ has returned some minimal conflict set $L$ as a label for $\mathsf{nd}$, Algorithm~\ref{algo:hs} goes to line~\ref{algoline:hs:generate_nodes} since $L\neq closed$ and $L\neq valid$ and adds an element $\mathsf{nd} \cup \setof{e}$ to $\Queue$ for each $e \in L$. Therefore, it holds that $|\mathsf{nd} \cup \setof{e}| = |\mathsf{nd}|+1$ for each successor of $\mathsf{nd}$. Hence,  $\mathsf{nd}_i \subset \mathsf{nd}_{i+1}$ and $|\mathsf{nd}_i|+1 = |\mathsf{nd}_{i+1}|$ holds for any path of nodes $\emptyset=\mathsf{nd}_1,\dots,\mathsf{nd}_k$ in $T$ starting from the root node.

The argumentation why each node must be a subset of $\mo$ is as follows:
Suppose $\mathsf{node}\cup\setof{e}$ is added to $\Queue$ in line~\ref{algoline:hs:generate_nodes} which is the only place in Algorithm~\ref{algo:hs} where nodes are added to $\Queue$. So, \textsc{label} must have returned neither $valid$ nor $closed$ for $\mathsf{node}$. Hence, $\mathsf{node}$ cannot be a diagnosis w.r.t.\ $\langle\mo,\mb,\Tp,\Tn\rangle_\RQ$ as otherwise \textsc{label} with argument $\mathsf{node}$ must have returned $valid$ in line~\ref{algoline:hs:return_valid}. Due to the fact that $\mathsf{node} = \mo$ is definitely a diagnosis w.r.t.\ $\langle\mo,\mb,\Tp,\Tn\rangle_\RQ$ as it must hit all minimal conflict sets w.r.t.\ $\langle\mo,\mb,\Tp,\Tn\rangle_\RQ$ which must all be subsets of $\mo$ (Definition~\ref{def:cs}), $\mathsf{node} \subset \mo$ must hold. 

(2): Suppose the \textsc{label} function is called for a node $\mathsf{nd}\in\Queue$ where $\mathsf{nd} \subset \md$ for some minimal diagnosis $\md$. 

First, there cannot be any $\mathsf{nd}' \in \mD_{calc}$ with $\mathsf{nd}' \subseteq \mathsf{nd}$ since $\mD_{calc}$ includes only diagnoses w.r.t.\ $\langle\mo,\mb,\Tp,\Tn\rangle_\RQ$ and $\mathsf{nd} \subset \md$ wherefore there would be a diagnosis $\mathsf{nd}' \subset \md$, contradiction. Due to the fact that $\mathsf{nd}$ is present only once in $\Queue$, there cannot be some $\mathsf{nd}' = \mathsf{nd}$ in $\Queue$. Thus, $closed$ cannot be returned for $\mathsf{nd}$ by \textsc{label}.

By the facts that a diagnosis must hit all minimal conflict sets (Proposition~\ref{prop:mindiag_mincs}) and that $\mathsf{nd}$ is a proper subset of a diagnosis, either the criterion checked in line~\ref{algoline:hs:test_cs_not_hit} must be true or $\scQX(\tuple{\mo\setminus\mathsf{nd},\mb,\Tp,\Tn}_\RQ)$ must return a minimal conflict set $L$, i.e.\ $L \neq$ 'no conflict'. In both cases, a minimal conflict set is returned by \textsc{label}.

There are no other labels that can be returned by \textsc{label}.
\end{proof}

\begin{lemma}\label{lem:each_min_diag_occurs_in_queue}
Each minimal diagnosis w.r.t.\ 
%the DPI 
$\langle\mo,\mb,\Tp,\Tn\rangle_\RQ$ 
%given as input to Algorithm~\ref{algo:hs} 
occurs as a node in $\Queue$ during the execution of Algorithm~\ref{algo:hs}, if the execution stops due to $\Queue = []$.
\end{lemma}
\begin{proof}
For Algorithm~\ref{algo:hs} it holds that 
\begin{enumerate}[(i)]
%\item the nodes $\emptyset=\mathsf{nd}_1,\dots,\mathsf{nd}_k\subseteq\mo$ along any path from the root node $\emptyset$ in the produced HS-Tree %produced by \textsc{dynamicHS} 
%satisfy $\mathsf{nd}_i \subset \mathsf{nd}_{i+1}$ and $|\mathsf{nd}_i|+1 = |\mathsf{nd}_{i+1}|$ by Lemma~\ref{lem:algo_hs_path+cs} and 
\item if $\mathsf{nd}$ is the last exemplar of some node in $\Queue$ which is a proper subset of some minimal diagnosis w.r.t.\ $\tuple{\mo,\mb,\Tp,\Tn}_\RQ$ and the \textsc{label} function is called for $\mathsf{nd}$, then it yields some minimal conflict set $\mc$ w.r.t.\ $\tuple{\mo,\mb,\Tp,\Tn}_\RQ$ with $\mathsf{nd} \cap \mc = \emptyset$ by Lemma~\ref{lem:algo_hs_path+cs} and 
\item each node $\mathsf{nd}$ that has been labeled by some minimal conflict set $\mc$ is deleted from $\Queue$ (line~\ref{algoline:dyn:delete_from_queue}) whereupon one successor node $\mathsf{nd}_{\tax} = \mathsf{nd} \cup \setof{\tax}$ for each element $\tax\in\mc$ is added to $\Queue$ (\textsc{insertSorted} in line~\ref{algoline:dyn:generate_nodes}) and
%the root node corresponds to the node $\emptyset$ and 
%\item all successor nodes of a node $\mathsf{nd}$ are added to $\Queue$ if $\mathsf{nd}$ was labeled by a minimal conflict set $\tuple{\mo,\mb,\Tp\cup\Tp',\Tn\cup\Tn'}_\RQ$ (line~\ref{algoline:dyn:generate_nodes}) and
\item each minimal diagnosis w.r.t.\ $\tuple{\mo,\mb,\Tp,\Tn}_\RQ$ is a superset of $\emptyset$ and a subset of $\mo$ (Definition~\ref{def:diagnosis}) which includes one element of each minimal conflict set w.r.t.\ $\tuple{\mo,\mb,\Tp,\Tn}_\RQ$ and includes only elements of minimal conflict sets (Proposition~\ref{prop:mindiag_mincs}).
\end{enumerate}
%Each minimal diagnosis w.r.t.\ $\tuple{\mo,\mb,\Tp,\Tn}_\RQ$ is a superset of $\emptyset$ and a subset of $\mo$ (Definition~\ref{def:diagnosis}) which includes one element of each minimal conflict set w.r.t.\ $\tuple{\mo,\mb,\Tp,\Tn}_\RQ$ and includes only elements of minimal conflict sets (Proposition~\ref{prop:mindiag_mincs}).

Let $\md$ be some minimal diagnosis w.r.t.\ $\langle\mo,\mb,\Tp,\Tn\rangle_\RQ$. Then, there is a path of nodes from the root node $\emptyset$ to $\md$ in the pHS-tree produced by Algorithm~\ref{algo:hs}, if the execution stops due to $\Queue = []$. 

This holds by the following argumentation: If $\md = \emptyset$, then the path is $\tuple{\emptyset}$. Now, suppose $\md \supset \emptyset$. Since $\md$ is a minimal diagnosis wherefore no other diagnosis can be equal to $\emptyset$, the root node $\mathsf{n}_0 := \emptyset$ of the constructed tree must be labeled by some minimal conflict set $\mc_1$.  
%$\mathsf{n}_0=\emptyset$. Otherwise, since $\md \supseteq \emptyset$, we conclude that $\md \supset \emptyset$. Since $\md$ is a minimal diagnosis wherefore no proper subset of it can be a diagnosis, there must be a minimal conflict set $\mc_1$ that is not hit by $\mathsf{n}_0$. 
%let $\mc_1$ be the minimal conflict set that labels the root node $\mathsf{n}_1$. 
Then, by (iii), there must be some $\tax_1\in\mc_1$ that is an element of $\md$. So, we define $\mathsf{n}_1 := \setof{\tax_1}$. If $\mathsf{n}_1 = \md$, then the path is $\tuple{\emptyset,\mathsf{n}_1}$. Otherwise, due to $\md \supset \mathsf{n}_1$ and (i), 
%is a minimal diagnosis wherefore no diagnosis can be equal to $\mathsf{n}_1$, 
node $\mathsf{n}_1$ in the pHS-tree must be labeled by some minimal conflict set $\mc_2$. Then, by (iii), there must be some $\tax_2\in\mc_2$ that is an element of $\md$. So, we define $\mathsf{n}_2 := \mathsf{n}_1 \cup \setof{\tax_2}$. If $\mathsf{n}_2 = \md$, then the path is $\tuple{\emptyset,\mathsf{n}_1,\mathsf{n}_2}$. Otherwise, due to $\md \supset \mathsf{n}_2$ and (i), 
%is a minimal diagnosis wherefore no diagnosis can be equal to $\mathsf{n}_1$, 
node $\mathsf{n}_2$ in the pHS-tree must be labeled by some minimal conflict set $\mc_3$. This reasoning can be continued until $\mathsf{n}_k = \md$ for some $k$. By (iii), $\md\subseteq\mo$ holds wherefore such $k$ must exist. %is an upper bound for the nodes.

Algorithm~\ref{algo:hs} cannot stop executing before $\mathsf{n}_k$ has been in $\Queue$ since each node $\mathsf{n}_i$ labeled by a minimal conflict set $\mc_{i+1}$ involves the addition of $|\mc_{i+1}|$ successor nodes to $\Queue$ by (ii). In particular, the successor node $\mathsf{n}_i \cup \setof{\tax_{i+1}}$ must be added to $\Queue$. As the execution stops due to $\Queue = []$, all nodes $\mathsf{n}_i$ for $i\leq k$ must be labeled before termination. Thus, $\md$ must be in $\Queue$ sometime. 
%
%$\emptyset=\mathsf{n}_1,\dots,\mathsf{n}_k,\md$ 
%
%$\md$ must occur as a node in $\Queue$ somewhen if we assume an entire execution of \textsc{dynamicHS} until $\Queue$ is empty. So, 
%
%there must be one path of nodes starting from the root node $\emptyset$ to $\md$ in the \emph{complete} HS-tree constructed by \textsc{dynamicHS}. In other words, $\md$ must occur as a node in $\Queue$ sometime if we assume an entire execution of \textsc{dynamicHS} until $\Queue$ is empty. 
%
%Moreover each 
%%path starting from the root node to any 
%node $\mathsf{nd}$ in the HS-tree of \textsc{dynamicHS} with $\mathsf{nd} \subset \mathsf{node}$ for some $\mathsf{node} \in \Queue$ must already
%
%As \textsc{dynamicHS} adds all successor nodes of some node $\mathsf{nd}$ to $\Queue$ after $\mathsf{nd}$ was labeled (line~\ref{algoline:dyn:generate_nodes}), 
\end{proof}
%The following corollary is a trivial consequence of Lemma~\ref{lem:each_min_diag_occurs_in_queue}.
%\begin{corollary}
%Let $T$ denote a pruned HS-tree produced by Algorithm~\ref{algo:hs}. Then each minimal diagnosis occurs as a path in $T$.
%\end{corollary}
\begin{proposition}[Completeness of Algorithm~\ref{algo:hs}]\label{prop:hs-tree_completeness}
If Algorithm~\ref{algo:hs} terminates due to $\Queue = []$,
%, i.e.\ the pruned HS-tree has been constructed, 
then the algorithm returns a set $\mD$ including all minimal diagnoses w.r.t.\ 
%the DPI 
$\langle\mo,\mb,\Tp,\Tn\rangle_\RQ$. 
%given as input to the algorithm.
\end{proposition}
\begin{proof}
Assume some minimal diagnosis $\md$ w.r.t.\ $\langle\mo,\mb,\Tp,\Tn\rangle_\RQ$ where $\md\notin\mD$ after Algorithm~\ref{algo:hs} has returned due to $\Queue = []$. First, each minimal diagnosis will occur in $\Queue$ throughout the execution of Algorithm~\ref{algo:hs} because it executes until $\Queue = []$ wherefore Lemma~\ref{lem:each_min_diag_occurs_in_queue} applies. 
Any node $\mathsf{nd}$ in $\Queue$ can only be deleted from $\Queue$ if \textsc{label} is called with the argument node $\mathsf{nd}$ (lines~\ref{algoline:hs:remove_from_queue} and \ref{algoline:hs:label}). There is no other point in Algorithm~\ref{algo:hs} where elements are removed from $\Queue$.
Since at the end $\Queue = []$, each minimal diagnosis, in particular $\md$, must be labeled.

Suppose $\md$ is the last exemplar of possibly multiple duplicates of it in $\Queue$. Then, the \textsc{label} function cannot return $closed$ for $\md$. 
This holds, on the one hand, because the duplicate criterion (lines~\ref{algoline:hs:duplicate_crit_start}-\ref{algoline:hs:duplicate_crit_end}) only removes possible duplicate nodes from $\Queue$, but never the last exemplar of a node in $\Queue$. On the other hand, $\md$ can never be closed due to the non-minimality criterion (lines~\ref{algoline:hs:non_min_crit_start}-\ref{algoline:hs:non_min_crit_end}) as $\mD_{calc}$ can only include diagnoses w.r.t.\ $\langle\mo,\mb,\Tp,\Tn\rangle_\RQ$ by Proposition~\ref{lem:algo_hs_only_diags_in_Dcalc}. Thus, due to the minimality of $\md$, $\mD_{calc}$ cannot comprise any diagnosis $\md'$ with $\md' \subseteq \md$, except for some $\md'$ which is equal to $\md$. This would however be a contradiction to the assumption that $\md\notin\mD$.

The reuse criterion (lines~\ref{algoline:hs:reuse_crit_start}-\ref{algoline:hs:reuse_crit_end}) cannot apply for $\md$ either since a minimal diagnosis is a hitting set of all minimal conflict sets (Proposition~\ref{prop:mindiag_mincs}) wherefore there cannot be a minimal conflict set in $\mC_{calc}$ which has an empty intersection with $\md$. So, the algorithm will come to line~\ref{algoline:hs:qx} where $\scQX(\tuple{\mo\setminus\md,\mb,\Tp,\Tn}_\RQ)$ will return 'no conflict' (Propositions~\ref{prop:qx_correctness} and \ref{prop:validonto_diag}). Therefore, $\md$ will be labeled by $valid$ and will be added to $\mD_{calc}$ in line~\ref{algoline:hs:update_Dcalc}.
\end{proof}
Next, we show the soundness of Algorithm~\ref{algo:hs} w.r.t.\ minimal diagnoses, i.e.\ that it computes \emph{only} minimal diagnoses w.r.t.\ the DPI it is given as input.
\begin{proposition}[Soundness of Algorithm~\ref{algo:hs}]\label{prop:hs-tree_soundness}
%Let $p(\tax) < 0.5$ for all $\tax \in \mo$. Then, if 
If 
%Algorithm~\ref{algo:hs} 
%using $p()$ 
%adds 
an element $\md$ is added to the set $\mD_{calc}$ during the execution of Algorithm~\ref{algo:hs}, $\md$ is a minimal diagnosis w.r.t.\ 
%the DPI 
$\langle\mo,\mb,\Tp,\Tn\rangle_\RQ$. 
%given as input to the algorithm.
\end{proposition}
\begin{proof}
Assume that some element $\mathsf{nd}$ is added to $\mD_{calc}$ which is not a diagnosis w.r.t.\ $\langle\mo,\mb,\Tp,\Tn\rangle_\RQ$. This immediately yields a contradiction due to Lemma~\ref{lem:algo_hs_only_diags_in_Dcalc}.
%
%since $\mD_{calc}$ is extended only in line~\ref{algoline:hs:update_Dcalc} and to get to this line the function \textsc{label} must have returned $valid$ as one part of its output. This can only be the case if the call to $\scQX$ in line~\ref{algoline:hs:qx} has returned 'no conflict' for the DPI $\tuple{\mo\setminus\mathsf{node},\mb,\Tp,\Tn}_\RQ$ which is equivalent to the fact that $\mathsf{node}$ is a diagnosis w.r.t.\ $\tuple{\mo,\mb,\Tp,\Tn}_\RQ$	by Proposition~\ref{prop:qx_correctness} and Corollary~\ref{cor:notions_equiv}.

Assume now that some element $\mathsf{nd}$ is added to $\mD_{calc}$ which is a diagnosis w.r.t.\ $\langle\mo,\mb,\Tp,\Tn\rangle_\RQ$, but not a minimal one. Now, since $\mathsf{nd}$ is a non-minimal diagnosis, there is some $\md \subset \mathsf{nd}$ which is a minimal diagnosis w.r.t.\ $\langle\mo,\mb,\Tp,\Tn\rangle_\RQ$.

Then, there are three cases to distinguish: (a)~$\md$ is in $\Queue$ and (b)~$\md$ is in $\mD_{calc}$ and (c)~$\md$ is neither in $\Queue$ nor in $\mD_{calc}$, i.e.\ the node $\md$ has not yet been generated.
%, i.e.\ has not yet been added to $\Queue$. 

Note that these are all possible cases as $\md$ is a \emph{minimal} diagnosis by assumption. So, $\md$ cannot have been ruled out, i.e.\ labeled by $closed$, by the non-minimality criterion (lines~\ref{algoline:hs:non_min_crit_start}-\ref{algoline:hs:non_min_crit_end}) before since only diagnoses can be added to $\mD_{calc}$ as argued in the first paragraph of this proof and there cannot be a diagnosis $\md'\in\mD_{calc}$ such that $\md' \subset \md$. The case $\md' = \md$ is already considered by case (b). The duplicate criterion (lines~\ref{algoline:hs:duplicate_crit_start}-\ref{algoline:hs:duplicate_crit_end}) does not need to be taken into account since it deletes duplicate nodes only. 

(a): To be added to $\mD_{calc}$, $\mathsf{nd}$ must have been the first element of the queue $\Queue$ by \textsc{getFirst} in line~\ref{algoline:hs:getfirst}. Since $\md \in \Queue$ by assumption and since $\Queue$ is sorted in descending order of node probability (\textsc{insertSorted} in line~\ref{algoline:hs:generate_nodes}), we conclude that $p_{nodes}(\md) \leq p_{nodes}(\mathsf{nd})$. However, as $p_{nodes}(X)$ for a node $X\subseteq \mo$ is defined by means of $p(\tax)$ where $p(\tax) \in (0,0.5)$ for all $\tax\in\mo$ as per Formula~\ref{eq:path_prob_calc} (Definition~\ref{def:p_node()}), Lemma~\ref{lem:superset_lower_prob} applies and establishes the truth of $p_{nodes}(S_1) > p_{nodes}(S_2)$ if $S_1 \subset S_2$ for $S_1,S_2 \subseteq \mo$. By $\md \subset \mathsf{nd}$, this implies $p_{nodes}(\md) > p_{nodes}(\mathsf{nd})$, contradiction.
%
%This implies that no subset $\mathsf{node}'$ of $\mathsf{node}$ can be in $\Queue$. The reasons for this are that $\Queue$ is sorted in descending order of node probability (\textsc{insertSorted} in line~\ref{algoline:hs:generate_nodes}) and that $p(\mathsf{node}') > p(\mathsf{node})$. Therefore, case (a) leads to a contradiction.

(b): Assuming case (b), we can derive a contradiction as follows. By the fact that $\mathsf{nd}$ is added to $\mD_{calc}$, it must hold that the \textsc{label} procedure called for $\mathsf{nd}$ in line~\ref{algoline:hs:label} returned $valid$ as part of its output in line~\ref{algoline:hs:return_valid}. However, as $\md \subset \mathsf{nd}$ is already an element of $\mD_{calc}$ by assumption, the \textsc{label} procedure must have already returned in line~\ref{algoline:hs:non_min_crit_end} wherefore it cannot have reached line~\ref{algoline:hs:return_valid}, contradiction.

(c): Suppose that $\md$ has not yet been generated as a node in $\Queue$. 
%Recall that a path from root in the HS-Tree produced by Algorithm~\ref{algo:hs} is a set of nodes $\emptyset=\mathsf{nd_1},\dots,\mathsf{nd_k}$ where $\mathsf{nd}_i \subset \mathsf{nd}_{i+1}$ and $|\mathsf{nd}_i|+1 = |\mathsf{nd}_{i+1}|$. 
By Lemma~\ref{lem:algo_hs_path+cs}, the nodes $\emptyset=\mathsf{nd_1},\dots,\mathsf{nd_k}$ along a path from the root node in the pHS-Tree produced by Algorithm~\ref{algo:hs} 
satisfy $\mathsf{nd}_i \subset \mathsf{nd}_{i+1}$ and $|\mathsf{nd}_i|+1 = |\mathsf{nd}_{i+1}|$.
So, by Lemma~\ref{lem:superset_lower_prob}, the node probabilities along any path from the root node are strictly monotonically decreasing.  Since $p_{nodes}(\md) > p_{nodes}(\mathsf{nd})$ holds by the same argumentation as in (a), we have that all nodes on the path from the root node to $\md$ have a higher probability than $\mathsf{nd}$. As $\Queue$ is sorted in descending order of node probability and in each iteration the first element in $\Queue$ is processed as explained in (a), we infer that $\md$ must have already been generated at the time $\mathsf{nd}$ is processed, contradiction. 
%
%Let $p(X)$ for a set of formulas $X\subseteq \mo$ be defined by means of $p(\tax), \tax\in\mo$ as per Formula~\ref{eq:path_prob_calc}. Then, by Lemma~\ref{lem:superset_lower_prob}, $p(S_1) \geq p(S_2)$ if $S_1 \subseteq S_2$ for $S_1,S_2 \subseteq \mo$ holds for the function $p()$.
%
%Now, since $\mathsf{node}$ is a non-minimal diagnosis, there is some $\md \subset \mathsf{node}$ which is a diagnosis w.r.t.\ $\langle\mo,\mb,\Tp,\Tn\rangle_\RQ$. Hence, $p(\md) > p(\mathsf{node})$. By the order of node expansion in descending order as per $p()$, $\md$ must have been processed before $\mathsf{node}$. Since the non-minimality criterion is tested in lines~\ref{algoline:hs:non_min_crit_start}-\ref{algoline:hs:non_min_crit_end} before calling $\scQX$ in line~\ref{algoline:hs:qx} to check whether $\mathsf{node}$ is a diagnosis, $valid$ cannot ever have been returned by \tetxsc{label} for $\mathsf{node}$. Consequently, $\mathsf{node}$ cannot ever have been added to $\mD_{calc}$, contradiction. 
%
%To show the first point, assume the algorithm adds a non-minimal diagnosis $\mathsf{node}$ w.r.t.\ $\langle\mo,\mb,\Tp,\Tn\rangle_\RQ$ to $\mD_{calc}$, i.e.\ there is a subset $\mathsf{node}'$ of $\mathsf{node}$ which is also a diagnosis w.r.t.\ $\langle\mo,\mb,\Tp,\Tn\rangle_\RQ$. 
\end{proof}
Next, we argue that Algorithm~\ref{algo:hs} computes minimal diagnoses in descending order of diagnosis probability according to the parameter $p()$ given as input to the algorithm.  
\begin{corollary}\label{cor:hs_tree_finds_most-prob_diags_first}
%Assume the DPI $\langle\mo,\mb,\Tp,\Tn\rangle_\RQ$ 
%and the function $p()$ to be the inputs of 
Let the probability $p(\md)$ of a diagnosis $\md$ in Algorithm~\ref{algo:hs} be computed from the given function $p(\tax),\tax\in\mo$ as per Formula~\ref{eq:diag_prob_calc}.
\begin{enumerate}
	\item At any point in time during the execution of Algorithm~\ref{algo:hs}, $\mD_{calc}$ comprises the $|\mD_{calc}|$ most probable minimal diagnoses w.r.t.\ 
	$\langle\mo,\mb,\Tp,\Tn\rangle_\RQ$.
%	the DPI given as input to the algorithm.
	\item %Let $p(\tax) < 0.5$ for all $\tax \in \mo$. Then, 
If Algorithm~\ref{algo:hs} 
%using $p()$ 
returns a set $\mD$ of cardinality $n$, then $\mD$ is the set of the $n$ most-probable minimal diagnoses w.r.t.\ $\langle\mo,\mb,\Tp,\Tn\rangle_\RQ$.
%the DPI given as input to the algorithm.
%
%Let $p()$ be specified in a way that its definition can be extended to sets of sentences such that $p(S_1) \geq p(S_2)$ iff $S_1 \subseteq S_2$ for $S_1,S_2 \subseteq \mo$. For example, $p(S) = \frac{1}{|S|}$ would satisfy this property.
%the first n diagnoses output by the algorithm with a p function that has the property to explore subsets first, are the most probable diagnoses w.r.t. the DPI.
\end{enumerate}
\end{corollary}
\begin{proof}
(1): By Propositions~\ref{prop:hs-tree_completeness} and \ref{prop:hs-tree_soundness}, it is a fact that Algorithm~\ref{algo:hs} 
%with $p(\tax)< 0.5$ for all $\tax \in \mo$ 
computes all and only minimal diagnoses w.r.t.\ $\langle\mo,\mb,\Tp,\Tn\rangle_\RQ$. 
What must still be shown is that minimal diagnoses are added to $\mD_{calc}$ in descending order of their probability $p()$ as per Formula~\ref{eq:diag_prob_calc}. 
The probability $p(\md)$ of some diagnosis $\md$ is equal to $p_{nodes}(\md)$ since a each diagnosis is a node and Formula~\ref{eq:diag_prob_calc} is a special case of Formula~\ref{eq:path_prob_calc} by which the probability $p_{nodes}(\mathsf{nd})$ of a node $\mathsf{nd}$ is calculated.

Let us denote by $\md_{pmax}$ the minimal diagnosis with maximum probability that has not yet been added to $\mD_{calc}$ and by $\md_{\lnot pmax}$ an arbitrary minimal diagnosis with non-maximal probability, that is $p_{nodes}(\md_{\lnot pmax}) < p_{nodes}(\md_{pmax})$. So, we need to demonstrate that each node $\mathsf{nd} \subset \md_{pmax}$ on a path from the root node to node $\md_{pmax}$ is processed before $\md_{\lnot pmax}$ is treated. By Lemma~\ref{lem:algo_hs_path+cs}, a path from the root node in the pHS-Tree produced by Algorithm~\ref{algo:hs} is a set of nodes $\emptyset=\mathsf{nd_1},\dots,\mathsf{nd_k}$ where $\mathsf{nd}_i \subset \mathsf{nd}_{i+1}$ and $|\mathsf{nd}_i|+1 = |\mathsf{nd}_{i+1}|$. Further recall that the probability $p_{nodes}(X)$ of a node $X\subseteq \mo$ in Algorithm~\ref{algo:hs} is defined as per Formula~\ref{eq:path_prob_calc}. So, by Lemma~\ref{lem:superset_lower_prob}, the node probabilities along any path from the root node are strictly monotonically decreasing. Hence, each node $\mathsf{nd}$ on a path from the root node to $\md_{pmax}$ has a probability $p_{nodes}(\mathsf{nd}) > p_{nodes}(\md_{pmax}) > p_{nodes}(\md_{\lnot pmax})$. By the insertion of new nodes into $\Queue$ (\textsc{insertSorted} in line~\ref{algoline:hs:generate_nodes}) in a way descending order of $\Queue$ as per $p_{nodes}()$ is always maintained, and by the selection of the first element of $\Queue$ (\textsc{getFirst} in line~\ref{algoline:hs:getfirst}) as next node to be processed, each node $\mathsf{nd}$ on a path to $\md_{pmax}$ must be processed before $\md_{\lnot pmax}$ is processed. Consequently, minimal diagnoses are added to $\mD_{calc}$ in descending order of their probability $p()$ as per Formula~\ref{eq:diag_prob_calc}. 
%since Formula~\ref{eq:diag_prob_calc} is a special case of Formula~\ref{eq:path_prob_calc}.

(2): This proposition follows directly from (1).
\end{proof}
\begin{proposition}\label{prop:hs_prob_correct}
%Let $p(\tax) < 0.5$ for all $\tax \in \mo$. Then, 
Algorithm~\ref{algo:hs} 
%with the same inputs as in Proposition~\ref{prop:hs_bfs_correct} (except for $p$) 
always terminates and returns a set $\mD$ of minimal diagnoses w.r.t.\ $\langle\mo$, $\mb,\Tp$, $\Tn\rangle_\RQ$ which is 
\begin{itemize}
\item the set of the $|\mD|$ most probable (w.r.t.\ $p()$ and Formula~\ref{eq:diag_prob_calc}) minimal diagnoses w.r.t.\ $\langle\mo,\mb,\Tp,\Tn\rangle_\RQ$ such that $n_{\min} \leq |\mD| \leq n_{\max}$, if at least $n_{\min}$ minimal diagnoses exist w.r.t.\ $\langle\mo,\mb,\Tp,\Tn\rangle_\RQ$, or 
\item the set of all minimal diagnoses w.r.t.\ $\langle\mo,\mb,\Tp,\Tn\rangle_\RQ$, otherwise.
\end{itemize}
\end{proposition}
\begin{proof}
The proposition is a direct consequence of Propositions~\ref{prop:hs:termination}, \ref{prop:hs-tree_completeness} and \ref{prop:hs-tree_soundness} and Corollary~\ref{cor:hs_tree_finds_most-prob_diags_first}.
\end{proof}

\subsection{Using Probabilities to Compute Minimum Cardinality Diagnoses}
\label{sec:UsingProbabilitiesToComputeMinimumCardinalityDiagnoses}

The function $p:\mo\rightarrow (0,0.5)$ can be defined in a way 
%the postulation is reflected 
that minimum cardinality instead of maximum probability diagnoses 
%should be 
are identified first. To this end, $p()$ is specified as a fixpoint function that maps each formula $\tax\in\mo$ to \emph{one and the same} constant value $p(\tax) := c$ where $c$ is an arbitrary real number such that $0 < c < 0.5$, e.g.\ $c := 0.3$. That in this setting diagnoses are found in order of ascending cardinality is a simple consequence of Corollary~\ref{cor:hs_tree_finds_most-prob_diags_first}.

\begin{example}\label{example:ax_prob_calc}
Let us now study how such formula and diagnosis probabilities would be constructed for the example DPI depicted by Table~\ref{tab:example2}. Let us suppose that the KB $\mo$ in the DPI was formulated by a single user $u$ for whom the personal fault probabilities of syntactical elements $\widetilde{\mo}\cup\overline{\mo}$ given by the first row of Table~\ref{tab:example:ax_prob} have been extracted from log data of the KB editing software applied by $u$. Then, the resulting probabilities of formulas $\tax\in\mo$ as per Formula~\ref{eq:ax_prob_calc} are as presented in the rightmost column of Table~\ref{tab:example:ax_prob}. The entries in the table from the second to the last but two column display the number of occurrences of the syntactical element given by the column label in the formula given by the row label. These values are required to compute the formula probabilities listed in the last but one column as per Formula~\ref{eq:ax_prob_calc}. The final probabilities that can ``safely'' be incorporated into Algorithm~\ref{algo:hs} under a guarantee that only minimal diagnoses will be output are shown in the last column. These result from an application of Formula~\ref{eq:adapt_ax_prob_to_get_min_diags} to the probabilities given in the last but one column with an adaptation parameter $c := 0.49$.

Notice that, for example, $p(\tax_5)$ is rather high since the predicates $A$ and $Y$ as well as the connective $\lnot$ occurring in $\tax_5$ have a comparably high fault probability in relation to syntactical elements appearing in other formulas. Formula $\tax_3$, on the other hand, comprises only two predicates which should be well-understood by $u$ and no connectives except for $\rightarrow$ which is not problematic for $u$ either. Therefore, its fault probability is rather low.\qed 
\end{example}

\begin{table*}
\small
%\rowcolors[]{2}{gray!8}{gray!16} 
%\renewcommand{\arraystretch}{1}
\centering
\setlength\tabcolsep{2pt}
\begin{tabular}{c @{\kern10pt } c c c c c c c c @{\kern15pt } c c c c @{\kern15pt} c @{\kern12pt} c}
\toprule
fault prob.   & 0.25 & 0.01 & 0.03 & 0.05 & 0.4 & 0.1 & 0.6 & 0.6 & 0.01 & 0.25 & 0.05 & 0.05 & & \\ \midrule
%\cellcolor{gray!8}
& \multicolumn{8}{c}{\hspace{-18pt}terms $\widetilde{\mo}$ } & \multicolumn{4}{c}{\hspace{-18pt} logical conn.\ $\overline{\mo}$} & after Eq.~\ref{eq:ax_prob_calc} & after Eq.~\ref{eq:adapt_ax_prob_to_get_min_diags}\\ 
\cmidrule(l{0em}r{1.5em}){2-9} \cmidrule(r{1.7em}){10-13} \cmidrule(l{-0.2em}r{1.3em}){14-14} \cmidrule(l{-0.2em}r{2pt}){15-15}
%fault prob.   & 0.25 & 0.01 & 0.03 & 0.05 & 0.4 & 0.1 & 0.6 & 0.6 & 0.01 & 0.25 & 0.05 & 0.05 & \\   \cmidrule(r{3em}){2-9} \cmidrule(r{3em}){10-13}
$\tax\in\mo$	& $A$ & $B$ & $E$ & $F$ & $G$ & $X$ & $Y$ & $Z$ & $\rightarrow$ & $\lnot$ & $\land$ & $\lor$ & $p(\tax)$ & $p(\tax)$\\ \midrule
						$\tax_1$ & 1 &   & 1 &   &   &   &   &   &       1       &         &         &        & 0.28	 &	0.14		  \\
						$\tax_2$ &   &   & 1 & 1 &   & 1 & 1 & 1 &       1       &         &     2   &    1   & 0.89  &	0.43			\\
						$\tax_3$ &   & 1 &   & 1 &   &   &   &   &       1       &         &         &        & 0.07  &  0.03			\\
						$\tax_4$ &   & 1 &   &   &   & 1 &   &   &       1       &         &         &        & 0.12	 &	0.06			\\
						$\tax_5$ & 1 &   &   &   &   &   & 1 &   &       1       &     1   &         &        & 0.78	 &	0.38			\\
						$\tax_6$ &   & 1 &   &   &   &   &   & 1 &       1       &         &         &        & 0.61	 &	0.30			\\
						$\tax_7$ &   &   &   &   & 1 &   &   & 1 &       1       &         &         &        & 0.76	 &	0.37			\\ 
						\bottomrule
\end{tabular}
\caption[Computation of Fault Probabilities]{Computing fault probabilities of formulas in $\mo$ given fault probabilities of syntactical elements $e \in\widetilde{\mo}\cup\overline{\mo}$ for the DPI given by Table~\ref{tab:example2}.} \label{tab:example:ax_prob}
\end{table*}

\section[Non-Interactive Knowledge Base Debugging Algorithm]{Non-Interactive Knowledge Base Debugging Algorithm%
\sectionmark{Non-Interactive KB Debugging Algorithm}}
\sectionmark{Non-Interactive KB Debugging Algorithm}
\label{sec:non_int_debug_procedure}
%%%%%%
%\section{Non-Interactive Knowledge Base Debugging Algorithm}
%\label{sec:non_int_debug_procedure}
%%%%%%
Algorithm~\ref{algo:non_int_debug} describes the procedure for non-interactive debugging of KBs. The algorithm requires as input all the parameters that are required by Algorithm~\ref{algo:hs} and an additional parameter $auto \in \setof{\true,\false}$ indicating either automatic ($\true$) or manual ($\false$) mode. If $auto=\false$, Algorithm~\ref{algo:non_int_debug} calls $\scHS$ (Algorithm~\ref{algo:hs}) with the parameters as provided. The set of minimal diagnoses $\mD$ returned by $\scHS$ is then presented to the user who can select a diagnosis manually after inspecting the diagnoses in $\mD$. Alternatively, in case of $auto = \true$, the system calls $\scHS$ with the parameters as provided, but with $n_{\min} = n_{\max} = 1$. Hence, only the most probable minimal diagnosis is computed by $\scHS$ and returned as an output of Algorithm~\ref{algo:non_int_debug} to the user.
%The algorithm simply calls $\scHS$ (Algorithm~\ref{algo:hs}) which returns a set of diagnoses $\mD$. If $auto=\false$, $\mD$ is presented to the user who can select a diagnosis manually after inspecting the diagnoses in $\mD$. Alternatively, the system outputs the diagnosis with highest probability within $\mD$ automatically. Thereby \textsc{mode}($\mD, p()$) computes the modal value of the probability distribution $p(\md), \md \in \mD$, i.e. $\arg\max_{\md\in\mD}(p(\md))$. 

If a user wants the algorithm to output the set of all minimal diagnoses w.r.t.\ $\langle\mo,\mb,\Tp,\Tn\rangle_\RQ$, then the parameter setting $auto = \false$ and $n_{\min} = \infty$ must be chosen. If, on the other hand, a fixed number $n$ of leading diagnoses should be computed (as long as there are at least $n$ minimal diagnoses for the DPI), then $n_{\min} := n =: n_{\max}$ are the correct parameter settings. Note that in both cases the specification of $t$ has no effect.

Of course, the user can also apply Algorithm~\ref{algo:non_int_debug} several times with varying parameters $t$, $n_{\min}$, $n_{\max}$ and $p()$. Or they can specify a test case, i.e.\ add a set of formulas $X$ either to $\Tp$ (if each $\tax\in X$ should be entailed by the correct KB) or to $\Tn$ (if \emph{the conjunction of} all formulas in $X$ must not be implied by the correct KB), and rerun the algorithm with this modified DPI. 

Anyway, the user must either find the correct diagnosis (if it is an element of the output set $\mD$ at all) by hand or be convinced that the returned minimum cardinality or respectively maximum probability diagnosis is indeed the one that yields a solution KB with the intended semantics. Moreover, when formulating test cases by hand, a user can be assumed to be as likely to specify something contradictory or faulty as during creation of the KB itself.

Unsurprisingly, application of Algorithm~\ref{algo:non_int_debug} will often lead to unsatisfying solution ontologies. Remedy for this is provided by Interactive KB Debugging which on the one hand requires higher effort of one (or several) user(s), but on the other hand ensures a high quality solution in terms of its semantics to the problem of Parsimonious KB Debugging (Problem Definition~\ref{prob_def:evidence_just}). 

\begin{algorithm}[tp]
\small
\caption{Non-Interactive KB Debugging} \label{algo:non_int_debug}
\begin{algorithmic}[1]
\Require a tuple $\tuple{ \langle\mo,\mb,\Tp,\Tn\rangle_\RQ, t, n_{\min}, n_{\max}, p(), auto}$ consisting of
\begin{itemize}
	\item an admissible DPI $\langle\mo,\mb,\Tp,\Tn\rangle_\RQ$,
	\item some computation timeout $t$,
	\item a desired minimal ($n_{\min}$) and maximal ($n_{\max}$) number of diagnoses to be returned,
	\item a function $p:\mo \rightarrow (0, 0.5)$ and
	\item a boolean parameter $auto \in \setof{\true,\false}$.
\end{itemize}
%an admissible DPI $\langle\mo,\mb,\Tp,\Tn\rangle_\RQ$, 
%some computation timeout $t$, a desired minimal ($n_{\min}$) and maximal ($n_{\max}$) number of diagnoses to be returned, a function $p(\tax) \in [0,1]$ that assigns to each $\tax \in \mo$ a fault probability, a parameter $auto \in \setof{true,false}$
\Ensure a set $\mD$ which is 
\begin{enumerate}[(a)]
	\item the set of the $|\mD|$ most probable minimal diagnoses w.r.t.\ $\langle\mo,\mb,\Tp,\Tn\rangle_\RQ$ such that $n_{\min} \leq |\mD| \leq n_{\max}$, if at least $n_{\min}$ minimal diagnoses exist w.r.t.\ $\langle\mo,\mb,\Tp,\Tn\rangle_\RQ$, or
	\item the set of all minimal diagnoses w.r.t.\ $\langle\mo,\mb,\Tp,\Tn\rangle_\RQ$ otherwise
\end{enumerate}
where ``most-probable'' refers to the probability measure $p_{nodes}()$ (cf.\ Definition~\ref{def:p_node()}) obtained from the given function $p()$. 
%(a)~a set of best (according to $p()$) minimal diagnoses w.r.t.\ $\langle\mo,\mb,\Tp,\Tn\rangle_\RQ$ such that $n_{\min} \leq |\mD| \leq n_{\max}$, if at least $n_{\min}$ minimal diagnoses exist w.r.t.\ $\langle\mo,\mb,\Tp,\Tn\rangle_\RQ$, or \newline
%(b)~the set of all minimal diagnoses w.r.t.\ $\langle\mo,\mb,\Tp,\Tn\rangle_\RQ$ otherwise
\vspace{10pt}
\If{$auto = \true$}
	\State $\mD \gets \Call{\scHS}{\langle\mo,\mb,\Tp,\Tn\rangle_\RQ, t, 1, 1, p()}$  \Comment{see Algorithm~\ref{algo:hs}}
\Else 
	\State $\mD \gets \Call{\scHS}{\langle\mo,\mb,\Tp,\Tn\rangle_\RQ, t, n_{\min}, n_{\max}, p()$}  \Comment{see Algorithm~\ref{algo:hs}}
\EndIf
\State \Return $\mD$
%\State \Return $\Call{mode}{\mD, p()}$
%\State \Return $\Call{userSelectedDiagnosis}{\mD}$
\end{algorithmic}
\normalsize
\end{algorithm}

\begin{figure*}[tp]
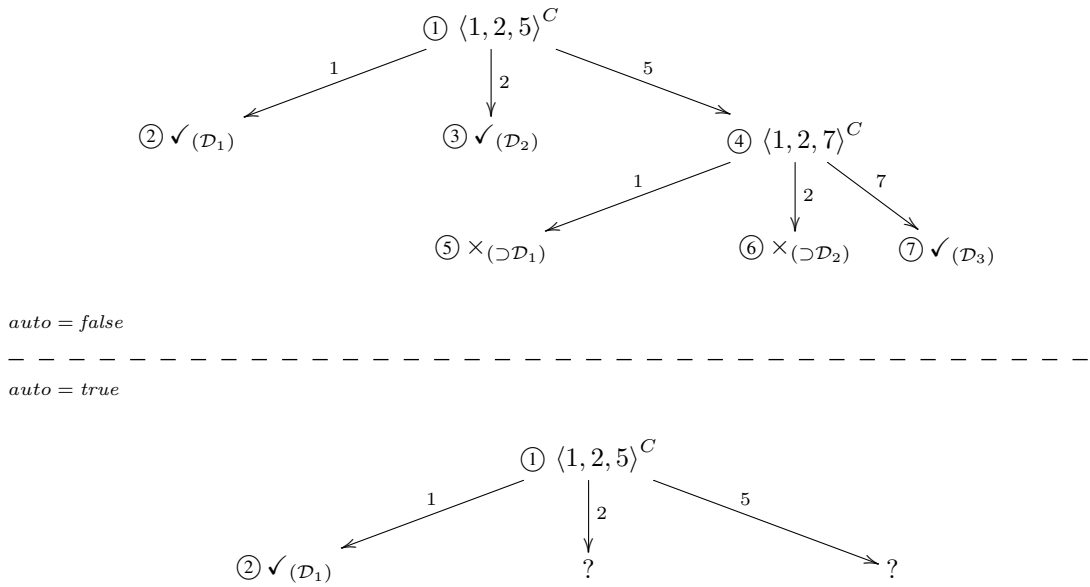

\centering
\begin{minipage}[c]{0.99\textwidth}
\xygraph{
!{<0cm,0cm>;<2cm,0cm>:<0cm,1.5cm>::}
%!~-{@[|(4)]}
%d=0
!{(1,4)}*+{{\textcircled{\scriptsize 1}}\,\tuple{1,2,5}^C}="c1c"
%d=1
!{(-1,3) }*+{{\textcircled{\scriptsize 2}}\,\checkmark_{(\md_1)}}="d1"
!{(1,3) }*+{{\textcircled{\scriptsize 3}}\,\checkmark_{(\md_2)}}="d2" 
!{(3,3) }*+{{\textcircled{\scriptsize 4}}\,\tuple{1,2,7}^C}="c2c"
%d=2 
%!{(0,2) }*+{\times}="inv_d1_q1"
%!{(1,2) }*+{\checkmark}="val_d2_q1"
!{(1,2) }*+{{\textcircled{\scriptsize 5}}\,\times_{(\supset\md_1)}}="nonmin"
!{(3,2) }*+{{\textcircled{\scriptsize 6}}\,\times_{(\supset\md_2)}}="nonmin1"
!{(4,2) }*+{{\textcircled{\scriptsize 7}}\,\checkmark_{(\md_3)}}="d3"
%%d=3
%!{(1,1) }*+{\times}="inv_d2_q2"
%!{(2,1) }*+{\times}="inv_d3_q2"
%!{(3,1) }*+{\checkmark}="d5"
%!{(4,1) }*+{\checkmark}="val_d3_q2"
%%d=4
%!{(1,0) }*+{\times}="inv_d2_q3"
%!{(3,0) }*+{\times}="inv_d5_q4"
%!{(4,0) }*+{\checkmark}="val_d4_q4"
%d0->d1
"c1c":"d1"_{1}
"c1c":"d2"^{2}
"c1c":"c2c"^{5}
%d1->d2
%"d1":@2{->}"inv_d1_q1"^{Q_1}
%"d2":@2{->}"val_d2_q1"^{Q_1}
"c2c":"nonmin"_{1}
"c2c":"nonmin1"^{2}
"c2c":"d3"^{7}
%%d2->d3
%"val_d2_q1":@2{->}"inv_d2_q2"^{Q_2}
%"d3":@2{->}"inv_d3_q2"^{Q_2}
%"nonmin1":@2{->}"d5"^{Q_3}
%"d3":@2{->}"val_d3_q2"^{Q_2}
%%d3->d4
%"d5":@2{->}"inv_d5_q4"^{Q_4}
%"val_d4_q3":@2{->}"val_d4_q4"^{Q_4}   
%"val_d2_q2":@2{->}"inv_d2_q3"^{Q_3}
}
\end{minipage}

\vspace{5pt}
\begin{flushleft}
\scriptsize $auto=\false$
\end{flushleft}
\vspace{-10pt} 
%\rule{15cm}{0.35pt}
\hdashrule{\textwidth}{0.5pt}{2mm}
%\hrule
\vspace{-15pt} 
\begin{flushleft}
\scriptsize $auto=\true$
\end{flushleft} 

\vspace{5pt}
\centering
\begin{minipage}[c]{0.99\textwidth} 
\xygraph{
!{<0cm,0cm>;<2cm,0cm>:<0cm,1.5cm>::}
%!~-{@[|(4)]}
%d=0
!{(1,4)}*+{{\textcircled{\scriptsize 1}}\,\tuple{1,2,5}^C}="c1c"
%d=1
!{(-1,3) }*+{{\textcircled{\scriptsize 2}}\,\checkmark_{(\md_1)}}="d1"
!{(1,3) }*+{?}="d2" 
!{(3,3) }*+{?}="c2c"
%d=2 
%!{(0,2) }*+{\times}="inv_d1_q1"
%!{(1,2) }*+{\checkmark}="val_d2_q1"
%%%%%%%!{(1,2) }*+{{\textcircled{\tiny 5}}\,\times_{(\supset\md_1)}}="nonmin"
%%%%%%%!{(3,2) }*+{{\textcircled{\tiny 6}}\,\times_{(\supset\md_2)}}="nonmin1"
%%%%%%%!{(4,2) }*+{{\textcircled{\tiny 7}}\,\checkmark_{(\md_3)}}="d3"
%%d=3
%!{(1,1) }*+{\times}="inv_d2_q2"
%!{(2,1) }*+{\times}="inv_d3_q2"
%!{(3,1) }*+{\checkmark}="d5"
%!{(4,1) }*+{\checkmark}="val_d3_q2"
%%d=4
%!{(1,0) }*+{\times}="inv_d2_q3"
%!{(3,0) }*+{\times}="inv_d5_q4"
%!{(4,0) }*+{\checkmark}="val_d4_q4"
%d0->d1
"c1c":"d1"_{1}
"c1c":"d2"^{2}
"c1c":"c2c"^{5}
%d1->d2
%"d1":@2{->}"inv_d1_q1"^{Q_1}
%"d2":@2{->}"val_d2_q1"^{Q_1}
%%%%%%%"c2c":"nonmin"_{1}
%%%%%%%"c2c":"nonmin1"^{2}
%%%%%%%"c2c":"d3"^{7}
%%d2->d3
%"val_d2_q1":@2{->}"inv_d2_q2"^{Q_2}
%"d3":@2{->}"inv_d3_q2"^{Q_2}
%"nonmin1":@2{->}"d5"^{Q_3}
%"d3":@2{->}"val_d3_q2"^{Q_2}
%%d3->d4
%"d5":@2{->}"inv_d5_q4"^{Q_4}
%"val_d4_q3":@2{->}"val_d4_q4"^{Q_4}   
%"val_d2_q2":@2{->}"inv_d2_q3"^{Q_3}
}
\end{minipage}

\vspace{10pt}
\caption[Non-Interactive KB Debugging Process without Fault Information]{Non-interactive KB debugging process without any given fault information applied to the DPI given by Table~\ref{tab:example2} with settings $auto = \false$ and $n_{\min} = \infty$ (above) and $auto = \true$ (below).} \label{fig:example:non-interactive_onto_debug_auto=false+nmin=infty_and_auto=true}
\end{figure*}

\begin{figure*}[tp]
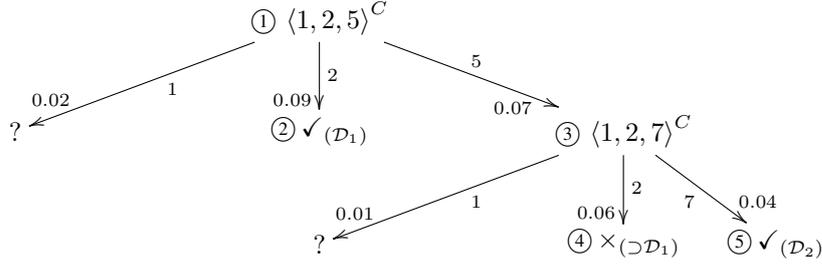

\centering
\begin{minipage}[c]{0.99\textwidth} 
\xygraph{
!{<0cm,0cm>;<2cm,0cm>:<0cm,1.5cm>::}
%!~-{@[|(4)]}
%d=0
!{(1,4)}*+{{\textcircled{\scriptsize 1}}\,\tuple{1,2,5}^C}="c1c"
%d=1
!{(-1,3) }*+{?}="d1"
!{(1,3) }*+{{\textcircled{\scriptsize 2}}\,\checkmark_{(\md_1)}}="d2" 
!{(3,3) }*+{{\textcircled{\scriptsize 3}}\,\tuple{1,2,7}^C}="c2c"
%d=2 
%!{(0,2) }*+{\times}="inv_d1_q1"
%!{(1,2) }*+{\checkmark}="val_d2_q1"
!{(1,2) }*+{?}="nonmin"
!{(3,2) }*+{{\textcircled{\scriptsize 4}}\,\times_{(\supset\md_1)}}="nonmin1"
!{(4,2) }*+{{\textcircled{\scriptsize 5}}\,\checkmark_{(\md_2)}}="d3"
%%d=3
%!{(1,1) }*+{\times}="inv_d2_q2"
%!{(2,1) }*+{\times}="inv_d3_q2"
%!{(3,1) }*+{\checkmark}="d5"
%!{(4,1) }*+{\checkmark}="val_d3_q2"
%%d=4
%!{(1,0) }*+{\times}="inv_d2_q3"
%!{(3,0) }*+{\times}="inv_d5_q4"
%!{(4,0) }*+{\checkmark}="val_d4_q4"
%d0->d1
"c1c":"d1"^{1}_(0.85){0.02}
"c1c":"d2"^{2}_(0.72){0.09}
"c1c":"c2c"^{5}_(0.67){0.07}
%d1->d2
%"d1":@2{->}"inv_d1_q1"^{Q_1}
%"d2":@2{->}"val_d2_q1"^{Q_1}
"c2c":"nonmin"^{1}_(0.85){0.01}
"c2c":"nonmin1"^{2}_(0.72){0.06}
"c2c":"d3"_{7}^(0.75){0.04}
%%d2->d3
%"val_d2_q1":@2{->}"inv_d2_q2"^{Q_2}
%"d3":@2{->}"inv_d3_q2"^{Q_2}
%"nonmin1":@2{->}"d5"^{Q_3}
%"d3":@2{->}"val_d3_q2"^{Q_2}
%%d3->d4
%"d5":@2{->}"inv_d5_q4"^{Q_4}
%"val_d4_q3":@2{->}"val_d4_q4"^{Q_4}   
%"val_d2_q2":@2{->}"inv_d2_q3"^{Q_3}
}
\end{minipage}

\vspace{10pt}
\caption[Non-Interactive KB Debugging Process with Fault Information]{Non-interactive KB debugging process with given fault information applied to the DPI given by Table~\ref{tab:example2} with settings $auto = \false$, $n_{\min} = 2$, $n_{\max} = 4$ and $t = 1$.} \label{fig:example:non-interactive_onto_debug_auto=false+nmin=2+nmax=4_with_probs}
\end{figure*}

\begin{example}\label{example:non_interactive_debugging_with_tabExDpi2_and_without_probs}
Assume a user wants to find a maximal solution KB for the example DPI $\tuple{\mo,\mb,\Tp,\Tn}_{\RQ}$ provided by Table~\ref{tab:example2} and that no data giving information about fault probabilities of syntactical constructs or formulas in $\mo$ is available. Therefore, let $p(\tax) := c$ for some fixed $c \in (0,0.5)$ (see Section~\ref{sec:probs_diag_comp} for an explanation of this choice of $c$). The non-interactive KB debugging algorithm presented by Algorithm~\ref{algo:non_int_debug} called with $\tuple{\mo,\mb,\Tp,\Tn}_{\RQ}$, the function $p()$, $n_{\min}=\infty$ and $auto = \false$ as inputs results in the hitting set tree given by the upper picture in Figure~\ref{fig:example:non-interactive_onto_debug_auto=false+nmin=infty_and_auto=true}. By $n_{\min}=\infty$ and $auto = \false$, the user signalizes that \emph{inspection of} \emph{all} minimal diagnoses w.r.t.\ the input DPI is desired. Hence, the (complete) breadth-first pHS-tree as per Algorithm~\ref{algo:hs} is constructed. So, the output is the set of all minimal diagnoses $\minD_{\tuple{\mo,\mb,\Tp,\Tn}_{\RQ}} = \setof{[1],[2],[5,7]}$.

In the shown hitting set tree, minimal diagnoses are indicated by nodes labeled by $\checkmark_{(\md)}$ where $\md$ is a name given to this diagnosis. A node closed due to non-minimality is denoted by $\times_{(\supset \md)}$ where $\md$ is some minimal diagnosis that is a subset of the set of edge labels along the path leading from the root node to this node. %Nodes in $\Queue$ are expressed by a question mark.
The label $\mc^C$ means that the minimal conflict set $\mc$ has been freshly computed by a call to $\scQX$. The label $\mc^R$, on the other hand, means that the minimal conflict set $\mc$ has been reused from the set of already computed minimal conflict sets. In this example, both minimal conflict sets are computed by $\scQX$ and no conflict sets are reused. The order of node labeling is indicated by the numbers $\textcircled{\scriptsize i}$ starting from 1. Open nodes, i.e.\ generated nodes that have not yet been labeled, are indicated by a question mark.

In case $auto = \true$ was given as an input to the algorithm instead, the partial pHS-tree depicted by the lower picture in Figure~\ref{fig:example:non-interactive_onto_debug_auto=false+nmin=infty_and_auto=true} would be constructed and the output would be $\mD = \setof{\md_1} = \setof{[1]}$ containing just the first found and thus most probable minimal diagnosis w.r.t.\ the input DPI. Note that $\md_1 = [1]$ and $\md_2 = [2]$ (which is not computed) have equal probability and whether the one or the other is computed first depends only on the ordering of equally probable (in this case: equal cardinality) nodes in $\Queue$. As already mentioned in Section~\ref{sec:probs_diag_comp}, in this example the most probable diagnosis is equivalent to a minimum cardinality diagnosis since all formula probabilities are equal.

Please notice that the internal ``flat'' representation used by Algorithm~\ref{algo:hs} which does not store a tree but only the set of open and closed nodes differs from the standard tree representation~\cite{Kalyanpur2006a, friedrich2005gdm, Suntisrivaraporn2008, Reiter87} we use to depict the hitting set tree \emph{graphically} in Figure~\ref{fig:example:non-interactive_onto_debug_auto=false+nmin=infty_and_auto=true}. Whereas within Algorithm~\ref{algo:hs} a node $\mathsf{node}$ stores the set of all the edge labels on the path leading from the root node to $\mathsf{node}$, in the figure we label each node in the tree by the respective label that is computed for this node by the \textsc{label} function, i.e.\ either by a minimal conflict set, by $\checkmark$ or by $\times$.\qed
\end{example}

\begin{example}\label{example:non_interactive_debugging_with_tabExDpi2_and_probs}
Recall Example~\ref{example:ax_prob_calc} which demonstrated how formula fault probabilities are constructed from fault probabilities of syntactical elements for the example DPI depicted by Table~\ref{tab:example2}. Now we want to show how the non-interactive KB debugging algorithm given by Algorithm~\ref{algo:non_int_debug} works when these formula probabilities are incorporated. 

Suppose the inputs to the algorithm are the DPI $\tuple{\mo,\mb,\Tp,\Tn}_{\RQ}$, the function $p(\tax)$ for $\tax\in\mo$ displayed by the rightmost column of Table~\ref{tab:example:ax_prob} and $auto = \false$. Further on, let the user of the debugging algorithm be willing to wait a maximum of one second for an output and let them postulate a minimum of two most probable minimal diagnoses to be returned, e.g.\ to have at least a second choice if the employed formula probabilities are not perfectly suitable and the most probable diagnosis is not the desired solution. These postulations are expressed by specifying the parameters $n_{\min} = 2$ and $t = 1$ (second). Additionally, assume the user expects the provided probabilities to be sufficiently reasonable such that the desired diagnosis will be among the best four diagnoses wherefore $n_{\max} = 4$ is chosen. Moreover, let us imagine that the time for each fresh computation of a minimal conflict plus generation of the (unlabeled) successor nodes of this node is $0.4$ seconds 
%cost of each call to $\scQX$, 
%i.e.\ , is , 
and the cost of computing any other label of a node is $0.1$ seconds. 

Then the partial wpHS-tree produced by Algorithm~\ref{algo:non_int_debug} initialized in this way is illustrated by Figure~\ref{fig:example:non-interactive_onto_debug_auto=false+nmin=2+nmax=4_with_probs}. The used notation is as described in Example~\ref{example:non_interactive_debugging_with_tabExDpi2_and_without_probs} with one additional attribute. Namely, each edge is not only labeled by one element of the conflict set from which it goes out, but also by a label $p \in (0,1)$ that is placed near the arrow head of the arrow that expresses the edge. This label $p$ gives 
%$(p)$ where $p \in (0,1)$ is the probability of the (partial) diagnosis that corresponds to the union of the non-bracketed edge labels from the root to and including the edge that is labeled by $(p)$.
%$p$ where $p \in (0,1)$ is 
the probability as per $p_{nodes}()$ (cf.\ Definition~\ref{def:p_node()}) of the (partial) diagnosis that corresponds to the union of the 
%non-bracketed 
edge labels along the path from the root to and including the edge that is labeled by $p$. 
For example, the label $0.06$ of the edge directed at the node number $\textcircled{\scriptsize 4}$ means that the probability of $\setof{2,5}$ is $0.06$. Further on, open, i.e.\ generated, but not yet labeled nodes, are designated by a question mark.

As outlined by the circled numbers $\textcircled{\scriptsize i}$, as a first action the root node is labeled by the newly computed minimal conflict set $\tuple{1,2,5}$, the computation time of which amounts to $0.4$. Then, the tree construction proceeds according to the (partial) diagnosis probabilities according to $p_{nodes}()$ computed from the formula probabilities $p(\tax), \tax\in\mo$ provided by the last column of Table~\ref{tab:example:ax_prob}. 
%by Formula~\ref{eq:diag_prob_calc}. 
Therefore, the most probable edge leading away from the root node is labeled next. This already leads to the finding of the first minimal diagnosis $\md_1 = [2]$ after overall computation time of $0.5$ seconds. Since $n_{\min} = 2$ diagnoses have not yet been computed and there are still unlabeled open nodes, namely those corresponding to paths $\setof{1}$ and $\setof{5}$, the algorithm continues the execution by labeling the next best node $\setof{5}$ with a probability of $0.07$ -- as opposed to $0.02$ for the other open node $\setof{1}$. Since $\setof{5}$ is neither a superset of an already computed minimal diagnosis nor a duplicate of another open node nor a diagnosis itself, it must be labeled by some minimal conflict set. Because the already established minimal conflict set $\tuple{1,2,5}$ is not disjoint with $\setof{5}$, no reuse is possible and $\scQX$ is called to determine a new minimal conflict set $\tuple{1,2,7}$ w.r.t.\ $\tuple{\mo,\mb,\Tp,\Tn}_{\RQ}$. All successor nodes of the newly labeled node $\textcircled{\scriptsize 3}$, i.e.\ the nodes corresponding to the paths $\setof{1,5}, \setof{2,5}$ and $\setof{5,7}$, are added to the list $\Queue$ of open nodes such that descending order of probabilities is maintained. The resulting queue is then $\Queue = [\setof{2,5},\setof{5,7}, \setof{1}, \setof{1,5}]$. As a next step, again the first and thus best open node $\setof{2,5}$ is chosen from $\Queue$ and labeled by $\times_{(\supset \md_1)}$ which means that the corresponding path is closed since it is a superset of an already found minimal diagnosis, namely $\md_1 = [2]$. At this point, the overall computation time amounts to $1$ second which corresponds to the time limit $t$. For that reason, the algorithm will go ahead searching for minimal diagnoses only until a minimal number $n_{\min}$ thereof is detected. 
The node processed next, corresponding to the path $\setof{5,7}$, is then determined to be a minimal diagnosis by the \textsc{label} procedure.
 
Thus, the output of the algorithm after $1.1$ seconds execution time is the set of minimal diagnoses $\mD = \setof{[2],[5,7]}$ which is a proper subset of all minimal diagnoses $\mD_{\tuple{\mo,\mb,\Tp,\Tn}_{\RQ}} = \setof{[1],[2],[5,7]}$. However, if we assume that the user's intended KB should entail $E \rightarrow G$, for instance, then none of the returned diagnoses can be used to compute a solution KB featuring this entailment when integrated with the background knowledge $\mb$. Hence, the true diagnosis $\dt$ would be missed in this case.

Also, when computing all minimal diagnoses w.r.t.\ a DPI -- if this is even possible in a concrete case due to the computational complexity -- and showing them to the user, a user might review just the most probable ones and make a decision on which one to choose only based on these. For instance, \cite{ksgf2010} reported on one DPI where computation of all minimal diagnoses, 1782 in number, is feasible. In such a case it is hard to expect that a user will be willing or will have the time to inspect more than a small fraction of these 1782 diagnoses. The consequence will be a wrong choice of diagnosis in many cases, also because a simple view on a diagnosis will often not lead to the certainty of a user that this one is or is not the desired one. The reason for this is that usually it is too complex for a human brain to perform the necessary mental reasoning to make oneself a picture of the implications of choosing one diagnosis as opposed to another one.

For our example DPI, a user getting the output $\mD = \minD_{\tuple{\mo,\mb,\Tp,\Tn}_{\RQ}} = \setof{[1],[2],[5,7]}$ with the computed probabilities $p([1]) = 12\%$, $p([2]) = 60\%$ and $p([5,7]) = 28\%$ might decide to just inspect the diagnoses that make the most probable $80\%$ fraction of diagnoses. In this case, either $[2]$ or $[5,7]$ would be selected, which corresponds to a wrong choice in case $E \rightarrow G$ should be entailed be the resulting solution KB after integration with the background KB $\mb$.\qed  
\end{example}

\chapter{Summary}
\label{sec:summary_part_prolog}
In this part, we profoundly introduced the topic of knowledge base debugging. We stated necessary properties of knowledge representation languages to be compatible with our approaches, namely that the entailment relation must be monotonic, idempotent and extensive. We gave precise definitions of the problems of KB debugging and parsimonious KB debugging. Both problems assume a given instance of a diagnosis problem (DPI). The former seeks any solution in line with the given requirements whereas the latter seeks a solution that preserves as much formulas as possible of the given faulty KB, i.e.\ aims at minimal changes. With the \emph{validity of a KB}, a \emph{solution KB}, a \emph{diagnosis} and a \emph{conflict set}, we have characterized central notions that will be extensively used throughout this work. We have studied the relationship between all these notions and proved that solving the problem of parsimonious KB debugging is equivalent to finding a minimal diagnosis w.r.t.\ a given DPI.

We established the relationship between conflict sets and justifications, a similar notion that is used concurrently to conflict sets in (prevalently DL, OWL or Semantic Web) literature, and provided evidence that conflict sets are the better choice for the debugging problems addressed here. In particular, conflict sets serve the purpose of reducing the search space for minimal diagnoses -- minimal hitting sets of all minimal conflict sets -- and help a debugging software to focus on the relevant and problematic parts of the faulty KB. A method for the efficient, polynomial time computation of a conflict set was detailed and its correctness was formally proven. Based on this method, we were able to depict a way of computing minimal diagnoses which is based on using a hitting set tree. Such a tree constitutes a systematic way of generating all minimal conflict sets and, in the course of this, also all minimal diagnoses. Depending on the particular situation, the presented algorithm can be configured to compute diagnoses in a predefined order, e.g.\ most probable diagnoses first or those diagnoses first that are minimally invasive in terms of the changes made to the faulty KB. 

Different ways of obtaining and incorporating meta (fault) information into the debugging process were elucidated. Such information, if reasonable, can facilitate and accelerate the debugging process significantly. However, even in the case of the availability of high-quality fault information, we discovered substantial drawbacks of the debugging system presented so far. That is, such a system either chooses automatically a solution (diagnosis) based on the given fault information in a solution space of (generally) exponential size or refers a subset of all solutions, e.g.\ the most probable solutions, to the user for manual inspection. In the former case, the probability of being presented a solution KB with undesired semantics is very high implying unwanted changes to the faulty KB and unexpected entailments and non-entailments as well as future errors. Such unexpected semantics can be critical or even fatal; one should imagine intelligent medical applications relying on such KBs, for instance. In the latter case, the burden is placed on the user(s) who must mentally anticipate the implications of applying different repairs (using the different submitted diagnoses) to the KB which is practically impossible for human beings both from the time/effort as well as from the mental perspective. Moreover, it is basically intractable to generate all possible solutions. Hence, it is not even sure that the manually investigated solutions include to correct one (with the postulated semantics). 

This leads us to the next part which deals with exactly these issues and proposes a solution.    
% of the central notions of a diagnosis problem instance (DPI) which is the input to a KB debugging problem

\part{Interactive Knowledge Base Debugging}
%\label{sec:InteractiveOntologyDebugging}
\label{part:InteractiveKBDebugging}

This part is organized as follows:\\

In Chapter~\ref{chap:MotivationAndProblemDefinitions}, we first discuss how disadvantages of non-interactive KB debugging procedures can be overcome by allowing a user to take part in the debugging process. Next, we define the problem of \emph{interactive static KB debugging} as well as the problem of \emph{interactive dynamic KB debugging} which ``naturally'' arise from the fact that the DPI in interactive KB debugging is always renewed after a new test case has been specified (a new query has been answered). The former problem searches for a solution KB \emph{w.r.t.\ the DPI given as input} such that this solution KB satisfies all test cases added during the debugging session and there is no other such solution KB. The latter problem searches for a solution KB \emph{w.r.t.\ the current DPI} (i.e.\ the input DPI including all new test cases added throughout the debugging session so far) such that there is no other solution KB w.r.t.\ the current DPI.

Next, in Chapter~\ref{chap:UserInteraction}, the central term of a \emph{query} is specified which constitutes the medium for user interaction. Queries are generated from a set of \emph{leading diagnoses} which is characterized thereafter. The set of leading diagnoses is uniquely partitioned into three subsets by each query. The tuple including these subsets is called \emph{q-partition}. Subsequently, the reader is given some explanations how the q-partition can be interpreted, and how it relates to a query. In fact, we will prove that the notion of a q-partition can serve as a criterion for checking whether a set of logical formulas is a query or not. After that, we will learn that a query exists for any set of (at least two) leading diagnoses which grants that the presented algorithms will definitely be able to come up with a query without the need to impose any restrictions on which (minimal) diagnoses are computed by the diagnosis engine in each iteration. 

Chapter~\ref{chap:QueryGeneration} shows a method for the generation of (a pool of) set-minimal queries (Algorithm~\ref{algo:query_gen}) aiming at stressing the interacting user as sparsely as possible, features in-depth discussions of this method's properties, proves its correctness, provides complexity results and gives some illustrating examples. Further on, drawbacks of this method are pointed out and possible solutions are discussed. 

Subsequently, Chapter~\ref{chap:WorkflowInInteractiveKBDebugging} deals with the presentation of the central algorithm of this work which implements an interactive KB debugging system (Algorithm~\ref{algo:inter_onto_debug}). First, an overview of the workflow of interactive KB debugging %(Section~\ref{sec:AlgorithmOverview}) 
is given, followed by a more comprehensive detailed specification of the algorithm. 
%(Section~\ref{sec:DetailedAlgorithmDescription}). 
Some query selection measures are discussed~\cite{Rodler2013,Shchekotykhin2012} 
%(Section~\ref{sec:query_selection_measures}) 
and optimization versions of the problems of interactive dynamic and static KB debugging are defined where the goal is to obtain the solution to these problems by asking the user a minimal number of queries. Finally, %Section~\ref{sec:CorrectnessOfAlgorithmInterOntoDebug}
we prove the correctness of the interactive KB debugging algorithm and provide a discussion of its complexity.

Non-theoretically-oriented readers might well skip Sections~\ref{sec:remarks_query_gen}, \ref{sec:SoundnessOfQueryMinimization}, \ref{sec:query_gen_complexity}, \ref{sec:query_gen_correctness} and \ref{sec:CorrectnessOfAlgorithmInterOntoDebug} in this part. Moreover, for the superficially interested reader, it may suffice to concentrate only on Chapter~\ref{chap:MotivationAndProblemDefinitions} and Sections~\ref{sec:Queries}, \ref{sec:LeadingDiagnoses} and \ref{sec:AlgorithmOverview} in this part.\footnote{Parts of Part~\ref{part:InteractiveKBDebugging} already appeared in \cite{Rodler2015}.}

\chapter{Motivation and Problem Definitions}
\label{chap:MotivationAndProblemDefinitions}
So far, we have learned that the problem of (parsimonious) KB debugging as defined in Problem Definitions~\ref{prob_def:onto_debug} and \ref{prob_def:evidence_just} in Chapter~\ref{chap:KBDebugging} can be solved by investigating minimal diagnoses w.r.t.\ a given DPI $\langle\mo,\mb,\Tp,\Tn\rangle_\RQ$. We have seen how minimal diagnoses can be computed, we have introduced a probability space over diagnoses and we have discussed how a-priori probability estimates for diagnoses can be established. Now, assume the situation where a DPI with say $100$ minimal diagnoses is given, among which there is one diagnosis $\md$ with highest estimated probability $p(\md) = 10\%$. By the definitions of a diagnosis and a solution KB (Definitions~\ref{def:target_ont} and \ref{def:diagnosis}), each of the $100$ diagnoses can be used to formulate a solution KB w.r.t.\ the DPI $\langle\mo,\mb,\Tp,\Tn\rangle_\RQ$. So, should the system output the solution KB $(\mo \setminus \md) \cup U_{\Tp}$ obtained from $\md$ as the optimal solution? Will a user be satisfied with a likeliness of $90\%$ of being offered a suboptimal solution? What if the diagnoses probabilities are bad estimates and another diagnosis $\md'$ should actually have a probability of $20\%$? 

Why not simply apply Algorithm~\ref{algo:non_int_debug} to show all $100$ minimal diagnoses to the user and let them select the preferred one by hand? First, due to the complexity of diagnosis calculation algorithms (cf.\ Chapter~\ref{chap:intro}), pre-computation of $100$ (or, generally, all) minimal diagnoses is usually not tractable within reasonable time. This makes such an approach quite unattractive in an interactive setting. Second, going through large sets of diagnoses can be time-consuming, tedious and error-prone. Third, human beings are normally not capable of (fully) realizing the semantic consequences of deleting a diagnosis from a KB, especially if the KB is large, complex and/or has been created by multiple engineers or automatic systems. Thus, applying a suboptimal diagnosis can result in unexpected entailments or unwanted changes, and thus an incorrect solution KB (incorrect in the sense of the semantics, \emph{not} in the sense of violating given requirements or test cases), which might cause unexpected new faults and contradictions when augmented by new formulas. Consequently, a solution diagnosis is only acceptable if the user has sufficiently scrutinized and approved its semantic effect to the KB.

This leads to the definition of two types of Interactive KB Debugging problems. First, there is the problem of \emph{Interactive Dynamic KB Debugging} which, given an input DPI, aims at the extension of this DPI by new test cases confirmed by a user such that there is only one minimal diagnosis left w.r.t.\ the extended DPI. Second, we specify the problem of \emph{Interactive Static KB Debugging} which, given an input DPI, aims at the formulation of new test cases confirmed by a user such that these new test cases rule out all but one minimal diagnosis w.r.t.\ the input DPI.\vspace{3pt}

\noindent\fcolorbox{black}{light-gray1}{\parbox[c][7.1em][c]{0.975\linewidth}{\vspace{-4pt}
\begin{prob_def}[Interactive Dynamic KB Debugging]\label{prob_def:dynamic}
Given a DPI $\langle\mo,\mb,\Tp,\Tn\rangle_\RQ$, the task is to find a maximal solution KB $(\mo\setminus\md) \cup U_{\Tp\cup\Tp'}$ w.r.t.\ a DPI $\langle\mo,\mb,\Tp\cup\Tp',\Tn\cup\Tn'\rangle_\RQ$ such that 
\begin{itemize}
\item $\md$ is the only minimal diagnosis w.r.t.\ $\langle\mo,\mb,\Tp\cup\Tp',\Tn\cup\Tn'\rangle_\RQ$ and 
\item a user has confirmed that each $\tp'\in\Tp'$ is a positive test case and that each $\tn'\in\Tn'$ is a negative test case.
\end{itemize}
%Given a DPI $\langle\mo,\mb,\Tp,\Tn\rangle_\RQ$, the task is to find sets of sets of logical sentences $\Tp'$ and $\Tn'$ such that there is only a single minimal diagnosis w.r.t.\ $\langle\mo,\mb,\Tp\cup\Tp',\Tn\cup\Tn'\rangle_\RQ$.
\end{prob_def}\vspace{-4pt}
}}

\vspace{3pt}
%\fixme{define D as the true diagnosis}
%Notice that the 
%single minimal diagnosis w.r.t.\ $\langle\mo,\mb,\Tp\cup\Tp',\Tn\cup\Tn'\rangle_\RQ$ 
%solution KB found 
\begin{remark}\label{rem:ad_int_dyn_KB_debug_problem}
The solution of an Interactive Dynamic KB Debugging problem given the DPI $\langle\mo,\mb,\Tp,\Tn\rangle_\RQ$ solves the problem of KB Debugging (Problem Defnition~\ref{prob_def:onto_debug}) as well as the problem of Parsimonious KB Debugging (Problem Defnition~\ref{prob_def:evidence_just}) for the DPI $\langle\mo,\mb,\Tp\cup\Tp',\Tn\cup\Tn'\rangle_\RQ$, but in general \emph{not} for the original DPI $\langle\mo,\mb,\Tp,\Tn\rangle_\RQ$. This is the reason why we term it ``dynamic'', since a solution is found for a version of the initial DPI that has been extended by test cases.\qed
\end{remark}
\vspace{4pt}
%However, we will also provide an algorithm that solves the Interactive Static KB Debugging problem given in the following:\vspace{4pt}

%\begin{framed}
\noindent\fcolorbox{black}{light-gray1}{\parbox[c][8.5em][c]{0.975\linewidth}{\vspace{-4pt}
\begin{prob_def}[Interactive Static KB Debugging]\label{prob_def:static}
Given a DPI $\langle\mo,\mb,\Tp,\Tn\rangle_\RQ$, the task is to find a maximal solution KB $(\mo\setminus\md) \cup U_\Tp$ w.r.t.\ $\langle\mo,\mb,\Tp,\Tn\rangle_\RQ$ such that 
\begin{itemize}
\item there are sets of positive test cases $\Tp'$ and negative test cases $\Tn'$ where a user has confirmed that each $\tp'\in\Tp'$ is a positive test case and that each $\tn'\in\Tn'$ is a negative test case, and 
\item $\md$ is the only minimal diagnosis w.r.t.\ $\langle\mo,\mb,\Tp,\Tn\rangle_\RQ$ that satisfies all positive and negative test cases $\Tp'$ and $\Tn'$, respectively.
\end{itemize}
%there is a single minimal diagnosis w.r.t.\ $\langle\mo,\mb,\Tp\cup\Tp',\Tn\cup\Tn'\rangle_\RQ$ and a user has confirmed that each $\tp'\in\Tp'$ is a a positive test case and that each $\tn'\in\Tn'$ is a a negative test case.
%Given a DPI $\langle\mo,\mb,\Tp,\Tn\rangle_\RQ$, the task is to find sets of sets of logical sentences $\Tp'$ and $\Tn'$ such that there is only a single minimal diagnosis w.r.t.\ $\langle\mo,\mb,\Tp,\Tn\rangle_\RQ$ that satisfies all positive ($\Tp\cup\Tp'$) as well as all negative ($\Tn\cup\Tn'$) test cases.
\end{prob_def}\vspace{-4pt}
}}

\vspace{3pt}
%\end{framed}
%The solution KB found by Interactive Static KB Debugging 
\begin{remark}\label{rem:ad_int_static_KB_debug_problem}
The solution of an Interactive Static KB Debugging problem given the DPI $\langle\mo,\mb,\Tp,\Tn\rangle_\RQ$ constitutes a solution to the problem of KB Debugging (Problem Defnition~\ref{prob_def:onto_debug}) as well as to the problem of Parsimonious KB Debugging (Problem Defnition~\ref{prob_def:evidence_just}) \emph{for the original DPI} $\langle\mo,\mb,\Tp,\Tn\rangle_\RQ$, therefore the term ``static''.\qed
\end{remark}
Now, we give a more formal definition of a true diagnosis (an informal characterization of which was given in Section~\ref{sec:DiagnosisProbabilitySpace}). If sufficiently many new test cases are specified and added to a given DPI such that there is only one remaining minimal diagnosis w.r.t.\ the input DPI (the input DPI extended by the new test cases) left, then this diagnosis is referred to as the true diagnosis w.r.t.\ Interactive Static (Dynamic) KB Debugging.
\begin{definition}[True Diagnosis]\label{def:true_diagnosis}
Let $\dt$ be equal to $\md$ in Problem Definition~\ref{prob_def:minimize_user_interact_static} (\ref{prob_def:minimize_user_interact_dynamic}). Then $\dt$ is called the \emph{true diagnosis w.r.t.\ Interactive Static KB Debugging (Interactive Dynamic KB Debugging)}.
\end{definition}
%If a user wants to solve the Interactive Static (Dynamic) KB Debugging problem, we define the \emph{true diagnosis} (an informal characterization of which was given in Section~\ref{sec:DiagnosisProbabilitySpace}) as the value of $\md$ in Problem Definition~\ref{prob_def:minimize_user_interact_static} (\ref{prob_def:minimize_user_interact_dynamic}).

\chapter{User Interaction}
\label{chap:UserInteraction}
The idea in interactive KB debugging is to iteratively consult a user asking them to give additional information as regards desired and undesired entailments of the correct KB. Thus, the principle of interactive KB debugging is based on that of \emph{Sequential Diagnosis} which has been suggested by \cite{dekleer1987} as an iterative way to localize the faulty components (among an initially large set of possibilities) in malfunctioning digital circuits by performing repeated (most informative) measurements. We have shown in our previous works \cite{ksgf2010,Shchekotykhin2012} how sequential diagnosis can be applied to KBs (ontologies). 

In our approach, for the selection of which question (of a pool of possible ones) to ask a user next, an active learning~\cite{settles2012} approach is applied.\footnote{Note that the minimal a-posteriori expected entropy of solution candidate probabilities as a means to select the best next measurement as used in \cite{dekleer1987} is only one of many possible active learning strategies \cite{settles2012}.} \emph{Active Learning} is an iterative supervised machine learning technique in which a learning algorithm is able to interactively query the user to obtain a label for a desired unlabeled instance. In the case of a KB debugging system, an unlabeled instance is a set of logical formulas and the label is whether the conjunction of these formulas should or should not be entailed by the correct KB. Since the learner can choose the instances to be labeled, the number of consultations of an interacting user required to learn a concept (in this case the one solution KB with the desired semantics w.r.t.\ a given DPI) can often be much lower than the number required in a standard supervised learning setting since the risk that the algorithm must deal with lots of uninformative examples is reduced. 

We suppose the user of an interactive KB debugger to be a single person or multiple persons, usually experts of the particular domain the faulty KB is dealing with or authors of the faulty KB.
%the term ``user'' in this context can refer to, e.g., (one of) the person(s) who authored the KB $\mo$, some expert in the domain modeled by $\mo$ or some information extraction system.
Moreover, we 
%make the following assumptions about the user (cf.\ Chapter~\ref{chap:intro}):
assume the interacting user to be able to answer concrete queries about the intended domain that should be modeled. Otherwise put, we suppose that a user can classify a given logical formula (or a conjunction of logical formulas) as a wanted or unwanted proposition in the intended domain, i.e.\ as an entailment or non-entailment of the correct domain model. We have already argued in Chapter~\ref{chap:intro} why this assumption is plausible. 
%The idea in interactive KB debugging is to apply supervised learning and consult a user asking them to give additional information as regards desired and undesired entailments of the correct KB; the term ``user'' in this context can refer to, e.g., (one of) the person(s) who authored the KB $\mo$, some expert in the domain modeled by $\mo$ or some information extraction system.

%\noindent\textbf{Query.} 
\section{Queries}
\label{sec:Queries}
In interactive KB debugging, a set of logical formulas $Q$ is presented to the user who should decide whether to assign $Q$ to the set of positive ($\Tp$) or negative ($\Tn$) test cases w.r.t. a given DPI $\langle\mo,\mb,\Tp,\Tn\rangle_\RQ$. In other words, the system asks the user ``should the KB you intend to model entail all formulas in $Q$?''. In that, $Q$ is generated by the debugging algorithm in a way that \emph{any} decision of the user 
\begin{enumerate}
	\item invalidates at least one minimal diagnosis (\emph{search space restriction}) and
	\item preserves validity of at least one minimal diagnosis (\emph{solution preservation}).
\end{enumerate}
%(1)~invalidates at least one minimal diagnosis and (2)~preserves validity of at least one minimal diagnosis. 
%A test case $Q$ is said to \emph{invalidate} a diagnosis $\md$ iff $\md \in \minD_{\langle\mo,\mb,\Tp,\Tn\rangle_\RQ}$ and $\md \notin \minD_{\langle\mo,\mb,\Tp \cup \setof{Q},\Tn\rangle_\RQ}$
We call a set of logical formulas $Q$ with these properties a \emph{query}. Successive classification of queries as entailments (all formulas in $Q$ must be entailed) or non-entailments (at least one formula in $Q$ must not be entailed) of the correct KB enables gradual restriction of the search space for (minimal) diagnoses. Further on, classification of sufficiently many queries guarantees the detection of a single correct solution diagnosis which can be used to determine a solution KB with the correct semantics w.r.t.\ a given DPI.\footnote{Correctness of the diagnosis must not be understood as a guarantee that all formulas in the KB which are not in the diagnosis are definitely correct. Instead, correctness must be seen with regard to other diagnoses and with the ``Principle of Parsimony'' in mind (cf.\ Section~\ref{sec:MinimallyInvasiveOntologyDebugging}). That is, all other possible diagnoses are ruled out by a present set of test cases wherefore the single remaining diagnosis is the one that is correct (in comparison with all other incorrect ones). And, there is no evidence (at the time the correct diagnosis is found) that any other formulas in the KB might be faulty. This might change however after new formulas are added to the KB.} 

\begin{definition}[Query]\label{def:query}
Let $\langle\mo,\mb,\Tp,\Tn\rangle_\RQ$ over $\mathcal{L}$ and $\mD \subseteq \minD_{\langle\mo,\mb,\Tp,\Tn\rangle_\RQ}$. % a set of minimal diagnoses. 
Then a set of logical formulas $Q\neq\emptyset$ over $\mathcal{L}$ is called a \emph{query w.r.t.\ $\mD$} iff there are diagnoses $\md, \md' \in \mD$ such that $\md \notin \minD_{\langle\mo,\mb,\Tp \cup \setof{Q},\Tn\rangle_\RQ}$ and $\md'\notin\minD_{\langle\mo,\mb,\Tp,\Tn \cup \setof{Q}\rangle_\RQ}$. The set of all queries w.r.t.\ $\mD$ and $\langle\mo,\mb,\Tp,\Tn\rangle_\RQ$ is denoted by $\mQ_{\mD,\langle\mo,\mb,\Tp,\Tn\rangle_\RQ}$.
\end{definition}
\begin{remark}
%\emph{Remark on Definition~\ref{def:query}:} 
Although Definition~\ref{def:query} only postulates that at least one diagnosis in $\mD$ is invalidated for whatever answer is given to the query, this implies that, for each answer to the query, there is also a diagnosis that remains valid after adding the corresponding test case to the DPI, as will be shown by Proposition~\ref{prop:query_dx_dnx}.\qed
\end{remark}

So, w.r.t. a set of minimal diagnoses $\mD \subseteq \minD_{\langle\mo,\mb,\Tp,\Tn\rangle_\RQ}$, a query $Q$ is a set of logical formulas that rules out at least one diagnosis in $\mD$ (and therefore in $\minD_{\langle\mo,\mb,\Tp,\Tn\rangle_\RQ}$) as a candidate to formulate a solution KB, regardless of whether $Q$ is classified as a positive or negative test case.
%So, regardless of whether a query $Q$ is added to the positive or negative test cases, there is at least one diagnosis $\md\in\mD$ that gets invalidated.

%\noindent\textbf{Leading Diagnoses.} 
\section{Leading Diagnoses}
\label{sec:LeadingDiagnoses}
Query generation requires a precalculated set of minimal diagnoses $\mD \subseteq \minD_{\langle\mo,\mb,\Tp,\Tn\rangle_\RQ}$ 
%for the given DPI $\langle\mo,\mb,\Tp,\Tn\rangle_\RQ$ 
that serves as a representative for all minimal diagnoses $\minD_{\langle\mo,\mb,\Tp,\Tn\rangle_\RQ}$. As already mentioned, computation of the entire set $\minD_{\langle\mo,\mb,\Tp,\Tn\rangle_\RQ}$ is generally not tractable within reasonable time. Usually, $\mD$ is defined as a set of most probable or minimum cardinality diagnoses (cf.\ Chapter~\ref{chap:DiagnosisComputation}). Therefore, $\mD$ is called the set of \emph{leading diagnoses w.r.t. $\langle\mo,\mb,\Tp,\Tn\rangle_\RQ$} \cite{Shchekotykhin2012}. 
%A reasonable number of diagnoses in $\mD$

The leading diagnoses $\mD$ are then exploited to determine a query $Q$ the answering of which enables a discrimination between the diagnoses in $\minD_{\langle\mo,\mb,\Tp,\Tn\rangle_\RQ}$. That is, a subset of $\minD_{\langle\mo,\mb,\Tp,\Tn\rangle_\RQ}$ which is not ``compatible'' with the new information obtained by adding the test case $Q$ to $\Tp$ or $\Tn$ is ruled out (see Proposition~\ref{prop:dpi_update} below). For the computation of the subsequent query only a leading diagnoses set $\mD_{new}$ w.r.t.\ the minimal diagnoses still compliant with the new sets of test cases $\Tp'$ and $\Tn'$ is taken into consideration, i.e.\ $\mD_{new} \subseteq \mD_{\langle\mo,\mb,\Tp',\Tn'\rangle_\RQ}$.
% and not taken into consideration for further query generation actions 

The number of precomputed leading diagnoses $\mD$ affects the quality of the obtained query. The higher $|\mD|$, the more representative is $\mD$ w.r.t. $\minD_{\langle\mo,\mb,\Tp,\Tn\rangle_\RQ}$, the more options there are to specify a query in a way that a user can easily comprehend and answer it, and the higher is the chance that a query that eliminates a high rate of diagnoses w.r.t. $\mD$ will also eliminate a high rate of all minimal diagnoses $\minD_{\langle\mo,\mb,\Tp,\Tn\rangle_\RQ}$. The selection of a lower $|\mD|$ on the other hand means better timeliness regarding the interaction with a user, first because fewer leading diagnoses might be computed much faster and second because the search space for an ``optimal'' query is smaller.\footnote{Roughly, a query $Q$ is ``optimal'' if the number of queries that still need to be answered to identify the desired solution KB after $Q$ is added to the (positive or negative) test cases is minimal. ``Optimality'' of a query can be captured by quantitative information theoretic measures studied in the field of active learning \cite{settles2012} that can be used to estimate the quality of a query beforehand, i.e.\ before an answer to it is known. See Section~\ref{sec:query_selection_measures} and \cite{Rodler2013, ksgf2010, Shchekotykhin2012} for details.}
So, the optimal number of leading diagnoses depends on the complexity of the particular DPI considered. One way to determine a suitable $|\mD|$ can be to first define an interval $[n_{\min},n_{\max}]$ that must comprise $|\mD|$ where the upper bound defines the desired number of leading diagnoses and the lower bound the minimally postulated number. Second, the search for minimal diagnoses is run at least as long as it takes to compute $n_{\min}$ diagnoses and at the longest until $n_{\max}$ diagnoses have been found or a timeout $t$ expires that is specified in a manner it enables frequent user interaction. Note that such parameters have already been taken into account in the non-interactive KB debugging Algorithm~\ref{algo:hs} (see Section~\ref{sec:non_int_debug_procedure}).\label{etc:leading_diag_params}

%A query $Q_j$ induces a \emph{partition} of $\mD$ into the set $\dx{j}$ of diagnoses that ``predict'' the classification of $Q_j$ as a positive test case, the set $\dnx{j}$ of diagnoses that ``predict'' assignment of $Q_j$ to the negative test cases, and the set $\dz{j}$ comprising diagnoses in $\mD$ that ``predict'' neither $Q_j \in \Tp$ nor $Q_j \in \Tn$. In other words, 
%%this means that 
%if the true diagnosis is element of $\dx{j}$ ($\dnx{j}$), then $Q_j$ will be answered positively (negatively). Note that the opposite does not hold, e.g. a positive classification of $Q_j$ does not necessarily mean that the true diagnosis is an element of $\dx{j}$. This is because $\mD$ is usually a strict subset of $\minD_{\langle\mo,\mb,\Tp,\Tn\rangle_\RQ}$, whereby the true diagnosis can also be an element of $\minD_{\langle\mo,\mb,\Tp,\Tn\rangle_\RQ} \setminus \mD$.
%
%\begin{definition}[Partition]
%Let $Q_j$ be a query w.r.t. $\mD\subseteq \minD_{\langle\mo,\mb,\Tp,\Tn\rangle_\RQ}$ and $\mo^{*}_i$ be defined as $(\mo \setminus \md_i) \cup \mb \cup U_\Tp$ for $\md_i\in\mD$. Further, let 
%\begin{itemize}
%\item $\dx{j}:=\setof{\md_i \in \mD\,|\,\mo^{*}_i \models Q_j}$, 
%\item $\dnx{j}:=\setof{\md_i \in \mD\,|\,\exists r\in\RQ: \mo^{*}_i \cup Q_j \text{ violates } r}$, and 
%\item $\dz{j} := \mD \setminus (\dx{j} \cup \dnx{j})$. 
%\end{itemize}
%Then $\Pt_j := \langle \dx{j}, \dnx{j}, \dz{j} \rangle$ is called \emph{partition} of query $Q_j$.
%\end{definition}

%\noindent\textbf{Q-Partition.} 
\section{Q-Partitions}
\label{sec:QPartitions}
Now we introduce the notion of a \emph{q-partition}, a partition of the leading diagnoses set $\mD$ induced by a query w.r.t.~$\mD$. A q-partition will be a helpful instrument in deciding whether a set of logical formulas is a query or not. It will facilitate an estimation of the impact a query answer has in terms of invalidation of minimal diagnoses. And, given fault probabilities, it will enable us to gauge the probability of getting a positive or negative answer to a query. 

From now on, given a DPI $\langle\mo,\mb,\Tp,\Tn\rangle_\RQ$ and some minimal diagnosis $\md_i$ w.r.t.\ $\langle\mo,\mb,\Tp,\Tn\rangle_\RQ$, we will use the following abbreviation for the solution KB obtained by deletion of $\md_i$ along with the given background knowledge $\mb$:
\begin{align} 
\mo^{*}_i \; := \; (\mo \setminus \md_i) \cup \mb \cup U_\Tp \label{eq:sol_ont_candidate} 
\end{align}
\begin{definition}[q-Partition\footnote{In existing literature, e.g.\ \cite{Shchekotykhin2012,Rodler2013,ksgf2010}, a q-partition is often simply referred to as partition. We call it q-partition to emphasize that not each partition of $\mD$ into three sets is necessarily a q-partition.}]\label{def:q-partition}
Let $\langle\mo,\mb,\Tp,\Tn\rangle_\RQ$ be a DPI over $\mathcal{L}$, $\mD\subseteq \minD_{\langle\mo,\mb,\Tp,\Tn\rangle_\RQ}$. 
%and $\mo^{*}_i$ be defined as $(\mo \setminus \md_i) \cup \mb \cup U_\Tp$ for $\md_i\in\mD$. 
Further, let $Q$ be a set of logical formulas over $\mathcal{L}$ and
\begin{itemize}
\item $\dx{}(Q):=\setof{\md_i \in \mD\,|\,\mo^{*}_i \models Q}$, 
\item $\dnx{}(Q):=\setof{\md_i \in \mD\,|\,\exists x\in\RQ\cup\Tn: \mo^{*}_i \cup Q \text{ violates } x}$,
\item $\dz{}(Q) := \mD \setminus (\dx{j} \cup \dnx{j})$. 
\end{itemize}
Then $\langle \dx{}(Q), \dnx{}(Q), \dz{}(Q) \rangle$ is called a \emph{q-partition} iff $Q$ is a query w.r.t.\ $\mD$ and $\langle\mo,\mb,\Tp,\Tn\rangle_\RQ$.
\end{definition}
\begin{remark}\label{rem:dnx_contains_exactly_these_diagnoses...}
The set $\dnx{}(Q)$ contains exactly those diagnoses $\md_i\in\mD$ where $\mo \setminus \md_i$ is invalid w.r.t.\ $\tuple{\cdot,\mb,\Tp\cup\setof{Q},\Tn}$ (cf.~Definition~\ref{def:valid_onto}).\qed
\end{remark}
\begin{proposition}\label{prop:q-partition_is_partition}
For each query $Q$ w.r.t. some $\mD \subseteq \minD_{\langle\mo,\mb,\Tp,\Tn\rangle_\RQ}$ it holds that $\langle \dx{}(Q)$, $\dnx{}(Q)$, $\dz{}(Q) \rangle$ is a partition of $\mD$.
\end{proposition}
\begin{proof}
%This follows from the fact that each diagnosis is element of exactly one set of $\dx{j}, \dnx{j}, \dz{j}$. 
First, by definition of $\dz{}(Q)$, we have that $\dx{}(Q) \cup \dnx{}(Q) \cup \dz{}(Q) = \mD$, $\dx{}(Q) \cap \dz{}(Q) = \emptyset$ and $\dnx{}(Q) \cap \dz{}(Q) = \emptyset$. Second, $\dx{}(Q) \cap \dnx{}(Q) = \emptyset$ since $\mo^{*}_i \models Q_j$ and $\exists x\in\RQ\cup\Tn: (\mo^{*}_i \cup Q_j \text{ violates } x)$ imply by idempotency of $\mathcal{L}$ that $\mo^{*}_i$ violates some $x\in\RQ\cup\Tn$ which is a contradiction to $\md_i$ being a diagnosis w.r.t. $\langle\mo,\mb,\Tp,\Tn\rangle_\RQ$. Thus, each diagnosis in $\mD$ is an element of exactly one set of $\dx{}(Q), \dnx{}(Q), \dz{}(Q)$ which is equivalent to the statement of the proposition.
\end{proof}
\begin{remark}\label{rem:query_partitions_any_set_of_diagnoses_into_dx_dnx_dz}
In fact, Proposition~\ref{prop:q-partition_is_partition} holds for any set $\mD \subseteq \allD_{\langle\mo,\mb,\Tp,\Tn\rangle_\RQ}$, i.e.\ for any subset of \emph{all} diagnoses w.r.t.\ $\langle\mo,\mb,\Tp,\Tn\rangle_\RQ$. This can be easily seen from the proof of Proposition~\ref{prop:q-partition_is_partition} which does not require minimality of diagnoses. That is, any set of diagnoses w.r.t.\ a DPI is partitioned into the three sets $\dx{}(Q)$, $\dnx{}(Q)$ and $\dz{}(Q)$ as per Definition~\ref{def:q-partition} by a query $Q$ w.r.t.\ this DPI.\qed 
\end{remark}
\begin{proposition}\label{prop:unique_q-partition}
For each query $Q$ w.r.t. some $\mD \subseteq \minD_{\langle\mo,\mb,\Tp,\Tn\rangle_\RQ}$ there is one and only one partition $\langle \dx{}(Q), \dnx{}(Q), \dz{}(Q) \rangle$.
\end{proposition}
\begin{proof}
The existence of a partition $\dx{}(Q), \dnx{}(Q), \dz{}(Q)$ follows directly from Proposition~\ref{prop:q-partition_is_partition}. Assume there are two different partitions $\langle \dx{1}(Q), \dnx{1}(Q), \dz{1}(Q) \rangle$ and $\langle \dx{2}(Q), \dnx{2}(Q), \dz{2}(Q) \rangle$. Then, (a)~$\dx{1}(Q) \neq \dx{2}(Q)$ or (b)~$\dnx{1}(Q)\neq\dnx{2}(Q)$ or (c)~$\dz{1}(Q)\neq\dz{2}(Q)$ must hold. If (a) is true, then there is one diagnosis $\md_i \in \mD$ such that $\mo^{*}_{i} \models Q$ and $\mo^{*}_{i} \not\models Q$ -- a contradiction. If (b) is true, then there is one diagnosis $\md_i \in \mD$ such that $\mo^{*}_{i} \cup Q$ violates some $x\in\RQ\cup\Tn$ and $\mo^{*}_{i} \cup Q$ does not violate any $y\in\RQ\cup\Tn$ -- a contradiction. If (c) is true, then $(\dx{1}(Q)\cup\dnx{1}(Q)) \neq (\dx{2}(Q)\cup\dnx{2}(Q))$ which implies that either (a) or (b) must be true. 
\end{proof}
Due to the uniqueness of a q-partition $\langle \dx{}(Q), \dnx{}(Q), \dz{}(Q) \rangle$ for a query $Q$, we denote this q-partition by $\Pt(Q)$.
As a consequence of Definition~\ref{def:q-partition} and Proposition~\ref{prop:unique_q-partition}, a query $Q$ is a set of common entailments of KBs $\mo^{*}_i$, each resulting from the deletion of a single minimal diagnosis $\md_i \in \dx{}(Q)$ from $\mo$. 
\begin{corollary}
For each query $Q\in\mQ_{\mD,\langle\mo,\mb,\Tp,\Tn\rangle_\RQ}$ there is a set of minimal diagnoses $\dx{}(Q) \subseteq \minD_{\langle\mo,\mb,\Tp,\Tn\rangle_\RQ}$ as defined by Definition~\ref{def:q-partition} such that $Q \subseteq \setof{e \,|\,\forall \md_i \in \dx{}(Q): \mo^{*}_i \models e}$.
\end{corollary}

%\noindent\textbf{Interpretation of a Q-Partition.}\label{etc:interpretation_q-partition} 
\section{Interpretation of Q-Partitions}
\label{sec:InterpretationOfQPartitions}
Since $\mo^{*}_i$ corresponds to the solution KB (along with $\mb$) obtained under the assumption that $\dt = \md_i$, i.e.\ the true diagnosis (cf.\ Definition~\ref{def:true_diagnosis}) corresponds to $\md_i$, the sets $\dx{}(Q)$ and $\dnx{}(Q)$ can be interpreted as those leading diagnoses that predict the classification of $Q$ as a positive and negative test case, respectively. In other words, if the true diagnosis $\dt$ is in $\dx{}(Q)$, then the true solution KB $\mo^{*}_t$ entails $Q$ by Definition~\ref{def:q-partition}. Therefore the user will answer $Q$ positively (cf.\ Definition~\ref{def:true_diagnosis}). If, conversely, $\dt$ is in $\dnx{}(Q)$, then the true solution KB $\mo^{*}_t$ would be invalidated if $Q$ was answered positively, since $\mo^{*}_t \cup Q = (\mo\setminus\dt)\cup\mb\cup U_{\Tp\cup\setof{Q}}$ violates some $x\in\RQ\cup\Tn$ and thus $\mo\setminus\dt$ is invalid w.r.t.\ $\tuple{\cdot,\mb,\Tp\cup\setof{Q},\Tn}_\RQ$, which implies that $\dt$ is not a diagnosis w.r.t.\ $\tuple{\mo,\mb,\Tp\cup\setof{Q},\Tn}_\RQ$ according to Proposition~\ref{prop:validonto_diag}. Hence, the user will answer $Q$ negatively (cf.\ Definition~\ref{def:true_diagnosis}).
%the set $\dnx{j}$ of diagnoses that ``predict'' assignment of $Q_j$ to the negative test cases, and 
Diagnoses in $\dz{}(Q)$ on the other hand neither predict $Q \in \Tp$ nor $Q \in \Tn$. This means that we do not know how the user will answer a query $Q$ for which the true diagnosis $\dt$ is in $\dz{}(Q)$. In this case, for any answer to $Q$, the true diagnosis $\dt$ is in the set of minimal diagnoses w.r.t.\ the new DPI including $Q$ as a test case. To summarize: If the true diagnosis $\dt$ is an element of $\dx{}(Q)$ ($\dnx{}(Q)$), then $Q$ will be answered positively (negatively). 
%Note that the opposite does not hold, i.e.\ a positive (negative) classification of $Q_j$ does not necessarily mean that the true diagnosis is an element of $\dx{}(Q)$ ($\dnx{}(Q)$). This is because $\mD$ is usually a strict subset of $\minD_{\langle\mo,\mb,\Tp,\Tn\rangle_\RQ}$, whereby the true diagnosis can also be an element of $\minD_{\langle\mo,\mb,\Tp,\Tn\rangle_\RQ} \setminus \mD$.

Conversely, this means that a q-partition $\Pt(Q)$ gives a prior indication which leading diagnoses would be invalidated by a user's answer. Diagnoses in $\dx{}(Q)$ are invalidated by the classification $Q \in \Tn$, and diagnoses in $\dnx{}(Q)$ in case of $Q \in \Tp$. Diagnoses in $\dz{}(Q)$ can never be invalidated by an answer to $Q$. Thus, intuitively, queries with $\dz{}(Q) = \emptyset$ are preferable over other queries (as per the information provided by the set of leading diagnoses $\mD$) as the number of (definitely) eliminated diagnoses in $\minD_{\langle\mo,\mb,\Tp,\Tn\rangle_\RQ}$ should be maximized.

The following proposition is a direct consequence of Corollary~\ref{cor:notions_equiv} and explicates the impact of the addition of a test case to a DPI regarding the set of minimal diagnoses for this DPI.
\begin{proposition}\label{prop:dpi_update}
Let $Q$ be a query w.r.t. $\mD \subseteq \minD_{\langle\mo,\mb,\Tp,\Tn\rangle_\RQ}$ and let the answer of a user to $Q$ be $u(Q) \in \setof{\true,\false}$.

If $u(Q) = \true$, then $\md_i \in \minD_{\langle\mo,\mb,\Tp,\Tn\rangle_\RQ}$ is a diagnosis w.r.t.\ $\langle\mo,\mb,\Tp\cup\setof{Q},\Tn\rangle_\RQ$ iff $\mo\setminus \md_i$ is valid w.r.t.\ $\langle\cdot,\mb,\Tp\cup\setof{Q},\Tn\rangle_\RQ$. 

In other words, both of the following conditions must hold:
\begin{align*}
\forall r \in \RQ & \;:\; \mo_i^* \cup Q \; \emph{does not violate } r\\
\forall \tn \in \Tn & \;:\; \mo_i^* \cup Q \not\models \tn
\end{align*}

If $u(Q) = \false$, then $\md_i \in \minD_{\langle\mo,\mb,\Tp,\Tn\rangle_\RQ}$ is a diagnosis w.r.t.\ $\langle\mo,\mb,\Tp,\Tn\cup\setof{Q}\rangle_\RQ$ iff $\mo\setminus \md_i$ is valid w.r.t.\ $\langle\cdot,\mb,\Tp,\Tn\cup\setof{Q}\rangle_\RQ$. 

In other words, both of the following conditions must hold:
\begin{align*}
\forall r \in \RQ & \;:\; \mo_i^* \; \emph{does not violate } r\\
\forall \tn \in (\Tn \cup \setof{Q}) & \;:\; \mo_i^* \not\models \tn
\end{align*}
\end{proposition}
\begin{remark}\label{rem:invalidated_sets_of_q-partition_for_query_answer}
From Proposition~\ref{prop:dpi_update} and Definition~\ref{def:q-partition} it is easy to see that at least $\md_i \in \dnx{}(Q) \subset \minD_{\langle\mo,\mb,\Tp,\Tn\rangle_\RQ}$ are eliminated by a positive answer to $Q$. Namely, $\dnx{}(Q)$
%$u(Q)=true$ means that $Q$ is a positive test case of the new DPI and thus a subset of $U_\Tp$ which 
comprises exactly those diagnoses $\md_i$ that imply the violation of some $r\in\RQ$ or the entailment of some $\tn\in\Tn$ if $Q$ is added to $\mo_i^*$. On the other hand, at least $\md_i \in \dx{}(Q) \subset \minD_{\langle\mo,\mb,\Tp,\Tn\rangle_\RQ}$ are discarded if $u(Q) = \false$ as all diagnoses in $\dx{}(Q)$ entail $Q$ which must not be entailed. 

Note that, in general, the addition of a query to the test cases of a DPI causes not only an invalidation of some leading minimal diagnoses in $\mD$, but also the elimination of minimal diagnoses that have not even been computed yet. On the other hand, an added test case might also introduce new \emph{minimal} diagnoses, i.e.\ ones that were no minimal diagnoses before this test case was added. However, the newly obtained DPI after the addition of any new test case can only exhibit a reduced set of \emph{all} (i.e.\ minimal and non-minimal) diagnoses compared with the DPI before the test case was added (we will prove this result by Proposition~\ref{prop:diag_for_new_dpi_is_diag_for_old_dpi}). 
%Examples of such situations are given in Section~\ref{inter debug examples}.
\qed
\end{remark}

%\noindent\textbf{Relation between Query and Q-Partition.}\label{etc:relation_query_q-partition} 

\section[The Relation between a Query and Its Q-Partition]{The Relation between a Query and Its Q-Partition%
\sectionmark{Relation between Query and Q-Partition}}
\sectionmark{Relation between Query and Q-Partition}
\label{sec:TheRelationBetweenAQueryAndItsQPartition}
%%%%%%
%\section{The Relation between a Query and Its Q-Partition}
%\label{sec:TheRelationBetweenAQueryAndItsQPartition}
%%%%%%
The following proposition shows the relationship between a query and its q-partition and provides a criterion that enables to check whether a set of logical formulas is a query w.r.t.\ some set of leading diagnoses or not.
\begin{proposition}\label{prop:query_dx_dnx}
Let $\langle\mo,\mb,\Tp,\Tn\rangle_\RQ$ be a DPI over $\mathcal{L}$ and $\mD \subseteq \minD_{\langle\mo,\mb,\Tp,\Tn\rangle_\RQ}$. Then a set of logical formulas $Q\neq\emptyset$ over $\mathcal{L}$ is a query w.r.t. $\mD$ iff $\dx{}(Q) \neq \emptyset$ and $\dnx{}(Q)~\neq~\emptyset$.
\end{proposition}
\begin{proof}
``$\Leftarrow$'': If $\dx{}(Q) \neq \emptyset$ and $\dnx{}(Q) \neq \emptyset$ holds, then a non-empty set of diagnoses $\dnx{}(Q)$ ($\dx{}(Q)$) becomes invalid for positive (negative) answer to $Q$. So, $Q$ is a query.

``$\Rightarrow$'': If $Q$ is a query, then there are diagnoses $\md, \md' \in \mD$ such that $\md \notin \minD_{\langle\mo,\mb,\Tp \cup \setof{Q},\Tn\rangle_\RQ}$ and $\md'\notin\minD_{\langle\mo,\mb,\Tp,\Tn \cup \setof{Q}\rangle_\RQ}$. Consequently, $\md \in \mD \setminus \minD_{\langle\mo,\mb,\Tp \cup \setof{Q},\Tn\rangle_\RQ}$ and $\md' \in \mD \setminus \minD_{\langle\mo,\mb,\Tp,\Tn \cup \setof{Q}\rangle_\RQ}$ holds. But, as the diagnoses in $\mD \setminus \minD_{\langle\mo,\mb,\Tp \cup \setof{Q},\Tn\rangle_\RQ}$ are exactly the diagnoses in $\mD$ that become invalid by the positive answer to $Q$, we obtain $\md\in\dnx{}(Q)$. The argumentation for $\md' \in \dx{}(Q)$ is analogous. Hence, $\dx{}(Q) \neq \emptyset$ and $\dnx{}(Q) \neq \emptyset$.  
\end{proof}
%%%%%%%%%%%%%%%%%%%%%%%%%%%%%%%%%%%%%%%%%%%%%%%%%%%%%%
\begin{corollary}\label{cor:q-partition_dx_dnx}
Let $\mD\subseteq \minD_{\langle\mo,\mb,\Tp,\Tn\rangle_\RQ}$. Then, for each q-partition $\Pt(Q) = \langle \dx{}(Q), \dnx{}(Q), \dz{}(Q)\rangle$ w.r.t. $\mD$ it holds that $\dx{}(Q) \neq \emptyset$ and $\dnx{}(Q)~\neq~\emptyset$.
\end{corollary}
\begin{proof}
Follows from Definition~\ref{def:q-partition} which grants the existence of a query for any q-partition and Proposition~\ref{prop:query_dx_dnx} which states that neither $\dx{}(Q)$ nor $\dnx{}(Q)$ must be empty sets for any query.
\end{proof}
So, by Proposition~\ref{prop:query_dx_dnx}, a query not only eliminates at least one leading diagnosis, but also leaves at least one leading diagnosis valid. Therefore, an admissible DPI can never get non-admissible by adding a query to the positive or negative test cases.
\begin{corollary}\label{cor:query_leaves_valid_diag}
Let $\langle\mo,\mb,\Tp,\Tn\rangle_\RQ$ be an admissible DPI, $\mD \subseteq \minD_{\langle\mo,\mb,\Tp,\Tn\rangle_\RQ}$ and $Q\in\mQ_{\mD,\langle\mo,\mb,\Tp,\Tn\rangle_\RQ}$. Then $\langle\mo,\mb,\Tp\cup\setof{Q},\Tn\rangle_\RQ$ as well as $\langle\mo,\mb,\Tp,\Tn\cup\setof{Q}\rangle_\RQ$ are admissible DPIs.
\end{corollary}
\begin{proof}
Assume that $\langle\mo,\mb,\Tp\cup\setof{Q},\Tn\rangle_\RQ$ is non-admissible. Then there is no valid diagnosis for this DPI. Since $\langle\mo,\mb,\Tp,\Tn\rangle_\RQ$ is an admissible DPI, this means that $Q$ invalidates each diagnosis $\md\in\allD_{\langle\mo,\mb,\Tp,\Tn\rangle_\RQ} \supseteq \minD_{\langle\mo,\mb,\Tp,\Tn\rangle_\RQ}\supset\mD$. By Proposition~\ref{prop:query_dx_dnx}, this is a contradiction to the fact that $Q$ is a query. The argumentation for $\langle\mo,\mb,\Tp,\Tn\cup\setof{Q}\rangle_\RQ$ is analogue.
\end{proof}
This means in particular that a query can never contain a conflict set or result in a violation of some requirement $r\in\RQ$ when added to $\mb\cup U_\Tp$ (cf. Proposition~\ref{prop:exist_diag}).

\section{Existence of Queries}
\label{sec:ExistenceOfQueries}
For any set of at least two leading minimal diagnoses the existence of a query is guaranteed, as the next proposition and corollary show. 
%In fact, $|\mD|$ is a lower bound for the number of queries w.r.t. $\mD$.
In particular, this implies that for arbitrary two minimal diagnoses $\md,\md'$ w.r.t. a DPI there is a query $Q$ that enables to differentiate between $\md$ and $\md'$, i.e.\ exactly one of these diagnoses is invalidated by each answer to $Q$.
%So far we know that a set of leading diagnoses $\mD$ is required to compute a query. 
%But do we need to take care which minimal diagnoses are taken into account when constructing the set $\mD$ to guarantee existence of a query w.r.t. $\mD$? The answer is no, as the next proposition and corollary show.
\begin{proposition}\label{prop:q1}
Let $\mD \subseteq \minD_{\langle\mo,\mb,\Tp,\Tn\rangle_\RQ}$ with $|\mD|\geq 2$ and $U_\mD$ be the union of all diagnoses in $\mD$. Then 
\begin{enumerate}[(I)]
	\item $Q:=(U_{\mD}\setminus \md_i)$ is a query w.r.t. $\mD$ for arbitrary $\md_i\in\mD$ and
	\item $\Pt(Q)=\langle\setof{\md_i},\mD\setminus\setof{\md_i},\emptyset\rangle$.
\end{enumerate}
%\textbf{(I)}~$Q:=(U_{\mD}\setminus \md_i)$ is a query w.r.t. $\mD$ for arbitrary $\md_i\in\mD$ and \textbf{(II)}~$\Pt(Q)=\langle\setof{\md_i},\mD\setminus\setof{\md_i},\emptyset\rangle$.
\end{proposition}
\begin{proof}
\textbf{Ad (I):} Assume that $Q$ is not a query. Then either (1)~$Q = \emptyset$ or (2)~$\dx{}(Q) = \emptyset$ or (3)~$\dnx{}(Q) = \emptyset$. In the following we prove that neither (1) nor (2) nor (3) can hold.

(1): $Q = \emptyset$ means that $\md_i \supseteq U_{\mD}$. Since any diagnosis $\md$ in $\mD$ is a subset of $U_{\mD}$, this implies that for each $\md \in \mD$, $\md \subseteq \md_i$ holds. As $|\mD| \geq 2$ is assumed, there is a $\md_k \neq \md_i \in \mD$ for which this property holds. This, however, is a contradiction to the 
%subset-minimality of diagnoses, in particular of $\md_i$. 
minimality of diagnosis $\md_i$.

(2): $\dx{}(Q) = \emptyset$ cannot hold, since $(\mo \setminus \md_i) \supseteq (U_{\mD}\setminus \md_i)$ and $U_{\mD}\setminus \md_i \models Q$ by monotonicity of description logics imply that $\mo^{*}_i = (\mo \setminus \md_i) \cup \mb \cup U_P \models Q$. Hence, there is at least one diagnosis, namely $\md_i$, in $\dx{}(Q)$.

(3): To prove that $\dnx{}(Q) \neq \emptyset$, we must show that there is a diagnosis $\md\in\mD$ such that $Y:=(\mo\setminus\md) \cup \mb \cup U_P \cup Q = (\mo\setminus\md) \cup \mb \cup U_P \cup (U_{\mD}\setminus \md_i)$ is incoherent. 
However, $(\mo\setminus\md) \cup (U_{\mD}\setminus \md_i) = \mo \setminus (\md\cap\md_i)$ by distributive and De Morgan laws which yields $Y = \mo \setminus (\md\cap\md_i)\cup \mb \cup U_P$.
%Let $Z:=Y\setminus [\mo\setminus (U_\mD\setminus \md)] = (Y\setminus \mo) \cup (U_\mD \setminus \md) = (U_{\mD}\setminus\md) \cup \mb \cup U_P \cup (U_{\mD}\setminus \md_i) = [U_{\mD}\setminus(\md\cap\md_i)] \cup \mb \cup U_P$. 
But, $\md\cap\md_i \subset \md$ must hold as $\md \not\subseteq \md_i$ by the subset-minimality of $\md_i$ whereby $\md$ must comprise a formula $\tax \notin \md_i$. Hence, $Y \supset (\mo\setminus\md) \cup \mb \cup U_P$ is incoherent by subset-minimality of $\md$.

\textbf{Ad (II):} We already know that $\md_i \in \dx{}(Q)$ by (2). Since $\md\in\mD$ in (3) can be chosen arbitrarily, we obtain that $\md\in\dnx{}(Q)$ for all diagnoses $\md\in\mD\setminus\setof{\md_i}$.
%(3): To prove that $\dnx{}(Q) \neq \emptyset$, we must show that there is a diagnosis $\md\in\mD$ such that $Y:=(\mo\setminus\md) \cup \mb \cup U_P \cup Q = (\mo\setminus\md) \cup \mb \cup U_P \cup (U_{\mD}\setminus \md_i)$ is incoherent. Due to monotonicity of description logics it suffices to demonstrate that a subset of $Y$, e.g.~$Z:=(U_{\mD}\setminus\md) \cup \mb \cup U_P \cup (U_{\mD}\setminus \md_i) = (U_{\mD}\setminus(\md\cap\md_i)) \cup \mb \cup U_P$, is incoherent. But, $\md\cap\md_i \subset \md$ must hold as $\md \not\subseteq \md_i$ by the subset-minimality of $\md_i$ whereby $\md$ must comprise an formula $\tax \notin \md_i$. Hence, $Z \supset (\mo\setminus\md) \cup \mb \cup U_P$ which means that $Z$ is incoherent by subset-minimality of $\md$.
\end{proof}
We immediately obtain a lower bound for the number of queries by Proposition~\ref{prop:q1}:
\begin{corollary}\label{cor:query_num_lower_bound}
%There is a query for any set of leading diagnoses $\mD$ with $|\mD|>1$.
Let $\mD \subseteq \minD_{\langle\mo,\mb,\Tp,\Tn\rangle_\RQ}$ with $|\mD|>1$. Then a lower bound for the number of queries w.r.t. $\mD$ is $|\mD|$.
\end{corollary}
\begin{remark}\label{rem:existence_of_queries_requires_minimal_leading_diagnoses}
Notice that the preceding proposition and corollary require a set of \emph{minimal} diagnoses. This means that subset-minimality of diagnoses is a necessary prerequisite for guaranteeing the possibility of discrimination between diagnoses. In other words, interactive debugging by means of (some or only) non-minimal diagnoses cannot be proven to work correctly (without making any further assumptions).\qed
\end{remark} 

\chapter{Query Generation}
\label{chap:QueryGeneration}
In this chapter we want to describe, discuss and prove the correctness of methods for the generation of queries which takes place in each iteration of an interactive KB debugging algorithm after a set of leading diagnoses has been determined.
% w.r.t.\ a set of leading diagnoses and a DPI. 
With Algorithm~\ref{algo:query_gen}, similar versions of which can be found in~\cite{Shchekotykhin2012, Rodler2013}, we present a way to compute a pool $\QP$ of queries and associated q-partitions w.r.t.\ a set of leading diagnoses $\mD$ and a DPI $\tuple{\mo,\mb,\Tp,\Tn}_\RQ$. The generation of this pool $\QP$ is the first stage of the query computation function used in the interactive debugging algorithm (Algorithm~\ref{algo:inter_onto_debug}) presented below. In a second stage, one particular query that meets certain criteria such as maximum expected information gain is selected from $\QP$ (see Section~\ref{sec:query_selection_measures}).
%a query in two stages: First, a pool of queries and associated q-partitions is generated and stored in a set $\QP$, and second, one particular query that meets certain criteria such as maximum expected information gain is selected from $\QP$.

Before we give a description of Algorithm~\ref{algo:query_gen}, let us have a look at some example by which we want to demonstrate the principle how a query w.r.t.\ some set of leading diagnoses for a DPI can be constructed. This should give the reader a first idea and an intuition of how the presented algorithm works.

\renewcommand{\arraystretch}{1.4} 
\begin{table*}
\footnotesize
	\centering
		\rowcolors[]{2}{gray!8}{gray!16} %\arrayrulecolor{black}
		%\extrarowheight10pt
		\begin{tabular}{ c c c c	} 
		%\hline\hline
			\rowcolor{gray!40}
			\toprule\addlinespace[0pt]
			$i$ & $\tax_i$ & $\mo$ & $\mb$  \\ \addlinespace[0pt]\midrule\addlinespace[0pt]
			%\hline
			1 & $\forall X a_1(X) \;\rightarrow\; a_2(X) \land m_1(X) \land m_2(X)$ & $\bullet$ & 	\\
			%\hline
			2 & $\forall X a_2(X) \;\rightarrow\; \lnot(\exists Y s(X,Y) \land m_3(Y)) \land \exists Z s(X,Z) \land m_2(Z)$ & $\bullet$ &  	\\
			%\hline
			3 & $\forall X m_1(X) \;\rightarrow\; \lnot a(X) \land b(X)$ & $\bullet$ &  	\\
			%\hline
			4 & $\forall X m_2(X) \;\rightarrow\; (\forall Y s(X,Y) \rightarrow a(Y)) \land d(X)$ & $\bullet$ & 	\\
			%\hline
			5 & $\forall X m_3(X) \;\leftrightarrow\; b(X) \lor c(X)$ & $\bullet$ & 	\\
			%\hline
			6 & $a_1(w)$ &  & $\bullet$   \\
			%\hline
			7 & $a_1(u)$ &  & $\bullet$  	\\
			%\hline
			8 & $s(u,w)$ &  & $\bullet$  \\ 
			%\hline\hline
			\addlinespace[0pt]\bottomrule
			\rowcolor{gray!40}
			$i$ & \multicolumn{3}{c}{$\tp_i\in\Tp$} \\ \addlinespace[0pt]\midrule\addlinespace[0pt]
			$\times$ & \multicolumn{3}{c}{$\times$} 	\\ \addlinespace[0pt]\toprule\addlinespace[0pt]
			\rowcolor{gray!40}
			$i$ & \multicolumn{3}{c}{$\tn_i\in\Tn$} \\ \addlinespace[0pt]\midrule\addlinespace[0pt]
			$\times$ & \multicolumn{3}{c}{$\times$} 	\\ \addlinespace[0pt]\toprule\addlinespace[0pt]
			\rowcolor{gray!40}
			$i$ & \multicolumn{3}{c}{$r_i\in\RQ$} \\ \addlinespace[0pt]\midrule\addlinespace[0pt]
			1 & \multicolumn{3}{c}{consistency} \\ %\addlinespace[0pt]\midrule\addlinespace[0pt]
			2 & \multicolumn{3}{c}{coherency} \\ \addlinespace[0pt]\bottomrule
		\end{tabular}
	\caption{First-Order Logic Example DPI}
	\label{tab:example1}
\end{table*}

\begin{example}\label{example:query_computation}
Consider the example FOL DPI given by Table~\ref{tab:example1}. The set of minimal conflict sets $\minC_{\tuple{\mo,\mb,\Tp,\Tn}_\RQ} = \setof{\mc_1,\mc_2} = \setof{\tuple{1,3,4},\tuple{1,2,3,5}}$ (like in previous examples, formulas $\tax_i$ in Table~\ref{tab:example1} are sometimes referred to just by their number $i$ if it is clear from the context what is meant). Let the set of leading diagnoses be the set of all minimal diagnoses, i.e.\ $\mD = \minD_{\tuple{\mo,\mb,\Tp,\Tn}_\RQ} = \setof{\md_1,\md_2,\md_3,\md_4} = \setof{[1],[3],[4,5],[2,4]}$. To enable a better understanding of this example, we first analyze why $\mc_1$ and $\mc_2$ are minimal conflict sets w.r.t.\ $\tuple{\mo,\mb,\Tp,\Tn}_\RQ$.

Why is $\mc_1$ a conflict set w.r.t.\ $\tuple{\mo,\mb,\Tp,\Tn}_\RQ$? In the following we underline the formulas $\tax_i$ and relevant parts of these formulas used in the derivation of the conflict set. First, there is the background KB $\mb$ including $\underline{a_1(w)}$ and $a_1(u)$. Due to $\underline{\tax_1}$, by substitution of $X$ by $w$ (written as $X/w$), we obtain $a_2(w), \underline{m_1(w)}$ and $m_2(w)$ from $a_1(w)$. Likewise, we can derive $a_2(u), m_1(u)$ and $\underline{m_2(u)}$ from $a_1(u)$ by $X/u$. Substituting $X$ by $w$ in $\underline{\tax_3}$ yields $\underline{m_1(w) \rightarrow \lnot a(w)} \land b(w)$. Thus, we obtain $\underline{\lnot a(w)}$. A substitution of $X$ by $u$ in $\underline{\tax_4}$ results in $m_2(u) \rightarrow (\forall Y s(u,Y) \rightarrow a(Y)) \land d(u)$. By $Y/w$, we have $\underline{m_2(u) \rightarrow (s(u,w) \rightarrow a(w))} \land d(u)$. Since $m_2(u)$ has already been deduced from the background formula $a_1(u)$ and $\underline{s(u,w)}$ is a background formula as well, we can conclude $\underline{a(w)}$ from $\tax_4$. All in all, we have derived $\lnot a(w)$ and $a(w)$, i.e.\ an inconsistency, by means of $\mb$ and $\mc_1$ (and $U_\Tp$ which is the empty set) wherefore $\mc_1$ is a conflict set w.r.t.\ $\tuple{\mo,\mb,\Tp,\Tn}_\RQ$ by Definition~\ref{def:cs}. The minimality of $\mc_1$ can be easily verified by the way we derived that it is a conflict set; namely, leaving out any of the formulas $\tax_1$, $\tax_3$ or $\tax_4$ does not allow to derive an inconsistency or incoherency (note that the set of negative test cases $\Tn$ is empty). 

Why is $\mc_2$ a conflict set w.r.t.\ $\tuple{\mo,\mb,\Tp,\Tn}_\RQ$? We argue as follows to deduce the inconsistency responsible for $\mc_2$ to be a conflict set (the relevant implications and used formulas are again underlined):
\begin{align*}
(1):\,a_1(w) \in \mb:&\quad \underline{a_1(w)}  \\
(2):\,X/w \mbox{ in } \underline{\tax_1}:&\quad  \underline{a_1(w) \;\rightarrow}\; a_2(w) \land \underline{m_1(w)} \land m_2(w) \\
(3):\,X/w \mbox{ in } \underline{\tax_3}:&\quad  \underline{m_1(w) \;\rightarrow}\; \lnot a(w) \land \underline{b(w)} \\
(4):\,\underline{\tax_5} \mbox{ and } X/w:& \quad \underline{b(w) \;\rightarrow\; m_3(w)} \\
(5):\,(1) - (4):& \quad \underline{m_3(w)} \\
(6):\,a_1(u) \in \mb:&\quad \underline{a_1(u)}  \\
(7):\,X/u \mbox{ in } \underline{\tax_1}:&\quad  \underline{a_1(u) \;\rightarrow\; a_2(u)} \land m_1(u) \land m_2(u) \\
(8):\,X/u \mbox{ in } \underline{\tax_2}:&\quad  \underline{a_2(u) \;\rightarrow\; \lnot (\exists Y s(u,Y) \land m_3(Y))} \\ 											 &\quad	\land (\exists Z s(u,Z) \land m_2(Z))  \\
(9):\,(6) - (8):& \quad \underline{\lnot (\exists Y s(u,Y) \land m_3(Y))} \\
(10):\,s(u,w) \in \mb:&\quad \underline{s(u,w)}  \\
(11):\,(5)\mbox{ and } (10):&\quad 	\underline{\exists Y s(u,Y) \land m_3(Y)}	\\
(9)\mbox{ and } (11):&\quad \mbox{\Lightning} \quad\qed
\end{align*}
Minimality of $\mc_2$ can again be verified by observing that, given any formula of $\mc_2$ is left out, no inconsistency or incoherency can be derived.
% and trying to derive that is again obvious from the argumentation why it is a conflict set.

Now we show how to construct a query manually. As suggested by Definition~\ref{def:q-partition} and Proposition~\ref{prop:query_dx_dnx} and discussed in Section~\ref{sec:TheRelationBetweenAQueryAndItsQPartition},
%\textbf{Relation between Query and Q-Partition} on page~\pageref{etc:relation_query_q-partition}, 
an obvious way of generating a query w.r.t.\ $\mD$ and $\tuple{\mo,\mb,\Tp,\Tn}_\RQ$ is via the notion of a q-partition. Definition~\ref{def:q-partition} states that $Q$ is a set of common entailments of KBs $\mo_i^*$ (Formula~\ref{eq:sol_ont_candidate}) where $\md_i \in \dx{}(Q)$, a subset of $\mD$. Hence, a first step towards query computation is to choose some non-empty subset $\mS$ of the leading diagnoses $\mD$ which we will call the \emph{seed} for query generation. For our manual construction, let $\mS = \setof{\md_3,\md_4} = \setof{[4,5],[2,4]}$. For each of the diagnoses $\md_i$ in $\mS$, we assemble the KB $\mo_i^*$ and use a reasoning engine to obtain a set of entailments $E_{\md_i}$ of $\mo_i^*$. For $\md_3$ we obtain $\mo_3^* := \setof{1,2,3,4,5}\setminus\setof{4,5} \cup \setof{6,7,8} \cup \setof{} = \setof{1,2,3,6,7,8}$. Similarly, we compute $\mo_4^* = \setof{1,3,5,6,7,8}$. 

Suppose that the reasoner invoked by the used \textsc{getEntailments} function produces only entailments of the type $\forall X p_1(X) \rightarrow p_2(X)$ for predicate names $p_1, p_2$ and of the type $p(a)$ where $p$ is a predicate name and $a$ is a constant (cf.\ Remark~\ref{rem:entailment_computation_finite_types_of_entailments}). For this purpose, DL and OWL reasoners, respectively, such as Pellet \cite{sirin2007pellet}, HermiT \cite{Shearer2008}, FaCT++ \cite{Tsarkov06} or KAON2\footnote{\url{http://kaon2.semanticweb.org/}} could be used with their classification and realization reasoning services. The reason why this is possible can be realized after a short analysis of the DPI $\tuple{\mo,\mb,\Tp,\Tn}_\RQ$ given by Table~\ref{tab:example1}. For, this DPI can be translated to DL similarly as demonstrated in Example~\ref{example:FOL_to_DL}. All the mentioned reasoners can deal with the expressivity of the resulting DL language.
%\fixme{cite DL-reasoners here and give explanation according to thesis horridge why such entailment types are relevant} 

Then, we obtain the sets $E_{\md_3}$ and $E_{\md_4}$, i.e.\ the sets of entailments of $\mo_3^*$ and $\mo_4^*$, respectively, as depicted by Table~\ref{tab:example:query_construction_entailments}. The set of common entailments $Q$, i.e.\ $Q = E_{\md_3} \cap E_{\md_4}$ is then the set containing all elements in the rows of Table~\ref{tab:example:query_construction_entailments} that are above the dashed line. 

Notice at this point that the set $\setof{a_1(w),a_1(u),s(u,w)} = \mb$ does not need to be computed or, respectively, included in $Q$ since none of these formulas can serve to discriminate between diagnoses (which is the only aim of a query). The simple reason for this is that $\mo_i^*$ for \emph{each} $\md_i \in \mD$ comprises these formulas and thus each $\mo_i^*$ entails these formulas by the extensiveness of FOL (cf.\ Chapter~\ref{chap:basics}). Since entailed by each potential solution KB $\mo_i^*$, these formulas cannot yield a violation of any requirements or test cases since none of the KBs $\mo_i^*$ violates any requirements or test cases (follows from Definitions~\ref{def:diagnosis} and \ref{def:target_ont}).   

Continuing with our query construction, we know by Proposition~\ref{prop:query_dx_dnx} that $Q$ is a query w.r.t.\ $\mD$ and $\tuple{\mo,\mb,\Tp,\Tn}_\RQ$ iff $\dx{}(Q) \neq \emptyset$ and $\dnx{}(Q) \neq \emptyset$. Whereas it is trivial that the former condition is met since $\dx{}(Q)$ contains (at least) the two diagnoses $\md_3$ and $\md_4$ that we used to compute $Q$ (cf.\ Definition~\ref{def:q-partition}), we still need to verify whether the latter condition is actually satisfied for $Q$. To this end, as per Definition~\ref{def:q-partition}, we must simply find some diagnosis $\md_j$ in $\mD \setminus \mS = \setof{\md_1,\md_2,\md_3,\md_4} \setminus \setof{\md_3,\md_4} = \setof{\md_1,\md_2}$ such that $\mo_j^* \cup Q$ violates some $x\in\Tn\cup\RQ$, i.e.\ whether some negative test case is entailed or whether this KB is incoherent or inconsistent. So, we start with $\md_1$, i.e.\ we examine $(\mo\setminus\md_1) \cup \mb \cup \Tp \cup Q = \setof{1,2,3,4,5}\setminus \setof{1} \cup \setof{6,7,8} \cup \setof{} \cup Q = \setof{2,3,4,5,6,7,8} \cup Q$. 

And, indeed, we are able to prove an inconsistency for this KB. To see that, verify that by $X/w$ in $e_2\in Q$ (see Table~\ref{tab:example:query_construction_entailments}) and $a_1(w) = \tax_6 \in \mo_1^*$ we can derive $m_1(w)$ which lets us conclude $\lnot a(w)$ by the substitution of $X$ by $w$ in $\tax_3\in \mo_1^*$. On the other hand, we obtain $a(w)$ by $X/u$ in $e_3 \in Q$, $\setof{X/u,Y/w}$ in $\tax_4 \in \mo_1^*$ and $s(u,w) = \tax_8 \in \mo_1^*$ as shown in the explanation for conflict set $\mc_1$ above. Thus, $\md_1 \in \dnx{}(Q)$.

That is, we have just proven that $Q$ is de facto a query w.r.t.\ $\mD$ and $\tuple{\mo,\mb,\Tp,\Tn}_\RQ$. And this, although we have not yet assigned each leading diagnosis to the respective set of the q-partition of $Q$. In a situation where just any query shall be asked to the user, this would suffice, and the query could be presented to the interacting user.

However, in case a ``best'' query according to some criterion shall be determined from a set of different competing queries, usually the computation of the full q-partition of each competing query is required. This is due to the fact that the q-partition provides information about several properties of queries that are considered by common query selection techniques (for details see Section~\ref{sec:query_selection_measures}). So, let us complete the q-partition for our query $Q$ by investigating $\mo_2^* \cup Q = \setof{1,2,4,5,6,7,8} \cup Q$. Also in this case we can derive an inconsistency which can be easily realized by reconsidering the argumentation why $\mc_2$ is a conflict set above and by using $e_4 \in Q$ instead of $\tax_3 \notin \mo_2^* \cup Q$. That means, the final q-partition $\Pt(Q)$ for $Q$ is given by $\tuple{\setof{\md_3,\md_4},\setof{\md_1,\md_2},\emptyset}$.

The next question that arises directly from the proofs that $\md_3, \md_4 \in \dnx{}(Q)$ is whether there is a (set-minimal) subset $Q_{\min}$ of $Q$ such that $Q_{\min}$ preserves the discrimination properties of $Q$, i.e.\ the q-partition $\Pt(Q_{\min}) = \Pt(Q)$. In fact, the answer is yes for the query $Q$ we computed, but also for the majority of other cases. This is a simple consequence of using the reasoning engine as a black-box which suggests a strategy we pursued in our query construction which relies on a precomputation of entailments and a final minimization part. Sticking to this black-box concept however does not allow to use some customized reasoning procedure that pointedly returns a set of common entailments $Q$ for a set of diagnoses $\mS \subset \mD$ where all formulas in $Q$ are necessary for a requirement or test case violation, respectively, of KBs $\mo_j^*$ for diagnoses in $\mD \setminus \mS$.

What militates for such a black-box approach is the generality and independence of a particular logic (for which an adequate glass-box reasoner exists), the easier implementation of the debugging system and potential performance issues with a glass-box approach~\cite{kalyanpur2005}. For a black-box algorithm to work, only a reasoner implementing a sound and complete inference procedure for the used logic $\mathcal{L}$ must be available.

In general, there is more than one minimized version of a query that preserves the q-partition. Theoretically, the number of such minimal queries w.r.t.\ one q-partition can be exponential in the size of the initially computed query that is provided as an input to the minimization procedure. For our query $Q$, for instance, 
%$Q_{\min,1}=\setof{a_2(u),b(w)} = \setof{e_7,e_{12}}$, $Q_{\min,2} =\setof{ \forall X a_1(X)\rightarrow a_2(X),b(w)} = \setof{e_1,e_{12}}$, $Q_{\min,3} =\setof{ \forall X a_1(X)\rightarrow a_2(X)$, $\forall X a_1(X)\rightarrow m_1(X)$, $\forall X m_1(X)\rightarrow b(X)}$ $= \setof{e_1,e_2,e_4}$ and $Q_{\min,4} = \setof{ \forall X a_1(X)\rightarrow m_1(X)$, $\forall X a_1(X)\rightarrow m_2(X)$, $\forall X m_1(X)\rightarrow b(X)}$ $= \setof{e_2,e_3,e_4}$ 
\begin{align*}
Q_{\min,1}=\{&a_2(u),b(w)\} = \{e_7,e_{12}\}, \\
Q_{\min,2}=\{&\forall X a_1(X)\rightarrow a_2(X),b(w)\} = \setof{e_1,e_{12}}, \\
Q_{\min,3}=\{ &\forall X a_1(X)\rightarrow a_2(X),\\
					    &\forall X a_1(X)\rightarrow m_1(X), \\ 					
					    &\forall X m_1(X)\rightarrow b(X)\}  
					= \{e_1,e_2,e_4\} \quad\mbox{   and} \\ 
Q_{\min,4}= \{& \forall X a_1(X)\rightarrow m_1(X), \\
									 &\forall X a_1(X)\rightarrow m_2(X), \\				 
									 & \forall X m_1(X)\rightarrow b(X)\} 
									= \{e_2,e_3,e_4\} 
\end{align*}
are set-minimal, q-partition preserving subqueries. Namely, each of the sets $Q_{\min,1}$, $Q_{\min,2}$ and $Q_{\min,3}$ together with $\setof{2,5,6,7,8}$ implies an inconsistency since $m_3(w)$ and $\lnot m_3(w)$ can be derived and $\setof{2,5,6,7,8}\subseteq \mo_1^*$ and $\setof{2,5,6,7,8}\subseteq \mo_2^*$. $\setof{e_2,e_3} \subset Q_{\min,4}$ yields an inconsistency when added to $\mo_1^*$, i.e.\ $a(w)$ and $\lnot a(w)$ are entailed, and $\setof{e_4} \subset Q_{\min,4}$ merged with $\mo_2^*$ yields an inconsistency, i.e.\ the derivation of $m_3(w)$ and $\lnot m_3(w)$. In order not to overwhelm the user we would of course ask them such a minimized version of a query rather than the full query that contains plenty of irrelevant formulas.

An example of a seed $\mS$ that does not lead to the discovery of a query is $\mS = \setof{\md_1,\md_2,\md_3}$ since the set of common entailments $E_{\md_1}\cap E_{\md_2}\cap E_{\md_3} = \emptyset$. Note that this holds when all $E_{\md_i}$ contain only entailments of the types we specified above. For other types of entailments, i.e.\ a different specification of the \textsc{getEntailments} function, this might no longer hold.
%
%------- difference if entailments of form $p(a)$ where this is a positive or negative literal are included --> then $\lnot m_3(w)$ and $\lnot a(u)$ and $\lnot a(w)$ are included in $Q$.
%
%-------- difference if only explicit entailments are used. 
\qed
\end{example}
%Then, the set $E_{\md_3}$, i.e.\ the entailments produced for $\mo_3^*$ is $\{\forall X a_1(X)\rightarrow a_2(X)$, $\forall X a_1(X)\rightarrow m_1(X)$, $\forall X a_1(X)\rightarrow m_2(X)$, $\forall X m_1(X)\rightarrow b(X)$, $\forall X a_1(X) \rightarrow b(X)$, $a_2(w)$, $a_2(u)$, $m_1(w)$, $m_1(u)$, $m_2(w)$, $m_2(u)$, $b(w)$, $b(u)\}$. For $\md_4$, we obtain $E_{\md_4} = \{\forall X a_1(X)\rightarrow a_2(X)$, $\forall X a_1(X)\rightarrow m_1(X)$, $\forall X a_1(X)\rightarrow m_2(X)$, $\forall X m_1(X)\rightarrow b(X)$, $\forall X a_1(X) \rightarrow b(X)$, $a_2(w)$, $a_2(u)$, $m_1(w)$, $m_1(u)$, $m_2(w)$, $m_2(u)$, $b(w)$, $b(u)$, $\forall X b(X) \rightarrow m_3(X)$, $\forall X c(X) \rightarrow m_3(X)$, $\forall X m_1(X) \rightarrow m_3(X)$, $\forall X a_1(X) \rightarrow m_3(X)$, $m_3(w)$, $m_3(u)\} = E_{\md_3} \cup \{\forall X b(X) \rightarrow m_3(X)$, $\forall X c(X) \rightarrow m_3(X)$, $\forall X m_1(X) \rightarrow m_3(X)$, $\forall X a_1(X) \rightarrow m_3(X)$, $m_3(w)$, $m_3(u)\}$.

\renewcommand{\arraystretch}{1.4} 
\begin{table}
\small
	\centering
		\rowcolors[]{2}{gray!8}{gray!16} %\arrayrulecolor{black}
		%\extrarowheight10pt
		\begin{tabular}{ c | c | c } 
		%\hline\hline
			\rowcolor{gray!40}
			\toprule\addlinespace[0pt]
			   & $E_{\md_3}$ & $E_{\md_4}$   \\ \addlinespace[0pt]\midrule\addlinespace[0pt]
			%\hline
			$e_1$ & $\forall X a_1(X)\rightarrow a_2(X)$  &  $\forall X a_1(X)\rightarrow a_2(X)$\\ 
			$e_2$ & $\forall X a_1(X)\rightarrow m_1(X)$  &  $\forall X a_1(X)\rightarrow m_1(X)$ \\
			$e_3$ & $\forall X a_1(X)\rightarrow m_2(X)$  &  $\forall X a_1(X)\rightarrow m_2(X)$ \\
			$e_4$ & $\forall X m_1(X)\rightarrow b(X)$    &  $\forall X m_1(X)\rightarrow b(X)$ \\
			$e_5$ & $\forall X a_1(X) \rightarrow b(X)$   &  $\forall X a_1(X) \rightarrow b(X)$\\
			$e_6$ & $a_2(w)$  &  $a_2(w)$ \\
			$e_7$ &  $a_2(u)$ &  $a_2(u)$\\
			$e_8$ &  $m_1(w)$ &  $m_1(w)$\\
			$e_9$ &  $m_1(u)$ &  $m_1(u)$\\
			$e_{10}$ &  $m_2(w)$ &  $m_2(w)$\\
			$e_{11}$ &  $m_2(u)$ &  $m_2(u)$ \\
			$e_{12}$ &  $b(w)$   &  $b(w)$\\
			$e_{13}$ &  $b(u)$   &  $b(u)$\\
			\hdashline
			$e_{14}$ &  & $\forall X b(X) \rightarrow m_3(X)$ \\
			$e_{15}$ &  & $\forall X c(X) \rightarrow m_3(X)$ \\
			$e_{16}$ &  & $\forall X m_1(X) \rightarrow m_3(X)$ \\
			$e_{17}$ &  & $\forall X a_1(X) \rightarrow m_3(X)$ \\
			$e_{18}$ &  & $m_3(w)$\\
			$e_{19}$ &  & $m_3(u)$ \\
			%\hline\hline
			\addlinespace[0pt]\bottomrule
		\end{tabular}
	\caption[(Example~\ref{example:query_computation}) Computing Entailments for Query Generation]{(Example~\ref{example:query_computation}) Entailments computed for KBs $\mo_3^*$ and $\mo_4^*$.}
	\label{tab:example:query_construction_entailments}
\end{table}

\begin{algorithm*}\label{algo:query_gen}
\small
\caption{\small Generation of Queries and Q-Partitions \normalsize} \label{algo:query_gen}
\begin{algorithmic}[1]
\Require an admissible DPI $\tuple{\mo,\mb,\Tp,\Tn}_\RQ$, a set of minimal diagnoses $\mD \subseteq \minD_{\tuple{\mo,\mb,\Tp,\Tn}_\RQ}$ such that $|\mD| \geq 2$, a desired number $q \in \mathbb{N} \cup \setof{\infty}, q \geq 1$ of queries w.r.t.\ $\tuple{\mo,\mb,\Tp,\Tn}_\RQ$ to be returned
\Ensure a set $\QP$ including tuples $\tuple{Q,\tuple{\dx{}(Q),\dnx{}(Q),\dz{}(Q)}}$ such that: %\newline
If $q \geq |\QP_{\max}|$, 
%(in particular $q=\infty$), 
then 
\begin{enumerate}
\item there are no two tuples $\tuple{Q,\Pt(Q)}, \tuple{Q',\Pt(Q')}$ in $\QP$ such that $Q = Q'$ \emph{or} $\Pt(Q)=\Pt(Q')$, and
\item $\QP$ includes a tuple $\tuple{Q,\tuple{\dx{}(Q),\dnx{}(Q),\dz{}(Q)}}$ only if $Q\in\mQ_{\mD,\tuple{\mo,\mb,\Tp,\Tn}_\RQ}$, and  
\item $\QP$ includes at most one tuple where $\dx{}(Q) = Y$ for each $Y \subset \mD$, and 
\item for each $Y \subset \mD$ for which a query $Q$ w.r.t.\ $\mD$ and $\tuple{\mo,\mb,\Tp,\Tn}_\RQ$ exists such that 
%\begin{enumerate}
	%\item 
	(a)~$Q$ includes only entailments computed by the used \textsc{getEntailments} function and
	%\item 
	(b)~$\Pt(Q)$ is such that $\dx{}(Q) = Y$,
%\end{enumerate}
$\QP$ includes a tuple $\tuple{Q',\Pt(Q')}$ such that $\dx{}(Q') = Y$, and
%\item for each $Y \subset \mD$ for which a q-partition $\Pt(Q)$ with $\dx{}(Q) = Y$ exists w.r.t.\ $\mD$ and $\tuple{\mo,\mb,\Tp,\Tn}_\RQ$, $\QP$ includes a tuple $\tuple{Q,\Pt(Q)}$, and
\item $\QP \neq \emptyset$.
\end{enumerate}
If $q < |\QP_{\max}|$, then $\QP$ includes $q$ tuples satisfying (1), (2) and (3). %\newline 
($|\QP_{\max}| \geq 0$ is the maximum number of tuples $\tuple{Q,\Pt(Q)}$ that can be computed by \textsc{getPoolOfQueries} by the used \textsc{getEntailments} function)
%%%%%%%%%%%% OLD start
%\Ensure a set $\QP$ including tuples $\tuple{Q,\tuple{\dx{}(Q),\dnx{}(Q),\dz{}(Q)}}$ such that: \newline If $q \geq |\QP|$ (in particular $q=\infty$), then 
%\begin{enumerate}
%\item there are no two tuples $\tuple{Q,\Pt(Q)}, \tuple{Q',\Pt(Q')}$ in $\QP$ such that $Q = Q'$ \emph{or} $\Pt(Q)=\Pt(Q')$, and
%\item $\QP$ includes a tuple $\tuple{Q,\tuple{\dx{}(Q),\dnx{}(Q),\dz{}(Q)}}$ only if $Q\in\mQ_{\mD,\tuple{\mo,\mb,\Tp,\Tn}_\RQ}$, and  
%\item $\QP$ includes at most one tuple where $\dx{}(Q) = Y$ for each $Y \subset \mD$, and 
%\item for each $Y \subset \mD$ for which a q-partition $\Pt(Q)$ with $\dx{}(Q) = Y$ exists, $\QP$ includes a tuple $\tuple{Q,\Pt(Q)}$.
%\end{enumerate}
%If $q < |\QP|$, then $\QP$ includes $q$ tuples satisfying (1), (2) and (3).
%%%%%%%%%%% OLD end
%a set $\QP$ of $q$ different queries and associated q-partitions, if $q$ different queries w.r.t.\ $\mD$ and $\tuple{\mo,\mb,\Tp,\Tn}_\RQ$ exist; a set $\QP$ including exactly one q-partition with $\dx{} = Y$ for $Y \subset \mD$ if such a q-partition exists

\vspace{10pt}

\Procedure{$\textsc{getPoolOfQueries}$}{$\langle\mo,\mb,\Tp,\Tn\rangle_\RQ, \mD, q$}
\State $E_\mD \gets \emptyset$
\For{$\md\in\mD$} \label{algoline:query:ent_start}
	\State $E_\md \gets \Call{getEntailments}{\md,\mo,\mb,\Tp}$     \Comment{$E_{\md_r}$ is the set of entailments of $\mo_r^*$}
	\State $E_\mD \gets E_\mD \cup \setof{\tuple{\md,E_\md}}$ \label{algoline:query:ent_end}
\EndFor
\For{$\emptyset\subset \mS \subset \mD$}  \label{algoline:query:seed}
	\State $isQuery \gets \false$
	\State $Q \leftarrow \Call{getCommonEntailments}{\mS, E_\mD}$  \label{algoline:query:common_ent}
	\If {$Q \neq \emptyset$} 
		\For{$\md_r \in \mD\setminus\mS$} 		\label{algoline:query:verify_CQ3_start}
			\If{$Q \subseteq E_{\md_r}$}					\Comment{Does $\mo^{*}_r \,\models Q$ ?}
					\State $\dx{} \leftarrow \dx{} \cup \left\{\md_r\right\}$     \label{algoline:query:dx}
			\ElsIf{$\lnot\Call{isKBValid}{\mo^{*}_r \cup Q, \tuple{\cdot,\emptyset,\emptyset,\Tn}_{\RQ}}$}  \label{algoline:query:is_ont_valid}  \Comment{\textsc{isKBValid} (see Algorithm~\ref{algo:qx})}
					\State $\dnx{} \leftarrow \dnx{} \cup \left\{\md_r\right\}$    \label{algoline:query:dnx}
					\State $isQuery \gets \true$    \label{algoline:query:is_query_true}
			\Else
					\State $\dz{} \leftarrow \dz{} \cup \left\{\md_r\right\}$     \label{algoline:query:dz}
			\EndIf
		\EndFor		\label{algoline:query:verify_CQ3_end}
		\If {$isQuery \land \lnot \Call{inclQPart}{\QP,\tuple{\dx{}, \dnx{}, \dz{}}}$} \label{algoline:query:add_QP_start}
				\State $Q' \gets \Call{minQ}{\emptyset,Q,\emptyset, \tuple{\dx{}, \dnx{}, \dz{}}, \tuple{\mo,\mb,\Tp,\Tn}_\RQ}$   \label{algoline:query:minQ}
				\State $\QP \leftarrow \QP \cup \setof{\tuple{Q', \tuple{\dx{}, \dnx{}, \dz{}}}}$  \label{algoline:query:add_QP_end}
				\If {$|\QP| = q$}					\label{algoline:query:test_QP}
						\State \Return $\QP$ \label{algoline:query:return_QP_1}
				\EndIf
		\EndIf
	\EndIf
\EndFor
\If{$|\QP| = 0$} \label{algoline:query:check_QP_empty}
	\State $\QP \gets \Call{addTrivialQueries}{\mD,\QP}$\label{algoline:query:addTrivialQueries}
\EndIf
\State \Return $\QP$
\EndProcedure

\vspace{10pt}

\Procedure{$\textsc{minQ}$}{$X,Q,QB,\tuple{\dx{}, \dnx{}, \dz{}},\langle\mo,\mb,\Tp,\Tn\rangle_\RQ$}
\If{$\ X \neq \emptyset \land \Call{isQPartConst}{QB,\tuple{\dx{}, \dnx{}, \dz{}}, \tuple{\mo,\mb,\Tp,\Tn}_\RQ}$}  \label{algoline:query:validitytest2}  
	\State \Return $\emptyset$    \label{algoline:query:return_emptyset}
\EndIf
\If{$|Q| = 1$}  \label{algoline:query:test_singleton}              
  \State \Return $Q$ \label{algoline:query:return_Q}
\EndIf
\State $k \gets \Call{split}{|Q|}$     \label{algoline:query:split}
\State $Q_1 \gets \Call{get}{Q, 1, k}$%\label{algoline:get1} 
\State $Q_2 \gets \Call{get}{Q, k + 1, |Q|}$%\label{algoline:get2}
\State $Q^{\min}_2 \gets \Call{\textsc{minQ}}{Q_1,Q_2,QB\cup Q_1,\tuple{\dx{}, \dnx{}, \dz{}},\langle\mo,\mb,\Tp,\Tn\rangle_\RQ}$ \label{algoline:query:recursive_call1}
\State $Q^{\min}_1 \gets \Call{\textsc{minQ}}{Q^{\min}_2,Q_1,QB\cup Q^{\min}_2	,\tuple{\dx{}, \dnx{}, \dz{}},\langle\mo,\mb,\Tp,\Tn\rangle_\RQ}$ \label{algoline:query:recursive_call2}
\State \Return $Q^{\min}_1 \cup Q^{\min}_2$  %\label{algoline:return_upwards}
\EndProcedure

\vspace{10pt}

\Procedure{\textsc{isQPartConst}}{$Q,\tuple{\dx{}, \dnx{}, \dz{}}, \tuple{\mo,\mb,\Tp,\Tn}_\RQ$}
\For{$\md_r \in \dnx{}$} 
			\If{$\Call{isKBValid}{\mo^{*}_r \cup Q, \tuple{\cdot,\emptyset,\emptyset,\Tn}_{\RQ}}$} \Comment{\textsc{isKBValid} (see Algorithm~\ref{algo:qx})}
					\State \Return \false
			\EndIf
\EndFor
\For{$\md_r \in \dz{}$} 
			\If{$\mo^{*}_r \models Q$}
					\State \Return \false
			\EndIf
\EndFor
\State \Return \true
\EndProcedure
\end{algorithmic}
\normalsize
\end{algorithm*}
%--> to show that minQ works correctly: analogous to the proof of QX + show that ``$\tuple{\dx{}, \dnx{}, \dz{}}$ is q-partition of $Q$'' is a monotonic function in $Q$
%
%where can a diagnosis ``move'' if $Q$ shrinks? --> from D- to D+ or D0 and from D0 to D+ AND can never move to D- from elsewhere and no diag can move from D+ to anywhere, thus in particular not to D0.
%
%--> remark at function getEntailments: Usually, the number of entailments of a set of formulas is not finite. However, the entailments of a certain type returned by a reasoner are finite. E.g.\, asked for entailments of $A \sqsubseteq B \sqcap C$, a reasoner that performing the classification reasoning service would give back $A \sqsubseteq B$ and $A \sqsubseteq C$, but not $A \sqsubseteq B \sqcup C$ or $A \sqsubseteq C \sqcap C \sqcap C$. That is, when we speak of entailments, then we mean entailments in the practical sense, i.e.\ w.r.t.\ a reasoning service such as classification for description logic which computes subsumptions $X \sqsubseteq Y$ such that $Y$ is the most specific concept that subsumes $X$, or forward checking for propositional logic which computes all and only atoms that are entailed by a knowledge base.
%

%\noindent\textbf{Generation of a pool of queries.} 
\section[Generation of a Pool of Queries]{Generation of a Pool of Queries%
\sectionmark{Query Pool Generation}}
\sectionmark{Query Pool Generation}
\label{sec:GenerationOfAPoolOfQueries}
%%%%%
%\section{Generation of a Pool of Queries}
%\label{sec:GenerationOfAPoolOfQueries}
%%%%%
The main function \textsc{getPoolOfQueries} of Algorithm~\ref{algo:query_gen} gets as inputs an admissible DPI $\tuple{\mo,\mb,\Tp,\Tn}_\RQ$ over $\mathcal{L}$, a set of leading (minimal) diagnoses $\mD \subseteq \minD_{\tuple{\mo,\mb,\Tp,\Tn}_\RQ}$ such that $|\mD|\geq 2$ and a parameter $q \in \mathbb{N}\cup\setof{\infty}, q\geq 1$ that indicates the number of queries in $\mQ_{\mD,\tuple{\mo,\mb,\Tp,\Tn}_\RQ}$ the algorithm is supposed to return (where $q := \infty$ signalizes that a maximum number of queries should be output). The way of generating a pool of queries is guided by Proposition~\ref{prop:query_dx_dnx} which says that a non-empty set $Q$ of formulas over $\mathcal{L}$ is a query w.r.t.\ $\mD$ and $\tuple{\mo,\mb,\Tp,\Tn}_\RQ$ if and only if $\dx{}(Q)$ as well as $\dnx{}(Q)$ are non-empty sets of diagnoses. That is, the necessary and sufficient criteria for $Q$ to be a query are 
\begin{enumerate}[(CQ1)]
	\item $\quad Q \neq \emptyset$ and
	\item $\quad\dx{}(Q) \neq \emptyset$ and
	\item $\quad\dnx{}(Q)\neq \emptyset$.
\end{enumerate}
%(CQ1) $Q \neq \emptyset$ and (CQ2) $\dx{}(Q) \neq \emptyset$ and (CQ3) $\dnx{}(Q)\neq \emptyset$.
Note, since the disjoint sets of diagnoses $\dx{}(Q) \subseteq \mD$ and $\dnx{}(Q) \subseteq \mD$ must not be empty, $|\mD|\geq 2$ must be postulated in order for any queries to exist w.r.t.\ $\mD$ and $\tuple{\mo,\mb,\Tp,\Tn}_\RQ$ (cf.\ Corollary~\ref{cor:query_num_lower_bound}).  

As a first action (lines \ref{algoline:query:ent_start}-\ref{algoline:query:ent_end}), the algorithm computes a set of entailments $E_{\md_i}$ 
%-- for instance, entailmentof certain types, e.g.\ classification and realization for description logics or atoms for propositional logic, 
for each $\mo_i^*$ (cf.\ Formula~\ref{eq:sol_ont_candidate})
%$\mo_i^* = (\mo\setminus\md_i)\cup\mb\cup U_\Tp$ 
where $\md_i \in \mD$ and stores these entailments along with the respective diagnosis as a tuple $\tuple{\md_i,E_{\md_i}}$ in a set $E_\mD$. \label{etc:definition_of_getEntailments_function} This is accomplished by the function \textsc{getEntailments} which gets a tuple $\tuple{X,Y,Z,W}$ of arguments where $X,Y,Z$ are sets of formulas over some logic $\mathcal{L}$ and $W$ is a set including sets of formulas over $\mathcal{L}$. Then, \textsc{getEntailments} computes a \emph{finite} (cf.\ Remark~\ref{rem:entailment_computation_finite_types_of_entailments}) set of entailments of certain types (cf.\ Examples~\ref{example:query_computation} and \ref{example:ad_Table_of_queries_partitions}) of the KB $(Y\setminus X) \cup Z \cup U_{W}$. 
%It is important to note that the \textsc{getEntailments} function must be fixed throughout the entire debugging session, i.e.\ it must deterministically compute the s

Then, the algorithm runs through all proper non-empty subsets $\mS$ of the leading diagnoses $\mD$ and, for each $\mS$, it computes the set of common entailments $Q$ of all KBs $\mo_i^*$ where $\md_i \in \mS$ (function \textsc{getCommonEntailments}) by means of the precomputed set $E_\mD$. That is, $Q := \bigcap_{\md \in \mS} E_{\md}$. If $Q$ is non-empty, then CQ1 and CQ2 are fulfilled for $Q$. CQ2 is met since $\mS \neq \emptyset$ and thus there is a diagnosis $\md_i\in\mD$ such that $\mo_i^* \models Q$ which implies that $\dx{}(Q) \neq \emptyset$. So, the algorithm proceeds to verify CQ3 (lines~\ref{algoline:query:verify_CQ3_start}-\ref{algoline:query:verify_CQ3_end}) in that it assigns the remaining diagnoses in $\mD$ that are not in $\mS$ to the according sets $\dx{}(Q)$, $\dnx{}(Q)$ or $\dz{}(Q)$ as per Definition~\ref{def:q-partition}. Note that the function \textsc{isKBValid} has been specified in Algorithm~\ref{algo:qx} on page~\pageref{algo:qx}. With the parameters given when called in line~\ref{algoline:query:is_ont_valid}, \textsc{isKBValid} checks whether $\mo_r^* \cup Q = (\mo\setminus\md_r) \cup\mb\cup U_{\Tp\cup\setof{Q}}$ does not violate any requirement in $\RQ$ and does not entail any test case in $\Tn$. Once the call to this function returns $\false$ for one diagnosis $\md_r \in \mD\setminus\mS$, it holds that $\md_r \in \dnx{}(Q)$ thus CQ3 is definitely met. Therefore, $isQuery$ is set to $\true$ in line~\ref{algoline:query:is_query_true}. If, on the other hand, $isQuery$ is not set to $\true$ for any diagnosis in $\mD \setminus \mS$, then the set $\dnx{}(Q) =\emptyset$ and thus $Q$ is not in $\mQ_{\mD,\tuple{\mo,\mb,\Tp,\Tn}_\RQ}$.

So far, we have proven the following proposition.
\begin{proposition}\label{prop:query_gen_isQuery_correct}
Let a DPI $\tuple{\mo,\mb,\Tp,\Tn}_\RQ$, a set of diagnoses $\mD\subseteq\minD_{\tuple{\mo,\mb,\Tp,\Tn}_\RQ}$ and a natural number $q \geq 1$ be the input to the function \textsc{getPoolOfQueries}. Then, a value stored in variable $Q$ at the time \textsc{getPoolOfQueries} executes line~\ref{algoline:query:add_QP_start} is a query w.r.t.\ $\mD$ and $\tuple{\mo,\mb,\Tp,\Tn}_\RQ$ iff the variable $isQuery$ stores the value $\true$.
\end{proposition}

If the purpose was only to find queries (and not q-partitions), the algorithm could stop processing for the current $Q$ and go to the next set $\mS$, given that $isQuery$ is set to $\true$ for some diagnosis. However, as the q-partition provides meaningful information to assess a query, e.g.\ it gives the number of diagnoses invalidated for each answer or the estimated probability of each answer (cf.\ Chapter~\ref{chap:UserInteraction}), the q-partition is a necessary input to the subsequently called function \textsc{selectBestQuery} (line~\ref{algoline:inter_onto_debug_continued:selectBestQuery} in Algorithm~\ref{algo:inter_onto_debug_continued}, see later in Sections~\ref{sec:Walkthrough} and \ref{sec:query_selection_measures}) that selects a query from the pool of queries $\QP$. For this reason, the algorithm continues until the computation of the q-partition for $Q$ is complete.

In a last step (lines~\ref{algoline:query:add_QP_start}-\ref{algoline:query:add_QP_end}), given that $isQuery$ is $\true$ and there is not yet a query with the same q-partition in $\QP$, the algorithm computes a set-minimal subset $Q_{\min}$ of $Q$ such that the q-partition of $Q_{\min}$ is the same as the one of $Q$ (function \textsc{minQ}). Finally, the tuple $\tuple{Q_{\min},\tuple{\dx{}, \dnx{}, \dz{}}}$ including the minimized query $Q_{\min}$ along with its q-partition $\tuple{\dx{}, \dnx{}, \dz{}}$ is added to $\QP$. If $|\QP|=q$, then $\QP$ is returned; otherwise, a further iteration for another $\mS$ is executed. If $|\QP|=q$ is not met until all seeds $\mS$ have been processed, the set $\QP$ is checked for emptiness in line~\ref{algoline:query:check_QP_empty}. If $\QP = \emptyset$, then the function \textsc{addTrivialQueries} (line~\ref{algoline:query:addTrivialQueries}) adds $|\mD| \geq 2$ queries as defined by $Q$ in Proposition~\ref{prop:q1} to $\QP$ (cf. Corollary~\ref{cor:query_num_lower_bound}) and then returns $\QP$; otherwise, $\QP$ is directly returned.

\begin{remark}\label{rem:guaranteeing_non-empty_QP_by_getPoolOfQueries}
Notice that lines~\ref{algoline:query:check_QP_empty} and \ref{algoline:query:addTrivialQueries} in Algorithm~\ref{algo:query_gen} aim at ensuring the non-emptiness of the pool of queries $\QP$ returned by \textsc{getPoolOfQueries} for \emph{any} \textsc{getEntailments} function (see Example~\ref{example:ad_Table_of_queries_partitions} for different specifications of the \textsc{getEntailments} function). This is a necessary criterion for the interactive KB debugging system (Algorithm~\ref{algo:inter_onto_debug}) to work in a sound way since it guarantees that the \textsc{calcQuery} function (line~\ref{algoline:inter_onto_debug:calc_query} in Algorithm~\ref{algo:inter_onto_debug}) \emph{always} returns a query w.r.t.\ the current set of leading diagnoses $\mD$ and the given DPI. 
%is to add the $|\mD|$ queries defined as per Proposition~\ref{prop:q1} in any case to $\QP$. 
%
Note that the $|\mD|$ queries generated and added to $\QP$ by \textsc{addTrivialQueries} can be \emph{trivially} obtained without the consultation of a reasoning service by extraction of the respective formulas from the KB $\mo$, as prescribed by Proposition~\ref{prop:q1}.\qed 
%
%In the sequel, let us assume that these trivial queries are added to $\QP$ 
%
%employed \textsc{getEntailments} function, besides computing some particular types of entailments (cf.\ Section~\ref{sec:remarks_query_gen} and Example~\ref{example:ad_Table_of_queries_partitions}), adds these $|\mD|$ trivial queries to $\QP$.
\end{remark}

\section{Discussion of Query Pool Generation}
\label{sec:remarks_query_gen}

%\noindent\textbf{Multiple Equal Q-Partitions.} 
\paragraph{Multiple Equal Q-Partitions.}
In the general case there is more than one query w.r.t.\ one and the same q-partition. For that reason alone that a minimized query is a set-minimal subset of an initially computed one where multiple such subsets may exist. 

\begin{example}\label{example:multiple_minimized_queries}
An example for such a query resulting in multiple minimized subqueries with identical q-partition can be found in Example~\ref{example:query_computation}.\qed
\end{example}

However, note that \textsc{getPoolOfQueries} is designed to compute a pool $\QP$ that includes at most one query with one and the same q-partition. The idea behind this is (1)~to minimize the calls to the expensive function \textsc{minQ} and (2)~that two queries with the same q-partition have exactly the same properties w.r.t.\ common query selection criteria such as maximum expected information gain or maximum worst case invalidation rate of diagnoses after the query answer is known. Such criteria have been shown to often lead to a reduction of debugging effort for the interacting user (cf. \cite{Shchekotykhin2012,Rodler2013}).
%well known information theoretic measures for query selection such as entropy or maximum worst case invalidation rate of diagnoses after the query answer is known (cf. \cite{Shchekotykhin2012,Rodler2013}). 
As the purpose of the computation of the pool of queries $\QP$ is to constitute an input to the query selection function that uses exactly such selection measures, the inclusion of only one query with a particular q-partition is reasonable, also (3)~to minimize computation time of the query selection function which needs to go through all elements of $\QP$ in order to pick the ``best'' one in the worst case.
 
On the other hand, regarding the comprehensibility of the query, i.e.\ the cognitive load on the user when it comes to understanding the meaning of the query, two queries with the same q-partition may well be significantly different. This however is beyond the scope of this work and considered a topic for future research. 

The following proposition gives evidence that the set $\QP$ returned by \textsc{getPoolOfQueries} is indeed duplicate-free w.r.t.\ the q-partitions in $\QP$. 
\begin{proposition}\label{prop:query_gen_duplicate_free}
Let a DPI $\tuple{\mo,\mb,\Tp,\Tn}_\RQ$, a set of diagnoses $\mD\subseteq\minD_{\tuple{\mo,\mb,\Tp,\Tn}_\RQ}$ and $q \in \mathbb{N} \cup \setof{\infty}, q \geq 1$ be the input to the function \textsc{getPoolOfQueries}. Then, the function \textsc{getPoolOfQueries} returns a set $\QP$ including tuples of the form $\tuple{Q,\Pt(Q)}$ where $Q\in\mQ_{\mD,\tuple{\mo,\mb,\Tp,\Tn}_\RQ}$ is a query and $\Pt(Q) = \tuple{\dx{}(Q),\dnx{}(Q),\dz{}(Q)}$ is the q-partition of $Q$ such that $\QP$ does not include any two equal queries and does not include any two equal q-partitions.  
\end{proposition}
\begin{proof}
%As the function \textsc{minQ} is q-partition-preserving, i.e.\ the q-partition $\tuple{\dx{},\dnx{},\dz{}}$ of the non-minimized query given as input to \textsc{minQ} is the q-partition of the minimized query returned as output of \textsc{minQ}, we have that 
The test of the criterion $\lnot \textsc{inclQPart}$ tested before the call to \textsc{minQ} will always return false for the q-partition $\tuple{\dx{},\dnx{},\dz{}}$ if $\tuple{\dx{},\dnx{},\dz{}}$ is already included in a tuple in $\QP$. Since \textsc{minQ} is q-partition-preserving, no q-partition that does not occur in a tuple in $\QP$ can become equal to some q-partition in $\QP$ by a call to \textsc{minQ}. Therefore, $\QP$ cannot include any two equal q-partitions. Since two equal queries have equal q-partitions, any two different q-partitions cannot be q-partitions of equal queries. Thus, $\QP$ cannot include any two equal queries either. 
%, i.e.\ equal input to this function means equal output of that function.
%show that by determinism of minQ equal output of this function means equal input, and since q-partition of input and output are the same, this means criterion in line if... will never be satisfied for a duplicate q-partition and thus the proposition holds
\end{proof}
Note that, on account of the q-partition preserving property of \textsc{minQ}, only such q-partitions are ruled out by the criterion in line~\ref{algoline:query:add_QP_start} that would lead to duplicates at the time they should be added to $\QP$ in line~\ref{algoline:query:add_QP_end}.
\paragraph{Computation of Entailments.}
Generally, the (\emph{theoretical}) number of entailments of a set of formulas is not finite. However, the entailments (of a certain type) returned by a reasoner are finite. For instance, asked for entailments of $\setof{A \sqsubseteq B \sqcap C}$, a reasoner performing the classification reasoning service would give back $A \sqsubseteq B$ and $A \sqsubseteq C$, but not entailments like $A \sqsubseteq B \sqcup C$ or $A \sqsubseteq C \sqcap C \sqcap C$. That is, when we speak of entailments, then we mean entailments in the \emph{practical} sense (cf.\ Remark~\ref{rem:entailment_computation_finite_types_of_entailments}), i.e.\ w.r.t.\ a reasoning service such as classification for DL KBs which computes all and only subsumptions $X \sqsubseteq Y$ such that $Y$ is the most specific concept that subsumes $X$, or forward-chaining for Datalog KBs which computes all and only atoms that are entailed by the KB.

\begin{example}\label{example:entailment_number} 
If we recall Example~\ref{example:query_computation}, we see that the number of computed entailments of $\mo_4^*$ and $\mo_3^*$ was 19 and 13 respectively, which are rather high numbers in the light of the small KBs, but importantly these numbers are necessarily finite. For, there cannot be more than $|Pred|^2$ entailments of the $\forall X p_1(X) \rightarrow p_2(X)$ type and not more than $|Pred|\,|Const|$ entailments of the $p(a)$ type for a KB whose signature includes the unary predicate symbols $Pred$ and constant symbols $Const$ and does not include any function symbols. In case of KB $\mo_3^*$, for example, the set $Pred = \setof{a_1,a_2,m_1,m_2,m_3,a,b}$ and $Const=\setof{u,w}$ which means that upper bounds for the number of entailments of the first and second type are $49$ and $14$, respectively.\qed
\end{example}

Further, note that the number of existing different q-partitions and which q-partitions there are at all w.r.t.\ some set of leading diagnoses $\mD$ and a DPI depends on the function \textsc{getEntailments}, i.e.\ on the set of entailments calculated by it. 

%\begin{example}\label{example:q-partition_depend_on_getEntailments} Assume a description logic KB. If \textsc{getEntailments} computes only (atomic) concept subsumption entailments of the KB, an equivalent to the first type of entailments we computed in Example~\ref{example:query_computation} for our first-order KB, then there might be fewer different queries and fewer different q-partitions than for a function that additionally returns atomic concept instantiation entailments, an equivalent to the second type of entailments we computed in Example~\ref{example:query_computation} for our first-order KB.
%\end{example} 

\begin{example}
Recall Example~\ref{example:query_computation} where we constructed a query $Q$ w.r.t.\ the set of all minimal diagnoses for the DPI given by Table~\ref{tab:example1}. Assume now that only entailments of the first type, i.e.\ those of the form $\forall X p_1(X) \rightarrow p_2(X)$, and none of the second type $p(a)$ are computed by \textsc{getEntailments} and denote the set of entailments of this form of $\mo_i^*$ by $E'_{\md_i}$. Then, $Q' = E'_{\md_3}\cup E'_{\md_4} = \setof{e_1,\dots,e_5}$ (cf.\ Table~\ref{tab:example:query_construction_entailments}), i.e.\ a subset of the query $Q$ computed for a \textsc{getEntailments} function producing entailments of both types. The q-partition of $Q'$ is the same as the q-partition of $Q$, namely $\tuple{\setof{\md_3,\md_4},\setof{\md_1,\md_2},\emptyset}$. However, the queries $Q_{\min,1}$ and $Q_{\min,2}$ are no longer obtained as minimized versions of $Q'$, unlike $Q_{\min,3}$ and $Q_{\min,4}$ which are subqueries of $Q'$, too.  \qed
\end{example}

%\noindent\textbf{Minimizing the Set $\dz{}$ in Q-Partitions.} 
\paragraph{Minimizing the Set $\dz{}$ in Q-Partitions.}
Recall that $\dz{} = \emptyset$ is a desirable property of a q-partition since a query with such q-partition may invalidate any leading diagnosis, depending on the answer to the query (cf.\ Chapter~\ref{chap:UserInteraction}). In other words, no leading diagnosis is guaranteed to be still valid \emph{for any answer} after the query is added as a test case to the DPI. 

In general, \textsc{getPoolOfQueries} computes q-partitions where $\dz{}$ may be a non-empty set. However, if the \textsc{getEntailments} function is specified to compute certain explicit entailments of $\mo$, then $\dz{} = \emptyset$ can be guaranteed.
\begin{definition}[Explicit Entailment]\label{def:explicit_entailment}
Let $\mo$ be a KB. Then, $\alpha$ is an explicit entailment of $\mo$ iff $\alpha \in \mo$.
\end{definition}
Now, if each set of entailments $E_\md$ computed by \textsc{getEntailments} includes all the formulas that occur in some diagnosis in $\mD$, but do not occur in $\md$, then \textsc{getPoolOfQueries} definitely returns a set $\QP$ of queries and associated q-partitions where $\dz{}(Q) = \emptyset$ holds for each tuple in $\QP$.
\begin{proposition}\label{prop:query_gen_explicit_entailments_dz_empty}
Let $\langle\mo,\mb,\Tp,\Tn\rangle_\RQ$ be a DPI and $\mD \subseteq \minD_{\langle\mo,\mb,\Tp,\Tn\rangle_\RQ}$. If the set $E_{\md}$ computed by \textsc{getEntailments} meets $E_{\md} \supseteq U_{\mD} \setminus \md$ for all $\md \in \mD$, then \textsc{getPoolOfQueries} computes only queries $Q$ with $\dz{}(Q) = \emptyset$.
\end{proposition}
\begin{proof}
Assume that $Q$ is some query computed by \textsc{getPoolOfQueries}. As \textsc{minQ} is a q-partition preserving transformation of $Q$, we can assume w.l.o.g.\ that $Q$ is a query computed by \textsc{getPoolOfQueries} \emph{before} \textsc{minQ} is called for $Q$. 
We have to show that for an arbitrary diagnosis $\md_i \in \mD$ either $\md_i$ is assigned to $\dx{}(Q)$ or to $\dnx{}(Q)$. 

So, let us assume that there is a diagnosis $\md_k$ which is assigned to $\dz{}(Q) = \mD\setminus (\dx{}(Q)\cup\dnx{}(Q))$ in line~\ref{algoline:query:dz}. Then, $Q \not\subseteq E_{\md_k}$ and $\mo_k^* \cup Q$ does not violate any $x \in \RQ\cup\Tn$ must hold, otherwise $\md_k$ would have already been assigned to $\dx{}(Q)$ in line~\ref{algoline:query:dx} or to $\dnx{}(Q)$ in line~\ref{algoline:query:dnx}. 
%Let $\mS \subset \mD$ be the seed selected in line~\ref{algoline:query:seed} for which $Q$ was computed in line~\ref{algoline:query:common_ent}. 
But $Q \not\subseteq E_{\md_k}$ implies $Q \not\subseteq U_{\mD} \setminus \md_k$ since $E_{\md_k} \supseteq U_{\mD} \setminus \md_k$ by precondition.
%, as otherwise $Q$ would be entailed by $\mo_k^*$. 
This in turn means that there is some formula $\tax$ in $Q$ which is not in $U_{\mD} \setminus \md_k$. Then $\tax\in\md_k$ must hold, as otherwise for all formulas $\tax'\in Q$ it would hold that $\tax'$ is an entailment of $\mo_k^* = (\mo\setminus\md_k)\cup \mb \cup U_\Tp$, i.e.\ an entailment of all formulas in $\mo \cup \mb \cup U_{\Tp}$ except for those in $\md_k$. However, all entailments of $\mo_k^*$ are stored in $E_{\md_k}$ by the implementation of the function \textsc{getEntailments}. Thus $Q \subseteq E_{\md_k}$ would hold which cannot be the case as shown before. Consequently, we have derived that $Q \cap \md_k \neq \emptyset$ which means by set-minimality of diagnoses in $\mD$, in particular of $\md_k$, that $\mo_k^* \cup Q$ must violate some $x\in\RQ\cup\Tn$ which is a contradiction to the assumption that $\md_k \in \dz{}(Q)$.
\end{proof}
%For KBs including a large set of logical formulas, the strategy of having all explicit entailments computed by \textsc{getEntailments}, can lead to a degradation of query selection time due to an overhead in functions \textsc{getCommonEntailments} (larger sets must be handled), \textsc{isOntValid} (significantly larger $Q$ can lead to significant overhead in reasoning) and \textsc{minQ} (overhead in reasoning time and number of calls to the reasoner).

\begin{example}
Let us come back to the example DPI given by Table~\ref{tab:example1}. The possibility of a query $Q$ constructed by Algorithm~\ref{algo:query_gen} with $\dz{}(Q) \neq \emptyset$ is witnessed by the selection of seed $\mS = \setof{\md_1}$ and the assumption that entailments of the two types given in Example~\ref{example:query_computation} are produced by \textsc{getEntailments}. The set of entailments $Q = E_{\md_1} = \setof{e_4,e_{14},e_{15},\forall X m_2(X) \rightarrow d(X)}$ (for $e_i$ cf.\ Table~\ref{tab:example:query_construction_entailments}). Then, $\md_2$ as well as $\md_3$ are assigned to $\dnx{}(Q)$ as both KBs $\mo_3^*\cup Q$, $\mo_4^* \cup Q$ entail $m_3(w)$ and $\lnot m_3(w)$ wherefore they are both inconsistent and thus violate $r_1 \in \RQ$. However, $\md_4 \in \dz{}(Q)$ since $\mo_i^* \not\models \forall X m_2(X) \rightarrow d(X)$ and hence does not entail $Q$ and since $\mo_i^* \cup Q$ does not violate consistency or coherency (recall that the set of negative test cases is empty in the DPI and thus must not be considered), i.e.\ does not contain a conflict set.

Applying Proposition~\ref{prop:query_gen_explicit_entailments_dz_empty}, we could use a modified \textsc{getEntailments} function that returns a minimal set of entailments just that the precondition of the proposition is met, i.e.\ $E'_{\md} = U_{\mD} \setminus \md$ for all $\md\in\mD$. With this function, for the seed $\mS = \setof{\md_1}$ we would get $Q' = E'_{\md_1} = \setof{2,3,4,5}$ (again, formulas in Table~\ref{tab:example1} are referred to just by their number). Let us now check whether $\dz{}(Q')$ is indeed empty. As explicit entailments are stronger than non-explicit ones, we must still have that $\md_2,\md_3 \in \dnx{}(Q')$. For $\md_4$, we have $\mo_4^* \cup Q' = \setof{1,3,5,6,7,8} \cup \setof{2,3,4,5} = \setof{1,2,3,4,5,6,7,8}$ which corresponds to the entire KB plus background knowledge of the given DPI and includes conflict sets $\mc_1 = \setof{1,3,4}$ and $\mc_2= \setof{1,2,3,5}$ wherefore it is inconsistent. Therefore, diagnosis $\md_4$ must also be an element of $\dnx{}(Q')$. 

Please note that making the entailments $Q = E_{\md_1}$ computed by the unmodified \textsc{getEntailments} function only slightly stronger would already suffice to force inclusion of $\md_4$ in $\dz{}(Q)$. In fact, including $\tax_4 := \forall X m_2(X) \rightarrow (\forall Y s(X,Y) \rightarrow a(Y)) \land d(X)$ in $Q$ instead of $\forall X m_2(X) \rightarrow d(X)$ would make $Q$ non-disjoint with $\md_4$ as both comprise $\tax_4$. Consequently, in line with the proof of Proposition~\ref{prop:query_gen_explicit_entailments_dz_empty}, $\mo_4^* \cup Q$ must include a conflict set ($\setof{1,3,4}$) wherefore $\md_4 \in \dnx{}(Q)$. 

Another point we want to mention is that empty $\dz{}$ \emph{could} also be achieved by making the query slightly weaker. For our concrete query $Q = E_{\md_1}$, this means that leaving out $\forall X m_2(X) \rightarrow d(X)$ would lead to empty $\dz{}(Q)$. However, the difference to the scenario above where we made $Q$ sightly stronger is that $\md_4$ would be an element of $\dx{}(Q)$ instead of $\dnx{}(Q)$ in this case, i.e.\ the q-partition would be $\tuple{\setof{\md_1,\md_4},\setof{\md_2,\md_3},\emptyset}$. 

A shortcoming of the strategy of making the query weaker is that it can be computationally expensive as perhaps a large number of subsets of $Q$ might need to be considered and tested for fulfillment of $\dz{}(Q)=\emptyset$. Each such test would involve calls to the reasoner which are usually expensive. A second drawback is that no guarantee is given to finally end up with an empty set $\dz{}(Q)$ since weakening of $Q$ might also involve the ``shift'' of some diagnosis from $\dnx{}(Q)$ to $\dz{}(Q)$. On the other hand, the strategy of computing stronger entailments is computationally more resource-saving as (trivially obtained) explicit entailments can be added to make the query stronger. Furthermore, making the query stronger -- in a controlled way, by adding formulas from $U_{\mD} \setminus U_{\dx{}(Q)}$ to $Q$ as suggested by Proposition~\ref{prop:query_gen_explicit_entailments_dz_empty} -- can never lead to non-empty $\dz{}(Q)$ as Proposition~\ref{prop:query_gen_explicit_entailments_dz_empty} substantiates. \qed
\end{example}

%\noindent\textbf{(Non-)Completeness of Query Pool $\QP$.} 
\paragraph{(Non-)Completeness of Query Pool $\QP$.}
Note that specifying $q:=\infty$ causes \textsc{getPoolOfQueries} to run through all $\mS\subset\mD$ and to compute a maximum number of queries. However, in general, not all theoretically possible queries are computed by \textsc{getPoolOfQueries}. One trivial reason for this is that only minimized, i.e.\ set-minimal, queries are contained in the returned set $\QP$. 

But, also queries $Q'$ with $\dx{}(Q') = Y \subset \mD$ will not be included in $\QP$ if there is some query $Q$ with $\dx{}(Q) = Y$ such that $|\dnx{}(Q)| > |\dnx{}(Q')|$ (and, equivalently, $|\dz{}(Q)| < |\dz{}(Q')|$). As we will learn in a moment, both mentioned reasons for the incompleteness of the output of \textsc{getPoolOfQueries} will even be desirable for reasons of efficiency. That is, the mentioned types of queries that are not taken into account in $\QP$ are ``non-preferred'' as non-set-minimal queries demand a non-necessary amount of user interaction and the answering of queries $Q$ with a non-necessarily large set $\dz{}(Q)$ involves a worse discrimination between leading minimal diagnoses (and, if these are ``good'' representatives of all minimal diagnoses, then of all minimal diagnoses) than other queries $Q'$ with $|\dz{}(Q')| < |\dz{}(Q)|$ and $\dx{}(Q) = \dx{}(Q')$. 

%In fact, however, the queries that are not found by the algorithm are ``non-preferred'' queries in the following sense: If the set $Q$ of common entailments of diagnoses in $\mS$ computed in line~\ref{algoline:query:common_ent} is not a query, then CQ3 is violated and $\dnx{}(Q) = \emptyset$. In this case, no subset $Q'$ of $Q$ can have $\dnx{}(Q') \neq \emptyset$. Thus, we do not miss any query in this case. 
%If $Q$ is a query, then it is one query with minimal $|\dx{}(Q)|$ and maximal $|\dnx{}(Q)|$ (w.r.t.\ the seed $\mS$ from which it was calculated). 
%This holds because a deletion of formulas from $Q$ can only increase the number of diagnoses in $\dnx{}$ and decrease the number of diagnoses in $\dnx{}$. The number of diagnoses in $\dz{}$ can increase or decrease (diagnoses from $\dnx{}$ can ``move'' to $\dz{}$ and diagnoses from $\dz{}$ can ``move'' to $\dx{}$). These facts where derived in the proof of Lemma~\ref{lem:isqpartconst_correct}. 
%In other words, given that \textsc{getEntailments} computes \emph{all} entailments of a certain type, then the algorithm detects all q-partitions where for each fixed $\mS \subset \mD$ it finds a query $Q'$ with minimal (w.r.t.\ $\mS$) $|\dx{}(Q')|$ and maximal $|\dnx{}(Q')|$. 

Still, \textsc{getPoolOfQueries} meets a completeness criterion for a subset of all queries $\mQ_{\mD,\tuple{\mo,\mb,\Tp,\Tn}_\RQ}$, elements of which cannot be \emph{trivially} detected to be ``non-preferred''. That is, \textsc{getPoolOfQueries} is complete w.r.t.\ the set $\dx{}$, as the following proposition states. In other words, for each subset $X \subset \mD$ it detects a q-partition with $\dx{} = X$, if one exists. 

\begin{proposition} \label{prop:query:dx_complete}
Let a DPI $\tuple{\mo,\mb,\Tp,\Tn}_\RQ$, $\mD \subseteq \minD_{\tuple{\mo,\mb,\Tp,\Tn}_\RQ}$ such that $|\mD| \geq 2$ and some $q \in \mathbb{N} \cup \setof{\infty}, q \geq 1$  be the inputs to \textsc{getPoolOfQueries} and let $|\QP_{\max}| \geq 0$ be the maximum number of tuples $\tuple{Q,\Pt(Q)}$ that can be computed by \textsc{getPoolOfQueries} by means of the used \textsc{getEntailments} function. Further, let $Y$ be an arbitrary subset of $\mD$. If there is some query $Q \in \mQ_{\mD,\tuple{\mo,\mb,\Tp,\Tn}_\RQ}$ that (1)~includes only entailments that are computed by \textsc{getEntailments} and (2)~has a q-partition such that $\dx{}(Q) = Y$, then \textsc{getPoolOfQueries} with parameter $q\geq |\QP_{\max}|$ returns a set $\QP$ including a query $Q'$ with $\dx{}(Q') = Y$. 
Moreover, this query $Q'$ is found in the iteration where the seed $\mS = Y$.
%If there is a q-partition with \dx = Y for queries that include only entailments computed by getEntailments
%(holds w.r.t.\ ONE implementation of the getEntailments function, e.g. one that computes only and all entailments of a certain type) 
%Let $Y \subset \mD$ an arbitrary set of diagnoses. If there is a query $Q$ with $\dx{}(Q) = Y$, then \textsc{getPoolOfQueries} with $q = \infty$ will return a set $\QP$ that includes a q-partition with $\dx{}(Q) = Y$.
%--> ``if there is a q-partition with dx = Y, then this q-partition is found by seed S = Y''
\end{proposition}
\begin{proof}
Since $q\geq |\QP_{\max}|$, \textsc{getPoolOfQueries} will arrive at a step where it selects the seed $\mS = Y$ in line~\ref{algoline:query:seed}. Now, let us assume that in this iteration no query $Q$ with $\dx{}(Q)=Y$ is found. Then, either (a)~no query is found at all, i.e.\ CQ1 or CQ2 or CQ3 are violated, or (b)~a query $Q$ with $\dx{}(Q) \neq Y$ is found. 

(a): Assume first that CQ1 is violated, i.e.\ \textsc{getCommonEntailments} called with argument $\mS$ returns $\emptyset$. This implies that the KBs $\mo_r^*$ for $\md_r \in Y$ have no common entailments, if entailments are computed by \textsc{getEntailments}. This however means that there cannot be a q-partition with $\dx{} \supseteq Y$ which is a contradiction to the precondition that there is some query $Q \in \mQ_{\mD,\tuple{\mo,\mb,\Tp,\Tn}_\RQ}$ that includes only entailments computed by \textsc{getEntailments} and has a q-partition such that $\dx{}(Q) = Y$.

Second, assume that CQ2 is violated, i.e.\ $\dx{}(Q) = \emptyset$. If \textsc{getCommonEntailments} with argument $\mS$ returned $Q \neq \emptyset$, then $\dx{}(Q) \supseteq \mS \supset \emptyset$ would hold. Thus, $Q = \emptyset$, i.e.\ CQ1 is violated. So, as shown before, this leads to a contradiction.

In case any of CQ1 or CQ2 is violated, we already derived a contradiction. So, we make the assumption that CQ1 and CQ2 are met. So, finally, let us assume that CQ3 is violated, i.e.\ that $\dnx{}(Q) = \emptyset$. That is, if $Q$ (which must be a non-empty set by CQ1) denotes \emph{all} common entailments (computable with \textsc{getEntailments}) of $\mo_r^*$ for $\md_r\in Y$, then $\mo_i^* \cup Q$ does not violate any $x\in\RQ\cup\Tn$ for any $\md_i \in \mD \setminus \mS$. Consequently, for all diagnoses $\md_i$ in $\mD$ we have that $\mo_i^* \cup Q$ does not violate any $x\in\RQ\cup\Tn$. But, as there is, by precondition, a query with $\dx{} = Y$, this query must be a subset of all possible common entailments (computable with \textsc{getEntailments}) of KBs $\mo_i^*$ for diagnoses in $Y$, i.e.\ this query must be a subset of $Q$. But, by monotonicity of $\mathcal{L}$, no $\mo_i^* \cup Q'$ for a subset $Q'$ of $Q$ can violate $x\in\RQ\cup\Tn$ if $Q$ does not. Again, we have a contradiction to the precondition as above.

(b): Here, a query $Q$ is found with $\dx{}(Q) \neq Y$ and $\dnx{}(Q) \neq \emptyset$. Since $Q$ is a query, $Q\neq \emptyset$ must hold. Since the seed $\mS = Y$, this means that $Q$ is the set of all common entailments (computable with \textsc{getEntailments}) of $\mo_i^*$ for $\md_i \in Y$, i.e.\ $\dx{}(Q) \supseteq Y$. By $\dx{}(Q) \neq Y$, we conclude that $\dx{}(Q) \supset Y$ must be true. The only way of achieving a smaller set $\dx{}(Q)$, namely $\dx{}(Q) = Y$, is to add some formulas to $Q$ as making $Q$ smaller can only increase $\dx{}(Q)$. This holds because postulating that, instead of $Q$, only a subset $Q'$ of $Q$ must be entailed by $\mo_i^*$, can cause a new KB $\mo_j^*$ for diagnosis $\md_j \notin\dx{}(Q)$ to entail $Q'$. However, as $Q$ is the set of \emph{all} entailments computable with \textsc{getEntailments} of KBs $\mo_i^*$ for $\md_i\in Y$, a superset $Q''$ of $Q$ computed by \textsc{getEntailments} with $\dx{}(Q'') = Y$ can never be obtained. Therefore, we have a contradiction to the precondition.

We have now proven the following: If there exists a q-partition as described in the proposition, then this q-partition is found in the iteration where the seed $\mS = Y$.  
%
%First observation: To yield such a q-partition, the seed $\mS$ must be a subset of $Y$ since at least all diagnoses in $\mS$ will be in $\dx{}$. Assume that $\mS = Y$ does not yield a query $Q$ with such a q-partition. This implies that $\dx{}(Q) \supset Y$.
%$Q$ is the set of entailments w.r.t. the getEntailments function that produces a minimal set $\dx{}(Q)$. 
%Assume that $\mS \subset Y$ yields a query $Q'$ with such a q-partition. ($Q' \supseteq Q$)
%Then $Q'$ must be an entailment of each diagnosis in $Y$, i.e.\ a subset of $Q$ by the completeness of getEntailments (i.e. this function computes all common entailments). ($Q \subseteq Q'$)
%=> $Q = Q'$
%This however implies that $\dx{}(Q') = \dx{}(Q)$ which is a contradiction to $\dx{}(Q) \supset Y$ and $\dx{}(Q') = Y$. 
\end{proof}
\begin{remark}\label{rem:ad_prop:query:dx_complete}
Regarding Proposition~\ref{prop:query:dx_complete}, note the following:
\begin{enumerate}[(a)]
	\item In fact, as one and the same q-partition must occur at most once in $\QP$, \textsc{getPoolOfQueries} must only keep assigning diagnoses in $\mD \setminus \mS$ to the respective sets of the q-partition as long as $\dx{} = \mS$. Because for $\dx{} = Z \supset \mS$, we know to find a query (if one exists) for the seed $\mS = Z$.
	\item A statement equivalent to the proposition is: If there is no query (including only entailments computed by the \textsc{getEntailments} function) with $\dx{} = Y$ found for seed $\mS = Y$, then such a query and q-partition, respectively, does not exist.\qed
\end{enumerate}
%(a)~In fact, as one and the same q-partition must occur at most once in $\QP$, \textsc{getPoolOfQueries} must only keep assigning diagnoses in $\mD \setminus \mS$ to the respective sets of the q-partition as long as $\dx{} = \mS$. Because for $\dx{} = Z \supset \mS$, we know to find a query (if one exists) for the seed $\mS = Z$.\\ 
%(b)~A statement equivalent to the proposition is: If there is no query with $\dx{} = Y$ found for seed $\mS = Y$, then such a query and q-partition, respectively, does not exist.
\end{remark}
The following proposition states that if a q-partition with one and the same set $\dx{}$ is found twice during the execution of \textsc{getPoolOfQueries}, then the queries for both q-partitions and thus both q-partitions must be equal. That is, for one set $\dx{}$, there is at most one tuple in $\QP$. 
\begin{proposition}\label{prop:query_gen_dx_unique}
Let $Q_i$ be a query with $\dx{}(Q_i) = Y$ in the set $\QP$ returned by \textsc{getPoolOfQueries} and found for seed $\mS_i = Y$ and
let $Q_j$ be a query with $\dx{}(Q_j) = Y$ in the set $\QP$ returned by \textsc{getPoolOfQueries} and found for some seed $\mS_j \subset Y$. Then $Q_i = Q_j$.  
\end{proposition}
\begin{proof} Let $Q'_i, Q'_j$ be the queries stored in the variable $Q$ in line~\ref{algoline:query:add_QP_start} for seeds $\mS_i$ and $\mS_j$, respectively; i.e. the supersets of the queries $Q_i,Q_j$ \emph{before} the minimization function \textsc{minQ} is called for each of them. $Q'_j \subseteq Q'_i$ holds by the fact that $Q'_i$ is the set of \emph{all} common entailments computable with \textsc{getEntailments} of $\mo_r^*$ for $\md_r \in Y$ and by the fact that $Q'_j$ must be a set of common entailments computed by \textsc{getEntailments} of exactly these KBs, because of $\dx{}(Q'_j) = Y$ and Definition~\ref{def:q-partition}. $Q'_j \supseteq Q'_i$ holds by the fact that $Q'_j$ is computed as intersection of $E_{\md_r}$ where $\md_r \in \mS_j$ and $Q'_i$ is computed as intersection of $E_{\md_s}$ where $\md_s \in \mS_i \supset \mS_j$. Thus, we can conclude that $Q'_i = Q'_j$.

As $Q'_i = Q'_j$, also $\Pt(Q'_i) = \Pt(Q'_j)$ must hold for the q-partitions by Proposition~\ref{prop:unique_q-partition}. That the minimized versions $Q_i, Q_j$ of $Q'_i, Q'_j$ output by \textsc{minQ} are equal, follows from the determinism of the \textsc{minQ} function, wherefore equal inputs, i.e.\ $(\emptyset,Q'_i,\emptyset,\Pt(Q'_i),\tuple{\mo,\mb,\Tp,\Tn}_\RQ) = (\emptyset,Q'_j,\emptyset,\Pt(Q'_j),\langle\mo,\mb,\Tp$, $\Tn\rangle_\RQ)$, must yield equal outputs.
\end{proof}
%
%\begin{proof} Let $Q'_i, Q'_j$ be the queries stored in the variable $Q$ in line~\ref{algoline:query:add_QP_start} for seeds $\mS_i$ and $\mS_j$, respectively; i.e. the supersets of the queries $Q_i,Q_j$ \emph{before} the minimization function \textsc{minQ} is called for each of them. $Q'_j \subseteq Q'_i$ holds by the fact that $Q'_i$ is the set of all common entailments computable with \textsc{getEntailments} of $\mo_r^*$ for $\md_r \in Y$ and by the fact that $Q'_j$ must be a common entailment computed by \textsc{getEntailments} of exactly these KBs, because of $\dx{}(Q'_j) = Y$. $Q'_j \supseteq Q'_i$ holds by the fact that $Q'_j$ is computed as intersection of $E_{\md_r}$ where $\md_r \in \mS_j$ and $Q'_i$ is computed as intersection of $E_{\md_s}$ where $\md_s \in \mS_i \supset \mS_j$. Thus, we can conclude that $Q'_i = Q'_j$.
%
%As $Q'_i = Q'_j$, also $\Pt(Q'_i) = \Pt(Q'_j)$ must hold for the q-partitions by Proposition~\ref{prop:unique_q-partition}. That the minimized versions $Q_i, Q_j$ of $Q'_i, Q'_j$ output by \textsc{minQ} are equal, follows from the determinism of the \textsc{minQ} function, wherefore equal inputs, i.e.\ $(\emptyset,Q'_i,\emptyset,\Pt(Q'_i),\tuple{\mo,\mb,\Tp,\Tn}_\RQ) = (\emptyset,Q'_j,\emptyset,\Pt(Q'_j),\tuple{\mo,\mb,\Tp,\Tn}_\RQ)$, must yield equal outputs.
%\end{proof}
\begin{remark}\label{rem:improvements_of_algo_QP_generation}
Proposition~\ref{prop:query_gen_dx_unique} hints at a possible improvement 
%means that an improvement 
of Algorithm~\ref{algo:query_gen}, namely to check in line~\ref{algoline:query:seed} whether the seed $\mS$ already occurs as a set $\dx{}$ in some tuple in $\QP$ and only continue the execution for $\mS$ if this does not hold (not shown in Algorithm~\ref{algo:query_gen}). In this vein, time and reasoning costs (line~\ref{algoline:query:dnx}) can be saved. 

Another improvement regarding line~\ref{algoline:query:seed} is to delete all remaining seeds $\mS'$ with the property $\mS' \supset \mS$ if $Q$ in line~\ref{algoline:query:common_ent} is the empty set (not shown in Algorithm~\ref{algo:query_gen}). Namely, all seeds $\mS'$ must also lead to $Q=\emptyset$ since the intersection of $E_{\md}$ for $\md\in\mS$ already returned $\emptyset$ wherefore the intersection of $E_{\md}$ for $\md\in\mS'$ must also return $\emptyset$.\qed
\end{remark}

By now, we know from Proposition~\ref{prop:query_gen_dx_unique} that, given a query with $\dx{}$ exists, one and only one q-partition with $\dx{}$ will be added to $\QP$, but which one?

W.r.t.\ one and the same set $\dx{}$, queries with a set $\dnx{}$ with higher cardinality are preferable over others as the cardinality of $\dz{}$ should be minimized (cf.\ Chapter~\ref{chap:UserInteraction}). So, preferable queries among those with equal set $\dx{}$ are those for which $\dnx{}$ is a set-maximal set. Exactly such a query is added to $\QP$ for each $\dx{}$ for which a query exists, as the following proposition shows.
\begin{proposition}\label{prop:given_dx_the_q-partition_with_min_dz_is_found}
If the set $\QP$ returned by \textsc{getPoolOfQueries} comprises a query $Q$ with $\dx{}(Q) = Y$, then $Q$ is a query with minimal $|\dz{}(Q)|$ among all queries $Q'$ with $\dx{}(Q') = Y$ computable with the function \textsc{getEntailments}.
\end{proposition} 
\begin{proof}
Assume that \textsc{getPoolOfQueries} finds a query $Q$ with $\dx{}(Q) = Y$ and $|\dz{}(Q)| = k$ and assume there is a query $Q'$ (consisting only of entailments computed by function \textsc{getEntailments}) with $\dx{}(Q') = Y$ and with $|\dz{}(Q')| < k$. This means that $|\dnx{}(Q)| < |\dnx{}(Q')|$. However, as $Q$ is computed for seed $\mS = Y$, $Q$ is a maximal set of entailments computable with \textsc{getEntailments} of $\mo_i^*$ for $\md_i \in Y$. Because $Q'$ is also a common entailment of $\mo_i^*$ for $\md_i \in Y$, we have that $Q' \subseteq Q$ must be true. 
%If $Q' = Q$, then $\dnx{}(Q') = \dnx{}(Q)$ and thus, by $\dx{}(Q') = \dx{}(Q)$, also $\dz{}(Q') = \dz{}(Q)$ which is a contradiction. 
%Otherwise, i.e.\ if $Q' \subset Q$, for each $\md_i \in \dnx{}(Q)$ it holds that either (a)~$\md_i \in \dnx{}(Q')$ or (b)~$\md_i \in \dz{}(Q')$ or (c)~$\md_i \in \dx{}(Q)$. 
Since the fact that $\mo_i^* \cup Q$ does not violate any $x\in\RQ\cup\Tn$, i.e.\ the fact that $\md_i \notin \dnx{}(Q)$, implies by monotonicity of $\mathcal{L}$ that $\mo_i^* \cup Q'$ for the subset $Q'$ of $Q$ cannot violate any $x\in\RQ\cup\Tn$ either, i.e.\ $\md_i \notin \dnx{}(Q')$, we conclude that $|\dnx{}(Q')| \leq |\dnx{}(Q)|$ must hold. This is a contradiction.
\end{proof}
%Since $|\dnx{}(Q)|$ can increase only for a proper superset $Q'$ of $Q$, we obtain a contradiction.  
%
%So, if $\dx{}$ is fixed, then the best q-partition to find for this $\dx{}$ is one with maximum $\dnx{}$, i.e. minimum $\dz{}$. Exactly such a q-partition is found.
%
%However, it cannot be granted that $\dz{}$ is the empty set, except for the setting where getEntailments function computes all explicit entailments.
%--> So, if a pool of a maximum number of queries should be computed, then go from low cardinality seeds to high cardinality ones --> this  

\section[Minimization of Queries]{Minimization of Queries%
\sectionmark{Query Minimization}}
\sectionmark{Query Minimization}
\label{sec:minimization_of_queries}
%%%%%%
%\section{Minimization of Queries}
%\label{sec:minimization_of_queries}
%%%%%%
%\noindent\textbf{\textsc{minQ}.} 
\paragraph{\textsc{minQ}.}
The minimization of the query $Q$ by \textsc{minQ} (see Algorithm~\ref{algo:query_gen}) while preserving the q-partition
% -- that is, the properties of $Q$ -- 
aims at simplifying the job of the answering user who only needs to go through a smaller set of logical formulas $Q_{\min}$ in order to come up with an answer to the query. Since the q-partition reflects the properties of a query w.r.t.\ the invalidation of (leading) diagnoses and two queries have equal such properties, then of course the one that is a subset of the other should be asked.

The concept of the function \textsc{minQ} is similar to the one of $\scQX$ (Algorithm~\ref{algo:qx}). Like $\scQX$, \textsc{minQ} carries out a divide-and-conquer strategy to find a set-minimal set with a monotonic property. In this case, the monotonic property is not the invalidity of a subset of the KB w.r.t.\ a DPI (as per Definition~\ref{def:valid_onto}) as it is for the computation of minimal conflict sets using $\scQX$, but the property of some $Q_{\min}\subset Q$ having the same q-partition as $Q$. So, the crucial difference between $\scQX$ and \textsc{minQ} is the function that checks this monotonic property. For \textsc{minQ}, this function -- that checks a subset of a query for constant q-partition -- is \textsc{isQPartConst}.	 

%\noindent\textbf{\textsc{minQ} -- Input Parameters.} 
\paragraph{\textsc{minQ} -- Input Parameters.}
\textsc{minQ} gets five parameters as input. The first three, namely $X, Q$ and $QB$, are relevant for the divide-and-conquer execution, whereas the last two, namely the original q-partition $\tuple{\dx{}, \dnx{}, \dz{}}$ of the query (i.e.\ the parameter $Q$) that should be minimized, and the DPI $\langle\mo,\mb,\Tp,\Tn\rangle_\RQ$ are both needed as an input to the function \textsc{isQPartConst}. Besides the latter two, another argument $QB$ is passed to this function where $QB$ is a subset of the original query $Q$. \textsc{isQPartConst} then checks whether the q-partition for the (potential) query $QB$ is equal to the q-partition $\tuple{\dx{}, \dnx{}, \dz{}}$ of the original query given as argument. The DPI is required as the parameters $\mo, \mb,\Tp,\Tn$ and $\RQ$ are necessary for these checks. 

%\noindent\textbf{\textsc{minQ} -- Testing Sub-Queries for Constant Q-Partition.} 
\paragraph{\textsc{minQ} -- Testing Sub-Queries for Constant Q-Partition.}
In particular, \textsc{isQPartConst} tests for each $\md_r\in\dnx{}$ whether $\mo_r^*\cup QB$ is valid (w.r.t.\ $\tuple{\cdot,\emptyset,\emptyset,N}_\RQ$). If so, this means that $\md_r \notin \dnx{}(QB)$ and thus that the q-partition of $QB$ is different to the one of $Q$ wherefore $\false$ is immediately returned. If $\true$ for all $\md_r\in\dnx{}$, it is tested for $\md_r\in\dz{}$ whether $\mo_r^* \models QB$. If so, this means that $\md_r \notin \dz{}(QB)$ and thus that the q-partition of $QB$ is different to the one of $Q$ wherefore $\false$ is immediately returned. If $\false$ is not returned for any $\md_r\in\dnx{}$ or $\md_r\in\dz{}$, then the conclusion is that $QB$ is a query w.r.t.\ to $\mD$ and $\langle\mo,\mb,\Tp,\Tn\rangle_\RQ$ and has the same q-partition as $Q$ wherefore the function returns $\true$. 

Note that, instead of calling a reasoner to answer whether $\mo_r^* \models QB$, the set of precalculated entailments $E_{\md_r}$ of $\mo_r^*$ for each $\md_r\in\mD$ can be given as an argument to \textsc{minQ} as well as to \textsc{isQPartConst} (not shown in Algorithm~\ref{algo:query_gen}). In this case an equivalent test is $QB \subseteq E_{\md_r}$. Such a strategy is particularly appropriate if reasoning is expensive for the DPI at hand.

Soundness of \textsc{isQPartConst} is proven by the following lemma.
\begin{lemma}\label{lem:isqpartconst_correct}
Let $\langle\mo,\mb,\Tp,\Tn\rangle_\RQ$ be a DPI, $\mD \subseteq \minD_{\langle\mo,\mb,\Tp,\Tn\rangle_\RQ}$, $Q\in\mQ_{\mD,\langle\mo,\mb,\Tp,\Tn\rangle_\RQ}$ with q-partition $\Pt(Q) = \tuple{\dx{}(Q),\dnx{}(Q),\dz{}(Q)}$. Then a non-empty set $QB \subset Q$ is a query in $\mQ_{\mD,\langle\mo,\mb,\Tp,\Tn\rangle_\RQ}$ with $\Pt(QB) = \Pt(Q)$ if 
\begin{enumerate}
	\item $\forall \md_r\in\dnx{}(Q): \; \mo_r^* \cup QB$ violates some $r\in\RQ$ or entails some $\tn\in\Tn$ and
	%\mo_r^* \cup QB$ does not violate any $r\in\RQ$ and does not entail any $\tn\in\Tn$ and
	\item $\forall \md_r\in\dz{}(Q): \; \mo_r^* \not\models QB$.
\end{enumerate}
\end{lemma}
\begin{proof}
Let $Q \in \mQ_{\mD,\langle\mo,\mb,\Tp,\Tn\rangle_\RQ}$ and $QB$ be an arbitrary proper subset of $Q$.
%A minimal diagnosis $\md_r$ w.r.t.\ $\langle\mo,\mb,\Tp,\Tn\rangle_\RQ$ is in $\dnx{}(Q)$ iff $\mo_r^* \cup Q$ violates some $r\in\RQ$ or entails some $\tn\in\Tn$. If we assume that $\mo_r^* \cup QB$ does not violate any $x\in\RQ\cup\Tn$, then (a)~$\mo_r^*$ does not entail $QB$
%%$\mo_r^* \cup QB$ meets all $r\in\RQ$ and does not entail any $\tn\in\Tn$ 
%or (b)~$\mo_r^*$ entails $QB$. If (a) is true, $\md_r \in \dz{}(QB)$ must hold; if (b) is true, then $\md_r \in \dx{}(QB)$ must hold. 
%%all $x\in\RQ\cup\Tn$ must be met 
%%and no $\tn\in\Tn$ can be entailed 
%%since $\mo_r^*$ does not include a conflict set by the definition of a diagnosis. 
%Thus, due to deletion of some sentences from $Q$, if not in $\dnx{}(QB)$, $\md_r \in \dnx{}(Q)$ can ``move'' to $\dz{}(QB)$ or to $\dx{}(QB)$. \\
If criterion 1) of this lemma is met, then we know that each diagnosis in $\dnx{}(Q)$ is in $\dnx{}(QB)$ as well, i.e.\ (I):~$\dnx{}(QB) \supseteq \dnx{}(Q)$ holds.

%A minimal diagnosis $\md_r$ w.r.t.\ $\langle\mo,\mb,\Tp,\Tn\rangle_\RQ$ is in $\dz{}(Q)$ iff $\mo_r^* \cup Q$ does not violate any $r\in\RQ$ and does not entail any $\tn\in\Tn$ and $\mo_r^*$ does not entail $Q$. Given that $\mo_r^* \cup Q$ does not violate any $r\in\RQ$ and does not entail any $\tn\in\Tn$, $\mo_r^* \cup QB$ cannot
%violate any $r\in\RQ$ and cannot entail any $\tn\in\Tn$ by monotonicity of $\mathcal{L}$. However, it is possible that $\mo_r^* \models QB$. Hence, due to deletion of some sentences from $Q$, if not in $\dz{}(QB)$, $\md_r \in \dz{}(Q)$ can ``move'' to $\dx{}(QB)$, but not to $\dnx{}(QB)$. \\
Assume a minimal diagnosis $\md_r\in\dz{}(Q)$. Then, $\mo_r^* \cup Q$ does not violate any $r\in\RQ$ and does not entail any $\tn\in\Tn$ and $\mo_r^*$ does not entail $Q$. This however implies that $\mo_r^* \cup QB$ cannot violate any $r\in\RQ$ and cannot entail any $\tn\in\Tn$ either by monotonicity of $\mathcal{L}$. But it is possible that $\mo_r^* \models QB$.
So, validity of criterion 2) of this lemma is sufficient to guarantee that each diagnosis in $\dz{}(Q)$ is in $\dz{}(QB)$ as well, i.e.\ (II):~$\dz{}(QB) \supseteq \dz{}(Q)$ holds.

As all diagnoses in $\dx{}(Q)$ entail all formulas in $Q$ by Definition~\ref{def:q-partition}, all diagnoses in $\dx{}(Q)$ must entail $QB$ as well. Consequently, due to deletion of some formulas from $Q$, no $\md_r \in \dx{}(Q)$ can ``move'' to any set $\dnx{}(QB)$ or $\dz{}(QB)$.
That is, (III):~$\dx{}(QB) \supseteq \dx{}(Q)$ must hold.

So, the overall conclusion is that, if criterion 1) and 2) are met, then (I), (II) and (III) hold. Assume that some $\supseteq$-relation in $i\in\setof{\text{(I), (II), (III)}}$ is a $\supset$-relation. This leads to a violation of some $j \in\setof{\text{(I), (II), (III)}}$ with $j\neq i$ since $\tuple{\dx{}(Q),\dnx{}(Q),\dz{}(Q)}$ and $\tuple{\dx{}(QB),\dnx{}(QB),\dz{}(QB)}$ are partitions of $\mD$. Therefore, all $\supseteq$-relations must be $=$-relations and we can derive that $\Pt(Q) = \Pt(QB)$.

Moreover, we have that $QB$ must be a query. This is due to the facts that $QB$ is non-empty, $Q$ is a query and the q-partitions of $Q$ and $QB$ are equal. Therefore, $\dx{}(QB) = \dx{}(Q) \geq 1$ and $\dnx{}(QB) = \dnx{}(Q) \geq 1$ which lets us conclude by Proposition~\ref{prop:query_dx_dnx} that $QB$ is a query.
%where can a diagnosis ``move'' if $Q$ shrinks? --> from D- to D+ or D0 and from D0 to D+ AND can never move to D- from elsewhere and no diag can move from D+ to anywhere, thus in particular not to D0.
\end{proof}

%\noindent\textbf{\textsc{minQ} -- The Divide-and-Conquer Strategy.} 
\paragraph{\textsc{minQ} -- The Divide-and-Conquer Strategy.}
Intuitively, \textsc{minQ} partitions the given query $Q$ in two parts $Q_1$ and $Q_2$ and first analyzes $Q_2$ while $Q_1$ is part of $QB$ (line~\ref{algoline:query:recursive_call1}). Note that in each iteration $QB$ is the subset of $Q$ that is currently assumed to be part of the sought minimized query (i.e.\ the one query that will finally be output by \textsc{minQ}). In other words, analysis of $Q_2$ while $Q_1$ is part of $QB$ means that all irrelevant formulas in $Q_2$ should be located and removed from $Q_2$ resulting in $Q^{\min}_2 \subseteq Q_2$. That is, $Q^{\min}_2$ must include only relevant formulas which means that $Q^{\min}_2$ along with $QB$ is a query with an equal q-partition as $Q$, but the deletion of any further formula from $Q^{\min}_2$ changes the q-partition.

After the relevant subset $Q^{\min}_2$ of $Q_2$, i.e.\ the subset that is part of the minimized query, has been returned, $Q_1$ is removed from $QB$, $Q^{\min}_2$ is added to $QB$ and $Q_1$ is analyzed for a relevant subset that is part of the minimized query (line~\ref{algoline:query:recursive_call2}). This relevant subset, $Q^{\min}_1$, together with $Q^{\min}_2$, then builds a set-minimal subset of the input $Q$ that is a query and has a q-partition equal to that of $Q$. Note that the argument $X$ of \textsc{minQ} is the subset of $Q$ that has most recently been added to $QB$.

For each call in line~\ref{algoline:query:recursive_call1} or line~\ref{algoline:query:recursive_call2}, the input $Q$ to \textsc{minQ} is recursively analyzed until a trivial case arises, i.e.\ (a)~until $Q$ is identified to be irrelevant for the computed minimized query wherefore $\emptyset$ is returned (lines~\ref{algoline:query:validitytest2} and \ref{algoline:query:return_emptyset}) or (b)~until $|Q|=1$ and $Q$ is not irrelevant for the computed minimized query wherefore $Q$ is returned (lines~\ref{algoline:query:test_singleton} and \ref{algoline:query:return_Q}).

\begin{example}\label{example:ad_Table_of_queries_partitions}
Let us reconsider the FOL DPI depicted by Table~\ref{tab:example1} on page~\pageref{tab:example1}. 
We recall that sets of minimal conflict sets and minimal diagnoses w.r.t.\ this DPI were given by $\minC_{\tuple{\mo,\mb,\Tp,\Tn}_\RQ} = \setof{\mc_1,\mc_2} = \setof{\tuple{1,3,4},\tuple{1,2,3,5}}$ as well as $\minD_{\tuple{\mo,\mb,\Tp,\Tn}_\RQ} = \setof{\md_1,\md_2,\md_3,\md_4} = \setof{[1],[3],[4,5],[2,4]}$. For this DPI, a set of minimized queries computed by \textsc{getPoolOfQueries} is presented by Table~\ref{tab:queries_partitions}. Note that these queries have been produced by different \textsc{getEntailments} functions (as indicated by the dashed lines in Table~\ref{tab:queries_partitions}). That is, $Q_i$ for $i \in \setof{1,\dots,5}$ have been produced by the same \textsc{getEntailments} function that is described in Example~\ref{example:query_computation}. For $i \in \setof{6,\dots,9}$, $Q_i$ has been computed from a \textsc{getEntailments} function that outputs only explicit entailments (cf.\ Definition~\ref{def:explicit_entailment}) and $Q_{10}$ from a \textsc{getEntailments} function that returns a finite set of entailments where each entailment is some FOL formula. This could be accomplished, for example, by some resolution-based reasoning procedure \cite{chang1973}.

It is important to realize that the results regarding Algorithm~\ref{algo:query_gen} established so far, most of which depend on the particular used \textsc{getEntailments} function, must only hold within one part of Table~\ref{tab:queries_partitions} (where different parts are separated by the dashed lines). For example, for $Q_2$ and $Q_9$ it holds that $\dx{}(Q_2) = \dx{}(Q_9)$, but $\dnx{}(Q_2) \neq \dnx{}(Q_9)$ and $\dz{}(Q_2) \neq \dz{}(Q_9)$. By application of one and the same \textsc{getEntailments} function, this case would be prohibited by Proposition~\ref{prop:query_gen_dx_unique}. Furthermore, by Proposition~\ref{prop:given_dx_the_q-partition_with_min_dz_is_found}, only $Q_9$ would be an element of the query pool $\QP$ in this case since $\dz{}(Q_9) \subset \dz{}(Q_2)$.

Moreover, we want to remark that $Q_7$, $Q_8$ and $Q_9$ can be seen as a proof that $Q_6$ is indeed set-minimal. Each $Q_i, i \in \setof{7,8,9}$ is a result of the removal of a single formula from $Q_6$. And, each such $Q_i$ features a q-partition different from the one of $Q_6$. This illustrates quite well the principle of \textsc{minQ} which performs tests of exactly this kind to verify minimality of a query or detect formulas that might be deleted from it under preservation of the q-partition, respectively.

Another essential note is that it is guaranteed that $\dz{}(Q_6) = \emptyset$. This holds due to the construction of $Q_6$ as $U_{\mD}\setminus \md_4 = \setof{1,2,3,4,5}\setminus [2,4] = \setof{1,3,5}$ (recall that we use squared brackets to denote diagnoses in spite of the fact that these are sets, cf.\ Table~\ref{tab:abbreviations}). So, $Q_6$ comprises all formulas occurring in minimal diagnoses except for the ones contained in $\md_4$. We have that for any two different minimal diagnoses $\md_i, \md_j$ w.r.t.\ one and the same DPI it must be true that $\md_i \setminus \md_j \neq \emptyset$ as well as $\md_j \setminus \md_i \neq \emptyset$ as otherwise one would be necessarily a subset of the other. From this, we can easily derive that $\mo^*_i \cup Q_6$ for $i \in \setof{1,\dots,3}$, i.e.\ for all minimal diagnoses $\md_i$ w.r.t.\ this DPI other than $\md_4$ which was used to build the query $Q_6$, must comprise a conflict set. This must be valid by the minimality of $\md_i$ and since by $Q_6$ at least one formula of $\md_i$ is readded to the KB. Note that a similar argumentation was used in the proof of Proposition~\ref{prop:query_gen_explicit_entailments_dz_empty}.\qed
\end{example}

\renewcommand{\arraystretch}{1.4} 
%\begin{table*}
	%\centering
		%\rowcolors[]{2}{gray!8}{gray!16} %\arrayrulecolor{black}
		%%\extrarowheight10pt
		%\begin{tabular}{ c c c c	} 
		%%\hline\hline
			%\rowcolor{gray!40}
			%\toprule\addlinespace[0pt]
			%$i$ & $\tax_i$ & $\mo$ & $\mb$  \\ \addlinespace[0pt]\midrule\addlinespace[0pt]
			%%\hline
			%1 & $\forall X a_1(X) \;\rightarrow\; a_2(X) \land m_1(X) \land m_2(X)$ & $\bullet$ & 	\\
			%%\hline
			%2 & $\forall X a_2(X) \;\rightarrow\; \lnot(\exists Y s(X,Y) \land m_3(Y)) \land \exists Z s(X,Z) \land m_2(Z)$ & $\bullet$ &  	\\
			%%\hline
			%3 & $\forall X m_1(X) \;\rightarrow\; \lnot a(X) \land b(X)$ & $\bullet$ &  	\\
			%%\hline
			%4 & $\forall X m_2(X) \;\rightarrow\; (\forall Y s(X,Y) \rightarrow a(Y)) \land d(X)$ & $\bullet$ & 	\\
			%%\hline
			%5 & $\forall X m_3(X) \;\leftrightarrow\; b(X) \lor c(X)$ & $\bullet$ & 	\\
			%%\hline
			%6 & $a_1(w)$ &  & $\bullet$   \\
			%%\hline
			%7 & $a_1(u)$ &  & $\bullet$  	\\
			%%\hline
			%8 & $s(u,w)$ &  & $\bullet$  \\ 
			%%\hline\hline
			%\addlinespace[0pt]\bottomrule
			%\rowcolor{gray!40}
			%$i$ & \multicolumn{3}{c}{$\tp_i\in\Tp$} \\ \addlinespace[0pt]\midrule\addlinespace[0pt]
			%$\times$ & \multicolumn{3}{c}{$\times$} 	\\ \addlinespace[0pt]\toprule\addlinespace[0pt]
			%\rowcolor{gray!40}
			%$i$ & \multicolumn{3}{c}{$\tn_i\in\Tn$} \\ \addlinespace[0pt]\midrule\addlinespace[0pt]
			%$\times$ & \multicolumn{3}{c}{$\times$} 	\\ \addlinespace[0pt]\toprule\addlinespace[0pt]
			%\rowcolor{gray!40}
			%\multicolumn{4}{c}{$r\in\RQ$} \\ \addlinespace[0pt]\midrule\addlinespace[0pt]
			%\multicolumn{4}{c}{consistency, coherency} \\ \addlinespace[0pt]\bottomrule
		%\end{tabular}
	%\caption{Example 1: DPI}
	%\label{tab:example1}
%\end{table*}
\begin{table*}[!htb]
\footnotesize
\centering
\begin{tabular}{@{\extracolsep{2pt}}lllll}
$i$ & Query $Q_i$ & $\dx{}(Q_i)$  & $\dnx{}(Q_i)$ &  $\dz{}(Q_i)$ \\ \hline
$1$ & $\{\forall X b(X) \rightarrow m_3(X)\}$ & $\{\md_1,\md_2,\md_4\}$ & $\{\md_3\}$ & $\emptyset$\\
$2$ & $\{b(w)\}$ & $\{\md_3, \md_4\}$ & $\{\md_2\}$ & $\{\md_1\}$ \\
$3$ & $\{\forall X m_1(X) \rightarrow b(X)\}$ & $\{\md_1,\md_3,\md_4\}$ & $\{\md_2\}$ & $\emptyset$\\
$4$ & $\{m_1(w), m_2(u)\}$ & $\{\md_2,\md_3,\md_4\}$ & $\{\md_1\}$ & $\emptyset$  \\
$5$ & $\{a(w)\}$ & $\{\md_2\}$ & $\{\md_3, \md_4\}$ & $\{\md_1\}$\\ \hdashline
$6$ & $\{\tax_1,\tax_3,\tax_5\}$ & $\{\md_4\}$ & $\{\md_1,\md_2, \md_3\}$ & $\emptyset$\\
$7$ & $\{\tax_3,\tax_5\}$ & $\{\md_1, \md_4\}$ & $\{\md_2,\md_3\}$ & $\emptyset$\\
$8$ & $\{\tax_1,\tax_5\}$ & $\{\md_2, \md_4\}$ & $\{\md_1,\md_3\}$ & $\emptyset$\\
$9$ & $\{\tax_1,\tax_3\}$ & $\{\md_3, \md_4\}$ & $\{\md_1,\md_2\}$ & $\emptyset$\\ \hdashline
\multirow{2}{*}{$10$} & $\{\forall X m_1(X) \rightarrow \lnot a(X),$ & \multirow{2}{*}{$\{\md_1\}$} & \multirow{2}{*}{$\{\md_2,\md_3, \md_4\}$} & \multirow{2}{*}{$\emptyset$}\\
    & $\forall X m_2(X) \rightarrow (\forall Y s(X,Y) \rightarrow a(Y))\}$ & \\
%$6$ & $\{\forall X m_2(X)\rightarrow d(X)\}$ & $\{\md_1,\md_2\}$ & $\emptyset$ & $\{\md_3, \md_4\}$\\
%$7$ & $\{m_3(u)\}$ & $\{\md_4\}$ & $\emptyset$ & $\{\md_1, \md_2, \md_3\}$\\
\hline
\end{tabular}
\caption[Queries and Associated Q-Partitions]{Some queries and associated q-partitions for the DPI given by Table~\ref{tab:example1}.}
\label{tab:queries_partitions}
    %\vspace{-15pt}
\end{table*}

%\noindent\textbf{Soundness of query minimization.} 
\section{Soundness of Query Minimization}
\label{sec:SoundnessOfQueryMinimization}
The following lemma shows that the function \textsc{isQPartConst} used by \textsc{minQ} is indeed a monotonic function (cf.\ Definition~\ref{def:monotonic}), which is a necessary prerequisite for versions of the $\scQX$ algorithm to work in a sound way. 
\begin{lemma}\label{lem:isqpartconst_monotonic}
Let $\langle\mo,\mb,\Tp,\Tn\rangle_\RQ$ be a DPI, $\mD \subseteq \minD_{\langle\mo,\mb,\Tp,\Tn\rangle_\RQ}$, $Q\in\mQ_{\mD,\langle\mo,\mb,\Tp,\Tn\rangle_\RQ}$ with q-partition $\Pt(Q)$. Further, let $f: 2^Q \rightarrow \setof{0,1}$ be a function that maps a subset $QB$ of $Q$ to 1 if $QB$ has q-partition $\Pt(QB) = \Pt(Q)$, to 0 otherwise. Then, $f$ is a monotonic function (as per Definition~\ref{def:monotonic}).
\end{lemma}
\begin{proof}
Assume a subset $Q'$ of $Q$ with $f(Q') = 1$, i.e.\ $Q'$ has q-partition $\Pt(Q') = \Pt(Q)$. Let $Q' \subset Q'' \subseteq Q$ and assume that $f(Q'') = 0$, i.e.\ $Q''$ has a q-partition $\Pt(Q'') \neq \Pt(Q)$. 

As shown in the proof of Lemma~\ref{lem:isqpartconst_correct}, $\dx{}(X_1) \supseteq \dx{}(X_2)$ holds for any $X_1 \subseteq X_2$. Therefore, we have $\dx{}(Q') \supseteq \dx{}(Q'') \supseteq \dx{}(Q)$ and by $\Pt(Q') = \Pt(Q)$ that $\dx{}(Q') = \dx{}(Q)$ and thus that all $\supseteq$-relations are $=$-relations. So, either $\dnx{}(Q'') \neq \dnx{}(Q)$ or $\dz{}(Q'') \neq \dz{}(Q)$ must hold.
%First, assume that $\dx{}(Q'') \neq \dx{}(Q)$. Then, $\dx{}(Q'') \supset \dx{}(Q)$ must hold as $\mo_r^* \models Q$ means that $\mo_r^* \models Q''$ because of $Q'' \subseteq Q$ and the monotonicity of $\mathcal{L}$. 

First, assume that $\dnx{}(Q'') \neq \dnx{}(Q)$. Then, as $\mo_r^* \cup Q'' \subset \mo_r^* \cup Q$ and by monotonicity of $\mathcal{L}$, it can only be the case that for some $\md_r \in \mD$ some $x \in \RQ \cup \Tn$ that is violated for $\mo_r^* \cup Q$ is \emph{not} violated for $\mo_r^* \cup Q''$. Hence, $\dnx{}(Q'') \subset \dnx{}(Q)$ must hold. By a similar argumentation -- \emph{without} the assumption that $\dnx{}(Q') \neq \dnx{}(Q'')$ holds -- we have that $\dnx{}(Q') \subseteq \dnx{}(Q'')$ and thus, altogether, that $\dnx{}(Q') \subset \dnx{}(Q)$ must be true. Due to $\Pt(Q') = \Pt(Q)$ we know that $\dnx{}(Q') = \dnx{}(Q)$ which is a contradiction.
% to the fact that $\dnx{}(Q') \subset \dnx{}(Q)$

Finally, assume that $\dz{}(Q'') \neq \dz{}(Q)$. Since $\mo_r^* \cup Q$ does not violate any $x \in \RQ \cup \Tn$ for $\md_r\in\dz{}(Q)$, $\mo_r^* \cup Q''$ cannot violate any $x \in \RQ \cup \Tn$ by monotonicity of $\mathcal{L}$. As a conclusion, the only possibility for $\dz{}(Q'') \neq \dz{}(Q)$ is that $\mo_r^* \models Q''$ for some $\md_r\in\dz{}(Q)$, i.e.\ that $\md_r \in \dx{}(Q'')$ which implies that $\dz{}(Q'') \subset \dz{}(Q)$. By a similar argumentation -- \emph{without} the assumption that $\dz{}(Q') \neq \dz{}(Q'')$ holds -- we have that $\dz{}(Q') \subseteq \dz{}(Q'')$ and thus, altogether, that $\dz{}(Q') \subset \dz{}(Q)$ must be true. Due to $\Pt(Q') = \Pt(Q)$ we know that $\dz{}(Q') = \dz{}(Q)$ which is a contradiction.  
%As shown in the proof of Lemma~\ref{lem:isqpartconst_correct}, it can only hold that $\dnx{}(Q'') \supset \dnx{}(Q)$ 

This completes the proof for monotonicity of the given function $f$.
\end{proof} 
\begin{proposition}[Correctness of \textsc{minQ}]\label{prop:minQ_correctness}
Given a query $Q\in\mQ_{\mD,\tuple{\mo,\mb,\Tp,\Tn}_\RQ}$ as input, \textsc{minQ} computes a subset $Q_{\min} \subseteq Q$ such that $\Pt(Q_{\min}) = \Pt(Q)$ and there is no $Q' \subset Q_{\min}$ such that $\Pt(Q') = \Pt(Q)$. 
\end{proposition}
\begin{proof}
This proposition is a consequence of the correctness of $\scQX$ shown by Proposition~\ref{prop:qx_correctness}, of the correctness of function \textsc{isQPartConst} established by Lemma~\ref{lem:isqpartconst_correct} and of the monotonicity of the property tested by the function \textsc{isQPartConst} guaranteed by Lemma~\ref{lem:isqpartconst_monotonic}.
\end{proof}

%More concretely, all entailments of certain types, e.g.\ classification and realization for description logics or atoms for propositional logic, are computed for each $\mo_i^*$

\section{Complexity of Query Pool Generation}
\label{sec:query_gen_complexity}
The complexity of query minimization, i.e.\ one call to \textsc{minQ}, in terms of calls to the \textsc{isQPartConst} function is directly obtained from the complexity results for the standard $\scQX$ algorithm given by Proposition~\ref{prop:qx_complexity}.
\begin{proposition}[Complexity of \textsc{minQ}]\label{prop:minQ_complexity}
Let $\langle\mo,\mb,\Tp,\Tn\rangle_\RQ$ be a DPI, $\mD\subseteq\minD_{\langle\mo,\mb,\Tp,\Tn\rangle_\RQ}$, $Q \in \mQ_{\mD,\langle\mo,\mb,\Tp,\Tn\rangle_\RQ}$ with $\Pt(Q) = \tuple{\dx{}(Q), \dnx{}(Q), \dz{}(Q)}$ and the function \textsc{split} (line~\ref{algoline:query:split} of Algorithm~\ref{algo:query_gen}) be defined as $\textsc{split}(n) = \lfloor \frac{n}{2}\rfloor$ where $n$ is a natural number. Then, the worst case number of calls to \textsc{isQPartConst} during one call to $\textsc{minQ}(\emptyset,Q,\emptyset,\Pt(Q),\langle\mo,\mb,\Tp,\Tn\rangle_\RQ)$ is in 
\begin{align*}
O\left(\left|Q_{\min}\right|\log \frac{|Q|}{\left|Q_{\min}\right|}\right)
\end{align*}
%$O(|Q_{\min}|\log \frac{|Q|}{|Q_{\min}|})$ 
where $Q_{\min}$ is the output of $\textsc{minQ}(\emptyset,Q,\emptyset,\Pt(Q),\langle\mo,\mb,\Tp,\Tn\rangle_\RQ)$.

For any other definition of the function \textsc{split}, the worst case number of calls to \textsc{isQPartConst} gets larger.
\end{proposition}
%
%--> complexity of function minQ is $O(|Q_{\min}|\log(\frac{|Q|}{|Q_{\min}|}))$, cf. Proposition~\ref{prop:qx_complexity}.
The overall complexity of \textsc{getPoolOfQueries} in terms of calls to functions that call the reasoner, i.e.\ functions \textsc{getEntailments}, \textsc{isKBValid} and \textsc{isQPartConst}, is established by the following proposition.
\begin{proposition}[Complexity of \textsc{getPoolOfQueries}]\label{prop:query_gen_complexity}
Let $\tuple{\mo,\mb,\Tp,\Tn}_\RQ$ be a DPI, $q$ a natural number and $\mD \subseteq \minD_{\langle\mo,\mb,\Tp,\Tn\rangle_\RQ}$. Then, the worst case number of calls to functions that call a reasoner during one call to $\textsc{getPoolOfQueries}(\langle\mo,\mb,\Tp,\Tn\rangle_\RQ, \mD, q)$ is in 
\begin{align*}
O\left(\left(|\mD|+\left|Q_{\min}^{(\max)}\right|\log \frac{\left|Q^{(\max)}\right|}{\left|Q_{\min}^{(\max)}\right|}\right)2^{|\mD|}\right)
\end{align*}
%$O((|\mD|+|Q_{\min}^{(\max)}|\log \frac{|Q^{(\max)}|}{|Q_{\min}^{(\max)}|})2^{|\mD|})$ 
where $\left|Q^{(\max)}\right|$ is the maximum size of a query before minimization, i.e.\ the size of the set of maximum cardinality that is stored in variable $Q$ in line~\ref{algoline:query:minQ} throughout all iterations, and $\left|Q_{\min}^{(\max)}\right|$ is the maximum size of a minimized query, i.e.\ the size of the set of maximum cardinality that is stored in variable $Q'$ in line~\ref{algoline:query:minQ} throughout all iterations.
\end{proposition}
\begin{proof}
During the execution of the for-loop over lines~\ref{algoline:query:ent_start}-\ref{algoline:query:ent_end} the function \textsc{getEntailments} is called $|\mD|$ times. During the execution of the for-loop over lines~\ref{algoline:query:seed}-\ref{algoline:query:return_QP_1} which may be executed at most $2^{|\mD|}-2$ times, \textsc{isKBValid} is called at most $|\mD|-1$ times since $|\mS| \geq 1$ and $\mS \subset \mD$ and thus $|\mD\setminus\mS| \leq |\mD|-1$ holds; furthermore, \textsc{minQ} may be called once, namely if the condition tested by the if-statement in line~\ref{algoline:query:add_QP_start} is true. During one execution of \textsc{minQ}, by Proposition~\ref{prop:minQ_complexity}, at most 
\begin{align*}
|Q_{\min}|\log \frac{|Q|}{|Q_{\min}|}
\end{align*}
%$|Q_{\min}|\log \frac{|Q|}{|Q_{\min}|}$ 
calls to \textsc{isQPartConst} are made where $Q_{\min}$ is the output of the call to $\textsc{minQ}$. So, an upper bound of the number of calls to \textsc{isQPartConst} performed by one call to \textsc{minQ} among all calls to \textsc{minQ} throughout the execution of \textsc{getPoolOfQueries}, is 
\begin{align*}
\left|Q_{\min}^{(\max)}\right|\log \frac{\left|Q^{(\max)}\right|}{\left|Q_{\min}^{(\max)}\right|}
\end{align*}
%$|Q_{\min}^{(\max)}|\log \frac{|Q^{(\max)}|}{|Q_{\min}^{(\max)}|}$ 
where $\left|Q_{\min}^{(\max)}\right|$ is the set of maximum cardinality that is stored in variable $Q'$ in line~\ref{algoline:query:minQ} throughout all iterations and $\left|Q^{(\max)}\right|$ is the set of maximum cardinality that is stored in variable $Q$ in line~\ref{algoline:query:minQ} throughout all iterations. 

So, all in all we know that functions that call a reasoner are invoked at most 
\begin{align*}
|\mD| + \left(|\mD|-1 + \left|Q_{\min}^{(\max)}\right|\log \frac{\left|Q^{(\max)}\right|}{\left|Q_{\min}^{(\max)}\right|}\right) (2^{|\mD|} - 2)
\end{align*}
%$|\mD| + (|\mD|-1 + |Q_{\min}^{(\max)}|\log \frac{|Q^{(\max)}|}{|Q_{\min}^{(\max)}|}) (2^{|\mD|} - 2)$ 
times during the execution of \textsc{getPoolOfQueries}. Since 
\begin{align*}
\left(|\mD|+\left|Q_{\min}^{(\max)}\right|\log \frac{\left|Q^{(\max)}\right|}{\left|Q_{\min}^{(\max)}\right|}\right)2^{|\mD|}
\end{align*}
%$(|\mD|+|Q_{\min}^{(\max)}|\log \frac{|Q^{(\max)}|}{|Q_{\min}^{(\max)}|})2^{|\mD|}$ 
is an upper bound of this number, the proposition holds.
\end{proof}
Note that none of the parameters that affect the complexity of the function \textsc{getPoolOfQueries} grows with the size of the DPI provided as an input to the interactive KB debugging problem. Merely the costs for reasoning, where a black-box debugging approach has no influence on, are affected by a higher complexity or larger size of the input DPI. Moreover, the size of the most relevant parameter influencing the worst case complexity, namely the exponent $|\mD|$, can be specified by the user to any value greater or equal to 2. In other words, minus reasoning time, the generation of a pool of queries is a fixed parameter tractable problem~\cite{downey1995} in the context of interactive KB debugging.

\section{Shortcomings of Query Pool Generation}
\label{sec:query_gen_shortcomings}
First, the exponential time complexity regarding the parameter $|\mD|$ is a problem arising from the paradigm of computing an optimal query w.r.t.\ a certain quantitative measure $qsm()$ such as information gain~\cite{Shchekotykhin2012, Rodler2013} by calculating a (generally exponentially large) pool $\QP$ of queries in a first stage, whereupon $qsm(Q) \in \mathbb{R}$ is evaluated for $Q\in\QP$ until the one $Q^*$ with optimal $qsm(Q^*)$ is found and selected as the query to be asked to the user.
 
A key to solving this issue is the use of a different paradigm that does not rely on the computation of the pool $\QP$. Instead, qualitative measures can be derived from quantitative measures that have been used in interactive debugging scenarios~\cite{Shchekotykhin2012, Rodler2013, ksgf2010}. These qualitative measures provide a way to estimate the $qsm()$ value of \emph{partial q-partitions}, i.e.\ ones where not all leading diagnoses have been assigned to the respective set in the q-partition yet. That way a \emph{direct} search for a query with (nearly) optimal properties is possible. A similar strategy called CKK has been employed in~\cite{Shchekotykhin2012} for the information gain measure (see Section~\ref{sec:query_selection_measures}). From such a technique we can expect to save a high number of reasoner calls. Because only a usually small subset of q-partitions included in the pool computed by \textsc{getPoolOfQueries} is required to find a query with desirable properties if the search is implemented by means of a heuristic that involves the exploration of seemingly favorable (potential) queries and (partial) q-partitions, respectively, first. This is a topic of future work.

Another shortcoming of \textsc{getPoolOfQueries} is the extensive use of reasoning services which may be computationally expensive (depending on the given DPI). Instead of computing a set of common entailments $Q$ of a set of KBs $\mo_i^*$ first and consulting a reasoner to fill up the (q-)partition for $Q$ in order to test whether $Q$ is a query at all, the idea enabling a significant reduction of reasoner dependence is to compute some kind of \emph{canonical query} without a reasoner and use simple set comparisons to decide whether the associated partition is a q-partition. Guided by qualitative properties mentioned before, a search for such q-partition with desirable properties can be accomplished \emph{without reasoning at all}. Also, a set-minimal version of the optimal canonical query can be computed without reasoning aid. Only for the optional enrichment of the identified optimal canonical query by additional entailments and for the subsequent minimization of the enriched query, the reasoner may be employed. This is also a topic of future work.

Another aspect that can be improved is that \emph{only one} minimized version of each query is computed by Algorithm~\ref{algo:query_gen}. That is, per q-partition $\Pt$, there might be some set-minimal queries which do not occur in the output set $\QP$. From the point of view of how well a query might be understood by an interacting user, of course not all minimized queries can be assumed equally good in general. Hence, in order to avoid a situation where a potentially best-understood query w.r.t.\ $\Pt$ is not included in $\QP$, the query minimization process (see Section~\ref{sec:minimization_of_queries}) might be adapted to take into account some information about faults the interacting user is prone to. This could be exploited to estimate how well this user might be able to understand and answer a query. For instance, given that the user frequently has problems to apply $\exists$ in a correct manner to express what they intend to express, but has never made any mistakes in formulating implications $\rightarrow$, then the query $Q_1 = \setof{\forall X\,p(X) \rightarrow q(X), r(a)}$ might be better comprehended than $Q_2 = \setof{\forall X \exists Y s(X,Y)}$. One way to achieve the finding of a well-understood query for some q-partition $\Pt$ is to run the query minimization \textsc{minQ} more than once, each time with a modified input (using a hitting set tree to accomplish this in a systematic manner -- cf.\ Chapter~\ref{chap:DiagnosisComputation}, where an analogue idea is used to compute different minimal conflict sets w.r.t.\ a DPI). In this way, different set-minimal queries for $\Pt$ can be identified and the process can be stopped when a suitable query is found.     
\section{Correctness of Query Pool Generation}
\label{sec:query_gen_correctness}
The following proposition confirms the correctness of Algorithm~\ref{algo:query_gen}, i.e.\ of the function \textsc{getPoolOfQueries}. Roughly, it states that the output of $\QP$ of the function is duplicate-free, i.e.\ no query or q-partition occurs twice in $\QP$, that $\QP$ includes \emph{only queries} and q-partitions, that tuples in $\QP$ are unique w.r.t.\ the set $\dx{}$ of a q-partition and that, given $q > |\QP|$, there is no subset $Y$ of $\mD$ for which a q-partition with $\dx{} = Y$ exists and for which no q-partition with $\dx{} = Y$ is an element of $\QP$. 
\begin{proposition}\label{prop:getPoolOfQueries_correctness}
Let a DPI $\tuple{\mo,\mb,\Tp,\Tn}_\RQ$, $\mD \subseteq \minD_{\tuple{\mo,\mb,\Tp,\Tn}_\RQ}$ such that $|\mD| \geq 2$ and some $q \in \mathbb{N} \cup \setof{\infty}, q \geq 1$  be the inputs to \textsc{getPoolOfQueries} and let $|\QP_{\max}| \geq 0$ be the maximum number of tuples $\tuple{Q,\Pt(Q)}$ that can be computed by \textsc{getPoolOfQueries} by means of the used \textsc{getEntailments} function. If $q \geq |\QP_{\max}|$ (in particular $q=\infty$), then 
\begin{enumerate}
\item there are no two tuples $\tuple{Q,\Pt(Q)}, \tuple{Q',\Pt(Q')}$ in $\QP$ such that $Q = Q'$ \emph{or} $\Pt(Q)=\Pt(Q')$, and
\item $\QP$ includes a tuple $\tuple{Q,\tuple{\dx{}(Q),\dnx{}(Q),\dz{}(Q)}}$ only if $Q\in\mQ_{\mD,\tuple{\mo,\mb,\Tp,\Tn}_\RQ}$, and  
\item $\QP$ includes at most one tuple where $\dx{}(Q) = Y$ for each $Y \subset \mD$, and 
\item for each $Y \subset \mD$ for which a query $Q$ w.r.t.\ $\mD$ and $\tuple{\mo,\mb,\Tp,\Tn}_\RQ$ exists such that 
\begin{enumerate}
	\item $Q$ includes only entailments computed by the used \textsc{getEntailments} function and
	\item $\Pt(Q)$ is such that $\dx{}(Q) = Y$,
\end{enumerate}
$\QP$ includes a tuple $\tuple{Q',\Pt(Q')}$ such that $\dx{}(Q') = Y$, and
%\item for each $Y \subset \mD$ for which a q-partition $\Pt(Q)$ with $\dx{}(Q) = Y$ exists w.r.t.\ $\mD$ and $\tuple{\mo,\mb,\Tp,\Tn}_\RQ$, $\QP$ includes a tuple $\tuple{Q,\Pt(Q)}$, and
\item $\QP \neq \emptyset$.
\end{enumerate}
If $q < |\QP_{\max}|$, then $\QP$ includes $q$ tuples satisfying (1), (2) and (3).
%$\textsc{getPoolOfQueries}(\langle\mo,\mb,\Tp,\Tn\rangle_\RQ, \mD, q)$ returns (a)~a set $\QP$ of $q$ different queries $Q\in\mQ_{\mD,\tuple{\mo,\mb,\Tp,\Tn}_\RQ}$ and associated q-partitions $\Pt(Q)$, if $q$ different queries w.r.t.\ $\mD$ and $\tuple{\mo,\mb,\Tp,\Tn}_\RQ$ exist. Otherwise, (b)~a set $\QP$ including exactly one query $Q\in\mQ_{\mD,\tuple{\mo,\mb,\Tp,\Tn}_\RQ}$ and associated q-partition with $\dx{} = Y$ for each $Y \subset \mD$ for which such a q-partition exists is returned.
\end{proposition}
\begin{proof}
Statement~(1) is a consequence of Proposition~\ref{prop:query_gen_duplicate_free}. Statement~(2) is an implication of Proposition~\ref{prop:query_gen_isQuery_correct} and Proposition~\ref{prop:minQ_correctness}. The former says that only sets $Q$ that are actually queries w.r.t.\ $\mD$ and $\tuple{\mo,\mb,\Tp,\Tn}_\RQ$ can pass line~\ref{algoline:query:add_QP_start}. Thus, only queries are passed to \textsc{minQ} as parameter $Q$. By the latter which states that \textsc{minQ} is correct, i.e.\ outputs a query if the input is a query, statement~(2) follows. Statement~(3) follows from Proposition~\ref{prop:query_gen_dx_unique}. 
If $q \geq |\QP_{\max}|$, the truth of statement~(4) is witnessed by Proposition~\ref{prop:query:dx_complete}. Statement~(5) is true by lines~\ref{algoline:query:check_QP_empty} and \ref{algoline:query:addTrivialQueries} and by Proposition~\ref{prop:q1} as well as Corollary~\ref{cor:query_num_lower_bound} and the premise that $|\mD| \geq 2$ which guarantee that the function \textsc{addTrivialQueries} always adds at least $|\mD| \geq 2 > 0$ queries to $\QP$. In case $q < |\QP_{\max}|$, only statements (1), (2) and (3) are satisfied in general (for the same reasons as given above for the case $q \geq |\QP_{\max}|$) and $\QP$ is returned in line~\ref{algoline:query:return_QP_1} by the definition of $|\QP_{\max}|$. Thence, the condition $|\QP| = q \geq 1$ tested in line~\ref{algoline:query:test_QP} must be valid for $\QP$.   
\end{proof}

\begin{algorithm*}
\small
\caption{Interactive KB Debugging}\label{algo:inter_onto_debug}
\begin{algorithmic}[1]
\Require a tuple $\tuple{ \langle\mo,\mb,\Tp,\Tn\rangle_\RQ, n_{\min}, n_{\max}, t, p_{\widetilde{\mo}\cup\overline{\mo}}, q, qsm(),\sigma, mode}$ consisting of
\begin{itemize}
\item an admissible DPI $\langle\mo,\mb,\Tp,\Tn\rangle_\RQ$, 
\item leading diagnoses computation parameters, natural numbers $n_{\min} \geq 2, n_{\max}, t$, 
\item a function $p_{\widetilde{\mo}\cup\overline{\mo}}: \widetilde{\mo}\cup\overline{\mo} \rightarrow (0,1]$,
\item a parameter $q \in \mathbb{N}\cup\setof{\infty}, q \geq 1$ that determines the size of the computed query pool,
%quality of generated queries (the higher $q$, the better query quality, the worse reaction time)%, i.e.\ time between two successive queries), 
\item a function $qsm(Q) \in \mathbb{R}$ used for query selection that assigns a real number to a query $Q$ to express the ``goodness'' of $Q$, 
\item a maximum fault tolerance $\sigma \in [0,1]$ and 
\item a mode $mode \in \setof{static,dynamic}$ that determines the used method for diagnosis computation. 
\end{itemize}
% OLD
%an admissible DPI $\langle\mo,\mb,\Tp,\Tn\rangle_\RQ$, leading diagnoses computation parameters $n_{\min} \geq 2, n_{\max}, t$, a function $p_{\widetilde{\mo}\cup\overline{\mo}}(e) \in [0,1]$ for $e\in\widetilde{\mo}\cup\overline{\mo}$, a parameter $q$ that determines the quality of generated queries (the higher $q$, the better query quality, the worse reaction time, i.e.\ time between two successive queries), a function $qsm(Q) \in \mathbb{R}$ assigning a real number to a query $Q$ used for query selection, a maximum fault tolerance $\sigma \in [0,1]$, a mode $mode \in \setof{static,dynamic}$ that determines pruning strategy of diagnosis computation
% OLD end 
\Ensure The output depends on $mode$ and $\sigma$:
\begin{itemize}
\item $mode = static$: 
a maximal solution KB w.r.t.\ the input DPI $\langle\mo,\mb,\Tp,\Tn\rangle_\RQ$ which is
\begin{itemize}
	\item an approximation of the solution to Interactive Static KB Debugging (Problem Def.~\ref{prob_def:static}) if $\sigma > 0$. %The smaller $\sigma$, the better the approximate solution tends to be.
	\item the (exact) solution to Interactive Static KB Debugging if $\sigma = 0$. 
\end{itemize}
\item $mode = dynamic$: 
a maximal solution KB w.r.t.\ the current DPI $\langle\mo,\mb,\Tp\cup\Tp',\Tn\cup\Tn'\rangle_\RQ$ which is
\begin{itemize}
	\item an approximation of the solution to Interactive Dynamic KB Debugging (Problem Def.~\ref{prob_def:dynamic}) if $\sigma > 0$. %The smaller $\sigma$, the better the approximate solution tends to be.
	\item the (exact) solution to Interactive Dynamic KB Debugging if $\sigma = 0$. 
\end{itemize}
\end{itemize}
(for a more formal and precise characterization of the output see Proposition~\ref{prop:correctness_of_interactive_KB_debugging_algo} on page~\pageref{prop:correctness_of_interactive_KB_debugging_algo}) 
%\textbf{Output} on page~\pageref{etc:algo:inter_onto_debug:output}).
%a solution ontology w.r.t.\ the input DPI $\langle\mo,\mb,\Tp,\Tn\rangle_\RQ$ which is an approximation of a solution to Interactive Static KB Debugging (Problem Definition~\ref{prob_def:static}) if $\sigma > 0$ and the solution to Interactive Static KB Debugging otherwise.
%
%the most probable solution ontology w.r.t.\ the input DPI with a probability greater or equal $1-\sigma$ \newline
%if $mode = dynamic$: the most probable solution ontology w.r.t.\ the current DPI with a probability greater or equal $1-\sigma$ \newline
\newline

\State $\Tp', \Tn', \mathbf{C}_{calc}, \mD_{\checkmark}, \mD_{\times}, \mD_{out}, \mD_{\supset}, qData \gets \emptyset$ \label{algoline:inter_onto_debug:var_inst_start}
%\State $\Queue \gets \setof{\emptyset}$ \label{algoline:inter_onto_debug:Q={{}}}
\State $\Queue_{dup}, QA \gets []$
\State $\Queue \gets [\emptyset]$ \label{algoline:inter_onto_debug:Q={{}}}
\State $answer \gets \false$ \label{algoline:inter_onto_debug:var_inst_end}
\State $p_{\mo}() \gets \Call{getFormulaProbs}{\mo, p_{\widetilde{\mo}\cup\overline{\mo}}()}$  \label{algoline:inter_onto_debug:getAxiomProbs}	\Comment{application of Formulas~\ref{eq:ax_prob_calc} and \ref{eq:adapt_ax_prob_to_get_min_diags}}
\While{\true}  \label{algoline:inter_onto_debug:while}
\If{$mode = static$} \Comment{see Algorithm~\ref{algo:inter_stat_hs}}
	%\State $\tuple{\mD_{\checkmark},\Queue, \mathbf{C}_{calc}, \mD_{\times}} \gets \Call{staticHS}{}(\langle\mo,\mb,\Tp,\Tn\rangle_\RQ, \Queue, t, n_{\min}, n_{\max}, \mathbf{C}_{calc},\mD_{\checkmark}, \mD_{\times}, p_{\mo}(), \Tp', \Tn')$\label{algoline:inter_onto_debug:staticHS}
	\State $\tuple{\mD_{\checkmark},\Queue, \mathbf{C}_{calc}, \mD_{\times}} \gets \Call{staticHS}{}(\langle\mo,\mb,\Tp,\Tn\rangle_\RQ, \Queue, t, n_{\min}, n_{\max}, $ 
	\Statex \qquad\qquad\qquad\qquad\quad\qquad\qquad\qquad\qquad\quad\; $\mathbf{C}_{calc}, \mD_{\checkmark}, \mD_{\times}, p_{\mo}(), \Tp', \Tn')$\label{algoline:inter_onto_debug:staticHS}
\Else \Comment{see Algorithms~\ref{algo:inter_dyn_hs}, \ref{algo:update_tree} and \ref{algo:prune}}
	%\State $\tuple{\mD_{\checkmark},\Queue, \mathbf{C}_{calc}, \mD_{\times}, \mD_{\supset},\Queue_{dup}} \gets  
	%\Call{dynamicHS}{}(\langle\mo,\mb,\Tp,\Tn\rangle_\RQ, \Queue, \Queue_{dup}, t, n_{\min}, n_{\max}, \mathbf{C}_{calc},\mD_{\checkmark}, \mD_{\times}, p_{\mo}(), \Tp', \Tn', \mD_{\supset})$\label{algoline:inter_onto_debug:dynamicHS}
\State $\tuple{\mD_{\checkmark},\Queue, \mathbf{C}_{calc}, \mD_{\times}, \mD_{\supset},\Queue_{dup}} \gets
%$ 
	%\Statex \qquad\quad $
	\Call{dynamicHS}{}(\langle\mo,\mb,\Tp,\Tn\rangle_\RQ, \Queue, \Queue_{dup}, t, n_{\min}, n_{\max},$  
	\Statex \qquad\qquad\qquad\qquad\quad\qquad\quad\qquad\qquad\qquad\qquad\qquad\qquad\; $\mathbf{C}_{calc},\mD_{\checkmark}, \mD_{\times}, p_{\mo}(), \Tp', \Tn', \mD_{\supset})$ \label{algoline:inter_onto_debug:dynamicHS}
\EndIf
\State $p_{\mD}() \gets \Call{getProbDist}{\mD_{\checkmark},p_{\mo}(),\langle\mo,\mb,\Tp,\Tn\rangle_\RQ,QA}$\label{algoline:inter_onto_debug:getProbDist}  \Comment{see Algorithm~\ref{algo:inter_onto_debug_continued}}
\State $\md_{\max} \gets \Call{getMode}{\mD_{\checkmark},p_{\mD}()}$  \label{algoline:inter_onto_debug:getMode}
\If{$p_{\mD}(\md_{\max}) \geq 1-\sigma$}     \label{algoline:inter_onto_debug:stop_crit} \Comment{stop criterion}
	\State \Return $\Call{getSolKB}{\md_{\max},\langle\mo,\mb,\Tp\cup\Tp',\Tn\cup\Tn'\rangle_\RQ, \Tp', mode}$\label{algoline:inter_onto_debug:return} \Comment{return solution KB}
\Else
	%\State $\tuple{Q,\Pt(Q)} \gets \Call{calcQuery}{}(\mD_{\checkmark},qData,p_{\mD}(),p_{\widetilde{\mo}\cup\overline{\mo}}(), qsm(),\langle\mo,\mb,\Tp\cup\Tp',\Tn\cup\Tn'\rangle_\RQ, q)$\label{algoline:inter_onto_debug:calc_query} \Comment{see Algorithm~\ref{algo:inter_onto_debug_continued}}
	\State $\tuple{Q,\Pt(Q)} \gets \Call{calcQuery}{}(\mD_{\checkmark},qData,p_{\mD}(),p_{\widetilde{\mo}\cup\overline{\mo}}(), qsm(),$
	\Statex \qquad\qquad\qquad\qquad\qquad\qquad\qquad\quad\;$\langle\mo,\mb,\Tp\cup\Tp',\Tn\cup\Tn'\rangle_\RQ, q)$\label{algoline:inter_onto_debug:calc_query} \Comment{see Algorithm~\ref{algo:inter_onto_debug_continued}}
	\State $answer \gets u(Q)$\label{algoline:inter_onto_debug:user_interaction}    \Comment{user interaction}
	\State $QA \gets \Call{append}{\tuple{Q,answer},QA} $\label{algoline:inter_onto_debug:param_update_start}
	\State $\mD_{out} \gets \Call{getInvalidDiags}{\Pt(Q), answer}$\label{algoline:inter_onto_debug:get_invalid_diags}
	\State $qData \gets \Call{updateQData}{\mD_{out}, \mD_{\checkmark}, answer}$
	\State $\mD_{\checkmark} \gets \mD_{\checkmark} \setminus \mD_{out}$\label{algoline:inter_onto_debug:update_D_checkmark}
	\State $\mD_{\times} \gets \mD_{\times} \cup \mD_{out}$\label{algoline:inter_onto_debug:update_D_times}
	\If{$answer = \true$}
		\State $\Tp' \gets \Tp' \cup \setof{Q}$\label{algoline:inter_onto_debug:add_pos_tc}
	\Else
		\State $\Tn' \gets \Tn' \cup \setof{Q}$\label{algoline:inter_onto_debug:param_update_end}
	\EndIf
\EndIf
\EndWhile
\algstore{save_algo_inter_onto_debug}
\end{algorithmic}
\normalsize
\end{algorithm*}
\restoregeometry
%%%%%%%%%%%%%%%%%%%%%%%%%%%%%%%%%%%%%%%% FULL PAGE format of inter_onto_debug algo %%%%%%%%%%%%%%%%%%%%%%%%%%%%%%%%%%%%%%%% END

\begin{algorithm*}
\small
\caption{Interactive KB Debugging (continued)}\label{algo:inter_onto_debug_continued}
\begin{algorithmic}[1]
\algrestore{save_algo_inter_onto_debug}
%%
%\vspace{10pt}
%%
\Procedure{\textsc{getProbDist}}{$\mD_{\checkmark},p_{\mo}(),\langle\mo,\mb,\Tp,\Tn\rangle_\RQ,QA$}
\State $\Tp'', \Tn'' \gets \emptyset$
\State $p_{\mD,prio}() \gets \Call{getPrioDiagProbs}{\mD_{\checkmark},p_{\mo}(),\langle\mo,\mb,\Tp,\Tn\rangle_\RQ}$ \label{algoline:get_prob_dist:getPrioDiagProbs} \Comment{application of Formula~\ref{eq:diag_prob_calc}}
\For{$\tuple{Q,u(Q)} \in QA$}\label{algoline:get_prob_dist:for-loop_start}  \Comment{run through chronologically sorted query-answer pairs}
	\If{$u(Q) = \true$}
		\For{$\md_r \in \mD_{\checkmark}$}   \Comment{function \textsc{getEntailments} is defined on page~\pageref{etc:definition_of_getEntailments_function}}
			\State $E_{\md_r} \gets \Call{getEntailments}{\md_r,\mo,\mb,\Tp\cup\Tp''}$   \Comment{$E_{\md_r}$ is a set of entailments of 
			%$(\mo\setminus\md_r) \cup \mb \cup U_{\Tp\cup\Tp'}$}
			 $\mo^{*}_r$}
			\If{$Q \not\subseteq E_{\md_r}$}   \Comment{$\md_r \in \dz{}(Q)$} \label{algoline:get_prob_dist:check_dz_1}
				\State $p_{\mD,prio}(\md_r) \gets \frac{1}{2}\, p_{\mD,prio}(\md_r)$
			\EndIf
		\EndFor
		\State $\Tp'' \gets \Tp'' \cup \setof{Q}$    \label{algoline:get_prob_dist:update_Tp''}
	\Else
		\For{$\md_r \in \mD_{\checkmark}$}   \Comment{\textsc{isKBValid} (see Algorithm~\ref{algo:qx})}
			\If{$\Call{isKBValid}{(\mo\setminus\md_r) \cup Q, \tuple{\cdot,\mb,\Tp\cup\Tp'',\Tn\cup\Tn''}_{\RQ}}$}   \Comment{$\md_r \in \dz{}(Q)$}     \label{algoline:get_prob_dist:check_dz_2}
				\State $p_{\mD,prio}(\md_r) \gets \frac{1}{2}\, p_{\mD,prio}(\md_r)$
			\EndIf
		\EndFor
		\State $\Tn'' \gets \Tn'' \cup \setof{Q}$   \label{algoline:get_prob_dist:update_Tn''}
	\EndIf
\EndFor\label{algoline:get_prob_dist:for-loop_end}
\State $sum \gets \sum_{\md_r \in \mD_{\checkmark}} p_{\mD,prio}(\md_r)$\label{algoline:get_prob_dist:summation} 
\For{$\md_r \in \mD_{\checkmark}$}\label{algoline:get_prob_dist:for-loop2_start}
	\State $p_{\mD,prio}(\md_r) \gets \frac{1}{sum}\, p_{\mD,prio}(\md_r)$ \label{algoline:get_prob_dist:normalize} \Comment{normalization}
\EndFor
\State \Return $p_{\mD,prio}()$
\EndProcedure
\vspace{10pt}
\Procedure{\textsc{calcQuery}}{$\mD_{\checkmark},qData,p_{\mD}(),p_{\widetilde{\mo}\cup\overline{\mo}}(),qsm(),\langle\mo,\mb,\Tp\cup\Tp',\Tn\cup\Tn'\rangle_\RQ, q$}
\State $\QP \gets \Call{getPoolOfQueries}{}(\langle\mo,\mb,\Tp\cup\Tp',\Tn\cup\Tn'\rangle_\RQ, \mD_{\checkmark}, q)$ \Comment{see Algorithm~\ref{algo:query_gen}}
\State \Return $\Call{selectBestQuery}{}(\QP, qData,p_{\mD}(),p_{\widetilde{\mo}\cup\overline{\mo}}(),qsm())$ \label{algoline:inter_onto_debug_continued:selectBestQuery} \Comment{see Section~\ref{sec:query_selection_measures}}
\EndProcedure
\end{algorithmic}
\normalsize
\end{algorithm*}

\chapter[An Algorithm for Interactive Knowledge Base Debugging]{An Algorithm for Interactive Knowledge Base Debugging%
\chaptermark{Interactive KB Debugging Algorithm}}
\chaptermark{Interactive KB Debugging Algorithm}
\label{chap:WorkflowInInteractiveKBDebugging}
%%%%%
%\chapter{An Algorithm for Interactive Knowledge Base Debugging}
%\label{chap:WorkflowInInteractiveKBDebugging}
%%%%%
In this chapter we will give a description of an algorithm for interactive KB debugging (Algorithm~\ref{algo:inter_onto_debug}) which implements the entire functionality required by an interactive debugging system. All other algorithms presented so far will be subroutines of Algorithm~\ref{algo:inter_onto_debug} which are either directly or indirectly called by it. Before we explain and discuss Algorithm~\ref{algo:inter_onto_debug} in detail, we give the reader a rough and informal overview of the algorithm's input, output and actions in the following section in order to make the details of the algorithm easier to digest. 

\begin{remark}\label{rem:what_we_mean_by_currentDPI_inputDPI_intermediateDPI}
Note, in the following, when we speak of \emph{the input DPI} we refer to the DPI $\tuple{\mo,\mb,\Tp,\Tn}_\RQ$ that is provided as an input to Algorithm~\ref{algo:inter_onto_debug}, by \emph{the current DPI} we mean the DPI $\tuple{\mo,\mb,\Tp\cup\Tp',\Tn\cup\Tn'}_\RQ$ where $\Tp'$ and $\Tn'$, respectively, are all positive and negative test cases added to the input DPI from the start of the algorithm's execution until the current point in time. Further on, \emph{an intermediate (or previous) DPI} denotes a DPI $\tuple{\mo,\mb,\Tp\cup\Tp'',\Tn\cup\Tn''}_\RQ$ which is not the current DPI and where $\emptyset \subseteq \Tp'' \subseteq \Tp'$ and $\emptyset \subseteq \Tn'' \subseteq \Tn'$. Finally, \emph{the last-but-one DPI} corresponds to an intermediate DPI $\tuple{\mo,\mb,\Tp\cup\Tp'',\Tn\cup\Tn''}_\RQ$ where 
%exactly one of the following statements holds: 
either $|\Tp'| = |\Tp''|+1$ or $|\Tn'| = |\Tn''|+1$ is true, but not both.\qed
\end{remark}

\section[Overview]{Interactive Debugging Algorithm: Overview}
\label{sec:AlgorithmOverview}
%\newline
%\textbf{Overview:}\newline
%\textit{INPUT:} 
\textbf{Input:}\\
%\begin{adjustwidth}%{\parindent}{}
\indent An admissible DPI and some meta information where the latter consists of 
\begin{itemize}
	\item fault probabilities of syntactical elements occurring in the KB,
	\item a minimal and desired number of leading diagnoses,
	\item a desired maximum reaction time (time between two successive queries presented to the user), 
	\item a maximum fault tolerance (roughly, the probability of being presented a non-desired solution KB as output),
	\item a measure for query selection (determines which query is the best query within a given set of queries),
	\item a parameter that determines the size of the computed pool of queries in each iteration and
	%influencing the quality of queries asked to the user w.r.t.\ the given query selection measure and
	\item a parameter specifying the way the hitting set tree for computation of leading diagnoses is constructed and updated.
\end{itemize}
%\end{adjustwidth}
%\textit{OUTPUT:}
\vspace{3pt}

\noindent\textbf{Output:}
\begin{adjustwidth}{\parindent}{}
A solution KB such that the diagnosis used to formulate the solution KB has a probability (w.r.t.\ the current leading diagnoses) greater than or equal to 1 minus the given maximum fault tolerance.
\end{adjustwidth}
%\textit{PROCEDURE:}
\vspace{6pt}

\noindent\textbf{Procedure:}
\begin{enumerate}
		\item \emph{Initialization:} Compute the fault probability of each formula in the KB by means of the given fault probabilities.
		\item \emph{Leading Diagnoses Computation:} Use a hitting set tree constructed and updated in a manner as specified in the input coupled with $\scQX$ to calculate a set of leading diagnoses. In that, the cardinality and computation time of the set of leading diagnoses is determined by the corresponding input parameters specifying minimal and desired number of leading diagnoses and desired reaction time. 
		%by use of \textsc{QX} and \textsc{HS-Tree} where $a$ is the number of all minimal diagnoses for the DPI.
    \item \emph{Probability Update and Stop Criterion:} Use the formula fault probabilities and the new information obtained by already specified test cases (answered queries) to compute updated (posterior) probabilities of the current leading diagnoses.
		If one diagnosis probability is greater than or equal to 1 minus the maximum fault tolerance, return the solution KB obtained by deletion of this diagnosis from the KB and subsequent addition of the union of all positive test cases.
		% as per the Bayesian Rule (cf.~\cite{Shchekotykhin2012}). If some diagnosis $\md \in \mD$ has probability greater or equal to $1-\sigma$, return $\md$.
		\item \emph{Query Generation and Selection:} Use the set of leading diagnoses (and possibly their fault probabilities) to generate a pool of queries, the size of which depends on the respective parameter provided as input. Given the pool of queries, select the best query according to the given query selection measure.
		\item \emph{User Interaction and Incorporation of New Information:} Ask the user the selected query and add it to the positive test cases in case of a positive answer and to the negative test cases otherwise.
    \item \emph{Hitting Set Tree Update:} Update the hitting set tree based on the new information given by the classification of the test case resulting from the query answer. In particular, this involves the deletion of all those minimal diagnoses that conflict with the new test case.
    \item Repeat from Step 2.
\end{enumerate}

\section[Detailed Description]{Interactive Debugging Algorithm: Detailed Description%
\sectionmark{Detailed Algorithm Description}}
\sectionmark{Detailed Algorithm Description}
\label{sec:DetailedAlgorithmDescription}
%%%%%
%\section{Interactive Debugging Algorithm: Detailed Description}
%\label{sec:DetailedAlgorithmDescription}
%%%%%
%\noindent\textbf{Detailed algorithm description.} 
To describe the detailed process of Algorithm~\ref{algo:inter_onto_debug}, we first characterize the input arguments, the output and the meaning of the variables used and then provide a step-by-step textual description of the actions taken by the algorithm.
%workflow in interactive KB debugging.

%\noindent\textbf{Input arguments.}
\subsection{Input Arguments}
\label{sec:InputArguments}
The input parameters of Algorithm~\ref{algo:inter_onto_debug} are the following:
\begin{itemize}
\item An admissible DPI $\langle\mo,\mb,\Tp,\Tn\rangle_\RQ$ (cf.\ Definition~\ref{def:admissible}).  
\item Natural numbers $n_{\min} \geq 2, n_{\max}, t$ for leading diagnoses calculation (see description in Chapter~\ref{chap:UserInteraction} on page~\pageref{etc:leading_diag_params}). 

\emph{Remark:} The postulation $n_{\min} \geq 2$ is necessary in order for the existence of queries w.r.t.\ any computed set of leading minimal diagnoses $\mD$ and $\langle\mo,\mb,\Tp,\Tn\rangle_\RQ$ to be guaranteed (see Proposition~\ref{prop:q1}).
\item A	 function $p_{\widetilde{\mo}\cup\overline{\mo}}: \widetilde{\mo}\cup\overline{\mo} \rightarrow (0,1]$ that assigns a fault probability $p_{\widetilde{\mo}\cup\overline{\mo}}(e)$ to each $e \in \widetilde{\mo}\cup\overline{\mo}$ reflecting the degree of belief that (one occurrence of) a syntactical element $e$ appearing in $\mo$ is faulty (see Section~\ref{sec:DiagnosisProbabilitySpace}). 

\emph{Remarks:} Forbidding a probability of zero for syntactical elements assures that no formula in $\mo$ can have a probability of zero (cf.\ Remark~\ref{rem:ax_prob_not_zero}).

Recall from Section~\ref{sec:prob_space_construction} that $\widetilde{\mo}$ refers to the signature of $\mo$ (cf.\ Chapter~\ref{chap:basics})
%, i.e.\ the set of all terms in $\mo$, 
and $\overline{\mo}$ denotes the set of all logical connectives occurring in $\mo$. From probabilities of logical connectives and elements of the signature, probabilities of formulas in $\mo$ and from those in turn probabilities of diagnoses w.r.t.\ the DPI can be derived as shown by Formulas~\ref{eq:ax_prob_calc} and \ref{eq:diag_prob_calc}.

Further note that in the description of the algorithms in this section, unlike in Section~\ref{sec:DiagnosisProbabilitySpace}, we use different denotations for probabilities of syntactical elements ($p_{\widetilde{\mo}\cup\overline{\mo}}$), formulas ($p_{\mo}()$) and diagnoses ($p_{\mD}()$) in order to make a clear distinction between these different functions.

%Moreover, we point out that Algorithm~\ref{algo:inter_onto_debug} requires a probability measure as input, as opposed to Algorithm~\ref{algo:hs} which can basically deal with any real-valued weights
%Note that weights only need to be assigned to sentences in $\mo$ and not to those in $\mb \cup U_\Tp$ since diagnoses can comprise only sentences from $\mo$ and weights serve to compare and assess diagnoses w.r.t.\ their likeliness of being the one the user is looking for. More on how such weights can be obtained and about their meaning can be read in Section~\ref{sec:DiagnosisProbabilitySpace}.
%A	 weight function $w:\mo \rightarrow \mathbb{R}^{[0,1]}$ that assigns a real-numbered weight $w(\tax)$ in the interval between 0 and 1 to each $\tax\in\mo$. \\
%Note that weights only need to be assigned to sentences in $\mo$ and not to those in $\mb \cup U_\Tp$ since diagnoses can comprise only sentences from $\mo$ and weights serve to compare and assess diagnoses w.r.t.\ their likeliness of being the one the user is looking for. More on how such weights can be obtained and about their meaning can be read in Section~\ref{sec:DiagnosisProbabilitySpace}.
\item A natural number $q \geq 1$ that denotes the number of queries that should be precomputed, i.e.\ the preferred size of the query pool $\QP$ (see Chapter~\ref{chap:QueryGeneration}), before the ``best'' tuple $\tuple{Q^*,\Pt(Q^*)}$ is selected from $\QP$. 

\emph{Remark:} In general, higher $q$ implies better quality of the selected query in terms of the query selection measure $qsm()$ (see next bullet point). The chance of locating a good query in a larger set of queries is higher. On the other hand, higher $q$ involves a worse reaction time, i.e.\ time between two successive queries. The more queries are computed, the more time the function \textsc{getPoolOfQueries} consumes.	
\item A query selection measure $qsm()$ where $qsm: \QP \rightarrow \mathbb{R}$ is a function that assigns a real-valued number $qsm(\tuple{Q,\Pt(Q)})$ to each tuple in $\QP$, often called the score of $\tuple{Q,\Pt(Q)}$. 

\emph{Remark:} $qsm()$ defines what is considered the ``best'' query in the set $\QP$, namely the query $Q^*$ in the tuple $\tuple{Q^*,\Pt(Q^*)}$ with best score among all tuples in the pool $\QP$.
% for which the value $qsm(Q*) \in \mathbb{R}$ is optimal
%Further inputs to Algorithm~\ref{inter_onto_debug} are a natural number $q$ and a query selection measure $qsm()$ where $q$ denotes the number of queries that should be precomputed, i.e.\ the preferred size of the query pool $\QP$ (see Section~\ref{chap:QueryGeneration}), before the query $Q* \in \QP$ for which the value $qsm(Q*) \in \mathbb{R}$ is optimal is selected from $\QP$. 
Diverse measures that can be used as a $qsm()$ function in this algorithm have been discussed and evaluated within the scope of interactive KB debugging in literature~\cite{Shchekotykhin2012,Rodler2013} (for details see Section~\ref{sec:query_selection_measures}). 
\item A maximum fault tolerance $\sigma$ that defines the stop criterion of the algorithm.
That is, for a current set of leading diagnoses, the stop criterion is satisfied iff the most probable leading diagnosis has an (updated) probability of at least $1 - \sigma$ (see below for a precise definition of what ``updated'' means). 

\emph{Remark:} The smaller $\sigma$ is chosen, the higher is the chance that a desired diagnosis is found. Selecting $\sigma:=0$, i.e.\ admitting zero fault tolerance, is the safest (but also most time-consuming) way to run a debugging session with Algorithm~\ref{algo:inter_onto_debug}, as in this case the session will stop only after all but one diagnosis have been invalidated by test cases. 
\item A mode $mode \in \setof{static,dynamic}$ that determines 
\begin{enumerate}[(i)]
	\item which type of leading diagnoses are computed, i.e.\ only minimal diagnoses w.r.t.\ the input DPI ($static$) or minimal diagnoses w.r.t.\ the current DPI 
	%that includes more test cases than the input DPI 
	($dynamic$),
	\item the hitting set tree pruning strategy after a query has been answered, i.e.\ conservative pruning ($static$) or invasive pruning ($dynamic$),
	\item the space and time complexity of diagnosis computation, i.e.\ not much affected by the asked queries ($static$) -- tree is almost monotonically growing, but cannot get larger in size than the complete non-interactive hitting set tree (the tree produced by Algorithm~\ref{algo:hs} with input $n_{\min} = \infty$) -- or significantly influenced by the asked queries ($dynamic$) -- tree may shrink significantly if new test cases do not introduce ``completely new'' minimal conflict sets (that are in no subset-relation with an existing one), or lead to a tree that is significantly larger than the complete non-interactive hitting set tree 
%	(Algorithm~\ref{algo:hs} with input $n_{\min} = \infty$) 
if many ``completely new'' minimal conflict sets result from the addition of new test cases. For an in-depth discussion and comparison of both strategies the reader may consult Part~\ref{part:IterativeDiagnosisComputation}.
\end{enumerate}
%(i)~which type of leading diagnoses are computed, i.e.\ only minimal diagnoses for the input DPI (static) or minimal diagnoses for the updated current DPI that includes more test cases than the input DPI (dynamic), (ii)~the HS-tree pruning strategy after a query has been answered, i.e.\ conservative pruning (static) or 
%invasive pruning (dynamic), (iii)~the space and time complexity of diagnosis computation, i.e.\ not much affected by the asked queries (static) -- tree is monotonically growing, but cannot get larger in size than the complete non-interactive hitting set tree (Algorithm~\ref{algo:hs} with input $n_{\min} = \infty$) -- or significantly influenced by the asked queries (dynamic) -- tree may shrink significantly if new test cases do not introduce new minimal conflict sets that are in no subset-relation with an existing one, or lead to a tree that is significantly larger than the complete non-interactive hitting set tree (Algorithm~\ref{algo:hs} with input $n_{\min} = \infty$) if many new minimal conflict sets are attributable to new test cases. For an in-depth discussion and comparison of both strategies the reader may consult Section~\ref{sec:IterativeDiagnosisComputation}.
\end{itemize}

%a parameter $q$ that determines the quality of generated queries (the higher $q$, the better query quality, the worse reaction time, i.e.\ time between two successive queries), a measure $mqs$ used for query selection, a maximum fault tolerance $\sigma$, a mode $mode \in \setof{static,dynamic}$ that determines pruning strategy of diagnosis computation 

%\noindent\textbf{Output.}\label{etc:algo:inter_onto_debug:output}
\subsection{Output}
\label{sec:Output}
The output of Algorithm~\ref{algo:inter_onto_debug} can be explained as follows by making a distinction between the two modes of the algorithm specified by input parameter $mode$: 

\begin{proposition}\label{prop:correctness_of_interactive_KB_debugging_algo}
If $mode = static$, then Algorithm~\ref{algo:inter_onto_debug} returns the (exact) solution of the Interactive Static KB Debugging problem (Problem Definition~\ref{prob_def:static}) if $\sigma = 0$ and an approximate solution of the problem if $\sigma > 0$ where the likeliness of finding the (exact) solution increases with decreasing $\sigma$. 

More concretely, a maximal solution KB $\ot = (\mo \setminus \md_{\max}) \cup U_\Tp$ w.r.t.\ the input DPI $\langle\mo,\mb,\Tp,\Tn\rangle_\RQ$ is returned such that 
%the probability $p_{\mD}(\md_{\max})$ of $\md_{\max}\in\mD \subseteq\minD_{\langle\mo,\mb,\Tp,\Tn\rangle_\RQ}\cap\minD_{\langle\mo,\mb,\Tp\cup\Tp',\Tn\cup\Tn'\rangle_\RQ}$ is greater or equal $1-\sigma$ where 
\begin{enumerate}
\item \label{prop:correctness_of_interactive_KB_debugging_algo:stat1} $\md_{\max} \in \mD$ \hfill ($\md_{\max}$ is an element of the current set of leading diagnoses) 
\item \label{prop:correctness_of_interactive_KB_debugging_algo:stat2} $\md_{\max} = \argmax_{\md\in\mD} p_{\mD}(\md)$ \hfill ($\md_{\max}$ is the a-posteriori most probable leading diagnosis)
\item \label{prop:correctness_of_interactive_KB_debugging_algo:stat3} $p_{\mD}(\md_{\max}) \geq 1 - \sigma$ \hfill (the a-posteriori probability of $\mD_{\max}$ exceeds the predefined threshold)
\item \label{prop:correctness_of_interactive_KB_debugging_algo:stat4} $\mD\subseteq\minD_{\langle\mo,\mb,\Tp,\Tn\rangle_\RQ}\cap\minD_{\langle\mo,\mb,\Tp\cup\Tp',\Tn\cup\Tn'\rangle_\RQ}$ comprises the $|\mD|$ most probable minimal diagnoses w.r.t.\ $\langle\mo,\mb,\Tp,\Tn\rangle_\RQ$ as per the diagnosis probability measure $p_{\mD,prio}()$ \newline 
(the set of leading diagnoses corresponds to the a-priori most probable minimal diagnoses w.r.t.\ the input DPI that satisfy all specified test cases),
\item \label{prop:correctness_of_interactive_KB_debugging_algo:stat5}  a-priori probability measure $p_{\mD,prio}()$ is computed from $p_{\widetilde{\mo}\cup\overline{\mo}}()$ as per 
\begin{enumerate}
	\item Formula~\ref{eq:ax_prob_calc} \hfill (computation of formula fault probabilities)
	\item Formula~\ref{eq:adapt_ax_prob_to_get_min_diags} \hfill (adaptation of formula fault probabilities)
	\item Formula~\ref{eq:diag_prob_calc} \hfill (computation of diagnoses probabilities from formula fault probabilities)
%	\item Formula~\ref{eq:diag_prob_norm} \hfill (normalization of diagnoses probabilities w.r.t.\ probability space with $\Omega = \mD$)
\end{enumerate}
\item \label{prop:correctness_of_interactive_KB_debugging_algo:stat6} the a-posteriori probability measure $p_{\mD}()$ is computed from $p_{\mD,prio}()$ as per Bayes' Theorem (Formula~\ref{eq:bayes}, for details see below) taking into account the new information given by the set of all answered queries so far, i.e.\ the collected sets of positive ($\Tp'$) and negative ($\Tn'$) test cases.
\end{enumerate}
If $mode = dynamic$, then Algorithm~\ref{algo:inter_onto_debug} returns the (exact) solution of the Interactive Dynamic KB Debugging problem (Problem Definition~\ref{prob_def:dynamic}) if $\sigma = 0$ and an approximate solution of the problem if $\sigma > 0$ where the likeliness of finding the (exact) solution increases with decreasing $\sigma$. 

More concretely, a maximal solution KB $\ot = (\mo \setminus \md_{\max}) \cup U_{\Tp\cup\Tp'}$ w.r.t.\ the current DPI $\langle\mo,\mb,\Tp\cup\Tp',\Tn\cup\Tn'\rangle_\RQ$ is returned such that 
%the probability $p_{\mD}(\md_{\max})$ of $\md_{\max}\in\mD \subseteq\minD_{\langle\mo,\mb,\Tp\cup\Tp',\Tn\cup\Tn'\rangle_\RQ}$ is greater or equal $1-\sigma$ where 
\begin{enumerate}
\item \label{prop:correctness_of_interactive_KB_debugging_algo:dyn1} $\md_{\max} \in \mD$ \hfill ($\md_{\max}$ is an element of the current set of leading diagnoses) 
\item \label{prop:correctness_of_interactive_KB_debugging_algo:dyn2} $\md_{\max} = \argmax_{\md\in\mD} p_{\mD}(\md)$ \hfill ($\md_{\max}$ is the a-posteriori most probable leading diagnosis)
\item \label{prop:correctness_of_interactive_KB_debugging_algo:dyn3} $p_{\mD}(\md_{\max}) \geq 1 - \sigma$ \hfill (the a-posteriori probability of $\mD_{\max}$ exceeds the predefined threshold)
\item \label{prop:correctness_of_interactive_KB_debugging_algo:dyn4} $\mD\subseteq\minD_{\langle\mo,\mb,\Tp\cup\Tp',\Tn\cup\Tn'\rangle_\RQ}$ comprises the $|\mD|$ most probable minimal diagnoses w.r.t.\ $\langle\mo,\mb,\Tp\cup\Tp',\Tn\cup\Tn'\rangle_\RQ$ as per the diagnosis probability measure $p_{\mD,prio}()$ \newline 
(the set of leading diagnoses corresponds to the a-priori most probable minimal diagnoses w.r.t.\ the current DPI),
\item \label{prop:correctness_of_interactive_KB_debugging_algo:dyn5} the a-priori probability measure $p_{\mD,prio}()$ is computed from $p_{\widetilde{\mo}\cup\overline{\mo}}()$ as per 
\begin{enumerate}
	\item Formula~\ref{eq:ax_prob_calc} \hfill (computation of formula fault probabilities)
	\item Formula~\ref{eq:adapt_ax_prob_to_get_min_diags} \hfill (adaptation of formula fault probabilities)
	\item Formula~\ref{eq:diag_prob_calc} \hfill (computation of diagnoses probabilities from formula fault probabilities)
	%\item \label{prop:correctness_of_interactive_KB_debugging_algo:dyn6} Formula~\ref{eq:diag_prob_norm} \hfill (normalization of diagnoses probabilities w.r.t.\ probability space with $\Omega = \mD$)
\end{enumerate}
\item \label{prop:correctness_of_interactive_KB_debugging_algo:dyn6} the a-posteriori probability measure $p_{\mD}()$ is computed from $p_{\mD,prio}()$ as per Bayes' Theorem (Formula~\ref{eq:bayes}, for details see below) taking into account the new information given by the set of all answered queries so far, i.e.\ the collected sets of positive ($\Tp'$) and negative ($\Tn'$) test cases.
\end{enumerate}
\end{proposition}
%Algorithm~\ref{algo:inter_onto_debug} outputs a solution ontology $(\mo \setminus \md') \cup \mb \cup U_{\Tp\cup\Tp'}$ where $\md' \in \mD \subseteq \minD_{\tuple{\mo,\mb,\Tp\cup\Tp',\Tn\cup\Tn'}_\RQ}$ is the leading diagnosis with maximum probability according to $p_{\mD}$ w.r.t.\ the current DPI $\tuple{\mo,\mb,\Tp\cup\Tp',\Tn\cup\Tn'}_\RQ$ and satisfies $p_{\mD}(\md') \geq 1-\sigma$. That is, among the current leading diagnoses $\mD$, $\md'$ has a probability of at least $1-\sigma$. In other words, under the assumption that the desired diagnosis is included in the current set of leading diagnoses, the chance that $\md'$ is not the desired minimal diagnosis w.r.t.\ the current DPI is smaller than the predefined input parameter $\sigma$.
\begin{remark}\label{rem:approximate_solution}
We still need to explain what we mean by ``approximate solution'' of the Interactive Static (Dynamic) KB Debugging problem. 
Roughly, an approximate solution is one constructed from a diagnosis which is not the only remaining minimal diagnosis.
More precisely, an \emph{approximate solution} of
\begin{itemize}
	\item the Interactive Static KB Debugging problem is a maximal solution KB $(\mo \setminus \md)\cup U_{\Tp}$ such that 
\begin{itemize}
	\item $\md$ is a minimal diagnosis w.r.t.\ the input DPI and w.r.t.\ the current DPI and 
	\item there is some $\md' \neq \md$ which is a minimal diagnosis w.r.t.\ the input DPI and w.r.t.\ the current DPI  
\end{itemize}
	\item the Interactive Dynamic KB Debugging problem is a maximal solution KB $(\mo \setminus \md)\cup U_{\Tp\cup\Tp'}$ such that
\begin{itemize}
	\item $\md$ is a minimal diagnosis w.r.t.\ the current DPI and 
	\item there is some $\md' \neq \md$ which is a minimal diagnosis w.r.t.\ the current DPI  
\end{itemize}
\end{itemize}
where the input DPI is given by $\langle\mo,\mb,\Tp,\Tn\rangle_\RQ$ and the currect DPI by $\langle\mo,\mb,\Tp\cup\Tp',\Tn\cup\Tn'\rangle_\RQ$.

So, as long as not all but one diagnosis candidate that enables the formulation of a solution KB has been ruled out by the classification of test cases, we speak of an approximate solution. Now, the lower a value for $\sigma$ is predefined, the longer Algorithm~\ref{algo:inter_onto_debug} will usually need to iterate and the more test cases will usually need to be specified until one diagnosis has a probability greater than or equal to $1 - \sigma$. Thence, at the time a diagnosis exceeds the probability $1-\sigma$ there will be usually fewer minimal diagnoses left than in case of the selection a higher value for $\sigma$. Therefore, the likeliness of picking the (exact) solution will usually be the higher, the lower $\sigma$ is.\qed 
% either problem is a maximal solution KB that is constructed from a minimal diagnosis that is not the only minimal diagnosis w.r.t.\ the input and current KB in the case of static debugging and, respectively,  w.r.t.\ the current DPI in the case of dynamic debugging. That is,  
\end{remark}

\begin{remark}\label{rem:absolute_vs_relative_fault_tolerance_AND_initial_vs_updated_probabilities}
Note that granting a maximum \emph{absolute} fault tolerance $\sigma$ that is independent of a set of leading diagnoses is generally computationally infeasible due to the high complexity of diagnosis computation (see Chapter~\ref{chap:intro}). Since, for an absolute fault tolerance to hold, \emph{all} minimal diagnoses w.r.t.\ the current DPI have to be computed in order to determine their probability and to decide whether the most probable diagnosis has a probability greater than or equal to $1-\sigma$. 

In fact, the fault tolerance used by Algorithm~\ref{algo:inter_onto_debug} which is \emph{relative} to the set of leading diagnoses, i.e.\ the (a-priori) most probable minimal diagnoses $\mD$ 
w.r.t.\ a DPI can be interpreted as follows. Under the assumption that the true diagnosis $\dt$ is included in $\mD$, the chance that the most probable minimal diagnosis $\md_{\max} \in \mD$ which satisfies the stop criterion is not equal to $\dt$ is smaller than the predefined threshold $\sigma$ (cf.\ Section~\ref{sec:DiagnosisProbabilitySpace}). Thus, under this assumption, the (a-posteriori) probability of being presented a non-desired solution KB as output of Algorithm~\ref{algo:inter_onto_debug} is smaller than $\sigma$. 

The a-priori diagnoses probability measure $p_{\mD,prio}()$ refers to the one that is computed \emph{directly} from the fault information provided as an input to Algorithm~\ref{algo:inter_onto_debug} whereas the a-posteriori diagnoses probability measure $p_{\mD}()$ is the one obtained from $p_{\mD,prio}()$ after incorporating the information given by the new test cases specified so far during the debugging session.
So, $p_{\mD,prio}()$ and $p_{\mD}()$ might differ in terms of the probability order of diagnoses. Incorporation of updated probabilities \emph{directly} into the hitting set tree algorithms to be used for the determination of leading diagnoses in the order prescribed by an updated probability measure is only possible if there is an additional update operator (besides Bayes' Theorem for adapting \emph{diagnoses} probabilities) that can be applied to \emph{formula} probabilities. For, the latter are exploited in the hitting set tree to assign probability weights to paths that are \emph{not yet diagnoses} (cf.\ $p_{nodes}()$ specified by Definition~\ref{def:p_node()} and the discussion of Formula~\ref{eq:path_prob_calc}) in order to guide the search for minimal diagnoses in best-first order. Updated \emph{diagnosis} probabilities are not helpful at all for this purpose. Devising a reasonable mechanism of updating formula probabilities seems to be hard mostly due to the lack of suitable data that might be collected during the debugging session to accomplish that. What would be imaginable during the debugging session is to try to learn something about the fault probability of \emph{syntactical elements} by examining the positive (all formulas are \emph{definitely} correct) and singleton negative (the single formula is \emph{definitely} incorrect) test cases. However, a drawback of such a strategy comes into effect when only syntactically very simple queries are used which is, for instance, the case in Example~\ref{example:query_computation} (see the definition of the \textsc{getEntailments} function there). From such queries not many useful insights concerning faulty syntactical elements might be gained. On the other hand, such queries are absolutely desirable from the point of view of how well a user might comprehend the formulas asked by the system. Hence, these two aspects seem to contradict each other. Still, it is a topic for future research to attempt to elaborate a solution for that issue.

A way to achieve that $p_{\mD}()$ coincides with $p_{\mD,prio}()$, at least in case $mode=static$, is to exclude queries $Q$ with $\dz{}(Q) \neq \emptyset$ (see Remark~\ref{rem:prob_update_obsolete_if_dz(Q)={}_for_all_queries}).
%(for details see \emph{Computing a Probability Distribution of Leading Diagnoses} on page~\pageref{etc:computing_prob_dist_of_leading_diags}). 
How this might be accomplished is stated by Proposition~\ref{prop:query_gen_explicit_entailments_dz_empty}. 
%That is, in this case $\md_{\max}$ is the most probable diagnosis w.r.t.\ both the initially given and the updated probability measures. 
Please notice that ignorance of queries with non-empty $\dz{}$ does not implicate any disadvantages for interactive debugging. On the contrary, it is even a desirable feature of a debugger and brings along higher computational efficacy of query generation and stronger test cases from the logical point of view (cf.\ Section~\ref{sec:remarks_query_gen}). For the scenario $mode=dynamic$, it is not possible in general to bypass the probability update by means of such queries (see Remark~\ref{rem:prob_update_obsolete_if_dz(Q)={}_for_all_queries}).  
%
%that is directly computed from the \emph{initially} given probabilities $p_{\widetilde{\mo}\cup\overline{\mo}}()$ as per Formulas~\ref{eq:ax_prob_calc}, \ref{eq:adapt_ax_prob_to_get_min_diags}, \ref{eq:diag_prob_calc} and \ref{eq:diag_prob_norm}. However, $\md_{\max}$ is not necessarily the most probable diagnosis based on the updated probability measure $p_{\mD}()$. Sincedefined for $\md \in \Omega = \minD_{\langle\mo,\mb,\Tp,\Tn\rangle_\RQ}$ in case $mode=static$ and for $\md \in \Omega = \minD_{\langle\mo,\mb,\Tp\cup\Tp',\Tn\cup\Tn'\rangle_\RQ}$ in case $mode=static$,  
\qed
\end{remark} %\newline
%a the minimal diagnosis with maximal weight according to $w_{\mD}$ w.r.t.\ the current DPI $\tuple{\mo,\mb,\Tp\cup\Tp',\Tn\cup\Tn'}_\RQ$ such that $w_{\mD}(\md^*) \geq 1-\sigma$. That is,  

%\noindent\textbf{Variables.} 
\subsection{Variables}
\label{sec:Variables}
The variables used by Algorithm~\ref{algo:inter_onto_debug} that are not input arguments to the algorithm are the following:
\begin{itemize}
\item $\Tp', \Tn'$ are the sets of positive and negative test cases, respectively, collected \emph{during} the execution of Algorithm~\ref{algo:inter_onto_debug} so far. That is, $\Tp'$ stores all positively answered queries, whereas $\Tn'$ stores all negatively answered ones.
\item $\mC_{calc}$ is the 
%cumulated 
set of all conflict sets computed by $\scQX$ during the execution of Algorithm~\ref{algo:inter_onto_debug} so far.
%the \textsc{sLabel} (for $mode=static$, called within \textsc{staticHS}) or \textsc{dLabel} (for $mode=dynamic$, called within \textsc{dynamicHS}) procedure. 

\emph{Remark:} In case of static debugging ($mode=static$), $\mC_{calc}$ includes exclusively minimal conflict sets w.r.t.\ the input DPI, whereas, in case of dynamic debugging ($mode=dynamic$), $\mC_{calc}$ may comprise minimal conflict sets w.r.t.\ the current or any intermediate DPI. 
%$\tuple{\mo,\mb,\Tp\cup\Tp'',\Tn\cup\Tn''}_\RQ$ that was onbtained from the input DPI $\tuple{\mo,\mb,\Tp,\Tn}_\RQ$ by addition of sets of test cases $\Tp'', \Tn''$ where $\tuple{\mo,\mb,\Tp\cup\Tp',\Tn\cup\Tn'}_\RQ$ is the current DPI and $\emptyset \subseteq \Tp'' \subseteq \Tp'$, $\emptyset \subseteq \Tn'' \subseteq \Tn'$.
%minimal conflict sets w.r.t.\ the input DPI that are computed by the algorithm.  
\item $\mD_{\checkmark}$ is the set of leading diagnoses returned by a call of \textsc{staticHS} in case of static debugging ($mode=static$) and by a call of \textsc{dynamicHS} in case of dynamic debugging ($mode=dynamic$). 

\emph{Remarks:} In case of dynamic debugging, $\mD_{\checkmark} \subseteq \minD_{\langle\mo,\mb,\Tp\cup\Tp',\Tn\cup\Tn'\rangle_\RQ}$ is the set of most probable minimal diagnoses w.r.t.\ the current DPI $\langle\mo,\mb,\Tp\cup\Tp',\Tn\cup\Tn'\rangle_\RQ$ as per the diagnosis probability measure $p_{\mD,prio}()$ computed from $p_{\widetilde{\mo}\cup\overline{\mo}}()$ by Formulas~\ref{eq:ax_prob_calc}, \ref{eq:adapt_ax_prob_to_get_min_diags}, \ref{eq:diag_prob_calc} and \ref{eq:diag_prob_norm} (cf.\ Sections~\ref{sec:DiagnosisProbabilitySpace} and \ref{sec:Output}).

In case of static debugging, $\mD_{\checkmark} \subseteq \minD_{\langle\mo,\mb,\Tp,\Tn\rangle_\RQ} \cap \minD_{\langle\mo,\mb,\Tp\cup\Tp',\Tn\cup\Tn'\rangle_\RQ}$, i.e.\ $\mD_{\checkmark}$ includes only diagnoses that are minimal diagnoses w.r.t.\ the input DPI $\langle\mo,\mb,\Tp,\Tn\rangle_\RQ$ as well as w.r.t.\ the current DPI $\langle\mo,\mb,\Tp\cup\Tp',\Tn\cup\Tn'\rangle_\RQ$. Moreover, $\mD_{\checkmark}$ comprises the most probable minimal diagnoses \emph{w.r.t.\ the input DPI} according to the diagnosis probability measure $p_{\mD,prio}()$ computed from $p_{\widetilde{\mo}\cup\overline{\mo}}()$ by Formulas~\ref{eq:ax_prob_calc}, \ref{eq:adapt_ax_prob_to_get_min_diags}, \ref{eq:diag_prob_calc} and \ref{eq:diag_prob_norm} (cf.\ Sections~\ref{sec:DiagnosisProbabilitySpace} and \ref{sec:Output}).
%$p_{\mD}$ where $p_{\mD}$ is the probability measure for the probability space with $\Omega = \mD_{\checkmark}$. 
%
%In case of dynamic debugging, $\mD_{\checkmark} \subseteq \minD_{\langle\mo,\mb,\Tp\cup\Tp',\Tn\cup\Tn'\rangle_\RQ}$ is the set of minimal diagnoses with maximum probability \emph{w.r.t.\ the current DPI} according to $p_{\mD}$ where $p_{\mD}$ is the probability measure for the probability space with $\Omega = \mD_{\checkmark}$. So, there may be diagnoses in $\mD_{\checkmark}$ that are not minimal diagnoses w.r.t.\ the input DPI.
\item $\mD_{\times}$ stores all minimal diagnoses w.r.t.\ the input DPI that have been invalidated by one of the collected positive and negative test cases $\Tp'$ and $\Tn'$, respectively ($mode=static$). 
$\mD_{\times}$ stores the minimal diagnoses w.r.t.\ the last-but-one DPI that have been invalidated by the most recently added test case ($mode=dynamic$). 
%(i.e.\ still stored in the hitting set tree), invalidated diagnoses, i.e.\ the set of subsets of $\mo$ that have been minimal diagnoses for some intermediate DPI, but that are not minimal diagnoses for the current DPI anymore. 
%
%That is, each $\md \in \mD_{\times}$ has been invalidated by one test case among the collected positive and negative test cases $\Tp'$ and $\Tn'$, respectively.
\item $\mD_{out}$ is the subset of the set of current leading diagnoses $\md_{\checkmark}$ that has been invalidated by the most recently added test case.
\item $\mD_{\supset}$ stores all diagnoses that are non-minimal w.r.t.\ the current DPI, i.e.\ for each diagnosis $\mathsf{nd} \in \mD_{\supset}$ there is some $\mathsf{nd}'\in\mD_{\checkmark}$ such that $\mathsf{nd} \supset \mathsf{nd}'$ ($mode=dynamic$). 
%$\mD_{\supset}$ comprises diagnoses (paths in the hitting set tree) that are non-minimal w.r.t.\ the current DPI, i.e.\ for each diagnosis $\mathsf{nd} \in \mD_{\supset}$ there is some $\mathsf{nd}'\in\mD_{\checkmark}$ such that $\mathsf{nd} \supset \mathsf{nd}'$. 

\emph{Remark:} $\mD_{\supset}$ is solely needed for dynamic and not for static debugging as the latter does not need to store non-minimal diagnoses (cf.\ rule~\ref{def:pruned_hs_tree:non_min_pruning_rule} of Definition~\ref{def:pruned_hs_tree} on page~\pageref{def:pruned_hs_tree}). Reason for this is the fact that only minimal diagnoses w.r.t.\ the input DPI are searched for. 
%Since minimal diagnoses can only grow and not shrink in size after specifying a new test case (cf.\ Proposition~\ref{prop:after_adding_testcase_new_min_diag_is_equal_or_superset_of_old_min_diag} given later) and since test case sets $\Tp'$ and $\Tn'$ are monotonically growing throughout a debugging session, each non-minimal diagnosis 
%
%w.r.t.\ any intermediate DPI considered by the algorithm is also a non-minimal diagnosis w.r.t.\ the input DPI. Thus, such a non-minimal diagnosis can never become relevant during static interactive debugging.
On the other hand, in case of dynamic debugging, non-minimal diagnoses might become minimal ones after some new test cases are specified since minimal diagnoses w.r.t.\ the (changing) current DPI are considered.
%computed by the algorithm (for some DPI prior to the current one) that are non-minimal w.r.t.\ the current DPI.
%some DPI that occurred during execution of the algorithm. 
\item\label{etc:var:metrics} $qData$ is an informal variable that comprehends any kind of data that might be taken into account by the query selection measure $qsm()$ and that might need to be adapted after a query has been answered (and diagnoses have been invalidated) in order to take the obtained new information into account. One can imagine $qData$ as a log specific to the particular function $qsm()$ that is used which records data of prior (query answering) iterations executed by the algorithm such as certain performance measures. An example of a $qsm()$ strategy using one such metric, namely the ratio of leading diagnoses invalidated by a test case, can be found in~\cite{Rodler2013}. 
\item \label{etc:QA} $QA := [\tuple{Q,u(Q)}]_{Q\in\Tp'\cup\Tn'}$ where $u(Q) \in \setof{\true,\false}$ is the chronologically ordered list of queries and user answers collected so far during the execution of Algorithm~\ref{algo:inter_onto_debug}.
\item $\Queue$ is the current queue of open nodes in the 
%existing partial 
hitting set tree maintained by Algorithm~\ref{algo:inter_onto_debug}.
\item The list $\Queue_{dup}$ roughly stores all duplicate nodes (that is, nodes for each of which there is a node in the hitting set tree that corresponds to an equal set of edge labels) computed so far during the execution of Algorithm~\ref{algo:inter_onto_debug}.
% sorted by cardinality (that is, number of edge labels contained) in ascending order.

\emph{Remark:} The list $\Queue_{dup}$ is only relevant in case $mode = dynamic$ and not needed if $mode=static$. The purpose of this set is to enable the ``replacement'' of pruned nodes which is necessary to guarantee the completeness of \textsc{dynamicHS} in terms of not missing any minimal diagnoses (for a detailed explanation, see Chapter~\ref{chap:DynamicHSTree}).
%, i.e.\ nodes that are not in any of the sets $\mD_{\checkmark}, \mD_{\supset}, \mD_{\times}, \mD_{out}$. That is, a node is an open node if it corresponds to a branch in the hitting set tree that is neither a minimal nor a non-minimal diagnosis nor an invalidated diagnosis.
%those nodes that correspond to paths that are neither minimal nor non-minimal diagnoses nor w.r.t.\
\newline
\end{itemize}

%\noindent\textbf{Process flow.} \newline
\subsection{Algorithm Walkthrough}
\label{sec:Walkthrough}
%\noindent\textbf{Initialization.} 
\paragraph{Initialization.}
In the first \ref{algoline:inter_onto_debug:var_inst_end} lines,
%~\ref{algoline:inter_onto_debug:var_inst_start}-\ref{algoline:inter_onto_debug:var_inst_end}, 
variable declarations take place. First, all variables that store sets of conflict sets, diagnoses or test cases, and $qData$ are initialized to the empty set. Further on, $\Queue_{dup}$ and $QA$ are initialized to an empty list. Finally, the queue $\Queue$ of open nodes used for the hitting set tree construction by \textsc{staticHS} ($mode=static$) or \textsc{dynamicHS} ($mode=dynamic$), respectively, is set to $[\emptyset]$ since it initially includes only a non-labeled root node.
% which is represented by the empty set.
\begin{remark}\label{rem:queue_root_node_emptyset_emptylist}
The non-labeled root node is denoted by $\emptyset$ since nodes in \textsc{staticHS} are associated with the set of edge labels along the path in the hitting set tree from the root node to this node (cf.\ Chapters~\ref{chap:DiagnosisComputation} and \ref{chap:StaticHSTree}). Hence, the root node itself corresponds to the empty path which includes no edges.

Notice that in case of \textsc{dynamicHS}, nodes will be (ordered) lists instead of (non-ordered) sets like in \textsc{staticHS} (cf.\ Chapter~\ref{chap:DynamicHSTree}). That is, to be precise, the unlabeled root node in this case corresponds to the empty list $[]$. For the ease of representation of Algorithm~\ref{algo:inter_onto_debug}, only one set $\Queue$ is initialized to be used with either \textsc{staticHS} or \textsc{dynamicHS}. Thence, by abuse of notation, we associate $\emptyset$ in this case with the empty list $[]$.\qed
\end{remark}
% -- where each node is represented as a set of formulas (cf.\ Section~\ref{chap:DiagnosisComputation}) -- is set to $\setof{\emptyset}$ since it initially includes only a non-labeled node, namely the empty set. 

%\noindent\textbf{Computing Formula Probabilities.} 
\paragraph{Computing Fault Probabilities of Formulas.}
Then, \textsc{getFormulaProbs} is called in line~\ref{algoline:inter_onto_debug:getAxiomProbs} with the KB $\mo$ and the function $p_{\widetilde{\mo}\cup\overline{\mo}}: \widetilde{\mo}\cup\overline{\mo} \rightarrow (0,1]$ as inputs. The function first applies Formula~\ref{eq:ax_prob_calc} to compute probabilities for each formula in $\mo$, then applies Formula~\ref{eq:adapt_ax_prob_to_get_min_diags} to these probabilities leading to the output $p_{\mo}: \mo \rightarrow (0,0.5)$, a function that assigns a value $p_{\mo}(\tax) \in (0,0.5)$ to each $\tax\in\mo$.

%\noindent\textbf{Computing Leading Diagnoses.} 
\paragraph{Computing Leading Diagnoses.}
At this point, all input arguments required by for the hitting set tree construction are instantiated. So, the algorithm enters the while loop in line~\ref{algoline:inter_onto_debug:while}. As a first step within the loop, either \textsc{staticHS}, if $mode = static$, or \textsc{dynamicHS}, otherwise, is called in order to obtain a tuple including a set of leading diagnoses along with variables that store the ``state'' of the (partial) hitting set tree constructed so far and facilitate the reuse of this tree in the next iteration.

In concrete terms, \textsc{staticHS} accepts the arguments $\langle\mo,\mb,\Tp,\Tn\rangle_\RQ$, $\Queue$, $t$, $n_{\min}$, $n_{\max}$, $\mathbf{C}_{calc}$, $\mD_{\checkmark}$, $\mD_{\times}$, $p_{\mo}()$, $\Tp'$ and $\Tn'$ and returns a tuple $\tuple{\mD,\Queue, \mathbf{C}_{calc}, \mD_{\times}}$ the elements of which are defined as follows:
\begin{itemize}
\item $\mD$ is the current set of leading diagnoses such that
\begin{enumerate}[(a)]
\item \label{etc:staticHS_output_bullet_a} $\mD \subseteq \minD_{\langle\mo,\mb,\Tp,\Tn\rangle_\RQ} \cap \minD_{\langle\mo,\mb,\Tp\cup\Tp',\Tn\cup\Tn'\rangle_\RQ}$ is the set of most probable 
%(as per $p_{\mo}(\tax),\tax\in\mo$ and Formula~\ref{eq:diag_prob_calc})
 minimal diagnoses w.r.t.\ $\langle\mo,\mb,\Tp,\Tn\rangle_\RQ$ that satisfy all test cases $\Tp'$ and $\Tn'$ such that 
\begin{enumerate}[(i)]
\item $n_{\min} \leq |\mD| \leq n_{\max}$ and 
\item \label{etc:staticHS_must_compute_at_least_one_further_min_diag_per_call} $\mD\supset\mD_{\checkmark}$,
\end{enumerate} 
if such a set $\mD$ 
%with (i) and (ii)
 exists;
or 
\item $\mD$ is equal to the set of all minimal diagnoses $\minD_{\langle\mo,\mb,\Tp,\Tn\rangle_\RQ} \cap \minD_{\langle\mo,\mb,\Tp\cup\Tp',\Tn\cup\Tn'\rangle_\RQ}$, otherwise;
\end{enumerate}
where ``most-probable'' refers to the diagnosis probability measure $p_{\mD,prio}()$ obtained from $p_{\mo}()$ 
%for $\tax\in\mo$ 
by application of Formulas~\ref{eq:diag_prob_calc} and \ref{eq:diag_prob_norm}.
%where ``most-probable'' refers to the probability measure $p_{\mD}(\md), \md\in\mD$ obtained from $p_{\mo}(\tax),\tax\in\mo$ which is transformed to some probability measure $p_{\mD,prio}(\md),\md\in\mD$ by Formula~\ref{eq:diag_prob_calc} which is in turn used to calculate $p_{\mD}(\md), \md\in\mD$ by Formula~\ref{eq:bayes} (see \textbf{Computing a Probability Distribution of Leading Diagnoses} on page~\pageref{etc:computing_prob_dist_of_leading_diags}). 
\item $\Queue$ is the current queue of open nodes of the hitting set tree.
%, i.e.\ a list of nodes $\mathsf{nd} \subseteq \mo$ that have not yet been processed and labeled. 
\item $\mathbf{C}_{calc} \subseteq \minC_{\langle\mo,\mb,\Tp,\Tn\rangle_\RQ}$ is the set of all computed minimal conflict sets w.r.t.\ the input DPI throughout all calls of \textsc{staticHS} during the execution of Algorithm~\ref{algo:inter_onto_debug} so far.
%\item $\mathbf{C}_{calc} \subseteq \minC_{\langle\mo,\mb,\Tp,\Tn\rangle_\RQ}$ is the overall set of computed minimal conflict sets throughout all calls to \textsc{staticHS} during the execution of Algorithm~\ref{algo:inter_onto_debug}.
\item $\mD_{\times}$ comprises all computed minimal diagnoses throughout all calls of \textsc{staticHS} during the execution of Algorithm~\ref{algo:inter_onto_debug} so far where each $\md\in\mD_{\times}$ has been invalidated by some test case in $\Tp'$ or $\Tn'$.
%\item $\mD_{\times}$ is the overall set of computed minimal diagnoses throughout all calls to \textsc{staticHS} during the execution of Algorithm~\ref{algo:inter_onto_debug} where each $\md\in\mD_{\times}$ has been invalidated by some test case in $\Tp'\cup\Tn'$.
\end{itemize} 
%\item $\Queue$ is the current queue of open nodes of the hitting set tree;
%\item $\mathbf{C}_{calc} \subseteq \minC_{\langle\mo,\mb,\Tp,\Tn\rangle_\RQ}$ is the overall set of computed minimal conflict sets; and
%\item $\mD_{\times}$ is the overall set of invalidated minimal diagnoses. 
%\end{itemize}

Similarly, \textsc{dynamicHS} accepts the arguments $\langle\mo,\mb,\Tp,\Tn\rangle_\RQ$, $\Queue$, $\Queue_{dup}$, $t$, $n_{\min}$, $n_{\max}$, $\mathbf{C}_{calc}$, $\mD_{\checkmark}$, $\mD_{\times}$, $p_{\mo}()$, $\Tp'$, $\Tn'$ and $\mD_{\supset}$ and returns a tuple $\tuple{\mD,\Queue, \mathbf{C}_{calc}, \mD_{\times}, \mD_{\supset}, \Queue_{dup}}$ the elements of which are defined as follows:
\begin{itemize}
\item $\mD$ is the current set of leading diagnoses such that
\begin{enumerate}[(a)]
\item \label{etc:dynamicHS_output_bullet_a} $\mD \subseteq \minD_{\langle\mo,\mb,\Tp\cup\Tp',\Tn\cup\Tn'\rangle_\RQ}$ is the set of most probable minimal diagnoses w.r.t.\ $\langle\mo,\mb,\Tp\cup\Tp',\Tn\cup\Tn'\rangle_\RQ$ such that 
\begin{enumerate}[(i)]
\item $n_{\min} \leq |\mD| \leq n_{\max}$ and 
\item \label{etc:dynamicHS_must_compute_at_least_one_further_min_diag_per_call} $\mD \setminus\mD_{\checkmark} \neq \emptyset$,
\end{enumerate}
if such a set $\mD$ exists, or 
\item $\mD$ is equal to the set of all minimal diagnoses $\minD_{\langle\mo,\mb,\Tp\cup\Tp',\Tn\cup\Tn'\rangle_\RQ}$, otherwise,
\end{enumerate}
where ``most-probable'' refers to the diagnosis probability measure $p_{\mD,prio}()$ obtained from $p_{\mo}()$ by application of Formulas~\ref{eq:diag_prob_calc} and \ref{eq:diag_prob_norm}.
%where ``most-probable'' refers to the probability measure $p_{\mD}(\md), \md\in\mD$ obtained from $p_{\mo}(\tax),\tax\in\mo$ which is transformed to some probability measure $p_{\mD,prio}(\md),\md\in\mD$ by Formula~\ref{eq:diag_prob_calc} which is in turn used to calculate $p_{\mD}(\md), \md\in\mD$ by Formula~\ref{eq:bayes} (see \textbf{Computing a Probability Distribution of Leading Diagnoses} on page~\pageref{etc:computing_prob_dist_of_leading_diags}). 
\item $\Queue$ is the current queue of open (non-labeled) nodes of the hitting set tree,
%\item $\Queue$ is the current queue of open nodes of the hitting set tree, i.e.\ a list of nodes $\mathsf{nd} \subseteq \mo$ that have not yet been processed and labeled. 
\item $\mathbf{C}_{calc}$ is a set of conflict sets w.r.t.\ the current DPI $\langle\mo,\mb,\Tp\cup\Tp',\Tn\cup\Tn'\rangle_\RQ$,
%\item $\mathbf{C}_{calc}$ is the overall set of computed minimal conflict sets throughout all calls to \textsc{dynamicHS} during the execution of Algorithm~\ref{algo:inter_onto_debug} where each $\mc\in\mathbf{C}_{calc}$ is a minimal conflict set w.r.t.\ some intermediate DPI (not necessarily one w.r.t.\ the input DPI).  
\item $\mD_{\times} = \emptyset$, 
\item $\mD_{\supset}$ is the set of all processed nodes so far throughout the execution of Algorithm~\ref{algo:inter_onto_debug} that are non-minimal diagnoses w.r.t.\ the current DPI $\langle\mo,\mb,\Tp\cup\Tp',\Tn\cup\Tn'\rangle_\RQ$ and
%\item $\mD_{\supset}$ is the overall set of nodes generated throughout all calls to \textsc{dynamicHS} during the execution of Algorithm~\ref{algo:inter_onto_debug} where for each $\mathsf{nd} \in \mD_{\supset}$ there is some $\md \in \mD$ with $\mathsf{nd} \supset \md$. In other words, $\mD_{\supset}$ stores all computed non-minimal diagnoses (paths) w.r.t.\ the current DPI.  
%a set $\mD$ of minimal diagnoses w.r.t.\ $\langle\mo,\mb,\Tp,\Tn\rangle_\RQ$ such that $n_{\min} \leq |\mD| \leq n_{\max}$ and $\mD \setminus \mD_{\checkmark} \neq \emptyset$
\item $\Queue_{dup}$ includes all duplicate nodes found so far throughout the execution of Algorithm~\ref{algo:inter_onto_debug} (for a detailed explanation see Chapter~\ref{chap:DynamicHSTree} and Algorithm~\ref{algo:inter_dyn_hs}).
\end{itemize}

\begin{remark}\label{rem:p_D,prio()_same_order_for_diags_as_p_nodes()}
It is very important to notice that the function $p_{nodes}()$ for $p() := p_{\mo}()$ as specified by Definition~\ref{def:p_node()} on page~\pageref{def:p_node()} imposes the same order on a set of minimal diagnoses as the a-priori probability measure $p_{\mD,prio}()$. That is $p_{nodes}(\md) = c \cdot p_{\mD,prio}(\md)$ for all minimal diagnoses $\md$ w.r.t.\ a DPI where $c$ is a constant (which is the same for all diagnoses $\md$). The difference between both functions is that $p_{nodes}()$ is defined for all $X \subseteq \mo$ whereas $p_{\mD,prio}()$ is only defined for (leading) minimal diagnoses $\md \subseteq \mo$. Further on $p_{\mD,prio}()$ is normalized whereas $p_{nodes}()$ is not which accounts for the (normalization) constant $c$. The function $p_{nodes}()$ is essential for the best-first construction of the hitting set tree in \textsc{staticHS} and \textsc{dynamicHS} since it allows for the assignment of a ``probability'' to non-diagnoses (cf.\ the discussion of Formula~\ref{eq:path_prob_calc} on page~\pageref{eq:path_prob_calc}). 
Since the input argument $p()$ (which is the same for all calls) to \textsc{staticHS} as well as \textsc{dynamicHS} is equal to $p_{\mo}()$ by lines~\ref{algoline:inter_onto_debug:staticHS} and \ref{algoline:inter_onto_debug:dynamicHS} in Algorithm~\ref{algo:inter_onto_debug}, the set $\mD$ returned by \textsc{staticHS} (\textsc{dynamicHS}) is also the set of most probable minimal diagnoses w.r.t.\ $\langle\mo,\mb,\Tp,\Tn\rangle_\RQ$ ($\langle\mo,\mb,\Tp\cup\Tp',\Tn\cup\Tn'\rangle_\RQ$) \emph{as per the function $p_{nodes}()$} (cf.\ Proposition~\ref{prop:static_hs_correctness} and Corollary~\ref{cor:dynamic_hs_correctness}).\qed
\end{remark}

\begin{remark}\label{rem:return_params_of_stat+dynHS}
Notice that the return parameter that is relevant for the main purpose of Algorithm~\ref{algo:inter_onto_debug}, namely to compute a query and thereby obtain a new test case classified by the user, is solely the set of leading diagnoses $\mD$. The other return parameters serve as a means to store the state of the hitting set tree that is gradually built up by successive calls of \textsc{staticHS} (if $mode=static$) and \textsc{dynamicHS} (if $mode=dynamic$), respectively. Whereas $\Queue$ and $\mC_{calc}$ (and $\mD_{\supset}$ and $\Queue_{dup}$ in case of \textsc{dynamicHS}) are never modified until the next call to \textsc{staticHS} or \textsc{dynamicHS}, the sets $\mD_{\checkmark}$ and $\mD_{\times}$ are only changed once, after the subset of invalidated leading diagnoses $\mD_{out}$ is known, in lines~\ref{algoline:inter_onto_debug:update_D_checkmark} and \ref{algoline:inter_onto_debug:update_D_times}.\qed
\end{remark}

At this moment, we do not go into detail regarding the way how leading diagnoses are computed by \textsc{staticHS} and \textsc{dynamicHS}. We simply suppose that both functions act in a manner that the outputs just specified are returned for the given inputs. An in-depth delineation of both functions will be given in Chapters~\ref{chap:StaticHSTree} and \ref{chap:DynamicHSTree} in Part~\ref{part:IterativeDiagnosisComputation}. Further note that the return parameter $\mD$ is stored in variable $\mD_{\checkmark}$ from line~\ref{algoline:inter_onto_debug:dynamicHS} on. 

%\noindent\textbf{Computing a Probability Distribution of Leading Diagnoses.}\label{etc:computing_prob_dist_of_leading_diags} 
\paragraph{Computing a Probability Distribution of Leading Diagnoses.}\label{etc:computing_prob_dist_of_leading_diags}
After the set of leading diagnoses $\mD_{\checkmark}$ has been computed, the variables $\mD_{\checkmark}$, $p_{\mo}()$, $\langle\mo,\mb,\Tp,\Tn\rangle_\RQ$ and $QA$ are used as arguments to the function \textsc{getProbDist} (see Algorithm~\ref{algo:inter_onto_debug_continued}) which computes a probability distribution of the leading diagnoses, i.e.\ a probability measure $p_{\mD}()$ for the probability space with sample space $\Omega = \mD_{\checkmark}$ (cf.\ Section~\ref{sec:DiagnosisProbabilitySpace}).
% and the set of events $2^{\md_{\checkmark}}$. 
As a first action to achieve this, the (a-priori) probabilities $p_{\mD,prio}(\md)$ for $\md\in\mD_{\checkmark}$ are computed from the (a-priori) probabilities $p_{\mo}(\tax)$ for formulas $\tax\in\mo$ as per Formula~\ref{eq:diag_prob_calc} (\textsc{getPrioDiagProbs} in line~\ref{algoline:get_prob_dist:getPrioDiagProbs}). Application of Formula~\ref{eq:diag_prob_norm} is not necessary at this point as probabilities are anyhow normalized at the end of \textsc{getProbDist} (line~\ref{algoline:get_prob_dist:normalize}). Notice that the function $p_{\mo}()$ remains constant, i.e.\ unmodified, throughout the entire execution of Algorithm~\ref{algo:inter_onto_debug}.  
%and normalized such that $\sum_{\md\in\mD_{\checkmark}}w_{\mD,prio}(\md) = 1$.

Now, since a-priori diagnosis probabilities assigned by $p_{\mD,prio}()$ directly rely upon $p_{\mo}()$ which in turn is computed directly from the initially given fault probabilities $p_{\widetilde{\mo}\cup\overline{\mo}}()$, the probability measure $p_{\mD,prio}()$ is adapted to yield \emph{a-posteriori diagnosis probabilities} $p_{\mD}()$ in order to reflect the new evidence provided by the collected test cases $\Tp'$ and $\Tn'$.
%a more or less suitable guess of (a-priori) fault probabilities (weights) $w_{\widetilde{\mo}\cup\overline{\mo}}$, the algorithm is designed to compute a-posteriori diagnosis weights $w_{\mD}()$ by means of the a-priori diagnosis weights $w_{\mD,prio}()$ and the new evidence provided by the collected test cases $\Tp'$ and $\Tn'$. 

%Let $QA := \tuple{\tuple{Q,u(Q)}}_{Q\in\Tp'\cup\Tn'}$ where $u(Q) \in \setof{\true,\false}$ is the chronologically ordered list of queries and user answers collected so far during the execution of Algorithm~\ref{algo:inter_onto_debug}.
%\in \setof{\true,\false}^{|\Tp'|+|\Tn'|}$ 
%the chronologically ordered list of user answers collected so far during the execution of Algorithm~\ref{algo:inter_onto_debug} for the chronologically ordered list of queries $\tuple{Q}_{Q\in\Tp'\cup\Tn'}$. 
The a-posteriori probability of a current leading diagnosis $\md$ in $\mD_{\checkmark}$ is $p_{\mD}(\md\,|\,QA)$ and can be computed by means of Bayes' Theorem (Formula~\ref{eq:bayes}) from $p_{\mD,prio}()$ as follows.
%To this end, Bayes' Theorem is employed to update a-priori diagnosis probabilities as follows.
\begin{align*} 
p_{\mD}(\md\,|\,QA) = \frac{p_{\mD,prio}(QA\,|\,\md)\;\,p_{\mD,prio}(\md)}{p_{\mD,prio}(QA)} 
%\quad\mbox{ where } v \in \{\textit{true},\textit{false}\}
%\label{eq:bayes}
\end{align*}
where $QA$ is the chronologically ordered list of queries and user answers collected so far during the execution of Algorithm~\ref{algo:inter_onto_debug} (see page~\pageref{etc:QA}).
We point out that $p_{\mD,prio}(QA)$ is only a normalization factor that is equal for each diagnosis and thus does not need to be explicitly computed. The crucial factor is 
\begin{align*}
p_{\mD,prio}(QA\,|\,\md) = p_{\mD,prio}(\forall \tuple{Q,u(Q)} \in QA: Q = u(Q) \,|\,\md)
%$p_{\mD,prio}(QA\,|\,\md) = p_{\mD,prio}(\forall \tuple{Q,u(Q)} \in QA: Q = u(Q) \,|\,\md)$ 
\end{align*}
which describes the probability of getting exactly the answer $u(Q)$ for each query $Q\in \Tp'\cup\Tn'$ under the assumption that $\md$ corresponds to the true diagnosis $\dt$, i.e.\ $\dt = \md$. In other words, $p_{\mD,prio}(QA\,|\,\md)$ is the probability of $QA$ under the assumption that the user answers in a way that $u(Q) = \true$ if $\md \in \dx{}(Q)$ and $u(Q) = \false$ if $\md \in \dnx{}(Q)$. 

For a single query $Q_i$, the probability $p_{\mD,prio}(Q_i=u(Q_i)\,|\,\md)$ is defined as (cf. \cite{dekleer1987})
\begin{equation}\label{eq:cond_query_prob1}
p_{\mD,prio}(Q_i=u(Q_i)\,|\,\md) = 
\begin{cases}
1, 						& \mbox{if } \md \in \dx{}(Q_i) \\   %\mo^*_{k} \models Q_i\\
0, 						& \mbox{if } \md \in \dnx{}(Q_i) \\
   %\mo^*_{k} \cup Q_i \mbox{ violates some } x \in \RQ\cup \Tn\cup\Tn'' \\
\frac{1}{2}, 	& \mbox{if } \md \in \dz{}(Q_i) %$\md_j \in \dzi{s}$
\end{cases} 
\end{equation} 
for $u(Q_i) = \true$ and 
\begin{equation}\label{eq:cond_query_prob2}
p_{\mD,prio}(Q_i=u(Q_i)\,|\,\md) = 
\begin{cases}
1, 						& \mbox{if } \md \in \dnx{}(Q_i) \\   %\mo^*_{k} \models Q_i\\
0, 						& \mbox{if } \md \in \dx{}(Q_i) \\
   %\mo^*_{k} \cup Q_i \mbox{ violates some } x \in \RQ\cup \Tn\cup\Tn'' \\
\frac{1}{2}, 	& \mbox{if } \md \in \dz{}(Q_i) %$\md_j \in \dzi{s}$
\end{cases} 
\end{equation} 
for $u(Q_i) = \false$ where $\dx{}(Q_i)$, $\dnx{}(Q_i)$ and $\dz{}(Q_i)$ are computed w.r.t.\ the DPI $\langle\mo$, $\mb,\Tp\cup\Tp'',\Tn\cup\Tn''\rangle$ where $\Tp''$ and $\Tn''$, respectively, include all test cases collected \emph{prior} to $Q_i$, i.e.\ $\Tp''\cup\Tn'' = \setof{Q_1,\dots,Q_{i-1}}$ if queries are numbered chronologically. That is, if $\md$ predicted the answer $u(Q_i)$ to $Q_i$ given by the user, the probability is 1, zero if $\md$ predicted the converse answer $\lnot u(Q_i)$ and $\frac{1}{2}$ if $\md$ did not predict any answer to $Q_i$.
% (\mo\setminus\md)\cup\mb\cup U_{\Tp \cup \Tp''}
% (\mo\setminus\md)\cup\mb\cup U_{\Tp \cup \Tp''} \cup Q_i

So, aside from the normalization factor (see above), $p_{\mD,prio}(Q_i=u(Q_i)\,|\,\md)$ is the factor by which the a-priori probability $p_{\mD,prio}(\md)$ must be multiplied to obtain the a-posteriori probability $p_{\mD}(\md)$ of a diagnosis $\md$ after a single query $Q_i$ has been answered and added as a test case to the DPI.

The intuitive explanation for the update by this factor
% of the probability $p(Q_i=u(Q_i)\,|\,\md)$ as we did 
is that if $\md$ predicted (at least) one answer $u(Q)$ conversely as given by the user, then $\md$ is a-posteriori impossible since it has already been invalidated by the addition of test case $Q$.
In case a diagnosis has never predicted the wrong answer, but did not predict any answer for many queries so far, then it is a-posteriori more unlikely than a diagnosis that did predict a correct answer more often. That is, our a-posteriori degree of belief that $\md$ is the correct diagnosis is the higher, the more often $\md$ had predicted answers to queries that were later actually given by the user 
(cf.\ Section~\ref{sec:InterpretationOfQPartitions} for an explanation what we mean by ``predict''). 
%(cf. \textbf{Interpretation of a Q-Partition} on page~\pageref{etc:interpretation_q-partition} for an explanation what we mean by ``predict''). 

The value of $p_{\mD,prio}(Q_i=u(Q_i)\,|\,\md)$ can be computed by use of $QA$ and the q-partitions $\Pt(Q_1)$, $\dots$, $\Pt(Q_{i-1})$ of the \emph{current} set of leading diagnoses $\mD_{\checkmark}$ (for which a-posteriori probabilities are to be computed) for all queries $Q_1,\dots,Q_{i-1}$ answered before query $Q_i$. Thereby, each $\Pt(Q_j)$ where $j\in\setof{1,\dots,i-1}$ must be computed for a DPI where only $Q_1,\dots,Q_{j-1}$ are incorporated as test cases.

Taking these thoughts into account, \textsc{getProbDist} (Algorithm~\ref{algo:inter_onto_debug_continued}) updates $p_{\mD,prio}(\md)$ for each diagnosis $\md\in\mD_{\checkmark}$ in that it runs through all query-answer pairs $\tuple{Q,u(Q)}$ in $QA$ chronologically and for each $\md\in\mD_{\checkmark}$ it multiplies $p_{\mD,prio}(\md)$ by $\frac{1}{2}$ if $\md \in \dz{}(Q)$ as per Formulas~\ref{eq:cond_query_prob1} and \ref{eq:cond_query_prob2}. For each check whether a diagnosis is in $\dz{}(Q)$ in lines~\ref{algoline:get_prob_dist:check_dz_1} and \ref{algoline:get_prob_dist:check_dz_2} a DPI is used that already incorporates all test cases $\Tp''$ and $\Tn''$ that have been added chronologically before $Q$ was asked. This is achieved by updating $\Tp''$ and $\Tn''$ successively (lines~\ref{algoline:get_prob_dist:update_Tp''} and \ref{algoline:get_prob_dist:update_Tn''}). After all elements of $QA$ have been processed, the updated diagnosis probabilities are finally normalized (line~\ref{algoline:get_prob_dist:normalize}, cf. Formula~\ref{eq:diag_prob_norm} on page \pageref{eq:diag_prob_norm}) and the resulting function $p_{\mD,prio}()$ is returned. 

\begin{remark}\label{rem:remarks_ad_function_getProbDist}
Note that the function \textsc{getProbDist} exploits the fact that all diagnoses in $\md_{\checkmark}$ are leading diagnoses w.r.t.\ the \emph{current} DPI $\langle\mo,\mb,\Tp\cup\Tp',\Tn\cup\Tn'\rangle_\RQ$ which guarantees that none of these diagnoses has been invalidated by any of the test cases in $\Tp'$ or in $\Tn'$ added throughout the execution of Algorithm~\ref{algo:inter_onto_debug} (cf.\ Proposition~\ref{prop:diag_for_new_dpi_is_diag_for_old_dpi} given later). Hence, it is clear that each $\md \in \mD_{\checkmark}$ must be in $\dx{}(Q) \cup \dz{}(Q)$ if $u(Q) = \true$ and in $\dnx{}(Q) \cup \dz{}(Q)$ if $u(Q) = \false$, and it is only tested whether $\md \notin \dx{}(Q)$ in the prior case (line~\ref{algoline:get_prob_dist:check_dz_1}) and whether $\md \notin \dnx{}(Q)$ in the latter (line~\ref{algoline:get_prob_dist:check_dz_2}). 
It must be further noted that, in case of $mode=dynamic$, diagnoses in $\mD_{\checkmark}$ are not necessarily minimal diagnoses w.r.t.\ the intermediate DPIs $\langle\mo$, $\mb,\Tp\cup\Tp'',\Tn\cup\Tn''\rangle$ that are used for the probability update. However, this is not problematic since any set of (minimal and/or non-minimal) diagnoses is partitioned into the three sets $\dx{}(Q)$, $\dnx{}(Q)$ and $\dz{}(Q)$ by a query $Q$ (cf.\ Remark~\ref{rem:query_partitions_any_set_of_diagnoses_into_dx_dnx_dz}) wherefore $\Pt(Q)$ exists for any set $\mD_{\checkmark}$. Thence, the correctness of \textsc{getProbDist} remains unaffected by the usage of the setting $mode=dynamic$.\qed
\end{remark}
\begin{remark}\label{rem:prob_update_obsolete_if_dz(Q)={}_for_all_queries}
We want to emphasize that an adaptation of $p_{\mD,prio}(\md)$ is only necessary in case $\md \in \dz{}(Q_j)$ for some query $Q_j$ answered so far during the execution of Algorithm~\ref{algo:inter_onto_debug} as otherwise a multiplication by 1 is required which does not change $p_{\mD,prio}(\md)$.

For the case of static debugging ($mode=static$), an immediate implication of this is the following: The restriction of asking the user \emph{only} queries $Q_j$ w.r.t.\ a DPI with the property that \emph{no minimal diagnosis} w.r.t.\ this DPI can be an element of $\dz{}(Q_j)$ makes the probability update for each diagnosis in $\mD_{\checkmark}$ equivalent to a multiplication by 1 and hence obsolete. This must be the case since each diagnosis in $\mD_{\checkmark}$ which is a subset of $\minD_{\langle\mo,\mb,\Tp,\Tn\rangle_\RQ} \cap \minD_{\langle\mo,\mb,\Tp\cup\Tp',\Tn\cup\Tn'\rangle_\RQ}$ (see 
%Subsection \emph{Output} in 
Section~\ref{sec:Output}) must be a \emph{minimal} diagnosis w.r.t.\ each intermediate DPI (which includes a superset of the test cases in the input DPI $\langle\mo,\mb,\Tp,\Tn\rangle_\RQ$ and a subset of the test cases in the current DPI $\langle\mo,\mb,\Tp\cup\Tp',\Tn\cup\Tn'\rangle_\RQ$) as will be substantiated by Proposition~\ref{prop:after_adding_testcase_new_min_diag_is_equal_or_superset_of_old_min_diag} given later.
Consequently, such a scenario implicates that the order of diagnoses computed by \textsc{staticHS} corresponds to the best-first order also w.r.t.\ the a-posteriori diagnosis probabilities (cf.\ Remark~\ref{rem:absolute_vs_relative_fault_tolerance_AND_initial_vs_updated_probabilities}).

The approach of only using queries with this property is feasible, e.g.\ by using a \textsc{getEntailments} function in conformity with Proposition~\ref{prop:query_gen_explicit_entailments_dz_empty} for the generation of the query pool (\textsc{getPoolOfQueries}). Such a type of queries is also favorable from the discrimination point of view, as we pointed out in Section~\ref{sec:remarks_query_gen}. An improvement of static debugging with this type of queries is to deactivate the probability update, i.e.\ replace line~\ref{algoline:inter_onto_debug:getProbDist} in Algorithm~\ref{algo:inter_onto_debug} by line~\ref{algoline:get_prob_dist:getPrioDiagProbs} of Algorithm~\ref{algo:inter_onto_debug_continued}. This improvement is not shown in Algorithm~\ref{algo:inter_onto_debug}.

In a dynamic debugging session ($mode = dynamic$), on the contrary, the usage of such queries does not guarantee the triviality of the probability update. For, also if no minimal diagnosis w.r.t.\ the DPI (for which a query $Q_j$ is computed) can be an element of $\dz{}(Q_j)$, there may be some non-minimal one which is. For example, for any admissible DPI $\langle\mo,\mb,\Tp,\Tn\rangle_\RQ$ is holds that $\md := \mo$ is a diagnosis (cf.\ Proposition~\ref{prop:exist_diag} and Definition~\ref{def:admissible}), albeit in most cases a non-minimal one. In such a case, $(\mo \setminus \md) \cup \mb \cup U_{\Tp}$ which is equal to $\mb \cup U_{\Tp}$ cannot entail $Q_j$. Because, were this the case, then all minimal diagnoses $\md_i \in \minD_{\langle\mo,\mb,\Tp,\Tn\rangle_\RQ}$ would be elements of $\dx{}(Q_j)$ as each $\mo^*_i \supseteq \mb \cup U_{\Tp}$ and thus each $\mo^*_i \models Q_j$ by the monotonicity of $\mathcal{L}$. Hence, this would be a contradiction to the fact that $Q_j$ is a query w.r.t.\ $\langle\mo,\mb,\Tp,\Tn\rangle_\RQ$ by Corollary~\ref{cor:q-partition_dx_dnx}. On the other hand, $(\mo \setminus \md) \cup \mb \cup U_{\Tp} \cup Q_j = \mb \cup U_{\Tp} \cup Q_j$ cannot violate any $x \in \Tn\cup\RQ$. Since, if this were the case, then adding $Q_j$ to the positive test cases would lead to a non-admissible DPI $\langle\mo,\mb,\Tp\cup\setof{Q_j},\Tn\rangle_\RQ$. By Corollary~\ref{cor:query_leaves_valid_diag}, this would be a contradiction to the fact that $Q_j$ is a query w.r.t.\ $\langle\mo,\mb,\Tp,\Tn\rangle_\RQ$. Thence, $\md \in \dz{}(Q_j)$ must hold for the assumed non-minimal diagnosis $\md$. From that we conclude that the probability update in dynamic debugging cannot be made obsolete in general by the usage of such a type of queries.
\qed 
\end{remark}

%\noindent\textbf{Stop Criterion and Output.} 
\paragraph{Stop Criterion and Output.}
The (a-posteriori) probability distribution $p_{\mD}()$ of leading diagnoses $\mD_{\checkmark}$ is then used in line~\ref{algoline:inter_onto_debug:getMode} of Algorithm~\ref{algo:inter_onto_debug} to compute the mode of this distribution, i.e.\ the one diagnosis $\md_{\max}\in\mD_{\checkmark}$ with maximum probability according to $p_{\mD}()$. 

In the sequel, $\md_{\max}$ is used to check the stop criterion (line~\ref{algoline:inter_onto_debug:stop_crit}), namely whether $\md_{\max}$ has a probability greater than or equal to $1-\sigma$. If this is the case and $mode = static$, the function \textsc{getSolKB} computes a maximal solution KB w.r.t.\ the input DPI as $(\mo \setminus \md_{\max}) \cup U_{\Tp}$ by means of the current DPI $\langle\mo,\mb,\Tp\cup\Tp',\Tn\cup\Tn'\rangle_\RQ$, $\Tp'$ and $\md_{\max}$. Given that $mode = dynamic$, \textsc{getSolKB} returns a maximal solution KB w.r.t.\ the current DPI as $(\mo \setminus \md_{\max}) \cup U_{\Tp\cup\Tp'}$ by means of the current DPI $\langle\mo,\mb,\Tp\cup\Tp',\Tn\cup\Tn'\rangle_\RQ$ and $\md_{\max}$.
% $(\mo \setminus \md_{\max}) \cup U_{\Tp\cup\Tp'}$ by means of the current DPI $\langle\mo,\mb,\Tp\cup\Tp',\Tn\cup\Tn'\rangle_\RQ$ and $\md_{\max}$. 
This solution KB is then returned as an output of Algorithm~\ref{algo:inter_onto_debug}. If, on the other hand, the stop criterion is not met, the algorithm continues the execution with the computation of another query. 

\begin{remark}\label{rem:mode=static_=>_returned_solution_onto_extensible_to_be_solution_onto_w.r.t._current_DPI}
Notice that the returned maximal solution KB $(\mo \setminus \md_{\max}) \cup U_{\Tp}$ w.r.t.\ the input DPI in case $mode = static$ can be easily extended to constitute a maximal solution KB w.r.t.\ the current DPI, namely by extending it by $U_{\Tp'}$.  
%is a solution KB w.r.t.\ \emph{the input DPI and the current DPI }in case a static hitting set tree \textsc{staticHS} is employed to compute diagnoses. 
%If a dynamic hitting set tree is the used for generation of leading diagnoses, 
If $mode = dynamic$, then the KB output in line~\ref{algoline:inter_onto_debug:return} is a maximal solution KB w.r.t.\ \emph{the current DPI}, but possibly a non-maximal solution KB w.r.t.\ the input DPI. %(cf.\ \textbf{Output} on page~\pageref{etc:algo:inter_onto_debug:output}).
\qed
\end{remark}

%\noindent\textbf{Query Computation and User Interaction.} 
\paragraph{Query Computation and User Interaction.}
In line~\ref{algoline:inter_onto_debug:calc_query}, the function \textsc{calcQuery} 
%(Algorithm~\ref{algo:get_prob_dist}) 
is applied to compute a query and the associated q-partition by means of the leading diagnoses $\mD_{\checkmark}$, (possibly) the collected data $qData$, the probability distribution $p_{\mD}()$ of the leading diagnoses, a query selection function $qsm()$ (which might exploit the function $p_{\widetilde{\mo}\cup\overline{\mo}}()$), a parameter $q$ determining the size of the computed query pool 
%goodness of the returned query, 
and the current DPI $\langle\mo,\mb,\Tp\cup\Tp',\Tn\cup\Tn'\rangle_\RQ$. 

As a first step within \textsc{calcQuery}, the function \textsc{getPoolOfQueries} computes a query pool $\QP$ as detailed in Chapter~\ref{chap:QueryGeneration} from $\mD_{\checkmark}$, $q$ and $\langle\mo,\mb,\Tp\cup\Tp',\Tn\cup\Tn'\rangle_\RQ$.
% where $q$ determines the size of the returned pool $\QP$ of queries and associated q-partitions.
%
Then, the best tuple $\tuple{Q,\Pt(Q)} \in \QP$ according to the function $qsm()$ is searched for and finally returned as the output of \textsc{calcQuery}. During the query selection process, the evaluation of the query selection measure $qsm(Q) \in \mathbb{R}$ for queries $Q$ where $\tuple{Q,\Pt(Q)}\in\QP$ may require $qData$, the fault probabilities $p_{\mD}()$ of leading diagnoses as well as the fault probabilities $p_{\widetilde{\mo}\cup\overline{\mo}}()$ of syntactical elements in $\mo$. This depends on which concrete measure $qsm()$ is employed (see Section~\ref{sec:query_selection_measures} which presents some possible measures). 

As a next step, the query $Q$ of the best tuple $\tuple{Q,\Pt(Q)} \in \QP$ is presented to the interacting user in line~\ref{algoline:inter_onto_debug:user_interaction} which is the only place in Algorithm~\ref{algo:inter_onto_debug} where user interaction takes place. The user is modeled as a \emph{deterministic} function $u: \mQ_{\mD,\tuple{\mo,\mb,\Tp\cup\Tp',\Tn\cup\Tn'}} \rightarrow \setof{\true,\false}$ that allocates a positive ($\true$) or negative ($\false$) answer to each query w.r.t.\ any set of leading diagnoses $\mD$ for some current DPI $\tuple{\mo,\mb,\Tp\cup\Tp',\Tn\cup\Tn'}$. The answer $u(Q)$ given by the user is stored in the variable $answer$.

\begin{remark}\label{rem:remarks_to_the_user_function_for_answering_queries}
We want to point out that the algorithm can be easily adapted to allow a user to reject queries, e.g.\ if they are not sure how to answer. That is, the user function might be modeled as $u: \mQ_{\mD,\tuple{\mo,\mb,\Tp\cup\Tp',\Tn\cup\Tn'}} \rightarrow \setof{\true,\false, unknown}$ where $u(Q) = unknown$ signifies the rejection of query $Q$. In this case, an accordingly modified version of Algorithm~\ref{algo:inter_onto_debug} would calculate an alternative query w.r.t.\ $\mD$ and $\langle\mo$, $\mb,\Tp\cup\Tp',\Tn\cup\Tn'\rangle$, e.g.\ the second best one according to the query selection measure $qsm()$ among all tuples in $\QP$ (this potential feature is not shown in Algorithm~\ref{algo:inter_onto_debug}). In this vein, a total of $|\QP|-1$ queries can be dismissed per set of leading diagnoses $\mD$.

We want to accentuate that the presented interactive algorithm might be easily adapted to cope with queries whose answer is unknown to the user, but a definite assumption for the algorithm to return a correct solution is a user that does not give wrong answers. In other words, the algorithm does not provide inherent mechanisms that allow for the detection of wrong answers or for the debugging of the KB debugging procedure (keyword ``garbage in, garbage out''). So, we suppose the function $u()$ to be \emph{deterministic} which prohibits the situation that a user might change their mind at a later point in time. Of course, this is still a possible scenario in practice, but in case it arises, a user has to revise, i.e.\ delete or edit, specified test cases they disagree with by hand before a new debugging session using the modified DPI might be started.

Another remark at this place concerns the way a user might choose to answer the query. A ``minimal'' feedback of a user that we regard as an answer to a query $Q$ is to merely say $\true$, i.e.\ each formula in $Q$ (or the conjunction of formulas in $Q$) must be entailed by the correct KB, or $\false$, i.e.\ at least one formula in $Q$ (or the conjunction of formulas in $Q$) must not be entailed by the correct KB. The presented algorithm (Algorithm~\ref{algo:inter_onto_debug}) is designed to deal with exactly this kind of an answer. However, imagine a user being presented $Q$ and think of how they might proceed in order to come up with an answer to $Q$. The first observation is that, in order to respond by $\true$, a user must definitely scrutinize each single formula in $Q$ because otherwise they could never decide for sure whether the conjunction of all formulas in $Q$ is correct. Another observation is that a user might cease to go through the rest of the formulas in case they have already identified one that must not be an entailment of the desired KB. For, in this situation, the overall query $Q$ is already $\false$. This however indicates that at least one formula must be known to be correct or false whatever answer is given to $Q$. Therefore, we can usually expect a user to be able to give exactly this information, namely one formula in $Q$ that must be incorrect, additionally to answering by $\false$. This extra piece of information can be exploited to achieve better space and time efficiency in the context of diagnosis computation. Proposing more efficient algorithms that exploit this information is a topic for future work.
% \fixme{We shall see in Section~\ref{sec:TypesAndPropertiesOfTestCases} how this extra piece of information can be exploited to achieve better space and time efficiency in the context of diagnosis computation.}
\qed
\end{remark}  

%\noindent\textbf{Incorporating the New information.} 
\paragraph{Incorporating the New Information.}
The new information represented by the answer $answer$ to $Q$ is incorporated (lines~\ref{algoline:inter_onto_debug:param_update_start}-\ref{algoline:inter_onto_debug:param_update_end}) by updating values of all relevant parameters. First, by means of the function \textsc{append}, the tuple consisting of the answered query $Q$ and the corresponding answer $answer$ given by the user is added as a last element to the chronological list of queries and answers $QA$ that is used for the next probability update (line~\ref{algoline:inter_onto_debug:getProbDist}). 

Then, the subset $\mD_{out}$ of the leading diagnoses $\mD_{\checkmark}$ that gets invalidated after adding $Q$ to the 
%(positive if $answer = \true$ and negative otherwise) 
positive or negative test cases of the DPI, respectively, is computed by the function \textsc{getInvalidDiags} that gets the q-partition $\Pt(Q) = \tuple{\dx{}(Q),\dnx{}(Q),\dz{}(Q)}$ of $Q$ and $answer$ as input arguments. $\mD_{out}$ then corresponds to the set $\dnx{}(Q)$ given that $answer$ is $\true$ and to $\dx{}(Q)$ otherwise 
%(cf. \textbf{Interpretation of a Q-Partition} on page~\pageref{etc:interpretation_q-partition}). 
(cf.\ Section~\ref{sec:InterpretationOfQPartitions}). 
Note that $\emptyset \subset \mD_{out} \subset \mD_{\checkmark}$ holds by Proposition~\ref{prop:query_dx_dnx} and since $Q$ is a query w.r.t.\ $\mD_{\checkmark}$ (since $\mD_{\checkmark}$ is given as an input to \textsc{calcQuery}).

As a next step, the data $qData$ is updated. As already pointed out in 
%Subsection \emph{Variables} in 
Section~\ref{sec:Variables}, the form of the variable $qData$ depends on the employed query selection measure $qsm()$ and so do the actions that are performed by \textsc{updateQData}.

In order to communicate the impact of the answered query to the hitting set tree algorithm (either \textsc{staticHS} or \textsc{dynamicHS}), the set of invalidated leading diagnoses $\mD_{out}$ is deleted from the leading diagnoses $\mD_{\checkmark}$ and added to $\mD_{\times}$. After this update, $\mD_{\checkmark}$ includes all diagnoses that have been computed by the hitting set tree algorithm so far that are minimal diagnoses w.r.t.\ the current DPI. 
%$\mD_{\times}$, on the other hand, comprises all nodes that have been computed as minimal diagnoses w.r.t.\ some intermediate DPI by the hitting set tree algorithm so far that are invalid w.r.t.\ the current DPI (and have not yet been pruned).

Finally, the new test case $Q$ is added to the new positive test cases $\Tp'$ if $answer$ is $\true$ and to the new negative test cases $\Tn'$ in case of $answer = \false$.

\section{Query Selection Measures}
\label{sec:query_selection_measures}
In this section, we give a brief introduction to some query selection measures $qsm()$ that have been suggested and evaluated in literature within the scope of KB or ontology debugging \cite{Shchekotykhin2012,Rodler2013}. Such query selection measures, when used as a parameter in an interactive KB debugging algorithm such as the one described by Algorithm~\ref{algo:inter_onto_debug}, aim at solving the following optimization problems.
In Interactive Dynamic KB Debugging, the problem is defined as follows:\vspace{3pt}

\noindent\fcolorbox{black}{light-gray1}{\parbox[c][2em][c]{0.975\linewidth}{
\begin{prob_def}\label{prob_def:minimize_user_interact_dynamic}
The task is to solve the problem specified by Problem Definition~\ref{prob_def:dynamic} in a way that $|\Tp'|+|\Tn'|$ is minimal.
\end{prob_def}
}}

\vspace{3pt}
In Interactive Static KB Debugging, the problem is defined as follows:\vspace{3pt}

\noindent\fcolorbox{black}{light-gray1}{\parbox[c][2em][c]{0.975\linewidth}{
\begin{prob_def}\label{prob_def:minimize_user_interact_static}
The task is to solve the problem specified by Problem Definition~\ref{prob_def:static} in a way that $|\Tp'|+|\Tn'|$ is minimal.
\end{prob_def}
}}

\vspace{3pt}
That is, these optimization problems aim at the minimization of user effort during interactive KB debugging. In other words, the goal is the minimization of the number of queries required to be asked to a user in order to solve the Interactive Static KB Debugging or the Interactive Dynamic KB Debugging Problem, respectively.

In our previous work \cite{Shchekotykhin2012}, we have discussed entropy-based ($\mathsf{ENT}()$) and split-in-half ($\mathsf{SPL}()$) query selection measures. 

%\noindent\textbf{Entropy-Based Query Selection.} 
\paragraph{Entropy-Based Query Selection.}
A best query $Q_{\mathsf{ENT}}$ according to $\mathsf{ENT}()$ has a maximal information gain among all queries $Q$ where $\tuple{Q,\Pt(Q)}\in\QP$. In other words, $Q_{\mathsf{ENT}}$ minimizes the expected entropy of the 
%a-posteriori probability distribution $p_{\mD}()$ 
probability distribution
of the leading diagnoses $\mD_{\checkmark}$ after $Q_{\mathsf{ENT}}$ has been added as a test case to the DPI based on the user's answer $u(Q_{\mathsf{ENT}})$. As shown in \cite{dekleer1987}, this leads to the definition %of  $\mathsf{ENT}(Q)$ as
%Concretely, $\mathsf{ENT}(Q)$ is defined by the following formula:
\begin{align*}
\mathsf{ENT}(Q) := \sum_{a\in\setof{\true,\false}} p(Q=a) \log p(Q=a) + p(\dz{}(Q))
\end{align*}
where $p()$ in the case of our algorithm corresponds to the leading diagnoses probability measure $p_{\mD}()$ computed in line~\ref{algoline:inter_onto_debug:getProbDist} in Algorithm~\ref{algo:inter_onto_debug} and 
\begin{align*}
p(Q=\true) &= p(\dx{}(Q))+ \frac{1}{2} p(\dz{}(Q)) \\
p(Q=\false) &= p(\dnx{}(Q))+ \frac{1}{2} p(\dz{}(Q))
\end{align*}
(cf.\ Section~\ref{sec:InterpretationOfQPartitions}) where
\begin{align*}
p(\dx{}(Q)) &= \sum_{\md \in \dx{}(Q)} p(\md) \\
p(\dnx{}(Q))&= \sum_{\md \in \dnx{}(Q)} p(\md) \\
p(\dz{}(Q)) &= \sum_{\md \in \dz{}(Q)} p(\md)
\end{align*} 
%$p(Q=\true) = p(\dx{}(Q))$, $p(Q=\false) = p(\dnx{}(Q))$ (cf.\ \textbf{Interpretation of a Q-Partition} on page~\pageref{etc:interpretation_q-partition}). We discussed in Section~\ref{sec:DiagnosisProbabilitySpace} how such probabilities are computed.
Then, the best query in a pool $\QP$ according to $qsm():=\mathsf{ENT}()$ is 
\begin{align*}
Q_{\mathsf{ENT}} = \argmin_{\setof{Q\,|\,\tuple{Q,\Pt(Q)}\in\QP}} \mathsf{ENT}(Q)
\end{align*}
%$Q_{\mathsf{ENT}} = \arg\min_{\setof{Q\,|\,\tuple{Q,\Pt(Q)\in\QP}}} \mathsf{ENT}(Q)$.
So, theoretically optimal w.r.t.\ $\mathsf{ENT}()$ is a query $Q$ whose positive and negative answers are equally likely and for which $\dz{}(Q)$ is the empty set. In other words, the best query has the property that the \emph{sum of probabilities} of leading diagnoses predicting the positive answer as well as the \emph{sum of probabilities} of leading diagnoses predicting the negative answer is $50\%$.

%\noindent\textbf{Split-In-Half Query Selection.} 
\paragraph{Split-In-Half Query Selection.}
For the selection criterion $qsm():=\mathsf{SPL}()$, on the other hand, the query 
\begin{align*}
Q_{\mathsf{SPL}} = \argmin_{\setof{Q\,|\,\tuple{Q,\Pt(Q)}\in\QP}} \mathsf{SPL}(Q)
\end{align*}
%$Q_{\mathsf{SPL}} = \arg\min_{\setof{Q\,|\,\tuple{Q,\Pt(Q)\in\QP}}} \mathsf{SPL}(Q)$ 
is preferred where 
%$\mathsf{SPL}(Q)$ is defined as 
\begin{align*}
\mathsf{SPL}(Q) := \left|\, |\dx{}(Q)| - |\dnx{}(Q)| \,\right| + |\dz{}(Q)|
\end{align*} 
Hence, this measure is optimized by queries $Q$ for which the \emph{number} of leading diagnoses predicting the positive answer is equal to the \emph{number} of leading diagnoses predicting the negative answer and for which $\dz{}(Q)$ is the empty set.

%\noindent\textbf{Risk-Optimized Query Selection.} 
\paragraph{Risk-Optimized Query Selection.}
For scenarios where a-priori probabilities are vague, we have presented another more complex query selection measure $\mathsf{RIO}()$ in \cite{Rodler2013} which uses a reinforcement learning strategy to constantly adapt some ``risk'' parameter that indicates the current amount of trust in the probabilities. Whereas $\mathsf{ENT}()$ and $\mathsf{SPL}()$ do not rely on $qData$, this learning strategy does so and requires the invalidation rate or ``performance'', i.e.\ $\frac{|\mD_{out}|}{|\mD_{\checkmark}|}$, of the previous iteration for the adaptation of the learning parameter. As long as the invalidation rate is ``good'', the trust in the current (a-posteriori) probabilities -- that strongly depend on the vague a-priori probabilities -- is high, but it is gradually decreased after observing ``worse'' performance, and so on. High trust in the probabilities means usage of $\mathsf{ENT}()$ which can exploit high quality fault information well as demonstrated in the experiments conducted in \cite{Shchekotykhin2012}, whereas low trust involves selection of queries that guarantee a higher worst case invalidation rate, i.e.\ have similar properties to queries $\mathsf{SPL}()$ would select. 
%In this vein, when applying $\mathsf{ENT}$ as measure $m$ for query computation, a user can profit from a good prior fault estimation w.r.t.\ the number of queries that need to be answered to identify the true diagnosis, but may at the same time have to put up with a serious overhead in answering effort in case of poor estimates that assign a low probability to the true diagnosis.
%%In this vein, $\mathsf{ENT}$ can profitize from a good prior fault estimation in terms of number of queries, but may at the same time seriously suffer from poor estimates that assign a low probability to the true diagnosis.
%%where the true diagnosis is assigned a low probability. 
%$\mathsf{SPL}$, to the opposite, refrains from using any probabilities and aims at maximizing the elimination rate by selecting a query that guarantees invalidation of the half set of leading diagnoses $\mD$. As a consequence, $\mathsf{SPL}$ is generally inferior to $\mathsf{ENT}$ for good estimates and superior to $\mathsf{ENT}$ for misleading probabilities, as experiments conducted in~\cite{Shchekotykhin2012,Rodler2013} indicate.
%\fixme{insert example showing different queries for example DPI 1, add these queries to the Table that shows the queries for example DPI 1}
\begin{example}\label{example:query_selection_measures}
Let us reconsider the queries and associated q-partitions for the example DPI of Table~\ref{tab:example1} that are depicted by Table~\ref{tab:queries_partitions} on page~\pageref{tab:queries_partitions}. Let us denote by $Q_i \prec_{M} Q_j$ that $Q_i$ is preferred over $Q_j$ and by $Q_i \prec\succ_{M} Q_j$ that $Q_i$ is equally preferable as $Q_j$ if the query selection measure $qsm() := M$ is used. Furthermore, we make the assumption that the probability distribution $p_{\mD}$ of the (leading) diagnoses $\mD_{\checkmark} = \setof{\md_1,\dots,\md_4}$ is as shown in Table~\ref{tab:example:query_selection_measures--->diag_probs}. 

Then, we make the following observations: 
\begin{itemize}
	\item $Q_6$ is the theoretically optimal query w.r.t.\ $\mathsf{ENT}()$ since $p_{\mD}(\dx{}(Q_6)) = 0.5$, $p_{\mD}(\dnx{}(Q_6)) = 0.5$ and $\dz{}(Q_6) = \emptyset$, i.e.\ the positive and the negative answer have equal probabilities of $50\%$ and thus $Q_6$ the highest theoretically possible information gain of 1 (bit). This can be compared with one toss of a coin where the information gain of tossing the coin and checking whether it is head or tail is highest in a case where the coin is fair. For a coin that shows head with a probability of $0.95$, conversely, the information gain of tossing the coin is rather small since we are already quite sure about the result in advance.
	\item $Q_9 \prec_{M} Q_5$ as well as $Q_9 \prec_{M} Q_2$ for $M \in \setof{\mathsf{SPL}(),\mathsf{ENT}()}$ because both $Q_5$ and $Q_2$ share one set in $\setof{\dx{},\dnx{}}$ with $Q_9$, but exhibit a non-empty set $\dz{}$ whereas $\dz{}(Q_9) = \emptyset$. This shows that both split-in-half and entropy-based query selection penalize a query $Q$ if there are leading diagnoses that are definitely not discriminated by it, i.e.\ $\dz{}(Q) \neq \emptyset$. This is perfectly desirable as we discussed.
	\item $Q_4 \prec\succ_{M} Q_{10}$ for $M \in \setof{\mathsf{SPL}(),\mathsf{ENT}()}$ since their q-partitions differ just by commutation of the sets $\dx{}$ and $\dnx{}$. This is what one would expect of such a measure, i.e.\ that it does not matter whether the positive or negative answer is more probable if the probability values are the same (in case of $\mathsf{ENT}()$) and whether the number of diagnoses predicting the positive or negative answer is higher if the numbers are the same (in case of $\mathsf{SPL}()$). However, notice that $Q_4$ might be much easier to comprehend and answer for the interacting user. Therefore, $Q_4$ might be preferred in a scenario where some second measure $qsm_2()$ comes into play to identify a best query among equally preferable queries w.r.t.\ some $qsm_1()$ that is used as a primary measure. For, example some ``query-easiness'' measure $qsm_2()$ might be employed after $qsm_1() \in \setof{\mathsf{SPL}(),\mathsf{ENT}()}$ has filtered out an equally preferable set of queries; in this case let this set be $\setof{Q_4,Q_{10}}$. The measure $qsm_2()$ could be defined to simply count the logical connectives and quantifiers occurring in a query $Q$ 
%(i.e.\ to determine $|\overline{Q}|$) 
and pick one for which this number is minimal. In this case, this number would be 0 for $Q_4$ and 7 for $Q_{10}$, wherefore $Q_4$ would be decisively better than $Q_{10}$ w.r.t.\ $qsm_2()$. 
	\item It holds that $Q_3 \prec_{\mathsf{ENT()}} Q_{10} \prec_{\mathsf{ENT()}} Q_1$, but $Q_3 \prec\succ_{\mathsf{SPL()}} Q_{10} \prec\succ_{\mathsf{SPL()}} Q_1$. The former holds since all three queries feature an empty set $\dz{}$, but the difference between $p(\dx{})$ and $p(\dnx{})$ is largest for $Q_1$ ($p(\dx{}(Q_1)) = 0.95$), second largest for $Q_{10}$ ($p(\dnx{}(Q_{10})) = 0.85$) and smallest for $Q_3$ ($p(\dx{}(Q_3)) = 0.7$).
	\item $Q_9$ is the second best query among those given in Table~\ref{tab:queries_partitions} because both answers of it are almost equally probable (positive answer has a probability of 0.55 and negative answer a probability of $0.45$).
	\item Queries $Q_7$, $Q_8$ and $Q_9$ are theoretically optimal w.r.t.\ the $\mathsf{SPL()}$ measure, since $\dz{} = \emptyset$ and $|\dx{}| = |\dnx{}|$ for all of them.
	\item Regarding the $\mathsf{RIO()}$ measure, queries $Q_7$, $Q_8$ and $Q_9$ are ``no risk'' queries since they feature the maximum possible worst case elimination rate of $50\%$. $Q_2$ and $Q_6$, for instance, have a ``higher risk'' as their minimal invalidation rate amounts to only $25\%$. That is, if $Q_2$ ($Q_6$) is answered positively (negatively), then only one of four leading diagnoses is invalidated.\qed
\end{itemize}
\end{example}

\begin{table}[tb]
	\centering
		\begin{tabular}{lcccc}
			$\md \in \mD_{\checkmark}$ & $\md_1$ & $\md_2$ & $\md_3$ & $\md_4$ \\\hline
			$p_{\mD}(\md)$             & 0.15  & 0.3  & 0.05  & 0.5  
		\end{tabular}
\caption[(Example~\ref{example:query_selection_measures}) Diagnoses Probabilities]{(Example~\ref{example:query_selection_measures}) Diagnosis probabilities for the example DPI given by Table~\ref{tab:example1}.}
\label{tab:example:query_selection_measures--->diag_probs}
\end{table}

\section[Correctness and Complexity]{Interactive Debugging Algorithm: Correctness and Complexity%
\sectionmark{Algorithm Correctness and Complexity}}
\sectionmark{Algorithm Correctness and Complexity}
\label{sec:CorrectnessOfAlgorithmInterOntoDebug}
%%%%%
%\section{Interactive Debugging Algorithm: Correctness and Complexity}
%\label{sec:CorrectnessOfAlgorithmInterOntoDebug}
%%%%%
First, we prove the correctness of Proposition~\ref{prop:correctness_of_interactive_KB_debugging_algo} on page~\pageref{prop:correctness_of_interactive_KB_debugging_algo} by using the results of Sections~\ref{sec:CorrectnessOfTextscStaticHS} and \ref{sec:CorrectnessOfTextscDynamicHS} which provide evidence for the correctness (soundness, completeness and optimality) of methods \textsc{staticHS} and \textsc{dynamicHS}:
%\begin{proposition}
%Algorithm~\ref{algo:inter_onto_debug} terminates and returns a solution KB $\ot$ as specified in Output on page~\pageref{etc:algo:inter_onto_debug:output}.
%\end{proposition}
\begin{proof}[Proof of Proposition~\ref{prop:correctness_of_interactive_KB_debugging_algo}]
First, we argue why Algorithm~\ref{algo:inter_onto_debug} must terminate. The function \textsc{getFormulaProbs} in line~\ref{algoline:inter_onto_debug:getAxiomProbs} terminates since it applies Formulas~\ref{eq:ax_prob_calc} and \ref{eq:adapt_ax_prob_to_get_min_diags} $|\mo|$ times and $|\mo|$ is finite by Definition~\ref{def:dpi}. If $mode=static$, then \textsc{staticHS} terminates due to Proposition~\ref{prop:static_hs_correctness}. If $mode=dynamic$, then \textsc{dynamicHS} terminates due to Corollary~\ref{cor:dynamic_hs_correctness}. \textsc{getProbDist} terminates since (1)~the number of already answered queries $|QA|$ is finite, (2)~$|\mD_{\checkmark}|$ is finite since diagnoses are subsets of $\mo$ and thus there is only a finite number of (minimal) diagnoses w.r.t.\ any DPI according to Definition~\ref{def:dpi} (since all sets included in the DPI are finite) and (3)~reasoning (\textsc{getEntailments} and \textsc{isKBValid}) is assumed to be decidable for the logic $\mathcal{L}$ over which the DPI is formulated as per Chapter~\ref{chap:basics}. Further, \textsc{getMode} clearly terminates due to the fact that $|\mD_{\checkmark}|$ is finite and returns the mode $\md_{\max}$ of the diagnoses probability distribution $p_{\mD}()$ over the diagnoses in $\mD_{\checkmark}$. Now, if the stop criterion $p_{\mD}(\md_{\max}) \geq 1 -\sigma$ is met, then \textsc{getSolKB} is called. \textsc{getSolKB} simply deletes the given diagnosis $\md_{\max}$ from the given KB $\mo$ and adds a finite set of formulas to it, and thence terminates.

If the stop criterion is not met, then $|\mD_{\checkmark}| \geq 2$ must hold as otherwise the single diagnosis $\md \in \mD_{\checkmark}$ would necessarily have fulfilled the stop criterion as its probability as per any probability measure over the sample space $\Omega := \mD_{\checkmark}$ must be equal to 1 and thus greater than or equal to $1 - \sigma$ where $\sigma \geq 0$. 

Due to $|\mD_{\checkmark}| \geq 2$, Proposition~\ref{prop:getPoolOfQueries_correctness} implies that
%since $\mD_{\checkmark}$ is a set of leading diagnoses w.r.t.\ some DPI $DPI$, there are at least $|\mD_{\checkmark}|$ queries w.r.t.\ $DPI$ due to Corollary~\ref{cor:query_num_lower_bound}. Note that the (explicit) entailments that must be computed to guarantee the validity of Corollary~\ref{cor:query_num_lower_bound} can be \emph{trivially} obtained by any \textsc{getEntailments} function (cf.\ Proposition~\ref{prop:q1}). Hence, 
\textsc{getPoolOfQueries} (called within \textsc{calcQuery}) terminates and yields a non-empty query pool as output. \textsc{selectBestQuery} (also called within \textsc{calcQuery}) terminates as well since it simply selects one query from the pool according to the measure $qsm()$ (cf.\ Section~\ref{sec:query_selection_measures}). Since we assume the interacting user to answer to a query or to reject it within finite time, $u(Q)$ also terminates. It is clear that \textsc{append} terminates. \textsc{getInvalidDiags} simply extracts one entry of the given q-partition and thus terminates. Finally, \textsc{updateQData} also terminates by assumption (no $qsm()$ must be used for which \textsc{updateQData} might not terminate). As a consequence, all functions called in Algorithm~\ref{algo:inter_onto_debug} terminate. What remains to be proven is that the stop criterion must be met after a finite number of iterations, i.e.\ after a finite number of test cases have been added to the input DPI.

In $mode=static$ the stop criterion must be satisfied after a finite number of iterations due to the following argumentation: 
\begin{itemize}
	\item There is a finite set of minimal diagnoses w.r.t.\ the input DPI $\tuple{\mo,\mb,\Tp,\Tn}_{\RQ}$ since each (minimal) diagnosis w.r.t.\ this DPI is a subset of $\mo$ according to Definition~\ref{def:diagnosis} and since $|\mo|$ is finite by Definition~\ref{def:dpi}.
	\item In each iteration, one test case is added either to $\Tp'$ or $\Tn'$.
	\item Each test case added to whatever set $\Tp'$ or $\Tn'$ invalidates at least one minimal diagnosis w.r.t.\ the input DPI in the set $\mD_{\checkmark}$ by the definition of a query (Definition~\ref{def:query}) and since each query is computed w.r.t.\ the leading diagnoses $\mD_{\checkmark}$ by the correctness of \textsc{getPoolOfQueries} (cf.\ Proposition~\ref{prop:getPoolOfQueries_correctness}).
	\item $\mD_{\checkmark}$ contains only minimal diagnoses w.r.t.\ the input DPI by Proposition~\ref{prop:static_hs_correctness}.
	\item Also by Proposition~\ref{prop:static_hs_correctness}, no invalidated minimal diagnosis w.r.t.\ the input DPI can be an element of some subsequent set of leading diagnoses $\mD_{\checkmark}$.
	\item Therefore, unless the stop criterion is met before due to a sufficiently high probability of one of multiple leading diagnoses as per $p_{\mD}()$, Algorithm~\ref{algo:inter_onto_debug} in $mode=static$ must arrive at a point where $|\mD_{\checkmark}| = 1$ after a finite number of iterations. Note that $|\mD_{\checkmark}| = 0$ is impossible due to the definition of a query (Definition~\ref{def:query}) which ensures that each added test case leaves valid at least one minimal diagnosis in $\mD_{\checkmark}$.
\end{itemize}

%Hence, terminate after a finite number of iterations (i.e.\ answered queries).

%Algorithm~\ref{algo:inter_onto_debug} terminates in $mode=dynamic$ since there is a finite set of DPIs that comprise a proper superset of the test cases included in the input DPI and for which there is more than one (minimal) diagnosis. Assume the opposite holds. Then we argue as follows to derive a contradiction: 
Algorithm~\ref{algo:inter_onto_debug} terminates in $mode=dynamic$ since for any sequence $QA$ of queries that are added to the positive or negative test cases $\Tp'$ or $\Tn'$, respectively, there is a finite number $k_{QA}$ such that there is no more than one minimal diagnosis w.r.t.\ $\tuple{\mo,\mb,\Tp\cup\Tp',\Tn\cup\Tn'}_{\RQ}$ for $|\Tp'| + |\Tn'| = k_{QA}$ wherefore the stop criterion must be met.
%there is a finite set of DPIs that comprise a proper superset of the test cases included in the input DPI and for which there is more than one (minimal) diagnosis. 
Now, let us assume that the opposite holds. That is, there is a sequence $QA^*$ of queries that are added to the positive or negative test cases $\Tp'$ or $\Tn'$, respectively, and for all natural numbers $k$ there is more than one minimal diagnosis w.r.t.\ $\tuple{\mo,\mb,\Tp\cup\Tp',\Tn\cup\Tn'}_{\RQ}$ for $|\Tp'| + |\Tn'| = k$. Then we argue as follows to derive a contradiction:
\begin{itemize}
	\item There is a finite set of (minimal) diagnoses w.r.t.\ any DPI $\tuple{\mo,\mb,\Tp\cup\Tp',\Tn\cup\Tn'}_{\RQ}$ obtained from the input DPI by the addition of test cases. This is true since $|\mo|$ is finite by Definition~\ref{def:dpi} and since each (minimal) diagnosis w.r.t.\ $\tuple{\mo,\mb,\Tp\cup\Tp',\Tn\cup\Tn'}_{\RQ}$ is a subset of $\mo$ according to Definition~\ref{def:diagnosis}.
	\item In each iteration, one test case is added either to $\Tp'$ or $\Tn'$.
	\item Each test case added to whatever set $\Tp'$ or $\Tn'$ invalidates at least one minimal diagnosis w.r.t.\ the current DPI in the set $\mD_{\checkmark}$ by the definition of a query (Definition~\ref{def:query}) and since each query is computed w.r.t.\ the leading diagnoses $\mD_{\checkmark}$ by the correctness of \textsc{getPoolOfQueries} (cf.\ Proposition~\ref{prop:getPoolOfQueries_correctness}).
	\item If $DPI$ denotes the current DPI at the time \textsc{dynamicHS} is called, then the set $\mD_{\checkmark}$ returned by \textsc{dynamicHS} is a subset of or equal to $\minD_{DPI}$, i.e.\ $\mD_{\checkmark}$ contains only minimal diagnoses w.r.t.\ $DPI$ by Corollary~\ref{cor:dynamic_hs_correctness}.
	%\item For each invalidated minimal diagnosis $\md$ w.r.t.\ the DPI $DPI_j$ there can be no minimal diagnosis $\md'$ w.r.t.\ any subsequent $DPI_{j'}$ such that $\md' \subseteq \md$ by Proposition~\ref{prop:adding_testcase_cannot_make_min_diags_shrink} where $DPI_{j'}$ includes a proper superset of the test case contained in $DPI_j$. 
	%\item In case a minimal diagnosis $\md$ w.r.t.\ $DPI_j$ is not invalidated by the test case added to $DPI_j$ to obtain $DPI_{j+1}$, $\md$ is a minimal diagnosis w.r.t.\ $DPI_{j+1}$. That $\md$ is a diagnosis w.r.t.\ $DPI_{j+1}$ follows from Proposition~\ref{prop:dpi_update} and that it must be minimal is a consequence of the fact that the existence of some minimal diagnosis $\md'' \subset \md$ w.r.t.\ $DPI_{j+1}$ would be a contradiction to the minimality of $\md$ w.r.t.\ $DPI_{j}$ by Proposition~\ref{prop:adding_testcase_cannot_make_min_diags_shrink}. 
	\item Let $\tuple{DPI_0, DPI_1,\dots}$ denote the sequence of DPIs encountered in the case of adding answered queries as test cases to the input DPI $DPI_0$ as per $QA^*$. Further, let $\tuple{\allD_0, \allD_1, \dots}$ be the sequence 
such that $\allD_i:= \allD_{DPI_i}, i=0,1,\dots$, i.e.\ $\allD_i$ is the set of all diagnoses w.r.t.\ $DPI_i$. Then $\allD_i \supset \allD_{i+1}$ for all $i \geq 0$ due to Corollary~\ref{cor:adding_query_to_DPI_implies_that_allD_wrt_new_DPI_is_proper_subset_of_allD_wrt_old_DPI}.
%Proposition~\ref{prop:diag_for_new_dpi_is_diag_for_old_dpi} and the fact that at least one (minimal) diagnosis in $\allD_i$ is invalidated by the test case that is added to $DPI_i$ to obtain $DPI_{i+1}$.	
	\item As each query added as a test case to $DPI_i$ leaves valid at least one (minimal) diagnosis w.r.t.\ $DPI_i$ due to Definition~\ref{def:query}, we have that $\allD_k \supset \emptyset$ for $k = 0,1,\dots$.
	\item Since $\allD_i$ is finite, there must be some finite number $k^*$ such that $|\allD_{k^*}| = 1$ wherefore $|\minD_{k^*}| = 1$ must also be valid. This is a contradiction.
\end{itemize}
Thence, Algorithm~\ref{algo:inter_onto_debug} terminates in any mode $mode$. 
Now, we show that propositions (\ref{prop:correctness_of_interactive_KB_debugging_algo:stat1})-(\ref{prop:correctness_of_interactive_KB_debugging_algo:stat6}) of Proposition~\ref{prop:correctness_of_interactive_KB_debugging_algo} hold for (i)~$mode=static$ and (ii)~$mode=dynamic$.

(i): First, by the proof so far, we have that Algorithm~\ref{algo:inter_onto_debug} in $mode=static$ given the input DPI $\langle\mo,\mb,\Tp,\Tn\rangle_\RQ$ terminates. Since the only point where the algorithm can terminate is line~\ref{algoline:inter_onto_debug:return}, \textsc{getSolKB} is called with arguments $\tuple{\md_{\max}, \langle\mo,\mb,\Tp\cup\Tp',\Tn\cup\Tn'\rangle_\RQ, \Tp', static}$. By the definition of \textsc{getSolKB} (see Section~\ref{sec:Walkthrough}), we have that $(\mo \setminus \md_{\max}) \cup U_{\Tp}$ is returned by the algorithm.

Propositions~(\ref{prop:correctness_of_interactive_KB_debugging_algo:stat1}) and (\ref{prop:correctness_of_interactive_KB_debugging_algo:stat2}) follow from the specification of the \textsc{getMode} function which is called with arguments $\tuple{\mD_{\checkmark},p_{\mD}()}$. Proposition~(\ref{prop:correctness_of_interactive_KB_debugging_algo:stat3}) is true since \textsc{getSolKB} can never be reached without $p_{\mD}(\md_{\max}) \geq 1 -\sigma$ being fulfilled. 
$\mD_{\checkmark} \subseteq \minD_{\langle\mo,\mb,\Tp,\Tn\rangle_\RQ} \cap \minD_{\langle\mo,\mb,\Tp\cup\Tp',\Tn\cup\Tn'\rangle_\RQ}$ is true due to Proposition~\ref{prop:static_hs_correctness}, Remark~\ref{rem:p_D,prio()_same_order_for_diags_as_p_nodes()} and the fact that $\mD_{\checkmark}$ is obtained as an output of \textsc{staticHS}. Hence, Proposition~(\ref{prop:correctness_of_interactive_KB_debugging_algo:stat4}) holds. Proposition~(\ref{prop:correctness_of_interactive_KB_debugging_algo:stat5}) is implied by Remark~\ref{rem:p_D,prio()_same_order_for_diags_as_p_nodes()} and by the specification of the \textsc{getFormulaProbs} function which computes $p_{\mo}()$ from $p_{\widetilde{\mo} \cup \overline{\mo}}()$ as per Formulas~\ref{eq:ax_prob_calc} and \ref{eq:adapt_ax_prob_to_get_min_diags} in line~\ref{algoline:inter_onto_debug:getAxiomProbs}. Finally, Proposition~(\ref{prop:correctness_of_interactive_KB_debugging_algo:stat6}) is a consequence of the definition of the \textsc{getProbDist} function which accounts for the computation of $p_{\mD}()$ from $p_{\mo}()$, the input DPI, $\mD_{\checkmark}$ and the chronological sequence of all queries and associated answers $QA$ so far. Therefore, Proposition~\ref{prop:correctness_of_interactive_KB_debugging_algo} is true for $mode=static$.

(ii): First, by the proof so far, we have that Algorithm~\ref{algo:inter_onto_debug} in $mode=dynamic$ given the input DPI $\langle\mo,\mb,\Tp,\Tn\rangle_\RQ$ terminates. Since the only point where the algorithm can terminate is line~\ref{algoline:inter_onto_debug:return}, \textsc{getSolKB} is called with arguments $\tuple{\md_{\max}, \langle\mo,\mb,\Tp\cup\Tp',\Tn\cup\Tn'\rangle_\RQ, \Tp', dynamic}$. By the definition of \textsc{getSolKB} (see Section~\ref{sec:Walkthrough}), we have that $(\mo \setminus \md_{\max}) \cup U_{\Tp\cup\Tp'}$ is returned by the algorithm.

Propositions~(\ref{prop:correctness_of_interactive_KB_debugging_algo:dyn1}) and (\ref{prop:correctness_of_interactive_KB_debugging_algo:dyn2}) follow from the specification of the \textsc{getMode} function which is called with arguments $\tuple{\mD_{\checkmark},p_{\mD}()}$. Proposition~(\ref{prop:correctness_of_interactive_KB_debugging_algo:dyn3}) is true since \textsc{getSolKB} can never be reached without $p_{\mD}(\md_{\max}) \geq 1 -\sigma$ being fulfilled. 
$\mD_{\checkmark} \subseteq \minD_{\langle\mo,\mb,\Tp\cup\Tp',\Tn\cup\Tn'\rangle_\RQ}$ is true due to Corollary~\ref{cor:dynamic_hs_correctness}, Remark~\ref{rem:p_D,prio()_same_order_for_diags_as_p_nodes()} and the fact that $\mD_{\checkmark}$ is obtained as an output of \textsc{dynamicHS}. Hence, Proposition~(\ref{prop:correctness_of_interactive_KB_debugging_algo:dyn4}) holds. Proposition~(\ref{prop:correctness_of_interactive_KB_debugging_algo:dyn5}) is implied by Remark~\ref{rem:p_D,prio()_same_order_for_diags_as_p_nodes()} and by the specification of the \textsc{getFormulaProbs} function which computes $p_{\mo}()$ from $p_{\widetilde{\mo} \cup \overline{\mo}}()$ as per Formulas~\ref{eq:ax_prob_calc} and \ref{eq:adapt_ax_prob_to_get_min_diags} in line~\ref{algoline:inter_onto_debug:getAxiomProbs}. Finally, Proposition~(\ref{prop:correctness_of_interactive_KB_debugging_algo:dyn6}) is a consequence of the definition of the \textsc{getProbDist} function which accounts for the computation of $p_{\mD}()$ from $p_{\mo}()$, the input DPI, $\mD_{\checkmark}$ and the chronological sequence of all queries and associated answers $QA$ so far. Therefore, Proposition~\ref{prop:correctness_of_interactive_KB_debugging_algo} is true for $mode=dynamic$. 

Next, we show that the solution to Interactive Static KB Debugging is found for $\sigma = 0$ in case $mode = static$:
\begin{enumerate}[(s1)]
	\item $\mD_{\checkmark} \subseteq \minD_{\langle\mo,\mb,\Tp,\Tn\rangle_\RQ} \cap \minD_{\langle\mo,\mb,\Tp\cup\Tp',\Tn\cup\Tn'\rangle_\RQ}$ holds for the output of \textsc{staticHS} in each iteration by Proposition~\ref{prop:static_hs_correctness}. Therefore, $\mD_{\checkmark}$ comprises only minimal diagnoses w.r.t.\ the input DPI that comply with all specified test cases in $\Tp'$ and $\Tn'$.
	\item \label{proof_int_debug_correct:bullet_2} By $p_{\widetilde{\mo} \cup \overline{\mo}}(): \widetilde{\mo} \cup \overline{\mo} \rightarrow (0,1]$ we derive by Formula~\ref{eq:ax_prob_calc} that each formula in $\mo$ must have a probability greater than zero. Further, by Formula~\ref{eq:adapt_ax_prob_to_get_min_diags}, no formula in $\mo$ can have a probability greater than or equal to $0.5$ (i.e.\ in particular a probability of 1 is not possible for a formula). Hence, we have that $p_{\mo}: \mo \rightarrow (0,0.5)$ for the measure $p_{\mo}()$ computed by \textsc{getFormulaProbs} in line~\ref{algoline:inter_onto_debug:getAxiomProbs} in Algorithm~\ref{algo:inter_onto_debug}. Thence, by the definition of $p_{nodes}()$ in \textsc{staticHS} based on $p() := p_{\mo}()$ (cf.\ Definition~\ref{def:p_node()} on page~\pageref{def:p_node()}) due to the fact that $p_{\mo}()$ is given as an input argument to $\textsc{staticHS}$ in line~\ref{algoline:inter_onto_debug:staticHS}, we have that no diagnosis can have an (a-priori) probability of zero. Since the function \textsc{getProbDist} might only perform some multiplications of a diagnosis probability by $\frac{1}{2}$, also the a-posteriori  probability of each diagnosis must be greater than zero.
	\item \label{proof_int_debug_correct:bullet_3} Hence, due to $\sigma = 0$, it must be necessarily be true that $|\mD_{\checkmark}| = 1$ before the algorithm terminates.
	\item By Problem Definition~\ref{prob_def:static} and the specification of the \textsc{getSolKB} function, the output solution KB must be the solution to Interactive Static KB Debugging.
\end{enumerate}
That a solution found for $\sigma > 0$ in case $mode = static$ might be an approximate solution to Interactive Static KB Debugging is a direct consequence of the definition of approximate solution given in Remark~\ref{rem:approximate_solution}.
 %Because, $\mD_{\checkmark} \subseteq \minD_{\langle\mo,\mb,\Tp,\Tn\rangle_\RQ} \cap \minD_{\langle\mo,\mb,\Tp\cup\Tp',\Tn\cup\Tn'\rangle_\RQ}$ holds for the output of \textsc{staticHS} in each iteration by Proposition~\ref{prop:static_hs_correctness}. Moreover, by $p_{\widetilde{\mo} \cup \overline{\mo}}(): \widetilde{\mo} \cup \overline{\mo} \rightarrow (0,1]$ we have that each formula in $\mo$ must have a probability 
%
%
%$\sigma = 0$ and the definition of $p_{\widetilde{\mo} \cup \overline{\mo}}(): \widetilde{\mo} \cup \overline{\mo} \rightarrow (0,1]$ imply that it must necessarily be true that $|\mD_{\checkmark}| = 1$ before the algorithm terminates. 

Finally, the proof that the solution to Interactive Dynamic KB Debugging is found for $\sigma = 0$ in case $mode = dynamic$ is analogue to the one for $mode = static$, just 
\begin{enumerate}[(d1)]
	\item $\mD_{\checkmark} \subseteq \minD_{\langle\mo,\mb,\Tp\cup\Tp',\Tn\cup\Tn'\rangle_\RQ}$ holds for the output of \textsc{dynamicHS} in each iteration by Corollary~\ref{cor:dynamic_hs_correctness}. Therefore, $\mD_{\checkmark}$ comprises only minimal diagnoses w.r.t.\ the current DPI.
	\item By (s\ref{proof_int_debug_correct:bullet_2}), (s\ref{proof_int_debug_correct:bullet_3}), Problem Definition~\ref{prob_def:dynamic} and the specification of the \textsc{getSolKB} function, the output solution KB must be the solution to Interactive Dynamic KB Debugging.
\end{enumerate}
That a solution found for $\sigma > 0$ in case $mode = dynamic$ might be an approximate solution to Interactive Dynamic KB Debugging is a direct consequence of the definition of approximate solution given in Remark~\ref{rem:approximate_solution}.

This completes the proof of Proposition~\ref{prop:correctness_of_interactive_KB_debugging_algo}. 
%\fixme{selves the problem given by prob. def. .....}
\end{proof}
Next, we examine the complexity of Algorithm~\ref{algo:inter_onto_debug}.\footnote{Considerations in the rest of this section rely on the assumption that $\textsc{P} \neq \textsc{NP}$ (cf.\ \url{http://bit.ly/1lIuNcP)}.} To this end, we denote in the following by \emph{expensive operation} a call of a (usually) expensive function such as one that internally consults a logical reasoner or another operation such as addition or multiplication that is the most time consuming algorithmic action within a certain part of an algorithm. We analyze Algorithm~\ref{algo:inter_onto_debug} in terms of the number $num$ of expensive operations that are required during its execution %of Algorithm~\ref{algo:inter_onto_debug} 
in the worst case. The worst case time required by Algorithm~\ref{algo:inter_onto_debug} is then the multiplication of the maximal worst case time consumption of any expensive operation throughout the algorithm by $num$.

The next propositions assume $|\mo|$ as an upper bound of $|\Tp'|+|\Tn'|$. This is plausible in the light of evaluations performed in e.g.\ \cite{Shchekotykhin2012,Rodler2013} which substantiate that usually the size of the faulty KB exceeds the number of queries that are necessary to solve the interactive debugging problem by several orders of magnitude.

We first investigate the complexity of the function \textsc{getProbDist} which is called once in each iteration of Algorithm~\ref{algo:inter_onto_debug}:
\begin{proposition}\label{prop:getProbDist_complexity}
Let $|\mo|$ be an upper bound of $|\Tp'|+|\Tn'|$. Then, the function \textsc{getProbDist} in Algorithm~\ref{algo:inter_onto_debug} requires a number of expensive operations that is linear in $|\mo|$.
\end{proposition}
\begin{proof}
The time complexity of \textsc{getProbDist} can be assessed by adding the complexities of (i)~\textsc{getPrioDiagProbs}, (ii)~the for-loop between line~\ref{algoline:get_prob_dist:for-loop_start} and \ref{algoline:get_prob_dist:for-loop_end}, (iii)~the summation in line~\ref{algoline:get_prob_dist:summation} and (iv)~the for-loop in lines~\ref{algoline:get_prob_dist:for-loop2_start} and \ref{algoline:get_prob_dist:normalize}. Time complexity of (i) is in $O(n_{\max}\,|\mo|)$ since $|\mD_{\checkmark}| \leq n_{\max}$ where $n_{\max}$ is a predefined constant and $|\mo|-1$ multiplications must be conducted per diagnosis in $\mD_{\checkmark}$. (ii)~requires $|QA|\,|\mD_{\checkmark}| \leq (|\Tp'|+|\Tn'|)\,n_{\max} \leq |\mo|\,n_{\max}$ many calls to functions \textsc{getEntailments} and \textsc{isKBValid}, respectively, that internally call a logic reasoner. 
%At this, we can regard $|\mo|$ as an upper bound of $|\Tp'|+|\Tn'|$, since usually the size of the faulty KB exceeds the number of queries that are necessary to solve the debugging problem by several orders of magnitude (cf.\ \cite{Shchekotykhin2012,Rodler2013}). 
Time requirements of (iii) amount to $O(|\mD_{\checkmark}|) = O(n_{\max})$ summations. Finally, (iv) involves $O(n_{\max})$ multiplications.

Thus, we obtain an overall time complexity of $O(n_{\max}\,|\mo| + n_{\max}\,|\mo| + n_{\max} + n_{\max}) = O(|\mo|)$ for \textsc{getProbDist}.
\end{proof}
The next proposition is based on this result and witnesses that Algorithm~\ref{algo:inter_onto_debug} requires only a quadratic number of expensive operations in the size of the KB $\mo$.
\begin{proposition} \label{prop:inter_debug_algo:complexity_without_hs}
Let $|\mo|$ be an upper bound of $|\Tp'|+|\Tn'|$ and let the function $qsm()$ given as input to Algorithm~\ref{algo:inter_onto_debug} be such that the time complexity of \textsc{updateQData} is in $O(|\mo|)$. Minus the time consumed by diagnosis computation (by \textsc{staticHS} in case of $mode = static$ or by \textsc{dynamicHS} otherwise), the time complexity in terms of number of required expensive operations of Algorithm~\ref{algo:inter_onto_debug} is quadratic in $|\mo|$.
\end{proposition}
\begin{proof}
Variable instatiation (lines~\ref{algoline:inter_onto_debug:var_inst_start}-\ref{algoline:inter_onto_debug:var_inst_end}) and variable update (lines~\ref{algoline:inter_onto_debug:param_update_start}-\ref{algoline:inter_onto_debug:param_update_end}) is in $O(1)$ where some query selection measure $qsm()$ is supposed to be used, for which the time complexity of \textsc{updateQData} is in $O(|\mo|)$ (this holds for all query selection measures described in Section~\ref{sec:query_selection_measures}). \textsc{getFormulaProbs} called in line~\ref{algoline:inter_onto_debug:getAxiomProbs} runs in $O(|\mo|\,|\tax_{\max}|)$ as Formula~\ref{eq:ax_prob_calc} is applied once to each formula in $\mo$ for each of which at most $|\tax_{\max}|$ multiplications are performed where $|\tax_{\max}|$ 
%:= \widetilde{\tax} \cup \overline{\tax}$ 
is the maximum size of a formula in $\mo$ in terms of included syntactical elements (multiple occurrences of one and the same symbol are counted multiply). 
%By the size of a formula we mean the sum of all syntactical and terminological elements in $\tax$, i.e.\ $|\widetilde{\tax}|+|\overline{\tax}|$. 
As shown by Proposition~\ref{prop:getProbDist_complexity}, the complexity of \textsc{getProbDist} called in line~\ref{algoline:inter_onto_debug:getProbDist} is in $O(|\mo|)$. Execution of \textsc{getMode} needs one iteration over all diagnoses in $\mD_{\checkmark}$ in order to determine the one with maximum probability, i.e.\ it runs in $O(|n_{\max}|)=O(1)$ time since $n_{\max}$ is a constant. Next, \textsc{getSolKB} which computes a solution KB from a given diagnosis $\md$ works in 
%$O(|\md|+1+1+(|\Tp|+|\Tp'|-1)) 
$O(|\md|+|\Tp|+|\Tp'|) \subseteq O(|\mo|)$ since $|\md|$ elements need to be deleted from a set of cardinality $\mo$ which can be accomplished in constant time per element (e.g., using a hashtable) and additionally at most $|\Tp|+|\Tp'|$ set union operations are required, namely the union of $(\mo\setminus\md)$ with $U_{\Tp\cup\Tp'}$ where the latter needs $|\Tp|+|\Tp'|-1$ set union operations.
As $|\Tp|$ is a constant $c$, $O(|\md|+|\Tp|+|\Tp'|) \subseteq O(2c\,|\mo|) \subseteq O(|\mo|)$.
% Note that, at this, an implementation of Algorithm~\ref{algo:inter_onto_debug} is assumed where $U_{\Tp\cup\Tp'}$ is computed once, stored in some variable and updated in each iteration by computing the set union of $U_{\Tp\cup\Tp'}$ and $\tp$ whenever a new test case $\tp$ has been added to $\Tp'$ (this process is not explicitly shown by Algorithm~\ref{algo:inter_onto_debug}).
%an upper bound for the number of positive test cases is usually $|\mo|$ since a revision of the entire KB takes $|\mo|$ steps. 
In Section~\ref{sec:query_gen_complexity}, we have already underlined that \textsc{getPoolOfQueries} is a fixed parameter tractable problem, i.e.\ it requires 
\begin{align*}
O\left(\left(n_{\max}+\left|Q_{\min}^{(\max)}\right|\log \frac{\left|Q^{(\max)}\right|}{\left|Q_{\min}^{(\max)}\right|}\right)2^{n_{\max}}\right) = O(1)
\end{align*}
%$O((n_{\max}+|Q_{\min}^{(\max)}|\log \frac{|Q^{(\max)}|}{|Q_{\min}^{(\max)}|})2^{n_{\max}}) = O(1)$ 
calls to a reasoner in the worst case (cf.\ Proposition~\ref{prop:query_gen_complexity}). Similarly, \textsc{selectQuery} involves $O(2^{n_{\max}})$ comparisons $qsm(Q_i) < qsm(Q_j)$ for $Q_i,Q_j\in\QP$ since the cardinality of the computed query pool is in $O(2^{n_{\max}})$. The latter holds due to Proposition~\ref{prop:getPoolOfQueries_correctness} which substantiates that the calculated query pool includes at most one query $Q$ for which $\dx{}(Q) = Y$ for each $Y \subset \mD_{\checkmark}$. And, an upper bound for the cardinality of $\mD_{\checkmark}$ is the constant $n_{\max}$. Therefore, the runtime of \textsc{selectQuery} is in $O(1)$, too.

Since adding up a number of time complexities each of which is at most in $O(|\mo|)$, we can conclude that the runtime of one iteration of Algorithm~\ref{algo:inter_onto_debug} minus the time needed for diagnosis computation is also in $O(|\mo|)$, i.e.\ linear in $|\mo|$ in terms of number of expensive operations needed. As there might be a maximum of $|\mo|$ iterations by the premise that $|\Tp'|+|\Tn'| \leq |\mo|$, we obtain an overall time complexity -- minus the complexity of diagnoses computation -- of $O(|\mo|^2)$ for Algorithm~\ref{algo:inter_onto_debug}.
\end{proof}
That is, Algorithm~\ref{algo:inter_onto_debug} requires only a quadratic number of expensive operations ``outside'' of the methods \textsc{staticHS} or \textsc{dynamicHS}, respectively, that account for diagnosis computation. That the substantial complexity of Algorithm~\ref{algo:inter_onto_debug} lies in the computation of diagnoses, is confirmed by the following results. 

The first result is based on the fact that determining minimal diagnoses w.r.t.\ a DPI is an MBD problem (cf.\ page~\pageref{etc:MBD_problem_is_abduction_problem}) which in turn can be regarded as an abduction problem as defined in \cite{Bylander1991}. More precisely, the problem of detecting minimal diagnoses w.r.t.\ a DPI is a \emph{monotonic} abduction problem \cite{Bylander1991}. Hence, the following proposition holds \cite[Theorem 4.3]{Bylander1991}:% for monotonic abduction problems
\begin{proposition}\label{prop:finding_next_min_diag_is_NP-complete}
Let $\langle\mo,\mb,\Tp,\Tn\rangle_\RQ$ be a DPI over $\mathcal{L}$ and let \textsc{isKBValid} (see Algorithm~\ref{algo:qx}) be a function computable for $\mathcal{L}$ in polynomial time w.r.t.\ the size of $\langle\mo,\mb,\Tp,\Tn\rangle_\RQ$ (cf.\ the description of the function $e$ in \cite[Section 3.3]{Bylander1991}). Then, given a set $\mD$ of minimal diagnoses w.r.t.\ $\langle\mo,\mb,\Tp,\Tn\rangle_\RQ$ such that $\emptyset \subset \mD$, it is $\textsc{NP}$-complete to determine whether there is a minimal diagnosis $\md$ w.r.t.\ $\langle\mo,\mb,\Tp,\Tn\rangle_\RQ$ such that $\md \notin \mD$.
\end{proposition}
\begin{remark}\label{rem:isKBValid_is_analogue_to_function_e_in_Bylander1991}
The function \textsc{isKBValid} in the case of KB debugging is analogue to the function $e$ used in \cite{Bylander1991}. Given the overall data $D_{all}$ that must be explained by a solution to an abduction problem, the function $e$ computes for a subset $H$ of $H_{all}$, the set of all individual hypotheses, the set $e(H) = D$ where $D \subseteq D_{all}$ is the data explained by $H$. $H$ is an explanation of the abduction problem iff it is set-minimal and $e(H) = D_{all}$ \cite{Bylander1991}. 

In the case of our KB debugging system, given a DPI $\langle\mo,\mb,\Tp,\Tn\rangle_\RQ$, $D_{all}$ corresponds to
the set of all requirements in $\RQ$ and all test cases in $\Tn$ violated by $\mo \cup \mb \cup U_{\Tp}$. $H_{all}$ corresponds to $\mo$. So, $e$ corresponds to \textsc{isKBValid} since \textsc{isKBValid} is given some $\mo \setminus \md$ and $\langle\cdot,\mb,\Tp,\Tn\rangle_\RQ$ (where $\md$ corresponds to some $H \subseteq H_{all}$) and checks whether $(\mo \setminus \md) \cup \mb \cup U_{\Tp}$ does not violate any requirement or test case, i.e.\ whether $e(H) = D_{all}$. Notice that \textsc{isKBValid} can easily be slightly modified to return the subset of $D_{all}$ that is explained by $H$, i.e.\ the subset of the initially violated requirements and test cases that are resolved by deletion of $\md$ from $\mo \cup \mb \cup U_{\Tp}$. To this end, the early termination in case of detected invalidity must simply be omitted.\qed  
\end{remark}
\begin{remark}\label{rem:parsimonious_KB_debugging_is_monotonic_abduction_problem}
An abduction problem is monotonic \cite{Bylander1991} iff for all $H,H' \subseteq H_{all}$ it holds that $H \subseteq H' \rightarrow e(H) \subseteq e(H')$. That parsimonious KB debugging (or the problems given by Problem Definitions~\ref{prob_def:evidence_just}, \ref{prob_def:static}, \ref{prob_def:dynamic}, \ref{prob_def:minimize_user_interact_static} and \ref{prob_def:minimize_user_interact_dynamic}) seen as an abduction problem is indeed monotonic is a simple consequence of the monotonicity of the logic $\mathcal{L}$ over which a DPI must be defined (as per the postulations of Chapter~\ref{chap:basics}). For, if $(\mo\setminus\md') \cup \mb \cup U_{\Tp} \models x$, then also $(\mo\setminus\md) \cup \mb \cup U_{\Tp} \models x$ for $\md \subseteq \md'$. Modeling requirements $r \in \RQ$ as unwanted entailments of the correct KB (see Remark~\ref{rem:reduce_conditions_of_dpi_def}), we immediately see that $\md$ cannot resolve more unwanted entailments $x \in \RQ \cup \Tn$ than $\md'$. Thence, parsimonious KB debugging is a monotonic abduction problem.\qed
\end{remark}
Unfortunately, \textsc{isKBValid} is not tractable (i.e.\ computable in polynomial time) for many logics $\mathcal{L}$. In particular, it is already in $\Delta_2^{\textsc{P}} = \textsc{P}^\textsc{NP}$ for PL (cf.\ the polynomial hierarchy defined by \cite{meyer1972}). This holds since propositional satisfiability checking is $\textsc{NP}$-complete \cite{cook1971, karp1972} and since \textsc{isKBValid}, in order to to check the validity (see Definition~\ref{def:valid_onto}) of a set of PL formulas $X$ w.r.t.\ some PL DPI $\langle\cdot,\mb,\Tp\cup\Tp',\Tn\cup\Tn'\rangle_\RQ$, requires a polynomial number of calls to a propositional satisfiability checker $Alg_{\mathsf{SAT}}$. 
%in the worst case. 
For, by the definition of \textsc{isKBValid} (see Algorithm~\ref{algo:qx}), one call of $Alg_{\mathsf{SAT}}$ is required for testing whether $X \cup \mb \cup U_{\Tp\cup\Tp'}$ is consistent and a maximum of $|\Tn|+|\Tn'|$ further calls are needed to verify whether $X \cup \mb \cup U_{\Tp\cup\Tp'} \cup \setof{\lnot \tn}$ is consistent for all $\tn \in \Tn\cup\Tn'$, i.e.\ whether $X \cup \mb \cup U_{\Tp\cup\Tp'} \not\models \tn$ for all $\tn \in \Tn\cup\Tn'$ (note that $\lnot \tn$ refers to the formula $\lnot \tax_1 \lor \dots \lor \lnot\tax_k$ if $\tn := \setof{\tax_1,\dots,\tax_k}$, cf.\ page~\pageref{etc:test_cases_are_sets_or_conjuntions_of_formulas}). Since we assume $|\Tp'|+|\Tn'| \leq |\mo|$ and since $|\Tn|$ is a constant throughout the execution of Algorithm~\ref{algo:inter_onto_debug}, we have that the number $|\Tn|+|\Tn'|+1 \leq |\Tn|+|\mo|+1$ of calls to $Alg_{\mathsf{SAT}}$ performed by \textsc{isKBValid} is bounded by a polynomial in $|\mo|$. 

As a conclusion of this discussion and Proposition~\ref{prop:finding_next_min_diag_is_NP-complete}, we have: 
%\fixme{show why each call of static and dynamic must compute a further min diag}
\begin{corollary}
Let $\langle\mo,\mb,\Tp,\Tn\rangle_\RQ$ be a PL DPI given as an input to Algorithm~\ref{algo:inter_onto_debug}. Then, each call of \textsc{staticHS} or \textsc{dynamicHS} within Algorithm~\ref{algo:inter_onto_debug} must solve (at least) an $\textsc{NP}$-complete problem by means of an oracle that requires a polynomial number of calls to another $\textsc{NP}$-complete oracle. 
%That is, each call of \textsc{staticHS} or \textsc{dynamicHS} within Algorithm~\ref{algo:inter_onto_debug} solves a problem in $\Sigma_3^P$ (cf.\ the polynomial hierarchy defined by \cite{meyer1972}).
%, i.e.\ a problem that is contained in the third level of the polynomial hierarchy~\cite{meyer1972}.
\end{corollary} 
\begin{proof}
Both \textsc{staticHS} and \textsc{dynamicHS} must return a set of at least $n_{\min} \geq 2$ minimal diagnoses each time they are called (given that $n_{\min}$ minimal diagnoses exist w.r.t.\ the given DPI) due to the specification of input parameter $n_{\min}$ in Algorithm~\ref{algo:inter_onto_debug} and the calls of \textsc{staticHS} and \textsc{dynamicHS} in lines~\ref{algoline:inter_onto_debug:staticHS} and \ref{algoline:inter_onto_debug:dynamicHS}, respectively. For the first call, this implies that at least two minimal diagnoses must be found. Hence, Proposition~\ref{prop:finding_next_min_diag_is_NP-complete} applies to the complexity of finding the second minimal diagnosis during the execution of the first call of both \textsc{staticHS} and \textsc{dynamicHS}, just that \textsc{isKBValid} does not terminate in polynomial time, but uses a polynomial number of calls to an $\textsc{NP}$-complete oracle (the propositional satisfiability checker). 

In each subsequent call of any of the two methods \textsc{staticHS} and \textsc{dynamicHS}, the existing set of leading diagnoses will contain at least one minimal diagnosis w.r.t.\ the current DPI (since each query leaves valid at least one leading diagnosis, cf.\ Definition~\ref{def:query}), and at least one further minimal diagnosis w.r.t.\ this DPI must be extracted (cf.\ bullet (\ref{etc:staticHS_must_compute_at_least_one_further_min_diag_per_call}) in the characterization of the outputs of \textsc{staticHS} and \textsc{dynamicHS} on page~\pageref{etc:staticHS_output_bullet_a} ff.). Thus, Proposition~\ref{prop:finding_next_min_diag_is_NP-complete} holds for the computation of the first diagnosis in any subsequent call of any of the two functions, just that \textsc{isKBValid} does not terminate in polynomial time, but uses a polynomial number of calls to an $\textsc{NP}$-complete oracle (the propositional satisfiability checker).  
\end{proof}
The general complexity of \textsc{isKBValid} is even worse if DPIs over more expressive logics such as OWL~2 are considered for which one single call of a reasoner invoked by \textsc{isKBValid} is already 2-$\textsc{NExpTime}$-complete~\cite{Grau2008a,Kazakov2008}. 

However, in spite of these discouraging \emph{theoretical} complexity results, debugging techniques similar to the ones discussed in this work have proven to perform reasonably \emph{in practice} for many real-world KB debugging problems over DL and OWL languages, respectively~\cite{Shchekotykhin2012, Rodler2013, Shchekotykhin2014} which are more expressive than PL. For instance, we have shown in \cite{Shchekotykhin2012} that faulty real-world OWL KBs with sizes of up to over 33000 formulas are efficiently interactively debuggable with similar methods as those presented in this work (reaction time of the system, i.e.\ time between two successive queries: only 1 minute; average query length: not more than 4 formulas; overall number of queries: at most 14). Moreover, we have demonstrated in \cite{Rodler2013} that a pair of real-world OWL KBs (the first including over 11000 formulas, the second almost 5000) that has been automatically integrated by diverse ontology matching systems resulting in a faulty aligned KB (see Chapter~\ref{chap:RelatedWork} for details; we also list some matching systems there) can be debugged with absolutely reasonable time and query answering effort for the interacting user. In concrete terms, the RIO debugging strategy proposed in \cite{Rodler2013} (which can also be plugged in as a query selection measure into the system described in this work, see Section~\ref{sec:query_selection_measures}) involved an average reaction time of no more than 13 seconds and required an average number of queries to be answered by the user of no more than nine.

\chapter{Summary}
\label{sec:summary_part_interactive_KB_debugging} 
In this part we dealt with how the process of KB debugging can be designed so as to enable a (group of) user(s) to interact with the debugging software in order to achieve high quality solutions. We defined the problem of \emph{interactive static KB debugging} as well as the problem of \emph{interactive dynamic KB debugging} which ``naturally'' arise from the fact that the DPI in interactive KB debugging is always renewed after a new test case has been specified (a new query has been answered). The former problem searches for a solution KB \emph{w.r.t.\ the original DPI given as input} such that this solution KB satisfies all test cases added during the debugging session and there is no other such solution KB. The latter problem searches for a solution KB \emph{w.r.t.\ the current DPI} (i.e.\ the original DPI including all new test cases added throughout the debugging session so far) such that there is no other solution KB w.r.t.\ the current DPI.

We specified the pivotal notion of a \emph{query} which constitutes the ``interface'' between the debugging system and the interacting user. Queries are sets of logical formulas satisfying the \emph{search space restriction} as well as the \emph{solution preservation} property. That is, incorporation of any answer to a particular query into the debugging process leads to a reduction of the search space for solutions on the one hand, but guarantees the existence of at least one remaining solution on the other hand. Queries are generated from a set of \emph{leading diagnoses} that act as a representative of all (minimal) diagnoses. We established that, for any set of at least two leading diagnoses, a query exists. The unique \emph{q-partition} of a query constitutes the relationship between a query and the set of leading diagnoses and can be used to decide for a set of logical formulas whether this set is or is not a query. Furthermore, the q-partition can be used to estimate the impact of a query answer on the (distribution of the) set of solutions and thence can be exploited to assess the (expected) quality of different queries which in turn can help to filter out a suitable query among a pool of possible queries. 

It was also presented how a pool of queries can be generated for a given set of leading diagnoses and a DPI. We showed how to minimize these queries in terms of the included number of logical formulas the aim of which is to strain the user(s) as little as possible when it comes to answering them. Moreover, we pointed out that query generation is a fixed parameter tractable problem due to the fact that the (maximum) number of leading diagnoses can be predefined and therefore constitutes a constant value (which is not growing as the diagnosis problem instance grows). We featured an in-depth discussion of the properties of the query generation algorithm, in the course of which we detected several drawbacks. The gave a hint to potential solutions that we will address in our future work. Additionally, we formally proved the correctness of the query generation method and derived complexity results. All of this was concretized by means of several illustrating examples.   

Finally, we explicated the central algorithm of this work which implements an interactive KB debugging system. First, an overview of the workflow of interactive KB debugging was given, followed by a more comprehensive detailed specification of the algorithm. Some query selection measures (all of which are later covered in more depth in Parts~\ref{part:JWS} and \ref{part:RIO}) were discussed
and optimization versions of the problems of interactive dynamic and static KB debugging were defined where the goal is to obtain the solution to these problems by asking the user a minimal number of queries. Finally, we formally proved the correctness of the interactive KB debugging algorithm and gave a discussion of its complexity.

\part{Iterative Diagnosis Computation}
\label{part:IterativeDiagnosisComputation}
In this part we introduce and discuss two methods, \textsc{staticHS} and \textsc{dynamicHS}, which are called in lines~\ref{algoline:inter_onto_debug:staticHS} and \ref{algoline:inter_onto_debug:dynamicHS} of Algorithm~\ref{algo:inter_onto_debug}, respectively. The former provides a method for solving the Interactive Static KB Debugging Problem (Problem Definition~\ref{prob_def:static}) whereas the latter aims at solving the Interactive Dynamic KB Debugging Problem (Problem Definition~\ref{prob_def:dynamic}). Both are methods for iterative diagnosis computation that are employed to compute a set of leading diagnoses in each iteration of the presented interactive KB debugging algorithm (Algorithm~\ref{algo:inter_onto_debug}). Each time a query has been answered by the interacting user and added to the respective set of test cases of the DPI, a subset of the leading diagnoses (and usually also a set of not-yet-computed minimal diagnoses) is invalidated. An iterative diagnosis computation method is then invoked to update the leading diagnoses set taking the new information into account that is given by the recently added test case.
%fill up the reduced leading diagnoses set by the 
That is, the $k \leq n_{\max}$ most probable ways of solving the Interactive Static (Dynamic) KB Debugging Problem in the light of the new evidence are extracted by \textsc{staticHS} (\textsc{dynamicHS}) after the search space has been suitably pruned. In this vein, if there is only one solution left, the (exact) solution of Interactive Static (Dynamic) KB Debugging has been found. 

Chapter~\ref{chap:StaticHSTree} provides an in-depth description of the $static$ method and proves its correctness.
%Used as a routine for leading diagnosis computation in Algorithm~\ref{algo:inter_onto_debug}, the $static$ method solves the problem of interactive static KB debugging.
%
Chapter~\ref{chap:DynamicHSTree} details the $dynamic$ method and demonstrates its correctness.
The practically oriented reader or the one that is willing to believe that the presented iterative diagnosis computation techniques in fact work as claimed might skip Sections~\ref{sec:CorrectnessOfTextscStaticHS} as well as \ref{sec:TextscDynamicHSDetailsAndCorrectness} in this part.\footnote{Parts of Part~\ref{part:IterativeDiagnosisComputation} already appeared in \cite{Rodler2015}. However, \cite{Rodler2015} includes a significantly less detailed presentation of the algorithms and does not give any proofs of correctness.}
%minimal diagnoses w.r.t.\ the updated DPI are extracted after the search space has been suitably pruned. 
%
%and a subset of the leading diagnoses has been invalidated, a set of new minimal diagnoses is calculated by such an iterative diagnosis computation algorithm and added to those leading diagnoses that are still valid. In this manner, a basis in terms of a ``representative'' diagnosis set for the construction of a further query is established. %It is important to notice that the main source of complexity within Algorithm~\ref{algo:inter_onto_debug} is located exactly in this diagnosis computation process. Whereas the rest of Algorithm~\ref{algo:inter_onto_debug} is linear w.r.t.\ the cardinality of the given KB by Proposition~\ref{prop:inter_debug_algo:complexity_without_hs}, we shall recognize in this section that the worst case complexity of both introduced methods \textsc{staticHS} and \textsc{dynamicHS} is not polynomial.
%exponential (where the base is the cardinality of the maximum cardinality minimal conflict set and the exponent is the number of minimal conflict sets).

\chapter[\textsc{staticHS}: A Static Iterative Diagnosis Computation Algorithm]{\textsc{staticHS}: A Static Iterative Diagnosis Computation Algorithm%
\chaptermark{Static Diagnosis Computation Algorithm}}
\chaptermark{Static Diagnosis Computation Algorithm}
\label{chap:StaticHSTree}
%%%%
%\chapter{\textsc{staticHS}: A Static Iterative Diagnosis Computation Algorithm}
%\chaptermark{Static HS}
%\label{chap:StaticHSTree}
%%%%
As the name already suggests, \textsc{staticHS} (Algorithm~\ref{algo:inter_stat_hs}) is a procedure that solves the problem of \emph{Interactive Static KB Debugging} defined by Problem Definition~\ref{prob_def:static} if used for leading diagnosis computation in Algorithm~\ref{algo:inter_onto_debug}. \textsc{staticHS} is sound, complete and optimal w.r.t.\ the set of solutions of the \emph{Interactive Static KB Debugging} problem (this will be proven in Section~\ref{sec:CorrectnessOfTextscStaticHS}). Optimality refers to the best-first computation of minimal diagnoses regarding a given probability measure.
  
\section{Overview and Intuition}
\label{sec:TheIntuition}
The \textsc{staticHS} algorithm is strongly related to the non-interactive hitting set algorithm \textsc{HS} (see Algorithm~\ref{algo:hs}) in that, at any stage during the execution of Algorithm~\ref{algo:inter_onto_debug}, the hitting set tree produced by \textsc{staticHS} corresponds to some part of the complete (non-interactive) wpHS-tree built-up by Algorithm~\ref{algo:hs}. This is achieved by the strategy to \emph{use new test cases only for the invalidation of diagnoses, and not for the computation of conflict sets (and thus diagnoses)}. That is, all minimal conflict sets are computed w.r.t.\ the input DPI. Thereby, the introduction of new diagnoses, i.e.\ ones that are not minimal diagnoses w.r.t.\ the input DPI, through addition of new test cases to the DPI is prohibited (cf.\ Proposition~\ref{prop:mindiag_mincs}).

So, what \textsc{staticHS} as a subroutine of Algorithm~\ref{algo:inter_onto_debug} does is gradually building up the standard (non-interactive) wpHS-tree in multiple phases. During each phase some new (not-yet-computed) minimal diagnoses w.r.t.\ the \emph{input DPI} are computed, in the order of their probability, most probable ones first. Before such a newly detected minimal diagnosis is added to the set of leading diagnoses ($\mD_{calc}\cup\mD_{\checkmark}$), a test is performed that verifies that this new diagnosis is consistent with all test cases added to the input DPI so far. In this vein, all answered queries so far not only serve to eliminate a subset of the set of leading diagnoses at the time when the respective query is answered, but also to eliminate incompatible minimal diagnoses w.r.t.\ the input DPI that are found at some later point in time. However, in order to be eliminated due to a specified test case, a minimal diagnosis must first be computed. That is, no partial diagnoses can be eliminated due to newly specified test cases.

Between each two phases of tree construction, a query computed on the basis of the current set of leading diagnoses is asked to the user (this is accomplished directly in Algorithm~\ref{algo:inter_onto_debug}). 
%Each query is computed based on a new set of leading, i.e.\ most probable, minimal diagnoses of the input DPI. 
After incorporating the user's answer, some leading diagnoses are eliminated (this is granted by the definition of a query, see Definition~\ref{def:query}). Moreover, the ``state'' of the tree is maintained during the execution of Algorithm~\ref{algo:inter_onto_debug} until \textsc{staticHS} is again called in order to calculate further leading diagnoses. The state of the current partial wpHS-tree is stored by variables 
\begin{itemize}
	\item $\mD_{calc} \cup \mD_{\checkmark}$ -- computed minimal diagnoses w.r.t.\ the input DPI consistent with all test cases specified so far,
	\item $\Queue$ -- the list of open, non-labeled nodes,
	\item $\mC_{calc}$ -- minimal conflict sets w.r.t.\ the input DPI computed so far and
	\item $\mD_{\times}$ -- computed minimal diagnoses w.r.t.\ the input DPI not consistent with all test cases specified so far.
\end{itemize} 

Each time a tree construction phase, i.e.\ the computation of new leading diagnoses, is finished, a new diagnosis probability distribution is obtained by the diagnosis probability update as per Bayes' Theorem described in Section~\ref{sec:DetailedAlgorithmDescription}. Once this distribution involves one highly probable diagnosis (the probability of which exceeds a predefined threshold $1-\sigma$) and else just highly improbable ones, the algorithm terminates. The output is a solution KB w.r.t.\ the \emph{input DPI} built from this highly probable minimal diagnosis. 
%Notice that the output is also a solution KB w.r.t.\ the current DPI, i.e.\ the input DPI incorporating all newly specified test cases. This holds as a minimal diagnosis w.r.t.\ the 

\begin{remark}\label{rem:sigma_in_staticHS}
In case $\sigma$ has a predefined value of zero, the output is the (exact) solution to the problem of \emph{Interactive Static KB Debugging} for the input DPI. In a scenario where some fault tolerance $\sigma > 0$ is given, the solution KB returned by Algorithm~\ref{algo:inter_onto_debug} is an approximation of the (exact) solution to \emph{Interactive Static KB Debugging} for the input DPI where a better approximation can be expected for smaller values of $\sigma$ (cf.\ Remark~\ref{rem:approximate_solution}). ``Better'' in this context refers to the satisfaction of desired semantic properties of the KB returned by Algorithm~\ref{algo:inter_onto_debug}, i.e.\ desired entailments and desired non-entailments of the KB. The intuition is that the specification of additional test cases $T$ guarantees the output of a KB complying with these test cases, whereas accepting 
%as solution KB of \emph{Interactive Static KB Debugging} 
one -- albeit highly probable -- of multiple solution KBs without having incorporated $T$ leaves open the possibility for this KB to not fulfill $T$.

However, answering queries is effort for an interacting user. Therefore, the approach that involves the ``early'' termination of the algorithm after a solution KB has a sufficiently high probability (lower than 1) constitutes a trade-off between exactness of the output and the effort of the user and overall execution time of the interactive KB debugging algorithm, respectively.\qed
\end{remark}

%\noindent\textbf{Constant ``Convergence'' towards the Solution.}
\paragraph{Constant ``Convergence'' towards the Solution.}
 As said, each added test case is an answered \emph{query} and thus eliminates at least one minimal diagnosis w.r.t.\ the input DPI. And, only minimal diagnoses w.r.t.\ the input DPI are computed by \textsc{staticHS}. Hence, by the fact that a solution to Interactive Static KB Debugging can only be constructed from a minimal diagnosis w.r.t.\ the input DPI, it is guaranteed that the number of solutions to Interactive Static KB Debugging is strictly monotonically decreasing throughout the execution of Algorithm~\ref{algo:inter_onto_debug}. That is, the initial number of (all) minimal diagnoses (w.r.t.\ the input DPI) is ``static'' which means that no ``new'' minimal diagnoses can be introduced when the input DPI is extended by new test cases. 

As a consequence of this, it is reasonable to employ \textsc{staticHS} in a situation where the (complete) wpHS-tree produced by the standard (non-interactive) algorithm \textsc{HS} is believed to be as compact as to fit into the available system memory. In this case, \textsc{staticHS} is also guaranteed to not exceed the available memory, even if an exact solution ($\sigma = 0$) is intended. 

Unfortunately, however, it will be generally the case that a complete enumeration of all minimal diagnoses is intractable, especially due to an overwhelming space complexity. In such a case, 
Algorithm~\ref{algo:inter_onto_debug} using \textsc{staticHS} will definitely run out of memory (given that \textsc{staticHS} is called sufficiently often). The reason is that the space consumption of \textsc{staticHS} will sooner or later definitely reach the huge extent of the wpHS-tree produced by \textsc{HS}. Nevertheless, \textsc{staticHS} might be used to (possibly) find some (approximate) solution. This might work in a scenario where the given probabilistic information in terms of $p_{\widetilde{\mo}\cup\overline{\mo}}()$ provided as an input to Algorithm~\ref{algo:inter_onto_debug} is ``reasonable'' in that the desired diagnosis is assigned a rather high probability and is thus figured out early, before the available memory is exhausted. 

A possible modification of the stop criterion in \textsc{staticHS} in a way that new leading diagnoses are not computed until a desired number of such is detected or a timeout is reached, but rather until a predefined maximum space is consumed, would not mitigate space complexity issues very much. An explanation for this is that stopping \textsc{staticHS} on account of no more available memory implies that no further call of \textsc{staticHS} will be able to execute. That is because, as mentioned before, an added test case can only invalidate already computed diagnoses, no other branches in the wpHS-tree, and each invalidated minimal diagnosis cannot be discarded, but must be stored (in $\mD_{\times}$) to avoid the usage of leading diagnoses that are non-minimal w.r.t.\ the input DPI (cf.\ lines~\ref{algoline:slabel:non_min_crit_start}-\ref{algoline:slabel:non_min_crit_end} in Algorithm~\ref{algo:inter_stat_hs}). 
%
%However, it will be definitely the case that \textsc{staticHS} runs out of memory.  The reason is that the space consumption of \textsc{staticHS} will sooner or later definitely reach the huge extent of the wpHS-tree produced by \textsc{HS}. 
%is strictly monotonically increasing throughout the execution of Algorithm~\ref{algo:inter_onto_debug}. 

%\noindent\textbf{Poor Tree Pruning.} 
\paragraph{Poor Search Tree Pruning.}
As we explained before, the preservation of a constantly shrinking set of minimal diagnoses comes at the cost of being able to exploit new test cases only partially, i.e.\ only for the invalidation of already computed minimal diagnoses w.r.t.\ the input DPI and not for the computation of minimal conflict sets and thus minimal diagnoses. % w.r.t.\ the current DPI. 
%The preservation of a fixed set of minimal diagnoses comes at the cost of another restriction of \textsc{staticHS}, namely the fact that only (minimal or non-minimal) \emph{diagnoses} w.r.t.\ the input DPI can be pruned. 
%
%That is, no pre-pruning of paths leading from the root to a node that is not yet a diagnosis w.r.t.\ the input DPI is possible. The cause of this is the ``principle of statics'' pursued by \textsc{staticHS} which means that only minimal diagnoses w.r.t.\ the input DPI must be computed and that the set of minimal diagnoses must not grow after the formulation of a new test case. To this end, as mentioned, new test cases are only exploited partially, i.e.\ to \emph{invalidate} diagnoses and not to compute minimal conflict sets (and thus minimal diagnoses). 
%
The incorporation of test cases into the DPI that is used to determine minimal conflict sets (line~\ref{algoline:slabel:qx} in Algorithm~\ref{algo:inter_stat_hs}) could, on the one hand, lead to new minimal conflict sets that are no minimal conflict sets w.r.t.\ the input DPI. As a consequence of this, minimal diagnoses might be determined by the algorithm which are no minimal diagnoses w.r.t.\ the input DPI, but w.r.t.\ the current DPI. Hence, the soundness of \textsc{staticHS} w.r.t.\ the set of solutions of the Interactive Static KB Debugging problem would be violated. Furthermore, such conflict sets 
%implicates a varying set of solutions in general and 
could lead to the missing of some minimal diagnoses w.r.t.\ the input DPI, a violation of the completeness of \textsc{staticHS} w.r.t.\ the set of solutions of the Interactive Static KB Debugging problem. 

On the other hand, the exploitation of new test cases for conflict set generation might give rise to the possibility of pre-pruning of \emph{any} tree branches, not just branches that already correspond to diagnoses w.r.t.\ the input DPI. Such a ``dynamic'' strategy which exploits the new information given by a test case not just partially, but for the invalidation \emph{and computation} of diagnoses and conflict sets, will be implemented be \textsc{dynamicHS} which we will detail in Chapter~\ref{chap:DynamicHSTree}.

Put another way, in \textsc{staticHS} only the standard pruning rules for the construction of a wpHS-tree are applicable, namely the deletion of duplicate nodes and the elimination of non-minimal diagnoses (cf.\ Definition~\ref{def:weighted_pruned_hs_tree}). Newly defined test cases only facilitate the deletion of tree branches from the leading diagnoses set $\mD_{calc} \cup \mD_{\checkmark}$, but not from memory (as invalidated minimal diagnoses must be stored in $\mD_{\times}$, as pointed out before).  

To summarize, \textsc{staticHS} on the one hand makes sure to only consider relevant solutions 
of the problem of Interactive Static KB Debugging,
%w.r.t.\ the input DPI 
%(which is the DPI a user usually wants to find a solution KB for), 
but on the other hand suffers from this conservative strategy in that tree pruning cannot be designed very effectively. So, on the positive side, uncontrolled growth of the produced wpHS-tree can be avoided, but, on the negative side, consultation of an interacting user cannot be taken advantage of in terms of reduction of the space complexity of \textsc{staticHS} compared to the construction of a wpHS-tree by a non-interactive procedure like Algorithm~\ref{algo:hs}.

\section{Algorithm Walkthrough}
\label{sec:AlgorithmWalkthrough_static}
%\noindent\textbf{Input Parameters.} 
\paragraph{Input Parameters.}
When \textsc{staticHS} (Algorithm~\ref{algo:inter_stat_hs}) is called for the first time in Algorithm~\ref{algo:inter_onto_debug}, the inputs $\mC_{calc}$, $\mD_{\checkmark}$, $\mD_{\times}$, $\Tp'$ and $\Tn'$ correspond to the empty set and $\Queue = [\emptyset]$ (cf.\ lines~\ref{algoline:inter_onto_debug:var_inst_start}-\ref{algoline:inter_onto_debug:var_inst_end} and \ref{algoline:inter_onto_debug:staticHS} in Algorithm~\ref{algo:inter_onto_debug}). Further on, $\mD_{calc}$ is defined to be the empty set at the beginning of each execution of \textsc{staticHS}. That is, \textsc{staticHS} starts the construction of the wpHS-tree from an initial tree consisting of a single unlabeled root node $\emptyset$ ($\in \Queue$). And, all collections that are later returned by \textsc{staticHS}, except for $\Queue$, are initially empty. Further input arguments are the DPI $\langle\mo,\mb,\Tp,\Tn\rangle_\RQ$ provided as an input to Algorithm~\ref{algo:inter_onto_debug}, the sets of positively ($\Tp'$) and negatively ($\Tn'$) answered queries since the start of Algorithm~\ref{algo:inter_onto_debug}, the leading diagnosis computation parameters $n_{\min},n_{\max},t$ (see the description in Chapter~\ref{chap:UserInteraction} on page~\pageref{etc:leading_diag_params}) and the probability measure $p() := p_{\mo}()$ that assigns a probability in the interval $(0,0.5)$ to each formula in $\mo$ (cf.\ line~\ref{algoline:inter_onto_debug:getAxiomProbs} in Algorithm~\ref{algo:inter_onto_debug}).

%\noindent\textbf{The Main Loop.} 
\paragraph{The Main Loop.}
During the repeat-loop, in each iteration the first node $\mathsf{node}$ in $\Queue$ is processed (\textsc{getFirst}, line~\ref{algoline:static:getfirst}). That is, $\mathsf{node}$ is deleted from $\Queue$ (\textsc{deleteFirst}, line~\ref{algoline:static_update_Q}) and the \textsc{sLabel} function is called given $\mathsf{node}$ (i.a.) as a parameter. Notice that elements are added to $\Queue$ (line~\ref{algoline:static:generate_nodes}) in a way that a sorting of $\Queue$ in descending order according to $p_{nodes}()$ (cf.\ Definition~\ref{def:p_node()}) is maintained throughout the execution of \textsc{staticHS}.
%$\Queue$ is always sorted in descending order according to $p_{nodes}()$ (cf.\ Definition~\ref{def:p_node()}) throughout \textsc{staticHS}.

%\noindent\textbf{Computation of a Node Label.} 
\paragraph{Computation of a Node Label.}
The \textsc{sLabel} function processes $\mathsf{node}$ as follows. First, the \emph{non-minimality criterion} (lines~\ref{algoline:slabel:non_min_crit_start}-\ref{algoline:slabel:non_min_crit_end}) is checked. That is, among all nodes in $\mD_{(\times,\checkmark,calc)} = \md_{\times} \cup \mD_{\checkmark} \cup \mD_{calc}$ one is searched which is a subset of $\mathsf{node}$. If such a node $\mathsf{nd}$ is found, then $\mathsf{node}$ must be a non-minimal diagnosis ($\mathsf{nd} \subset \mathsf{node}$) or a duplicate diagnosis ($\mathsf{nd} = \mathsf{node}$) w.r.t.\ $\langle\mo,\mb,\Tp,\Tn\rangle_\RQ$ since all sets $\md_{\times}$, $\mD_{\checkmark}$ and $\mD_{calc}$ contain only minimal diagnoses w.r.t.\ $\langle\mo,\mb,\Tp,\Tn\rangle_\RQ$. In this case, the branch in the wpHS-tree corresponding to $\mathsf{node}$ can be dismissed which is taken account of by returning the label $closed$ for $\mathsf{node}$. 

In case the non-minimality criterion is not satisfied, the \emph{duplicate criterion} (lines~\ref{algoline:slabel:dup_crit_start}-\ref{algoline:slabel:dup_crit_end}) is checked next. Here, $\Queue$ is browsed for a node that is equal to $\mathsf{node}$. If such a one is found, $\mathsf{node}$ can be discarded because it suffices to consider only one tree branch among multiple tree branches in the wpHS-tree featuring one and the same set of edge labels. Hence, $closed$ is returned as a label for $\mathsf{node}$. Altogether, this means that only the last processed exemplar of a node corresponding to one and the same set of edge labels is labeled, all others are discarded.

If the duplicate criterion is not met, the \emph{reuse criterion} (lines~\ref{algoline:slabel:reuse_crit_start}-\ref{algoline:slabel:reuse_crit_end}) is checked next. That is, $\mC_{calc}$ is browsed for a set $\mc$ ($\mC_{calc}$ comprises only minimal conflict sets w.r.t.\ $\langle\mo,\mb,\Tp,\Tn\rangle_\RQ$) such that $\mc$ and $\mathsf{node}$ are disjoint sets. If such a $\mc$ is detected, then $\mc$ can be used to label $\mathsf{node}$ since the set of edge labels along the path in the wpHS-tree leading from the root node to $\mathsf{node}$ does not hit $\mc$. In this case, the label $\mc$ is returned for $\mathsf{node}$ by \textsc{sLabel}.

Given that the reuse criterion fails, $\scQX$ is called given the DPI $\langle\mo\setminus\mathsf{node},\mb,\Tp,\Tn\rangle_\RQ$ as an argument (line~\ref{algoline:slabel:qx}). If the output $L$ is equal to 'no conflict', then we know by Proposition~\ref{prop:qx_correctness} that $\mathsf{node}$ is a diagnosis w.r.t.\ $\langle\mo,\mb,\Tp,\Tn\rangle_\RQ$, wherefore the label $valid$ is returned for $\mathsf{node}$. Otherwise, the output $L$ must be a minimal conflict set w.r.t.\ $\langle\mo,\mb,\Tp,\Tn\rangle_\RQ$ that has an empty set-intersection with $\mathsf{node}$. Since the reuse criterion failed, i.e.\ there is no set in $\mC_{calc}$ that does not intersect with $\mathsf{node}$, $L$ must be a fresh minimal conflict set w.r.t.\ $\langle\mo,\mb,\Tp,\Tn\rangle_\RQ$ in the sense that $L \notin \mC_{calc}$ must hold. Therefore the label $L$ is first added to $\mC_{calc}$ and then returned by \textsc{sLabel} as a label for $\mathsf{node}$. 

%\noindent\textbf{Processing of a Node Label.} 
\paragraph{Processing of a Node Label.}
Back in the main procedure, $\mC_{calc}$ is updated (line~\ref{algoline:static:update_Ccalc}) and then the label $L$ returned by the \textsc{sLabel} function is processed as follows. If $L = valid$, then it is a fact that $\mathsf{node}$ is a minimal diagnosis w.r.t.\ $\langle\mo,\mb,\Tp,\Tn\rangle_\RQ$, but it is not certain that $\mathsf{node}$ also meets all positive test cases $\Tp'$ and all negative test cases $\Tn'$ that have been specified and added to $\langle\mo,\mb,\Tp,\Tn\rangle_\RQ$ so far. Thus, according to Proposition~\ref{prop:dpi_update}, the validity of the KB $\mo \setminus \mathsf{node}$ w.r.t.\ $\langle\cdot,\mb,\Tp\cup\Tp',\Tn\cup\Tn'\rangle_\RQ$ must still be checked (line~\ref{algoline:static:isOntValid}). If successful, $\mathsf{node}$ is added to the set $\mD_{calc}$ of calculated minimal diagnoses w.r.t.\ the input DPI that comply with all answered queries so far. Otherwise, $\mathsf{node}$ is added to the set $\mD_{\times}$  of minimal diagnoses w.r.t.\ the input DPI that have been invalidated by some answered query. 

Roughly, the minimality of diagnoses added to $\mD_{calc}$ is assured by the pruning rule (lines~\ref{algoline:slabel:non_min_crit_start}-\ref{algoline:slabel:non_min_crit_end}) which eliminates non-minimal nodes and the fact that $p_{nodes}()$ sorts a node $\mathsf{nd}'$ corresponding to a superset of some node $\mathsf{nd}$ behind $\mathsf{nd}$ in $\Queue$. 

If, on the other hand, $L=closed$ is the label returned by $\textsc{sLabel}$, then $\mathsf{node}$ must simply be removed from $\Queue$ which has already been executed in line~\ref{algoline:static_update_Q}. Thence, no actions are necessary (cf.\ line~\ref{algoline:static:if_L_closed}).

In the third case, if a minimal conflict set $L$ is returned by \textsc{sLabel}, then $L$ is a label for $\mathsf{node}$ meaning that $|L|$ successor nodes of $\mathsf{node}$, namely a node $\mathsf{node}\cup\setof{e}$ for all elements $e \in L$, need to be added to $\Queue$ in sorted order using the function $p_{nodes}()$ (\textsc{insertSorted}, line~\ref{algoline:static:generate_nodes}).
%
%This indeed implies the deletion of $\mathsf{node}$ since $\mathsf{node}$ has been removed from $\Queue$ in line~\ref{algoline:static_update_Q} and line~\ref{algoline:st}

%\noindent\textbf{Stop Criterion.} 
\paragraph{Stop Criterion.}
The first criterion causing \textsc{staticHS} to terminate is $\Queue = []$ which means that the complete wpHS-tree has been constructed and no further nodes can be labeled. In this case, $\mD_{calc} \cup \mD_{\checkmark}$ comprises all minimal diagnoses w.r.t.\ $\langle\mo,\mb,\Tp,\Tn\rangle_\RQ$ that are compliant with all the specified positive and negative test cases $\Tp'$ and $\Tn'$.

If the first criterion is not met, then the second criterion is checked. That is, a test is performed which checks whether the number of leading minimal diagnoses w.r.t.\ $\langle\mo,\mb,\Tp,\Tn\rangle_\RQ$ in $\mD_{calc} \cup \mD_{\checkmark}$ amounts to at least $n_{\min}$ and either $|\mD_{calc} \cup \mD_{\checkmark}| = n_{\max}$ or more than $t$ time has passed since the start of the execution of \textsc{staticHS}. In the latter case, $n_{\min} \leq |\mD_{calc} \cup \mD_{\checkmark}| < n_{\max}$ holds. In the former case, $|\mD_{calc} \cup \mD_{\checkmark}| = n_{\max}$ is satisfied.

%\noindent\textbf{Processing of the Leading Diagnoses Returned by \textsc{staticHS}.} 
\paragraph{Processing of the Leading Diagnoses Returned by \textsc{staticHS}.}
When a call of \textsc{staticHS} in Algorithm~\ref{algo:inter_onto_debug} returns $\tuple{\mD_{calc} \cup \mD_{\checkmark},\Queue, \mathbf{C}_{calc}, \mD_{\times}}$, the set $\mD_{calc} \cup \mD_{\checkmark}$ is stored in the variable $\mD_{\checkmark}$ in Algorithm~\ref{algo:inter_onto_debug}. Between two successive calls of \textsc{staticHS} in Algorithm~\ref{algo:inter_onto_debug}, only this set $\mD_{\checkmark}$ as well as $\mD_{\times}$ are modified. The list $\Queue$ and the set $\mC_{calc}$ remain unchanged until they are used as input parameters to the next call of \textsc{staticHS} in Algorithm~\ref{algo:inter_onto_debug}.

In case one diagnosis $\md_{\max}$ of the current leading diagnoses in $\mD_{\checkmark}$ has a probability greater or equal $1 - \sigma$ as per the probability measure $p_{\mD}()$ (see Section~\ref{sec:DetailedAlgorithmDescription}), the stop criterion of interactive KB debugging is met and a solution KB w.r.t.\ $\langle\mo,\mb,\Tp,\Tn\rangle_\RQ$ constructed from the input DPI $\langle\mo,\mb,\Tp,\Tn\rangle_\RQ$ as well as from $\md_{\max}$ is returned to the user. Thereafter, Algorithm~\ref{algo:inter_onto_debug} terminates and no more calls of \textsc{staticHS} take place.

Otherwise, if no leading diagnosis satisfies the stop criterion, a query $Q$ together with its q-partition $\Pt(Q)$ is computed, as was detailed in Chapter~\ref{chap:QueryGeneration} and Section~\ref{sec:DetailedAlgorithmDescription}. An answer $u(Q)$ to this query is submitted by the interacting user (line~\ref{algoline:inter_onto_debug:user_interaction} in Algorithm~\ref{algo:inter_onto_debug}). Then $u(Q)$ along with $\Pt(Q)$ is exploited to figure out the subset $\mD_{out}$ of $\mD_{\checkmark}$ that does not comply with $u(Q)$. This set $\mD_{out}$ is then deleted from $\mD_{\checkmark}$ and added to $\mD_{\times}$. Additionally, $Q$ is added to the positive test cases $\Tp'$ if $u(Q) = \true$ and to the negative test cases $\Tn'$ otherwise. Subsequently, \textsc{staticHS} is called again given 
\begin{itemize}
	\item the updated parameters $\mD_{\checkmark}$, $\mD_{\times}$, $\Tp'$ and $\Tn'$ (which are modified within and outside of \textsc{staticHS} during the execution of Algorithm~\ref{algo:inter_onto_debug}),
	\item the unchanged parameters $\Queue$, $\mC_{calc}$ (which are modified only within \textsc{staticHS} during the execution of Algorithm~\ref{algo:inter_onto_debug}) and
	\item the constant parameters $\langle\mo,\mb,\Tp,\Tn\rangle_\RQ$, $t$, $n_{\min}$, $n_{\max}$ and $p_{\mo}()$ (which are not modified within or outside of \textsc{staticHS} during the execution of Algorithm~\ref{algo:inter_onto_debug}).
%fixed throughout all calls of \textsc{staticHS} made in Algorithm~\ref{algo:inter_onto_debug}.
\end{itemize}
The execution of this next and any subsequent call to \textsc{staticHS} runs in analogue way as described.  

\begin{remark}\label{rem:staticHS_query_computed_from_P_cup_P'_and_N_cup_N'}
We want to emphasize that queries are computed w.r.t.\ the current DPI $\langle\mo,\mb,\Tp\cup\Tp',\Tn\cup\Tn'\rangle_\RQ$ although \textsc{staticHS} focuses on solutions to the problem of Interactive Static KB Debugging which involves exclusively minimal diagnoses w.r.t.\ the input DPI $\langle\mo,\mb,\Tp,\Tn\rangle_\RQ$. However, a minimal diagnosis w.r.t.\ $\langle\mo,\mb,\Tp,\Tn\rangle_\RQ$ that satisfies all positive test cases $\Tp'$ as well as all negative test cases $\Tn'$ is also a minimal diagnosis w.r.t.\ $\langle\mo,\mb,\Tp\cup\Tp',\Tn\cup\Tn'\rangle_\RQ$. And, a minimal diagnosis w.r.t.\ $\langle\mo,\mb,\Tp,\Tn\rangle_\RQ$ that does not satisfy all positive test cases $\Tp'$ as well as all negative test cases $\Tn'$ is not a minimal diagnosis w.r.t.\ $\langle\mo,\mb,\Tp\cup\Tp',\Tn\cup\Tn'\rangle_\RQ$. These two facts are guaranteed by Proposition~\ref{prop:after_adding_testcase_new_min_diag_is_equal_or_superset_of_old_min_diag} that will be given on page~\pageref{prop:after_adding_testcase_new_min_diag_is_equal_or_superset_of_old_min_diag}. 

Hence, it holds that
\begin{itemize}
\item $\md$ is a minimal diagnosis w.r.t.\ $\langle\mo,\mb,\Tp,\Tn\rangle_\RQ$ that satisfies $\Tp'\cup\setof{Q}$ as well as $\Tn'$ if and only if $\md$ is a minimal diagnosis w.r.t.\ $\langle\mo,\mb,\Tp\cup\Tp'\cup\setof{Q},\Tn\cup\Tn'\rangle_\RQ$ and
\item $\md$ is a minimal diagnosis w.r.t.\ $\langle\mo,\mb,\Tp,\Tn\rangle_\RQ$ that satisfies $\Tp'$ as well as $\Tn'\cup\setof{Q}$ if and only if $\md$ is a minimal diagnosis w.r.t.\ $\langle\mo,\mb,\Tp\cup\Tp',\Tn\cup\Tn'\cup\setof{Q}\rangle_\RQ$.
\end{itemize}
Therefore, each query constructed during Algorithm~\ref{algo:inter_onto_debug} with $mode = static$ must be a query w.r.t.\ the current set of leading diagnoses $\mD_{\checkmark}$ and the \emph{current} DPI $\langle\mo,\mb,\Tp\cup\Tp',\Tn\cup\Tn'\rangle_\RQ$ (cf.\ Equation~\ref{eq:sol_ont_candidate}, Definition~\ref{def:q-partition} and Proposition~\ref{prop:dpi_update} on pages~\pageref{eq:sol_ont_candidate}-\pageref{prop:dpi_update}).

As a consequence of this, no additional test is required in order to ascertain that each diagnosis in the set $\mD_{\checkmark}$ that is given as a parameter to the next call of \textsc{staticHS} does in fact satisfy all answered queries so far.\qed
%That is, the new positive test cases $\Tp'$ are incorporated into the KBs $\ot_i$ (see Equation~\ref{eq:sol_ont_candidate} on page~\pageref{eq:sol_ont_candidate}) that are used to 
\end{remark} 

\section[Examples]{Illustrating Examples%
\sectionmark{Examples}}
\sectionmark{Examples}
\label{sec:TextscStaticHSExamples}
%%%%%
%\section{\textsc{staticHS}: Examples}
%\label{sec:TextscStaticHSExamples}
%%%%%
In this section we will give two examples of how interactive KB debugging using \textsc{staticHS} (Algorithm~\ref{algo:inter_onto_debug} with parameter $mode=static$) works. The first one will show the similarities and differences between the usage of \textsc{staticHS} (within Algorithm~\ref{algo:inter_onto_debug}) and \textsc{HS} (within Algorithm~\ref{algo:non_int_debug}) since it will depict the application of \textsc{staticHS} on the same example DPI (see Table~\ref{tab:example2}) that was used to show the functionality of \textsc{HS} in examples~\ref{example:non_interactive_debugging_with_tabExDpi2_and_without_probs} and \ref{example:non_interactive_debugging_with_tabExDpi2_and_probs}. At the same time, the first example will provide evidence that solving the problem of Interactive Static KB Debugging can be more efficient than solving the problem of Interactive Dynamic KB Debugging in terms of the number of query answers required from an interacting user. This will be discussed in more detail in Chapter~\ref{chap:TextscStaticHSVersusTextscDynamicHS}.

The second example is supposed to deepen the reader's understanding of the way \textsc{staticHS} works. To this end, the example DPI provided by Table~\ref{tab:example3} will be used which constitutes a significantly harder (interactive) debugging task than the DPI investigated in the first example. This example will involve the construction of a relatively large hitting set tree and thereby give a presentiment of the space and time complexity problems caused by the poor tree pruning inherent in the \textsc{staticHS} algorithm. In addition, this example will draw a reverse image of the first example in that it will stress the advantage of the decision to search for a solution of Interactive Dynamic KB Debugging rather than for a solution of Interactive Static KB Debugging (more on that in Chapter~\ref{chap:TextscStaticHSVersusTextscDynamicHS}).

\begin{example}\label{example:staticHS_simple_example_using_tabExDpi2}
In this example we assume that the author (called user throughout this example) of the (admissible) DPI $\langle\mo,\mb,\Tp,\Tn\rangle_\RQ$ given by Table~\ref{tab:example2} applies Algorithm~\ref{algo:inter_onto_debug} with $mode = static$ to interactively debug $\langle\mo,\mb,\Tp,\Tn\rangle_\RQ$. Further, suppose the following user requirements:

%  using the following parameters: 
In order to guarantee a fast reaction time of the system (the time between two successive queries to the user), the user wants each query to be computed from the minimally necessary number of leading diagnoses. Thus, in each iteration exactly two leading diagnoses should be computed by \textsc{staticHS} (cf.\ Proposition~\ref{prop:q1}). This postulation is reflected by setting $n_{\min} = n_{\max} = 2$. Notice that the time limit $t$ is irrelevant in this case. 

Moreover, the user desires to get just any query, i.e.\ they do not demand any particular properties -- such as optimal information gain among a pool of queries -- to be satisfied by a query. This can be ensured by choosing $q := 1$ (cf.\ Chapter~\ref{chap:QueryGeneration}) and $qsm()$ equal to any query selection measure described in Section~\ref{sec:query_selection_measures}. 

The user is new to KB debugging and has neither an idea of faults they frequently make nor access to any kind of data that would indicate their tendency to certain types of faults. Thence, $p_{\mo}(\tax) := c < 0.5$ for all $\tax \in \mo$, i.e.\ all formula fault probabilities are specified to be equal (to some constant $c$). In such a case, if a formula fault probability measure $p_{\mo}()$ is given as an input to Algorithm~\ref{algo:inter_onto_debug}, then line~\ref{algoline:inter_onto_debug:getAxiomProbs} in Algorithm~\ref{algo:inter_onto_debug} is omitted. Please notice that this aspect is not shown in Algorithm~\ref{algo:inter_onto_debug}.

Finally, the user's intention is to get the (exact) solution to the problem of Interactive Static KB Debugging. This can be taken into account by specifying $\sigma := 0$.

The tree constructed and parameters computed and used by Algorithm~\ref{algo:inter_onto_debug} using \textsc{staticHS} are visualized by Figure~\ref{fig:example:inter_onto_debug_staticHS_TabExDpi2}. 
We use the same notation as in Figures~\ref{fig:example:non-interactive_onto_debug_auto=false+nmin=infty_and_auto=true} and \ref{fig:example:non-interactive_onto_debug_auto=false+nmin=2+nmax=4_with_probs} which is described in Examples~\ref{example:non_interactive_debugging_with_tabExDpi2_and_without_probs} and \ref{example:non_interactive_debugging_with_tabExDpi2_and_probs}. The only new notational element here is the $\Longrightarrow$ labeled by some designator of a query. That is, $\checkmark_{(\md_i)} \stackrel{Q_j}{\Longrightarrow} \checkmark$ means that $\md_i$ is still a minimal diagnosis after $Q_j$ has been answered and added to the respective set of test cases of the DPI. On the other hand, $\checkmark_{(\md_i)} \stackrel{Q_j}{\Longrightarrow} \times$ signifies that the minimal diagnosis $\md_i$ is invalidated through the addition of the answered query $Q_j$ to the respective set of test cases of the DPI. Please notice that $\Longrightarrow$ does not point at a node of the wpHS-tree. Instead, the label at which $\Longrightarrow$ points is to be understood as the new label of the node originally labeled by $\checkmark_{(\md_i)}$ from which the (first of possibly multiple) $\Longrightarrow$ goes out. This notation should help to keep track of the evolution of node labels in the wpHS-tree without needing to overload a single node by multiple different successive labels. 

In the first iteration, i.e.\ during the execution of the first call of \textsc{staticHS} during Algorithm~\ref{algo:inter_onto_debug}, the root node (initially the empty set) is labeled by the minimal conflict set $\tuple{1,2,5}$ w.r.t.\ $\langle\mo,\mb,\Tp,\Tn\rangle_\RQ$ and three successor nodes, namely $\setof{1}$, $\setof{2}$ as well as $\setof{5}$, are added to the queue of open nodes $\Queue$. Since all formulas have been assigned an equal fault probability, \textsc{staticHS} conducts a breadth-first tree construction (as displayed by the numbers \textcircled{\scriptsize i} that give the order of node labeling). That is, $\Queue$ in this case is a first-in-first-out queue. In this vein, first $[1]$ and then $[2]$ are identified as minimal diagnoses w.r.t.\ the given DPI. Since $\mD_{\checkmark}\cup\mD_{calc} = \emptyset \cup \setof{[1],[2]}$ has a cardinality of $n_{\min} = n_{\max} = 2$, the stop criterion of \textsc{staticHS} causes it to terminate and return $\tuple{\mD_{calc}\cup\mD_{\checkmark},\mC_{calc},\Queue,\mD_{\times}} = \tuple{\setof{[1],[2]},\setof{\tuple{1,2,5}},[\setof{5}],\emptyset}$ (because $\mD_{\checkmark}$ and $\mD_{\times}$ are initially empty sets), as shown in the upper right column in Figure~\ref{fig:example:inter_onto_debug_staticHS_TabExDpi2}. 

Then, in Algorithm~\ref{algo:inter_onto_debug}, outside of the \textsc{staticHS} procedure, the first query $Q_1 = \setof{E \rightarrow \lnot A}$ is computed from the leading diagnoses set $\setof{[1],[2]}$. The q-partition $\Pt(Q_1)$ associated with $Q_1$ is $\langle\setof{[1]},\setof{[2]}$, $\emptyset\rangle$. The user's answer $u(Q_1)$ to $Q_1$ is then $\false$. Thence, the set $\mD_{out}$ is calculated from $\Pt(Q_1)$ as $\dx{}(Q_1) = \setof{[1]}$ (due to negative answer, cf.\ Remark~\ref{rem:invalidated_sets_of_q-partition_for_query_answer}), deleted from $\mD_{\checkmark} := \mD_{\checkmark} \cup \mD_{calc}$ to yield $\mD_{\checkmark} = \setof{[2]}$ and added to $\mD_{\times}$ to yield $\mD_{\times} = \setof{[1]}$.
The set $\mD_{\checkmark}$ corresponds to the set of all already computed minimal diagnoses w.r.t.\ the input DPI that satisfy all queries answered so far. The set $\mD_{\times}$ comprises all already computed minimal diagnoses w.r.t.\ the input DPI that do not satisfy all queries answered so far. 
These sets $\mD_{\checkmark}$ and $\mD_{\times}$ along with the collections $\Queue$ and $\mC_{calc}$ which are unmodified outside of \textsc{staticHS} are used as input arguments for the second call of \textsc{staticHS}. Notice that, in the figure, the resulting values of operations performed within \textsc{staticHS} are given in the righthand column above the dashed line whereas values computed outside of \textsc{staticHS} are given below the dashed line. 

After the modifications caused by the addition of the query $Q_1$ to the negative test cases of $\langle\mo,\mb,\Tp$, $\Tn\rangle_\RQ$ have been taken into account in step \textcircled{\scriptsize 4}, 
%throughout the second call of \textsc{staticHS}, 
the partial wpHS-tree built in iteration 1 is further constructed in iteration 2 resulting in the tree depicted by the middle picture in the lefthand column of Figure~\ref{fig:example:inter_onto_debug_staticHS_TabExDpi2}. Whereas the branches with edge labels $\setof{5,1}$ and $\setof{5,2}$ correspond to proper supersets of the minimal diagnoses $[1]$ and $[2]$, respectively, w.r.t.\ the input DPI $\tuple{\mo,\mb,\Tp,\Tn}_\RQ$ and are thus closed by the non-minimality criterion tested in the \textsc{sLabel} function, the branch with edge labels $\setof{5,7}$ is identified as a minimal diagnosis $\md_3 := [5,7]$ w.r.t.\ $\tuple{\mo,\mb,\Tp,\Tn}_\RQ$. However, $\md_3$ is not directly added to the set $\mD_{calc}$. In fact, the validity of the KB $\mo \setminus \md_3$ w.r.t.\ the \emph{current} DPI $\tuple{\mo,\mb,\Tp,\Tn\cup\setof{Q_1}}_\RQ$ is tested beforehand. As this test is successful, meaning that $\md_3 \in \minD_{\tuple{\mo,\mb,\Tp,\Tn}_\RQ} \cap \minD_{\tuple{\mo,\mb,\Tp,\Tn\cup\setof{Q_1}}_\RQ}$, $\md_3$ can be safely added to $\mD_{calc}$ implying the set of leading diagnoses $\mD_{\checkmark}\cup\mD_{calc} = \setof{\md_2,\md_3}$ with cardinality two. Due to $n_{\min} = n_{\max} = 2$, \textsc{staticHS} terminates.
%%$\md_3 \in \minD_{\tuple{\mo,\mb,\Tp,\Tn}_\RQ} \cap \minD_{\tuple{\mo,\mb,\Tp,\Tn\cup\setof{Q_1}}_\RQ}$ \\ 
		%since $(\mo\setminus\md_3)$ is $valid$ w.r.t.\ $\tuple{\mo,\mb,\Tp,\Tn}_\RQ$ (lines~\ref{algoline:slabel:qx}-\ref{algoline:slabel:return_valid}) \\ 
		%and w.r.t.\ $\tuple{\mo,\mb,\Tp,\Tn\cup\setof{Q_1}}_\RQ$ (lines~\ref{algoline:static:isOntValid}-\ref{algoline:static:add_to_Dcalc}) \\

After the second query $Q_2$ has been answered negatively involving the dismissal of the leading diagnosis $\md_2$, \textsc{staticHS} ends up with an empty queue $\Queue$ of open nodes in iteration 3 (see the tree in the lower left column of Figure~\ref{fig:example:inter_onto_debug_staticHS_TabExDpi2}). Hence, \textsc{staticHS} returns a singleton set including the leading diagnosis $\md_3$. Now, independently of the specified formula probabilities, $p_{\mD}(\md_3) = 1 \geq 1 - \sigma = 1$ is satisfied since the probability space considered by the probability measure $p_{\mD}()$ focuses on the sample space $\Omega = \setof{\md_3}$ (cf.\ Sections~\ref{sec:DiagnosisProbabilitySpace} and \ref{sec:DetailedAlgorithmDescription}). Thus, the stop condition of Algorithm~\ref{algo:inter_onto_debug} is met wherefore the solution KB $\mo_{sol} := (\mo \setminus \md_3) \cup U_{\Tp} = (\mo \setminus \md_3) \cup \emptyset = \mo \setminus \md_3$ is returned to the user. This solution KB $\mo_{sol}$ is the (exact) solution to Interactive Static KB Debugging given the DPI $\tuple{\mo,\mb,\Tp,\Tn}_\RQ$ of Table~\ref{tab:example2} as an input because $\md_3$ is the only minimal diagnosis w.r.t.\ $\tuple{\mo,\mb,\Tp,\Tn}_\RQ$ that conforms with all answered queries $Q_1 = \false$ and $Q_2 = \false$.

All in all, the execution of Algorithm~\ref{algo:inter_onto_debug} in this example performs 
\begin{itemize}
	\item 2 full $\scQX$ calls, i.e.\ calls of $\scQX$ that actually return a minimal conflict set (there are two minimal conflict sets labeled by $C$ in the picture at the bottom of the lefthand column in Figure~\ref{fig:example:inter_onto_debug_staticHS_TabExDpi2}) and
	\item 6 validity checks, i.e.\ calls of $\scQX$ that return 'no conflict' (one check for each of the three found minimal diagnoses; notice that $\scQX$ does only perform a single KB validity check by \textsc{isKBValid} in case it returns 'no conflict', see Algorithm~\ref{algo:qx}) or calls of \textsc{isKBValid} in line~\ref{algoline:static:isOntValid} in \textsc{staticHS} (one call for each of the three found minimal diagnoses),
\end{itemize}
computes
\begin{itemize}
	\item 3 minimal diagnoses w.r.t.\ the input DPI,
	\item 2 minimal conflict sets w.r.t.\ the input DPI and
	\item 2 queries and asks the user 2 logical formulas (1 per query)  
\end{itemize}
and stores
\begin{itemize}
	\item a maximum of 5 nodes (where node refers to the internal representation of a node in \textsc{staticHS} as a set of edge labels along a path from the root node to a leaf node; there are even more nodes in the sense of tree nodes in the picture at the bottom of the lefthand column in Figure~\ref{fig:example:inter_onto_debug_staticHS_TabExDpi2}).\qed
%\begin{itemize}
	%\item $n_{\min} = n_{\max} = 2$, i.e.\ in each iteration exactly two leading diagnoses should be computed by \textsc{staticHS}. Notice that the time limit $t$ is irrelevant in this case.
	%\item $p_{\mo}(\tax) = c < 0.5$ for all $\tax \in \mo$, i.e.\ all formula fault probabilities are assumed to be equal (to some constant $c$). This
\end{itemize}
\end{example}

\begin{figure}[t]
%%%%%%%%%%%%%%%%%%%%%%%%%%%%%%%%%%%%%%%%%%%% 1   
\begin{adjustwidth}{-1.5cm}{-2cm}
\begin{minipage}[c]{0.472\linewidth} %0.472 
\small
\xygraph{
!{<0cm,0cm>;<1.7cm,0cm>:<0cm,1.2cm>::}
%!~-{@[|(4)]}
%d=0
!{(1,4)}*+{\textcircled{\scriptsize 1}\tuple{1,2,5}^C}="c1c"
%d=1
!{(0,3) }*+{\textcircled{\scriptsize 2}\checkmark_{(\md_1)}}="d1" 
!{(1,3) }*+{\textcircled{\scriptsize 3}\checkmark_{(\md_2)}}="d2" 
!{(3,3) }*+{?}="c2c"
%d0->d1
"c1c":"d1"_{1}
"c1c":"d2"^{2}
"c1c":"c2c"^{5}
}
\vspace{5pt}
\begin{center}
\small \hspace{-12pt} Iteration 1
\end{center}
\end{minipage}
\begin{minipage}[c]{5pt}
\hspace{-14pt} $\Bigg>$ 
\end{minipage}
\begin{minipage}[c]{0.41\linewidth}
\small \hspace{-4pt}
\begin{tabular}{l}                                          
    %DPI: \\ $\tuple{\mt,\ma,\setof{\setof{B(w)}},\setof{\setof{\neg C(w)}}}$ \\
		$\mD_{\checkmark}\cup\mD_{calc} = \emptyset \cup \setof{\md_1,\md_2} = \setof{[1],[2]}$ \\
		$\Queue = [\setof{5}]$ \\
		$\mC_{calc} = \setof{\tuple{1,2,5}}$ \\
		$\mD_{\times} = \emptyset$ \vspace{-8pt} \\
    \hspace{-8pt}\hdashrule{1\textwidth}{0.5pt}{2mm}\vspace{-4pt} \\
		%$\rightarrow\quad 
		$\tuple{Q_1,\Pt(Q_1)} = \tuple{\setof{E \rightarrow \lnot A},\tuple{\setof{\md_1},\setof{\md_2},\emptyset}}$ \\
    %$\quad\quad\;\, 
		$u(Q_1) = \false$ \\
		%$\quad\quad\;\,
		$\mD_{\checkmark} = \setof{\md_2}$ \\
		%$\quad\quad\;\,
		$\mD_{out} = \mD_{\times} = \setof{\md_1}$
\end{tabular}
\end{minipage}
\begin{minipage}[c]{10pt}
$\Bigg>$ 
\end{minipage} 

\vspace{10pt}
%%%%%%%%%%%%%%%%%%%%%%%%%%%%%%%%%%%%%%%%%%%% 2
\begin{minipage}[c]{0.45\linewidth} 
\small
\xygraph{
!{<0cm,0cm>;<1.7cm,0cm>:<0cm,1.2cm>::}
%!~-{@[|(4)]}
%d=0
!{(1,4)}*+{\textcircled{\scriptsize 1}\tuple{1,2,5}^C}="c1c"
%d=1
!{(0,3) }*+{\textcircled{\scriptsize 2}\checkmark_{(\md_1)}}="d1"
!{(1,3) }*+{\textcircled{\scriptsize 3}\checkmark_{(\md_2)}}="d2" 
!{(3,3) }*+{\textcircled{\scriptsize 5}\tuple{1,2,7}^C}="c2c"
%d=2 
!{(0,2) }*+{\textcircled{\scriptsize 4}\times}="inv_d1_q1"
!{(1,2) }*+{\textcircled{\scriptsize 4}\checkmark}="val_d2_q1"
!{(2,2) }*+{\textcircled{\scriptsize 6}\times_{(\supset\md_1)}}="nonmin"
!{(3,2) }*+{\textcircled{\scriptsize 7}\times_{(\supset\md_2)}}="nonmin1"
!{(4,2) }*+{\textcircled{\scriptsize 8}\checkmark_{(\md_3)}}="d3"
%%d=3
%!{(1,1) }*+{\checkmark}="val_d2_q2"
%!{(2,1) }*+{\times}="inv_d3_q2"
%!{(3,1) }*+{\checkmark}="d5"
%!{(4,1) }*+{\checkmark}="val_d4_q3"
%%d=4
%!{(1,0) }*+{\times}="inv_d2_q3"
%!{(3,0) }*+{\times}="inv_d5_q4"
%!{(4,0) }*+{\checkmark}="val_d4_q4"
%d0->d1
"c1c":"d1"_{1}
"c1c":"d2"^{2}
"c1c":"c2c"^{5}
%d1->d2
"d1":@2{->}"inv_d1_q1"^{Q_1}
"d2":@2{->}"val_d2_q1"^{Q_1}
"c2c":"nonmin"_{1}
"c2c":"nonmin1"^{2}
"c2c":"d3"^{7}
%%d2->d3
%"val_d2_q1":@2{->}"val_d2_q2"^{Q_2}
%"d3":@2{->}"inv_d3_q2"^{Q_2}
%"nonmin1":@2{->}"d5"^{Q_3}
%"d4":@2{->}"val_d4_q3"^{Q_3}
%%d3->d4
%"d5":@2{->}"inv_d5_q4"^{Q_4}
%"val_d4_q3":@2{->}"val_d4_q4"^{Q_4}   
%"val_d2_q2":@2{->}"inv_d2_q3"^{Q_3}
}
\vspace{5pt}
\begin{center}
\small Iteration 2
\end{center}
\end{minipage}
\begin{minipage}[c]{15pt}
$\Bigg>$ 
\end{minipage}
\begin{minipage}[c]{0.41\linewidth}
\small\vspace{7pt}
\begin{tabular}{l}            
    %DPI: \\ $\tuple{\mt,\ma,\setof{\setof{B(w)}},\setof{\setof{\neg C(w)}}}$ \\
		%
		%$\md_3 \in \minD_{\tuple{\mo,\mb,\Tp,\Tn}_\RQ} \cap \minD_{\tuple{\mo,\mb,\Tp,\Tn\cup\setof{Q_1}}_\RQ}$ \\ 
		%since $(\mo\setminus\md_3)$ is $valid$ w.r.t.\ $\tuple{\mo,\mb,\Tp,\Tn}_\RQ$ (lines~\ref{algoline:slabel:qx}-\ref{algoline:slabel:return_valid}) \\ 
		%and w.r.t.\ $\tuple{\mo,\mb,\Tp,\Tn\cup\setof{Q_1}}_\RQ$ (lines~\ref{algoline:static:isOntValid}-\ref{algoline:static:add_to_Dcalc}) \\
		%
		$\mD_{\checkmark}\cup\mD_{calc} = \setof{\md_2} \cup \setof{\md_3} = \setof{[2],[5,7]}$ \\
		$\Queue = []$ \\
		$\mC_{calc} = \setof{\tuple{1,2,5}, \tuple{1,2,7}}$ \\
		$\mD_{\times} = \setof{\md_1} = \setof{[1]}$ \vspace{-8pt} \\
    \hspace{-8pt}\hdashrule{1\textwidth}{0.5pt}{2mm}\vspace{-4pt} \\
		%$\rightarrow\quad 
		$\tuple{Q_2,\Pt(Q_2)} = \tuple{\setof{Y \rightarrow \lnot A},\tuple{\setof{\md_2},\setof{\md_3},\emptyset}}$ \\
    %$\quad\quad\;\, 
		$u(Q_2) = \false$ \\
		%$\quad\quad\;\,
		$\mD_{\checkmark} = \setof{\md_3}$ \\
		%$\quad\quad\;\,
		$\mD_{out} = \setof{\md_2}$ \\
		%$\quad\quad\;\,
		$\mD_{\times} = \setof{\md_1,\md_2}$
		\end{tabular}
\end{minipage}
\begin{minipage}[c]{10pt}
$\Bigg>$ 
\end{minipage} 

\vspace{10pt}
%%%%%%%%%%%%%%%%%%%%%%%%%%%%%%%%%%%%%%%%%%%% 3
\begin{minipage}[c]{0.45\linewidth} 
\small
\xygraph{
!{<0cm,0cm>;<1.7cm,0cm>:<0cm,1.2cm>::}
%!~-{@[|(4)]}
%d=0
!{(1,4)}*+{\textcircled{\scriptsize 1}\tuple{1,2,5}^C}="c1c"
%d=1
!{(0,3) }*+{\textcircled{\scriptsize 2}\checkmark_{(\md_1)}}="d1"
!{(1,3) }*+{\textcircled{\scriptsize 3}\checkmark_{(\md_2)}}="d2" 
!{(3,3) }*+{\textcircled{\scriptsize 5}\tuple{1,2,7}^C}="c2c"
%d=2 
!{(0,2) }*+{\textcircled{\scriptsize 4}\times}="inv_d1_q1"
!{(1,2) }*+{\textcircled{\scriptsize 4}\checkmark}="val_d2_q1"
!{(2,2) }*+{\textcircled{\scriptsize 6}\times_{(\supset\md_1)}}="nonmin"
!{(3,2) }*+{\textcircled{\scriptsize 7}\times_{(\supset\md_2)}}="nonmin1"
!{(4,2) }*+{\textcircled{\scriptsize 8}\checkmark_{(\md_3)}}="d3"
%%d=3
!{(1,1) }*+{\textcircled{\scriptsize 9}\times}="inv_d2_q2"
%!{(2,1) }*+{\times}="inv_d3_q2"
%!{(3,1) }*+{\checkmark}="d5"
!{(4,1) }*+{\textcircled{\scriptsize 9}\checkmark}="val_d3_q2"
%%d=4
%!{(1,0) }*+{\times}="inv_d2_q3"
%!{(3,0) }*+{\times}="inv_d5_q4"
%!{(4,0) }*+{\checkmark}="val_d4_q4"
%d0->d1
"c1c":"d1"_{1}
"c1c":"d2"^{2}
"c1c":"c2c"^{5}
%d1->d2
"d1":@2{->}"inv_d1_q1"^{Q_1}
"d2":@2{->}"val_d2_q1"^{Q_1}
"c2c":"nonmin"_{1}
"c2c":"nonmin1"^{2}
"c2c":"d3"^{7}
%%d2->d3
"val_d2_q1":@2{->}"inv_d2_q2"^{Q_2}
%"d3":@2{->}"inv_d3_q2"^{Q_2}
%"nonmin1":@2{->}"d5"^{Q_3}
"d3":@2{->}"val_d3_q2"^{Q_2}
%%d3->d4
%"d5":@2{->}"inv_d5_q4"^{Q_4}
%"val_d4_q3":@2{->}"val_d4_q4"^{Q_4}   
%"val_d2_q2":@2{->}"inv_d2_q3"^{Q_3}
}
\vspace{5pt}
\begin{center}
\small Iteration 3
\end{center}
\end{minipage}
\begin{minipage}[c]{15pt}
$\Bigg>$ 
\end{minipage}
\begin{minipage}[c]{0.45\linewidth}
\small
\begin{tabular}{l}                                          
    %DPI: \\ $\tuple{\mt,\ma,\setof{\setof{B(w)}},\setof{\setof{\neg C(w)}}}$ \\
		$\mD_{\checkmark}\cup\mD_{calc} = \setof{\md_3} \cup \emptyset = \setof{[5,7]}$ \\
		$\Queue = []$ \\
		$\mC_{calc} = \setof{\tuple{1,2,5}, \tuple{1,2,7}}$ \\
		$\mD_{\times} = \setof{\md_1,\md_2} = \setof{[1],[2]}$\vspace{-8pt} \\
    \hspace{-8pt}\hdashrule{0.95\textwidth}{0.5pt}{2mm}\vspace{-4pt} \\
    %$\rightarrow\quad 
		$p_{\mD}(\md_3) = 1 $ \\
		 %\quad\quad\,\,
		$\Rightarrow\quad$ return the solution KB $(\mo\setminus\md_3) \qed$
		\end{tabular}
\end{minipage}
%\begin{minipage}[c]{10pt}
%$\Bigg>$ 
%\end{minipage} 

\vspace{10pt}
\caption[(Example~\ref{example:staticHS_simple_example_using_tabExDpi2}) Solving the Problem of Interactive Static KB Debugging]{(Example~\ref{example:staticHS_simple_example_using_tabExDpi2}) Solving the problem of Interactive Static KB Debugging (Problem Definition~\ref{prob_def:static}) for the \newline example DPI given by Table~\ref{tab:example2} by means of Algorithm~\ref{algo:inter_onto_debug} and \textsc{staticHS}.} 
\label{fig:example:inter_onto_debug_staticHS_TabExDpi2}
\end{adjustwidth} 
\end{figure}

\begin{example}\label{example:staticHS_complex_example_using_tabExDpi3}
Let us now consider the (admissible) DPI $\langle\mo,\mb,\Tp,\Tn\rangle_\RQ$ given by Table~\ref{tab:example3}. We assume an expert (called user throughout this example) in the domain $Dom$ modeled by $\mo$ who wants to find a solution to Interactive Static KB Debugging for the given DPI $\langle\mo,\mb,\Tp,\Tn\rangle_\RQ$ by means of Algorithm~\ref{algo:inter_onto_debug} with $mode = static$. 
Further, we suppose the following requirements:

%  using the following parameters: 
The user wants each query to be computed from three leading diagnoses. Thus, after each iteration of \textsc{staticHS}, the set $\mD_{\checkmark} \cup \mD_{calc}$ should comprise exactly three elements. This postulation is reflected by setting $n_{\min} = n_{\max} = 3$. Notice that the time limit $t$ is irrelevant in this case. 

Moreover, as in example~\ref{example:staticHS_simple_example_using_tabExDpi2}, we assume no demand for queries satisfying special properties which is reflected by choosing $q := 1$ (cf.\ Chapter~\ref{chap:QueryGeneration}) and $qsm()$ equal to any query selection measure described in Section~\ref{sec:query_selection_measures}.

Let there be several documentations of past debugging sessions (e.g.\ in terms of formula change logs) involving KBs in the domain $Dom$ of the author $auth$ of $\mo$ accessible to the user. Further, let the user have extracted term and logical construct probabilities $p_{\widetilde{\mo}\cup\overline{\mo}}(\tax)\in[0,1]$ for $\tax\in\mo$ for $auth$ from this data. This function $p_{\widetilde{\mo}\cup\overline{\mo}}: \widetilde{\mo}\cup\overline{\mo} \rightarrow [0,1]$ is then provided as an input to Algorithm~\ref{algo:inter_onto_debug}. 

Finally, the user's intention is to get the (exact) solution to the problem of Interactive Static KB Debugging. This can be taken into account by specifying $\sigma := 0$.

The tree constructed and parameters computed and used by Algorithm~\ref{algo:inter_onto_debug} using \textsc{staticHS} are visualized by Figures~\ref{fig:example:inter_onto_debug_staticHS_TabExDpi3} as well as \ref{fig:example:inter_onto_debug_staticHS_TabExDpi3_continued}. 
We use the same notation as in Figures~\ref{fig:example:non-interactive_onto_debug_auto=false+nmin=infty_and_auto=true}, \ref{fig:example:non-interactive_onto_debug_auto=false+nmin=2+nmax=4_with_probs} and \ref{fig:example:inter_onto_debug_staticHS_TabExDpi2} which is described in Examples~\ref{example:non_interactive_debugging_with_tabExDpi2_and_without_probs}, \ref{example:non_interactive_debugging_with_tabExDpi2_and_probs} and \ref{example:staticHS_simple_example_using_tabExDpi2}. 

After the initialization of variables, Algorithm~\ref{algo:inter_onto_debug} calls the function \textsc{getFormulaProbs} in line~\ref{algoline:inter_onto_debug:getAxiomProbs} which exploits $p_{\widetilde{\mo}\cup\overline{\mo}}()$ to calculate the function $p_{\mo}()$ giving the fault probabilities of formulas in $\mo$ (cf.\ Sections~\ref{sec:prob_space_construction}, \ref{sec:DetailedAlgorithmDescription} and Example~\ref{example:ax_prob_calc}). Let the resulting probabilities be as depicted by Table~\ref{tab:example:staticHS_complex--->axiom_probs}.
%\begin{table}[h]
	%\centering
		%\begin{tabular}{cc}
			%$\tax \in \mo$ & $p_{\mo}(\tax)$ \\
			%1 & 0.26 \\
			%2 & 0.18 \\
			%3 & 0.21 \\
			%4 & 0.41 \\
			%5 & 0.18 \\
			%6 & 0.40 \\
			%8 & 0.18 \\
		%\end{tabular}
%\end{table}
\begin{table}[tb]
	\centering
		\begin{tabular}{lccccccc}
			$\tax \in \mo$ & 1 & 2 & 3 & 4 & 5 & 6 & 8 \\\hline
			$p_{\mo}(\tax)$ & 0.26 & 0.18 & 0.21 & 0.41 & 0.18 & 0.40 & 0.18 
		\end{tabular}
\caption[(Example~\ref{example:staticHS_complex_example_using_tabExDpi3}) Formula Fault Probabilities]{(Example~\ref{example:staticHS_complex_example_using_tabExDpi3}) Computed formula fault probabilities for the example DPI given by Table~\ref{tab:example3}.}
\label{tab:example:staticHS_complex--->axiom_probs}
\end{table}

Then, \textsc{staticHS} is called for the first time, resulting in the wpHS-tree given in the first picture in Figure~\ref{fig:example:inter_onto_debug_staticHS_TabExDpi3}. Contrary to Example~\ref{example:staticHS_simple_example_using_tabExDpi2}, where the tree was built up in breadth-first order, in this example the formula probabilities $p() := p_{\mo}()$ given by Table~\ref{tab:example:staticHS_complex--->axiom_probs} are used to assign a probability $p_{nodes}(\mathsf{n})$ to each path $\mathsf{n}$ in the wpHS-tree starting from the root node (cf.\ Formula~\ref{eq:path_prob_calc} and Definition~\ref{def:p_node()}). In this vein, as outlined by the numbers \textcircled{\scriptsize i} indicating when a node is labeled, after the root node has been labeled by $\mc_1 := \tuple{1,2,5}$, the node corresponding to the outgoing edge of $\mc_1$ labeled by the formula with the largest fault probability among all formulas in $\mc_1$ is labeled first. That is, the node $\setof{1}$ with $p_{nodes}(\setof{1}) = 0.41$ (as opposed to the nodes $\setof{2}$ and $\setof{5}$ with $0.25$ each) is labeled first. The \textsc{sLabel} procedure, after checking whether $\setof{1}$ is a non-minimal diagnosis w.r.t.\ $\langle\mo,\mb,\Tp,\Tn\rangle_\RQ$ or a duplicate of some other node in $\Queue$ (both checks negative), computes another minimal conflict set $\mc_2 := \tuple{2,4,6}$ such that $\setof{1}\cap\mc_2 = \emptyset$ ($\mc_2$ is not hit by the node $\setof{1}$) to constitute a label for node $\setof{1}$. The successor nodes $\setof{1,2}$, $\setof{1,4}$ and $\setof{1,6}$ of $\setof{1}$ are generated and added to the list $\Queue$ in a way that the sorting of $\Queue$ in descending order of $p_{nodes}()$ is maintained.  

Since $\setof{1,4}$ (0.28) as well as $\setof{1,6}$ (0.27) have a larger probability (as per $p_{nodes}()$) than the nodes $\setof{2}$ (0.25) and $\setof{5}$ (0.25), $\Queue$ is given by $[\setof{1,4},\setof{1,6},\setof{2},\setof{5},\setof{1,2}]$ when it comes to the processing of the next node. Since \textsc{staticHS} always treats the first node of $\Queue$ next, it identifies the first minimal diagnosis $\md_1 := [1,4]$ w.r.t.\ $\langle\mo,\mb,\Tp,\Tn\rangle_\RQ$ in step \textcircled{\scriptsize 3}. In steps \textcircled{\scriptsize 4} and \textcircled{\scriptsize 8}, two further minimal diagnoses $\md_2 := [1,6]$ and $\md_3 := [5,4]$ are detected. Altogether, the union of $\mD_{\checkmark}$ (initially the empty set) and $\mD_{calc}$ (comprising the three computed diagnoses) now contains $3 = n_{\min} = n_{\max}$ elements wherefore \textsc{staticHS} terminates and outputs the tuple $\tuple{\mD_{calc}\cup\mD_{\checkmark},\mC_{calc},\Queue,\mD_{\times}}$ where the sets in this tuple are given under the wpHS-tree of iteration 1 in Figure~\ref{fig:example:inter_onto_debug_staticHS_TabExDpi3}.

From this set of leading diagnoses $\mD_{\checkmark} := \mD_{\checkmark} \cup \mD_{calc}$, the probability measure $p_{\mD}: \mD_{\checkmark} \rightarrow [0,1]$ is computed by the function \textsc{getProbDist} (cf.\ Algorithm~\ref{algo:inter_onto_debug_continued} and Section~\ref{sec:DetailedAlgorithmDescription}). The result is $\tuple{p_{\mD}(\md_1),p_{\mD}(\md_2),p_{\mD}(\md_3)} = \tuple{0.38,0.37,0.25}$. The mode $\md_{\max} := \md_1$ of this probability distribution is then computed by \textsc{getMode}. As $\sigma = 0$, $p_{\mD}(\md_{\max}) = 0.38 \not\geq 1$ wherefore the stop criterion of Algorithm~\ref{algo:inter_onto_debug} is not satisfied. 

Consequently, Algorithm~\ref{algo:inter_onto_debug} proceeds to generate the first query $Q_1 = \setof{B \sqsubseteq K}$ (based on the current set of leading diagnoses $\mD_{\checkmark}$) along with its associated q-partition $\Pt(Q_1) = \tuple{\setof{\md_1,\md_2},\setof{\md_3},\emptyset}$. The diagnosis $\md_1$ is in $\dx{}(Q_1)$ because $\ot_1 = (\mo\setminus\md_1) \cup \mb \cup U_{\Tp}$ (recall Formula~\ref{eq:sol_ont_candidate} for a definition of $\ot_i$) comprises formulas 2, 3, 5, 6, 7, 8 and 9 as well as $\tp_1$ (cf.\ Table~\ref{tab:example3}) wherefore $\ot_1 \models \setof{B \sqsubseteq K} = Q_1$ (due to the set of formulas $\setof{2,3} = \setof{B\sqsubseteq G, G \sqsubseteq K}$). That $\md_2$ belongs to $\dx{}(Q_1)$ as well follows analogously. On the other hand, $\md_3 \in \dnx{}(Q_1)$ must be true since $\ot_3 \cup Q_1$ includes i.a.\ $A \sqsubseteq B$ (formula 1) and $B \sqsubseteq K$ ($\in Q_1$) wherefore $\setof{A \sqsubseteq K} = \tn_1$ is an entailment of $\ot_3$. Thus, the negative test case $\tn_1$ is violated.    

The positive user answer $u(Q_1) = \true$ is incorporated in that $Q_1$ is appended to the set of positive test cases $\Tp$ yielding $\Tp \cup \setof{Q_1} = \setof{\setof{r(x,y)},\setof{B \sqsubseteq K}}$. Step~\textcircled{\scriptsize 9} shows the impact of this test case addition on the set of leading diagnoses, i.e.\ all diagnoses in the set $\mD_{out} = \dnx{}(Q_1) = \setof{\md_3}$ (due to positive answer, cf.\ Remark~\ref{rem:invalidated_sets_of_q-partition_for_query_answer}) are re-labeled by $\times$ whereas all other leading diagnoses ($\md_1, \md_2$) are still labeled by $\checkmark$.

In the same fashion, further node labelings are conducted in iteration 2 until $|\mD_{\checkmark} \cup \mD_{calc}| = |\setof{\md_1, \md_2} \cup \setof{[2,1]}| = 3 = n_{\min} = n_{\max}$ holds again. These actions are displayed by the tree at the bottom of Figure~\ref{fig:example:inter_onto_debug_staticHS_TabExDpi3}. 

Notice that, after step~\textcircled{\tiny 12}, two nodes corresponding to the same set are elements of the list $\Queue$. At step~\textcircled{\tiny 13}, the duplicate criterion checked by \textsc{sLabel} comes into play. Since the node $\setof{1,2}$ (the leftmost branch in the tree) is ranked first in $\Queue$ (we assume a first-in-first-out ordering of nodes corresponding to equal sets of edge labels in $\Queue$), the \textsc{sLabel} procedure is called given $\mathsf{node} := \setof{1,2}$ as an argument and detects the node $\setof{2,1}$ (the fourth leftmost branch in the tree) in $\Queue$. Hence, $\mathsf{node} = \setof{1,2}$ is closed as a duplicate node which finds expression in the label $\times_{(dup)}$. When $\setof{2,1}$ (which must have the same probability as $\setof{1,2}$ due to set-equality) is processed at step~\textcircled{\tiny 14}, it is discovered to be a minimal diagnosis ($\md_5$) w.r.t.\ $\langle\mo,\mb,\Tp,\Tn\rangle_\RQ$.

Moreover, we want to point out that another minimal diagnosis ($\md_4 = [2,4,6]$) is found in iteration 2 before $\md_5$ is detected. However, $\md_4$ is immediately ruled out and added to $\mD_{\times}$ (cf.\ line~\ref{algoline:static:add_to_Dtimes} in \textsc{staticHS}) due to the fact that $\mo \setminus \md_4$ is invalid w.r.t.\ the \emph{current} DPI $\langle\cdot,\mb,\Tp\cup\setof{Q_1},\Tn\rangle_\RQ$ (cf.\ Definition~\ref{def:valid_onto}). The explanation why this holds is as follows: 

By Definition~\ref{def:valid_onto}, $\mo\setminus\md_4$ is valid w.r.t.\ $\langle\cdot,\mb,\Tp\cup\setof{Q_1},\Tn\rangle_\RQ$ iff $\ot_4 = (\mo \setminus \md_4) \cup \mb \cup U_{(\Tp\cup\setof{Q_1})}$ (recall Formula~\ref{eq:sol_ont_candidate} for a definition of $\ot_i$) does not violate any $r\in\RQ = \setof{\text{consistency},\text{coherency}}$ and does not entail any $\tn \in \Tn = \setof{\tn_1,\tn_2} = \setof{\setof{A \sqsubseteq K},\setof{L \sqsubseteq \exists r.F, B(x), G \sqsubseteq K}}$. Applying the diagnosis $\md_4$ to $\mo$ yields $\mo \setminus \md_4 = \setof{1,3,5,8}$ which includes in particular formula $1$ which is equal to $A \sqsubseteq B$ (see Table~\ref{tab:example3}). However, there is also the negative test case $\tn_1$ indicating that $A \sqsubseteq K$ must not be entailed by $\ot_4$. That is, $B \sqsubseteq K \in \ot_4$ (due to $Q_1$) and $A \sqsubseteq B \in \ot_4$ which implies that $\ot_4 \models \setof{A \sqsubseteq K} = \tn_1$ wherefore $\ot_4$ is invalid w.r.t.\ $\langle\cdot,\mb,\Tp\cup\setof{Q_1},\Tn\rangle_\RQ$.  

Such a direct dismissal of a discovered diagnosis $\md_i$ due to a newly added test case $Q_j$ is indicated by $\textcircled{\scriptsize k} \checkmark_{(\md_i)} \stackrel{Q_j}{\Longrightarrow}  \textcircled{\scriptsize k}\times$, i.e.\ the step number \textcircled{\scriptsize k} at the shaft of the $\Longrightarrow$ is equal to the step number at the head of $\Longrightarrow$. In case of the invalidation of a \emph{leading} diagnosis (i.e.\ one that was utilized in the computation of $Q_j$), on the contrary, the step number at the shaft is lower than the step number at the arrow head.

As shown at the top of Figure~\ref{fig:example:inter_onto_debug_staticHS_TabExDpi3_continued}, the second query $Q_2$ computed from the leading diagnosis set $\mD_{\checkmark} \cup \mD_{calc} = \setof{\md_1,\md_2,\md_5}$ is then answered by $u(Q_2) = \true$ as well, wherefore the leading diagnoses $\md_2, \md_5$ are ruled out and added to $\mD_{\times}$. So, the input argument $\mD_{\checkmark}$ given to the next call of \textsc{staticHS} in Algorithm~\ref{algo:inter_onto_debug} consists of the single diagnosis $\md_1$. 

In the third iteration (see the picture given in Figure~\ref{fig:example:inter_onto_debug_staticHS_TabExDpi3_continued}), \textsc{staticHS} again executes in order to complete the leading diagnosis set to contain three elements. However, as we can say in advance, $\md_1$ is the only minimal diagnosis w.r.t.\ the input DPI $\langle\mo,\mb,\Tp,\Tn\rangle_\RQ$ which is also a diagnosis w.r.t.\ the current DPI $\langle\mo,\mb,\Tp\cup\setof{Q_1,Q_2},\Tn\rangle_\RQ$. Nevertheless, \textsc{staticHS} continues expanding the wpHS-tree until it has verified that this is the case ($\Queue = []$). This is equivalent to finishing the construction of the non-interactive wpHS-tree that is generated by \textsc{HS} with parameters $n_{\min} = n_{\max} = \infty$. We want to stress that the construction of the entire wpHS-tree w.r.t.\ $\langle\mo,\mb,\Tp,\Tn\rangle_\RQ$ and $p() := p_{\mo}()$ is inevitable in a debugging scenario where the (exact) solution to the Interactive Static KB Debugging problem is sought (the probability w.r.t.\ $p_{\mD}()$ of a diagnosis can only be equal to 1 if there is only a single leading diagnosis returned by \textsc{staticHS}). 

In fact, there are five further diagnoses $\md_6,\dots,\md_{10}$ w.r.t.\ $\langle\mo,\mb,\Tp,\Tn\rangle_\RQ$ that are detected in iteration 3 and directly dismissed (added to $\mD_{\times}$) after the validity check in line~\ref{algoline:static:isOntValid} of \textsc{staticHS}. All other tree branches are closed due to the non-minimality (label $\times_{(\supset \md_i)}$) or duplicate criterion (label $\times_{(dup)}$).  
Due to $\sigma = 0$ and the associated necessity to grow the wpHS-tree until all leaf nodes are labeled, the final tree (19 labeled leaf nodes) depicted in Figure~\ref{fig:example:inter_onto_debug_staticHS_TabExDpi3_continued} is relatively large in comparison to the small size $|\mo| = 7$. 

This example might already give an idea of the potential explosion of the wpHS-tree produced by \textsc{staticHS} in case the (exact) solution to the Interactive Static KB Debugging problem is desired. This is why it will usually make sense in practice to specify a fault tolerance $\sigma > 0$ which enables Algorithm~\ref{algo:inter_onto_debug} with $mode = static$ to escape from the generally intractable complexity of the complete investigation of all minimal diagnoses w.r.t.\ the input DPI (full construction of the wpHS-tree). However, in this concrete example, allowing a small fault tolerance $\sigma$ has no effect either. Actually, $\sigma \geq 0.56$ is necessary to achieve a premature termination of the tree construction. This holds due to the fact that the probability distributions of leading diagnoses are $\tuple{p_{\mD}(\md_1),p_{\mD}(\md_2),p_{\mD}(\md_3)} = \tuple{0.38,0.37,0.25}$ (after iteration 1) and $\tuple{p_{\mD}(\md_1),p_{\mD}(\md_2),p_{\mD}(\md_5)} = \tuple{0.44,0.42,0.14}$ (after iteration 2). Now, given say $\sigma := 0.6$, the stop criterion of Algorithm~\ref{algo:inter_onto_debug} would be met after iteration 2 because $p_{\mD}(\md_{\max}) = p_{\mD}(\md_1) = 0.44 \geq 0.4 = 1 - 0.6 = 1-\sigma$. Nate that, in this case, the same (exact) solution would be returned as for the setting $\sigma := 0$. The (significant) difference is just that the final tree in this case has only 14 leaf nodes, of which only 7 are labeled (the labeling of a node is in general significantly more costly than the mere generation of a node). As opposed to this, the full tree comprises 19 labeled nodes. On the other side of the coin, choosing a value of $\sigma > 0.5$, for example, means that -- from the point of view of the knowledge at the time Algorithm~\ref{algo:inter_onto_debug} terminates -- a solution to Interactive Static KB Debugging is returned by Algorithm~\ref{algo:inter_onto_debug} which has a higher probability of not being the (exact) solution than of being the (exact) solution.

All in all, the execution of Algorithm~\ref{algo:inter_onto_debug} in this example performs 
\begin{itemize}
	\item 4 full $\scQX$ calls, i.e.\ calls of $\scQX$ that actually return a minimal conflict set (there are four minimal conflict sets labeled by $C$ in the tree in Figure~\ref{fig:example:inter_onto_debug_staticHS_TabExDpi3_continued}) and
	\item 20 validity checks, i.e.\ calls of $\scQX$ that return 'no conflict' (one check for each of the 10 found minimal diagnoses; notice that $\scQX$ does only perform a single KB validity check by \textsc{isKBValid} in case it returns 'no conflict', see Algorithm~\ref{algo:qx}) or calls of \textsc{isKBValid} in line~\ref{algoline:static:isOntValid} in \textsc{staticHS} (one call for each of the 10 found minimal diagnoses),
	%\item 20 validity checks, i.e.\ calls of $\scQX$ that return 'no conflict' (the 10 found minimal diagnoses w.r.t.\ the input DPI) or calls of \textsc{isOntValid} for a found diagnosis (line~\ref{algoline:static:isOntValid} in \textsc{staticHS}),
\end{itemize}
computes
\begin{itemize}
	\item 10 minimal diagnoses w.r.t.\ the input DPI,
	\item 4 minimal conflict sets w.r.t.\ the input DPI and
	\item 2 queries and asks the user 2 logical formulas (1 per query)  
\end{itemize}
and stores
\begin{itemize}
	\item a maximum of 19 nodes (where node refers to the internal representation of a node in \textsc{staticHS} as a set of edge labels along a path from the root node to a leaf node; there are even more nodes in the sense of tree nodes in the picture in Figure~\ref{fig:example:inter_onto_debug_staticHS_TabExDpi3_continued}).\qed
\end{itemize}
%analysis: how many calls to QX etc.
%no pruning -- problematic, explosion
%all unseen minimal diagnoses w.r.t.\ input DPI must be computed in order to let staticHS terminate if exact solution is desired, might be must faster and space-saving with a positive, but in this case sigma must be set to higher than 0.5 to achieve that 
%
%if the user is lucky to select some sigma that lets staticHS terminate before no more unseen minimal diagnoses w.r.t.\ the input DPI exists, it may work much better... 
\end{example}

\newgeometry{margin=2cm}

\begin{sidewaysfigure*}
%%%%%%%%%%%%%%%%%%%%%%%%%%%%%%%%%%%%%%%%%%%% 1
\begin{minipage}[c]{0.95\textwidth} 
\xygraph{
!{<0cm,0cm>;<1.5cm,0cm>:<0cm,1.5cm>::}
%!~-{@[|(4)]}
%d=0
!{(4,4)}*+{\textcircled{\footnotesize 1}\tuple{1,2,5}^C}="c1c"
%d=1
!{(1,3) }*+{\textcircled{\footnotesize 2}\tuple{2,4,6}^C}="c2c" 
!{(4,3) }*+{\textcircled{\footnotesize 5}\tuple{1,3,4}^C}="c3c" 
!{(11,3) }*+{\textcircled{\footnotesize 6}\tuple{2,4,6}^R}="c2r"
%d=2
!{(0,2) }*+{?}="d4"
!{(1,2) }*+{\textcircled{\footnotesize 3}\checkmark_{(\md_1)}}="d1"
!{(2,2) }*+{\textcircled{\footnotesize 4}\checkmark_{(\md_2)}}="d2"
!{(3,2) }*+{?}="dup1"
!{(5,2) }*+{?}="?2-2"
!{(8,2) }*+{\textcircled{\footnotesize 7}\tuple{1,5,6,8}^C}="c4c"
!{(10,2) }*+{?}="?2-3"
!{(12,2) }*+{\textcircled{\footnotesize 8}\checkmark_{(\md_3)}}="d3"
!{(14,2) }*+{?}="?2-4"
%d=3
!{(7,1) }*+{?}="?3-1"
!{(8,1) }*+{?}="?3-2"
!{(9,1) }*+{?}="?3-3"
!{(10,1) }*+{?}="?3-4"
%!{(4,0) }*+{?}="?3-1"
%!{(5,0) }*+{?}="?3-2"
%!{(6,0) }*+{?}="?3-3"
%!{(7,0) }*+{?}="?3-4"
%d0->d1
"c1c":"c2c"_{1}^(0.65){0.41}
"c1c":"c3c"_{2}^(0.7){0.25}
"c1c":"c2r"_{5}^(0.87){0.25}
%d1->d2
"c2c":"d4"_{2}_(0.9){0.09}
"c2c":"d1"_{4}^(0.75){0.28}
"c2c":"d2"^{6}^(0.85){0.27}
"c3c":"dup1"_{1}^(0.6){0.09}
"c3c":"?2-2"_{3}^(0.88){0.07}
"c3c":"c4c"_{4}^(0.75){0.18}
"c2r":"?2-3"_{2}^(0.7){0.06}
"c2r":"d3"_(0.4){4}^(0.8){0.18}
"c2r":"?2-4"^{6}^(0.88){0.17}
%d2->d3
"c4c":"?3-1"_{1}_(0.85){0.06}
"c4c":"?3-2"_{5}_(0.78){0.04}
"c4c":"?3-3"_{6}^(0.9){0.11}
"c4c":"?3-4"^{8}^(0.9){0.04}
}
\vspace{5pt}
\begin{center}
\small Iteration 1
\end{center}
\end{minipage}
\begin{minipage}[c]{20pt}
$\Bigg>$ 
\end{minipage}

\vspace{20pt}
\begin{minipage}[c]{0.95\textwidth}
\small  
\begin{tabular}{l}                                          
    %DPI: \\ $\tuple{\mt,\ma,\setof{\setof{B(w)}},\setof{\setof{\neg C(w)}}}$ \\
		%$\mD_{\times} = \emptyset$   \\
		%$\mD_{\checkmark} = \emptyset$   \\
		$\mD_{\checkmark}\cup\mD_{calc} = \emptyset\cup\setof{\md_1, \md_2, \md_3} = \setof{[1,4],[1,6],[5,4]}$, 
		$\quad\Queue = [\setof{5,6},\setof{2,4,6},\setof{1,2},\setof{2,1},\setof{2,3}, %$\\
		%$\quad\quad\;\;\, 
		\setof{5,2},\setof{2,4,1},\setof{2,4,5},\setof{2,4,8}]$, 
		$\quad\mD_{\times} = \emptyset$\\
		%$\quad\mD_{\times} = \emptyset$, \\
		$\mC_{calc} = \setof{\tuple{1,2,5},\tuple{2,4,6},\tuple{1,3,4},\tuple{1,5,6,8}}$, \vspace{-8pt} \\
    \hspace{-8pt}\hdashrule{0.95\textwidth}{0.5pt}{2mm}\vspace{-4pt} \\
		%$\rightarrow\quad
		$\tuple{Q_1,\Pt(Q_1)} = \tuple{\setof{B \sqsubseteq K},\tuple{\setof{\md_1,\md_2},\setof{\md_3},\emptyset}}$, 
    $\quad u(Q_1) = \true$, 
		$\quad\mD_{\checkmark} = \setof{\md_1,\md_2}$,  $\quad \mD_{out} = \mD_{\times} = \setof{\md_3}$
		%$\mD_{calc} = \setof{\md_1, \md_2, \md_3} = \setof{[1,4],[1,6],[5,4]}$ \\
		%$\Queue = \{[5,6],[2,4,6],[1,2],[2,3],[5,2], $\\
		%$\quad\quad\;\;\, [2,4,1],[2,4,5],[2,4,8]\}$ \\
		%$\mC_{calc} = \setof{\tuple{1,2,5},\tuple{2,4,6},\tuple{1,3,4},\tuple{1,5,6,8}}$ \\
		%$\mD_{\supset} = \emptyset$ \\
		%$\Queue_{dup} = \setof{[2,1]}$ \\
    %$\tuple{Q_1,\Pt(Q_1)} = \tuple{\setof{B \sqsubseteq K},\tuple{\setof{\md_1,\md_2},\setof{\md_3},\emptyset}}$ \\
    %$u(Q_1) = \true$ \\
		%$\mD_{\checkmark} = \setof{\md_1,\md_2}$, $\mD_{\times} = \setof{\md_3}$
		\end{tabular}
\end{minipage}
\begin{minipage}[c]{10pt}
$\Bigg>$ 
\end{minipage} 

\vspace{10pt}
%%%%%%%%%%%%%%%%%%%%%%%%%%%%%%%%%%%%%%%%%%%%% 2
%\begin{minipage}[c]{20pt}
%$\Bigg>$ 
%\end{minipage}
\begin{minipage}[c]{0.95\textwidth} 
\xygraph{
!{<0cm,0cm>;<1.45cm,0cm>:<0cm,1.5cm>::}
%!~-{@[|(4)]}
%d=0
!{(4,4)}*+{\textcircled{\footnotesize 1}\tuple{1,2,5}^C}="c1c"
%d=1
!{(1,3) }*+{\textcircled{\footnotesize 2}\tuple{2,4,6}^C}="c2c" 
!{(4,3) }*+{\textcircled{\footnotesize 5}\tuple{1,3,4}^C}="c3c" 
!{(11,3) }*+{\textcircled{\footnotesize 6}\tuple{2,4,6}^R}="c2r"
%d=2
!{(0,2) }*+{\textcircled{\tiny 13}\times_{(dup)}}="dup1"
!{(1,2) }*+{\textcircled{\footnotesize 3}\checkmark_{(\md_1)}}="d1"
!{(2,2) }*+{\textcircled{\footnotesize 4}\checkmark_{(\md_2)}}="d2"
!{(3,2) }*+{\textcircled{\tiny 14}\checkmark_{(\md_5)}}="d5"
!{(5,2) }*+{?}="?2-2"
!{(8,2) }*+{\textcircled{\footnotesize 7}\tuple{1,5,6,8}^C}="c4c"
!{(10,2) }*+{?}="?2-3"
!{(12,2) }*+{\textcircled{\footnotesize 8}\checkmark_{(\md_3)}}="d3"
!{(14,2) }*+{\textcircled{\tiny 10}\tuple{1,3,4}^R}="c3r"
%d=3
!{(7,1) }*+{?}="?3-1"
!{(8,1) }*+{?}="?3-2"
!{(10,1) }*+{?}="?3-4"
!{(9,1) }*+{\textcircled{\tiny 11}\checkmark_{(\md_4)}}="d4"
!{(12,1) }*+{\textcircled{\footnotesize 9}\times}="inv_d3_q1"
!{(1,1) }*+{\textcircled{\footnotesize 9}\checkmark}="val_d1_q1"
!{(2,1) }*+{\textcircled{\footnotesize 9}\checkmark}="val_d2_q1"
!{(13,1) }*+{?}="?3-5"
!{(14,1) }*+{?}="?3-6"
!{(15,1) }*+{\textcircled{\tiny 12}\times_{(\supset\md_3)}}="sup_d3"
%d=4
!{(9,0) }*+{\textcircled{\tiny 11}\times}="inv_d4_q1"
%!{(15,0) }*+{\textcircled{\tiny }\times}="inv_d5_q1"
%d0->d1
"c1c":"c2c"_{1}^(0.65){0.41}
"c1c":"c3c"_{2}^(0.7){0.25}
"c1c":"c2r"_{5}^(0.87){0.25}
%d1->d2
"c2c":"dup1"_{2}_(0.85){0.09}
"c2c":"d1"_{4}^(0.75){0.28}
"c2c":"d2"^{6}^(0.85){0.27}
"c3c":"d5"_{1}^(0.6){0.09}
"c3c":"?2-2"_{3}^(0.88){0.07}
"c3c":"c4c"_{4}^(0.75){0.18}
"c2r":"?2-3"_{2}^(0.7){0.06}
"c2r":"d3"_{4}^(0.8){0.18}
"c2r":"c3r"^{6}^(0.88){0.17}
%d2->d3
"c4c":"?3-1"_{1}_(0.85){0.06}
"c4c":"?3-2"_{5}_(0.78){0.04}
"c4c":"?3-4"^{8}^(0.9){0.04}
"c4c":"d4"_{6}^(0.9){0.11}
"d3":@2{->}"inv_d3_q1"^{Q_1}
"d1":@2{->}"val_d1_q1"^{Q_1}
"d2":@2{->}"val_d2_q1"^{Q_1}
"c3r":"?3-5"_{1}_(0.75){0.06}
"c3r":"?3-6"_{3}^(0.75){0.04}
"c3r":"sup_d3"^(0.55){4}^(0.85){0.11}
%d3->d4
"d4":@2{->}"inv_d4_q1"^{Q_1}
%"d5":@2{->}"inv_d5_q1"^{Q_1}
}
\vspace{5pt}
\begin{center}
\small Iteration 2
\end{center}
\end{minipage}
\begin{minipage}[c]{20pt}
$\Bigg>$ 
\end{minipage} 

\vspace{10pt}
\caption[(Example~\ref{example:staticHS_complex_example_using_tabExDpi3}) Solving the Problem of Interactive Static KB Debugging]{(Example~\ref{example:staticHS_complex_example_using_tabExDpi3}) Solving the problem of Interactive Static KB Debugging (Problem Definition~\ref{prob_def:static}) for the example DPI given by Table~\ref{tab:example3} by means of %\newline 
Algorithm~\ref{algo:inter_onto_debug} and \textsc{staticHS}.} 
\label{fig:example:inter_onto_debug_staticHS_TabExDpi3}
\end{sidewaysfigure*}

\begin{sidewaysfigure*}
\begin{minipage}[c]{0.95\textwidth}
\small  
\begin{tabular}{l}                                          
    %DPI: \\ $\tuple{\mt,\ma,\setof{\setof{B(w)}},\setof{\setof{\neg C(w)}}}$ \\
		%$\mD_{\times} = \emptyset$   \\
		%$\mD_{\checkmark} = \emptyset$   \\
		$\mD_{\checkmark} \cup \mD_{calc} = \setof{\md_1, \md_2} \cup \setof{\md_5} = \setof{[1,4],[1,6],[2,1]}$, 
		$\quad\Queue = [\setof{2,4,6},\setof{2,3},\setof{5,2},\setof{2,4,1},\setof{5,6,1},\setof{2,4,5},\setof{5,6,3}]$,
		$\quad\mD_{\times} = \setof{\md_3,\md_4} = \setof{[5,4],[2,4,6]}$   \\
		$\mC_{calc} = \setof{\tuple{1,2,5},\tuple{2,4,6},\tuple{1,3,4},\tuple{1,5,6,8}}$, \vspace{-8pt} \\
    \hspace{-8pt}\hdashrule{0.95\textwidth}{0.5pt}{2mm}\vspace{-4pt} \\		
    %$\rightarrow\quad
		$\tuple{Q_2,\Pt(Q_2)} = \tuple{\setof{B \sqsubseteq \exists r.F},\tuple{\setof{\md_1},\setof{\md_2,\md_5},\emptyset}}$, 
    $\quad u(Q_2) = \true$,
		$\quad \mD_{\checkmark} = \setof{\md_1}$, $\quad\mD_{out} = \setof{\md_2,\md_5}$, $\quad\mD_{\times} = \setof{\md_3,\md_4,\md_2,\md_5}$
		%$\mD_{calc} = \setof{\md_1, \md_2, \md_3} = \setof{[1,4],[1,6],[5,4]}$ \\
		%$\Queue = \{[5,6],[2,4,6],[1,2],[2,3],[5,2], $\\
		%$\quad\quad\;\;\, [2,4,1],[2,4,5],[2,4,8]\}$ \\
		%$\mC_{calc} = \setof{\tuple{1,2,5},\tuple{2,4,6},\tuple{1,3,4},\tuple{1,5,6,8}}$ \\
		%$\mD_{\supset} = \emptyset$ \\
		%$\Queue_{dup} = \setof{[2,1]}$ \\
    %$\tuple{Q_1,\Pt(Q_1)} = \tuple{\setof{B \sqsubseteq K},\tuple{\setof{\md_1,\md_2},\setof{\md_3},\emptyset}}$ \\
    %$u(Q_1) = \true$ \\
		%$\mD_{\checkmark} = \setof{\md_1,\md_2}$, $\mD_{\times} = \setof{\md_3}$
		\end{tabular}
\end{minipage}
\begin{minipage}[c]{10pt}
$\Bigg>$ 
\end{minipage} 

\vspace{10pt}
%%%%%%%%%%%%%%%%%%%%%%%%%%%%%%%%%%%%%%%%%%%%% 2
\begin{minipage}[c]{0.95\textwidth} 
\xygraph{
!{<0cm,0cm>;<1.27cm,0cm>:<0cm,1.5cm>::}
%!~-{@[|(4)]}
%d=0
!{(4,4)}*+{\textcircled{\footnotesize 1}\tuple{1,2,5}^C}="c1c"
%d=1
!{(1,3) }*+{\textcircled{\footnotesize 2}\tuple{2,4,6}^C}="c2c" 
!{(6,1) }*+{\textcircled{\footnotesize 5}\tuple{1,3,4}^C}="c3c" 
!{(13,4) }*+{\textcircled{\footnotesize 6}\tuple{2,4,6}^R}="c2r"
%d=2
%\framebox[1.1\width]{Guess I’m framed now!}
!{(0,2) }*+{\textcircled{\tiny 13}\times_{(dup)}}="dup1"
!{(1,2) }*+{\textcircled{\footnotesize 3}\checkmark_{(\md_1)}}="d1"
!{(2,2) }*+{\textcircled{\footnotesize 4}\checkmark_{(\md_2)}}="d2"
!{(3,0) }*+{\textcircled{\tiny 14}\checkmark_{(\md_5)}}="d5"
!{(7,0) }*+{\textcircled{\tiny 16}\tuple{1,5,6,8}^R}="c4r"
!{(13,0) }*+{\textcircled{\footnotesize 7}\tuple{1,5,6,8}^C}="c4c"
!{(9,3) }*+{\textcircled{\tiny 17}\tuple{1,3,4}^R}="c3r1"
!{(12,3) }*+{\textcircled{\footnotesize 8}\checkmark_{(\md_3)}}="d3"
!{(15,3) }*+{\textcircled{\tiny 10}\tuple{1,3,4}^R}="c3r"
%d=3
!{(10,-1) }*+{\textcircled{\tiny 18}\times_{(\supset\md_5)}}="sup_d5"
!{(12,-1) }*+{\textcircled{\tiny 20}\times_{(\supset\md_3)}}="sup_d3_1"
!{(16,-1) }*+{\textcircled{\tiny 21}\checkmark_{(\md_6)}}="d6"
!{(14,-1) }*+{\textcircled{\tiny 11}\checkmark_{(\md_4)}}="d4"
!{(12,2) }*+{\textcircled{\footnotesize 9}\times}="inv_d3_q1"
!{(1,1) }*+{\textcircled{\footnotesize 9}\checkmark}="val_d1_q1"
!{(2,1) }*+{\textcircled{\footnotesize 9}\checkmark}="val_d2_q1"
!{(3,-1) }*+{\textcircled{\tiny 15}\times}="inv_d5_q2"
!{(13,2) }*+{\textcircled{\tiny 19}\times_{(\supset\md_2)}}="sup_d2"
!{(15,2) }*+{\textcircled{\tiny 22}\checkmark_{(\md_7)}}="d7"
!{(17,2) }*+{\textcircled{\tiny 12}\times_{(\supset\md_3)}}="sup_d3"
!{(7,2) }*+{\textcircled{\tiny 28}\times_{(\supset\md_5)}}="sup_d5_1"
%!{(12,2) }*+{\textcircled{\tiny }\checkmark_{(\md_9)}}="d9"
!{(11,2) }*+{\textcircled{\tiny 24}\times_{(\supset\md_3)}}="sup_d3_2"
!{(3,-2) }*+{\textcircled{\tiny 25}\times_{(\supset\md_5)}}="sup_d5_2"
!{(5,-2) }*+{\textcircled{\tiny 26}\times_{(dup)}}="dup2"
!{(7,-2) }*+{\textcircled{\tiny 23}\checkmark_{(\md_8)}}="d8"
!{(9,-2) }*+{\textcircled{\tiny 27}\checkmark_{(\md_{9})}}="d9"
!{(9,2) }*+{\textcircled{\tiny 29}\checkmark_{(\md_{10})}}="d10"
%d=4
!{(14,-2) }*+{\textcircled{\tiny 11}\times}="inv_d4_q1"
!{(15,1) }*+{\textcircled{\tiny 22}\times}="inv_d7_q1"
!{(1,0) }*+{\textcircled{\tiny 15}\checkmark}="val_d1_q2"
!{(2,0) }*+{\textcircled{\tiny 15}\times}="inv_d2_q2"
!{(16,-2) }*+{\textcircled{\tiny 21}\times}="inv_d6_q1"
!{(7,-3) }*+{\textcircled{\tiny 23}\times}="inv_d8_q2"
!{(9,-3) }*+{\textcircled{\tiny 27}\times}="inv_d9_q1"
!{(9,1) }*+{\textcircled{\tiny 29}\times_{(\md_{10})}}="inv_d10_q1"
%d0->d1
"c1c":"c2c"_{1}^(0.65){0.41}
"c1c":"c3c"_{2}^(0.9){0.25}
"c1c":"c2r"_{5}^(0.9){0.25}
%d1->d2
"c2c":"dup1"_{2}_(0.8){0.09}
"c2c":"d1"_{4}^(0.75){0.28}
"c2c":"d2"^{6}^(0.85){0.27}
"c3c":"d5"_{1}^(0.7){0.09}
"c3c":"c4r"_{3}^(0.7){0.07}
"c3c":"c4c"_{4}^(0.84){0.18}
"c2r":"c3r1"_{2}^(0.7){0.06}
"c2r":"d3"_{4}^(0.6){0.18}
"c2r":"c3r"_(0.4){6}^(0.75){0.17}
%d2->d3
"c4c":"sup_d5"_{1}^(0.68){0.06}
"c4c":"sup_d3_1"^{5}_(0.85){0.04}
"c4c":"d6"^{8}^(0.8){0.04}
"c4c":"d4"_{6}^(0.87){0.11}
"d3":@2{->}"inv_d3_q1"^{Q_1}
"d1":@2{->}"val_d1_q1"^{Q_1}
"d2":@2{->}"val_d2_q1"^{Q_1}
"c3r":"sup_d2"_{1}^(0.6){0.06}
"c3r":"d7"_{3}^(0.75){0.04}
"c3r":"sup_d3"_{4}^(0.75){0.11}
"c4r":"sup_d5_2"_{1}^(0.75){0.02}
"c4r":"dup2"_{5}^(0.8){0.02}
"c4r":"d8"_{6}^(0.85){0.04}
"c4r":"d9"_{8}^(0.88){0.02}
"c3r1":"sup_d3_2"_{4}^(0.75){0.04}
"c3r1":"sup_d5_1"_{1}^(0.6){0.02}
"c3r1":"d10"_{3}^(0.75){0.02}
%d3->d4
"d4":@2{->}"inv_d4_q1"^{Q_1}
"d5":@2{->}"inv_d5_q2"^{Q_2}
"val_d1_q1":@2{->}"val_d1_q2"^{Q_2}
"val_d2_q1":@2{->}"inv_d2_q2"^{Q_2}
"d6":@2{->}"inv_d6_q1"^{Q_1}
"d8":@2{->}"inv_d8_q2"^{Q_2}
"d7":@2{->}"inv_d7_q1"^{Q_1}
"d9":@2{->}"inv_d9_q1"^{Q_1}
"d10":@2{->}"inv_d10_q1"^{Q_1}
}
\vspace{5pt}
\begin{center}
\small Iteration 3
\end{center}
\end{minipage}
\begin{minipage}[c]{20pt}
$\Bigg>$ 
\end{minipage}

\vspace{10pt}
\begin{minipage}[c]{0.95\textwidth}
\small  
\begin{tabular}{l}                        
    $\mD_{\checkmark} \cup \mD_{calc} = \setof{\md_1} \cup \emptyset = \setof{\md_1}$, 
		$\quad \Queue = []$, 
		$\quad \mC_{calc} = \setof{\tuple{1,2,5},\tuple{2,4,6},\tuple{1,3,4},\tuple{1,5,6,8}}$, \\
		$\mD_{\times} = \setof{\md_3,\md_4,\md_2,\md_5,\md_6,\md_7,\md_8,\md_9,\md_{10}} = \setof{[5,4],[2,4,6],[1,6],[2,1],[2,4,8],[5,6,3],[2,3,6],[2,3,8],[5,2,3]}$, \vspace{-8pt} \\
    \hspace{-8pt}\hdashrule{0.95\textwidth}{0.5pt}{2mm}\vspace{-4pt} \\
		%$\rightarrow\quad 
		$p_{\mD}(\md_1) = 1  \quad\Rightarrow\quad$ return the solution KB $(\mo\setminus\md_1) \cup p_1 \quad$ ($p_1$: cf.\ Table~\ref{tab:example3}) $\qed$
\end{tabular}
\end{minipage}
\caption[(Example~\ref{example:staticHS_complex_example_using_tabExDpi3} continued) Solving the Problem of Interactive Static KB Debugging]{(Example~\ref{example:staticHS_complex_example_using_tabExDpi3} continued) Solving the problem of Interactive Static KB Debugging (Problem Definition~\ref{prob_def:static}) for the example DPI given by Table~\ref{tab:example3} by means of \newline Algorithm~\ref{algo:inter_onto_debug} and \textsc{staticHS}.} 
\label{fig:example:inter_onto_debug_staticHS_TabExDpi3_continued}
\end{sidewaysfigure*}
\restoregeometry

%%%%%%%%%%%%%%%%%%%%%%%%%%%%%%%%%%%%%%%%%%%%%%%%%%%%%%%%%%%%%%%%%%%%%%%%%%%%%%%%%%
%%%%%%%%%%%%%%%%%%%%%%%%%%%%%%%%%%%%%%%%%%%%%%%%%%%%%%%%%%%%%%%%%%%%%%%%%%%%%%%%%%
%%%%%%%%%%%%%%%%%%%%%%%%%%%%%%%%%%%%%%%%%%%%%%%%%%%%%%%%%%%%%%%%%%%%%%%%%%%%%%%%%%
%%%%%%%%%%%%%%%%%%%%%%%%%%%%%%%%%%%%%%%%%%%%%%%%%%%%%%%%%%%%%%%%%%%%%%%%%%%%%%%%%%
%%%%%%%%%%%%%%%%%%%%%%%%%%%%%%%     END                  %%%%%%%%%%%%%%%%%%%%%%%%%
%%%%%%%%%%%%%%%%%%%%%%%%%%%%%%%%%%%%%%%%%%%%%%%%%%%%%%%%%%%%%%%%%%%%%%%%%%%%%%%%%%
%%%%%%%%%%%%%%%%%%%%%%%%%%%%%%%%%%%%%%%%%%%%%%%%%%%%%%%%%%%%%%%%%%%%%%%%%%%%%%%%%%
%%%%%%%%%%%%%%%%%%%%%%%%%%%%%%%%%%%%%%%%%%%%%%%%%%%%%%%%%%%%%%%%%%%%%%%%%%%%%%%%%%
%%%%%%%%%%%%%%%%%%%%%%%%%%%%%%%%%%%%%%%%%%%%%%%%%%%%%%%%%%%%%%%%%%%%%%%%%%%%%%%%%%
%%%%%%%%%%%%%%%%%%%%%%%%%%%%%%%%%%%%%%%%%%%%%%%%%%%%%%%%%%%%%%%%%%%%%%%%%%%%%%%%%%

\section[Correctness]{Correctness of the Algorithm%
\sectionmark{Algorithm Correctness}}
\sectionmark{Algorithm Correctness}
\label{sec:CorrectnessOfTextscStaticHS}
%%%%%
%\section{\textsc{staticHS}: Correctness}
%\label{sec:CorrectnessOfTextscStaticHS}
%%%%%
In this section we will demonstrate the correctness of \textsc{staticHS}. That is, we will prove that \textsc{staticHS}, given the inputs described in Algorithm~\ref{algo:inter_stat_hs}, yields the outputs enumerated in Algorithm~\ref{algo:inter_stat_hs}. Used in Algorithm~\ref{algo:inter_onto_debug} to iteratively compute a set of leading diagnoses for query generation, \textsc{staticHS} in this way serves to solve the problem of Interactive Static KB Debugging approximately (parameter $\sigma > 0$ in Algorithm~\ref{algo:inter_onto_debug}) or exactly ($\sigma = 0$).

After each call to \textsc{staticHS} during Algorithm~\ref{algo:inter_onto_debug}, the hitting set tree produced by \textsc{staticHS} is a (partial) wpHS-tree w.r.t.\ the DPI $\langle\mo,\mb,\Tp,\Tn\rangle_\RQ$ given as an input to Algorithm~\ref{algo:inter_onto_debug} and $p_{nodes}()$ which can be directly obtained from the function $p()$ given as input to \textsc{staticHS}. This proposition is made by Lemma~\ref{lem:static_hs_each_call}.

In order to be able to prove this proposition, we formulate and prove two lemmata, Lemma~\ref{lem:static_hs_first_call} and \ref{lem:static_hs_second_or_later_call}. The former, which is given next, shows that this proposition holds for the very first call of \textsc{staticHS} during the execution of Algorithm~\ref{algo:inter_onto_debug}. The latter assures that this proposition holds for any further call of \textsc{staticHS} during Algorithm~\ref{algo:inter_onto_debug} for an adequate set of input parameters to \textsc{staticHS}. Finally, Lemma~\ref{lem:static_hs_each_call} exploits these results to ascertain that this proposition is satisfied for all calls of \textsc{staticHS}.
%The only difference may be that there are some nodes that would be marked as $valid$ in the standard pruned HS-tree -- e.g.\ resulting from application of Algorithm~\ref{algo:hs} given the input DPI -- which might by not in $\mD_{calc}$, but in $\mD_{\times}$ instead. This just means that the respective diagnosis has been ruled out by some new test case in $\Tp'$ or $\Tn'$.
%XXXXXXXXXXXXXXXXXXXXXXXXXXXXXXXXXXXXXXXXXXXXXXXXXXXXXXXXXXXXXXXXXXXXXXXXXXXXXXXXXXXXXXXXXXXXXXXXXXXXXXXXXXXXXXXXXXXXXXXXXXXXXXXXXXXXXXXXXXXXXXXXXXXXXXXXXXXXXXXXXXXXXXXXXXXXXXXXXXXXXXXXX
\begin{lemma}\label{lem:static_hs_first_call}
Let the following be the input parameters to the \textsc{staticHS} function:
\begin{itemize}
\item $\langle\mo,\mb,\Tp,\Tn\rangle_\RQ$ is the DPI given as input to Algorithm~\ref{algo:inter_onto_debug},
% and to \textsc{staticHS},
\item $n_{\min},n_{\max},t\in\mathbb{N}$ where $n_{\min}\geq 2$,
% are the parameters for leading diagnoses computation,
\item a function $p:\mo\rightarrow (0,0.5)$,
\item $\Queue = [\emptyset]$, 
\item $\Tp'=\Tn'=\mD_{\times}=\mD_{\checkmark}=\mC_{calc}=\emptyset$.
\end{itemize}
%Let $p_{nodes}()$ be defined by means of Formula~\ref{eq:path_prob_calc} and $p(\tax),\tax\in\mo$. 
Then, \textsc{staticHS} creates a (partial) wpHS-tree $T$ w.r.t.\ $\langle\mo,\mb,\Tp,\Tn\rangle_\RQ$ and $p_{nodes}()$ (cf.\ Definition~\ref{def:p_node()}) equivalent to one produced by Algorithm~\ref{algo:hs} with input parameters $\langle\mo,\mb,\Tp,\Tn\rangle_\RQ$, $n_{\min}$, $n_{\max}$, $t$ and $p()$ and returns $\tuple{\mD,\Queue,\mathbf{C}_{calc},\mD_{\times}}$ where
%\begin{enumerate}[(1)]
%\item 
$\tuple{\mD\cup\mD_{\times},\Queue,\mathbf{C}_{calc}}$ is the relevant data of $T$. 
%and
%\item $\mD_{\times} = \emptyset$.
%\end{enumerate}
\end{lemma}
\begin{proof}
Since all input parameters $\Tp'$, $\Tn'$, $\mD_{\times}$, $\mD_{\checkmark}$ and $\mC_{calc}$ are equal to the empty set, $\mD_{calc} = \emptyset$ and $\Queue$ includes only the node $\emptyset$, we might regard $\tuple{\mD_{calc},\Queue,\mC_{calc}}$ as the initial relevant data of some (partial) wpHS-tree which includes only an unlabeled root node. The root node $\emptyset$ cannot be labeled as otherwise it would be necessarily an element of $\mD_{calc}$ if $\emptyset$ is a diagnosis w.r.t.\ $\langle\mo,\mb,\Tp,\Tn\rangle_\RQ$ or the set $\mC_{calc}$ would include the conflict set that labels the root node. 

$\mD_{\times}$ can never be extended during the execution of \textsc{staticHS} since line~\ref{algoline:static:add_to_Dtimes} can never be reached. This holds because the test made in line~\ref{algoline:static:isOntValid} can never be negative. Namely, as $\Tp'=\Tn'=\emptyset$, this test actually checks whether $\mo \setminus \mathsf{node}$ is valid w.r.t.\ $\langle\cdot,\mb,\Tp,\Tn\rangle_\RQ$. Due to the fact that $L=valid$ has been output as a label for $\mathsf{node}$ (line~\ref{algoline:static:if_L_valid}) by the \textsc{sLabel} function called in line~\ref{algoline:static:slabel}, it must hold that $\scQX(\langle\mo\setminus\mathsf{node},\mb,\Tp,\Tn\rangle_\RQ)$ yielded 'no conflict'. By Proposition~\ref{prop:qx_correctness}, this implies that $\mo \setminus \mathsf{node}$ is valid w.r.t.\ $\langle\cdot,\mb,\Tp,\Tn\rangle_\RQ$. Thence, $\mD_{\times} = \emptyset$ definitely holds whenever \textsc{staticHS} terminates. 

Moreover, each node with the label $valid$ is added to $\mD_{calc}$ since line~\ref{algoline:static:add_to_Dtimes} can never be reached. As a consequence, with the given input parameters, the execution of the code between line~2 and line~\ref{algoline:static:until} of Algorithm~\ref{algo:inter_stat_hs} has exactly the same effect as executing the code between line~2 and line~\ref{algoline:hs:until} of Algorithm~\ref{algo:hs}. 

$\mD_{\checkmark}$ can never be extended as there is no such modification operation at all in \textsc{staticHS}. Thus, $\mD_{\checkmark} = \emptyset$ holds throughout the execution of \textsc{staticHS}.

Now, the \textsc{sLabel} procedure is equivalent to the \textsc{label} procedure of Algorithm~\ref{algo:hs}, except for the first line of the non-minimality criterion. That is, in \textsc{staticHS} (line~\ref{algoline:slabel:non_min_crit_start}) some $\mathsf{nd}$ is searched for in $\mD_{(\times,\checkmark,calc)}$ whereas in Algorithm~\ref{algo:hs} (line~\ref{algoline:hs:non_min_crit_start}) such $\mathsf{nd}$ is searched in $\mD_{calc}$. However, we point out that $\mD_{(\times,\checkmark,calc)}$ in the \textsc{sLabel} procedure corresponds to the set $\mD_{\times} \cup \mD_{\checkmark} \cup \mD_{calc}$ in \textsc{staticHS} (cf.\ the call to \textsc{sLabel} in line~\ref{algoline:static:slabel}), where $\mD_{\checkmark} = \mD_{\times} = \emptyset$ is an invariant, as argued above. %On the other hand, $\mD_{\times}$ in \textsc{sLabel} is equal to $\mD_{\times}$ in \textsc{staticHS}, where $\mD_{\times} = \emptyset$ is an invariant, as argued above. 
Taking these arguments into account, we have that $\mD_{(\times,\checkmark,calc)}$ in \textsc{sLabel} in line~\ref{algoline:slabel:non_min_crit_start} is equal to $\mD_{calc}$, just as in Algorithm~\ref{algo:hs}.

Hence, with the given input parameters, we have verified that \textsc{staticHS} acts equivalently to Algorithm~\ref{algo:hs}. As Algorithm~\ref{algo:hs} produces a (partial) wpHS-tree $T$ w.r.t.\ the input DPI $\langle\mo,\mb,\Tp,\Tn\rangle_\RQ$ and $p_{nodes}()$ by Lemma~\ref{lem:algo_hs_produces_weighted_pruned_hs_tree}, we infer that \textsc{staticHS} also does so. 

As opposed to Algorithm~\ref{algo:hs} which returns only $\mD_{calc}$, \textsc{staticHS} returns $\tuple{\mD,\Queue,\mathbf{C}_{calc},\mD_{\times}}$ where $\mD := \mD_{calc} \cup \mD_{\checkmark} = \mD_{calc}$ since $\mD_{\checkmark} = \emptyset$, as argued above. In that, $\mD_{calc}$, $\Queue$ and $\mC_{calc}$ correspond exactly to the equally named collections in Algorithm~\ref{algo:hs} and $\mD_{\times} = \emptyset$, as argued above.
Therefore, by Corollary~\ref{cor:algo_hs_returns_relevant_data_of_weighted_(partial)_pruned_hs_tree}, $\tuple{\mD\cup\mD_{\times},\Queue,\mathbf{C}_{calc}} = \tuple{\mD_{calc},\Queue,\mathbf{C}_{calc}}$ is the relevant data of the (partial) wpHS-tree $T$ w.r.t.\ $\langle\mo,\mb,\Tp,\Tn\rangle_\RQ$ and $p_{nodes}()$ produced by Algorithm~\ref{algo:hs}. 
\end{proof}
The next lemma manifests that \textsc{staticHS}, given such parameters that $\tuple{\mD_{\times}\cup\mD_{\checkmark},\Queue,\mC_{calc}}$ is the relevant data of a (partial) wpHS-tree w.r.t.\ $\langle\mo,\mb,\Tp,\Tn\rangle_\RQ$ and $p_{nodes}()$, again yields a (partial) wpHS-tree w.r.t.\ $\langle\mo,\mb,\Tp,\Tn\rangle_\RQ$ and $p_{nodes}()$. 
\begin{lemma}\label{lem:static_hs_second_or_later_call}
Let the following be the input parameters to the \textsc{staticHS} function:
\begin{itemize}
\item $\langle\mo,\mb,\Tp,\Tn\rangle_\RQ$ is the DPI given as input to Algorithm~\ref{algo:inter_onto_debug},
% and to \textsc{staticHS},
\item $\Tp'$ is the set of positive and $\Tn'$ is the set of negative test cases specified since the start of Algorithm~\ref{algo:inter_onto_debug} where $\Tp'\cup\Tn' \supset\emptyset$,
\item $n_{\min},n_{\max},t\in\mathbb{N}$ where $n_{\min}\geq 2$, 
%are the parameters for leading diagnoses computation,
\item a function $p:\mo\rightarrow (0,0.5)$,
\item $\mD_{\times} \neq \emptyset$, $\mD_{\checkmark}\neq \emptyset$, $\mC_{calc}\neq \emptyset$ and $\Queue$ such that $\tuple{\mD_{\times}\cup\mD_{\checkmark},\Queue,\mC_{calc}}$ is the relevant data of a (partial) wpHS-tree w.r.t.\ $\langle\mo,\mb,\Tp,\Tn\rangle_\RQ$ and $p_{nodes}()$ produced by Algorithm~\ref{algo:hs} with input parameters $\langle\mo,\mb,\Tp,\Tn\rangle_\RQ$ and $p()$.
\end{itemize}
%Let $p_{nodes}()$ be defined by means of Formula~\ref{eq:path_prob_calc} and $p(\tax),\tax\in\mo$.
Then, \textsc{staticHS} creates a (partial) wpHS-tree $T$ w.r.t.\ $\langle\mo,\mb,\Tp,\Tn\rangle_\RQ$ and $p_{nodes}()$ equivalent to one produced by Algorithm~\ref{algo:hs} with input parameters $\langle\mo,\mb,\Tp,\Tn\rangle_\RQ$ and $p()$ 
and returns $\tuple{\mD,\Queue,\mathbf{C}_{calc},\mD_{\times}}$ where
%\begin{enumerate}[(1)]
%\item 
$\tuple{\mD\cup\mD_{\times},\Queue,\mathbf{C}_{calc}}$ is the relevant data of $T$. 
%and
%\item $\mD_{\times} = \emptyset$.
%\end{enumerate}
\end{lemma}
\begin{proof}
Since $\tuple{\mD_{\times}\cup\mD_{\checkmark},\Queue,\mC_{calc}}$ is the relevant data of a (partial) wpHS-tree $T$ w.r.t.\ $\langle\mo,\mb,\Tp,\Tn\rangle_\RQ$ and $p_{nodes}()$ produced by Algorithm~\ref{algo:hs} with input parameters $\langle\mo,\mb,\Tp,\Tn\rangle_\RQ$ and $p()$, it is clear that, if the construction of $T$ is continued by an algorithm working equivalently to Algorithm~\ref{algo:hs} and using this relevant data, the relevant data of a (partial) wpHS-tree $T'$ w.r.t.\ $\langle\mo,\mb,\Tp,\Tn\rangle_\RQ$ and $p_{nodes}()$ will be stored by this algorithm (Corollary~\ref{cor:algo_hs_returns_relevant_data_of_weighted_(partial)_pruned_hs_tree}). Therefore, we show that \textsc{staticHS} is such an algorithm. 

In Algorithm~\ref{algo:hs}, the set of all already computed minimal diagnoses w.r.t.\ $\langle\mo,\mb,\Tp,\Tn\rangle_\RQ$ is denoted by $\mD_{calc}$. 
%By Proposition~\ref{prop:hs_prob_correct}, 
%By Corollary~\ref{cor:hs_tree_finds_most-prob_diags_first}, 
%these are exactly the $|\mD_{calc}|$ most probable minimal diagnoses w.r.t.\ $\langle\mo,\mb,\Tp,\Tn\rangle_\RQ$ and $p_{nodes}()$. 
%Additionally, $\mD_{calc}$ includes only minimal diagnoses w.r.t.\ $\langle\mo,\mb,\Tp,\Tn\rangle_\RQ$. 
Nodes labeled by $valid$ are added to $\mD_{calc}$ (line~\ref{algoline:hs:update_Dcalc}) and $\mD_{calc}$ is used in the non-minimality criterion in the \textsc{label} function (line~\ref{algoline:hs:non_min_crit_start}). If Algorithm~\ref{algo:hs} should be used to continue construction of $T$ using the relevant data $\tuple{\mD_{\times}\cup\mD_{\checkmark},\Queue,\mC_{calc}}$, the required setting is just to use $\mD_{calc} := \mD_{\times}\cup\mD_{\checkmark}$ and use $\Queue$ and $\mC_{calc}$ for the equally named variables in Algorithm~\ref{algo:hs}. If then a new node $\mathsf{nd}$ labeled by $valid$ were added to $\mD_{calc}$, we would have that $\mD_{calc} := \mD_{\times}\cup\mD_{\checkmark} \cup \setof{\mathsf{nd}}$. By Corollary~\ref{cor:hs_tree_finds_most-prob_diags_first}, this set $\mD_{calc}$ used by Algorithm~\ref{algo:hs} would at each point in time comprise exactly the $|\mD_{calc}|$ most probable minimal diagnoses w.r.t.\ $\langle\mo,\mb,\Tp,\Tn\rangle_\RQ$ and $p_{nodes}()$.

In \textsc{staticHS}, each node $\mathsf{node}$ labeled by $valid$ is added either to $\mD_{calc}$, which is initially the empty set in \textsc{staticHS}, or to $\mD_{\times}$ (lines~\ref{algoline:static:add_to_Dcalc} and \ref{algoline:static:add_to_Dtimes}), i.e.\ $\mathsf{node}$ is added to $\mD_{calc}\cup\mD_{\times}$. Thus, it is also true to say that $\mathsf{node}$ is added to $\mD_{calc}\cup\mD_{\times}\cup\mD_{\checkmark}$. So, the first new node $\mathsf{nd}$ labeled by $valid$ is added to this set which is then equal to $\mD_{\times}\cup\mD_{\checkmark}\cup\setof{\mathsf{nd}}$. This set is equal to the set $\mD_{calc}$ that would be used by Algorithm~\ref{algo:hs} to further construct the (partial) wpHS-tree $T$. 

In the non-minimality criterion in function \textsc{sLabel}, $\mD_{\times,\checkmark,calc}$ is used which is equal to the set $\mD_{calc}\cup\mD_{\times}\cup\mD_{\checkmark}$ in \textsc{staticHS} (cf.\ the call to \textsc{sLabel} in line~\ref{algoline:static:slabel}). Hence, $\mD_{calc}\cup\mD_{\times}\cup\mD_{\checkmark}$ is used and modified in \textsc{staticHS} in exactly the same way as $\mD_{calc}$ is used and modified in Algorithm~\ref{algo:hs}. 

Apart from this, as can be easily verified, the labeling function \textsc{sLabel} in \textsc{staticHS} is identical to \textsc{label} in Algorithm~\ref{algo:hs} and the way $\Queue$ and $\mC_{calc}$ are used and modified in \textsc{staticHS} is exactly equivalent to the way these are used and modified in Algorithm~\ref{algo:hs}.  
%, as argued in the proof of Lemma~\ref{lem:static_hs_first_call}.

What remains to be shown is that $\mD_{calc}\cup\mD_{\times}\cup\mD_{\checkmark}$, as $\mD_{calc}$ in Algorithm~\ref{algo:hs}, always contains all already computed minimal diagnoses w.r.t.\ $\langle\mo,\mb,\Tp,\Tn\rangle_\RQ$ which are the $|\mD_{calc}\cup\mD_{\times}\cup\mD_{\checkmark}|$ most probable minimal diagnoses w.r.t.\ $\langle\mo,\mb,\Tp,\Tn\rangle_\RQ$. 

Since $\mD_{\times}\cup\mD_{\checkmark}$ is the first set in the relevant data of a (partial) wpHS-tree $T$ w.r.t.\ $\langle\mo,\mb,\Tp,\Tn\rangle_\RQ$ and $p_{nodes}()$ produced by Algorithm~\ref{algo:hs} with input parameters $\langle\mo,\mb,\Tp,\Tn\rangle_\RQ$ and $p()$, by 
%Proposition~\ref{prop:hs_prob_correct} and 
Corollaries~\ref{cor:algo_hs_returns_relevant_data_of_weighted_(partial)_pruned_hs_tree} and \ref{cor:hs_tree_finds_most-prob_diags_first}, it must be valid that $\mD_{\times}\cup\mD_{\checkmark}$ comprises the $|\mD_{\times}\cup\mD_{\checkmark}|$ most probable minimal diagnoses w.r.t.\ $\langle\mo,\mb,\Tp,\Tn\rangle_\RQ$. Since $\mD_{calc}$ is initially defined to be the empty set in \textsc{staticHS}, it is also true to say that $\mD_{\times}\cup\mD_{\checkmark}\cup\mD_{calc}$ comprises the $|\mD_{\times}\cup\mD_{\checkmark}\cup\mD_{calc}|$ most probable minimal diagnoses w.r.t.\ $\langle\mo,\mb,\Tp,\Tn\rangle_\RQ$ when \textsc{staticHS} starts executing. Since, by assumption, the same $p()$ is used by \textsc{staticHS} as was used for the construction of the (partial) wpHS-tree $T$ so far, the same ordering of $\Queue$ is used by \textsc{staticHS} as would be used by Algorithm~\ref{algo:hs} to further construct the (partial) wpHS-tree $T$. Therefore, $\mD_{\times}\cup\mD_{\checkmark}\cup\mD_{calc}$ must indeed comprise the $|\mD_{\times}\cup\mD_{\checkmark}\cup\mD_{calc}|$ most probable minimal diagnoses w.r.t.\ $\langle\mo,\mb,\Tp,\Tn\rangle_\RQ$ at each point in time.

The set $\mD$ in the tuple $\tuple{\mD,\Queue,\mathbf{C}_{calc},\mD_{\times}}$ returned by \textsc{staticHS} corresponds exactly to $\mD_{calc} \cup \mD_{\checkmark}$. So, $\mD \cup \mD_{\times} = \mD_{\times}\cup\mD_{\checkmark}\cup\mD_{calc}$.

To summarize, \textsc{staticHS} acts exactly equivalently to Algorithm~\ref{algo:hs}. As a consequence, Corollary~\ref{cor:algo_hs_returns_relevant_data_of_weighted_(partial)_pruned_hs_tree} regarding Algorithm~\ref{algo:hs} applies to \textsc{staticHS} as well. This means that the tuple consisting of the set of nodes labeled by $valid$, i.e.\ $\mD_{\times}\cup\mD_{\checkmark}\cup\mD_{calc}$, the list of open nodes $\Queue$ and the set of minimal conflict sets w.r.t.\ $\langle\mo,\mb,\Tp,\Tn\rangle_\RQ$ in \textsc{staticHS} store the relevant data of a (partial) wpHS-tree $T$ as it could have been generated by Algorithm~\ref{algo:hs}. This completes the proof.
\end{proof}

\begin{lemma}\label{lem:static_hs_each_call}
Any call to \textsc{staticHS} within Algorithm~\ref{algo:inter_onto_debug} yields an output $\tuple{\mD,\Queue,\mC_{calc},\mD_{\times}}$ where 
\begin{itemize}
	\item $\tuple{\mD\cup\mD_{\times},\Queue,\mathbf{C}_{calc}}$ is the relevant data of $T$ and
	\item $T$ is a (partial) wpHS-tree w.r.t.\ $\langle\mo,\mb,\Tp,\Tn\rangle_\RQ$ and $p_{nodes}()$ equivalent to one produced by Algorithm~\ref{algo:hs} with input parameters $\langle\mo,\mb,\Tp,\Tn\rangle_\RQ$ and $p()$.
\end{itemize}
%
%Any call to \textsc{staticHS} within Algorithm~\ref{algo:inter_onto_debug} yields an output $\tuple{\mD,\Queue,\mathbf{C}_{calc},\mD_{\times}}$ where $\tuple{\mD\cup\mD_{\times},\Queue,\mathbf{C}_{calc}}$ is the relevant data of $T$ where $T$ is a (partial) weighted pruned HS-tree w.r.t.\ $\langle\mo,\mb,\Tp,\Tn\rangle_\RQ$ and $p_{nodes}()$ equivalent to one produced by Algorithm~\ref{algo:hs} with input parameters $\langle\mo,\mb,\Tp,\Tn\rangle_\RQ$ and $p()$.
\end{lemma}
\begin{proof}
As can be easily verified, the arguments given to \textsc{staticHS} at the first time it is called throughout the execution of Algorithm~\ref{algo:inter_onto_debug} correspond exactly to the input parameters to \textsc{staticHS} assumed in Lemma~\ref{lem:static_hs_first_call} (cf.\ the variable instantiations in lines~\ref{algoline:inter_onto_debug:var_inst_start}-\ref{algoline:inter_onto_debug:var_inst_end} of Algorithm~\ref{algo:inter_onto_debug}). Thus, by Lemma~\ref{lem:static_hs_first_call}, we conclude that the first call to \textsc{staticHS} during the runtime of Algorithm~\ref{algo:inter_onto_debug} yields the output $\tuple{\mD,\Queue,\mathbf{C}_{calc},\mD_{\times}}$ where
$\tuple{\mD\cup\mD_{\times},\Queue,\mathbf{C}_{calc}}$ is the relevant data of $T$ and $T$ is a (partial) wpHS-tree w.r.t.\ $\langle\mo,\mb,\Tp,\Tn\rangle_\RQ$ and $p_{nodes}()$ equivalent to one produced by Algorithm~\ref{algo:hs} with input parameters $\langle\mo,\mb,\Tp,\Tn\rangle_\RQ$ and $p()$.

When this first call to \textsc{staticHS} returns in Algorithm~\ref{algo:inter_onto_debug}, $\mD$
%, which corresponds to the set $\mD_{calc} \cup \mD_{\checkmark}$ within \textsc{staticHS}, 
is renamed to become $\mD_{\checkmark}$ in Algorithm~\ref{algo:inter_onto_debug} (line~\ref{algoline:inter_onto_debug:staticHS}). $\Queue$, $\mathbf{C}_{calc}$ and $\mD_{\times}$ bear unmodified names within Algorithm~\ref{algo:inter_onto_debug}. We point out that $\Queue$ and $\mathbf{C}_{calc}$ are not modified anywhere in Algorithm~\ref{algo:inter_onto_debug}. $\mD_{\checkmark}$ and $\mD_{\times}$ are modified only in lines~\ref{algoline:inter_onto_debug:update_D_checkmark} and \ref{algoline:inter_onto_debug:update_D_times}. In these lines, a subset $\mD_{out}$ of $\mD_{\checkmark}$ is deleted from $\mD_{\checkmark}$ and added to $\mD_{\times}$.

$\mD_{out}$ must be a subset of $\mD_{\checkmark}$. This holds, first, because $\tuple{Q,\Pt(Q)}$ is a query $Q$ w.r.t.\ the leading diagnoses $\mD_{\checkmark}$ and the DPI $\langle\mo,\mb,\Tp\cup\Tp',\Tn\cup\Tn'\rangle_\RQ$ together with its q-partition $\Pt(Q)$ (\textsc{calcQuery} in line~\ref{algoline:inter_onto_debug:calc_query}, cf.\ Section~\ref{sec:DetailedAlgorithmDescription}). Second, $\mD_{out}$ corresponds either to $\dx{}(Q)$ (if the answer $u(Q) = \false$) or to $\dnx{}(Q)$ (if the answer $u(Q) = \true$) where both sets must be subsets of the set of leading diagnoses $\mD_{\checkmark}$ by Definition~\ref{def:q-partition} (\textsc{getInvalidDiags} in line~\ref{algoline:inter_onto_debug:get_invalid_diags}, cf.\ Section~\ref{sec:DetailedAlgorithmDescription}).

Hence, $\mD_{\checkmark} \cup \mD_{\times}$ remains unchanged throughout Algorithm~\ref{algo:inter_onto_debug}. 
By the renaming of $\mD$ to become $\mD_{\checkmark}$ in Algorithm~\ref{algo:inter_onto_debug} (see the argumentation above), $\mD_{\checkmark} \cup \mD_{\times}$ is equal to the set $\mD \cup \mD_{\times}$ where $\tuple{\mD,\Queue,\mathbf{C}_{calc},\mD_{\times}}$ is the output of the first call to \textsc{staticHS} in Algorithm~\ref{algo:inter_onto_debug}. Therefore, the relevant data $\tuple{\mD\cup\mD_{\times},\Queue,\mathbf{C}_{calc}}$ of $T$ is unmodified until the second call to \textsc{staticHS} within Algorithm~\ref{algo:inter_onto_debug} is made.

So, we have that the arguments given to \textsc{staticHS} at the second time it is called throughout the execution of Algorithm~\ref{algo:inter_onto_debug} correspond exactly to the input parameters to \textsc{staticHS} assumed in Lemma~\ref{lem:static_hs_second_or_later_call}. Notice that the probability measure $p_{\mo}()$ which corresponds to the probability measure $p()$ in \textsc{staticHS} is never changed throughout the while-loop in Algorithm~\ref{algo:inter_onto_debug} (cf.\ Section~\ref{sec:DetailedAlgorithmDescription}).

Thus, by Lemma~\ref{lem:static_hs_second_or_later_call}, we conclude that the second call to \textsc{staticHS} during the runtime of Algorithm~\ref{algo:inter_onto_debug} yields the output $\tuple{\mD,\Queue,\mathbf{C}_{calc},\mD_{\times}}$ where
$\tuple{\mD\cup\mD_{\times},\Queue,\mathbf{C}_{calc}}$ is the relevant data of $T'$ and $T'$ is a (partial) wpHS-tree w.r.t.\ $\langle\mo,\mb,\Tp,\Tn\rangle_\RQ$ and $p_{nodes}()$ equivalent to one produced by Algorithm~\ref{algo:hs} with input parameters $\langle\mo,\mb,\Tp,\Tn\rangle_\RQ$ and $p()$.

By means of the same line of argument we used so far and further applications of Lemma~\ref{lem:static_hs_second_or_later_call} it can be derived that the proposition of this lemma holds for any call to \textsc{staticHS} throughout Algorithm~\ref{algo:inter_onto_debug}.  
\end{proof}
By means of the just proven Lemma~\ref{lem:static_hs_each_call}, we are now able to show by the next lemma that \textsc{staticHS} computes minimal diagnoses w.r.t.\ the DPI $\langle\mo,\mb,\Tp,\Tn\rangle_\RQ$ given as an input to Algorithm~\ref{algo:inter_onto_debug} in most-probable-first order. Further on, the next lemma will reveal that \emph{only} minimal diagnoses w.r.t.\ the DPI $\langle\mo,\mb,\Tp,\Tn\rangle_\RQ$ are computed by \textsc{staticHS} which assures the soundness of \textsc{staticHS} concerning the (input) DPI $\langle\mo,\mb,\Tp,\Tn\rangle_\RQ$. The soundness of \textsc{staticHS} as regards the (current) DPI $\langle\mo,\mb,\Tp\cup\Tp',\Tn\cup\Tn'\rangle_\RQ$ will be considered in Lemma~\ref{lem:staticHS_Dcalc_includes_diags_wrt_current_DPI} below.
\begin{lemma}\label{lem:static_hs_most_prob_diags}
Any call to \textsc{staticHS} within Algorithm~\ref{algo:inter_onto_debug} yields an output $\tuple{\mD,\Queue,\mathbf{C}_{calc},\mD_{\times}}$ where $\mD\cup\mD_{\times}$ is the set of $|\mD\cup\mD_{\times}|$ most probable 
%(w.r.t.\ $p()$ and Formula~\ref{eq:diag_prob_calc}) 
(w.r.t.\ $p_{nodes}()$) 
minimal diagnoses w.r.t.\ $\langle\mo,\mb,\Tp,\Tn\rangle_\RQ$.
\end{lemma}
\begin{proof}
Let $T$ be the (partial) wpHS-tree $T$ produced by any call to \textsc{staticHS} within Algorithm~\ref{algo:inter_onto_debug}. Then, by Lemma~\ref{lem:static_hs_each_call}, 
\begin{itemize}
	\item $T$ is equal to a (partial) wpHS-tree produced by Algorithm~\ref{algo:hs} with input parameters $\langle\mo,\mb,\Tp,\Tn\rangle_\RQ$ and $p()$ and
	\item the first set $\mD_{calc}$ in the relevant data $\tuple{\mD_{calc},\Queue,\mathbf{C}_{calc}}$ of $T$ produced by Algorithm~\ref{algo:hs} corresponds to $\mD\cup\mD_{\times}$.%, the first set of the relevant data of $T$ output by \textsc{staticHS}.
\end{itemize}
So, by Corollary~\ref{cor:hs_tree_finds_most-prob_diags_first}, the proposition of this lemma follows. 
%
%By Lemma~\ref{lem:static_hs_each_call}, Corollary~\ref{cor:hs_tree_finds_most-prob_diags_first} applies to any call to \textsc{staticHS} within Algorithm~\ref{algo:inter_onto_debug}. 
%Since $\mD_{calc}$ (used in Corollary~\ref{cor:hs_tree_finds_most-prob_diags_first}) is the first set in the relevant data tuple of the (partial) weighted pruned HS-tree $T$ produced by Algorithm~\ref{algo:hs}, $\mD_{calc}$ corresponds to $\mD\cup\mD_{\times}$, the first set in the relevant data tuple of $T$ produced by \textsc{staticHS}. 
%
%corresponds to the set of the $|\mD_{calc}|$ most probable minimal diagnoses w.r.t.\ $\langle\mo,\mb,\Tp,\Tn\rangle_\RQ$ and $\mD\cup\mD_{\times}$ corresponds to the first set of the relevant data of $T$ output by \textsc{staticHS}, 
%
%Since the first set $\mD_{calc}$ in the tuple $\tuple{\mD_{calc},\Queue,\mathbf{C}_{calc}}$ of the relevant data of the (partial) weighted pruned HS-tree $T$ produced by Algorithm~\ref{algo:hs} corresponds to the set of the $|\mD_{calc}|$ most probable minimal diagnoses w.r.t.\ $\langle\mo,\mb,\Tp,\Tn\rangle_\RQ$ and $\mD\cup\mD_{\times}$ corresponds to the first set of the relevant data of $T$ output by \textsc{staticHS}, 
%%From this, 
%the proposition of this lemma follows.
\end{proof}
Moreover, Lemma~\ref{lem:static_hs_each_call} provides the basis for showing the completeness of \textsc{staticHS}. That is, Lemma \ref{lem:static_hs_completeness} will manifest that \emph{all} minimal diagnoses w.r.t.\ the DPI $\langle\mo,\mb,\Tp,\Tn\rangle_\RQ$ given as an input to Algorithm~\ref{algo:inter_onto_debug} will be found by \textsc{staticHS} given that it keeps executing for a sufficiently long period of time.  
\begin{lemma}\label{lem:static_hs_completeness}
Any call to \textsc{staticHS} within Algorithm~\ref{algo:inter_onto_debug} where the execution of \textsc{staticHS} terminates due to $\Queue = []$ yields an output $\tuple{\mD,\Queue,\mathbf{C}_{calc},\mD_{\times}}$ where $\mD\cup\mD_{\times}$ is the set of all minimal diagnoses w.r.t.\ $\langle\mo,\mb,\Tp,\Tn\rangle_\RQ$.
\end{lemma}
\begin{proof}
The proposition of this lemma follows from Lemma~\ref{lem:static_hs_each_call} and Proposition~\ref{prop:hs-tree_completeness} by an analogue argumentation as in the proof of Lemma~\ref{lem:static_hs_most_prob_diags}.
\end{proof}
The following lemma proves that \textsc{staticHS} is sound w.r.t.\ the finding of minimal diagnoses w.r.t.\ the current DPI $\langle\mo,\mb,\Tp \cup \Tp',\Tn \cup \Tn'\rangle_\RQ$, i.e.\ the DPI $\langle\mo,\mb,\Tp,\Tn\rangle_\RQ$ given as an input to Algorithm~\ref{algo:inter_onto_debug} extended by all new positive and negative test cases $\Tp'$ and $\Tn'$, respectively, that have been collected so far. 
\begin{lemma}\label{lem:staticHS_Dcalc_includes_diags_wrt_current_DPI}
%[Soundness of \textsc{staticHS}]
If any call to \textsc{staticHS} adds an element $\md$ to the set $\mD_{calc}$ during the execution of Algorithm~\ref{algo:inter_onto_debug}, $\md$ is a minimal diagnosis w.r.t.\ $\langle\mo,\mb,\Tp \cup \Tp',\Tn \cup \Tn'\rangle_\RQ$.
% where $\langle\mo,\mb,\Tp,\Tn\rangle_\RQ$ is the DPI given as input to Algorithm~\ref{algo:inter_onto_debug} and $\Tp'$ are all positive and $\Tn'$ all negative test cases that have been added to the input DPI during the execution of Algorithm~\ref{algo:inter_onto_debug}.
\end{lemma}
\begin{proof}
By Lemma~\ref{lem:static_hs_most_prob_diags} we know that each node $\mathsf{node}$ that is added to $\mD_{calc}$ by \textsc{staticHS} is a minimal diagnosis w.r.t.\ the input DPI $\langle\mo,\mb,\Tp,\Tn\rangle_\RQ$. Through the test for validity of $\mo \setminus \mathsf{node}$ w.r.t.\ $\langle\cdot,\mb,\Tp \cup \Tp',\Tn \cup \Tn'\rangle_\RQ$ (cf.\ Definition~\ref{def:valid_onto}) which must be successful before $\mathsf{node}$ is added to $\mD_{calc}$ (\textsc{isKBValid} in line~\ref{algoline:static:isOntValid}), we have that $\mathsf{node}$ must also be a diagnosis w.r.t.\ $\langle\mo,\mb,\Tp \cup \Tp',\Tn \cup \Tn'\rangle_\RQ$ by Proposition~\ref{prop:validonto_diag}. Since $\mathsf{node}$ is a minimal diagnosis w.r.t.\ $\langle\mo,\mb,\Tp,\Tn\rangle_\RQ$ as argued and due to Proposition~\ref{prop:adding_testcase_cannot_make_min_diags_shrink} (see page~\pageref{prop:adding_testcase_cannot_make_min_diags_shrink}), there cannot be a minimal diagnosis w.r.t.\ $\langle\mo,\mb,\Tp \cup \Tp',\Tn \cup \Tn'\rangle_\RQ$ which is a proper subset of $\mathsf{node}$. Thence, $\mathsf{node}$ must be a minimal diagnosis w.r.t.\ $\langle\mo,\mb,\Tp \cup \Tp',\Tn \cup \Tn'\rangle_\RQ$.
\end{proof}
We are now in a position to bring to proof that the first set $\mD$ in the tuple output by any call of \textsc{staticHS} in Algorithm~\ref{algo:inter_onto_debug} contains only these minimal diagnoses w.r.t.\ the (input) DPI $\langle\mo,\mb,\Tp,\Tn\rangle_\RQ$ 
%given as input to Algorithm~\ref{algo:inter_onto_debug} 
that are also minimal diagnoses w.r.t.\ the (current) DPI $\langle\mo,\mb,\Tp \cup \Tp',\Tn \cup \Tn'\rangle_\RQ$. In other words, this means that the set of leading diagnoses used for query generation in Algorithm~\ref{algo:inter_onto_debug} consists only of minimal diagnoses w.r.t.\ the input DPI that are in agreement with the additional information given by all query answers so far.   
\begin{lemma}\label{lem:static_hs_soundness}
Any call to \textsc{staticHS} within Algorithm~\ref{algo:inter_onto_debug} yields an output $\tuple{\mD,\Queue,\mathbf{C}_{calc},\mD_{\times}}$ where $\mD \subseteq \minD_{\langle\mo,\mb,\Tp,\Tn\rangle_\RQ} \cap \minD_{\langle\mo,\mb,\Tp \cup \Tp',\Tn \cup \Tn'\rangle_\RQ}$. 
\end{lemma}
\begin{proof}
The output set $\mD$ of any call to \textsc{staticHS} during the execution of Algorithm~\ref{algo:inter_onto_debug} corresponds to the set $\mD_{calc} \cup \mD_{\checkmark}$ in \textsc{staticHS}. As per Lemma~\ref{lem:staticHS_Dcalc_includes_diags_wrt_current_DPI}, $\mD_{calc}$ includes only minimal diagnoses w.r.t.\ $\langle\mo,\mb,\Tp \cup \Tp',\Tn \cup \Tn'\rangle_\RQ$. By Lemma~\ref{lem:static_hs_most_prob_diags}, $\mD_{calc}$ includes only minimal diagnoses w.r.t.\ $\langle\mo,\mb,\Tp,\Tn\rangle_\RQ$. Therefore, we can conclude that $\mD_{calc} \subseteq \minD_{\langle\mo,\mb,\Tp,\Tn\rangle_\RQ} \cap \minD_{\langle\mo,\mb,\Tp \cup \Tp',\Tn \cup \Tn'\rangle_\RQ}$. So, we must show that $\mD_{\checkmark}\subseteq \minD_{\langle\mo,\mb,\Tp,\Tn\rangle_\RQ} \cap \minD_{\langle\mo,\mb,\Tp \cup \Tp',\Tn \cup \Tn'\rangle_\RQ}$ holds when any call to \textsc{staticHS} during the execution of Algorithm~\ref{algo:inter_onto_debug} terminates. We will perform an induction proof.

Base Case: At the first call of \textsc{staticHS} during the execution of Algorithm~\ref{algo:inter_onto_debug}, the argument $\mD_{\checkmark}$ passed to \textsc{staticHS} is the empty set. As argued in the proof of Lemma~\ref{lem:static_hs_first_call}, $\mD_{\checkmark}$ is never modified throughout \textsc{staticHS}. Thus, $\mD_{\checkmark} = \emptyset \subseteq \minD_{\langle\mo,\mb,\Tp,\Tn\rangle_\RQ} \cap \minD_{\langle\mo,\mb,\Tp \cup \Tp',\Tn \cup \Tn'\rangle_\RQ}$ holds for the output of the first call to \textsc{staticHS}. Therefore, the proposition of this lemma holds for the output of the first call of \textsc{staticHS}.

Induction Step: Assume that 
%$\mD_{\checkmark}\subseteq \minD_{\langle\mo,\mb,\Tp,\Tn\rangle_\RQ} \cap \minD_{\langle\mo,\mb,\Tp \cup \Tp',\Tn \cup \Tn'\rangle_\RQ}$ 
the proposition of this lemma holds for the last-but-one call to \textsc{staticHS} during the execution of Algorithm~\ref{algo:inter_onto_debug} (Induction Hypothesis). Consider the last, i.e.\ most recent, call to \textsc{staticHS} during the execution of Algorithm~\ref{algo:inter_onto_debug}.

First, the set $\mD_{\checkmark}$ given as an input argument to \textsc{staticHS} at the last call of \textsc{staticHS} is unmodified throughout the entire execution of \textsc{staticHS}, as already mentioned. Second, $\mD_{\checkmark} = \mD'\setminus\mD_{out} \subseteq \mD'$ holds where $\mD'$ is the output of the last-but-one call of \textsc{staticHS} by Algorithm~\ref{algo:inter_onto_debug} since the only modification to the set $\mD'$ (which is denoted by $\mD_{\checkmark}$ in Algorithm~\ref{algo:inter_onto_debug}) during Algorithm~\ref{algo:inter_onto_debug} is the deletion (line~\ref{algoline:inter_onto_debug:update_D_checkmark}) of exactly those diagnoses $\mD_{out}$ in $\mD'$ that are invalidated by the addition of the most recent test case (\textsc{getInvalidDiags} in line~\ref{algoline:inter_onto_debug:get_invalid_diags}). That is, the input $\mD_{\checkmark}$ to the most recent call to \textsc{staticHS} includes only diagnoses that comply with the most recently added test case. Call the most recently added test case $tc$. By the Induction Hypothesis, $\mD' \subseteq \minD_{\langle\mo,\mb,\Tp,\Tn\rangle_\RQ} \cap \minD_{\langle\mo,\mb,\Tp \cup (\Tp'\setminus\setof{tc}),\Tn \cup (\Tn'\setminus\setof{tc})\rangle_\RQ}$. Notice that either $tc\in\Tp'$ or $tc\in\Tn'$ holds, but not both. 
%where $\Tp'' \cup \Tn''$ corresponds to the set of all test cases $\Tp' \cup \Tn'$ minus the most recently added one.
As $\mD_{\checkmark} \subseteq \mD'$, it must be true that $\mD_{\checkmark} \subseteq \minD_{\langle\mo,\mb,\Tp,\Tn\rangle_\RQ} \cap \minD_{\langle\mo,\mb,\Tp \cup (\Tp'\setminus\setof{tc}),\Tn \cup (\Tn'\setminus\setof{tc})\rangle_\RQ}$ and $\mD_{\checkmark}$ complies with the test case $tc$. Hence, we infer that $\mD_{\checkmark} \subseteq \minD_{\langle\mo,\mb,\Tp,\Tn\rangle_\RQ} \cap \minD_{\langle\mo,\mb,\Tp \cup \Tp',\Tn \cup \Tn'\rangle_\RQ}$. Consequently, the proposition of this lemma must hold for each call of \textsc{staticHS} during the execution of Algorithm~\ref{algo:inter_onto_debug}.
\end{proof}
The results proven so far in this section facilitate the proof of correctness of \textsc{staticHS}:
\begin{proposition}[Correctness of \textsc{staticHS}]\label{prop:static_hs_correctness}
Any call to \textsc{staticHS} (given the inputs described in Algorithm~\ref{algo:inter_stat_hs}) within Algorithm~\ref{algo:inter_onto_debug} terminates and yields an output $\tuple{\mD,\Queue,\mathbf{C}_{calc},\mD_{\times}}$ where 
\begin{enumerate}[(1)]
\item it holds for $\mD$ that 
%is the current set of leading diagnoses such that
\begin{enumerate}[(a)]
\item $\mD \subseteq \minD_{\langle\mo,\mb,\Tp,\Tn\rangle_\RQ} \cap \minD_{\langle\mo,\mb,\Tp\cup\Tp',\Tn\cup\Tn'\rangle_\RQ}$ is the set of most probable 
 minimal diagnoses w.r.t.\ $\langle\mo,\mb,\Tp,\Tn\rangle_\RQ$ that satisfy all test cases $\Tp'$ and $\Tn'$ such that 
\begin{enumerate}[(i)]
\item $n_{\min} \leq |\mD| \leq n_{\max}$ and 
\item $\mD\supset\mD_{\checkmark}$,
\end{enumerate} 
if such a set $\mD$ exists, or 
\item $\mD$ is equal to the set of all minimal diagnoses $\minD_{\langle\mo,\mb,\Tp,\Tn\rangle_\RQ} \cap \minD_{\langle\mo,\mb,\Tp\cup\Tp',\Tn\cup\Tn'\rangle_\RQ}$, otherwise, 
\end{enumerate}
where ``most-probable'' refers to the probability measure $p_{nodes}()$ (cf.\ Definition~\ref{def:p_node()}) obtained from the given function $p()$;
\item $\Queue$ is the current queue of open (non-labeled) nodes of the produced (partial) wpHS-tree, 
\item $\mC_{calc}$ is the set of all minimal conflict sets w.r.t.\ $\langle\mo,\mb,\Tp,\Tn\rangle_\RQ$ computed so far and 
\item $\mD_{\times}$ is the set of all minimal diagnoses w.r.t.\ $\langle\mo,\mb,\Tp,\Tn\rangle_\RQ$ computed so far where each diagnosis in $\mD_{\times}$ does not satisfy all test cases $\Tp'$ and $\Tn'$.
% OLD
%where ``most-probable'' refers to the probability measure $p(\md), \md\in\minD_{\langle\mo,\mb,\Tp,\Tn\rangle_\RQ}$ obtained from $p(\tax),\tax\in\mo$ by application of Formula~\ref{eq:diag_prob_calc};
%% and \ref{eq:diag_prob_norm}.
%\item $\Queue$ is the current queue of open nodes of the (partial) weighted pruned HS-tree, i.e.\ a set of nodes $\mathsf{nd} \subseteq \mo$ that have not yet been processed and labeled. 
%\item $\mathbf{C}_{calc} \subseteq \minC_{\langle\mo,\mb,\Tp,\Tn\rangle_\RQ}$ is the overall set of computed minimal conflict sets throughout all calls to \textsc{staticHS} during the execution of Algorithm~\ref{algo:inter_onto_debug}.
%\item $\mD_{\times}$ is the overall set of computed minimal diagnoses throughout all calls to \textsc{staticHS} during the execution of Algorithm~\ref{algo:inter_onto_debug} where each $\md\in\mD_{\times}$ has been invalidated by some test case in $\Tp'\cup\Tn'$.
% end OLD
\end{enumerate} 
\end{proposition}
\begin{proof}
Termination of any call to \textsc{staticHS} within Algorithm~\ref{algo:inter_onto_debug} is granted by the fact that each node is a subset of $\mo$ wherefore $2^{|\mo|}$ is a finite upper bound of the overall number of nodes that might be elements of $\Queue$ during the execution of any call of \textsc{staticHS}. Moreover, in each iteration of the repeat-loop in \textsc{staticHS}, one element is removed from $\Queue$ (line~\ref{algoline:static_update_Q}) and no once removed element can ever be readded to $\Queue$. The latter is satisfied due to the non-minimality criterion (lines~\ref{algoline:slabel:non_min_crit_start}-\ref{algoline:slabel:non_min_crit_end}) that deletes all but one nodes set-equal to some set $X \subseteq \mo$ before the first node set-equal to $X$ is processed and due to the fact that no once labeled nodes, i.e.\ those nodes that are elements of $\mD_{calc}$, $\mD_{\checkmark}$ or $\mD_{\times}$, are ever added to $\Queue$ again (because there is no line of code in \textsc{staticHS} that does so). 
%
%Algorithm~\ref{algo:hs} terminates (Proposition~\ref{prop:hs:termination}) wherefore the relevant data of the (partial) weighted pruned HS-tree $T$ produced by Algorithm~\ref{algo:hs} includes finite sets. Since \tectsc{staticHS} outputs the same relevant data as Algorithm~\ref{algo:hs} by Lemma~\ref{lem:static_hs_each_call} and in each iteration of the repeat-loop in \textsc{staticHS} at least $\Queue$ is modified, \textsc{staticHS} must terminate either due to $\Queue =\emptyset$ or because the finite second set in the relevant data of $T$  

Proposition~(1): During the execution of Algorithm~\ref{algo:inter_onto_debug} (and \textsc{staticHS}), diagnoses are added to $\mD_{\times}$ only in line~\ref{algoline:inter_onto_debug:update_D_times}. In this line, only and all diagnoses not complying with the most recent test case are added to $\mD_{\times}$ (\textsc{getInvalidDiags} in line~\ref{algoline:inter_onto_debug:get_invalid_diags}, cf.\ Section~\ref{sec:DetailedAlgorithmDescription}). Hence, no diagnosis in $\mD_{\times}$ can be in $\minD_{\langle\mo,\mb,\Tp,\Tn\rangle_\RQ} \cap \minD_{\langle\mo,\mb,\Tp \cup \Tp',\Tn \cup \Tn'\rangle_\RQ}$. Now, by Lemmata~\ref{lem:static_hs_most_prob_diags} and \ref{lem:static_hs_soundness}, we deduce that $\mD \subset \minD_{\langle\mo,\mb,\Tp,\Tn\rangle_\RQ} \cap \minD_{\langle\mo,\mb,\Tp\cup\Tp',\Tn\cup\Tn'\rangle_\RQ}$ is the set of most probable 
 minimal diagnoses w.r.t.\ $\langle\mo,\mb,\Tp,\Tn\rangle_\RQ$ that satisfy all test cases $\Tp'$ and $\Tn'$. If \textsc{staticHS} does not terminate due to $\Queue = []$, properties (a)-(i) and (a)-(ii) of $\mD$ are direct consequences of the stop criterion in line~\ref{algoline:static:until} in \textsc{staticHS}. Otherwise, we infer by Lemma~\ref{lem:static_hs_completeness} that (b) must be true.

Propositions~(2) and (3) hold by Lemma~\ref{lem:static_hs_each_call} and the definition of relevant data of a (partial) wpHS-tree (cf.\ Remark~\ref{rem:algo_hs:internal_representation}).
%
%3): This holds by Lemma~\ref{lem:static_hs_each_call} and the definition of relevant data of a (partial) weighted pruned HS-tree (cf.\ Remark~\ref{rem:algo_hs:internal_representation}).

Proposition~(4): This proposition follows from the line of argument in the proof of proposition~(1) above.
\end{proof} 

\newgeometry{margin=2cm}

\begin{algorithm*}
\small
\caption{Iterative Construction of a Static Hitting Set Tree} \label{algo:inter_stat_hs}
\begin{algorithmic}[1]
\Require 
a tuple $\tuple{ \langle\mo,\mb,\Tp,\Tn\rangle_\RQ, \Queue, t, n_{\min}, n_{\max}, \mC_{calc}, \mD_{\checkmark}, \mD_{\times}, p(), \Tp', \Tn' }$ consisting of
\begin{itemize}
	\item the DPI $\langle\mo,\mb,\Tp,\Tn\rangle_\RQ$ given as input to Algorithm~\ref{algo:inter_onto_debug},
	\item the overall sets of positively ($\Tp'$) and negatively ($\Tn'$) answered queries added as test cases to $\langle\mo,\mb,\Tp,\Tn\rangle_\RQ$ so far, 
	\item the current queue $\Queue$ of open (non-labeled) nodes of a (partial) wpHS-tree,
	%a queue $\Queue$ of open (non-labeled) nodes, 
	\item some desired computation timeout $t$,
	\item a desired minimal ($n_{\min}\geq2$) and maximal ($n_{\max}$) number of minimal diagnoses to be returned, 
	\item the set $\mC_{calc}$ of all minimal conflict sets w.r.t.\ $\langle\mo,\mb,\Tp,\Tn\rangle_\RQ$ computed so far,
	%of calculated conflict sets so far (each $\mc \in \mC_{calc}$ is a conflict set w.r.t.\ the current DPI 
	%minimal conflict set w.r.t.\ some intermediate DPI), 
	\item the set $\mD_{\checkmark}$ of all minimal diagnoses w.r.t.\ $\langle\mo,\mb,\Tp,\Tn\rangle_\RQ$ computed so far that satisfy all test cases $\Tp'$ and $\Tn'$,
	\item the set $\mD_{\times}$ of all minimal diagnoses w.r.t.\ $\langle\mo,\mb,\Tp,\Tn\rangle_\RQ$ computed so far 
	that do not satisfy all test cases $\Tp'$ and $\Tn'$.
	%where each diagnosis in $\mD_{\times}$ does not satisfy all test cases $\Tp'$ and $\Tn'$ and
	\item a function $p: \mo \rightarrow (0,0.5)$.
\end{itemize}
%%a DPI $\langle\mo,\mb,\Tp,\Tn\rangle_\RQ$, 
%%queues of open ($\Queue$) and closed ($Q_{closed}$) nodes computed by prior calls to $\scHS$, some computation timeout $t$, a set $\mathbf{C}_{calc}$ of already calculated minimal conflict sets throughout prior calls to $\scHS$, a set of known valid diagnoses (nodes) $\mD_{\checkmark}$, a desired minimal ($n_{\min}$) and maximal ($n_{\max}$) number of diagnoses to be returned per call to $\scHS$, a weight function $w(\tax) \in \mathbb{R}$ that assigns to each $\tax \in \mo$ a weight and thereby determines the search strategy (breadth-first or uniform-cost)
%a DPI $\langle\mo,\mb,\Tp,\Tn\rangle_\RQ$, the overall sets of newly collected positive ($\Tp'$) and negative ($\Tn'$) test cases, a queue $\Queue$ of open nodes, some computation timeout $t$, a desired minimal ($n_{\min}$) and maximal ($n_{\max}$) number of minimal diagnoses to be returned, the set $\mC_{calc} \subseteq \minC_{\langle\mo,\mb,\Tp,\Tn\rangle_\RQ}$ of all already calculated minimal conflict sets, a set of known valid ($\mD_{\checkmark}$) and invalid ($\mD_{\times}$) minimal diagnoses, a function $p(\tax) \in (0,0.5)$ for $\tax\in\mo$
%%, a mode $mode$ (static or dynamic) that influences tree pruning
\Ensure 
a tuple $\tuple{\mD,\Queue, \mathbf{C}_{calc}, \mD_{\times}}$ where
\begin{itemize}
\item $\mD$ is the current set of leading diagnoses such that
\begin{enumerate}[(a)]
\item $\mD \subseteq \minD_{\langle\mo,\mb,\Tp,\Tn\rangle_\RQ} \cap \minD_{\langle\mo,\mb,\Tp\cup\Tp',\Tn\cup\Tn'\rangle_\RQ}$ is the set of most probable 
%(as per $p_{\mo}(\tax),\tax\in\mo$ and Formula~\ref{eq:diag_prob_calc})
 minimal diagnoses w.r.t.\ $\langle\mo,\mb,\Tp,\Tn\rangle_\RQ$ that satisfy all test cases $\Tp'$ and $\Tn'$ such that 
\begin{enumerate}[(i)]
\item $n_{\min} \leq |\mD| \leq n_{\max}$ and 
\item $\mD\supset\mD_{\checkmark}$,
\end{enumerate} 
if such a set $\mD$ 
%with (i) and (ii)
 exists,
or 
\item $\mD$ is equal to the set of all minimal diagnoses $\minD_{\langle\mo,\mb,\Tp,\Tn\rangle_\RQ} \cap \minD_{\langle\mo,\mb,\Tp\cup\Tp',\Tn\cup\Tn'\rangle_\RQ}$, otherwise,
\end{enumerate}
where ``most-probable'' refers to the probability measure $p_{nodes}()$ (cf.\ Definition~\ref{def:p_node()}) obtained from the given function $p()$;
%$p(\md), \md\in\minD_{\langle\mo,\mb,\Tp,\Tn\rangle_\RQ}$ obtained from $p(\tax),\tax\in\mo$ by application of Formula~\ref{eq:diag_prob_calc}. 
%and \ref{eq:diag_prob_norm}.
%where ``most-probable'' refers to the probability measure $p_{\mD}(\md), \md\in\mD$ obtained from $p_{\mo}(\tax),\tax\in\mo$ which is transformed to some probability measure $p_{\mD,prio}(\md),\md\in\mD$ by Formula~\ref{eq:diag_prob_calc} which is in turn used to calculate $p_{\mD}(\md), \md\in\mD$ by Formula~\ref{eq:bayes} (see \emph{Computing a Probability Distribution of Leading Diagnoses} on page~\pageref{etc:computing_prob_dist_of_leading_diags}). 
\item $\Queue$ is the current queue of open (non-labeled) nodes of the produced (partial) wpHS-tree, 
%\item $\Queue$ is the current queue of open (non-labeled) nodes of the hitting set tree, 
%$\Queue$ is the current queue of open nodes of the hitting set tree, i.e.\ a set of nodes $\mathsf{nd} \subseteq \mo$ that have not yet been processed and labeled. 
\item $\mC_{calc}$ is the set of all minimal conflict sets w.r.t.\ $\langle\mo,\mb,\Tp,\Tn\rangle_\RQ$ computed so far and 
%$\mathbf{C}_{calc} \subseteq \minC_{\langle\mo,\mb,\Tp,\Tn\rangle_\RQ}$ is the overall set of computed minimal conflict sets throughout all calls to \textsc{staticHS} during the execution of Algorithm~\ref{algo:inter_onto_debug}.
\item $\mD_{\times}$ comprises those minimal diagnoses w.r.t.\ $\langle\mo,\mb,\Tp,\Tn\rangle_\RQ$ computed so far that do not satisfy all test cases $\Tp'$ and $\Tn'$.
%\item $\mD_{\times}$ is the set of all minimal diagnoses w.r.t.\ $\langle\mo,\mb,\Tp,\Tn\rangle_\RQ$ computed so far where each diagnosis in $\mD_{\times}$ does not satisfy all test cases $\Tp'$ and $\Tn'$.
%$\mD_{\times}$ is the overall set of computed minimal diagnoses throughout all calls to \textsc{staticHS} during the execution of Algorithm~\ref{algo:inter_onto_debug} where each $\md\in\mD_{\times}$ has been invalidated by some test case in $\Tp'\cup\Tn'$.
\end{itemize}  
\vspace{10pt}
\Procedure{staticHS}{$\langle\mo,\mb,\Tp,\Tn\rangle_\RQ, \Queue, t, n_{\min}, n_{\max}, \mathbf{C}_{calc}, \mD_{\checkmark}, \mD_{\times}, p(), \Tp', \Tn'$}
\State $t_{start} \gets \Call{getTime}{ }$
\State $\mD_{calc} \gets \emptyset$
\Repeat
\State $\mathsf{node} \gets \Call{getFirst}{\Queue}$ \label{algoline:static:getfirst}
%\State $\Queue \gets \Queue \setminus \setof{\mathsf{node}}$\label{algoline:static_update_Q}
\State $\Queue \gets \Call{deleteFirst}{\Queue}$\label{algoline:static_update_Q}
\State $\tuple{L,\mathbf{C}} \gets \Call{sLabel}{\langle\mo,\mb,\Tp,\Tn\rangle_\RQ, \mathsf{node},\mathbf{C}_{calc},\mD_{\times} \cup \mD_{\checkmark} \cup \mD_{calc}, \Queue}$ \label{algoline:static:slabel}
\State $\mathbf{C}_{calc} \gets \mathbf{C}$ \label{algoline:static:update_Ccalc}
\If{$L = valid$} \label{algoline:static:if_L_valid}     \Comment{$\mathsf{node}$ is minimal diagnosis w.r.t.\ $\langle\mo,\mb,\Tp,\Tn\rangle_\RQ$}
	%\State $\Queue \gets \Queue \setminus \setof{\mathsf{node}}$
	\If{\Call{isKBValid}{$\mo\setminus\mathsf{node}, \langle\cdot,\mb,\Tp\cup\Tp',\Tn\cup\Tn'\rangle_\RQ$}}\label{algoline:static:isOntValid}  \Comment{\textsc{isKBValid} (see Algorithm~\ref{algo:qx})}
	%\fixme{\Tn can be omitted here since addition of automatically generated positive tcs cannots violate negative tcs}
		\State $\mD_{calc} \gets \mD_{calc} \cup \setof{\mathsf{node}}$\label{algoline:static:add_to_Dcalc}  \Comment{$\mathsf{node}$ does satisfy all test cases $\Tp'$ and $\Tn'$}
	\Else
		\State $\mD_{\times} \gets \mD_{\times} \cup \setof{\mathsf{node}}$\label{algoline:static:add_to_Dtimes} \Comment{$\mathsf{node}$ does not satisfy all test cases $\Tp'$ and $\Tn'$}
	\EndIf
\ElsIf{$L = closed$} \label{algoline:static:if_L_closed}
				\Comment{do nothing, no need to store non-minimal diagnoses}
	%\State $\Queue \gets \Queue \setminus \setof{\mathsf{node}}$
\Else 	\Comment{$L$ must be a minimal conflict set}
	%\State $\Queue \gets \Queue \setminus \setof{\mathsf{node}}$
	\For{$e \in L$}
		\State $\Queue \gets \Call{insertSorted}{ \mathsf{node} \cup \setof{e}, \Queue, p_{nodes}(), descending}$  \label{algoline:static:generate_nodes}
\EndFor
\EndIf
\Until{$\Queue=[] \lor [|\mD_{calc}| \neq \emptyset \land \left|\mD_{calc} \cup \mD_{\checkmark}\right| \geq n_{\min} \land ( |\mD_{calc} \cup \mD_{\checkmark}| = n_{\max} \lor \Call{getTime}{ } - t_{start} > t)]$}\label{algoline:static:until}
\State \Return $\tuple{\mD_{calc} \cup \mD_{\checkmark},\Queue, \mathbf{C}_{calc}, \mD_{\times}}$
\EndProcedure
\vspace{10pt}
\Procedure{\textsc{sLabel}}{$\langle\mo,\mb,\Tp,\Tn\rangle_\RQ,\mathsf{node},\mathbf{C}_{calc}, \mD_{(\times,\checkmark,calc)}, \Queue$}
\For{$\mathsf{nd} \in \mD_{(\times,\checkmark,calc)}$}\label{algoline:slabel:non_min_crit_start}
	\If{$\mathsf{node} \supseteq \mathsf{nd}$}    \Comment{$\mathsf{node}$ is a non-minimal diagnosis}
			\State \Return $\tuple{closed,\mathbf{C}_{calc}}$
	\EndIf
\EndFor\label{algoline:slabel:non_min_crit_end}
\For{$\mathsf{nd} \in \Queue$}\label{algoline:slabel:dup_crit_start}
	\If{$\mathsf{node} = \mathsf{nd}$}      \Comment{$\mathsf{node}$ is a duplicate node}
			\State \Return $\tuple{closed,\mathbf{C}_{calc}}$
	\EndIf
\EndFor\label{algoline:slabel:dup_crit_end}
\For{$\mc \in \mathbf{C}_{calc}$}   \label{algoline:slabel:reuse_crit_start}
	\If{$\mc \cap \mathsf{node} = \emptyset$}    \Comment{the minimal conflict set $\mc$ can be reused to label $\mathsf{node}$}
		\State \Return $\tuple{\mc,\mathbf{C}_{calc}}$  \label{algoline:slabel:reuse_crit_end}
	\EndIf
\EndFor
\State $L\gets \Call{QX}{\langle\mo\setminus\mathsf{node},\mb,\Tp,\Tn\rangle_\RQ}$\label{algoline:slabel:qx} \Comment{see Algorithm~\ref{algo:qx} (page~\pageref{algo:qx})}
\If{$L$ = \text{'no conflict'}}						\Comment{$\mathsf{node}$ is a diagnosis}
	\State \Return $\tuple{valid,\mathbf{C}_{calc}}$\label{algoline:slabel:return_valid}
\Else						\Comment{$L$ is a \emph{new} minimal conflict set ($\notin \mathbf{C}_{calc}$)}
	\State $\mathbf{C}_{calc} \gets \mathbf{C}_{calc} \cup \setof{L}$
	\State \Return $\tuple{L,\mathbf{C}_{calc}}$
\EndIf
\EndProcedure
\end{algorithmic}
\normalsize
\end{algorithm*}
\restoregeometry

\chapter[\textsc{dynamicHS}: A Dynamic Iterative Diagnosis Computation Algorithm]{\textsc{dynamicHS}: A Dynamic Iterative Diagnosis Computation Algorithm%
\chaptermark{Dynamic Diagnosis Computation Algorithm}}
\chaptermark{Dynamic Diagnosis Computation Algorithm}
\label{chap:DynamicHSTree}
%%%%%%
%\chapter{\textsc{dynamicHS}: A Dynamic Iterative Diagnosis Computation Algorithm}
%\chaptermark{Dynamic HS}
%\label{chap:DynamicHSTree}
%%%%%%
As the name already suggests, \textsc{dynamicHS} (Algorithm~\ref{algo:inter_dyn_hs}) is a procedure that solves the problem of \emph{Interactive Dynamic KB Debugging} defined by Problem Definition~\ref{prob_def:dynamic} if used for leading diagnosis computation in Algorithm~\ref{algo:inter_onto_debug}. \textsc{dynamicHS} is sound, complete and optimal w.r.t.\ the set of solutions of the \emph{Interactive Dynamic KB Debugging} problem (this will be proven in Section~\ref{sec:CorrectnessOfTextscDynamicHS}). Optimality refers to the best-first computation of minimal diagnoses regarding a given probability measure.

\section{Overview and Intuition}
\label{sec:OverviewAndIntuition}
%\fixme{emphasize the terms current, last-but-one DPI}
%\noindent\textbf{Synoptic View of the Algorithm.} 
\paragraph{Synoptic View of the Algorithm.}
\textsc{dynamicHS} (Algorithm~\ref{algo:inter_dyn_hs}) is employed as a subroutine in Algorithm~\ref{algo:inter_onto_debug} with $mode = dynamic$ to build up a hitting set tree iteratively. That is, each time \textsc{dynamicHS} is called in Algorithm~\ref{algo:inter_onto_debug}, it expands the existing tree only to a sufficient extent in order to determine a desired number of new leading diagnoses used for the generation of the next query. Then, the leading diagnoses set is returned. 

Outside of the \textsc{dynamicHS} method in Algorithm~\ref{algo:inter_onto_debug}, a new diagnosis probability distribution is obtained by the diagnosis probability update (cf.\ Section~\ref{sec:DetailedAlgorithmDescription}). Once this distribution involves one diagnosis, the probability of which exceeds a predefined threshold $1-\sigma$, the algorithm terminates. The output is a solution KB w.r.t.\ the \emph{current DPI} built from this highly probable minimal diagnosis.

\begin{remark}\label{rem:sigma_in_dynamicHS}
In case $\sigma$ has a predefined value of zero, the output is the (exact) solution to the problem of \emph{Interactive Dynamic KB Debugging} for the input DPI. In a scenario where some fault tolerance $\sigma > 0$ is given, the solution KB returned by Algorithm~\ref{algo:inter_onto_debug} is an approximation of the (exact) solution to \emph{Interactive Dynamic KB Debugging} for the input DPI where a better approximation can be expected for smaller values of $\sigma$ (cf.\ Remark~\ref{rem:approximate_solution}). ``Better'' in this context refers to the satisfaction of desired semantic properties of the KB returned by Algorithm~\ref{algo:inter_onto_debug}, i.e.\ desired entailments and desired non-entailments of the KB. The intuition is that specification of additional test cases $T$ guarantees the output of a KB complying with these test cases, whereas accepting 
%as solution KB of \emph{Interactive Static KB Debugging} 
one -- albeit highly probable -- of multiple solution KBs without having incorporated $T$ leaves open the possibility for this KB to not fulfill $T$.

However, answering queries is effort for an interacting user. Therefore, the approach that involves the ``early'' termination of the algorithm after a solution KB has a sufficiently high probability (lower than 1) constitutes a trade-off between exactness of the output and the effort of the user and overall execution time of the interactive KB debugging algorithm, respectively.\qed
\end{remark}
%After incorporating the user's answer, some leading diagnoses are eliminated (this is granted by the definition of a query, see Definition~\ref{def:query}). Moreover, 

In case there is no highly probable leading diagnosis, a query constructed from the current set of leading diagnoses is asked to the user. The user's answer is incorporated into the current DPI resulting in a new DPI. Thereafter, \textsc{dynamicHS} is invoked again given this new DPI 
%(and parameters that store the ``state'' of the current hitting set tree) 
as an argument.

%\noindent\textbf{Storage of the Tree.} 
\paragraph{Storage of the Search Tree.}
Between each two calls of \textsc{dynamicHS} in Algorithm~\ref{algo:inter_onto_debug}, the ``state'' of the current hitting set tree is stored by variables 
\begin{itemize}
	\item $\mD_{calc}$ -- computed minimal diagnoses w.r.t.\ the current DPI,
	\item $\Queue$ -- the list of open, non-labeled nodes,
	\item $\mC_{calc}$ -- (not necessarily minimal) conflict sets w.r.t.\ the current DPI computed so far,
	\item $\mD_{\supset}$ -- non-minimal diagnoses w.r.t.\ the current DPI computed so far,
	\item $\Queue_{dup}$ -- non-labeled duplicate nodes (i.e.\ nodes corresponding to tree branches with the same set of edge labels as branches that are already present in the tree)
	%that correspond to branches that are already represented in the tree),
	\item $\mD_{\times}$ -- the empty set (is filled up during Algorithm~\ref{algo:inter_onto_debug} between two calls of \textsc{dynamicHS} with diagnoses from $\mD_{calc}$ that have been invalidated by an answered query)
\end{itemize}
where nodes in the tree again store (among others) the edge labels on the path from the root node to themselves. 

%\noindent\textbf{Tree Update.} 
\paragraph{Search Tree Update.}
It is immediately apparent from the enumeration given above that, in comparison to \textsc{staticHS}, additional collections, i.e.\ $\mD_{\supset}$ as well as $\Queue_{dup}$, need to be maintained in order to ``remember'' the current tree while Algorithm~\ref{algo:inter_onto_debug} is processing outside of the method \textsc{dynamicHS}. The cause for these additional variables is the \emph{tree update} necessary after each addition of a test case to a DPI. For, each iteration of \textsc{dynamicHS} considers a different DPI in terms of the test cases. And, any two different DPIs in general lead to a different hitting set tree and to different sets of minimal diagnoses and conflict sets. Hence, the idea of the tree update is the following: Reuse the partial hitting set tree $T$ (stored by the variables described above) constructed before the new test case was added to the current DPI $DPI_j$ and perform suitable modifications to $T$ 
%(which is what we call \emph{tree update}) 
in order to obtain a tree $T'$ such that the further expansion of $T'$ allows to identify \emph{all} minimal diagnoses w.r.t.\ the new DPI $DPI_{j+1}$ resulting from the addition of the new test case to $DPI_j$. In other words, the tree update seeks to establish a tree that is equivalent to one built by execution of \textsc{dynamicHS} using the new DPI $DPI_{j+1}$ starting from an empty tree.

%\noindent\textbf{Node Storage.} 
\paragraph{Node Storage.}
Notice that, unlike in \textsc{staticHS} or \textsc{HS}, it is crucial to store nodes not as sets in \textsc{dynamicHS}, but as \emph{ordered lists of formulas}. That is, each node $\mathsf{nd}$ stores a list of all the edge labels along the (directed) path in the hitting set tree from the root node to $\mathsf{nd}$ where the order of formulas in the list is given by the order of traversing the edge labels along this path. Additionally, \textsc{dynamicHS} stores the attribute $\mathsf{nd.cs}$ for each node $\mathsf{nd}$ which is an ordered list including the node labels, i.e.\ the conflict sets, along the path from the root node to $\mathsf{nd}$ in analogous way. Associating a node with these two lists instead of one set is necessary from the point of view of the tree update.
% that will be made at the beginning of each iteration of \textsc{dynamicHS}. 
Because this facilitates the differentiation between two nodes corresponding to an equal (partial) diagnosis. For example, there could be some node $\mathsf{nd}_1$ that is ``redundant'' after some query $Q$ has been answered, but there is a set-equal node $\mathsf{nd}_2$ which is still ``relevant'' (set-equality refers to equal \emph{sets}, not lists, of edge labels stored by two nodes). 
%, i.e.\ non-redundant. 
In this case, the algorithm should get rid of $\mathsf{nd}_1$ (in order to save time and space) while preserving node $\mathsf{nd}_2$ (in order to maintain completeness). Associating set-equal nodes with each other might thus either lead to unnecessary tree expansion steps (if none is deleted) or incompleteness of the algorithm concerning the consideration of all minimal diagnoses (in case both are deleted).
\paragraph{Addition of a Test Case Changes Set of Solutions.}
Unlike the \textsc{staticHS} algorithm, which is strongly related to the non-interactive hitting set algorithm \textsc{HS} (Algorithm~\ref{algo:hs}) as outlined in Section~\ref{sec:TheIntuition},
% in terms of the produced hitting set tree, 
the hitting set tree produced by \textsc{dynamicHS} will usually differ significantly from the non-interactive hitting set tree produced by \textsc{HS}. 
The reason for this is that in \textsc{dynamicHS} the initial DPI $DPI_0$ is not fixed (in that conflict sets and diagnoses are calculated only w.r.t.\ $DPI_0$), but \emph{new test cases are also used for the computation of minimal conflict sets (and thus minimal diagnoses) and not only for the invalidation of diagnoses}. Hence, every time a query has been answered and a respective test case has been incorporated into the DPI, the minimal conflict sets computed for the old DPI $DPI_j$ might not be minimal conflict sets w.r.t.\ the current DPI $DPI_{j+1}$ anymore (see Examples~\ref{example:dynamicHS_small_example_using_tabExDpi2} and \ref{example:dynamicHS_large_example_using_tabExDpi3}). 
On the one hand, a minimal conflict set $\mc$ w.r.t.\ $DPI_j$ might be a non-minimal conflict set w.r.t.\ $DPI_{j+1}$ (since there is a new minimal conflict set $\mc' \subset \mc$ w.r.t.\ $DPI_{j+1}$). On the other hand, there might be also ``completely new'' minimal conflict sets $\mc_k$ w.r.t.\ $DPI_{j+1}$ which are in no set-relationship with any minimal conflict set w.r.t.\ $DPI_j$. 

Due to this changing set of minimal conflict sets, the set of minimal diagnoses is variable as well (cf.\ Proposition~\ref{prop:mindiag_mincs}). To see this, let $\md$ be a minimal diagnosis w.r.t.\ $DPI_j$. Then $\md$ hits all minimal conflict sets $\mc_k$ in $\minC_{DPI_j}$. Now, assume that $\md$ comprises (only) the element $\tax$ from $\mc_k$, but there is a minimal conflict set $\mc'_k$ in $\minC_{DPI_{j+1}}$ such that $\mc'_k \subseteq \mc_k \setminus \setof{\tax}$. In this case, $\md$ is not a (minimal) hitting set of all minimal conflict sets in $\minC_{DPI_{j+1}}$ (since $\md$ does not hit $\mc'_k$), i.e.\ $\md$ is not a (minimal) diagnosis w.r.t.\ $DPI_{j+1}$. That means, $\md$ needs to be extended (by a hitting set of all minimal conflict sets in $\minC_{DPI_{j+1}}$ it does not hit) in order to become a diagnosis w.r.t.\ $DPI_{j+1}$. After extending $\md$, both situations might arise, either that $\md$ is a minimal diagnosis w.r.t.\ $DPI_{j+1}$ or that $\md$ is a non-minimal diagnosis w.r.t.\ $DPI_{j+1}$. When the latter case occurs, \textsc{dynamicHS} might often be able to figure out that (the tree branch corresponding to) $\md$ is simply \emph{redundant} (w.r.t.\ the new DPI $DPI_{j+1}$) and does not need to be considered during the further expansion of the hitting set tree (which searches for minimal diagnoses w.r.t.\ $DPI_{j+1}$ and not w.r.t.\ $DPI_{j}$). That is, such redundant tree branches are unnecessary in order to explore all minimal diagnoses w.r.t.\ $DPI_{j+1}$ (cf.\ Sections~\ref{sec:OverviewAndIntuition} and \ref{sec:RedundantNodesInTextscDynamicHS} for an explanation and precise characterization of redundancy). 

As a consequence, the nice property of \textsc{staticHS} that the set of minimal diagnoses that needs to be taken into account given $DPI_{j+1}$ is a proper subset of the minimal diagnoses set that needed to be considered given $DPI_{j}$ in no longer valid for \textsc{dynamicHS}. That is, the set of remaining solution candidates in \textsc{dynamicHS} is not guaranteed to ``converge'' \emph{constantly} towards a singleton comprising only one solution. The DPI, the minimal conflict sets as well as the minimal diagnoses are ``dynamic''. What holds for both \textsc{dynamicHS} and \textsc{staticHS} is the guarantee that the set of all (i.e.\ minimal and non-minimal) diagnoses is constantly shrinking, i.e.\ $\allD_{DPI_j} \supset \allD_{DPI_{j+1}}$ (as well will later prove by Corollary~\ref{cor:adding_query_to_DPI_implies_that_allD_wrt_new_DPI_is_proper_subset_of_allD_wrt_old_DPI}).

%\noindent\textbf{Tree Pruning.} 
\paragraph{Search Tree Pruning.}
Let $T$ be the hitting set tree produced in the $j$-th iteration of \textsc{dynamicHS} (i.e.\ $T$ is the tree that was used to search for minimal diagnoses w.r.t.\ $DPI_{j}$). Then, after a new test case has been added to $DPI_{j}$, there are often redundant subtrees in $T$ that can be pruned. The resulting tree $T'$ can then be used in the $(j+1)$-th iteration of \textsc{dynamicHS} to identify minimal diagnoses w.r.t.\ the new DPI $DPI_{j+1}$. Using $T$ instead of $T'$ might lead to a significant time and (more severely) space overhead, due to the unnecessary expansion of redundant branches that are known to give no new information at all. Another approach could be to simply discard the entire tree $T$ and start to construct a new one w.r.t.\ $DPI_{j+1}$ from scratch. This strategy, however, will usually also suffer from a non-negligible time overhead since most of the tree $T$ can be safely reused in iteration $j+1$ and only parts of it must be revised. In particular, this strategy would potentially involve many additional calls of $\scQX$ (which internally calls an expensive reasoner) as, in the worst case (when no pruning is possible), the entire existing tree might be rebuilt. 
%This would, in particular, involve the recomputation of all already computed minimal conflict sets () one for each node labeled by a minimal conflict set in $T$ that is also a node in $T'$. For,  

As we shall see in Remark~\ref{rem:finding_all_redundant_nodes_makes_algo_inefficient}, Section~\ref{sec:TextscDynamicHSDetailsAndCorrectness} and Examples~\ref{example:dynamicHS_small_example_using_tabExDpi2} as well as \ref{example:dynamicHS_large_example_using_tabExDpi3}, the overhead in terms of (expensive) calls to a reasoner (i.e.\ calls of $\scQX$) due to tree pruning (compared to its impact on the tree) is absolutely reasonable. In fact, only one call of a ``fast version'' of $\scQX$ (see Section~\ref{sec:HittingSetTreePruningInTextscDynamicHS}) might already lead to the deletion of $75\%$ of the tree branches as one can see in the first pruning step in Example~\ref{example:dynamicHS_large_example_using_tabExDpi3}.
%
%For the reason of pruning, Qdup Dsupset must be stored and nodes are not sets but lists of sentences and nd.cs is given as an additional attibute of a node......

The evolution of the hitting set tree produced by Algorithm~\ref{algo:inter_onto_debug} using \textsc{dynamicHS} is thus characterized by \emph{alternating expansion and pruning phases}. Also for very complex problems, in case that expansion phases are ``short enough'' such that tree pruning can take place ``often enough'', one might be able to keep the hitting set tree ``small enough'' to handle it efficiently. The extent of the expansion phase can be steered by the specification of the leading diagnosis parameters $n_{\min}$, $n_{\max}$ and $t$ (cf.\ Section~\ref{sec:DetailedAlgorithmDescription}). In the extreme case, these can be defined in a way ($n_{\min}=n_{\max}=2$) the algorithm will allow only the computation of a single further minimal diagnosis (in the first expansion phase: two diagnoses) before \textsc{dynamicHS} (i.e.\ the tree expansion phase) terminates and a further pruning phase might take place. 

However, it is not automatically warranted that tree pruning is possible after each expansion phase. Similarly, no certainty is given that the transition from $DPI_j$ to $DPI_{j+1}$ just causes the deletion of parts of the tree and no additional expansion of the tree. 
In fact, this depends on certain properties of the test case that is added after an expansion phase (i.e.\ properties of the generated query).

%\noindent\textbf{Test Cases Affect Tree Pruning.} 
\paragraph{Test Cases Affect Tree Pruning.}
Some added test case might give rise to some pruning steps as well as it might induce the construction of new subtrees (where ``new'' means that these would be no subtress of a hitting set tree w.r.t.\ the previous DPI $DPI_j$). The latter situation occurs when ``completely new'' minimal conflict sets (see above) are introduced by the addition of a test case. If this is the only impact of a test case, then this test case has only a negative influence on the time and space complexity. In other words, none of the invalidated minimal diagnoses (and no other nodes in the tree) are redundant; but all of them must additionally hit the set of ``completely new'' minimal conflict sets (in order to become diagnoses w.r.t.\ $DPI_{j+1}$). Hence, in this case, the transition from $DPI_j$ to $DPI_{j+1}$ results only in monotonic growth of the tree. If possible, such ``negative-impact test cases'' must be avoided. On the other hand, one must strive for the usage of ``positive-impact test cases'', i.e.\ those that only trigger tree pruning, but no tree expansion.  
Defining and studying properties that constitute such ``positive-impact test cases'' and developing specialized algorithms for extracting exactly those types of queries that enable as substantial and effective pruning as possible is a topic of future research. 

An idea pertinent to this issue could for example be to attempt to extract a query by means of the conflict set $\mc$ that labels the root node of the tree. More concretely, if any answer to a query yields a new test case that leads to the introduction of a minimal conflict set that is a proper subset of $\mc$, then it is for sure that significant pruning can take place (since \emph{entire subtrees starting from the root of the tree} can be deleted). For instance, the first query $Q_1$ in Example~\ref{example:dynamicHS_large_example_using_tabExDpi3} features this property. Roughly, the reasons for that are that $Q_1$ is an entailment of a proper subset $\mc_{sub}$ of $\mc$ (i.e.\ $\mc_{sub}$ is a justification of $Q_1$, cf.\ Section~\ref{sec:ConflictSetsVersusJustifications}) and $Q_1$ 
%is ``relevant'' for this conflict set $\mc$ to hold. 
is ``relevant'' for this conflict set $\mc$ to be a conflict set. In other words, the latter means that $Q_1$ can be used to ``replace'' the part $\mc_{sub}$ of $\mc$, i.e.\ $(\mc \setminus \mc_{sub}) \cup Q_1$ is invalid w.r.t.\ the given DPI. 
That is, addition of $Q_1$ to the positive test cases asserts the correctness of one part of $\mc$, namely $\mc_{sub}$ (cf.\ Example~\ref{example:dynamicHS_large_example_using_tabExDpi3}), wherefore the other part must be incorrect (because some part of a conflict set must be definitely incorrect). On the other hand, assignment of $Q_1$ to the negative test cases asserts exactly the incorrectness of $\mc_{sub}$ wherefore the formulas $\mc \setminus \mc_{sub}$ become obsolete in the minimal conflict set $\mc$ yielding the new minimal conflict set $\mc' := \mc_{sub}$.
%must not be further considered as a part of the minimal conflict set $$. 
Another desirable property of $Q_1$ is that addition of $Q_1$ to either set of test cases does not imply the origination of any ``completely new'' conflict sets (see above) which result in additional growth of the tree.

That is, in its original form (without assuring only the usage of ``positive-impact test cases''), the time and
%, more critically, the 
space complexity of \textsc{dynamicHS} is a function of the generated queries. There is a potential to perform significant pruning, but also the risk of significant tree growth. In case mostly ``positive-impact queries'' are generated and asked to the user, the performance might be very nice and significantly superior to the one of \textsc{staticHS}. In the reverse case, the performance might be also worse than the one of \textsc{staticHS}. In the case of \textsc{staticHS}, there is no chance for significant pruning, but also no chance for a tree growth that goes beyond the size of the non-interactive tree produced by \textsc{HS}.  

In \textsc{staticHS}, there are only expansion phases (in case the tree pruning described by Definition~\ref{def:pruned_hs_tree} is considered part of an expansion phase) which means that the tree constructed by \textsc{staticHS} will constantly grow (apart from the deleted duplicate nodes and non-minimal diagnoses). All the user can do is hope that Algorithm~\ref{algo:inter_onto_debug} applying \textsc{staticHS} will not run out of memory (cf.\ Section~\ref{sec:TheIntuition}). 

The idea is now to be able to use \textsc{dynamicHS} instead of \textsc{staticHS} particularly if the latter runs out of memory soon. If the leading diagnosis parameters are specified small enough to prevent the hitting set tree produced during one expansion phase from becoming too large and test cases are not chosen unfavorably, the \textsc{dynamicHS} method should be able to outperform \textsc{staticHS} significantly, as Examples~\ref{example:staticHS_complex_example_using_tabExDpi3} and \ref{example:dynamicHS_large_example_using_tabExDpi3} suggest.

%
%
%
%redundant subtrees of the hitting set tree resulting from the  which used
%
%
%
 %in that a minimal diagnoses w.r.t.\ the old DPI might need to be extended by further logical sentences to become a minimal diagnosis w.r.t.\ the current DPI (since there is now a minimal conflict set that it does not hit). In addition, minimal diagnoses w.r.t.\ the old DPI might become redundant, i.e.\ the tree branches corresponding to these diagnoses might be unnecessary in order to explore all minimal diagnoses w.r.t.\ the current DPI. 
%
%
%
%
%
 %and the minimal diagnoses w.r.t.\ $DPI_j$ might not be minimal diagnoses w.r.t.\ $DPI_{j+1}$. 
%
%
%In this vein, each time \textsc{dynamicHS} is executed after getting called from Algorithm~\ref{algo:inter_onto_debug}, it computes minimal diagnoses w.r.t.\ a new DPI that is different from the DPI that was used for diagnosis calculation one iteration before. 
%
%
%
%
%
%

\section{Algorithm Walkthrough}
\label{sec:AlgorithmWalkthrough_dynamic}
%\noindent\textbf{Input parameters.} 
\paragraph{Input Parameters.}
When \textsc{dynamicHS} (Algorithm~\ref{algo:inter_dyn_hs}) is called for the first time in Algorithm~\ref{algo:inter_onto_debug}, the inputs $\mC_{calc}$, $\mD_{\checkmark}$, $\mD_{\times}$, $\Tp'$ and $\Tn'$ correspond to the empty set and $\Queue = [\emptyset]$ (cf.\ lines~\ref{algoline:inter_onto_debug:var_inst_start}-\ref{algoline:inter_onto_debug:var_inst_end} and \ref{algoline:inter_onto_debug:dynamicHS} in Algorithm~\ref{algo:inter_onto_debug}). Further on, $\mD_{calc}$ is defined to be the empty set at the beginning of each execution of \textsc{dynamicHS}. That is, \textsc{dynamicHS} starts the construction of the hitting set tree from an initial tree consisting of a single unlabeled root node $\emptyset$ ($\in \Queue$). And, all collections that are later returned by \textsc{dynamicHS} in line~\ref{algoline:dyn:return}, except for $\Queue$, are initially empty. Further input arguments are the DPI $\langle\mo,\mb,\Tp,\Tn\rangle_\RQ$ provided as an input to Algorithm~\ref{algo:inter_onto_debug}, the sets of positively ($\Tp'$) and negatively ($\Tn'$) answered queries since the start of Algorithm~\ref{algo:inter_onto_debug} (both sets initially empty), the leading diagnosis computation parameters $n_{\min},n_{\max},t$ (see description in Chapter~\ref{chap:UserInteraction} on page~\pageref{etc:leading_diag_params}) and the probability measure $p() := p_{\mo}()$ that assigns a probability in the interval $(0,0.5)$ to each formula in $\mo$ (see line~\ref{algoline:inter_onto_debug:getAxiomProbs} in Algorithm~\ref{algo:inter_onto_debug}).

%\noindent\textbf{Tree Update during First Iteration of \textsc{dynamicHS}.} 
\paragraph{Tree Update during First Iteration of \textsc{dynamicHS}.}
Before the repeat-loop in \textsc{dynamicHS} is entered, the \textsc{updateTree} function is called (line~\ref{algoline:dyn:update_tree}), but has no effect. This holds since \textsc{updateTree} first iterates over all elements in $\mD_{\times}$, then over all elements in $\mD_{\supset}$ and finally over all elements in $\mD_{\checkmark}$ where $\mD_{\times}=\mD_{\supset}=\mD_{\checkmark}=\emptyset$, as pointed out before.

%\noindent\textbf{The Main Loop.} 
\paragraph{The Main Loop.}
During the repeat-loop, in each iteration the first node $\mathsf{node}$ in the queue $\Queue$ of open (non-labeled) nodes is processed (\textsc{getFirst}, line~\ref{algoline:dyn:get_first}). Notice that, anywhere throughout \textsc{dynamicHS}, nodes are added to $\Queue$ in a way that a sorting of $\Queue$ in descending order according to $p_{nodes}()$ (cf.\ Definition~\ref{def:p_node()}) is maintained (cf.\ \textsc{insertSorted} in lines~\ref{algoline:static:generate_nodes}, \ref{algoline:update:insert_sorted_0}, \ref{algoline:update:insert_sorted_0.5}, \ref{algoline:update:insert_sorted_1}, \ref{algoline:prune:insert_alternative_equal_node_into_S'} and \ref{algoline:prune:insert_same_node_into_S'}). Hence, the most probable node (according to $p_{nodes}()$) is always processed next.

So, when $\mathsf{node}$ is processed, it is first deleted from $\Queue$ (\textsc{deleteFirst}, line~\ref{algoline:dyn:delete_from_queue}). Then a test is performed whether $\mathsf{node} \in \mD_{\checkmark}$, i.e.\ whether $\mathsf{node}$ is already known to be a minimal diagnosis w.r.t.\ the current DPI $\langle\mo,\mb,\Tp\cup\Tp',\Tn\cup\Tn'\rangle_\RQ$. In case this test is positive, $\mathsf{node}$ is directly added to $\mD_{calc}$, the set of leading diagnoses that will be output by the current call of \textsc{dynamicHS}. Otherwise, the \textsc{dLabel} function is called given $\mathsf{node}$ (i.a.) as a parameter (line~\ref{algoline:dyn:dlabel}). 

%\noindent\textbf{Computation of a Node Label.} 
\paragraph{Computation of a Node Label.}
The \textsc{dLabel} function processes $\mathsf{node}$ as follows. First, the \emph{non-minimality criterion} (lines~\ref{algoline:dlabel:non-min_crit_start}-\ref{algoline:dlabel:non-min_crit_end}) is checked. That is, among all nodes in $\mD_{calc}$, one is searched which is a proper subset of $\mathsf{node}$. If such a node $\mathsf{nd}$ is found, then $\mathsf{node}$ must be a non-minimal diagnosis w.r.t.\ the current DPI since, anytime throughout the execution of \textsc{dynamicHS}, 
%given the DPI $\langle\mo,\mb,\Tp,\Tn\rangle_\RQ$ as well as the newly specified positive ($\Tp'$) as well as negative ($\Tn'$) test cases, 
$\mD_{calc}$ contains only minimal diagnoses w.r.t.\ the current DPI $\langle\mo,\mb,\Tp\cup\Tp',\Tn\cup\Tn'\rangle_\RQ$ (this will be proven later by Proposition~\ref{prop:dyn_soundness}). In this case, unlike in \textsc{staticHS}, the branch in the hitting set tree corresponding to $\mathsf{node}$ cannot be simply discarded, but needs to be still stored (in the set $\mD_{\supset}$). It is necessary to store non-minimal diagnoses as these might become minimal diagnoses w.r.t.\ the new DPI obtained after the subsequent addition of a new test case to the current DPI (cf.\ Proposition~\ref{prop:after_adding_testcase_new_min_diag_is_equal_or_superset_of_old_min_diag}). 

In case the non-minimality criterion is not satisfied, the \emph{reuse criterion} (lines~\ref{algoline:dlabel:reuse_start}-\ref{algoline:dlabel:reuse_end}) is checked next. That is, the set $\mC_{calc}$ containing (not necessarily minimal) conflict sets w.r.t.\ the current DPI is browsed for a set $\mc$ such that $\mc$ and $\mathsf{node}$ are disjoint sets. If such a set $\mc$ is found, there must be some set $X \subseteq \mc$ which is a \emph{minimal} conflict set w.r.t.\ the current DPI. This minimal conflict set $X$ can then be 
%is determined to be a minimal conflict set w.r.t.\ the current DPI, it can be 
used to label $\mathsf{node}$ since the set of edge labels along the path in the tree leading from the root node to $\mathsf{node}$ does not hit $X$ (because it does not hit $\mc$). 

The minimality of $\mc$ is verified by a call of $\scQX(\langle\mc,\mb,\Tp\cup\Tp',\Tn\cup\Tn'\rangle_\RQ)$ that yields $X$, a minimal conflict set w.r.t.\ the current DPI (cf.\ Proposition~\ref{prop:qx_correctness}; notice that $X$ must be a non-empty set due to Proposition~\ref{prop:if_DPI_cs_neq_emptyset_then_DPI+1_cs_neq_emptyset}, for details see Section~\ref{sec:TextscDynamicHSDetailsAndCorrectness}). In case $X \subset \mc$ (line~\ref{algoline:dlabel:if_X=C}), before $X$ is returned as a 
%minimal conflict set w.r.t.\ the current DPI suitable to 
label for $\mathsf{node}$, the following tree pruning steps are performed:
\begin{itemize}
	\item All the conflict sets $\mc_i$ used as node labels in the hitting set tree or in duplicate tree branches so far (i.e.\ $\mc_i \in \mathsf{nd.cs}$ for a node $\mathsf{nd} \in \Queue \cup \mD_{\supset} \cup \Queue_{dup}$) such that $X \subset \mc_i$ are replaced by $X$ (\textsc{pruneQdup} and \textsc{prune} in lines~\ref{algoline:dlabel:call_prune_Qdup}-\ref{algoline:dlabel:call_prune_Dsupset}),
	\item any subtree is pruned if its root node is linked to a node now labeled by $X$ (replacing some $\mc_i \supset X$) by an edge with label $\tax$ where $\tax$ is in $\mc_i \setminus X$ (\textsc{pruneQdup} and \textsc{prune} in lines~\ref{algoline:dlabel:call_prune_Qdup}-\ref{algoline:dlabel:call_prune_Dsupset}) and
	\item for each pruned node $\mathsf{nd}$, if there is a non-pruned node in $\Queue_{dup}$ suited to construct a node $\mathsf{nd}'$ that can replace $\mathsf{nd}$, $\mathsf{nd}'$ is added to the collection of nodes from which $\mathsf{nd}$ was deleted (\textsc{pruneQdup} and \textsc{prune} in lines~\ref{algoline:dlabel:call_prune_Qdup}-\ref{algoline:dlabel:call_prune_Dsupset}),
	%$\Queue$ (if $\mathsf{nd}$ was deleted from $\Queue$) or to $\Queue_{dup}$ (if $\mathsf{nd}$ was deleted from $\Queue_{dup}$), respectively,  
	%
	%all subtrees starting from outgoing edges of replaced sets $\mc_i$ labeled by an element not in $X$ are pruned (\textsc{pruneQdup} and \textsc{prune} in lines~\ref{algoline:dlabel:call_prune_Qdup}-\ref{algoline:dlabel:call_prune_Dsupset}) and
	\item all the conflict sets $\mc_i \in \mC_{calc}$ that are proper supersets of $X$ are deleted from $\mC_{calc}$ and $X$ is added to $\mC_{calc}$ (\textsc{addSetDelSupsets} in line~\ref{algoline:dlabel:add_set_del_supset}). 
	%\item all the conflict sets $\mc_i$ used as node labels in duplicate tree branches (i.e.\ $\mc_i \in \mathsf{nd.cs}$ for a node $\mathsf{nd} \in \Queue_{dup}$)
	%, i.e.\ those nodes stored in $\Queue_{dup}$ 
	%that are proper supersets of $X$ are replaced by $X$ (\textsc{prune} in lines~\ref{algoline:dlabel:call_prune_Q}-\ref{algoline:dlabel:call_prune_Dsupset})
\end{itemize}
Otherwise, $\mc$ ($= X$) is directly returned by \textsc{dLabel} without performing any tree pruning because the reused conflict set $\mc$ is (still) a \emph{minimal} conflict set w.r.t.\ the current DPI $\langle\mo,\mb,\Tp\cup\Tp',\Tn\cup\Tn'\rangle_\RQ$ (notice that each element of $\mC_{calc}$ was added to $\mC_{calc}$ as a minimal conflict set w.r.t.\ some DPI $\langle\mo,\mb,\Tp\cup\Tp'',\Tn\cup\Tn''\rangle_\RQ$ where $\Tp'' \subseteq \Tp'$ and $\Tn'' \subseteq \Tn'$ during the execution of this or a previous call of \textsc{dynamicHS}). For an in-depth explanation of the pruning functions \textsc{prune} and \textsc{pruneQdup} the reader is kindly referred to Section~\ref{sec:HittingSetTreePruningInTextscDynamicHS}.  

\begin{remark}\label{rem:no_tree_pruning_in_first_iteration_of_dynamicHS}
During the execution of the first call of \textsc{dynamicHS} in Algorithm~\ref{algo:inter_onto_debug}, no tree pruning can take place (neither within the scope of \textsc{dLabel} nor anywhere else) since all elements of $\mC_{calc}$ (initially the empty set) must be minimal conflict sets w.r.t.\ the input DPI which is at the same time the current DPI. Pruning of the hitting set tree is only possible in case some non-leaf nodes of the tree are labeled by conflict sets that are \emph{not minimal} w.r.t.\ the current DPI.\qed
\end{remark}

Given that the reuse criterion fails, $\scQX$ is called given the \emph{current} DPI $\langle\mo\setminus\mathsf{node},\mb,\Tp\cup\Tp',\Tn\cup\Tn'\rangle_\RQ$ as an argument (line~\ref{algoline:dlabel:qx_2}). If the output $L$ is equal to 'no conflict', then we know by Proposition~\ref{prop:qx_correctness} that $\mathsf{node}$ is a diagnosis w.r.t.\ the current DPI, wherefore the label $valid$ is returned for $\mathsf{node}$. Otherwise, the output $L$ must be a minimal conflict set w.r.t.\ $\langle\mo,\mb,\Tp\cup\Tp',\Tn\cup\Tn'\rangle_\RQ$ that has an empty set-intersection with $\mathsf{node}$. Since the reuse criterion failed, i.e.\ there is no set in $\mC_{calc}$ that does not intersect with $\mathsf{node}$, $L$ must be a fresh minimal conflict set w.r.t.\ $\langle\mo,\mb,\Tp\cup\Tp',\Tn\cup\Tn'\rangle_\RQ$ in the sense that $L \notin \mC_{calc}$ must hold. Therefore the label $L$ is first added to $\mC_{calc}$ and then returned by \textsc{dLabel} as a label for $\mathsf{node}$. 

\begin{remark}\label{rem:qx_to_compute_min_cs_key_difference_between_staticHS_and_dynamicHS}
Please notice that this call of $\scQX$ to label a node is one of the key differences between \textsc{staticHS} and \textsc{dynamicHS}. Whereas the former uses $\scQX$ exclusively for the computation of minimal conflict sets w.r.t.\ the (static) input DPI exploiting just the initial sets of positive and negative test cases $\Tp$ and $\Tn$, respectively, the latter employs $\scQX$ to compute minimal conflict sets w.r.t.\ the (dynamic) current DPI which includes all new test cases ($\Tp'$ and $\Tn'$) resulting from answered queries in the ongoing interactive debugging session so far.\qed
\end{remark}

%\noindent\textbf{Processing of a Node Label.} 
\paragraph{Processing of a Node Label.}
Back in the main procedure, the label $L$ returned by the \textsc{dLabel} function is processed as follows. If $L = valid$, then it is a fact that $\mathsf{node}$ is a minimal diagnosis w.r.t.\ the current DPI (cf.\ Proposition~\ref{prop:dyn_soundness} in Section~\ref{sec:SoundnessOfTextscDynamicHS}) wherefore $\mathsf{node}$ is added to the set $\mD_{calc}$. Otherwise, if $nonmin$ is the returned label for $\mathsf{node}$, $\mathsf{node}$ is added to the set $\mD_{\supset}$ of non-minimal diagnoses w.r.t.\ the current DPI. Otherwise, i.e.\ if $L \notin \setof{valid,nonmin}$, then $L$ must be a minimal conflict set w.r.t.\ the current DPI (see the description of node label computation above). In this case, $|L|$ successor nodes of $\mathsf{node}$ are generated (lines~\ref{algoline:dyn:add_ax_to_node} and \ref{algoline:dyn:add_cs_to_node.cs}). For each logical formula $e \in L$, a new node is computed from $\mathsf{node}$ (and $\mathsf{node.cs}$) as $\mathsf{node}_e := \textsc{add}(\mathsf{node},e)$ and $\mathsf{node}_e.\mathsf{cs} := \textsc{add}(\mathsf{node.cs},L)$ which means that $e$ is appended to the end of the list $\mathsf{node}$ and $L$ is appended to the end of the list $\mathsf{node.cs}$.

If there is already a node $\mathsf{nd}\in\Queue$ such that $\mathsf{nd} = \mathsf{node}_e$ (line~\ref{algoline:dyn:check_node_already_in_Q}), where '$=$' applied to these lists means that the list $\mathsf{nd}$ \emph{interpreted as a set} is equal to the list $\mathsf{node}_e$ \emph{interpreted as a set} (cf.\ Section~\ref{sec:DefinitionsAndNotation} for an explication of this notation), then there is already a branch in the existing tree which includes the same set of edge labels as the new node $\mathsf{node}_e$. Note that the tree branch corresponding to $\mathsf{nd}$ will differ from the one corresponding to $\mathsf{node}_e$ in terms of the order of edge labels or (the order of) the node labels visited when traversed starting from the root node. As it makes no sense to expand two branches with equal sets of edge labels in a hitting set tree (cf.\ rule~\ref{def:pruned_hs_tree:rule6} in Definition~\ref{def:pruned_hs_tree}) for time and space complexity reasons and the fact that the sought diagnoses are sets -- and not lists -- of edge labels in the tree, such a duplicate node $\mathsf{node}_e$ is stored in the separate list $\Queue_{dup}$. This list $\Queue_{dup}$ is always kept sorted by ascending node-cardinality (\textsc{insertSorted} in line~\ref{algoline:dyn:add_to_Qdup}). 

The purpose of storing and not deleting such nodes is the possibility that the now ``active'' branch $\mathsf{nd}$ might be pruned after the addition of some test case whereas $\mathsf{node}_e$ might be unaffected by that pruning step. In this case, $\mathsf{node}_e$, given it meets certain properties (see Section~\ref{sec:TextscDynamicHSDetailsAndCorrectness} for details), can be reactivated and incorporated into the tree in order to replace $\mathsf{nd}$. Had $\mathsf{node}_e$ just been discarded instead of being stored, the completeness of Algorithm~\ref{algo:inter_onto_debug} with $mode = dynamic$ would be violated in general. That is, we would not have any guarantee that all minimal diagnoses w.r.t.\ the current DPI are actually explored by the algorithm.

Otherwise, if there is no node in $\Queue$ that is set-equal to $\mathsf{node}_e$, then $\mathsf{node}_e$ is added to the $k$-th position in $\Queue$ (\textsc{insertSorted} in line~\ref{algoline:dyn:generate_nodes}) if there are (exactly) $k-1$ nodes in $\Queue$ that have a probability as per $p_{nodes}()$ that is greater than or equal to $p_{nodes}(\mathsf{node}_e)$.  

%\noindent\textbf{Stop Criterion.} 
\paragraph{Stop Criterion.}
The repeat-loop of \textsc{dynamicHS} is executed until the stop criterion in line~\ref{algoline:dyn:until} is satisfied. The first criterion causing \textsc{dynamicHS} to terminate is $\Queue = []$ which means that the complete hitting set tree has been constructed and no further nodes can be labeled. In this case, $\mD_{calc}$ comprises all minimal diagnoses w.r.t.\ the current DPI $\langle\mo,\mb,\Tp\cup\Tp',\Tn\cup\Tn'\rangle_\RQ$ (cf.\ Proposition~\ref{prop:dyn_completeness}).

If the first criterion is not met, then the second criterion is checked. That is, a test is performed which checks first whether there is at least one \emph{new} diagnosis w.r.t.\ the current DPI in $\mD_{calc}$ which was not returned by the last-but-one call of \textsc{dynamicHS} (i.e.\ which is not an element of $\mD_{\checkmark}$). Notice that this criterion or $\Queue = []$ will be definitely met after finite execution time of \textsc{dynamicHS} since either new nodes in $\Queue$ will be processed (and labeled) until there is some new diagnosis w.r.t.\ the current DPI identified or the $\Queue$ will become empty.

Additionally, the second criterion involves a test that checks whether the cardinality of $\mD_{calc}$ amounts to at least $n_{\min}$ and either $|\mD_{calc}| = n_{\max}$ or more than $t$ time has passed since the start of the execution of \textsc{dynamicHS}. In the latter case, $n_{\min} \leq |\mD_{calc}| < n_{\max}$ holds. In the former case, $|\mD_{calc}| = n_{\max}$ is satisfied.

%\noindent\textbf{Processing of the Leading Diagnoses Returned by \textsc{dynamicHS}.} 
\paragraph{Processing of the Leading Diagnoses Returned by \textsc{dynamicHS}.}
When a call of \textsc{dynamicHS} in Algorithm~\ref{algo:inter_onto_debug} returns $\tuple{\mD_{calc} ,\Queue, \mathbf{C}_{calc}, \mD_{\times}, \mD_{\supset}, \Queue_{dup}}$, the set $\mD_{calc}$ is stored in the variable $\mD_{\checkmark}$ in Algorithm~\ref{algo:inter_onto_debug}. Between two successive calls of \textsc{dynamicHS} in Algorithm~\ref{algo:inter_onto_debug}, only this set $\mD_{\checkmark}$ as well as $\mD_{\times}$ are modified. The collections $\Queue$, $\mC_{calc}$, $\mD_{\supset}$ as well as $\Queue_{dup}$ remain unchanged until they are used as input parameters when it comes to the next call of \textsc{dynamicHS} in Algorithm~\ref{algo:inter_onto_debug}.

In case one diagnosis $\md_{\max}$ of the current leading diagnoses in $\mD_{\checkmark}$ has a probability greater than or equal to $1 - \sigma$ as per the probability measure $p_{\mD}()$ (see Section~\ref{sec:DetailedAlgorithmDescription}), the stop criterion of interactive KB debugging is met and the solution KB $(\mo\setminus\md_{\max}) \cup U_{\Tp \cup \Tp'}$ w.r.t.\ the current DPI $\langle\mo,\mb,\Tp\cup\Tp',\Tn\cup\Tn'\rangle_\RQ$ 
%constructed from the current DPI $\langle\mo,\mb,\Tp\cup\Tp',\Tn\cup\Tn'\rangle_\RQ$ as well as from $\md_{\max}$ 
is returned to the user (\textsc{getSolKB} in line~\ref{algoline:inter_onto_debug:return}, cf.\ Section~\ref{sec:DetailedAlgorithmDescription}). Thereafter, Algorithm~\ref{algo:inter_onto_debug} terminates and no more calls of \textsc{dynamicHS} take place.

Otherwise, if no leading diagnosis satisfies the stop criterion, a query $Q$ together with its q-partition $\Pt(Q)$ is computed as has been detailed in Chapter~\ref{chap:QueryGeneration} and Section~\ref{sec:DetailedAlgorithmDescription}. An answer $u(Q)$ to this query is submitted by the interacting user (line~\ref{algoline:inter_onto_debug:user_interaction} in Algorithm~\ref{algo:inter_onto_debug}). Then $u(Q)$ along with $\Pt(Q)$ is exploited to figure out the subset $\mD_{out}$ of $\mD_{\checkmark}$ that does not comply with $u(Q)$. This set $\mD_{out}$ is then deleted from $\mD_{\checkmark}$ and added to $\mD_{\times}$. Additionally, $Q$ is added to the positive test cases $\Tp'$ if $u(Q) = \true$ and to the negative test cases $\Tn'$ otherwise. Subsequently, \textsc{dynamicHS} is called again given 
\begin{itemize}
	\item the \emph{updated parameters} $\mD_{\checkmark}$, $\mD_{\times}$, $\Tp'$ and $\Tn'$ (which are modified within and outside of \textsc{dynamicHS} during the execution of Algorithm~\ref{algo:inter_onto_debug}),
	\item the \emph{unchanged parameters} $\Queue$, $\mC_{calc}$, $\mD_{\supset}$ and $\Queue_{dup}$ (which are modified only within \textsc{dynamicHS} during the execution of Algorithm~\ref{algo:inter_onto_debug}) and
	\item the \emph{constant parameters} $\langle\mo,\mb,\Tp,\Tn\rangle_\RQ$, $t$, $n_{\min}$, $n_{\max}$ and $p_{\mo}()$ (which are not modified within or outside of \textsc{dynamicHS} during the execution of Algorithm~\ref{algo:inter_onto_debug}).
%fixed throughout all calls of \textsc{staticHS} made in Algorithm~\ref{algo:inter_onto_debug}.
\end{itemize}
The execution of this next and any subsequent call to \textsc{dynamicHS} runs in analogue way as described so far, except for the effect of the \textsc{updateTree} function called at the very beginning of each execution of \textsc{dynamicHS} (recall that the execution of \textsc{updateTree} had no effect during the \emph{first} execution of \textsc{dynamicHS}). We shall now explicate how this function works in all other executions of \textsc{dynamicHS}, except for the first one.

%The \textsc{updateTree} function
%\noindent\textbf{Tree Update.} 
\paragraph{Tree Update.}
Between line~\ref{algoline:update:process_Dtimes_start} and line~\ref{algoline:update:process_Dtimes_end}, \textsc{updateTree} goes through all nodes $\mathsf{nd}\in\mD_{\times}$ (recall that $\mD_{\times}$ includes exactly these diagnoses that have been ruled out by the most recently answered query) and first performs the \emph{Quick Redundancy Check} (QRC, lines~\ref{algoline:update:qx}-\ref{algoline:update:quickPC_end}) for $\mathsf{nd}$. If the QRC is not successful, it additionally performs the \emph{Complete Redundancy Check} (CRC, lines~\ref{algoline:update:completePC_start}-\ref{algoline:update:completePC_end}) for $\mathsf{nd}$. 

The QRC (for details see Lemma~\ref{lem:quick_prune_check}) aims at identifying whether $\mathsf{nd}$ is redundant and can be pruned, i.e.\ it attempts to find a witness of redundancy of $\mathsf{nd}$. Informally, a \emph{redundant node} in (redundant subtree of) the tree is a node (subtree) such that the further expansion of the current tree without this node (subtree) still yields to the detection of all minimal diagnoses w.r.t.\ the current DPI. A \emph{witness of redundancy of $\mathsf{nd}$} is a minimal conflict set $\mc'$ w.r.t.\ the current DPI such that a superset $\mc \supset \mc'$ was used as a node label on the tree path $\mathsf{nd}$ represents (that is, 
there is some $i \leq |\mathsf{nd.cs}|$ such that $\mc$ is the $i$-th element of $\mathsf{nd.cs}$, i.e.\ $\mc = \mathsf{nd.cs}[i]$)
%$\mc \in \mathsf{nd.cs}$) 
and the label ($\mathsf{nd}[i]$) of the outgoing edge of $\mc$ on the path represented by $\mathsf{nd}$ is an element not in $\mc'$ (that is, an element in $\mc\setminus \mc'$). Formal and precise characterizations of redundancy of nodes and the witness of redundancy of a node are given by Definition~\ref{def:redundant_node} in Section~\ref{sec:RedundantNodesInTextscDynamicHS}. 

To this end, the QRC involves the call of $\scQX(\tuple{U_{\mathsf{nd.cs}}\setminus \mathsf{nd},\mb,\Tp\cup\Tp',\Tn\cup\Tn'}_\RQ)$ which returns $X$. If $X$ is a set (and not 'no conflict'), then $X$ is a minimal conflict set w.r.t.\ the current DPI $\langle\mo,\mb,\Tp\cup\Tp',\Tn\cup\Tn'\rangle_\RQ$ (as $U_{\mathsf{nd.cs}}\setminus \mathsf{nd} \subseteq \mo$, cf.\ Proposition~\ref{prop:qx_correctness}). To check if $X$ is in fact a witness of redundancy of $\mathsf{nd}$, $X \subset \mc$ (line~\ref{algoline:update:X_subset_C_(QRC)}) is tested for all $\mc \in \mathsf{nd.cs}$. If such a $\mc$ is located, $X$ is a witness of redundancy of $\mathsf{nd}$ and the QRC is successful (expressed by $quickRC \gets \true$ in line~\ref{algoline:update:qrc_gets_true}). In this case, the execution is resumed at line~\ref{algoline:update:if_qrc_or_crc_true}.

The QRC bears its name due to the fact that it requires \emph{at most one call of $\scQX$} (which internally performs expensive calls to a reasoner). Moreover, it passes to $\scQX$ a (DPI including a) KB of a size that is generally significantly smaller than $|\mo|$ where $|\mo|$ is roughly the size of the KB used in the (more expensive) calls of $\scQX$ made in the \textsc{dLabel} function. Hence, the QRC will be usually very fast (cf.\ Proposition~\ref{prop:qx_complexity}).

Otherwise, since the negative outcome of the QRC (which is sound, but not complete w.r.t.\ the finding of a witness of redundancy of $\mathsf{nd}$) does not imply the non-existence of a witness of redundancy of $\mathsf{nd}$, the CRC (for details see Lemma~\ref{lem:complete_prune_check}) must be performed. As the name already suggests, the CRC is sound \emph{and complete} and will therefore be positive and yield a witness of redundancy if and only if there is some. The CRC involves multiple calls of $\scQX(\tuple{\mathsf{nd.cs}[i]\setminus \setof{\mathsf{nd}[i]},\mb,\Tp\cup\Tp',\Tn\cup\Tn'}_\RQ)$, one for each conflict set $\mathsf{nd.cs}[i]$ in $\mathsf{nd.cs}$. 
%(where $\mathsf{nd}[i]$ and $\mathsf{nd.cs}[i]$ denote the $i$-th element in the lists $\mathsf{nd}$ and $\mathsf{nd.cs}$, respectively). 
It is straightforward from the characterization of a witness of redundancy given before that, given the CRC returns a set $X$, $X$ is a witness of redundancy of $\mathsf{nd}$. 

If $\mathsf{nd}$ is non-redundant, there cannot be any witness of redundancy of $\mathsf{nd}$. Hence, the complete and sound method CRC will not find such a one. Therefore, $quickRC = \false$ and $completeRC = \false$ must hold in line~\ref{algoline:update:if_qrc_or_crc_true}. In this case, the for-loop in line~\ref{algoline:update:process_Dtimes_start} continues with the next node in $\mD_{\times}$.

On the other hand, if $\mathsf{nd}$ is redundant, due to the completeness of CRC, either $quickRC = \true$ or $completeRC = \true$ must hold when it comes to the execution of the if-statement in line~\ref{algoline:update:if_qrc_or_crc_true}. At this point, it is guaranteed that the variable $X$ stores a witness of redundancy of $\mathsf{nd}$.

The CRC, contrary to the QRC, generally requires \emph{multiple} (at most $|\mathsf{nd}|$) \emph{calls of $\scQX$} (which internally performs expensive calls to a reasoner). But, like the QRC, it passes to $\scQX$ a (DPI including a) KB of a size that is generally significantly smaller than $|\mo|$. Furthermore, at most one call of $\scQX$ will involve more than one call of \textsc{isKBValid} (see Algorithm~\ref{algo:qx}), i.e.\ the function that calls the reasoner. This must be true since CRC only requires an additional call of $\scQX$ if a witness of redundancy has not yet been found. And, each call of $\scQX$ that does not find a witness of redundancy of $\mathsf{nd}$ returns 'no conflict' which necessitates only a single invocation of \textsc{isKBValid}. Hence, each execution of the CRC will be very fast in general as well (cf.\ Proposition~\ref{prop:qx_complexity}).

What comes next is the pruning of all redundant nodes in the tree for which $X$ is a witness of redundancy. Essentially, the same pruning steps are performed here as in the \emph{reuse criterion} described in 'Computation of a node label' above. A detailed discussion of the pruning functions \textsc{prune} as well as \textsc{pruneQdup} can be found in Section~\ref{sec:HittingSetTreePruningInTextscDynamicHS}. 

Notice that a redundant node is guaranteed to be a redundant node in any further iteration of \textsc{dynamicHS} (using a new current DPI that incorporates new test cases). We will prove this by Lemma~\ref{lem:redundant_node_remains_redundant_after_adding_new_testcases} in Section~\ref{sec:RedundantNodesInTextscDynamicHS}. So, nodes pruned by \textsc{prune} or \textsc{pruneQdup} can be deleted for good and do not need to be stored any longer. Moreover, it should be noted that only redundant nodes are pruned at any pruning step in \textsc{dynamicHS}. For, as long as a node in \textsc{dynamicHS} is not known to be redundant, 
%despite the fact that it might have been ruled out by a specified test case, 
some successor node of this node might be a minimal diagnosis w.r.t.\ the current DPI. Thus, the deletion of such a node could perhaps prevent the algorithm from finding a particular minimal diagnosis which would implicate the algorithm's incompleteness. 

\begin{remark}\label{rem:Dtimes_might_change_in_for_loop_in_updateTree} 
Since the removal of a node from a collection $S \in \setof{\mD_{\times},\Queue,\Queue_{dup},\mD_{\supset}}$ within the scope of \textsc{prune} or \textsc{pruneQdup} can be followed by the re-addition to $S$ of a suitable duplicate node constructed from a node stored in $\Queue_{dup}$ (see Section~\ref{sec:HittingSetTreePruningInTextscDynamicHS} for a precise explanation of node replacements), $\mD_{\times}$ might be changed both in that nodes are deleted from it and added to it during the for-loop (line~\ref{algoline:update:process_Dtimes_start}). Therefore, the '$\mathbf{for}\; \mathsf{nd}\in\mD_{\times}$'-statement must be read as 'if $\mathsf{nd}$ is a node in the \emph{current} set $\mD_{\times}$ which has not yet been processed'. For a better code readability, we abstained from using a programmatically precise representation of this issue in Algorithm~\ref{algo:update_tree}.\qed 
\end{remark}

Due to the soundness and completeness of QRC paired with CRC concerning the identification of a witness of redundancy for a given node and the accomplished pruning of (at least) all nodes in $\mD_{\times}$ for which a witness of redundancy has been extracted, all nodes that are in $\mD_{\times}$ when the algorithm reaches line~\ref{algoline:update:reinsert_D_of_Dx_to_Q} are \emph{non-redundant} nodes. 
Consequently, there is no evidence to exclude the remaining nodes in $\mD_{\times}$ from the further search for minimal diagnoses. For this reason, each of these nodes is reinserted into $\Queue$ by \textsc{insertSorted} in line~\ref{algoline:update:insert_sorted_0} such that the sorting of $\Queue$ in descending order of $p_{nodes}()$ is maintained. Then these nodes are deleted from $\mD_{\times}$. Thus, $\mD_{\times} = \emptyset$ holds after each execution of \textsc{updateTree}. 

So, in \textsc{dynamicHS}, unlike in \textsc{staticHS}, diagnoses (and nodes in general) are not ruled out due to the fact that they contradict an answered query, but only if they are (found to be) redundant. Nevertheless, a diagnosis that contradicts an answered query is a ``hot candidate'' for finding some witness of redundancy. For that reason, \textsc{updateTree} searches for witnesses of redundancy (only) by means of $\mD_{\times}$ which includes the most ``suspicious'' nodes. Namely, it comprises those nodes that were minimal diagnoses w.r.t.\ the last-but-one DPI, but have been invalidated by the most recently answered query. The two possible reasons for a diagnosis $\mathsf{nd}$ to be invalidated are its redundancy as defined above or that it does not hit a \emph{new} minimal conflict set (which is not a subset of one in $\mathsf{nd.cs}$) that has been introduced by the addition of the test case resulting from the user's query answer. Thus, it is likely to detect witnesses of redundancy by investigating nodes in $\mD_{\times}$, as the QRC and the CRC do. Throughout the pruning steps performed in lines~\ref{algoline:update:call_prune_Qdup}-\ref{algoline:update:call_prune_Dsupset}, witnesses of redundancy extracted from nodes in $\mD_{\times}$ are exploited to remove redundant nodes in the other collections $\Queue_{dup}$, $\mD_{\supset}$ and $\Queue$ as well.
%the set of diagnoses $\mD_{\times}$ that have been invalidated by the most recently answered query provides ``hot candidates'' for finding witnesses of redundancy (which can be generally exploited to delete not only ``hot candidates'', but any node stored by \textsc{dynamicHS}). 

\begin{remark}\label{rem:finding_all_redundant_nodes_makes_algo_inefficient}
It should be noted that the collections $\Queue$ as well as $\mD_{\supset}$ are not necessarily cleaned from all redundant nodes after all pruning steps in \textsc{updateTree} are finished. At this point, all those redundant nodes are still elements of these collections for which no witness of redundancy was found (there might exist one, though) throughout the redundancy checks (QRC and CRC) performed. 

Assuring the non-existence of redundant nodes in $\Queue$ and $\mD_{\supset}$ might involve extensive usage of the (expensive) reasoner. In the worst case, one call of $\scQX$ for each non-leaf node along each path from the root node to a leaf node labeled by $nonmin$ or to a leaf node that has no label would be necessary. However, the number of these non-leaf nodes is generally \emph{exponential} in the maximum length of such a path in the tree. In comparison, the number of calls of $\scQX$ for investigating all nodes in $\mD_{\times}$ by QRC and CRC is \emph{polynomial (linear)} in the maximum length of a tree path labeled by $\times$. For, the number of $\scQX$-calls cannot get larger than $(n_{\max}-1) (|\mathsf{nd}_{\max}|+1)$ where the constant $n_{\max}$ is the maximum number of desired leading diagnoses predefined by the user and $|\mathsf{nd}_{\max}|$ is the maximum cardinality of some $\mathsf{nd} \in \mD_{\times}$. This holds since $|\mD_{\times}| \leq n_{\max} - 1$ (cf.\ Corollary~\ref{cor:query_leaves_valid_diag}) and QRC requires at most one and CRC at most $|\mathsf{nd}_{\max}|$ $\scQX$-calls. 
%size of the minimal diagnosis with maximum cardinality w.r.t.\ the last-but-one DPI (because \textsc{updateTree} is applied to the tree constructed for the last-but-one DPI). 
%, i.e.\ in $O(|\mc_{\max}|^{|\md_{\max}|})$ where $|\mc_{\max}|$ ($|\md_{\max}|$) is the minimal conflict set w.r.t.\ the input DPI (minimal diagnosis w.r.t.\ the last-but-one DPI) with maximum cardinality.
 
Other than that, the chance of locating new witnesses of redundancy by means of investigating nodes in $\Queue$ and $\mD_{\supset}$ can be assumed to be smaller than for nodes in $\mD_{\times}$ since there is no indication or evidence that these nodes might be redundant. So, cleaning $\Queue$ and $\mD_{\supset}$ from all redundant nodes might be significant effort with negligible impact. Therefore, \textsc{dynamicHS} is designed to focus the search for witnesses of redundancy only on the ``suspicious nodes'' in $\mD_{\times}$.\qed   
%
%the computation of \emph{all} minimal conflict sets w.r.t.\ the current DPI (in order to verify that there is no witness of redundancy for a node) which is, in terms of computational complexity, more or less equivalent to the construction of the complete hitting set tree. Since a complete tree construction may not be necessary to find an adequate solution in many practical cases, this would generally imply heavy performance losses or even the non-applicability of the algorithm to find a solution of an Interactive Dynamic KB Debugging problem at all. Besides, it would contradict the basic principle of iterative tree construction, with changing phases of expansion and pruning, that can make Algorithm~\ref{algo:inter_onto_debug} with $mode=dynamic$ such a powerful device (cf.\ Example~\ref{example:dynamicHS_large_example_using_tabExDpi3}). 
\end{remark}

As mentioned above, when the execution arrives at line~\ref{algoline:update:process_Dsupset_start}, only nodes that are definitely redundant (because they were deleted due to some witness of redundancy) have been deleted from the sets $\Queue$, $\mD_{\times}$, $\mD_{\supset}$ and $\Queue_{dup}$. 

In lines~\ref{algoline:update:process_Dsupset_start}-\ref{algoline:update:process_Dsupset_end}, each node $\mathsf{nd}\in\mD_{\supset}$ which has not been deleted throughout the pruning operations in line~\ref{algoline:update:call_prune_Dsupset} is processed as follows: If there is no minimal diagnosis $\md\in\mD_{\checkmark}$ such that $\mathsf{nd} \supset \md$, then $\mathsf{nd}$ is removed from $\mD_{\supset}$ and reinserted into $\Queue$ (lines~\ref{algoline:update:insert_sorted_0.5} and \ref{algoline:update:delete_from_Dsupset}) in a way the sorting of $\Queue$ in descending order according to $p_{nodes}()$ is maintained (\textsc{insertSorted}). This re-insertion is plausible since there is no more evidence 
%in terms of a minimal diagnosis that is a proper subset 
of $\mathsf{nd}$ (which is a non-minimal diagnosis w.r.t.\ the last-but-one DPI) being a non-minimal diagnosis w.r.t.\ the current DPI (non-minimal diagnoses might become minimal diagnoses by the addition of test cases, cf.\ Section~\ref{sec:ImpactOfAnsweredQueriesOnConflictSets} and Proposition~\ref{prop:after_adding_testcase_new_min_diag_is_equal_or_superset_of_old_min_diag}).

Otherwise, $\mathsf{nd}$ remains an element of the set of non-minimal diagnoses $\mD_{\supset}$ w.r.t.\ the current DPI as $\mD_{\checkmark}$ comprises exclusively minimal diagnoses w.r.t.\ the current DPI and one of these is a proper subset of $\mathsf{nd}$. 

In lines~\ref{algoline:update:process_Dcheckmark_start}-\ref{algoline:update:process_Dcheckmark_end}, all elements in $\mD_{\checkmark}$, each of which is a minimal diagnosis w.r.t.\ the current DPI, are added to $\Queue$ in a way the sorting of $\Queue$ in descending order according to $p_{nodes}()$ is maintained. 

\begin{remark}\label{rem:elements_of_Dcheckmark_are_added_to_Q}
Please notice that the elements of $\mD_{\checkmark}$, although they are known to be minimal diagnoses w.r.t.\ the current DPI, are not directly added to the set of found leading diagnoses $\mD_{calc}$ w.r.t.\ the current DPI, but to $\Queue$. The reason for this is that there might be (not-yet-found) minimal diagnoses w.r.t.\ the current DPI (nodes in $\Queue$ or successor nodes thereof) which were not minimal diagnoses w.r.t.\ the last-but-one DPI (and thus are no elements of $\mD_{\checkmark}$) that have a higher probability as per $p_{nodes}()$ than elements of $\mD_{\checkmark}$. For instance, such diagnoses might have been added to $\Queue$ from the set $\mD_{\supset}$ in line~\ref{algoline:update:insert_sorted_0.5}. 

In this way, since always the first (and most probable) node in $\Queue$ is processed next, a guarantee is given that $\mD_{calc}$ always comprises the $|\mD_{calc}|$ most probable minimal diagnoses w.r.t.\ the current DPI as per $p_{nodes}()$. 
The knowledge of the validity of minimal diagnoses in $\mD_{\checkmark}$ w.r.t.\ the current DPI is however not forgotten, but exploited in line~\ref{algoline:dyn:if_L_valid} (i.e.\ no call of \textsc{dLabel} and $\scQX$ is necessary for a node in $\mD_{\checkmark}$ to be added to $\mD_{calc}$), as elucidated in 'The main loop' above.\qed
\end{remark}

\section[Examples]{Illustrating Examples%
\sectionmark{Examples}}
\sectionmark{Examples}
\label{sec:TextscDynamicHSExamples}
%%%%%
%\section{\textsc{dynamicHS}: Examples}
%\label{sec:TextscDynamicHSExamples}
%%%%%
In this section we will give two examples of how interactive KB debugging using \textsc{dynamicHS} (Algorithm~\ref{algo:inter_onto_debug} with parameter $mode=dynamic$) works. The first one will show the similarities and differences between the usage of \textsc{dynamicHS} (within Algorithm~\ref{algo:inter_onto_debug}) and \textsc{HS} (within Algorithm~\ref{algo:non_int_debug}) since it will depict the application of \textsc{staticHS} on the same example DPI (see Table~\ref{tab:example2}) that was used to show the functionality of \textsc{HS} in examples~\ref{example:non_interactive_debugging_with_tabExDpi2_and_without_probs} and \ref{example:non_interactive_debugging_with_tabExDpi2_and_probs}. At the same time, the first example will provide evidence that solving the problem of Interactive Dynamic KB Debugging can be less efficient than solving the problem of Interactive Static KB Debugging in terms of the number of query answers required from an interacting user. This will be discussed in more detail in Chapter~\ref{chap:TextscStaticHSVersusTextscDynamicHS}.

The second example is supposed to deepen the reader's understanding of the way \textsc{dynamicHS} works. To this end, the example DPI provided by Table~\ref{tab:example3} will be used which constitutes a significantly harder (interactive) debugging task than the DPI investigated in the first example. This example will involve the construction of a relatively large hitting set tree in the first iteration of \textsc{dynamicHS} (which behaves very similarly to \textsc{staticHS} as well as \textsc{HS} and constructs the same wpHS-tree as these methods), but will then show the power of the tree pruning that can be exploited in Interactive Dynamic KB Debugging in that the tree will shrink rapidly after the addition of test cases. Hence, this example will emphasize the advantage of the decision to search for a solution of Interactive Dynamic KB Debugging rather than for a solution of Interactive Static KB Debugging (more on that in Chapter~\ref{chap:TextscStaticHSVersusTextscDynamicHS}).

\begin{figure*}
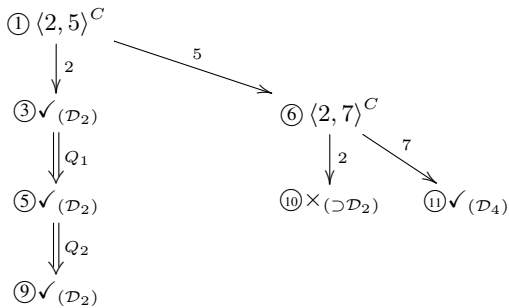

%%%%%%%%%%%%%%%%%%%%%%%%%%%%%%%%%%%%%%%%%%%% 1
\begin{minipage}[c]{0.45\textwidth} 
\small
\xygraph{
!{<0cm,0cm>;<1.8cm,0cm>:<0cm,1.2cm>::}
%!~-{@[|(4)]}
%d=0
!{(1,4)}*+{\textcircled{\scriptsize 1}\tuple{1,2,5}^C}="c1c"
%d=1
!{(0,3) }*+{\textcircled{\scriptsize 2}\checkmark_{(\md_1)}}="d1" 
!{(1,3) }*+{\textcircled{\scriptsize 3}\checkmark_{(\md_2)}}="d2" 
!{(3,3) }*+{?}="c2c"
%d0->d1
"c1c":"d1"_{1}
"c1c":"d2"^{2}
"c1c":"c2c"^{5}
}
\vspace{5pt}
\begin{center}
\small Iteration 1
\end{center}
\end{minipage}
\begin{minipage}[c]{20pt}
$\Bigg>$ 
\end{minipage}
\begin{minipage}[c]{0.45\textwidth}
\small  
\begin{tabular}{l}                                          
    %DPI: \\ $\tuple{\mt,\ma,\setof{\setof{B(w)}},\setof{\setof{\neg C(w)}}}$ \\
		$\mD_{calc} = \setof{\md_1,\md_2} = \setof{[1],[2]}$ \\
		$\Queue = [[5]]$ \\
		$\mC_{calc} = \setof{\tuple{1,2,5}}$ \\
		$\mD_{\supset} = \emptyset$ \\
		$\Queue_{dup} = []$\vspace{-7pt} \\
    \hspace{-8pt}\hdashrule{0.95\textwidth}{0.5pt}{2mm}\vspace{-4pt} \\
		$\tuple{Q_1,\Pt(Q_1)} = \tuple{\setof{E \rightarrow \lnot A},\tuple{\setof{\md_1},\setof{\md_2},\emptyset}}$ \\
    $u(Q_1) = \false$ \\
		$\mD_{\checkmark} = \setof{\md_2}$, $\mD_{\times} = \setof{\md_1}$
		\end{tabular}
\end{minipage}
\begin{minipage}[c]{10pt}
$\Bigg>$ 
\end{minipage} 

\vspace{10pt}
%%%%%%%%%%%%%%%%%%%%%%%%%%%%%%%%%%%%%%%%%%%% 2
\begin{minipage}[c]{0.45\textwidth}
\small  
\begin{tabular}{l}                        
    %DPI: \\ $\tuple{\mt,\ma,\setof{\setof{B(w)}},\setof{\setof{\neg C(w)}}}$ \\
		\textsc{updateTree}: \\
		QRC ($\md_1$): 
		$\scQX(\tuple{\setof{2,5},\mb,\Tp,\Tn\cup\setof{Q_1}}) = \tuple{2,5}$ \\
		$\Rightarrow\;$ \textsc{prune}: $\tuple{1,2,5} \rightarrow \tuple{2,5}$ \\
		%PRUNE($\tuple{1,2,5},1$): 
		$\quad \bullet\;$ prune all subtrees starting  
		from nodes $\tuple{1,2,5}$ \\ 
		$\quad\;\;\;$ by outgoing edge with label $1$  \\ 
		%$\Rightarrow\;$ REP($\tuple{2,5}$): 
		$\quad \bullet\;$ replace by $\tuple{2,5}$ all node labels in the tree \\ 
		$\quad\;\;\;$ that are proper supersets of $\tuple{2,5}$ \\
		$\Rightarrow\;$
		$\mD_{\supset} = \emptyset$, 
		$\mC_{calc} = \setof{\tuple{2,5}}$, \\
		$\quad\;$ $\Queue = [[2],[5]]$, 
		$\Queue_{dup} = []$,
\end{tabular}
\end{minipage}
\begin{minipage}[c]{20pt}
$\Bigg>$ 
\end{minipage}
\begin{minipage}[c]{0.45\textwidth}
\small 
\xygraph{
!{<0cm,0cm>;<1.8cm,0cm>:<0cm,1.2cm>::}
%!~-{@[|(4)]}
%d=0
!{(1,4)}*+{\textcircled{\scriptsize 1}\tuple{\xcancel{1},2,5}^C}="c1c"
%d=1
!{(0,3) }*+{\textcircled{\scriptsize 2}\checkmark_{(\md_1)}}="d1" 
!{(1,3) }*+{\textcircled{\scriptsize 3}\checkmark_{(\md_2)}}="d2" 
!{(3,3) }*+{?}="c2c"
%d=2 
!{(0,2) }*+{\textcircled{\scriptsize 4}\times}="inv_d1_q1" 
!{(1,2) }*+{\textcircled{\scriptsize 4} ?}="val_d2_q1"
%!{(2,2) }*+{\checkmark_{(\md_3)}}="d3"
%!{(3,2) }*+{?}="nonmin1"
%!{(4,2) }*+{?}="d4"
%%d=3
%!{(1,1) }*+{\checkmark}="val_d2_q2"
%!{(2,1) }*+{\times}="inv_d3_q2"
%!{(3,1) }*+{\checkmark}="d5"
%!{(4,1) }*+{\checkmark}="val_d4_q3"
%%d=4
%!{(1,0) }*+{\times}="inv_d2_q3"
%!{(3,0) }*+{\times}="inv_d5_q4"
%!{(4,0) }*+{\checkmark}="val_d4_q4"
%d0->d1
"c1c":@{|.>}"d1"_{1}
"c1c":"d2"^{2}
"c1c":"c2c"^{5}
%d1->d2
"d1":@2{.>}"inv_d1_q1"^{Q_1}
"d2":@2{->}"val_d2_q1"^{Q_1}
%"c2c":"d3"_{1}
%"c2c":"nonmin1"^{2}
%"c2c":"d4"^{7}
%%d2->d3
%"val_d2_q1":@2{->}"val_d2_q2"^{Q_2}
%"d3":@2{->}"inv_d3_q2"^{Q_2}
%"nonmin1":@2{->}"d5"^{Q_3}
%"d4":@2{->}"val_d4_q3"^{Q_3}
%%d3->d4
%"d5":@2{->}"inv_d5_q4"^{Q_4}
%"val_d4_q3":@2{->}"val_d4_q4"^{Q_4}   
%"val_d2_q2":@2{->}"inv_d2_q3"^{Q_3}
}
\vspace{5pt}
\begin{center}
\small Updated Tree
\end{center}
\end{minipage}
\begin{minipage}[c]{10pt}
$\Bigg>$ 
\end{minipage} 

\vspace{10pt}
%%%%%%%%%%%%%%%%%%%%%%%%%%%%%%%%%%%%%%%%%%%% 3
\begin{minipage}[c]{0.45\textwidth}
\small 
\xygraph{
!{<0cm,0cm>;<1.8cm,0cm>:<0cm,1.2cm>::}
%!~-{@[|(4)]}
%d=0
!{(1,4)}*+{\textcircled{\scriptsize 1}\tuple{2,5}^C}="c1c"
%d=1
!{(1,3) }*+{\textcircled{\scriptsize 3}\checkmark_{(\md_2)}}="d2" 
!{(3,3) }*+{\textcircled{\scriptsize 6}\tuple{1,2,7}^C}="c2c"
%d=2 
!{(1,2) }*+{\textcircled{\scriptsize 5}\checkmark_{(\md_2)}}="val_d2_q1"
!{(2,2) }*+{\textcircled{\scriptsize 7}\checkmark_{(\md_3)}}="d3"
!{(3,2) }*+{?}="nonmin1"
!{(4,2) }*+{?}="d4"
%%d=3
%!{(1,1) }*+{\checkmark}="val_d2_q2"
%!{(2,1) }*+{\times}="inv_d3_q2"
%!{(3,1) }*+{\checkmark}="d5"
%!{(4,1) }*+{\checkmark}="val_d4_q3"
%%d=4
%!{(1,0) }*+{\times}="inv_d2_q3"
%!{(3,0) }*+{\times}="inv_d5_q4"
%!{(4,0) }*+{\checkmark}="val_d4_q4"
%d0->d1
"c1c":"d2"^{2}
"c1c":"c2c"^{5}
%d1->d2
"d2":@2{->}"val_d2_q1"^{Q_1}
"c2c":"d3"_{1}
"c2c":"nonmin1"^{2}
"c2c":"d4"^{7}
%%d2->d3
%"val_d2_q1":@2{->}"val_d2_q2"^{Q_2}
%"d3":@2{->}"inv_d3_q2"^{Q_2}
%"nonmin1":@2{->}"d5"^{Q_3}
%"d4":@2{->}"val_d4_q3"^{Q_3}
%%d3->d4
%"d5":@2{->}"inv_d5_q4"^{Q_4}
%"val_d4_q3":@2{->}"val_d4_q4"^{Q_4}   
%"val_d2_q2":@2{->}"inv_d2_q3"^{Q_3}
}
\vspace{5pt}
\begin{center}
\small Iteration 2
\end{center}
\end{minipage}
\begin{minipage}[c]{20pt}
$\Bigg>$ 
\end{minipage}
\begin{minipage}[c]{0.45\textwidth}
\small
\begin{tabular}{l}                                          
    %DPI: \\ $\tuple{\mt,\ma,\setof{\setof{B(w)}},\setof{\setof{\neg C(w)}}}$ \\
		$\mD_{calc} = \setof{\md_2,\md_3} = \setof{[2],[5,1]}$ \\
		$\Queue = [[5,2], [5,7]]$ \\
		$\mC_{calc} = \setof{\tuple{2,5}, \tuple{1,2,7}}$ \\
		$\mD_{\supset} = \emptyset$ \\
		$\Queue_{dup} = []$ \vspace{-7pt} \\
    \hspace{-8pt}\hdashrule{0.95\textwidth}{0.5pt}{2mm}\vspace{-4pt} \\
    $\tuple{Q_2,\Pt(Q_2)} = \tuple{\setof{E \rightarrow G},\tuple{\setof{\md_3},\setof{\md_2},\emptyset}}$ \\
    $u(Q_2) = \false$ \\
		$\mD_{\checkmark} = \setof{\md_2}$, $\mD_{\times} = \setof{\md_3}$
		%$\mD_{\checkmark} = \setof{\md_2,\md_3} = \setof{[2],[1,5]}$ \\
		%$\Queue = \setof{\setof{2,5}, \setof{2,7}}$ \\
		%$\mC_{calc} = \setof{\tuple{2,5}, \tuple{1,2,7}}$ \\
		%$\mD_{\supset} = \emptyset$
                         %\\
    %$\tuple{Q_2,\Pt(Q_2)} = \tuple{\setof{E \sqsubseteq G},\tuple{\setof{[1,5]},\setof{[2]},\emptyset}}$ \\
    %$u(Q_2) = \false$ \\
		%$\mD_{out} = \setof{[1,5]}$
		\end{tabular}
\end{minipage}
\begin{minipage}[c]{10pt}
$\Bigg>$ 
\end{minipage} 

\vspace{10pt}
%%%%%%%%%%%%%%%%%%%%%%%%%%%%%%%%%%%%%%%%%%%% 4
\begin{minipage}[c]{0.45\textwidth}
\small
\begin{tabular}{l}                                          
    %DPI: \\ $\tuple{\mt,\ma,\setof{\setof{B(w)}},\setof{\setof{\neg C(w)}}}$ \\
		\textsc{updateTree}: \\
		QRC ($\md_3$): 
		$\scQX(\tuple{\setof{2,7},\mb,\Tp,\Tn\cup\setof{Q_1,Q_2}}) = \tuple{2,7}$ \\
		$\Rightarrow\;$ \textsc{prune}: $\tuple{1,2,7} \rightarrow \tuple{2,7}$ \\
		$\Rightarrow\;$
		$\mD_{\supset} = \emptyset$, 
		$\mC_{calc} = \setof{\tuple{2,5}, \tuple{2,7}}$ \\
		$\quad\;$ $\Queue = [[2], [5,2], [5,7]]$, 
		$\Queue_{dup} = []$
		%PC ($\md_3$): \\
		%$\scQX(\tuple{\setof{2,7},\mb,\Tp,\Tn\cup\setof{Q_1,Q_2}}) = \tuple{2,7}$ \\
		%$\Rightarrow\;$ prune all subtrees starting from nodes \\
		%$\tuple{1,2,7}$ by outgoing edge with label $1$ \\  
		%Update: \\
		%$\mD_{\supset} = \emptyset$ \\
		%$\mC_{calc} = \setof{\tuple{2,5}, \tuple{2,7}}$
\end{tabular}
\end{minipage}
\begin{minipage}[c]{20pt}
$\Bigg>$ 
\end{minipage}
\begin{minipage}[c]{0.45\textwidth}
\small 
\xygraph{
!{<0cm,0cm>;<1.8cm,0cm>:<0cm,1.2cm>::}
%!~-{@[|(4)]}
%d=0
!{(1,4)}*+{\textcircled{\scriptsize 1}\tuple{2,5}^C}="c1c"
%d=1
!{(1,3) }*+{\textcircled{\scriptsize 3}\checkmark_{(\md_2)}}="d2" 
!{(3,3) }*+{\textcircled{\scriptsize 6}\tuple{\xcancel{1},2,7}^C}="c2c"
%d=2 
!{(1,2) }*+{\textcircled{\scriptsize 5}\checkmark_{(\md_2)}}="val_d2_q1"
!{(2,2) }*+{\textcircled{\scriptsize 7}\checkmark_{(\md_3)}}="d3"
!{(3,2) }*+{?}="nonmin1"
!{(4,2) }*+{?}="d4"
%%d=3
!{(1,1) }*+{\textcircled{\scriptsize 8} ?}="val_d2_q2"
!{(2,1) }*+{\textcircled{\scriptsize 8}\times}="inv_d3_q2"
%!{(3,1) }*+{\checkmark}="d5"
%!{(4,1) }*+{\checkmark}="val_d4_q3"
%%d=4
%!{(1,0) }*+{\times}="inv_d2_q3"
%!{(3,0) }*+{\times}="inv_d5_q4"
%!{(4,0) }*+{\checkmark}="val_d4_q4"
%d0->d1
"c1c":"d2"^{2}
"c1c":"c2c"^{5}
%d1->d2
"d2":@2{->}"val_d2_q1"^{Q_1}
"c2c":@{|.>}"d3"_{1}
"c2c":"nonmin1"^{2}
"c2c":"d4"^{7}
%%d2->d3
"val_d2_q1":@2{->}"val_d2_q2"^{Q_2}
"d3":@2{.>}"inv_d3_q2"^{Q_2}
%"nonmin1":@2{->}"d5"^{Q_3}
%"d4":@2{->}"val_d4_q3"^{Q_3}
%%d3->d4
%"d5":@2{->}"inv_d5_q4"^{Q_4}
%"val_d4_q3":@2{->}"val_d4_q4"^{Q_4}   
%"val_d2_q2":@2{->}"inv_d2_q3"^{Q_3}
}
\vspace{5pt}
\begin{center}
\small Updated Tree
\end{center}
\end{minipage}
\begin{minipage}[c]{10pt}
$\Bigg>$ 
\end{minipage} 

\vspace{10pt}
%%%%%%%%%%%%%%%%%%%%%%%%%%%%%%%%%%%%%%%%%%%% 5
\begin{minipage}[c]{0.45\textwidth}
\small 
\xygraph{
!{<0cm,0cm>;<1.8cm,0cm>:<0cm,1.2cm>::}
%!~-{@[|(4)]}
%d=0
!{(1,4)}*+{\textcircled{\scriptsize 1}\tuple{2,5}^C}="c1c"
%d=1
!{(1,3) }*+{\textcircled{\scriptsize 3}\checkmark_{(\md_2)}}="d2" 
!{(3,3) }*+{\textcircled{\scriptsize 6}\tuple{2,7}^C}="c2c"
%d=2 
!{(1,2) }*+{\textcircled{\scriptsize 5}\checkmark_{(\md_2)}}="val_d2_q1"
!{(3,2) }*+{\textcircled{\tiny 10}\times_{(\supset\md_2)}}="nonmin1"
!{(4,2) }*+{\textcircled{\tiny 11}\checkmark_{(\md_4)}}="d4"
%%d=3
!{(1,1) }*+{\textcircled{\scriptsize 9}\checkmark_{(\md_2)}}="val_d2_q2"
%!{(3,1) }*+{\checkmark}="d5"
%!{(4,1) }*+{\checkmark}="val_d4_q3"
%%d=4
%!{(1,0) }*+{\times}="inv_d2_q3"
%!{(3,0) }*+{\times}="inv_d5_q4"
%!{(4,0) }*+{\checkmark}="val_d4_q4"
%d0->d1
"c1c":"d2"^{2}
"c1c":"c2c"^{5}
%d1->d2
"d2":@2{->}"val_d2_q1"^{Q_1}
"c2c":"nonmin1"^{2}
"c2c":"d4"^{7}
%%d2->d3
"val_d2_q1":@2{->}"val_d2_q2"^{Q_2}
%"nonmin1":@2{->}"d5"^{Q_3}
%"d4":@2{->}"val_d4_q3"^{Q_3}
%%d3->d4
%"d5":@2{->}"inv_d5_q4"^{Q_4}
%"val_d4_q3":@2{->}"val_d4_q4"^{Q_4}   
%"val_d2_q2":@2{->}"inv_d2_q3"^{Q_3}
}
\vspace{5pt}
\begin{center}
\small Iteration 3
\end{center}
\end{minipage}
\begin{minipage}[c]{20pt}
$\Bigg>$ 
\end{minipage}
\begin{minipage}[c]{0.45\textwidth}
\small
\begin{tabular}{l}                                          
    %DPI: \\ $\tuple{\mt,\ma,\setof{\setof{B(w)}},\setof{\setof{\neg C(w)}}}$ \\
		$\mD_{calc} = \setof{\md_2,\md_4} = \setof{[2],[5,7]}$ \\
		$\Queue = []$ \\
		$\mC_{calc} = \setof{\tuple{2,5}, \tuple{2,7}}$ \\
		$\mD_{\supset} = \setof{[5,2]}$ \\
		$\Queue_{dup} = []$ \vspace{-7pt} \\
    \hspace{-8pt}\hdashrule{0.95\textwidth}{0.5pt}{2mm}\vspace{-4pt} \\
    $\tuple{Q_3,\Pt(Q_3)} = \tuple{\setof{Y \rightarrow \lnot A},\tuple{\setof{\md_2},\setof{\md_4},\emptyset}}$ \\
    $u(Q_3) = \false$ \\
		$\mD_{\checkmark} = \setof{\md_4}$, $\mD_{\times} = \setof{\md_2}$
		%$\mD_{\checkmark} = \setof{\md_2,\md_4} = \setof{[2],[5,7]}$ \\
		%$\Queue = \emptyset$ \\
		%$\mC_{calc} = \setof{\tuple{2,5}, \tuple{2,7}}$ \\
		%$\mD_{\supset} = \setof{\setof{2,5}}$
                         %\\
    %$\tuple{Q_3,\Pt(Q_3)} = \tuple{\setof{Y \sqsubseteq \lnot A},\tuple{\setof{[2]},\setof{[5,7]},\emptyset}}$ \\
    %$u(Q_3) = \false$ \\
		%$\mD_{out} = \setof{[2]}$
		\end{tabular}
\end{minipage}
\begin{minipage}[c]{10pt}
$\Bigg>$ 
\end{minipage} 

\vspace{10pt}
\caption[(Example~\ref{example:dynamicHS_small_example_using_tabExDpi2}) Solving the Problem of Interactive Dynamic KB Debugging]{(Example~\ref{example:dynamicHS_small_example_using_tabExDpi2}) Solving the problem of Interactive Dynamic KB Debugging (Problem Definition~\ref{prob_def:dynamic}) for the example DPI given by Table~\ref{tab:example2} by means of Algorithm~\ref{algo:inter_onto_debug} and \textsc{dynamicHS}.} 
\label{fig:example:inter_onto_debug_dynamicHS_TabExDpi2}
\end{figure*}

Notice that, in the following examples, whenever some tuple or list occurs in an expression using set operators, it is interpreted as a set.
\begin{example}\label{example:dynamicHS_small_example_using_tabExDpi2}
In this example we assume that the author (called user throughout this example) of the (admissible) DPI $\langle\mo,\mb,\Tp,\Tn\rangle_\RQ$ given by Table~\ref{tab:example2} applies Algorithm~\ref{algo:inter_onto_debug} with $mode = dynamic$ to interactively debug $\langle\mo,\mb,\Tp,\Tn\rangle_\RQ$. Further, the same scenario and parameter settings as in Example~\ref{example:staticHS_simple_example_using_tabExDpi2} are supposed. That is, $n_{\min} = n_{\max} = 2$ (notice that the time limit $t$ is irrelevant in this case), $q := 1$ (cf.\ Chapter~\ref{chap:QueryGeneration}), $qsm()$ is equal to any query selection measure described in Section~\ref{sec:query_selection_measures}, $p_{\mo}(\tax) := c < 0.5$ for all $\tax \in \mo$, i.e.\ all formula fault probabilities are specified to be equal (to some constant $c$) and $\sigma := 0$.

The tree constructed and parameters computed and used by Algorithm~\ref{algo:inter_onto_debug} using \textsc{dynamicHS} are visualized by Figures~\ref{fig:example:inter_onto_debug_dynamicHS_TabExDpi2} and \ref{fig:example:inter_onto_debug_dynamicHS_TabExDpi2_continued}. 
We use the same notation as in Figures~\ref{fig:example:non-interactive_onto_debug_auto=false+nmin=infty_and_auto=true}, \ref{fig:example:non-interactive_onto_debug_auto=false+nmin=2+nmax=4_with_probs}, \ref{fig:example:inter_onto_debug_staticHS_TabExDpi2}, \ref{fig:example:inter_onto_debug_staticHS_TabExDpi3} and \ref{fig:example:inter_onto_debug_staticHS_TabExDpi3_continued} which is described in Examples~\ref{example:non_interactive_debugging_with_tabExDpi2_and_without_probs}, \ref{example:non_interactive_debugging_with_tabExDpi2_and_probs}, \ref{example:staticHS_simple_example_using_tabExDpi2} and \ref{example:staticHS_complex_example_using_tabExDpi3}. 
%
%The only new notational element here is the $\Longrightarrow$ labeled by some designator of a query. That is, $\checkmark_{(\md_i)} \stackrel{Q_j}{\Longrightarrow} \checkmark$ means that $\md_i$ is still a minimal diagnosis after $Q_j$ has been answered and added to the respective set of test cases of the DPI. On the other hand, $\checkmark_{(\md_i)} \stackrel{Q_j}{\Longrightarrow} \times$ signifies that the minimal diagnosis $\md_i$ is invalidated through the addition of the answered query $Q_j$ to the respective set of test cases of the DPI. Please notice that $\Longrightarrow$ does not point at a node of the wpHS-tree. Instead, the label at which $\Longrightarrow$ points is to be understood as the new label of the node originally labeled by $\checkmark_{(\md_i)}$ from which the (first of possibly multiple) $\Longrightarrow$ goes out. This notation should help to keep track of the evolution of node labels in the wpHS-tree without needing to overload a single node by multiple different successive labels. 

In the first iteration, i.e.\ during the execution of the first call of \textsc{dynamicHS} during Algorithm~\ref{algo:inter_onto_debug}, the root node (initially the empty set) is labeled by the minimal conflict set $\tuple{1,2,5}$ w.r.t.\ $\langle\mo,\mb,\Tp,\Tn\rangle_\RQ$ and three successor nodes, namely $\mathsf{nd}_1 := [1]$, $\mathsf{nd}_2 := [2]$ as well as $\mathsf{nd}_3 := [5]$ with $\mathsf{nd}_1.\mathsf{cs} = \mathsf{nd}_2.\mathsf{cs} = \mathsf{nd}_3.\mathsf{cs} = [\tuple{1,2,5}]$, are added to the queue of open nodes $\Queue$. Since all formulas have been assigned an equal fault probability, \textsc{dynamicHS} conducts a breadth-first tree construction (as displayed by the numbers \textcircled{\scriptsize i} that give the order of node labeling). That is, $\Queue$ in this case is a first-in-first-out queue. In this vein, first $[1]$ and then $[2]$ are identified as minimal diagnoses w.r.t.\ the given DPI. 

Since $\mD_{calc} = \setof{[1],[2]}$ has a cardinality of $n_{\min} = n_{\max} = 2$, the stop criterion of \textsc{dynamicHS} causes it to terminate and return $\tuple{\mD_{calc},\Queue,\mC_{calc},\Queue,\mD_{\times},\mD_{\supset},\Queue_{dup}} = \langle\ \setof{[1],[2]}$, $[[5]]$, $\setof{\tuple{1,2,5}}$, $\emptyset,\emptyset,[]\rangle$, as shown in the upper right column in Figure~\ref{fig:example:inter_onto_debug_dynamicHS_TabExDpi2}. 

Then, in Algorithm~\ref{algo:inter_onto_debug}, outside of the \textsc{dynamicHS} procedure, the first query $Q_1 = \setof{E \rightarrow \lnot A}$ is computed from the leading diagnoses set $\setof{[1],[2]}$. The q-partition $\Pt(Q_1)$ associated with $Q_1$ is $\tuple{\setof{[1]},\setof{[2]},\emptyset}$. The user's answer $u(Q_1)$ to $Q_1$ is then $\false$. Thence, the set $\mD_{out}$ is calculated from $\Pt(Q_1)$ as $\dx{}(Q_1) = \setof{[1]}$ (due to negative answer, cf.\ Remark~\ref{rem:invalidated_sets_of_q-partition_for_query_answer}), deleted from $\mD_{\checkmark} := \mD_{\checkmark} \cup \mD_{calc}$ to yield $\mD_{\checkmark} = \setof{[2]}$ and added to $\mD_{\times}$ to yield $\mD_{\times} = \setof{[1]}$.
Now, the set $\mD_{\checkmark}$ corresponds to the set of all computed (i.e.\ added to $\mD_{calc}$) minimal diagnoses w.r.t.\ the last-but-one DPI $\langle\mo,\mb,\Tp,\Tn\rangle_\RQ$ that are minimal diagnoses w.r.t.\ current DPI $\langle\mo,\mb,\Tp,\Tn\cup\setof{Q_1}\rangle_\RQ$, i.e.\ that satisfy the most recently answered query $Q_1$. The set $\mD_{\times}$ comprises all
computed (i.e.\ added to $\mD_{calc}$) minimal diagnoses w.r.t.\ the last-but-one DPI $\langle\mo,\mb,\Tp,\Tn\rangle_\RQ$ that are not minimal diagnoses w.r.t.\ current DPI $\langle\mo,\mb,\Tp,\Tn\cup\setof{Q_1}\rangle_\RQ$, i.e.\ that do not satisfy the most recently answered query $Q_1$. 

These sets $\mD_{\checkmark}$ and $\mD_{\times}$ along with the collections $\Queue$, $\Queue_{dup}$, $\mD_{\supset}$ and $\mC_{calc}$ which are unmodified outside of \textsc{dynamicHS} are used as input arguments for the second call of \textsc{dynamicHS}. Notice that, in Figures~\ref{fig:example:inter_onto_debug_dynamicHS_TabExDpi2} and \ref{fig:example:inter_onto_debug_dynamicHS_TabExDpi2_continued}, the resulting values of operations performed within \textsc{dynamicHS} are given in the righthand column above the dashed line whereas values computed outside of \textsc{dynamicHS} are given below the dashed line. 

The execution of the second call of \textsc{dynamicHS} starts with a call of the \textsc{updateTree} function. The purpose of this function is to transform the hitting set tree $T$ that was constructed by the first call of \textsc{dynamicHS} into an updated hitting set tree $T'$. Whereas the tree $T$ was used to locate minimal diagnoses w.r.t.\ the last-but-one DPI $\langle\mo,\mb,\Tp,\Tn\rangle_\RQ$, the modified tree $T'$ should serve to generate minimal diagnoses w.r.t.\ the current DPI $\langle\mo,\mb,\Tp,\Tn\cup\setof{Q_1}\rangle_\RQ$. The parameters $\mD_{\checkmark}$, $\mD_{\times}$, $\Queue$, $\Queue_{dup}$, $\mD_{\supset}$ and $\mC_{calc}$ that represent the tree $T$ (given at the top of the lefthand column in Figure~\ref{fig:example:inter_onto_debug_dynamicHS_TabExDpi2}), where $\mD_{\checkmark} \cup \mD_{\times}$ is equal to the set $\mD_{calc}$ produced by the first call of \textsc{dynamicHS}, are i.a.\ given as input arguments to the \textsc{updateTree} function. 

As a first step within \textsc{updateTree}, a redundancy check is performed for each diagnosis in $\mD_{\times}$. In this case $\mD_{\times} = \setof{\md_1}$ since $\md_1$ is the only minimal diagnosis that has been ruled out by the most recently added negative test case $Q_1$. The purpose of the redundancy check is to figure out whether $\md_1$ is redundant w.r.t.\ the current DPI and must be pruned or whether it might be extended to become a minimal diagnosis w.r.t.\ the current DPI.
  
First, the Quick Redundancy Check (QRC) $\scQX(\tuple{\setof{2,5},\mb,\Tp,\Tn\cup\setof{Q_1}}_\RQ) = \tuple{2,5}$ (line~\ref{algoline:update:qx} in \textsc{dynamicHS}) is executed for $\md_1$ which detects (line~\ref{algoline:update:X_subset_C_(QRC)} in \textsc{dynamicHS}) that $\md_1$ (and possibly some further nodes) is redundant and can be pruned. This holds since the minimal conflict set $\tuple{1,2,5}$ w.r.t.\ the last-but-one DPI $\langle\mo,\mb,\Tp,\Tn\rangle_\RQ$ is not a minimal conflict set w.r.t.\ the current DPI $\langle\mo,\mb,\Tp,\Tn\cup\setof{Q_1}\rangle_\RQ$ because $\tuple{2,5}$ returned by $\scQX$ is already a minimal conflict set w.r.t.\ the current DPI (cf.\ Proposition~\ref{prop:qx_correctness}). We call the minimal conflict set $\tuple{2,5}$ a \emph{witness of redundancy} for $\md_1$. Hence, all branches in the hitting set tree starting from the outgoing edge of $\tuple{1,2,5}$ labeled by $1$ can be safely deleted from all collections representing the new tree $T'$ (warranted that all minimal diagnoses w.r.t.\ the current DPI can still be generated from the pruned tree $T'$).

Please notice that the QRC involves only a single call of $\scQX$ using a KB of a size (here: 2) that is generally significantly smaller than $|\mo|$ (here: 7) which is roughly the size of the KB used in calls of $\scQX$ made in the \textsc{dLabel} function. Hence, the QRC will be usually very fast.

An illustration why $\tuple{2,5}$ ``replaces'' $\tuple{1,2,5}$ as a minimal conflict set w.r.t.\ the current DPI can be given as follows: First, $\tuple{1,2,5}$ is a minimal conflict set w.r.t.\ $\langle\mo,\mb,\Tp,\Tn\rangle_\RQ$ as it is a set-minimal subset of $\mo$ that entails $\setof{\lnot A} = \tn_1 \in \Tn$, there is no other negative test case in $\Tn$ except for $\tn_1$ and there is no proper subset $\mc'$ of $\tuple{1,2,5}$ where $\mc' \cup \mb \cup U_{\Tp}$ violates any $r \in \RQ$ (see example~\ref{example:analysis_TabExDpi2} for a detailed explanation). Second, formula $2$ implies in particular $E \rightarrow Y$ which, along with formula $5$ ($Y \rightarrow \lnot A$), yields $E \rightarrow \lnot A$. As the negative answer to $Q_1$ is equivalent to postulating that $\setof{E \rightarrow \lnot A}$ must not be entailed by the KB desired by the user, we have that $\tuple{2,5}$ is a conflict set w.r.t.\ $\langle\mo,\mb,\Tp,\Tn\cup\setof{Q_1}\rangle_\RQ$. As neither $\setof{2}$ nor $\setof{5}$ is a invalid KB w.r.t.\ $\langle\cdot,\mb,\Tp,\Tn\cup\setof{Q_1}\rangle_\RQ$ (cf.\ Corollary~\ref{cor:validonto_cs} and Definition~\ref{def:cs}), we have that $\tuple{2,5}$ is a \emph{minimal} conflict set w.r.t.\ $\langle\mo,\mb,\Tp,\Tn\cup\setof{Q_1}\rangle_\RQ$.

Because the QRC has been successful, yielding some witness of redundancy of $\md_1$, the Complete Redundancy Check (CRC) is no more necessary and the collections $\Queue_{dup}$, $\Queue$, $\mD_{\times}$ as well as $\mD_{\supset}$ are processed by the \textsc{prune} and \textsc{pruneQdup} functions, respectively, which involve the removal of all nodes in these collections that are redundant due to the witness $\tuple{2,5}$. In other words, all nodes are eliminated which correspond to a path in the tree that includes a node label $\mc_{old} \supset \tuple{2,5}$ and the label $e$ of the outgoing edge of $\mc_{old}$ on this path is an element of $\mc_{old}\setminus\tuple{2,5}$. Moreover, all the supersets of $\tuple{2,5}$ in $\mC_{calc}$ (here, only $\tuple{1,2,5}$) are replaced by $\tuple{2,5}$ since they are not minimal conflict sets anymore (\textsc{addSetDelSupsets}).

The pruning of nodes is expressed by dashed arrows in the pictures labeled by 'Updated Tree' in Figures~\ref{fig:example:inter_onto_debug_dynamicHS_TabExDpi2} and \ref{fig:example:inter_onto_debug_dynamicHS_TabExDpi2_continued} where the location of cutting a branch is marked by a crossline at the shaft of a dashed arrow. Furthermore, the elements of ``old'' minimal conflict sets that are no more elements of known (i.e.\ already computed) current minimal conflict sets are crossed out. As shown by the picture 'Updated Tree' in the righthand column of Figure~\ref{fig:example:inter_onto_debug_dynamicHS_TabExDpi2}, $\md_1$ is the only removed node during the pruning steps using the witness of redundancy $\tuple{2,5}$. 

Since $\mD_{\supset} = \emptyset$, \textsc{updateTree} directly jumps to the last three lines where all elements of $\mD_{\checkmark}$ are re-added to $\Queue$ in sorted order (but at the same time remain elements of $\mD_{\checkmark}$). In the figure, this is displayed by the $\stackrel{Q_1}{\Longrightarrow}$ pointing to a question mark (which stands for an open node) instead of a checkmark as in the case of the \textsc{staticHS} algorithm. Notice that, although it is a fact that all elements of $\mD_{\checkmark}$ are minimal diagnoses w.r.t.\ the current DPI, this step is necessary in order to make sure the set $\mD_{calc}$ returned by any call of \textsc{dynamicHS} actually comprises the $|\mD_{calc}|$ most probable minimal diagnoses w.r.t.\ the current DPI. For, there might be, for instance, some node that is a non-minimal diagnosis w.r.t.\ the last-but-one DPI (and is thus not an element of $\mD_{\checkmark}$), but becomes a minimal diagnosis w.r.t.\ the current DPI and has a higher probability than some node in $\mD_{\checkmark}$. Additionally, we want to point out that no calls of the \textsc{dLabel} procedure are needed for diagnoses in $\mD_{\checkmark}$ as we know their label must be $valid$. This is reflected by the test in line~\ref{algoline:dyn:node_in_Dcheckmark} in \textsc{dynamicHS}.

In the figure, all the updated collections $\mD_{\supset}$, $\mC_{calc}$, $\Queue$ as well as $\Queue_{dup}$, after being processed by \textsc{updateTree} are shown at the bottom of fields labeled by \textsc{updateTree}. We want to remark that $\mD_{\times}$ is always the empty set at the end of the execution of \textsc{updateTree} since each node in $\mD_{\times}$ gets either pruned or is reinserted into $\Queue$ as an open node. These updated collections represent the new pruned hitting set tree that can be further constructed in order to detect all and only minimal diagnoses w.r.t.\ the current DPI $\langle\mo,\mb,\Tp,\Tn\cup\setof{Q_1}\rangle_\RQ$. Note that the actions carried out by \textsc{updateTree} take place between steps \textcircled{\scriptsize 4} and \textcircled{\scriptsize 5}.

The expansion of this tree during the repeat-loop in \textsc{dynamicHS} is depicted by the picture named 'Iteration 2' in Figure~\ref{fig:example:inter_onto_debug_dynamicHS_TabExDpi2}. Namely, first (step \textcircled{\scriptsize 5}) the node $[2]$ is directly labeled by $valid$ (line~\ref{algoline:dyn:node_in_Dcheckmark}) since it is a known minimal diagnosis w.r.t.\ the current DPI (as explained before). In the sixth step, $[5]$ is labeled by the minimal conflict set $\tuple{1,2,7}$ w.r.t.\ the current DPI and three further nodes ($[5,1]$, $[5,2]$ and $[5,7]$, all with $\mathsf{nd.cs} = [\tuple{2,5}, \tuple{1,2,7}]$) are generated as successor nodes of $[5]$ and are added to $\Queue$. Now, $[5,1]$ (first-in-first-out) is the foremost node in $\Queue$ and is thus processed next and found to be a minimal diagnosis w.r.t.\ the current DPI. Therefore, \textsc{dynamicHS} terminates and returns i.a.\ the new set of leading diagnoses $\mD_{calc} = \setof{[2],[5,1]}$.

Please notice the difference here to Example~\ref{example:staticHS_simple_example_using_tabExDpi2} where the node $\setof{5,1}$ never became part of $\Queue$ in \textsc{staticHS} due to the existence of a minimal diagnosis $[1]$ w.r.t.\ the input DPI $\langle\mo,\mb,\Tp,\Tn\rangle_\RQ$ which is a proper subset of this node (and due to the fact that \textsc{staticHS} must only consider minimal diagnoses \emph{w.r.t.\ the input DPI}). In the current example, this node can only become relevant \emph{w.r.t.\ the current DPI} if all (known) diagnoses (here, only $[1]$) that are proper subsets of it have already been pruned. It should now be clear to the reader why non-minimal nodes cannot be deleted for good as in \textsc{staticHS} and why the set $\mD_{\supset}$ is necessary in \textsc{dynamicHS}.

This leading diagnosis $[5,1]$ is also the reason why the second query $Q_2 = \setof{E \rightarrow G}$ is different from the second query ($Y \rightarrow \lnot A$) calculated in Example~\ref{example:staticHS_simple_example_using_tabExDpi2}.

The execution of the algorithm continues in an analogue manner as explained so far. In the following, we just want to explain some interesting aspects in the rest of its execution:
\begin{itemize}
	\item After the query $Q_3 = \setof{Y \rightarrow \lnot A}$ (the same query as the second query in Example~\ref{example:staticHS_simple_example_using_tabExDpi2}) is answered negatively and $Q_3$ is added to $\Tn'$ yielding the current DPI $\langle\mo,\mb,\Tp,\Tn\cup\setof{Q_1,Q_2,Q_3}\rangle_\RQ$, the \textsc{updateTree} function not only prunes $[2] = \md_2 \in \mD_{\times}$ and adds $[5,7] = \md_4 \in \mD_{\checkmark}$ to $\Queue$ as we delineated above for the first query $Q_1$, but adds $[5,2] \in \mD_{\supset}$ to $\Queue$ as well. The reason for that is the deletion of the minimal diagnosis $[2]$ w.r.t.\ the last-but-one DPI $\langle\mo,\mb,\Tp,\Tn\cup\setof{Q_1,Q_2}\rangle_\RQ$ wherefore the last evidence for the non-minimality of node $[5,2]$ has been deleted. Hence, the status of $[5,2]$ as a non-minimal diagnosis is no more justified wherefore it must be added to the queue to preserve the completeness of the algorithm w.r.t.\ the finding of all minimal diagnoses w.r.t.\ the current DPI. And, indeed, $[5,2]$ is identified as minimal diagnosis ($\md_5$) in iteration 4.
	\item For each element of $\mD_{\times}$ during each execution of \textsc{updateTree} throughout the execution of Algorithm~\ref{algo:inter_onto_debug}, the Quick Redundancy Check (QRC) is successful. That is, each witness of redundancy used for pruning throughout the entire runtime of the algorithm could be determined very fast. Namely, as it is easy to see from line~\ref{algoline:update:qx} in \textsc{dynamicHS}, the KB used in the call of $\scQX$ in the QRC for some node $\mathsf{nd}$ has a size in $O((|\mathsf{nd}|-1) |\mc_{\max}|)$ where $\mc_{\max}$ is the minimal conflict set of maximum cardinality in $\mC_{calc}$. In most of the cases, $|\mathsf{nd}| \ll |\mo|$ as well as $|\mc_{\max}| \ll |\mo|$ will hold. The (usually more expensive) Complete Redundancy Check (CRC), which requires $O(|\mathsf{nd}|)$ calls to $\scQX$ with a KB of size $O(|\mc_{\max}|-1)$, is thus never employed.
	\item In this example, the same minimal diagnosis $[5,7]$ is used to compute the finally returned solution KB as in Example~\ref{example:staticHS_simple_example_using_tabExDpi2}. The only difference between both outputs is that the KB $(\mo \setminus [5,7]) \cup Q_4$ returned by \textsc{dynamicHS} in this example contains the new positive test case $Q_4 \in \Tp'$. The output by \textsc{staticHS} in	Example~\ref{example:staticHS_simple_example_using_tabExDpi2} does not contain any newly specified positive test case in $\Tp'$ (cf.\ Remark~\ref{rem:mode=static_=>_returned_solution_onto_extensible_to_be_solution_onto_w.r.t._current_DPI}), just the union of the ``original'' positive test cases in $\Tp$ (apart from that, there is not even a newly specified positive test case in Example~\ref{example:staticHS_simple_example_using_tabExDpi2}).
	\item In spite of finding the same solution diagnosis, \textsc{staticHS} requires fewer queries than \textsc{dynamicHS}. Notably, \textsc{dynamicHS} even needs a proper superset of the queries asked by \textsc{staticHS} ($Q_1, Q_2$ in Example~\ref{example:staticHS_simple_example_using_tabExDpi2} are equal to $Q_1, Q_3$ in our current example) in this case. Such a proposition however cannot be made in general since the queries formulated by \textsc{staticHS} generally differ from those formulated by \textsc{dynamicHS}. In this vein, it might just as well be the case that it takes \textsc{dynamicHS} fewer queries to finish than it takes \textsc{staticHS}, due to its advantages in tree pruning. 
	%	any diagnosis that might arise from the expansion of a subtree that has been pruned in \textsc{dynamicHS} cannot be a diagnosis
\end{itemize}
%Since $\scQX$ returns a set and the KB $\setof{2,5}$ used in $\scQX$ is the set union of all minimal conflict sets along the tree branch corresponding to $\md_1$ minus the formulas in $\md_1$, we know that the returned set $\tuple{2,5}$ must be a proper subset of some minimal conflict set along the tree branch corresponding to $\md_1$.
%
%, after performing the Quick Redundancy Check (QRC) $\scQX(\tuple{\setof{2,5},\mb,\Tp,\Tn\cup\setof{Q_1}}) = \tuple{2,5}$ for the ruled out minimal diagnosis $\md_1 \in \mD_{\times}$, \textsc{updateTree} finds out that the minimal conflict set $\tuple{1,2,5}$ w.r.t.\ the last-but-one DPI is not a minimal conflict set w.r.t.\ the current DPI because $\tuple{2,5}$ is already a minimal conflict set w.r.t.\ the current DPI. 
%
%the only justification (cf.\ Definition~\ref{def:just_set}) in $\mathsf{Just}(Q_1, \mo \cup \mb \cup U_{\Tp})$ is $\tuple{2,5}$ where $\tuple{2,5} \subseteq \mo$ and $\tuple{2,5} \cap U_{\Tp} = \emptyset$ (cf.\ Proposition~\ref{prop:cs_just}) since sentence $2$ implies in particular $E \rightarrow Y$ which, along with sentence $5$ ($Y \rightarrow \lnot A$) yields $E \rightarrow \lnot A$. 
 %
%as it is a justification in $\textsc{Just}(\tn_1, \mo \cup \mb \cup U_{\Tp})$ and $(\tuple{1,2,5} \cap \mo) \setminus U_{\Tp}$ (cf.\ Proposition~\ref{}) of the entailment 
%
%That is, the negative test case $Q_1$, which means that $\setof{E \rightarrow \lnot A}$ must not be entailed by the KB desired by the user,

All in all, the execution of Algorithm~\ref{algo:inter_onto_debug} in this example performs 
\begin{itemize}
	\item 2 full $\scQX$ calls, i.e.\ calls of $\scQX$ using the KB $\mo \setminus \mathsf{node}$ for a node $\mathsf{node}$ that actually return a minimal conflict set (there are two minimal conflict sets labeled by $C$ in Figures~\ref{fig:example:inter_onto_debug_dynamicHS_TabExDpi2} and \ref{fig:example:inter_onto_debug_dynamicHS_TabExDpi2_continued} which do not result from QRC, CRC or the minimality test of a conflict set in line~\ref{algoline:dlabel:qx_1} of \textsc{dynamicHS}),
	\item 4 fast $\scQX$ calls, i.e.\ executions of $\scQX$ within the scope of the QRC (one call of $\scQX$ each for the QRC of $\md_1$, $\md_3$, $\md_2$ and $\md_5$), 
	\item 5 validity checks, i.e.\ calls of $\scQX$ that return 'no conflict' (one check for each of the five found minimal diagnoses where the identification of diagnoses $\md_2$ at step \textcircled{\scriptsize 5}, $\md_2$ at step \textcircled{\scriptsize 9}, $\md_4$ at step \textcircled{\tiny 14} and $\md_4$ at step \textcircled{\tiny 16} does not require any call to a reasoning service by means of $\mD_{\checkmark}$, see line~\ref{algoline:dyn:node_in_Dcheckmark} in \textsc{dynamicHS}; notice that $\scQX$ does only perform a single KB validity check by \textsc{isKBValid} in case it returns 'no conflict', see Algorithm~\ref{algo:qx}) and
	\item 4 tree update processes involving 4 pruned nodes (1 per tree update), 
\end{itemize}
computes
\begin{itemize}
	\item 5 minimal diagnoses ($\md_1$, $\md_2$, $\md_4$ w.r.t.\ the input DPI and $\md_3$ and $\md_5$ w.r.t.\ some DPI resulting from the input DPI by addition of new test cases),
	\item 6 minimal conflict sets ($\tuple{1,2,5}$ as well as $\tuple{1,2,7}$ w.r.t.\ the input DPI and the subsets thereof $\tuple{2,5}$, $\tuple{2,7}$, $\tuple{5}$ and $\tuple{7}$ w.r.t.\ some DPI resulting from the input DPI by addition of new test cases) and 
	\item 4 queries and asks the user 4 logical formulas (1 per query)  
\end{itemize}
and stores
\begin{itemize}
	\item a maximum of 4 nodes (where node refers to the internal representation of a node $\mathsf{nd}$ in \textsc{dynamicHS} as a list of edge labels ($\mathsf{nd}$) and a list of node labels ($\mathsf{nd.cs}$) along a path from the root node to a leaf node).\qed
\end{itemize}	
%Finally, we want to emphasize that, in all executions of \textsc{updateTree} throughout this example, the usually very efficient QRC was successful right off and the usually more time-consuming CRC was never required.\qed
\end{example}

\newgeometry{margin=2cm}

\begin{figure*}
%%%%%%%%%%%%%%%%%%%%%%%%%%%%%%%%%%%%%%%%%%%% 6
\begin{minipage}[c]{0.45\textwidth}
\small
\begin{tabular}{l}                                          
    %DPI: \\ $\tuple{\mt,\ma,\setof{\setof{B(w)}},\setof{\setof{\neg C(w)}}}$ \\
		\textsc{updateTree}: \\
		QRC ($\md_2$): 
		$\scQX(\tuple{\setof{5},\mb,\Tp,\Tn\cup\setof{Q_1,Q_2,Q_3}}) = \tuple{5}$ \\
		$\Rightarrow\;$ \textsc{prune}: $\tuple{2,5} \rightarrow \tuple{5}$ \\
		$\Rightarrow\;$
		$\mD_{\supset} = \emptyset$, 
		$\mC_{calc} = \setof{\tuple{5}, \tuple{2,7}}$ \\
		$\quad\;$ $\Queue = [[5,2],[5,7]]$, 
		$\Queue_{dup} = []$
		%PC ($\md_2$): \\
		%$\scQX(\tuple{\setof{5},\mb,\Tp,\Tn\cup\setof{Q_1,Q_2,Q_3}}) = \tuple{5}$ \\
		%$\Rightarrow\;$ prune all subtrees starting from nodes \\
		%$\tuple{2,5}$ by outgoing edge with label $2$ \\  
		%Update: \\
		%$\mD_{\supset} = \emptyset$ \\
		%$\Queue = \setof{\setof{2,5}}$ \\
		%$\mC_{calc} = \setof{\tuple{5}, \tuple{2,7}}$
\end{tabular}
\end{minipage}
\begin{minipage}[c]{20pt}
$\Bigg>$ 
\end{minipage}
\begin{minipage}[c]{0.45\textwidth}
\small 
\xygraph{
!{<0cm,0cm>;<1.8cm,0cm>:<0cm,1.2cm>::}
%!~-{@[|(4)]}
%d=0
!{(1,4)}*+{\textcircled{\scriptsize 1}\tuple{\xcancel{2},5}^C}="c1c"
%d=1
!{(1,3) }*+{\textcircled{\scriptsize 3}\checkmark_{(\md_2)}}="d2" 
!{(3,3) }*+{\textcircled{\scriptsize 6}\tuple{2,7}^C}="c2c"
%d=2 
!{(1,2) }*+{\textcircled{\scriptsize 5}\checkmark_{(\md_2)}}="val_d2_q1"
!{(3,2) }*+{\textcircled{\tiny 10}\times_{(\supset\md_2)}}="nonmin1"
!{(4,2) }*+{\textcircled{\tiny 11}\checkmark_{(\md_4)}}="d4"
%%d=3
!{(1,1) }*+{\textcircled{\scriptsize 9}\checkmark_{(\md_2)}}="val_d2_q2"
!{(3,1) }*+{\textcircled{\tiny 12}?}="d5"
!{(4,1) }*+{\textcircled{\tiny 12}?}="val_d4_q3"
%%d=4
!{(1,0) }*+{\textcircled{\tiny 12}\times}="inv_d2_q3"
%!{(3,0) }*+{\times}="inv_d5_q4"
%!{(4,0) }*+{\checkmark}="val_d4_q4"
%d0->d1
"c1c":@{|.>}"d2"^{2}
"c1c":"c2c"^{5}
%d1->d2
"d2":@2{.>}"val_d2_q1"^{Q_1}
"c2c":"nonmin1"^{2}
"c2c":"d4"^{7}
%%d2->d3
"val_d2_q1":@2{.>}"val_d2_q2"^{Q_2}
"nonmin1":@2{->}"d5"^{Q_3}
"d4":@2{->}"val_d4_q3"^{Q_3}
%%d3->d4
%"d5":@2{->}"inv_d5_q4"^{Q_4}
%"val_d4_q3":@2{->}"val_d4_q4"^{Q_4}   
"val_d2_q2":@2{.>}"inv_d2_q3"^{Q_3}
}
\vspace{5pt}
\begin{center}
\small Updated Tree
\end{center}
\end{minipage}
\begin{minipage}[c]{10pt}
$\Bigg>$ 
\end{minipage} 

\vspace{10pt}
%%%%%%%%%%%%%%%%%%%%%%%%%%%%%%%%%%%%%%%%%%%% 7
\begin{minipage}[c]{0.45\textwidth}
\small 
\xygraph{
!{<0cm,0cm>;<1.8cm,0cm>:<0cm,1.2cm>::}
%!~-{@[|(4)]}
%d=0
!{(1,4)}*+{\textcircled{\scriptsize 1}\tuple{5}^C}="c1c"
%d=1
!{(3,3) }*+{\textcircled{\scriptsize 6}\tuple{2,7}^C}="c2c"
%d=2 
!{(3,2) }*+{\textcircled{\tiny 10}\times_{(\supset\md_2)}}="nonmin1"
!{(4,2) }*+{\textcircled{\tiny 11}\checkmark_{(\md_4)}}="d4"
%%d=3
!{(3,1) }*+{\textcircled{\tiny 13}\checkmark_{(\md_5)}}="d5"
!{(4,1) }*+{\textcircled{\tiny 14}\checkmark_{(\md_4)}}="val_d4_q3"
%%d=4
%!{(3,0) }*+{\times}="inv_d5_q4"
%!{(4,0) }*+{\checkmark}="val_d4_q4"
%d0->d1
"c1c":"c2c"^{5}
%d1->d2
"c2c":"nonmin1"^{2}
"c2c":"d4"^{7}
%%d2->d3
"nonmin1":@2{->}"d5"^{Q_3}
"d4":@2{->}"val_d4_q3"^{Q_3}
%%d3->d4
%"d5":@2{->}"inv_d5_q4"^{Q_4}
%"val_d4_q3":@2{->}"val_d4_q4"^{Q_4}   
}
\vspace{5pt}
\begin{center}
\small Iteration 4
\end{center}
\end{minipage}
\begin{minipage}[c]{20pt}
$\Bigg>$ 
\end{minipage}
\begin{minipage}[c]{0.45\textwidth}
\small
\begin{tabular}{l}                                          
    %DPI: \\ $\tuple{\mt,\ma,\setof{\setof{B(w)}},\setof{\setof{\neg C(w)}}}$ \\
		$\mD_{calc} = \setof{\md_4,\md_5} = \setof{[5,7],[5,2]}$ \\
		$\Queue = []$ \\
		$\mC_{calc} = \setof{\tuple{5}, \tuple{2,7}}$ \\
		$\mD_{\supset} = \emptyset$ \\
		$\Queue_{dup} = []$ \vspace{-7pt} \\
    \hspace{-8pt}\hdashrule{0.95\textwidth}{0.5pt}{2mm}\vspace{-4pt} \\
    $\tuple{Q_4,\Pt(Q_4)} = \tuple{\setof{E \rightarrow Z},\tuple{\setof{\md_4},\setof{\md_5},\emptyset}}$ \\
    $u(Q_4) = \true$ \\
		$\mD_{\checkmark} = \setof{\md_4}$, $\mD_{\times} = \setof{\md_5}$
		%$\mD_{\checkmark} = \setof{\md_4,\md_5} = \setof{[5,7],[2,5]}$ \\
		%$\Queue = \emptyset$ \\
		%$\mC_{calc} = \setof{\tuple{5}, \tuple{2,7}}$ \\
		%$\mD_{\supset} = \emptyset$
                         %\\
    %$\tuple{Q_4,\Pt(Q_4)} = \tuple{\setof{E \sqsubseteq Z},\tuple{\setof{[5,7]},\setof{[2,5]},\emptyset}}$ \\
    %$u(Q_4) = \true$ \\
		%$\mD_{out} = \setof{[2,5]}$
		\end{tabular}
\end{minipage}
\begin{minipage}[c]{10pt}
$\Bigg>$ 
\end{minipage} 

\vspace{10pt}
%%%%%%%%%%%%%%%%%%%%%%%%%%%%%%%%%%%%%%%%%%%% 8
\begin{minipage}[c]{0.45\textwidth}
\small
\begin{tabular}{l}                                          
    %DPI: \\ $\tuple{\mt,\ma,\setof{\setof{B(w)}},\setof{\setof{\neg C(w)}}}$ \\
		\textsc{updateTree}: \\
		QRC ($\md_5$): \\
		$\scQX(\tuple{\setof{7},\mb,\Tp\cup\setof{Q_4},\Tn\cup\setof{Q_1,Q_2,Q_3}}) = \tuple{7}$ \\
		$\Rightarrow\;$ \textsc{prune}: $\tuple{2,7} \rightarrow \tuple{7}$ \\
		$\Rightarrow\;$
		$\mD_{\supset} = \emptyset$, 
		$\mC_{calc} = \setof{\tuple{5}, \tuple{7}}$ \\
		$\quad\;$ $\Queue = [[5,7]]$, 
		$\Queue_{dup} = []$
		%PC ($\md_5$): \\
		%$\scQX(\tuple{\setof{7},\mb,\Tp\cup\setof{Q_4},\Tn\cup\setof{Q_1,Q_2,Q_3}}) = \tuple{7}$ \\
		%$\Rightarrow\;$ prune all subtrees starting from nodes \\
		%$\tuple{2,7}$ by outgoing edge with label $2$ \\  
		%Update: \\
		%$\mD_{\supset} = \emptyset$ \\
		%$\Queue = \emptyset$ \\
		%$\mC_{calc} = \setof{\tuple{5}, \tuple{7}}$
\end{tabular}
\end{minipage}
\begin{minipage}[c]{20pt}
$\Bigg>$ 
\end{minipage}
\begin{minipage}[c]{0.45\textwidth} 
\small
\xygraph{
!{<0cm,0cm>;<1.8cm,0cm>:<0cm,1.2cm>::}
%!~-{@[|(4)]}
%d=0
!{(1,4)}*+{\textcircled{\scriptsize 1}\tuple{5}^C}="c1c"
%d=1
!{(3,3) }*+{\textcircled{\scriptsize 6}\tuple{\xcancel{2},7}^C}="c2c"
%d=2 
!{(3,2) }*+{\textcircled{\tiny 10}\times_{(\supset\md_2)}}="nonmin1"
!{(4,2) }*+{\textcircled{\tiny 11}\checkmark_{(\md_4)}}="d4"
%%d=3
!{(3,1) }*+{\textcircled{\tiny 13}\checkmark_{(\md_5)}}="d5"
!{(4,1) }*+{\textcircled{\tiny 14}\checkmark_{(\md_4)}}="val_d4_q3"
%%d=4
!{(3,0) }*+{\textcircled{\tiny 15}\times}="inv_d5_q4"
!{(4,0) }*+{\textcircled{\tiny 15} ?}="val_d4_q4"
%d0->d1
"c1c":"c2c"^{5}
%d1->d2
"c2c":@{|.>}"nonmin1"^{2}
"c2c":"d4"^{7}
%%d2->d3
"nonmin1":@2{.>}"d5"^{Q_3}
"d4":@2{->}"val_d4_q3"^{Q_3}
%%d3->d4
"d5":@2{.>}"inv_d5_q4"^{Q_4}
"val_d4_q3":@2{->}"val_d4_q4"^{Q_4}   
}
\vspace{5pt}
\begin{center}
\small Updated Tree
\end{center}
\end{minipage}
\begin{minipage}[c]{10pt}
$\Bigg>$ 
\end{minipage} 

\vspace{10pt}
%%%%%%%%%%%%%%%%%%%%%%%%%%%%%%%%%%%%%%%%%%%% 9
\begin{minipage}[c]{0.45\textwidth} 
\small
\xygraph{
!{<0cm,0cm>;<1.8cm,0cm>:<0cm,1.2cm>::}
%!~-{@[|(4)]}
%d=0
!{(1,4)}*+{\textcircled{\scriptsize 1}\tuple{5}^C}="c1c"
%d=1
!{(3,3) }*+{\textcircled{\scriptsize 6}\tuple{7}^C}="c2c"
%d=2 
!{(4,2) }*+{\textcircled{\tiny 11}\checkmark_{(\md_4)}}="d4"
%%d=3
!{(4,1) }*+{\textcircled{\tiny 14}\checkmark_{(\md_4)}}="val_d4_q3"
%%d=4
%!{(3,0) }*+{\times}="inv_d5_q4"
!{(4,0) }*+{\textcircled{\tiny 16}\checkmark_{(\md_4)}}="val_d4_q4"
%d0->d1
"c1c":"c2c"^{5}
%d1->d2
"c2c":"d4"^{7}
%%d2->d3
"d4":@2{->}"val_d4_q3"^{Q_3}
%%d3->d4
"val_d4_q3":@2{->}"val_d4_q4"^{Q_4}   
}
\vspace{5pt}
\begin{center}
\small Iteration 5
\end{center}
\end{minipage}
\begin{minipage}[c]{20pt}
$\Bigg>$ 
\end{minipage}
\begin{minipage}[c]{0.45\textwidth}
\small
\begin{tabular}{l}                                          
    %DPI: \\ $\tuple{\mt,\ma,\setof{\setof{B(w)}},\setof{\setof{\neg C(w)}}}$ \\
		$\mD_{calc} = \setof{\md_4} = \setof{[5,7]}$ \\
		$\Queue = []$ \\
		$\mC_{calc} = \setof{\tuple{5}, \tuple{7}}$ \\
		$\mD_{\supset} = \emptyset$ \\
		$\Queue_{dup} = []$ \vspace{-7pt} \\
    \hspace{-8pt}\hdashrule{0.95\textwidth}{0.5pt}{2mm}\vspace{-4pt} \\
    $p_{\mD}(\md_4) = 1 $ \\ % \quad\Rightarrow\quad$ return $\md_4 \qed$
		$\Rightarrow\quad$ return the solution KB $(\mo\setminus\md_4) \cup Q_4 \quad \qed$
		%$\mD_{\checkmark} = \setof{\md_4} = \setof{[5,7]}$ \\
		%$\Queue = \emptyset$ \\
		%$\mC_{calc} = \setof{\tuple{5}, \tuple{7}}$ \\
		%$\mD_{\supset} = \emptyset$ \\
    %$p_{\mD}(\md_4) = 1 \geq 1-\sigma \;\Rightarrow\;$ return $\md_4 \qed$
		\end{tabular}
\end{minipage}
%\begin{minipage}[c]{10pt}
%$\Bigg>$ 
%\end{minipage} 

\vspace{10pt}
\caption[(Example~\ref{example:dynamicHS_small_example_using_tabExDpi2} continued) Solving the Problem of Interactive Dynamic KB Debugging]{(Example~\ref{example:dynamicHS_small_example_using_tabExDpi2} continued) Solving the problem of Interactive Dynamic KB Debugging (Problem Definition~\ref{prob_def:dynamic}) for the example DPI given by Table~\ref{tab:example2} by means of Algorithm~\ref{algo:inter_onto_debug} and \textsc{dynamicHS}.} 
\label{fig:example:inter_onto_debug_dynamicHS_TabExDpi2_continued}
\end{figure*}
\restoregeometry

\begin{example}\label{example:dynamicHS_large_example_using_tabExDpi3}
Let us now consider the (admissible) DPI $\langle\mo,\mb,\Tp,\Tn\rangle_\RQ$ given by Table~\ref{tab:example3}. We assume an expert (called user throughout this example) in the domain $Dom$ modeled by $\mo$ who wants to find a solution to Interactive Dynamic KB Debugging for the given DPI $\langle\mo,\mb,\Tp,\Tn\rangle_\RQ$ by means of Algorithm~\ref{algo:inter_onto_debug} with $mode = dynamic$.
Further, the same scenario and parameter settings as in Example~\ref{example:staticHS_complex_example_using_tabExDpi3} are supposed. That is, $n_{\min} = n_{\max} = 3$ (notice that the time limit $t$ is irrelevant in this case), $q := 1$ (cf.\ Chapter~\ref{chap:QueryGeneration}), $qsm()$ is equal to any query selection measure described in Section~\ref{sec:query_selection_measures}, $p_{\widetilde{\mo}\cup\overline{\mo}}: \widetilde{\mo}\cup\overline{\mo} \rightarrow [0,1]$ is given such that $p_{\mo}(\tax)$ for $\tax \in \mo$ resulting from the application of \textsc{getAxiomsProbs} is as given by Table~\ref{tab:example:staticHS_complex--->axiom_probs} and $\sigma := 0$.

The tree constructed and parameters computed and used by Algorithm~\ref{algo:inter_onto_debug} using \textsc{dynamicHS} are visualized by Figures~\ref{fig:example:inter_onto_debug_dynamicHS_TabExDpi3} and \ref{fig:example:inter_onto_debug_dynamicHS_TabExDpi3_continued}. 
We use the same notation as in Figures~\ref{fig:example:non-interactive_onto_debug_auto=false+nmin=infty_and_auto=true}, \ref{fig:example:non-interactive_onto_debug_auto=false+nmin=2+nmax=4_with_probs}, \ref{fig:example:inter_onto_debug_staticHS_TabExDpi2}, \ref{fig:example:inter_onto_debug_staticHS_TabExDpi3}, \ref{fig:example:inter_onto_debug_staticHS_TabExDpi3_continued}, \ref{fig:example:inter_onto_debug_dynamicHS_TabExDpi2} and \ref{fig:example:inter_onto_debug_dynamicHS_TabExDpi2_continued} which is described in Examples~\ref{example:non_interactive_debugging_with_tabExDpi2_and_without_probs}, \ref{example:non_interactive_debugging_with_tabExDpi2_and_probs}, \ref{example:staticHS_simple_example_using_tabExDpi2}, \ref{example:staticHS_complex_example_using_tabExDpi3} and \ref{example:dynamicHS_small_example_using_tabExDpi2}. 
%
%The only new notational element here is the $\Longrightarrow$ labeled by some designator of a query. That is, $\checkmark_{(\md_i)} \stackrel{Q_j}{\Longrightarrow} \checkmark$ means that $\md_i$ is still a minimal diagnosis after $Q_j$ has been answered and added to the respective set of test cases of the DPI. On the other hand, $\checkmark_{(\md_i)} \stackrel{Q_j}{\Longrightarrow} \times$ signifies that the minimal diagnosis $\md_i$ is invalidated through the addition of the answered query $Q_j$ to the respective set of test cases of the DPI. Please notice that $\Longrightarrow$ does not point at a node of the wpHS-tree. Instead, the label at which $\Longrightarrow$ points is to be understood as the new label of the node originally labeled by $\checkmark_{(\md_i)}$ from which the (first of possibly multiple) $\Longrightarrow$ goes out. This notation should help to keep track of the evolution of node labels in the wpHS-tree without needing to overload a single node by multiple different successive labels. 

After the initialization of variables, Algorithm~\ref{algo:inter_onto_debug} calls the function \textsc{getFormulaProbs} in line~\ref{algoline:inter_onto_debug:getAxiomProbs} which exploits $p_{\widetilde{\mo}\cup\overline{\mo}}()$ to calculate the function $p_{\mo}()$ giving the fault probabilities of formulas in $\mo$ (cf.\ Sections~\ref{sec:prob_space_construction}, \ref{sec:DetailedAlgorithmDescription} and Example~\ref{example:ax_prob_calc}).

Then, \textsc{dynamicHS} is called for the first time, resulting in the hitting set tree given in the first picture in Figure~\ref{fig:example:inter_onto_debug_dynamicHS_TabExDpi3}. 
As outlined by the numbers \textcircled{\scriptsize i} indicating at which point in time a node is labeled, the root node (initially the empty set) is labeled first by $\mc_1 := \tuple{1,2,5}$ and three successor nodes, namely $\mathsf{nd}_1 := [1]$, $\mathsf{nd}_2 := [2]$ as well as $\mathsf{nd}_3 := [5]$ with $\mathsf{nd}_1.\mathsf{cs} = \mathsf{nd}_2.\mathsf{cs} = \mathsf{nd}_3.\mathsf{cs} = [\tuple{1,2,5}]$, are added to the queue of open nodes $\Queue$.
Contrary to Example~\ref{example:dynamicHS_small_example_using_tabExDpi2}, where the tree was built up in breadth-first order, in this example the formula probabilities $p() := p_{\mo}()$ given by Table~\ref{tab:example:staticHS_complex--->axiom_probs} are used to assign a probability $p_{nodes}(\mathsf{n})$ to each path $\mathsf{n}$ in the tree starting from the root node (cf.\ Formula~\ref{eq:path_prob_calc} and Definition~\ref{def:p_node()}). 
In this vein, the node corresponding to the outgoing edge of $\mc_1$ labeled by the formula with the largest fault probability among all formulas in $\mc_1$ is processed next. That is, the node $[1]$ with $p_{nodes}([1]) = 0.41$ (as opposed to the nodes $[2]$ and $[5]$ with $0.25$ each) is labeled next. The \textsc{dLabel} procedure, after checking whether $[1]$ is a non-minimal diagnosis w.r.t.\ $\langle\mo,\mb,\Tp,\Tn\rangle_\RQ$ (check is negative), computes another minimal conflict set $\mc_2 := \tuple{2,4,6}$ such that $[1]\cap\mc_2 = \emptyset$ ($\mc_2$ is not hit by the node $[1]$) to constitute a label for node $[1]$. The successor nodes $[1,2]$, $[1,4]$ and $[1,6]$ of $[1]$ are generated and added to the list $\Queue$ in a way that the sorting of $\Queue$ in descending order of $p_{nodes}()$ is maintained. 

Since $[1,4]$ (0.28) as well as $[1,6]$ (0.27) have a larger probability (as per $p_{nodes}()$) than the nodes $[2]$ (0.25) and $[5]$ (0.25), $\Queue$ is given by $[[1,4],[1,6],[2],[5],[1,2]]$ when it comes to the processing of the next node. Since \textsc{dynamicHS} always treats the first node of $\Queue$ next, it identifies the first minimal diagnoses $\md_1 := [1,4]$ and $\md_2 := [1,6]$ w.r.t.\ $\langle\mo,\mb,\Tp,\Tn\rangle_\RQ$ at steps \textcircled{\scriptsize 3} and \textcircled{\scriptsize 4}, respectively. At step \textcircled{\scriptsize 5}, when node $[2]$ is processed, a minimal conflict set $\mc_3 := \tuple{1,3,4}$ is computed and set as a label for $[2]$, giving rise to the generation of three further nodes $[2,1]$, $[2,3]$ and $[2,4]$, all with $\mathsf{nd}_i.\mathsf{cs} = [\tuple{1,2,5},\tuple{1,3,4}]$. 

However, notice that not all of these new nodes are added to $\Queue$, contrary to \textsc{staticHS} (cf.\ Example~\ref{example:staticHS_complex_example_using_tabExDpi3}). For, there is already a node $[1,2]$ corresponding to the set $\setof{1,2}$ in $\Queue$. Due to the test performed in line~\ref{algoline:dyn:check_node_already_in_Q}, this duplicate node $[2,1]$ is assigned to the list $\Queue_{dup}$ which is expressed in the figure by $dup$. Since diagnoses are sets, not lists, $[1,2,\tax_1,\dots,\tax_k]$ and $[2,1,\tax_1,\dots,\tax_k]$ constitute one and the same diagnosis and it is irrelevant whether the one or the other is found. Hence, the nodes $[1,2]$ and $[2,1]$ are regarded as duplicates. Nevertheless, $\mathsf{nd}_i := [2,1]$ (with $\mathsf{nd}_i.\mathsf{cs} = [\tuple{1,2,5},\tuple{1,3,4}]$) must not be completely deleted as it might be the case that (some successor node of) $\mathsf{nd}_j := [1,2]$ (with $\mathsf{nd}_j.\mathsf{cs} = [\tuple{1,2,5},\tuple{2,4,6}]$) becomes redundant due to the eventual addition of some test case. For example, in case the reason for the redundancy of $\mathsf{nd}_j$ is given (only) by a witness of redundancy that is a subset of $\tuple{2,4,6}$, $\mathsf{nd}_j$ is pruned and replaced by the node $\mathsf{nd}_i$ which is still non-redundant.

Thence, only $[2,3]$ and $[2,4]$ are added to $\Queue$ as successor nodes of the processed node $[2]$. Next, the minimal conflict set $\mc_2 = \tuple{2,4,6}$ is reused (lines~\ref{algoline:dlabel:reuse_start}-\ref{algoline:dlabel:reuse_end} in \textsc{dLabel}) as a label for node $[5]$ with $p_{nodes}([5]) = 0.25$ and the three new nodes $[5,2]$, $[5,4]$ as well as $[5,6]$ are generated and assigned to $\Queue$ at step \textcircled{\scriptsize 7}. Then, the fourth minimal conflict set $\mc_4 := \tuple{1,5,6,8}$ is computed to label the node $[2,4]$ with $p_{nodes}([2,4]) = 0.18$ and the four new nodes $[2,4,1]$, $[2,4,5]$, $[2,4,6]$ as well as $[2,4,8]$ are generated and assigned to $\Queue$ st step \textcircled{\scriptsize 8}. At step \textcircled{\scriptsize 9}, the third minimal diagnosis $\md_3 := [5,4]$ w.r.t.\ $\langle\mo,\mb,\Tp,\Tn\rangle_\RQ$ is eventually found and added to $\mD_{calc}$ which now has reached a cardinality of $3 = n_{\min} = n_{\max}$ wherefore \textsc{dynamicHS} stops and returns i.a.\ the set of leading diagnoses $\mD_{calc} = \setof{[1,4],[1,6],[5,4]}$. The returned values are given in the lefthand column in Figure~\ref{fig:example:inter_onto_debug_dynamicHS_TabExDpi3}.

As in Example~\ref{example:staticHS_complex_example_using_tabExDpi3}, where a debugging session for the same DPI using \textsc{staticHS} is presented, the first query $Q_1$ is computed as $\setof{B \sqsubseteq K}$ and answered by $\true$ by the user. The assignment of $Q_1$ to the positive test cases of the DPI $\langle\mo,\mb,\Tp,\Tn\rangle_\RQ$ brings the opportunity to perform some significant pruning actions (within the function \textsc{updateTree} called at the beginning of the second call of \textsc{dynamicHS}). These are shown in the tree with the caption 'Updated Tree' and in the righthand column in Figure~\ref{fig:example:inter_onto_debug_dynamicHS_TabExDpi3}.

As a first step within \textsc{updateTree}, a redundancy check is performed for each diagnosis in $\mD_{\times}$. In this case $\mD_{\times} = \setof{\md_3} = \setof{[5,4]}$ since $\md_3$ is the only minimal diagnosis that has been ruled out by the most recently added positive test case $Q_1$. The purpose of the redundancy check is to figure out whether $\md_3$ is redundant w.r.t.\ the current DPI and must be pruned or whether it might be extended to become a minimal diagnosis w.r.t.\ the current DPI.
  
First, the Quick Redundancy Check (QRC) $\scQX(\tuple{\setof{1,2,6},\mb,\Tp\cup\setof{Q_1},\Tn}) = \tuple{1}$ (line~\ref{algoline:update:qx} in \textsc{dynamicHS}) is executed for $\md_3$ where the KB $\setof{1,2,6}$ used in this call of $\scQX$ is obtained by deletion of $\mathsf{node} := \md_3$ from the union of all conflict sets (the elements of $\mathsf{node.cs}$) along the path that corresponds to $\md_3$, i.e.\ $\setof{1,2,6} = (\tuple{1,2,5} \cup \tuple{2,4,6}) \setminus [5,4]$. By means of the QRC it is figured out (line~\ref{algoline:update:X_subset_C_(QRC)} in \textsc{dynamicHS}) that $\md_3$ (and possibly some further nodes) is redundant and can be pruned. This holds since the minimal conflict set $\tuple{1,2,5}$ w.r.t.\ the last-but-one DPI $\langle\mo,\mb,\Tp,\Tn\rangle_\RQ$ is not a minimal conflict set w.r.t.\ the current DPI $\langle\mo,\mb,\Tp\cup\setof{Q_1},\Tn\rangle_\RQ$ because $\tuple{1}$ returned by $\scQX$ is already a minimal conflict set w.r.t.\ the current DPI (cf.\ Proposition~\ref{prop:qx_correctness}). We call this minimal conflict set $\tuple{1}$ a \emph{witness of redundancy} for $\md_3$. Hence, all branches in the hitting set tree starting from an outgoing edge of $\tuple{1,2,5}$ labeled by $2$ or by $5$ can be safely deleted from all collections storing nodes in \textsc{dynamicHS}.

An illustration why $\tuple{1}$ ``replaces'' $\tuple{1,2,5}$ as a minimal conflict set w.r.t.\ the current DPI can be given as follows: First, $\tuple{1,2,5}$ is a minimal conflict set w.r.t.\ $\langle\mo,\mb,\Tp,\Tn\rangle_\RQ$ as it is a set-minimal subset of $\mo$ that entails $\setof{A \sqsubseteq K} = \tn_1 \in \Tn$ and there is no proper subset $\mc'$ of $\tuple{1,2,5}$ where $\mc' \cup \mb \cup U_{\Tp}$ violates any $r \in \RQ$ or entails any $\tn \in \Tn$ (see example~\ref{example:analysis_TabExDpi3} for a detailed explanation). Second, considering the current DPI $\langle\mo,\mb,\Tp\cup\setof{Q_1},\Tn\rangle_\RQ$, we have that $\tuple{1,2,5} \cup \mb \cup U_{\Tp\cup\setof{Q_1}} \models \tn_1$, too. However, $\setof{2,5} = \setof{B \sqsubseteq G, G \sqsubseteq K} \models \setof{B \sqsubseteq K} = Q_1$ implies that $\mb \cup U_{\Tp \cup \setof{Q_1}} \supseteq Q_1$ can replace the subset $\setof{2,5}$ of the conflict set $\tuple{1,2,5}$. For, formula $1$ ($A \sqsubseteq B$) along with $Q_1$ ($B \sqsubseteq K$) already entails $\tn_1$. Further, $\mb \cup U_{\Tp \cup \setof{Q_1}}$ cannot violate any negative test case $\tn_i \in \Tn$ or requirement $r_j \in \RQ$ by the admissibility of the input DPI $\langle\mo,\mb,\Tp,\Tn\rangle_\RQ$, the fact that $Q_1$ is a query, Corollary~\ref{cor:query_leaves_valid_diag}, Definition~\ref{def:admissible} and Proposition~\ref{prop:exist_diag}. Thus, by Definition~\ref{def:cs}, $\tuple{1}$ is in fact a minimal conflict set w.r.t.\ the current DPI $\langle\mo,\mb,\Tp\cup\setof{Q_1},\Tn\rangle_\RQ$.

Now, the first nice thing at this point is that $\tuple{1}$ is not only a witness of redundancy of nodes $\mathsf{nd}$ where $\tuple{1,2,5} \in \mathsf{nd.cs}$, but of each $\mathsf{nd}$ (in the tree or in the set $\Queue_{dup}$ of duplicate nodes) where $\mathsf{nd.cs}$ contains a conflict set that is a proper superset of $\tuple{1}$.
% and has already been used as a node label in the tree (i.e.\ occurs in $\mathsf{node.cs}$ for some node in the tree or in the set $\Queue_{dup}$ of duplicate nodes). 
That is, $\tuple{1}$ also replaces $\tuple{1,3,4}$ as well as $\tuple{1,5,6,8}$. This implicates that two outgoing edges (those labeled by $2$ or $5$) of $\tuple{1,2,5}$, two outgoing edges (those labeled by $3$ or $4$) of $\tuple{1,3,4}$ and three outgoing edges (those labeled by $5$, $6$ or $8$) of $\tuple{1,5,6,8}$ can be pruned.

The second nice thing that has an even more significant bearing on tree pruning than the first thing is that $\tuple{1}$ is a witness of redundancy of the conflict set that labels the root node. That is, pruning can take place at the very top of the tree and two of three subtrees rooted at successor nodes of the root node can be pruned.
That is, for instance, \emph{within} the rightmost subtree of the root node in the picture with caption 'Updated Tree' in Figure~\ref{fig:example:inter_onto_debug_dynamicHS_TabExDpi3} no pruning is possible at all since the conflict set $\tuple{2,4,6}$ labels the root node of this subtree and $\tuple{1}$ is not a subset of $\tuple{2,4,6}$. However, this subtree is still redundant since it is connected with the root node by a ``redundant'' edge labeled by $5$. As a consequence, we can observe the pruning of a total of 9 nodes (of altogether 12 nodes in the tree) in only one execution of \textsc{updateTree}. 

Now, to receive an impression of the power of tree pruning in \textsc{dynamicHS}, the reader is invited to compare the trees used in iterations 2 and 3 in the current example (the bottom left pictures in Figure~\ref{fig:example:inter_onto_debug_dynamicHS_TabExDpi3} and Figure~\ref{fig:example:inter_onto_debug_dynamicHS_TabExDpi3_continued}) with the trees used in iterations 2 and 3 in Example~\ref{example:staticHS_complex_example_using_tabExDpi3} (the bottom picture in Figure~\ref{fig:example:inter_onto_debug_staticHS_TabExDpi3} and the picture in Figure~\ref{fig:example:inter_onto_debug_staticHS_TabExDpi3_continued}) which deals with the debugging of the same DPI (just by means of \textsc{staticHS} instead of \textsc{dynamicHS}), uses the same sets of leading diagnoses in each iteration, thus the same queries, and of course the same user (that gives the same answers in both examples).

After all diagnoses of $\mD_{\checkmark}$ are added to $\Queue$ as a final action within \textsc{updateTree}, the repeat-loop of the second iteration of \textsc{dynamicHS} is entered. Here, the minimal diagnoses $\md_1$ ($p_{nodes}(\md_1) = 0.28$, step \textcircled{\tiny 11}), $\md_2$ ($0.27$, \textcircled{\tiny 12}) and $\md_4$ ($0.09$, \textcircled{\tiny 13}) are found and assigned to the empty set $\mD_{calc}$ before \textsc{dynamicHS} terminates again. Notice that only one call of the \textsc{dLabel} procedure is required in the second iteration (for node $[1,2]$) due to the test in line~\ref{algoline:dyn:node_in_Dcheckmark} of \textsc{dynamicHS} which is positive for $\md_1$ and $\md_2$ (since $\md_1, \md_2 \in \mD_{\checkmark}$). 

Once the second query $Q_2 = \setof{B \sqsubseteq \exists r.F}$ is added to the positive test cases resulting in the DPI $\langle\mo,\mb,\Tp\cup\setof{Q_1,Q_2},\Tn\rangle_\RQ$, the \textsc{updateTree} function causes the pruning of two further nodes ($\md_2 = [1,6]$ and $\md_4 = [1,2]$) leading to the continuance of only a single node ($\md_1 = [1,4]$) in the memory of \textsc{dynamicHS} (see the picture with caption 'Updated Tree' in Figure~\ref{fig:example:inter_onto_debug_dynamicHS_TabExDpi3_continued}). The reason for this is that $Q_2$ can ``replace'' the part $\setof{2,6} = \setof{B \sqsubseteq G, G \sqsubseteq \exists r.F}$ (which entails $Q_2$) of the minimal conflict set $\tuple{2,4,6}$ w.r.t.\ the last-but-one DPI $\langle\mo,\mb,\Tp\cup\setof{Q_1},\Tn\rangle_\RQ$ such that $\tuple{2,4,6} \setminus \setof{2,6} = \tuple{4}$ is already a minimal conflict set w.r.t.\ the current DPI $\langle\mo,\mb,\Tp\cup\setof{Q_1,Q_2},\Tn\rangle_\RQ$ (cf.\ the analysis of the minimal conflict set $\mc_2 = \tuple{2,4,6}$ in Example~\ref{example:analysis_TabExDpi3}). 

Since, by now, all minimal conflict sets $\tuple{1,2,5}$, $\tuple{2,4,6}$, $\tuple{1,5,6,8}$ as well as $\tuple{1,3,4}$ w.r.t.\ the input DPI $\langle\mo,\mb,\Tp,\Tn\rangle_\RQ$ have ``shrunk'' as much as to constitute only two different set-minimal sets $\tuple{1}$ and $\tuple{4}$, it is clear by Proposition~\ref{prop:mindiag_mincs} that there can be only a single minimal diagnosis $[1,4]$ w.r.t.\ the current DPI $\langle\mo,\mb,\Tp\cup\setof{Q_1,Q_2},\Tn\rangle_\RQ$. Therefore, the third iteration of \textsc{dynamicHS} terminates due to $\Queue = []$ and returns the singleton set $\mD_{calc} = \setof{[1,4]}$. Consequently, the probability $p_{\mD}([1,4]) = 1$ wherefore Algorithm~\ref{algo:inter_onto_debug} also stops executing and returns $(\mo \setminus [1,4]) \cup \tp_1 \cup Q_1 \cup Q_2$ as the (exact) solution to the Interactive Dynamic KB Debugging problem for the DPI $\langle\mo,\mb,\Tp,\Tn\rangle_\RQ$.

The advantage of \textsc{dynamicHS} in this example over \textsc{staticHS} in Example~\ref{example:staticHS_complex_example_using_tabExDpi3} in iterations 2 and 3 is that the pruning of nodes lets the algorithm automatically focus on the still relevant (i.e.\ non-redundant) parts of the tree. \textsc{staticHS}, on the other hand, is doomed to spend most of the execution time for investigating nodes that turn out to be already invalidated by some specified test case(s). As already mentioned in Example~\ref{example:staticHS_complex_example_using_tabExDpi3}, the inability of \textsc{staticHS} to ``early-prune'' incomplete branches of the tree is especially unfavorable in the last iteration of \textsc{staticHS} in case $\sigma = 0$ since all irrelevant minimal diagnoses w.r.t.\ the input DPI must first be computed before they can be ruled out.
%(1)~that $Q_2$ is an element of the entailments closure $EC(\tuple{2,4,6})$ of $\tuple{2,4,6}$ (the set of all logical sentences derivable from $\tuple{2,4,6}$) because $\setof{2,6} \models Q_2$ and (2)~that $Q_2$ is ``necessary'' for $\tuple{2,4,6}$ to be a conflict set w.r.t.\ $\langle\mo,\mb,\Tp\cup\setof{Q_1,Q_2},\Tn\rangle_\RQ$, i.e.\ $(EC(\tuple{2,4,6})\setminus Q_2)$ is a valid KB w.r.t.\ $\langle\cdot,\mb,\Tp\cup\setof{Q_1,Q_2},\Tn\rangle_\RQ$ and thus not a conflict set w.r.t.\ $\langle\mo,\mb,\Tp\cup\setof{Q_1,Q_2},\Tn\rangle_\RQ$.
%
%, i.e.\ the deletion of $Q_2$ from. an entailment of $\tuple{2,4,6} \cup \mb$ and $Q_2$ is ``necessary'' for $\tuple{2,4,6} \cup \mb \cup U_{\Tp}$ to be incoherent the minimal conflic 
%
%$\setof{2,6}$ is a justification for $Q_2$, i.e.\ $\setof{2,6} \in \mathsf{Just}(Q_2,\mo \cup \mb \cup U_{\Tp \cup \setof{Q_1,Q_2}})$

This immense upside of \textsc{dynamicHS} over \textsc{staticHS} (see the analysis in the end of Example~\ref{example:staticHS_complex_example_using_tabExDpi3}) also finds expression in the quantitative analysis of this example given next. 
All in all, the execution of Algorithm~\ref{algo:inter_onto_debug} in this example performs 
\begin{itemize}
	\item 4 full $\scQX$ calls, i.e.\ calls of $\scQX$ using the KB $\mo \setminus \mathsf{node}$ for a node $\mathsf{node}$ that actually return a minimal conflict set (there are four minimal conflict sets labeled by $C$ in Figures~\ref{fig:example:inter_onto_debug_dynamicHS_TabExDpi3} and \ref{fig:example:inter_onto_debug_dynamicHS_TabExDpi3_continued} which do not result from QRC, CRC or the minimality test of a conflict set in line~\ref{algoline:dlabel:qx_1} of \textsc{dynamicHS}),
	\item 2 fast $\scQX$ calls, i.e.\ executions of $\scQX$ within the scope of the QRC (one call of $\scQX$ each for the QRC of $\md_3$ and $\md_2$), 
	\item 4 validity checks, i.e.\ calls of $\scQX$ that return 'no conflict' (one check for each of the four found minimal diagnoses where the identification of diagnoses $\md_1$ at step \textcircled{\tiny 11}, $\md_2$ at step \textcircled{\tiny 12} and $\md_1$ at step \textcircled{\tiny 15} does not require any call to a reasoning service by means of $\mD_{\checkmark}$, see line~\ref{algoline:dyn:node_in_Dcheckmark} in \textsc{dynamicHS}; notice that $\scQX$ does only perform a single KB validity check by \textsc{isKBValid} in case it returns 'no conflict', see Algorithm~\ref{algo:qx}) and
	\item 2 tree update processes involving 11 pruned nodes (9 nodes during the first update between steps \textcircled{\tiny 10} and \textcircled{\tiny 11} and 2 nodes during the second between steps \textcircled{\tiny 14} and \textcircled{\tiny 15}), 
\end{itemize}
computes
\begin{itemize}
	\item 4 minimal diagnoses ($\md_1$, $\md_2$, $\md_3$ and $\md_4$, all w.r.t.\ the input DPI),
	\item 6 minimal conflict sets ($\tuple{1,2,5}$, $\tuple{2,4,6}$, $\tuple{1,3,4}$ and $\tuple{1,5,6,8}$ w.r.t.\ the input DPI and the subsets thereof $\tuple{1}$ and $\tuple{4}$ w.r.t.\ some DPI resulting from the input DPI by addition of new test cases) and 
	\item 2 queries and asks the user 2 logical formulas (1 per query)  
\end{itemize}
and stores
\begin{itemize}
	\item a maximum of 12 nodes (where node refers to the internal representation of a node $\mathsf{nd}$ in \textsc{dynamicHS} as a list of edge labels ($\mathsf{nd}$) and a list of node labels ($\mathsf{nd.cs}$) along a path from the root node to a leaf node).
\end{itemize}
Finally, we want to emphasize that, in all executions of \textsc{updateTree} throughout this example, the usually very efficient QRC was successful right off and the usually more time-consuming CRC was never required.\qed
\end{example}

%%%%%%%%%%%%%%%%%%%%%%%% LARGER EXAMPLE start %%%%%%%%%%%%%%%%%%%%%%%%%%
%%%%%%%%%%%%%%%%%%%%%%%%%%%%%%%%%%%%%%%%%%%%%%%%%%%%%%%%%%%%%%%%%%
%%%%%%%%%%%%%%%%%%%%%%%%%%%%%%%%%%%%%%%%%%%%%%%%%%%%%%%%%%%%%%%%%%

\newgeometry{margin=1.8cm}

\begin{figure*}
%%%%%%%%%%%%%%%%%%%%%%%%%%%%%%%%%%%%%%%%%%%% 1
\begin{minipage}[c]{0.95\textwidth} 
\small
\xygraph{
!{<0cm,0cm>;<2.0cm,0cm>:<0cm,1.0cm>::}
%!~-{@[|(4)]}
%d=0
!{(4,4)}*+{\textcircled{\scriptsize 1}\tuple{1,2,5}^C}="c1c"
%d=1
!{(1,3) }*+{\textcircled{\scriptsize 2}\tuple{2,4,6}^C}="c2c" 
!{(4,3) }*+{\textcircled{\scriptsize 5}\tuple{1,3,4}^C}="c3c" 
!{(7,3) }*+{\textcircled{\scriptsize 7}\tuple{2,4,6}^R}="c2r"
%d=2
!{(0,2) }*+{?}="d4"
!{(1,2) }*+{\textcircled{\scriptsize 3}\checkmark_{(\md_1)}}="d1"
!{(2,2) }*+{\textcircled{\scriptsize 4}\checkmark_{(\md_2)}}="d2"
!{(3,2) }*+{\textcircled{\scriptsize 6}\, dup}="dup1"
!{(4,2) }*+{?}="?2-2"
!{(5,2) }*+{\textcircled{\scriptsize 8}\tuple{1,5,6,8}^C}="c4c"
!{(6,2) }*+{?}="?2-3"
!{(7,2) }*+{\textcircled{\scriptsize 9}\checkmark_{(\md_3)}}="d3"
!{(8,2) }*+{?}="?2-4"
%d=3
!{(3,0) }*+{?}="?3-1"
!{(4,0) }*+{?}="?3-2"
!{(5,0) }*+{?}="?3-3"
!{(6,0) }*+{?}="?3-4"
%d0->d1
"c1c":"c2c"_{1}^(0.75){0.41}
"c1c":"c3c"_{2}^(0.5){0.25}
"c1c":"c2r"_{5}^(0.75){0.25}
%d1->d2
"c2c":"d4"_{2}^(0.6){0.09}
"c2c":"d1"_{4}^(0.6){0.28}
"c2c":"d2"_{6}^(0.75){0.27}
"c3c":"dup1"_{1}^(0.6){0.09}
"c3c":"?2-2"_{3}^(0.6){0.07}
"c3c":"c4c"_{4}^(0.75){0.18}
"c2r":"?2-3"_{2}^(0.6){0.06}
"c2r":"d3"_{4}^(0.6){0.18}
"c2r":"?2-4"_{6}^(0.75){0.17}
%d2->d3
"c4c":"?3-1"_{1}^(0.75){0.06}
"c4c":"?3-2"_{5}^(0.82){0.04}
"c4c":"?3-3"_{6}^(0.91){0.11}
"c4c":"?3-4"_{8}^(0.93){0.04}
}
\vspace{5pt}
\begin{center}
\small Iteration 1
\end{center}
\end{minipage}
\begin{minipage}[c]{20pt}
$\Bigg>$ 
\end{minipage}

\vspace{20pt}
\begin{minipage}[c]{0.45\textwidth}
\small  
\begin{tabular}{l}                                          
    %DPI: \\ $\tuple{\mt,\ma,\setof{\setof{B(w)}},\setof{\setof{\neg C(w)}}}$ \\
		$\mD_{calc} = \setof{\md_1, \md_2, \md_3} = \setof{[1,4],[1,6],[5,4]}$ \\
		$\Queue = [[5,6],[2,4,6],[1,2],[2,3],[5,2], $\\
		$\quad\quad\;\;\, [2,4,1],[2,4,5],[2,4,8]]$ \\
		$\mC_{calc} = \setof{\tuple{1,2,5},\tuple{2,4,6},\tuple{1,3,4},\tuple{1,5,6,8}}$ \\
		$\mD_{\supset} = \emptyset$ \\
		$\Queue_{dup} = [[2,1]]$ \\
    $\tuple{Q_1,\Pt(Q_1)} = \tuple{\setof{B \sqsubseteq K},\tuple{\setof{\md_1,\md_2},\setof{\md_3},\emptyset}}$ \\
    $u(Q_1) = \true$ \\
		$\mD_{\checkmark} = \setof{\md_1,\md_2}$, $\mD_{\times} = \setof{\md_3}$
		\end{tabular}
\end{minipage}
\begin{minipage}[c]{10pt}
$\Bigg>$ 
\end{minipage} 
%
%\vspace{10pt}
%%%%%%%%%%%%%%%%%%%%%%%%%%%%%%%%%%%%%%%%%%%%% 2
\begin{minipage}[c]{0.45\textwidth}
\small  
\begin{tabular}{l}                        
    %DPI: \\ $\tuple{\mt,\ma,\setof{\setof{B(w)}},\setof{\setof{\neg C(w)}}}$ \\
		\textsc{updateTree}: \\
		QRC ($\md_3$): 
		$\scQX(\tuple{\setof{1,2,6},\mb,\Tp\cup\setof{Q_1},\Tn}) = \tuple{1}$ \\
		$\Rightarrow\;$ \textsc{pruneQdup}/\textsc{prune}: $\tuple{1,2,5} \rightarrow \tuple{1}$, $\tuple{1,3,4} \rightarrow \tuple{1}$ \\
		%$\Rightarrow\;$ \textsc{prune}: $\tuple{1,2,5} \rightarrow \tuple{2,5}$ \\
		%PRUNE($\tuple{1,2,5},1$): 
		$\quad \bullet\;$ prune all subtrees starting  
		from nodes $\tuple{1,2,5}$ \\
		$\quad\;\;\;$ by an outgoing edge with label $2$ or $5$  \\ 
		$\quad \bullet\;$ prune all subtrees starting  
		from nodes $\tuple{1,3,4}$\\
		$\quad\;\;\;$ by an outgoing edge with label $3$ or $4$  \\ 
		%$\Rightarrow\;$ REP($\tuple{2,5}$): 
		$\quad \bullet\;$ replace by $\tuple{1}$ all node labels in the tree \\ 
		$\quad\;\;\;$ that are proper supersets of $\tuple{1}$ \\
		$\Rightarrow\;$
		$\mD_{\supset} = \emptyset$, 
		$\mC_{calc} = \setof{\tuple{1},\tuple{2,4,6}}$, \\
		$\quad\;$ $\Queue = [[1,4],[1,6],[1,2]]$, $\Queue_{dup} = []$
\end{tabular}
\end{minipage}
\begin{minipage}[c]{20pt}
$\Bigg>$ 
\end{minipage}

\vspace{20pt}
\begin{minipage}[c]{0.95\textwidth} 
\small
\xygraph{
!{<0cm,0cm>;<2.0cm,0cm>:<0cm,1.0cm>::}
%!~-{@[|(4)]}
%d=0
!{(4,4)}*+{\textcircled{\scriptsize 1}\tuple{1,\xcancel{2},\xcancel{5}}^C}="c1c"
%d=1
!{(1,3) }*+{\textcircled{\scriptsize 2}\tuple{2,4,6}^C}="c2c" 
!{(4,3) }*+{\textcircled{\scriptsize 5}\tuple{1,\xcancel{3},\xcancel{4}}^C}="c3c" 
!{(7,3) }*+{\textcircled{\scriptsize 7}\tuple{2,4,6}^R}="c2r"
%d=2
!{(0,2) }*+{?}="d4"
!{(1,2) }*+{\textcircled{\scriptsize 3}\checkmark_{(\md_1)}}="d1"
!{(2,2) }*+{\textcircled{\scriptsize 4}\checkmark_{(\md_2)}}="d2"
!{(3,2) }*+{\textcircled{\scriptsize 6}\, dup}="dup1"
!{(4,2) }*+{?}="?2-2"
!{(5,2) }*+{\textcircled{\scriptsize 8}\tuple{1,\xcancel{5},\xcancel{6},\xcancel{8}}^C}="c4c"
!{(6,2) }*+{?}="?2-3"
!{(7,2) }*+{\textcircled{\scriptsize 9}\checkmark_{(\md_3)}}="d3"
!{(8,2) }*+{?}="?2-4"
%d=3
!{(1,1) }*+{\textcircled{\tiny 10} ?}="val_d1_q1"
!{(2,1) }*+{\textcircled{\tiny 10} ?}="val_d2_q1"
!{(7,1) }*+{\textcircled{\tiny 10} \times}="inv_d3_q1"
!{(3,0) }*+{?}="?3-1"
!{(4,0) }*+{?}="?3-2"
!{(5,0) }*+{?}="?3-3"
!{(6,0) }*+{?}="?3-4"
%d0->d1
"c1c":"c2c"_{1}^(0.75){0.41}
"c1c":@{|.>}"c3c"_{2}^(0.5){0.25}
"c1c":@{|.>}"c2r"_{5}^(0.75){0.25}
%d1->d2
"c2c":"d4"_{2}^(0.6){0.09}
"c2c":"d1"_{4}^(0.6){0.28}
"c2c":"d2"_{6}^(0.75){0.27}
"c3c":"dup1"_{1}^(0.6){0.09}
"c3c":@{|.>}"?2-2"_{3}^(0.6){0.07}
"c3c":@{|.>}"c4c"_{4}^(0.75){0.18}
"c2r":"?2-3"_{2}^(0.6){0.06}
"c2r":"d3"_{4}^(0.6){0.18}
"c2r":"?2-4"_{6}^(0.75){0.17}
%d2->d3
"d1":@2{->}"val_d1_q1"^{Q_1}
"d2":@2{->}"val_d2_q1"^{Q_1}
"d3":@2{->}"inv_d3_q1"^{Q_1}
"c4c":"?3-1"_{1}^(0.75){0.06}
"c4c":@{|.>}"?3-2"_{5}^(0.82){0.04}
"c4c":@{|.>}"?3-3"_{6}^(0.91){0.11}
"c4c":@{|.>}"?3-4"_{8}^(0.93){0.04}
}
\vspace{5pt}
\begin{center}
\small Updated Tree
\end{center}
\end{minipage}
\begin{minipage}[c]{10pt}
$\Bigg>$ 
\end{minipage} 

\vspace{10pt}
%\hrule

\vspace{10pt}
%%%%%%%%%%%%%%%%%%%%%%%%%%%%%%%%%%%%%%%%%%%% 3
\begin{minipage}[c]{0.45\textwidth} 
\small
\xygraph{
!{<0cm,0cm>;<2.0cm,0cm>:<0cm,1.0cm>::}
%!~-{@[|(4)]}
%d=0
!{(3,4)}*+{\textcircled{\scriptsize 1}\tuple{1}^C}="c1c"
%d=1
!{(1,3) }*+{\textcircled{\scriptsize 2}\tuple{2,4,6}^C}="c2c" 
%d=2
!{(0,2) }*+{\textcircled{\tiny 13}\checkmark_{(\md_4)}}="d4"
!{(1,2) }*+{\textcircled{\scriptsize 3}\checkmark_{(\md_1)}}="d1"
!{(2,2) }*+{\textcircled{\scriptsize 4}\checkmark_{(\md_2)}}="d2"
%d=3
!{(1,1) }*+{\textcircled{\tiny 11} \checkmark_{(\md_1)}}="val_d1_q1"
!{(2,1) }*+{\textcircled{\tiny 12} \checkmark_{(\md_2)}}="val_d2_q1"
%d0->d1
"c1c":"c2c"_{1}^(0.6){0.41}
%d1->d2
"c2c":"d4"_{2}^(0.6){0.09}
"c2c":"d1"_{4}^(0.6){0.28}
"c2c":"d2"_{6}^(0.75){0.27}
%d2->d3
"d1":@2{->}"val_d1_q1"^{Q_1}
"d2":@2{->}"val_d2_q1"^{Q_1}
}
\vspace{5pt}
\begin{center}
\small Iteration 2
\end{center}
\end{minipage}
\begin{minipage}[c]{20pt}
$\Bigg>$ 
\end{minipage}
\begin{minipage}[c]{0.45\textwidth}
\small
\begin{tabular}{l}                                          
    %DPI: \\ $\tuple{\mt,\ma,\setof{\setof{B(w)}},\setof{\setof{\neg C(w)}}}$ \\
		$\mD_{calc} = \setof{\md_1, \md_2, \md_4} = \setof{[1,4],[1,6],[1,2]}$ \\
		$\Queue = []$ \\
		$\mC_{calc} = \setof{\tuple{1},\tuple{2,4,6}}$ \\
		$\mD_{\supset} = \emptyset$ \\
		$\Queue_{dup} = []$ \\
    $\tuple{Q_2,\Pt(Q_2)} = \tuple{\setof{B \sqsubseteq \exists r.F},\tuple{\setof{\md_1},\setof{\md_2,\md_4},\emptyset}}$ \\
    $u(Q_2) = \true$ \\
		$\mD_{\checkmark} = \setof{\md_1}$, $\mD_{\times} = \setof{\md_2,\md_4}$
		%$\mD_{\checkmark} = \setof{\md_2,\md_3} = \setof{[2],[1,5]}$ \\
		%$\Queue = \setof{\setof{2,5}, \setof{2,7}}$ \\
		%$\mC_{calc} = \setof{\tuple{2,5}, \tuple{1,2,7}}$ \\
		%$\mD_{\supset} = \emptyset$
                         %\\
    %$\tuple{Q_2,\Pt(Q_2)} = \tuple{\setof{E \sqsubseteq G},\tuple{\setof{[1,5]},\setof{[2]},\emptyset}}$ \\
    %$u(Q_2) = \false$ \\
		%$\mD_{out} = \setof{[1,5]}$
		\end{tabular}
\end{minipage}
\begin{minipage}[c]{10pt}
$\Bigg>$ 
\end{minipage}
\vspace{20pt}
\caption[(Example~\ref{example:dynamicHS_large_example_using_tabExDpi3}) Solving the Problem of Interactive Dynamic KB Debugging]{(Example~\ref{example:dynamicHS_large_example_using_tabExDpi3}) Solving the problem of Interactive Dynamic KB Debugging (Problem Definition~\ref{prob_def:dynamic}) for the example DPI given by Table~\ref{tab:example3} by means of Algorithm~\ref{algo:inter_onto_debug} and \textsc{dynamicHS}.}  
\label{fig:example:inter_onto_debug_dynamicHS_TabExDpi3}
\end{figure*}
\restoregeometry

\newgeometry{margin=2cm, top=-10cm}
\vspace{-12cm}

\begin{figure*}
%%%%%%%%%%%%%%%%%%%%%%%%%%%%%%%%%%%%%%%%%%%% 6
\begin{minipage}[c]{0.45\textwidth}
\small
\begin{tabular}{l}                                          
    %DPI: \\ $\tuple{\mt,\ma,\setof{\setof{B(w)}},\setof{\setof{\neg C(w)}}}$ \\
		\textsc{updateTree}: \\
		QRC ($\md_2$): 
		$\scQX(\tuple{\setof{2,4},\mb,\Tp\cup\setof{Q_1,Q_2},\Tn}) = \tuple{4}$ \\
		$\Rightarrow\;$ \textsc{prune}: $\tuple{2,4,6} \rightarrow \tuple{4}$ \\
		$\Rightarrow\;$
		$\mD_{\supset} = \emptyset$, 
		$\mC_{calc} = \setof{\tuple{1}, \tuple{4}}$ \\
		$\quad\;$ $\Queue = [[1,4]]$, 
		$\Queue_{dup} = []$
		%PC ($\md_3$): \\
		%$\scQX(\tuple{\setof{2,7},\mb,\Tp,\Tn\cup\setof{Q_1,Q_2}}) = \tuple{2,7}$ \\
		%$\Rightarrow\;$ prune all subtrees starting from nodes \\
		%$\tuple{1,2,7}$ by outgoing edge with label $1$ \\  
		%Update: \\
		%$\mD_{\supset} = \emptyset$ \\
		%$\mC_{calc} = \setof{\tuple{2,5}, \tuple{2,7}}$
\end{tabular}
\end{minipage}
\begin{minipage}[c]{20pt}
$\Bigg>$ 
\end{minipage}
\begin{minipage}[c]{0.45\textwidth} 
\small
\xygraph{
!{<0cm,0cm>;<2.0cm,0cm>:<0cm,1.0cm>::}
%!~-{@[|(4)]}
%d=0
!{(3,4)}*+{\textcircled{\scriptsize 1}\tuple{1}^C}="c1c"
%d=1
!{(1,3) }*+{\textcircled{\scriptsize 2}\tuple{\xcancel{2},4,\xcancel{6}}^C}="c2c" 
%d=2
!{(0,2) }*+{\textcircled{\tiny 13}\checkmark_{(\md_4)}}="d4"
!{(1,2) }*+{\textcircled{\scriptsize 3}\checkmark_{(\md_1)}}="d1"
!{(2,2) }*+{\textcircled{\scriptsize 4}\checkmark_{(\md_2)}}="d2"
%d=3
!{(1,1) }*+{\textcircled{\tiny 11} \checkmark_{(\md_1)}}="val_d1_q1"
!{(2,1) }*+{\textcircled{\tiny 12} \checkmark_{(\md_2)}}="val_d2_q1"
!{(0,1) }*+{\textcircled{\tiny 14} \times}="inv_d4_q2"
%d=4
!{(1,0) }*+{\textcircled{\tiny 14} ?}="val_d1_q2"
!{(2,0) }*+{\textcircled{\tiny 14} \times}="inv_d2_q2"
%d0->d1
"c1c":"c2c"_{1}^(0.6){0.41}
%d1->d2
"c2c":@{|.>}"d4"_{2}^(0.6){0.09}
"c2c":"d1"_{4}^(0.6){0.28}
"c2c":@{|.>}"d2"_{6}^(0.75){0.27}
%d2->d3
"d1":@2{->}"val_d1_q1"^{Q_1}
"d2":@2{.>}"val_d2_q1"^{Q_1}
"d4":@2{.>}"inv_d4_q2"^{Q_2}
%d3->d4
"val_d1_q1":@2{->}"val_d1_q2"^{Q_2}
"val_d2_q1":@2{.>}"inv_d2_q2"^{Q_2}
}
\vspace{5pt}
\begin{center}
\small Updated Tree
\end{center}
\end{minipage}
\begin{minipage}[c]{10pt}
$\Bigg>$ 
\end{minipage} 

\vspace{20pt}
%%%%%%%%%%%%%%%%%%%%%%%%%%%%%%%%%%%%%%%%%%%% 5
\begin{minipage}[c]{0.45\textwidth}
\small 
\xygraph{
!{<0cm,0cm>;<2.0cm,0cm>:<0cm,1.0cm>::}
%!~-{@[|(4)]}
%d=0
!{(3,4)}*+{\textcircled{\scriptsize 1}\tuple{1}^C}="c1c"
%d=1
!{(1,3) }*+{\textcircled{\scriptsize 2}\tuple{4}^C}="c2c" 
%d=2
!{(1,2) }*+{\textcircled{\scriptsize 3}\checkmark_{(\md_1)}}="d1"
%d=3
!{(1,1) }*+{\textcircled{\tiny 11} \checkmark_{(\md_1)}}="val_d1_q1"
%d=4
!{(1,0) }*+{\textcircled{\tiny 15} \checkmark_{(\md_1)}}="val_d1_q2"
%d0->d1
"c1c":"c2c"_{1}^(0.6){0.41}
%d1->d2
"c2c":"d1"_{4}^(0.6){0.28}
%d2->d3
"d1":@2{->}"val_d1_q1"^{Q_1}
%d3->d4
"val_d1_q1":@2{->}"val_d1_q2"^{Q_2}
}
\vspace{5pt}
\begin{center}
\small Iteration 3
\end{center}
\end{minipage}
\begin{minipage}[c]{20pt}
$\Bigg>$ 
\end{minipage}
\begin{minipage}[c]{0.45\textwidth}
\small
\begin{tabular}{l}                                          
    %DPI: \\ $\tuple{\mt,\ma,\setof{\setof{B(w)}},\setof{\setof{\neg C(w)}}}$ \\
		$\mD_{calc} = \setof{\md_1} = \setof{[1,4]}$ \\
		$\Queue = []$ \\
		$\mC_{calc} = \setof{\tuple{1}, \tuple{4}}$ \\
		$\mD_{\supset} = \emptyset$ \\
		$\Queue_{dup} = []$ \\
    $p_{\mD}(\md_1) = 1$  \\
		$\Rightarrow\quad$ return the solution KB $(\mo\setminus\md_1) \cup p_1 \cup Q_1 \cup Q_2$ \\
		$\qquad$ ($p_1$: cf.\ Table~\ref{tab:example3}) $\qed$
		%return $\md_1 \qed$
		%$\mD_{\checkmark} = \setof{\md_2,\md_4} = \setof{[2],[5,7]}$ \\
		%$\Queue = \emptyset$ \\
		%$\mC_{calc} = \setof{\tuple{2,5}, \tuple{2,7}}$ \\
		%$\mD_{\supset} = \setof{\setof{2,5}}$
                         %\\
    %$\tuple{Q_3,\Pt(Q_3)} = \tuple{\setof{Y \sqsubseteq \lnot A},\tuple{\setof{[2]},\setof{[5,7]},\emptyset}}$ \\
    %$u(Q_3) = \false$ \\
		%$\mD_{out} = \setof{[2]}$
		\end{tabular}
\end{minipage}

\vspace{20pt}
\caption[(Example~\ref{example:dynamicHS_large_example_using_tabExDpi3} continued) Solving the Problem of Interactive Dynamic KB Debugging]{(Example~\ref{example:dynamicHS_large_example_using_tabExDpi3} continued) Solving the problem of Interactive Dynamic KB Debugging (Problem Definition~\ref{prob_def:dynamic}) for the example DPI given by Table~\ref{tab:example3} by means of Algorithm~\ref{algo:inter_onto_debug} and \textsc{dynamicHS}.} \label{fig:example:inter_onto_debug_dynamicHS_TabExDpi3_continued}
\end{figure*}
\restoregeometry
%%%%%%%%%%%%%%%%%%%%%%%% LARGER EXAMPLE end %%%%%%%%%%%%%%%%%%%%%%%%%%
%%%%%%%%%%%%%%%%%%%%%%%%%%%%%%%%%%%%%%%%%%%%%%%%%%%%%%%%%%%%%%%%%%
%%%%%%%%%%%%%%%%%%%%%%%%%%%%%%%%%%%%%%%%%%%%%%%%%%%%%%%%%%%%%%%%%%

\section[Details and Correctness]{Algorithm Details and Correctness%
\sectionmark{Algorithm Details and Correctness}}
\sectionmark{Algorithm Details and Correctness}
\label{sec:TextscDynamicHSDetailsAndCorrectness}
%%%%%
%\section{\textsc{dynamicHS}: Details and Correctness}
%\label{sec:TextscDynamicHSDetailsAndCorrectness}
%%%%%
In this section we will discuss \textsc{dynamicHS} in a detailed way and give proofs of its completeness and soundness. To this end, we first give some definitions and some hints regarding the notation used in this section.

\subsection{Definitions and Notation}
\label{sec:DefinitionsAndNotation}
The \textsc{dynamicHS} algorithm will require a different storage of nodes than \textsc{staticHS} and Algorithm~\ref{algo:hs} since it will not interpret different branches with the same set of edge labels in the hitting set tree to be equivalent. 
So, \textsc{dynamicHS}, as opposed to \textsc{staticHS} and Algorithm~\ref{algo:hs}, will not discard any branch that is a duplicate branch in terms of its edge labels. Instead, a set storing these duplicate branches will be consulted each time a branch is found to be ``redundant'' and thus needs to be pruned. This strategy enables the substitution of a ``redundant'' branch by a ``non-redundant'' branch featuring an equal set of edge labels. 

%For diverse reasons that will become obvious throughout this section, \textsc{dynamicHS} will require a different storage of nodes than \textsc{staticHS} and Algorithm~\ref{algo:hs}. 
That is why a node $\mathsf{nd}$ in (the hitting set tree produced by) \textsc{dynamicHS} corresponds to the \emph{ordered} list of edge labels visited when traversing a path from the root node to some leaf node. As an attribute of $\mathsf{nd}$, $\mathsf{nd.cs}$ corresponds to the \emph{ordered} list of node labels visited when traversing a path from the root node to some leaf node. 
\begin{definition}\label{def:dyn:node_node.cs}
Let $\langle\mo,\mb,\Tp,\Tn\rangle_\RQ$ be the DPI and $\Tp'$ and $\Tn'$ the sets of positively and negatively answered queries given as an input to \textsc{dynamicHS}. Let further $\Tp''_1, \dots, \Tp''_k$ and $\Tn''_1, \dots, \Tn''_k$ such that $\Tp''_j \subseteq \Tp'$ and $\Tn''_j\subseteq\Tn'$ for $j \in \setof{1,\dots,k}$. Then we define in \textsc{dynamicHS} 
\begin{itemize}
\item a \emph{node} $\mathsf{nd} = [\tax_1, \dots, \tax_k]$ to be an (ordered) list of elements $\tax_j \in \mo$ 
\end{itemize}
where each node $\mathsf{nd}$ stores as an attribute
\begin{itemize}
\item the (ordered) list $\mathsf{nd.cs} = [\mc_1, \dots, \mc_k]$ such that 
%$\mc_j \in \minC_{\langle\mo,\mb,\Tp\cup\Tp''_j,\Tn\cup\Tn''_j\rangle_\RQ}$ 
%\fixme{say that min cs w.r.t.\ extended input dpi} 
$\mc_j$ is a minimal conflict set w.r.t.\ $\langle\mo,\mb,\Tp\cup\Tp''_j,\Tn\cup\Tn''_j\rangle_\RQ$ and $\tax_j \in \mc_j$ for all $j \in \setof{1,\dots,k}$ corresponding to the set of node labels on the path from the root node to $\mathsf{nd}$.
%\item the (ordered) list $\mathsf{nd.cs} = [\mc_1, \dots, \mc_k]$ such that $\mc_j \in \allC_{\langle\mo,\mb,\Tp,\Tn\rangle_\RQ}$ \fixme{say that min cs w.r.t.\ extended input dpi} and $\tax_j \in \mc_j$ for all $j \in \setof{1,\dots,k}$ corresponding to the set of node labels on the path from the root node to $\mathsf{nd}$.
\end{itemize}
Further, $\mathsf{nd}[i]$ refers to the $i$-th element in $\mathsf{nd}$, i.e.\ to $\tax_i$, and $\mathsf{nd.cs}[i]$ refers to the $i$-th element in $\mathsf{nd.cs}$, i.e.\ to $\mc_i$.
%\begin{itemize}
%\item each node $\mathsf{nd}$ as an \emph{ordered} list of sentences in $\mo$, i.e.\ edge labels, on the path from the root node to $\mathsf{nd}$,
%\item $\mathsf{nd}[i]$ to refer to the $i$-th element in $\mathsf{nd}$, i.e.\ the label of the outgoing edge of the $i$-th conflict set on the path from the root node to $\mathsf{nd}$,
%\item $\mathsf{nd.cs}$ to be the \emph{ordered} list of conflict sets on the path from the root node to $\mathsf{nd}$, and 
%\item $\mathsf{nd.cs}[i]$ be the $i$-th conflict set in $\mathsf{nd.cs}$ where $\mathsf{nd.cs}[1]$ is the label of the root node and $\mathsf{nd.cs}[|\mathsf{nd.cs}|]$ is the conflict set that labels the predecessor of $\mathsf{nd}$. 
%\end{itemize} 
%such that $\mathsf{nd}[i] \in \mathsf{nd.cs}[i]$ for all $i \in \setof{1,\dots,|\mathsf{nd}|}$.
Notice that conflict sets $\mathsf{nd.cs}[i]$ itself are (non-ordered) sets.

Moreover, we define
\begin{itemize}
\item $|\mathsf{nd}|$ and $|\mathsf{nd.cs}|$ to denote the number of elements in the lists $\mathsf{nd}$ and $\mathsf{nd.cs}$,
\item $\mathsf{nd}[i..k] := [\mathsf{nd}[i],\dots,\mathsf{nd}[k]]$ for $i \leq k$ and $|\mathsf{nd}|\geq k$,
\item $\mathsf{nd.cs}[i..k] := [\mathsf{nd.cs}[i],\dots,\mathsf{nd.cs}[k]]$ for $i \leq k$ and $|\mathsf{nd.cs}|\geq k$,
\item 
%terms of the form 
nodes $\mathsf{nd}$ 
and $\mathsf{nd}[i..k]$ 
appearing on the left or right side of expressions using the following set operators to be considered as (non-ordered) sets: $\supset, \supseteq, \subset, \subseteq, =, \setminus$
\end{itemize} 
We call 
\begin{itemize}
\item $\mathsf{nd}[1..k]$ 
%(with $\mathsf{nd.cs}[1..k]$) 
where ($k < |\mathsf{nd}|$) $k \leq |\mathsf{nd}|$ a \emph{(proper) subnode of $\mathsf{nd}$} 
%(with $\mathsf{nd.cs}$)} 
and 
\item $\mathsf{nd}''$ a \emph{successor (node) of $\mathsf{nd}'$} iff 
%$|\mathsf{nd}''|>|\mathsf{nd}'|$ and 
$\mathsf{nd}'$ is a proper subnode of $\mathsf{nd}''$.
\item \emph{$\mathsf{nd}$ the same node as $\mathsf{nd}'$} iff 
\begin{itemize}
	\item $|\mathsf{nd}| = |\mathsf{nd}'|$ and
	\item $\mathsf{nd}[i] = \mathsf{nd}'[i]$ for $i \in \setof{1, \dots, |\mathsf{nd}|}$ and 
	\item $\mathsf{nd.cs}[i] = \mathsf{nd}'.\mathsf{cs}[i]$ for $i \in \setof{1, \dots, |\mathsf{nd}|}$.
\end{itemize}
 
\end{itemize} 
%Further on, a node $\mathsf{nd}$ with $\mathsf{nd.cs}$ is called 
%\begin{itemize}
%\item \emph{generated} iff it is built in lines~\ref{algoline:dyn:add_ax_to_node} and \ref{algoline:dyn:add_cs_to_node.cs},
%\item \emph{processed} iff lines~\ref{algoline:dyn:get_first}-\ref{algoline:dyn:add_to_Dsupset} have been executed for $\mathsf{node} := \mathsf{nd}$,
%\end{itemize} 	
%during the execution of \textsc{dynamicHS} at any call to \textsc{dynamicHS} during the execution of Algorithm~\ref{algo:inter_onto_debug}.
\end{definition}
\begin{example}
For instance, in line~\ref{algoline:dyn:check_node_already_in_Q} of Algorithm~\ref{algo:inter_dyn_hs}, the test $\mathsf{node}_e \in \Queue$ checks whether there is some set $\mathsf{nd}$ in $\Queue$ such that $\mathsf{node}_e$ and $\mathsf{nd}$ interpreted as sets are equal. That is, $\mathsf{node}_e:=\setof{1,3,2}$ is equal to $\mathsf{nd}:=\setof{2,1,3}$ although the order of formulas is different and the ordered sets of conflict sets $\mathsf{node}_e.\mathsf{cs}$ and $\mathsf{nd.cs}$ might be different as well. Another example of this interpretation of nodes as sets can be found in line~\ref{algoline:update:qx} where $U_{\mathsf{nd.cs}}\setminus\mathsf{nd}$ refers to the set difference of the union of all sets in $\mathsf{nd.cs}$ and the set $\mathsf{nd}$. If, e.g.\ $U_{\mathsf{nd.cs}} := \setof{1,2,3,4}$ and $\mathsf{nd}:=\setof{4,2}$, the result of this set difference is $\setof{1,3}$ or, equivalently, $\setof{3,1}$. 

On the other hand, if the operator is not one of those listed above, then $\mathsf{node}$ is interpreted as an ordered set. For example, consider line~\ref{algoline:dyn:add_cs_to_node.cs} where the \textsc{add} operator is used to append a logical formula $e$ to the end of the ordered set of formulas $\mathsf{node}$. Suppose, e.g.\ $\mathsf{node} := [3,1,2]$ and $e := 4$, then the result is $[3,1,2,4]$ which is not equal to $[1,2,3,4]$.
\qed
\end{example}
%The following definition characterizes alternative paths in a (partial) wpHS-tree or, equivalently, alternative equal nodes in the representation used by \textsc{dynamicHS}. 
The following definition characterizes alternative paths in a hitting set tree produced by \textsc{dynamicHS}, i.e.\ different paths leading to the same (leaf) node in the tree.
\begin{definition}\label{def:alternative_equal_node}
Let $\mathsf{nd}$ and $\mathsf{nd}'$ be nodes in \textsc{dynamicHS} such that 
\begin{itemize}
	\item $|\mathsf{nd}'| \leq |\mathsf{nd}|$,
	\item $\mathsf{nd}' = \mathsf{nd}[1..|\mathsf{nd}'|]$ and
	\item there is some $j \in \setof{1,\dots,|\mathsf{nd}'|}$ with the property that $\mathsf{nd}'[j] \neq \mathsf{nd}[j]$ or $\mathsf{nd}'.\mathsf{cs}[j] \neq \mathsf{nd.cs}[j]$.
\end{itemize}
  
Further, let $\textsc{add}(L_1,L_2)$ be the function that outputs the list $[a_1,\dots,a_n,b_1,\dots,b_m]$ given two lists $L_1 := [a_1,\dots,a_n]$ and $L_2 := [b_1,\dots,b_m]$. 

Then we call 
\begin{itemize}
\item $\mathsf{nd}'$ an \emph{alternative subnode of $\mathsf{nd}$},
\item $\mathsf{nd}'$ a \emph{proper alternative subnode of $\mathsf{nd}$} if $|\mathsf{nd}'| < |\mathsf{nd}|$ and
\item $\mathsf{node}$ where 
\begin{itemize}
	\item $\mathsf{node}:=\textsc{add}(\mathsf{nd}', \mathsf{nd}[|\mathsf{nd}'|+1..|\mathsf{nd}|])$ and 
	\item $\mathsf{node.cs}:=\textsc{add}(\mathsf{nd}'.\mathsf{cs},\mathsf{nd.cs}[|\mathsf{nd}'.\mathsf{cs}|+1..|\mathsf{nd.cs}|])$
\end{itemize}
an \emph{alternative equal node of $\mathsf{nd}$}. 
\item In a context where $\mathsf{nd}'$ is relevant, we call $\mathsf{node}$ the \emph{alternative equal node of $\mathsf{nd}$ constructed from $\mathsf{nd}'$}.
\end{itemize}
\end{definition}
%Regarded as a set, an alternative equal node $\mathsf{node}$ of some node $\mathsf{nd}$ is equal to $\mathsf{nd}$, just the order of elements in $\mathsf{nd}$ may be different from the order of elements in $\mathsf{node}$ and the elements in the list $\mathsf{nd.cs}$ or their order may be different from $\mathsf{node.cs}$.
Regarded as a set, an alternative equal node $\mathsf{node}$ of some node $\mathsf{nd}$ is equal to $\mathsf{nd}$. There is just at least one difference between $\mathsf{node}$ and $\mathsf{nd}$ with regard to the order of elements in $\mathsf{nd}$ as opposed to the order of elements in $\mathsf{node}$ or with regard to the (order of) elements in $\mathsf{nd.cs}$ as opposed to the (order of) elements in $\mathsf{node.cs}$.
\begin{example}\label{example:alternative_equal_node_alternative_subnode}
Let $\mathsf{nd} := [1,2,3,4]$ with $\mathsf{nd.cs} := [\tuple{1,2,3},\tuple{2,6},\tuple{3,6,7},\tuple{4,5}]$. Then, $\mathsf{nd}_1 := [2,1]$ with $\mathsf{nd}_1.\mathsf{cs} := [\tuple{1,2,3},\tuple{1,4}]$ as well as $\mathsf{nd}_2 := [3,2,1]$ with $\mathsf{nd}_2.\mathsf{cs} := [\tuple{1,2,3},\tuple{2,6},\tuple{1,4}]$ are alternative subnodes of $\mathsf{nd}$. To see that $\mathsf{nd}_1$ is an alternative subnode of $\mathsf{nd}$, observe that the set-equality between $\mathsf{nd}_1 = [2,1]$ and $\mathsf{nd}[1..|\mathsf{nd}_1|] = [1,2]$ holds and $2 = \mathsf{nd}_1[j] \neq \mathsf{nd}[j] = 1$ for $j:=1$ holds. Similarly, for $\mathsf{nd}_2$, we have that the set equality between $[1,2,3]$ and $[3,2,1]$ holds and the elements on the $j$-th position for, e.g.\ $j:= 1$, are different, i.e.\ $1 \neq 3$.

These alternative subnodes of $\mathsf{nd}$ can be used to construct the following alternative equal nodes of $\mathsf{nd}$: The one obtained from $\mathsf{nd}_1$ is $\mathsf{node}_1 := [2,1,3,4]$ with $\mathsf{node}_1.\mathsf{cs} := [\tuple{1,2,3},\tuple{1,4},\tuple{3,6,7},\tuple{4,5}]$ and the one obtained from $\mathsf{nd}_2$ is $\mathsf{node}_2 := [3,2,1,4]$ with $\mathsf{node}_1.\mathsf{cs} := [\tuple{1,2,3},\tuple{2,6},\tuple{1,4},\tuple{4,5}]$.\qed
\end{example}
The following definition introduces the terminology that will be used throughout this section to refer to nodes in \textsc{dynamicHS} with certain properties.
\begin{definition}\label{def:node_denotations_1}
In \textsc{dynamicHS}, a node $\mathsf{nd}$ with $\mathsf{nd.cs}$ is called 
\begin{itemize}
\item \emph{generated} iff it is built in lines~\ref{algoline:dyn:add_ax_to_node} and \ref{algoline:dyn:add_cs_to_node.cs},
%\item \emph{active} iff it currently is an element of any of the sets $\Queue$, $\mD_{calc}$, $\mD_{\checkmark}$, $\mD_{\times}$ or $\mD_{\supset}$, 
%\item \emph{active subnode of $\mathsf{node}$} iff $\mathsf{node}$ is an active node and $\mathsf{nd}$ is a subnode of $\mathsf{node}$,
\item \emph{processed} iff lines~\ref{algoline:dyn:get_first}-\ref{algoline:dyn:add_to_Dsupset} have been executed for $\mathsf{node} := \mathsf{nd}$,
\item \emph{pruned} iff 
\begin{itemize}
\item it is found to be redundant in line~\ref{algoline:prune:redundancy_check_part2} and no node $\mathsf{nd}''=\mathsf{nd}$ is added to $S'$ in line~\ref{algoline:prune:insert_alternative_equal_node_into_S'} or
%\item deleted from the set $S$ in line~\ref{algoline:prune:delete_from_S} and no alternative equal node of $\mathsf{nd}$ is added to $S$ in line~\ref{algoline:prune:insert_alternative_equal_node_into_S} or
\item it is found to be redundant in line~\ref{algoline:pruneQdup:redundancy_check_part2} and no node $\mathsf{nd}''=\mathsf{nd}$ is added to $Dup_{new}$ in line~\ref{algoline:pruneQdup:insert_alternative_equal_node_into_Dupnew}
\end{itemize}
%\item \emph{replaced} iff it is deleted from the set $S$ in line~\ref{algoline:prune:delete_from_S} and some alternative equal node $\mathsf{nd}_{rep}$ of $\mathsf{nd}$ is added to $S$ in line~\ref{algoline:prune:insert_alternative_equal_node_into_S} 
\item \emph{replaced} iff it is found to be redundant in line~\ref{algoline:prune:redundancy_check_part2} and some node $\mathsf{nd}_{rep}=\mathsf{nd}$ is added to $S'$ in line~\ref{algoline:prune:insert_alternative_equal_node_into_S'}
\item \emph{combined-replaced} iff it is found to be redundant in line~\ref{algoline:pruneQdup:redundancy_check_part2} and some node $\mathsf{nd}_{comb,rep}=\mathsf{nd}$ is added to $Dup_{new}$ in line~\ref{algoline:pruneQdup:insert_alternative_equal_node_into_Dupnew} 
%\item \emph{reactivated} iff it is added to the set $React$ in line~\ref{algoline:prune:add_to_React},
\end{itemize} 
at any point in time during the execution of \textsc{dynamicHS} at any call to \textsc{dynamicHS} during the execution of Algorithm~\ref{algo:inter_onto_debug}.

The node $\mathsf{nd}_{rep}$ is referred to as \emph{replacement node (of $\mathsf{nd}$)} and the node $\mathsf{nd}_{comb,rep}$ is referred to as \emph{combined replacement node (of $\mathsf{nd}$)}.
\end{definition}

%---------------------------- BLOCK START  (dLabel)-----------------------------------
\subsection[The Labeling Function]{The Labeling Function in \textsc{dynamicHS}}
\label{sec:TheLabelingFunctionInTextscDynamicHS}
The following two lemmata provide an analysis of the \textsc{dLabel} function and characterize the output given by this function independently of when it is called during the execution of Algorithm~\ref{algo:inter_onto_debug}. 

The first one analyzes the case where \textsc{dLabel} returns $valid$ or $nonmin$ which means that the node for which \textsc{dLabel} was called is a diagnosis or a non-minimal diagnosis w.r.t.\ the current DPI, respectively. Further on, it states that only diagnoses w.r.t.\ the current DPI can be stored in the set $\mD_{calc}$ and only diagnoses for whose non-minimality there is evidence in terms of a diagnosis in $\mD_{calc}$ can be labeled by $nonmin$.
\begin{lemma}\label{lem:if_dlabel_returns_valid_nonmin_then}
Let the \textsc{dLabel} procedure be called at any point in time during the execution of \textsc{dynamicHS} given i.a.\ some node $\mathsf{node}$, some DPI $\tuple{\mo,\mb,\Tp,\Tn}_\RQ$, some set of positive test cases $\Tp'$ and some set of negative test cases $\Tn'$ as argument. Then the following holds: 
\begin{enumerate}[(1)]
\item If \textsc{dLabel} returns $valid$, $\mathsf{node}$ is a diagnosis w.r.t.\ the current DPI $\tuple{\mo,\mb,\Tp\cup\Tp',\Tn\cup\Tn'}_\RQ$.
\item During this execution of \textsc{dynamicHS}, $\mD_{calc}$ comprises only diagnoses w.r.t.\ the current DPI $\langle\mo,\mb$, $\Tp\cup\Tp',\Tn\cup\Tn'\rangle_\RQ$.
\item If \textsc{dLabel} returns $nonmin$, $\mathsf{node}$ is a non-minimal diagnosis w.r.t.\ the current DPI $\langle\mo$, $\mb$, $\Tp\cup\Tp',\Tn\cup\Tn'\rangle_\RQ$. 
\item At the time the label $nonmin$ is returned for $\mathsf{node}$, there is some diagnosis $\md'$ w.r.t.\ the current DPI $\tuple{\mo,\mb,\Tp\cup\Tp',\Tn\cup\Tn'}_\RQ$ such that $\md'\in\mD_{calc}$ and $\mathsf{node} \supset \md'$.
\end{enumerate}
\end{lemma}
\begin{proof}
(1): Assume that \textsc{dLabel} returns $valid$ for $\mathsf{node}$. Then, by Proposition~\ref{prop:qx_correctness}, Remark~\ref{rem:qx_with_O_setminus_node_yields_cs_wrt_O}, Corollary~\ref{cor:notions_equiv}, Corollary~\ref{cor:query_leaves_valid_diag} and the fact that the DPI $\tuple{\mo,\mb,\Tp,\Tn}_\RQ$ used in \textsc{dynamicHS} as an input to \textsc{dLabel} is the same DPI as the admissible one given as an input to Algorithm~\ref{algo:inter_onto_debug}, $\mathsf{node}$ must be a diagnosis w.r.t.\ $\tuple{\mo,\mb,\Tp\cup\Tp',\Tn\cup\Tn'}_\RQ$. This proves proposition~(1).

(2): This is a direct conclusion from proposition~(1) and the facts that nodes labeled by valid are added to the set $\mD_{calc}$ in line~\ref{algoline:dyn:add_to_Dcalc}, at the beginning of the execution of \textsc{dynamicHS}, $\mD_{calc} = \emptyset$ holds (line~\ref{algoline:dyn:Dcalc_gets_emptyset}) and $\mD_{calc}$ is modified only in line~\ref{algoline:dyn:add_to_Dcalc} throughout \textsc{dynamicHS}.

(3): At the beginning of the execution of \textsc{dynamicHS}, $\mD_{calc} = \emptyset$ (line~\ref{algoline:dyn:Dcalc_gets_emptyset}) and $\mD_{calc}$ is modified only in line~\ref{algoline:dyn:add_to_Dcalc} throughout \textsc{dynamicHS}. In line~\ref{algoline:dyn:add_to_Dcalc}, exactly those nodes are added to $\mD_{calc}$ for which the \textsc{dLabel} function returns $valid$. By the correctness of proposition~(1), only diagnoses w.r.t.\ the current DPI $\tuple{\mo,\mb,\Tp\cup\Tp',\Tn\cup\Tn'}_\RQ$ can be added to $\mD_{calc}$. 

Now, assume \textsc{dLabel} returns $nonmin$ for $\mathsf{node}$. Then, due to the fact that $\mD_{calc}$ can only comprise diagnoses w.r.t.\ the current DPI $\tuple{\mo,\mb,\Tp\cup\Tp',\Tn\cup\Tn'}_\RQ$ and $\mathsf{node}\supset \md'$ for some $\md'\in\mD_{calc}$ by line~\ref{algoline:dlabel:non-min_crit_start}, 
$\mathsf{node}$ must be a non-minimal diagnosis w.r.t.\ the current DPI $\tuple{\mo,\mb,\Tp\cup\Tp',\Tn\cup\Tn'}_\RQ$.

(4): This is a direct consequence of proposition~(3).
\end{proof}
The following lemma states that the set $\mC_{calc}$ given as an input to \textsc{dLabel} must include only minimal conflict sets, each w.r.t.\ the current DPI or some DPI including only a subset of the test cases the current DPI comprises. Moreover, it provides evidence that, in case \textsc{dLabel} returns a set, this set is a minimal conflict set w.r.t.\ the current DPI which is not hit by the node given as input to \textsc{dLabel}.
\begin{lemma}\label{lem:Ccalc_in_dlabel}
Let the \textsc{dLabel} procedure be called at any point in time during the execution of \textsc{dynamicHS} given i.a.\ some node $\mathsf{node}$, a set of sets $\mC_{calc}$, some DPI $\tuple{\mo,\mb,\Tp,\Tn}_\RQ$, some set of positive test cases $\Tp'$ and some set of negative test cases $\Tn'$ as argument. Then, 
\begin{enumerate}[(1)]
\item each element in $\mC_{calc}$ is a minimal conflict set w.r.t.\ some DPI $\tuple{\mo,\mb,\Tp\cup\Tp'',\Tn\cup\Tn''}_\RQ$ where $\Tp'' \subseteq \Tp'$ and $\Tn'' \subseteq \Tn'$ and 
\item if \textsc{dLabel} returns a set $L$, then this set $L$ is a minimal conflict set w.r.t.\ the current DPI $\langle\mo$, $\mb$, $\Tp\cup\Tp',\Tn\cup\Tn'\rangle_\RQ$ and $\mathsf{node} \cap L = \emptyset$.
\end{enumerate}
\end{lemma}
\begin{proof}
(1): At the first call to \textsc{dynamicHS}, $\mC_{calc} = \emptyset$ is given as an input argument to \textsc{dynamicHS} (lines~\ref{algoline:inter_onto_debug:var_inst_start} and \ref{algoline:inter_onto_debug:dynamicHS} in Algorithm~\ref{algo:inter_onto_debug}). The only places throughout \textsc{dynamicHS} where $\mC_{calc}$ is modified are lines~\ref{algoline:dlabel:add_set_del_supset}, \ref{algoline:dlabel:add_new_cs} and \ref{algoline:update:add_set_del_supset}. However, modifications to $\mC_{calc}$ in lines~\ref{algoline:dlabel:add_set_del_supset} and \ref{algoline:update:add_set_del_supset} can only take place in case there is already some element in $\mC_{calc}$. That is, the first element must be added to $\mC_{calc}$ in line~\ref{algoline:dlabel:add_new_cs}. 
%
%Moreover, \textsc{addSetDelSupset} only adds a sub
%only in line~\ref{algoline:dlabel:add_new_cs}, new sets which are not subsets of other sets in $\mC_{calc}$ are added
%

In line~\ref{algoline:dlabel:add_new_cs}, only minimal conflict sets w.r.t.\ some DPI $\tuple{\mo,\mb,\Tp\cup\Tp'',\Tn\cup\Tn''}_\RQ$ are added to $\mC_{calc}$ where $\Tp'' \subseteq \Tp'$ and $\Tn'' \subseteq \Tn'$ since the call to \textsc{dLabel} might have taken place during some prior execution of \textsc{dynamicHS} during the execution of Algorithm~\ref{algo:inter_onto_debug}. In order to reach line~\ref{algoline:dlabel:add_new_cs}, $\scQX$ called with the DPI $\tuple{\mo\setminus\mathsf{node},\mb,\Tp\cup\Tp'',\Tn\cup\Tn''}_\RQ$ as argument must not return 'no conflict' (line~\ref{algoline:dlabel:qx_2}). That is, a minimal conflict set $L\neq\emptyset$ w.r.t.\ $\tuple{\mo,\mb,\Tp\cup\Tp'',\Tn\cup\Tn''}_\RQ$ is computed in line~\ref{algoline:dlabel:qx_2} by Propostition~\ref{prop:qx_correctness}, Remark~\ref{rem:qx_with_O_setminus_node_yields_cs_wrt_O}, Corollary~\ref{cor:query_leaves_valid_diag} and the fact that the DPI $\tuple{\mo,\mb,\Tp,\Tn}_\RQ$ used in \textsc{dynamicHS} as an input to \textsc{dLabel} is the same DPI as the admissible one given as an input to Algorithm~\ref{algo:inter_onto_debug}.

In lines~\ref{algoline:dlabel:add_set_del_supset} and \ref{algoline:update:add_set_del_supset}, the following is true: (*) Only minimal conflict sets that are proper subsets of elements already in $\mC_{calc}$ can be added to $\mC_{calc}$. 
In the case of line~\ref{algoline:dlabel:add_set_del_supset}, (*) is true due to the following reasons: 
In order to reach line~\ref{algoline:dlabel:add_set_del_supset}, $\scQX(\tuple{\mc,\mb,\Tp\cup\Tp'',\Tn\cup\Tn''}_\RQ) = X \neq \mc$ must hold for some element $\mc\in\mC_{calc}$. Since $\mC_{calc}$ is never changed in Algorithm~\ref{algo:inter_onto_debug} between two calls to \textsc{dynamicHS}, $\mC_{calc}$ comprises only conflict sets w.r.t.\ the current DPI or previous DPIs (including fewer test cases than the current one). Moreover, a minimal conflict set $\mc$ can only shrink after the addition of new test cases to the DPI for which it was computed by Proposition~\ref{prop:changes_in_conflict_sets_after_testcase_added}. Hence, the newly added element $X$ must be a proper subset of the existing element $\mc$ in $\mC_{calc}$. That $X$ is a minimal conflict set w.r.t.\ the DPI $\tuple{\mo,\mb,\Tp\cup\Tp'',\Tn\cup\Tn''}_\RQ$ follows from $\scQX(\tuple{\mc,\mb,\Tp\cup\Tp'',\Tn\cup\Tn''}_\RQ) = X$, Propostition~\ref{prop:qx_correctness}, Remark~\ref{rem:qx_with_O_setminus_node_yields_cs_wrt_O}, Corollary~\ref{cor:query_leaves_valid_diag} and the fact that the DPI $\tuple{\mo,\mb,\Tp,\Tn}_\RQ$ used in \textsc{dynamicHS} as an input to \textsc{dLabel} is the same DPI as the admissible one given as an input to Algorithm~\ref{algo:inter_onto_debug}.

In the case of line~\ref{algoline:update:add_set_del_supset}, (*) is true due to the following reasons: Due to Lemmata~\ref{lem:quick_prune_check} and \ref{lem:complete_prune_check}, $quickPC = \true$ or $completePC = \true$ can only hold if $X$ is a witness of redundancy of $\mathsf{nd}$. By Definition~\ref{def:redundant_node}, a witness of redundancy is a conflict set w.r.t.\ the current DPI which is a proper subset of some conflict set that has been used as a label in $\mathsf{nd.cs}$. However, each label in $\mathsf{nd.cs}$ must be an element of $\mC_{calc}$ due to lines~\ref{algoline:dlabel:reuse_start}, \ref{algoline:dlabel:add_new_cs} and \ref{algoline:dyn:add_cs_to_node.cs}. 

(2): That, in case \textsc{dLabel} returns a set $L$, it returns a minimal conflict set w.r.t.\ the current DPI is a consequence from the inference in the proof of proposition~(1). We still need to show that $L \cap \mathsf{node} = \emptyset$. 

If \textsc{dLabel} returns in line~\ref{algoline:dlabel:return_new_cs}, we can derive from the fact that $L$ is the output of the call $\scQX(\langle\mo\setminus$ $\mathsf{node}$, $\mb,\Tp\cup\Tp',\Tn\cup\Tn'\rangle_\RQ)$, Proposition~\ref{prop:qx_correctness} and Definition~\ref{def:cs} that $L \subseteq \mo \setminus \mathsf{node}$ which implies that $L \cap \mathsf{node} = \emptyset$. 

If \textsc{dLabel} returns in line~\ref{algoline:dlabel:return_C} or line~\ref{algoline:dlabel:return_X}, then the return can be executed only if the check $\mc \cap \mathsf{node} = \emptyset$ is true in line~\ref{algoline:dlabel:if_C_cap_node=emptyset}. By the argumentation in the proof of proposition~(1), for the returned set $L$ it must hold that $L \subseteq \mc$. Hence, $L \cap \mathsf{node} = \emptyset$ is satisfied. 
\end{proof}
As a simple conclusion from Lemma~\ref{lem:Ccalc_in_dlabel}, we have that the argument $X$ passed to the \textsc{prune} function called within \textsc{dLabel} is a minimal conflict set w.r.t.\ the current DPI:
\begin{corollary}\label{cor:prune_only_called_with_min_cs_in_dlabel}
Assume the execution of some call to \textsc{dynamicHS} during the execution of Algorithm~\ref{algo:inter_onto_debug} using the current DPI $DPI$.
Anytime \textsc{prune} is called within \textsc{dLabel}, the input $X$ given to it is a minimal conflict set w.r.t.\ $DPI$.
\end{corollary}
\begin{proof}
Assume the execution of some call to \textsc{dynamicHS} during the execution of Algorithm~\ref{algo:inter_onto_debug} using the current DPI $DPI$. Then, Lemma~\ref{lem:Ccalc_in_dlabel} says that the set $X$ returned in line~\ref{algoline:dlabel:return_X} is a minimal conflict set w.r.t.\ $DPI$. Since $X$ is not modified by any of the functions \textsc{prune} and \textsc{addSetDelSupsets}, we obtain the proposition of this corollary.
\end{proof}
From this we derive that the input $X$ passed to \textsc{pruneQdup} called within \textsc{dLabel} must be a minimal conflict set w.r.t.\ the current DPI:
\begin{corollary}\label{cor:pruneQdup_only_called_with_min_cs_in_dlabel}
Assume the execution of some call to \textsc{dynamicHS} during the execution of Algorithm~\ref{algo:inter_onto_debug} using the current DPI $DPI$.
Anytime \textsc{pruneQdup} is called within \textsc{dLabel}, the input $X$ given to it is a minimal conflict set w.r.t.\ $DPI$.
\end{corollary}
\begin{proof}
This corollary is a direct consequence of Corollary~\ref{cor:prune_only_called_with_min_cs_in_dlabel} and the fact that the argument $X$ given to \textsc{pruneQdup} is the same argument $X$ that is given to \textsc{pruneQdup} (none of these functions modifies $X$).
\end{proof}
%---------------------------- BLOCK END (dLabel) -----------------------------------
%---------------------------- BLOCK START  (changes in cs)----------------------------------
\subsection{Impact of Answered Queries on Conflict Sets}
\label{sec:ImpactOfAnsweredQueriesOnConflictSets}
After one call to \textsc{dynamicHS} in Algorithm~\ref{algo:inter_onto_debug} returns, the set $\mD_{calc}$ (called $\mD_{\checkmark}$ in Algorithm~\ref{algo:inter_onto_debug}) returned by \textsc{dynamicHS} is used as a set of leading diagnoses w.r.t.\ the current DPI in order to compute a query. After the answered query is incorporated into the DPI, a new call to \textsc{dynamicHS} for this new current DPI is made.

As we have learned from Lemmata~\ref{lem:if_dlabel_returns_valid_nonmin_then} and \ref{lem:Ccalc_in_dlabel}, the new call to \textsc{dynamicHS} considers only minimal diagnoses and minimal conflict sets w.r.t.\ the new current DPI. Therefore, the next proposition investigates the impact of the addition of the answered query as a new test case on the set of minimal conflict sets w.r.t.\ the new current DPI. Concretely, it claims that the transition from a DPI to a new DPI extended by a test case does change the set of minimal conflict sets, that each (minimal) conflict set remains a (not necessarily minimal) conflict set and that minimal conflict sets cannot grow in size. 

It is however important to notice that some ``new'' minimal conflict set might emerge in the course of this DPI-transition which is not in a subset-relationship with any existing minimal conflict set.
\begin{proposition}\label{prop:changes_in_conflict_sets_after_testcase_added}
Let $\mD$ be a set of minimal diagnoses w.r.t.\ $\langle\mo,\mb,\Tp,\Tn\rangle_\RQ$ and $Q \in \mQ_{\mD,\langle\mo,\mb,\Tp,\Tn\rangle_\RQ}$. Further, let either $\Tp' = \Tp \cup \setof{Q}$ or $\Tn' = \Tn \cup \setof{Q}$. Then it holds that
\begin{enumerate}[(1)]
\item $\minC_{\langle\mo,\mb,\Tp,\Tn\rangle_\RQ} \neq \minC_{\langle\mo,\mb,\Tp',\Tn'\rangle_\RQ}$,
\item each conflict set w.r.t.\ $\langle\mo,\mb,\Tp,\Tn\rangle_\RQ$ is a conflict set w.r.t.\ $\langle\mo,\mb,\Tp',\Tn'\rangle_\RQ$,
\item each minimal conflict set w.r.t.\ $\langle\mo,\mb,\Tp,\Tn\rangle_\RQ$ is a conflict set w.r.t.\ $\langle\mo,\mb,\Tp',\Tn'\rangle_\RQ$,
\item there are no $\mc \in \minC_{\langle\mo,\mb,\Tp,\Tn\rangle_\RQ}$ and $\mc' \in \minC_{\langle\mo,\mb,\Tp',\Tn'\rangle_\RQ}$ such that $\mc \subset \mc'$,
\item if there is a subset-relationship between $\mc \in \minC_{\langle\mo,\mb,\Tp,\Tn\rangle_\RQ}$ and $\mc' \in \minC_{\langle\mo,\mb,\Tp',\Tn'\rangle_\RQ}$, then $\mc' = \mc$ or $\mc' \subset \mc$.
\end{enumerate}
\end{proposition}
\begin{proof}
(1): Assume the opposite, namely that $\minC_{\langle\mo,\mb,\Tp,\Tn\rangle_\RQ} = \minC_{\langle\mo,\mb,\Tp',\Tn'\rangle_\RQ}$. Then, by Proposition~\ref{prop:mindiag_mincs}, $\minD_{\langle\mo,\mb,\Tp,\Tn\rangle_\RQ} = \minD_{\langle\mo,\mb,\Tp',\Tn'\rangle_\RQ}$ must be true. This however is a contradiction to Definition~\ref{def:query} and the fact that $Q$ is a query.

(2): Let $\mc$ be a conflict set w.r.t.\ $\langle\mo,\mb,\Tp,\Tn\rangle_\RQ$. Then $\mc \cup \mb \cup U_{\Tp}$ violates some $x \in \RQ \cup \Tn$. If $\Tp' = \Tp \cup \setof{Q}$ holds, then, by monotonicity of $\mathcal{L}$, $\mc \cup \mb \cup U_{\Tp \cup \setof{Q}}$ violates some $x \in \RQ \cup \Tn$, i.e.\ $\mc$ is a conflict set w.r.t.\ $\langle\mo,\mb,\Tp',\Tn'\rangle_\RQ$. Otherwise, if $\Tn' = \Tn \cup \setof{Q}$ is given, then $\mc \cup \mb \cup U_{\Tp}$ violates some $x \in \RQ \cup \Tn \subset \RQ \cup \Tn'$, i.e.\ $\mc$ is a conflict set w.r.t.\ $\langle\mo,\mb,\Tp',\Tn'\rangle_\RQ$.

(3): This is a direct consequence of (2), since each minimal conflict set w.r.t.\ $\langle\mo,\mb,\Tp,\Tn\rangle_\RQ$ is a conflict set w.r.t.\ $\langle\mo,\mb,\Tp,\Tn\rangle_\RQ$.

(4): Since, by (3), each minimal conflict set w.r.t.\ $\langle\mo$, $\mb$, $\Tp,\Tn\rangle_\RQ$ is also a conflict set w.r.t.\ $\langle\mo$, $\mb,\Tp',\Tn'\rangle_\RQ$, there cannot be a minimal conflict set $\mc'$ w.r.t.\ $\langle\mo,\mb,\Tp',\Tn'\rangle_\RQ$ which is a proper superset of a minimal conflict set w.r.t.\ $\langle\mo,\mb,\Tp,\Tn\rangle_\RQ$ as this would imply non-minimality of $\mc'$ w.r.t.\ $\langle\mo,\mb,\Tp',\Tn'\rangle_\RQ$.

(5): This proposition is a direct consequence of (4).
\end{proof}
Given the existence of some non-empty minimal conflict set w.r.t.\ an admissible DPI $DPI$, the extension of the test cases of $DPI$ by a query yields a new DPI $DPI'$ for which all minimal conflict sets are non-empty:
%In the following we prove that the extension of the test cases of an admissible DPI $DPI$ by a query yields a new DPI $DPI'$ for which all minimal conflict sets are non-empty if there is some non-empty minimal conflict set w.r.t.\ $DPI$. 
%
%extended by a query w.r.t.\ $DPI$
%and whose test cases are extended by a single query w.r.t.\ some set of minimal diagnoses w.r.t.\ $DPI$ yields a new DPI $DPI'$ such that each minimal conflict set
\begin{proposition}\label{prop:if_DPI_cs_neq_emptyset_then_DPI+1_cs_neq_emptyset}
Let $\langle\mo,\mb,\Tp,\Tn\rangle_\RQ$ and $\langle\mo,\mb,\Tp',\Tn'\rangle_\RQ$ be two DPIs such that $\langle\mo,\mb,\Tp,\Tn\rangle_\RQ$ is admissible and $\Tp' \supseteq \Tp$ and $\Tn' \supseteq \Tn$ and $|\Tp' \cup \Tn'| = |\Tp \cup \Tn| + 1$. Let further $\mc \neq \emptyset$ be a minimal conflict set w.r.t.\ $\langle\mo,\mb,\Tp,\Tn\rangle_\RQ$ and $Q \in (\Tp' \cup \Tn')\setminus(\Tp \cup \Tn)$ be a query w.r.t.\ some $\mD \subseteq \minD_{\langle\mo,\mb,\Tp,\Tn\rangle_\RQ}$ and $\langle\mo,\mb,\Tp,\Tn\rangle_\RQ$. Then, for each minimal conflict set $\mc'$ w.r.t.\ $\langle\mo,\mb,\Tp',\Tn'\rangle_\RQ$ it holds that $\mc' \neq \emptyset$.
\end{proposition}
\begin{proof}
Assume there is some minimal conflict set $\mc'$ w.r.t.\ $\langle\mo,\mb,\Tp',\Tn'\rangle_\RQ$ such that $\mc' = \emptyset$. This implies that there cannot be a minimal conflict set $\mc''$ w.r.t.\ $\langle\mo,\mb,\Tp',\Tn'\rangle_\RQ$ which is not the empty set because $\mc'$ would be a proper subset of $\mc''$, which would be a contradiction to the minimality of $\mc''$.

Due to Corollary~\ref{cor:query_leaves_valid_diag} and the fact that a query $Q$ w.r.t.\ some $\mD \subseteq \minD_{\langle\mo,\mb,\Tp,\Tn\rangle_\RQ}$ and $\langle\mo,\mb,\Tp,\Tn\rangle_\RQ$ is added to $\langle\mo,\mb,\Tp,\Tn\rangle_\RQ$ in order to obtain $\langle\mo,\mb,\Tp',\Tn'\rangle_\RQ$, we have that $\langle\mo,\mb,\Tp',\Tn'\rangle_\RQ$ must be admissible.

By Corollary~\ref{cor:notions_equiv}, $\mo$ cannot be valid w.r.t.\ $\langle\cdot,\mb,\Tp,\Tn\rangle_\RQ$ since $\emptyset$ cannot be a diagnosis w.r.t.\ $\langle\mo$, $\mb,\Tp,\Tn\rangle_\RQ$ by Proposition~\ref{prop:mindiag_mincs} and the fact that $\mc$ is a non-empty minimal conflict set w.r.t.\ $\langle\mo,\mb,\Tp,\Tn\rangle_\RQ$. From this we can infer that $\mo$ cannot be valid w.r.t.\ $\langle\cdot,\mb,\Tp',\Tn'\rangle_\RQ$ as $\Tp' \supseteq \Tp$ and $\Tn' \supseteq \Tn$.

Now, by Proposition~\ref{prop:cs_admissible}, there must be some minimal conflict set w.r.t.\ $\langle\mo,\mb,\Tp',\Tn'\rangle_\RQ$ which is not the empty set, contradiction.
\end{proof}
%---------------------------- BLOCK END  (changes in cs)----------------------------------
%---------------------------- BLOCK START  (changes in diags)----------------------------------
\subsection{Impact of Answered Queries on Diagnoses}
\label{sec:ImpactOfAnsweredQueriesOnDiagnoses}
Next, we analyze what influence answered queries that are added as new test cases to the current DPI have on the (minimal) diagnoses w.r.t.\ this DPI. The first lemma assures that each DPI constructed during the execution of Algorithm~\ref{algo:inter_onto_debug} must be admissible as a consequence of the postulated admissibility of the DPI given as an initial input to Algorithm~\ref{algo:inter_onto_debug}.
\begin{lemma}\label{lem:current_dpi_admissible}
Let $\langle\mo,\mb,\Tp,\Tn\rangle_\RQ$ be the DPI and $\Tp'$ and $\Tn'$ the sets of positively and negatively answered queries given as an input to \textsc{dynamicHS}. Then, the DPI $\langle\mo,\mb,\Tp\cup\Tp',\Tn\cup\Tn'\rangle_\RQ$ is admissible.
\end{lemma}
\begin{proof}
The admissibility of $\langle \mo,\mb,\Tp\cup\Tp',\Tn\cup\Tn'\rangle_\RQ$ follows from the fact that $\langle\mo,\mb,\Tp,\Tn\rangle_\RQ$ is the (coercively) admissible input DPI of Algorithm~\ref{algo:inter_onto_debug}, Corollary~\ref{cor:query_leaves_valid_diag} which reveals that admissibility of a DPI is preserved under the addition of a query to the test cases of the DPI and the fact that $\Tp'$ as well as $\Tn'$ are sets of queries. The latter holds because \textsc{calcQuery} (Algorithm~\ref{algo:inter_onto_debug}, line~\ref{algoline:inter_onto_debug:calc_query}) computes only queries and the only place where $\Tp'$ and $\Tn'$ are modified is lines~\ref{algoline:inter_onto_debug:add_pos_tc}-\ref{algoline:inter_onto_debug:param_update_end} where only sets returned by \textsc{calcQuery} are added to $\Tp'$ and $\Tn'$. 
%This holds by Proposition~\ref{prop:qx_correctness} and the admissibility of $\langle \mathsf{nd.cs}[i] \setminus \mathsf{nd}[i],\mb,\Tp\cup\Tp',\Tn\cup\Tn'\rangle_\RQ$. The latter is equivalent to the admissibility of $\langle \mo,\mb,\Tp\cup\Tp',\Tn\cup\Tn'\rangle_\RQ$ since admissibility does not depend on the parameter $\mo$ of a DPI (\cf.\ Definition~\ref{def:admissible}). Finally, the admissibility of $\langle \mo,\mb,\Tp\cup\Tp',\Tn\cup\Tn'\rangle_\RQ$ follows from the fact that $\langle\mo,\mb,\Tp,\Tn\rangle_\RQ$ is the (coercively) admissible input DPI of Algorithm~\ref{algo:inter_onto_debug}, Corollary~\ref{cor:query_leaves_valid_diag} which reveals that admissibility of a DPI is preserved under the addition of a query to the test cases of the DPI and the fact that $\Tp'$ as well as $\Tn'$ are sets of queries.
\end{proof}

The next proposition confirms the restrictive character of test cases. That is, any extension of a current DPI through the addition of a test case cannot lead to a set of (all) diagnoses w.r.t.\ the new DPI that is a superset of the set of (all) diagnoses w.r.t.\ the current DPI. We want to point out that this is not necessarily true for the set of \emph{minimal} diagnoses. 
\begin{proposition}\label{prop:diag_for_new_dpi_is_diag_for_old_dpi}
Let $\langle\mo,\mb,\Tp,\Tn\rangle_\RQ$ and $\langle\mo,\mb,\Tp',\Tn'\rangle_\RQ$ be two DPIs such that $\Tp' \supseteq \Tp$ and $\Tn' \supseteq \Tn$. Then, each diagnosis w.r.t.\ $\langle\mo,\mb,\Tp',\Tn'\rangle_\RQ$ is also a diagnosis w.r.t.\ $\langle\mo,\mb,\Tp,\Tn\rangle_\RQ$.
\end{proposition}
\begin{proof}
Let $\md'\in \allD_{\langle\mo,\mb,\Tp',\Tn'\rangle_\RQ}$. 
%and assume that $\md'\notin \allD_{\langle\mo,\mb,\Tp,\Tn\rangle_\RQ}$. 
Then, by Corollary~\ref{cor:notions_equiv} and Definition~\ref{def:target_ont}, $(\mo \setminus \md') \cup \mb \cup U_{\Tp'}$ does not violate any $x\in\RQ\cup\Tn'$.
Since however formulas, in particular those in $U_{\Tp' \setminus \Tp}$, that are \emph{added} to a KB cannot invalidate any (unwanted) entailments, in particular those in $\Tn'$, and cannot resolve any inconsistencies or incoherencies by the monotonicity of $\mathcal{L}$, we can conclude that $(\mo \setminus \md') \cup \mb \cup U_{\Tp}$ does not violate any $x\in\RQ\cup\Tn'$ either.
Since $\Tn' \supseteq \Tn$, non-violation of any test case in $\Tn'$ implies non violation of any test case in $\Tn$ also. Consequently, $(\mo \setminus \md') \cup \mb \cup U_{\Tp}$ does not violate any $x\in\RQ\cup\Tn$ and entails all $\tp \in \Tp$ (due to $U_{\Tp}$) wherefore $\md' \in \allD_{\langle\mo,\mb,\Tp,\Tn\rangle_\RQ}$ due to Corollary~\ref{cor:notions_equiv} and Definition~\ref{def:target_ont}. 
\end{proof}
As a consequence of this, each minimal diagnosis w.r.t.\ the new DPI is a diagnosis w.r.t.\ the current DPI, i.e.\ either a minimal or a non-minimal diagnosis w.r.t.\ the current DPI.
\begin{corollary}\label{cor:min_diag_for_new_dpi_is_diag_for_old_dpi}
Let $\langle\mo,\mb,\Tp,\Tn\rangle_\RQ$ and $\langle\mo,\mb,\Tp',\Tn'\rangle_\RQ$ be two DPIs such that $\Tp' \supseteq \Tp$ and $\Tn' \supseteq \Tn$. Then, each minimal diagnosis w.r.t.\ $\langle\mo,\mb,\Tp',\Tn'\rangle_\RQ$ is also a diagnosis w.r.t.\ $\langle\mo,\mb,\Tp,\Tn\rangle_\RQ$.
\end{corollary}
\begin{proof}
Since Proposition~\ref{prop:diag_for_new_dpi_is_diag_for_old_dpi} holds for all diagnoses w.r.t.\ $\langle\mo,\mb,\Tp',\Tn'\rangle_\RQ$, it also holds for all minimal diagnoses w.r.t.\ $\langle\mo,\mb,\Tp',\Tn'\rangle_\RQ$ since each minimal diagnosis is a diagnosis.
\end{proof}
Adding a test case to a DPI cannot make minimal diagnoses shrink:
\begin{proposition}\label{prop:adding_testcase_cannot_make_min_diags_shrink}
Let $\langle\mo,\mb,\Tp,\Tn\rangle_\RQ$ and $\langle\mo,\mb,\Tp',\Tn'\rangle_\RQ$ be two DPIs such that $\Tp' \supseteq \Tp$ and $\Tn' \supseteq \Tn$ and let $\md \in \minD_{\langle\mo,\mb,\Tp,\Tn\rangle_\RQ}$. Then, for all $\md' \in \minD_{\langle\mo,\mb,\Tp',\Tn'\rangle_\RQ}$, it holds that $\md' \not\subset \md$.
\end{proposition}
\begin{proof}
Let $\md \in \minD_{\langle\mo,\mb,\Tp,\Tn\rangle_\RQ}$ and let $\md' \in \minD_{\langle\mo,\mb,\Tp',\Tn'\rangle_\RQ}$ such that $\Tp' \supseteq \Tp, \Tn' \supseteq \Tn$ and suppose $\md' \subset \md$. 
%Then, $(\mo \setminus \md') \cup \mb \cup U_{\Tp'}$ does not violate any $x\in\RQ\cup\Tn'$. Since however sentences, in particular those in $U_{\Tp' \setminus \Tp}$, that are \emph{added} to a KB cannot invalidate any (unwanted) entailments, in particular those in $\Tn'$, and cannot resolve any inconsistencies or incoherences by monotonicity of $\mathcal{L}$, we can conclude that $(\mo \setminus \md') \cup \mb \cup U_{\Tp}$ does not violate any $x\in\RQ\cup\Tn'$ either.
%Since $\Tn' \supseteq \Tn$, non violation of any test case in $\Tn'$ implies non violation of any test case in $\Tn$ also. Consequently, $(\mo \setminus \md') \cup \mb \cup U_{\Tp}$ does not violate any $x\in\RQ\cup\Tn$ wherefore $\md'$ is a diagnosis w.r.t.\ $\langle\mo,\mb,\Tp,\Tn\rangle_\RQ$. 
By Proposition~\ref{prop:diag_for_new_dpi_is_diag_for_old_dpi}, $\md'$ must be a diagnosis w.r.t.\ $\langle\mo,\mb,\Tp,\Tn\rangle_\RQ$.
By $\md' \subset \md$, this is a contradiction to the premise that $\md \in \minD_{\langle\mo,\mb,\Tp,\Tn\rangle_\RQ}$, i.e.\ that $\md$ is minimal.
\end{proof}
In fact, it even holds that each ``new'' minimal diagnosis (which is not a minimal diagnosis w.r.t.\ the current DPI) resulting from the addition of a test case to the current DPI
%a minimal diagnosis w.r.t.\ the new DPI which is not a minimal diagnosis w.r.t.\ the current DPI, 
must be a proper superset of some minimal diagnosis w.r.t.\ the current DPI. In other words, a minimal diagnosis w.r.t.\ the new DPI is either a minimal diagnosis w.r.t.\ the current DPI or a proper superset of some minimal diagnosis w.r.t.\ the current DPI.
\begin{proposition}\label{prop:after_adding_testcase_new_min_diag_is_equal_or_superset_of_old_min_diag}
Let $\langle\mo,\mb,\Tp,\Tn\rangle_\RQ$ and $\langle\mo,\mb,\Tp',\Tn'\rangle_\RQ$ be two DPIs such that $\Tp' \supseteq \Tp$ and $\Tn' \supseteq \Tn$ and let $\md' \in \minD_{\langle\mo,\mb,\Tp',\Tn'\rangle_\RQ}$ and $\md' \notin \minD_{\langle\mo,\mb,\Tp,\Tn\rangle_\RQ}$. Then, there is some $\md \in \minD_{\langle\mo,\mb,\Tp,\Tn\rangle_\RQ}$ such that $\md \subset \md'$.
\end{proposition}
\begin{proof}
By Corollary~\ref{cor:min_diag_for_new_dpi_is_diag_for_old_dpi}, we know that $\md' \in \minD_{\langle\mo,\mb,\Tp',\Tn'\rangle_\RQ}$ is a diagnosis w.r.t.\ $\langle\mo,\mb,\Tp,\Tn\rangle_\RQ$. If $\md'$ is already a minimal diagnosis w.r.t.\ $\langle\mo,\mb,\Tp,\Tn\rangle_\RQ$, then the proposition holds. Otherwise, there must be some $\md \subset \md'$ such that $\md$ is a minimal diagnosis w.r.t.\ $\langle\mo,\mb,\Tp,\Tn\rangle_\RQ$.
\end{proof}
Addition of a query to whatever test case set of a DPI $DPI$ implies that the set of all diagnoses w.r.t.\ the new DPI is a proper subset of all diagnoses w.r.t.\ $DPI$:
\begin{corollary}\label{cor:adding_query_to_DPI_implies_that_allD_wrt_new_DPI_is_proper_subset_of_allD_wrt_old_DPI}
Let $\langle\mo,\mb,\Tp,\Tn\rangle_\RQ$ and $\langle\mo,\mb,\Tp',\Tn'\rangle_\RQ$ be two DPIs such that 
\begin{itemize}
	\item $\Tp' \supseteq \Tp$ and $\Tn' \supseteq \Tn$,
	\item $|\Tp'| = |\Tp| + 1$ or $|\Tn'| = |\Tn| + 1$, but not both, and
	\item $(\Tp' \cup \Tn') \setminus (\Tp \cup \Tn) = \setof{Q}$ where $Q$ is a query w.r.t.\ some set $\mD \subseteq \minD_{\langle\mo,\mb,\Tp,\Tn\rangle_\RQ}$ and $\langle\mo,\mb,\Tp,\Tn\rangle_\RQ$.
\end{itemize}
Then, $\allD_{\langle\mo,\mb,\Tp',\Tn'\rangle_\RQ} \subset \allD_{\langle\mo,\mb,\Tp,\Tn\rangle_\RQ}$ holds.
\end{corollary}
\begin{proof}
By Proposition~\ref{prop:diag_for_new_dpi_is_diag_for_old_dpi} we have that $\allD_{\langle\mo,\mb,\Tp',\Tn'\rangle_\RQ} \subseteq \allD_{\langle\mo,\mb,\Tp,\Tn\rangle_\RQ}$. Since $\langle\mo,\mb,\Tp',\Tn'\rangle_\RQ$ results from $\langle\mo,\mb,\Tp,\Tn\rangle_\RQ$ by the addition of the query $Q$ w.r.t.\ some set $\mD$ and $\langle\mo,\mb,\Tp,\Tn\rangle_\RQ$ to either $\Tp$ or $\Tn$, we conclude by Definition~\ref{def:query} that at least one minimal diagnosis $\md$ w.r.t.\ $\langle\mo,\mb,\Tp,\Tn\rangle_\RQ$ in $\mD$ is not a minimal diagnosis w.r.t.\ $\langle\mo,\mb,\Tp',\Tn'\rangle_\RQ$. 
Assume, $\md$ is a non-minimal diagnosis w.r.t.\ $\langle\mo,\mb,\Tp',\Tn'\rangle_\RQ$. In this case, there must be some $\md' \subset \md$ such that $\md'$ is a minimal diagnosis w.r.t.\ $\langle\mo,\mb,\Tp',\Tn'\rangle_\RQ$. This is a contradiction to Proposition~\ref{prop:adding_testcase_cannot_make_min_diags_shrink}. Consequently, $\md \notin \allD_{\langle\mo,\mb,\Tp',\Tn'\rangle_\RQ}$. Hence, $\md \in \allD_{\langle\mo,\mb,\Tp,\Tn\rangle_\RQ} \setminus \allD_{\langle\mo,\mb,\Tp',\Tn'\rangle_\RQ}$. By $\allD_{\langle\mo,\mb,\Tp',\Tn'\rangle_\RQ} \subseteq \allD_{\langle\mo,\mb,\Tp,\Tn\rangle_\RQ}$, the proposition of the corollary follows.
\end{proof}
\subsection[Redundant Nodes]{Redundant Nodes in \textsc{dynamicHS}}
\label{sec:RedundantNodesInTextscDynamicHS}
The following result constitutes the basis for the definition of a redundant node we give in the next section. It is already stated in \cite{Reiter87}, but without a proof. It testifies that the set of all minimal hitting sets of a collection $F$ of sets remains steady if elements that are not set-minimal sets in $F$ are deleted from $F$. By Proposition~\ref{prop:mindiag_mincs}, the same must hold for the set of all minimal diagnoses of the collection of all minimal conflict sets w.r.t.\ some DPI $DPI$. 
That is, considering only minimal hitting sets of minimal conflict sets w.r.t.\ $DPI$ is sufficient for completeness of a hitting set tree algorithm concerning the finding of all minimal diagnoses w.r.t.\ $DPI$. 

However, we proved by Proposition~\ref{prop:changes_in_conflict_sets_after_testcase_added} that existing conflict sets will tend to shrink gradually through the specification of new test cases. This implicates that more and more nodes $\mathsf{nd}_i$ stored by \textsc{dynamicHS} will have the property that $\mathsf{nd}_i.\mathsf{cs}$ will include non-minimal conflict sets w.r.t.\ the current DPI which constitutes the first of two criteria that are together sufficient for a safe pruning of $\mathsf{nd}_i$. By safe pruning we mean the deletion of a node without eliminating any minimal diagnoses w.r.t.\ the current DPI.
%%
%This implicates that certain nodes corresponding to a subtree starting from a non-minimal
%
 %does not affect the completeness of 
%That is, considering no hitting sets of non-minimal conflict sets w.r.t.\ $DPI$ does not affect the completeness of a hitting set algorithm concerning the finding of all minimal diagnoses w.r.t.\ $DPI$.
%only minimal hitting sets of all minimal conflict sets w.r.t.\ $DPI$
\begin{proposition}\label{prop:leave_out_nonmin_cs_same_min_hs}%\cite{Reiter87}
If $F$ is a collection of sets, and if $S \in F$ and $S' \in F$ such that $S \subset S'$, then $F_{sub} := F \setminus \setof{S'}$ has the same minimal hitting sets as $F$.
\end{proposition}
\begin{proof}
Let $\md$ be a minimal hitting set of $F_{sub}$, then $\md$ is a hitting set of $F$ since $\md \cap S \neq \emptyset$ holds which implies by $S \subset S'$ that $\md \cap S' \neq \emptyset$. Assume that $\md$ is a non-minimal hitting set of $F$, i.e.\ that a subset $\md' \subset \md$ is a hitting set of $F$. Then, however, by minimality of $\md$ w.r.t.\ $F_{sub}$ we have that not all sets in $F_{sub}$ are hit by $\md'$ and thus, by $F_{sub} \subset F$, that not all sets in $F$ can be hit by $\md'$, contradiction. Thus, each minimal hitting set of $F_{sub}$ is also a minimal hitting set of $F$.
  
Let $\md$ be a minimal hitting set of $F$, then $\md$ is clearly a hitting set of $F_{sub} \subset F$. Suppose that $\md$ is a non-minimal hitting set of $F_{sub}$, i.e.\ that a proper subset of $\md$ is a hitting set of $F_{sub}$. Let $\md' \subset \md$ be a subset-minimal such subset of $\md$. That is, $\md'$ is a minimal hitting set of $F_{sub}$. Since $\md$ is a minimal hitting set of $F$, $\md'$ is not a (minimal) hitting set of $F$, but a minimal hitting set of $F_{sub}$. This is a contradiction to the already proven fact that any minimal hitting set of $F_{sub}$ is also a minimal hitting set of $F$.
\end{proof}
%---------------------------- BLOCK END  (changes in diags)----------------------------------
%
%---------------------------- BLOCK START  (redundancy)----------------------------------
%In the following, we characterize redundant nodes, i.e.\ paths in a (partial) wpHS-tree, 
%%or nodes in the algorithms' internal representations, respectively, 
%that can be pruned without affecting completeness, i.e.\ the finding of all minimal diagnoses w.r.t.\ the given DPI (cf.\ \cite{Reiter87}).
Assume the first criterion for a safe pruning of a node $\mathsf{nd}_i$, namely the existence of some non-minimal conflict set w.r.t.\ the current DPI in $\mathsf{nd}_i.\mathsf{cs}$, is met. Then, we have not yet any evidence that $\mathsf{nd}_i$ is obsolete since for each of the non-minimal conflict sets in $\mathsf{nd}_i.\mathsf{cs}$ there must be one (or multiple) proper subset(s) which is a minimal conflict set w.r.t.\ the current DPI. Let $\mc_{\lnot min}$ be one particular non-minimal conflict set in $\mathsf{nd}_i.\mathsf{cs}$ and let $\mc$ be the particular proper subset of $\mc_{\lnot min}$ that is the first ``witness'' found by \textsc{dynamicHS} which documents the non-minimality of $\mc_{\lnot min}$. Then $\mc_{\lnot min}$ can be split into two disjoint parts, namely 
%the set of sentences $\mc$ that $\mc_{\lnot\min}$ shares with 
$\mc$ and the set of formulas $\overline{\mc}$ that $\mc_{\lnot min}$ does not share with $\mc$. 

Now, the second criterion for a safe pruning of $\mathsf{nd}_i$ is about whether $\mathsf{nd}_i$ hits $\overline{\mc}$. If so, then $\mathsf{nd}_i$ is not a (partial) hitting set of \emph{only} minimal conflict sets w.r.t.\ the current DPI. Put another way, this means that, under the assumption that a wpHS-tree was constructed using only the ``static'' current DPI, then the label $\mc_{\lnot min}$ would have never been produced and hence the node $\mathsf{nd}_i$ could have never been generated. Eventually, by the considerations made in Sections~\ref{sec:CorrectnessOfAlgorithmHs} and \ref{sec:CorrectnessOfTextscStaticHS}, we know that such a static hitting set tree algorithm is complete although not taking into account nodes like $\mathsf{nd}_i$. 

These thoughts motivate the following definition of a redundant node\footnote{We adopt the term ``redundant'' from \cite{Reiter87} where is was informally used in the same context.}. 
\begin{definition}\label{def:redundant_node}
Let $\langle\mo,\mb,\Tp,\Tn\rangle_\RQ$ be the DPI and $\Tp'$ and $\Tn'$ the sets of positively and negatively answered queries given as an input to \textsc{dynamicHS}. Further, let 
%Let $\langle\mo,\mb,\Tp,\Tn\rangle_\RQ$ be a DPI and 
$\mathsf{nd}$ be a node in \textsc{dynamicHS}. Then we call $\mathsf{nd}$ a \emph{redundant node w.r.t.\ $\langle\mo,\mb,\Tp\cup\Tp',\Tn\cup\Tn'\rangle_\RQ$} iff there is
\begin{itemize}
	\item some $r \in\setof{1,\dots,|\mathsf{nd}|}$ and
	\item some minimal conflict set $\mc$ w.r.t.\ $\langle\mo,\mb,\Tp\cup\Tp',\Tn\cup\Tn'\rangle_\RQ$
\end{itemize}
  such that 
\begin{itemize}
	\item $\mc \subset \mathsf{nd.cs}[r]$ and
	\item $\mathsf{nd}[r] \in \mathsf{nd.cs}[r]\setminus \mc$.
\end{itemize}
Moreover, $\mc$ is called a \emph{witness of redundancy of $\mathsf{nd}$}.
%Let $\langle\mo,\mb,\Tp,\Tn\rangle_\RQ$ be a DPI, $\mathsf{nd} = [\tax_1, \dots, \tax_j]$ be a list of elements $\tax_k \in \mo$ and $\mathsf{nd.cs} = [\mc_1, \dots, \mc_j]$ be a list of elements such that $\mc_k \in \allC_{\langle\mo,\mb,\Tp,\Tn\rangle_\RQ}$ and $\tax_k \in \mc_k$ for all $k \in \setof{1,\dots,j}$. Further, let $\mathsf{nd}[i]$ be the $i$-th element of $\mathsf{nd}$ and $\mathsf{nd.cs}[i]$ be the $i$-th element of $\mathsf{nd.cs}$.
%
%Then we call $\mathsf{nd}$ a \emph{redundant node w.r.t.\ $\langle\mo,\mb,\Tp,\Tn\rangle_\RQ$} iff there is some $r \in\setof{1,\dots,j}$ and some minimal conflict set $\mc$ w.r.t.\ $\langle\mo,\mb,\Tp,\Tn\rangle_\RQ$ such that $\mc$ is a proper subset of $\mathsf{nd.cs}[r]$ and $\mathsf{nd}[r] \in \mathsf{nd.cs}[r]\setminus \mc$. Moreover, $\mc$ is called a \emph{witness of redundancy of $\mathsf{nd}$}.
\end{definition}
A node $\mathsf{node}$ in \textsc{dynamicHS} can be only redundant w.r.t.\ a DPI $DPI$ if $\mathsf{node.cs}$ comprises some non-minimal conflict set w.r.t.\ $DPI$:
\begin{corollary}\label{cor:node_with_only_min_cs_on_path_is_not_redundant}
Let $\langle\mo,\mb,\Tp,\Tn\rangle_\RQ$ be the DPI and $\Tp'$ and $\Tn'$ the sets of positively and negatively answered queries given as an input to \textsc{dynamicHS}. Further, let $\mathsf{nd}$ be a node in \textsc{dynamicHS}
%Let $\langle\mo,\mb,\Tp,\Tn\rangle_\RQ$ be a DPI and $\mathsf{nd}$ a node in \textsc{dynamicHS} 
such that $\mathsf{nd}[i] \in \minC_{\langle\mo,\mb,\Tp\cup\Tp',\Tn\cup\Tn'\rangle_\RQ}$ for all $i \in \setof{1,\dots,|\mathsf{nd}|}$. Then $\mathsf{nd}$ is not a redundant node w.r.t.\ $\langle\mo,\mb,\Tp\cup\Tp',\Tn\cup\Tn'\rangle_\RQ$.
%Let $\langle\mo,\mb,\Tp,\Tn\rangle_\RQ$ be a DPI, $\mathsf{nd} = [\tax_1, \dots, \tax_j]$ be a list of elements $\tax_k \in \mo$ and $\mathsf{nd.cs} = [\mc_1, \dots, \mc_j]$ be a list of elements such that $\mc_k \in \minC_{\langle\mo,\mb,\Tp,\Tn\rangle_\RQ}$ and $\tax_k \in \mc_k$ for all $k \in \setof{1,\dots,j}$. Then $\mathsf{nd}$ is not a redundant node w.r.t.\ $\langle\mo,\mb,\Tp,\Tn\rangle_\RQ$.
\end{corollary}
\begin{proof}
Since $\mathsf{nd.cs}$ comprises only minimal conflict sets w.r.t.\ $\langle\mo,\mb,\Tp,\Tn\rangle_\RQ$, there cannot be any $\mc \in \minC_{\langle\mo,\mb,\Tp,\Tn\rangle_\RQ}$ such that $\mc \subset \mathsf{nd.cs}[i]$ for some $i$.
\end{proof}
A node that is redundant w.r.t.\ some DPI $DPI$ remains redundant w.r.t.\ any $DPI'$ that includes a superset of the test cases $DPI$ includes: 
\begin{lemma}\label{lem:redundant_node_remains_redundant_after_adding_new_testcases}
Let $\langle\mo,\mb,\Tp,\Tn\rangle_\RQ$ and $\langle\mo,\mb,\Tp',\Tn'\rangle_\RQ$ be two DPIs such that $\Tp' \supseteq \Tp$ and $\Tn' \supseteq \Tn$. Further, let $\mathsf{nd}$
% := [\tax_1, \dots, \tax_k]$ 
be a redundant node w.r.t.\ $\langle\mo,\mb,\Tp,\Tn\rangle_\RQ$. Then, $\mathsf{nd}$ is a redundant node w.r.t.\ $\langle\mo,\mb,\Tp',\Tn'\rangle_\RQ$.
\end{lemma}
\begin{proof}
By Proposition~\ref{prop:changes_in_conflict_sets_after_testcase_added}, if $\langle\mo,\mb,\Tp',\Tn'\rangle_\RQ$ results from the addition of a single new positive or negative test case to $\langle\mo,\mb,\Tp,\Tn\rangle_\RQ$, there cannot be any minimal conflict set w.r.t.\ $\langle\mo,\mb,\Tp',\Tn'\rangle_\RQ$ that is a proper superset of a minimal conflict w.r.t.\ $\langle\mo,\mb,\Tp,\Tn\rangle_\RQ$. By Definition~\ref{def:redundant_node}, we can derive that any redundant node w.r.t.\ $\langle\mo,\mb,\Tp,\Tn\rangle_\RQ$ must be a redundant node w.r.t.\ $\langle\mo,\mb,\Tp',\Tn'\rangle_\RQ$. The proposition of this lemma is a consequence of further applications of Proposition~\ref{prop:changes_in_conflict_sets_after_testcase_added}. 
\end{proof}
This implies that a redundant node that is deleted during the execution of \textsc{dynamicHS} using the current DPI $DPI$ cannot become non-redundant throughout the entire remaining execution of the interactive debugging session, i.e.\ the execution of Algorithm~\ref{algo:inter_onto_debug}. Reason for this is that the sets of test cases in a DPI can only be extended and not reduced in the course of debugging. 

\begin{remark}\label{rem:mind_changes_in_dynamicHS} Note that this has consequences on the way how ``mind-changes'' of a user might be handled by the interactive algorithm. It implies that the current state of \textsc{dynamicHS} (stored in the output variables of \textsc{dynamicHS}) cannot be exploited in case a user decides to discard some already answered query or to switch the already submitted answer of some query, resulting in some modified DPI $DPI'$. In such a situation a new construction of a hitting set tree by \textsc{dynamicHS} using the DPI $DPI'$ is indicated. Otherwise, some already pruned redundant node w.r.t.\ $DPI$ might become a relevant node for $DPI'$ which would lead to a violation of the postulated completeness of \textsc{dynamicHS} w.r.t.\ each current DPI, in this case the DPI $DPI'$.\qed
\end{remark}
The following result is straightforward and claims that each successor node of a redundant node $\mathsf{nd}_i$ w.r.t.\ $DPI$ is a redundant node w.r.t.\ $DPI$. So, if $r$ is the minimal value such that both criteria of Definition~\ref{def:redundant_node} hold for $\mathsf{nd}_i$, all successor nodes of the subnode $\mathsf{nd}_i[1..r]$ of $\mathsf{nd}_i$ can be deleted. In other words, the entire subtree (of the hitting set tree produced by \textsc{dynamicHS}) rooted at an outgoing edge $e$ of a non-minimal conflict set where $e$ is labeled by an element $\tax$ which is not an element of a given witness of redundancy is obsolete.  
\begin{lemma}\label{lem:each_supernode_of_redundant_node_is_redundant}
Let $\langle\mo,\mb,\Tp,\Tn\rangle_\RQ$ be a DPI, $\mathsf{nd}$ be a redundant node w.r.t.\ $\langle\mo,\mb,\Tp,\Tn\rangle_\RQ$ and $\mathsf{nd}'$ be a successor node of $\mathsf{nd}$. Then, $\mathsf{nd}'$ is a redundant node w.r.t.\ $\langle\mo,\mb,\Tp,\Tn\rangle_\RQ$.
%Let $\langle\mo,\mb,\Tp,\Tn\rangle_\RQ$ be a DPI and $\mathsf{nd} := [\tax_1, \dots, \tax_k]$ be a redundant node w.r.t.\ $\langle\mo,\mb,\Tp,\Tn\rangle_\RQ$ with $\mathsf{nd.cs} := [\mc_1, \dots, \mc_k]$. Further, let $\mc_i \in \allC_{\langle\mo,\mb,\Tp,\Tn\rangle_\RQ}$ and $\tax_{i} \in \mc_i$ for all $i\in\setof{k+1,\dots,k+m}$. Then, $\mathsf{nd}' := [\tax_1, \dots, \tax_{k+m}]$ with $\mathsf{nd}'.\mathsf{cs} := [\mc_1, \dots, \mc_{k+m}]$ is a redundant node.
\end{lemma}
\begin{proof}
The proposition of this lemma is a direct consequence of Definition~\ref{def:redundant_node}.
\end{proof}
%---------------------------- BLOCK END  (redundancy)----------------------------------
%-------------------------------- BLOCK START (prune) -------------------------
\subsection[Hitting Set Tree Pruning]{Hitting Set Tree Pruning in \textsc{dynamicHS}}
\label{sec:HittingSetTreePruningInTextscDynamicHS}
The main pruning operations performed by \textsc{dynamicHS} take place in the scope of the \textsc{updateTree} function which is called right at the beginning of the execution of each call to \textsc{dynamicHS}. Assume a call to \textsc{dynamicHS} during Algorithm~\ref{algo:inter_onto_debug} given i.a.\ the DPI $\langle\mo,\mb,\Tp,\Tn\rangle_\RQ$ and the test cases $\Tp'$ and $\Tn'$ as arguments and suppose the last-but-one call to \textsc{dynamicHS} was given $\Tp''$ and $\Tn''$ as arguments. The job of \textsc{updateTree} is to restore the parameters that store the state of \textsc{dynamicHS} (for DPI $\langle\mo,\mb,\Tp\cup\Tp'',\Tn\cup\Tn''\rangle_\RQ$) in a way that they include at least all nodes that would be included by the respective parameters produced by a call to \textsc{dynamicHS} for the static DPI $\langle\mo,\mb,\Tp\cup\Tp',\Tn\cup\Tn'\rangle_\RQ$.

Roughly speaking, this involves the following actions: 
\begin{itemize}
	\item \emph{Pruning:} That is, only nodes that are definitely redundant w.r.t.\ $\langle\mo,\mb,\Tp\cup\Tp',\Tn\cup\Tn'\rangle_\RQ$ are deleted. A node is definitely redundant if a witness of redundancy of it is known.
	\item \emph{Replacement:} A deleted redundant node is replaced by an alternative equal node of it which is non-redundant w.r.t.\ $\langle\mo,\mb,\Tp\cup\Tp',\Tn\cup\Tn'\rangle_\RQ$, if there is such a one. Alternative equal nodes are constructed from the list of duplicate nodes $\Queue_{dup}$.
	\item \emph{Rearrangement:} the reassignation of nodes to $\Queue$ that ``survived'' all pruning steps or were introduced in the course of a replacement step and for which no evidence w.r.t.\ $\langle\mo,\mb,\Tp\cup\Tp',\Tn\cup\Tn'\rangle_\RQ$ is given that it should be assigned to any other set.  
\end{itemize}

More concretely, \textsc{updateTree} has the following effect on the collections $\Queue$, $\mD_{\times}$, $\mD_{\supset}$, $\Queue_{dup}$ which are, together with $\mD_{\checkmark}$, the only node-storing collections of \textsc{dynamicHS} at the beginning of the execution of each call to \textsc{dynamicHS}:
\begin{enumerate}[(a)]
\item If $\mathsf{nd}$ is in $\Queue_{dup}$, then $\mathsf{nd}$ is removed from $\Queue_{dup}$ only if there is a known witness of redundancy of $\mathsf{nd}$ w.r.t.\ $\langle\mo,\mb,\Tp\cup\Tp',\Tn\cup\Tn'\rangle_\RQ$. If there is an alternative equal replacement node $\mathsf{nd}'$ of $\mathsf{nd}$ which is constructable from some node in $\Queue_{dup}$, then $\mathsf{nd}'$ is added to $\Queue_{dup}$. 
\item %(a)~
If $\mathsf{nd}$ is in $\Queue$, then $\mathsf{nd}$ is removed from $\Queue$ only if there is a known witness of redundancy of $\mathsf{nd}$ w.r.t.\ $\langle\mo,\mb,\Tp\cup\Tp',\Tn\cup\Tn'\rangle_\RQ$. If there is an alternative equal replacement node $\mathsf{nd}'$ of $\mathsf{nd}$ which is constructable from some node in $\Queue_{dup}$, then $\mathsf{nd}'$ is added to $\Queue$. 
%(b)~for each node $\mathsf{nd}$ that is deleted from $\mD_{\times}$, $\mD_{\supset}$ or $\Queue$, an alternative equal node $\mathsf{nd}'$ of $\mathsf{nd}$ is added to the respective set if an alternative subnode of $\mathsf{nd}$ exists in $\Queue_{dup}$ from which $\mathsf{nd}'$ can be constructed 
\item %(b)~
If $\mathsf{nd}$ is in $\mD_{\times}$ and there is no known witness of redundancy of $\mathsf{nd}$ w.r.t.\ $\langle\mo,\mb,\Tp\cup\Tp',\Tn\cup\Tn'\rangle_\RQ$, then $\mathsf{nd}$ is added to $\Queue$.
\item %(c)~
If $\mathsf{nd}$ is in $\mD_{\times}$ and $\mathsf{nd}$ is redundant w.r.t.\ $\langle\mo,\mb,\Tp\cup\Tp',\Tn\cup\Tn'\rangle_\RQ$, then, if there is some alternative equal replacement node $\mathsf{nd}'$ of $\mathsf{nd}$ which is constructable from some node in $\Queue_{dup}$, then $\mathsf{nd}'$ is added to $\Queue$.
\item %(d)~
If $\mathsf{nd}$ is in $\mD_{\supset}$, there is no known witness of redundancy of $\mathsf{nd}$ w.r.t.\ $\langle\mo,\mb,\Tp\cup\Tp',\Tn\cup\Tn'\rangle_\RQ$ and there is no known minimal diagnosis w.r.t.\ $\langle\mo,\mb,\Tp\cup\Tp',\Tn\cup\Tn'\rangle_\RQ$ which is a proper subset of $\mathsf{nd}$, then $\mathsf{nd}$ is added to $\Queue$.
\item %(e)~
All nodes $\mathsf{nd}$ in $\mD_{\checkmark}$ are added to $\Queue$.
\end{enumerate}
Some comments: Step (a) is conducted by \textsc{pruneQdup} before \textsc{prune} is called, for each witness of redundancy $X$ of some node detected during the execution of \textsc{updateTree}. \textsc{prune} is the function that prunes or replaces nodes 
that are elements of any other collection than $\Queue_{dup}$, i.e.\ $\Queue$, $\mD_{\times}$ or $\mD_{\supset}$, and for which $X$ is a witness of redundancy. In this vein, the \textsc{prune} function just needs to perform a test whether there is \emph{any} node in $\Queue_{dup}$ that enables the construction of a replacement node of a deleted node. No check for redundancy of nodes in $\Queue_{dup}$ is necessary at this stage since $\Queue_{dup}$ has already been processed and cleaned from all redundant nodes w.r.t.\ $\langle\mo,\mb,\Tp\cup\Tp',\Tn\cup\Tn'\rangle_\RQ$.

Under the assumption that the deletion of a node redundant w.r.t.\ $\langle\mo,\mb,\Tp\cup\Tp',\Tn\cup\Tn'\rangle_\RQ$ is safe in terms of completeness of \textsc{dynamicHS} as to finding all minimal diagnoses w.r.t.\ $\langle\mo,\mb,\Tp\cup\Tp',\Tn\cup\Tn'\rangle_\RQ$ (which we will prove throughout this section), \textsc{updateTree} acts safely. That is, deletion actions are performed just on the basis of given evidence in the form of a witness of redundancy. However, it must be accentuated that this does not necessarily imply the pruning or replacement of \emph{all} redundant nodes w.r.t.\ $\langle\mo,\mb,\Tp\cup\Tp',\Tn\cup\Tn'\rangle_\RQ$. This is quite desired as guaranteeing complete pruning might be very costly concerning execution time since it would involve the precomputation of all not-yet-computed minimal conflict sets w.r.t.\ the current DPI \emph{at once}. In the bad case, since these computations would take place online, i.e.\ between two successive queries shown to the user, this would be anything but beneficial for an interactive algorithm whose usability and usefulness depends greatly on its timeliness. Apart from that, a single newly added test case can be expected to lead to the introduction of only a small number of minimal conflict sets w.r.t.\ the current DPI that are no minimal conflict sets w.r.t.\ the last-but-one DPI.

Which nodes are pruned throughout \textsc{updateTree} depends on which witnesses of redundancy are found, i.e.\ which minimal conflict sets are computed. The \textsc{updateTree} function is implemented to search targeted for witnesses of redundancy of stored nodes. That is, instead of just computing any minimal conflict set w.r.t.\ $\langle\mo,\mb,\Tp\cup\Tp',\Tn\cup\Tn'\rangle_\RQ$, it focuses on the set of nodes $\mD_{\times}$ which includes the subset of all minimal diagnoses $\mD_{calc}$ computed in the last-but-one iteration of \textsc{dynamicHS} w.r.t. the last-but-one DPI $\langle\mo,\mb,\Tp\cup\Tp'',\Tn\cup\Tn''\rangle_\RQ$, which are no diagnoses w.r.t.\ the current DPI $\langle\mo,\mb,\Tp\cup\Tp',\Tn\cup\Tn'\rangle_\RQ$. Note that we will prove later in this section that $\mD_{calc}$, and thus $\mD_{\times}$ and $\mD_{\checkmark}$ which are subsets thereof, will indeed comprise only minimal diagnoses. So, \textsc{updateTree} looks for witnesses of redundancy by means of exactly these minimal diagnoses that have been invalidated through the addition of the most recent answered query to the test cases of the DPI. Each diagnosis $\mathsf{nd}$ w.r.t.\ the last-but-one DPI can be invalidated only because it does not hit some minimal conflict set w.r.t.\ the current DPI and not because it is a non-minimal hitting set of all minimal conflict sets w.r.t.\ the current DPI. This can be directly inferred from Proposition~\ref{prop:adding_testcase_cannot_make_min_diags_shrink} which manifests that minimal diagnoses cannot shrink by the addition of a new test case i.e.\ there cannot be any minimal diagnosis w.r.t.\ the current DPI which is a proper subset of $\mathsf{nd}$. 

Now, two cases can be identified for a minimal conflict set $\mc$ w.r.t.\ the current DPI that is not hit by $\mathsf{nd}$:
\begin{enumerate}[C1:]
	\item $\mc$ is not in a subset-relationship with any minimal conflict set in $\mathsf{nd.cs}$. That is, $\mc$ is definitely not a witness of redundancy of $\mathsf{nd}$.
	\item $\mc$ is in a subset-relationship with some minimal conflict set in $\mathsf{nd.cs}$. That is, $\mc$ satisfies the first criterion of a witness of redundancy of $\mathsf{nd}$ (cf.\ Definition~\ref{def:redundant_node}). Thence, $\mc$ might be a witness of redundancy of $\mathsf{nd}$.
\end{enumerate}
Now, the idea is to try to figure out very fast some $\mc$ for a node $\mathsf{nd}\in\mD_{\times}$ such that $\mc$ is a witness of redundancy of $\mathsf{nd}$. This idea is implemented in the so-called \emph{Quick Redundancy Check (QRC)} which 
\begin{itemize}
	\item calls $\scQX$ just once given the DPI $\langle U_{\mathsf{nd.cs}} \setminus \mathsf{nd},\mb,\Tp,\Tn\rangle_\RQ$ with the usually very small KB $U_{\mathsf{nd.cs}} \setminus \mathsf{nd} \subseteq\mo$ in order to calculate just one minimal conflict set $\mc$ w.r.t.\ the current DPI
	\item and then verifies whether $\mc$ is a witness of redundancy of $\mathsf{nd}$ by conducting at most $|\mathsf{nd}|$ subset-relationship checks.
\end{itemize}
The following lemma confirms that QRC (lines~\ref{algoline:update:qx}-\ref{algoline:update:quickPC_end} in Algorithm~\ref{algo:update_tree}), if successful, indeed computes a witness of redundancy of $\mathsf{nd}$ and thus gives evidence that $\mathsf{nd}$ is redundant w.r.t.\ the current DPI. 
%\textsc{dynamicHS} in case it would have been called for the current DPI $\langle\mo,\mb,\Tp\cup\Tp',\Tn\cup\Tn'\rangle_\RQ$.
%, namely $\Queue$, $\mD_{\supset}$, $\Queue_{dup}$, 
%Explanation of update tree function:
%update tree restores the hitting set tree parameters Q, ..., Ccalc in a way they correspond to the parameters of dynamicHS called with CURRENT parameters P', N'
\begin{lemma}[Quick Redundancy Check -- QRC]\label{lem:quick_prune_check} 
Let $\langle\mo,\mb,\Tp,\Tn\rangle_\RQ$ be the DPI and $\Tp'$ and $\Tn'$ the sets of positively and negatively answered queries given as an input to \textsc{dynamicHS}.
%Let $\langle\mo,\mb,\Tp,\Tn\rangle_\RQ$ be a DPI and 
Further, let $\mathsf{nd}$ be some node in \textsc{dynamicHS}.
%, $\mathsf{nd} = [\tax_1, \dots, \tax_j]$ be a list of elements $\tax_k \in \mo$ and $\mathsf{nd.cs} = [\mc_1, \dots, \mc_j]$ be a list of elements such that $\mc_k \in \allC_{\langle\mo,\mb,\Tp,\Tn\rangle_\RQ}$ and $\tax_k \in \mc_k$ for all $k \in \setof{1,\dots,j}$. 
Then the following holds: 

If $\scQX(\langle U_{\mathsf{nd.cs}} \setminus \mathsf{nd} ,\mb,\Tp\cup\Tp',\Tn\cup\Tn'\rangle_\RQ)$ returns a set $\mc$ such that $\mc \subset \mathsf{nd.cs}[i]$ for some $i \in \setof{1,\dots,|\mathsf{nd.cs}|}$, then
\begin{itemize}
\item $\mathsf{nd}$ is a redundant node w.r.t.\ $\langle\mo,\mb,\Tp\cup\Tp',\Tn\cup\Tn'\rangle_\RQ$ and
\item $\mc$ is a witness of redundancy of $\mathsf{nd}$.
\end{itemize}
\end{lemma}
\begin{proof}
First, $U_{\mathsf{nd.cs}} \setminus \mathsf{nd}$ includes all elements in the union of all conflict sets in $\mathsf{nd.cs}$ except for the elements occurring in $\mathsf{nd}$. So, if $\scQX(\langle U_{\mathsf{nd.cs}} \setminus \mathsf{nd},\mb,\Tp\cup\Tp',\Tn\cup\Tn'\rangle_\RQ)$ returns a set $\mc$, then $\mc$ is a minimal conflict set w.r.t.\ $\langle U_{\mathsf{nd.cs}} \setminus \mathsf{nd},\mb,\Tp\cup\Tp',\Tn\cup\Tn'\rangle_\RQ$ by Proposition~\ref{prop:qx_correctness}. 

By Definition~\ref{def:cs}, $\mc \subseteq U_{\mathsf{nd.cs}} \setminus \mathsf{nd}$ holds wherefore $\mc \cap \mathsf{nd} = \emptyset$. By $\mo \supseteq U_{\mathsf{nd.cs}} \setminus \mathsf{nd}$ and Remark~\ref{rem:qx_with_O_setminus_node_yields_cs_wrt_O}, $\mc$ is a minimal conflict set w.r.t.\ $\langle\mo,\mb,\Tp\cup\Tp',\Tn\cup\Tn'\rangle_\RQ$. 

If $\mc \subset \mathsf{nd.cs}[i]$ for some $i \in \setof{1,\dots,|\mathsf{nd.cs}|}$, then we have that $\mc$ is a minimal conflict set w.r.t.\ $\langle\mo,\mb,\Tp\cup\Tp',\Tn\cup\Tn'\rangle_\RQ$ which is a proper subset of $\mathsf{nd.cs}[i]$. Since $\mc \cap \mathsf{nd} = \emptyset$ implies that $\mathsf{nd}[i] \notin \mc$ for all $i \in \setof{1,\dots,|\mathsf{nd}|}$, we conclude that $\mathsf{nd}[i] \in \mathsf{nd.cs}[i] \setminus \mc$. Now, by Definition~\ref{def:redundant_node}, $\mathsf{nd}$ is a redundant node w.r.t.\ $\langle\mo,\mb,\Tp\cup\Tp',\Tn\cup\Tn'\rangle_\RQ$ and $\mc$ is a witness of redundancy of $\mathsf{nd}$. 
\end{proof}

\begin{remark}\label{rem:quick_prune_check}
Please notice that the opposite direction does not necessarily hold. That is, if the node $\mathsf{nd}$ is redundant w.r.t.\ $\langle\mo,\mb,\Tp\cup\Tp',\Tn\cup\Tn'\rangle_\RQ$, $\scQX(\langle U_{\mathsf{nd.cs}} \setminus \mathsf{nd} ,\mb,\Tp\cup\Tp',\Tn\cup\Tn'\rangle_\RQ)$ might return
\begin{itemize}
\item some $\mc$ which is not a subset of any conflict set in $\mathsf{nd.cs}$ or 
%For instance, there might be some $i,j \in \setof{1,\dots,|\mathsf{nd.cs}|}$ such that $\mc \cap \mathsf{nd.cs}[i] \neq \emptyset$ and $\mc \cap \mathsf{nd.cs}[j] \neq \emptyset$
\item 'no conflict'.\qed 
\end{itemize}
\end{remark}
As an illustration of that remark, we give the following example:
\begin{example}
For instance, assume a node $\mathsf{nd} = [1,2]$ with $\mathsf{nd.cs} = [\tuple{1,2,3},\tuple{2,4,5}]$ and that $\tuple{2,3}$ is a minimal conflict set w.r.t.\ the current DPI $\langle\mo,\mb,\Tp\cup\Tp',\Tn\cup\Tn'\rangle_\RQ$ wherefore $\mathsf{nd}$ is redundant by Definition~\ref{def:redundant_node}. Then $U_{\mathsf{nd.cs}} \setminus \mathsf{nd} = \setof{3,4,5}$.

Suppose that $\tuple{3,5}$ is a minimal conflict set w.r.t.\ the current DPI as well. So, in this case, $\scQX(\langle \{3$, $4$, $5\}$, $\mb,\Tp\cup\Tp',\Tn\cup\Tn'\rangle_\RQ)$ might return $\tuple{3,5}$. However, $\tuple{3,5}$ is neither a subset of $\tuple{1,2,3}$ nor a subset of $\tuple{2,4,5}$ wherefore $\tuple{3,5}$ is no witness of redundancy of $\mathsf{nd}$. 

On the other hand, if we suppose that $\tuple{2,3}$ and $\tuple{2,4,5}$ are the only minimal conflict sets w.r.t.\ the current DPI that are subsets of $U_{\mathsf{nd.cs}} = \setof{1,2,3,4,5}$, then 'no conflict' is the output of the call to $\scQX$. This holds since $\mathsf{nd}[2] = 2$ is an element of both $\tuple{2,3}$ and $\tuple{2,4,5}$ and hence not an element of $U_{\mathsf{nd.cs}} \setminus \mathsf{nd} = \setof{3,4,5}$. Therefore, neither $\tuple{2,3}$ nor $\tuple{2,4,5}$ is returned by $\scQX$ since $\scQX(\langle \setof{3,4,5},\mb,\Tp\cup\Tp',\Tn\cup\Tn'\rangle_\RQ)$ can only return a set that is a subset of $\setof{3,4,5}$ by Proposition~\ref{prop:qx_correctness} and Definition~\ref{def:cs}.%. that is a does not include all elements of $\mathsf{nd.cs}[1]\setminus\mathsf{nd}[1]$.
\qed
\end{example}
In both cases of the previous example, an existing witness of redundancy of $\mathsf{nd}$ is not detected by QRC. In this situation, i.e.\ when QRC is negative, a \emph{Complete Redundancy Check (CRC)} is performed which involves $\scQX$ investigating all the DPIs $\tuple{\mathsf{nd.cs}[i]\setminus\mathsf{nd}[i],\mb,\Tp\cup\Tp',\Tn\cup\Tn'}_\RQ$ for $i \in \setof{1,\dots,|\mathsf{nd}|}$ separately. CRC, as substantiated by the following lemma, does find a witness of redundancy if the node $\mathsf{nd}$ is redundant w.r.t.\ the current DPI; and, if CRC does not find a witness of redundancy w.r.t.\ the current DPI, then $\mathsf{nd}$ is non-redundant w.r.t.\ the current DPI.
\begin{lemma}[Complete Redundancy Check -- CRC]\label{lem:complete_prune_check} 
Let $\langle\mo,\mb,\Tp,\Tn\rangle_\RQ$ be the DPI and $\Tp'$ and $\Tn'$ the sets of positively and negatively answered queries given as an input to \textsc{dynamicHS}.
Further, let $\mathsf{nd}$ be some node in \textsc{dynamicHS}.
%
%Let $\langle\mo,\mb,\Tp,\Tn\rangle_\RQ$ be an admissible DPI, $\mathsf{nd} = [\tax_1, \dots, \tax_j]$ be a list of elements $\tax_k \in \mo$ and $\mathsf{nd.cs} = [\mc_1, \dots, \mc_j]$ be a list of elements such that $\mc_k \in \allC_{\langle\mo,\mb,\Tp,\Tn\rangle_\RQ}$ and $\tax_k \in \mc_k$ for all $k \in \setof{1,\dots,j}$. 
Then, the following holds:
\begin{enumerate}[(1)]
	\item $\mathsf{nd}$ is redundant w.r.t.\ $\langle\mo,\mb,\Tp\cup\Tp',\Tn\cup\Tn'\rangle_\RQ$ iff there is some $i \in \setof{1,\dots,|\mathsf{nd}|}$ such that \\ 
	$\scQX(\langle \mathsf{nd.cs}[i] \setminus \setof{\mathsf{nd}[i]},\mb,\Tp\cup\Tp',\Tn\cup\Tn'\rangle_\RQ) = X$ where $X \neq \text{'no conflict'}$.
	\item If there is some $i \in \setof{1,\dots,|\mathsf{nd}|}$ such that $\scQX(\langle \mathsf{nd.cs}[i] \setminus \setof{\mathsf{nd}[i]},\mb,\Tp\cup\Tp',\Tn\cup\Tn'\rangle_\RQ) = X$ where $X \neq \text{'no conflict'}$, then $X$ is a witness of redundancy of $\mathsf{nd}$.
\end{enumerate}
\end{lemma}
\begin{proof}
(1): ``$\Leftarrow$'': Assume there is some $i \in \setof{1,\dots,|\mathsf{nd}|}$ such that $\scQX(\langle \mathsf{nd.cs}[i] \setminus \setof{\mathsf{nd}[i]},\mb,\Tp\cup\Tp',\Tn\cup\Tn'\rangle_\RQ) = X$ where $X \neq \text{'no conflict'}$. Then, by Proposition~\ref{prop:qx_correctness}, we have that $X$ is a minimal conflict set w.r.t.\ $\langle \mathsf{nd.cs}[i] \setminus \setof{\mathsf{nd}[i]},\mb,\Tp\cup\Tp',\Tn\cup\Tn'\rangle_\RQ$ such that $X \subseteq \mathsf{nd.cs}[i] \setminus \setof{\mathsf{nd}[i]}$. By Definition~\ref{def:cs}, $X$ is a minimal conflict set w.r.t.\ $\langle \mo,\mb,\Tp\cup\Tp',\Tn\cup\Tn'\rangle_\RQ$. Hence, we can conclude that $\mathsf{nd}[i] \notin X$. By Definition~\ref{def:dyn:node_node.cs} and since $\mathsf{nd}$ is a node in \textsc{dynamicHS}, it holds that $\mathsf{nd}[i] \in \mathsf{nd.cs}[i]$. As a consequence, $\mathsf{nd}[i] \in \mathsf{nd.cs}[i] \setminus X$ holds. By Definition~\ref{def:redundant_node}, $\mathsf{nd}$ is redundant w.r.t.\ $\langle\mo,\mb,\Tp\cup\Tp',\Tn\cup\Tn'\rangle_\RQ$ (and $X$ is a witness of redundancy of $\mathsf{nd}$).

``$\Rightarrow$'': Suppose $\mathsf{nd}$ is a redundant node w.r.t.\ $\langle\mo,\mb,\Tp\cup\Tp',\Tn\cup\Tn'\rangle_\RQ$. Then, by Definition~\ref{def:redundant_node}, there must be some $r \in\setof{1,\dots,|\mathsf{nd}|}$ and some minimal conflict set $X$ w.r.t.\ $\langle\mo,\mb,\Tp\cup\Tp',\Tn\cup\Tn'\rangle_\RQ$ such that (i)~$X \subset \mathsf{nd.cs}[r]$ and (ii)~$\mathsf{nd}[r] \in \mathsf{nd.cs}[r]\setminus X$. By (ii), 
$\mathsf{nd}[r] \notin X$. By Definition~\ref{def:dyn:node_node.cs} and the fact that $\mathsf{nd}$ is a node in \textsc{dynamicHS}, we obtain that $\mathsf{nd}[r] \in \mathsf{nd.cs}[r]$ must be true. Hence, by (i), we derive that $X \subseteq \mathsf{nd.cs}[r] \setminus \setof{\mathsf{nd}[r]}$. By Proposition~\ref{prop:qx_correctness}, $\scQX$ given some DPI $DPI$ outputs a minimal conflict set w.r.t.\ $DPI$ iff there is a minimal conflict set w.r.t.\ $DPI$. Therefore and since $\scQX(\langle\mathsf{nd.cs}[i] \setminus \setof{\mathsf{nd}[i]},\mb,\Tp\cup\Tp',\Tn\cup\Tn'\rangle_\RQ)$ is called for each $i \in \setof{1,\dots,|\mathsf{nd}|}$, it must also be called for $i := r$ since $r \in \setof{1,\dots,|\mathsf{nd}|}$. So, some minimal conflict set $X'$, and not 'no conflict', must be returned by $\scQX(\langle\mathsf{nd.cs}[r] \setminus \setof{\mathsf{nd}[r]},\mb,\Tp\cup\Tp',\Tn\cup\Tn'\rangle_\RQ)$ since there is at least one minimal conflict set w.r.t.\ $\langle\mathsf{nd.cs}[r] \setminus \setof{\mathsf{nd}[r]},\mb,\Tp\cup\Tp',\Tn\cup\Tn'\rangle_\RQ$, namely $X$.

(2): This proposition follows directly from (1) (``$\Leftarrow$'').
\end{proof}

At the point where some witness of redundancy $X$ of some node $\mathsf{nd}\in\mD_{\times}$ is found by QRC or CRC in \textsc{updateTree}, the next steps (lines~\ref{algoline:update:call_prune_Qdup}-\ref{algoline:update:call_prune_Dsupset}) involve the pruning of $\Queue_{dup}$, $\Queue$, $\mD_{\times}$ and $\mD_{\supset}$. As already mentioned, $\Queue_{dup}$ is the first collection to be cleaned from redundant nodes (w.r.t.\ the witness $X$) in \textsc{pruneQdup} in order to constitute an input to the \textsc{prune} function that does not include any redundant nodes (w.r.t.\ the witness $X$) and can be used ``blindly'' to construct replacement nodes of redundant nodes (w.r.t.\ the witness $X$) deleted from $\Queue$, $\mD_{\times}$ or $\mD_{\supset}$.

Before any pruning steps have ever been executed during the execution of Algorithm~\ref{algo:inter_onto_debug}, $\Queue_{dup}$ comprises all generated nodes $\mathsf{nd}_{dup}$ for which, at generation time, there was one node $\mathsf{nd}\in\Queue$ such that $\mathsf{nd}_{dup} = \mathsf{nd}$. That means, $\mathsf{nd}_{dup}$ is stored in $\Queue_{dup}$ in order to be available as an alternative equal node of $\mathsf{nd}$ or as an alternative subnode of some successor of $\mathsf{nd}$ in case $\mathsf{nd}$ is found to be redundant w.r.t.\ some current DPI.

If some node $\mathsf{nd}_{dup}$ in $\Queue_{dup}$ is found to be redundant w.r.t.\ the current DPI, there might be other nodes in $\Queue_{dup}$ from which a non-redundant alternative equal node $\mathsf{nd}'_{dup}$ of $\mathsf{nd}_{dup}$ w.r.t.\ the current DPI can be constructed. By Definition~\ref{def:node_denotations_1}, we call such a node $\mathsf{nd}'_{dup}$ a combined replacement node of $\mathsf{nd}_{dup}$. The name stems from the fact that $\mathsf{nd}'_{dup}$ is generated as a combination of existing nodes in $\Queue_{dup}$. Combining two nodes $\mathsf{nd}_1,\mathsf{nd}_2 \in \Queue_{dup}$ such that $\mathsf{nd}_1$ is a proper alternative subnode of $\mathsf{nd}_2$ yields $\mathsf{nd}_3$ with $\mathsf{nd}_2=\mathsf{nd}_3$. $\mathsf{nd}_3$ is constructed in that the first (redundant) part of $\mathsf{nd}_2$ (and $\mathsf{nd}_2.\mathsf{cs}$) is replaced by the (non-redundant) part $\mathsf{nd}_1$ (and $\mathsf{nd}_1.\mathsf{cs}$). 

Such a combination is ``legitimate'' since it gives a node $\mathsf{nd}_3$ that would have been constructed if all duplicate nodes would have been added to $\Queue$ and processed regularly instead of being added to $\Queue_{dup}$. The strategy to store duplicate nodes (where ``duplicate'' refers to the \emph{set} a node represents) in a separate collection $\Queue_{dup}$ as soon as they are found is part of the space-saving policy the \textsc{dynamicHS} algorithm pursues. For, in general, this prevents the algorithm to generate and store exponentially many nodes corresponding to equal sets. Since diagnoses are sets and not lists like nodes, it suffices to find \emph{only one} node corresponding to a diagnosis. Only if some active node (one that is not in $\Queue_{dup}$) becomes redundant, some other set-equal node, if available, is constructed from the stored duplicate nodes. This idea is very similar to the way pruning is handled in the directed acyclic graph described in \cite{greiner1989correction}.

The idea of node combination is formalized by the following definition.
\begin{definition}\label{def:comb_node}
Let $S$ be a collection of nodes in \textsc{dynamicHS} and let $S_i$ be the set of nodes of cardinality $i$ in $S$. Further, let the set $Comb_1(S) := S_1$ and let $Comb_i(S)$ comprise
\begin{itemize}
\item all nodes in $S_i$ and
%\item only nodes of cardinality $i$
\item all nodes $\mathsf{nd}$ such that $\mathsf{nd}$ is an alternative equal node of some node in $S_i$ constructed from some node in $\bigcup_{j=1}^{i-1} Comb_{j}(S)$.  
%%%%%%%%%%%%%%%%%%%%%%%%%
%nodes $\mathsf{nd} := \textsc{add}(\mathsf{node},\mathsf{node}'[|\mathsf{node}|+1..i])$ where $\mathsf{node}\in \bigcup_{j=1}^{i-1} Comb_{j}(S)$, $\mathsf{node}'\in S_i$ and $\mathsf{node}$ is an alternative subnode of $\mathsf{node}'$.
%%%%%%%%%%%%%%%%%%%%%%%%%
%$\mathsf{node} = \mathsf{node}'[1..j]$ and $j<i$ is the maximal number such that $S_j \neq \emptyset$. 
\end{itemize}
Then, $Comb(S) := \bigcup_{i=1}^{\infty}Comb_i(S)$ is called \emph{the set of combined nodes of $S$} and a node in $Comb_i(S)$ is called a \emph{combined node of cardinality $i$ in $S$}.

Further, let $\mathsf{node}$ be a node in \textsc{dynamicHS} and $X$ be a minimal conflict set w.r.t.\ the current DPI. Then, 
\begin{itemize}
\item $Comb_{\mathsf{node}}(S) := \setof{\mathsf{nd}\,|\,\mathsf{nd}\in Comb(S), \mathsf{nd} = \mathsf{node}}$ is \emph{the set of combined equal nodes of $\mathsf{nd}$ of $S$} and
\item $Comb_{\mathsf{node},X}(S) \subseteq Comb_{\mathsf{node}}(S)$ is the set of combined equal nodes of $\mathsf{nd}$ of $S$ for which $X$ is not a witness of redundancy.
\end{itemize}
\end{definition}
The following corollary summarizes some simple consequences of Definition~\ref{def:comb_node}.
\begin{corollary}\label{cor:consequences_of_def_comb}
Let $S$ be a set of nodes in \textsc{dynamicHS} and let $S_i$ be the set of nodes of cardinality $i$ in $S$. Then:
\begin{enumerate}[(1)]
\item $Comb_i(S) = \emptyset$ iff $S_i = \emptyset$.
\item\label{cor:consequences_of_def_comb:enum:Combi_only_cardinality_i} $Comb_i(S)$ includes only nodes of cardinality $i$.
\item $Comb_{\mathsf{node}}(S) = \emptyset$ iff there is no node $\mathsf{nd} \in S$ such that $\mathsf{nd} = \mathsf{node}$.
%\item If $\mathsf{nd} \in Comb_i(S)$, then there is some $k \geq 0$ and some $\mathsf{nd}' \in S_i$ such that $\mathsf{nd}[|\mathsf{nd}|-k..|\mathsf{nd}|] = \mathsf{nd}'[|\mathsf{nd}|-k..|\mathsf{nd}|]$
\item\label{cor:consequences_of_def_comb:enum:partial_nodes} If $\mathsf{nd} \in Comb_i(S)$ and $\mathsf{nd} \notin S$, then 
\begin{itemize}
	\item there is some $\mathsf{nd}' \in Comb_j(S)$ for some $j \in \setof{1,\dots,i-1}$ and
	\item some $\mathsf{nd}'' \in S_i$
\end{itemize}
such that 
\begin{itemize}
	\item $\mathsf{nd}'$ is an alternative subnode of $\mathsf{nd}''$ and
	\item $\mathsf{nd} = \textsc{add}(\mathsf{nd}',\mathsf{nd}''[j+1..i])$ and 
	\item $\mathsf{nd.cs} = \textsc{add}(\mathsf{nd}'.\mathsf{cs},\mathsf{nd}''.\mathsf{cs}[j+1..i])$.
\end{itemize}
%\item If $\mathsf{nd} \in Comb_i(S)$ and $\mathsf{nd} \notin S_i$, then there is some $\mathsf{nd}' \in Comb_j(S)$ for some $j \in \setof{1,\dots,i-1}$ such that $\mathsf{nd} = \textsc{add}(\mathsf{nd}',\mathsf{nd}[|\mathsf{nd}'|+1..|\mathsf{nd}|])$ and $\mathsf{nd.cs} = \textsc{add}(\mathsf{nd}'.\mathsf{cs},\mathsf{nd.cs}[|\mathsf{nd}'|+1..|\mathsf{nd}|])$.
\end{enumerate}
\end{corollary}
The example we give next illustrates Definition~\ref{def:comb_node}. 
\begin{example}
Recall the nodes $\mathsf{nd}, \mathsf{nd}_1, \mathsf{nd}_2$, $\mathsf{node}_1$ and $\mathsf{node}_2$  of Example~\ref{example:alternative_equal_node_alternative_subnode} and let $\mathsf{nd}_3 := [1,2,6,4]$ with $\mathsf{nd}_3.\mathsf{cs} := [\tuple{1,2,3},\tuple{2,6},\tuple{3,6,7},\tuple{4,5}]$ and $S := \setof{\mathsf{nd},\mathsf{nd}_1,\mathsf{nd}_2, \mathsf{nd}_3}$. Then, 
\begin{align*}
S_1 &= \emptyset \\
S_2 &= \setof{\mathsf{nd}_1} \\
S_3 &= \setof{\mathsf{nd}_2} \\
S_4 &= \setof{\mathsf{nd},\mathsf{nd}_3} \\
S_i &= \emptyset \quad \forall i>4 \\
Comb_1(S) &= \emptyset \\
Comb_2(S) &= \setof{\mathsf{nd}_1} \\
Comb_3(S) &= \setof{\mathsf{nd}_2} \\
Comb_4(S) &= \setof{\mathsf{nd},\mathsf{node}_1,\mathsf{node}_2,\mathsf{nd}_3,\mathsf{nd}_4} \\
Comb_i(S) &= \emptyset \quad \forall i>4 \\
Comb_{\mathsf{nd}}(S) &= \setof{\mathsf{nd},\mathsf{node}_1,\mathsf{node}_2} \\
Comb_{\mathsf{node}_1}(S) &= Comb_{\mathsf{node}_2}(S) = Comb_{\mathsf{nd}}(S)\\
Comb_{\mathsf{nd}_3}(S) &= \setof{\mathsf{nd}_3,\mathsf{nd}_4} \\
Comb_{\mathsf{nd}_4}(S) &= Comb_{\mathsf{nd}_3}(S)
\end{align*}
%$S_1 = \emptyset$, $S_2 = \setof{\mathsf{nd}_1}$, $S_3 = \setof{\mathsf{nd}_2}$, $S_4 = \setof{\mathsf{nd}}$ and $Comb_1(S) = \emptyset$, $Comb_2(S) = \setof{\mathsf{nd}_1}$, $Comb_3(S) = \setof{\mathsf{nd}_2}$ and $Comb_4(S) = \setof{\mathsf{nd},\mathsf{nd}_{alt,1},\mathsf{nd}_{alt,2}}$ 
%%%%%%%%%%%%%%%%%%%%%%%%
where 
\begin{align*}
\mathsf{nd}_{4} &:= [2,1,6,4] \\
\mathsf{nd}_{4}.\mathsf{cs} &:= [\tuple{1,2,3},\tuple{1,4},\tuple{3,6,7},\tuple{4,5}]
\end{align*}
is the alternative equal node of $\mathsf{nd}_3$ constructed from $\mathsf{nd}_1$. 
%where 
%\begin{align*}
%\mathsf{nd}_{alt,1} &:= [2,1,3,4] \\
%\mathsf{nd}_{alt,1}.\mathsf{cs} &:= [\tuple{1,2,3},\tuple{1,4},\tuple{3,6,7},\tuple{4,5}]
%\end{align*}
%is the alternative equal node of $\mathsf{nd}$ constructed from $\mathsf{nd}_1$ and 
%\begin{align*}
%\mathsf{nd}_{alt,2} &:= [3,2,1,4] \\
%\mathsf{nd}_{alt,2}.\mathsf{cs} &:= [\tuple{1,2,3},\tuple{2,6},\tuple{1,4},\tuple{4,5}]
%\end{align*}
%is the alternative equal node of $\mathsf{nd}$ constructed from $\mathsf{nd}_2$.
%%%%%%%%%%%%%%%%%%%%%%%%%% 
%$\mathsf{nd}_{alt,1} := [2,1,3,4]$ with $\mathsf{nd}_{alt,1}.\mathsf{cs} := [\tuple{1,2,3},\tuple{1,4},\tuple{3,6,7},\tuple{4,5}]$ and $\mathsf{nd}_{alt,2} := [3,2,1,4]$ with $\mathsf{nd}_{alt,2}.\mathsf{cs} := [\tuple{1,2,3},\tuple{2,6},\tuple{1,4},\tuple{4,5}]$.	
\qed
\end{example}
The \textsc{pruneQdup} function is always called given the current list $\Queue_{dup}$ which is anytime sorted in ascending order by node cardinality. This holds by lines~\ref{algoline:dyn:add_to_Qdup}, \ref{algoline:pruneQdup:insert_alternative_equal_node_into_Dupnew} and \ref{algoline:pruneQdup:insert_node_into_Dupnew} which are the only places where nodes are added to $\Queue_{dup}$ throughout \textsc{dynamicHS} and where nodes are inserted into $\Queue_{dup}$ such that the order by node cardinality is preserved. Now, the next lemma substantiates that \textsc{pruneQdup}, given some minimal conflict set $X$ w.r.t.\ the current DPI, updates $\Queue_{dup}$ in a way that all redundant nodes w.r.t.\ the witness $X$ are deleted, each deleted node is replaced by one non-redundant combined replacement node w.r.t.\ the witness $X$ if such a one is constructable (cf.\ Definition~\ref{def:comb_node}), 
and for each remaining node $\mathsf{nd}$, i.e.\ $\mathsf{nd}$ is a non-deleted node or a combined replacement node of some deleted node, each superset of $X$ in $\mathsf{nd.cs}$ is replaced by $X$. 
%and for each remaining node $\mathsf{nd}$ each superset of $X$ in $\mathsf{nd.cs}$ is replaced by $X$. 

This leads to a new list $\Queue_{dup}$ returned by \textsc{pruneQdup} which includes only non-redundant nodes w.r.t.\ the witness $X$. Furthermore, the new list $\Queue_{dup}$ contains a node corresponding to each set (path) $S$ for which there was a corresponding node in the old list $\Queue_{dup}$ if there would be a non-redundant (w.r.t.\ $X$) node corresponding to $S$ in a hitting set tree equal to the one produced by \textsc{dynamicHS} except that all duplicate nodes corresponding to equal sets (paths) would be regularly processed and expanded. 
%it holds that $\mathsf{nd.cs}$ does not include a conflict set that is a proper superset of $X$.
%it includes all nodes in the input set $\Queue_{dup}$ that are non-redundant w.r.t.\ the witness $X$ and for each deleted redundant 
\begin{lemma}\label{lem:pruneQdup} 
Let $\langle\mo,\mb,\Tp,\Tn\rangle_\RQ$ be a DPI and let the input parameters to the \textsc{pruneQdup} function be:
\begin{itemize}
\item $X$ is a minimal conflict set w.r.t.\ $\langle\mo,\mb,\Tp,\Tn\rangle_\RQ$,
\item $Dup$ is a set of nodes sorted ascending by node cardinality.
\end{itemize}
%Let further $Comb'_{\mathsf{nd}}(Dup) \subseteq Comb_{\mathsf{nd}}(Dup)$ denote the set of nodes in $Comb_{\mathsf{nd}}(Dup)$ for which $X$ is not a witness of redundancy.

Then, \textsc{pruneQdup} returns $Dup_{new}$ where $Dup_{new}$ includes 
\begin{enumerate}[(1)]
\item all nodes in $Dup$ for which $X$ is not a witness of redundancy, 
\item at least one node in $Comb_{\mathsf{nd},X}(Dup)$ for each node $\mathsf{nd} \in Dup$ for which $X$ is a witness of redundancy, if $Comb_{\mathsf{nd},X}(Dup) \neq \emptyset$ and
\item only nodes $\mathsf{nd}$ such that there is no $r \in \setof{1,\dots,|\mathsf{nd}|}$ for which $\mathsf{nd.cs}[r] \supset X$. 
\end{enumerate}
\end{lemma}
\begin{proof}
The function \textsc{pruneQdup} walks through all nodes $\mathsf{ndi}$ in the set $Dup$.
%, from low cardinality nodes to high cardinality nodes by lines~\ref{algoline:pruneQdup:for_i_in_Dup} and \ref{algoline:pruneQdup:ndi_gets_Dup[i]} and the precondition that $Dup$ is sorted ascending by node cardinality. 
%
If $X$ is not a witness of redundancy of $\mathsf{ndi}$, tested in lines~\ref{algoline:pruneQdup:redundancy_check_part1} and \ref{algoline:pruneQdup:redundancy_check_part2} exactly as prescribed by Definition~\ref{def:redundant_node}, then $k = 0$ must hold in line~\ref{algoline:pruneQdup:if_k>0} by lines~\ref{algoline:pruneQdup:k_gets_0}-\ref{algoline:pruneQdup:endfor_m_gets_1_to_ndi.cs}. Thus, line~\ref{algoline:pruneQdup:insert_node_into_Dupnew} is executed and $\mathsf{ndi}$ added to $Dup_{new}$. Since no nodes are removed from $Dup_{new}$ throughout \textsc{pruneQdup}, proposition~(1) is valid.

Otherwise, i.e.\ if $X$ is a witness of redundancy of $\mathsf{ndi}$, then line~\ref{algoline:pruneQdup:k_gets_m} must have been executed at least once before line~\ref{algoline:pruneQdup:if_k>0} is reached. This implies that $k>0$ must hold in line~\ref{algoline:pruneQdup:if_k>0}. At this point, $k$ stores the maximum position in (the list) $\mathsf{ndi}$ at which the redundancy criterion of lines~\ref{algoline:pruneQdup:redundancy_check_part1} and \ref{algoline:pruneQdup:redundancy_check_part2} is satisfied. So, in line~\ref{algoline:pruneQdup:for_ndj_in_Dupnew}, nodes in $Dup_{new}$ are tested successively until some $\mathsf{ndj} \in Dup_{new}$ meets $|\mathsf{ndj}|\geq k$ and $\mathsf{ndi}[1..|\mathsf{ndj}|] = \mathsf{ndj}$. 
%This means that the subnode $\mathsf{ndi}[1..|\mathsf{ndj}|]$ ($\mathsf{ndi.cs}[1..|\mathsf{ndj}|]$) of $\mathsf{ndi}$ ($\mathsf{ndi.cs}$) can be replaced by $\mathsf{ndj}$ ($\mathsf{ndj.cs}$) to yield an alternative equal node $\mathsf{ndi}_{new}$ (with $\mathsf{ndi}_{new}.\mathsf{cs}$) of $\mathsf{ndi}$ (lines~\ref{algoline:pruneQdup:construct_alternative_equal_node} and \ref{algoline:pruneQdup:construct_alternative_equal_node.cs}).
This means that the subnode $\mathsf{ndi}[1..|\mathsf{ndj}|]$ of $\mathsf{ndi}$ can be replaced by $\mathsf{ndj}$ (and $\mathsf{ndi.cs}[1..|\mathsf{ndj}|]$ by $\mathsf{ndj.cs}$) to yield an alternative equal node $\mathsf{ndi}_{new}$ of $\mathsf{ndi}$ (lines~\ref{algoline:pruneQdup:construct_alternative_equal_node} and \ref{algoline:pruneQdup:construct_alternative_equal_node.cs}).

We still have to show that $X$ cannot be a witness of redundancy of $\mathsf{ndi}_{new}$. For this to hold it is sufficient that $X$ is not a witness of redundancy of $\mathsf{ndj}$ by $|\mathsf{ndj}| \geq k$. So, we must verify that $Dup_{new}$ can comprise only nodes of which $X$ is not a witness of redundancy. We prove this by induction. 

Since $Dup_{new}$ is initialized to be the empty set when the function \textsc{pruneQdup} starts executing, we just need to investigate which nodes are added to $Dup_{new}$ within \textsc{pruneQdup}. Addition of nodes to $Dup_{new}$ happens at lines~\ref{algoline:pruneQdup:insert_alternative_equal_node_into_Dupnew} and \ref{algoline:pruneQdup:insert_node_into_Dupnew}.

Base case: When line~\ref{algoline:pruneQdup:insert_alternative_equal_node_into_Dupnew} executed for the first time during the execution of \textsc{pruneQdup}, $Dup_{new}$ can only comprise nodes which have been added to it in line~\ref{algoline:pruneQdup:insert_node_into_Dupnew}. By the argumentation used to prove proposition~(1) of this lemma, it holds that $X$ is not a witness of redundancy of any node added to $Dup_{new}$ in line~\ref{algoline:pruneQdup:insert_node_into_Dupnew}. Thus, there cannot be a witness of redundancy of the very first node added to $Dup_{new}$ in line~\ref{algoline:pruneQdup:insert_alternative_equal_node_into_Dupnew}. 

Induction step: Let us assume that $Dup_{new}$ comprises only nodes such that $X$ is not a witness of redundancy of any of them. Further, suppose that $\mathsf{ndi}_{new}$ is added to $Dup_{new}$ when line~\ref{algoline:pruneQdup:insert_alternative_equal_node_into_Dupnew} is executed for the $k$-th time where $k > 1$. Then, by the same line of argument as in the base case, we can conclude that $X$ is not a witness of redundancy of $\mathsf{ndi}_{new}$.

Each node $\mathsf{ndi}_{new}$ added to $Dup_{new}$ in line~\ref{algoline:pruneQdup:insert_alternative_equal_node_into_Dupnew} is an element of $Comb_{\mathsf{ndi},X}(Dup)$. Namely, $\mathsf{ndj}$ satisfies the criterion in line~\ref{algoline:pruneQdup:check_if_alternative_subnode} and thus $\mathsf{ndi}_{new}$ is an element of $Comb_{\mathsf{ndi}}(Dup)$ by Definition~\ref{def:comb_node}. And, as shown before, $X$ is not a witness of redundancy of $\mathsf{ndi}_{new}$, wherefore $\mathsf{ndi}_{new} \in Comb_{\mathsf{ndi},X}(Dup)$ by the definition of $Comb_{\mathsf{ndi},X}(Dup)$ (Definition~\ref{def:comb_node}).

Thence, if $Comb_{\mathsf{ndi},X}(Dup) \neq \emptyset$, there must be at least one node $\mathsf{nd}$ added to $Dup_{new}$ such that $\mathsf{nd} \in Comb_{\mathsf{ndi},X}(Dup)$ and $X$ is not a witness of redundancy of $\mathsf{nd}$. 
Consequently, proposition~(2) holds.

Proposition~(3): First, observe that each node in $Dup$ is definitely processed as $\mathsf{ndi}$
by the for-loop in line~\ref{algoline:pruneQdup:for_i_in_Dup} and the fact that there is no criterion that can cause a preliminary break of this for-loop. Each time the first part of the redundancy check (line~\ref{algoline:pruneQdup:redundancy_check_part1}) is successful for $\mathsf{ndi}$,	we know that some conflict set $\mathsf{ndi.cs}[m]$ is non-minimal w.r.t.\ $\langle\mo,\mb,\Tp,\Tn\rangle_\RQ$. If the second part of the redundancy check (line~\ref{algoline:pruneQdup:redundancy_check_part2}) is negative, then $\mathsf{ndi}[m] \in X$, wherefore there is -- at least so far -- no evidence that $\mathsf{ndi}$ is redundant w.r.t.\ $\langle\mo,\mb,\Tp,\Tn\rangle_\RQ$. In this case, $\mathsf{ndi}$ might later be inserted to $Dup_{new}$ (in case $X$ is not a witness of redundancy of $\mathsf{ndi}$) and hence the set $\mathsf{ndi.cs}[m]$ is replaced by the minimal conflict set $X$ w.r.t.\ $\langle\mo,\mb,\Tp,\Tn\rangle_\RQ$ in line~\ref{algoline:pruneQdup:ndi.cs[m]_gets_X}. If the second part of the redundancy check in line~\ref{algoline:pruneQdup:redundancy_check_part2} is positive, then it is guaranteed that $\mathsf{ndi}$ is either combined-replaced or pruned. This holds due to lines~\ref{algoline:pruneQdup:if_k>0}-\ref{algoline:pruneQdup:break} and since $k > 0$ must be true due to line~\ref{algoline:pruneQdup:k_gets_m}. That a combined replacement node that might be found for some redundant $\mathsf{ndi}$ throughout lines~\ref{algoline:pruneQdup:if_k>0}-\ref{algoline:pruneQdup:break} meets proposition~(3) can be shown by induction in a very similar way as proposition~(2) was shown. 
\end{proof}
The following corollary is a direct consequence of Lemma~\ref{lem:pruneQdup} and states that the updated list $\Queue_{dup}$ (if interpreted as a set) is a subset of the set of combined nodes of the old list $\Queue_{dup}$. In other words, no nodes corresponding to sets (paths) that are not represented by a node in the old list $\Queue_{dup}$ can be introduced throughout \textsc{pruneQdup}. The introduction of such nodes corresponding to ``new'' sets (paths) can only take place in line~\ref{algoline:dyn:add_to_Qdup} where newly generated nodes are added to $\Queue_{dup}$.
\begin{corollary}\label{cor:pruneQdup_outputs_subset_of_Comb(Qdup)}
Given the same preconditions as in Lemma~\ref{lem:pruneQdup}, \textsc{pruneQdup} returns $Dup_{new}$ where $Dup_{new} \subseteq Comb(Dup)$.
\end{corollary}
The following result provides sufficient and necessary criteria for a node $\mathsf{nd}$ to be a combined node of $\Queue_{dup}$. Roughly, these criteria involve the existence of a sequence of nodes $\mathsf{nd}_1,\dots,\mathsf{nd}_k \in \Queue_{dup}$ where each node in this sequence is a proper alternative subnode of the next node and $\mathsf{nd}$ is constructed from this sequence of nodes in that $\mathsf{nd}$ is an alternative equal node of $\mathsf{nd}_k$ constructed from $\mathsf{nd}'_{k-1}$.
%an alternative equal node $\mathsf{nd}'_{k-1}$ of $\mathsf{nd}_{k-1}$. 
$\mathsf{nd}'_{k-1}$ in turn is an alternative equal node of $\mathsf{nd}_{k-1}$ constructed from $\mathsf{nd}'_{k-2}$, and so on. 
%$\mathsf{nd}_{k-1}$ is an alternative equal node (of some node) constructed from $\mathsf{nd}_{k-2}$, and so on. 
Finally, $\mathsf{nd}'_{2}$ is an alternative equal node of $\mathsf{nd}_{2}$ constructed from $\mathsf{nd}_{1}$ and $\mathsf{nd}_{1} \in \Queue_{dup}$.
\begin{lemma}\label{lem:necessary_and_sufficient_for_nd_in_Comb(Qdup)}
Let $\mathsf{nd}$ be a node in \textsc{dynamicHS}. Then, $\mathsf{nd} \in Comb(\Queue_{dup})$ 
%There is a node $\mathsf{nd} \in Comb(\Queue_{dup})$ 
%that is non-redundant w.r.t.\ the DPI $DPI$ 
iff there are nodes $\mathsf{nd}_1$, $\dots$, $\mathsf{nd}_k \in \Queue_{dup}$ for $k \geq 1$ such that 
\begin{enumerate}[(1)]
\item $|\mathsf{nd}_1| < \dots < |\mathsf{nd}_k| = |\mathsf{nd}|$,
%\item $\mathsf{nd}_1 \in \Queue_{dup}$,
%\item $\textsc{add}(\mathsf{nd}_i,\mathsf{nd}_{i+1}[|\mathsf{nd}_i|+1..|\mathsf{nd}_{i+1}|]$ is a node non-redundant w.r.t.\ $DPI$ and an element of $Comb(\Queue_{dup})$ for all $i \in \setof{1,\dots,k-1}$ and
\item it holds that
%\begin{itemize}
	%\item $\mathsf{nd}[i_1] = \mathsf{nd}_1[i_1]$ for $i_1 \in \setof{1,\dots,|\mathsf{nd}_1|}$,
	%\item $\mathsf{nd}[i_2] = \mathsf{nd}_2[i_2]$ for $i_2 \in \setof{|\mathsf{nd}_1|+1,\dots,|\mathsf{nd}_2|}$,
	%\item $\dots$,
	%\item $\mathsf{nd}[i_k] = \mathsf{nd}_k[i_k]$ for $i_k \in \setof{|\mathsf{nd}_{k-1}|+1,\dots,|\mathsf{nd}_k|}$.
%\end{itemize}
\begin{align*}
	\mathsf{nd}[i_1] &= \mathsf{nd}_1[i_1] \,\mbox{  for  }\, i_1 \in \setof{1,\dots,|\mathsf{nd}_1|} \\
	\mathsf{nd}[i_2] &= \mathsf{nd}_2[i_2] \,\mbox{  for  }\, i_2 \in \setof{|\mathsf{nd}_1|+1,\dots,|\mathsf{nd}_2|} \\
	&\dots \\
	\mathsf{nd}[i_k] &= \mathsf{nd}_k[i_k] \,\mbox{  for  }\, i_k \in \setof{|\mathsf{nd}_{k-1}|+1,\dots,|\mathsf{nd}_k|} %\mbox{    and}
\end{align*}
and
%\item $\mathsf{nd}_i[1..|\mathsf{nd}_i|] = \mathsf{nd}_{i+1}[1..|\mathsf{nd}_i|]$ for $i\in\setof{1,\dots,k-1}$.
\item $\mathsf{nd}_i$ is an alternative subnode of $\mathsf{nd}_{i+1}$ for $i\in\setof{1,\dots,k-1}$.
\end{enumerate}
\end{lemma}
\begin{proof}
``$\Rightarrow$'': Suppose $\mathsf{nd} \in Comb(\Queue_{dup})$ 
%is non-redundant w.r.t.\ $DPI$ 
and that $|\mathsf{nd}|=i$. Then, there are two cases, either $\mathsf{nd} \in \Queue_{dup}$ or $\mathsf{nd} \notin \Queue_{dup}$. 

In the former case, we can define $\mathsf{nd}_1$ as $\mathsf{nd}$ and the proposition of the lemma holds.

%and assuming that $S := \Queue_{dup}$
In the latter case, by proposition~\ref{cor:consequences_of_def_comb:enum:Combi_only_cardinality_i} of Corollary~\ref{cor:consequences_of_def_comb}, Definition~\ref{def:comb_node} and $|\mathsf{nd}|=i$, it holds that $\mathsf{nd} \in Comb_i(\Queue_{dup})$. By proposition~\ref{cor:consequences_of_def_comb:enum:partial_nodes} of
%the construction of $Comb(\Queue_{dup})$ (Definition~\ref{def:comb_node}), 
Corollary~\ref{cor:consequences_of_def_comb} and the fact that $\mathsf{nd} \in Comb_i(\Queue_{dup})$, there is some $\mathsf{nd}' \in Comb_j(\Queue_{dup})$ for some $j \in \setof{1,\dots,i-1}$ and some $\mathsf{nd}'' \in \Queue_{dup}$ with $|\mathsf{nd}''|=i$ such that $\mathsf{nd} = \textsc{add}(\mathsf{nd}',\mathsf{nd}''[j+1..i])$ and $\mathsf{nd.cs} = \textsc{add}(\mathsf{nd}'.\mathsf{cs},\mathsf{nd}''.\mathsf{cs}[j+1..i])$. Moreover, proposition~\ref{cor:consequences_of_def_comb:enum:partial_nodes} of Corollary~\ref{cor:consequences_of_def_comb} states that $\mathsf{nd}'$ is an alternative subnode of $\mathsf{nd}''$. 
%That is, in particular we have that $\mathsf{nd}'[1..|\mathsf{nd}'|] = \mathsf{nd}''[1..|\mathsf{nd}'|]$.   

So, set $\mathsf{nd}_k$ to $\mathsf{nd}''$ and $\mathsf{nd}_{k-1}$ to $\mathsf{nd}'$. Then, we obtain that $|\mathsf{nd}_{k-1}| < |\mathsf{nd}_{k}|$ by $j < i$, that $\mathsf{nd}_{k-1}$ is an alternative subnode of $\mathsf{nd}_k$
%, that $\mathsf{nd}_{k-1}[1..|\mathsf{nd}_{k-1}|] = \mathsf{nd}_k[1..|\mathsf{nd}_{k-1}|]$ 
and that $\mathsf{nd}[i_k] = \mathsf{nd}_k[i_k] \,\mbox{  for  }\, i_k \in \setof{|\mathsf{nd}_{k-1}|+1,\dots,|\mathsf{nd}_k|}$ must be true. That is, propositions (1), (2) and (3) hold for $\mathsf{nd}_k$ and $\mathsf{nd}_{k-1}$. 

Now, again, there are two cases for $\mathsf{nd}_{k-1}$, i.e.\ either $\mathsf{nd}_{k-1} \in \Queue_{dup}$ or $\mathsf{nd}_{k-1} \notin \Queue_{dup}$. 

In the former case, we can define $\mathsf{nd}_1$ as $\mathsf{nd}_{k-1}$ and the proposition of the lemma holds.

In the latter case, the same argumentation as for $\mathsf{nd}$ can be applied to show the existence of some $\mathsf{nd}_{k-2}$ that meets propositions (1), (2) and (3). Due to the fact that the cardinality of $\mathsf{nd}_{k-i-1}$ is strictly smaller than the cardinality of $\mathsf{nd}_{k-i}$ for all $i$ and the fact that $Comb_1(\Queue_{dup}) = \Queue_{dup}$, the case $\mathsf{nd}_{k-m} \in \Queue_{dup}$ must finally arise for some $m$.
%
%
%$\mathsf{nd}' \in Comb_j(\Queue_{dup})$ for some $j \in \setof{1,\dots,i-1}$ such that $\mathsf{nd} = \textsc{add}(\mathsf{nd}',\mathsf{nd}[|\mathsf{nd}'|+1..|\mathsf{nd}|])$ and $\mathsf{nd.cs} = \textsc{add}(\mathsf{nd}'.\mathsf{cs},\mathsf{nd.cs}[|\mathsf{nd}'|+1..|\mathsf{nd}|])$.
%By the second proposition of Corollary~\ref{cor:consequences_of_def_comb}, $j < i$ and $\mathsf{nd}' \in Comb_j(\Queue_{dup})$ it holds that $|\mathsf{nd}'| < |\mathsf{nd}|$. 
%%Due to the non-redundancy of $\mathsf{nd}$ w.r.t.\ $DPI$, we can deduce that $\mathsf{nd}'$ must be also non-redundant w.r.t.\ $DPI$. 
%By the fourth proposition of Corollary~\ref{cor:consequences_of_def_comb}, there must be some $\mathsf{nd}'' \in \Queue_{dup}$ such that 
%
%by Corollary~\ref{cor:consequences_of_def_comb}, there must be some node $\mathsf{nd}' \in \Queue_{dup}$ such that $\mathsf{nd}[|\mathsf{nd}|-k..|\mathsf{nd}|] = \mathsf{nd}'[|\mathsf{nd}|-k..|\mathsf{nd}|]$ for some $k \geq 0$. Since $\mathsf{nd} \notin \Queue_{dup}$ by assumption, $k < |\mathsf{nd}|-1$ must hold. Thus, there must be some $k'$ with $1\leq k' \leq |\mathsf{nd}|-1$ such that $\mathsf{nd}[k'] \neq \mathsf{nd}'[k']$. As $\mathsf{nd} \in Comb(\Queue_{dup})$, there must 

``$\Leftarrow$'': Suppose there are nodes $\mathsf{nd}_1,\dots,\mathsf{nd}_k \in \Queue_{dup}$ such that propositions~(1)-(3) are satisfied. 
%Then, there are two cases, either $|\mathsf{nd}_1| = |\mathsf{nd}|$ or $|\mathsf{nd}_1| \neq |\mathsf{nd}|$.
% is the same node as $\mathsf{nd}$ or not. 
Let $k = 1$. Then, by propositions~(1) and (2) of this lemma, we have that $\mathsf{nd}$ is the same node as $\mathsf{nd}_1$. Since $\mathsf{nd}_1\in\Queue_{dup}$ and by Definition~\ref{def:comb_node}, we have that $\mathsf{nd}\in Comb(\Queue_{dup})$. So, the lemma holds for $k=1$. 
%proposition~(2) of the lemma holds by simply defining $\mathsf{nd}_1$ to be the same node as $\mathsf{nd}$. Propositions~(1) and (3) of the lemma are trivially met. 

Now, assume that the lemma holds for $k = m$ for some natural number $m$. That is, assume that there is a node $\mathsf{nd} \in Comb(\Queue_{dup})$ if there are nodes $\mathsf{nd}_1,\dots,\mathsf{nd}_{m} \in \Queue_{dup}$ such that 
$|\mathsf{nd}_1| < \dots < |\mathsf{nd}_{m}|=|\mathsf{nd}|$,
\begin{align*}
	\mathsf{nd}[i_1] &= \mathsf{nd}_1[i_1] \,\mbox{  for  }\, i_1 \in \setof{1,\dots,|\mathsf{nd}_1|} \\
	\mathsf{nd}[i_2] &= \mathsf{nd}_2[i_2] \,\mbox{  for  }\, i_2 \in \setof{|\mathsf{nd}_1|+1,\dots,|\mathsf{nd}_2|} \\
	&\dots \\
	\mathsf{nd}[i_m] &= \mathsf{nd}_m[i_m] \,\mbox{  for  }\, i_m \in \setof{|\mathsf{nd}_{m-1}|+1,\dots,|\mathsf{nd}_m|} %\mbox{    and}
\end{align*}
and $\mathsf{nd}_i$ is an alternative subnode of $\mathsf{nd}_{i+1}$ for $i\in\setof{1,\dots,m-1}$.

Let now $k = m + 1$. That is, assume that there are nodes $\mathsf{nd}_1,\dots,\mathsf{nd}_{m+1} \in \Queue_{dup}$ such that $|\mathsf{nd}_1| < \dots < |\mathsf{nd}_{m+1}|=|\mathsf{nd}|$, 
\begin{align*}
	\mathsf{nd}'[i_1] &= \mathsf{nd}_1[i_1] \,\mbox{  for  }\, i_1 \in \setof{1,\dots,|\mathsf{nd}_1|} \\
	\mathsf{nd}'[i_2] &= \mathsf{nd}_2[i_2] \,\mbox{  for  }\, i_2 \in \setof{|\mathsf{nd}_1|+1,\dots,|\mathsf{nd}_2|} \\
	&\dots \\
	\mathsf{nd}'[i_{m+1}] &= \mathsf{nd}_{m+1}[i_{m+1}] \,\mbox{  for  }\, i_{m+1} \in \setof{|\mathsf{nd}_{m}|+1,\dots,|\mathsf{nd}_{m+1}|} %\mbox{    and}
\end{align*}
and $\mathsf{nd}_i$ is an alternative subnode of $\mathsf{nd}_{i+1}$ for $i\in\setof{1,\dots,m}$. What we need to show is that $\mathsf{nd}' \in Comb(\Queue_{dup})$.

If $\mathsf{nd}' \in \Queue_{dup}$, then, by Definition~\ref{def:comb_node}, the lemma is true. So suppose $\mathsf{nd}' \notin \Queue_{dup}$.

By the definition of an alternative subnode (Definition~\ref{def:alternative_equal_node}), $\mathsf{node}_{sub} \subseteq \mathsf{node}$ in case $\mathsf{node}_{sub}$ is an alternative subnode of $\mathsf{node}$. So, because $\mathsf{nd}_i$ is an alternative subnode of $\mathsf{nd}_{i+1}$ and $|\mathsf{nd}_i| < |\mathsf{nd}_{i+1}|$ for $i\in\setof{1,\dots,m}$, we have that $\mathsf{nd}_i \subset \mathsf{nd}_{i+1}$ for $i\in\setof{1,\dots,m}$. Consequently, $\mathsf{nd}_i \subseteq \mathsf{nd}'$ for $i\in\setof{1,\dots,m+1}$ and $\mathsf{nd}_i \subseteq \mathsf{nd}$ for $i\in\setof{1,\dots,m}$ must hold. Due to $|\mathsf{nd}_{m}|=|\mathsf{nd}|$ we obtain the set-equality between $\mathsf{nd}_{m}$ and $\mathsf{nd}$. This result along with $\mathsf{nd}_m \subseteq \mathsf{nd}'$ and $|\mathsf{nd}_{m}| < |\mathsf{nd}_{m+1}| = |\mathsf{nd}'|$ implies that $\mathsf{nd} \subset \mathsf{nd}'$. However, since  
\begin{align*}
	\mathsf{ndx}[i_1] &= \mathsf{nd}_1[i_1] \,\mbox{  for  }\, i_1 \in \setof{1,\dots,|\mathsf{nd}_1|} \\
	\mathsf{ndx}[i_2] &= \mathsf{nd}_2[i_2] \,\mbox{  for  }\, i_2 \in \setof{|\mathsf{nd}_1|+1,\dots,|\mathsf{nd}_2|} \\
	&\dots \\
	\mathsf{ndx}[i_m] &= \mathsf{nd}_m[i_m] \,\mbox{  for  }\, i_m \in \setof{|\mathsf{nd}_{m-1}|+1,\dots,|\mathsf{nd}_m|} %\mbox{    and}
\end{align*}
is met for $\mathsf{ndx}$ being the same node as $\mathsf{nd}$ as well as for $\mathsf{ndx}$ being the same node as $\mathsf{nd}'$, we can conclude that $\mathsf{nd}[i] = \mathsf{nd}'[i]$ for $i \in \setof{1, \dots, |\mathsf{nd}|}$. 
%So, either (a)~$\mathsf{nd}$ is a subnode of $\mathsf{nd}'$ or (b)~$\mathsf{nd}$ is an alternative subnode of $\mathsf{nd}'$.

Moreover, we have that $\mathsf{nd}'[i_{m+1}] = \mathsf{nd}_{m+1}[i_{m+1}] \,\mbox{  for  }\, i_{m+1} \in \setof{|\mathsf{nd}|+1,\dots,|\mathsf{nd}'|}$ since $|\mathsf{nd}| = |\mathsf{nd}_{m}|$ and $|\mathsf{nd}'| = |\mathsf{nd}_{m+1}|$.

Since $\mathsf{nd}' \notin \Queue_{dup}$ and $\mathsf{nd}_{m+1} \in \Queue_{dup}$ by assumption, we have that $\mathsf{nd}_{m+1}$ which is set-equal to $\mathsf{nd}'$ (as argued before) must be an alternative equal node of $\mathsf{nd}'$. That is, there must be some $j$ such that $\mathsf{nd}_{m+1}[j] \neq \mathsf{nd}'[j]$ or $\mathsf{nd}_{m+1}.\mathsf{cs}[j] \neq \mathsf{nd}'.\mathsf{cs}[j]$ for some $j \in \setof{1,\dots,|\mathsf{nd}|}$. Hence, $\mathsf{nd}_{m+1}[j] \neq \mathsf{nd}[j]$ or $\mathsf{nd}_{m+1}.\mathsf{cs}[j] \neq \mathsf{nd}.\mathsf{cs}[j]$ wherefore $\mathsf{nd}$ must be an alternative subnode of $\mathsf{nd}_{m+1}$. Because $\mathsf{nd} \in Comb(\Queue_{dup})$ and $\mathsf{nd}_{m+1} \in \Queue_{dup}$, we infer by Definition~\ref{def:comb_node} that $\mathsf{nd}' \in Comb(\Queue_{dup})$. 
\end{proof} 
The \textsc{prune} function (lines~\ref{algoline:update:call_prune_Q}-\ref{algoline:update:call_prune_Dsupset}) is called given a collection $S \in \setof{\Queue, \mD_{\times}, \mD_{\supset}}$, a minimal conflict set $X$ w.r.t.\ the current DPI and $\Queue_{dup}$ which has already been updated and cleaned from redundant nodes (w.r.t.\ the witness $X$) by the \textsc{pruneQdup} function. 
%
%That means, $\mathsf{nd}_{dup}$ is stored in $\Queue_{dup}$ in order to be available as an alternative equal node of $\mathsf{nd}$ or as an alternative subnode of some successor of $\mathsf{nd}$ in case $\mathsf{nd}$ is found to be redundant w.r.t.\ some current DPI.
%
So, %at the time \textsc{prune} is called, 
let $\mathsf{nd}_{dup}\in\Queue_{dup}$ be a (not necessarily proper) alternative subnode of some node $\mathsf{node}$ that is stored in $S$.
%some of the sets $\Queue$, $\mD_{\times}$ and $\mD_{\supset}$. 
Assume 
%some minimal conflict set $X$ w.r.t.\ the current DPI is found which 
$X$ is a witness of redundancy of $\mathsf{node}$. By Lemma~\ref{lem:pruneQdup} and since $\mathsf{nd}_{dup}\in\Queue_{dup}$, $X$ cannot be a witness of redundancy of $\mathsf{nd}_{dup}$. Further, let $r \in\setof{1,\dots,|\mathsf{node}|}$ be the highest number such that %there is some minimal conflict set $X$ w.r.t.\ $\langle\mo,\mb,\Tp\cup\Tp',\Tn\cup\Tn'\rangle_\RQ$ such that 
$X \subset \mathsf{node}.\mathsf{cs}[r]$ and $\mathsf{node}[r] \in \mathsf{node}.\mathsf{cs}[r]\setminus X$. Now, in case $r \leq |\mathsf{nd}_{dup}|$ holds, $\mathsf{nd}_{dup}$ (and $\mathsf{nd}_{dup}.\mathsf{cs}$) can be used to replace the first $|\mathsf{nd}_{dup}|$ elements of $\mathsf{node}$ (and $\mathsf{node}.\mathsf{cs}$). The result is an alternative equal node of $\mathsf{node}$ which is non-redundant w.r.t.\ the current DPI and which can be added to $S$ after deletion of $\mathsf{node}$ as a representative of the set (path) $\mathsf{node}$ has represented.
%So, let $\mathsf{nd}$ be a (not necessarily proper) subnode of $\mathsf{nd}_{suc}$ that is stored in some of the sets $\Queue$, $\mD_{\times}$ and $\mD_{\supset}$ and assume some minimal conflict set $X$ w.r.t.\ the current DPI is found which is a witness of redundancy of $\mathsf{nd}_{suc}$, but not a witness of redundancy of $\mathsf{nd}_{dup}$. Further, let $r \in\setof{1,\dots,|\mathsf{nd}|}$ be the highest number such that %there is some minimal conflict set $X$ w.r.t.\ $\langle\mo,\mb,\Tp\cup\Tp',\Tn\cup\Tn'\rangle_\RQ$ such that 
%$X \subset \mathsf{nd}_{suc}.\mathsf{cs}[r]$ and $\mathsf{nd}_{suc}[r] \in \mathsf{nd}_{suc}.\mathsf{cs}[r]\setminus X$. Then, $\mathsf{nd}_{dup}$ (and $\mathsf{nd.cs}$) can be used to replace the first $|\mathsf{nd}|$ elements of $\mathsf{nd}_{suc}$ (and $\mathsf{nd}_{suc}.\mathsf{cs}$). The result is an alternative equal node of $\mathsf{nd}_{suc}$ which is non-redundant w.r.t.\ the current DPI.

Now, the next lemma substantiates that \textsc{prune} updates $S$ in a way that all redundant nodes w.r.t.\ the witness $X$ are deleted, each deleted node is replaced by one non-redundant replacement node w.r.t.\ the witness $X$ if such a one is constructable from $\Queue_{dup}$ and for each remaining node $\mathsf{nd}$, i.e.\ $\mathsf{nd}$ is a non-deleted node or a replacement node of some deleted node,  each superset of $X$ in $\mathsf{nd.cs}$ is replaced by $X$. 

This leads to a new set $S$ returned by \textsc{prune} which includes only non-redundant nodes w.r.t.\ the witness $X$. Furthermore, the new set $S$ contains a node corresponding to each set (path) $Y$ for which there was a corresponding node in the old set $S$ if there would be a non-redundant (w.r.t.\ $X$) node corresponding to $Y$ in a hitting set tree equal to the one produced by \textsc{dynamicHS} except that all duplicate nodes corresponding to equal sets (paths) would be regularly processed and expanded.

\begin{lemma}\label{lem:prune} 
Let $\langle\mo,\mb,\Tp,\Tn\rangle_\RQ$ be a DPI and let the following be the input parameters to the \textsc{prune} function:
\begin{itemize}
\item $X$ is a minimal conflict set w.r.t.\ $\langle\mo,\mb,\Tp,\Tn\rangle_\RQ$,
\item $S$ is a set of nodes in \textsc{dynamicHS},
\item $Dup$ is a set of nodes %sorted from low to high cardinality 
where 
\begin{itemize}
\item $X$ is not a witness of redundancy of any node in $Dup$ and 
\item for each $\mathsf{nd}\in S$ there might be some $\mathsf{nd}' \in Dup$ such that $\mathsf{nd}'$ is an alternative subnode of $\mathsf{nd}$ and
\item for each node $\mathsf{nd} \in Dup$ there is no $r \in \setof{1,\dots,|\mathsf{nd}|}$ for which $\mathsf{nd.cs}[r] \supset X$.
\end{itemize}
\item $p_{nodes}$ is as defined by Definition~\ref{def:p_node()}. 
\end{itemize}
Then, \textsc{prune} returns $S'$ where the following holds: 
\begin{enumerate}[(1)]
\item $S'$ is a set such that $S\setminus S'$ includes exactly these nodes in $S$ for which $X$ is a witness of redundancy and $S\cap S'$ includes exactly these nodes in $S$ for which $X$ is not a witness of redundancy.
\item Each element $\mathsf{nd}\in S'\setminus S$ is an alternative equal node of some node in $S\setminus S'$ constructed from some node in $Dup$ such that $X$ is not a witness of redundancy of $\mathsf{nd}$.
\item Let $\mathsf{nd}\in S\setminus S'$ and $Alt_{\mathsf{nd}}$ denote the set of all alternative equal nodes of $\mathsf{nd}$, each of which can be constructed from some node in $Dup$ and for each of which $X$ is not a witness of redundancy. Then there is some $\mathsf{nd}' \in Alt_{\mathsf{nd}}$ such that $\mathsf{nd}' \in S'\setminus S$.
%\item If $S'\setminus S$ includes an alternative equal node $\mathsf{node}$ of a node $\mathsf{nd}\in S\setminus S'$, then $X$ is not a witness of redundancy of $\mathsf{node}$. 
\item $S'$ includes only nodes $\mathsf{nd}$ such that there is no $r \in \setof{1,\dots,|\mathsf{nd}|}$ for which $\mathsf{nd.cs}[r] \supset X$.  
%\end{itemize}
\end{enumerate}
\end{lemma}
\begin{proof}
%The set $S'$ is equal to the (modified) set $S$ at the end of the execution of \textsc{prune}. 
The \textsc{prune} procedure runs through all nodes $\mathsf{nd}\in S$ and for each $\mathsf{nd}$ runs through all sets in $\mathsf{nd.cs}$ 
%(starting from the root node) 
(lines~\ref{algoline:prune:for_nd_in_S} and \ref{algoline:prune:for_cs_in_nd.cs}). Lines~\ref{algoline:prune:redundancy_check_part1} and \ref{algoline:prune:redundancy_check_part2} perform a check whether $X$ is a witness of redundancy of $\mathsf{nd}$, implementing exactly the criteria given by Definition~\ref{def:redundant_node}. If the check is not successful for any 
%$\mathsf{nd.cs}[i]$ for 
$i \in \setof{1,\dots,|\mathsf{nd}|}$, i.e.\ $X$ is not a witness of redundancy of $\mathsf{nd}$, then $k=0$ must hold when line~\ref{algoline:prune:if_k>0} is reached. Hence, $\mathsf{nd}$ is added to $S'$ in line~\ref{algoline:prune:insert_same_node_into_S'} in this case. As only nodes different from $\mathsf{nd}$ can be added to $S'$ in line~\ref{algoline:prune:insert_alternative_equal_node_into_S'} and as there are no other ways nodes might be added to $S'$, we have that $S\setminus S'$ includes exactly these nodes in $S$ for which $X$ is a witness of redundancy and $S\cap S'$ includes exactly these nodes in $S$ for which $X$ is not a witness of redundancy. So, proposition~(1) is true.
%
 %(at least once), then $k > 0$ holds by line~\ref{algoline:prune:k_gets_i} and $\mathsf{nd}$ is not added to $S'$ in line~\ref{algoline:prune:insert_same_node_into_S'}. Otherwise,  Since line~\ref{algoline:prune:delete_from_S} is the only place where nodes are deleted from $S$ and no other nodes are deleted from $S$ throughout \textsc{prune}, 
%%in the end of the procedure's execution, $S$ does not include any nodes for which $X$ is a witness of redundancy, i.e.\ 
%it holds that   

The truth of proposition~(2) can be derived as follows: 
By the proof of proposition~(1), line~\ref{algoline:prune:insert_alternative_equal_node_into_S'} is the only place where nodes that are not elements of $S$ are added to $S'$. Hence, each node in $S'\setminus S$ must be added to $S'$ in line~\ref{algoline:prune:insert_alternative_equal_node_into_S'}. 
%Line~\ref{algoline:prune:insert_alternative_equal_node_into_S'} can only be reached if $k > 0$ for $\mathsf{nd}$ which implies by Definition~\ref{def:redundant_node} that $X$ is a witness of redundancy of $\mathsf{nd}$. By lines~\ref{algoline:prune:for_cs_in_nd.cs}-\ref{algoline:prune:nd.cs[i]_gets_X}, $k$
%
%By the truth of proposition~(1), it holds for
Thus, only nodes $\mathsf{node}_{new} := \textsc{add}(\mathsf{node},\mathsf{nd}[|\mathsf{node}|+1..|\mathsf{nd}|])$ with $\mathsf{node}_{new}.cs := \textsc{add}(\mathsf{node.cs},\mathsf{nd.cs}[|\mathsf{node}|+1..|\mathsf{nd}|])$ constructed exactly as per Definition~\ref{def:alternative_equal_node} in lines~\ref{algoline:prune:construct_alternative_equal_node} and \ref{algoline:prune:construct_alternative_equal_node.cs} where $\mathsf{nd}\in S$ can be added to $S'$. 
%Clearly $\mathsf{nd} \in S \setminus S'$ since $\mathsf{nd}$ must have been deleted from $S$ in line~\ref{algoline:prune:delete_from_S} in order for line~\ref{algoline:prune:construct_alternative_equal_node} to be reached.

Now, we still have to show that $\mathsf{node}$ is an alternative subnode of $\mathsf{nd}$. From the precondition that $X$ is not a witness of redundancy of any node in $Dup$, $X$ cannot be a witness of redundancy of $\mathsf{node}$. Moreover, $|\mathsf{node}| \geq k$ must hold as line~\ref{algoline:prune:check_if_alternative_subnode} has been passed. So, we have that $X$ must be a witness of redundancy for $\mathsf{nd}[1..|\mathsf{node}|]$ since $k > 0$ (line~\ref{algoline:prune:if_k>0}) and by the way $k$ is constructed (lines~\ref{algoline:prune:k_gets_0}-\ref{algoline:prune:k_gets_i}). Hence, there must be some $j \in \setof{1,\dots,|\mathsf{node}|}$ with the property that $\mathsf{node}[j] \neq \mathsf{nd}[j]$ or $\mathsf{node}.\mathsf{cs}[j] \neq \mathsf{nd.cs}[j]$ wherefore $\mathsf{node}$ is indeed an alternative subnode of $\mathsf{nd}$. Thus, $\mathsf{node}_{new}$ is an alternative equal node of $\mathsf{nd}$ by Definition~\ref{def:alternative_equal_node}.

That $\mathsf{nd}\in S\setminus S'$ must be true can be explained as follows. By the argumentation to prove proposition~(1) and (2) so far, we know that only nodes can be added to $S'$ in line~\ref{algoline:prune:insert_alternative_equal_node_into_S'} and line~\ref{algoline:prune:insert_same_node_into_S'} for which $X$ is not a witness of redundancy. Moreover, we have shown that line~\ref{algoline:prune:insert_alternative_equal_node_into_S'} can only be reached for some node $\mathsf{nd}\in S$ for which $X$ is a witness of redundancy. Consequently, $\mathsf{nd} \notin S'$ must hold.

That $X$ is not a witness of redundancy of $\mathsf{node}_{new}$ can be derived as follows: From the precondition that $X$ is not a witness of redundancy of any node in $Dup$, $X$ cannot be a witness of redundancy of $\mathsf{node}_{new}[1..|\mathsf{node}|]$ with $\mathsf{node}_{new}.\mathsf{cs}[1..|\mathsf{node}|]$ since $\mathsf{node}_{new}[j] = \mathsf{node}[j]$ and $\mathsf{node}_{new}.\mathsf{cs}[j] = \mathsf{node.cs}[j]$ for all $j \in \setof{1,\dots,|\mathsf{node}|}$. $k$ is the maximum index such that $X\subset\mathsf{nd.cs}[k]$ and $\mathsf{nd}[k]\in \mathsf{nd.cs}[k] \setminus X$ by lines~\ref{algoline:prune:k_gets_0}-\ref{algoline:prune:k_gets_i}. Since $|\mathsf{node}| \geq k$, $X$ cannot be a witness of redundancy of $\mathsf{node}_{new}[|\mathsf{node}|+1..|\mathsf{nd}|]$ with $\mathsf{node}_{new}.\mathsf{cs}[|\mathsf{node}|+1..|\mathsf{nd}|]$ either since $\mathsf{node}_{new}[j] = \mathsf{nd}[j]$ and $\mathsf{node}_{new}.\mathsf{cs}[j] = \mathsf{nd.cs}[j]$ for all $j \in \setof{|\mathsf{node}|+1,\dots,|\mathsf{nd}|}$. Therefore, $X$ cannot be a witness of redundancy of $\mathsf{node}_{new}$.

Proposition~(3): As already argued, for each node $\mathsf{nd} \in S \setminus S'$, line~\ref{algoline:prune:check_for_alternative_paths_start} must be reached. 
Then, in line~\ref{algoline:prune:check_for_alternative_paths_start}, \emph{all} nodes in $Dup$ are investigated in order to find an alternative subnode of $\mathsf{nd}$. So, if there is such a one, then it must be found.

Proposition~(4): For a node $\mathsf{nd}$ that is added to $S'$ in line~\ref{algoline:prune:insert_same_node_into_S'}, the for-loop in line~\ref{algoline:prune:for_cs_in_nd.cs} must have been executed. Since, as already shown, line~\ref{algoline:prune:k_gets_i} cannot be executed for a node that is added to $S'$ in line~\ref{algoline:prune:insert_same_node_into_S'}, line~\ref{algoline:prune:nd.cs[i]_gets_X} must have been executed for all $i \in \setof{1, \dots, |\mathsf{nd}|}$. Hence, proposition~(4) holds for all nodes inserted into $S'$ in line~\ref{algoline:prune:insert_same_node_into_S'}. 

For nodes 
\begin{align*}
\mathsf{node}_{new} &:= \textsc{add}(\mathsf{node},\mathsf{nd}[|\mathsf{node}|+1..|\mathsf{nd}|]) \\
\mathsf{node}_{new}.cs &:= \textsc{add}(\mathsf{node.cs},\mathsf{nd.cs}[|\mathsf{node}|+1..|\mathsf{nd}|])
\end{align*}
%$\mathsf{node}_{new} := \textsc{add}(\mathsf{node},\mathsf{nd}[|\mathsf{node}|+1..|\mathsf{nd}|])$ with $\mathsf{node}_{new}.cs := \textsc{add}(\mathsf{node.cs},\mathsf{nd.cs}[|\mathsf{node}|+1..|\mathsf{nd}|])$ 
inserted into $S'$ in line~\ref{algoline:prune:insert_alternative_equal_node_into_S'}, proposition~(4) follows from the precondition that $Dup$ includes only nodes $\mathsf{n}$ such that there is no $r \in \setof{1,\dots,|\mathsf{n}|}$ for which $\mathsf{n.cs}[r] \supset X$, from the fact that $\mathsf{node} \in Dup$ and the fact that line~\ref{algoline:prune:nd.cs[i]_gets_X} must have been executed for all indices $i > k$.
%
%This follows from the fact that, in lines~\ref{algoline:prune:check_for_alternative_paths_start} and \ref{algoline:prune:check_if_alternative_subnode}, for each node $\mathsf{nd}$ deleted from $S$ in line~\ref{algoline:prune:delete_from_S}, a check whether $\mathsf{nd}'$ enables the construction of an alternative equal node $\mathsf{node}_{new}$ of $\mathsf{nd}$ such that $X$ is not a witness of redundancy of $\mathsf{node}_{new}$ is made for \emph{all} nodes $\mathsf{nd}' \in Dup$.
\end{proof}
\subsection[De-Facto Non-Redundant Nodes]{De-Facto Non-Redundant Nodes in \textsc{dynamicHS}}
\label{sec:DeFactoNonRedundantNodesInTextscDynamicHS}
The following definition introduces a notion that is of rather theoretical use for the proof of completeness of \textsc{dynamicHS} we will give later. The definition assumes a fixed $DPI$ and characterizes as active sublabel of a particular conflict set $\mathsf{nd.cs}[r]$ in $\mathsf{nd.cs}$ the subset of $\mathsf{nd.cs}[r]$ that ``survives'' all the pruning steps, i.e.\ \textsc{pruneQdup} and \textsc{prune} calls,  during all executions of \textsc{dynamicHS} up to the one with a current DPI $DPI$. Notice that the shape of the active sublabel can never be known in advance as we do not know which witnesses of redundancy might be found. This makes up the theoretical nature of this definition. However, we will be able to show that no active sublabel of a node can be the empty set under certain preconditions that are met for \textsc{dynamicHS}.
\begin{definition}\label{def:active_sublabel}
Let
\begin{itemize} 
	\item $\mathsf{nd}$ be a node in \textsc{dynamicHS},
	\item $r\in \setof{1,\dots,|\mathsf{nd}|}$ fixed,
	\item $DPI_1, \dots, DPI_n$ be a sequence of DPIs where $DPI_j$ includes a proper subset of the test cases $DPI_{j+1}$ includes for $j \in \setof{1,\dots,n-1}$,
	\item $DPI_n$ is equal to $DPI$ or includes a proper subset of the test cases $DPI$ includes,
	\item $\mc_1,\dots,\mc_n$ be the chronological sequence of all sets $X$ given as an argument to \textsc{prune} and \textsc{pruneQdup} during all executions of \textsc{dynamicHS} up to and including the one with current DPI $DPI$ where 
	\begin{itemize}
		\item each $\mc_i$ is a minimal conflict set w.r.t.\ $DPI_i$ for $i \in \setof{1,\dots,n}$
		\item $\mc_k \supset \mc_{k+1}$ for $k \in \setof{1,\dots,n-1}$,
		\item $\mathsf{nd.cs}[r] \supset \mc_1$. 
	\end{itemize}
\end{itemize}
Then, we call $\mc_n$ the active sublabel of $\mathsf{nd.cs}[r]$ w.r.t.\ $DPI$.
\end{definition}
The next definition of a de-facto non-redundant node is based on Definition~\ref{def:active_sublabel}. A de-facto non-redundant node w.r.t.\ $DPI$ includes at each position an element that hits the active sublabel w.r.t.\ $DPI$ at this position. Again, this definition is of theoretical rather than practical use, but crucial for the proof of completeness of \textsc{dynamicHS}. In fact, we will be able to show that for each minimal diagnosis w.r.t.\ $DPI$ there must be -- anytime during any execution of \textsc{dynamicHS} with a current DPI including a subset of the test cases in $DPI$ -- a de-facto non-redundant node corresponding to a subset of this diagnosis. In further consequence, this will allow us to derive the algorithm's completeness concerning the detection of all minimal diagnoses w.r.t.\ $DPI$.
\begin{definition}\label{def:de-facto_non-redundant}
%We call a node $\mathsf{nd}$ in \textsc{dynamicHS} de-facto non-redundant w.r.t.\ $DPI$ iff $\mathsf{nd.cs}[r]$ is an active sublabel w.r.t.\ $DPI$ and $\mathsf{nd}[r] \in \mathsf{nd.cs}[r]$ for all $r \in \setof{1,\dots,|\mathsf{nd}|}$.
We call a node $\mathsf{nd}$ in \textsc{dynamicHS} de-facto non-redundant w.r.t.\ $DPI$ iff 
$\mathsf{nd}[r]$ is an element of an active sublabel w.r.t.\ $DPI$ for all $r \in \setof{1,\dots,|\mathsf{nd}|}$.
\end{definition}
A de-facto non-redundant node w.r.t.\ a DPI $DPI$ ``survives'' all pruning steps at least until the execution of \textsc{dynamicHS} with current DPI $DPI$:
\begin{proposition}\label{prop:de-facto_non_redundant_node_cannot_be_pruned_or_replaced}
Let $\mathsf{nd}$ be a node which is de-facto non-redundant w.r.t.\ $DPI$. Then, $\mathsf{nd}$ cannot be pruned or replaced during any execution of \textsc{dynamicHS} up to and including the one with current DPI $DPI$.
\end{proposition}
\begin{proof}
By Definitions~\ref{def:active_sublabel} and \ref{def:de-facto_non-redundant}, \textsc{prune} and \textsc{pruneQdup} cannot be called given a witness of redundancy of $\mathsf{nd}$ during any execution of \textsc{dynamicHS} up to and including the one with current DPI $DPI$. By Lemmata~\ref{lem:pruneQdup} and \ref{lem:prune}, only nodes can be pruned or replaced for which the input set $X$ given to \textsc{prune} and \textsc{pruneQdup} is a witness of redundancy.
\end{proof}

\begin{example}\label{example:de-facto_non-redundant}
Let $\mo = \setof{1,\dots,10}$ be the KB of the (admissible) input DPI $DPI_0$ to Algorithm~\ref{algo:inter_onto_debug} and let $\mathsf{nd} := [1,2,3,4]$ with $\mathsf{nd.cs} := [\tuple{1,5,7},\tuple{2,4,6},\tuple{3,6,7},\tuple{4,5}]$ be a node 
%in $\Queue$ 
stored by \textsc{dynamicHS} during the execution of some call to \textsc{dynamicHS} during Algorithm~\ref{algo:inter_onto_debug}. Moreover, let $DPI$ be a fixed DPI constructed during the execution Algorithm~\ref{algo:inter_onto_debug} that includes a (not necessarily proper) superset of the test cases in $DPI_0$. 
%where each test case in $DPI$ that is not in $DPI_0$ is a query computed by Algorithm~\ref{}. 
Assume that the chronological sequence of all inputs $X$ to \textsc{prune} and \textsc{pruneQdup} throughout all executions of \textsc{dynamicHS} up to and including the one with current DPI $DPI$ during Algorithm~\ref{algo:inter_onto_debug} and after $\mathsf{nd}$ has been generated is given by $\tuple{1,6},\tuple{3,7},\tuple{1,3,8},\tuple{2},\tuple{4},\tuple{1,5}$. 

Then $\mathsf{nd.cs}$ %for $\mathsf{nd}$
%$\mathsf{nd} \in \Queue$ 
undergoes the transition depicted by Table~\ref{tab:example:de-facto_non-redundant} induced by this sequence of $X$ arguments to \textsc{prune}/\textsc{pruneQdup}.
\begin{table}[h]
\centering
\begin{tabular}{c|l}
%input $X$ to \textsc{prune}/\textsc{pruneQdup} & $\mathsf{nd.cs}$ after execution of \textsc{prune}/\textsc{pruneQdup} \\
$X$ & $\mathsf{nd.cs}$ after \textsc{prune}/\textsc{pruneQdup} with argument $X$ \\
\hline
 \tuple{1,6}& \quad [\tuple{1,5,7},\tuple{2,4,6},\tuple{3,6,7},\tuple{4,5}] \\
 \tuple{3,7}& \quad [\tuple{1,5,7},\tuple{2,4,6},\tuple{3,7},\tuple{4,5}] \\
 \tuple{1,3,8}& \quad [\tuple{1,5,7},\tuple{2,4,6},\tuple{3,7},\tuple{4,5}] \\
 \tuple{2}& \quad [\tuple{1,5,7},\tuple{2},\tuple{3,7},\tuple{4,5}] \\
 \tuple{4}& \quad [\tuple{1,5,7},\tuple{2},\tuple{3,7},\tuple{4}] \\
 \tuple{1,5}& \quad [\tuple{1,5},\tuple{2},\tuple{3,7},\tuple{4}] \\
\hline 
\end{tabular}
\caption[Transition of Node Labels due to Tree Pruning]{Transition of $\mathsf{nd.cs}$ induced by multiple calls to \textsc{prune}.}
\label{tab:example:de-facto_non-redundant}
\vspace{-10pt}
\end{table}
%\begin{tabular}
%\mbox{input } X \mbox{ to \textsc{prune}} & \mathsf{nd.cs} after execution of \textsc{prune} \\
%\hrule
 %\tuple{1,4}:& \quad [\tuple{1,5,7},\tuple{2,6},\tuple{3,6,7},\tuple{4,5}] \\
 %\tuple{3,7}:& \quad [\tuple{1,5,7},\tuple{2,6},\tuple{3,6,7},\tuple{4,5}] \\
 %\tuple{1,3,8}:& \quad [\tuple{1,5,7},\tuple{2,6},\tuple{3,6,7},\tuple{4,5}] \\
 %\tuple{2}:& \quad [\tuple{1,5,7},\tuple{2,6},\tuple{3,6,7},\tuple{4,5}] \\
 %\tuple{4}:& \quad [\tuple{1,5,7},\tuple{2,6},\tuple{3,6,7},\tuple{4,5}] \\
 %\tuple{1,5}:& \quad [\tuple{1,5,7},\tuple{2,6},\tuple{3,6,7},\tuple{4,5}] \\
%\hrule 
%\end{tabular}
We can observe in Table~\ref{tab:example:de-facto_non-redundant} that each proper superset of some argument $X$ of \textsc{prune}/\textsc{pruneQdup} that occurs in $\mathsf{nd.cs}$ is replaced by $X$ (cf.\ Lemmata~\ref{lem:pruneQdup} and \ref{lem:prune}). This is the case, for instance, for $X = \tuple{3,7}$ in the second row of the table which replaces $\mathsf{nd.cs}[3] = \tuple{3,6,7}$. Similar situations can be found in rows 4-6. No changes to $\mathsf{nd.cs}$ are triggered for $X = \tuple{1,6}$ or $X = \tuple{1,3,8}$ in rows 1 and 3, respectively, because at this stage $\mathsf{nd.cs}$ does not include any superset of $X$.

We learn from the last row of the table that $\mathsf{nd}$ is de-facto non-redundant w.r.t.\ $DPI$. This holds, first, since we considered the chronological sequence of \emph{all} inputs $X$ to \textsc{prune} and \textsc{pruneQdup} throughout \emph{all} executions of \textsc{dynamicHS} up to and including the one with current DPI $DPI$ during Algorithm~\ref{algo:inter_onto_debug}. Second, we have that
\begin{align*}
\mathsf{nd}[1] = 1 \quad\in&\quad \tuple{1,5} = \mathsf{nd.cs'}[1] \\
\mathsf{nd}[2] = 2 \quad\in&\quad \tuple{2} = \mathsf{nd.cs'}[2] \\
\mathsf{nd}[3] = 3 \quad\in&\quad \tuple{3,7} = \mathsf{nd.cs'}[3] \\
\mathsf{nd}[4] = 4 \quad\in&\quad \tuple{4} = \mathsf{nd.cs'}[4]
\end{align*}
where $\mathsf{nd.cs'}$ is the value of $\mathsf{nd.cs}$ given by the last row of the table which is the ``current'' value of $\mathsf{nd.cs}$ during the execution of \textsc{dynamicHS} with current DPI $DPI$. By Definition~\ref{def:active_sublabel}, $\mathsf{nd.cs'}[i]$ is the active sublabel of $\mathsf{nd.cs}[i]$ w.r.t.\ $DPI$ for $i \in \setof{1,\dots,4}$. That is, for example, $\tuple{3,7}$ is the active sublabel of $\mathsf{nd.cs}[3]$. As we realized that each element of $\mathsf{nd}$ is an element of an active sublabel w.r.t.\ $DPI$, we obtain the de-facto non-redundancy of $\mathsf{nd}$ w.r.t.\ $DPI$ as per Definition~\ref{def:de-facto_non-redundant}.

Notice that the sole definition of redundancy of a node w.r.t.\ $DPI$ (Definition~\ref{def:redundant_node}) does not perfectly serve our purposes as it does not take into account the order in which new conflict sets emerge and are used for pruning. 

For instance, consider $\mathsf{nd.cs}[2] = \tuple{2,4,6}$ which includes $2$ as well as $4$. Both values $\tuple{2}$ and $\tuple{4}$ of $X$ in rows 4 and 5 of Table~\ref{tab:example:de-facto_non-redundant} must be conflict sets w.r.t.\ $DPI$ by Proposition~\ref{prop:changes_in_conflict_sets_after_testcase_added}, which says that conflict sets cannot grow after the addition of a test case to a DPI, and the fact that each $X$ must be a minimal conflict set w.r.t.\ some DPI including a subset of the test cases in $DPI$. 
%Hence, $\tuple{2}$ and $\tuple{4}$ are two conflict sets w.r.t.\ $DPI$. 
In fact, by Proposition~\ref{prop:if_DPI_cs_neq_emptyset_then_DPI+1_cs_neq_emptyset} and the admissibility of $DPI_0$, $\tuple{2}$ and $\tuple{4}$ are even \emph{minimal} conflict sets w.r.t.\ $DPI$. Thus, application of Definition~\ref{def:redundant_node} yields that $\mathsf{nd}$ is redundant w.r.t.\ $DPI$ because $\tuple{4} \subset \tuple{2,4,6}$ and $\mathsf{nd}[2] = 2 \in \tuple{2,4,6} \setminus \tuple{4}$ (cf.\ Definition~\ref{def:redundant_node}). 
However, bearing in mind that $\tuple{2}$ was known to the algorithm before $\tuple{4}$, or, $\tuple{2}$ was used for pruning before $\tuple{4}$, we have that the set $\mathsf{nd.cs}[2]$, after being modified by \textsc{prune} or \textsc{pruneQdup}, is not redundant w.r.t.\ $DPI$. This is true since the new set $\mathsf{nd.cs}[2] = \tuple{2}$ which is not a superset of $\tuple{4}$.

So, to summarize, a node is (theoretically) redundant w.r.t.\ $DPI$ as per Definition~\ref{def:redundant_node} iff there is a minimal conflict set w.r.t.\ $DPI$ which is a witness of redundancy of this node. As however the example above has shown, whether a node is found to be redundant or not depends on the order of conflict sets used for pruning. This fact is also mentioned in \cite{greiner1989correction}. And, a (theoretically) redundant node w.r.t.\ $DPI$ does not necessarily need to be discovered by \textsc{dynamicHS} and might be modified by \textsc{prune} or \textsc{pruneQdup} in a way that it becomes non-redundant w.r.t.\ $DPI$. 

On the other hand, the definition of de-facto non-redundancy w.r.t.\ $DPI$ (Definition~\ref{def:de-facto_non-redundant}) incorporates exactly these thoughts and declares only nodes as de-facto non-redundant w.r.t.\ $DPI$ which are \emph{actually} not found to be redundant w.r.t.\ $DPI$. \qed
%
%Whether it is or is not redundant depends on ``the choice of the algorithm'', i.e.\ which of these two minimal conflict sets is detected first. 
%
 %If $\tuple{4}$ had been called before $\tuple{2}$, then we would face the situation where $\mathsf{nd}$ is redundant w.r.t.\ $DPI$
%
 %
%%-- which might be a subnode of another node --   
\end{example}
The criteria for a node $\mathsf{nd}$ to be a combined node of $\Queue_{dup}$ given by Lemma~\ref{lem:necessary_and_sufficient_for_nd_in_Comb(Qdup)} will facilitate the proof of the next lemma. This lemma states that a combined node of $\Queue_{dup}$ which is non-redundant w.r.t.\ some DPI $\langle\mo,\mb,\Tp\cup\Tp',\Tn\cup\Tn'\rangle_\RQ$ cannot be pruned during \textsc{dynamicHS} given i.a.\ the DPI $\langle\mo,\mb,\Tp,\Tn\rangle_\RQ$ and sets of positively and negatively answered queries $\Tp''$ and $\Tn''$ as input where $\Tp'' \subseteq \Tp'$ and $\Tn'' \subseteq \Tn'$. This result will constitute an essential prerequisite for the proof of completeness of \textsc{dynamicHS}.
\begin{lemma}\label{lem:non-redundant_node_in_Comb(Qdup)_persists} 
Let $\mathsf{nd} \in Comb(\Queue_{dup})$ be some node that is de-facto non-redundant w.r.t.\ the DPI $DPI$ and let $DPI'$ be some DPI that is either equal to $DPI$ or includes only a subset of the test cases of $DPI$. Then, throughout any execution of \textsc{dynamicHS} using the current DPI $DPI'$, $\mathsf{nd} \in Comb(\Queue_{dup})$ holds.
\end{lemma}
\begin{proof}
%Let $DPI''$ be the DPI that was the current DPI when $\mathsf{nd}$ was generated. Then, $DPI'$ must include a (not necessarily proper) superset of the test cases in $DPI''$ because $\mathsf{nd}$ must have been already generated in order to be an element of 
%
First, we show that there cannot be a minimal conflict set $\mc$ w.r.t.\ $DPI'$ such that \textsc{pruneQdup} is called with $X := \mc$ and there is some $q \in \setof{1,\dots,|\mathsf{nd}|}$ with the property that $\mc \subset \mathsf{nd.cs}[q]$ and $\mathsf{nd}[q] \in \mathsf{nd.cs}[q] \setminus \mc$. 

So, assume that \textsc{pruneQdup} is called with $X := \mc$ and there is some $\mc$ w.r.t.\ $DPI'$ such that there is some $q \in \setof{1,\dots,|\mathsf{nd}|}$ with the property that $\mc \subset \mathsf{nd.cs}[q]$ and $\mathsf{nd}[q] \in \mathsf{nd.cs}[q] \setminus \mc$. Let now $\mc_1,\dots,\mc_n$ be the (arbitrary actual) chronological sequence of all sets $X$ given as an argument to \textsc{prune} and \textsc{pruneQdup} during all executions of \textsc{dynamicHS} up to and including the one with current DPI $DPI$ where 
\begin{itemize}
		\item $\mathsf{nd.cs}[q] \supset \mc_1$,
		\item each $\mc_i$ is a minimal conflict set w.r.t.\ $DPI_i$ for $i \in \setof{1,\dots,n}$
		\item $\mc_k \supset \mc_{k+1}$ for $k \in \setof{1,\dots,n-1}$,
		\item $DPI_j$ includes a proper subset of the test cases $DPI_{j+1}$ includes for $j \in \setof{1,\dots,n-1}$,
		\item $DPI_n$ is equal to $DPI$ or includes a proper subset of the test cases $DPI$ includes.
\end{itemize}
Then, $\mc_n$ is the active sublabel of $\mathsf{nd.cs}[q]$ w.r.t.\ $DPI$. 
Since $\mc \subset \mathsf{nd.cs}[q]$ and $X := \mc$ is an argument of \textsc{pruneQdup} during $DPI'$, we have that $\mc$ must be equal to some set $\mc_j$ in the sequence $\mc_1,\dots,\mc_n$. By Definition~\ref{def:de-facto_non-redundant} and the de-facto non-redundancy of $\mathsf{nd}$ w.r.t.\ $DPI$, $\mathsf{nd.cs}[q] \in \mc_n$ must hold. By $\mc_n \subseteq \mc_j = \mc$, we finally obtain $\mathsf{nd.cs}[q] \in \mc$, which is a contradiction to $\mathsf{nd}[q] \in \mathsf{nd.cs}[q] \setminus \mc$.

Lemma~\ref{lem:necessary_and_sufficient_for_nd_in_Comb(Qdup)} and $\mathsf{nd} \in Comb(\Queue_{dup})$ guarantee the existence of nodes $\mathsf{nd}_1,\dots,\mathsf{nd}_k \in \Queue_{dup}$ for $k \geq 1$ such that 
\begin{enumerate}[(1)]
\item $|\mathsf{nd}_1| < \dots < |\mathsf{nd}_k| = |\mathsf{nd}|$,
\item it holds that
\begin{align*}
	\mathsf{nd}[i_1] &= \mathsf{nd}_1[i_1] \,\mbox{  for  }\, i_1 \in \setof{1,\dots,|\mathsf{nd}_1|} \\
	\mathsf{nd}[i_2] &= \mathsf{nd}_2[i_2] \,\mbox{  for  }\, i_2 \in \setof{|\mathsf{nd}_1|+1,\dots,|\mathsf{nd}_2|} \\
	&\dots \\
	\mathsf{nd}[i_k] &= \mathsf{nd}_k[i_k] \,\mbox{  for  }\, i_k \in \setof{|\mathsf{nd}_{k-1}|+1,\dots,|\mathsf{nd}_k|} 
\end{align*}
and
\item $\mathsf{nd}_i$ is an alternative subnode of $\mathsf{nd}_{i+1}$ for $i\in\setof{1,\dots,k-1}$.
\end{enumerate}

So, let us assume that $\mathsf{nd} \notin Comb(\Queue_{dup})$ at some point in time during the execution of \textsc{dynamicHS} using the current DPI $DPI'$. That is, some node $\mathsf{nd}_j$ for some $j \in \setof{1,\dots,k}$ must have been deleted from $\Queue_{dup}$. Nodes can only be deleted from $\Queue_{dup}$ in the scope of the function \textsc{pruneQdup}. 
By Lemma~\ref{lem:pruneQdup} and Corollary~\ref{cor:pruneQdup_only_called_with_min_cs_in_dlabel}, only nodes for which $X$ is a witness of redundancy can be deleted from $\Queue_{dup}$ by the function \textsc{pruneQdup} where $X$ is the minimal conflict set given to \textsc{pruneQdup}.
%A node deleted from $\Queue_{dup}$ is either combined replaced or pruned. 
%It is combined replaced by some node in $Comb_{\mathsf{nd},X}(Dup)$ if $Comb_{\mathsf{nd},X}(Dup) \neq \emptyset$ where $X$ is the minimal conflict set w.r.t.\ $DPI'$ given to \textsc{pruneQdup}. 

Thus, assume that $\mathsf{nd}_j$ for some $j \in \setof{1,\dots,k}$ is the first node among $\mathsf{nd}_1,\dots,\mathsf{nd}_k \in \Queue_{dup}$ deleted from $\Queue_{dup}$ by \textsc{pruneQdup} given the minimal conflict set $X$ w.r.t.\ $DPI'$ as an argument. 
%Then, as it holds that
%\begin{itemize}
	%\item $X$ is a witness of redundancy of $\mathsf{nd}_j$,
	%\item $\mathsf{nd}[i_j] = \mathsf{nd}_j[i_j] \,\mbox{ for }\, i_j \in \setof{|\mathsf{nd}_{j-1}|+1,\dots,|\mathsf{nd}_j|}$ where $|\mathsf{nd}_{j-1}| \geq 0$ and $|\mathsf{nd}_j| \leq |\mathsf{nd}|$ and
	%\item there can be no $q \in \setof{1,\dots,|\mathsf{nd}|}$ such that $X \subset \mathsf{nd.cs}[q]$ and $\mathsf{nd}[q] \in \mathsf{nd.cs}[q] \setminus X$,
%\end{itemize}
%we have that there is some $m \in \setof{1,\dots,|\mathsf{nd}_{j-1}|}$ such that $X \subset \mathsf{nd}_j.\mathsf{cs}[m]$ and $\mathsf{nd}_j[m] \in \mathsf{nd}_j.\mathsf{cs}[m] \setminus X$.
Then, as $X$ must be a witness of redundancy of $\mathsf{nd}_j$, we have that there is some $m \in \setof{1,\dots,|\mathsf{nd}_{j}|}$ such that $X \subset \mathsf{nd}_j.\mathsf{cs}[m]$ and $\mathsf{nd}_j[m] \in \mathsf{nd}_j.\mathsf{cs}[m] \setminus X$.

Since Lemma~\ref{lem:necessary_and_sufficient_for_nd_in_Comb(Qdup)} holds also for $j \leq k$ and $\mathsf{nd}_j$ is the first node among $\mathsf{nd}_1,\dots,\mathsf{nd}_k \in \Queue_{dup}$ deleted from $\Queue_{dup}$, we deduce that there is some node $\mathsf{node} \in Comb(\Queue_{dup})$ such that $|\mathsf{node}| = |\mathsf{nd}_{j}|$ and $\mathsf{node}[r] = \mathsf{nd}[r]$ for $r \in \setof{1, \dots, |\mathsf{nd}_{j}|}$ where $|\mathsf{nd}_{j}| \leq |\mathsf{nd}|$. As pointed out before, there cannot be any $q \in \setof{1,\dots,|\mathsf{nd}|}$ such that $X \subset \mathsf{nd.cs}[q]$ and $\mathsf{nd}[q] \in \mathsf{nd.cs}[q] \setminus X$.
This, however, is a contradiction that there is some $m \in \setof{1,\dots,|\mathsf{nd}_{j}|}$ such that $X \subset \mathsf{nd}_j.\mathsf{cs}[m]$ and $\mathsf{nd}_j[m] \in \mathsf{nd}_j.\mathsf{cs}[m] \setminus X$.  

Hence, none of the nodes $\mathsf{nd}_1,\dots,\mathsf{nd}_k \in \Queue_{dup}$ can be deleted throughout the execution of \textsc{dynamicHS} using the current DPI $DPI'$. Consequently, by Lemma~\ref{lem:necessary_and_sufficient_for_nd_in_Comb(Qdup)}, $\mathsf{nd} \in Comb(\Queue_{dup})$ must be preserved.
\end{proof}
The finding of the next lemma is that a node $\mathsf{nd}$ in \textsc{dynamicHS} cannot be processed before all nodes that are set-equal to $\mathsf{nd}$ or proper subsets of $\mathsf{nd}$ have been generated.
\begin{lemma}\label{lem:node_not_processed_before_all_subset_nodes_generated}
Let $GenNodes$ be the set of all nodes generated throughout the execution of all calls to \textsc{dynamicHS} during the execution of Algorithm~\ref{algo:inter_onto_debug}.
Then, a node $\mathsf{nd}$ cannot be processed before each node $\mathsf{nd}'\in GenNodes$ where $\mathsf{nd}' \subseteq \mathsf{nd}$ is generated.
\end{lemma}
\begin{proof}
Let $\mathsf{nd}' \in GenNodes$ such that $\mathsf{nd}' \subseteq \mathsf{nd}$. Assume that $\mathsf{nd}$ is processed, but $\mathsf{nd}'$ has not yet been generated. In order to be processed, $\mathsf{nd}$ must be an element of $\Queue$. By the fact that $\mathsf{nd}' \in GenNodes$, $\mathsf{nd}'$ must be generated at some point in time. In order for $\mathsf{nd}'$ to be generated, some node $\mathsf{nd}''$ with $\mathsf{nd}'' \subset \mathsf{nd}'$ must be an element of $\Queue$. This follows from 
\begin{itemize}
\item the fact that each generated node is a superset of some node in $\Queue$ (cf.\ lines~\ref{algoline:dyn:get_first}, \ref{algoline:dyn:add_ax_to_node} and \ref{algoline:dyn:generate_nodes} and Definition~\ref{def:node_denotations_1}), 
\item the fact that $\Queue$ can only be modified by (a)~deleting from $\Queue$ some node and adding a set of successor nodes of it to $\Queue$ (lines~\ref{algoline:dyn:get_first}, \ref{algoline:dyn:delete_from_queue} and \ref{algoline:dyn:generate_nodes}) or by (b)~deleting from $\Queue$ some node and possibly adding to $\Queue$ a replacement node of it in the function \textsc{prune} and 
\item the fact that for any replacement node $\mathsf{nd}_{rep}$ of $\mathsf{nd}$ it holds that $\mathsf{nd}_{rep} = \mathsf{nd}$.
\end{itemize} 

By Lemma~\ref{lem:superset_lower_prob}, each node which is a proper subset of another node has a higher probability as per $p_{nodes}()$. Since $\mathsf{nd}$ is processed before $\mathsf{nd}'$ is generated and nodes in $\Queue$ are processed in descending order of $p_{nodes}()$ (lines~\ref{algoline:dyn:generate_nodes} and \ref{algoline:dyn:get_first}), $p_{nodes}(\mathsf{nd}) > p_{nodes}(\mathsf{nd}'')$ where $\mathsf{nd}'' \subset \mathsf{nd}' \subseteq \mathsf{nd}$, contradiction.
\end{proof}

The purpose of the following definition is to refer to a node that results from another node $\mathsf{nd}$ by several replacements conducted by \textsc{prune} as a node in a transitive replaces-relation with $\mathsf{nd}$. This will simplify the notation used in the following two lemmata. 
\begin{definition}
Let $\mathsf{nd}_i \sim_{Rep} \mathsf{nd}_j$ iff $\mathsf{nd}_i$ is a replacement node of $\mathsf{nd}_j$ computed so far by \textsc{prune} at any time during the execution of any call to \textsc{dynamicHS} during the execution of Algorithm~\ref{algo:inter_onto_debug}. Further, let the set $Rep := \setof{\tuple{\mathsf{nd}_i,\mathsf{nd}_j}\,|\, \mathsf{nd}_i\sim_{Rep}\mathsf{nd}_j}$. Then we say that \emph{$\mathsf{nd}_1$ is in a transitive replaces-relation with $\mathsf{nd}_k$} iff there is a sequence of nodes $\mathsf{nd}_1,\mathsf{nd}_2, \dots,\mathsf{nd}_{k-1}, \mathsf{nd}_k$ such that $\tuple{\mathsf{nd}_i,\mathsf{nd}_{i+1}} \in Rep$ for all $i \in \setof{1,\dots,k-1}$.
\end{definition}

\subsection[Completeness]{Completeness of \textsc{dynamicHS}}
\label{sec:CompletenessOfTextscDynamicHS}
Lemmata~\ref{lem:successor_node_subset_diag_exists_1} and \ref{lem:successor_node_subset_diag_exists_2} constitute the key results towards proving the completeness of \textsc{dynamicHS} in terms of finding the complete set of minimal diagnoses w.r.t.\ any current DPI $DPI$ in case the execution of \textsc{dynamicHS} with current DPI $DPI$ terminates on account of $\Queue = []$. In other words, if there are no more open nodes in the hitting set tree constructed by \textsc{dynamicHS} with current DPI $DPI$, all minimal diagnoses w.r.t.\ $DPI$ have been labeled by $valid$ and are thus elements of the set $\mD_{calc}$.

The completeness proof (Lemma~\ref{prop:dyn_completeness}) will be a proof by induction where Lemma~\ref{lem:successor_node_subset_diag_exists_1} will serve to derive the base case of the induction, whereas Lemma~~\ref{lem:successor_node_subset_diag_exists_2} will be exploited to establish the induction step.

Lemma~\ref{lem:successor_node_subset_diag_exists_1} assumes an arbitrary fixed ``current'' DPI $DPI$ such that \textsc{dynamicHS} with this ``current'' DPI $DPI$ returns due to $\Queue = []$. Further on, it assumes an arbitrary minimal diagnosis $\md$ w.r.t.\ $DPI$ and a de-facto non-redundant node $\mathsf{nd}$ w.r.t.\ $DPI$ which is a proper subset of $\md$ generated anytime throughout all executions of \textsc{dynamicHS} during the execution of Algorithm~\ref{algo:inter_onto_debug} up to the one with the current DPI $DPI$. 

Given these preconditions, the lemma establishes the existence of a node $\mathsf{nd}_{suc}$ that corresponds to a superset of $\mathsf{nd}$ and to a subset of $\md$, includes one element more than the set $\mathsf{nd}$ and is generated anytime throughout all executions of \textsc{dynamicHS} during the execution of Algorithm~\ref{algo:inter_onto_debug} up to the one with the current DPI $DPI$. Moreover, it states that the node $\mathsf{nd}'_{suc}$ set-equal to this generated node that is an element of $\Queue$ cannot be pruned. However, it might be replaced. In case there is only one potential replacement node of $\mathsf{nd}'_{suc}$ constructable from (the combined nodes of) $\Queue_{dup}$, this replacement node is de-facto non-redundant w.r.t.\ $DPI$. 
%A replacement node of $\mathsf{nd}'_{suc}$ or 
Any node $\mathsf{nd}'_{suc,rep}$ in a transitive replaces relation with $\mathsf{nd}'_{suc}$ cannot be pruned either. It might again be replaced. In case there is only one potential replacement node of $\mathsf{nd}'_{suc,rep}$ constructable from (the combined nodes of) $\Queue_{dup}$, this replacement node is de-facto non-redundant w.r.t.\ $DPI$. 

Figuratively, with respect to the hitting set tree constructed by \textsc{dynamicHS}, this lemma predicates the following: Let the hitting set tree produced by \textsc{dynamicHS} be completely constructed for an arbitrary DPI $DPI$. In case there is any tree branch whose edge labels correspond to a part of the minimal diagnosis $\md$ w.r.t.\ $DPI$ and which is known to be definitely not pruned during this tree construction, 
%until $DPI$ is the current DPI, 
then this branch must be extended by one edge labeled by an element of $\md$ and this extended path is known to be definitely not pruned during this tree construction.
%until $DPI$ is the current DPI.

Notice that during this tree construction, in practice, we will generally never be able to say that a \emph{concrete} branch corresponding to a partial minimal diagnosis will definitely not be pruned. For, this depends on the answers to queries submitted by the interacting user. Nevertheless, for the proof of completeness of \textsc{dynamicHS}, it suffices to just know that there is \emph{any} such branch in the tree. 
%
%In other words, it is enough to have evidence that not all  
%The concrete path, however, is not required to be known. Because it suffices to just know that there is \emph{any} such path. In other words,  
\begin{lemma}\label{lem:successor_node_subset_diag_exists_1}
Assume the execution of \textsc{dynamicHS} with the current DPI $DPI$ and assume that the execution stops due to $\Queue = []$. Let 
\begin{itemize}
\item $GenNodes$ be the set of all nodes generated throughout the execution of all calls to \textsc{dynamicHS} during the execution of Algorithm~\ref{algo:inter_onto_debug},
\item $\md$ be some minimal diagnosis w.r.t.\ $DPI$, 
%used throughout the execution of the current call to \textsc{dynamicHS},
\item $\mathsf{nd} \in GenNodes$ such that $\mathsf{nd}$ is de-facto non-redundant w.r.t.\ $DPI$ and $\mathsf{nd} \subset \md$ and 
%\item \fixme{$\mD_{\checkmark}$ in \textsc{dynamicHS} include the set of all known minimal diagnoses w.r.t.\ the DPI $DPI'$ that is the current DPI of a particular call to \textsc{dynamicHS}.} 
\end{itemize}
Then there are nodes $\mathsf{nd}_{suc}$ and $\mathsf{nd}'_{suc}$ such that the following holds:
\begin{enumerate}[(1)]
\item $\mathsf{nd} \subset \mathsf{nd}_{suc} \subseteq \md$. 
\item $|\mathsf{nd}_{suc}| = |\mathsf{nd}| + 1$. 
\item $\mathsf{nd}_{suc} \in GenNodes$.
\item $\mathsf{nd}'_{suc} = \mathsf{nd}_{suc}$ is an element of $\Queue$ immediately after $\mathsf{nd}_{suc}$ has been generated.
%\item \fixme{change to: If \textsc{prune} is called given a witness of redundancy of $\mathsf{nd}'_{suc}$, then, if only one replacement node of $\mathsf{nd}'_{suc}$ is found, then this replacement node is non-redundant w.r.t.\ $DPI$.} Let $DPI''$ be the current DPI of any call to \textsc{dynamicHS} that is either equal to $DPI$ or includes fewer test cases than $DPI$. Then, if $\mathsf{nd}'_{suc}$ is redundant w.r.t.\ $DPI''$, there is a replacement node $\mathsf{nd}''_{suc}$ of $\mathsf{nd}'_{suc}$ 
%such that $\mathsf{nd}''_{suc}$ is non-redundant w.r.t.\ $DPI''$.
%
\item If \textsc{prune} is called given a witness of redundancy of $\mathsf{nd}'_{suc}$, then some replacement node of $\mathsf{nd}'_{suc}$ is found. If only one replacement node of $\mathsf{nd}'_{suc}$ is found, then this replacement node is de-facto non-redundant w.r.t.\ $DPI$. 
\item Let $\mathsf{nd}'_{suc,rep}$ be in a transitive replaces-relation with $\mathsf{nd}'_{suc}$. If \textsc{prune} is called given a witness of redundancy of $\mathsf{nd}'_{suc,rep}$, then some replacement node of $\mathsf{nd}'_{suc,rep}$ is found. If only one replacement node of $\mathsf{nd}'_{suc,rep}$ is found, then this replacement node is de-facto non-redundant w.r.t.\ $DPI$.
%
%if $\mathsf{nd}'_{suc}$ is redundant w.r.t.\ some DPI $DPI''$ that is the current DPI of any call to \textsc{dynamicHS} and is either equal to $DPI$ or includes fewer test cases than $DPI$, then 
%$\mathsf{nd}\in\Queue_{dup}$ and 
%
%constructed from $\mathsf{nd}$ which is non-redundant w.r.t.\ $DPI''$. 
\end{enumerate}
\end{lemma}
\begin{proof}
%Assume no such node $\mathsf{nd}_{suc}$ exists. 
%
Now, since $\mathsf{nd} \in GenNodes$, we know that $\mathsf{nd}$ must be generated at some point in time during the execution of any call to \textsc{dynamicHS} during the execution of Algorithm~\ref{algo:inter_onto_debug}. As the execution $Ex_{curr}$ of the call to \textsc{dynamicHS} using $DPI$ is assumed to terminate due to $\Queue=[]$ and no more nodes can be generated after $\Queue = []$ (each generated node is constructed by extending a node in $\Queue$), $\mathsf{nd}$ must be generated the latest during $Ex_{curr}$. 

So, let us consider exactly the point in time when $\mathsf{nd}$ is generated. Since this point in time might not arise during the execution $Ex_{curr}$ of \textsc{dynamicHS}, but during some execution $Ex_{prev}$ taking place before $Ex_{curr}$ which uses some ``current'' DPI which includes fewer test cases than the current DPI $DPI$ of $Ex_{curr}$, we call the ``current'' DPI in $Ex_{prev}$ in the following $DPI_{prev}$. That is, $DPI_{prev}$ might be equal to $DPI$ or comprise a subset of the test cases $DPI$ includes.

First, we observe that \emph{immediately} after $\mathsf{nd}$ has been generated, there is some node $\mathsf{nd}'\in\Queue$ such that $\mathsf{nd}'=\mathsf{nd}$. If $\mathsf{nd}'$ is not the same node as $\mathsf{nd}$, then $\mathsf{nd}\in\Queue_{dup}$. This follows from lines~\ref{algoline:dyn:check_node_already_in_Q}-\ref{algoline:dyn:generate_nodes}.

Second, we have that $\mathsf{nd}'\in\Queue$ cannot be pruned before it is processed. 
In case $\mathsf{nd}'$ is the same node as $\mathsf{nd}$, this follows from Proposition~\ref{prop:de-facto_non_redundant_node_cannot_be_pruned_or_replaced} %from Lemmata~\ref{lem:redundant_node_remains_redundant_after_adding_new_testcases} and \ref{lem:prune}, Corollary~\ref{cor:prune_only_called_with_min_cs_in_dlabel} 
and the precondition that $\mathsf{nd}$ is de-facto non-redundant w.r.t.\ $DPI$. Notice that in this case $\mathsf{nd}\in\Queue$ cannot even be replaced (also by Proposition~\ref{prop:de-facto_non_redundant_node_cannot_be_pruned_or_replaced}).
%as there cannot be any witness of redundancy of $\mathsf{nd}$ w.r.t.\ $DPI_{prev}$.

Otherwise, if $\mathsf{nd}'$ is not the same node as $\mathsf{nd}$, we argue as follows:
Assume that $\mathsf{nd}'$ is redundant w.r.t.\ $DPI_{prev}$ and that the \textsc{prune} function is called with arguments $\Queue$, $\Queue_{dup}$ and some minimal conflict set $X$ w.r.t.\ $DPI_{prev}$ which is a witness of redundancy of $\mathsf{nd}'$. Then, since $\mathsf{nd}$ is de-facto non-redundant w.r.t.\ $DPI$, since $DPI_{prev}$ includes a subset of the test cases $DPI$ comprises and by Proposition~\ref{prop:de-facto_non_redundant_node_cannot_be_pruned_or_replaced}, $\mathsf{nd}$ cannot have been deleted from $\Queue_{dup}$ during any pruning step. Thence, by Lemma~\ref{lem:prune}, $\mathsf{nd}$ (or some other node set-equal to $\mathsf{nd}'$ for which $X$ is not a witness of redundancy) must be constructed and added to $\Queue$ in lines~\ref{algoline:prune:check_for_alternative_paths_start}-\ref{algoline:prune:check_for_alternative_paths_end} during the execution of the \textsc{prune} function.
% after $\mathsf{nd}'$ has been deleted in line~\ref{algoline:prune:delete_from_S}.
%and thus cannot be redundant w.r.t.\ $DPI_{prev}$ or any other DPI including fewer test cases than $DPI$ by Lemma~\ref{lem:redundant_node_remains_redundant_after_adding_new_testcases}, no witness of redundancy of $\mathsf{nd}$ can have been found until this call to \textsc{prune} (which takes place during the execution of \textsc{dynamicHS} using $DPI_{prev}$). 
%
%Therefore, by Lemma~\ref{lem:pruneQdup}, $\mathsf{nd}$ cannot have been deleted from $\Queue_{dup}$ during \textsc{pruneQdup} which is always called before \textsc{prune} is called (lines~\ref{algoline:dlabel:call_prune_Qdup}-\ref{algoline:dlabel:call_prune_Dsupset} and lines~\ref{algoline:update:call_prune_Qdup}-\ref{algoline:update:call_prune_Dsupset}). Thence, by Lemma~\ref{lem:prune}, $\mathsf{nd}$ (or some other node set-equal to $\mathsf{nd}'$) must be constructed and added to $\Queue$ in lines~\ref{algoline:prune:check_for_alternative_paths_start}-\ref{algoline:prune:check_for_alternative_paths_end} during the execution of the \textsc{prune} function after $\mathsf{nd}'$ has been deleted in line~\ref{algoline:prune:delete_from_S}.
%
%Because the call to \textsc{dynamicHS} using $DPI$ is assumed to terminate due to $\Queue=\emptyset$, each node must be processed the latest during the execution of this call to \textsc{dynamicHS}. This holds as the only alternative way to achieve the deletion of a node from $\Queue$ (line~\ref{algoline:dyn:delete_from_queue}) is to process it.

That is, before any node set-equal to $\mathsf{nd}$ is processed, any number of calls to \textsc{prune} with arguments $\Queue$, $\Queue_{dup}$ and some minimal conflict set $X$ w.r.t.\ \emph{any} DPI $DPI_{prev}$ imply that $\Queue$ includes some node that is set-equal to $\mathsf{nd}$. Let us denote by $\mathsf{node}$ the node set-equal to $\mathsf{nd}$ that is finally processed. 

There must be some execution of \textsc{dynamicHS} with some DPI (which might be equal to $DPI$ or include a subset of the test cases in $DPI$) during which $\mathsf{node}$ is processed. This holds as the execution of \textsc{dynamicHS} with $DPI$ is assumed to stop because of $\Queue = []$, since not all nodes set-equal to $\mathsf{node}$ can be pruned, as just argued before, and because the only alternative way, except for pruning, to achieve the deletion of a node from $\Queue$ (line~\ref{algoline:dyn:delete_from_queue}) is to process it. 
Let $DPI_{prev}$ now be the ``current'' DPI of the execution of \textsc{dynamicHS} during which $\mathsf{node}$ is processed. Further, we denote the DPI considered by the immediate subsequent execution of \textsc{dynamicHS} by $DPI_{prev+1}$, and so on.

When $\mathsf{node}$ is processed, it is either 
\begin{itemize}
	\item (a)~labeled by a set (\textsc{dLabel} returns in line~\ref{algoline:dlabel:return_X}, \ref{algoline:dlabel:return_new_cs} or \ref{algoline:dlabel:return_C}) or
	\item (b)~not labeled by a set (\textsc{dLabel} returns in line~\ref{algoline:dlabel:non-min_crit_end} or \ref{algoline:dlabel:return_valid}).
\end{itemize}
  
%\textbf{Case~(b):} 
\paragraph{Case~(b):}
In this case, \textsc{dLabel} returns either 

\begin{itemize}
\item (i)~$nonmin$ or 
\item (ii)~$valid$.
\end{itemize}
%By Lemma~\ref{lem:if_dlabel_returns_valid_nonmin_then}, this is a contradiction to the assumption that $\md$ is a minimal diagnosis w.r.t.\ the current DPI as both outputs $nonmin$ and $valid$ imply the opposite.
%\textbf{In (i)}, 
\paragraph{Case~(i):}
By Lemma~\ref{lem:if_dlabel_returns_valid_nonmin_then}, $\mathsf{node}$ must be a non-minimal diagnosis w.r.t.\ $DPI_{prev}$. By line~\ref{algoline:dyn:add_to_Dsupset}, $\mathsf{node}$ is then added to the set $\mD_{\supset}$. $\mD_{\supset}$ is never modified throughout Algorithm~\ref{algo:inter_onto_debug} and is given as an input argument to each subsequent call to \textsc{dynamicHS} by line~\ref{algoline:inter_onto_debug:dynamicHS} in Algorithm~\ref{algo:inter_onto_debug}. During the execution of some subsequent call to \textsc{dynamicHS} using the DPI $DPI_{prev+i}$ for $i \geq 1$, the set $\mD_{\supset}$ might be modified by the \textsc{updateTree} function (line~\ref{algoline:update:call_prune_Dsupset} and lines~\ref{algoline:update:process_Dsupset_start}-\ref{algoline:update:process_Dsupset_end}) or in the \textsc{dLabel} function (line~\ref{algoline:dlabel:call_prune_Dsupset}) called for $DPI_{prev+i}$. Because $\mathsf{node} = \mathsf{nd}$ and $\mathsf{nd}$ is de-facto non-redundant w.r.t.\ $DPI$, we infer by the same argumentation as used above that $\mathsf{node} \in \mD_{\supset}$ cannot be pruned, i.e.\ $\mathsf{node}$ considered as a set cannot be deleted from $\mD_{\supset}$ in line~\ref{algoline:update:call_prune_Dsupset} or line~\ref{algoline:dlabel:call_prune_Dsupset}. The truth of this is supported by Corollary~\ref{cor:prune_only_called_with_min_cs_in_dlabel} and Lemmata~\ref{lem:quick_prune_check} and \ref{lem:complete_prune_check} which say that \textsc{prune} can only be called given some minimal conflict set $X$ w.r.t.\ $DPI_{prev+i}$. 
So, after any number of calls to \textsc{prune}, we have that either $\mathsf{node} \in \mD_{\supset}$ or, otherwise, there is some node in $\mD_{\supset}$ which is set-equal to $\mathsf{node}$ and which is in a transitive replaces-relation with $\mathsf{node}$. We keep calling this (possibly replacement) node $\mathsf{node}$ in the following.
%So, $\mathsf{node}\in \mD_{\supset}$ might just be replaced by some node which is in a transitive replaces-relation with $\mathsf{node}$.
%It might just be replaced by some replacement node (and this replacement node might again be replaced by another replacement node, and so on). 

%%%%%%%%%%%%%%%%%%%% OLD start %%%%%%%%%%%%%%%%%%%%
%By the precondition that $\md$ is a minimal diagnosis w.r.t.\ $DPI$ and $\mathsf{node} \subset \md$ (since $\mathsf{nd} \subset \md$ and $\mathsf{node} = \mathsf{nd}$), $\mathsf{node}$ cannot be a diagnosis w.r.t.\ $DPI$. Hence, there must be some DPIs $DPI_{prev+k}$ and $DPI_{prev+k+1}$ such that $\mathsf{node}$ is a non-minimal diagnosis w.r.t.\ $DPI_{prev+k}$ and $\mathsf{node}$ is not a non-minimal diagnosis w.r.t.\ $DPI_{prev+k+1}$. That is, there is no minimal diagnosis $\md'$ w.r.t.\ $DPI_{prev+k+1}$ such that $\md' \subset \mathsf{node}$. Therefore, by the fact that $\mD_{\checkmark}$ is assumed to store all known minimal diagnoses w.r.t.\ $DPI_{prev+k+1}$ during the execution of \textsc{dynamicHS} with the DPI $DPI_{prev+k+1}$, $\mathsf{node}$ will be reinserted into $\Queue$ in lines~\ref{algoline:update:process_Dsupset_start}-\ref{algoline:update:process_Dsupset_end} throughout the execution of \textsc{updateTree} for $DPI_{prev+k+1}$. 
%%Hence, at some execution of \textsc{dynamicHS}, case~(a) must emerge for $\mathsf{node}$.
%%%%%%%%%%%%%%%%% OLD end %%%%%%%%%%%%%%%%%%%%%%%%%%%%
%%%%%%%%%%%%%%%%%%%% NEW start %%%%%%%%%%%%%%%%%%%%
By Lemma~\ref{lem:if_dlabel_returns_valid_nonmin_then}, at the time $\mathsf{node}$ was processed, there must be some diagnosis $\md'$ w.r.t.\ $DPI_{prev}$ such that $\md'\in \mD_{calc}$ and $\mathsf{node} \supset \md'$. 
Additionally, by Lemma~\ref{lem:if_dlabel_returns_valid_nonmin_then}, the set $\mD_{calc}$ computed during \textsc{dynamicHS} for some ``current'' DPI $DPI_j$ comprises only diagnoses w.r.t.\ $DPI_j$. 
Now, we have $\mathsf{node} \subset \md$ since $\mathsf{nd} \subset \md$ and $\mathsf{node} = \mathsf{nd}$, and $\md'\subset\mathsf{node}$. That is, $\md'\subset\md$. By the precondition that $\md$ is a minimal diagnosis w.r.t.\ $DPI$, $\md'$ cannot be a diagnosis w.r.t.\ $DPI$. Thus, there cannot be any such $\md'$ in $\mD_{calc}$ computed during \textsc{dynamicHS} for $DPI$.

All nodes in $\mD_{calc}$ returned by some call to \textsc{dynamicHS} using DPI $DPI_1$ that are no diagnoses w.r.t.\ $DPI_2$, the extension of $DPI_1$ by a new query added as a positive or negative test case, are added to the set $\mD_{\times}$ (and not to $\mD_{\checkmark}$) in line~\ref{algoline:inter_onto_debug:update_D_times} of Algorithm~\ref{algo:inter_onto_debug} and are thus no elements of the set $\mD_{\checkmark}$ given as an argument to \textsc{dynamicHS} at the next call to \textsc{dynamicHS}. 
The elements of $\mD_{\checkmark}$ given as an argument to \textsc{dynamicHS} at the next call to \textsc{dynamicHS} using $DPI_2$ are definitely added to $\Queue$ again in lines~\ref{algoline:update:process_Dcheckmark_start}-\ref{algoline:update:process_Dcheckmark_end} as $\mD_{\checkmark}$ is not modified elsewhere in \textsc{dynamicHS} before lines~\ref{algoline:update:process_Dcheckmark_start}-\ref{algoline:update:process_Dcheckmark_end} are reached. 

Therefore, we need to differentiate between two cases: Either 
\begin{itemize}
	\item (x1)~$\md'\in\mD_{\times}$ never holds for the input argument $\mD_{\times}$ to any call to \textsc{dynamicHS} or
	\item (x2)~$\md'\in\mD_{\times}$ holds at least once for the input argument $\mD_{\times}$ to some call to \textsc{dynamicHS}.
\end{itemize}

%\textbf{Case~(x1):} 
\paragraph{Case~(x1):}
Since $\md'\in\mD_{calc}$ holds after the execution of \textsc{dynamicHS} using $DPI_{prev}$ stops, we have that $\md'\in\mD_{\checkmark}$ must hold for the argument $\mD_{\checkmark}$ given to \textsc{dynamicHS} using $DPI_{prev+1}$. After \textsc{updateTree} returns during \textsc{dynamicHS} using $DPI_{prev+1}$, $\md' \in \Queue$ holds as argued. Subsequently, $\md'$ might be added again to $\mD_{calc}$ and then to $\mD_{\checkmark}$ again in line~\ref{algoline:inter_onto_debug:update_D_checkmark} of Algorithm~\ref{algo:inter_onto_debug} and to $\Queue$ again in line~\ref{algoline:update:insert_sorted_1} during \textsc{dynamicHS} using $DPI_{prev+2}$, and so forth. But, when a test case is added to some DPI $DPI_{prev+i}$ in Algorithm~\ref{algo:inter_onto_debug} that invalidates the diagnosis $\md'$ (yielding the DPI $DPI_{prev+i+1}$), $\md' \notin \mD_{calc}$ is assumed to hold (otherwise it would be an element of $\mD_{\times}$ against our assumption). Such a test case must be added sometime as argued above. By Proposition~\ref{prop:diag_for_new_dpi_is_diag_for_old_dpi}, $\md'$ cannot be a (minimal) diagnosis w.r.t.\ any DPI including a superset of the test cases in $DPI_{prev+i+1}$ either.
Notice that the case $\md' \notin \mD_{calc}$ can emerge in spite of the fact that $\md'$ is a minimal diagnosis w.r.t.\ $DPI_{prev+i}$ because there may be minimal diagnoses w.r.t.\ $DPI_{prev+i}$ that have a higher probability as per $p_{nodes}()$ than $\md'$. For $DPI_{prev+i+1}$ and all DPIs including more test cases than $DPI_{prev+i+1}$, $\md'$ cannot be added to $\mD_{calc}$ anymore due to Lemma~\ref{lem:if_dlabel_returns_valid_nonmin_then} since only \emph{diagnoses} w.r.t.\ the currently used DPI can be added to $\mD_{calc}$.

%\textbf{Case~(x2):} 
\paragraph{Case~(x2):}
Here, $\md'\in\mD_{\times}$ holds at least once for the input argument $\mD_{\times}$ to some call to \textsc{dynamicHS} using the DPI $DPI_{prev+i}$. Then, \textsc{dynamicHS} using the DPI $DPI_{prev+i-1}$ must have returned a set $\mD_{calc}$ including $\md'$ as otherwise $\md'$ cannot be added to $\mD_{\times}$. Hence, $\md'$ must be a diagnosis w.r.t.\ $DPI_{prev+i-1}$ by Lemma~\ref{lem:if_dlabel_returns_valid_nonmin_then}. Since $\md'$ is added to $\mD_{\times}$, it cannot be a diagnosis w.r.t.\ $DPI_{prev+i}$. This must hold 
\begin{itemize}
	\item by Remark~\ref{rem:invalidated_sets_of_q-partition_for_query_answer},
	\item since the set added to $\mD_{\times}$ in Algorithm~\ref{algo:inter_onto_debug} is exactly the set $\mD_{out}$ returned by \textsc{getInvalidDiags} in line~\ref{algoline:inter_onto_debug:get_invalid_diags} of Algorithm~\ref{algo:inter_onto_debug} and
	\item $\mD_{out} = \dx{}(Q)$ in case the user answer $u(Q)$ to the query $Q$ w.r.t.\ $\mD_{calc}$ and $DPI_{prev+i-1}$ is $\false$ and $\mD_{out}=\dnx{}(Q)$ otherwise (notice that $\mD_{calc}$ is called $\mD_{\checkmark}$ in Algorithm~\ref{algo:inter_onto_debug}).
\end{itemize}
So, by Proposition~\ref{prop:diag_for_new_dpi_is_diag_for_old_dpi}, $\md'$ cannot be a (minimal) diagnosis w.r.t.\ any DPI including more test cases than $DPI_{prev+i}$ either.
 
Each element in $\mD_{\times}$ is processed by the \textsc{updateTree} function (lines~\ref{algoline:update:process_Dtimes_start}-\ref{algoline:update:process_Dtimes_end}) called for the DPI $DPI_{prev+i}$. In lines~\ref{algoline:update:process_Dtimes_start}-\ref{algoline:update:process_Dtimes_end}, each node $\mathsf{ndx}$ in $\mD_{\times}$ can only be pruned or either $\mathsf{ndx}$ or a node in a transitive replaces-relation with $\mathsf{ndx}$ is added to $\Queue$ in line~\ref{algoline:update:insert_sorted_0}.
%Then, either $\md'$ is redundant w.r.t.\ $DPI_{prev+i}$ and \textsc{prune} in line~\ref{algoline:update:call_prune_Dtimes} is called given some witness of redundancy $X$ of $\md'$ or the opposite holds. 
%
%In the former case, 
%%by Lemma~\ref{lem:prune} and Corollary~\ref{cor:prune_only_called_with_min_cs_in_dlabel}, 
%$\md'$ is either pruned or replaced. If it is pruned, it is deleted from $\mD_{\times}$. If it is replaced, then $\mD_{\times}$ might include a node in a transitive replaces-relation with $\md'$ when line~\ref{algoline:update:reinsert_D_of_Dx_to_Q} is reached or not (i.e.\ the node in a transitive replaces-relation with $\md'$ has been pruned). If there is a node in a transitive replaces-relation with $\md'$ when line~\ref{algoline:update:reinsert_D_of_Dx_to_Q} is reached, then this node is added to $\Queue$ in line~\ref{algoline:update:insert_sorted_0}. Otherwise, no node
%
%In the latter case, $\md'$ is reinserted into $\Queue$ in line~\ref{algoline:update:insert_sorted_0}.
%Thus, in all cases $\md' \notin \mD_{\checkmark}$ and $\md' \notin \mD_{calc}$ holds. 
$\mD_{calc}$ is not modified by \textsc{updateTree} and $\mD_{calc} = \emptyset$ holds at the beginning of the execution of each call to \textsc{dynamicHS}. (A node set-equal to) $\md'$ cannot ever be readded to $\mD_{calc}$ by Lemma~\ref{lem:if_dlabel_returns_valid_nonmin_then} and since $\md'$ is not a diagnosis w.r.t\ any DPI including more test cases than $DPI_{prev+i}$. Hence, $\md'\in\mD_{calc}$ can never hold for any DPI including more test cases than $DPI_{prev+i}$.  

Hence, there must be some DPI $DPI_{prev+k}$ such that $\mD_{\checkmark}$ given as input to the \textsc{dynamicHS}-call for $DPI_{prev+k}$ does not include \emph{any} diagnosis $\md'\subset \mathsf{node}$.
So, during the execution of the call to \textsc{dynamicHS} using DPI $DPI_{prev+k}$, $\mathsf{node}$ must be deleted from $\mD_{\supset}$ and be reinserted into $\Queue$ by lines~\ref{algoline:update:process_Dsupset_start}-\ref{algoline:update:process_Dsupset_end} in \textsc{updateTree} which is called at the beginning of the execution of \textsc{dynamicHS} at any call to \textsc{dynamicHS}. This must hold since all nodes $\mathsf{ndx}$ in $\mD_{\supset}$ that have not yet been pruned and for which there is no diagnosis in $\mD_{\checkmark}$ which is a proper subset of $\mathsf{ndx}$, are added to $\Queue$ throughout lines~\ref{algoline:update:process_Dsupset_start}-\ref{algoline:update:process_Dsupset_end}. As shown, both criteria are met for $\mathsf{node}$ during the execution of the call to \textsc{dynamicHS} using DPI $DPI_{prev+k}$.
\paragraph{Case~(ii):}
By Lemma~\ref{lem:if_dlabel_returns_valid_nonmin_then}, we know that $\mathsf{node}$ is a diagnosis w.r.t.\ $DPI_{prev}$ and that $\mathsf{node}$ is added to $\mD_{calc}$. Since $\mathsf{node} \subset \md$ and $\md$ is a minimal diagnosis w.r.t.\ $DPI$, we obtain, by the same argumentation as in (i), 
that there must be some DPI $DPI_{prev+k}$ such that $\mD_{\checkmark}$ given as input to the \textsc{dynamicHS}-call for $DPI_{prev+k}$ does not include $\mathsf{node}$.

If $\mathsf{node} \notin \mD_{\times}$, then it cannot ever be added to $\mD_{calc}$ again, as argued in case (i). Otherwise, during the execution of \textsc{updateTree} which is called at the beginning of the execution of each call to \textsc{dynamicHS}, $\mD_{\times}$ is modified in lines~\ref{algoline:update:process_Dtimes_start}-\ref{algoline:update:process_Dtimes_end}.

Now, we differentiate between two cases, namely $\mathsf{node}$ is either 
\begin{itemize}
	\item ($\lnot$r)~non-redundant w.r.t.\ $DPI$ or
	\item (r)~redundant w.r.t.\ $DPI$.
\end{itemize}

%\textbf{Case~($\lnot$r):} 
\paragraph{Case~($\lnot$r):}
Due to the non-redundancy of $\mathsf{node}$ w.r.t.\ $DPI$, Lemma~\ref{lem:redundant_node_remains_redundant_after_adding_new_testcases}, Lemma~\ref{lem:prune} and Corollary~\ref{cor:prune_only_called_with_min_cs_in_dlabel}, $\mathsf{node}$ cannot be replaced or pruned throughout lines~\ref{algoline:update:process_Dtimes_start}-\ref{algoline:update:add_set_del_supset}. Thus, $\mathsf{node}$ is reinserted into $\Queue$ in line~\ref{algoline:update:insert_sorted_0}.

%\textbf{Case~(r):} 
\paragraph{Case~(r):}
Since $\mathsf{node}$ is redundant w.r.t.\ $DPI$, it may or may not be redundant w.r.t.\ $DPI_{prev+k+1}$. So, during the \textsc{updateTree} function called in \textsc{dynamicHS} for $DPI_{prev+k+1}$, there may or may not be some call to \textsc{prune} given some $X$ as argument which is a witness of redundancy of $\mathsf{node}$. In the latter case, $\mathsf{node}$ will not be replaced or pruned during any \textsc{prune} execution and will be reinserted into $\Queue$ in line~\ref{algoline:update:insert_sorted_0}. In the former case, $\mathsf{node}$ might be replaced, but it cannot be pruned due to the same reasoning as given above in case (i). 
So, either $\mathsf{node}$ or some node in a transitive replaces-relation with $\mathsf{node}$ must be in $\mD_{\times}$ at the time line~\ref{algoline:update:reinsert_D_of_Dx_to_Q} is reached. This node is then added to $\Queue$ in line~\ref{algoline:update:insert_sorted_0}. 

Now, both cases (i) and (ii) identified for case (b) lead to the reinsertion of $\mathsf{node}$ or some node set-equal to $\mathsf{node}$ into $\Queue$. Notice that this node has the same properties as $\mathsf{node}$ before one of the cases (i) or (ii) emerged. That is, 
%it is either non-redundant w.r.t.\ $DPI$ or, 
if \textsc{prune} is called given a witness of redundancy of $\mathsf{node}$, then a replacement node of $\mathsf{node}$ is found. And, if only one replacement node of $\mathsf{node}$ is found, this replacement node is de-facto non-redundant w.r.t.\ $DPI$. 

If $\mathsf{node}$ is the same node as $\mathsf{nd}$, this holds since there cannot be a witness of redundancy of $\mathsf{nd}$ due to the de-facto non-redundancy of $\mathsf{nd}$ w.r.t.\ $DPI$ and Proposition~\ref{prop:de-facto_non_redundant_node_cannot_be_pruned_or_replaced}. Otherwise, this holds by Lemma~\ref{lem:prune} and since $\mathsf{node} = \mathsf{nd}$ and $\mathsf{nd} \in \Queue_{dup}$ must hold due the de-facto non-redundancy of $\mathsf{nd}$ w.r.t.\ $DPI$ and Proposition~\ref{prop:de-facto_non_redundant_node_cannot_be_pruned_or_replaced}. So, we call this reinserted node again $\mathsf{node}$.

Furthermore, $\mathsf{node}$ can be neither labeled by $valid$ nor by $nonmin$ during the execution of \textsc{dynamicHS} for $DPI$. This holds by Lemma~\ref{lem:if_dlabel_returns_valid_nonmin_then} and since $\mathsf{node}$ can be neither a diagnosis nor a non-minimal diagnosis w.r.t.\ $DPI$ due to $\mathsf{node} \subset \md$ and the fact that $\md$ is a minimal diagnosis w.r.t.\ $DPI$. As a consequence of this and the assumption that the \textsc{dynamicHS}-call for $DPI$ terminates due to $\Queue = []$, case~(a) must arise at some point in time for $\mathsf{node}$ during some execution of \textsc{dynamicHS} for some (previous) DPI not-necessarily equal to $DPI$.
%%%%%%%%%%%%%%%%%%%%%%%%%%%%%% NEW end %%%%%%%%%%%%%%%%%%%%%%%%%%

%\textbf{Case~(a):} 
\paragraph{Case~(a):}
In this case, by Lemma~\ref{lem:Ccalc_in_dlabel}, \textsc{dLabel} returns a minimal conflict set $L$ w.r.t.\ $DPI_{prev}$ as a label for $\mathsf{node}$ where $L$ has the property that $L \cap \mathsf{node} = \emptyset$.

It must hold that $L \neq \emptyset$. Otherwise, by Proposition~\ref{prop:cs_admissible}, either 
\begin{itemize}
	\item (v1)~$\mo$ is valid w.r.t.\ $\tuple{\cdot,\mb,\Tp_{prev},\Tn_{prev}}_\RQ$ where $DPI_{prev} = \tuple{\mo,\mb,\Tp_{prev},\Tn_{prev}}_\RQ$ or
	\item (v2)~$DPI_{prev}$ is non-admissible.
\end{itemize}

In the former \textbf{case (v1)}, we know by Corollary~\ref{cor:notions_equiv} that the only (minimal) diagnosis w.r.t.\ $DPI_{prev}$ is $\emptyset$. If $DPI_{prev}$ is equal to $DPI$, this is a contradiction to the existence of some minimal diagnosis w.r.t.\ $DPI$, namely $\md$, which is not the empty set. $\md \supset \emptyset$ must hold since, by precondition, there is a node $\mathsf{nd}$ such that $\mathsf{nd} \subset \md$ and since $\emptyset \subseteq \mathsf{nd}$. 

Otherwise, if $DPI_{prev}$ includes a proper subset of the test cases $DPI$ includes, $DPI$ can never be a current DPI during any execution of \textsc{dynamicHS} during the same execution of Algorithm~\ref{algo:inter_onto_debug} during which there is an execution of \textsc{dynamicHS} using $DPI_{prev}$ as a current DPI. This holds as there must be at least two diagnoses in $\mD_{\checkmark}$ (which is the set $\mD_{calc}$ returned by \textsc{dynamicHS} for $DPI_{prev}$) in line~\ref{algoline:inter_onto_debug:stop_crit} of Algorithm~\ref{algo:inter_onto_debug} in order for \textsc{dynamicHS} to be called again with an extended DPI. For, in case there is only one diagnosis, i.e.\ $\emptyset$, then the probability of this diagnosis is 1 which is greater or equal $1 - \sigma$ for any choice of $\sigma$ due to $\sigma \geq 0$. Consequently, Algorithm~\ref{algo:inter_onto_debug} would return in line~\ref{algoline:inter_onto_debug:return}. This is a contradiction to the assumption that there is an execution of \textsc{dynamicHS} using $DPI$ as a current DPI. 

In the latter \textbf{case (v2)}, we can infer by Corollary~\ref{cor:query_leaves_valid_diag}, which states that adding queries as test cases to an admissible DPI can never yield a non-admissible DPI, that the DPI given as an input to Algorithm~\ref{algo:inter_onto_debug} must be non-admissible, contradiction. 
%
%Thence, $L \neq \emptyset$. 
%if \textsc{dLabel} returns in line~\ref{algoline:dlabel:add_new_cs}, by analogue argumentation as used in the proof of Lemma~\ref{lem:Ccalc_in_dlabel}, we can derive that \textsc{dLabel} returns a minimal conflict set w.r.t.\ the current DPI. If, on the other hand, \textsc{dLabel} returns in line~\ref{algoline:dlabel:return_C}, by Lemma~\ref{lem:Ccalc_in_dlabel} which states that $\mC_{calc}$ contains only minimal conflict sets w.r.t.\ the current or some previous DPI (including fewer test cases than the current one)
%

Thence, $L \neq \emptyset$ and \textsc{dynamicHS} will execute lines~\ref{algoline:dyn:for_e_in_L}-\ref{algoline:dyn:generate_nodes} and generate one node $\mathsf{node}_{e} :=$ $\textsc{add}(\mathsf{node}$, $e)$ with $\mathsf{node}_{e}.\mathsf{cs} := \textsc{add}(\mathsf{node.cs},L)$ for each $e\in L$ (cf.\ Definition~\ref{def:alternative_equal_node} for an explanation of the function \textsc{add}). 

Now, we have that there must be some non-empty active sublabel of $L = \mathsf{node}_{e}.\mathsf{cs}[r]$ w.r.t.\ $DPI$ where $r := |\mathsf{node}_{e}|$ by Definition~\ref{def:active_sublabel}. This holds by the following argumentation: 

The first observation is that $\mathsf{node}_{e}.\mathsf{cs}[r]$ cannot be reduced twice during one and the same execution of \textsc{dynamicHS} using one and the same DPI $DPI_{prev+j}$ which results from $DPI_{prev}$ by addition of test cases. For, by Corollaries~\ref{cor:prune_only_called_with_min_cs_in_dlabel} and \ref{cor:pruneQdup_only_called_with_min_cs_in_dlabel} and Lemmata~\ref{lem:quick_prune_check} and \ref{lem:complete_prune_check}, \textsc{prune} as well as \textsc{pruneQdup} can only be called given some minimal conflict set $X$ w.r.t.\ $DPI_{prev+j}$. By Lemmata~\ref{lem:prune} and \ref{lem:pruneQdup}, all nodes $\mathsf{ndx}$ that are in the set returned by \textsc{prune} and \textsc{pruneQdup}, respectively, have the property that there are no proper supersets of $X$ in $\mathsf{ndx.cs}$. Moreover, there are no proper subsets of $X$ in $\mathsf{ndx.cs}$. Because each $\mathsf{ndx.cs}[m]$ for $m \in \setof{1,\dots,|\mathsf{ndx.cs}|}$ must be a minimal conflict set w.r.t.\ some DPI equal to $DPI_{prev+j}$ or including a subset of the test cases in $DPI_{prev+j}$. Otherwise, $\mathsf{ndx}$ could not be a node during the execution of \textsc{dynamicHS} where $DPI_{prev+j}$ is the current DPI. By Proposition~\ref{prop:changes_in_conflict_sets_after_testcase_added}, there cannot be any $m \in \setof{1,\dots,|\mathsf{ndx.cs}|}$ such that $\mathsf{ndx.cs}[m] \subset X$ as $X$ is a minimal conflict set w.r.t.\ $DPI_{prev+j}$.
As two minimal conflict sets w.r.t.\ $DPI_{prev+j}$ can never be in a proper subset-relationship with one another, $L = \mathsf{node}_{e}.\mathsf{cs}[r]$ can be modified at most once by \textsc{prune} or \textsc{pruneQdup} for the DPI $DPI_{prev+j}$.

Second, by Proposition~\ref{prop:changes_in_conflict_sets_after_testcase_added}, each minimal conflict set w.r.t.\ $DPI_{prev}$ is a conflict set w.r.t.\ any DPI $DPI_{prev+j}$ that results from $DPI_{prev}$ by addition of test cases, that is, in particular, w.r.t.\ $DPI$. So, there must be some minimal conflict set $\mc_j$ w.r.t.\ each $DPI_{prev+j}$ such that $\mc_j \subseteq L$ and there cannot be any minimal conflict set w.r.t.\ $DPI_{prev+j}$ that is a proper superset of $L$. 

Third, we have that $L \neq \emptyset$, $L$ is a minimal conflict set w.r.t.\ $DPI_{prev}$, and $DPI_{prev+j}$ includes a superset of the test cases in $DPI_{prev}$. Thus, by Proposition~\ref{prop:if_DPI_cs_neq_emptyset_then_DPI+1_cs_neq_emptyset}, each minimal conflict set w.r.t.\ $DPI_{prev+j}$ must be non-empty. In particular, this implies that all minimal conflict sets w.r.t.\ $DPI$ that are subsets of $L$ must be non-empty.

By these three observations, the criteria 
%for an active sublabel w.r.t.\ $DPI$ given by 
of Definition~\ref{def:active_sublabel} can be applied to analyze the active subnode of $\mathsf{node}_{e}.\mathsf{cs}[r]$  w.r.t.\ $DPI$. 
That is, if $\mc_1,\dots,\mc_n$ is the (arbitrary actual) chronological sequence of all sets $X$ given as an argument to \textsc{prune} and \textsc{pruneQdup} during all executions of \textsc{dynamicHS} from the one with current DPI $DPI_{prev}$ up to and including the one with current DPI $DPI$ where 
\begin{itemize}
		\item $\mathsf{node}_{e}.\mathsf{cs}[r] \supset \mc_1$,
		\item each $\mc_i$ is a minimal conflict set w.r.t.\ $DPI_i$ for $i \in \setof{1,\dots,n}$
		\item $\mc_k \supset \mc_{k+1}$ for $k \in \setof{1,\dots,n-1}$,
		\item $DPI_j$ includes a proper subset of the test cases $DPI_{j+1}$ includes for $j \in \setof{1,\dots,n-1}$,
		\item $DPI_n$ is equal to $DPI$ or includes a proper subset of the test cases $DPI$ includes and 
		\item $DPI_{prev}$ includes a proper subset of the test cases $DPI_1$ includes,
	\end{itemize}
then $\mc_n$ is the active sublabel of $\mathsf{node}_{e}.\mathsf{cs}[r]$ w.r.t.\ $DPI$. However, as argued before, the minimal conflict set $\mc_n$ w.r.t.\ $DPI_n$ cannot be the empty set. As a consequence, we obtain that there must be a non-empty active sublabel of $\mathsf{node}_{e}.\mathsf{cs}[r]$ w.r.t.\ $DPI$. 

By Propositions~\ref{prop:changes_in_conflict_sets_after_testcase_added} and \ref{prop:if_DPI_cs_neq_emptyset_then_DPI+1_cs_neq_emptyset}, there is a non-empty minimal conflict set $\mc'$ w.r.t.\ $DPI$ such that $\mc' \subseteq \mc_n$. Due to $\mc_n \subset \dots \subset \mc_1 \subset \mathsf{node}_{e}.\mathsf{cs}[r] = L$ we conclude that $\mc_n \subset L$. Therefore, $\emptyset \subset \mc' \subset L$ holds.

By Proposition~\ref{prop:mindiag_mincs}, each minimal diagnosis w.r.t.\ $DPI$ is a minimal hitting set of all minimal conflict sets w.r.t.\ $DPI$. Thence, we have that 
%$\md \cap \mc \neq \emptyset$ for each minimal conflict set $\mc$ w.r.t.\ the current DPI. As $\mc'$ is a minimal conflict set w.r.t.\ the current DPI, it must hold that 
$\mc' \cap \md \neq \emptyset$. 
%So, by $\mc' \subset L$, we have that $L \cap \md \neq \emptyset$. 
So, by $\mc' \subset L$, we have that $\emptyset \subset \mc' \cap \md \subseteq L \cap \md \subseteq L$. 
Consequently, we define $\mathsf{nd}_{suc} := \mathsf{node}_{x} = \textsc{add}(\mathsf{node},x)$ with $\mathsf{nd}_{suc}.\mathsf{cs} := \mathsf{node}_{x}.\mathsf{cs} =\textsc{add}(\mathsf{node.cs},L)$ for some $x \in \mc' \cap \md \subseteq L$. Then, $\mathsf{nd}_{suc} \subseteq \md$ because $\mathsf{node} \subset \md$ and $x \in \md$. It is clear from the inference so far that $\mathsf{nd} \subset \mathsf{nd}_{suc}$, $|\mathsf{nd}_{suc}| = |\mathsf{nd}| + 1$ and $\mathsf{nd}_{suc} \in GenNodes$. This shows the truth of propositions (1)-(3).

Proposition~(4) must hold by lines~\ref{algoline:dyn:check_node_already_in_Q}-\ref{algoline:dyn:generate_nodes}.

Now we argue why propositions~(5) and (6) must hold. Assume that $\mathsf{nd}'_{suc} \in \Queue$ is redundant w.r.t.\ some DPI $DPI''_{prev}$ which is equal to $DPI$ or includes a subset of the test cases in $DPI$.  
Then, there must be some minimal conflict set $\mc''$ w.r.t.\ $DPI''_{prev}$ which is a witness of redundancy of $\mathsf{nd}'_{suc}$. Suppose that \textsc{prune} is called given $X := \mc''$ as an argument.

Now, we have to distinguish two cases: Either 
\begin{itemize}
	\item (q1)~$\mathsf{nd}_{suc}$ was added to $\Queue$ after it was generated or 
	\item (q2)~$\mathsf{nd}_{suc}$ was added to $\Queue_{dup}$ after it was generated
\end{itemize}
(there are no other possibilities, see lines~\ref{algoline:dyn:for_e_in_L}-\ref{algoline:dyn:generate_nodes}). 

For each of these two cases, there are two more cases to discriminate between: 
\begin{itemize}
	\item (c1)~$\mc'' \subset \mathsf{nd}'_{suc}.\mathsf{cs}[|\mathsf{nd}'_{suc}|]$ and   
$\mathsf{nd}'_{suc}[|\mathsf{nd}'_{suc}|] \in \mathsf{nd}'_{suc}.\mathsf{cs}[|\mathsf{nd}'_{suc}|] \setminus \mc''$ or
	\item (c2)~$\mc'' \subset \mathsf{nd}'_{suc}.\mathsf{cs}[j]$ and 
$\mathsf{nd}'_{suc}[j] \in \mathsf{nd}'_{suc}.\mathsf{cs}[j] \setminus \mc''$ for some $j\in\setof{1,\dots,|\mathsf{nd}'_{suc}|-1}$. 
\end{itemize}
 
%\textbf{Case~(q1):} 
\paragraph{Case~(q1):}
Here, we have that $\mathsf{nd}'_{suc}$ is the same node as $\mathsf{nd}_{suc}$ since $\mathsf{nd}_{suc}$ was added to $\Queue$ after generation and no node replacement can have taken place because $\mathsf{nd}'_{suc}$ is defined as the node set-equal to $\mathsf{nd}_{suc}$ that is an element of $\Queue$ \emph{immediately} after $\mathsf{nd}_{suc}$ has been generated. And, only one node corresponding to one and the same set can be in $\Queue$ at the same time. 

%\textbf{In (c1)}, 
\paragraph{Case~(c1):}
We have that $\mc''$ must be equal to some minimal conflict set $\mc_j$ in the sequence $\mc_1,\dots,\mc_n$. This must be \true since, first, $DPI''_{prev}$ is equal to $DPI$ or includes a subset of the test cases in $DPI$ and $DPI_{prev}$ includes a proper subset of the test cases in $DPI''_{prev}$.
 
To understand why the latter must hold, recall that $DPI_{prev}$ is the DPI of the call to \textsc{dynamicHS} where $\mathsf{nd}_{suc}$ was generated and the minimal conflict set $L$ was computed. By assumption, however, there is some minimal conflict set w.r.t.\ $DPI''_{prev}$, namely $\mc''$, such that $\mc'' \subset \mathsf{nd}'_{suc}.\mathsf{cs}[|\mathsf{nd}'_{suc}|] = L$. Hence, it cannot be \true that both $L$ and $\mc''$ are minimal conflict sets w.r.t.\ the same DPI. Otherwise, we would have a contradiction to the minimality of $L$. By Proposition~\ref{prop:changes_in_conflict_sets_after_testcase_added}, which states that minimal conflict sets cannot grow by the addition of new test cases to the DPI, we obtain the claimed fact that $DPI_{prev}$ includes a proper subset of the test cases in $DPI''_{prev}$.

Second, the sequence $\mc_1,\dots,\mc_n$ comprises \emph{all} sets $X$ given as an argument to \textsc{prune} and \textsc{pruneQdup} during \emph{all} executions of \textsc{dynamicHS} from the one with current DPI $DPI_{prev}$ up to and including the one with current DPI $DPI$ where $\mathsf{nd}'_{suc}.\mathsf{cs}[|\mathsf{nd}'_{suc}|] \supset \mc_1 \supset \dots \supset \mc_n$ holds. Reason for this to be valid is the fact that $\mathsf{nd}'_{suc}$ is the same node as $\mathsf{nd}_{suc}$ in the currently considered case~(q1).

Now, recall that $\mc'$ is a minimal conflict set w.r.t.\ $DPI$ such that $x \in \mc' \cap \md \subset L$. Further, by $\mathsf{nd}'_{suc}=\mathsf{node}_x$, we have that $\mathsf{nd}'_{suc}[|\mathsf{nd}'_{suc}|] = x$.
Due to $\mc' \subseteq \mc_n$ and $\mc_n \subseteq \mc_j$, we have that $\mc' \subseteq \mc_j$.
% \subset L$, 
Therefore, we can infer by $\mc'' = \mc_j$ that $\mc' \subseteq \mc''$
% \subset L$ 
is \true. 
Now, $x \in \mc'$ implies that $x \in \mc''$ wherefore $x \notin \mathsf{nd}'_{suc}.\mathsf{cs}[|\mathsf{nd}'_{suc}|] \setminus \mc''$. By $x = \mathsf{nd}'_{suc}[|\mathsf{nd}'_{suc}|]$, this is a contradiction to the assumption of case~(c1). Hence, case~(c2) must arise.
%%%%%%%%%%%%%%%%%%%%%%%%%%%%%%% NEW end %%%%%%%%%%%%%%%%%%%%%%%%%%%%%%%%%%%%%

%\textbf{In (c2)}, 
\paragraph{Case~(c2):}
We have that $\mathsf{nd}'_{suc}[1..|\mathsf{nd}'_{suc}|-1]$ is the same node as $\mathsf{node}$ since $\mathsf{nd}'_{suc} = \mathsf{node}_x$. 
Then, there are two cases: Either
\begin{itemize}
	\item (s1)~$\mathsf{node}$ is the same node as $\mathsf{nd}$ or
	\item (s2)~$\mathsf{node}$ is not the same node as $\mathsf{nd}$.
\end{itemize}

%\textbf{Case~(s1):}
\paragraph{Case~(s1):} 
If $\mathsf{node}$ is the same node as $\mathsf{nd}$, then $\mathsf{node}$ is de-facto non-redundant w.r.t.\ $DPI$ since $\mathsf{nd}$ is de-facto non-redundant w.r.t.\ $DPI$ by precondition. Moreover, $x$ is an element of the active sublabel of $\mathsf{nd}'_{suc}[|\mathsf{nd}'_{suc}|]$ w.r.t.\ $DPI$, as specified before. Thus, by Definition~\ref{def:de-facto_non-redundant}, $\mathsf{nd}'_{suc}$ is de-facto non-redundant w.r.t.\ $DPI$. Hence, \textsc{prune} cannot be given an argument $\mc''$ which is a witness of redundancy of $\mathsf{nd}'_{suc}$ where $\mc''$ is a minimal conflict set w.r.t.\ $DPI''_{prev}$. This holds due to 
\begin{itemize}
	\item the fact that $DPI''_{prev}$ comprises a (not necessarily proper) subset of the test cases in $DPI$, 
	\item Proposition~\ref{prop:de-facto_non_redundant_node_cannot_be_pruned_or_replaced} which states that a de-facto non-redundant node w.r.t.\ $DPI$ cannot be pruned or replaced during any execution of \textsc{dynamicHS} with a current DPI that includes a (not necessarily proper) subset of the test cases in $DPI$ and
	\item Lemma~\ref{lem:prune} which says that $\mathsf{nd}'_{suc}$ would be replaced or pruned in case that \textsc{prune} is called given a witness of redundancy of $\mathsf{nd}'_{suc}$.
\end{itemize}
So, we have derived a contradiction to the assumption that \textsc{prune} is called given a minimal conflict set $X := \mc''$ w.r.t.\ $DPI''_{prev}$ which is a witness of redundancy of $\mathsf{nd}'_{suc}$. Hence, case~(s2) must be \true.
\paragraph{Case~(s2):}
If $\mathsf{node}$ is not the same node as $\mathsf{nd}$, then $\mathsf{node}$ may or may not be de-facto non-redundant w.r.t.\ $DPI$. In the former case, the same argumentation as in case~(s1) applies and yields a contradiction. In the latter case, we know that $\mc'' \subset \mathsf{nd}'_{suc}.\mathsf{cs}[j]$ as well as $\mathsf{nd}'_{suc}[j] \in \mathsf{nd}'_{suc}.\mathsf{cs}[j] \setminus \mc''$ must be \true for some $j\in\setof{1,\dots,|\mathsf{nd}'_{suc}|-1}$. So, by Lemma~\ref{lem:prune}, $\mathsf{nd}'_{suc}$ is not an element of the returned list $\Queue'$ of the call to \textsc{prune} given the arguments $\Queue$ (which includes $\mathsf{nd}'_{suc}$), $X := \mc''$ and $\Queue_{dup}$. 

However, at least one replacement node of $\mathsf{nd}'_{suc}$ must be found by \textsc{prune}. This must hold by the following reasoning:

First, $\mathsf{nd}\in\Queue_{dup}$ must hold at the time this call to \textsc{prune} is made. This is satisfied since
\begin{itemize}
\item the entire (current) list $\Queue_{dup}$ is browsed for an alternative subnode of $\mathsf{nd}'_{suc}$, 
\item $\mathsf{nd}\in\Queue_{dup}$ holds at some point in time during the execution of \textsc{dynamicHS} with the current DPI $DPI_{prev}$ due to the fact that $\mathsf{node}$ is not the same node as $\mathsf{nd}$ and the argumentation at the beginning of this proof,
\item $DPI_{prev}$ includes a subset of the test cases in $DPI''_{prev}$,
\item $DPI''_{prev}$ includes a subset of the test cases in $DPI$,
%as shown at the beginning of this proof (because otherwise $\mathsf{node}$ must be the same node as $\mathsf{nd}$ which would be a contradiction),
%
%\item due to Lemma~\ref{lem:pruneQdup} which states that only elements can be deleted from $\Queue_{dup}$ in \textsc{pruneQdup} for which a witness of redundancy w.r.t.\ the currently considered DPI exists, i.e.\ which are redundant w.r.t.\ this DPI and
%, $\mathsf{nd}$ cannot have been deleted from $\Queue_{dup}$ during the execution of \textsc{pruneQdup} which is executed before \textsc{prune} and 
\item Proposition~\ref{prop:de-facto_non_redundant_node_cannot_be_pruned_or_replaced} states that a de-facto non-redundant node w.r.t.\ $DPI$ cannot be pruned or replaced during the execution of \textsc{dynamicHS} with a current DPI that includes a subset of the test cases in $DPI$,
\item nodes can only be deleted from $\Queue_{dup}$ by being pruned and
\item $\mathsf{nd}$ is de-facto non-redundant w.r.t.\ $DPI$. 
\end{itemize}

Second, by line~\ref{algoline:dyn:add_to_Qdup} and \textsc{pruneQdup}, which are the only places in \textsc{dynamicHS} where $\Queue_{dup}$ is modified, $\Queue_{dup}$ is sorted in ascending order by node cardinality at any time during the execution of any call to \textsc{dynamicHS}. 

Third, in order to construct a replacement node of $\mathsf{nd}'_{suc}$, \textsc{prune} first determines the maximal $k$ such that $\mc'' \subset \mathsf{nd}'_{suc}.\mathsf{cs}[k]$ and $\mathsf{nd}'_{suc}[k] \in \mathsf{nd}'_{suc}.\mathsf{cs}[k] \setminus \mc''$. As case~(c1) was proven to be false, we conclude that $k \leq |\mathsf{nd}'_{suc}|-1$ must hold. Then, in line~\ref{algoline:prune:check_for_alternative_paths_start}, an alternative subnode of $\mathsf{nd}'_{suc}$ 
\begin{itemize}
	\item which has cardinality $k + z$ where $z \geq 0$ is minimal and 
	\item from which a replacement node of $\mathsf{nd}'_{suc}$ can be constructed
\end{itemize}
is searched for in $\Queue_{dup}$. To see this, observe that elements in $\Queue_{dup}$ -- which is sorted in ascending order of node cardinality, as argued -- are visited in order starting from the lowest cardinality node (line~\ref{algoline:prune:check_for_alternative_paths_start}).  

Fourth, $\mathsf{nd}\in\Queue_{dup}$ is an alternative equal node of $\mathsf{node}$. Since $\mathsf{nd}'_{suc} = \mathsf{node}_x$, we have that $\mathsf{nd}$ is an alternative subnode of $\mathsf{nd}'_{suc}$ such that $k \leq |\mathsf{nd}'_{suc}|-1 = |\mathsf{nd}|$.

Thus, we have that one replacement node of $\mathsf{nd}'_{suc}$ is definitely found by \textsc{prune}. And, in case there is only one replacement node of $\mathsf{nd}'_{suc}$ constructable during \textsc{prune}, then this replacement node is given by $\mathsf{nd}'_{suc,new} := \textsc{add}(\mathsf{nd},x)$ with $\mathsf{nd}'_{suc,new}.\mathsf{cs} := \textsc{add}(\mathsf{nd.cs},L)$. By the de-facto non-redundancy of $\mathsf{nd}$ and since $x$ is specified as an element of the active sublabel of $\mathsf{nd}'_{suc}.\mathsf{cs}[|\mathsf{nd}'_{suc}|]$ w.r.t.\ $DPI$ (see above), we obtain by Definition~\ref{def:de-facto_non-redundant} that $\mathsf{nd}'_{suc,new}$ is a de-facto non-redundant node w.r.t.\ $DPI$.
Thence, proposition~(5) is \true.

Due to $|\mathsf{nd}| = |\mathsf{node}| = |\mathsf{nd}'_{suc}|-1$, the alternative subnode of $\mathsf{nd}'_{suc}$ \emph{actually} found by \textsc{prune} 
%given the witness of redundancy $\mc''$ 
cannot have a cardinality greater than $|\mathsf{nd}'_{suc}|-1$. 
So, let $\mathsf{nd}_{alt}$ be the found alternative subnode of $\mathsf{nd}'_{suc}$. Since $|\mathsf{nd}_{alt}|\leq |\mathsf{nd}'_{suc}|-1$, we obtain that the replacement node $\mathsf{nd}'_{suc,new,1}$ of $\mathsf{nd}'_{suc}$ constructed from $\mathsf{nd}_{alt}$ must meet $\mathsf{nd}'_{suc,new,1}[|\mathsf{nd}'_{suc}|] = \mathsf{nd}'_{suc}[|\mathsf{nd}'_{suc}|] = x$ as well as $\mathsf{nd}'_{suc,new,1}.\mathsf{cs}[|\mathsf{nd}'_{suc}|] = \mathsf{nd}'_{suc}.\mathsf{cs}[|\mathsf{nd}'_{suc}|] = L$.
%So, for the replacement node it holds, by the reasoning conducted, that the element at the $|\mathsf{nd}'_{suc}|$-th position must be $x$, as in $\mathsf{nd}'_{suc}$. 
That is, the first $|\mathsf{nd}| = |\mathsf{node}| = |\mathsf{nd}'_{suc}|-1$ positions of $\mathsf{nd}'_{suc,new,1}$ as a set correspond to a node in a transitive replaces-relation with $\mathsf{nd}$. 

Therefore, the same line of argument as used for $\mathsf{nd}'_{suc}$ can be applied to any node $\mathsf{nd}'_{suc,rep}$ in a transitive replaces-relation with $\mathsf{nd}'_{suc}$. That is, the following must be valid for any node $\mathsf{nd}'_{suc,rep}$ in a transitive replaces-relation with $\mathsf{nd}'_{suc}$: 
\begin{itemize}
	\item $\mathsf{nd}'_{suc,rep}[|\mathsf{nd}'_{suc}|] = x$ and $\mathsf{nd}'_{suc,rep}.\mathsf{cs}[|\mathsf{nd}'_{suc}|] = L$.
	\item If \textsc{prune} is called given a witness of redundancy of $\mathsf{nd}'_{suc,rep}$, then some replacement node of $\mathsf{nd}'_{suc,rep}$ is found. And, if only one replacement node of $\mathsf{nd}'_{suc,rep}$ is constructable, then this replacement node is de-facto non-redundant w.r.t.\ $DPI$.
\end{itemize}
After once a replacement node of $\mathsf{nd}'_{suc}$ or of some node in a transitive replaces-relation with $\mathsf{nd}'_{suc}$ is found which is de-facto non-redundant w.r.t.\ $DPI$, this replacement node cannot be replaced or pruned by Proposition~\ref{prop:de-facto_non_redundant_node_cannot_be_pruned_or_replaced}. Therefore, by Lemma~\ref{lem:prune}, no witness of redundancy of this replacement node can exist w.r.t.\ any DPI including a (not necessarily proper) subset of the test cases in $DPI$. Thence, proposition~(6) is \true.
\paragraph{Case~(q2):}
Here, we have that $\mathsf{nd}'_{suc}$ is not the same node as $\mathsf{nd}_{suc}$. This must be valid as $\mathsf{nd}'_{suc}$ is defined as the node set-equal to $\mathsf{nd}_{suc}$ that is an element of $\Queue$ \emph{immediately} after $\mathsf{nd}_{suc}$ was generated and $\mathsf{nd}_{suc}$ is assumed to be added to $\Queue_{dup}$ after being generated.
% and there can be only one node corresponding to one and the same set in $\Queue$ at the same time.
%Assume now, legitimately, that $\mathsf{nd}_{suc}$ is defined as $\mathsf{node}_{x}$ as shown above.

Now, independently of whether \textbf{(c1) or (c2)} occurs, the following holds: If \textsc{prune} is called given a witness of redundancy $\mc''$ of $\mathsf{nd}'_{suc}$ w.r.t.\ $DPI''_{prev}$, then a replacement node of $\mathsf{nd}'_{suc}$ is found. And, if only one replacement node of $\mathsf{nd}'_{suc}$ is constructable, then this replacement node 
%found in lines~\ref{algoline:prune:check_if_alternative_subnode}-\ref{algoline:prune:insert_alternative_equal_node_into_S} 
is de-facto non-redundant w.r.t.\ $DPI$. 

To understand why this must hold, first recall that $\mathsf{nd}_{suc}$ is a successor of $\mathsf{node}$, i.e.\ $\mathsf{nd}_{suc}[1..|\mathsf{nd}_{suc}|-1]$ is the same node as $\mathsf{node}$. Furthermore, $\mathsf{node}$ is the node set-equal to $\mathsf{nd}$ that is processed. That is, $\mathsf{node}$ is either the same node as $\mathsf{nd}$ or it is in a transitive replaces-relation with $\mathsf{nd}$. Then, the same two cases~(s1) and (s2) can be distinguished as in case~(q1)(c2) where (s1) leads to a contradiction. So, case~(s2) must be \true. That is, $\mathsf{node}$ is not the same node as $\mathsf{nd}$. Hence, by the argumentation in case~(q1)(c2)(s2), $\mathsf{nd} \in \Queue_{dup}$ must hold during the execution of any call to \textsc{dynamicHS} with a current DPI that comprises a (not necessarily proper) superset of the test cases in $DPI_{prev}$ -- which is the current DPI at the time $\mathsf{nd}$ is generated -- and a (not necessarily proper) subset of the test cases in $DPI$.
In particular, this implies that $\mathsf{nd} \in \Queue_{dup}$ at the time \textsc{prune} is called given the witness of redundancy $\mc''$ of $\mathsf{nd}'_{suc}$ w.r.t.\ $DPI''_{prev}$ as an argument.

By assumption, $\mathsf{nd}_{suc}$ has been added to $\Queue_{dup}$ after being generated. Now, suppose \textsc{pruneQdup} is called given a witness of redundancy $\mc'$ of $\mathsf{nd}_{suc} \in \Queue_{dup}$ w.r.t.\ some DPI $DPI'_{prev}$ as an argument. Then $DPI'_{prev}$ must comprise a (not necessarily proper) superset of the test cases in $DPI_{prev}$. This can be concluded from Lemma~\ref{lem:node_not_processed_before_all_subset_nodes_generated} which implies that $\mathsf{nd}_{suc}$ cannot have been generated during an execution of \textsc{dynamicHS} with a current DPI including a proper subset of the test cases in $DPI_{prev}$. Hence, the argumentation before implicates that $\mathsf{nd} \in \Queue_{dup}$ at the time \textsc{pruneQdup} is called given the witness of redundancy $\mc'$ of $\mathsf{nd}_{suc}$ w.r.t.\ $DPI'_{prev}$ as an argument.

Thus, $\mathsf{nd}_{suc}$ cannot be pruned on account of Lemma~\ref{lem:pruneQdup} which says that a node can only be pruned from $\Queue_{dup}$ if the set $Comb_{\mathsf{nd}_{suc}}(\Queue_{dup})$ of combined equal nodes of $\mathsf{nd}_{suc}$ of $\Queue_{dup}$ (cf.\ Definition~\ref{def:comb_node}) is the empty set.

However, $Comb_{\mathsf{nd}_{suc}}(\Queue_{dup}) \neq \emptyset$ must be valid. Because we demonstrated that
\begin{itemize}
	\item $\mathsf{nd} \in \Queue_{dup}$,
	\item $\mathsf{nd}_{suc} \in \Queue_{dup}$,
	\item $\mathsf{nd}_{suc}$ is the same node as $\mathsf{node}_x = \textsc{add}(\mathsf{node},x)$ with $\mathsf{nd}_{suc}.\mathsf{cs}$ being equal to $\mathsf{node}_x.\mathsf{cs} = \textsc{add}(\mathsf{node.cs},L)$,
	\item $\mathsf{nd} = \mathsf{node}$ and
	\item $x$ is specified as an element of the active sublabel of $\mathsf{nd}_{suc}.\mathsf{cs}[|\mathsf{nd}_{suc}|]$ w.r.t.\ $DPI$ (see above) wherefore $x \notin \mathsf{nd}_{suc}.\mathsf{cs}[|\mathsf{nd}_{suc}|] \setminus \mc'$.
\end{itemize}
Therefore, 
\begin{align*}
\mathsf{nd}_{comb} &:= \textsc{add}(\mathsf{nd},\mathsf{nd}_{suc}[|\mathsf{nd}|+1..|\mathsf{nd}_{suc}|]) = \textsc{add}(\mathsf{nd},x) \\
\mathsf{nd}_{comb}.\mathsf{cs} &:= \textsc{add}(\mathsf{nd}.\mathsf{cs},\mathsf{nd}_{suc}.\mathsf{cs}[|\mathsf{nd}|+1..|\mathsf{nd}_{suc}|]) = \textsc{add}(\mathsf{nd}.\mathsf{cs},L)
\end{align*}
%$\mathsf{nd}_{comb} := \textsc{add}(\mathsf{nd},\mathsf{nd}_{suc}[|\mathsf{nd}|+1..|\mathsf{nd}_{suc}|]) = \textsc{add}(\mathsf{nd},x)$ with $\mathsf{nd}_{comb}.\mathsf{cs} := \textsc{add}(\mathsf{nd}.\mathsf{cs},\mathsf{nd}_{suc}.\mathsf{cs}[|\mathsf{nd}|+1..|\mathsf{nd}_{suc}|]) = \textsc{add}(\mathsf{nd}.\mathsf{cs},L)$ 
is a combined equal node of $\mathsf{nd}_{suc}$ of $\Queue_{dup}$, i.e.\ $\mathsf{nd}_{comb} \in Comb_{\mathsf{nd}_{suc}}(\Queue_{dup})$. The node $\mathsf{nd}_{comb}$ is de-facto non-redundant w.r.t.\ $DPI$ as $\mathsf{nd}$ is de-facto non-redundant w.r.t.\ $DPI$ and since $x$ is an element of the active sublabel of $\mathsf{nd}_{suc}.\mathsf{cs}[|\mathsf{nd}_{suc}|]$ w.r.t.\ $DPI$.
 
By Definition~\ref{def:comb_node}, any combined equal node of $\mathsf{nd}_{suc}$ must share the element at the $|\mathsf{nd}_{suc}|$-th position with $\mathsf{nd}_{suc}$ and $\mathsf{nd}_{suc}.\mathsf{cs}$, respectively. Hence, the first $|\mathsf{nd}_{suc}| - 1$ elements of a combined equal node of $\mathsf{nd}_{suc}$ are set-equal to the first $|\mathsf{nd}_{suc}| - 1$ elements of $\mathsf{nd}_{suc}$. So, there exists a combined equal node, namely $\mathsf{nd}_{comb}$, of any (redundant) node that results from $\mathsf{nd}_{suc}$ by a set of combined replacements.
%As argued in case~(q1)(c2), this node $\mathsf{nd}_{comb}$ (denoted by $\mathsf{nd}'_{suc,new}$ in case~(q1)(c2)) is de-facto non-redundant w.r.t.\ $DPI$.

By Lemma~\ref{lem:non-redundant_node_in_Comb(Qdup)_persists}, the fact that $\mathsf{nd}_{comb} \in Comb_{\mathsf{nd}_{suc}}(\Queue_{dup}) \subseteq Comb(\Queue_{dup})$ at some point in time during the execution of \textsc{dynamicHS} with current DPI $DPI'_{prev}$ and the de-facto non-redundancy of $\mathsf{nd}_{comb}$ w.r.t.\ $DPI$, we conclude that, during any execution of \textsc{dynamicHS} with a current DPI that includes a (not necessarily proper) superset of the test cases in $DPI'_{prev}$ and includes a (not necessarily proper) subset of the test cases in $DPI$, $\mathsf{nd}_{comb} \in Comb(\Queue_{dup})$ must hold. 
Because $DPI'_{prev}$ is an arbitrary DPI that comprises a (not necessarily proper) superset of the test cases in $DPI_{prev}$, we derive that $\mathsf{nd}_{comb} \in Comb(\Queue_{dup})$ must be \true particularly during the execution of \textsc{dynamicHS} with the current DPI $DPI''_{prev}$.

If $\mc''$ is a witness of redundancy of $\mathsf{nd}_{suc} \in \Queue_{dup}$, then the updated list $\Queue_{dup}$ returned by \textsc{pruneQdup} must include a combined replacement node of $\mathsf{nd}_{suc}$, either $\mathsf{nd}_{comb}$ or some other node. Otherwise, i.e.\ if $\mc''$ is not a witness of redundancy of $\mathsf{nd}_{suc} \in \Queue_{dup}$, the updated list $\Queue_{dup}$ returned by \textsc{pruneQdup} must include $\mathsf{nd}_{suc}$.
 
\textsc{prune} is always called immediately after \textsc{pruneQdup} and thus uses the updated list $\Queue_{dup}$ which comprises a node set-equal to $\mathsf{nd}_{suc}$ and thus set-equal to $\mathsf{nd}'_{suc}$.
%$\mathsf{nd}_{comb}$ and because $\mathsf{nd}_{comb} = \mathsf{nd}_{suc} = \mathsf{nd}'_{suc}$, 
Consequently, we have that one replacement node of $\mathsf{nd}'_{suc}$ is definitely found by \textsc{prune}. And, in case there is only one replacement node of $\mathsf{nd}'_{suc}$ constructable during \textsc{prune}, this replacement node is given by $\mathsf{nd}_{comb}$. Thence, proposition~(5) is \true.
%
%Further on, $\mathsf{nd}_{comb} = \mathsf{nd}'_{suc}$ is \true.

Independently of which replacement node of $\mathsf{nd}'_{suc}$ is \emph{actually} found by \textsc{prune}, a set-equality between this replacement node and $\mathsf{nd}_{comb}$ will hold. This is \true since $\mathsf{nd}_{comb} = \mathsf{nd}'_{suc}$ and since each replacement node, by definition, is set-equal to the node it replaces. Consequently, this set-equality holds for any node in a transitive replaces-relation with $\mathsf{nd}'_{suc}$. 
So, we have that one replacement node of any node $\mathsf{nd}'_{suc,rep}$ in a transitive replaces-relation with $\mathsf{nd}'_{suc}$ is definitely found by \textsc{prune}. And, in case there is only one replacement node of $\mathsf{nd}'_{suc,rep}$ constructable during \textsc{prune}, this replacement node is given by $\mathsf{nd}_{comb}$ which is de-facto non-redundant w.r.t.\ $DPI$. 

That $\mathsf{nd}_{comb}$, after it has been used as a replacement node of $\mathsf{nd}'_{suc}$ or of some node in a transitive replaces-relation with $\mathsf{nd}'_{suc}$, cannot be pruned or replaced, follows from
Proposition~\ref{prop:de-facto_non_redundant_node_cannot_be_pruned_or_replaced} and the fact that $\mathsf{nd}_{comb}$ is de-facto non-redundant w.r.t.\ $DPI$. Therefore, by Lemma~\ref{lem:prune}, no witness of redundancy of $\mathsf{nd}_{comb}$ can exist w.r.t.\ any DPI including a (not necessarily proper) subset of the test cases in $DPI$. Thence, proposition~(6) is \true.
\end{proof}

The next result, Lemma~\ref{lem:successor_node_subset_diag_exists_2}, assumes an arbitrary fixed ``current'' DPI $DPI$ such that \textsc{dynamicHS} with this ``current'' DPI $DPI$ returns due to $\Queue = []$. Further on, it assumes an arbitrary minimal diagnosis $\md$ w.r.t.\ $DPI$ and a node $\mathsf{nd}$ which is a proper subset of $\md$ such that $\mathsf{nd}$ is an element of $\Queue$ anytime throughout all executions of \textsc{dynamicHS} during the execution of Algorithm~\ref{algo:inter_onto_debug} up to the one with the current DPI $DPI$. Additionally, $\mathsf{nd}$ cannot be pruned. It might be replaced; and in case there is only one potential replacement node of $\mathsf{nd}$ constructable from (the combined nodes of) $\Queue_{dup}$, this replacement node is de-facto non-redundant w.r.t.\ $DPI$. Any node $\mathsf{nd}'$ in a transitive replaces relation with $\mathsf{nd}$ cannot be pruned either. It might again be replaced. In case there is only one potential replacement node of $\mathsf{nd}'$ constructable from (the combined nodes of) $\Queue_{dup}$, this replacement node is de-facto non-redundant w.r.t.\ $DPI$.

Given these preconditions, the lemma establishes the existence of a node $\mathsf{nd}_{suc}$ that corresponds to a superset of $\mathsf{nd}$ and to a subset of $\md$, includes one element more than the set $\mathsf{nd}$ and is generated anytime throughout all executions of \textsc{dynamicHS} during the execution of Algorithm~\ref{algo:inter_onto_debug} up to the one with the current DPI $DPI$. Moreover, it states that the node $\mathsf{nd}'_{suc}$ set-equal to this generated node that is an element of $\Queue$ cannot be pruned. However, it might be replaced. In case there is only one potential replacement node of $\mathsf{nd}'_{suc}$ constructable from (the combined nodes of) $\Queue_{dup}$, this replacement node is de-facto non-redundant w.r.t.\ $DPI$. 
%A replacement node of $\mathsf{nd}'_{suc}$ or 
Any node $\mathsf{nd}'_{suc,rep}$ in a transitive replaces relation with $\mathsf{nd}'_{suc}$ cannot be pruned either. It might again be replaced. In case there is only one potential replacement node of $\mathsf{nd}'_{suc,rep}$ constructable from (the combined nodes of) $\Queue_{dup}$, this replacement node is de-facto non-redundant w.r.t.\ $DPI$. 

Pictured, with respect to the hitting set tree constructed by \textsc{dynamicHS}, this lemma purports the following: Let the hitting set tree produced by \textsc{dynamicHS} be completely constructed for an arbitrary DPI $DPI$. 
In case there is any tree branch whose edge labels correspond to a part of the minimal diagnosis $\md$ w.r.t.\ $DPI$ and which is known to be definitely not pruned during this tree construction, 
%until $DPI$ is the current DPI, 
then this branch must be extended by one edge labeled by an element of $\md$ and this extended path is known to be definitely not pruned during this tree construction.
%In case there is any path corresponding to a partial minimal diagnosis $\md$ w.r.t.\ $DPI$ which is known to be definitely not pruned during this construction, 
%%until $DPI$ is the current DPI, 
%then this path must be extended by one edge labeled by an element of $\md$ and this extended path is known to be definitely not pruned during this construction.
\begin{lemma}\label{lem:successor_node_subset_diag_exists_2}
Assume the execution of \textsc{dynamicHS} with the current DPI $DPI$ and assume that the execution stops due to $\Queue = []$. Let 
\begin{itemize}
\item $GenNodes$ be the set of all nodes generated throughout the execution of all calls to \textsc{dynamicHS} during the execution of Algorithm~\ref{algo:inter_onto_debug},
\item $\md$ be some minimal diagnosis w.r.t.\ $DPI$, 
%\item $DPI'_{prev}$ and $DPI''_{prev}$ be two DPIs each of which is either equal to $DPI$ or includes fewer test cases than $DPI$ and each of which is a current DPI during any call to \textsc{dynamicHS},
\item $DPI'_{prev}$ be a DPI which is either equal to $DPI$ or includes fewer test cases than $DPI$ and which is the current DPI during any particular call to \textsc{dynamicHS},
%used throughout the execution of the current call to \textsc{dynamicHS},
\item $\mathsf{nd}$ be some node such that the following holds: 
\begin{itemize}
%\item $\mathsf{nd}$ is redundant w.r.t.\ $DPI$, 
\item $\mathsf{nd} \subset \md$. 
\item There is some execution of \textsc{dynamicHS} with current DPI $DPI'_{prev}$ during which it holds at some point in time that $\mathsf{nd}\in\Queue$.
%and $\mathsf{nd}$ is processed during the execution of \textsc{dynamicHS} with current DPI $DPI$ 
%\item if \textsc{prune} is called given a witness of redundancy of $\mathsf{nd}$, then some replacement node of $\mathsf{nd}$ is found which is non-redundant w.r.t.\ $DPI$,
\item If \textsc{prune} is called given a witness of redundancy of $\mathsf{nd}$, then some replacement node of $\mathsf{nd}$ is found. If only one replacement node of $\mathsf{nd}$ is found, then this replacement node is de-facto non-redundant w.r.t.\ $DPI$. 
\item Let $\mathsf{nd}'$ be in a transitive replaces-relation with $\mathsf{nd}$. If \textsc{prune} is called given a witness of redundancy of $\mathsf{nd}'$, then some replacement node of $\mathsf{nd}'$ is found. If only one replacement node of $\mathsf{nd}'$ is found, then this replacement node is de-facto non-redundant w.r.t.\ $DPI$.
\end{itemize}
%\item $\mD_{\checkmark}$ in \textsc{dynamicHS} include the set of all known minimal diagnoses w.r.t.\ the DPI $DPI'$ that is the current DPI of a particular call to \textsc{dynamicHS}. 
\end{itemize}
Then there are nodes $\mathsf{nd}_{suc}$ and $\mathsf{nd}'_{suc}$ such that the following holds:
\begin{enumerate}[(1)]
\item $\mathsf{nd} \subset \mathsf{nd}_{suc} \subseteq \md$. 
\item $|\mathsf{nd}_{suc}| = |\mathsf{nd}| + 1$. 
\item $\mathsf{nd}_{suc} \in GenNodes$.
\item $\mathsf{nd}'_{suc} = \mathsf{nd}_{suc}$ is an element of $\Queue$ immediately after $\mathsf{nd}_{suc}$ has been generated.
%\item If $\mathsf{nd}'_{suc}$ is redundant w.r.t.\ $DPI''_{prev}$, there is a replacement node $\mathsf{nd}''_{suc}$ of $\mathsf{nd}'_{suc}$ such that $\mathsf{nd}''_{suc}$ is non-redundant w.r.t.\ $DPI''$.
\item If \textsc{prune} is called given a witness of redundancy of $\mathsf{nd}'_{suc}$, then some replacement node of $\mathsf{nd}'_{suc}$ is found. If only one replacement node of $\mathsf{nd}'_{suc}$ is found, then this replacement node is de-facto non-redundant w.r.t.\ $DPI$. 
%\item If \textsc{prune} is called given a witness of redundancy of $\mathsf{nd}'_{suc}$, then some replacement node of $\mathsf{nd}'_{suc}$ is found which is non-redundant w.r.t.\ $DPI$.
%
%\item Each node $\mathsf{nd}'_{suc,t-rep}$ in a transitive replaces-relation with $\mathsf{nd}'_{suc}$ satisfies $\mathsf{nd}'_{suc}[|\mathsf{nd}'_{suc}|] = \mathsf{nd}'_{suc,t-rep}[|\mathsf{nd}'_{suc}|]$ and $\mathsf{nd}'_{suc}.\mathsf{cs}[|\mathsf{nd}'_{suc}|] = \mathsf{nd}'_{suc,t-rep}.\mathsf{cs}[|\mathsf{nd}'_{suc}|]$.
\item Let $\mathsf{nd}'_{suc,rep}$ be in a transitive replaces-relation with $\mathsf{nd}'_{suc}$. If \textsc{prune} is called given a witness of redundancy of $\mathsf{nd}'_{suc,rep}$, then some replacement node of $\mathsf{nd}'_{suc,rep}$ is found. If only one replacement node of $\mathsf{nd}'_{suc,rep}$ is found, then this replacement node is de-facto non-redundant w.r.t.\ $DPI$.
%
%if $\mathsf{nd}'_{suc}$ is redundant w.r.t.\ some DPI $DPI''$ that is the current DPI of any call to \textsc{dynamicHS} and is either equal to $DPI$ or includes fewer test cases than $DPI$, then 
%$\mathsf{nd}\in\Queue_{dup}$ and 
%
%constructed from $\mathsf{nd}$ which is non-redundant w.r.t.\ $DPI''$. 
\end{enumerate}
\end{lemma}
\begin{proof}
Since $\mathsf{nd}\in\Queue$ holds at some point in time during the execution of some call to \textsc{dynamicHS} with current DPI $DPI'_{prev}$ and since the execution of \textsc{dynamicHS} with $DPI$ terminates due to $\Queue = []$, we have that some node set-equal to $\mathsf{nd}$ must be processed. This must be satisfied because nodes can only be deleted from $\Queue$ in that they are processed or pruned, and $\mathsf{nd}$ cannot be pruned from $\Queue$. For, by precondition, if \textsc{prune} is called given a witness of redundancy of $\mathsf{nd}$, then a replacement node of $\mathsf{nd}$ is found. And, if only one replacement node $\mathsf{nd}_{rep}$ of $\mathsf{nd}$ is found, $\mathsf{nd}_{rep}$ is de-facto non-redundant w.r.t.\ $DPI$.

Now, let $\mathsf{nd}_1$ be a replacement node of $\mathsf{nd}$ found by \textsc{prune} called with some witness of redundancy of $\mathsf{nd}$. Then, by precondition, what holds for $\mathsf{nd}$ also holds for $\mathsf{nd}_1$. That is, if \textsc{prune} is called given a witness of redundancy of $\mathsf{nd}_1$, then a replacement node of $\mathsf{nd}_1$ is found. And, if only one replacement node $\mathsf{nd}_{1,rep}$ of $\mathsf{nd}_1$ is found, $\mathsf{nd}_{1,rep}$ is de-facto non-redundant w.r.t.\ $DPI$.

The same holds for \emph{any} $\mathsf{nd}_i$ which is in a transitive replaces-relation with $\mathsf{nd}$. So, anytime \textsc{prune} is called for a node set-equal to $\mathsf{nd}$, at least one replacement node is found by \textsc{prune}. And, in case $\mathsf{nd}_i$ is de-facto non-redundant w.r.t.\ $DPI$ -- which must be the case sooner or later for some node in a transitive replaces-relation with $\mathsf{nd}$, by the given preconditions -- then, by Proposition~\ref{prop:de-facto_non_redundant_node_cannot_be_pruned_or_replaced}, $\mathsf{nd}_i$ cannot be pruned or replaced. 
Hence, let us denote by $\mathsf{node}$ the node set-equal to $\mathsf{nd}$ that is finally processed. Let $DPI_{prev}$ now be the ``current'' DPI of the execution of \textsc{dynamicHS} during which $\mathsf{node}$ is processed. Further, we denote the DPI of the immediate subsequent execution of \textsc{dynamicHS} by $DPI_{prev+1}$, and so on.

%Then, we have two cases, namely $\mathsf{node}$ is either ($\lnot$r)~non-redundant or (r)~redundant w.r.t.\ $DPI$.

%Case~($\lnot$r): There are two more cases, namely (g)~$\mathsf{node}\in GenNodes$ and ($\lnot$g)~$\mathsf{node}\notin GenNodes$. In the former case, i.e.\ (g), the preconditions of Lemma~\ref{lem:non-redundant_node_has_suc_node_subset_min_diag} are satisfied wherefore the proposition of this lemma follows.

%Case~($\lnot$g): In this case, since $\mathsf{node}$ is processed, it is either (s)~labeled by a set (\textsc{dLabel} returns in line~\ref{algoline:dlabel:return_X}, \ref{algoline:dlabel:return_new_cs} or \ref{algoline:dlabel:return_C}) or ($\lnot$s)~not labeled by a set (\textsc{dLabel} returns in line~\ref{algoline:dlabel:non-min_crit_end} or \ref{algoline:dlabel:return_valid}). 

%Case~($\lnot$r): 
Since $\mathsf{node}$ is processed, it is either 
\begin{itemize}
	\item (s)~labeled by a set (\textsc{dLabel} returns in line~\ref{algoline:dlabel:return_X}, \ref{algoline:dlabel:return_new_cs} or \ref{algoline:dlabel:return_C}) or 
	\item ($\lnot$s)~not labeled by a set (\textsc{dLabel} returns in line~\ref{algoline:dlabel:non-min_crit_end} or \ref{algoline:dlabel:return_valid}).
\end{itemize}

%\textbf{Case~($\lnot$s):} 
\paragraph{Case~($\lnot$s):}
In this case, \textsc{dLabel} returns 
\begin{itemize}
	\item (i)~$nonmin$ or
	\item (ii)~$valid$.
\end{itemize}  
%By Lemma~\ref{lem:if_dlabel_returns_valid_nonmin_then}, this is a contradiction to the assumption that $\md$ is a minimal diagnosis w.r.t.\ the current DPI as both outputs $nonmin$ and $valid$ imply the opposite.

%\textbf{In (i)}, 
\paragraph{Case~(i):}
By Lemma~\ref{lem:if_dlabel_returns_valid_nonmin_then}, $\mathsf{node}$ must be a non-minimal diagnosis w.r.t.\ $DPI_{prev}$. By line~\ref{algoline:dyn:add_to_Dsupset}, $\mathsf{node}$ is then added to the set $\mD_{\supset}$. $\mD_{\supset}$ is never modified throughout Algorithm~\ref{algo:inter_onto_debug} and is given as an input argument to each subsequent call to \textsc{dynamicHS} by line~\ref{algoline:inter_onto_debug:dynamicHS} in Algorithm~\ref{algo:inter_onto_debug}. During the execution of some subsequent call to \textsc{dynamicHS} using the DPI $DPI_{prev+i}$ for $i \geq 1$, the set $\mD_{\supset}$ might be modified by the \textsc{prune} function called during \textsc{updateTree} (line~\ref{algoline:update:call_prune_Dsupset} and lines~\ref{algoline:update:process_Dsupset_start}-\ref{algoline:update:process_Dsupset_end}) or during \textsc{dLabel} (line~\ref{algoline:dlabel:call_prune_Dsupset}). 

Recall that $\mathsf{node}$ is either the same node as $\mathsf{nd}$ or in a transitive replaces-relation with $\mathsf{nd}$. 
%And, since the execution of \textsc{dynamicHS} with current DPI $DPI$ terminates due to $\Queue = \emptyset$
Hence, by the argumentation given before, we have that, if \textsc{prune} is called given a witness of redundancy of $\mathsf{node}$, then there is a replacement node of $\mathsf{node}$ found by \textsc{prune}. And, if there is only one replacement node of $\mathsf{node}$ found by \textsc{prune}, then this replacement node is de-facto non-redundant w.r.t.\ $DPI$.  
%
%By precondition, if \textsc{prune} is called given a witness of redundancy of $\mathsf{nd}$, then there is a replacement node of $\mathsf{nd}$ found by \textsc{prune}. And, if there is only one replacement node of $\mathsf{nd}$ found by \textsc{prune}, then this replacement node is de-facto non-redundant w.r.t.\ $DPI$. Further, $\mathsf{node} = \mathsf{nd}$ must hold by the definition of $\mathsf{node}$. Hence, the found replacement node of $\mathsf{nd}$ which is de-facto non-redundant w.r.t.\ $DPI$ is also a replacement node of $\mathsf{node}$. By Lemma~\ref{lem:redundant_node_remains_redundant_after_adding_new_testcases}, as argued above, this replacement node is also non-redundant w.r.t.\ $DPI_{prev+i}$. 

Therefore, $\mathsf{node} \in \mD_{\supset}$ cannot be pruned, i.e.\ $\mathsf{node}$ considered as a set cannot be deleted from $\mD_{\supset}$ in line~\ref{algoline:update:call_prune_Dsupset} or line~\ref{algoline:dlabel:call_prune_Dsupset}. 
%The truth of this is supported by Corollary~\ref{cor:prune_only_called_with_min_cs_in_dlabel} and Lemmata~\ref{lem:quick_prune_check} and \ref{lem:complete_prune_check} which say that \textsc{prune} can only be called given some minimal conflict set $X$ w.r.t.\ $DPI_{prev+i}$. 
So, after any number of calls to \textsc{prune}, we have that either $\mathsf{node} \in \mD_{\supset}$ or, otherwise, there is some node in $\mD_{\supset}$ which is set-equal to $\mathsf{node}$ and which is in a transitive replaces-relation with $\mathsf{node}$. We keep calling this (possibly replacement) node $\mathsf{node}$ in the following.

By Lemma~\ref{lem:if_dlabel_returns_valid_nonmin_then}, at the time $\mathsf{node}$ was processed, there must be some diagnosis $\md'$ w.r.t.\ $DPI_{prev}$ such that $\md'\in \mD_{calc}$ and $\mathsf{node} \supset \md'$. 
Additionally, by Lemma~\ref{lem:if_dlabel_returns_valid_nonmin_then}, the set $\mD_{calc}$ computed during \textsc{dynamicHS} for some ``current'' DPI $DPI_j$ comprises only diagnoses w.r.t.\ $DPI_j$. 
Now, we have $\mathsf{node} \subset \md$ since $\mathsf{nd} \subset \md$ and $\mathsf{node} = \mathsf{nd}$, and $\md'\subset\mathsf{node}$. That is, $\md'\subset\md$. By the precondition that $\md$ is a minimal diagnosis w.r.t.\ $DPI$, $\md'$ cannot be a diagnosis w.r.t.\ $DPI$. Thus, there cannot be any such $\md'$ in $\mD_{calc}$ computed during \textsc{dynamicHS} for $DPI$.

All nodes in $\mD_{calc}$ returned by some call to \textsc{dynamicHS} using DPI $DPI_1$ that are no diagnoses w.r.t.\ $DPI_2$, the extension of $DPI_1$ by a new query added as a positive or negative test case, are added to the set $\mD_{\times}$ (and not to $\mD_{\checkmark}$) in line~\ref{algoline:inter_onto_debug:update_D_times} of Algorithm~\ref{algo:inter_onto_debug} and are thus no elements of the set $\mD_{\checkmark}$ given as an argument to \textsc{dynamicHS} at the next call to \textsc{dynamicHS}. 
The elements of $\mD_{\checkmark}$ given as an argument to \textsc{dynamicHS} at the next call to \textsc{dynamicHS} using $DPI_2$ are definitely added to $\Queue$ again in lines~\ref{algoline:update:process_Dcheckmark_start}-\ref{algoline:update:process_Dcheckmark_end} as $\mD_{\checkmark}$ is not modified elsewhere in \textsc{dynamicHS} before lines~\ref{algoline:update:process_Dcheckmark_start}-\ref{algoline:update:process_Dcheckmark_end} are reached. 

Therefore, we need to differentiate between two cases: Either 
\begin{itemize}
	\item (x1)~$\md'\in\mD_{\times}$ never holds for the input argument $\mD_{\times}$ to any call to \textsc{dynamicHS} or
	\item (x2)~$\md'\in\mD_{\times}$ holds at least once for the input argument $\mD_{\times}$ to some call to \textsc{dynamicHS}.
\end{itemize}
%one possible case involves that $\md'\in\mD_{\times}$ never holds for $\md'$. 

%\textbf{Case~(x1):} 
\paragraph{Case~(x1):}
Since $\md'\in\mD_{calc}$ holds after the execution of \textsc{dynamicHS} using $DPI_{prev}$ stops, we have that $\md'\in\mD_{\checkmark}$ must hold for the argument $\mD_{\checkmark}$ given to \textsc{dynamicHS} using $DPI_{prev+1}$. After \textsc{updateTree} returns during \textsc{dynamicHS} using $DPI_{prev+1}$, $\md' \in \Queue$ holds as argued. Subsequently, $\md'$ might be added again to $\mD_{calc}$ and then to $\mD_{\checkmark}$ again in line~\ref{algoline:inter_onto_debug:update_D_checkmark} of Algorithm~\ref{algo:inter_onto_debug} and to $\Queue$ again in line~\ref{algoline:update:insert_sorted_1} during \textsc{dynamicHS} using $DPI_{prev+2}$, and so forth. But, when a test case is added to some DPI $DPI_{prev+i}$ in Algorithm~\ref{algo:inter_onto_debug} that invalidates the diagnosis $\md'$ (yielding the DPI $DPI_{prev+i+1}$), $\md' \notin \mD_{calc}$ is assumed to hold (otherwise it would be an element of $\mD_{\times}$ against our assumption). Such a test case must be added sometime as argued above. By Proposition~\ref{prop:diag_for_new_dpi_is_diag_for_old_dpi}, $\md'$ cannot be a (minimal) diagnosis w.r.t.\ any DPI including more test cases than $DPI_{prev+i+1}$ either.
Notice that the case $\md' \notin \mD_{calc}$ can emerge in spite of the fact that $\md'$ is a minimal diagnosis w.r.t.\ $DPI_{prev+i}$ because there may be minimal diagnoses w.r.t.\ $DPI_{prev+i}$ that have a higher probability than $\md'$. For $DPI_{prev+i+1}$ and all DPIs including more test cases than $DPI_{prev+i+1}$, $\md'$ cannot be added to $\mD_{calc}$ anymore due to Lemma~\ref{lem:if_dlabel_returns_valid_nonmin_then} which claims that only diagnoses w.r.t.\ the currently used DPI can be added to $\mD_{calc}$.
\paragraph{Case~(x2):}
Here, $\md'\in\mD_{\times}$ holds at least once for the input argument $\mD_{\times}$ to some call to \textsc{dynamicHS} using the DPI $DPI_{prev+i}$. Then, \textsc{dynamicHS} using the DPI $DPI_{prev+i-1}$ must have returned a set $\mD_{calc}$ including $\md'$ as otherwise $\md'$ cannot be added to $\mD_{\times}$. Hence, $\md'$ must be a diagnosis w.r.t.\ $DPI_{prev+i-1}$ by Lemma~\ref{lem:if_dlabel_returns_valid_nonmin_then}. Since $\md'$ is added to $\mD_{\times}$, it cannot be a diagnosis w.r.t.\ $DPI_{prev+i}$. This must hold 
\begin{itemize}
	\item by Remark~\ref{rem:invalidated_sets_of_q-partition_for_query_answer},
	\item since the set added to $\mD_{\times}$ in Algorithm~\ref{algo:inter_onto_debug} is exactly the set $\mD_{out}$ returned by \textsc{getInvalidDiags} in line~\ref{algoline:inter_onto_debug:get_invalid_diags} of Algorithm~\ref{algo:inter_onto_debug} and
	\item $\mD_{out} = \dx{}(Q)$ in case the user answer $u(Q)$ to the query $Q$ w.r.t.\ $\mD_{calc}$ and $DPI_{prev+i-1}$ is $\false$ and $\mD_{out}=\dnx{}(Q)$ otherwise (notice that $\mD_{calc}$ is referred to as $\mD_{\checkmark}$ in Algorithm~\ref{algo:inter_onto_debug}).
\end{itemize}
So, by Proposition~\ref{prop:diag_for_new_dpi_is_diag_for_old_dpi}, $\md'$ cannot be a (minimal) diagnosis w.r.t.\ any DPI including more test cases than $DPI_{prev+i}$ either.

%%%%%%%%%%%%%%%%%%%%%%% OLD start %%%%%%%%%%%%%%%%%%%%%% 
%Each element in $\mD_{\times}$ is processed by the \textsc{updateTree} function called for the DPI $DPI_{prev+i}$. Then, either $\md'$ is redundant w.r.t.\ $DPI_{prev+i}$ and \textsc{prune} in line~\ref{algoline:update:call_prune_Dtimes} is called given some witness of redundancy $X$ of $\md'$ or the opposite holds. In the former case, by Lemma~\ref{lem:prune} and Corollary~\ref{cor:prune_only_called_with_min_cs_in_dlabel}, $\md'$ is pruned. 
%%after which it is not an element of $\mD_{\times}$ anymore. 
%In the latter case, $\md'$ is reinserted into $\Queue$ in line~\ref{algoline:update:insert_sorted_0}.
%% and deleted from $\mD_{\times}$.
%Thus, in both cases $\md' \notin \mD_{calc}$ holds. Moreover, $\md'$ cannot ever be readded to $\mD_{calc}$ by Lemma~\ref{lem:if_dlabel_returns_valid_nonmin_then} and since $\md'$ is not a diagnosis w.r.t\ any DPI including more test cases than $DPI_{prev+i}$.
%%%%%%%%%%%%%%%%%%%%%%% OLD end %%%%%%%%%%%%%%%%%%%%%%
%%%%%%%%%%%%%%%%%%%%%% NEW start %%%%%%%%%%%%%%%%%%%%%% 
Each element in $\mD_{\times}$ is processed by the \textsc{updateTree} function (lines~\ref{algoline:update:process_Dtimes_start}-\ref{algoline:update:process_Dtimes_end}) called for the DPI $DPI_{prev+i}$. In lines~\ref{algoline:update:process_Dtimes_start}-\ref{algoline:update:process_Dtimes_end}, each node $\mathsf{ndx}$ in $\mD_{\times}$ can only be pruned or either $\mathsf{ndx}$ or a node in a transitive replaces-relation with $\mathsf{ndx}$ is added to $\Queue$ in line~\ref{algoline:update:insert_sorted_0}.
$\mD_{calc}$ is not modified by \textsc{updateTree} and $\mD_{calc} = \emptyset$ holds at the beginning of the execution of each call to \textsc{dynamicHS}. (A node set-equal to) $\md'$ cannot ever be readded to $\mD_{calc}$ by Lemma~\ref{lem:if_dlabel_returns_valid_nonmin_then} and since $\md'$ is not a diagnosis w.r.t\ any DPI including more test cases than $DPI_{prev+i}$. Hence, $\md'\in\mD_{calc}$ can never hold for any DPI including more test cases than $DPI_{prev+i}$.
%%%%%%%%%%%%%%%%%%%%%% NEW end %%%%%%%%%%%%%%%%%%%%%%

Hence, there must be some DPI $DPI_{prev+k}$ such that $\mD_{\checkmark}$ given as input to the \textsc{dynamicHS}-call for $DPI_{prev+k}$ does not include \emph{any} diagnosis $\md'\subset \mathsf{node}$.
%which is a diagnosis w.r.t.\ $DPI_{prev+k}$,  but is not a diagnosis w.r.t.\ $DPI_{prev+k+1}$. 
So, during the execution of the call to \textsc{dynamicHS} using DPI $DPI_{prev+k}$, $\mathsf{node}$ must be deleted from $\mD_{\supset}$ and be reinserted into $\Queue$ by lines~\ref{algoline:update:process_Dsupset_start}-\ref{algoline:update:process_Dsupset_end} in \textsc{updateTree} which is called at the beginning of the execution of \textsc{dynamicHS} at any call to \textsc{dynamicHS}. This must hold since all nodes $\mathsf{ndx}$ in $\mD_{\supset}$ that have not yet been pruned and for which there is no diagnosis in $\mD_{\checkmark}$ which is a proper subset of $\mathsf{ndx}$, are added to $\Queue$ throughout lines~\ref{algoline:update:process_Dsupset_start}-\ref{algoline:update:process_Dsupset_end}. As shown, both criteria are met for $\mathsf{node}$ during the execution of the call to \textsc{dynamicHS} using DPI $DPI_{prev+k}$.

%\textbf{In (ii)}, 
\paragraph{Case~(ii):}
By Lemma~\ref{lem:if_dlabel_returns_valid_nonmin_then}, we know that $\mathsf{node}$ is a diagnosis w.r.t.\ $DPI_{prev}$ and that $\mathsf{node}$ is added to $\mD_{calc}$. Since $\mathsf{node} \subset \md$ and $\md$ is a minimal diagnosis w.r.t.\ $DPI$, we obtain, by the same argumentation as in (i), 
%there must be some DPIs $DPI_{prev+k}$ and $DPI_{prev+k+1}$ such that $\mD_{\checkmark}$ given as input to the \textsc{dynamicHS}-call for $DPI_{prev+k}$ includes at least one diagnosis $\mathsf{node}$, but at the \textsc{dynamicHS}-call for $DPI_{prev+k+1}$, $\mD_{\times}$ (and not $\mD_{\checkmark}$) includes $\mathsf{node}$.
that there must be some DPI $DPI_{prev+k}$ such that $\mD_{\checkmark}$ given as input to the \textsc{dynamicHS}-call for $DPI_{prev+k}$ does not include $\mathsf{node}$.

If $\mathsf{node} \notin \mD_{\times}$, then it cannot ever be added to $\mD_{calc}$ again, as argued in case (i). Otherwise, during the execution of \textsc{updateTree} which is called at the beginning of the execution of each call to \textsc{dynamicHS}, $\mD_{\times}$ is modified in lines~\ref{algoline:update:process_Dtimes_start}-\ref{algoline:update:process_Dtimes_end}.

Now, we differentiate between two cases, namely $\mathsf{node}$ is either 
\begin{itemize}
	\item ($\lnot$r)~non-redundant w.r.t.\ $DPI$ or
	\item (r)~redundant w.r.t.\ $DPI$.
\end{itemize}

%\textbf{Case~($\lnot$r):} 
\paragraph{Case~($\lnot$r):}
Due to the non-redundancy of $\mathsf{node}$ w.r.t.\ $DPI$, Lemma~\ref{lem:redundant_node_remains_redundant_after_adding_new_testcases}, Lemma~\ref{lem:prune} and Corollary~\ref{cor:prune_only_called_with_min_cs_in_dlabel}, $\mathsf{node}$ cannot be replaced or pruned throughout lines~\ref{algoline:update:process_Dtimes_start}-\ref{algoline:update:add_set_del_supset}. Thus, $\mathsf{node}$ is reinserted into $\Queue$ in line~\ref{algoline:update:insert_sorted_0}.

%\textbf{Case~(r):} 
\paragraph{Case~(r):}
Since $\mathsf{node}$ is redundant w.r.t.\ $DPI$, it may or may not be redundant w.r.t.\ $DPI_{prev+k+1}$. So, during the \textsc{updateTree} function called in \textsc{dynamicHS} for $DPI_{prev+k+1}$, there may or may not be some call to \textsc{prune} given some $X$ as argument which is a witness of redundancy of $\mathsf{node}$. In the latter case, $\mathsf{node}$ will not be replaced or pruned during any \textsc{prune} execution and will be reinserted into $\Queue$ in line~\ref{algoline:update:insert_sorted_0}. In the former case, $\mathsf{node}$ might be replaced, but it cannot be pruned due to the same reasoning as given in the second paragraph of case (i). 
So, either $\mathsf{node}$ or some node in a transitive replaces-relation with $\mathsf{node}$ must be in $\mD_{\times}$ at the time line~\ref{algoline:update:reinsert_D_of_Dx_to_Q} is reached. This node is then added to $\Queue$ in line~\ref{algoline:update:insert_sorted_0}. 
%Let us refer to this node also by $\mathsf{node}$.
%since, by assumption, there if \textsc{prune} is called given a witness of redundancy of $\mathsf{nd}$, then some replacement node of $\mathsf{nd}$ is found which is non-redundant w.r.t.\ $DPI$

Now, both cases (i) and (ii) identified for case ($\lnot$s) lead to the reinsertion of $\mathsf{node}$ or some node in a transitive replaces-relation with $\mathsf{node}$ -- which is thus set-equal to $\mathsf{nd}$ -- into $\Queue$. Notice that this node has the same properties as $\mathsf{node}$ before one of the cases (i) or (ii) emerged (by analogue reasoning as conducted above). That is, 
%it is either non-redundant w.r.t.\ $DPI$ or, 
if \textsc{prune} is called given a witness of redundancy of $\mathsf{node}$, then a replacement node of $\mathsf{node}$ is found. And, if only one replacement node of $\mathsf{node}$ is found, this replacement node is de-facto non-redundant w.r.t.\ $DPI$. So, we call this reinserted node again $\mathsf{node}$.

Furthermore, $\mathsf{node}$ can be neither labeled by $valid$ nor by $nonmin$ during the execution of \textsc{dynamicHS} for $DPI$. This holds by Lemma~\ref{lem:if_dlabel_returns_valid_nonmin_then} and since $\mathsf{node}$ can be neither a diagnosis nor a non-minimal diagnosis w.r.t.\ $DPI$ due to $\mathsf{node} \subset \md$ and the fact that $\md$ is a minimal diagnosis w.r.t.\ $DPI$. As a consequence of this and the assumption that the \textsc{dynamicHS}-call for $DPI$ terminates due to $\Queue = []$, case~(s) must arise at some point in time for $\mathsf{node}$ during some execution of \textsc{dynamicHS} for some (previous) DPI not-necessarily equal to $DPI$.

%\textbf{Case~(s):} 
\paragraph{Case~(s):}
In this case, by 
%the second proposition of 
Lemma~\ref{lem:Ccalc_in_dlabel}, \textsc{dLabel} returns a minimal conflict set $L$ w.r.t.\ $DPI_{prev}$ as a label for $\mathsf{node}$ where $L$ has the property that $L \cap \mathsf{node} = \emptyset$.

It must hold that $L \neq \emptyset$. Otherwise, by Proposition~\ref{prop:cs_admissible}, either 
\begin{itemize}
	\item (v1)~$\mo$ is valid w.r.t.\ $\tuple{\cdot,\mb,\Tp_{prev},\Tn_{prev}}_\RQ$ where $DPI_{prev} = \tuple{\mo,\mb,\Tp_{prev},\Tn_{prev}}_\RQ$ or
	\item (v2)~$DPI_{prev}$ is non-admissible.
\end{itemize}

In the former \textbf{case (v1)}, we know by Corollary~\ref{cor:notions_equiv} that the only (minimal) diagnosis w.r.t.\ $DPI_{prev}$ is $\emptyset$. If $DPI_{prev}$ is equal to $DPI$, this is a contradiction to the existence of some minimal diagnosis w.r.t.\ $DPI$, namely $\md$, which is not the empty set. $\md \supset \emptyset$ must hold since, by precondition, there is a node $\mathsf{nd}$ such that $\mathsf{nd} \subset \md$ and since $\emptyset \subseteq \mathsf{nd}$. 

Otherwise, if $DPI_{prev}$ includes a proper subset of the test cases $DPI$ includes, $DPI$ can never be a current DPI during any execution of \textsc{dynamicHS} during the same execution of Algorithm~\ref{algo:inter_onto_debug} during which there is an execution of \textsc{dynamicHS} where $DPI_{prev}$ is the current DPI. This holds as there must be at least two diagnoses in $\mD_{\checkmark}$ in line~\ref{algoline:inter_onto_debug:stop_crit} of Algorithm~\ref{algo:inter_onto_debug} in order for \textsc{dynamicHS} to be called again with a DPI including a proper superset of the test cases in $DPI_{prev}$ (notice that, in Algorithm~\ref{algo:inter_onto_debug}, the name of the set $\mD_{calc}$ returned by \textsc{dynamicHS} for $DPI_{prev}$ is $\mD_{\checkmark}$). For, in case there is only one diagnosis, i.e.\ $\emptyset$, then the probability of this diagnosis is 1 which is greater or equal $1 - \sigma$ for any choice of $\sigma$ due to $\sigma \geq 0$. Consequently, Algorithm~\ref{algo:inter_onto_debug} would return in line~\ref{algoline:inter_onto_debug:return}. This is a contradiction to the assumption that there is an execution of \textsc{dynamicHS} where $DPI$ is the current DPI. 

In the latter \textbf{case (v2)}, we can infer by Corollary~\ref{cor:query_leaves_valid_diag}, which states that adding queries as test cases to an admissible DPI can never yield a non-admissible DPI, that the DPI given as an input to Algorithm~\ref{algo:inter_onto_debug} must be non-admissible, contradiction. 
%
%Thence, $L \neq \emptyset$. 
%if \textsc{dLabel} returns in line~\ref{algoline:dlabel:add_new_cs}, by analogue argumentation as used in the proof of Lemma~\ref{lem:Ccalc_in_dlabel}, we can derive that \textsc{dLabel} returns a minimal conflict set w.r.t.\ the current DPI. If, on the other hand, \textsc{dLabel} returns in line~\ref{algoline:dlabel:return_C}, by Lemma~\ref{lem:Ccalc_in_dlabel} which states that $\mC_{calc}$ contains only minimal conflict sets w.r.t.\ the current or some previous DPI (including fewer test cases than the current one)
%

Thence, $L \neq \emptyset$ and \textsc{dynamicHS} will execute lines~\ref{algoline:dyn:for_e_in_L}-\ref{algoline:dyn:generate_nodes} and generate one node $\mathsf{node}_{e} :=$ $\textsc{add}(\mathsf{node}$, $e)$ with $\mathsf{node}_{e}.\mathsf{cs} := \textsc{add}(\mathsf{node.cs},L)$ for each $e\in L$ (cf.\ Definition~\ref{def:alternative_equal_node} for an explanation of the function \textsc{add}). 

Now, we have that there must be some non-empty active sublabel of $L = \mathsf{node}_{e}.\mathsf{cs}[r]$ w.r.t.\ $DPI$ where $r := |\mathsf{node}_{e}|$ by Definition~\ref{def:active_sublabel}. Definition~\ref{def:active_sublabel} is applicable by the following argumentation: 

The first observation is that $\mathsf{node}_{e}.\mathsf{cs}[r]$ cannot be reduced twice during one and the same execution of \textsc{dynamicHS} using one and the same DPI $DPI_{prev+j}$ which results from $DPI_{prev}$ by addition of test cases. For, by Corollaries~\ref{cor:prune_only_called_with_min_cs_in_dlabel} and \ref{cor:pruneQdup_only_called_with_min_cs_in_dlabel} and Lemmata~\ref{lem:quick_prune_check} and \ref{lem:complete_prune_check}, \textsc{prune} as well as \textsc{pruneQdup} can only be called given some minimal conflict set $X$ w.r.t.\ $DPI_{prev+j}$. By Lemmata~\ref{lem:prune} and \ref{lem:pruneQdup}, all nodes $\mathsf{ndx}$ that are in the set returned by \textsc{prune} and \textsc{pruneQdup}, respectively, have the property that there are no proper supersets of $X$ in $\mathsf{ndx.cs}$. Moreover, there are no proper subsets of $X$ in $\mathsf{ndx.cs}$. Because each $\mathsf{ndx.cs}[m]$ for $m \in \setof{1,\dots,|\mathsf{ndx.cs}|}$ must be a minimal conflict set w.r.t.\ some DPI equal to $DPI_{prev+j}$ or including a subset of the test cases in $DPI_{prev+j}$. Otherwise, $\mathsf{ndx}$ could not be a node during the execution of \textsc{dynamicHS} where $DPI_{prev+j}$ is the current DPI. By Proposition~\ref{prop:changes_in_conflict_sets_after_testcase_added}, there cannot be any $m \in \setof{1,\dots,|\mathsf{ndx.cs}|}$ such that $\mathsf{ndx.cs}[m] \subset X$ as $X$ is a minimal conflict set w.r.t.\ $DPI_{prev+j}$.
As two minimal conflict sets w.r.t.\ $DPI_{prev+j}$ can never be in a proper subset-relationship with one another, $L = \mathsf{node}_{e}.\mathsf{cs}[r]$ can be modified at most once by \textsc{prune} or \textsc{pruneQdup} for the DPI $DPI_{prev+j}$.

Second, by Proposition~\ref{prop:changes_in_conflict_sets_after_testcase_added}, each minimal conflict set w.r.t.\ $DPI_{prev}$ is a conflict set w.r.t.\ any DPI $DPI_{prev+j}$ that results from $DPI_{prev}$ by addition of test cases; that is, in particular, w.r.t.\ $DPI$. So, there must be some minimal conflict set $\mc_j$ w.r.t.\ each $DPI_{prev+j}$ such that $\mc_j \subseteq L$ and there cannot be any minimal conflict set w.r.t.\ $DPI_{prev+j}$ that is a proper superset of $L$. 

Third, we have that $L \neq \emptyset$, $L$ is a minimal conflict set w.r.t.\ $DPI_{prev}$, and $DPI_{prev+j}$ includes a superset of the test cases in $DPI_{prev}$. Thus, by Proposition~\ref{prop:if_DPI_cs_neq_emptyset_then_DPI+1_cs_neq_emptyset}, each minimal conflict set w.r.t.\ $DPI_{prev+j}$ must be non-empty. In particular, Proposition~\ref{prop:if_DPI_cs_neq_emptyset_then_DPI+1_cs_neq_emptyset} implies that all minimal conflict sets w.r.t.\ $DPI$ that are subsets of $L$ must be non-empty.

By these three observations, the criteria 
%for an active sublabel w.r.t.\ $DPI$ given by 
of Definition~\ref{def:active_sublabel} can be applied to analyze the active subnode of $\mathsf{node}_{e}.\mathsf{cs}[r]$  w.r.t.\ $DPI$. 
That is, if $\mc_1,\dots,\mc_n$ is the (arbitrary actual) chronological sequence of all sets $X$ given as an argument to \textsc{prune} and \textsc{pruneQdup} during all executions of \textsc{dynamicHS} from the one with current DPI $DPI_{prev}$ up to and including the one with current DPI $DPI$ where 
\begin{itemize}
		\item $\mathsf{node}_{e}.\mathsf{cs}[r] \supset \mc_1$,
		\item each $\mc_i$ is a minimal conflict set w.r.t.\ $DPI_i$ for $i \in \setof{1,\dots,n}$
		\item $\mc_k \supset \mc_{k+1}$ for $k \in \setof{1,\dots,n-1}$,
		\item $DPI_j$ includes a proper subset of the test cases $DPI_{j+1}$ includes for $j \in \setof{1,\dots,n-1}$,
		\item $DPI_n$ is equal to $DPI$ or includes a proper subset of the test cases $DPI$ includes and 
		\item $DPI_{prev}$ includes a proper subset of the test cases $DPI_1$ includes,
	\end{itemize}
then $\mc_n$ is the active sublabel of $\mathsf{node}_{e}.\mathsf{cs}[r]$ w.r.t.\ $DPI$. However, as argued before, the minimal conflict set $\mc_n$ w.r.t.\ $DPI_n$ cannot be the empty set. As a consequence, we obtain that there must be a non-empty active sublabel of $\mathsf{node}_{e}.\mathsf{cs}[r]$ w.r.t.\ $DPI$. 

By Propositions~\ref{prop:changes_in_conflict_sets_after_testcase_added} and \ref{prop:if_DPI_cs_neq_emptyset_then_DPI+1_cs_neq_emptyset}, there is a non-empty minimal conflict set $\mc'$ w.r.t.\ $DPI$ such that $\mc' \subseteq \mc_n$. Due to $\mc_n \subset \dots \subset \mc_1 \subset \mathsf{node}_{e}.\mathsf{cs}[r] = L$ we conclude that $\mc_n \subset L$. Therefore, $\emptyset \subset \mc' \subset L$ holds.

By Proposition~\ref{prop:mindiag_mincs}, each minimal diagnosis w.r.t.\ $DPI$ is a minimal hitting set of all minimal conflict sets w.r.t.\ $DPI$. Thence, we have that 
%$\md \cap \mc \neq \emptyset$ for each minimal conflict set $\mc$ w.r.t.\ the current DPI. As $\mc'$ is a minimal conflict set w.r.t.\ the current DPI, it must hold that 
$\mc' \cap \md \neq \emptyset$. 
%So, by $\mc' \subset L$, we have that $L \cap \md \neq \emptyset$. 
So, by $\mc' \subset L$, we have that $\emptyset \subset \mc' \cap \md \subseteq L \cap \md \subseteq L$. 
Consequently, we define $\mathsf{nd}_{suc} := \mathsf{node}_{x} = \textsc{add}(\mathsf{node},x)$ with $\mathsf{nd}_{suc}.\mathsf{cs} := \mathsf{node}_{x}.\mathsf{cs} =\textsc{add}(\mathsf{node.cs},L)$ for some $x \in \mc' \cap \md \subseteq L$. Then, $\mathsf{nd}_{suc} \subseteq \md$ because $\mathsf{node} \subset \md$ and $x \in \md$. It is clear from the inference so far that $\mathsf{nd} \subset \mathsf{nd}_{suc}$, $|\mathsf{nd}_{suc}| = |\mathsf{nd}| + 1$ and $\mathsf{nd}_{suc} \in GenNodes$. This shows the truth of propositions (1)-(3).

Proposition~(4) must hold by lines~\ref{algoline:dyn:check_node_already_in_Q}-\ref{algoline:dyn:generate_nodes}.

Now we argue why propositions~(5) and (6) must hold. Assume that $\mathsf{nd}'_{suc} \in \Queue$ is redundant w.r.t.\ some DPI $DPI''_{prev}$ which is equal to $DPI$ or includes fewer test cases than $DPI$.  
%Since $\mathsf{nd}'_{suc}$ cannot become redundant before it has been generated, $DPI''$ must include at least as many test cases as the DPI which was the ``current'' DPI at the time $\mathsf{nd}_{suc}$ was generated. Since $\mathsf{nd}_{suc}$ is redundant, 
Then, there must be some minimal conflict set $\mc''$ w.r.t.\ $DPI''_{prev}$ which is a witness of redundancy of $\mathsf{nd}'_{suc}$. Suppose that \textsc{prune} is called given $X := \mc''$ as an argument.
%By Proposition~\ref{prop:changes_in_conflict_sets_after_testcase_added}, $\mc''$ cannot be a proper superset of any conflict set in $\mathsf{nd}_{suc}.\mathsf{cs}$.

Now, we have to distinguish two cases: Either 
\begin{itemize}
	\item (q1)~$\mathsf{nd}_{suc}$ was added to $\Queue$ after it was generated or 
	\item (q2)~$\mathsf{nd}_{suc}$ was added to $\Queue_{dup}$ after it was generated
\end{itemize}
(there are no other possibilities, see lines~\ref{algoline:dyn:for_e_in_L}-\ref{algoline:dyn:generate_nodes}). 

For each of these two cases, there are two more cases to discriminate between: 
\begin{itemize}
	\item (c1)~$\mc'' \subset \mathsf{nd}'_{suc}.\mathsf{cs}[|\mathsf{nd}'_{suc}|]$ and   
$\mathsf{nd}'_{suc}[|\mathsf{nd}'_{suc}|] \in \mathsf{nd}'_{suc}.\mathsf{cs}[|\mathsf{nd}'_{suc}|] \setminus \mc''$ or
	\item (c2)~$\mc'' \subset \mathsf{nd}'_{suc}.\mathsf{cs}[j]$ and 
$\mathsf{nd}'_{suc}[j] \in \mathsf{nd}'_{suc}.\mathsf{cs}[j] \setminus \mc''$ for some $j\in\setof{1,\dots,|\mathsf{nd}'_{suc}|-1}$. 
\end{itemize}
 
%\textbf{Case~(q1):} 
\paragraph{Case~(q1):}
Here, we have that $\mathsf{nd}'_{suc}$ is the same node as $\mathsf{nd}_{suc}$ since $\mathsf{nd}_{suc}$ was added to $\Queue$ after generation and no node replacement can have taken place because $\mathsf{nd}'_{suc}$ is defined as the node set-equal to $\mathsf{nd}_{suc}$ that is an element of $\Queue$ \emph{immediately} after $\mathsf{nd}_{suc}$ has been generated. And, only one node corresponding to one and the same set can be in $\Queue$ at the same time. 

%\textbf{In (c1)}, 
\paragraph{Case~(c1):}
We have that $\mc''$ must be equal to some minimal conflict set $\mc_j$ in the sequence $\mc_1,\dots,\mc_n$. This must be \true since, first, $DPI''_{prev}$ is equal to $DPI$ or includes a subset of the test cases in $DPI$ and $DPI_{prev}$ includes a proper subset of the test cases in $DPI''_{prev}$.
 
To understand why the latter must hold, recall that $DPI_{prev}$ is the DPI of the call to \textsc{dynamicHS} where $\mathsf{nd}_{suc}$ was generated and the minimal conflict set $L$ was computed. By assumption, however, there is some minimal conflict set w.r.t.\ $DPI''_{prev}$, namely $\mc''$, such that $\mc'' \subset \mathsf{nd}'_{suc}.\mathsf{cs}[|\mathsf{nd}'_{suc}|] = L$. Hence, it cannot be \true that both $L$ and $\mc''$ are minimal conflict sets w.r.t.\ the same DPI. Otherwise, we would have a contradiction to the minimality of $L$. By Proposition~\ref{prop:changes_in_conflict_sets_after_testcase_added}, which states that minimal conflict sets cannot grow by the addition of new test cases to the DPI, we obtain the claimed fact that $DPI_{prev}$ includes a proper subset of the test cases in $DPI''_{prev}$.

Second, the sequence $\mc_1,\dots,\mc_n$ comprises \emph{all} sets $X$ given as an argument to \textsc{prune} and \textsc{pruneQdup} during \emph{all} executions of \textsc{dynamicHS} from the one with current DPI $DPI_{prev}$ up to and including the one with current DPI $DPI$ where $L = \mathsf{nd}'_{suc}.\mathsf{cs}[|\mathsf{nd}'_{suc}|] \supset \mc_1 \supset \dots \supset \mc_n$ holds. Reason for this to be valid is the fact that $\mathsf{nd}'_{suc}$ is the same node as $\mathsf{nd}_{suc}$ in the currently considered case~(q1).

Now, recall $\mc'$ is a minimal conflict set w.r.t.\ $DPI$ such that $x \in \mc' \cap \md \subset L$. Further, by $\mathsf{nd}'_{suc}=\mathsf{node}_x$, we have that $\mathsf{nd}'_{suc}[|\mathsf{nd}'_{suc}|] = x$.
Since $\mc' \subseteq \mc_n$, we have that $\mc' \subseteq \mc_j$ must hold due to $\mc_n \subseteq \mc_j$.
% \subset L$, 
Therefore, we can infer by $\mc'' = \mc_j$ that $\mc' \subseteq \mc''$
% \subset L$ 
is \true. 
Now, $x \in \mc'$ implies that $x \in \mc''$ wherefore $x \notin \mathsf{nd}'_{suc}.\mathsf{cs}[|\mathsf{nd}'_{suc}|] \setminus \mc''$. By $x = \mathsf{nd}'_{suc}[|\mathsf{nd}'_{suc}|]$, this is a contradiction to the assumption of case~(c1). Hence, case~(c2) must arise.

%\textbf{In (c2)}, 
\paragraph{Case~(c2):}
We have that $\mathsf{nd}'_{suc}[1..|\mathsf{nd}'_{suc}|-1]$ must be redundant w.r.t.\ $DPI''_{prev}$. The subnode $\mathsf{nd}'_{suc}[1..$ $|\mathsf{nd}'_{suc}|-1]$ of $\mathsf{nd}'_{suc}$ is the same node as $\mathsf{node}$ by $\mathsf{nd}'_{suc}=\mathsf{node}_x$. So, suppose \textsc{prune} is called with arguments $\Queue$ (which inlcudes $\mathsf{nd}'_{suc}$), $X:=\mc''$ and $\Queue_{dup}$ during the execution of \textsc{dynamicHS} with current DPI $DPI''_{prev}$. 

Recall that $\mathsf{node}$ is the node set-equal to $\mathsf{nd}$ that is processed. That is, $\mathsf{node}$ is either the same node as $\mathsf{nd}$ or it is in a transitive replaces-relation with $\mathsf{nd}$. Therefore, by the preconditions of this lemma, the following holds: If \textsc{prune} is called given a witness of redundancy of $\mathsf{node}$, then a replacement node of $\mathsf{node}$ is found. And, if only one replacement node $\mathsf{node}_{rep}$ of $\mathsf{node}$ is found, then $\mathsf{node}_{rep}$ is de-facto non-redundant w.r.t.\ $DPI$.

So, at the time \textsc{prune} might be called given a witness of redundancy of $\mathsf{node}$, $Comb(\Queue_{dup})$ must include a (non-necessarily proper) alternative subnode $\mathsf{node}_{rep,sub}$ of $\mathsf{node}$ from which the de-facto non-redundant node $\mathsf{node}_{rep}$ w.r.t.\ $DPI$ can be constructed as 
\begin{align*}
\mathsf{node}_{rep} &:= \textsc{add}(\mathsf{node}_{rep,sub}, \mathsf{node}[|\mathsf{node}_{rep,sub}|+1..|\mathsf{node}|]) \\
\mathsf{node}_{rep}.\mathsf{cs} &:= \textsc{add}(\mathsf{node}_{rep,sub}.\mathsf{cs},\mathsf{node}.\mathsf{cs}[|\mathsf{node}_{rep,sub}|+1..|\mathsf{node}|])
\end{align*}
%$\mathsf{node}_{rep} := \textsc{add}(\mathsf{node}_{rep,sub}$, $\mathsf{node}[|\mathsf{node}_{rep,sub}|+1..$ $|\mathsf{node}|])$ and $\mathsf{node}_{rep}.\mathsf{cs} := \textsc{add}(\mathsf{node}_{rep,sub}.\mathsf{cs},\mathsf{node}.\mathsf{cs}[|\mathsf{node}_{rep,sub}|+1..|\mathsf{node}|])$. 
This holds due to
\begin{itemize}
	\item Corollary~\ref{cor:pruneQdup_outputs_subset_of_Comb(Qdup)}, which says that each call to \textsc{pruneQdup} returns the list $\Queue_{dup}$, a subset of \newline \mbox{$Comb(\Queue_{dup})$},
	\item the fact that \textsc{pruneQdup} is always called immediately before \textsc{prune} is called and
	\item the fact that \textsc{prune} searches for alternative subnodes for the construction of a replacement node of a redundant node exactly in the output set of \textsc{pruneQdup}.
\end{itemize}
By Definition~\ref{def:de-facto_non-redundant}, this is implies that $\mathsf{node}_{rep,sub}$ must be de-facto non-redundant w.r.t.\ $DPI$ as otherwise the de-facto non-redundancy w.r.t.\ $DPI$ could not hold for $\mathsf{node}_{rep}$.

Consequently, by Lemma~\ref{lem:non-redundant_node_in_Comb(Qdup)_persists}, $\mathsf{node}_{rep,sub}\in Comb(\Queue_{dup})$ must always be satisfied during \emph{any} execution of \textsc{dynamicHS} using a DPI that is equal to $DPI$ or includes a subset of the test cases in $DPI$. Hence, in particular, this must hold for the DPI $DPI''_{prev}$.
%
 %and , we know that $\mathsf{nd}_{rep}$ must still be an element of $\Queue_{dup}$. This holds by Lemma~\ref{lem:pruneQdup} which says that only nodes can be deleted from $\Queue_{dup}$ in \textsc{pruneQdup} for which there is a witness of redundancy, the fact that \textsc{pruneQdup} is the only place in \textsc{dynamicHS} where nodes are deleted from $\Queue_{dup}$ and the fact that there cannot be a witness of redundancy of $\mathsf{nd}_{rep}$ w.r.t.\ any DPI equal to $DPI''_{prev}$ or including fewer test cases than $DPI''_{prev}$. The latter fact is a consequence of the non-redundancy of $\mathsf{nd}_{rep}$ w.r.t.\ $DPI$ and Lemma~\ref{lem:redundant_node_remains_redundant_after_adding_new_testcases}. 

By line~\ref{algoline:dyn:add_to_Qdup} and \textsc{pruneQdup}, which are the only places in \textsc{dynamicHS} where $\Queue_{dup}$ is modified, $\Queue_{dup}$ is sorted in ascending order by node cardinality at any time during the execution of any call to \textsc{dynamicHS}.	

In order to construct a replacement node of $\mathsf{nd}'_{suc}$, \textsc{prune} first determines the maximal $k$ such that $\mc'' \subset \mathsf{nd}'_{suc}.\mathsf{cs}[k]$ and $\mathsf{nd}'_{suc}[k] \in \mathsf{nd}'_{suc}.\mathsf{cs}[k] \setminus \mc''$. As case~(c1) was proven to be false, we conclude that $k \leq |\mathsf{nd}'_{suc}|-1$ must hold. Due to the fact that $\mathsf{nd}'_{suc}[1..|\mathsf{nd}'_{suc}|-1]$ is the same node as $\mathsf{node}$, as reasoned above, and the fact that a de-facto non-redundant alternative equal node $\mathsf{node}_{rep}$ (see above) of $\mathsf{node}$ can be constructed from $\mathsf{node}_{rep,sub} \in Comb(\Queue_{dup})$, we obtain that $k \leq |\mathsf{node}_{rep,sub}|$. This holds because the truth of both $\mathsf{node.cs}[m] \supset \mc''$ and $\mathsf{node}[m] \in \mathsf{node.cs}[m]\setminus\mc''$ for some $m \in \setof{|\mathsf{node}_{rep,sub}|+1,\dots,|\mathsf{node}|}$ would be a contradiction to the de-facto non-redundancy of $\mathsf{node}_{rep}$ w.r.t.\ $DPI$.  
%As $\mathsf{nd}'_{suc}[|\mathsf{nd}'_{suc}|] = x$ where $x \in \mc''$ by the same argumentation as in case~(c1), we have that $\mathsf{nd}'_{suc}[|\mathsf{nd}'_{suc}|] \notin \mathsf{nd}'_{suc}.\mathsf{cs}[|\mathsf{nd}'_{suc}|] \setminus \mc''$ wherefore $k \leq |\mathsf{nd}'_{suc}|-1$ must hold.

Then, in line~\ref{algoline:prune:check_for_alternative_paths_start}, an alternative subnode of $\mathsf{nd}'_{suc}$ 
\begin{itemize}
	\item which has cardinality $k + z$ where $z \geq 0$ is minimal and 
	\item from which a replacement node of $\mathsf{nd}'_{suc}$ can be constructed
\end{itemize}
is searched for in $\Queue_{dup}$. To see this, observe that elements in $\Queue_{dup}$ -- which is sorted in ascending order of node cardinality, as argued -- are visited in order starting from the lowest cardinality node (line~\ref{algoline:prune:check_for_alternative_paths_start}). 

%However, $\mathsf{nd}'_{suc,new} := \textsc{add}(\mathsf{node}_{rep,sub},\mathsf{node}_x[|\mathsf{node}_{rep,sub}|+1..|\mathsf{node}_x|]) = \textsc{add}(\mathsf{node}_{rep},x)$ with $\mathsf{nd}'_{suc,new}.\mathsf{cs} := \textsc{add}(\mathsf{node}_{rep,sub}.\mathsf{cs},\mathsf{node}_x.\mathsf{cs}[|\mathsf{node}_{rep,sub}|+1..|\mathsf{node}_x|]) = \textsc{add}(\mathsf{node}_{rep}.\mathsf{cs},L)$ is a definitely a replacement node that is detectable by \textsc{prune}.
However, there is an alternative subnode $\mathsf{node}_{rep,sub}$ of $\mathsf{node}$ such that $k \leq |\mathsf{node}_{rep,sub}| \leq |\mathsf{node}| = |\mathsf{nd}'_{suc}|-1$ and 
%$\mathsf{node}_{rep,sub}$ is de-facto non-redundant w.r.t.\ $DPI$ as shown.
$\mathsf{node}_{rep,sub}$ is an element of the argument $\Queue_{dup}$ given to \textsc{prune}, as shown above. 
As $\mathsf{nd}'_{suc}$ is the same node as $\mathsf{node}_x$, $\mathsf{node}$ is a subnode of $\mathsf{nd}'_{suc}$. Therefore, $\mathsf{node}_{rep,sub}$ is an alternative subnode of $\mathsf{nd}'_{suc}$. 

Thus, we have that one replacement node of $\mathsf{nd}'_{suc}$ is definitely found by \textsc{prune}. And, in case there is only one replacement node of $\mathsf{nd}'_{suc}$ constructable during \textsc{prune}, then this replacement node is given by $\mathsf{nd}'_{suc,new} := \textsc{add}(\mathsf{node}_{rep,sub},\mathsf{node}_x[|\mathsf{node}_{rep,sub}|+1..|\mathsf{node}_x|]) = \textsc{add}(\mathsf{node}_{rep},x)$ with $\mathsf{nd}'_{suc,new}.\mathsf{cs} := \textsc{add}(\mathsf{node}_{rep,sub}.\mathsf{cs},\mathsf{node}_x.\mathsf{cs}[|\mathsf{node}_{rep,sub}|+1..|\mathsf{node}_x|]) = \textsc{add}(\mathsf{node}_{rep}.\mathsf{cs},L)$. As it is straightforward from the deductions above, $\mathsf{nd}'_{suc,new}$ is de-facto non-redundant w.r.t.\ $DPI$. Thence, proposition~(5) is \true.

%Moreover, as proven above, the replacement node of $\mathsf{node}$ given by $\mathsf{node}_{rep} = \textsc{add}(\mathsf{node}_{rep,sub},\mathsf{node}[|\mathsf{node}_{rep,sub}|+1..|\mathsf{node}|])$ with $\mathsf{node}_{rep}.\mathsf{cs} = \textsc{add}(\mathsf{node}_{rep,sub}.\mathsf{cs},\mathsf{node}.\mathsf{cs}[|\mathsf{node}_{rep,sub}|+1..|\mathsf{node}|])$ which is constructed from $\mathsf{nd}_{rep,sub}$ is de-facto non-redundant w.r.t.\ $DPI$.
%
%\begin{itemize}
	%\item there is an alternative subnode $\mathsf{nd}_{rep,sub}$ of $\mathsf{node}$ such that $|\mathsf{nd}_{rep,sub}| \leq |\mathsf{node}| = |\mathsf{nd}'_{suc}|-1$ and
	%\item $\mathsf{nd}'_{suc}$ is the same node as $\mathsf{node}_x$ and 
	%\item $\textsc{add}(\mathsf{node}_{rep,sub},\mathsf{node}_x[|\mathsf{node}_{rep,sub}|+1..|\mathsf{node}_x|]) = \textsc{add}(\mathsf{node}_{rep},x)$ with $\textsc{add}(\mathsf{node}_{rep,sub}.\mathsf{cs},\mathsf{node}_x.\mathsf{cs}[|\mathsf{node}_{rep,sub}|+1..|\mathsf{node}_x|]) = \textsc{add}(\mathsf{node}_{rep}.\mathsf{cs},L)$ is de-facto non-redundant w.r.t.\ $DPI$ because $\mathsf{node}_{rep}$ is de-facto non-redundant w.r.t.\ $DPI$ and by the definition of $x$ (see above).
%\end{itemize}
Due to $|\mathsf{node}_{rep,sub}| \leq |\mathsf{node}| = |\mathsf{nd}'_{suc}|-1$, the alternative subnode of $\mathsf{nd}'_{suc}$ \emph{actually} found by \textsc{prune} 
%given the witness of redundancy $\mc''$ 
cannot have a cardinality greater than $|\mathsf{nd}'_{suc}|-1$. 
So, let $\mathsf{nd}_{alt}$ be the found alternative subnode of $\mathsf{nd}'_{suc}$. Since $|\mathsf{nd}_{alt}|\leq |\mathsf{nd}'_{suc}|-1$, we obtain that the replacement node $\mathsf{nd}'_{suc,new,1}$ of $\mathsf{nd}'_{suc}$ constructed from $\mathsf{nd}_{alt}$ must meet $\mathsf{nd}'_{suc,new,1}[|\mathsf{nd}'_{suc}|] = \mathsf{nd}'_{suc}[|\mathsf{nd}'_{suc}|] = x$ as well as $\mathsf{nd}'_{suc,new,1}.\mathsf{cs}[|\mathsf{nd}'_{suc}|] = \mathsf{nd}'_{suc}.\mathsf{cs}[|\mathsf{nd}'_{suc}|] = L$.
%So, for the replacement node it holds, by the reasoning conducted, that the element at the $|\mathsf{nd}'_{suc}|$-th position must be $x$, as in $\mathsf{nd}'_{suc}$. 
That is, the first $|\mathsf{node}| = |\mathsf{nd}'_{suc}|-1$ positions as a set correspond to a node in a transitive replaces-relation with $\mathsf{nd}$. 

Now, we have the following precondition of this lemma: Let $\mathsf{nd}'$ be in a transitive replaces-relation with $\mathsf{nd}$. If \textsc{prune} is called given a witness of redundancy of $\mathsf{nd}'$, then some replacement node of $\mathsf{nd}'$ is found. If only one replacement node of $\mathsf{nd}'$ is found, then this replacement node is de-facto non-redundant w.r.t.\ $DPI$.

Therefore, the same line of argument as used for $\mathsf{nd}'_{suc}$ can be applied to any node $\mathsf{nd}'_{suc,rep}$ in a transitive replaces-relation with $\mathsf{nd}'_{suc}$. That is, the following must be valid for any node $\mathsf{nd}'_{suc,rep}$ in a transitive replaces-relation with $\mathsf{nd}'_{suc}$: 
\begin{itemize}
	\item $\mathsf{nd}'_{suc,rep}[|\mathsf{nd}'_{suc}|] = x$ and $\mathsf{nd}'_{suc,rep}.\mathsf{cs}[|\mathsf{nd}'_{suc}|] = L$.
	\item If \textsc{prune} is called given a witness of redundancy of $\mathsf{nd}'_{suc,rep}$, then some replacement node of $\mathsf{nd}'_{suc,rep}$ is found. And, if only one replacement node of $\mathsf{nd}'_{suc,rep}$ is constructable, then this replacement node is de-facto non-redundant w.r.t.\ $DPI$.
\end{itemize}
After once a replacement node of $\mathsf{nd}'_{suc}$ or of some node in a transitive replaces-relation with $\mathsf{nd}'_{suc}$ is found which is de-facto non-redundant w.r.t.\ $DPI$, this replacement node cannot be replaced or pruned by Proposition~\ref{prop:de-facto_non_redundant_node_cannot_be_pruned_or_replaced}. Therefore, by Lemma~\ref{lem:prune}, no witness of redundancy of this replacement node can exist w.r.t.\ any DPI including a (not necessarily proper) subset of the test cases in $DPI$. Thence, proposition~(6) is \true.
\paragraph{Case~(q2):}
Here, we have that $\mathsf{nd}'_{suc}$ is not the same node as $\mathsf{nd}_{suc}$. This must be valid as $\mathsf{nd}'_{suc}$ is defined as the node set-equal to $\mathsf{nd}_{suc}$ that is an element of $\Queue$ \emph{immediately} after $\mathsf{nd}_{suc}$ was generated and $\mathsf{nd}_{suc}$ is assumed to be added to $\Queue_{dup}$ after being generated.
% and there can be only one node corresponding to one and the same set in $\Queue$ at the same time.
%Assume now, legitimately, that $\mathsf{nd}_{suc}$ is defined as $\mathsf{node}_{x}$ as shown above.

Now, independently of whether \textbf{(c1) or (c2)} occurs, the following holds: If \textsc{prune} is called given a witness of redundancy of $\mathsf{nd}'_{suc}$, then a replacement node of $\mathsf{nd}'_{suc}$ is found. And, if only one replacement node of $\mathsf{nd}'_{suc}$ is constructable, then this replacement node 
%found in lines~\ref{algoline:prune:check_if_alternative_subnode}-\ref{algoline:prune:insert_alternative_equal_node_into_S} 
is de-facto non-redundant w.r.t.\ $DPI$. 

To understand why this must hold, first recall that $\mathsf{nd}_{suc}$ is a successor of $\mathsf{node}$, i.e.\ $\mathsf{nd}_{suc}[1..|\mathsf{nd}_{suc}|-1]$ is the same node as $\mathsf{node}$. Furthermore, $\mathsf{node}$ is the node set-equal to $\mathsf{nd}$ that is processed. That is, $\mathsf{node}$ is either the same node as $\mathsf{nd}$ or it is in a transitive replaces-relation with $\mathsf{nd}$. 

Therefore, by the preconditions of this lemma, the following holds: If \textsc{prune} is called given a witness of redundancy of $\mathsf{node}$, then a replacement node of $\mathsf{node}$ is found. And, if only one replacement node $\mathsf{node}_{rep}$ of $\mathsf{node}$ is constructable, then $\mathsf{node}_{rep}$ is de-facto non-redundant w.r.t.\ $DPI$.

As argued in case (q1)(c2), $Comb(\Queue_{dup})$ must include a subnode $\mathsf{node}_{rep,sub}$ of $\mathsf{node}_{rep}$ that is de-facto non-redundant w.r.t.\ $DPI$ and from which $\mathsf{node}_{rep}$ is constructed. This must be satisfied during any execution of \textsc{dynamicHS} using a DPI that is equal to $DPI$ or includes a subset of the test cases in $DPI$. Hence, in particular, this must hold for the DPI $DPI''_{prev}$.

%By precondition, we have: If \textsc{prune} is called for $\mathsf{nd}$, then at least one replacement node of $\mathsf{nd}$ is found. And, if only one replacement node $\mathsf{nd}_{rep}$ of $\mathsf{nd}$ is found in lines~\ref{algoline:prune:check_if_alternative_subnode}-\ref{algoline:prune:insert_alternative_equal_node_into_S}, then $\mathsf{nd}_{rep}$ is non-redundant w.r.t.\ $DPI$. 
%To understand why this must hold, first recall that $\mathsf{nd}_{suc}$ is a successor of $\mathsf{node}$, i.e.\ $\mathsf{nd}_{suc}[1..|\mathsf{nd}_{suc}|-1]$ is the same node as $\mathsf{node}$. By precondition, we have: If \textsc{prune} is called for $\mathsf{nd}$, then at least one replacement node of $\mathsf{nd}$ is found. And, if only one replacement node $\mathsf{nd}_{rep}$ of $\mathsf{nd}$ is found in lines~\ref{algoline:prune:check_if_alternative_subnode}-\ref{algoline:prune:insert_alternative_equal_node_into_S}, then $\mathsf{nd}_{rep}$ is non-redundant w.r.t.\ $DPI$. 
%
%Because, as already argued above, only redundant nodes might be deleted from $\Queue_{dup}$ by \textsc{pruneQdup}, we conclude that $\mathsf{nd}_{rep}$ cannot be deleted from $\Queue_{dup}$ (or, respectively, 
%%remains constructable as a combined replacement node of some node in $\Queue_{dup}$
%must still be an element of $Comb(\Queue_{dup})$) during any execution of \textsc{dynamicHS} using a DPI that is equal or has fewer test cases than $DPI$. Due to $\mathsf{nd} = \mathsf{node}$, we have that $\mathsf{nd}_{rep}$ is also an alternative equal node of $\mathsf{node}$.

Since $\mathsf{nd}_{suc}$ has been added to $\Queue_{dup}$ by assumption, it might be found to be redundant w.r.t.\ some DPI (either equal to $DPI$ or including a subset of the test cases in $DPI$) during some execution of \textsc{pruneQdup}. If so, $\mathsf{nd}_{suc}$ cannot be pruned on account of Lemma~\ref{lem:pruneQdup} which says that a node can only be pruned from $\Queue_{dup}$ if the set $Comb_{\mathsf{nd}_{suc}}(\Queue_{dup})$ of combined equal nodes of $\mathsf{nd}_{suc}$ of $\Queue_{dup}$ (cf.\ Definition~\ref{def:comb_node}) is the empty set. 

However, $Comb_{\mathsf{nd}_{suc}}(\Queue_{dup}) \neq \emptyset$ must be valid. Because we demonstrated that
\begin{itemize}
	\item $\mathsf{node}_{rep,sub} \in Comb(\Queue_{dup})$,
	\item $\mathsf{nd}_{suc}\in\Queue_{dup}$,
	\item $\mathsf{nd}_{suc}$ is the same node as $\mathsf{node}_x = \textsc{add}(\mathsf{node},x)$ with $\mathsf{nd}_{suc}.\mathsf{cs}$ being equal to $\mathsf{node}_x.\mathsf{cs} = \textsc{add}(\mathsf{node.cs},L)$ and
	\item $x \notin \mathsf{nd}_{suc}.\mathsf{cs}[|\mathsf{nd}_{suc}|] \setminus \mc''$ (see case (q1)(c1)) wherefore $\mc''$ must be a witness of redundancy of $\mathsf{node}$.
\end{itemize}
Therefore, $\mathsf{nd}_{comb} := \textsc{add}(\mathsf{node}_{rep,sub},\mathsf{node}_x[|\mathsf{node}_{rep,sub}|+1..|\mathsf{node}_x|]) = \textsc{add}(\mathsf{node}_{rep},x)$ with $\mathsf{nd}_{comb}.\mathsf{cs} := \textsc{add}(\mathsf{node}_{rep,sub}.\mathsf{cs},\mathsf{node}_x.\mathsf{cs}[|\mathsf{node}_{rep,sub}|+1..|\mathsf{node}_x|]) = \textsc{add}(\mathsf{node}_{rep}.\mathsf{cs},L)$ is a combined equal node of $\mathsf{nd}_{suc}$ of $\Queue_{dup}$, i.e.\ $\mathsf{nd}_{comb} \in Comb_{\mathsf{nd}_{suc}}(\Queue_{dup})$. As argued in case~(q1)(c2), this node $\mathsf{nd}_{comb}$ (denoted by $\mathsf{nd}'_{suc,new}$ in case~(q1)(c2)) is de-facto non-redundant w.r.t.\ $DPI$.

Because \textsc{prune} is called immediately after \textsc{pruneQdup} and thus uses the updated list $\Queue_{dup}$ which comprises $\mathsf{nd}_{comb}$ and because $\mathsf{nd}_{comb} = \mathsf{nd}_{suc} = \mathsf{nd}'_{suc}$, we have that one replacement node of $\mathsf{nd}'_{suc}$ is definitely found by \textsc{prune}. And, in case there is only one replacement node of $\mathsf{nd}'_{suc}$ constructable during \textsc{prune}, this replacement node is given by $\mathsf{nd}_{comb}$. Thence, proposition~(5) is \true.

By Proposition~\ref{prop:de-facto_non_redundant_node_cannot_be_pruned_or_replaced}, the fact that $\mathsf{nd}_{comb} \in Comb_{\mathsf{nd}_{suc}}(\Queue_{dup}) \subseteq Comb(\Queue_{dup})$ at some point in time during the execution of \textsc{dynamicHS} with current DPI $DPI''_{prev}$ and the de-facto non-redundancy of $\mathsf{nd}_{comb}$ w.r.t.\ $DPI$, we conclude that, during any execution of \textsc{dynamicHS} with a current DPI that includes a (not necessarily proper) superset of the test cases in $DPI''_{prev}$ and includes a (not necessarily proper) subset of the test cases in $DPI$, $\mathsf{nd}_{comb} \in Comb(\Queue_{dup})$ must hold. Further on, $\mathsf{nd}_{comb} = \mathsf{nd}'_{suc}$ is \true.

Hence, independently of which replacement node of $\mathsf{nd}'_{suc}$ is \emph{actually} found by \textsc{prune}, a set-equality between this replacement node and $\mathsf{nd}_{comb}$ will hold. This is \true since each replacement node, by definition, is set-equal to the node it replaces. Consequently, this set-equality holds for any node in a transitive replaces-relation with $\mathsf{nd}'_{suc}$. 
So, we have that one replacement node of any node $\mathsf{nd}'_{suc,rep}$ in a transitive replaces-relation with $\mathsf{nd}'_{suc}$ is definitely found by \textsc{prune}. And, in case there is only one replacement node of $\mathsf{nd}'_{suc,rep}$ constructable during \textsc{prune}, this replacement node is given by $\mathsf{nd}_{comb}$ which is de-facto non-redundant w.r.t.\ $DPI$. 

That $\mathsf{nd}_{comb}$, after it has been used as a replacement node of $\mathsf{nd}'_{suc}$ or of some node in a transitive replaces-relation with $\mathsf{nd}'_{suc}$, cannot be pruned or replaced, follows from
Proposition~\ref{prop:de-facto_non_redundant_node_cannot_be_pruned_or_replaced} and the fact that $\mathsf{nd}_{comb}$ is de-facto non-redundant w.r.t.\ $DPI$. Therefore, by Lemma~\ref{lem:prune}, no witness of redundancy of $\mathsf{nd}_{comb}$ can exist w.r.t.\ any DPI including a (not necessarily proper) subset of the test cases in $DPI$. Thence, proposition~(6) is \true.
\end{proof}
In the following we prove the completeness of \textsc{dynamicHS}. Given an arbitrary minimal diagnosis $\md$ w.r.t.\ to an arbitrary fixed DPI $DPI$, Proposition~\ref{prop:dyn_completeness} testifies that there must be some node set-equal to $\md$ that is processed during the execution of \textsc{dynamicHS} with current DPI $DPI$ in case this execution terminates by reason of $\Queue = []$. Second, the proposition demonstrates that the set $\mD_{calc}$ returned by this execution of \textsc{dynamicHS} comprises all minimal diagnoses w.r.t.\ $DPI$. Additionally, the proposition shows that, at any point in time during the execution of Algorithm~\ref{algo:inter_onto_debug}, some node that corresponds to a subset of $\md$ must be stored by \textsc{dynamicHS}.

In terms of the hitting set tree produced by \textsc{dynamicHS}, the proposition states that, after all branches in the tree have been closed or pruned, there is a closed branch labeled by $valid$ for each minimal diagnosis w.r.t.\ $DPI$. And, for any minimal diagnosis $\md$ w.r.t.\ $DPI$, at any time during the tree construction, there is some branch that corresponds to a part of $\md$.

This proposition will be proven by deriving the existence of a de-facto non-redundant node $\mathsf{nd}_\md$ w.r.t.\ $DPI$ for any minimal diagnosis $\md$ w.r.t.\ $DPI$ such that $\mathsf{nd}_\md \subseteq \md$. In case $\mathsf{nd}_\md = \md$, we will deduce directly that the proposition must be \true. Otherwise, i.e.\ if $\mathsf{nd}_\md \subset \md$, then Lemmata~\ref{lem:successor_node_subset_diag_exists_1} and \ref{lem:successor_node_subset_diag_exists_2} will be exploited.
\begin{proposition}[Completeness of \textsc{dynamicHS}]\label{prop:dyn_completeness}
Let $\langle\mo,\mb,\Tp,\Tn\rangle_\RQ$ be the DPI and $\Tp'$ and $\Tn'$ the sets of positively and negatively answered queries given as an input to \textsc{dynamicHS} and assume that \textsc{dynamicHS} terminates due to $\Queue = []$. Let further $DPI:=\langle\mo,\mb,\Tp\cup\Tp',\Tn\cup\Tn'\rangle_\RQ$ and $\md$ be some minimal diagnosis w.r.t.\ $DPI$. 
%Assume the execution of \textsc{dynamicHS} with the current DPI $DPI$ and assume that the execution stops due to $\Queue = \emptyset$. Let $\md$ be some minimal diagnosis w.r.t.\ $DPI$. 
Then the following holds:
\begin{enumerate}[(1)]
	\item At some point in time during the execution of \textsc{dynamicHS} with current DPI $DPI$, there is a node $\mathsf{nd}$ such that $\mathsf{nd} = \md$ and $\mathsf{nd}$ is processed.
	\item The execution of \textsc{dynamicHS} with current DPI $DPI$ returns a set $\mD_{calc}$ that comprises all minimal diagnoses w.r.t.\ $DPI$.
	\item Let $DPI'$ be an arbitrary DPI that includes a (not necessarily proper) subset of the test cases in $DPI$. Then, at any point in time 
%	before some node $\mathsf{nd}$ with $\mathsf{nd} = \md$ is processed 
	during the execution of \textsc{dynamicHS} with current DPI $DPI'$, there is some node $\mathsf{nd}'$ such that $\mathsf{nd'} \subseteq \md$ and $\mathsf{nd'}$ is an element of one of the collections $\Queue, \mD_{calc}, \mD_{\checkmark}, \mD_{\times}$ or $\mD_{\supset}$.
%	$\mathsf{nd'} \in \Queue$ or $\mathsf{nd'} \in \mD_{calc}$ or $\mathsf{nd'} \in \mD_{\checkmark}$ or $\mathsf{nd'} \in \mD_{\times}$ or $\mathsf{nd'} \in \mD_{\supset}$.
\end{enumerate}
%\begin{itemize}
%%\item $GenNodes$ be the set of all nodes generated throughout the execution of all calls to \textsc{dynamicHS} during the execution of Algorithm~\ref{algo:inter_onto_debug},
%\item $\md$ be some minimal diagnosis w.r.t.\ $DPI$, 
%\item $\mD_{\checkmark}$ in \textsc{dynamicHS} include the set of all known minimal diagnoses w.r.t.\ the DPI $DPI'$ that is the current DPI of a particular call to \textsc{dynamicHS}.
%\end{itemize}
%Then there is some node $\mathsf{nd}$ such that 
%%\begin{enumerate}[(1)]
%%\item 
%$\mathsf{nd} = \md$ and 
%%\item 
%$\mathsf{nd}$ is processed.
%%\end{enumerate}
\end{proposition}
%PROP: If a redundant node nd w.r.t. some previous DPI is deleted during the execution of some previous call to \textsc{dynamicHS} such that nd subseteq md for some minimal diagnosis w.r.t. the current DPI, then there is a non-redundant node nd' w.r.t.\ the current DPI such thatnd' subseteq md. 
\begin{proof}
Let $GenNodes$ be the set of all nodes generated throughout the execution of all calls to \textsc{dynamicHS} during the execution of Algorithm~\ref{algo:inter_onto_debug}. 

Assume first that $\md = \emptyset$. This means that $DPI$ must be the input DPI of Algorithm~\ref{algo:inter_onto_debug}. Assume the opposite. 

A query is only generated and added as a new test case to the DPI in lines~\ref{algoline:inter_onto_debug:calc_query} and \ref{algoline:inter_onto_debug:add_pos_tc} or \ref{algoline:inter_onto_debug:param_update_end} of Algorithm~\ref{algo:inter_onto_debug} if there are at least two diagnoses in the set $\mD_{calc}$ (called $\mD_{\checkmark}$ in Algorithm~\ref{algo:inter_onto_debug}) returned by \textsc{dynamicHS}. Otherwise, line~\ref{algoline:inter_onto_debug:calc_query} cannot be reached since there must be exactly one diagnosis in $\mD_{\checkmark}$ when it comes to the execution of line~\ref{algoline:inter_onto_debug:stop_crit} wherefore the probability of this diagnosis must be equal to 1 which is greater or equal to $1 - \sigma$ for any choice of $\sigma$ (recall that $\sigma$ is positive). Please notice that $\mD_{\checkmark} = \emptyset$ cannot hold in line~\ref{algoline:inter_onto_debug:stop_crit} since this would imply the non-admissibility of the input DPI given to Algorithm~\ref{algo:inter_onto_debug} by Corollary~\ref{cor:query_leaves_valid_diag} and Definition~\ref{def:admissible}. By precondition, however, the DPI provided as an input to Algorithm~\ref{algo:inter_onto_debug} must be admissible.
	
Now, since $DPI$ is assumed to be not equal to the input DPI of Algorithm~\ref{algo:inter_onto_debug}, we have, by the argumentation given, that there must have been at least two diagnoses w.r.t.\ the input DPI. 

Let us first assume that $\mo$ is valid w.r.t.\ $\tuple{\cdot,\mb,\Tp,\Tn}_\RQ$ where $\tuple{\mo,\mb,\Tp,\Tn}_\RQ$ is the input DPI. Then, by Corollary~\ref{cor:notions_equiv}, $\emptyset$ is a diagnosis w.r.t.\ the input DPI. Obviously, it must be a minimal diagnosis and the only minimal diagnosis w.r.t.\ the input DPI, contradiction.

Second, suppose that $\mo$ is invalid w.r.t.\ $\tuple{\cdot,\mb,\Tp,\Tn}_\RQ$. By Proposition~\ref{prop:mindiag_mincs} which says that a diagnosis w.r.t.\ some DPI is a hitting set of all minimal conflict sets w.r.t.\ this DPI, we conclude that there must be at least one minimal conflict set $\mc$ w.r.t.\ the input DPI. Now, by Proposition~\ref{prop:changes_in_conflict_sets_after_testcase_added}, there must be a minimal conflict set $\mc'$ w.r.t.\ $DPI$ such that $\mc' \subseteq \mc$. By Proposition~\ref{prop:cs_admissible}, the fact that $\mo$ is invalid w.r.t.\ $\tuple{\cdot,\mb,\Tp,\Tn}_\RQ$, the fact that the input DPI is admissible and Corollary~\ref{cor:query_leaves_valid_diag} which states that the addition of queries as test cases cannot make an admissible DPI non-admissible, we obtain that $\emptyset \subset \mc'$. By Proposition~\ref{prop:mindiag_mincs}, this is a contradiction to $\md = \emptyset$ and the fact that $\md$ is a diagnosis w.r.t.\ $DPI$.

So, $DPI$ is the input DPI. Hence, the first call to \textsc{dynamicHS} throughout the execution of Algorithm~\ref{algo:inter_onto_debug} considers this DPI. During the execution of the first call to \textsc{dynamicHS}, $\Queue = [\emptyset]$ holds by lines~\ref{algoline:inter_onto_debug:Q={{}}} and \ref{algoline:inter_onto_debug:dynamicHS} of Algorithm~\ref{algo:inter_onto_debug}. The function \textsc{updateTree} has no effect during the execution of the first call to \textsc{dynamicHS} in Algorithm~\ref{algo:inter_onto_debug}. That is, in particular, it does not modify $\Queue$. For, \textsc{updateTree} first iterates over all elements in $\mD_{\times}$, then over all elements in $\mD_{\supset}$ and finally over all elements in $\mD_{\checkmark}$ where $\mD_{\times}=\mD_{\supset}=\mD_{\checkmark}=\emptyset$ by lines~\ref{algoline:inter_onto_debug:var_inst_start} and \ref{algoline:inter_onto_debug:dynamicHS} in Algorithm~\ref{algo:inter_onto_debug}. 
Hence, $\Queue = [\emptyset]$ holds when \textsc{dynamicHS} reaches line~\ref{algoline:dyn:get_first} wherefore $\emptyset$ is processed.
%
%
 %since the DPI can only change by the addition of test cases and no test cases are added unless there are at least two diagnoses. So, $\emptyset$ is the only minimal diagnosis w.r.t.\ the current DPI which is a contradiction to $\md \neq \emptyset$ and $\md$ is a minimal diagnosis w.r.t.\ the current DPI.
%
%
%Then, $\md$ is processed since the root node $\emptyset$ must be processed during the first call to \textsc{dynamicHS}.

Now, assume $\md \neq \emptyset$. In this case, the root node must be labeled by some minimal conflict set $L$ w.r.t.\ the DPI given as input to Algorithm~\ref{algo:inter_onto_debug}. To see this, suppose the opposite, i.e.\ that the root node is labeled by (i)~$nonmin$ or (ii)~$valid$.

Case~(i): This leads to a contradiction. For, $\mD_{calc} = \emptyset$ holds at the beginning of each execution of \textsc{dynamicHS} (line~\ref{algoline:dyn:Dcalc_gets_emptyset}). The root node $\emptyset$ must be the first node that is processed throughout all executions of \textsc{dynamicHS} during the execution of Algorithm~\ref{algo:inter_onto_debug} since it holds for each other node $\mathsf{node}$ that $\mathsf{node} \supset \emptyset$. Thus, the non-minimality criterion (lines~\ref{algoline:dlabel:non-min_crit_start}-\ref{algoline:dlabel:non-min_crit_end}) cannot be satisfied because $\mD_{calc} = \emptyset$ must hold in line~\ref{algoline:dlabel:non-min_crit_start} when \textsc{dLabel} is executed for the root node. Hence, the label $nonmin$ is impossible for the node $\emptyset$.

Case~(ii): By Lemma~\ref{lem:if_dlabel_returns_valid_nonmin_then}, we can deduce that $\emptyset$ is a diagnosis w.r.t.\ the input DPI. The fact that there cannot be any diagnosis w.r.t.\ the input DPI which is a proper subset of $\emptyset$ implies that $\emptyset$ is a minimal diagnosis w.r.t.\ the input DPI. By the reasoning applied before (in the case $\md = \emptyset$), we obtain that $DPI$ is equal to the input DPI and that $ \emptyset$ is the only minimal diagnosis w.r.t.\ $DPI$. This is a contradiction to the existence of a minimal diagnosis w.r.t.\ $DPI$, namely $\md$, which is non-empty.  
%
%Thence, by Proposition~\ref{prop:mindiag_mincs}, there cannot be a non-empty minimal conflict set w.r.t.\ the input DPI. 
%
%This is a contradiction to the fact that there is a minimal diagnosis w.r.t. 
%
%Moreover, it means that the current DPI must be the input DPI since the DPI can only change by the addition of test cases and no test cases are added unless there are at least two diagnoses. So, $\emptyset$ is the only minimal diagnosis w.r.t.\ the current DPI which is a contradiction to $\md \neq \emptyset$ and $\md$ is a minimal diagnosis w.r.t.\ the current DPI. 

Consequently, the root node must be labeled by some minimal conflict set $L$ w.r.t.\ the input DPI. Hence, \textsc{dynamicHS} will execute lines~\ref{algoline:dyn:for_e_in_L}-\ref{algoline:dyn:generate_nodes} and generate one node $\mathsf{node}_{e} := \textsc{add}(\emptyset,e) = [e]$ with $\mathsf{node}_{e}.\mathsf{cs} := \textsc{add}(\emptyset,L) = [L]$ for each $e\in L$ (cf.\ Definition~\ref{def:alternative_equal_node} for an explanation of the function \textsc{add}). This means that $\mathsf{node}_{e}\in GenNodes$ for each $e\in L$. As $L$ is a set and thus comprises only one exemplar of each element, there cannot be a set-equal node $\mathsf{node}'_{e}$ of $\mathsf{node}_{e}$ in $\Queue$ at the time $\mathsf{node}_{e}$ is generated. So, each $\mathsf{node}_{e}$ must be added to $\Queue$ in line~\ref{algoline:dyn:generate_nodes}. 

By Proposition~\ref{prop:changes_in_conflict_sets_after_testcase_added}, there must be some minimal conflict set $\mc$ w.r.t.\ $DPI$ such that $\mc \subseteq L$. Since $\md$ is a diagnosis w.r.t.\ $DPI$, we have that $\mc \cap \md \neq \emptyset$ by Proposition~\ref{prop:mindiag_mincs}. Thence, $L \cap \md \neq \emptyset$ must be \true. Therefore, in particular, $L \neq \emptyset$ must hold. 

Assume that $|\md|=1$. This implies by Proposition~\ref{prop:mindiag_mincs} that each minimal conflict set w.r.t.\ $DPI$ includes $x$. Further, there is some $x \in L$ such that $\md = \setof{x} = \mathsf{node}_x$. By Corollary~\ref{cor:prune_only_called_with_min_cs_in_dlabel} and Lemmata~\ref{lem:quick_prune_check} and \ref{lem:complete_prune_check}, \textsc{prune} is only called given some minimal conflict set $X$ w.r.t.\ the current DPI $DPI_{prev}$ as argument. As \textsc{dynamicHS} using $DPI$ is assumed to terminate due to $\Queue = []$, $DPI_{prev}$ must be equal to $DPI$ or include only a subset of the test cases $DPI$ includes. By Proposition~\ref{prop:changes_in_conflict_sets_after_testcase_added}, it must hold for $X$ that it is equal to or a superset of some minimal conflict set w.r.t.\ $DPI$. Hence $x \in X$ must hold wherefore $X$ cannot be a witness of redundancy of $\mathsf{node}_x$. So, $\mathsf{node}_x$ can never be pruned and must be finally processed as $DPI$ terminates due to $\Queue = []$ and nodes can only be deleted from $\Queue$ by being pruned or processed.
So far, we have established the truth of the lemma for $|\md| \leq 1$.

Now, suppose $|\md| \geq 2$. In the following, we argue that there must be some node $\mathsf{node}_y \subset \md$ for some $y \in L$ which is de-facto non-redundant w.r.t.\ $DPI$.

As \textsc{dynamicHS} using $DPI$ is assumed to terminate due to $\Queue = []$, each node $\mathsf{node}_e$ for $e \in L$ must have been generated (and $L$ must have been computed) during \textsc{dynamicHS} with some current DPI $DPI_{prev}$ which is equal to $DPI$ or includes only a subset of the test cases $DPI$ includes. Let $DPI_{prev+i}$ be any DPI which includes a proper superset of the test cases $DPI_{prev}$ includes and is either equal to $DPI$ or comprises a subset of the test cases $DPI$ comprises. Then, Proposition~\ref{prop:changes_in_conflict_sets_after_testcase_added} manifests that there must be some minimal conflict set $\mc_i$ w.r.t.\ $DPI_{prev+i}$ such that $\mc_i \subseteq L$. Since we proved above that $L\neq \emptyset$ must hold, we deduce by Proposition~\ref{prop:if_DPI_cs_neq_emptyset_then_DPI+1_cs_neq_emptyset} that $\mc_i \neq \emptyset$ must be valid. 

From Corollaries~\ref{cor:prune_only_called_with_min_cs_in_dlabel}, \ref{cor:pruneQdup_only_called_with_min_cs_in_dlabel} and Lemmata~\ref{lem:quick_prune_check} and \ref{lem:complete_prune_check} we infer that \textsc{prune} as well as \textsc{pruneQdup} are always called with a minimal conflict set $X$ w.r.t.\ the current DPI given as an argument. Lemma~\ref{lem:pruneQdup} and the fact that \textsc{prune} is always called immediately after \textsc{pruneQdup} given the argument $\Queue_{dup}$ which is the output list of \textsc{pruneQdup}, we have that the list $\Queue_{dup}$ includes only nodes $\mathsf{nd}$ such that there is no $r \in \setof{1,\dots,|\mathsf{nd}|}$ for which $\mathsf{nd.cs}[r] \supset X$. As a consequence of this, we have by Lemma~\ref{lem:prune} that for all nodes $\mathsf{nd}$ in the collection $S'$ returned by \textsc{prune} there is no $r \in \setof{1,\dots,|\mathsf{nd}|}$ for which $\mathsf{nd.cs}[r] \supset X$. 
%All nodes for which $X$ is a witness of redundancy have been pruned and are thus not elements of $S'$.

Thence, the first time \textsc{prune} is called with some $X_1 \subset L$, $X_1$ is a minimal conflict set w.r.t.\ some DPI $DPI_{prev+i}$. Thus, as argued, $X_1 \supset \emptyset$ must hold. 
So, after \textsc{prune} has finished executing, for each node $\mathsf{node}$ in its output set there will be no $r \in \setof{1,\dots,|\mathsf{node}|}$ such that $\mathsf{node.cs}[r] \supset X_1$. For any further minimal conflict set $X_2$ w.r.t.\ some $DPI_{prev+i+k}$ for which \textsc{prune} is called, we have that $X_2 \supset \emptyset$ and for each node $\mathsf{node}$ in its output set there will be no $r \in \setof{1,\dots,|\mathsf{node}|}$ such that $\mathsf{node.cs}[r] \supset X_2$, and so on.

For $L$, in particular, there is some (possibly empty) sequence of minimal conflict sets $X_1, \dots, X_n$ w.r.t.\ DPIs $DPI_{prev+i_{1}}, \dots, DPI_{prev+i_{n}}$ ($i_j < i_{j+1}$ for $j \in \setof{1,\dots,n-1}$) such that $L \supset X_1$ and $X_i \supset X_{i+1}$ for $i \in \setof{1, \dots, n}$ where this sequence includes all such conflict sets which restrict a conflict set used to label nodes that was initially given by $L$. Since $X_n$ is a minimal conflict set w.r.t.\ $DPI_{prev+i_{n}}$ which is equal to $DPI$ or includes only a subset of the test cases $DPI$ includes, we have that there must be some minimal conflict set $\mc$ w.r.t.\ $DPI$ such that $\mc \subseteq X_n$, as already argued. As $\md$ must hit $\mc$ by Proposition~\ref{prop:mindiag_mincs}, we obtain that $\md \cap X_n \neq \emptyset$. 

So, by the inference given, there must be some $y \in L$ such that $y \in X_1 \cap \dots \cap X_n$ and $y \in \md$. That is, $\mathsf{node}_y \subset \md$.

%and some node $\mathsf{node}$ (either the same node as $\mathsf{node}_e$ or some successor of $\mathsf{node}_e$) such that $\mathsf{node.cs}[1] = L$ and $\mathsf{node}[1] \in X_1$. For
Since $|\mathsf{node}_e| = 1$ and $\mathsf{node}_e.\mathsf{cs}[1] = L$ for all $e \in L$, in particular for $e = y$, we obtain by Definitions~\ref{def:active_sublabel} and \ref{def:de-facto_non-redundant} that $\mathsf{node}_y$ is de-facto non-redundant w.r.t.\ $DPI$.

%Let us now refer to $\mathsf{node}_y$ by $\mathsf{nd}$. 
So, the preconditions of Lemma~\ref{lem:successor_node_subset_diag_exists_1} are met for $\mathsf{node}_y$. As a consequence, there must be a node $\mathsf{nd}'_{suc}$ such that $|\mathsf{nd}'_{suc}| = |\mathsf{node}_y| + 1$, $\mathsf{nd}'_{suc} \subseteq \md$, $\mathsf{nd}'_{suc}$ is an element of $\Queue$ immediately after $\mathsf{node}_y$ has been processed and $\mathsf{nd}'_{suc}$ satisfies the postulations to the node $\mathsf{nd}$ in the preconditions of Lemma~\ref{lem:successor_node_subset_diag_exists_2}. Hence, if $\mathsf{nd}'_{suc} \subset \md$, there must be a node $\mathsf{nd}''_{suc}$ such that $|\mathsf{nd}''_{suc}| = |\mathsf{nd}'_{suc}| + 1$, $\mathsf{nd}''_{suc} \subseteq \md$, $\mathsf{nd}''_{suc}$ is an element of $\Queue$ immediately after a node set-equal to $\mathsf{nd}'_{suc}$ has been processed and $\mathsf{nd}''_{suc}$ satisfies the postulations to the node $\mathsf{nd}$ in the preconditions of Lemma~\ref{lem:successor_node_subset_diag_exists_2}. 

This reasoning by means of Lemma~\ref{lem:successor_node_subset_diag_exists_2} can be further applied to finally derive that some node $\mathsf{nd} = \md$ must be generated and some node $\mathsf{nd}'$ set-equal to $\mathsf{nd}$ must be an element of $\Queue$. 
By Lemma~\ref{lem:successor_node_subset_diag_exists_2}, either $\mathsf{nd}'$ or a node set-equal to $\mathsf{nd}'$ which is in a transitive replaces-relation with $\mathsf{nd}'$ must finally be processed. Reason for this is that $\mathsf{nd}'\in \Queue$ cannot be pruned, but can only be replaced, and each replacement node is set-equal to $\mathsf{nd}'$ and thus to $\md$. Moreover, 
%because of $\mathsf{nd}' \in \Queue$ and the termination of 
the execution of \textsc{dynamicHS} with current DPI $DPI$ terminates due to $\Queue = []$ wherefore each node in $\Queue$ must be either pruned or processed as these are the only two ways nodes might be eliminated from $\Queue$. 

If some node $\mathsf{nd} = \md$ is processed during an execution of \textsc{dynamicHS} with current DPI some DPI $DPI'$ that includes a proper subset of the test cases in $DPI$, then \textsc{dLabel} cannot return a set $L$. This holds by Lemma~\ref{lem:Ccalc_in_dlabel} and Proposition~\ref{prop:changes_in_conflict_sets_after_testcase_added}. The former says that $\mathsf{nd} \cap L = \emptyset$ and $L$ is a minimal conflict set w.r.t.\ $DPI'$. The latter asserts that each conflict set w.r.t.\ $DPI$ is a conflict set w.r.t.\ $DPI$. Moreover, we can deduce that $L \neq \emptyset$ must hold if a set $L$ is returned by \textsc{dLabel} by a similar argumentation as used in the proof of Lemma~\ref{lem:successor_node_subset_diag_exists_2}. That is, by Proposition~\ref{prop:mindiag_mincs}, we have that $\md$ cannot be a diagnosis w.r.t.\ $DPI$, contradiction.

Hence, \textsc{dLabel} must return $nonmin$ or $valid$ for $\mathsf{nd}$. In the former case, it would be added to $\mD_{\supset}$, in the latter to $\mD_{calc}$. Similarly as done in the proof of Lemma~\ref{lem:successor_node_subset_diag_exists_2}, we can show that $\mathsf{nd}$ must be reinserted into $\Queue$ the latest during the execution of \textsc{dynamicHS} with current DPI $DPI$ and, in particular, $\mathsf{nd}$ must be an element of $\Queue$ when the repeat-loop during the execution of \textsc{dynamicHS} with current DPI $DPI$ is entered. Thus, $\mathsf{nd}$ must be (again) processed during the execution of \textsc{dynamicHS} with current DPI $DPI$. This proves proposition~(1).

Proposition~(2): At the beginning of each execution of \textsc{dynamicHS}, it holds that $\mD_{calc} = \emptyset$. This is \true in particular for the execution of \textsc{dynamicHS} with current DPI $DPI$. Now, proposition~(1) reveals that, for \emph{each} diagnosis $\md$ w.r.t.\ $DPI$, at some point in time during the execution of \textsc{dynamicHS} with current DPI $DPI$, there is a node $\mathsf{nd}$ such that $\mathsf{nd} = \md$ and $\mathsf{nd}$ is processed. When $\mathsf{nd}$ is processed, the \textsc{dLabel} function is called for $\mathsf{nd}$. The \textsc{dLabel} function might return (a)~a set $L$, (b)~$nonmin$ or (c)~$valid$. There are no other possible return values of \textsc{dLabel}.

Case~(a): By Lemma~\ref{lem:Ccalc_in_dlabel}, $L$ must be a minimal conflict set w.r.t.\ $DPI$ such that $\mathsf{nd} \cap L = \emptyset$. According to Proposition~\ref{prop:mindiag_mincs}, it must hold for $\md$ that $\md \cap L \neq \emptyset$ since $\md$ is a minimal diagnosis w.r.t.\ $DPI$. Since $\md = \mathsf{nd}$, we obtain a contradiction.

Case~(b): By Lemma~\ref{lem:if_dlabel_returns_valid_nonmin_then}, $\mD_{calc}$ can comprise only diagnoses w.r.t.\ $DPI$. By line~\ref{algoline:dlabel:non-min_crit_start}, this yields that there is a diagnosis w.r.t.\ $DPI$ that is a proper subset of $\mathsf{nd}$. This however is a contradiction to the set-equality of $\mathsf{nd}$ with the minimal diagnosis $\md$ w.r.t.\ $DPI$.

Consequently, case~(c) must arise. This implies that $\mathsf{nd}$ is added to $\mD_{calc}$ in line~\ref{algoline:dyn:add_to_Dcalc}. 
%Since \textsc{dynamicHS} with current DPI $DPI$ is assumed to return by reason of $\Queue = \emptyset$

Proposition~(3) is a direct consequence of the reasoning in this proof and in the proofs of Lemmata~\ref{lem:successor_node_subset_diag_exists_1} and \ref{lem:successor_node_subset_diag_exists_2}.
\end{proof}

\subsection[Soundness]{Soundness of \textsc{dynamicHS}}
\label{sec:SoundnessOfTextscDynamicHS}
Having established the completeness of each call to \textsc{dynamicHS}
%, i.e.\ all minimal diagnoses w.r.t.\ $DPI$ are added to the set $\mD_{calc}$ 
concerning the minimal diagnoses w.r.t.\ the current DPI $DPI$ at this call, we are now able to prove the soundness of each call to \textsc{dynamicHS}. That is, we will demonstrate that \emph{only} minimal diagnoses w.r.t.\ $DPI$ can be added to the set $\mD_{calc}$ 
%(which is finally returned by \textsc{dynamicHS} and used for query computation during Algorithm~\ref{algo:inter_onto_debug}) 
during \textsc{dynamicHS} with the current DPI $DPI$. Necessary condition for the proof of the following proposition is the completeness of \textsc{dynamicHS}, i.e.\ Proposition~\ref{prop:dyn_completeness}.
\begin{proposition}[Soundness of \textsc{dynamicHS}]\label{prop:dyn_soundness}
Let $\langle\mo,\mb,\Tp,\Tn\rangle_\RQ$ be the DPI and $\Tp'$ and $\Tn'$ the sets of positively and negatively answered queries given as an input to \textsc{dynamicHS}. Let further $DPI:=\langle\mo,\mb,\Tp\cup\Tp',\Tn\cup\Tn'\rangle_\RQ$.
%Assume the execution of \textsc{dynamicHS} with the current DPI $DPI$. 
Then, the following holds: 
\begin{enumerate}[(1)]
	\item At any point in time during the execution of \textsc{dynamicHS} with current DPI $DPI$, each node in $\mD_{calc}$ is a minimal diagnosis w.r.t.\ $DPI$.
	\item At any point in time during the execution of \textsc{dynamicHS} with current DPI $DPI$, $\mD_{calc}$ comprises the $|\mD_{calc}|$ most-probable minimal diagnoses w.r.t.\ $DPI$.
\end{enumerate}
\end{proposition}
\begin{proof}
Proposition~(1): At the beginning of any execution of \textsc{dynamicHS}, the set $\mD_{calc}$ is the empty set (line~\ref{algoline:dyn:Dcalc_gets_emptyset}). So, it suffices to show that only minimal diagnoses w.r.t.\ $DPI$ can be added to $\mD_{calc}$ during the execution of \textsc{dynamicHS} with the current DPI $DPI$.

A node $\mathsf{node}$ can be added to $\mD_{calc}$ exclusively in line~\ref{algoline:dyn:add_to_Dcalc}. In order for this line to be reached, by the criterion that is checked in line~\ref{algoline:dyn:if_L_valid}, $\mathsf{node}$ must be processed and labeled by $valid$. By Lemma~\ref{lem:if_dlabel_returns_valid_nonmin_then}, if $\mathsf{node}$ gets labeled by $valid$, then it is a diagnosis w.r.t.\ $DPI$. 

So, assume that $\mathsf{node}$ is added to $\mD_{calc}$ where $\mathsf{node}$ is a non-minimal diagnosis w.r.t.\ $DPI$. Since $\mathsf{node}$ must have been processed and labeled by $valid$, the \textsc{dLabel} function must have been executed given $\mathsf{node}$ as an argument and must have returned in line~\ref{algoline:dlabel:return_valid}. Hence, there can be no node $\mathsf{nd} \in \mD_{calc}$ such that $\mathsf{nd} \subset \mathsf{node}$ holds, as otherwise \textsc{dLabel} would have already returned in line~\ref{algoline:dlabel:non-min_crit_end}. 

However, since $\mathsf{node}$ is a non-minimal diagnosis w.r.t.\ $DPI$ there must be some minimal diagnosis $\md$ w.r.t.\ $DPI$ such that $\md \subset \mathsf{node}$. 
Moreover, by Proposition~\ref{prop:dyn_completeness}, at any point in time before $\md$ is added to $\mD_{calc}$, 
%during the execution of \textsc{dynamicHS} with the current DPI $DPI$, 
there must be some node $\mathsf{nd}$ such that $\mathsf{nd} \subseteq \md$ and $\mathsf{nd}$ is an element of one of the collections (a)~$\mD_{calc}$, (b)~$\mD_{\checkmark}$, (c)~$ \mD_{\times}$, (d)~$\mD_{\supset}$ or (e)~$\Queue$. So, let us consider these cases in sequence.

Case~(a): First, 
%By case~(a), we have that
$\mathsf{nd} \subseteq \md$ and $\md \subset \mathsf{node}$ implies that 
$\mathsf{nd} \subset \mathsf{node}$ must be valid. As mentioned above, there can be no node in $\mD_{calc}$ which is a proper subset of $\mathsf{node}$, contradiction.

Case~(b): In this case, $\mathsf{nd}$ must be also an element of $\Queue$ since all nodes in $\mD_{\checkmark}$ are inserted into $\Queue$ during \textsc{updateTree} which is executed before the repeat-loop is entered, i.e.\ before it can come to the assumed addition of $\mathsf{node}$ to $\mD_{calc}$ which can only take place within the repeat-loop. So, in fact case~(e) applies here.

Case~(c): As can be easily seen from lines~\ref{algoline:update:reinsert_D_of_Dx_to_Q}-\ref{algoline:update:process_Dtimes_end} in \textsc{updateTree}, $\mD_{\times}$ must be the empty set at the time $\mathsf{node}$ might be added to $\mD_{calc}$ by analogue argumentation as in case~(b), contradiction.

Case~(d): By lines~\ref{algoline:update:process_Dsupset_start}-\ref{algoline:update:process_Dsupset_end} in \textsc{updateTree} and the fact that \textsc{updateTree} must have been executed before the assumed addition of $\mathsf{node}$ to $\mD_{calc}$ can take place as argued in case~(b), we have that there must be some node $\mathsf{nd}_{sub} \in \mD_{\checkmark}$ such that $\mathsf{nd}_{sub} \subset \mathsf{nd}$. Otherwise, $\mathsf{nd}$ would have been deleted from $\mD_{\supset}$ in line~\ref{algoline:update:delete_from_Dsupset}. By $\mathsf{nd} \subset \mathsf{node}$ as per case~(a), we deduce that $\mathsf{nd}_{sub} \subset \mathsf{node}$. Due to $\mathsf{nd} \subseteq \md$, it must be \true that $\mathsf{nd}_{sub} \subseteq \md$. 
Thus, we have derived that case case~(b) holds for the node $\mathsf{nd}_{sub}$. By the deductions in case~(b) above, we eventually know that case~(e) must hold.

%Thence, all cases either lead to a contradiction or imply the truth of another case for which a contradiction can be derived. So, at the bottom line we have got a contradiction.
Thence, assumption of cases~(a) and (c) is contradictory. Cases~(b) and (d) imply the truth of case~(e). Therefore, case~(e) must occur.
% for which a contradiction can be derived. So, at the bottom line we have got a contradiction.

Case~(e): Due to the facts that all nodes are inserted into $\Queue$ in a manner that descending order of nodes in $\Queue$ by $p_{nodes}()$ is maintained (cf.\ lines~\ref{algoline:dyn:generate_nodes}, \ref{algoline:prune:insert_alternative_equal_node_into_S'} and \ref{algoline:prune:insert_same_node_into_S'}) and always the first node in $\Queue$ is processed next (cf.\ line~\ref{algoline:dyn:get_first}), we conclude that $p_{nodes}(\mathsf{nd}) \leq p_{nodes}(\mathsf{node})$ must be valid. However, due to $\mathsf{nd} \subseteq \md \subset \mathsf{node}$ we have that $\mathsf{nd} \subset \mathsf{node}$. Now, by Lemma~\ref{lem:superset_lower_prob}, $p_{nodes}(\mathsf{n}) > p_{nodes}(\mathsf{n}')$ holds for any two nodes $\mathsf{n}$ and $\mathsf{n}'$ such that $\mathsf{n} \subset \mathsf{n}'$. Therefore, $p_{nodes}(\mathsf{nd}) > p_{nodes}(\mathsf{node})$, contradiction.

Proposition~(2): By proposition~(1), each node added to $\mD_{calc}$ must be a minimal diagnosis w.r.t.\ $DPI$. 

Assume any point in time $t$ during the execution of \textsc{dynamicHS} with the current DPI $DPI$. Then, $|\mD_{calc}| = m \geq 0$ must hold. We use induction by $m$ to prove proposition~(2). 

Base Case: Suppose that $m = 0$ and some minimal diagnosis $\md$ w.r.t.\ $DPI$ is added to $\mD_{calc}$ where $\md$ is not the most probable minimal diagnosis w.r.t.\ $DPI$. This implies that $\md$ is processed and that $\md$ has the highest probability as per $p_{nodes}()$ among all nodes that are elements of $\Queue$ at time $t$, as argued in the proof of proposition~(1). 

Let us denote by $\md_1$ the most probable minimal diagnosis w.r.t.\ $DPI$. That is, $p_{nodes}(\md_1) > p_{nodes}(\md)$ holds. 

Then, by Proposition~\ref{prop:dyn_completeness}, at any point in time during the execution of \textsc{dynamicHS} with the current DPI $DPI$, there must be some node $\mathsf{nd}_1$ such that
$\mathsf{nd}_1 \subseteq \md_1$ and $\mathsf{nd}_1$ is an element of one of the collections (a)~$\mD_{calc}$, (b)~$\mD_{\checkmark}$, (c)~$ \mD_{\times}$, (d)~$\mD_{\supset}$ or (e)~$\Queue$. 

Case~(a) can be ruled out due to the assumption that $\mD_{calc} = \emptyset$. Cases~(b)-(d) can be treated analogously as above in the proof of proposition~(1). Hence, case~(e) must hold.

%So, let us call the node for which case~(e) holds also $\mathsf{nd}_1$. 
That is, $\mathsf{nd}_1\in\Queue$ at time $t$ and $\mathsf{nd}_1$ is equal to or a subset of $\md_1$. 
As $p_{nodes}(\mathsf{nd}_1) \geq p_{nodes}(\md_1) > p_{nodes}(\md)$ holds by Lemma~\ref{lem:superset_lower_prob}, we can infer that $\md$ has not the highest probability as per $p_{nodes}()$ among all nodes that are elements of $\Queue$ at time $t$, contradiction.
%Then, by Proposition~\ref{prop:dyn_completeness}, at any point in time before $\md_1$ is added to $\mD_{calc}$ during the execution of \textsc{dynamicHS} with the current DPI $DPI$, there must be some node in $\Queue$ which is equal to or a subset of $\md_1$. By the assumption that $\md_1$ is not an element of $\mD_{calc}$ at time $t$, let $\mathsf{nd}_1 \in \Queue$ such that $\mathsf{nd}_1 \subseteq \md_1$ at time $t$. As $p_{nodes}(\mathsf{nd}_1) \geq p_{nodes}(\md_1) > p_{nodes}(\md)$ holds by Lemma~\ref{lem:superset_lower_prob}, we can infer that $\md$ has not the highest probability as per $p_{nodes}()$ among all nodes that are elements of $\Queue$ at time $t$, contradiction.

Inductive Step: Now, let $m > 0$ and assume that the $m$ most probable minimal diagnoses w.r.t.\ $DPI$ are already elements of $\mD_{calc}$. Suppose further that some minimal diagnosis $\md$ w.r.t.\ $DPI$ is added to $\mD_{calc}$ where $\md$ is not the $(m+1)$-th most probable minimal diagnosis w.r.t.\ $DPI$. 
This implies that $\md$ is processed and that $\md$ has the highest probability as per $p_{nodes}()$ among all nodes that are elements of $\Queue$ at time $t$.

Let us denote by $\md_{m+1}$ the $(m+1)$-th most probable minimal diagnosis w.r.t.\ $DPI$. That is, $p_{nodes}(\md_{m+1}) > p_{nodes}(\md)$ holds since the $m$ most probable minimal diagnoses w.r.t.\ $DPI$ are already elements of $\Queue$. 

Then, by Proposition~\ref{prop:dyn_completeness}, at any point in time during the execution of \textsc{dynamicHS} with the current DPI $DPI$, there must be some node $\mathsf{nd}_{m+1}$ such that $\mathsf{nd}_{m+1} \subseteq \md_{m+1}$ and $\mathsf{nd}_{m+1}$ is an element of one of the collections (a)~$\mD_{calc}$, (b)~$\mD_{\checkmark}$, (c)~$ \mD_{\times}$, (d)~$\mD_{\supset}$ or (e)~$\Queue$. 

Case~(a) can be ruled out due to proposition~(1) which affirms that only minimal diagnoses w.r.t.\ $DPI$ can be elements of $\mD_{calc}$. As $\md_{m+1}$ is not an element of $\mD_{calc}$ per assumption, a node $\mathsf{nd}_{m+1} = \md_{m+1}$ cannot be an element of $\mD_{calc}$. Furthermore, by the fact that $\md_{m+1}$ is a minimal diagnosis w.r.t.\ $DPI$, any node $\mathsf{nd}_{m+1} \subset \md_{m+1}$ cannot be a (minimal) diagnosis w.r.t.\ $DPI$ and thus cannot be an element of $\mD_{calc}$. Cases~(b)-(d) can be treated analogously as above in the proof of proposition~(1). Hence, case~(e) must hold.

That is, $\mathsf{nd}_{m+1} \in \Queue$ at time $t$ and $\mathsf{nd}_{m+1}$ is equal to or a subset of $\md_{m+1}$. As $p_{nodes}(\mathsf{nd}_{m+1}) \geq p_{nodes}(\md_{m+1}) > p_{nodes}(\md)$ holds by Lemma~\ref{lem:superset_lower_prob}, we can infer that $\md$ has not the highest probability as per $p_{nodes}()$ among all nodes that are elements of $\Queue$ at time $t$, contradiction.
%
%Then, at any point in time before $\md_{m+1}$ is added to $\mD_{calc}$ during the execution of \textsc{dynamicHS} with the current DPI $DPI$, there must be some node in $\Queue$ which is equal to or a subset of $\md_{m+1}$. By the assumption that $\md_{m+1}$ is not an element of $\mD_{calc}$ at time $t$, let $\mathsf{nd}_{m+1} \in \Queue$ such that $\mathsf{nd}_{m+1} \subseteq \md_{m+1}$ at time $t$. As $p_{nodes}(\mathsf{nd}_{m+1}) \geq p_{nodes}(\md_{m+1}) > p_{nodes}(\md)$ holds by Lemma~\ref{lem:superset_lower_prob}, we can infer that $\md$ has not the highest probability as per $p_{nodes}()$ among all nodes that are elements of $\Queue$ at time $t$, contradiction.
\end{proof}

\subsection[Correctness]{Correctness of \textsc{dynamicHS}}
\label{sec:CorrectnessOfTextscDynamicHS}
Now, we are able to prove that \textsc{dynamicHS} terminates and yields an output complying with the assertions given in Algorithm~\ref{algo:inter_dyn_hs}:
\begin{corollary}\label{cor:dynamic_hs_correctness}
Any call to \textsc{dynamicHS} (given the inputs described in Algorithm~\ref{algo:inter_dyn_hs}) within Algorithm~\ref{algo:inter_onto_debug} terminates and yields an output $\tuple{\mD_{calc},\Queue, \mathbf{C}_{calc}, \mD_{\times}, \mD_{\supset},\Queue_{dup}}$ where
\begin{enumerate}[(1)]
\item $\mD_{calc}$ is the current set of leading diagnoses such that
\begin{enumerate}[(a)]
\item $\mD_{calc} \subseteq \minD_{\langle\mo,\mb,\Tp\cup\Tp',\Tn\cup\Tn'\rangle_\RQ}$ is the set of most probable minimal diagnoses w.r.t.\ $\langle\mo,\mb,\Tp\cup\Tp',\Tn\cup\Tn'\rangle_\RQ$ such that 
\begin{enumerate}[(i)]
\item $n_{\min} \leq |\mD_{calc}| \leq n_{\max}$ and 
\item $\mD_{calc}\setminus\mD_{\checkmark} \neq \emptyset$,
\end{enumerate}
if such a set $\mD_{calc}$ exists;
or 
\item $\mD_{calc}$ is equal to the set of all minimal diagnoses $\minD_{\langle\mo,\mb,\Tp\cup\Tp',\Tn\cup\Tn'\rangle_\RQ}$, otherwise;
\end{enumerate}
where ``most-probable'' refers to the probability measure $p_{nodes}()$ given by Definition~\ref{def:p_node()} and obtained from the function $p()$ given as an input argument to \textsc{dynamicHS}.
\item $\Queue$ is the current queue of open (non-labeled) nodes of the produced hitting set tree,
\item $\mathbf{C}_{calc}$ is a set of conflict sets w.r.t.\ the current DPI $\langle\mo,\mb,\Tp\cup\Tp',\Tn\cup\Tn'\rangle_\RQ$,
\item $\mD_{\times} = \emptyset$, 
\item $\mD_{\supset}$ is the set of all processed nodes so far throughout the execution of Algorithm~\ref{algo:inter_onto_debug} that are non-minimal diagnoses w.r.t.\ the current DPI $\langle\mo,\mb,\Tp\cup\Tp',\Tn\cup\Tn'\rangle_\RQ$ and
%, i.e.\ for each $\mathsf{nd} \in \mD_{\supset}$ there is some $\md \in \mD_{calc}$ satisfying $\mathsf{nd} \supset \md$ and 
%$\mD_{\supset}$ is a set of non-minimal diagnoses w.r.t.\ the current DPI $\langle\mo,\mb,\Tp\cup\Tp',\Tn\cup\Tn'\rangle_\RQ$ such that for each $\mathsf{nd} \in \mD_{\supset}$ there is some $\md \in \mD_{calc}$ satisfying $\mathsf{nd} \supset \md$ and 
\item $\Queue_{dup}$ includes a node set-equal to $X$ for a set $X \subseteq \mo$ iff 
\begin{itemize}
	\item $\mathsf{nd} = X$ is a generated node that is de-facto non-redundant w.r.t.\ the current DPI $\langle\mo,\mb,\Tp\cup\Tp',\Tn\cup\Tn'\rangle_\RQ$, such that, at generation time of $\mathsf{nd}$, there was a node set-equal to $X$ in $\Queue$ or 
	\item there is a de-facto non-redundant node $\mathsf{nd}' = X$ w.r.t.\ the current DPI $\langle\mo,\mb,\Tp\cup\Tp',\Tn\cup\Tn'\rangle_\RQ$ which is a combined equal node of some generated node $\mathsf{nd}''$ that has been added to $\Queue_{dup}$. 
\end{itemize}
%%there is a generated node $\mathsf{nd}' = X$ which 
%
%is the set of all non-pruned nodes $\mathsf{nd}$ for each of which, at generation time, there was one node $\mathsf{nd}'\in\Queue$ such that $\mathsf{nd} = \mathsf{nd}'$.
%
%includes all de-facto non-redundant nodes $\mathsf{nd}$ w.r.t.\ the current DPI $\langle\mo,\mb,\Tp\cup\Tp',\Tn\cup\Tn'\rangle_\RQ$ for each of which, at generation time, there was one node $\mathsf{nd}'\in\Queue$ such that $\mathsf{nd} = \mathsf{nd}'$.
%
%$\Queue_{dup}$ is a set of nodes such that for each $\mathsf{nd} \in \Queue_{dup}$ either 
\end{enumerate}
\end{corollary}
\begin{proof}
First, we prove that any call to \textsc{dynamicHS} within Algorithm~\ref{algo:inter_onto_debug} terminates.
%Termination of any call to \textsc{dynamicHS} within Algorithm~\ref{algo:inter_onto_debug} is granted by the following argumentation:
To this end, assume that a call to \textsc{dynamicHS} executes infinitely. That is, $\Queue = []$ must not be satisfied at any time during the execution of \textsc{dynamicHS} due to the stop criterion of \textsc{dynamicHS} in line~\ref{algoline:dyn:until}.

However, the overall number of nodes that might be elements of $\Queue$ during the processing of the repeat-loop of any call to \textsc{dynamicHS} is finite. This is satisfied since each node $\mathsf{nd}$ in \textsc{dynamicHS} is a list corresponding to a subset of $\mo$ and each element of the list $\mathsf{nd.cs}$ is a subset of $\mo$ as well. For, a node can never correspond to a proper superset of $\mo$ by Proposition~\ref{prop:qx_correctness} which says that $\scQX(\langle\mo\setminus\md$, $\mb,\Tp\cup\Tp',\Tn\cup\Tn'\rangle_\RQ)$ returns 'no conflict' in case $\mo\setminus\md$ is valid w.r.t.\ $\tuple{\cdot,\mb,\Tp\cup\Tp',\Tn\cup\Tn'}_\RQ$ which is equivalent to $\md$ being a diagnosis w.r.t.\ $\tuple{\mo\setminus\md,\mb,\Tp\cup\Tp',\Tn\cup\Tn'}_\RQ$ by Corollary~\ref{cor:notions_equiv}. Now, the DPI $\tuple{\mo,\mb,\Tp\cup\Tp',\Tn\cup\Tn'}_\RQ$ is admissible which follows from the admissibility of the input DPI $\tuple{\mo,\mb,\Tp,\Tn}_\RQ$ and Corollary~\ref{cor:query_leaves_valid_diag}. That $\md := \mo$ must be a diagnosis w.r.t.\ $\tuple{\mo,\mb,\Tp\cup\Tp',\Tn\cup\Tn'}_\RQ$ is a direct consequence of the admissibility of $\tuple{\mo,\mb,\Tp\cup\Tp',\Tn\cup\Tn'}_\RQ$ and Definition~\ref{def:admissible}. Therefore \textsc{dLabel} must return $valid$ for each node the latest when the node becomes set-equal to $\mo$. A node that was assigned the label $valid$ and added to $\mD_{calc}$ can never be processed again during this execution of \textsc{dynamicHS} wherefore no successors of such a node can be added to $\Queue$. The same holds for some node that is labeled by $nonmin$ and added to $\mD_{\supset}$. 

Thence, the assumption that $\Queue \neq []$ forever implies that there is (at least) one node $\mathsf{node}$ that is never removed from $\Queue$. 

By Lemma~\ref{lem:node_not_processed_before_all_subset_nodes_generated}, 
each node that is a subset of or set-equal to a once processed node $\mathsf{nd}$ must have been generated before $\mathsf{nd}$ is processed.
%no node that is a subset of or set-equal to a once processed node can be (re)generated at some later point in time. 
That is, after a node is processed, it is guaranteed that no proper subsets of it can ever be processed and no subsets of it can ever be added to $\Queue$. After a node $\mathsf{nd}$ is processed and is not labeled by $valid$ or $nonmin$, $\mathsf{nd}$ is not an element of $\Queue$ anymore (cf.\ line~\ref{algoline:dyn:delete_from_queue}) and $\Queue$ comprises a set of successor nodes of $\mathsf{nd}$ where each such node corresponds to a proper superset of $\mathsf{nd}$ (cf.\ line~\ref{algoline:dyn:generate_nodes}). Consequently, a node in $\Queue$ that is processed can either be deleted whereupon no successor thereof is added to $\Queue$ (in case of pruning or labeling a node by $valid$ or $nonmin$) or be deleted whereupon proper supersets of it are added to $\Queue$ (in case of labeling a node by a conflict set). 

A (combined) replacement of a node involves the substitution of this node by another node set-equal to it. However, there can be only finitely many possibilities to construct a replacement or combined replacement node of some node since $Comb(\Queue_{dup}) \supseteq \Queue_{dup}$ also includes only nodes, i.e.\ finitely many elements. Therefore, each node in $\Queue$ can be replaced only finitely many times.

Since in each iteration of the repeat-loop in \textsc{dynamicHS} one node is processed, the cardinality of the nodes that are elements of $\Queue$ is strictly monotonically increasing.

As $\mathsf{node}$ is supposed to be never processed, we have that in each iteration of the repeat-loop, one of the other nodes in $\Queue$ must by processed. By the given argumentation, we know that after finitely many iterations, $\Queue = [\mathsf{node}]$ must be given (since all other nodes must be already pruned or labeled). Hence, $\mathsf{node}$ will be processed in the next iteration as \textsc{getFirst} in line~\ref{algoline:dyn:get_first} must catch $\mathsf{node}$, contradiction.
%
%each of the other nodes in $\Queue$ must be pruned or an element of either $\mD_{c}$  we conclude that $\mathsf{node}$ is not processed
%
%Moreover, in each iteration of the repeat-loop in \textsc{dynamicHS}, one element is removed from $\Queue$ (line~\ref{algoline:dyn:get_first}).
%
%$\md$ is a diagnosis w.r.t.\ $\tuple{\mo,\mb,\Tp,\Tn}_\RQ$ and Proposition~\ref{} which says that $\mo$ is 
%
%%
%%Algorithm~\ref{algo:hs} terminates (Proposition~\ref{prop:hs:termination}) wherefore the relevant data of the (partial) weighted pruned HS-tree $T$ produced by Algorithm~\ref{algo:hs} includes finite sets. Since \tectsc{staticHS} outputs the same relevant data as Algorithm~\ref{algo:hs} by Lemma~\ref{lem:static_hs_each_call} and in each iteration of the repeat-loop in \textsc{staticHS} at least $\Queue$ is modified, \textsc{staticHS} must terminate either due to $\Queue =\emptyset$ or because the finite second set in the relevant data of $T$  

Proposition~(1): This proposition is a direct consequence of Proposition~\ref{prop:dyn_soundness}-(2) and the stop criterion of \textsc{dynamicHS} in line~\ref{algoline:dyn:until}. 

Proposition~(2) is clear. Proposition~(3) follows from Lemma~\ref{lem:Ccalc_in_dlabel} which asserts that each element of $\mC_{calc}$ is a minimal conflict set w.r.t.\ some DPI $\tuple{\mo,\mb,\Tp\cup\Tp'',\Tn\cup\Tn''}_\RQ$ where $\Tp'' \subseteq \Tp'$ and $\Tn'' \subseteq \Tn'$. By Proposition~\ref{prop:changes_in_conflict_sets_after_testcase_added}, we obtain that each element of $\mC_{calc}$ is a conflict set w.r.t.\ the current DPI $\tuple{\mo,\mb,\Tp\cup\Tp',\Tn\cup\Tn'}_\RQ$. 

Proposition~(4): This proposition is $\true$ since \textsc{updateTree} is called at the beginning of each execution of \textsc{dynamicHS} and all elements in $\mD_{\times}$ that have not been deleted from $\mD_{\times}$ before are deleted in lines~\ref{algoline:update:reinsert_D_of_Dx_to_Q}-\ref{algoline:update:delete_from_Dtimes}. After \textsc{updateTree} has finished processing, there is no other place in \textsc{dynamicHS} where nodes can be added to $\mD_{\times}$. Hence, $\mD_{\times} = \emptyset$ must hold when \textsc{dynamicHS} terminates.

Proposition~(5): The elements of $\mD_{\supset}$ after \textsc{updateTree} at the beginning of the execution of \textsc{dynamicHS} has returned must be non-minimal diagnoses w.r.t.\ the current DPI $\tuple{\mo,\mb,\Tp\cup\Tp',\Tn\cup\Tn'}_\RQ$ by lines~\ref{algoline:update:process_Dsupset_start}-\ref{algoline:update:process_Dsupset_end} and the fact that $\mD_{\checkmark}$ comprises only diagnoses w.r.t.\ the current DPI. The latter holds by lines~\ref{algoline:inter_onto_debug:get_invalid_diags} and \ref{algoline:inter_onto_debug:update_D_checkmark} of Algorithm~\ref{algo:inter_onto_debug} where only diagnoses w.r.t.\ the current DPI $\tuple{\mo,\mb,\Tp\cup\Tp',\Tn\cup\Tn'}_\RQ$ are added to $\mD_{\checkmark}$. That only non-minimal diagnoses w.r.t.\ the current DPI can be added to $\mD_{\supset}$ during the execution of the repeat-loop is a simple implication of Lemma~\ref{lem:if_dlabel_returns_valid_nonmin_then}-(4). 

Proposition~(6) is a consequence of lines~\ref{algoline:dyn:check_node_already_in_Q}-\ref{algoline:dyn:add_to_Qdup}, the definition of de-facto non-redundancy (Definition~\ref{def:de-facto_non-redundant}) and Lemma~\ref{lem:pruneQdup}.
\end{proof} 
\newgeometry{margin=2cm}

\begin{algorithm*}
\small
\caption{Iterative Construction of a Dynamic Hitting Set Tree} \label{algo:inter_dyn_hs}
\begin{algorithmic}[1]
\Require a tuple $\tuple{ \langle\mo,\mb,\Tp,\Tn\rangle_\RQ, \Queue, \Queue_{dup}, t, n_{\min}, n_{\max}, \mathbf{C}_{calc}, \mD_{\checkmark}, \mD_{\times}, p(), \Tp', \Tn', \mD_{\supset} }$ consisting of
\begin{itemize}
	\item the DPI $\langle\mo,\mb,\Tp,\Tn\rangle_\RQ$ given as input to Algorithm~\ref{algo:inter_onto_debug},
	\item the overall sets of positively ($\Tp'$) and negatively ($\Tn'$) answered queries added as test cases to $\langle\mo,\mb,\Tp,\Tn\rangle_\RQ$ so far, 
	\item a queue $\Queue$ of open (non-labeled) nodes, 
	\item some computation timeout $t$,
	\item a desired minimal ($n_{\min}\geq2$) and maximal ($n_{\max}$) number of minimal diagnoses to be returned, 
	\item a set $\mathbf{C}_{calc}$ of conflict sets w.r.t.\ the current DPI $\langle\mo,\mb,\Tp\cup\Tp',\Tn\cup\Tn'\rangle_\RQ$,
	%of calculated conflict sets so far (each $\mc \in \mC_{calc}$ is a conflict set w.r.t.\ the current DPI 
	%minimal conflict set w.r.t.\ some intermediate DPI), 
	\item a set $\mD_{\checkmark}$ of minimal diagnoses w.r.t.\ the current DPI $\langle\mo,\mb,\Tp\cup\Tp',\Tn\cup\Tn'\rangle_\RQ$,
	\item a set $\mD_{\times}$ of minimal diagnoses w.r.t.\ the last-but-one DPI that are invalidated by the most recently added test case, 
	\item a function $p: \mo \rightarrow (0,0.5)$,  
	\item a set $\mD_{\supset}$ of non-minimal diagnoses w.r.t.\ the last-but-one DPI and
	\item a set $\Queue_{dup}$ of stored (duplicate) nodes $\mathsf{nd}$ that can be used when it comes to constructing a replacement node of a pruned node $\mathsf{nd}' \supseteq \mathsf{nd}$ after tree pruning.
\end{itemize}
 
%

%

%
 
%

%
%(where $\mc\in\mathbf{C}_{calc}$ implies $\mc\in\minC_{\langle\mo,\mb,\Tp\cup\Tp'',\Tn\cup\Tn''\rangle_\RQ}$ for $\emptyset \subseteq \Tp'' \subseteq \Tp'$ and $\emptyset \subseteq \Tn'' \subseteq \Tn'$), 
%

%

%

%

%$\langle\mo,\mb,\Tp\cup\Tp',\Tn\cup\Tn'\rangle_\RQ$ 
%that are nodes (paths) in the existing partial hitting set tree
%
%a DPI $\langle\mo,\mb,\Tp,\Tn\rangle_\RQ$, 
%queues of open ($\Queue$) and closed ($Q_{closed}$) nodes computed by prior calls to $\scHS$, some computation timeout $t$, a set $\mathbf{C}_{calc}$ of already calculated minimal conflict sets throughout prior calls to $\scHS$, a set of known valid diagnoses (nodes) $\mD_{\checkmark}$, a desired minimal ($n_{\min}$) and maximal ($n_{\max}$) number of diagnoses to be returned per call to $\scHS$, a weight function $w(\tax) \in \mathbb{R}$ that assigns to each $\tax \in \mo$ a weight and thereby determines the search strategy (breadth-first or uniform-cost), a mode $mode$ (static or dynamic) that influences tree pruning
\Ensure
a tuple $\tuple{\mD_{calc},\Queue, \mathbf{C}_{calc}, \mD_{\times}, \mD_{\supset}}$ where
\begin{itemize}
\item $\mD_{calc}$ is the current set of leading diagnoses such that
\begin{enumerate}[(a)]
\item $\mD_{calc} \subseteq \minD_{\langle\mo,\mb,\Tp\cup\Tp',\Tn\cup\Tn'\rangle_\RQ}$ is the set of most probable minimal diagnoses w.r.t.\ $\langle\mo,\mb,\Tp\cup\Tp',\Tn\cup\Tn'\rangle_\RQ$ such that 
\begin{enumerate}[(i)]
\item $n_{\min} \leq |\mD_{calc}| \leq n_{\max}$ and 
\item $\mD_{calc}\setminus\mD_{\checkmark} \neq \emptyset$,
\end{enumerate}
if such a set $\mD_{calc}$ 
%with (i) and (ii)
 exists,
or 
\item $\mD_{calc}$ is equal to the set of all minimal diagnoses $\minD_{\langle\mo,\mb,\Tp\cup\Tp',\Tn\cup\Tn'\rangle_\RQ}$, otherwise,
\end{enumerate}
where ``most-probable'' refers to the probability measure $p_{nodes}()$ (cf.\ Definition~\ref{def:p_node()}) obtained from the given function $p()$;
%
%$p(\md), \md\in\minD_{\langle\mo,\mb,\Tp\cup\Tp',\Tn\cup\Tn'\rangle_\RQ}$ obtained from $p(\tax),\tax\in\mo$ by application of Formulas~\ref{eq:diag_prob_calc} and \ref{eq:diag_prob_norm}.
%where ``most-probable'' refers to the probability measure $p_{\mD}(\md), \md\in\mD$ obtained from $p_{\mo}(\tax),\tax\in\mo$ which is transformed to some probability measure $p_{\mD,prio}(\md),\md\in\mD$ by Formula~\ref{eq:diag_prob_calc} which is in turn used to calculate $p_{\mD}(\md), \md\in\mD$ by Formula~\ref{eq:bayes} (see \emph{Computing a Probability Distribution of Leading Diagnoses} on page~\pageref{etc:computing_prob_dist_of_leading_diags}). 
\item $\Queue$ is the current queue of open (non-labeled) nodes of the hitting set tree,
%, i.e.\ a set of nodes that are not labeled.
%have not yet been processed and labeled. 
\item $\mathbf{C}_{calc}$ is a set of conflict sets w.r.t.\ the current DPI $\langle\mo,\mb,\Tp\cup\Tp',\Tn\cup\Tn'\rangle_\RQ$,
%is the overall set of computed minimal conflict sets throughout all calls to \textsc{dynamicHS} during the execution of Algorithm~\ref{algo:inter_onto_debug} where each $\mc\in\mathbf{C}_{calc}$ is a minimal conflict set w.r.t.\ some intermediate DPI (not necessarily one w.r.t.\ the input DPI).  
\item $\mD_{\times} = \emptyset$, 
%is the overall set of computed minimal diagnoses throughout all calls to \textsc{dynamicHS} during the execution of Algorithm~\ref{algo:inter_onto_debug} where each $\md\in\mD_{\times}$ has been invalidated by some test case in $\Tp'\cup\Tn'$, but has not been pruned, i.e.\ completely deleted from all sets storing nodes in \textsc{dynamicHS}.
%%%%%%%%%%%%%%%%%%%
%\item $\mD_{\supset}$ is a set of non-minimal diagnoses w.r.t.\ the current DPI $\langle\mo,\mb,\Tp\cup\Tp',\Tn\cup\Tn'\rangle_\RQ$ such that for each $\mathsf{nd} \in \mD_{\supset}$ there is some $\md \in \mD_{calc}$ satisfying $\mathsf{nd} \supset \md$ 
%%%%%%%%%%%%%%%%%%%
\item $\mD_{\supset}$ is the set of all processed nodes so far throughout the execution of Algorithm~\ref{algo:inter_onto_debug} that are non-minimal diagnoses w.r.t.\ the current DPI $\langle\mo,\mb,\Tp\cup\Tp',\Tn\cup\Tn'\rangle_\RQ$ and
\item $\Queue_{dup}$ includes a node set-equal to $X$ for a set $X \subseteq \mo$ iff 
\begin{itemize}
	\item $\mathsf{nd} = X$ is a generated node that is de-facto non-redundant w.r.t.\ the current DPI $\langle\mo,\mb,\Tp\cup\Tp',\Tn\cup\Tn'\rangle_\RQ$, such that, at generation time of $\mathsf{nd}$, there was a node set-equal to $X$ in $\Queue$ or 
	\item there is a de-facto non-redundant node $\mathsf{nd}' = X$ w.r.t.\ the current DPI $\langle\mo,\mb,\Tp\cup\Tp',\Tn\cup\Tn'\rangle_\RQ$ which is a combined equal node of some generated node $\mathsf{nd}''$ that has been added to $\Queue_{dup}$.
\end{itemize}
\end{itemize}
%%%%%%%%%%%%%%%%%%%%%%%%%%%%%%%%%%% OLD - start %%%%%%%%%%%%%%%%%%%%%%%%%%%%
%a tuple $\tuple{\mD,\Queue, \mathbf{C}_{calc}, \mD_{\times}, \mD_{\supset}}$ where 
%\begin{itemize}
%\item (a)~$\mD \subset \minD_{\langle\mo,\mb,\Tp\cup\Tp',\Tn\cup\Tn'\rangle_\RQ}$ is the set of most probable (according to $p()$) minimal diagnoses w.r.t.\ $\langle\mo,\mb,\Tp\cup\Tp',\Tn\cup\Tn'\rangle_\RQ$ such that (i)~$n_{\min} \leq |\mD| \leq n_{\max}$ and (ii)~$\mD\supset\mD_{\checkmark}$, if such a set $\mD$ with (i) and (ii) exists;
%or \newline (b)~the set 
%%of all minimal diagnoses 
%$\minD_{\langle\mo,\mb,\Tp\cup\Tp',\Tn\cup\Tn'\rangle_\RQ}$, otherwise
%\item $\Queue$ is the current queue of open nodes
%\item $\mathbf{C}_{calc}$ is the overall set of computed minimal conflict sets (where $\mc\in\mathbf{C}_{calc}$ implies $\mc\in\minC_{\langle\mo,\mb,\Tp\cup\Tp'',\Tn\cup\Tn''\rangle_\RQ}$ for $\Tp \subseteq \Tp'' \subseteq \Tp'$ and $\Tn \subseteq \Tn'' \subseteq \Tn'$)
%\item $\mD_{\times}$ is the overall set of non-pruned, invalidated minimal diagnoses 
%\item $\mD_{\supset}$ the overall set of computed non-minimal nodes (paths)  
%%a set $\mD$ of minimal diagnoses w.r.t.\ $\langle\mo,\mb,\Tp,\Tn\rangle_\RQ$ such that $n_{\min} \leq |\mD| \leq n_{\max}$ and $\mD \setminus \mD_{\checkmark} \neq \emptyset$
%\end{itemize}
%%%%%%%%%%%%%%%%%%%%%%%%%%%%%%% OLD - end %%%%%%%%%%%%%%%%%%%%%%%%%%%%%%%%%%
\vspace{10pt}
\Procedure{dynamicHS}{$\langle\mo,\mb,\Tp,\Tn\rangle_\RQ, \Queue, \Queue_{dup}, t, n_{\min}, n_{\max}, \mathbf{C}_{calc}, \mD_{\checkmark}, \mD_{\times}, p(), \Tp', \Tn', \mD_{\supset}$}
\State $t_{start} \gets \Call{getTime}{ }$
\State $\mD_{calc} \gets \emptyset$\label{algoline:dyn:Dcalc_gets_emptyset}
\State $\tuple{\Queue, \mD_{\times}, \mD_{\supset}, \mC_{calc}, \Queue_{dup}} \gets \Call{updateTree}{\langle\mo,\mb,\Tp,\Tn\rangle_\RQ, \mD_{\times}, \Queue, \Queue_{dup}, \mD_{\supset},\mD_{\checkmark}, \mC_{calc}, p(), \Tp', \Tn'}$\label{algoline:dyn:update_tree}     
\Repeat	\label{algoline:dyn:repeat}																																	\Comment{\textsc{updateTree} (see Algorithm~\ref{algo:update_tree})}
\State $\mathsf{node} \gets \Call{getFirst}{\Queue}$\label{algoline:dyn:get_first}			  \Comment{$\mathsf{node}$ is processed}		
%\State $\Queue \gets \Queue \setminus \setof{\mathsf{node}}$\label{algoline:dyn:delete_from_queue}
\State $\Queue \gets \Call{deleteFirst}{\Queue}$\label{algoline:dyn:delete_from_queue}
\If{$\mathsf{node} \in \mD_{\checkmark}$}\label{algoline:dyn:node_in_Dcheckmark}  \Comment{$\mD_{\checkmark}$ includes only minimal diagnoses w.r.t.\ current DPI}
	\State $L \gets valid$\label{algoline:dyn:set_L_to_valid}               
\Else
	\State $\tuple{L,\mathbf{C}_{calc},\Queue_{dup}} \gets \Call{dLabel}{\langle\mo,\mb,\Tp,\Tn\rangle_\RQ, \mathsf{node}, \mathbf{C}_{calc}, \mD_{calc}, \Queue, \Queue_{dup}, p(), \Tp', \Tn'}$\label{algoline:dyn:dlabel}
	%\State $\mathbf{C}_{calc} \gets \mathbf{C}$
\EndIf
\If{$L = valid$}\label{algoline:dyn:if_L_valid}  \Comment{\textsc{dLabel} (see Algorithm~\ref{algo:update_tree})}
	\State $\mD_{calc} \gets \mD_{calc} \cup \setof{\mathsf{node}}$\label{algoline:dyn:add_to_Dcalc}      \Comment{$\mathsf{node}$ is a minimal diagnosis w.r.t.\ current DPI}
\ElsIf{$L = nonmin$}								
	\State $\mD_{\supset} \gets \mD_{\supset} \cup \setof{\mathsf{node}}$ \label{algoline:dyn:add_to_Dsupset}  \Comment{$\mathsf{node}$ is a non-minimal diagnosis w.r.t.\ current DPI}
%\ElsIf{$L = dup$}
%				\Comment{do nothing}
\Else 	
	\For{$e \in L$}\label{algoline:dyn:for_e_in_L}            \Comment{$L$ is a minimal conflict set w.r.t.\ current DPI}
		\State $\mathsf{node}_e \gets \Call{add}{\mathsf{node},e}$ \label{algoline:dyn:add_ax_to_node}       \Comment{$\mathsf{node}_e$ is generated}   
		\State $\mathsf{node}_{e}.\mathsf{cs} \gets \Call{add}{\mathsf{node.cs},L}$ \label{algoline:dyn:add_cs_to_node.cs}
		\If{$\mathsf{node}_e \in \Queue$}   \label{algoline:dyn:check_node_already_in_Q}                      \Comment{$\mathsf{node}_e$ is a (set-equal) duplicate of a node in $\Queue$}
			\State $\Queue_{dup} \gets \Call{insertSorted}{ \mathsf{node}_e, \Queue_{dup}, cardinality, ascending}$ \label{algoline:dyn:add_to_Qdup}
		\Else
			\State $\Queue \gets \Call{insertSorted}{ \mathsf{node}_e, \Queue, p_{nodes}(), descending}$\label{algoline:dyn:generate_nodes}
		\EndIf
			%
		%\For{$\mathsf{nd} \in \Queue$}
			%\State $dup \gets \false$
			%\If{$\mathsf{nd} = \mathsf{node}_e$}
				%\State $\Queue_{dup} \gets \Queue_{dup} \cup \setof{\mathsf{node}_e}$
				%\State $dup \gets \\true$
				%\State \textbf{break}
			%\EndIf
		%\EndFor
			%\If{$dup = \false$}
				%\State $\Queue \gets \Call{insertSorted}{ \mathsf{node}_e, \Queue, p()}$\label{algoline:dyn:generate_nodes}
			%\EndIf
%
	\EndFor
\EndIf
\Until{$\Queue= [] \lor [\mD_{calc}\setminus\mD_{\checkmark} \neq \emptyset \land \left|\mD_{calc}\right| \geq n_{\min} \land ( |\mD_{calc}| = n_{\max} \lor \Call{getTime}{ } - t_{start} > t)]$}\label{algoline:dyn:until}
\State \Return $\tuple{\mD_{calc} ,\Queue, \mathbf{C}_{calc}, \mD_{\times}, \mD_{\supset}, \Queue_{dup}}$ \label{algoline:dyn:return}
\EndProcedure
\algstore{save_algo_dyn_hs}
\end{algorithmic}
\normalsize
\end{algorithm*}

\begin{algorithm*}
\small
\caption{Iterative Construction of a Dynamic Hitting Set Tree (continued)} \label{algo:update_tree}
\begin{algorithmic}[1]
\algrestore{save_algo_dyn_hs}
\Procedure{\textsc{dLabel}}{$\langle\mo,\mb,\Tp,\Tn\rangle_\RQ,\mathsf{node},\mathbf{C}_{calc},\mD_{calc}, \Queue, \Queue_{dup}, p(), \Tp', \Tn'$} \Comment{\textsc{dLabel} (see page~\pageref{lem:if_dlabel_returns_valid_nonmin_then})}
\For{$\mathsf{nd} \in \mD_{calc}$}\label{algoline:dlabel:non-min_crit_start}
	\If{$\mathsf{node} \supset \mathsf{nd}$}    \Comment{$\mathsf{node}$ is a non-minimal diagnosis}
			\State \Return $\tuple{nonmin,\mathbf{C}_{calc},\Queue_{dup}}$
	\EndIf
\EndFor\label{algoline:dlabel:non-min_crit_end}
%\For{$\mathsf{nd} \in \Queue \cup \mD_{\checkmark}$}\label{algoline:dlabel:dup_crit_start}
	%\If{$\mathsf{node} = \mathsf{nd}$}      \Comment{no duplicates}
			%\State \Return $\tuple{dup,\mathbf{C}_{calc}}$
	%\EndIf
%\EndFor\label{algoline:dlabel:dup_crit_end}
\For{$\mc \in \mathbf{C}_{calc}$}\label{algoline:dlabel:reuse_start} \Comment{$\mC_{calc}$ includes only conflict sets w.r.t.\ current DPI}
	\If{$\mc \cap \mathsf{node} = \emptyset$}\label{algoline:dlabel:if_C_cap_node=emptyset}    \Comment{reuse (a subset of) $\mc$ to label $\mathsf{node}$} 
		\State $X \gets \Call{\scQX}{\langle\mc,\mb,\Tp\cup\Tp',\Tn\cup\Tn'\rangle_\RQ}$\label{algoline:dlabel:qx_1} \Comment{Algorithm~\ref{algo:qx} (page~\pageref{algo:qx}) to test if $\mc$ is minimal w.r.t.\ current DPI} 
		\If{$X = \mc$} \label{algoline:dlabel:if_X=C}
			\State \Return $\tuple{\mc,\mathbf{C}_{calc},\Queue_{dup}}$\label{algoline:dlabel:return_C}
		\Else      \Comment{$X \subset \mc$}
			%\State $\tuple{\Queue_{dup},\_} \gets \Call{prune}{X,\Queue_{dup},\emptyset,\emptyset}$\label{algoline:dlabel:call_prune_Qdup}
			\State $\Queue_{dup} \gets \Call{pruneQdup}{X,\Queue_{dup}}$\label{algoline:dlabel:call_prune_Qdup}     \Comment{\textsc{pruneQdup} (see Algorithm~\ref{algo:prune})}
			%\State $\tuple{\Queue,React\_\Queue} \gets \Call{prune}{X,\Queue,\Queue_{dup},p()}$\label{algoline:dlabel:call_prune_Q}
			\State $\Queue \gets \Call{prune}{X,\Queue,\Queue_{dup},p_{nodes}()}$\label{algoline:dlabel:call_prune_Q}     \Comment{\textsc{prune} (see Algorithm~\ref{algo:prune})}
			%\State $\tuple{\mD_{\supset},React\_\mD_{\supset}} \gets \Call{prune}{X,\mD_{\supset},\Queue_{dup},\emptyset}$\label{algoline:dlabel:call_prune_Dsupset}
			\State $\mD_{\supset} \gets \Call{prune}{X,\mD_{\supset},\Queue_{dup},\emptyset}$\label{algoline:dlabel:call_prune_Dsupset}
			%
			%\State $\tuple{\Queue,\mD_{\times},\mD_{\supset}} \gets \Call{prune}{X,\Queue,\mD_{\times},\mD_{\supset}}$\label{algoline:dlabel:call_prune1}
			%
			%\State $\Queue_{dup} \gets \Queue_{dup} \setminus (React\_\Queue \cup React\_\mD_{\supset})$
			\State $\mathbf{C}_{calc} \gets \Call{addSetDelSupsets}{X, \mathbf{C}_{calc}}$\label{algoline:dlabel:add_set_del_supset} \Comment{add $X$ to $\mC_{calc}$ and delete all its supersets from $\mC_{calc}$}
			\State \Return $\tuple{X,\mathbf{C}_{calc},\Queue_{dup}}$ \label{algoline:dlabel:return_X}
		\EndIf
	%\fixme{after reuse, call $\scQX(\mc,\mb,\Tp\cup\Tp',\Tn\cup\Tn'\rangle_\RQ)$ to check whether $\mc$ is indeed a \emph{minimal} CS w.r.t. the new DPI, if not, update $\mC_{calc}$ -----
	%Prop: Any min CS for some DPI remains a CS for a new DPI (with test cases added) by monotonicity, but not necessarily a minimal one -----
	%Thus reuse and then check for minimality by QX, if minimal, then OK, otherwise}
		%\State \Return $\tuple{\mc,\mathbf{C}_{calc}}$
	\EndIf
\EndFor\label{algoline:dlabel:reuse_end}
\State $L\gets \Call{QX}{\langle\mo\setminus\mathsf{node},\mb,\Tp\cup\Tp',\Tn\cup\Tn'\rangle_\RQ}$\label{algoline:dlabel:qx_2} \Comment{Algorithm~\ref{algo:qx} (page~\pageref{algo:qx}) to test if $\mathsf{node}$ is a diagnosis}
\If{$L$ = \text{'no conflict'}}						\Comment{$\mathsf{node}$ is a diagnosis}
	\State \Return $\tuple{valid,\mathbf{C}_{calc},\Queue_{dup}}$\label{algoline:dlabel:return_valid}
\Else						\Comment{$L$ is a \emph{new} minimal conflict set ($\notin \mathbf{C}_{calc}$)}
	\State $\mathbf{C}_{calc} \gets \mathbf{C}_{calc} \cup \setof{L}$\label{algoline:dlabel:add_new_cs}
	\State \Return $\tuple{L,\mathbf{C}_{calc},\Queue_{dup}}$\label{algoline:dlabel:return_new_cs}
\EndIf
\EndProcedure
\vspace{10pt}
\Procedure{\textsc{updateTree}}{$\langle\mo,\mb,\Tp,\Tn\rangle_\RQ,\mD_{\times}, \Queue, \Queue_{dup}, \mD_{\supset}, \mD_{\checkmark}, \mC_{calc}, p(), \Tp', \Tn'$}
\For{$\mathsf{nd} \in \mD_{\times}$} \label{algoline:update:process_Dtimes_start}
	%\If{}
		\State $quickRC, completeRC \gets \false$
		\State $X \gets \Call{QX}{\langle U_{\mathsf{nd.cs}}\setminus\mathsf{nd},\mb,\Tp\cup\Tp',\Tn\cup\Tn'\rangle_\RQ}$\label{algoline:update:qx} \Comment{QRC begin}
		\For{$\mc \in \mathsf{nd.cs}$} % if a subset X of a CS in the path of nd is found, then nd is DEFINITELY pruned, since X cannot include any formulas of nd.path since QX is called with ``setminus nd.path''
		% if some X is found that is not a subset of a CS, then X must be a subset of the union of the CSs in nd.cs --> this might still yield new conflict sets for the new DPI --> so put nd to the open nodes again
		% if ``no conflicts'' is returned by QX, then put nd to the open nodes
			\If{$X \subset \mc$}\label{algoline:update:X_subset_C_(QRC)}												\Comment{QRC (see page~\pageref{lem:quick_prune_check})}
				\State $quickRC \gets \true$ \label{algoline:update:qrc_gets_true}
				\State \textbf{break}  \Comment{QRC end}
			\EndIf
		\EndFor\label{algoline:update:quickPC_end} 
		\If{$quickRC = \false$}											\Comment{CRC begin}
			\For{$i \leftarrow 1,\dots,|\mathsf{nd}|$}\label{algoline:update:completePC_start} 
				\State $X \gets \Call{QX}{\langle \mathsf{nd.cs}[i]\setminus \setof{\mathsf{nd}[i]},\mb,\Tp\cup\Tp',\Tn\cup\Tn'\rangle_\RQ}$\label{algoline:update:qx1}  \Comment{CRC (see page~\pageref{lem:complete_prune_check})}
				\If{$X \neq \text{'no conflict'}$}      
					\State $completeRC \gets \true$     
					\State \textbf{break} \Comment{CRC end}
				\EndIf
			\EndFor\label{algoline:update:completePC_end} 
		\EndIf
		\If{$quickRC = \true \;\lor\; completeRC = \true$}\label{algoline:update:if_qrc_or_crc_true}  \Comment{condition $\true$ iff $\mathsf{nd}$ redundant w.r.t.\ current DPI}
			%\State $\tuple{\Queue_{dup},\_} \gets \Call{prune}{X,\Queue_{dup},\emptyset,\emptyset}$\label{algoline:update:call_prune_Qdup}
			\State $\Queue_{dup} \gets \Call{pruneQdup}{X,\Queue_{dup}}$\label{algoline:update:call_prune_Qdup}     \Comment{\textsc{pruneQdup} (see Algorithm~\ref{algo:prune})}
			%\State $\tuple{\Queue,React\_\Queue} \gets \Call{prune}{X,\Queue,\Queue_{dup},p()}$\label{algoline:update:call_prune_Q}
			\State $\Queue \gets \Call{prune}{X,\Queue,\Queue_{dup},p_{nodes}()}$\label{algoline:update:call_prune_Q} \Comment{\textsc{prune} (see Algorithm~\ref{algo:prune})}
			%\State $\tuple{\mD_{\times},React\_\mD_{\times}} \gets \Call{prune}{X,\mD_{\times},\Queue_{dup},\emptyset}$\label{algoline:update:call_prune_Dtimes}
			\State $\mD_{\times} \gets \Call{prune}{X,\mD_{\times},\Queue_{dup},\emptyset}$\label{algoline:update:call_prune_Dtimes}
			%\State $\tuple{\mD_{\supset},React\_\mD_{\supset}} \gets \Call{prune}{X,\mD_{\supset},\Queue_{dup},\emptyset}$\label{algoline:update:call_prune_Dsupset}
			\State $\mD_{\supset} \gets \Call{prune}{X,\mD_{\supset},\Queue_{dup},\emptyset}$\label{algoline:update:call_prune_Dsupset}
			%
			%\State $\tuple{\Queue,\mD_{\times},\mD_{\supset}} \gets \Call{prune}{X,\Queue,\mD_{\times},\mD_{\supset}}$\label{algoline:dlabel:call_prune2} 
			%
			%\State $\Queue_{dup} \gets \Queue_{dup} \setminus (React\_\Queue \cup React\_\mD_{\times} \cup React\_\mD_{\supset})$
			\State $\mC_{calc} \gets \Call{addSetDelSupsets}{X, \mathbf{C}_{calc}}$\label{algoline:update:add_set_del_supset} \Comment{add $X$ to $\mC_{calc}$ and delete all its supersets from $\mC_{calc}$} 
				%\fixme{update $\mC_{calc}$ also, i.e. delete all supersets of $X$ from it and add $X$}
				%\If{$\mathsf{nd} \notin \mD_{\times}$}
					%\State $pruned \gets \true$
					%\State \textbf{break}
				%\Else 
				
				%\EndIf
		\EndIf
		%\EndFor
	\EndFor
\For{$\mathsf{nd} \in \mD_{\times}$}\label{algoline:update:reinsert_D_of_Dx_to_Q}
	%\If{$pruned = \false$} 
	\Comment{add all (non-pruned) nodes in $\mD_{\times}$ to $\Queue$}
		\State $\Queue \gets \Call{insertSorted}{ \mathsf{nd}, \Queue, p_{nodes}(), descending}$\label{algoline:update:insert_sorted_0}
		\State $\mD_{\times} \gets \mD_{\times} \setminus \setof{\mathsf{nd}}$ \label{algoline:update:delete_from_Dtimes}
	%\EndIf
\EndFor \label{algoline:update:process_Dtimes_end}
% for all nd in \mD_{\supset} check if there is still a diag in D_checkmark that is a subset of nd --> if yes, then do nothing, if no, then delete nd from \mD_{\supset} and add it to queue
\For{$\mathsf{nd} \in \mD_{\supset}$}\label{algoline:update:process_Dsupset_start} \Comment{update $\mD_{\supset}$: add all nodes to $\Queue$ which are not proper supersets of a diagnosis in $\mD_{\checkmark}$}
	\State $nonmin \gets \false$
	\For{$\mathsf{nd}' \in \mD_{\checkmark}$}
		\If{$\mathsf{nd} \supset \mathsf{nd}'$}    
			\State $nonmin \gets \true$
			\State \textbf{break} 
		\EndIf
	\EndFor
	\If{$nonmin = \false$}
		\State $\Queue \gets \Call{insertSorted}{ \mathsf{nd}, \Queue, p_{nodes}(), descending}$\label{algoline:update:insert_sorted_0.5}
		\State $\mD_{\supset} \gets \mD_{\supset} \setminus \setof{\mathsf{nd}}$\label{algoline:update:delete_from_Dsupset}
	\EndIf
\EndFor \label{algoline:update:process_Dsupset_end}
\For{$\md \in \mD_{\checkmark}$}\label{algoline:update:process_Dcheckmark_start}  \Comment{reinsert known minimal diagnoses to $\Queue$ to find diagnoses in order of descending $p_{nodes}()$}
	\State $\Queue \gets \Call{insertSorted}{ \md, \Queue, p_{nodes}(), descending}$\label{algoline:update:insert_sorted_1}
\EndFor \label{algoline:update:process_Dcheckmark_end}
\State \Return $\tuple{\Queue, \mD_{\times}, \mD_{\supset}, \mC_{calc}, \Queue_{dup}}$
\EndProcedure
\vspace{10pt}
%
%\Procedure{\textsc{prune}}{$X,\Queue,\mD_{\times},\mD_{\supset}$}
%\For{$\mathsf{nd} \in \Queue \cup \mD_{\times} \cup \mD_{\supset}$}
	%\For{$i=1$ to $|\mathsf{nd.cs}|$}
	%%\For{$CS \in \mathsf{nd.cs}$}
		%\If{$\mathsf{nd.cs}[i]\supset X \land \mathsf{nd}[i] \in \mathsf{nd.cs}[i]\setminus X$}    \Comment{Lemma~\ref{lem:pruning}}
			%\State $\Queue \gets \Queue \setminus \setof{\mathsf{nd}}$
			%\State $\mD_{\times} \gets \mD_{\times} \setminus \setof{\mathsf{nd}}$ \label{algoline:prune:delete_from_Dtimes}
			%\State $\mD_{\supset} \gets \mD_{\supset} \setminus \setof{\mathsf{nd}}$
		%\EndIf
	%\EndFor
%\EndFor
%\State \Return $\tuple{\Queue,\mD_{\times},\mD_{\supset}}$
%\EndProcedure
\algstore{save_algo_dyn_hs1}
\end{algorithmic}
\normalsize
\end{algorithm*}

\begin{algorithm*}
\small
\caption{Iterative Construction of a Dynamic Hitting Set Tree (continued)} \label{algo:prune}
\begin{algorithmic}[1]
\algrestore{save_algo_dyn_hs1}
\Procedure{\textsc{prune}}{$X,S,Dup,sort\_measure$}    \Comment{\textsc{prune} (see page~\pageref{lem:prune})}
\If{$S$ is a list}
	\State $S' \gets []$
\Else
	\State $S' \gets \emptyset$
\EndIf
\For{$\mathsf{nd} \in S$}\label{algoline:prune:for_nd_in_S}
	\State $k \gets 0$ \label{algoline:prune:k_gets_0}
	\For{$i=1$ to $|\mathsf{nd.cs}|$}\label{algoline:prune:for_cs_in_nd.cs}
		\If{$\mathsf{nd.cs}[i]\supset X$}\label{algoline:prune:redundancy_check_part1}     \Comment{check first redundancy criterion (Definition~\ref{def:redundant_node} on page~\pageref{def:redundant_node})}  
			\If{$\mathsf{nd}[i] \in \mathsf{nd.cs}[i]\setminus X$}\label{algoline:prune:redundancy_check_part2}    \Comment{check second redundancy criterion (Definition~\ref{def:redundant_node} on page~\pageref{def:redundant_node})}
				\State $k \gets i$\label{algoline:prune:k_gets_i}
			\Else
				\State $\mathsf{nd.cs}[i] \gets X$\label{algoline:prune:nd.cs[i]_gets_X}          \Comment{replace each superset of $X$ in $\mathsf{nd.cs}$ by $X$}
			\EndIf
		\EndIf
	\EndFor
	\If{$k > 0$}\label{algoline:prune:if_k>0}                                    \Comment{$\mathsf{nd}$ is redundant}
		%\State $S \gets S \setminus \setof{\mathsf{nd}}$ \label{algoline:prune:delete_from_S}
		\For{$\mathsf{node} \gets Dup[1],\dots,Dup[|Dup|]$}\label{algoline:prune:check_for_alternative_paths_start}
				\If{$|\mathsf{node}| \geq k \land \mathsf{nd}[1..|\mathsf{node}|] = \mathsf{node}$}\label{algoline:prune:check_if_alternative_subnode}
					\State $\mathsf{nd}_{new} \gets \Call{add}{\mathsf{node},\mathsf{nd}[|\mathsf{node}|+1..|\mathsf{nd}|]}$\label{algoline:prune:construct_alternative_equal_node}     \Comment{construct replacement node $\mathsf{nd}_{new}$ of $\mathsf{nd}$}
					\State $\mathsf{nd}_{new}.cs \gets \Call{add}{\mathsf{node.cs},\mathsf{nd.cs}[|\mathsf{node}|+1..|\mathsf{nd}|]}$\label{algoline:prune:construct_alternative_equal_node.cs}
					\State $S' \gets \Call{insertSorted}{ \mathsf{nd}_{new}, S', sort\_measure, descending}$\label{algoline:prune:insert_alternative_equal_node_into_S'}
					\State \textbf{break}\label{algoline:prune:break1}
				\EndIf
		\EndFor \label{algoline:prune:check_for_alternative_paths_end}
	\Else     \Comment{$X$ is not a witness of redundancy of $\mathsf{nd}$}
		\State $S' \gets \Call{insertSorted}{ \mathsf{nd}, S', sort\_measure, descending}$\label{algoline:prune:insert_same_node_into_S'}
	\EndIf
\EndFor
\State \Return $S'$ 
\EndProcedure
\vspace{10pt}
\Procedure{\textsc{pruneQdup}}{$X,Dup$}    \Comment{\textsc{pruneQdup} (see page~\pageref{lem:pruneQdup})}
%\State $Dup \gets \Call{sort}{Dup, cardinality, ascending}$\label{algoline:pruneQdup:sort}
\State $Dup_{new} \gets []$
\For{$i \gets 1$ to $|Dup|$}\label{algoline:pruneQdup:for_i_in_Dup}
	\State $\mathsf{ndi} \gets Dup[i]$\label{algoline:pruneQdup:ndi_gets_Dup[i]}
	\State $k \gets 0$\label{algoline:pruneQdup:k_gets_0}
	\For{$m\gets 1$ to $|\mathsf{ndi.cs}|$}\label{algoline:pruneQdup:for_m_gets_1_to_ndi.cs_start}%\label{algoline:pruneQdup:}
		%\If{$\mathsf{ndi.cs}[m]\supset X \land \mathsf{ndi}[m] \in \mathsf{ndi.cs}[m]\setminus X$}\label{algoline:pruneQdup:redundancy_check}    %\Comment{Lemma~\ref{lem:pruning}}
			%\State $k \gets m$\label{algoline:pruneQdup:k_gets_m}
		%\EndIf
		\If{$\mathsf{ndi.cs}[m]\supset X$}\label{algoline:pruneQdup:redundancy_check_part1}    \Comment{check first redundancy criterion (Definition~\ref{def:redundant_node} on page~\pageref{def:redundant_node})}
			\If{$\mathsf{ndi}[m] \in \mathsf{ndi.cs}[m]\setminus X$}\label{algoline:pruneQdup:redundancy_check_part2}    \Comment{check second redundancy criterion (Definition~\ref{def:redundant_node} on page~\pageref{def:redundant_node})}
				\State $k \gets m$\label{algoline:pruneQdup:k_gets_m}
			\Else
				\State $\mathsf{ndi.cs}[m] \gets X$\label{algoline:pruneQdup:ndi.cs[m]_gets_X}    \Comment{replace each superset of $X$ in $\mathsf{ndi.cs}$ by $X$}
			\EndIf
		\EndIf
	\EndFor\label{algoline:pruneQdup:endfor_m_gets_1_to_ndi.cs}
	\If{$k > 0$}\label{algoline:pruneQdup:if_k>0}                          \Comment{$\mathsf{ndi}$ is redundant}
		%\State $Dup \gets Dup \setminus \setof{\mathsf{ndi}}$
		\For{$\mathsf{ndj} \in Dup_{new}$}\label{algoline:pruneQdup:for_ndj_in_Dupnew}
			%\State $\mathsf{ndj} \gets Dup[i]$
			\If{$|\mathsf{ndj}| \geq k \land \mathsf{ndi}[1..|\mathsf{ndj}|] = \mathsf{ndj}$}\label{algoline:pruneQdup:check_if_alternative_subnode}
				\State $\mathsf{ndi}_{new} \gets \Call{add}{\mathsf{ndj},\mathsf{ndi}[|\mathsf{ndj}|+1..|\mathsf{ndi}|]}$\label{algoline:pruneQdup:construct_alternative_equal_node}   \Comment{construct combined replacement node $\mathsf{ndi}_{new}$ of $\mathsf{ndi}$}
				\State $\mathsf{ndi}_{new}.cs \gets \Call{add}{\mathsf{ndj.cs},\mathsf{ndi.cs}[|\mathsf{ndj}|+1..|\mathsf{ndi}|]}$\label{algoline:pruneQdup:construct_alternative_equal_node.cs}
				\State $Dup_{new} \gets \Call{insertSorted}{ \mathsf{ndi}_{new}, Dup_{new}, cardinality, ascending}$\label{algoline:pruneQdup:insert_alternative_equal_node_into_Dupnew}
				%\State $Dup_{new} \gets Dup_{new} \cup \setof{\mathsf{ndi}_{new}}$\label{algoline:pruneQdup:insert_alternative_equal_node_into_Dupnew}
				\State \textbf{break}\label{algoline:pruneQdup:break}
			\EndIf
		\EndFor
	\Else                  \Comment{$X$ is not a witness of redundancy of $\mathsf{ndi}$}
	\State $Dup_{new} \gets \Call{insertSorted}{ \mathsf{ndi}, Dup_{new}, cardinality, ascending}$\label{algoline:pruneQdup:insert_node_into_Dupnew}
		%\State $Dup_{new} \gets Dup_{new} \cup \setof{\mathsf{ndi}}$\label{algoline:pruneQdup:insert_node_into_Dupnew}
	\EndIf
\EndFor
%\State $Dup_{new} \gets Dup_{new} \cup Dup$
\State \Return $Dup_{new}$
\EndProcedure
\end{algorithmic}
\normalsize
\end{algorithm*}
\restoregeometry

\chapter{Discussion of Iterative Diagnosis Computation}
\label{chap:TextscStaticHSVersusTextscDynamicHS}
In this chapter we want to summarize properties of and differences between \textsc{staticHS} and \textsc{dynamicHS} that we already pointed out in previous sections and, additionally, we want to shed light on some further interesting aspects of these iterative diagnosis computation methods in the scope of interactive KB debugging (Algorithm~\ref{algo:inter_onto_debug}). Table~\ref{tab:comparison_static_vs_dynamic} provides an overview of what we did discuss or will discuss below.

%\noindent\textbf{First Segment of Table~\ref{tab:comparison_static_vs_dynamic} -- Addressed Problem and Properties w.r.t.\ Solutions.} 
\paragraph{First Segment of Table~\ref{tab:comparison_static_vs_dynamic} -- Addressed Problem and Properties w.r.t.\ Solutions.}
The first row of the table has been proven by Proposition~\ref{prop:correctness_of_interactive_KB_debugging_algo} on page~\pageref{prop:correctness_of_interactive_KB_debugging_algo}. Results given by the second up to the fourth row of the table are substantiated by Proposition~\ref{prop:static_hs_correctness} (\textsc{staticHS}) and Corollary~\ref{cor:dynamic_hs_correctness} (\textsc{dynamicHS}). We have discussed in Section~\ref{sec:TheIntuition} that Algorithm~\ref{algo:inter_onto_debug} with $mode = static$ can artificially fix the search space for possible solutions initially. This is an inherent property of the Interactive Static KB Debugging Problem which the algorithm aims to solve in static mode. For, a minimal diagnosis w.r.t.\ the input DPI which satisfies all answered queries added as test cases throughout the debugging session must be detected (see left column of category ``diagnoses'' in Table~\ref{tab:comparison_static_vs_dynamic}). Hence, the solution space is given by $|\minD_{inputDPI}|$. ``Initially fixed search space'' in this case means that, given the fault tolerance $\sigma = 0$, Algorithm~\ref{algo:inter_onto_debug} in static mode must compute all minimal diagnoses w.r.t.\ the input DPI, i.e.\ the entire set $\minD_{inputDPI}$. In case of dynamic mode, on the other hand, the solution space (i.e.\ minimal diagnoses w.r.t.\ the current DPI, see right column of Table~\ref{tab:comparison_static_vs_dynamic} in category ``diagnoses'') that needs to be explored by Algorithm~\ref{algo:inter_onto_debug} for a given value of zero for $\sigma$ is not known in advance. It rather depends on which test cases are specified or, respectively, which queries the user is asked. In case of the usage of mainly ``positive-impact queries'', the search space might have significantly smaller cardinality than $\minD_{inputDPI}$ whereas it might grow significantly beyond the cardinality of $\minD_{inputDPI}$ in a scenario where many unfavorable ``negative-impact queries'' are generated (cf.\ Section~\ref{sec:OverviewAndIntuition}). The maximum theoretically possible cardinality of the search space for \textsc{dynamicHS} is given by $|\allD_{inputDPI}|$ due to Corollary~\ref{cor:adding_query_to_DPI_implies_that_allD_wrt_new_DPI_is_proper_subset_of_allD_wrt_old_DPI}. 
%
%\noindent\textbf{Second Segment of Table~\ref{tab:comparison_static_vs_dynamic} -- Impact of new Test Cases and Computation Focus.} 
\paragraph{Second Segment of Table~\ref{tab:comparison_static_vs_dynamic} -- Impact of New Test Cases and Computation Focus.}
The properties given in the category ``computes'' in Table~\ref{tab:comparison_static_vs_dynamic} are confirmed by Proposition~\ref{prop:static_hs_correctness} (\textsc{staticHS}) and Corollary~\ref{cor:dynamic_hs_correctness} (\textsc{dynamicHS}). Hence, other than \textsc{dynamicHS} which analyzes the current DPI in terms of minimal conflict sets and diagnoses in each iteration, \textsc{staticHS} must only consider minimal conflict sets w.r.t.\ the \emph{input DPI} (see categories ``diagnoses'' and ``conflict sets'' in Table~\ref{tab:comparison_static_vs_dynamic}). This is sufficient for the exploration of all minimal diagnoses w.r.t.\ the input DPI by Proposition~\ref{prop:mindiag_mincs}. 
%In the case of \textsc{dynamicHS}, minimal conflict sets w.r.t.\ the current DPI are computed 
In this vein, new test cases in static KB debugging are not taken into account in the computation of minimal conflict sets. Instead, new test cases are just exploited to invalidate \emph{already computed} minimal diagnoses w.r.t.\ the input DPI. 
Thus, test cases specified \emph{during} static KB debugging are treated somewhat inferior to test cases already present in the input DPI. Because, the newly gained information given by these test cases is not utilized to reveal new faults in the KB or to lay the focus on just the \emph{now} relevant parts of existing faults, but only for the purpose of constraining the search space for minimal diagnoses w.r.t.\ the input DPI $\langle\mo,\mb,\Tp,\Tn\rangle_\RQ$. We might thus call test cases added during the execution of Algorithm~\ref{algo:inter_onto_debug} with $mode = static$ pure \emph{differentiation test cases} (see category ``purpose of test cases'' in Table~\ref{tab:comparison_static_vs_dynamic}). 

Of course, seen from the point of view of a current DPI, i.e.\ the input DPI extended by differentiation test cases, \textsc{staticHS} does not guarantee completeness w.r.t.\ this current DPI, but only w.r.t.\ the initial one. This however does not mean that, after the (exact) solution $\ot := (\mo\setminus\md) \cup U_{\Tp}$ of the Interactive Static KB Debugging problem has been localized by means of \textsc{staticHS}, the differentiation test cases ($\Tp'$ and $\Tn'$) cannot be simply added to the DPI. In this case, $\ot$ is \emph{still} a maximal solution KB w.r.t.\ the extended input DPI $\langle\mo,\mb,\Tp\cup\Tp',\Tn\cup\Tn'\rangle_\RQ$. In other words, there is no conflict set (and thus no diagnosis) w.r.t.\ $\langle\mo\setminus\md,\mb,\Tp\cup\Tp',\Tn\cup\Tn'\rangle_\RQ$ and $\mo\setminus\md$ is valid w.r.t.\ $\langle\cdot,\mb,\Tp\cup\Tp',\Tn\cup\Tn'\rangle_\RQ$. However, in spite of using the (exact) solution KB of the Interactive Static KB Debugging problem, it is not ensured that this solution is the optimal one w.r.t.\ the \emph{extended DPI}, i.e.\ of the Interactive Dynamic KB Debugging problem. This is because user interaction is just exploited to the extent that the best solution w.r.t.\ the input DPI is crystallized out. It is not used to have the solution verified by the user in the light of the extended DPI. 

On the other hand, test cases assigned throughout dynamic KB debugging by means of Algorithm~\ref{algo:inter_onto_debug} with $mode = dynamic$ are treated equally as test cases already given in the input DPI. They are used to prune the search space and to pinpoint new faults that arise from added test cases resulting from answered queries. The dynamic algorithm assists the user in filtering out a solution and verifying in a thorough manner that this solution is the desired one w.r.t.\ the extended DPI, among \emph{all} existing solutions w.r.t.\ the extended DPI. Due to these aspects we might regard Algorithm~\ref{algo:inter_onto_debug} with mode $mode = dynamic$ as the \emph{standard method for Interactive KB Debugging}.
% and because the user is assisted in order to verify in a thorough manner that a solution is the desired one w.r.t.\ the extended DPI, 

In Sections~\ref{sec:TheIntuition}, \ref{sec:OverviewAndIntuition}, \ref{sec:ImpactOfAnsweredQueriesOnConflictSets} and \ref{sec:ImpactOfAnsweredQueriesOnDiagnoses} we have thoroughly investigated the impact of new test cases (answered queries) added to the DPI on the set of minimal (all) diagnoses and the set of minimal conflict sets considered by the respective method \textsc{staticHS} or \textsc{dynamicHS}. 
For the former, we have shown that (for arbitrary iteration $i$ of Algorithm~\ref{algo:inter_onto_debug}) $\minD_i \supset \minD_{i+1}$ and $\allD_i \supset \allD_{i+1}$ where $\minD_i$ and $\allD_i$ denote the set of all minimal diagnoses and the set of all diagnoses, respectively, that are relevant (for the DPI considered) during iteration $i$. That is, the set of minimal as well as the set of all diagnoses (w.r.t.\ the input DPI) is reduced to a proper subset after a new test case has been added.
For the latter, (for arbitrary iteration $i$ of Algorithm~\ref{algo:inter_onto_debug}) we have argued that generally $\minD_i \not\supset \minD_{i+1}$, but still $\allD_i \supset \allD_{i+1}$, where $\minD_i$ and $\allD_i$ are defined as above. 
That is, not only might some minimal diagnoses (w.r.t.\ the last-but-one DPI) be invalidated, but also some new ones (w.r.t.\ the current DPI) might originate from the incorporation of the information given by a query answer.

Concerning minimal conflict sets, the set of all (or: relevant) minimal conflict sets does not change throughout a debugging session by means of \textsc{staticHS}, i.e.\ $\minC_i = \minC_{i+1}$ (for arbitrary iteration $i$ of Algorithm~\ref{algo:inter_onto_debug}) where $\minC_i$ is the set of minimal conflict sets relevant (for the DPI considered) during iteration $i$. This holds since the minimal conflict sets w.r.t.\ the input DPI are artificially fixed (see above). 
On the contrary, the assignment of a new test case using \textsc{dynamicHS} involves the reduction of some minimal conflict sets (w.r.t.\ the last-but-one DPI) to smaller subset conflict sets (w.r.t.\ the current DPI) and/or the introduction of some ``completely new'' minimal conflict sets (which are in no subset-relation with existing ones, cf.\ Section~\ref{sec:OverviewAndIntuition}). These results are summarized by the categories ``set of all $X$ upon addition of a test case'' in Table~\ref{tab:comparison_static_vs_dynamic}.  

%\noindent\textbf{Third Segment of Table~\ref{tab:comparison_static_vs_dynamic} -- Hitting Set Tree Construction, Pruning and Complexity.} 
\paragraph{Third Segment of Table~\ref{tab:comparison_static_vs_dynamic} -- Hitting Set Tree Construction, Pruning and Complexity.}
Regarding the constructed hitting set tree, we have explained that \textsc{staticHS} builds a wpHS-tree (see Definition~\ref{def:weighted_pruned_hs_tree} on page~\pageref{def:weighted_pruned_hs_tree} and the argumentation in Section~\ref{sec:CorrectnessOfTextscStaticHS}) just as the \textsc{HS} method which is employed for diagnosis computation in the presented non-interactive KB debugging scenario (Algorithm~\ref{algo:non_int_debug}). The main differences between Algorithm~\ref{algo:inter_onto_debug} in static mode and Algorithm~\ref{algo:non_int_debug} are, first, that the former constructs the wpHS-tree step-by-step in multiple phases. Between each two phases a query is generated and presented to the user. The latter, by contrast, finishes the tree construction (to the extent as prescribed by the given parameters $n_{\min}$, $n_{\max}$ and $t$, see Section~\ref{sec:non_int_debug_procedure}) before a single most probable automatically selected solution or a set of solutions is displayed to the user. Second, the tree constructed by the interactive static algorithm exhibits a different labeling of leaf nodes than the one built up be the non-interactive algorithm. In the former, some leaf nodes might be labeled by $\times$ indicating that the path to this node is a minimal diagnosis w.r.t.\ the input DPI, but one which is not in accordance with all answered queries. Notice that such invalidated diagnoses cannot be simply deleted in favor of memory savings, but must be stored in order for the non-minimality criterion (lines~\ref{algoline:slabel:non_min_crit_start}-\ref{algoline:slabel:non_min_crit_end}) to function properly which is necessary to preserve the property of \textsc{staticHS} to compute only \emph{minimal} diagnoses (cf.\ Lemma~\ref{lem:static_hs_soundness}). 
In the non-interactive wpHS-tree, on the other hand, all minimal diagnoses w.r.t.\ the input DPI 
%(and not only the ones consistent with the new test cases) 
are labeled by $\checkmark$. 

What the interactive static and the non-interactive tree have in common is the usage of only minimal conflict sets w.r.t.\ the input DPI as labels of internal (i.e.\ non-leaf) nodes and the adherence to the ``standard'' pruning rules \cite{Reiter87} as per Definition~\ref{def:pruned_hs_tree} on page~\pageref{def:pruned_hs_tree}, i.e.\ the immediate deletion of non-minimal and duplicate tree paths. Except for the standard pruning actions that take place during tree expansion, no separate pruning phases are performed by \textsc{staticHS}. The reason for this is the fixation of the minimal conflict sets, i.e.\ the consideration of only minimal conflict sets w.r.t.\ the input DPI. Incorporation of new minimal conflict sets resulting from answered queries would generally negate completeness of \textsc{staticHS} w.r.t.\ the exploration of all minimal diagnoses w.r.t.\ the input DPI. Integration of new conflict sets that are subsets of existing ones, however, is the key to more substantial pruning actions carried out by \textsc{dynamicHS}. 

Due to the more or less equivalent construction of both the tree built up by \textsc{staticHS} and the one constructed by the \textsc{HS} method in the non-interactive algorithm, it is straightforward to recognize that the worst case time and space complexity of both \emph{tree} computations (without taking into the account other actions performed by the interactive algorithm like probability updates and query generations) are equal. By worst case complexity we refer to the complexity of the search for the (exact) solution of the Interactive Static KB Debugging Problem on the one hand and the complexity of enumerating all minimal diagnoses w.r.t.\ the input DPI on the other hand. In particular, the complexity of tree construction in static KB debugging is independent of given parameters such as the ones for leading diagnoses computation ($n_{\min}$, $n_{\max}$ and $t$) and of the test cases that are classified positively or negatively, respectively, during the debugging session. 

To sum up, due to the artificial fixation of the solution set, there is no possibility of tree pruning in static KB debugging except for the standard pruning rules and hence no way to escape the generally immense worst case complexity for diagnosis search in case $\sigma = 0$.

The hitting set tree constructed by \textsc{dynamicHS}, on the other hand, might differ significantly from the wpHS-tree produced by the non-interactive algorithm. First, it uses minimal conflict sets w.r.t.\ the \emph{current DPI} to label internal nodes in the tree during each expansion stage. Since minimal conflict sets can only ``shrink'' and not ``grow'' due to the integration of test cases into a DPI as stated by Proposition~\ref{prop:changes_in_conflict_sets_after_testcase_added}, the finding that by now a subset of a former minimal conflict set (w.r.t.\ some previous DPI) is already a minimal conflict set (w.r.t.\ the current DPI) gives rise to very powerful ways of tree pruning, as we detailed in Section~\ref{sec:HittingSetTreePruningInTextscDynamicHS} and illustrated by Example~\ref{example:dynamicHS_large_example_using_tabExDpi3}. In this vein, the evolution of the tree produced by \textsc{dynamicHS} can be characterized by alternating expansion and pruning stages. A pruning stage takes place after a test case has been added to the last-but-one DPI in order to modify the tree $T_i$ used to search for minimal diagnoses w.r.t.\ the last-but-one DPI to obtain a tree $T_{i+1}$ that enables the discovery of all minimal diagnoses w.r.t.\ the current DPI. Concretely, both pre-pruning as well as post-pruning is possible during a pruning phase. Pre-pruning refers to the deletion of tree paths ending in an open leaf node, i.e.\ paths corresponding to partial diagnoses, and post-pruning refers to the deletion of tree paths ending in a closed node, i.e.\ paths corresponding to (minimal or non-minimal) diagnoses. Both pre- and post-pruning are not possible in \textsc{staticHS}. The ability for significant tree pruning comes at the cost of not being able to exploit the standard pruning rules as \textsc{staticHS} does. For, non-minimal diagnoses and duplicate tree paths must be stored to guarantee the proper working of tree pruning and in further consequence the completeness of minimal diagnoses search for each current DPI (see Section~\ref{sec:TextscDynamicHSDetailsAndCorrectness}). 

As we pointed out in Section~\ref{sec:OverviewAndIntuition}, the test cases specified during the dynamic debugging session and the defined leading diagnoses computation parameters $n_{\min}$, $n_{\max}$ and $t$ might have a material influence on the extent of possible tree pruning on the one hand and the extent of undesired tree growth on the other. Thence, worst case time and space complexity of the tree generation by means of \textsc{dynamicHS} cannot be initially (at least theoretically) quantified as in the case of \textsc{staticHS}. Consequently, significant savings as well as a substantial overhead compared to \textsc{staticHS} are possible. Careful ``control'' of certain properties of asked queries (added test cases) might help to keep considerable unwanted tree growth within bounds, as we touched upon in Section~\ref{sec:OverviewAndIntuition} and will elaborate on in future work. 

Nevertheless, we want to mention a shortcoming of \textsc{staticHS} compared to \textsc{dynamicHS}. Namely, for $\sigma = 0$, \textsc{staticHS} must \emph{enumerate all minimal diagnoses w.r.t.\ the input DPI} (otherwise no diagnosis can have a probability of 1, see the proof of Proposition~\ref{prop:correctness_of_interactive_KB_debugging_algo} in Section~\ref{sec:CorrectnessOfAlgorithmInterOntoDebug}) whereas \textsc{dynamicHS} might be able to obtain some extended DPI (by the addition of test cases) soon for which only one minimal diagnosis exists. This might require the computation of only a small fraction of the number of $|\minD_{inputDPI}|$ minimal diagnoses that \textsc{staticHS} must determine and therefore might be substantially more time and space saving than figuring out all minimal diagnoses w.r.t.\ some DPI. This is quite well illustrated by Examples~\ref{example:staticHS_complex_example_using_tabExDpi3} and \ref{example:dynamicHS_large_example_using_tabExDpi3}.

%\noindent\textbf{Fourth Segment of Table~\ref{tab:comparison_static_vs_dynamic} -- Query Generation and Bias.} 
\paragraph{Fourth Segment of Table~\ref{tab:comparison_static_vs_dynamic} -- Query Generation and Bias.}
We explained in Remark~\ref{rem:staticHS_query_computed_from_P_cup_P'_and_N_cup_N'} on page~\pageref{rem:staticHS_query_computed_from_P_cup_P'_and_N_cup_N'} that queries in \textsc{staticHS} are computed w.r.t.\ the current DPI albeit only minimal diagnoses w.r.t.\ the input DPI (which are at the same time minimal diagnoses w.r.t.\ the current DPI, cf.\ bullet (\ref{etc:staticHS_output_bullet_a}) on page~\pageref{etc:staticHS_output_bullet_a}) are considered and calculated by Algorithm~\ref{algo:inter_onto_debug} with $mode = static$. In the case of dynamic debugging it is clear that queries are computed w.r.t.\ the current DPI since only minimal diagnoses w.r.t.\ the current DPI are taken into account.

Another important property of an interactive KB debugging algorithm is whether it is biased or unbiased. Intuitively, we call an interactive KB debugging algorithm biased w.r.t.\ some current DPI $DPI$ encountered during its execution iff there might be a minimal diagnosis $\md$ w.r.t.\ $DPI$ such that $\md$ might be definitely invalidated independently of the answers a user gives. In other words, an interactive KB debugging algorithm is unbiased iff for each minimal diagnosis $\md$ w.r.t.\ $DPI$ there is a set $QA_{\md}$ including query answer-pairs such that the addition of the positive queries in $QA_{\md}$ to the positive test cases of $DPI$ and the addition of the negative queries in $QA_{\md}$ to the negative test cases of $DPI$ yields an extended DPI $DPI'$ such that $\md$ is the only minimal diagnosis w.r.t.\ $DPI'$. This means that unbiasedness implies that any solution w.r.t.\ any encountered current DPI during the debugging session might be found as the finally remaining (exact) solution diagnosis. So, all solutions are treated equitably by an unbiased algorithm and \emph{only} the user may decide by their given answers which solutions are and which are not ruled out.

More formally, we define unbiasedness of an interactive KB debugging algorithm as follows:
\begin{definition}
Let $\langle\mo,\mb,\Tp,\Tn\rangle_\RQ$ be the input DPI given to an algorithm $Alg_{X}$ that solves the Interactive $X$ Debugging Problem for $X \in \setof{static, dynamic}$. Let $\Tp' \supseteq \emptyset$ and $\Tn' \supseteq \emptyset$ be the sets of test cases specified so far during the execution of $Alg_X$ and let $\mD \subseteq \minD_{\langle\mo,\mb,\Tp\cup\Tp',\Tn\cup\Tn'\rangle_\RQ}$ be the current set of leading diagnoses. Then, we call $Alg_X$ \emph{biased w.r.t.\ $\langle\mo,\mb,\Tp\cup\Tp',\Tn\cup\Tn'\rangle_\RQ$} iff there is a diagnosis $\md \in \minD_{\langle\mo,\mb,\Tp\cup\Tp',\Tn\cup\Tn'\rangle_\RQ}$ and a query $Q \in \mQ_{\mD,\langle\mo,\mb,\Tp\cup\Tp',\Tn\cup\Tn'\rangle_\RQ}$ such that $\md\notin\minD_{\langle\mo,\mb,\Tp\cup\Tp'\cup\setof{Q},\Tn\cup\Tn'\rangle_\RQ}$ and $\md \notin \minD_{\langle\mo,\mb,\Tp\cup\Tp',\Tn\cup\Tn'\cup\setof{Q}\rangle_\RQ}$.

Further, we call $Alg_X$ \emph{unbiased} iff there cannot be any sets of test cases $\Tp' \supseteq \emptyset$ and $\Tn' \supseteq \emptyset$ during any execution of $Alg_X$ such that $Alg_X$ is biased w.r.t.\ $\langle\mo,\mb,\Tp\cup\Tp',\Tn\cup\Tn'\rangle_\RQ$.
\end{definition}  
%
%only due to the way the algorithm works and not due to   
%
%there cannot be a query $Q$ and a minimal diagnosis $\md$ w.r.t.\ a DPI such that any answer to it rules out $\md$:
\begin{remark}
It is important to notice the difference between completeness (which has already been established for Algorithm~\ref{algo:inter_onto_debug} using any of the methods \textsc{staticHS} or \textsc{dynamicHS}, see Lemma~\ref{lem:static_hs_completeness} and Proposition~\ref{prop:dyn_completeness}) and unbiasedness of an algorithm. Completeness refers to the guarantee that the algorithm explores all minimal diagnoses w.r.t.\ any DPI $DPI$. However, it does not say anything about what might happen after a new test case $Q$ is added to $DPI$. Although it does state that all minimal diagnoses w.r.t.\ the new DPI $DPI'$ are explored, it leaves us unclear about what effect the addition of the query $Q$ to the test cases might have had on the minimal diagnoses. So, there might be a minimal diagnosis w.r.t.\ $DPI$ that would have been ruled out by both answers to $Q$ thereby violating unbiasedness, but not completeness. To sum up, completeness gives us guarantees about what happens during the diagnosis computation phase whereas unbiasedness gives us guarantees about what happens during the transition from one DPI to a new DPI.\qed
\end{remark}
In the following, we show that Algorithm~\ref{algo:inter_onto_debug} in both static and dynamic mode is unbiased.
\begin{proposition}
Assume the execution of Algorithm~\ref{algo:inter_onto_debug} with $mode \in \setof{static, dynamic}$ given the input DPI $\langle\mo,\mb,\Tp,\Tn\rangle_\RQ$. Further, let $\mD := \mD_{calc}$ be the set of minimal diagnoses w.r.t.\ $\langle\mo,\mb,\Tp\cup\Tp',\Tn\cup\Tn'\rangle_\RQ$ returned by a call of \textsc{dynamicHS} in case of $mode = dynamic$ and $\mD := \mD_{calc} \cup \mD_{\checkmark}$ be the set of minimal diagnoses w.r.t.\ $\langle\mo,\mb,\Tp\cup\Tp',\Tn\cup\Tn'\rangle_\RQ$ returned by a call of \textsc{staticHS} in case of $mode = static$. 
Moreover, let $\md \in \minD_{\langle\mo,\mb,\Tp\cup\Tp',\Tn\cup\Tn'\rangle_\RQ}$.

Then, no query $Q$ w.r.t.\ $\mD$ and $\langle\mo,\mb,\Tp\cup\Tp',\Tn\cup\Tn'\rangle_\RQ$ can be computed by Algorithm~\ref{algo:inter_onto_debug} 
%with $mode \in \setof{static, dynamic}$ 
such that $\md\notin\minD_{\langle\mo,\mb,\Tp\cup\Tp'\cup\setof{Q},\Tn\cup\Tn'\rangle_\RQ}$ and $\md \notin \minD_{\langle\mo,\mb,\Tp\cup\Tp',\Tn\cup\Tn'\cup\setof{Q}\rangle_\RQ}$.
%Let $\mD \subseteq \minD_{\langle\mo,\mb,\Tp,\Tn\rangle_\RQ}$ and $\md \in \minD_{\langle\mo,\mb,\Tp,\Tn\rangle_\RQ}$. Then, no query $Q$ w.r.t.\ $\mD$ and $\langle\mo,\mb,\Tp,\Tn\rangle_\RQ$ can be computed by Algorithm~\ref{algo:inter_onto_debug} with $mode \in \setof{static, dynamic}$ such that $\md$ is not a minimal diagnosis w.r.t.\ $\langle\mo,\mb,\Tp\cup\setof{Q},\Tn\rangle_\RQ$ and not a minimal diagnosis w.r.t.\ $\langle\mo,\mb,\Tp,\Tn\cup\setof{Q}\rangle_\RQ$.
\end{proposition}
\begin{proof}
Let us consider the q-partition $\Pt(Q) = \tuple{\dx{}(Q),\dnx{}(Q),\dz{}(Q)}$ of the query $Q$ that is computed by Algorithm~\ref{algo:inter_onto_debug} for the set of leading diagnoses $\mD$. By Proposition~\ref{prop:q-partition_is_partition}, we have that $\dx{}(Q) \cup \dnx{}(Q) \cup \dz{}(Q) = \mD$ and $\dx{}(Q)$, $\dnx{}(Q)$ and $\dz{}(Q)$ are pairwise disjoint sets, i.e.\ the sets $\dx{}(Q)$, $\dnx{}(Q)$ and $\dz{}(Q)$ constitute a partition of the set $\mD$. Let us now assume that each diagnosis in $\minD_{\langle\mo,\mb,\Tp\cup\Tp',\Tn\cup\Tn'\rangle_\RQ}$ is assigned to its respective set in $\Pt(Q)$ as per Definition~\ref{def:q-partition} yielding the tuple $\tuple{\dx{m}(Q),\dnx{m}(Q),\dz{m}(Q)}$ where $\dx{m}(Q) \cup \dnx{m}(Q) \cup \dz{m}(Q) = \minD_{\langle\mo,\mb,\Tp\cup\Tp',\Tn\cup\Tn'\rangle_\RQ}$. Then, by analogue argumentation as in the proof of Proposition~\ref{prop:q-partition_is_partition}, we obtain that $\dx{m}(Q)$, $\dnx{m}(Q)$ and $\dz{m}(Q)$ are pairwise disjoint sets. That is, $\tuple{\dx{m}(Q),\dnx{m}(Q),\dz{m}(Q)}$ is the (extended) q-partition of $Q$ w.r.t.\ the leading diagnoses set $\minD_{\langle\mo,\mb,\Tp\cup\Tp',\Tn\cup\Tn'\rangle_\RQ}$.

By Remark~\ref{rem:invalidated_sets_of_q-partition_for_query_answer}, we have that $\mD_{pos} := \dx{m}(Q) \cup \dz{m}(Q)$ are minimal diagnoses w.r.t.\ the DPI $\langle\mo,\mb,\Tp\cup\Tp'\cup\setof{Q},\Tn\cup\Tn'\rangle_\RQ$ (positive answer $u(Q)$) and $\mD_{neg} :=\dnx{m}(Q) \cup \dz{m}(Q)$ are minimal diagnoses w.r.t.\ the DPI $\langle\mo,\mb,\Tp\cup\Tp',\Tn\cup\Tn'\cup\setof{Q}\rangle_\RQ$ (negative answer $u(Q)$). Since $\mD_{pos}\cup\mD_{neg} \supseteq \minD_{\langle\mo,\mb,\Tp\cup\Tp',\Tn\cup\Tn'\rangle_\RQ}$, we have that each diagnosis in $\minD_{\langle\mo,\mb,\Tp\cup\Tp',\Tn\cup\Tn'\rangle_\RQ}$ is either in $\mD_{pos}$ or in $\mD_{neg}$ (or in both). Hence, for each diagnosis $\md\in\minD_{\langle\mo,\mb,\Tp\cup\Tp',\Tn\cup\Tn'\rangle_\RQ}$ there is some answer $u(Q) \in \setof{\true,\false}$ to the query $Q$ such that $\md$ is a diagnosis w.r.t.\ the DPI resulting from $\langle\mo,\mb,\Tp\cup\Tp',\Tn\cup\Tn'\rangle_\RQ$ by addition of the new test case $Q$ to the respective set ($\Tp \cup \Tp'$ for positive and $\Tn \cup \Tn'$ for negative answer). Consequently, the claimed proposition holds. 
\end{proof}
\begin{corollary}
Algorithm~\ref{algo:inter_onto_debug} with $mode \in \setof{static, dynamic}$ is unbiased for any given input DPI $\langle\mo,\mb,\Tp,\Tn\rangle_\RQ$.
\end{corollary}
%
%By the completeness of \textsc{staticHS} and \textsc{dynamicHS} it follows that all these (still valid) diagnoses are considered.

\newgeometry{margin=2cm}

\begin{table}[h]
\small
\centering
\rowcolors[]{2}{gray!8}{gray!16}
\begin{tabular}{p{0.23\textwidth} p{0.33\textwidth} p{0.33\textwidth}}
%input $X$ to \textsc{prune}/\textsc{pruneQdup} & $\mathsf{nd.cs}$ after execution of \textsc{prune}/\textsc{pruneQdup} \\
\rowcolor{gray!40}
\toprule\addlinespace[0pt] 
&\textsc{staticHS} & \textsc{dynamicHS} \\
\hline
\textbf{is used to solve} &
Interactive Static KB Debugging problem (Problem Definition~\ref{prob_def:static}) & 
Interactive Dynamic KB Debugging problem (Problem Definition~\ref{prob_def:dynamic})\\
\textbf{soundness} &
yes & 
yes \\
\textbf{completeness} &
yes & 
yes \\
\textbf{optimality} &
yes & 
yes \\
%
%
%\textbf{number of solutions that must be considered} &
%$\bullet\,$ initially fixed \newline $\bullet\,$ upper bound: $|\minD_{inputDPI}|$ & 
%$\bullet\,$ not initially fixed, depends on specified test cases (answered queries)  \newline $\bullet\,$ upper bound: $|\allD_{inputDPI}|$ \\
%%
\textbf{number of solutions that must be considered (but not necessarily computed)} &
\vspace{-12pt}
\begin{itemize}
	\item initially fixed
	\item upper bound: $|\minD_{inputDPI}|$\vspace{-12pt}
\end{itemize}
&
\vspace{-12pt}
\begin{itemize}
	\item not initially fixed, depends on specified test cases (answered queries)
	\item upper bound: $|\allD_{inputDPI}|$\vspace{-10pt}
\end{itemize}
 \\ \hline
\textbf{diagnoses} &
considers only minimal diagnoses w.r.t.\ the input DPI which satisfy all answered queries added as test cases so far & 
considers only minimal diagnoses w.r.t.\ the current DPI\\
\textbf{conflict sets} &
computes only minimal conflict sets w.r.t.\ the input DPI  & 
computes minimal conflict sets w.r.t.\ the current DPI\\
\textbf{computes} &
a set $\mD$ including the $|\mD| \leq n_{\max}$ (a-priori) most probable minimal diagnoses w.r.t.\ the input DPI which satisfy all answered queries added as test cases so far & 
a set $\mD$ including the $|\mD| \leq n_{\max}$ (a-priori) most probable minimal diagnoses w.r.t.\ the current DPI\\
\textbf{purpose of test cases} &
differentiation between minimal diagnoses of fixed DPI 
& 
obtaining a new DPI with fewer minimal diagnoses
\\
\textbf{set of all minimal diagnoses upon addition of a test case} &
is reduced to a proper subset %\newline $\rightarrow$ constant ``convergence'' towards (exact) solution  
& 
some are invalidated, some new ones might be introduced 
%\newline $\rightarrow$ no constant ``convergence'' towards (exact) solution
\\
\textbf{set of all diagnoses upon addition of a test case} &
is reduced to a proper subset  & 
is reduced to a proper subset \\
\textbf{set of all minimal conflict sets upon addition of a test case} &
constant  & 
some minimal conflict sets are reduced to smaller sets and/or some new minimal conflict sets (in no subset-relation with existing ones) are introduced \\ \hline
\textbf{constructed tree: \newline comparison to non-interac-tive wpHS-tree (Alg.~\ref{algo:hs})} &
equivalent (except for labels of leaf nodes) & 
might differ significantly \\
\textbf{non-leaf-node labels} &
only minimal conflict sets w.r.t.\ the input DPI  & 
(not necessarily minimal) conflict sets w.r.t.\ the current DPI  \\
\textbf{non-minimal and duplicate tree paths} &
deleted  & 
stored \\
\textbf{evolution of produced tree} &
only expansion (except for deletion of non-minimal and duplicate tree paths)  & 
alternating tree expansion and pruning phases \\
\textbf{pre-pruning (deletion of partial diagnoses)} &
only duplicate tree paths  & 
any \\
\textbf{post-pruning (deletion of complete diagnoses)} &
only non-minimal diagnoses (all invalidated minimal diagnoses are stored)  & 
any \\
\textbf{overall tree pruning} &
poor  & 
significant \\
%
%\textbf{tree construction: worst case time and space complexity} &
%$\bullet\,$ independent of specified test cases \newline $\bullet\,$ upper and lower bound is time and space required by non-interactive wpHS-tree (Alg.~\ref{algo:hs})  & 
%$\bullet\,$ a function of the specified test cases \newline $\bullet\,$ best case: significant savings compared to \textsc{staticHS}, \newline $\bullet\,$ worst case: significant overhead compared to \textsc{staticHS} \\
%%
\textbf{tree construction: worst case time and space complexity} &
\vspace{-12pt}
\begin{itemize}
	\item independent of specified test cases
	\item upper and lower bound is time and space required by non-interactive wpHS-tree (Alg.~\ref{algo:hs}) \vspace{-12pt}
\end{itemize}
&
\vspace{-12pt}
\begin{itemize}
	\item a function of the specified test cases and the leading diagnosis computation parameters $n_{\min}, n_{\max}, t$
	\item best case: significant savings compared to \textsc{staticHS}
	\item worst case: significant overhead compared to \textsc{staticHS} \vspace{-10pt}
\end{itemize}
 \\ \hline
\textbf{query generation} &
w.r.t.\ the current DPI  & 
w.r.t.\ the current DPI \\
\textbf{unbiased} &
yes  & 
yes \\
%\hline
\addlinespace[0pt]\bottomrule 
\end{tabular}
\caption{Comparison: \textsc{staticHS} versus \textsc{dynamicHS}.}
\label{tab:comparison_static_vs_dynamic}
\vspace{-10pt}
\end{table}

\restoregeometry

\part{Two Query Strategies for Efficient Fault Localization in Interactive Ontology Debugging}
\label{part:JWS}
In this part, we suggest and extensively analyze different methods for the selection of an ``optimal'' query. The material dealt with in Part~\ref{part:JWS} is based on the publications \cite{Shchekotykhin2012, ksgf2010} where the former was published in the journal \emph{Web
Semantics: Science, Services and Agents on the World Wide Web} and the latter in the \emph{Proceedings of the 9th International Semantic Web Conference (ISWC 2010)}.
\chapter{Introduction to the Problem}
Ontology acquisition and maintenance are important prerequisites for the successful application of semantic systems in areas such as the Semantic Web. 
However, as state of the art ontology extraction methods cannot automatically acquire ontologies in a complete and error-free fashion, users of such systems must formulate and correct logical descriptions on their own. In most of the cases these users are domain experts who have little or no experience in expressing knowledge in representation languages like OWL~2~DL~\cite{Grau2008a}.
Studies in cognitive psychology, e.g.~\cite{Ceraso71,Johnson1999}, indicate that humans make systematic errors while formulating or interpreting logical descriptions, with the results presented in~\cite{Rector2004,Roussey2009} confirming that these observations also apply to ontology development. 
Moreover, the problem gets even more if an ontology is developed by a group of users, such as OBO Foundry\footnote{\url{http://www.obofoundry.org}} or NCI Thesaurus\footnote{\url{http://ncit.nci.nih.gov}}, is based on a set of imported third-party ontologies, etc. In this case inconsistencies might appear if some user does not understand or accept the \emph{context} in which shared ontological descriptions are used. Therefore, identification of erroneous ontological definitions is a difficult and time-consuming task.

Several ontology debugging methods ~\cite{Schlobach2007,Kalyanpur.Just.ISWC07,friedrich2005gdm,Horridge2008} were proposed to simplify ontology development and maintenance. Usually the main aim of debugging is to obtain a consistent and, optionally, coherent ontology. These basic requirements can be extended with additional ones, such as test cases~\cite{friedrich2005gdm}, which must be fulfilled by the \emph{target ontology} $\jwsmo_t$. Any ontology that does not fulfill the requirements is \emph{faulty} regardless of how it was created. For instance, an ontology might be created by an expert 
%it was created by experts by 
specializing descriptions of the imported ontologies (top-down) or 
%it might be generated 
by an inductive learning algorithm from a set of examples (bottom-up).

Note that
even if all requirements are completely specified, 
%the user specifies a complete set of requirements, there may exist
%, in real-world scenarios it is impossible for a user to specify a complete set of requirements since the target ontology is unknown in the development phase; otherwise there is no need for debugging. Moreover, from a logical point of view, 
many logically equivalent target ontologies might exist. 
%, although 
They may differ in aspects such as the complexity of consistency checks, size or readability. However, selecting between logically equivalent theories based on such measures is out of the scope of this work. 
Furthermore, although target ontologies may evolve as requirements change over time, we assume that the target ontology remains stable throughout a debugging session. 

Given an set of requirements (e.g.\ formulated by a user) and a faulty ontology, the task of an ontology debugger is to identify the set of alternative diagnoses, where each diagnosis corresponds to a set of possibly faulty axioms. More concretely, a \emph{diagnosis} $\jwsmd$ is a subset of an ontology $\jwsmo$ such that one should remove (change) all the axioms of a diagnosis from the ontology (i.e.\ $\jwsmo \setminus \jwsmd$) in order to formulate an ontology $\jwsmo'$ that fulfills all the given requirements. 
Only if the set of requirements is complete the only possible ontology $\jwsmo'$ corresponds to the target ontology $\jwsmo_t$. In the following we refer to the removal of a diagnosis from the ontology as a \emph{trivial application} of a diagnosis.
Moreover, in practical applications it might be inefficient to consider all possible diagnoses. Therefore, modern ontology debugging approaches focus on the computation of minimal diagnoses. 
A set of axioms $\jwsmd_i$ is a \emph{minimal diagnosis} iff there is no proper subset $\jwsmd'_i \subset \jwsmd_i$ which is a diagnosis. 
%,i.e.\ for each $\jwsmd_i$, no $\jwsmd'_i$ exists such $\jwsmd'_i \subset \jwsmd_i$. 
Thus, minimal diagnoses constitute minimal required changes to the ontology.
%At a minimum, a user must remove (change) all of the axioms of a minimal diagnosis in order to formulate the intended target ontology. 

Application of diagnosis methods can be problematic in the cases for which many alternative minimal diagnoses exist for a given set of test cases and requirements. A sample study of real-world incoherent ontologies, which were used in~\cite{Kalyanpur.Just.ISWC07}, showed that hundreds or even thousands of minimal diagnoses may exist. In the case of the Transportation ontology the diagnosis method was able to identify 1782 minimal diagnoses \footnote{In Chapter \ref{chap:jws:eval}, we will give a detailed characterization of these ontologies.}. In such situations a simple visualization of all alternative sets of modifications to the ontology is ineffective. 
Thus an efficient debugging method should be able to discriminate between the diagnoses in order to select the \emph{target diagnosis} $\jwsmd_t$. Trivial application of $\jwsmd_t$ to the ontology $\jwsmo$ allows a user to extend $(\jwsmo \setminus \jwsmd_t)$ with a set of additional axioms $EX$ and, thus, to formulate the target ontology $\jwsmo_t$, i.e.\ $\jwsmo_t = (\jwsmo \setminus \jwsmd_t) \cup EX$.

One possible solution to the diagnosis discrimination problem would be to order the set of diagnoses by various preference criteria. For instance, Kalyanpur et al.~\cite{Kalyanpur2006} suggest a measure to rank the axioms of a diagnosis depending on their structure, usage in test cases, provenance, and impact in terms of entailments. Only the top ranking diagnoses are then presented to the user.
Of course this set of diagnoses will contain the target diagnosis only in cases where the faulty ontology, the given requirements and test cases provide sufficient data to the appropriate heuristic. However, it is difficult to identify which information, e.g. test cases, is really required to identify the target diagnosis. That is, a user does not know a priori which and how many tests should be provided to the debugger to ensure that it will return the target diagnosis.

In this part we present an approach for the acquisition of additional information by \emph{generating} a sequence of queries, the answers of which can be used to reduce the set of diagnoses and ultimately identify the target diagnosis. These queries should be answered by an oracle such as a user or an information extraction system. In order to construct queries we exploit the property that different ontologies resulting from trivial applications of different diagnoses entail unequal sets of axioms. 
Consequently, we can differentiate between diagnoses by asking the oracle if the target ontology should entail a set of logical sentences or not. These entailed logical sentences can be generated by the classification and realization services provided in description logic reasoning systems~\cite{sirin2007pellet,Haarslev2001,Motik2009}. In particular, the classification process computes a subsumption hierarchy (sometimes also called ``inheritance hierarchy'' of parents and children) for each concept description mentioned in a TBox. For each individual mentioned in an ABox, the realization computes all the concept names of which the individual is an instance~\cite{sirin2007pellet}. 
%For each pair of individuals, the realization returns all the role names of which this pair is an instance.
%the atomic classes (or concept names) of which the individual is an instance~\cite{Sirin2007Pellet}. 

We propose two methods for selecting the next query of the set of possible queries:
The first method employs a greedy approach that selects queries which try to cut the number of diagnoses in half. The second method exploits the fact that some diagnoses are more likely than others because of typical user errors~\cite{Rector2004,Roussey2009}. Beliefs for an error to occur in a given part of a knowledge base, represented as a probability, can be used to estimate the change in entropy of the set of diagnoses if a particular query is answered. In our evaluation the fault probabilities of axioms are estimated by the type and number of the logical operators employed. For example, roughly speaking, the greater the number of logical operators and the more complex these operators are, the greater the fault probability of an axiom. For assigning prior fault probabilities to diagnoses we employ the fault probabilities of axioms. Of course other methods for guessing prior fault probabilities, e.g. based on context of concept descriptions, measures suggested in the previous work~\cite{Kalyanpur2006}, etc., can be easily integrated in our framework. 
Given a set of diagnoses and their probabilities the method selects a query which minimizes the expected entropy of a set of diagnoses after an oracle answers a query, i.e.\ maximizes the information gain. An oracle should answer such queries until a diagnosis is identified whose probability is significantly higher than those of all other diagnoses. This diagnosis is most likely to be the target diagnosis. 

In 
the first evaluation scenario 
%an extensive evaluation  
we compare the performance of both methods in terms of the number of queries needed to identify the target diagnosis.
The evaluation is performed using generated examples as well as real-world ontologies presented in Tables~\ref{tab:motivation} and~\ref{tab:bigstats}. In the first case we alter a consistent and coherent ontology with additional axioms to generate conflicts that result in a predefined number of diagnoses of a required length. Each faulty ontology is then analyzed by the debugging algorithm using entropy, greedy and ``random'' strategies, where the latter selects queries at random. The evaluation results show that in some cases the entropy-based approach is almost $60\%$ better than the greedy one whereas both approaches clearly outperformed the random strategy.
%the random strategy is clearly inferior to both the greedy and the entropy-based approach. 

In the second evaluation scenario we investigate the robustness of the entropy-based strategy with respect to variations in the prior fault probabilities. We analyze the performance of entropy-based and greedy strategies on real-world ontologies by simulating different types of prior fault probability distributions as well as the ``quality'' of these probabilities that might occur in practice. In particular, we identify the cases where all prior fault probabilities are (1) equal, 
%(i.e.\ where the fault probabilities of an axiom depend only on the number of operators and not on their types)
(2) ``moderately'' varied or (3) ``extremely'' varied. 
%have moderate or high variance
Regarding the ``quality'' of the probabilities we investigate cases where the guesses based on the prior diagnosis probabilities are good, average or bad. 
The results show that the entropy method outperforms ``split-in-half'' in almost all of the cases, namely when the target diagnosis is located in the more likely two thirds of the minimal diagnoses. 
In some situations the entropy-based approach achieves even twice the performance of the greedy one. 
Only in cases where the initial guess of the prior probabilities is very vague (the bad case), \emph{and} the number of queries needed to identify the target diagnosis is low, ``split-in-half'' may save on average one query. 
However, if the number of queries increases, the performance of the entropy-based query selection increases compared to the ``split-in-half'' strategy. We observed that if the number of queries is greater than 10, the entropy-based method is preferable even if the initial guess of the prior probabilities is bad. This is due to the effect that the initial bad guesses are improved by the Bayes-update of the diagnoses probabilities as well as an ability of the entropy-based method to stop in the cases when a probability of some diagnosis is above an acceptance threshold predefined by the user. 
Consequently, entropy-based query selection is robust enough to handle different prior fault probability distributions. 
%This is also often the case when all initial fault probabilities are equal, i.e.\ no estimates about typical user errors are available. 

Additional experiments performed on big real-world ontologies demonstrate the scalability of the suggested approach. In our experiments we were able to identify the target diagnosis in an ontology with over 33000 axioms using entropy-based query selection in only 190 seconds using an average of five queries.

The remainder of Part~\ref{part:JWS} is organized as follows: Chapter~\ref{chap:jws:example} presents two introductory examples as well as the basic concepts. The details of the entropy-based query selection method are given in Chapter~\ref{chap:jws:theory}. Chapter~\ref{chap:jws:impl} describes the implementation of the approach and is followed by evaluation results in Chapter~\ref{chap:jws:eval}. An overview of related work is given in Chapter~\ref{chap:jws:related} and conclusions are drawn in Chapter~\ref{chap:jws:conclusions}.

%%%%%%%%%%%%%%%%%%%%%%%%%%%%%%%%%%%%%%%%%%%%%%%%%%%%%%%%%%%%%%%%%%%%%%%
%%%%%%%%%%%%%%%%%%%%%%%%%%%%%%%%%%%%%%%%%%%%%%%%%%%%%%%%%%%%%%%%%%%%%%%
%%%%%%%%%%%%%%%%%%%%%%%%%%%%%%%%%%%%%%%%%%%%%%%%%%%%%%%%%%%%%%%%%%%%%%%
%%%%%%%%%%%%%%%%%%%%%%%%%%%%%%%%%%%%%%%%%%%%%%%%%%%%%%%%%%%%%%%%%%%%%%%
%%%%%%%%%%%%%%%%%%%%%%%%%%%%%%%%%%%%%%%%%%%%%%%%%%%%%%%%%%%%%%%%%%%%%%%
%%%%%%%%%%%%%%%%%%%%%%%%%%%%%%%%%%%%%%%%%%%%%%%%%%%%%%%%%%%%%%%%%%%%%%%
%%%%%%%%%%%%%%%%%%%%%%%%%%%%%%%%%%%%%%%%%%%%%%%%%%%%%%%%%%%%%%%%%%%%%%%
%%%%%%%%%%%%%%%%%%%%%%%%%%%%%%%%%%%%%%%%%%%%%%%%%%%%%%%%%%%%%%%%%%%%%%%
%%%%%%%%%%%%%%%%%%%%%%%%%%%%%%%%%%%%%%%%%%%%%%%%%%%%%%%%%%%%%%%%%%%%%%%
%%%%%%%%%%%%%%%%%%%%%%%%%%%%%%%%%%%%%%%%%%%%%%%%%%%%%%%%%%%%%%%%%%%%%%%
%%%%%%%%%%%%%%%%%%%%%%%%%%%%%%%%%%%%%%%%%%%%%%%%%%%%%%%%%%%%%%%%%%%%%%%

\chapter{Motivating Examples and Basic Concepts}\label{chap:jws:example}

We begin by presenting the fundamentals of ontology diagnosis and then show how queries and answers can be generated and employed to differentiate between sets of diagnoses. 

\section*{Description Logics}

Since the underlying knowledge representation method of ontologies in the Semantic Web is based on description logics, we start by briefly introducing the main concepts, employing the usual definitions as in \cite{borgida1996,Baader2003c}. A knowledge base is comprised of two components, namely a TBox (denoted by $\jwsmT$) and a ABox ($\jwsmA$). The TBox defines the terminology whereas the ABox contains assertions about named individuals in terms of the vocabulary defined in the TBox. The vocabulary consists of concepts, denoting sets of individuals, and roles, denoting binary relationships between individuals. These concepts and roles may be either atomic or complex, the latter being obtained by employing description operators. The language of descriptions is defined recursively by starting from a schema $S = (\jwsmCN, \jwsmRN, \jwsmIN)$ of disjoint sets of names for concepts, roles, and individuals. Typical operators for the construction of complex descriptions are $C \sqcup D$ (disjunction), $C \sqcap D$ (conjunction), $\neg C$ (negation), $\forall R.C$ (concept value restriction), and $\exists R.C$(concept exists restriction), where $C$ and $D$ are elements of $\jwsmCN$ and $R \in \jwsmRN$.

Knowledge bases are defined by a finite set of logical sentences. Sentences regarding the TBox are called terminological axioms whereas sentences regarding the ABox are called assertional axioms. Terminological axioms are expressed by $C \sqsubseteq D$ (Generalized Concept Inclusion) which corresponds to the logical implication. Let $a,b \in \jwsmIN$ be individual names. $C(a)$ and $R(a,b)$ are thus assertional axioms.

Concepts (rsp.\ roles) can be regarded as unary (rsp.\ binary) predicates. Roughly speaking description logics can be seen as fragments of first-order predicate logic (without considering transitive closure or special fixpoint semantics). These fragments are specifically designed to ensure decidability or favorable computational costs.

The semantics of description terms are usually given using an interpretation $\jwsmI = \langle \Delta^\jwsmI, (\cdot)^\jwsmI \rangle$, where $\Delta^\jwsmI$ is a domain (non-empty universe) of values, and $(\cdot)^\jwsmI$ is a function that maps every concept description to a subset of $\Delta^\jwsmI$, and every role name to a subset of $\Delta^\jwsmI \times \Delta^\jwsmI$. The mapping also associates a value in $\Delta^\jwsmI$ with every individual name in $\jwsmIN$.
An interpretation $\jwsmI$ is a model of a knowledge base iff it satisfies all terminological axioms and assertional axioms. A knowledge base is satisfiable iff a model exists.
A concept description $C$ is coherent (satisfiable) w.r.t.\ a TBox $\jwsmT$, if  a model $\jwsmI$ of $\jwsmT$ exists such that $C^\jwsmI \neq \emptyset$. A
TBox is incoherent iff an incoherent concept description exists.

\section*{Diagnosis of Ontologies}

\begin{example} \label{ex:simple} Consider a simple ontology $\jwsmo$ with the  terminology $\jwsmt$:
\begin{center}
\begin{tabular}{cc}
$\jwstax_1 : A \sqsubseteq B$ & 
$\jwstax_2 : B \sqsubseteq C$  \\
$\jwstax_3 : C \sqsubseteq D$ &
$\jwstax_4 : D \sqsubseteq R$ 
\end{tabular}
\end{center}
and assertions $\jwsma :\{A(w), \lnot R(w), A(v)\}$. 

Assume that the user explicitly states that the three assertional axioms should be considered as correct, i.e. these axioms are added to a background theory $\jwsmb$. The introduction of a background theory ensures that the diagnosis method focuses purely on the potentially faulty axioms.

Furthermore, assume that the user requires the currently inconsistent ontology $\jwsmo \cup \jwsmb$ to be consistent. The only irreducible set of non-background axioms (minimal conflict set) that preserves the inconsistency is $CS:\langle\jwstax_1,\jwstax_2$, $\jwstax_3,\jwstax_4\rangle$. That is, one has to modify or remove the axioms of at least one of the following diagnoses
\begin{equation*}
\jwsmd_1:\left[\jwstax_1\right]\quad \jwsmd_2:\left[\jwstax_2\right]\quad \jwsmd_3:\left[\jwstax_3\right]\quad \jwsmd_4:\left[\jwstax_4\right]
\end{equation*}
to restore the consistency of the ontology. However, it is unclear which of the ontologies $\jwsmo_i = \jwsmo \setminus \jwsmd_i$ obtained by application of diagnoses from the set ${\bf D}: \{\jwsmd_1,\dots, \jwsmd_4\}$ is the target one.\qed
\end{example}

\begin{definition}\label{def:targetontology}

A target ontology $\jwsmo_t$ is a set of logical sentences characterized by a set of background axioms $\jwsmb$, a set of sets of logical sentences $\jwsTe$ that must be entailed by $\jwsmo_t$ and the set of sets of logical sentences $\jwsTne$ that must not be entailed by $\jwsmo_t$. 

A target ontology $\jwsmo_t$ must fulfill the following necessary requirements:
\begin{itemize}
	\item $\jwsmo_t$ must be satisfiable (optionally coherent)
	\item $\jwsmb \subseteq \jwsmo_t$
	\item $\jwsmo_t \models \jwste\quad \forall \jwste \in \jwsTe$
	\item $\jwsmo_t \not\models \jwstne\quad \forall \jwstne \in \jwsTne$  
\end{itemize}
\end{definition}
Given $\jwsmb,$ $\jwsTe,$ and $\jwsTne$, an ontology $\jwsmo$ is faulty iff $\jwsmo$ does not fulfill all the necessary requirements of the target ontology.
%An ontology $\jwsmo$ for $\jwsmb,$ $\jwsTe,$ and $\jwsTne$ is faulty iff $\jwsmo$ does not fulfill all the necessary requirements of the target ontology.

Note that the approach presented in this work can be used with any knowledge representation language for which there exists a sound and complete procedure to decide whether $\jwsmo\models\jwstax$ and the entailment operator $\models$ is extensive, monotone and idempotent. For instance, these requirements are fulfilled by all subsets of OWL 2 which are interpreted under OWL Direct Semantics.

Definition~\ref{def:targetontology} allows a user to identify the target diagnosis $\jwsmd_t$ by providing sufficient information about the target ontology in the sets $\jwsmb, \jwsTe$ and $\jwsTne$. For instance, if in Example~\ref{ex:simple} the user provides the information that $\jwsmo_t \models \setof{B(w)}$ and $\jwsmo_t \not\models \setof{C(w)}$, the debugger will return only one diagnosis, namely $\jwsmd_2$. Application of this diagnosis results in a consistent ontology $\jwsmo_2 = \jwsmo\setminus\jwsmd_2$ that -- integrated with the background knowledge $\jwsmb$ -- entails $\setof{B(w)}$ because of $\jwstax_1$ and the assertion $A(w)$. In addition, $\jwsmo_2 \cup \jwsmb$ does not entail $\setof{C(w)}$ since $\jwsmo_2 \cup \jwsmb \cup \{\neg C(w)\}$ is consistent and, moreover, $ \{ \lnot R(w), \jwstax_4, \jwstax_3 \} \models \setof{\lnot C(w)}$. All other ontologies $\jwsmo_i = (\jwsmo\setminus\jwsmd_i)$ obtained by the application of the diagnoses $\jwsmd_1,\jwsmd_3$ and $\jwsmd_4$ do not fulfill the given requirements, since $\jwsmo_1 \cup \jwsmb \cup \setof{B(w)}$ is inconsistent and therefore any consistent extension of $\jwsmo_1 \cup \jwsmb$ cannot entail $\setof{B(w)}$. As both $\jwsmo_3 \cup \jwsmb$ and $\jwsmo_4 \cup \jwsmb$ entail $\setof{C(w)}$, $\jwsmo_2 \cup \jwsmb$ corresponds to the target ontology $\jwsmo_t$.

\begin{definition}\label{def:diag}
Let $\tuple{\jwsmo, \jwsmb,\jwsTe,\jwsTne}$ be a diagnosis problem instance, where $\jwsmo$ is an ontology, $\jwsmb$ a background theory, $\jwsTe$ a set of sets of logical sentences which must be entailed by the target ontology $\jwsmo_t$, and $\jwsTne$ a set of sets of logical sentences which must \emph{not} be entailed by $\jwsmo_t$.

A set of axioms $\jwsmd\subseteq\jwsmo$ is a diagnosis iff the set of axioms $\jwsmo \setminus \jwsmd$ can be extended by a logical description $EX$ such that:
\begin{enumerate}
    \item $(\jwsmo \setminus \jwsmd) \cup \jwsmb \cup EX$ is consistent (and coherent if required)
	\item $(\jwsmo \setminus \jwsmd) \cup \jwsmb \cup EX \models \jwste\quad \forall\jwste \in \jwsTe$
    \item $(\jwsmo \setminus \jwsmd) \cup \jwsmb \cup EX \not\models \jwstne\quad \forall\jwstne \in \jwsTne$
\end{enumerate}
\end{definition}

A diagnosis $\jwsmd_i$ defines a partition of the ontology $\jwsmo$ where  each axiom $\jwstax_j \in \jwsmd_i$ is a candidate for changes by the user and each axiom $\jwstax_k \in \jwsmo\setminus\jwsmd_i$ is correct. 
If $\jwsmd_t$ is the set of axioms of $\jwsmo$ to be changed (i.e.\ $\jwsmd_t$ is the target diagnosis) then the target ontology $\jwsmo_t$ is $(\jwsmo \setminus \jwsmd_t) \cup \jwsmb \cup EX$ for some $EX$ defined by the user. 

In the following we assume the background theory $\jwsmb$ together with the sets of logical sentences in the sets $\jwsTe$ and $\jwsTne$ always allow formulation of the target ontology. Moreover, a diagnosis exists iff a target ontology exists.

\begin{proposition}
A diagnosis $\jwsmd$ for a diagnosis problem instance $\tuple{\jwsmo,\jwsmb,\jwsTe,\jwsTne}$ exists iff 

\begin{equation*}
\jwsmb \cup \bigcup_{\jwste \in \jwsTe}\jwste
\end{equation*}
is consistent (coherent) and 
\begin{equation*}
\forall \jwstne \in \jwsTne \;:\;  \jwsmb \cup \bigcup_{\jwste \in \jwsTe}\jwste \not\models \jwstne
\end{equation*}
\end{proposition}

The set of all diagnoses is complete in the sense that at least one diagnosis exists where the ontology resulting from the trivial application of a diagnosis is a subset of the target ontology:

\begin{proposition}
Let $\jwsmD \neq \emptyset$ be the set of all diagnoses for a diagnosis problem instance $\tuple{\jwsmo,\jwsmb,\jwsTe,\jwsTne}$ and $\jwsmo_t$ the target ontology. Then a diagnosis $\jwsmd_t \in \jwsmD$ exists s.t.\ $(\jwsmo \setminus \jwsmd_t) \subseteq \jwsmo_t$. 
\end{proposition}

The set of all diagnoses can be characterized by the set of minimal diagnoses. 

\begin{definition}
A diagnosis $\jwsmd$ for a diagnosis problem instance $\tuple{\jwsmo, \jwsmb,\jwsTe,\jwsTne}$ is a \emph{minimal diagnosis} iff there is no $\jwsmd^\prime \subset \jwsmd$ such that $\jwsmd^\prime$ is a diagnosis. 
\end{definition}

%With the requirement that entailment is monotonic, the following property holds:

\begin{proposition}
Let $\tuple{\jwsmo, \jwsmb,\jwsTe,\jwsTne}$ be a diagnosis problem instance. For every diagnosis $\jwsmd$ there is a minimal diagnosis $\jwsmd'$ s.t. $\jwsmd' \subseteq \jwsmd$. 
\end{proposition}

\begin{definition}
A diagnosis $\jwsmd$ for a diagnosis problem instance $\tuple{\jwsmo, \jwsmb,\jwsTe,\jwsTne}$ is a \emph{minimum cardinality diagnosis} iff there is no diagnosis $\jwsmd^\prime$ such that $|\jwsmd^\prime|<|\jwsmd|$. 
\end{definition}

To summarize, a diagnosis describes which axioms are candidates for modification. Despite the fact that multiple diagnoses may exist, some are more preferable than others. E.g.\ minimal diagnoses require minimal changes, i.e.\ axioms are not considered for modification unless there is a reason. Minimal cardinality diagnoses require changing a minimal number of axioms. The actual type of error contained in an axiom is irrelevant as the concept of diagnosis defined here does not make any assumptions about errors themselves. There can, however, be instances where an ontology is faulty and the empty diagnosis is the only minimal diagnosis, e.g. if some axioms are missing and nothing must be changed. 

The extension $EX$ plays an important role in the ontology repair process, suggesting axioms that should be added to the ontology. For instance, in Example~\ref{ex:simple} the user requires that the target ontology \emph{must not} entail $\setof{B(w)}$ but has to entail $\setof{B(v)}$, that is $\jwsTne=\{\{B(w)\}\}$ and $\jwsTe=\{\{B(v)\}\}$. Because, the example ontology $\jwsmo$ is inconsistent some sentences must be changed. The consistent ontology $\jwsmo_1=\jwsmo \setminus \jwsmd_1$ (along with the background axioms $\jwsmb$) neither entails $\setof{B(v)}$ nor $\setof{B(w)}$ (in particular $\jwsmo_1 \cup \jwsmb \models \setof{\lnot B(w)}$). Consequently, $\jwsmo_1$ has to be extended with a set $EX$ of logical sentences in order to entail $\setof{B(v)}$. This set of logical sentences can be approximated with $EX=\{B(v)\}$. $\jwsmo_1 \cup \mb \cup EX$ is satisfiable, entails $\setof{B(v)}$ but does not entail $\setof{B(w)}$. 
All other ontologies $O_i = \jwsmo \setminus \jwsmd_i, \; i=2,3,4$ (integrated with $\jwsmb$) are consistent but entail $\setof{B(w), B(v)}$ and must be rejected because of the monotonic semantics of description logic. That is, there is no such extension $EX$ that $(\jwsmo_i \cup \jwsmb \cup EX) \not\models \setof{B(w)}$. Therefore, the diagnosis $\jwsmd_1$ is the minimum cardinality diagnosis which allows the formulation of the target ontology. 
Note that formulation of the complete extension is impossible, since our diagnosis approach deals with changes to existing axioms and does not learn new axioms.

The following corollary characterizes diagnoses without employing the true extension $EX$ to formulate the target ontology. The idea is to use the sentences which must be entailed by the target ontology to approximate $EX$ as shown above.
\begin{corollary}
Given a diagnosis problem instance $\tuple{\jwsmo, \jwsmb,\jwsTe,\jwsTne}$, a set of axioms $\jwsmd \subseteq \jwsmo$ is a diagnosis iff 
\begin{flalign*}
(\jwsmo \setminus \jwsmd) \cup \jwsmb \cup \bigcup_{\jwste \in \jwsTe}\jwste \qquad & \text{(Condition~1)} \\
\text{is satisfiable (coherent) and}\qquad\qquad & \\
%\begin{equation*}
\forall \jwstne \in \jwsTne\;:\; (\jwsmo \setminus \jwsmd) \cup \jwsmb \cup \bigcup_{\jwste \in \jwsTe}\jwste \not\models \jwstne \qquad& \text{(Condition~2)}
\end{flalign*}
 \end{corollary}

\noindent \textbf{Proof sketch:} $(\Rightarrow)$ Let $\jwsmd \subseteq \jwsmo$ be a diagnosis for $\tuple{\jwsmo, \jwsmb,\jwsTe,\jwsTne}$. Since there is an $EX$ s.t.\ $(\jwsmo \setminus \jwsmd) \cup \jwsmb \cup EX$ is satisfiable (coherent) and $(\jwsmo \setminus \jwsmd) \cup \jwsmb \cup EX \models \jwste$ for all $\jwste \in \jwsTe$, it follows that $(\jwsmo \setminus \jwsmd) \cup \jwsmb \cup EX \cup \bigcup_{\jwste \in \jwsTe}\jwste$ is satisfiable (coherent) and therefore $(\jwsmo \setminus \jwsmd) \cup \jwsmb \cup \bigcup_{\jwste \in \jwsTe}\jwste$ is satisfiable (coherent). Consequently, the first condition of the corollary is fulfilled. Since $(\jwsmo \setminus \jwsmd) \cup \jwsmb \cup EX \models \jwste$ for all $\jwste \in \jwsTe$ and $(\jwsmo \setminus \jwsmd) \cup \jwsmb \cup EX \not\models \jwstne$ for all $\jwstne \in \jwsTne$ it follows that $(\jwsmo \setminus \jwsmd) \cup \jwsmb \cup EX \cup \bigcup_{\jwste \in \jwsTe}\jwste \not\models \jwstne$ for all $\jwstne \in \jwsTne$. Consequently, $(\jwsmo \setminus \jwsmd) \cup \jwsmb \cup \bigcup_{\jwste \in \jwsTe}\jwste \not\models \jwstne$ for all $\jwstne \in \jwsTne$ and the second condition of the corollary is fulfilled. 

$(\Leftarrow)$ Let $\jwsmd \subseteq \jwsmo$ and $\tuple{\jwsmo, \jwsmb,\jwsTe,\jwsTne}$ be a diagnosis problem instance. Without limiting generality let $EX = \jwsTe$. By Condition 1 of the corollary $(\jwsmo \setminus \jwsmd) \cup \jwsmb \cup \bigcup_{\jwste \in \jwsTe}\jwste$ is satisfiable (coherent). Therefore, for $EX = \jwsTe$ the sentences $(\jwsmo \setminus \jwsmd) \cup \jwsmb \cup EX$ are satisfiable (coherent), i.e.\ the first condition for a diagnosis is fulfilled and these sentences entail $\jwste$ for all $\jwste \in \jwsTe$ which corresponds to the second condition a diagnosis must fulfill. Furthermore, by Condition 2 of the corollary $(\jwsmo \setminus \jwsmd) \cup \jwsmb \cup EX \not\models \jwstne$ for all $\jwstne \in \jwsTne$ holds and therefore the third condition for a diagnosis is fulfilled. Consequently, $\jwsmd \subseteq \jwsmo$ is a diagnosis for $\tuple{\jwsmo, \jwsmb,\jwsTe,\jwsTne}$.\qed

\emph{Conflict sets}, which are the parts of the ontology that preserve the inconsistency/incoherency, are usually employed to constrain the search space during computation of diagnoses.
\begin{definition}
Given a diagnosis problem instance $\tuple{\jwsmo,\jwsmb,\jwsTe,\jwsTne}$,  a set of axioms $CS \subseteq \jwsmo$ is a conflict set iff  $CS \cup \jwsmb \cup \bigcup_{\jwste \in \jwsTe}\jwste$ is inconsistent (incoherent) or $\jwstne \in \jwsTne$ exists s.t.\ $CS \cup \jwsmb \cup \bigcup_{\jwste \in \jwsTe}\jwste \models \jwstne$.
\end{definition}

\begin{definition}
A conflict set $CS$ for an instance $\tuple{\jwsmo,\jwsmb,\jwsTe,\jwsTne}$ is minimal iff there is no $CS^\prime \subset CS$ such that $CS^\prime$ is a conflict set.
\end{definition}
A set of minimal conflict sets can be used to compute the set of minimal diagnoses as shown in~\cite{Reiter87}. The idea is that each diagnosis must include at least one element of each minimal conflict set. 
\begin{proposition}\label{prop:hittingset}
$\jwsmd$ is a minimal diagnosis for the diagnosis problem instance $\tuple{\jwsmo, \jwsmb,\jwsTe,\jwsTne}$ iff $\jwsmd$ is a minimal hitting set for the set of all minimal conflict sets of $\tuple{\jwsmo, \jwsmb,\jwsTe,\jwsTne}$.
\end{proposition}
Given a set of sets $\overline S$, a set $H$ is a hitting set of $\overline S$  iff $H \cap S_i \neq \emptyset$ for all $S_i \in \overline S$ and $H \subseteq \bigcup_{S_i \in \overline S} S_i$. 
Most modern ontology diagnosis methods~\cite{Schlobach2007,Kalyanpur.Just.ISWC07,friedrich2005gdm,Horridge2008} are implemented according to Proposition~\ref{prop:hittingset} and differ only in details, such as how and when (minimal) conflict sets are computed, the order in which hitting sets are generated, etc.

\section*{Differentiating between Diagnoses}\label{sect:discrimination}

The diagnosis method usually generates a set of diagnoses for a given diagnosis problem instance. Thus, in Example~\ref{ex:simple} an ontology debugger returns a set of four minimal diagnoses $\{\jwsmd_1, \dots, \jwsmd_4\}$. 
As explained in the previous section, additional information, i.e.\ sets of sets of logical sentences $\jwsTe$ and $\jwsTne$, can be used by the debugger to reduce the set of diagnoses. However, in the general case the user does not know which sets $\jwsTe$ and $\jwsTne$ to provide to the debugger such that the target diagnosis will be identified. Therefore, the debugger should be able to identify sets of logical sentences on its own and only ask the user or some other oracle, whether these sentences \emph{must} or \emph{must not} be entailed by the target ontology. To generate these sentences the debugger can apply each of the diagnoses in $\jwsmD=\{\jwsmd_1,\dots,\jwsmd_n\} $ and obtain a set of ontologies $\jwsmo_i = \jwsmo \setminus \jwsmd_i\:,\: i=1,\dots, n$ that fulfill the user requirements. For each ontology $\jwsmo_i$ a description logic reasoner can generate a set of entailments such as entailed subsumptions provided by the classification service and sets of class assertions provided by the realization service. 
%In fact, the intention of the classification is that a model for a specific application domain can be verified by exploiting the subsumption hierarchy~\cite{DLHandbook}.
These entailments can be used to discriminate between the diagnoses, as different ontologies entail different sets of sentences due to extensivity of the entailment relation. 
Note that in the examples provided in this section we consider only two types of entailments, namely subsumption and class assertion.
%In the following example we consider only two types of entailments that can be computed by a description logic reasoner, namely subsumptions and class assertions. 
In general, the approach presented in this work is not limited to these types and can use all of the entailment types supported by a reasoner. 

\renewcommand{\arraystretch}{1} 
\begin{table}%
\centering
\begin{tabular}{cl}
Ontology & Entailments \\ \hline
$\jwsmo_1$&$\emptyset$ \\
$\jwsmo_2$&$\{B(w)\}$ \\
$\jwsmo_3$&$\{B(w),C(w)\}$ \\
$\jwsmo_4$&$\{B(w),C(w), D(w)\}$ \\\hline
\end{tabular}
\caption[Entailments of Ontologies Repaired by Different Diagnoses]{Entailments of ontologies $\jwsmo_i = (\jwsmo \setminus \jwsmd_i)\:, \: i=1,\dots, 4$ (integrated with $\mb$) in Example~\ref{ex:simple} returned by realization.}
\label{tab:entailex1}
\end{table}

For instance, in Example~\ref{ex:simple} for each ontology $\jwsmo_i = (\jwsmo \setminus \jwsmd_i) \:,\: i=1\dots 4$ (integrated with $\mb$) the realization service of a reasoner returns the set of class assertions presented in Table~\ref{tab:entailex1}.
Without any additional information the debugger cannot decide which of these sentences must be entailed by the target ontology. To obtain this information the diagnosis method must query an oracle that can specify whether the target ontology entails some set of sentences or not. E.g.\ the debugger could ask an oracle if $\setof{D(w)}$ is entailed by the target ontology ($\jwsmo_t \models \setof{D(w)}$). 
If the answer is \emph{yes}, then $\setof{D(w)}$ is added to $\jwsTe$ and $\jwsmd_4$ is considered as the target diagnosis. All other diagnoses are rejected because $(\jwsmo \setminus \jwsmd_i) \cup \jwsmb \cup \{D(w)\}$ for $i=1,2,3$ is inconsistent.  If the answer is \emph{no}, then $\setof{D(w)}$ is added to $\jwsTne$ and $\jwsmd_4$ is rejected as $(\jwsmo \setminus \jwsmd_4) \cup \jwsmb \models \setof{D(w)}$ and we have to ask the oracle another question. In the following we consider a query $\jwsqry$ as a set of logical sentences such that $\jwsmo_t \models \jwsqry$ holds iff $\jwsmo_t \models q_i$ for all $q_i \in \jwsqry$.

\begin{property}
Given a diagnosis problem instance $\tuple{\jwsmo, \jwsmb,\jwsTe,\jwsTne}$, a set of diagnoses $\jwsmD$, a set of logical sentences $\jwsqry$
representing the query $(\jwsmo_t \models \jwsqry)\,$ and an oracle able to evaluate the query:

If the oracle  answers \emph{yes} then every diagnosis $\jwsmd_i \in \jwsmD$ is a diagnosis for $\jwsTe \cup \{ \jwsqry \}$ iff both conditions hold:
\begin{align*}
(\jwsmo \setminus \jwsmd_i)& \cup \jwsmb \cup \bigcup_{\jwste \in \jwsTe}\jwste \cup \jwsqry \; \emph{is consistent (coherent)}\\
\forall \jwstne \in \jwsTne & \;:\;(\jwsmo \setminus \jwsmd_i)  \cup \jwsmb \cup \bigcup_{\jwste \in \jwsTe}\jwste \cup \jwsqry \not\models \jwstne
\end{align*}

If the oracle answers \emph{no} then every diagnosis $\jwsmd_i \in \jwsmD$ is a diagnosis for $\jwsTne \cup \{ \jwsqry \}$ iff both conditions hold:
\begin{align*}
(\jwsmo \setminus \jwsmd_i)& \cup \jwsmb \cup \bigcup_{\jwste \in \jwsTe}\jwste\; \emph{is consistent (coherent)} \\
\forall \jwstne \in 
(\jwsTne \cup \{ \jwsqry \}) & \;:\; (\jwsmo \setminus \jwsmd_i) \cup \jwsmb \cup \bigcup_{\jwste \in \jwsTe}\jwste \not\models \jwstne
\end{align*}
\end{property}

In particular, a query partitions the set of diagnoses $\jwsmD$ into three disjoint subsets.

\begin{definition}\label{def:partition}
For a query $\jwsqry$, each diagnosis $\jwsmd_i \in \jwsmD$ of a diagnosis problem instance $\tuple{\jwsmo, \jwsmb,\jwsTe,\jwsTne}$ can be assigned to one of the three sets $\jwsdx$, $\jwsdnx$ or $\jwsdz$
where
\begin{itemize}
\item $\jwsmd_i \in \jwsdx$ iff it holds that
\begin{equation*}
(\jwsmo \setminus \jwsmd_i) \cup \jwsmb \cup \bigcup_{\jwste \in \jwsTe}\jwste \models \jwsqry
\end{equation*}
\item $\jwsmd_i \in \jwsdnx$ iff it holds that
\begin{equation*}
(\jwsmo \setminus \jwsmd_i) \cup \jwsmb \cup \bigcup_{\jwste \in \jwsTe}\jwste \cup  \jwsqry
\end{equation*}
is inconsistent (incoherent). 
\item $\jwsmd_i \in \jwsdz$ iff $\jwsmd_i \in \jwsmD\setminus\left(\jwsdx \cup \jwsdnx\right)$
\end{itemize}
\end{definition}

Given a diagnosis problem instance we say that the diagnoses in $\jwsdx$ predict a positive answer (\emph{yes}) as a result of the query $\jwsqry$, diagnoses in $\jwsdnx$ predict a negative answer (\emph{no}), and diagnoses in $\jwsdz$ do not make any predictions. 

\begin{property}\label{prop:removeDiag}
Given a diagnosis problem instance $\tuple{\jwsmo, \jwsmb,\jwsTe,\jwsTne}$, a set of diagnoses $\jwsmD$, a query $\jwsqry$ and an oracle:

If the oracle answers \emph{yes} then the set of rejected diagnoses is $\jwsdnx$ and the set of remaining diagnoses is $\jwsdx\cup \jwsdz$.

If the oracle answers \emph{no} then the set of rejected diagnoses is $\jwsdx$ and the set of remaining diagnoses is $\jwsdnx \cup \jwsdz$.
\end{property}

Consequently, given a query $\jwsqry$ either $\jwsdx$ or $\jwsdnx$ is eliminated but $\jwsdz$ always remains after the query is answered. For generating queries we have to investigate for which subsets $\jwsdx, \jwsdnx \subseteq \jwsmD$ a query exists that can differentiate between these sets. A straight forward approach is to investigate all possible subsets of $\jwsmD$. In our evaluation we show that this is feasible if we limit the number $n$ of minimal diagnoses to be considered during query generation and selection. E.g.\ for $n=9$, the algorithm has to verify $512$ possible partitions in the worst case.

Given a set of diagnoses $\jwsmD$ for the ontology $\jwsmo$, a set $\jwsTe$ of sets of sentences that must be entailed by the target ontology $\jwsmo_t$ and a set of background axioms $\jwsmb$, the set of partitions $\bf{PR}$ for which a query exists can be computed as follows:
\begin{enumerate}
	\item Generate the power set $\pset{\jwsmD}$, $\bf{PR}\leftarrow\emptyset$
    \item Assign an element of $\pset{\jwsmD}$ to the set $\jwsdxi{i}$ and generate a set of common entailments $\jwsent_{i}$ of all ontologies $(\jwsmo \setminus \jwsmd_j) \cup \jwsmb \cup \bigcup_{\jwste \in \jwsTe}\jwste$, where $\jwsmd_j \in \jwsdxi{i}$
    \item If $\jwsent_{i} = \emptyset$, then reject the current element $\jwsdxi{i}$, 
    %remove it from 
    i.e.\ set $\pset{\jwsmD} \leftarrow \pset{\jwsmD} \setminus \{\jwsdxi{i}\}$ and goto Step 2. Otherwise set $\jwsqry_i \leftarrow \jwsent_i$.
    \item Use Definition~\ref{def:partition} and the query $\jwsqry_i$ to classify the diagnoses $\jwsmd_k \in \jwsmD \setminus \jwsdxi{i}$ into the sets $\jwsdxi{i}$, $\jwsdnxi{i}$ and $\jwsdzi{i}$. The generated partition is added to the set of partitions $\mathbf{PR} \leftarrow \mathbf{PR} \cup \{\tuple{\jwsqry_i,\jwsdxi{i},\jwsdnxi{i},\jwsdzi{i}}\}$ and set $\pset{\jwsmD} \leftarrow \pset{\jwsmD}\setminus \{\jwsdxi{i}\}$. If $\pset{\jwsmD} \neq \emptyset$ then go to Step 2.
\end{enumerate}

In Example~\ref{ex:simple} the set of diagnoses $\jwsmD$ of the ontology $\jwsmo$ contains 4 elements. Therefore, the power set $\pset{\jwsmD}$ includes 15 elements $\setof{\{\jwsmd_1\},\setof{\jwsmd_2},\dots, \setof{\jwsmd_1,\jwsmd_2,\jwsmd_3,\jwsmd_4}}$, assuming we omit the element corresponding to $\emptyset$ as it does not contain any diagnoses to be evaluated. Moreover, assume that $\jwsTe$ and $\jwsTne$ are empty. 
In each iteration an element of $\pset{\jwsmD}$ is assigned to the set $\jwsdxi{i}$. For instance, the algorithm assigns $\jwsdxi{1}=\{\jwsmd_1,\jwsmd_2\}$. In this case the set of common entailments is empty as $(\jwsmo\setminus\jwsmd_1) \cup \jwsmb$ has no entailed 
sentences 
%except for the axioms in $(\jwsmo\setminus\jwsmd_1) \cup \jwsmb$
%instances 
%(in addition to the given class assertions, 
(see Table~\ref{tab:entailex1}). Therefore, the set $\{\jwsmd_1,\jwsmd_2\}$ is rejected and removed from $\pset{\jwsmD}$. Assume that in the next iteration the algorithm selects $\jwsdxi{2}=\{\jwsmd_2,\jwsmd_3\}$. 
In this case the set of common entailments $\jwsent_2 = \setof{B(w)}$ is not empty and so $\jwsqry_2=\{B(w)\}$. 
The remaining diagnoses $\jwsmd_1$ and $\jwsmd_4$ are classified according to Definition~\ref{def:partition}. That is, the algorithm selects the first diagnosis $\jwsmd_1$ and verifies whether $(\jwsmo \setminus \jwsmd_1) \cup \jwsmb \models \setof{B(w)}$. 
Given the negative answer of the reasoner, the algorithm checks if $(\jwsmo \setminus \jwsmd_1) \cup \jwsmb \cup \{B(w)\}$ is inconsistent. Since the condition is satisfied the diagnosis $\jwsmd_1$ is added to the set $\jwsdnxi{2}$. 
The second diagnosis $\jwsmd_4$ is added to the set $\jwsdxi{2}$ as it satisfies the first requirement $(\jwsmo \setminus \jwsmd_4) \cup \jwsmb \models \{B(w)\}$. 
The resulting partition $\tuple{\{B(w)\},\{\jwsmd_2,\jwsmd_3,\jwsmd_4\},\{\jwsmd_1\}, \emptyset}$ is added to the set $\bf{PR}$. 

However, a query need not include all of the entailed sentences. If a query $\jwsqry$ partitions the set of diagnoses into $\jwsdx$, $\jwsdnx$ and $\jwsdz$ and an (irreducible) subset $\jwsqry' \subset \jwsqry$ exists which preserves the partition then it is sufficient to query $\jwsqry'$. In our example, $\jwsqry_2:\{B(w),C(w)\}$ can be reduced to its subset $\jwsqry'_2:\{C(w)\}$.  If there are multiple irreducible subsets that preserve the partition then we select one of them.

All of the queries and their corresponding partitions generated in Example~\ref{ex:simple} are presented in Table~\ref{tab:example1}. 
Given these queries the debugger has to decide which one should be asked first in order to minimize the number of queries to be answered. A popular query selection heuristic (called ``split-in-half'') prefers queries which allow half of the diagnoses to be removed from the set $\jwsmD$ regardless of the answer of an oracle. 

Using the data presented in Table~\ref{tab:example1}, the ``split-in-half'' heuristic determines that asking the oracle if $(\jwsmo_t \models \{C(w)\})$ is the best query (i.e.\ the reduced query $\jwsqry_2$), as two diagnoses from the set $\jwsmD$ are removed regardless of the answer. Assuming that $\jwsmd_1$ is the target diagnosis, then an oracle will answer $no$ to our question (i.e.\ $\jwsmo_t \not\models \setof{C(w)}$). Based on this feedback, the diagnoses $\jwsmd_3$ and $\jwsmd_4$ are removed according to Property~\ref{prop:removeDiag}. Given the updated set of diagnoses $\jwsmD$ and $\jwsTe=\setof{\setof{C(w)}}$ the partitioning algorithm returns the only partition $\tuple{\setof{B(w)},\setof{\jwsmd_2},\setof{\jwsmd_1}, \emptyset}$. The heuristic then selects the query $\setof{B(w)}$, which is also answered with $no$ by the oracle. 
Consequently, $\jwsmd_1$ is identified as the only remaining minimal diagnosis. 

\renewcommand{\arraystretch}{1} 
\begin{table*}[tb]
\centering
\begin{tabular}{@{\extracolsep{8pt}} llll}
Query & $\jwsdx$  & $\jwsdnx$ &  $\jwsdz$ \\\hline
$\jwsqry_1:\{B(w)\}$ & $\{\jwsmd_2,\jwsmd_3,\jwsmd_4\}$ & $\{\jwsmd_1\}$ & $\emptyset$ \\
$\jwsqry_2:\{B(w),C(w)\}$ & $\{\jwsmd_3, \jwsmd_4 \}$   & $\{\jwsmd_1, \jwsmd_2\}$ & $\emptyset$ \\
$\jwsqry_3:\{B(w),C(w),Q(w)\}$ & $\{\jwsmd_4\}$         & $\{\jwsmd_1, \jwsmd_2, \jwsmd_3 \}$ &
$\emptyset$ \\\hline
\end{tabular}
\caption[(Example~\ref{ex:simple}) Possible Queries]{Possible queries in Example~\ref{ex:simple}}\label{tab:example1}
\end{table*}

In general, if $n$ is the number of diagnoses and we can split the set of diagnoses in half with each query, then the minimum number of queries is
$log_2{n}$. Note that this minimum number of queries can only be achieved when all minimal diagnoses are considered at once, which is intractable even for relatively small values of $n$. 
%However, if the 

However, in case probabilities of diagnoses are known we
can reduce the number of queries by utilizing two effects: 
\begin{enumerate}
	\item We can exploit diagnoses probabilities to assess the likelihood of each answer and the expected value of the information contained in the set of diagnoses after an answer is given. 
    \item Even if multiple diagnoses remain, further query generation may not be required if one diagnosis is highly probable and all other remaining diagnoses are highly improbable.
\end{enumerate}

\begin{example}\label{ex:complex} Consider an ontology $\jwsmo$ with the terminology $\jwsmt$:
%\vspace{-3pt}
\begin{center}
\begin{tabular}{ll}
$\jwstax_1 : A_1 \sqsubseteq A_2 \sqcap M_1 \sqcap M_2$ & 
$\jwstax_4 : M_2 \sqsubseteq \forall s.A \sqcap D$ \\
$\jwstax_2 : A_2 \sqsubseteq \lnot\exists s.M_3 \sqcap \exists s.M_2$ & 
$\jwstax_5 : M_3 \equiv B \sqcup C$ \\
$\jwstax_3 : M_1 \sqsubseteq \lnot A \sqcap B$ & \\
\end{tabular}
\end{center}
%\vspace{-2pt}
and the background theory containing the assertions $\jwsma :\{A_1(w), A_1(u), s(u,w)\}$. 

The ontology along with the background theory is inconsistent and 
the set of minimal conflict sets $CS =\{\left<\jwstax_1,\jwstax_3,\jwstax_4\right>, $ $\langle\jwstax_1,\jwstax_2$, $\jwstax_3,\jwstax_5\rangle\}$.
%includes two minimal conflict sets:  $\{\left<\jwstax_1,\jwstax_3,\jwstax_4\right>, $ $\left<\jwstax_1,\jwstax_2,\jwstax_3,\jwstax_5\right>\}$. 
To restore consistency, the user should modify all axioms of at least one minimal diagnosis: 
%\vspace{-2pt}
\begin{align*}
\jwsmd_1&:\left[\jwstax_1\right]  &\jwsmd_3&:\left[\jwstax_4,\jwstax_5\right]\\
\jwsmd_2&:\left[\jwstax_3\right]  &\jwsmd_4&:\left[\jwstax_4,\jwstax_2\right] 
\end{align*}\qed
\end{example}

\begin{figure}[b]
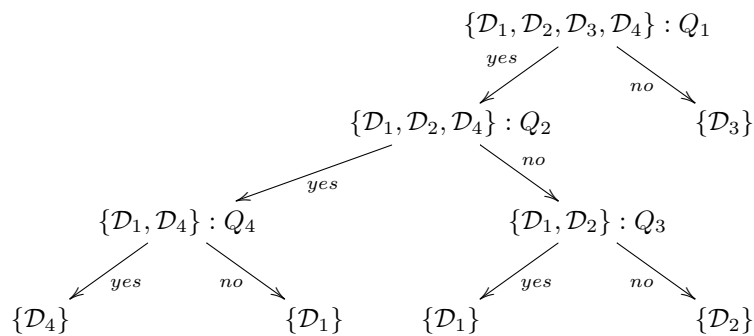

\centerline{
\xygraph{
!{<0cm,0cm>;<1.8cm,0cm>:<0cm,1.3cm>::}
!{(0,0) }*+{\{\jwsmd_4\}}="d4"
!{(2,0) }*+{\{\jwsmd_1\}}="d14"
!{(3,0) }*+{\{\jwsmd_1\}}="d13"
!{(5,0) }*+{\{\jwsmd_2\}}="d2"
!{(1,1) }*+{\{\jwsmd_1, \jwsmd_4\} : \jwsqry_4}="x4"
!{(4,1) }*+{\{\jwsmd_1, \jwsmd_2\} : \jwsqry_3}="x3"
!{(3,2) }*+{\{\jwsmd_1, \jwsmd_2, \jwsmd_4\} : \jwsqry_2}="x2"
!{(5,2) }*+{\{\jwsmd_3\}}="d3"
!{(4,3)}*+{\{\jwsmd_1, \jwsmd_2, \jwsmd_3, \jwsmd_4\} : \jwsqry_1}="x1"
"x4":"d4"^{yes}
"x4":"d14"_{no}
"x3":"d13"^{yes}
"x3":"d2"_{no}
"x2":"x4"^{yes}
"x2":"x3"^{no}
"x1":"d3"_{no}
"x1":"x2"_{yes}
}
}
\caption[The Search Tree of the Greedy Algorithm]{The search tree of the greedy algorithm} \label{fig:greedy:ex}
    %\vspace{-20pt}
\end{figure}

Following the same approach as in Example~\ref{ex:simple}, we compute a set of possible queries and corresponding partitions using the algorithm presented above.
A set of possible irreducible queries for Example~\ref{ex:complex} and their partitions are presented in Table~\ref{tab:example2}.
These queries partition the set of diagnoses ${\bf D}$ in a way that makes the application of myopic strategies, such as ``split-in-half'', inefficient.
A greedy algorithm based on such a heuristic would first select the first query $\jwsqry_1$,
% as the next query, 
since there is no query that cuts the set of diagnoses in half. If $\jwsmd_4$ is the target diagnosis then $\jwsqry_1$ will be answered with $yes$ by an oracle (see Figure~\ref{fig:greedy:ex}). In the next iteration the algorithm would also choose a suboptimal query, the first untried query $\jwsqry_2$, since there is no partition that divides the diagnoses $\jwsmd_1$, $\jwsmd_2$, and $\jwsmd_4$ into two groups of equal size. Once again, the oracle answers $yes$, and the algorithm identifies query $\jwsqry_4$ to differentiate between $\jwsmd_1$ and $\jwsmd_4$.

\renewcommand{\arraystretch}{1} 
\begin{table*}[!htb]
\centering
\begin{tabular}{@{\extracolsep{8pt}}llll}
Query & $\jwsdx$  & $\jwsdnx$ &  $\jwsdz$ \\ \hline
$\jwsqry_1:\{B \sqsubseteq M_3\}$ & $\{\jwsmd_1,\jwsmd_2,\jwsmd_4\}$ & $\{\jwsmd_3\}$ & $\emptyset$\\
$\jwsqry_2:\{B(w)\}$ & $\{\jwsmd_3, \jwsmd_4\}$ & $\{\jwsmd_2\}$ & $\{\jwsmd_1\}$ \\
$\jwsqry_3:\{M_1 \sqsubseteq B\}$ & $\{\jwsmd_1,\jwsmd_3,\jwsmd_4\}$ & $\{\jwsmd_2\}$ & $\emptyset$\\
$\jwsqry_4:\{M_1(w), M_2(u)\}$ & $\{\jwsmd_2,\jwsmd_3,\jwsmd_4\}$ & $\{\jwsmd_1\}$ & $\emptyset$  \\
$\jwsqry_5:\{A(w)\}$ & $\{\jwsmd_2\}$ & $\{\jwsmd_3, \jwsmd_4\}$ & $\{\jwsmd_1\}$\\
$\jwsqry_6:\{M_2\sqsubseteq D\}$ & $\{\jwsmd_1,\jwsmd_2\}$ & $\emptyset$ & $\{\jwsmd_3, \jwsmd_4\}$\\
$\jwsqry_7:\{M_3(u)\}$ & $\{\jwsmd_4\}$ & $\emptyset$ & $\{\jwsmd_1, \jwsmd_2, \jwsmd_3\}$\\\hline
\end{tabular}
\caption[(Example~\ref{ex:complex}) Possible Queries]{Possible queries in Example~\ref{ex:complex}}\label{tab:example2}
    %\vspace{-15pt}
\end{table*}

However, in real-world settings the assumption that all axioms fail with the same probability is rarely the case. For example, Roussey et al.~\cite{Roussey2009} present a list of ``anti-patterns'' where an anti-pattern is a set of axioms, such as $\{C1\sqsubseteq\forall R.C2, C1\sqsubseteq\forall R.C3, C2\equiv\lnot C3\}$ that corresponds to a minimal conflict set. The study performed by~\cite{Roussey2009} shows that such conflict sets often occur in practice due to frequent misuse of certain language constructs like quantification or disjointness. Such studies are ideal sources for estimating prior fault probabilities. However, this is beyond the scope of our work presented in this part. 
 
Our approach for computing the prior fault probabilities of axioms is inspired by \cite{Rector2004} and considers the syntax of a knowledge representation language, such as restrictions, conjunction, negation, etc. 
For instance, if a user frequently changes the universal to the existential quantifier and vice versa in order to restore coherency, then we can assume that axioms including such restrictions are more likely to fail than the other ones. 
In~\cite{Rector2004} the authors report that in most cases inconsistent ontologies are created because users (a)~mix up $\forall r.S$ and $\exists r.S$,  (b)~mix up $\lnot\exists r.S$ and $\exists r.\lnot S$, (c)~mix up $\sqcup$ and $\sqcap$, (d)~wrongly assume that classes are disjoint by default or overuse disjointness, or (e)~wrongly apply negation. Observing that misuses of quantifiers are more likely than other failure patterns one might find that the axioms $\jwstax_2$ and $\jwstax_4$ are more likely to be faulty than $\jwstax_3$ (because of the use of quantifiers), whereas $\jwstax_3$ is  more likely to be faulty than $\jwstax_5$ and $\jwstax_1$ (because of the use of negation). 

Detailed justifications of diagnoses probabilities are given in the next section. However, let us assume some probability distribution of the faults according to the observations presented above such that: 
(a)~the diagnosis $\jwsmd_2$ is the most probable one, i.e.\ single fault diagnosis of an axiom containing a negation; 
(b)~although $\jwsmd_4$ is a double fault diagnosis, it follows $\jwsmd_2$ closely as its axioms contain quantifiers; 
(c)~$\jwsmd_1$ and $\jwsmd_3$ are significantly less probable than $\jwsmd_4$ because conjunction/disjunction in $\jwstax_1$ and $\jwstax_5$ have a significantly lower fault probability than negation in $\jwstax_3$. 
Taking this information into account asking query $\jwsqry_1$ is essentially useless because it is highly probable that the target diagnosis is either $\jwsmd_2$ or $\jwsmd_4$ and, therefore, it is highly probable that the oracle will respond with $yes$. 
Instead, asking $\jwsqry_3$ is more informative because regardless of the answer we can exclude one of the highly probable diagnoses, i.e.\ either $\jwsmd_2$ or $\jwsmd_4$. 
If the oracle responds to $\jwsqry_3$ with $no$ then $\jwsmd_2$ is the only remaining diagnosis. 
However, if the oracle responds with $yes$, diagnoses $\jwsmd_4$, $\jwsmd_3$, and $\jwsmd_1$ remain, where $\jwsmd_4$ is significantly more probable compared to diagnoses $\jwsmd_3$ and $\jwsmd_1$. 
If the difference between the probabilities of the diagnoses is high enough such that $\jwsmd_4$ can be accepted as the target diagnosis, no additional questions are required. 
Obviously this strategy can lead to a substantial reduction in the number of queries compared to myopic approaches as we demonstrate in our evaluation.

Note that in real-world application scenarios failure patterns and their probabilities can be discovered by analyzing the debugging actions of a user in an ontology editor, like Prot\'eg\'e. Learning of fault probabilities can be used to ``personalize'' the query selection algorithm to prefer user-specific faults. However, as our evaluation shows, even a rough estimate of the probabilities is capable of outperforming the ``split-in-half'' heuristic. 

%%%%%%%%%%%%%%%%%%%%%%%%%%%%%%%%%%%%%%%%%%%%%%%%%%%%%%%%%%%%%%%%%%%%%%%
%%%%%%%%%%%%%%%%%%%%%%%%%%%%%%%%%%%%%%%%%%%%%%%%%%%%%%%%%%%%%%%%%%%%%%%
%%%%%%%%%%%%%%%%%%%%%%%%%%%%%%%%%%%%%%%%%%%%%%%%%%%%%%%%%%%%%%%%%%%%%%%
%%%%%%%%%%%%%%%%%%%%%%%%%%%%%%%%%%%%%%%%%%%%%%%%%%%%%%%%%%%%%%%%%%%%%%%
%%%%%%%%%%%%%%%%%%%%%%%%%%%%%%%%%%%%%%%%%%%%%%%%%%%%%%%%%%%%%%%%%%%%%%%
%%%%%%%%%%%%%%%%%%%%%%%%%%%%%%%%%%%%%%%%%%%%%%%%%%%%%%%%%%%%%%%%%%%%%%%
%%%%%%%%%%%%%%%%%%%%%%%%%%%%%%%%%%%%%%%%%%%%%%%%%%%%%%%%%%%%%%%%%%%%%%%
%%%%%%%%%%%%%%%%%%%%%%%%%%%%%%%%%%%%%%%%%%%%%%%%%%%%%%%%%%%%%%%%%%%%%%%
%%%%%%%%%%%%%%%%%%%%%%%%%%%%%%%%%%%%%%%%%%%%%%%%%%%%%%%%%%%%%%%%%%%%%%%
%%%%%%%%%%%%%%%%%%%%%%%%%%%%%%%%%%%%%%%%%%%%%%%%%%%%%%%%%%%%%%%%%%%%%%%
%%%%%%%%%%%%%%%%%%%%%%%%%%%%%%%%%%%%%%%%%%%%%%%%%%%%%%%%%%%%%%%%%%%%%%%

\chapter{Entropy-Based Query Selection}\label{chap:jws:theory}

To select the best query we exploit a-priori failure probabilities of each axiom derived from the syntax of description logics or some other knowledge representation language, such as OWL. 
That is, the user is able to specify own beliefs in terms of the probability of syntax element such as $\forall$, $\exists$, $\sqcap$, etc. being erroneous; alternatively, the debugger can compute these probabilities by analyzing the frequency of various syntax elements in the target diagnoses of different debugging sessions. If no failure information is available then the debugger can initialize all of the probabilities with some small value. Compared to statistically well-founded probabilities, the latter approach provides a suboptimal but useful diagnosis discrimination process, as discussed in the evaluation.  

Given the failure probabilities of all syntax elements $se \in {\bf S}$ of a knowledge representation language used in $\jwsmo$, we can compute the failure probability of an axiom $\jwstax_i \in \jwsmo$
\begin{align*}
p(\jwstax_i) = p(F_{se_1} \cup F_{se_2} \cup \dots \cup F_{se_n})
\end{align*}
where $F_{se_1} \dots F_{se_n}$ represent the events that the occurrence of a syntax element $se_j$ in $\jwstax_i$ is faulty. E.g.\ for $ax_2$ of Example~\ref{ex:complex} $p(ax_2) = p( F_\sqsubseteq \cup F_\lnot \cup F_\exists \cup F_\sqcap \cup F_\exists)$.  Assuming that each occurrence of a syntax element fails independently, i.e.\ an erroneous usage of a syntax element $se_k$ makes it neither more nor less probable that an occurrence of syntax element $se_j$ is faulty, the failure probability of an axiom is computed as:
\begin{align}
\label{eq:axiom:prob}
p(\jwstax_i) = 1 - \prod_{se \in {\bf S}} (1-F_{se})^{c(se)}
\end{align}
where $c(se_j)$ returns number of occurrences of the syntax element $se_j$ in an axiom $\jwstax_i$.
If among other failure probabilities the user states that $p(F_\sqsubseteq)=0.001, p(F_\lnot)=0.01, p(F_\exists)=0.05$ and $p(F_\sqcap)=0.001$ then $p(\jwstax_2) = p(F_\sqsubseteq \cup F_\lnot \cup F_\exists \cup F_\sqcap \cup F_\exists) = 0.108$.

Given the failure probabilities $p(\jwstax_i)$ of axioms, the diagnosis algorithm first calculates the a-priori probability $p(\jwsmd_j)$ that $\jwsmd_j$ is the target diagnosis. Since all axioms fail independently, this  probability can be computed as~\cite{dekleer1987}:
\begin{align}
p(\jwsmd_j) = \prod_{\jwstax_n\ \in \jwsmd_j}{p(\jwstax_n)} \prod_{\jwstax_m\ \in \jwsmo\setminus\jwsmd_j}{1-p(\jwstax_m)}
\label{eq:dprob}
\end{align}

The prior probabilities for diagnoses are then used to initialize an iterative algorithm that includes two main steps: (a) the selection of the best query and (b) updating the diagnoses probabilities given query feedback.

According to information theory the best query is the one that, given the answer of an oracle, minimizes the expected entropy of the set of diagnoses~\cite{dekleer1987}. Let $p(\jwsqry_i = yes)$ be the probability that query $\jwsqry_i$ is answered with $yes$ and $p(\jwsqry_i = no)$ be the probability for the answer $no$. Furthermore, let $p(\jwsmd_j | \jwsqry_i = yes)$ be the probability of diagnosis $\jwsmd_j$ after the oracle answers $yes$ and $p(\jwsmd_j | \jwsqry_i = no)$ be the probability after the oracle answers $no$. The expected entropy after querying $\jwsqry_i$ is:
\begin{align*}
H_e(\jwsqry_i) =   \sum_{v \in \setof{yes,no}} p(\jwsqry_i = v)  
 \sum_{\jwsmd_j \in \jwsmD} -p(\jwsmd_j | \jwsqry_i = v) \log_2 p(\jwsmd_j | \jwsqry_i = v)
\end{align*}

Based on a one-step-look-ahead information theoretic measure, the query which minimizes the expected entropy is considered best. This formula can be simplified to the following score function~\cite{dekleer1987} which we use to evaluate all available queries and select the one with the minimum score to maximize information gain:
\begin{align}
%\begin{split}
sc(\jwsqry_i) = \sum_{v \in \setof{yes,no}} \bigl[p(\jwsqry_i=v)\log_2&{p(\jwsqry_i=v)}\bigr] + p(\jwsdzi{i}) + 1
%\end{split}
\label{eq:score}
\end{align}
where $v \in \setof{yes,no}$ is a feedback of an oracle and ${\bf D^{\emptyset}_i}$ is the set of diagnoses which do not make any predictions for the query $\jwsqry_i$. The probability of the set of diagnoses $p(\jwsdzi{i})$ as well as of any other set of diagnoses ${\bf D_{i}}$ like $\jwsdxi{i}$ and $\jwsdnxi{i}$ is computed as: 
\begin{align*}
p({\bf D_{i}}) = \sum_{\jwsmd_j \in {\bf D_{i}}}{p(\jwsmd_j)}
\end{align*}
because by Definition~\ref{def:diag}, each diagnosis uniquely partitions all of the axioms of an ontology $\jwsmo$ into two sets, correct and faulty, and thus all diagnoses are mutually exclusive events. 

Since, for a query $\jwsqry_i$, the set of diagnoses $\jwsmD$ can be partitioned into the sets $\jwsdxi{i}$, $\jwsdnxi{i}$ and $\jwsdzi{i}$, the probability that an oracle will answer a query $\jwsqry_i$ with either $yes$ or $no$ can be computed as:
\begin{align}
\begin{split}
p&(\jwsqry_i=yes)= p(\jwsdxi{i}) + p(\jwsdzi{i})/2 \\
p&(\jwsqry_i=no)= p(\jwsdnxi{i}) + p(\jwsdzi{i})/2
\end{split}
\label{eq:dupdate:prob}
\end{align}

Clearly this assumes that for each diagnosis of $\jwsdzi{i}$ \emph{both outcomes are equally likely} and thus the probability that the set of diagnoses $\jwsdzi{i}$ predicts either $\jwsqry_i=yes$ or $\jwsqry_i=no$ is $p(\jwsdzi{i})/2$.

Following feedback $v$ for a query $\jwsqry_s$, i.e.\ $\jwsqry_s = v$, the probabilities of the diagnoses must be updated to take the new information into account. The update is made using Bayes' rule for each $\jwsmd_j \in \jwsmD$:
\begin{align}
p(\jwsmd_j|\jwsqry_s=v) = \frac{p(\jwsqry_s=v|\jwsmd_j)p(\jwsmd_j)}{p(\jwsqry_s = v)}
\label{eq:dupdate}
\end{align}
where the denominator $p(\jwsqry_s = v)$ is known from the query selection step (Equation~\ref{eq:dupdate:prob}) and $p(\jwsmd_j)$ is either a prior probability (Equation~\ref{eq:dprob}) or is a probability calculated using Equation~\ref{eq:dupdate} after a previous iteration of the debugging algorithm. We assign $p(\jwsqry_s=v|\jwsmd_j)$ as follows:
\begin{align*}
p(\jwsqry_s=v|\jwsmd_j) = 
\begin{cases}
1, & \mbox{if $\jwsmd_j$ predicted $\jwsqry_s=v$;}\\
0, & \mbox{if $\jwsmd_j$ is rejected by $\jwsqry_s=v$;}\\
\frac{1}{2}, & \mbox{if $\jwsmd_j \in \jwsdzi{s}$}
\end{cases} 
\end{align*}

%\vspace{5pt}\noindent\textbf{Example 1 (continued)} 
\begin{example} (Example~\ref{ex:simple} continued)
Suppose that the debugger is not provided with any information about possible failures and therefore assumes that all syntax elements fail with the same probability $0.01$ and therefore $p(\jwstax_i)=0.01$ for all $\jwstax_i \in \jwsmo$. Using Equation~\ref{eq:dprob} we can calculate probabilities for each diagnosis. 
For instance, $\jwsmd_1$ suggests that only one axiom $\jwstax_1$ should be modified by the user. Hence, we can calculate the probability of diagnosis $D_1$ as  
$p(\jwsmd_1) = p(\jwstax_1)(1-p(\jwstax_2))(1-p(\jwstax_3))(1-p(\jwstax_4)) = 0.0097$.
All other minimal diagnoses have the same probability, since every other minimal diagnosis suggests the modification of one axiom. To simplify the discussion we only consider minimal diagnoses for query selection. Therefore, the prior probabilities of the diagnoses can be normalized to $p(\jwsmd_j)=p(\jwsmd_j)/\sum_{\jwsmd_j \in \jwsmD}{p(\jwsmd_j)}$ and are equal to $0.25$.

Given the prior probabilities of the diagnoses and a set of queries (see Table~\ref{tab:example1}) we evaluate the score function (Equation~\ref{eq:score}) for each query. E.g. for the first query $\jwsqry_1:\{B(w)\}$ the probability $p(\jwsdz)=0$ and the probabilities of both the positive and negative outcomes are: 
$p(\jwsqry_1=1)=p(\jwsmd_2)+p(\jwsmd_3)+p(\jwsmd_4) = 0.75$ and $p(\jwsqry_1=0)=p(\jwsmd_1) = 0.25$. Therefore the query score is $sc(\jwsqry_1)=0.1887$. 

The scores computed during the initial stage (see Table~\ref{tab:example1:costs1}) suggest that $\jwsqry_2$ is the best query. Taking into account that $\jwsmd_1$ is the target diagnosis the oracle answers $no$ to the query. The additional information obtained from the answer is then used to update the probabilities of diagnoses using the Equation~\ref{eq:dupdate}. 
Since $\jwsmd_1$ and $\jwsmd_2$ predicted this answer, their probabilities are updated, $p(\jwsmd_1) = p(\jwsmd_2) = 1/p(\jwsqry_2=1)=0.5$. 
The probabilities of diagnoses $\jwsmd_3$ and $\jwsmd_4$ which are rejected by the oracle's answer are also updated, $p(\jwsmd_3) = p(\jwsmd_4) = 0$.

In the next iteration the algorithm recomputes the scores using the updated probabilities. The results show that $\jwsqry_1$ is the best query. The other two queries $\jwsqry_2$ and $\jwsqry_3$ are irrelevant since no information will be gained if they are asked. 
Given the oracle's negative feedback to $\jwsqry_1$, we update the probabilities $p(\jwsmd_1) = 1$ and $p(\jwsmd_2) = 0$. 
In this case the target diagnosis $\jwsmd_1$ was identified using the same number of steps as the ``split-in-half'' heuristic. 

However, if the user specifies that the first axiom is more likely to fail, e.g.\ $p(\jwstax_1) = 0.025$, then $\jwsqry_1:\{B(w)\}$ will be selected first (see Table~\ref{tab:example1:costs2}). The recalculation of the probabilities given the negative outcome $\jwsqry_1=0$ sets $p(\jwsmd_1) = 1$ and $p(\jwsmd_2)=p(\jwsmd_3)=p(\jwsmd_4)=0$. Therefore the debugger identifies the target diagnosis in only one step.\qed
\end{example}

\renewcommand{\arraystretch}{1} 
\begin{table}[tb]
\centering
\begin{tabular}{@{\extracolsep{8pt}}lcc}
Query & Initial score& $\; \jwsqry_2 = yes \;$  \\ \hline
$\jwsqry_1:\{B(w)\}$ & 0.1887 & \textbf{0}  \\
$\jwsqry_2:\{C(w)\}$ & \textbf{0}      & 1  \\
$\jwsqry_3:\{Q(w)\}$ & 0.1887 & 1  \\ \hline
\end{tabular}
\caption[Expected Scores for Minimized Queries (All Axioms Have Equal Fault Probability)]{Expected scores for minimized queries ($p(\jwstax_i)=0.01$)}\label{tab:example1:costs1}
\end{table}
\begin{table}[tb]
\centering
		\begin{tabular}{@{\extracolsep{8pt}}lc}
Query & Initial score   \\ \hline
$\jwsqry_1:\{B(w)\}$ & \textbf{0.250}   \\
$\jwsqry_2:\{C(w)\}$ & 0.408   \\
$\jwsqry_3:\{Q(w)\}$ & 0.629  \\ \hline
		\end{tabular}
        \caption[Expected Scores for Minimized Queries (One Axiom Has Greater Fault Probability than the Rest)]{Expected scores for minimized queries $(p(\jwstax_1)=0.025$, $p(\jwstax_2)=p(\jwstax_3)=p(\jwstax_4) = 0.01)$}\label{tab:example1:costs2}
\end{table}

%\vspace{5pt}\noindent\textbf{Example 2 (continued)} 
%Consider the second example presented in Section~\ref{chap:jws:example}.
\begin{example} (Example~\ref{ex:complex} continued) 
Suppose that in $\jwstax_4$ the user specified $\forall s.A$ instead of $\exists s.A$ and $\lnot\exists s.M_3$ instead of $\exists s.\lnot M_3$ in $\jwstax_2$. Therefore $\jwsmd_4$ is the target diagnosis. Moreover, assume that the debugger is provided with observations of three types of faults: (1) conjunction/disjunction occurs with probability $p_1 = 0.001$, (2) negation $p_2=0.01$, and (3) restrictions $p_3=0.05$. 
Using Equation~\ref{eq:axiom:prob} we can calculate the probability of the axioms containing an error: $p(\jwstax_1)=0.0019$, $p(\jwstax_2)=0.1074$, $p(\jwstax_3)=0.012$, $p(\jwstax_4)=0.051$, and $p(\jwstax_5)=0.001$. These probabilities are exploited to calculate the prior probabilities of the diagnoses (see Table~\ref{tab:example2:diagnoses}) and to initialize the query selection process. To simplify matters we focus on the set of minimal diagnoses.

In the first iteration the algorithm determines that $\jwsqry_3$ is the best query and asks the oracle whether $\jwsmo_t \models \setof{M_1 \sqsubseteq B}$ is true or not (see Table~\ref{tab:example2:costs}). The obtained information is then used to recalculate the probabilities of the diagnoses and to compute the next best subsequent query, i.e.\ $\jwsqry_4$, and so on. The query process stops after the third query, since $\jwsmd_4$ is the only diagnosis that has the probability $p(\jwsmd_4) > 0$. 

\renewcommand{\arraystretch}{1} 
\begin{table*}[tb]
	\centering
    	\begin{tabular}{lcccc}
Answers	&	$\jwsmd_1$	&	$\jwsmd_2$	&	$\jwsmd_3$	&	$\jwsmd_4$	\\ \hline
Prior	&	0.0970	&	0.5874	&	0.0026	&	0.3130	\\
$\jwsqry_3=yes$	&	0.2352	&	0	&	0.0063	&	0.7585	\\
$\jwsqry_3=yes$, $\jwsqry_4=yes$	&	0	&	0	&	0.0082	&	0.9918	\\
$\jwsqry_3=yes$, $\jwsqry_4=yes$, $\jwsqry_1=yes \quad$	&	0	&	0	&	0	&	1	\\ \hline
\end{tabular}
\caption[Probabilities of Diagnoses after Answers]{Probabilities of diagnoses after answers}\label{tab:example2:diagnoses}
\end{table*}

\begin{table*}[tb]
\centering
\begin{tabular}{lccc}
Queries	&	Initial	&	$\jwsqry_3 = yes$	&	$\jwsqry_3 = yes$, $\jwsqry_4 = yes$	\\ \hline
$\jwsqry_1:\{B \sqsubseteq M_3\}$	&	0.974	&	0.945	& \textbf{0.931} \\
$\jwsqry_2:\{B(w)\}$	&	0.151	&	0.713	&	1	\\
$\jwsqry_3:\{M_1 \sqsubseteq B\}$	& \textbf{0.022}&	1	&	1	\\
$\jwsqry_4:\{M_1(w), M_2(u)\}$	&	0.540	& \textbf{0.213}&	1	\\
$\jwsqry_5:\{A(w)\}$	&	0.151	&	0.713	&	1	\\
$\jwsqry_6:\{M_2\sqsubseteq D\}$	&	0.686	&	0.805	&	1	\\
$\jwsqry_7:\{M_3(u)\}$	&	0.759	&	0.710	&	0.970	\\\hline
\end{tabular}
 \caption[Expected Scores for Queries]{Expected scores for queries}\label{tab:example2:costs}
\end{table*}

Given the feedback of the oracle $\jwsqry_4=yes$ for the second query, the updated probabilities of the diagnoses show that the target diagnosis has a probability of $p(\jwsmd_4) = 0.9918$ whereas $p(\jwsmd_3)$ is only $0.0082$. 
In order to reduce the number of queries a user can specify a threshold, e.g. $\sigma=0.95$. 
If the absolute difference in probabilities of two most probable diagnoses is greater than this threshold, the query process stops and returns the most probable diagnosis. 
Therefore, in this example the debugger based on the entropy query selection requires less queries than the ``split-in-half'' heuristic. 
Note that already after the first answer $\jwsqry_3=yes$ the most probable diagnosis $\jwsmd_4$ is three times more likely than the second most probable diagnosis $\jwsmd_1$. Given such a great difference we could suggest to stop the query process after the first answer if the user would set $\sigma=0.65$.\qed 
\end{example}

%%%%%%%%%%%%%%%%%%%%%%%%%%%%%%%%%%%%%%%%%%%%%%%%%%%%%%%%%%%%%%%%%%%%%%%
%%%%%%%%%%%%%%%%%%%%%%%%%%%%%%%%%%%%%%%%%%%%%%%%%%%%%%%%%%%%%%%%%%%%%%%
%%%%%%%%%%%%%%%%%%%%%%%%%%%%%%%%%%%%%%%%%%%%%%%%%%%%%%%%%%%%%%%%%%%%%%%
%%%%%%%%%%%%%%%%%%%%%%%%%%%%%%%%%%%%%%%%%%%%%%%%%%%%%%%%%%%%%%%%%%%%%%%
%%%%%%%%%%%%%%%%%%%%%%%%%%%%%%%%%%%%%%%%%%%%%%%%%%%%%%%%%%%%%%%%%%%%%%%
%%%%%%%%%%%%%%%%%%%%%%%%%%%%%%%%%%%%%%%%%%%%%%%%%%%%%%%%%%%%%%%%%%%%%%%
%%%%%%%%%%%%%%%%%%%%%%%%%%%%%%%%%%%%%%%%%%%%%%%%%%%%%%%%%%%%%%%%%%%%%%%
%%%%%%%%%%%%%%%%%%%%%%%%%%%%%%%%%%%%%%%%%%%%%%%%%%%%%%%%%%%%%%%%%%%%%%%
%%%%%%%%%%%%%%%%%%%%%%%%%%%%%%%%%%%%%%%%%%%%%%%%%%%%%%%%%%%%%%%%%%%%%%%
%%%%%%%%%%%%%%%%%%%%%%%%%%%%%%%%%%%%%%%%%%%%%%%%%%%%%%%%%%%%%%%%%%%%%%%
%%%%%%%%%%%%%%%%%%%%%%%%%%%%%%%%%%%%%%%%%%%%%%%%%%%%%%%%%%%%%%%%%%%%%%%

\chapter{Implementation Details}\label{chap:jws:impl}

The iterative ontology debugger (Algorithm~\ref{algo:general}) takes a faulty  ontology $\jwsmo$ as input. Optionally, a user can provide a set of axioms $\jwsmb$ that are known to be correct as well as a set $\jwsTe$ of axioms that must be entailed by the target ontology and a set $\jwsTne$ of axioms that must not. If these sets are not given, the corresponding input arguments are initialized with $\emptyset$. Moreover, the algorithm takes a set $FP$ of fault probabilities for axioms $\jwstax_i \in \jwsmo$, which can be computed as described in Chapter~\ref{chap:jws:theory} by exploiting knowledge about typical user errors. Alternatively, if no estimates of such probabilities are available, all probability values can be initialized using a small constant. We show the results of such a strategy in our evaluation section. 
The two other arguments $\sigma$ and $n$ are used to improve the performance of the algorithm. $\sigma$ specifies the diagnosis acceptance threshold, i.e. the minimum difference in probabilities between the most likely and second-most likely diagnoses.  The parameter $n$ defines the maximum number of most probable diagnoses that should be considered by the algorithm during each iteration.
A further performance gain in Algorithm~\ref{algo:general} can be achieved if we approximate the set of the $n$ most probable diagnoses with the set of the $n$ most probable \emph{minimal} diagnoses, i.e.\ we neglect non-minimal diagnoses. We call this set of at most $n$ most probable minimal diagnoses the \emph{leading diagnoses}. Note, under the reasonable assumption that the fault probability of each axiom $p(\jwstax_i)$ is less than $0.5$, for every non-minimal diagnosis $ND$ a minimal diagnosis $\jwsmd \subset ND$ exists which from  Equation~\ref{eq:dprob} is more probable than $ND$.
Consequently the query selection algorithm presented here operates on the set of minimal diagnoses instead of all diagnoses (i.e. non-minimal diagnoses are excluded). 
However, the algorithm can be adapted with moderate effort to also consider non-minimal diagnoses.

We use the approach proposed by Friedrich et al.~\cite{friedrich2005gdm} to compute diagnoses and employ the combination of two algorithms, \textsc{QuickXplain}~\cite{junker04} and \textsc{HS-Tree}~\cite{Reiter87}. In a standard implementation the latter is a breadth-first search algorithm that takes an ontology $\jwsmo$, sets $\jwsTe$ and $\jwsTne$, and the maximum number of most probable minimal diagnoses $n$ as an input. The algorithm generates minimal hitting sets using minimal conflict sets, which are computed on-demand. This is motivated by the fact that in some circumstances a subset of all minimal conflict sets is sufficient for generating a subset of all required minimal diagnoses. 
For instance, in Example~\ref{ex:complex} the user wants to compute only $n=2$ leading minimal diagnoses and a minimal conflict search algorithm returns $CS_1$. In this case \textsc{HS-Tree} identifies two required minimal diagnoses $\jwsmd_1$ and $\jwsmd_2$ and avoiding the computation of the minimal conflict set $CS_2$.
Of course, in the worst case, when all minimal diagnoses have to be computed the algorithm should compute all minimal conflict sets.
In addition, the \textsc{HS-Tree} generation reuses minimal conflict sets in order to avoid unnecessary computations. Thus, in the real-world scenarios we evaluated (see Table~\ref{tab:motivation}), less than 10 minimal conflict sets were contained in the faulty ontologies having at most 13 elements while the maximal cardinality of minimal diagnoses was observed to be at most 9. Therefore, space limitations were not a problem for the breadth-first generation. However, for scenarios involving diagnoses of greater cardinalities iterative-deepening strategies could be applied. 

In our implementation of \textsc{HS-Tree} we use the uniform-cost search strategy. Given additional information in terms of axiom fault probabilities $FP$, the algorithm expands a leaf node in a search-tree if it is an element of the path corresponding to the maximum probability hitting set of minimal conflict sets computed so far.
The probability of each minimal hitting set can be computed using Equation~\ref{eq:dprob}. Consequently, the algorithm computes a set of diagnoses ordered by their probability starting from the most probable one.
\textsc{HS-Tree} terminates if either the $n$ most probable minimal diagnoses are identified or no further minimal diagnoses can be found. Thus the algorithm computes at most $n$ minimal diagnoses regardless of the number of all minimal diagnoses. 

\textsc{HS-Tree} uses \textsc{QuickXplain} to compute required minimal conflicts. 
This algorithm, given a set of axioms $AX$ and a set of correct axioms $\jwsmb$ returns a minimal conflict set $CS \subseteq AX$, or $\emptyset$ if axioms $AX \cup \jwsmb$ are consistent.  
In the worst case, to compute a minimal conflict \textsc{QuickXplain} performs $2k (\log (s/k) + 1)$ consistency checks, where $k$ is the size of the generated minimal conflict set and $s$ is the number of axioms in the ontology. 
In the best case only $\log (s/k) + 2k$ are performed~\cite{junker04}. 
Importantly, the size of the ontology is contained in the $\log$ function. Therefore, the time needed for consistency checks in our test ontologies remained below $0.2$ seconds, even for real world knowledge bases with thousands of axioms. The maximum time to compute a minimal conflict was observed in the Sweet-JPL ontology and took approx.\ 5 seconds (see Table~\ref{tab:stats}).

In order to take past answers into account the \textsc{HS-Tree} updates the prior probabilities of the diagnoses by evaluating Equation~\ref{eq:dupdate}. All required data is stored in the query history $QH$ as well as in the sets $\jwsTe$ and $\jwsTne$. When complete, \textsc{HS-Tree} returns a set of tuples of the form $\tuple{\jwsmd_i, p(\jwsmd_i)}$ where $\jwsmd_i$ is contained in the set of the $n$ most probable minimal diagnoses (leading diagnoses) and $p(\jwsmd_i)$ is its probability calculated using Equation~\ref{eq:dprob} and Equation~\ref{eq:dupdate}.

\begin{algorithm*}[tb]
\small
\caption{\textsc{ontoDebugging$(\jwsmo, \jwsmb, P, N, FP, n, \sigma)$}}\label{algo:general}
\begin{algorithmic}[1]
\Require 
ontology $\jwsmo$, set $\jwsmb$ of background axioms, set $\jwsTe$ of sets of logical sentences to be entailed, set $\jwsTne$ of sets of logical sentences not to be entailed, set $FP$ of fault probabilities for axioms, maximum number $n$ of most probable minimal diagnoses, acceptance threshold $\sigma$
\Ensure 
a diagnosis $\jwsmd$
\vspace{10pt}
%%%%%%%%%%%%%%%%%%
%%%%%%%%%%%%%%%%%%
\State $DP \gets \emptyset$
\State $QH \gets \emptyset$
\State $T \gets \tuple{\emptyset,\emptyset,\emptyset,\emptyset}$
\While{$\Call{belowThreshold}{DP,\sigma} \land \Call{getScore}{T}\neq 1$} 
	\State $DP \gets \Call{HS-Tree}{\jwsmo, \jwsmb, \jwsTe, \jwsTne, FP, QH, n}$
	\State $T \gets \Call{selectQuery}{DP, \jwsmo, \jwsmb, \jwsTe}$
	\State $Q  \gets \Call{getQuery}{T}$
	\If {$Q = \emptyset$} 
		\State \textbf{exit loop}
	\EndIf
	\If{$\Call{getAnswer}{\jwsmo_t \models \jwsqry}$}
		\State $\jwsTe \leftarrow \jwsTe \cup \setof{\jwsqry}$
	\Else 
		\State $\jwsTne \leftarrow \jwsTne \cup \setof{\jwsqry}$
	\EndIf
	\State $QH  \leftarrow QH  \cup \setof{T}$\;
\EndWhile		
\State \Return $\Call{mostProbableDiagnosis}{DP}$
\end{algorithmic}
\normalsize
\end{algorithm*}

%%%%%%%%%%%%%%%%%%%%%%%%%%%%%%%%%%%%%%%%%%%%%%%%%%%%%%%%%%%%%%%%%%%%%%%%%%%%%%%%%

\begin{algorithm*}[tb]
\small
\caption{\textsc{selectQuery$(DP, \jwsmo, \jwsmb, \jwsTe)$}} \label{algo:computeQuery}
\begin{algorithmic}[1]
\Require 
set $DP$ of tuples $\tuple{\jwsmd_i, p(\jwsmd_i)}$, ontology $\jwsmo$, set of background axioms $\jwsmb$, set $\jwsTe$ of sets of logical sentences that must be entailed by the target ontology 
\Ensure 
a tuple $\tuple{\jwsqry,\jwsdx,\jwsdnx,\jwsdz}$
\vspace{10pt}
%%%%%%%%%%%%%%%%%%
%%%%%%%%%%%%%%%%%%
\State $\jwsmD \gets \Call{getDiagnoses}{DP}$
\State $T \gets \Call{generate}{\emptyset, \jwsmD, \jwsmo, \jwsmb, \jwsTe, DP}$
\State \Return $\Call{minimizeQuery}{T}$

\vspace{10pt}

\Procedure{$\textsc{generate}$}{$\jwsdx, D, \jwsmo, \jwsmb, \jwsTe, DP$} \textbf{returns} a tuple $\tuple{\jwsqry,\jwsdx,\jwsdnx,\jwsdz}$  
\If{$D = \emptyset$} 
	\State $\jwsmD \gets \Call{getDiagnoses}{DP}$\;  
  \State \Return $\Call{createQuery}{\jwsdx,\jwsmo,\jwsmb,P, \jwsmD}$
\EndIf
\State $\jwsmd \gets \Call{pop}{D}$
\State $\mathit{left} \gets \Call{generate}{\jwsdx, D, \jwsmo, \jwsmb, \jwsTe, DP}$
\State $\mathit{right} \gets \Call{generate}{\jwsdx \cup \setof{\jwsmd}, D, \jwsmo, \jwsmb, \jwsTe, DP}$
\If{$\Call{getScore}{\mathit{left}, DP} < \Call{getScore}{\mathit{right}, DP}$}
	\State \Return $\mathit{left}$
\Else
	\State \Return $\mathit{right}$
\EndIf
\EndProcedure
\end{algorithmic}
\normalsize
\end{algorithm*}

In the query-selection phase Algorithm~\ref{algo:general} calls \textsc{selectQuery} function (Algorithm~\ref{algo:computeQuery}) to generate a tuple $T=\tuple{\jwsqry,\jwsdx,\jwsdnx,\jwsdz}$, where $\jwsqry$ is the minimum score query (Equation~\ref{eq:score}) and $\jwsdx,\jwsdnx$ and $\jwsdz$ the sets of diagnoses constituting the partition. The generation algorithm carries out a depth-first search, removing the top element of the set $D$ and calling itself recursively to generate all possible subsets of the leading diagnoses. The set of leading diagnoses $\jwsmD$ is extracted from the set of tuples $DP$ by the \textsc{getDiagnoses} function.
In each leaf node of the search tree the \textsc{generate} function calls \textsc{createQuery}  creates a query given a set of diagnoses $\jwsdx$ by computing common entailments and partitioning the set of diagnoses $\jwsmD \setminus \jwsdx$, as described in Section~\ref{sect:discrimination}. If a query for the set $\jwsdx$ does not exist (i.e.\ there are no common entailments) or $\jwsdx=\emptyset$ then \textsc{createQuery} returns an empty tuple $T=\tuple{\emptyset,\emptyset,\emptyset,\emptyset}$. 
In all inner nodes of the tree the algorithm selects a tuple that corresponds to a query with the minimum score as found using the \textsc{getScore} function. This function may implement the entropy-based measure (Equation~\ref{eq:score}), ``split-in-half'' or any other preference criteria. Given an empty tuple $T=\tuple{\emptyset,\emptyset,\emptyset,\emptyset}$ the function returns the highest possible score of a used measure. 
In general, \textsc{createQuery} is called $2^n$ times, where we set $n=9$ in our evaluation. Furthermore, for each leading diagnosis not in $\jwsdx$, \textsc{createQuery} has to check if the associated query is entailed. If a query is not entailed, a consistency check has to be performed. Entailments are determined by classification/realization and a subset check of the generated sentences. Common entailments are computed by exploiting the intersection of entailments for each diagnosis contained in $\jwsdx$. Note that the entailments for each leading diagnosis are computed just once and reused in for subsequent calls of \textsc{createQuery}. 

In the function \textsc{minimizeQuery}, the query $\jwsqry$ of the resulting tuple $\tuple{\jwsqry,\jwsdx,\jwsdnx,\jwsdz}$ is iteratively reduced by applying \textsc{QuickXplain} such that sets $\jwsdx$, $\jwsdnx$ and $\jwsdz$ are preserved.  
This is implemented by replacing the consistency checks performed by \textsc{QuickXplain} with checks that ensure that the reduction of the query preserves the partition. In order to check if a partition is preserved, a consistency/entailment check is performed for each element in $\jwsdnx$ and $\jwsdz$. Elements of $\jwsdx$ need not be checked because these elements entail the query and therefore any reduction. In the worst case $n (2k \log (s/k) + 2k)$ consistency checks have to be performed in \textsc{minimizeQuery} where $k$ is the length of the minimized query. Entailments of leading diagnoses are reused. 

Algorithm~\ref{algo:general} invokes the function \textsc{getQuery} to obtain the query from the tuple stored in $T$ and calls \textsc{getAnswer} to query the oracle. 
Depending on the answer, Algorithm~\ref{algo:general} extends either the set $\jwsTe$ or the set $\jwsTne$ and thus excludes diagnoses not compliant with the query answer from the results of \textsc{HS-Tree} in further iterations. Note, the algorithm can be easily adapted to allow the oracle to reject a query if the answer is unknown. In this case the algorithm proceeds with the next best query (w.r.t. the \textsc{getScore} function) until no further queries are available. 

Algorithm~\ref{algo:general} stops if the difference in the probabilities of the top two diagnoses is greater than the acceptance threshold $\sigma$ or if no query can be used to differentiate between the remaining diagnoses (i.e.\ the score of the minimum score query equals to the maximum score of the used measure).
The most probable diagnosis is then returned to the user. If it is impossible to differentiate between a number of highly probable minimal diagnoses, the algorithm returns a set that includes all of them. Moreover, in the first case (termination due to $\sigma$), the algorithm can continue if the user is not satisfied with the returned diagnosis and at least one further query exists. 
%One has to provide the sets $\jwsTe$ and $\jwsTne$ of the last iteration to the input of Algorithm~\ref{algo:general} in order to continue.

Additional performance improvements can be achieved by using greedy strategies in Algorithm~\ref{algo:computeQuery}. The idea is to guide the search such that a leaf node of the left-most branch of a search tree contains a set of diagnoses $\jwsdx$ that might result in a tuple $\tuple{\jwsqry,\jwsdx,\jwsdnx,\jwsdz}$ with a low-score query. This method is based on the property of Equation~\ref{eq:score} that $sc(\jwsqry)=0$ if 
\begin{align*}
\sum_{\jwsmd_i \in \jwsdx}p(\jwsmd_i)=\sum_{\jwsmd_j \in \jwsdnx}p(\jwsmd_j)=0.5\quad \textnormal{ and } \quad p(\jwsdz)=0
\end{align*} 
Consequently, the query selection problem can be presented as a two-way number partitioning problem: given a set of numbers, divide them into two sets such that the difference between the sums of the numbers in each set is as small as possible. The Complete Karmarkar-Karp (CKK) algorithm~\cite{Korf1998}, which is one of the best algorithms developed for the two-way partitioning problem, corresponds to an extension of the Algorithm~\ref{algo:computeQuery} with a set differencing heuristic~\cite{kk1986}. The algorithm stops if the optimal solution to the two-way partitioning problem is found or if there are no further subsets to be investigated. In the latter case the best found solution is returned. 

The main drawback of applying CKK to the query selection process is that none of the pruning techniques can be used. Also even if the algorithm finds an optimal solution to the two-way partitioning problem there just might be no query for a found set of diagnoses $\jwsdx$. 
Moreover, since the algorithm is complete it still has to investigate all subsets of the set of diagnoses in order to find the minimum score query. 
To avoid this exhaustive search we extended CKK with an additional termination criterion: the search stops if a query is found with a score below some predefined threshold $\gamma$. In our evaluation section we demonstrate substantial savings by applying the CKK partitioning algorithm. 

To sum up, the proposed method depends on the efficiency of the classification/realization system and consistency/coherency checks given a particular ontology. The number of calls to a reasoning system can be reduced by decreasing the number of leading diagnoses $n$. However, the more leading diagnoses provide the more data for generating the next best query. Consequently, by varying the number of leading diagnoses it is possible to balance runtime with the number of queries needed to isolate the target diagnosis.\footnote{The source code as well as precompiled binaries can be downloaded from \url{http://rmbd.googlecode.com}. The package also includes a Prot\'{e}g\'{e}-plugin implementing the methods as described.} 

%%%%%%%%%%%%%%%%%%%%%%%%%%%%%%%%%%%%%%%%%%%%%%%%%%%%%%%%%%%%%%%%%%%%%%%
%%%%%%%%%%%%%%%%%%%%%%%%%%%%%%%%%%%%%%%%%%%%%%%%%%%%%%%%%%%%%%%%%%%%%%%
%%%%%%%%%%%%%%%%%%%%%%%%%%%%%%%%%%%%%%%%%%%%%%%%%%%%%%%%%%%%%%%%%%%%%%%
%%%%%%%%%%%%%%%%%%%%%%%%%%%%%%%%%%%%%%%%%%%%%%%%%%%%%%%%%%%%%%%%%%%%%%%
%%%%%%%%%%%%%%%%%%%%%%%%%%%%%%%%%%%%%%%%%%%%%%%%%%%%%%%%%%%%%%%%%%%%%%%
%%%%%%%%%%%%%%%%%%%%%%%%%%%%%%%%%%%%%%%%%%%%%%%%%%%%%%%%%%%%%%%%%%%%%%%
%%%%%%%%%%%%%%%%%%%%%%%%%%%%%%%%%%%%%%%%%%%%%%%%%%%%%%%%%%%%%%%%%%%%%%%
%%%%%%%%%%%%%%%%%%%%%%%%%%%%%%%%%%%%%%%%%%%%%%%%%%%%%%%%%%%%%%%%%%%%%%%
%%%%%%%%%%%%%%%%%%%%%%%%%%%%%%%%%%%%%%%%%%%%%%%%%%%%%%%%%%%%%%%%%%%%%%%
%%%%%%%%%%%%%%%%%%%%%%%%%%%%%%%%%%%%%%%%%%%%%%%%%%%%%%%%%%%%%%%%%%%%%%%
%%%%%%%%%%%%%%%%%%%%%%%%%%%%%%%%%%%%%%%%%%%%%%%%%%%%%%%%%%%%%%%%%%%%%%%

\chapter{Evaluation}\label{chap:jws:eval}
 
\renewcommand{\arraystretch}{1} 
\begin{table*}[tb]
  %\vspace{-5pt}
\small 
\centering
      \begin{tabular}{@{\extracolsep{-3.4pt}} llcccccl} 
 &  Ontology  &     DL &      Axioms &  \#C/\#P/\#I   &  \#CS/min/max  & \#D/min/max    &  Domain \\ \hline
1.  &  Chemical   &  $\mathcal{ALCHF}^{(D)}$  & 144     &  48/20/0 &  6/5/6 & 6/1/3 &  Chemical elements     \\ 
2.  &   Koala    &  $\mathcal{ALCON}^{(D)}$  &    44 &   21/5/6  &   3/4/4   &   10/1/3  &   Training \\ 
3.  &  Sweet-JPL   &  $\mathcal{ALCHOF}^{(D)}$ &    2579   &  1537/121/50   &  1/13/13 &  13/1/1   &  Earthscience  \\ 
4.  &  miniTambis     &  $\mathcal{ALCN}$     &    173     &  183/44/0   &  3/2/6 & 48/3/3   &  Biological science  \\ 
5.  &  University   &  $\mathcal{SOIN}^{(D)}$   &   49   &  30/12/4 &  4/3/5 & 90/3/4   &  Training \\  
6.  &  Economy   &  $\mathcal{ALCH}^{(D)}$   &   1781   &  339/53/482     &  8/3/4 & 864/4/8 &  Mid-level   \\ 
7.  &  Transportation&    $\mathcal{ALCH}^{(D)}$   &     1300   &  445/93/183     &  9/2/6 & 1782/6/9   &  Mid-level  \\ \hline
      \end{tabular}
 \caption[Diagnosis Results for Several of the Real-World Ontologies]{Diagnosis results for several of the real-world ontologies presented in~\cite{Kalyanpur.Just.ISWC07}. \#C/\#P/\#I are the number of concepts, properties and individuals in each ontology. \#CS/min/max are the number of conflict sets, and their minimum and maximum cardinality. The same notation is used for diagnoses \#D/min/max. The ontologies are available upon request.} 
 \label{tab:motivation}
\end{table*}

\renewcommand{\arraystretch}{1.2} 
\begin{table*}[tb] 
\small
\begin{tabular}{@{\extracolsep{-2.5pt}}llccc||ccc} 
   & \multicolumn{1}{c}{}&   \multicolumn{3}{c}{Leading diagnoses}  &  \multicolumn{3}{c}{All diagnoses} \\
Ontology  &  \multicolumn{2}{r}{Consistency}   &  Conflicts    &  Diagnoses &  Consistency    &  Conflicts    &  Diagnoses   \\  \hline
Chemical   &  time   &  0/3/8 &  90/107/128     &  1/97/326   &  0/3/18   & 105/130/179   &  2/126/402  \\  
 &  calls    &  264     &  6   &  7   &  262     &  6   &  7   \\ \cline{2-8}
 &  \multicolumn{3}{c}{runtime: 723}   &   &  \multicolumn{2}{r}{runtime: 892}  &   \\  \hline
Koala &  time   &  0/1/3 &  19/25/30   &  0/11/70 &  0/2/4 &  24/30/37 &  0/12/105   \\  
 &  calls    &  74   &  3   &  10   &  75   &  3   &  11   \\ \cline{2-8}
 &  \multicolumn{3}{c}{runtime: 120}   &   &  \multicolumn{2}{r}{runtime: 148}  &   \\  \hline
Sweet-JPL   &  time   &  1/31/112   &  5185/5185/5185 &  0/586/5332     & 31/106/195     &  5192/5192/5192 &  1/438/5319     \\  
 &  calls    &  187     &  1   &  10   &  195     &  1   &  14   \\ \cline{2-8}
 &  \multicolumn{3}{c}{runtime: 5991}    &   &  \multicolumn{2}{r}{runtime: 6312}      &   \\  \hline
miniTambis     &  time   &  0/5/14   &  84/157/210     &  0/57/504   &  1/5/15   & 88/167/225     &  3/19/537   \\  
 &  calls    &  111     &  3   &  10   &  189     &  3   &  49   \\ \cline{2-8}
 &  \multicolumn{3}{c}{runtime: 586}   &   &  \multicolumn{2}{r}{runtime: 1027}      &   \\  \hline
University   &  time   &  0/2/3 &  31/41/54   &  0/20/157   &  0/2/5 & 37/46/60   &  2/5/200 \\  
 &  calls    &  126     &  4   &  10   &  283     &  4   &  91   \\ \cline{2-8}
 &  \multicolumn{3}{c}{runtime: 205}   &   &  \multicolumn{2}{r}{runtime: 536}  &   \\  \hline
Economy  &  time   &  1/12/26 &  410/460/569   &  0/282/2085     &  1/9/80   & 418/510/681   &  16/25/1929     \\  
 &  calls    &  239     &  6   &  10   &  2064   &  8   &  865     \\ \cline{2-8}
 &  \multicolumn{3}{c}{runtime: 2857}    &   &  \multicolumn{2}{r}{runtime: 25369}    &   \\  \hline
Transportaton   &  time   &  0/11/58 &  237/438/683   &  0/352/3176     &  1/9/130 &  222/429/636   &  16/29/6394     \\  
 &  calls    &  337     &  7   &  10   &  3966   &  9   &  1783   \\ \cline{2-8}
 &  \multicolumn{3}{c}{runtime: 3671}    &   &  \multicolumn{2}{r}{runtime: 65010}    &   \\  \hline
\end{tabular}
 \caption[Minimum, Average and Maximum Time and Calls Required to Compute Leading Most Probable Diagnoses as well as All Diagnoses for Real-World Ontologies]{Min/avg/max time and calls required to compute the nine leading most probable diagnoses as well as all diagnoses for the real-world ontologies. Values are given for each stage, i.e. consistency checking, computation of minimal conflicts and minimal diagnoses, together with the total runtime needed to compute the diagnoses. All time values are 15 trial averages and are given in milliseconds.}
 \label{tab:stats}
\end{table*}

\begin{figure*}[tb]
 \centering
 	\includegraphics[width=0.7\linewidth]{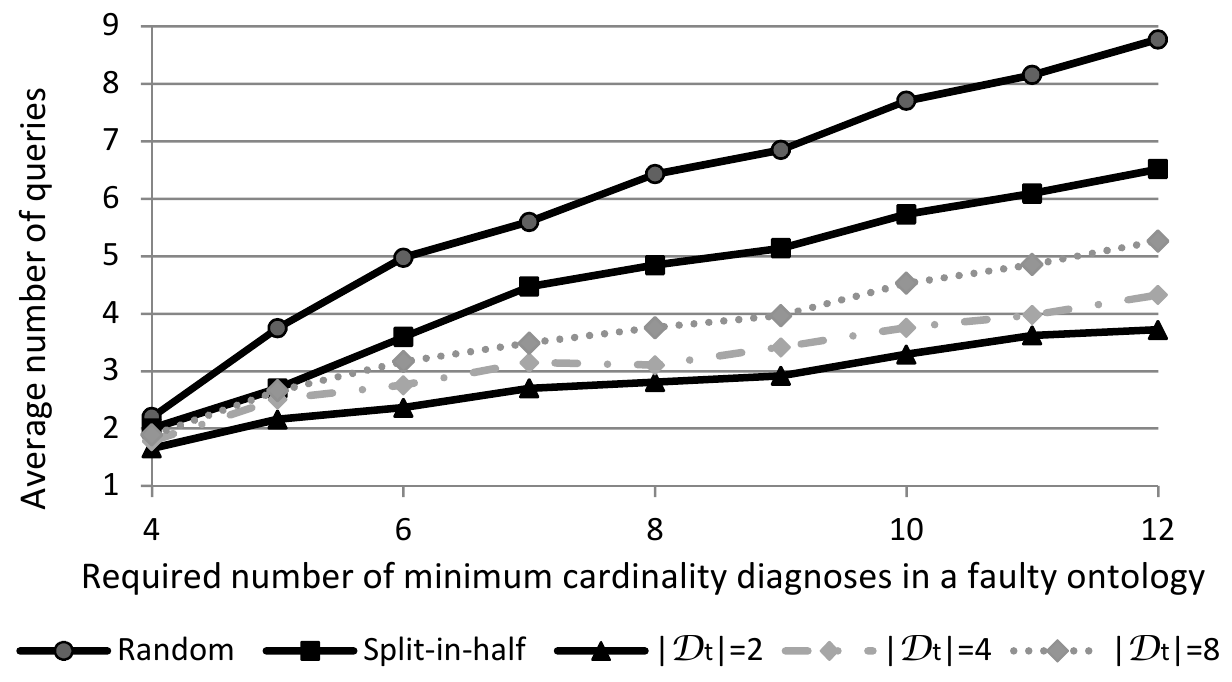}
 \caption[Average Number of Queries Required to Select the Target Diagnosis]{Average number of queries required to select the target diagnosis $\jwsmd_t$ with threshold $\sigma=0.95$. Random and ``split-in-half'' are shown for the cardinality of minimal diagnoses $|\jwsmd_t|=2$.}
 \label{fig:results}
\end{figure*}

We evaluated our approach using the real-world ontologies presented in Table~\ref{tab:motivation} with the aim of demonstrating its applicability real-world settings. In addition, we employed generated examples to perform controlled experiments where the number of minimal diagnoses and their cardinality could be varied to make the identification of the target diagnosis more difficult. Finally, we carried out a set of tests using randomly modified large real-world ontologies to provide some insights on the scalability of the suggested debugging method.

For the first test we created a generator which takes a consistent and coherent ontology, a set of fault patterns together with their probabilities, the minimum number of minimum cardinality diagnoses $m$, and the required cardinality $|\jwsmd_t|$ of these minimum cardinality diagnoses as inputs. We also assumed that the target diagnosis has cardinality $|\jwsmd_t|$. 
The output of the generator is an alteration of the input ontology for which at least the given number of minimum cardinality diagnoses with the required cardinality exist. Furthermore, to introduce inconsistencies (incoherencies), the generator applies fault patterns randomly to the input ontology depending on their probabilities. 

In this experiment we took five fault patterns from a case study reported by Rector et al.~\cite{Rector2004} and assigned fault probabilities according to their observations of typical user errors. Thus we assumed that in cases (a) and (b) (see Section~\ref{sect:discrimination}), where an axiom includes some roles (i.e.\ property assertions), axiom descriptions are faulty with a probability of $0.025$, in cases (c) and (d) $0.01$ and in case (e) $0.001$. In each iteration, the generator randomly selected an axiom to be altered and applied a fault pattern. Following this, another axiom was selected using the concept taxonomy and altered correspondingly to introduce an inconsistency (incoherency). The fault patterns were randomly selected in each step using the probabilities provided above.

For instance, given the description of a randomly selected concept $A$ and the fault pattern ``misuse of negation'', we added the construct $\sqcap \neg X$ to the description of $A$, where $X$ is a new concept name. Next, we randomly selected concepts $B$ and $S$ such that $S\sqsubseteq A$ and $S \sqsubseteq B$ and added $\sqcap X$ to the description of $B$.
During the generation process, we applied the \textsc{HS-Tree} algorithm after each introduction of an incoherency/inconsistency to control two parameters: the minimum number of minimal cardinality diagnoses in the ontology and their cardinality. The generator continues to introduce incoherences/inconsistencies until the specified parameter values are reached. For instance, if the minimum number of minimum cardinality diagnoses is equal to $m=6$ and their cardinality is $|\jwsmd_t|=4$, then the generated ontology will include at least $6$ diagnoses of cardinality $4$ and possibly some additional number of minimal diagnoses of higher cardinalities.

The resulting faulty ontology as well as the fault patterns and their probabilities were inputs for the ontology debugger.  The acceptance threshold $\sigma$ was set to $0.95$ and the number of most probable minimal diagnoses $n$ was set to $9$.
In addition, one of the minimal diagnoses with the required cardinality was randomly selected as the target diagnosis. 
Note, the target ontology is not equal to the original ontology, but rather a corrected version of the altered one in which the faulty axioms were repaired by replacing them with their original (correct) versions according to the target diagnosis.
The tests were performed using the ontologies bike2 to bike9, bcs3, galen and galen2 from Racer's benchmark suite\footnote{Available at \url{http://www.racer-systems.com/products/download/benchmark.phtml}}.

The average results of the evaluation performed on each test ontology (presented in Figure~\ref{fig:results}) show that the entropy-based approach outperforms the ``split-in-half'' heuristic  as well as the random query selection strategy by more than 50\% for the $|\jwsmd_t|=2$ case due to its ability to estimate the probabilities of diagnoses and to stop once the target diagnosis crossed the acceptance threshold. On average the algorithm required $8$ seconds to generate a query. In addition, Figure~\ref{fig:results} shows that the number of queries required increases as the cardinality of the target diagnosis increases, regardless of the method. Despite this, the entropy-based approach remains better than the ``split-in-half'' method for diagnoses with increasing cardinality. The approach did however require more queries to discriminate between high cardinality diagnoses because in such cases more minimal conflicts were generated. Consequently, the debugger should consider more minimal diagnoses in order to identify the target one.

For the next test we selected seven real-world ontologies described in Tables~\ref{tab:motivation} and~\ref{tab:stats}\footnote{All experiments were performed on a PC with Core2 Duo (E8400), 3 Ghz with 8 Gb RAM, running Windows 7 and Java 6.}. Performance of both the entropy-based and ``split-in-half'' selection strategies was evaluated using a variety of different prior fault probabilities to investigate under which conditions the entropy-based method should be preferred.

\begin{figure}[tb]
 \centering
 	\includegraphics[width=0.7\linewidth]{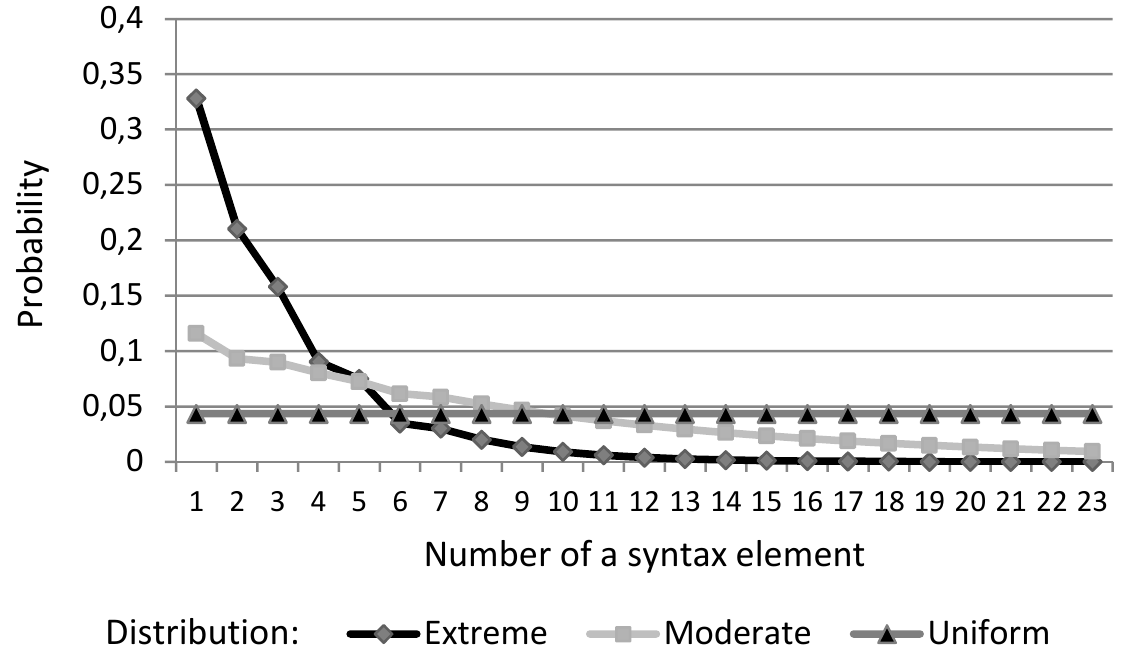}
 \caption[Example of Prior Fault Probabilities of Syntax Elements Sampled from Extreme, Moderate and Uniform Distributions]{Example of prior fault probabilities of syntax elements sampled from extreme, moderate and uniform distributions.}
 \label{fig:priors}
\end{figure}

In our experiments we distinguished between three different distributions of prior fault probabilities: extreme, moderate and uniform (see Figure~\ref{fig:priors} for an example). The \emph{extreme distribution} simulates a situation in which very high failure probabilities are assigned to a small number of syntax elements. That is, the provider of the estimates is quite sure that exactly these elements are causing a fault. For instance, it may be well known that a user has problems formulating restrictions in OWL whereas all other elements, such as subsumption and conjunction, are well understood. In the case of a \emph{moderate distribution} the estimates provide a slight bias towards some syntax elements. This distribution has the same motivation as the extreme one, however, in this case the probability estimator is less sure about the sources of possible errors in axioms. Both extreme and moderate distributions correspond to the exponential distribution with $\lambda=1.75$ and $\lambda=0.5$ respectively. The \emph{uniform distribution} models the situation where no prior fault probabilities are provided and the system assigns equal probabilities to all syntax elements found in a faulty ontology. Of course the prior probabilities of diagnoses may not reflect the actual situation. Therefore, for each of the three distributions we differentiate between good, average and bad cases. In the \emph{good case} the estimates of the prior fault probabilities are correct and the target diagnosis is assigned a high probability. The \emph{average case} corresponds to the situation when the target diagnosis is neither favored nor penalized by the priors. In the \emph{bad case} the prior distribution is unreasonable and disfavors the target diagnosis by assigning it a low probability. 

We executed 30 tests for each of the combinations of the distributions and cases with an acceptance threshold $\sigma=0.85$ and a required number of most probable minimal diagnoses $n=9$. Each iteration started with the generation of a set of prior fault probabilities of syntax elements by sampling from a selected distribution (extreme, moderate or uniform). Given the priors we computed the set of all minimal diagnoses $\jwsmD$ of a given ontology and selected the target one according to the chosen case (good, average or bad). In the good case the prior probabilities favor the target diagnosis and, therefore, it should be selected from the diagnoses with high probability. The set of diagnoses was ordered according to their probabilities and the algorithm iterated through the set starting from the most probable element. 
In the first iteration the most probable minimal diagnosis $\jwsmd_1$ is added to the set $G$. In next iteration $j$ a diagnosis $\jwsmd_j$ was added to the set $G$ if $\sum_{i \leq j}{p(\jwsmd_i)}\leq \frac{1}{3}$ and to the set $A$ if $\sum_{i\leq j}{p(\jwsmd_i)} \leq \frac{2}{3}$. The obtained set $G$ contained all most probable diagnoses which we considered as good. All diagnoses in the set $A\setminus G$ were classified as average and the remaining diagnoses $\jwsmD \setminus A$ as bad. Depending on the selected case we randomly selected one of the diagnoses as the target from the appropriate set. 

The results of the evaluation presented in Table~\ref{tab:results} show that the entropy-based query selection approach clearly outperforms ``split-in-half'' in good and average cases for the three probability distributions. The average time required by the debugger to perform such basic operations as consistency checking, computation of minimal conflicts and diagnoses is presented in Table~\ref{tab:diagstats}. The results indicate that on average at most 17 seconds required to compute up to 9 minimal diagnoses and a query. Moreover, the number of axioms in a query remains reasonable in most of the cases stays bounds, i.e. between 1 and 4 axioms per query. 
 
\renewcommand{\arraystretch}{1} 
\begin{table*}[p] 
\centering
\begin{tabular}{l|c||ccc|ccc|ccc} 
\multicolumn{11}{c}{Entropy-based query selection} \\ \hline 
\multicolumn{1}{c}{Ontology}  & Case & \multicolumn{9}{c}{Distribution} \\ \cline{4-10}
\multicolumn{1}{c}{} & & \multicolumn{3}{c}{Extreme} 	 	 & \multicolumn{3}{c}{Moderate} 	 	 & \multicolumn{3}{c}{Uniform} \\ 	 	 
\multicolumn{1}{c}{}& & min & avg & \multicolumn{1}{c}{max} & min & avg & \multicolumn{1}{c}{max} & min & avg & \multicolumn{1}{c}{max} \\ \hline\hline
 & Good & 1 & \textbf{ 1.63 } & 3 & 1 & \textbf{ 1.7 } & 2 & 1 & \textbf{ 1.83 } & 2 \\ 
Chemical & Avg. & 1 & \textbf{ 1.87 } & 4 & 1 & \textbf{ 1.73 } & 3 & 1 & \textbf{ 1.7 } & 2 \\ 
 & Bad & 2 & 	3.03 	& 4 & 2 & 	3.03 	& 4 & 2 & 	3.17 	& 4 \\ \hline
 	 	 	 	 	 	 	 	 	 	 	 	 	 	
 & Good & 1 & \textbf{ 1.7 } & 3 & 1 & \textbf{ 2.4 } & 4 & 1 & \textbf{ 2.67 } & 3 \\ 
Koala & Avg. & 1 & \textbf{ 1.8 } & 3 & 1 & \textbf{ 2.37 } & 4 & 1 & \textbf{ 2.4 } & 3 \\ 
 & Bad & 1 & 	3.5 	& 6 & 2 & 	4.33 	& 7 & 3 & 	4.13 	& 5 \\ \hline
 	 	 	 	 	 	 	 	 	 	 	 	 	 	
 & Good & 1 & \textbf{ 3.27 } & 7 & 2 & \textbf{ 3.43 } & 7 & 3 & 	\textbf{ 3.87 } 	& 7 \\ 
Sweet-JPL & Avg. & 1 & \textbf{ 3.5 } & 6 & 1 & 	4.03 	& 7 & 3 & 	4.07 	& 6 \\ 
 & Bad & 3 & 	3.93 	& 6 & 2 & 	4.03 	& 6 & 3 & \textbf{ 3.37 } & 4 \\ \hline
 	 	 	 	 	 	 	 	 	 	 	 	 	 	
 & Good & 1 & \textbf{ 2.37 } & 4 & 2 & \textbf{ 2.73 } & 4 & 2 & \textbf{ 2.77 } & 3 \\ 
miniTambis & Avg. & 1 & \textbf{ 2.53 } & 4 & 2 & \textbf{ 4.03 } & 8 & 3 & \textbf{ 4.53 } & 7 \\ 
 & Bad & 3 & 	6.43 	& 11 & 3 & 	7.93 	& 17 & 5 & 	9.03 	& 13 \\ \hline
 	 	 	 	 	 	 	 	 	 	 	 	 	 	
 & Good & 1 & \textbf{ 2.7 } & 4 & 3 & \textbf{ 3.83 } & 7 & 3 & \textbf{ 4.4 } & 8 \\ 
University & Avg. & 1 & \textbf{ 3.4 } & 6 & 3 & 	7.03 	& 12 & 4 & \textbf{ 7.27 } & 10 \\ 
 & Bad & 5 & 	9.13 	& 15 & 5 & 	9.7 	& 14 & 6 & 	10.03 	& 14 \\ \hline
 	 	 	 	 	 	 	 	 	 	 	 	 	 	
 & Good & 1 & \textbf{ 3.2 } & 11 & 3 & \textbf{ 3.1 } & 4 & 3 & \textbf{ 3.93 } & 6 \\ 
Economy & Avg. & 1 & \textbf{ 4.63 } & 14 & 3 & \textbf{ 5.57 } & 12 & 5 & \textbf{ 6.5 } & 8 \\ 
 & Bad & 8 & 	12.3 	& 19 & 6 & 	11.5 	& 21 & 7 & 	11.67 	& 19 \\ \hline
 	 	 	 	 	 	 	 	 	 	 	 	 	 	
 & Good & 1 & \textbf{ 5.63 } & 14 & 1 & \textbf{ 6.97 } & 12 & 3 & \textbf{ 9.5 } & 14 \\ 
Transportation & Avg. & 1 & \textbf{ 6.9 } & 16 & 1 & \textbf{ 7.73 } & 12 & 3 & \textbf{ 8.73 } & 14 \\ 
 & Bad & 3 & \textbf{ 12.4 } & 18 & 8 & \textbf{ 12.8 } & 20 & 3 & \textbf{ 12.1 } & 18 \\ \hline

    \multicolumn{11}{c}{ } \\
 \multicolumn{11}{c}{ ``Split-in-half'' query selection} \\\hline\hline
 & Good & 2 & 	2.63 	& 3 & 2 & 	2.7 	& 3 & 2 & 	2.53 	& 3 \\ 
Chemical & Avg. & 2 & 	2.63 	& 3 & 2 & 	2.67 	& 3 & 2 & 	2.77 	& 3 \\ 
 & Bad & 2 & \textbf{ 2.63 } & 3 & 2 & \textbf{ 2.6 } & 3 & 2 & \textbf{ 2.4 } & 3 \\ \hline
 	 	 	 	 	 	 	 	 	 	 	 	 	 	
 & Good & 3 & 	3.3 	& 4 & 3 & 	3.3 	& 4 & 3 & 	3.47 	& 4 \\ 
Koala & Avg. & 3 & 	3.33 	& 4 & 3 & 	3.2 	& 4 & 3 & 	3.23 	& 4 \\ 
 & Bad & 3 & \textbf{ 3.43 } & 4 & 3 & \textbf{ 3.4 } & 4 & 3 & \textbf{ 3.5 } & 4 \\ \hline
 	 	 	 	 	 	 	 	 	 	 	 	 	 	
 & Good & 3 & 	3.83 	& 4 & 3 & 	3.8 	& 4 & 4 &  4  & 4 \\ 
Sweet-JPL & Avg. & 3 & 	3.57 	& 4 & 3 & \textbf{ 3.8 } & 4 & 3 & \textbf{ 3.47 } & 4 \\ 
 & Bad & 3 & \textbf{ 3.87 } & 4 & 3 & \textbf{ 3.8 } & 4 & 3 & 	3.8 	& 4 \\ \hline
 	 	 	 	 	 	 	 	 	 	 	 	 	 	
 & Good & 4 & 	5.33 	& 6 & 4 & 	5 	& 6 & 4 & 	4 	& 4 \\ 
miniTambis & Avg. & 4 & 	5.1 	& 6 & 4 & 	4.93 	& 7 & 5 & 	5.43 	& 7 \\ 
 & Bad & 5 & \textbf{ 5.93 } & 8 & 4 & \textbf{ 5.8 } & 7 & 5 & \textbf{ 6.3 } & 7 \\ \hline
 	 	 	 	 	 	 	 	 	 	 	 	 	 	
 & Good & 4 & 	5.93 	& 8 & 4 & 	6 	& 8 & 4 & 	5.43 	& 8 \\ 
University & Avg. & 4 & 	5.87 	& 7 & 5 & \textbf{ 6.73 } & 9 & 6 & 	7.37 	& 8 \\ 
 & Bad & 5 & \textbf{ 6.97 } & 9 & 5 & \textbf{ 7.2 } & 9 & 5 & \textbf{ 7 } & 8 \\ \hline
 	 	 	 	 	 	 	 	 	 	 	 	 	 	
 & Good & 6 & 	7.87 	& 11 & 6 & 	7.4 	& 10 & 6 & 	7.5 	& 10 \\ 
Economy & Avg. & 6 & 	8 	& 12 & 5 & 	7.63 	& 12 & 6 & 	8.73 	& 13 \\ 
 & Bad & 9 & \textbf{ 11.50 } & 14 & 6 & \textbf{ 11.1 } & 14 & 8 & \textbf{ 11.3 } & 15 \\ \hline
 	 	 	 	 	 	 	 	 	 	 	 	 	 	
 & Good & 5 & 	8.03 	& 13 & 5 & 	7.3 	& 11 & 6 & 	11.43 	& 18 \\ 
Transportation & Avg. & 3 & 	9 	& 16 & 5 & 	9.4 	& 13 & 5 & 	11.43 	& 18 \\ 
 & Bad & 10 & 	12.67 	& 19 & 7 & 	13 	& 19 & 6 & 	13.8 	& 20 \\ \hline

\end{tabular}
\caption[Minimum, Average and Maximum Number of Queries Required by the Entropy-Based and ``Split-In-Half'' Query Selection Methods]{Minimum, average and maximum number of queries required by the entropy-based and ``split-in-half'' query selection methods to identify the target diagnosis in real-world ontologies. Ontologies are ordered by the number of diagnoses.}
\label{tab:results}
\end{table*}

\begin{table*}
\small
	\centering
		\begin{tabular}{l||ccc|ccc|ccc}
  \multicolumn{1}{c}{Ontology}	&	\multicolumn{3}{c}{Good} &	\multicolumn{3}{c}{Average}&\multicolumn{3}{c}{Bad} \\ \cline{2-10}
  \multicolumn{1}{c}{}		&	DT	&	QT	&	QL	&	DT	&	QT	&	QL	&	DT	&	QT	&	QL	\\	\hline
Chemical	&	459.33	&	117.67	&	3	&	461.33	&	121	&	3.34	&	256.67	&	75.67	&	2.19	\\	
Koala 	&	88.33	&	1308.33	&	3.47	&	92	&	1568.67	&	3.90	&	56.33	&	869.33	&	2.36	\\	
Sweet-JPL	&	2387.33	&	691.67	&	1.48	&	2272	&	926	&	1.61	&	2103	&	1240.33	&	1.57	\\	
miniTabmis	&	481.33	&	2764.33	&	3.27	&	398.33	&	2892	&	2.53	&	238.67	&	3223	&	1.76	\\	
University	&	189.33	&	822.67	&	3.91	&	145	&	903.33	&	2.82	&	113	&	872	&	2.11	\\	
Economy	&	2953.33	&	6927	&	3.06	&	3239	&	8789	&	3.80	&	3083	&	8424.67	&	1.58	\\	
Transportation	&	6577.33	&	9426.33	&	2.37	&	7080.67	&	10135.33	&	2.29	&	7186.67	&	9599.67	&	1.64	\\	\hline
 
		\end{tabular}
	\caption[Average Time Required to Compute a Set of Leading Diagnoses and a Query in Each Iteration]{Average time required to compute at most nine minimal diagnoses (DT) and a query (QT) in each iteration, as well as the average number of axioms in a query after minimization (QL). The averages are shown for extreme, moderate and uniform distributions using the entropy-based query selection method. Time is measured in milliseconds.}
	\label{tab:diagstats}
\end{table*}

\begin{figure}[h!]
 \centering
 	\includegraphics[width=0.77\linewidth]{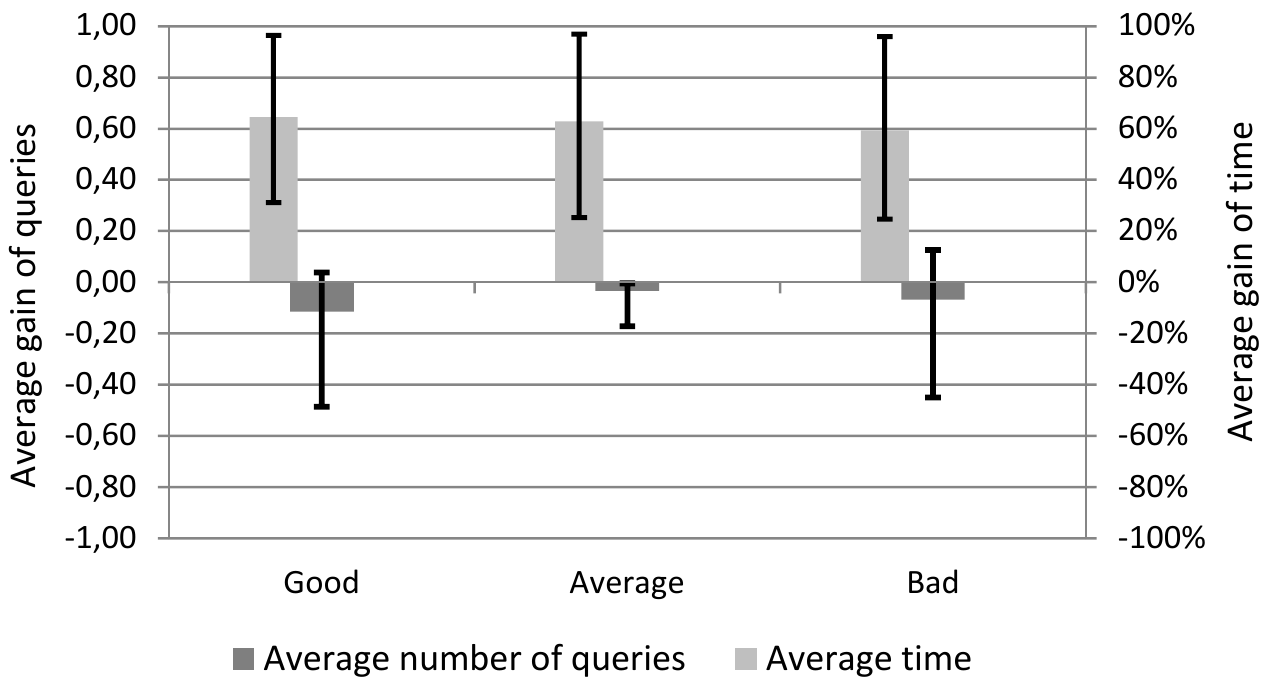}
 \caption[Average Time/Query Gain Resulting from the Application of the Extended CKK Partitioning Algorithm]{Average time/query gain resulting from the application of the extended CKK partitioning algorithm. The whiskers indicate the maximum and minimum possible average gain of queries/time using extended CKK.}
 \label{fig:greedy}
\end{figure}

In the uniform case better results were observed since the diagnoses have different cardinality and structure, i.e. they include different syntax elements. Consequently, even if equal probabilities for all syntax elements (uniform distribution) are given, the probabilities of diagnoses are different. Axioms with a greater number of syntax elements receive a higher fault probability. Also, diagnoses with a smaller cardinality in many cases receive a higher probability. This information provides enough bias to favor the entropy-based method.

In the bad case, where the target diagnosis received a low probability and no information regarding the prior fault probabilities was given, we observed that the performance of the entropy-method improved as more queries were posed. In particular, in the University ontology the performance is essentially similar (7.27 vs.\ 7.37) whereas in the Economy and Transportation ontology the entropy-based method can save and average of two queries. 

``Split-in-half'' appears to be particularly inefficient in all good, average and bad cases when applied to ontologies with a large number of minimal diagnoses, such as Economy and Transportation. The main problem is that no stop criteria can be used with the greedy method as it is unable to provide any ordering on the set of diagnoses. Instead, the method continues until no further queries can be generated, i.e.\ only one minimal diagnosis exists or there are no discriminating queries. Conversely, the entropy-based method is able to improve its probability estimates using Bayes-updates as more queries are answered and to exploit the differences in the probabilities in order to decide when to stop. 

The most significant gains are achieved for ontologies with many minimal diagnoses and for the average and good cases, e.g. the target diagnosis is within the first or second third of the minimal diagnoses ranked by their prior probability. In these cases the entropy-based method can save up to 60\% of the queries. 

Therefore, we can conclude that even rough estimates of the prior fault probabilities are sufficient, provided that the target diagnosis is not significantly penalized. Even if no fault probabilities are available and there are many minimal diagnoses, the entropy-based method is advantageous.  The differences between probabilities of individual syntax elements appears not to influence the results of the query selection process and affect only the number of outliers, i.e.\ cases in which the diagnosis approach required either few or many queries compared to the average. 

Another interesting observation is that often both methods eliminated more than $n$ diagnoses in one iteration. For instance, in the case of the Transportation ontology both methods were able to remove hundreds of minimal diagnoses with a small number of queries. This behavior appears to stem from relations between the diagnoses. That is, the addition of a query to either $\jwsTe$ or $\jwsTne$ allows the method to remove not only the diagnoses in sets $\jwsdx$ or $\jwsdnx$, but also some unobserved diagnoses that were not in any of the sets of $n$ leading diagnoses computed by \textsc{HS-Tree}. Given the sets $\jwsTe$ and $\jwsTne$, \textsc{HS-Tree} automatically invalidates all diagnoses which do not fulfill the requirements (see Definition~\ref{def:diag}). 

\begin{figure}
 \centering
 	\includegraphics[width=.78\linewidth]{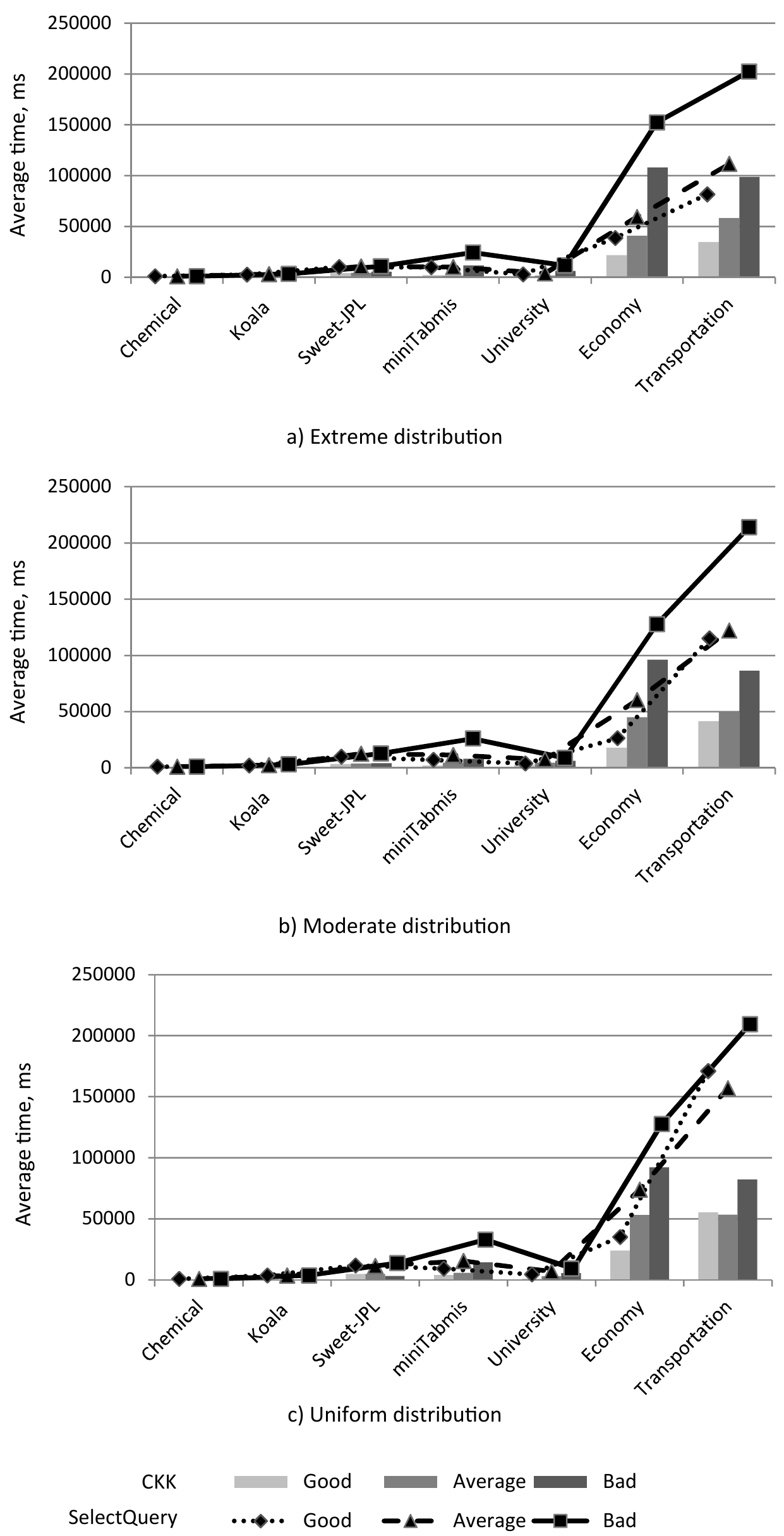}
 \caption[Average Time Required to Identify the Target Diagnosis Using CKK and Brute Force Query Selection]{Average time required to identify the target diagnosis using CKK and brute force query selection algorithms.}
 \label{fig:averages}
\end{figure}

\begin{table*}[tb]%[bt]
	\centering
		\begin{tabular} {l|cc} 
  \multicolumn{1}{c}{Ontology}	&	Cton	&	Opengalen-no-propchains	\\ \hline	
Axioms & 33203 & 9664  \\
DL	&	$\mathcal{SHF}$	&	$\mathcal{ALEHIF}^{(D)}$ \\	

\#CS/min/max  	&	6/3/7	&	9/5/8	\\	
\#D/min/max	&	15/1/5	&	110/2/6	\\	
Consistency	&	5/209/1078	&	1/98/471	\\	
QuickXplain	&	17565/20312/38594	&	7634/10175/12622	\\	
Diagnosis	&	1/5285/38594	&	10/1043/19543	\\	
Overall runtime	&	146186	&	119973	\\	\hline
   
		\end{tabular}
	\caption[Statistics for the Real-World Ontologies Used in the Stress-Tests]{Statistics for the real-world ontologies used in the stress-tests measured for a single random alteration. \#CS/min/max are the number of minimal conflict sets, and their minimum and maximum cardinality. The same notation is used for diagnoses \#D/min/max.  The minimum/average/maximum time required to make a consistency check (Consistency), compute a minimal conflict set (QuickXplain) and a minimal diagnosis are measured in milliseconds. Overall runtime indicates the time required to compute all minimal diagnoses in milliseconds.}
	\label{tab:bigstats}
\end{table*}

\begin{table*}[tb]
	\centering
		\begin{tabular} {l|ccccc} 
	 \multicolumn{6}{c}{Good} \\ \hline
					
  \multicolumn{1}{c|}{Ontology}	&	\#Query	&	Overall	&	QT	&	DT	&	QL \\ \hline
Cton	&	3	&	176828	&	6918	&	52237	&	4 \\
Opengalen-no-propchains	&	8	&	154145	&	2349	&	22905	&	4 \\\hline
		\multicolumn{6}{c}{Average}		\\\hline						
Cton	&	4	&	177383	&	6583	&	52586	&	3 \\
Opengalen-no-propchains	&	7	&	151048	&	3752	&	21344	&	4 \\\hline
		\multicolumn{6}{c}{Bad} 								\\\hline
Cton	&	5	&	190407	&	5742	&	35608	&	1 \\
Opengalen-no-propchains	&	14	&	177728	&	1991	&	11319	&	3 \\\hline

\end{tabular}
	\caption[Average Values Measured for Extreme, Moderate and Uniform Distributions in Each of the Good, Average and Bad Cases]{Average values measured for extreme, moderate and uniform distributions in each of the good, average and bad cases. \#Query is the number of queries required to find the target diagnosis. Overall runtime as well as the time required to compute a query (QT) and at least nine minimal diagnoses (DT) are given in milliseconds. Query length (QL) shows the average number of axioms in a query.}
	\label{tab:bigres}
\end{table*}

The extended CKK method presented in Chapter~\ref{chap:jws:impl} was evaluated in the same settings as the complete Algorithm~\ref{algo:computeQuery} with acceptance threshold $\gamma=0.1$.  The obtained results presented in Figure~\ref{fig:greedy} show that the extended CKK method decreases the length of a debugging session by at least 60\% while requiring on average $0.1$ queries more than Algorithm~\ref{algo:computeQuery}. In some cases (mostly for the uniform distribution) the debugger using CKK search required even fewer queries than Algorithm~\ref{algo:computeQuery} because of the inherent uncertainty of the domain. The plot of the average time required by Algorithm~\ref{algo:computeQuery} and CKK to identify the target diagnosis presented in Figure~\ref{fig:averages} shows that the application of the latter can reduce runtime significantly.

In the last experiment we tried to simulate an expert developing large real-world ontologies\footnote{The ontologies taken from TONES repository \url{http://owl.cs.manchester.ac.uk/repository}} as described in Table \ref{tab:bigstats}. Often in such settings an expert makes small changes to the ontology and then runs the reasoner to verify that the changes are valid, i.e. the ontology is consistent and its entailments are correct. To simulate this scenario we used the generator described in the first experiment to introduce 1 to 3 random changes that would make the ontology incoherent. Then, for each modified ontology, we performed 15 tests using the fault distributions as in the second test. The results obtained by the entropy-based query selection method using CKK for query computation are presented in Table~\ref{tab:bigres}. These results show that the method can be used for analysis of large ontologies with over 33000 axioms while requiring a user to wait for only a minute to compute the next query.

%%%%%%%%%%%%%%%%%%%%%%%%%%%%%%%%%%%%%%%%%%%%%%%%%%%%%%%%%%%%%%%%%%%%%%%
%%%%%%%%%%%%%%%%%%%%%%%%%%%%%%%%%%%%%%%%%%%%%%%%%%%%%%%%%%%%%%%%%%%%%%%
%%%%%%%%%%%%%%%%%%%%%%%%%%%%%%%%%%%%%%%%%%%%%%%%%%%%%%%%%%%%%%%%%%%%%%%
%%%%%%%%%%%%%%%%%%%%%%%%%%%%%%%%%%%%%%%%%%%%%%%%%%%%%%%%%%%%%%%%%%%%%%%
%%%%%%%%%%%%%%%%%%%%%%%%%%%%%%%%%%%%%%%%%%%%%%%%%%%%%%%%%%%%%%%%%%%%%%%
%%%%%%%%%%%%%%%%%%%%%%%%%%%%%%%%%%%%%%%%%%%%%%%%%%%%%%%%%%%%%%%%%%%%%%%
%%%%%%%%%%%%%%%%%%%%%%%%%%%%%%%%%%%%%%%%%%%%%%%%%%%%%%%%%%%%%%%%%%%%%%%
%%%%%%%%%%%%%%%%%%%%%%%%%%%%%%%%%%%%%%%%%%%%%%%%%%%%%%%%%%%%%%%%%%%%%%%
%%%%%%%%%%%%%%%%%%%%%%%%%%%%%%%%%%%%%%%%%%%%%%%%%%%%%%%%%%%%%%%%%%%%%%%
%%%%%%%%%%%%%%%%%%%%%%%%%%%%%%%%%%%%%%%%%%%%%%%%%%%%%%%%%%%%%%%%%%%%%%%
%%%%%%%%%%%%%%%%%%%%%%%%%%%%%%%%%%%%%%%%%%%%%%%%%%%%%%%%%%%%%%%%%%%%%%%

\chapter{Related Work}\label{chap:jws:related}
Despite the range of ontology diagnosis methods available (see ~\cite{Schlobach2007,Kalyanpur.Just.ISWC07,friedrich2005gdm}), to the best of our knowledge no interactive ontology debugging methods, such as our ``split-in-half'' or entropy-based methods, have been proposed so far. The idea of ranking of diagnoses and proposing a target diagnosis is presented in~\cite{Kalyanpur2006}. This method uses a number of measures such as: (a) the frequency with which an axiom appears in conflict sets, (b) impact on an ontology in terms of its ``lost'' entailments when an axiom is modified or removed, (c) ranking of test cases, (d) provenance information about axioms, and (e) syntactic relevance. For each axiom in a conflict set, these measures are evaluated and combined to produce a rank value. These ranks are then used by a modified \textsc{HS-Tree} algorithm to identify diagnoses with a minimal rank. 
However, the method fails when a target diagnosis cannot be determined reliably with the given a-priori knowledge. In our work required information is acquired until the target diagnosis can be identified with confidence.
In general, the work of~\cite{Kalyanpur2006} can be combined with the ideas presented in our work as axiom ranks can be taken into account together with other observations for calculating the prior probabilities of the diagnoses.

The idea of selecting the next best query based on the expected entropy was exploited in the generation of decisions trees in~\cite{Quinlan1986} and further refined for selecting measurements in the model-based diagnosis of circuits in~\cite{dekleer1987}. We extend these methods to  query selection in the domain of ontology debugging.  

In the area of debugging logic programs, Shapiro~\cite{Shapiro83} developed debugging methods based on query answering. Roughly speaking, Shapiro's method aims to detect one fault at a time by querying an oracle about the intended behavior of a Prolog program at hand. In our terminology, for each answer that must not be entailed this diagnosis approach generates one conflict at a time by exploiting the proof tree of a Prolog program. The method then identifies a query that splits the conflict in half.  Our approach can deal with multiple diagnoses and conflicts simultaneously which can be exploited by query generation strategies such as ``split-in-half'' and entropy-based methods. Whereas the ``split-in-half'' strategy splits the set of diagnoses in half, Shapiros's method focuses on one conflict. Furthermore, the exploitation of failure probabilities is not considered in~\cite{Shapiro83}. However, Shapiro's method includes the learning of new clauses in order to cover not entailed answers. Interleaving discrimination of diagnoses and learning of descriptions is currently not considered in our approach because of their additional computational costs.

From a general point of view Shapiro's method can be seen as a prominent example of inductive logic programming (ILP) including systems such as~\cite{MuggletonBu88,Muggleton1995}. In particular, \cite{Muggleton1995} proposes inverse entailments combined with general to specific search through a refinement graph with the goal of generating a theory (hypothesis) which covers the examples and fulfills additional properties. 
%For example~\cite{MuggletonST09} exploits the concept of relative least generalizations. 
Compared to ILP, the focus of our work lies on the theory revision. However, our knowledge representation languages are variants of description logics and not logic programs. 
%
%Therefore, we cannot exploit the syntactical properties of Horn clauses including functional symbols, etc. 
%
Moreover, our method aims to discover axioms which must be changed while minimizing user interaction. Preferences  of theory changes are expressed by probabilities which are updated through Bayes' rule. Other preferences based on plausible extensions of the theory were not considered, again because of their computational costs. 

Although model-based diagnosis has also been applied to logic programs~\cite{ConsoleFD93}, constraint knowledge bases~\cite{Felfernig2004213} and hardware descriptions~\cite{Friedrich1999}, none of these approaches propose a query generation method to discriminate between diagnoses.

%%%%%%%%%%%%%%%%%%%%%%%%%%%%%%%%%%%%%%%%%%%%%%%%%%%%%%%%%%%%%%%%%%%%%%%
%%%%%%%%%%%%%%%%%%%%%%%%%%%%%%%%%%%%%%%%%%%%%%%%%%%%%%%%%%%%%%%%%%%%%%%
%%%%%%%%%%%%%%%%%%%%%%%%%%%%%%%%%%%%%%%%%%%%%%%%%%%%%%%%%%%%%%%%%%%%%%%
%%%%%%%%%%%%%%%%%%%%%%%%%%%%%%%%%%%%%%%%%%%%%%%%%%%%%%%%%%%%%%%%%%%%%%%
%%%%%%%%%%%%%%%%%%%%%%%%%%%%%%%%%%%%%%%%%%%%%%%%%%%%%%%%%%%%%%%%%%%%%%%
%%%%%%%%%%%%%%%%%%%%%%%%%%%%%%%%%%%%%%%%%%%%%%%%%%%%%%%%%%%%%%%%%%%%%%%
%%%%%%%%%%%%%%%%%%%%%%%%%%%%%%%%%%%%%%%%%%%%%%%%%%%%%%%%%%%%%%%%%%%%%%%
%%%%%%%%%%%%%%%%%%%%%%%%%%%%%%%%%%%%%%%%%%%%%%%%%%%%%%%%%%%%%%%%%%%%%%%
%%%%%%%%%%%%%%%%%%%%%%%%%%%%%%%%%%%%%%%%%%%%%%%%%%%%%%%%%%%%%%%%%%%%%%%
%%%%%%%%%%%%%%%%%%%%%%%%%%%%%%%%%%%%%%%%%%%%%%%%%%%%%%%%%%%%%%%%%%%%%%%
%%%%%%%%%%%%%%%%%%%%%%%%%%%%%%%%%%%%%%%%%%%%%%%%%%%%%%%%%%%%%%%%%%%%%%%

\chapter{Summary and Conclusions}
\label{chap:jws:conclusions}

In this part we presented an approach to the interactive debugging of ontologies. This approach is applicable to any knowledge representation language with monotonic semantics.
We showed that the axioms generated by classification and realization reasoning services can be exploited to generate queries which differentiate between diagnoses. For selecting the best next query we proposed two strategies: The ``split-in-half'' strategy prefers queries which allow eliminating a half of the leading diagnoses. The entropy-based strategy employs information theoretic concepts to exploit knowledge about the likelihood of axioms to be faulty. %needing to be changed because the ontology at hand is faulty. 
Based on the probability of an axiom containing an error we predict the (expected) information gain produced by a query result, enabling us to select the best subsequent query according to a one-step-lookahead entropy-based scoring function. We described the implementation of an interactive debugging algorithm and compared the entropy-based method with the ``split-in-half'' strategy. Our experiments showed a significant reduction in the number of queries required to identify the target diagnosis when the entropy-based method is applied. Depending on the quality of the given prior fault probabilities the required number of queries could be reduced by up to 60\%. 

In order to evaluate the robustness of the entropy-based method we experimented with different prior fault probability distributions as well as different qualities of the prior probabilities. Furthermore, we investigated cases where knowledge about failure probabilities is missing or inaccurate. In case such knowledge is unavailable, the entropy-based methods ranks the diagnoses based on the number of syntax elements contained in an axiom and the number of axioms in a diagnosis. Given that this is a reasonable guess (i.e.\ the target diagnosis is not at the lower end of the diagnoses ranked by their prior probabilities), the entropy-based method outperformed ``split-in-half''. Moreover, even if the initial guess is not reasonable, the entropy-based method improves the accuracy of the probabilities as more questions are asked.
Furthermore, the applicability of the approach to real-world ontologies containing thousands of axioms was demonstrated by an extensive set of evaluations which are publicly available.
%%%%%%%%%%%%%%%
\part{Minimizing User Interaction in Ontology Debugging}
\label{part:RIO}
A reinforcement learning query selection strategy (RIO) that makes the presented debugging system robust against the usage of low-quality fault information is presented and thoroughly analyzed in this part which is based on the publications \cite{Rodler2013, Rodler2012OM, Rodler2011, Shchekotykhin2011} published in \emph{Web Reasoning and Rule Systems (RR-2013)}, in the \emph{Proceedings of the 7th International Workshop on Ontology Matching (OM-2012)}, in the \emph{Proceedings of the Joint Workshop on Knowledge Evolution and Ontology Dynamics 2011 (EvoDyn2011)} and in \emph{DX 2011 - 22nd International Workshop on Principles of Diagnosis}, respectively.
\chapter{Introduction to the Problem} 
%Support of ontology development and maintenance is an important requirement for the extensive use of Semantic Web technologies. 

%%%%%%%%%%%%%%%%%%%%
%Approach allows to minimize user interaction in interactive ontology debugging on average.
%%%%%%%%%%%%%%%%%%%%%%

The foundation for widespread adoption of Semantic Web technologies is a broad community of ontology developers which is not restricted to experienced knowledge engineers. Instead, domain experts from diverse fields should be able to create ontologies incorporating their knowledge as autonomously as possible. The resulting ontologies are required to fulfill some minimal quality criteria, usually consistency, coherency and no undesired entailments, in order to grant successful deployment. However, the correct formulation of logical descriptions in ontologies is an error-prone task which accounts for a need for assistance in ontology development in terms of ontology debugging tools. Usually, such tools~\cite{Schlobach2007,Kalyanpur.Just.ISWC07,friedrich2005gdm,Horridge2008} use model-based diagnosis~\cite{Reiter87} to identify sets of faulty axioms, called diagnoses, that need to be modified or deleted in order to meet the imposed quality requirements. The major challenge inherent in the debugging task is often a substantial number of alternative diagnoses. 

In \cite{Shchekotykhin2012} this issue is tackled by letting the user take action during the debugging session by answering queries about entailments and non-entailments of the desired ontology. These answers pose constraints to the validity of diagnoses and thus help to sort out incompliant diagnoses step-by-step. In addition, a Bayesian approach is used to continuously readjust the fault probabilities by means of the additional information given by the user. The user effort in this interactive debugging procedure is strongly affected by the quality of the initially provided meta information, i.e.~prior knowledge about fault probabilities of a user w.r.t.\ particular logical operators. To get this under control, the selection of queries shown to the user can be varied correspondingly. To this end, two essential paradigms for choosing the next ``best'' query have been proposed, split-in-half and entropy-based.
%
%This problem has been addressed in \cite{jws12} by proposing an entropy-based active learning debugging method (ENT) which exploits additional information in terms of queries to a user about the intended ontology. Thereby, the selection of queries is guided by the specification of some meta information, i.e. prior knowledge about fault probabilities of a user w.r.t.\ particular logical operators. 

In order to opt for the optimal strategy, however, the quality of the meta information, i.e.~good or bad (which means high or low probability of the correct solution), must be known in advance. This would, however, implicate the pre-knowledge of the initially unknown solution. Entropy-based methods can make optimal profit from exploiting properly adjusted initial fault probabilities (high potential), whereas they can completely fail in the case of weak prior information (high risk). The split-in-half technique, on the other hand, manifests constant behavior independently of the probabilities given (no risk), but lacks the ability to leverage appropriate fault information (no potential). 
%So, entropy-based strategies have high potential at high risk whereas split-in-half is risk-free at the cost of having no potential. 
This matter of fact is witnessed by the evaluation we conducted, which shows that an unsuitable combination of meta information and query selection strategy can result in a substantial increase of more than $2000\%$ w.r.t.\ number of queries to a user. So, there is a need to either (1) guarantee a sufficiently suited choice of prior fault information, or (2) to manage the ``risk'' of unsuitable method selection. 
The task of (1) might not be a severe problem in a debugging scenario involving a faulty ontology developed by a single expert, since the meta information might be extracted from the logs of previous sessions, if available, or specified by the expert based on their experience w.r.t.\ own faults. However, realization of task (1) is a major issue in scenarios involving automatized systems producing (parts of) ontologies, e.g. ontology alignment and ontology learning, or numerous users collaborating in modeling an ontology, where the choice of reasonable meta information is rather unclear. Therefore, we focus on accomplishing task (2).

The contribution of this part is a new \emph{RI}sk \emph{O}ptimization reinforcement learning method (RIO), which allows to minimize user interaction throughout a debugging session on average compared to existing strategies, for any quality of meta information (high potential at low risk). By virtue of its learning capability, our approach is optimally suited for debugging ontologies where only vague or no meta information is available.
A learning parameter is constantly adapted based on the information gathered so far. On the one hand, our method takes advantage of the given meta information as long as good performance is achieved. On the other hand, it gradually gets more independent of meta information if suboptimal behavior is measured.

Experiments on two datasets of faulty real-world ontologies show the feasibility, efficiency and scalability of RIO. The evaluation will indicate that, on average, RIO is the best choice of strategy for both good and bad meta information with savings as to user interaction of up to 80\%. 

The problem specification, basic concepts and a motivating example are provided in Chapter~\ref{chap:riobasics}. Chapter~\ref{chap:riotheory} explains the suggested approach and gives implementation details. Evaluation results are described in Chapter~\ref{chap:rioeval}. Related work is discussed in Chapter~\ref{chap:rioRelatedWork}. Chapter~\ref{chap:rioconclusion} concludes.

\chapter{Motivation and Basic Concepts} 
\label{chap:riobasics}
First we provide an informal introduction to ontology debugging, particularly addressing readers unfamiliar with the topic. Later we introduce precise formalizations. We assume the reader to be familiar with description logics~\cite{Baader2007}.
 
Ontology debugging deals with the following problem: Given is an ontology $\riomo$ which does not meet postulated requirements $\rioRQ$, e.g. $\rioRQ=\{\text{coherency},\text{consistency}\}$. $\riomo$ is a set of axioms formulated in some monotonic knowledge representation language, e.g. OWL DL.
%, which does not meet postulated requirements $\rioRQ$, e.g. $\rioRQ=\{\text{coherency},\text{consistency}\}$. 
The task is to find a subset of axioms in $\riomo$, called diagnosis, that needs to be altered or eliminated from the ontology in order to meet the given requirements. 
%To this end, our 
The presented approach to ontology debugging does not rely upon a specific knowledge representation formalism, it solely presumes that it is logic-based and monotonic. Additionally, the existence of sound and complete procedures for deciding logical consistency and for calculating logical entailments is assumed. These procedures are used as a black box. For OWL DL, e.g., both functionalities are provided by a standard DL-reasoner.

A diagnosis is a hypothesis about the state of each axiom in $\riomo$ of being either correct or faulty. Generally, there are many diagnoses for one and the same faulty ontology $\riomo$. The problem is then to figure out the single diagnosis, called target diagnosis $\riodt$, 
that complies with the knowledge to be modeled by the intended ontology.
%facts intended to be modeled by the ontology. 
In interactive ontology debugging we assume a user, e.g. the author of the faulty ontology or a domain expert, interacting with an ontology debugging system by answering queries about entailments of the desired ontology, called the target ontology $\rioot$. The target ontology can be understood as $\riomo$ minus the axioms of $\riodt$ plus a set of axioms needed to preserve the desired entailments, called positive test cases.
%i.e. the positive test cases that have been specified.
	%additional axioms $EX_{\riodt}$ which can be added in order to regain desired entailments which might have been eliminated together with axioms in $\riodt$. 
Note that the user is not expected to know $\rioot$ explicitly (in which case there would be no need to consult an ontology debugger), but implicitly in that they are able to answer queries about $\rioot$. 
%
%Roughly speaking,
 
A query is a set of axioms and the user is asked whether the conjunction of these axioms is entailed by $\rioot$. Every positively (negatively) answered query constitutes a positive (negative) test case fulfilled by $\rioot$. The set of positive (entailed) and negative (non-entailed) test cases is denoted by $\rioTp$ and $\rioTn$, respectively. So, $\rioTp$ and $\rioTn$ are sets of sets of axioms, which can be, but do not need to be, initially empty. Test cases can be seen as constraints $\rioot$ must satisfy and are therefore used to gradually reduce the search space for valid diagnoses. Roughly, the overall procedure consists of (1)~computing a predefined number of diagnoses, (2)~gathering additional information by querying the user, (3)~incorporating this information to prune the search space for diagnoses, and so forth, until a stopping criterion is fulfilled, e.g. one diagnosis $\riodt$
%in the search space 
has overwhelming probability. 
%We call this diagnosis the target diagnosis and denote it by $\riodt$.

The general debugging setting we consider also envisions the opportunity for the user to specify some background knowledge $\riomb$, i.e.~a set of axioms that are known to be correct.
%, which is 
$\riomb$ is then incorporated in the calculations throughout the ontology debugging procedure, but no axiom in $\riomb$ may take part in a diagnosis. 
For example, in case the user knows that a subset of axioms in $\riomo$ is definitely sound, all axioms in this subset are added to $\riomb$ before initiating the debugging session. The advantage of this over simply not considering the axioms in $\riomb$ at all is, that the semantics of axioms in $\riomb$ is not lost and can be exploited, e.g., in query generation. $\riomb$ and $\riomo~\setminus~\riomb$ partition the original ontology into a set of correct and possibly incorrect axioms, respectively. In the debugging session, only $\riomo := \riomo \setminus \riomb$ is used to search for diagnoses. This can reduce the search space for diagnoses substantially. Another application of background knowledge could be the reuse of an existing ontology to support successful debugging. For example, when formulating an ontology about medical terms, a thoroughly curated reference ontology $\riomb$ could be leveraged to find own formulations contradicting the correct ones in $\riomb$, which would not be found without integration of $\riomb$ into the debugging procedure.     

More formally, ontology debugging can be defined in terms of a diagnosis problem instance, for which we search for solutions, i.e. diagnoses, that enable to formulate the target ontology:
%conditions a target ontology must fulfill, which leads to the definition of a diagnosis problem instance, for which we search for solutions, i.e. diagnoses:
%
\begin{definition}[Diagnosis Problem Instance, Target Ontology] Let $\riomo = \riomt \cup \rioma$ be an ontology with terminological axioms $\riomt$ and assertional axioms $\rioma$, $\riomb$ a set of axioms which are assumed to be correct (background knowledge), $\rioRQ$ a set of requirements
%\footnote{In this work we assume $\rioRQ=\{\text{coherence}\}$.} 
to $\riomo$, $\rioTp$ and $\rioTn$ respectively a set of positive and negative test cases, where each test case $p\in\rioTp$ and $n\in\rioTn$ is a set of axioms. Then we call the tuple $\langle\riomo,\riomb,\rioTp,\rioTn\rangle_\rioRQ$ a \emph{diagnosis problem instance (DPI)}.
An ontology $\rioot$ is called \emph{target ontology} w.r.t.\ $\langle\riomo,\riomb,\rioTp,\rioTn\rangle_\rioRQ$ iff all the following conditions hold:
%\vspace{-6pt}
\begin{eqnarray*}
		 \forall \, r  \in \rioRQ&:& \;\rioot \cup \riomb \,\text{ fulfills }\, r  \\
		 \forall \,\riotp \in \rioTp&:& \;\rioot \cup \riomb \,\models\, \riotp				\\
		 \forall \,\riotn \in \rioTn&:& \;\rioot \cup \riomb \,\not\models\, \riotn .
\end{eqnarray*}
%\vspace{-14pt}
\end{definition}

\begin{definition}[Diagnosis]\label{def:riodiagnosis}
We call $\riomd \subseteq \riomo$ a \emph{diagnosis} w.r.t.\ a DPI $\langle\riomo,\riomb,\rioTp,\rioTn\rangle_\rioRQ$ iff 
%there exists a set of axioms $EX_\riomd$ such that 
$(\riomo\setminus\riomd)\cup (\bigcup_{\riotp \in \rioTp} \riotp)$ is a target ontology w.r.t.\ $\langle\riomo,\riomb,\rioTp,\rioTn\rangle_\rioRQ$.
A diagnosis $\riomd$ w.r.t.\ a DPI is \emph{minimal} iff there is no $\riomd' \subset \riomd$ such that \ $\riomd'$ is a diagnosis w.r.t.\ this DPI. The set of minimal diagnoses w.r.t.\ a DPI is denoted by $\riominD$.
\end{definition}
Note that a diagnosis $\riomd$ gives complete information about the correctness of each axiom $ax_k \in \riomo$, i.e. all $ax_i \in \riomd$ are assumed to be faulty and all $ax_j \in \riomo\setminus\riomd$ are assumed to be correct. 
%The identification of an extension $EX_\riomd$, accomplished e.g. by some learning approach, is a crucial part of the ontology repair process. However, the formulation of a complete extension is outside the scope of this work where we focus on computing diagnoses. Following the approach suggested in~\cite{jws12}, we approximate $EX_\riomd$ by the set $\bigcup_{\riotp \in \rioTp} \riotp$.

%\noindent\textbf{Example:} 
\begin{example}\label{example:rio_ex_1}
Consider $\riomo:=\riomt\cup\rioma$ with terminological axioms $\riomt:=\riomo_1\cup\riomo_2\cup\rioAl_{12}$:
\begin{center}
\normalsize
\begin{tabular}{crl}
$\riomo_1$ \qquad\quad & $\riotax_1:$ &  $PhD \sqsubseteq Researcher$ \\ 
        & $\riotax_2:$ &  $Researcher \sqsubseteq DeptEmployee$ \vspace{1pt}\\ 
        \hline\vspace{-8pt}\\
$\riomo_2$ \qquad\quad & $\riotax_3:$ &  $PhDStudent \sqsubseteq Student$  \\ 
        & $\riotax_4:$ &  $Student \sqsubseteq \lnot DeptMember$ \vspace{1pt}\\ 
        \hline\vspace{-8pt}\\
$\rioAl_{12}$ \qquad\quad & $\riotax_5:$ &  $PhDStudent \sqsubseteq PhD$ \\
        & $\riotax_6:$ &  $DeptEmployee \sqsubseteq DeptMember$
\end{tabular}
\end{center}
and an assertional axiom $\rioma=\setof{PhDStudent(s)}$, where $\rioAl_{12}$ is an automatically generated set of axioms serving as semantic links between $\riomo_1$ and $\riomo_2$. The given ontology $\riomo$ is inconsistent since it describes $s$ as both a $DeptMember$ and not. 

Let us assume that the assertion $PhDStudent(s)$ is considered as correct and is thus added to the background theory, i.e. $\riomb := \rioma$, and that no test cases are initially specified, i.e. the sets $\rioTp$ and $\rioTn$ are empty. For the resulting DPI $\tuple{\riomt,\rioma,\emptyset,\emptyset}_{\setof{\text{coherence}}}$ the set of minimal diagnoses $\riominD=\{\riomd_1 : [\riotax_1], \riomd_2 : [\riotax_2], \riomd_3 : [\riotax_3], \riomd_4 : [\riotax_4], \riomd_5 : [\riotax_5],\riomd_6 : [\riotax_6]\}$. 
%for the given problem instance $\tuple{\riomt,\rioma,\emptyset,\emptyset}$. 
$\riominD$ can be computed by a diagnosis algorithm such as the one presented in~\cite{friedrich2005gdm}.\qed 
\end{example}

With six minimal diagnoses for only six ontology axioms, this example already gives an idea that in many cases 
%the number of minimal diagnoses 
$|\riominD|$ can get very large. 
Note that generally the computation of \emph{all} minimal diagnoses w.r.t.\ a given DPI is not feasible within reasonable time due to the complexity of the underlying algorithms. Therefore, in practice, especially in an interactive scenario where reaction time is essential, a set of \emph{leading diagnoses} $\riomD \subseteq \riominD$ is considered as a representative for $\riominD$.\footnote{So, we will speak of $\riomD$ instead of $\riominD$ 
%in the rest of 
throughout this work. Note that the restriction to a subset of $\riominD$ does not necessarily have implications on the completeness of the associated ontology debugging algorithm. E.g., the algorithm can be iterative and recompute new diagnoses on demand and nevertheless guarantee completeness (as the algorithm presented in this work).} Concerning the optimal number of leading diagnoses, a trade-off between representativeness and complexity of associated computations w.r.t.\ $\riomD$ needs to be found.

Without any prior knowledge in terms of diagnosis fault probabilities or specified test cases, each diagnosis in $\riomD$ is equally likely to be the target diagnosis $\riodt$. In other words, for each $\riomd \in \riomD$ w.r.t.\ the DPI $\tuple{\riomt,\rioma,\emptyset,\emptyset}_{\setof{\text{coherence}}}$, the ontology $(\riomo \setminus \riomd) \cup (\bigcup_{\riotp \in \rioTp} \riotp)$ meets all the conditions defining a target ontology. However, besides postulating coherence the user might want the target ontology to entail that $s$ is a student as well as a researcher, i.e. $\rioot\models t_1$ where $t_1:=\{Researcher(s),Student(s)\}$. Formulating $t_1$ as a positive test case yields the DPI $\tuple{\riomt,\rioma,\{t_1\},\emptyset}_{\setof{\text{coherence}}}$, for which only diagnoses $\riomd_2,\riomd_4,\riomd_6 \in \riomD$ are valid and enable to formulate a corresponding $\rioot$. All other diagnoses in $\riomD$ are ruled out by the fact that $t_1 \in \rioTp$, which means they have a probability of zero of being the target diagnosis. If $t_1 \in \rioTn$, in contrast, this would imply that $\riomd_2,\riomd_4,\riomd_6$ had to be rejected.

So, it depends on the test cases specified by a user
%, i.e.~answers to the queries asked to the user, 
which diagnosis will finally be identified as target diagnosis. Also, the order in which test cases are specified, is crucial. For instance, consider the test cases $t_1:=\{PhD(s)\}$ and $t_2:=\{Student(s)\}$. If $t_1\in\rioTp$ is specified before $t_2\in\rioTn$, then $t_1\in\rioTp$ is redundant, since the only diagnosis agreeing with $t_2\in\rioTn$ is $\riomd_3$ which preserves also the entailment $t_1$ in the resulting target ontology $\rioot = (\riomo\setminus\riomd_3) \cup \emptyset$ without explicating it as a positive test case.

Since it is by no means trivial to get the right -- in the sense of most informative -- test cases formulated in the proper order such that the number of test cases necessary to detect the target diagnosis is minimized, interactive debugging systems offer the functionality to automatize selection of test cases. The benefit is that the user can just concentrate on ``answering'' the provided test cases which means assigning them to either $\rioTp$ or $\rioTn$. We call such automatically generated test cases queries.
%The test cases, however, represent properties, i.e. entailments and non-entailments, of the target ontology $\rioot:=(\riomo \setminus \riodt) \cup EX_{\riodt}$ and thus allow to constrain the candidates for $\riodt$.
The theoretical foundation for the application of queries is the fact that $\riomo\setminus\riomd_i$ and $\riomo\setminus\riomd_j$ for $\riomd_i \neq \riomd_j \in \riomD$\, entail different sets of axioms.   
%In order to define a query~\cite{jws12}, the fact is exploited that ontologies $\riomo\setminus\riomd_i$ and $\riomo\setminus\riomd_j$ resulting in application of different diagnoses $\riomd_i,\riomd_j \in \riomD$\, ($\riomd_i \neq \riomd_j$) entail different sets of logical descriptions. When we speak of entailments, we address the output computed by the classification and realization services of a reasoner. Formally, a query is defined as follows:
%
\begin{definition}[Query, Partition]
Let $\riomD$ be a set of minimal diagnoses w.r.t.\ a DPI $\langle\riomo,\riomb,\rioTp,\rioTn\rangle_\rioRQ$ and $\riomo^{*}_i := (\riomo \setminus \riomd_i) \cup \riomb \cup (\bigcup_{\riotp\in\rioTp} \riotp)$ for $\riomd_i\in\riomD$. Then a set of axioms $X_j\neq\emptyset$ is called a \emph{query} w.r.t.\ $\riomD$ iff $\riodx{j}:=\setof{\riomd_i \in \riomD\,|\,\riomo^{*}_i \models X_j}\neq \emptyset$ and 
$\riodnx{j}:=\setof{\riomd_i \in \riomD\,|\,\exists x\in\rioTn\cup\rioRQ: \riomo^{*}_i \cup X_j \text{ violates } x}\neq \emptyset$.
%\riomo^{*}_i \riomodels \lnot X_j}\neq \emptyset$.
The (unique) \emph{partition} of a query $X_j$ is denoted by $\langle \riodx{j}, \riodnx{j}, \riodz{j} \rangle$ where $\riodz{j} = \riomD \setminus (\riodx{j} \cup \riodnx{j})$.
%$\riomX_\riomD$ terms the set of all queries and associated partitions w.r.t.\ $\riomD$.
$\riomX_\riomD$ terms a set of queries and associated partitions w.r.t.\ $\riomD$ in which one and the same partition of $\riomD$ occurs at most once and only if there is an associated query for this partition. 
\end{definition}
%The (complete) set of queries 
Note that, in general, there can be $n_q$ queries for a particular partition of $\riomD$ where $n_q$ can be zero or some positive integer. We are interested in (1)~only those partitions for each of which $n_q\geq 1$
%there is a query which implies this partition 
and (2)~only one query for each such partition. The set $\riomX_\riomD$ includes elements such that (1) and (2) holds. 
$\riomX_\riomD$ for a given set of minimal diagnoses $\riomD$ w.r.t.\ a DPI can be generated as shown in Algorithm~\ref{rioalgo_query_gen}. In each iteration, given a set of diagnoses $\riodx{k} \subset \riomD$, common entailments\footnote{Note, when we speak of entailments throughout this work, we address (only) the \emph{finite} set of entailments computed by the classification and realization services of a DL-reasoner.} $X_k:=\setof{e\,|\,\forall\riomd_i\in\riodx{k}:\riomo^*_i\models e}$ are computed (\textsc{getEntailments}) and used to classify the remaining diagnoses in $\riomD\setminus\riodx{k}$ to obtain the partition $\langle\riodx{k},\riodnx{k},\riodz{k}\rangle$ associated with~$X_k$. 
Then, if the partition $\langle\riodx{k},\riodnx{k},\riodz{k}\rangle$ does not already occur in $\riomX_\riomD$ (\textsc{includesPartition}), the query $X_k$ is minimized~\cite{Shchekotykhin2012} (\textsc{minimizeQuery}) such that its partition is preserved, yielding a query $X'_k\subseteq X_k$ such that any $X''_k \subset X'_k$ is not a query or has not the same partition. Finally, $X'_k$ is added to 
%the set of queries 
$\riomX_\riomD$ together with its partition $\langle\riodx{k},\riodnx{k},\riodz{k}\rangle$. 
Function \textsc{reqViolated}($arg$) returns $true$ if $arg$ violates some requirement in $\rioRQ$ or entails some negative test case in $\rioTn$.
%is inconsistent or incoherent.

Asking the user a query $X_j$ means asking them $(\rioot \models X_j ?)$.
Let the answering of queries by a user be modeled as function $u: \riomX_\riomD \rightarrow \{\textit{t},\textit{f}\}$. If $u_j := u(X_j) = \textit{t}$, then $\rioTp \leftarrow \rioTp \cup \setof{X_j}$ and $\riomD \leftarrow \riomD\setminus\riodnx{j}$. Otherwise, $\rioTn \leftarrow \rioTn \cup \setof{X_j}$ and $\riomD \leftarrow \riomD\setminus\riodx{j}$.
Prospectively, according to Definition~\ref{def:riodiagnosis}, only those diagnoses are considered in the set $\riomD$ that comply with the new DPI obtained by the 
%specification 
addition
of a test case. 
% then the following holds: If $u(X_j) = \textit{t}$, then $X_j$ is added to the positive test cases, i.e. $\rioTp \leftarrow \rioTp \cup \setof{X_j}$, and all diagnoses in $\riodnx{j}$ are \emph{rejected}. Given that $u(X_j) = \textit{f}$, then $\rioTn \leftarrow \rioTn \cup \setof{X_j}$ and all diagnoses in $\riodx{j}$ are \emph{rejected}.
%
This allows us to formalize the problem we address in this work:
\vspace{3pt}

\noindent\fcolorbox{black}{light-gray1}{\parbox[c][5.4em][c]{0.975\linewidth}{\vspace{-4pt}
\begin{prob_def}[Query Selection] \label{prob_def:rio} 
\textbf{Given}~$\riomD$ w.r.t.\ a DPI $\langle\riomo,\riomb,\rioTp,\rioTn\rangle_\rioRQ$, a stopping criterion $stop:\riomD\rightarrow\{t,f\}$ and a user~$u$, \textbf{find} a next query $X_j\in\riomX_\riomD$ such that 
(1)~$(X_j,\dots,X_q)$ is a query sequence of minimal length and 
(2)~there exists a $\riodt \in \riomD$ w.r.t.\ $\langle\riomo,\riomb,\rioTp',\rioTn'\rangle_\rioRQ$ such that $stop(\riodt) = t$, where $\rioTp':= \rioTp \cup \{X_i\,|\,X_i\in\{X_j,\dots,X_q\},u_i=t\}$ and $\rioTn':= \rioTn \cup \{X_i\,|\,X_i\in\{X_j,\dots,X_q\},u_i=f\}$.
\end{prob_def}\vspace{-4pt}
}}

\vspace{3pt}
\begin{algorithm*}[tb]
\small
\caption{Generation of Queries and Partitions} \label{rioalgo_query_gen}
\begin{algorithmic}[1]
\Require 
DPI $\tuple{\riomo,\riomb,\rioTp,\rioTn}_\rioRQ$, set of minimal diagnoses $\riomD$ w.r.t.\ $\tuple{\riomo,\riomb,\rioTp,\rioTn}_\rioRQ$
\Ensure 
a set of queries and associated partitions $\riomX_\riomD$
\vspace{10pt}
\State $\riomX_\riomD \gets \emptyset$
\For{$\riodx{k} \subset \riomD$}
	\State $X_k \gets \Call{getEntailments}{\riomo, \riomb, \rioTp, \riodx{k}}$
	\If{$X_k \neq \emptyset$}
		\For{$\riomd_r \in \riomD\setminus\riodx{k}$}
			\If{$\riomo^{*}_r \,\models X_k$}
				\State $\riodx{k} \gets \riodx{k} \cup \left\{\riomd_r\right\}$
			\ElsIf{$\Call{reqViolated}{\riomo^{*}_r \cup X_k}$}
				\State $\riodnx{k} \gets \riodnx{k} \cup \left\{\riomd_r\right\}$
			\Else 
				\State $\riodz{k} \gets \riodz{k} \cup \left\{\riomd_r\right\}$
			\EndIf
		\EndFor
		\If{$\lnot \Call{includesPartition}{\riomX_\riomD,\tuple{\riodx{k}, \riodnx{k}, \riodz{k}}}$}
			\State $\riomX_\riomD \gets \riomX_\riomD \cup \Call{minimizeQuery}{\tuple{X_k, \tuple{\riodx{k}, \riodnx{k}, \riodz{k}}}}$
		\EndIf
	\EndIf
\EndFor
\State \Return $\riomX_\riomD$
\end{algorithmic}
\normalsize
\end{algorithm*}

Two strategies for selecting the ``best'' next query have been proposed~\cite{Shchekotykhin2012}:

%\noindent\textbf{Split-in-half strategy (SPL)}, 
\paragraph{Split-In-Half Strategy (SPL)}
selects the query $X_j$ which minimizes the following scoring function: 
%\begin{equation*}
%$sc_{split}(X_j) := \left| |\riodx{j}| - |\riodnx{j}| \right| + |\riodz{j}|$.
%\end{equation*}
\begin{align*}
sc_{split}(X_j) := \left| |\riodx{j}| - |\riodnx{j}| \right| + |\riodz{j}|
\end{align*}
So, SPL prefers queries which eliminate half of the diagnoses independently of the query outcome.
%
%The \textbf{entropy-based approach} uses meta information about probabilities $p_t$ that the user makes a fault when using a syntactical construct of type $t \in \mathit{CT}$ where $\mathit{CT}$ is the set of construct types available in the used ontology expression language. For example, $\forall$, $\exists$, $\sqsubseteq$, $\neg$, $\sqcup$, $\sqcap$ are some OWL DL construct types.

%\noindent\textbf{Entropy-based strategy (ENT)}, 
\paragraph{Entropy-Based Strategy (ENT)}
uses information about prior probabilities $p_t$ for the user to make a mistake when using a syntactical construct of type $t \in \mathit{CT}(\mathcal{L})$, where $\mathit{CT}(\mathcal{L})$ is the set of constructors available in the used knowledge representation language $\mathcal{L}$, e.g.~$\setof{\forall, \exists, \sqsubseteq, \neg, \sqcup, \sqcap} \subset \mathit{CT}(\text{OWL DL})$.
%a fault when using a syntactical construct of type $t \in \mathit{CT}$ where $\mathit{CT}$ is the set of construct types available in the used logical description language. E.g., $\forall$, $\exists$, $\sqsubseteq$, $\neg$, $\sqcup$, $\sqcap$ are some OWL DL construct types. 
These fault probabilities $p_t$ are assumed to be independent and used to calculate fault probabilities of axioms $ax_k$ as 
%\begin{equation}
%\label{eq:prob_axiom}
%$p(\riotax_k) = 1 - \prod_{t \in \mathit{CT}} (1-p_t)^{n(t)}$
%\end{equation}
\begin{align*}
p(\riotax_k) = 1 - \prod_{t \in \mathit{CT}} (1-p_t)^{n(t)}
\end{align*}
 where $n(t)$ is the number of occurrences of construct type $t$ in $ax_k$.
%The same approach can also be applied in ontology matching. However, in many application scenarios the matched ontologies are assumed to be correct whereas the alignment $M$ is considered as faulty. In this case one can assign some small prior fault probability, e.g. $p(\riotax_i)=0.001$, to all axioms $\riotax_i$ of the matched ontologies. The prior fault probability of an axiom $\riotax_j \in M$ can be computed as $p(\riotax_j) = 1-v_j$, where $v_j$ is the confidence value of the correspondence underlying $\riotax_j$. 
The probabilities of axioms can in turn be used to determine fault probabilities of diagnoses $\riomd_i \in \riomD$ as 
%\vspace{-3pt}
\begin{align}
\label{eq:rioprob_diagnosis}
p(\riomd_i) = \prod_{\riotax_r \in \riomd_i} p(\riotax_r) \prod_{\riotax_s \in \riomo\setminus\riomd_i} (1-p(\riotax_s)).
\end{align}
%\vspace{-9pt}\\
ENT selects the query $X_j\in\riomX_\riomD$ with highest expected information gain, i.e. that minimizes the following scoring function \cite{Shchekotykhin2012}:
\begin{align*} 
    sc_{ent}(X_j) &= \sum_{a\in\{t,f\}} p(u_j=a)\log_2{p(u_j=a)} + p(\riodz{j}) + 1 \\
		\intertext{where}
		p(u_j=t) &= \sum_{\riomd_r \in \riodx{j}} p(\riomd_r) + \frac{1}{2} p(\riodz{j}) \\
		\intertext{and}
		p(\riodz{j}) &=\sum_{\riomd_r \in \riodz{j}} p(\riomd_r)
%\label{eq:score}
\end{align*} 
% $sc_{ent}(X_j)$ defined as:
%\begin{equation*}
%%H_e(X_i) = 
%sc_{ent}(X_j)\;:=\;\sum_{a\in\{t,f\}} p(u_j = a) \sum_{\riomd_k \in {\bf D}} -p(\riomd_k | u_j = a) \log_2 p(\riomd_k | u_j = a)
%\vspace{2pt}
%\end{equation*}
%where $p(u_j=t) = \sum_{\riomd_r \in \riodx{j}} p(\riomd_r) + \frac{1}{2} p(\riodz{j})$ and $p(\riodz{j})=\sum_{\riomd_r \in \riodz{j}} p(\riomd_r)$.
 %and $p(u_j=f)=1-p(u_j=t)$.
The answer $u_j=a$ is used to update probabilities $p(\riomd_k)$ for $\riomd_k \in \riomD$ according to the Bayesian formula, yielding $p(\riomd_k | u_j = a)$.

The result of the evaluation in~\cite{Shchekotykhin2012} shows that ENT reveals better performance than SPL in most of the cases. However, SPL proved to be the best strategy in situations when misleading prior information is provided, i.e.~the target diagnosis $\riodt$ has low probability. So, one can regard ENT as a high risk strategy with high potential to perform well, depending on the priorly unknown quality of the given fault information. SPL, in contrast, can be seen as a no-risk strategy without any potential to leverage good meta information.
Therefore, selection of the proper combination of prior probabilities $\setof{p_t\,|\,t\in\mathit{CT}(\mathcal{L})}$ and query selection strategy is crucial for successful 
diagnosis discrimination and minimization of user interaction.

\begin{table}[tb]
\centering
\small 
\begin{tabular}{@{\extracolsep{0pt}}l|l|l|c}
Query & $\riodx{i}$  & $\riodnx{i}$ &  $\riodz{i}$ \\\hline
$X_1:\{DeptEmployee(s),$ & $\riomd_4,\riomd_6$ & $\riomd_1,\riomd_2,\riomd_3,\riomd_5$ & $\emptyset$ \\
$\qquad\quad Student(s)\}$ & & & \\
$X_2:\{PhD(s)\}$ & $\riomd_1, \riomd_2,\riomd_3,\riomd_4,\riomd_6$   & $\riomd_5$ & $\emptyset$ \\
$X_3:\{Researcher(s)\}$ & $\riomd_2, \riomd_3, \riomd_4, \riomd_6$  & $\riomd_1, \riomd_5$ & $\emptyset$ \\
$X_4:\{Student(s)\}$ & $\riomd_1, \riomd_2, \riomd_4, \riomd_5, \riomd_6$ & $\riomd_3$ & $\emptyset$ \\
$X_5:\{Researcher(s),$ & $\riomd_2, \riomd_4, \riomd_6$ & $ \riomd_1, \riomd_3, \riomd_5$ &
$\emptyset$ \\
$\qquad\quad Student(s)\}$ & & & \\
$X_6:\{DeptMember(s)\}$ & $\riomd_3, \riomd_4$ & $ \riomd_1, \riomd_2, \riomd_5, \riomd_6$ & \\
$X_7:\{PhD(s),$ & $\riomd_1, \riomd_2, \riomd_4, \riomd_6$ & $ \riomd_3, \riomd_5$ &
$\emptyset$ \\
$\qquad\quad Student(s)\}$ & & & \\
$X_8:\{DeptMember(s),$ & $\riomd_2$ & $ \riomd_1, \riomd_3, \riomd_4, \riomd_5, \riomd_6$ &
$\emptyset$ \\
$\qquad\quad Student(s)\}$ & & & \\
$X_9:\{DeptEmployee(s)\}$ & $\riomd_3, \riomd_4, \riomd_6$ & $ \riomd_1, \riomd_2, \riomd_5$ &
$\emptyset$ \\\hline
\end{tabular} \vspace{2pt}
\normalsize
\caption[A Set of Queries and Associated Partitions w.r.t.\ the Initial DPI of the Example Ontology]{A set $\riomX_\riomD$ of queries and associated partitions w.r.t.\ the initial DPI $\tuple{\riomt,\rioma,\emptyset,\emptyset}_{\setof{\text{coherence}}}$ of the example ontology $\riomo$.
%Nine queries computed with respect to entailed assertional axioms for diagnoses $\riomd_i \in \riomD$ of the sample ontology $\riomo$. Given that no diagnoses have been eliminated yet, $X_2, X_4, X_8$ are high-risk queries, $X_5, X_9$ are no-risk queries.
}\label{tab:rioqueries_ex}
%\vspace{-20pt}
\end{table}

%\noindent\textbf{Example (continued):}
\begin{example}\label{example:rio_ex_2} (Example~\ref{example:rio_ex_1} continued)
%To illustrate this, let the user decide to debug our example ontology $\riomo$. To that end, the user specifies the two merged ontologies as correct  $p(\riotax_i) = 0.001$ for all $\riotax_i \in \riomo_1\cup\riomo_2$ and uses confidence values provided by a matching system to compute the probabilities $p(\riotax_5)=0.1$ and $p(\riotax_6)=0.15$. Assume that $\riomd_2$ corresponds to the target diagnosis $\riodt$, i.e. the settings provided by the user are incorrect.
To illustrate this, let a user who wants to debug our example ontology $\riomo$ 
%regard the two matched ontologies $\riomo_1,\riomo_2$ as correct and thus 
set $p(\riotax_i) := 0.001$ for $ax_{i(i=1,\dots,4)}$ and $p(\riotax_5):=0.1, p(\riotax_6):=0.15$, e.g. because the user doubts the correctness of $\riotax_5,\riotax_6$ while being quite sure that $ax_{i(i=1,\dots,4)}$ are correct.
%$\riotax_i \in \riomo_1\cup\riomo_2$ and for the alignment axioms $p(\riotax_5):=0.1, p(\riotax_6):=0.15$ according to the confidence values provided by the matching system. 
Assume that $\riomd_2$ corresponds to the target diagnosis $\riodt$, i.e. the settings provided by the user are inept.
Application of ENT starts with computation of prior fault probabilities of diagnoses\label{ex:rioprobs} $p(\riomd_1) = p(\riomd_2) = p(\riomd_3) = p(\riomd_4) = 0.003$, $p(\riomd_5) = 0.393$, $p(\riomd_6) = 0.591$ (Formula~\ref{eq:rioprob_diagnosis}). Then $(\rioot \models X_1?)$ with $X_1 := \{DeptEmployee(s),Student(s)\}$, will be identified as the optimal query since it has the minimal score $sc_{ent}(X_1)=0.02$ (see Table~\ref{tab:rioqueries_ex} for queries and partitions w.r.t.\ the example ontology). 
%However, note that this query could eliminate only two diagnoses $\riomd_4$ and $\riomd_6$, if the unfavorable answer \textit{no} is given. Hence, we call $X_1$ a \emph{high-risk query} as its elimination rate = 2/6.
However, since the unfavorable answer $u_1 = f$ is given, this query eliminates only two of six diagnoses $\riomd_4$ and $\riomd_6$. 
%(worst case elimination rate $e_{wc}(X_1) = \frac{2}{6}$). 
%Hence, we call $X_1$ a \emph{high-risk query} as its elimination rate = 2/6. 
%
The Bayesian probability update 
%given by Formula~\ref{eq:bayes} 
then yields $p(\riomd_2) = p(\riomd_3) = p(\riomd_4) = 0.01$ and $p(\riomd_5) = 0.97$. As next query $X_2$ with $sc_{ent}(X_2)=0.811$ is selected and answered unfavorably ($u_2 = t$) as well which results in the elimination of only one of four diagnoses $\riomd_5$. 
%($e_{wc}(X_2) = \frac{1}{4}$). 
%Since the worst case elimination rate $e_{wc}(X_2)$ is minimal, we call $X_2$ a \emph{high-risk query}. %because the minimal elimination rate is only $\frac{1}{4}$. 
By querying $X_3$ ($sc_{ent}(X_3)=0.082$, $u_3 = t$) and $X_4$ ($sc(X_4)=0$, $u_4 = t$), the further execution of this procedure finally leads to the target diagnosis $\riomd_2$. So, application of ENT requires four queries to find $\riodt$.
If SPL is used instead, only three queries are required. The algorithm can select one of the two queries $X_5$ or $X_9$ because each eliminates half of all diagnoses in any case. %($e_{wc}(X)=\frac{1}{2}$ for $X\in\{X_5,X_9\}$). 
%We call such a query a \emph{no-risk query}. 
Let the strategy select $X_5$ which is answered positively ($u_5 = t$). As successive queries, $X_6$ ($u_6=f$) and $X_1$ ($u_1=f$) are selected, which leads to the revelation of $\riodt = \riomd_2$.\qed
\end{example}

This scenario demonstrates that the no-risk strategy SPL (\textit{three queries}) is more suitable than ENT (\textit{four queries}) for fault probabilities which disfavor the target diagnosis. Let us suppose, on the other hand, that probabilities are assigned more reasonably in our example, e.g. $\riodt=\riomd_6$. Then it will take ENT only \textit{two queries} $(X_1,X_6)$ to find $\riodt$ while SPL will still require \textit{three queries}, e.g. $(X_5,X_1,X_6)$. 
%The complexity of $sc_{ent}$ in terms of required queries varies between $O(1)$ in the best and $O(|\riomD|)$ in the worst case depending on the appropriateness of the fault probabilities. In contrast, $sc_{split}$ always requires $O(\log_2 |\riomD|)$ queries.

This example indicates that, unless the target diagnosis is known in advance, one can never be sure to select the best strategy from SPL and ENT. In Chapte~\ref{chap:riotheory} we present a learning query selection algorithm that combines the benefits of both SPL and ENT. It adapts the way of selecting the next query depending on the elimination rate (like SPL) and on information gain (like ENT). Thereby its performance approaches the performance of the better of both SPL and ENT.

\chapter{RIO: Risk Optimization for Query Selection}
%\section{Risk Optimization Strategy for Query Selection}
\label{chap:riotheory} 
The proposed Risk Optimization Algorithm (RIO) extends ENT strategy with a dynamic learning procedure that learns by reinforcement how to select the next query. Its behavior is determined by the achieved performance in terms of diagnosis elimination rate w.r.t.\ the set of leading diagnoses $\riomD$. Good performance causes similar behavior to ENT, whereas aggravation of performance leads to a gradual neglect of the given meta information, and thus to a behavior akin to SPL. Like ENT, RIO continually improves the prior fault probabilities based on new knowledge obtained through queries to a user.

RIO learns a ``cautiousness'' parameter $\riouc$ whose admissible values are captured by the user-defined interval $[\underline{\riouc},\overline{\riouc}]$. The relationship between $\riouc$ and queries is as follows:
%The relationship between cautiousness $\riouc$ and queries is formalized by the following definitions:
%
\begin{definition}[Cautiousness of a Query]\label{def:rioquery_cautiousness}
We define the \emph{cautiousness} $\rioqc(X_i)$ \emph{of a query} $X_i \in \riomX_\riomD$ as follows:
%
%\vspace{-7pt}
\begin{align*}\label{eq:rioquery_cautiousness}
\rioqc(X_i) := \frac{ \min\left\{|\riodx{i}| ,|\riodnx{i}| \right\} }{|\riomD|} \in \left[0,\frac{\left\lfloor \frac{|\riomD|}{2}\right\rfloor}{|\riomD|}\right] =: [\underline{\rioqc},\overline{\rioqc}]
%\vspace{-3pt}
\end{align*}
A query $X_i$ is called \emph{braver} than query $X_j$ iff $\rioqc(X_i) < \rioqc(X_j)$. Otherwise $X_i$ is called \emph{more cautious} than $X_j$.
A query with 
%highest possible 
maximum cautiousness $\overline{\rioqc}$ is called \emph{no-risk query}.
\end{definition}
\begin{definition}[Elimination Rate]\label{def:rioelimination_rate}
Given a query $X_i$ and the corresponding answer $u_i \in \{t,f\}$, the \emph{elimination rate} 
%$e(X_i,u_i)=\frac{|\riodnx{i}|}{|\riomD|}$ if $u_i = t$ and $e(X_i,u_i)=\frac{|\riodx{i}|}{|\riomD|}$ if $u_i = f$. 
\begin{align*}
e(X_i,u_i)&=\frac{|\riodnx{i}|}{|\riomD|} \qquad\text{ if }\; u_i = t \\
\intertext{and}
e(X_i,u_i)&=\frac{|\riodx{i}|}{|\riomD|} \qquad\text{ if }\; u_i = f
\end{align*}
%is defined as follows:
%\[
%e(X_i,u_i)=\begin{cases}
%  \frac{|\riodnx{i}|}{|\riomD|}  	& \text{if }\;u_i = t   \vspace{0.2cm}\\ 
%  \frac{|\riodx{i}|}{|\riomD|} 		& \text{if }\;u_i = f
%\end{cases}
%\]
The answer $u_i$ to a query $X_i$ is called \emph{favorable} iff it maximizes the elimination rate $e(X_i,u_i)$. Otherwise $u_i$ is called \emph{unfavorable}. The minimal or \emph{worst case elimination rate} $\min_{u_i \in \{t,f\}}(e(X_i,u_i))$ of $X_i$ is denoted by $e_{wc}(X_i)$.
\end{definition}
So, the cautiousness $\rioqc(X_i)$ of a query $X_i$ is exactly the worst case elimination rate, i.e. $\rioqc(X_i) = e_{wc}(X_i) = e(X_i,u_i)$ given that $u_i$ is the unfavorable query result.
Intuitively, parameter $\riouc$ characterizes the minimum proportion of diagnoses in $\riomD$ which should be eliminated by the successive query.
%For braver queries the interval between minimum and maximum elimination rate is 
%larger than for more cautious queries. For no-risk queries it is minimal. 
%
\begin{definition}[High-Risk Query]\label{def:riohigh-risk_query}
Given a query $X_i$ and cautiousness $\riouc$, $X_i$ is called a \emph{high-risk query} iff $\rioqc(X_i) < \riouc$, i.e. the cautiousness of the query is lower than the algorithm's current cautiousness value $c$.
Otherwise, $X_i$ is called \emph{non-high-risk query}. By $\rioNHR_c(\riomX_\riomD)\subseteq \riomX_\riomD$ we denote the set of non-high-risk queries w.r.t.\ $c$. For given cautiousness $c$, the set of queries $\riomX_\riomD$ can be partitioned in high-risk queries and non-high-risk queries.
\end{definition}

\begin{example}\label{example:rio_ex_3} (Example~\ref{example:rio_ex_2} continued)
%\noindent\textbf{Example (continued):} 
%Reconsider the example given in Section~\ref{sec:basics} with $|\riomD|=6$. 
Let the user specify $\riouc := 0.3$ for the set $\riomD$ with $|\riomD|=6$. 
Given these settings, $X_1:=\{DeptEmployee(s),Student(s)\}$ is a non-high-risk query since its partition $\langle \riodx{1}, \riodnx{1}, \riodz{1} \rangle = \langle \setof{\riomd_4,\riomd_6}, \setof{\riomd_1,\riomd_2,\riomd_3,\riomd_5}, \emptyset\rangle$ and thus its cautiousness $\rioqc(X_1) = 2/6 \geq 0.3 = \riouc$. The query $X_2 :=\{PhD(s)\}$ with the partition $\langle \setof{\riomd_1,\riomd_2,\riomd_3,\riomd_4,\riomd_6}, \setof{\riomd_5}, \emptyset\rangle$ is a high-risk query because $\rioqc(X_2) = 1/6 < 0.3 = \riouc$ and $X_3:=\{Researcher(s),Student(s)\}$ with $\langle \{\riomd_2,\riomd_4,\riomd_6\}$, $\{\riomd_1,\riomd_3,\riomd_5\}$, $\emptyset\rangle$ is a no-risk query due to $\rioqc(X_3) = 3/6 = \overline{\rioqc}$.\qed
\end{example}

Given a user's answer $u_s$ to a query $X_s$, the cautiousness $\riouc$ is updated depending on the elimination rate $e(X_s,u_s)$ by $\riouc \leftarrow \riouc + \riouc_{adj}$
%as follows:
%
%\begin{equation}\label{eq:risk_update}
	%\riouc \;\leftarrow\; \riouc + \riouc_{adj}
%\end{equation}
%
where 
%$\riouc_{adj}$ denotes 
the cautiousness adjustment factor $\riouc_{adj} :=\; 2\,(\overline{\riouc} - \underline{\riouc})\mathit{adj}$.
%which is defined as follows:
%
%\begin{equation}
%	\riouc_{adj} :=\; 2\,(\overline{\riouc} - \underline{\riouc}) \mathit{adj} 	
%\label{eq:adjust}
%\end{equation}
%
The scaling factor $2 \, (\overline{\riouc} - \underline{\riouc})$ 
%in Formula \ref{eq:adjust} 
%is a scaling factor that simply 
regulates the extent of the cautiousness adjustment depending on the interval length $\overline{\riouc} - \underline{\riouc}$. 
More crucial is the factor $\mathit{adj}$ that indicates the sign and magnitude of the cautiousness adjustment:  
%\vspace{-3pt}
\begin{align*}
\mathit{adj} :=  \frac{\left\lfloor \frac{|\riomD|}{2}-\varepsilon\right\rfloor}{|\riomD|} - e(X_s,u_s)
%\vspace{-3pt}
\end{align*}
where $\varepsilon \in (0,\frac{1}{2})$ is a constant which prevents the algorithm from getting stuck in a no-risk strategy for even $|\riomD|$. E.g., given $\riouc=0.5$ and $\varepsilon=0$, the elimination rate of a no-risk query $e(X_s,u_s) = \frac{1}{2}$ resulting always in $adj=0$. 
The value of $\varepsilon$ can be set to an arbitrary real number, e.g. $\varepsilon := \frac{1}{4}$. 
%
%Note that $\mathit{adj} < 0$ holds for any favorable query result, independently of the cautiousness $\rioqc(X_s)$ of the asked query $X_s$. This has the following reasons: For even $|\riomD|$ and any favorable query outcome $a_s$, the resulting elimination rate $e(X_s,a_s) \geq \lfloor \frac{|\riomD|}{2}\rfloor / |\riomD|$, which is the maximal value the minimal elimination rate of a query can take (see Definition \ref{def:query_cautiousness}). So, $\lfloor \frac{|\riomD|}{2} - \epsilon\rfloor / |\riomD| < \lfloor \frac{|\riomD|}{2}\rfloor / |\riomD| \leq e(X_s,a_s)$ holds since $|\riomD|$ is even and $\epsilon \in (0,\frac{1}{2})$. In case of odd $|\riomD|$, for any favorable query result $a_s$: $e(X_s,a_s) > \lfloor \frac{|\riomD|}{2}\rfloor / |\riomD| = \lfloor \frac{|\riomD|}{2} - \epsilon\rfloor / |\riomD|$. 
%
If $\riouc + \riouc_{adj}$
%the value $\riouc$ computed in (\ref{eq:risk_update}) 
is outside the user-defined cautiousness interval $[\underline{\riouc},\overline{\riouc}]$, it is set to $\underline{\riouc}$ if $\riouc < \underline{\riouc}$ and to $\overline{\riouc}$ if $\riouc > \overline{\riouc}$. Positive $\riouc_{adj}$ is a penalty telling the algorithm to get more cautious, whereas negative $\riouc_{adj}$ is a bonus resulting in a braver behavior of the algorithm.
Note, for the user-defined interval $[\underline{\riouc},\overline{\riouc}] \subseteq [\underline{\rioqc},\overline{\rioqc}]$ must hold. 
$\underline{\riouc} - \underline{\rioqc}$ and $\overline{\rioqc} - \overline{\riouc}$ represent the minimal desired difference in performance to a high-risk (ENT) and no-risk (SPL) query selection, respectively.
By expressing trust (disbelief) in the prior fault probabilities through specification of lower (higher) values for $\underline{\riouc}$ and/or $\overline{\riouc}$, the user can take influence on the behavior of RIO.
 
%\noindent\textbf{Example (continued):}  
\begin{example}\label{example:rio_ex_4} (Example~\ref{example:rio_ex_3} continued)
Assume $p(\riotax_i) := 0.001$ for $ax_{i(i=1,\dots,4)}$ and $p(\riotax_5):=0.1$, $p(\riotax_6):=0.15$ and the user 
%is quite unsure about the location of the fault 
rather disbelieves these fault probabilities and thus
%has doubts whether the fault is in the alignment and 
sets $\riouc=0.4$, $\underline{\riouc}=0$ and $\overline{\riouc}=0.5$. In this case RIO selects a no-risk query $X_3$ just as SPL would do. Given 
%the answer 
$u_3=t$ and $|\riomD|=6$, the algorithm computes the elimination rate $e(X_3,t)=0.5$ and adjusts the cautiousness by $\riouc_{adj}=-0.17$ which yields $\riouc=0.23$. This allows RIO to select a higher-risk query in the next iteration, whereupon the
%. The algorithm finds the 
target diagnosis $\riodt=\riomd_2$ is found after asking three queries. In the same situation, ENT (starting with high-risk query $X_1$) would require four queries.\qed
\end{example}

\begin{algorithm*}[tb]
\small
\caption{Risk Optimization Algorithm (RIO)} \label{rioalgo_main}
\begin{algorithmic}[1]
\Require 
DPI $\langle\riomo,\riomb,\rioTp,\rioTn\rangle_\rioRQ$, fault probabilities of diagnoses $DP$, cautiousness $C=(\riouc,\underline{\riouc},\overline{\riouc})$, number of leading diagnoses $n$ to be considered, acceptance threshold $\sigma$
\Ensure 
a minimal diagnosis $\riomd$ w.r.t.\ $\langle\riomo,\riomb,\rioTp,\rioTn\rangle_\rioRQ$
\vspace{10pt}
%$\riomD \leftarrow \emptyset$ 
%\Repeat{$(\aboveThresh(DP,\sigma) \lor \getScore(X_s) = 0$)}{
        %$\riomD \leftarrow \getDiagnoses(\riomD, n, \riomo, \riomb, \rioTp, \rioTn)$\;
				%$DP \leftarrow \compPriors(DP,\riomD, \rioTp, \rioTn)$\;
				%$\riomX_\riomD \leftarrow \genQs(\riomo, \riomb, \rioTp, \riomD)$\;
				%$X_s \leftarrow \getMinScQ(DP,\riomX_\riomD)$\;    
        %%$s\leftarrow\getScore(X_s,sc_{ent})$\;
				%\lIf {$\getPercent(X_s,\riomD) < \riouc$}{        
						%$X_s \leftarrow \getBestQ(\riouc,\riomX_\riomD, DP, \riomD)$
				%}  
				%\lIf {$ \getAnswer(X_s) = \textit{yes}$}{        
						%$\rioTp \leftarrow \rioTp \cup \{X_s\}$  
				%}
				%\lElse{
						%$\rioTn \leftarrow \rioTn \cup \{X_s\}$
				%} 
				%$c \leftarrow \updateRisk(\riomD,\rioTp,\rioTn,X_s,\riouc,\underline{\riouc},\overline{\riouc})$\;
%}
%\Return $\mostProbDiag(\riomD, DP)$\;
%%%%%%%%%%%%%%%%%%
\State $\riomD \gets \emptyset$
\Repeat
\State $\riomD \gets \Call{getDiagnoses}{\riomD, n, \riomo, \riomb, \rioTp, \rioTn}$
\State $DP \gets \Call{getProbabilities}{DP,\riomD, \rioTp, \rioTn}$
\State $\riomX_\riomD \gets \Call{generateQueries}{\riomo, \riomb, \rioTp, \riomD}$
\State $X_s \gets \Call{getMinScoreQuery}{DP,\riomX_\riomD}$
\If{$\Call{getQueryCautiousness}{X_s,\riomD} < \riouc$}
	\State $X_s \gets \Call{getAlternativeQuery}{\riouc,\riomX_\riomD, DP, \riomD}$
\EndIf
\If{$\Call{getAnswer}{X_s} = \textit{yes}$}
	\State $\rioTp \gets \rioTp \cup \{X_s\}$
\Else
	\State $\rioTn \gets \rioTn \cup \{X_s\}$
\EndIf
\State $c \gets \Call{updateCautiousness}{\riomD,\rioTp,\rioTn,X_s,\riouc,\underline{\riouc},\overline{\riouc}}$
\Until{$\Call{aboveThreshold}{DP,\sigma} \lor \Call{eliminationRate}{X_s} = 0$}
\State \Return $\Call{mostProbableDiag}{\riomD, DP}$
\end{algorithmic}
\normalsize
\end{algorithm*}

RIO, described in Algorithm~\ref{rioalgo_main}, starts with the computation of minimal diagnoses. \textsc{getDiagnoses} function implements a combination of HS-Tree and QuickXPlain algorithms~\cite{Shchekotykhin2012}. Using uniform-cost search, the algorithm extends the set of leading diagnoses $\riomD$ with a maximum number of most probable minimal diagnoses such that $|\riomD| \leq n$. 

Then the \textsc{getProbabilities} function calculates the fault probabilities $p(\riomd_i)$ for each diagnosis $\riomd_i$ of the set of leading diagnoses $\riomD$ using Formula~(\ref{eq:rioprob_diagnosis}). Next it adjusts the probabilities as per the Bayesian theorem taking into account all previous query answers which are stored in $\rioTp$ and $\rioTn$. Finally, the resulting probabilities $p_{adj}(\riomd_i)$ are normalized.
%. In order to take into account all information gathered by querying an oracle so far the algorithm 
%adjusts fault probabilities $p(\riomd_i)$ as follows: $p_{adj}(\riomd_i)=(1/2)^{z}\, p(\riomd_i)$, where $z$ is the number of precedent queries $X_k$ for which $\riomd_i \in \riodz{k}$.
%Afterwards the probabilities $p_{adj}(\riomd_i)$ are normalized. 
%Note that $z$ can be computed from $\rioTp$ and $\rioTn$ which comprise all query answers. This way of updating probabilities is exactly in compliance with the Bayesian theorem.
% given by Formula~\ref{eq:bayes}.
Based on the set of leading diagnoses $\riomD$, \textsc{generateQueries} generates queries according to Algorithm~\ref{rioalgo_query_gen}.
	\textsc{getMinScoreQuery} determines the best query $\rioXsc \in \riomX_\riomD$ according to $sc_{ent}$: 	%$\rioXsc = \argmin_{X_k \in \riomX_\riomD}(sc_{ent}(X_k))$.
\begin{align*}
\rioXsc = \argmin_{X_k \in \riomX_\riomD}(sc_{ent}(X_k))
\end{align*}
%			\begin{equation*}
%				\rioXsc = \argmin_{X_k \in \riomX_\riomD}(sc_{ent}(X_k)) 
%			\end{equation*}	
%\label{step_standard} 
If $\rioXsc$ is a non-high-risk query, i.e. $\riouc \leq \rioqc(\rioXsc)$ (determined by \textsc{getQueryCautiousness}), $\rioXsc$ is selected. In this case, $\rioXsc$ is the query with best information gain 
%among all queries 
in $\riomX_\riomD$ and moreover guarantees the required elimination rate specified by $\riouc$.
	
\label{step_alternative} Otherwise, \textsc{getAlternativeQuery} selects the query $\rioXalt \in \riomX_\riomD$\; $(\rioXalt \neq \rioXsc)$ which has minimal score $sc_{ent}$ among all least cautious non-high-risk queries $L_c$. That is, 
%$\rioXalt = \argmin_{X_k \in \mathit{L}_c}(sc_{ent}(X_k)) $
\begin{align*}
\rioXalt &= \argmin_{X_k \in \mathit{L}_c}(sc_{ent}(X_k)) \\
\intertext{where}
\mathit{L_c} &:= \{X_r \in \rioNHR_c(\riomX_\riomD) \;|\; \forall X_t \in \rioNHR_c(\riomX_\riomD):\, \rioqc(X_r) \leq \rioqc(X_t)\}
\end{align*}
%\begin{equation*}
%	\rioXalt = \argmin_{X_k \in \mathit{L}_c}(sc_{ent}(X_k)) 
%\end{equation*}
%where $\mathit{L_c} := \{X_r \in \rioNHR_c(\riomX_\riomD) \;|\; \forall X_t \in \rioNHR_c(\riomX_\riomD):\, \rioqc(X_r) \leq \rioqc(X_t)\}$. 
If there is no such query $\rioXalt\in\riomX_\riomD$, then $\rioXsc$ is selected.
	
Given the 
%positive answer of the oracle, 
user's answer $u_s$, the selected query $X_s \in \setof{\rioXsc,\rioXalt}$ is added to $\rioTp$ or $\rioTn$ accordingly (see Chapter~\ref{chap:riobasics}). 
%the set of positive test cases $\rioTp$ or, otherwise, to the set of negative test cases $\rioTn$. 
In the last step of the main loop the algorithm updates the cautiousness value~$\riouc$ (function \textsc{updateCautiousness}) as described above.

Before the next query selection iteration starts, a stop condition test is performed. The algorithm evaluates whether the most probable diagnosis is at least $\sigma\%$ more likely than the second most probable diagnosis (\textsc{aboveThreshold}) or none of the leading diagnoses has been eliminated by the previous query, i.e.\ \textsc{getEliminationRate} returns zero for $X_s$. 
%In case that one of the stop conditions is fulfilled, 
If a stop condition is met, the presently most likely diagnosis is returned (\textsc{mostProbableDiag}).
%%%%%%%%%%%%%%%%%%%%%%%%%%%%%%%%%%%%%%%%%%%%%%%%%%%%

%%%%%%%%%%%%%%%%%%%%%%%%%%%%%%%%%%%%%%%%%%%%%%%%%%%%%%%%%%%%%%%%%%%%%%%%%%%%%%%%%%
%%%%%%%%%%%%%%%%%%%%%%%%%%%%%%%%%%%%%%%%%%%%%%%%%%%%%%%%%%%%%%%%%%%%%%%%%%%%%%%%%%
%%%%%%%%%%%%%%%%%%%%%%%%%%%%%%%%%%%%%%%%%%%%%%%%%%%%%%%%%%%%%%%%%%%%%%%%%%%%%%%%%%
%%%%%%%%%%%%%%%%%%%%%%%%%%%%%%%%%%%%%%%%%%%%%%%%%%%%%%%%%%%%%%%%%%%%%%%%%%%%%%%%%%
%%%%%%%%%%%%%%%%%%%%%%%%%%%%%%%%%%%%%%%%%%%%%%%%%%%%%%%%%%%%%%%%%%%%%%%%%%%%%%%%%%
%%%%%%%%%%%%%%%%%%%%%%%%%%%%%%%%%%%%%%%%%%%%%%%%%%%%%%%%%%%%%%%%%%%%%%%%%%%%%%%%%%
%%%%%%%%%%%%%%%%%%%%%%%%%%%%%%%%%%%%%%%%%%%%%%%%%%%%%%%%%%%%%%%%%%%%%%%%%%%%%%%%%%
%%%%%%%%%%%%%%%%%%%%%%%%%%%%%%%%%%%%%%%%%%%%%%%%%%%%%%%%%%%%%%%%%%%%%%%%%%%%%%%%%%
%%%%%%%%%%%%%%%%%%%%%%%%%%%%%%%%%%%%%%%%%%%%%%%%%%%%%%%%%%%%%%%%%%%%%%%%%%%%%%%%%%
%%%%%%%%%%%%%%%%%%%%%%%%%%%%%%%%%%%%%%%%%%%%%%%%%%%%%%%%%%%%%%%%%%%%%%%%%%%%%%%%%%

\chapter{Evaluation} 
\label{chap:rioeval}
\paragraph{Goals.}%\textbf{Goals.}
This evaluation should demonstrate that (1) there is a significant discrepancy between existing strategies SPL and ENT concerning user effort where the winner depends on the quality of meta information, (2) RIO
exhibits superior average behavior compared to ENT and SPL w.r.t.\ the amount of user interaction required, irrespective of the quality of specified fault information, (3) RIO scales well and (4) its reaction time is well suited for an interactive debugging approach.

%\noindent\textbf{Provenance of Test Data.} 
\paragraph{Provenance of Test Data.}
As data source for the evaluation we used faulty real-world ontologies produced by automatic ontology matching systems (cf.\ Example~\ref{example:rio_ex_1}). 
% in Section~\ref{sec:riobasics}). 
%Ontology Matching is defined as:
Matching of two ontologies $\riomo_i$ and $\riomo_j$ is understood as detection of correspondences between elements of these ontologies~\cite{Shvaiko2012}: 
\begin{definition}[Ontology matching]\label{def:rioalign}%~\cite{Shvaiko2012}
%Let $Q(\riomo_i)\subseteq \rioSig(\riomo_i)$ and $Q(\riomo_j)\subseteq \rioSig(\riomo_j)$ 
Let $Q(\riomo)\subseteq \rioSig(\riomo)$ denote the set of matchable elements in an ontology $\riomo$, where $\rioSig(\riomo)$ denotes the signature of $\riomo$. An ontology matching operation determines an \emph{alignment} $\rioAl_{ij}$, which is a set of correspondences between matched ontologies $\riomo_i$ and $\riomo_j$. Each \emph{correspondence} is a 4-tuple $\tuple{x_i,x_j,r,v}$, such that $x_i \in Q(\riomo_i)$, $x_j \in Q(\riomo_j)$, $r$ is a semantic relation and $v \in [0,1]$ is a confidence value. We call $\riomo_{i\rioAl j} := \riomo_i \cup \rioN(\rioAl_{ij}) \cup \riomo_j$ the \emph{aligned ontology} for $\riomo_i$ and $\riomo_j$ where $\rioN$ maps each correspondence to an axiom.
\end{definition}
Let in the following $Q(\riomo)$ be the restriction to atomic concepts and roles in $\rioSig(\riomo)$, $r \in \setof{\sqsubseteq, \sqsupseteq, \equiv}$ and $\rioN$ the natural alignment semantics~\cite{MeilickeStuck2009} that maps correspondences one-to-one to axioms of the form $x_i~r~x_j$.
We evaluate RIO using aligned ontologies by the following reasons: (1) Matching results often cause inconsistency/incoherence of ontologies. (2) The (fault) structure of different ontologies obtained through matching generally varies due to different authors and matching systems involved in the genesis of these ontologies. (3) For the same reasons, it is hard to estimate the quality of fault probabilities, i.e. it is unclear which of the existing query selection strategies to chose for best performance. (4) Available reference mappings can be used as correct solutions of the debugging procedure.

%\noindent\textbf{Test Datasets.} 
\paragraph{Test Datasets.}
We used two datasets D1 and D2: Each faulty aligned ontology $\riomo_{i\rioAl j}$ in D1 is the result of applying one of four ontology matching systems to a set of six independently created ontologies in the domain of conference organization. For a given pair of ontologies $\riomo_i\neq\riomo_j$, each system produced an alignment $\rioAl_{ij}$. The average size of $\riomo_{i\rioAl j}$ per matching system was between $312$ and $377$ axioms. D1 is a superset of the dataset used in \cite{Stuckenschmidt2008} for which all debugging systems under evaluation manifested correctness or scalability problems. D2, used to assess the scalability of RIO, is the set of ontologies
%\footnote{The raw data representing the output of matching systems was downloaded from http://bit.ly/Koh1NB. The reference alignment as well as the source ontologies Mouse and Human were downloaded from http://bit.ly/MU5Ca9.} 
from the ANATOMY track in the Ontology Alignment Evaluation Initiative\footnote{\url{http://oaei.ontologymatching.org}} (OAEI) 2011.5 \cite{Shvaiko2012}, which comprises two input ontologies $\riomo_1$ (11545 axioms) and $\riomo_2$ (4838 axioms). The size of the aligned ontologies generated by results of seven different matching systems was between 17530 and 17844 axioms.
\footnote{Source ontologies, produced alignments by each matcher, and reference alignments were downloaded from \url{http://bit.ly/Zffkow} (D1) and \url{http://bit.ly/Koh1NB} as well as \url{http://bit.ly/MU5Ca9} (D2).}
%Thus, D2 served to assess the scalability of RIO. 
%Note that the aligned ontologies output by five matching systems, i.e. CODI, CSA, MaasMtch, MapEVO and Aroma, could not be analyzed in the experiments. This was due to a coherent output produced by CODI and the problem that the reasoner was not able to find a model within acceptable time (2 hours) in the case of CSA, MaasMtch, MapEVO and Aroma. Similar reasoning problems were also reported in \cite{Ferrara2011}.

%\noindent\textbf{Reference Solutions.}
\paragraph{Reference Solutions.}
For the dataset D1, based on a manually produced reference alignment $\mathcal{R}_{ij} \subseteq \rioAl_{ij}$ for ontologies $\riomo_i,\riomo_j$ (cf. \cite{Meilicke2008a}), we were able to fix a target diagnosis $\riodt:=\rioN(\rioAl_{ij}\setminus\mathcal{R}_{ij})$ for each incoherent $\riomo_{i\rioAl j}$. In cases where 
%$\mathcal{R}_{ij}$ 
$\riodt$
represented a non-minimal diagnosis, 
it was randomly redefined 
as 
%the 
a
minimal diagnosis
%which was a subset of 
$\riodt \subset \rioN(\rioAl_{ij}\setminus \mathcal{R}_{ij})$.
%we defined $\riodt$ as the minimum cardinality diagnosis which was a subset of $\rioAl_{ij}\setminus \mathcal{R}_{ij}$. 
%
In case of D2, given the ontologies $\riomo_1$ and $\riomo_2$, the output $\rioAl_{12}$ of a matching system, and the correct reference alignment $\mathcal{R}_{12}$, we fixed $\riodt$ as follows: We carried out (prior to the actual experiment) a debugging session with DPI 
$\langle\rioN(\rioAl_{12}\setminus \mathcal{R}_{12})$, $\riomo_1\cup\riomo_2\cup \rioN(\rioAl_{12} \cap \mathcal{R}_{12})$, $\emptyset$, $\emptyset\rangle_{\setof{\text{coherence}}}$
and randomly chose one of the identified diagnoses as $\riodt$.
\paragraph{Test Settings.}
We conducted 4 experiments EXP-$i$ ($i=1,\dots,4$), the first two with dataset D1 and the other two with D2. In experiments 1 and 3 we simulated good fault probabilities by setting $p(\riotax_k) := 0.001$ for $\riotax_k \in \riomo_i \cup \riomo_j$ and $p(\riotax_m):= 1-v_m$ for $\riotax_m \in \rioAl_{ij}$, where $v_m$ is the confidence of the correspondence underlying $\riotax_m$. Unreasonable fault information was used in experiments 2 and 4. In EXP-4 the following probabilities were defined: $p(\riotax_k) := 0.01$ for $\riotax_k \in \riomo_i \cup \riomo_j$ and $p(\riotax_m):= 0.001$ for $\riotax_m \in \rioAl_{ij}$. In EXP-2, in contrast, we used probability settings of EXP-1, but altered the target diagnosis $\riodt$ in that we precomputed (before the actual experiment started) the 30 most probable minimal diagnoses, and from these we selected the diagnosis with the highest number of axioms $\riotax_k \in \riomo_{i\rioAl j} \setminus \rioN(\rioAl_{ij})$ as $\riodt$. 

Throughout all four experiments, we set $|\riomD|:=9$ (which proved to be a good trade-off between computation effort and representativeness of the leading diagnoses), $\sigma:=85\%$ and as input parameters for RIO we set $\riouc:= 0.25$ and $[\underline{c},\overline{c}] := [\riouc_{\min},\riouc_{\max}] = [0,\frac{4}{9}]$. To let tests constitute the highest challenge for the evaluated methods, the initial DPI was specified as $\tuple{\riomo_{i \rioAl j},\emptyset,\emptyset,\emptyset}_{\setof{\text{coherence}}}$, i.e. the entire search space was explored without adding parts of $\riomo_{i \rioAl j}$ to 
%the background knowledge
$\riomb$, although 
%the target diagnosis 
$\riodt$ was always a subset of the alignment $\rioAl_{ij}$ only. In practice, given such prior knowledge, the search space could be severely restricted and debugging greatly accelerated.
All tests were executed on a Core-i7 (3930K) 3.2Ghz, 32GB RAM with Ubuntu Server 11.04 and Java 6 installed.%installed.
\footnote{See \url{http://code.google.com/p/rmbd/wiki} for code and details.} 
%The number $|\riomD|$ of leading diagnoses was set to 9 and $\sigma:=85\%$. As input parameters for RIO we set $\riouc:= 0.25$ and $[\underline{c},\overline{c}] := [\riouc_{\min},\riouc_{\max}] = [0,\frac{4}{9}]$. For the tests we considered the most general setting, i.e. $\riodt \subset \riomo_{i \rioAl j}$. 

%\noindent\textbf{Metrics.} 
\paragraph{Metrics.}
Each experiment involved a debugging session of ENT, SPL as well as RIO for each ontology in the respective dataset. In each debugging run we measured the number of required queries ($q$) until $\riodt$ was identified, the overall debugging time (\textit{debug}) assuming that queries are answered instantaneously and the reaction time (\textit{react}), i.e. the average time between two successive queries. The queries generated in the tests were answered by an automatic oracle by means of the target ontology $\riomo_{i\rioAl j}\setminus \riodt$.

\begin{figure}
	\centering
		\includegraphics[width=0.5\textwidth, trim=0mm 0mm 0mm 0mm]{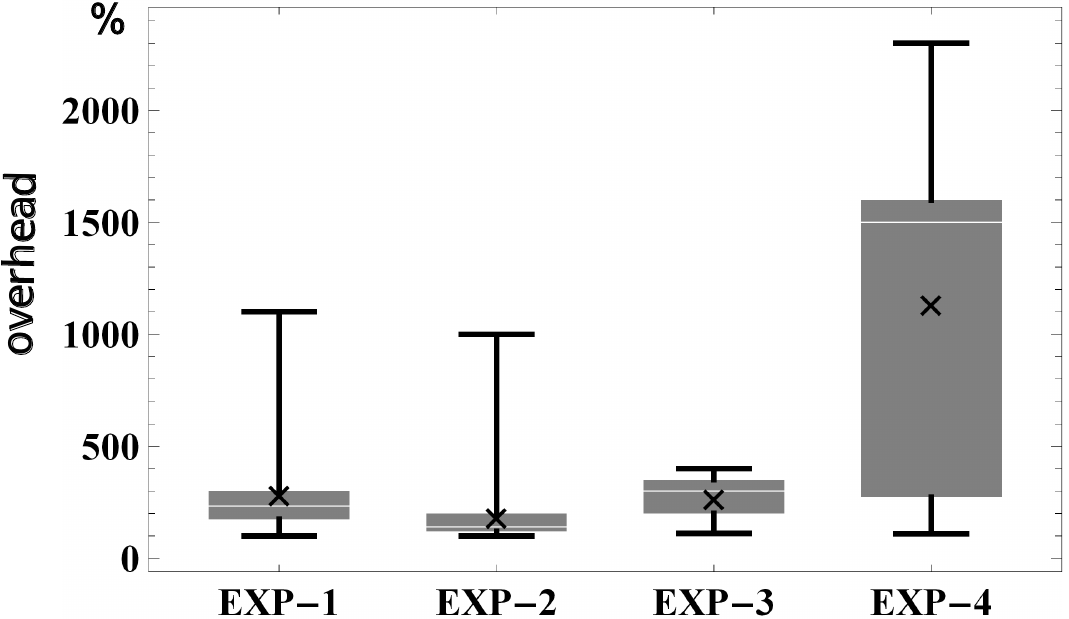}
    \caption[Box-Whisker Plots Illustrating the Performance Discrepancy between Better and Worse Query Selection Strategy]{Box-Whisker Plots presenting the distribution of overhead $(q_w-q_b)/q_b*100$ (in \%) per debugging session of the worse strategy $q_w := \max(q_\text{SPL},q_\text{ENT})$ compared to the better strategy $q_b := \min(q_\text{SPL},q_\text{ENT})$. Mean values are depicted by a cross.}
		\label{fig:riodiff_SPL_ENT}
\end{figure}

%\noindent\textbf{Observations.}
\paragraph{Observations.} 
The difference w.r.t.\ the number of queries per test run between the better and the worse strategy in \{SPL,ENT\} was absolutely significant, with a maximum of 2300\% in EXP-4 and averages of 190\% to 1145\% throughout all four experiments
(Figure~\ref{fig:riodiff_SPL_ENT}). 
Moreover, results show that varying quality of fault probabilities in \{\text{EXP-1},\text{EXP-3}\} compared to \{\text{EXP-2},\text{EXP-4}\} clearly affected the performance of ENT and SPL (see first two rows in 
Figure~\ref{tab:riobest_strategies}). This perfectly motivates the application of RIO.

\begin{table}[tb]
\small
\centering
\begin{tabular}{@{\extracolsep{-1pt}} l|c|c|c||c|c|c||c|c|c||c|c|c}
		&  \multicolumn{3}{c||}{EXP-1}								  &  \multicolumn{3}{c||}{EXP-2}										  														     &  \multicolumn{3}{c||}{EXP-3}					         &  \multicolumn{3}{c}{EXP-4} \\ 
		\hline
 		& \textit{debug}					& \textit{react} 			& $q$ 					& 	\textit{debug} 				& \textit{react} 			& $q$ 					& \textit{debug} 				& \textit{react} 				& $q$ 					&  \textit{debug}				&  	\textit{react}				& $q$	\\\hline
ENT &  1860					& 262 				& 3.67 					&  	1423					& 204 				& 5.26					& \textbf{60928}& 12367					& 5.86 					& 74463 				& 5629 					& 11.86	\\
SPL &  \textbf{1427}& \textbf{159}& 5.70 					&  	\textbf{1237}	& \textbf{148}& 5.44					& 104910				&	\textbf{4786}	& 19.43 				& 98647 				& \textbf{4781}	& 18.29	\\
RIO &  1592					& 286 				& \textbf{3.00}	&  	1749					&	245  				& \textbf{4.37}	& 62289					&	12825					& \textbf{5.43}	& \textbf{66895}& 8327 					& \textbf{8.14}	\\ \hline
\end{tabular}
\vspace{5pt}
\caption[Average Time for Debugging Session, between Two Successive Queries and Average Number of Queries Required by Each Strategy]{Average time (ms) for the entire debugging session (\textit{debug}), average time (ms) between two successive queries (\textit{react}), and average number of queries ($q$) required by each strategy.} 
\label{tab:riotime_query}
\end{table}

\begin{figure}[ht]
\centering
\subfigure[]{
    %\vspace{-15pt}
    \includegraphics[width=0.46\textwidth,trim=7mm -12mm 1mm 5mm]{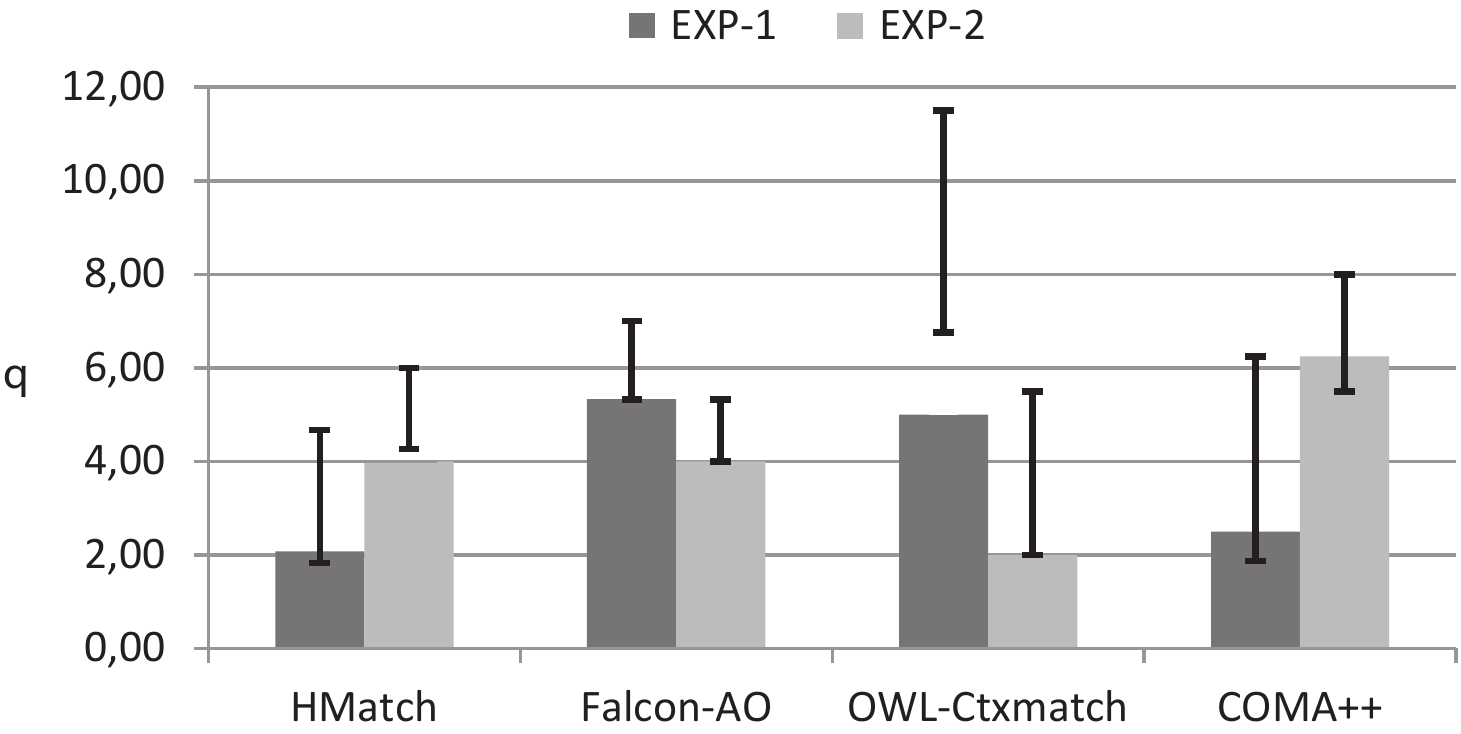}
		\label{fig:rioquery_conference}
}
\subfigure[]{
    %\rule{4cm}{3cm}
    \includegraphics[width=0.46\textwidth,trim=5mm 7mm 1mm 10mm]{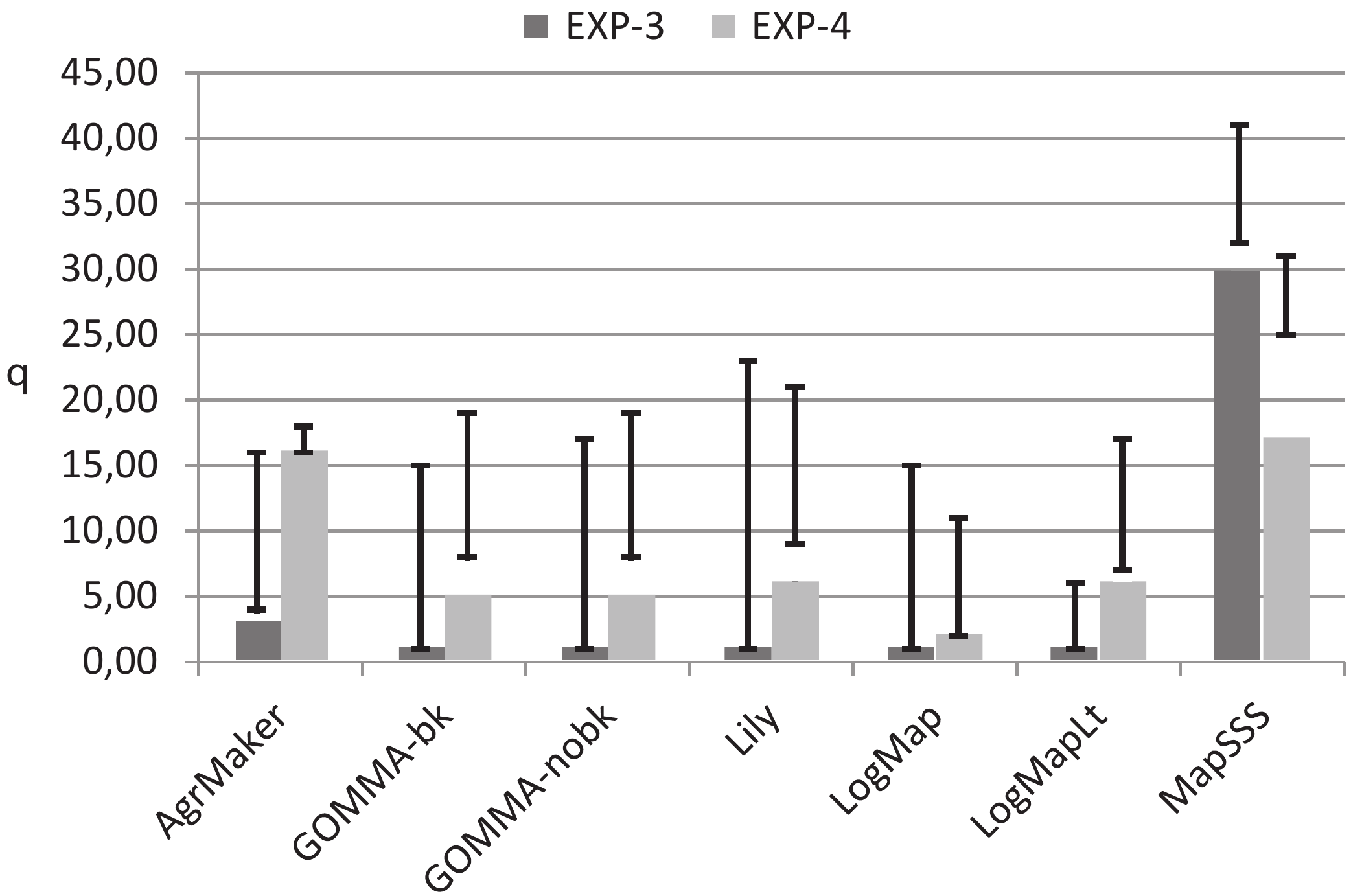}
    \label{fig:rioquery_anatomy}
}
%\vspace{-9pt}
\caption[Average Number of Queries Required by RIO Compared to Other Query Selection Approaches]{The bars show the avg. number of queries ($q$) needed by RIO, grouped by matching tools. The distance from the bar to the lower (upper) end of the whisker indicates the avg. difference of RIO to the queries needed by the per-session better (worse) strategy of SPL and ENT, respectively.}
\label{fig:rioqueries}
\vspace{10pt}
\end{figure}

\begin{table}
\small
\centering
\begin{tabular}[b]{@{\extracolsep{0pt}} l|c|c||c|c} 
																												& EXP-1		& EXP-2 	& EXP-3			& EXP-4			\\	\hline
			$q_\text{SPL} < q_\text{ENT}$ 										& 11\%	  & 37\% 		& 	0\%			&   29\%		\\
			$q_\text{ENT} < q_\text{SPL}$ 										& 81\%    &	56\% 	  & 	100\%		&		71\%		\\
			$q_\text{SPL} = q_\text{ENT}$											& 7\%     &	7\% 	  & 	0\%			&		0\%			\\
			$q_\text{RIO} < \min$															& 4\%	  	& 26\% 	  & 	29\%		&		71\%		\\
			$q_\text{RIO} \leq \min$															& 74\%		& 74\% 	  & 	100\%		&		100\%		\\	\hline 
				\multicolumn{5}{c}{}	\\																							
\end{tabular}
\caption[Percentage Rates Indicating How Often Which Query Selection Strategy Performed Best]{Percentage rates indicating which strategy performed best/better w.r.t.\ the required user interaction, i.e. number of queries. EXP-1 and EXP-2 involved 27, EXP-3 and EXP-4 seven debugging sessions each. $q_{str}$ denotes the number of queries needed by strategy $str$ and $\min$ is an abbreviation for $\min(q_\text{SPL},q_\text{ENT})$.}
\label{tab:riobest_strategies}
\end{table}

Results of both experimental sessions, $\langle\text{EXP-1,EXP-2}\rangle$ and $\langle\text{EXP-3,EXP-4}\rangle$, are summarized in Figures~\ref{fig:rioquery_conference} and \ref{fig:rioquery_anatomy}, respectively. 
The figures show the (average) number of queries asked by RIO and the (average) differences to the number of queries needed by the per-session better and worse strategy in \{SPL,ENT\}, respectively. 
The results illustrate clearly that the average performance achieved by RIO was always substantially closer to the better than to the worse strategy. In both EXP-1 and EXP-2, throughout 74\% of 27 debugging sessions, RIO worked as efficiently as the best strategy (Figure \ref{tab:riobest_strategies}).  
In 26\% of the cases in EXP-2, RIO even outperformed both other strategies; in these cases, RIO could save more than 20\% of user interaction on average compared to the best other strategy. In one scenario in EXP-1, it took ENT 31 and SPL 13 queries to finish, whereas RIO required only 6 queries, which amounts to an improvement of more than 80\% and 53\%, respectively. In $\langle\text{EXP-3,EXP-4}\rangle$, the savings achieved by RIO were even more substantial. RIO manifested superior behavior to both other strategies in 29\% and 71\% of cases, respectively. Not less remarkable, in 100\% of the tests in EXP-3 and EXP-4, RIO was at least as efficient as the best other strategy. Recalling Figure~\ref{fig:riodiff_SPL_ENT}, this means that RIO can avoid query overheads of over 2000\%.
Table~\ref{tab:riotime_query}, which provides average values for $q$, \textit{react} and \textit{debug} per strategy,
demonstrates that RIO is the best choice in all experiments w.r.t.\ $q$. Consequently, RIO is suitable for both good 
and poor meta information.

As to time aspects, RIO manifested good performance, too. Since times consumed in $\langle\text{EXP-1,EXP-2}\rangle$ are almost negligible, consider the more meaningful results obtained in $\langle\text{EXP-3,EXP-4}\rangle$. While the best reaction time in both experiments was achieved by SPL, we can clearly see that SPL was significantly inferior to both ENT and RIO concerning $q$ and \textit{debug}. RIO revealed the best debugging time in EXP-4, and needed only $2.2\%$ more time than the best strategy (ENT) in EXP-3. However, if we assume the user being capable of reading and answering a query in, e.g., 30 sec on average, which is already quite fast, then the overall time savings of RIO compared to ENT in EXP-3 would already account for $5\%$. Doing the same thought experiment for EXP-4, RIO would save $25\%$ (w.r.t.\ ENT) and $50\%$ (w.r.t.\ SPL) of debugging time on average. All in all, the measured times confirm that RIO is well suited for \emph{interactive} debugging.

\chapter{Related Work}
\label{chap:rioRelatedWork}

A similar interactive technique was presented in~\cite{Nikitina2011}, where a user is successively asked single ontology axioms in order to 
%to assign each axiom 
obtain a partition
of a given ontology 
%either to 
into a set of desired and a set of undesired consequences. However, given an inconsistent/incoherent ontology, this technique starts from an empty set of desired consequences aiming at adding to this set only axioms which preserve coherence, whereas our approach starts from the complete ontology aiming at finding a minimal set of axioms responsible for the violation of pre-specified requirements.    
%
%A similar interactive technique was presented in~\cite{Nikitina2011}, where a user is successively asked single ontology axioms in order to 
%%to assign each axiom 
%obtain a partition
%of a given ontology 
%%either to 
%into a set of desired and a set of undesired consequences. However, given an inconsistent ontology, this technique starts from an empty ontology aiming at adding only axioms which preserve consistence to the desired consequences, whereas our approach starts from the complete ontology aiming at finding a minimal set of axioms responsible for the violation of pre-specified requirements.    
%
%A similar interactive technique can be found in~\cite{Nikitina2011}, where queries to a user are incorporated to revise an ontology. However, ontology revision must be strictly differentiated from ontology debugging. Ontology revision aims at partitioning a given ontology into a set of correct axioms and a set of incorrect ones. The system~\cite{Nikitina2011} can deal with inconsistent/incoherent ontologies only after a union of all axioms causing these problems is identified and added to the initial set of incorrect axioms. Computation of these axioms, however, requires some ontology debugging mechanism.

An approach for alignment debugging was proposed in \cite{meilicke2011}. This work describes approximate algorithms for computing a ``local optimal diagnosis'' and complete methods to discover a ``global optimal diagnosis''. Optimality in this context refers to the maximum sum of confidences in the resulting coherent alignment. In contrast to our framework, diagnoses are determined automatically without support for user interaction. Instead, techniques for manual revision of the alignment as a procedure \emph{independent} from debugging are demonstrated. 
%
%
%Note that the comparison of RIO with techniques integrated in ontology matching systems such as CODI~\cite{noessner2010} or LogMap~\cite{Jimenez-Ruiz2011} is inappropriate, since all these systems use greedy diagnosis techniques (e.g.~\cite{MeilickeStuck2009}), whereas the method presented in this paper is complete.

\chapter{Summary and Conclusions}
\label{chap:rioconclusion}

We have shown problems of state-of-the-art interactive ontology debugging strategies w.r.t.\ the usage of unreliable meta information. To tackle this issue, we proposed a learning strategy which combines the benefits of existing approaches, i.e.\ high potential and low risk. Depending on the performance of the diagnosis discrimination actions, the trust in the a-priori information is adapted. Tested under various conditions, our algorithm 
%required minimal amount of user interaction on average in all experiments
revealed good scalability and reaction time as well as superior average performance to two common approaches in the field in all tested cases w.r.t.\ required user interaction. Highest achieved savings amounted to more than 80\% and user interaction overheads resulting from the wrong choice of strategy of up to 2300\% could be saved. In the hardest test cases, the new strategy was not only on average, but in 100\% of the test cases at least as good as the best other strategy.

%%%%%%%%%%%%%%%
\part{A Direct Approach to Sequential Diagnosis of High Cardinality Faults in Knowledge Bases}
\label{part:DX}
In this part we cover the topic of efficiently dealing with KB debugging problems involving high cardinality faults. This part relies on material \cite{Shchekotykhin2014, Shchekotykhin2014a, Shchekotykhin2014b} published in the \emph{Proceedings of the 21st European Conference on Artificial Intelligence (ECAI 2014)}, in \emph{DX 2014 - 25th International Workshop on Principles of Diagnosis} and in the \emph{Proceedings of the Third International Workshop on Debugging Ontologies and Ontology Mappings (WoDOOM14)}, respectively.\footnote{We are glad to report that the publication \cite{Shchekotykhin2014a} was awarded the Best Paper Award at the DX Workshop that took place in Graz, Austria in September 2014 (see \url{http://dx-2014.ist.tugraz.at}).}
\chapter{Introduction to the Problem}
Model-based diagnosis (MBD)~\cite{Reiter87} is a general method which can be used to find errors in hardware, software, knowledge-bases (KBs), orchestrated web-services, configurations, etc. In particular, ontology (KB) debugging tools~\cite{Kalyanpur.Just.ISWC07,friedrich2005gdm,Horridge2008} can localize a (potential) fault by finding sets of axioms $\dxmd \subseteq \dxmo$ called \emph{diagnoses} for the KB $\dxmo$. Diagnoses are generated using minimal conflict sets, i.e.\ irreducible sets of axioms $CS \subseteq \dxmo$ that violate some requirements, by using a consistency checker (black-box approach).
At least all axioms of a minimal diagnosis must be modified or deleted in order to formulate a fault-free knowledge-base $\dxot$. A knowledge-base $\dxmo$ is faulty if some requirements, such as consistency of $\dxmo$, presence or absence of specific entailments, are violated.

Sequential MBD methods~\cite{dekleer1987} applied to KB debugging acquire additional information in order to discriminate between diagnoses~\cite{Shchekotykhin2012}. 
Generated queries are answered by some oracle providing additional observations about the entailments of a valid KB. 
As various applications show, the standard methods work very satisfactorily for cases where the number of faults (minimal conflict sets) is low (single digit number), consistency checking is fast (single digit number of seconds), and sufficient possibilities for observations are available.% for discriminating between diagnoses.  

%All the discrimination and diagnosis approaches listed above follow the standard model-based diagnosis technique~\cite{Reiter87} and 
%Furthermore, diagnoses are ordered and filtered by some preference criteria, e.g.\ probability or cardinality, in order to focus debugging on the most likely cases. 

%In the common KB development scenario where a user develops an ontology manually, the changes between validation steps are rather small. Therefore, the number of faulty axioms is in a range where standard sequential model-based methods are applicable~\cite{Shchekotykhin2012}. 

However, there are situations when KBs comprise a large number of faults. For example, in ontology matching scenarios two KBs with several thousands of axioms are merged into a single one. High quality matchers (e.g.\ \cite{Jimenez-Ruiz2011}) require the diagnosis of such substantially extended KBs, but could not apply standard diagnosis methods because of the large number of minimal  diagnoses and their high cardinality. E.g.\ there are cases when the minimum cardinality of diagnoses is greater than 20. 

%??? Do we need this $\Rightarrow$ Moreover, excessive use of calls to a theorem prover for generating minimal conflicts creates additional runtime problems. 

In order to deal with hard diagnosis instances, we propose to relax the requirement for sequential diagnosis to compute a set of \textit{preferred} minimal diagnoses, such as a set of most probable diagnoses.
% minimum cardinality diagnoses or most probable diagnoses. %in order to locate the faults. 
Instead, we compute just \textit{some} set of minimal diagnoses which can be used for query generation. % which discriminate between diagnoses.
This allows to use direct computation of diagnoses~\cite{satoh2006} \textit{without} computing conflict sets. The direct approach was applied for non-interactive diagnosis of ontologies~\cite{Du2011,Baader2012} and constraints~\cite{Felfernig2011}. 
%
%The computation of a minimal diagnosis by a variant of \textsc{QuickXPlain}~\cite{junker04} requires $O(|\dxmd|\log(\frac{|\dxoo|}{|\dxmd|}))$ consistency checks, where $|\dxmd|$ is the cardinality of the minimal diagnosis and $|\dxoo|$ the size of the knowledge base. If $m$ minimal diagnoses are required for query generation, then only $m$ calls to a direct diagnosis generator are needed.
%
A recent approach~\cite{Stern2012} does not generate the standard \textsc{HS-Tree}, but still depends on the minimization of conflict sets, i.e.\ $|\dxmd|$ minimized conflicts have to be discovered. Consequently, if $|\dxmd| \gg m$, substantially more consistency checks are required, where $|\dxmd|$ is the cardinality of the minimal diagnosis and $m$ is the number of minimal diagnoses required for query generation.

Since we are replacing the set of most probable diagnoses by just a set of minimal diagnoses, some important practical questions have to be addressed. (1) Is a substantial number of additional queries needed, (2) is this approach able to locate the faults, and (3) how efficient is this approach? 

In order to answer these questions we have exploited the most difficult diagnosis problems of the ontology alignment competition~\cite{Ferrara2011}. 
Our evaluation shows that sequential diagnosis by direct diagnosis generation needs approximately the same number of queries ($\pm 1$) in order to identify the faults. This evaluation was carried out for cases where the standard sequential diagnosis method was applicable. Furthermore, the evaluation shows that our proposed method is able to locate faults in all cases correctly, particularly in those cases where debugging sessions by means of the standard method are not successful (due to overwhelming time or space consumption). %Moreover, the computation costs which are introduced in addition to the computational costs of theorem proving are less than 7\% of the total runtime.,\prnote{The last sentence is simply not true...only for one single case!! (I already mentioned that last time...) If somebody spends two minutes to think on this sentence they will find out it is not true. I would not write it like that.} 
Moreover, for the hardest cases (i.e., more than 4 minutes overall debugging time), the additional computation costs introduced by the direct method apart from the costs needed for theorem proving are less than 50\%, i.e. reasoning costs amount to more than two thirds of overall computation time.

The rest of Part~\ref{part:DX} is organized as follows: Chapter~\ref{chap:dx:diag} gives a brief introduction to the main notions of sequential KB diagnosis. The details of the suggested algorithms are presented in Chapter~\ref{chap:dx:details}. In Chapter~\ref{chap:dx:eval} we provide evaluation results whereupon Chapter~\ref{chap:dx:conc} gives a conclusion.

%%%%%%%%%%%%%%%%%%%%%%%%%%%%%%%%%%%%%%%%%%%%%%%%%%%%%%%%%%%%%%%%%%%%%%%%%%%%%%%%%%%%%%%%%%%%%%%%%
%%%%%%%%%%%%%%%%%%%%%%%%%%%%%%%%%%%%%%%%%%%%%%%%%%%%%%%%%%%%%%%%%%%%%%%%%%%%%%%%%%%%%%%%%%%%%%%%%
%%%%%%%%%%%%%%%%%%%%%%%%%%%%%%%%%%%%%%%%%%%%%%%%%%%%%%%%%%%%%%%%%%%%%%%%%%%%%%%%%%%%%%%%%%%%%%%%%
%%%%%%%%%%%%%%%%%%%%%%%%%%%%%%%%%%%%%%%%%%%%%%%%%%%%%%%%%%%%%%%%%%%%%%%%%%%%%%%%%%%%%%%%%%%%%%%%%
%%%%%%%%%%%%%%%%%%%%%%%%%%%%%%%%%%%%%%%%%%%%%%%%%%%%%%%%%%%%%%%%%%%%%%%%%%%%%%%%%%%%%%%%%%%%%%%%%
%%%%%%%%%%%%%%%%%%%%%%%%%%%%%%%%%%%%%%%%%%%%%%%%%%%%%%%%%%%%%%%%%%%%%%%%%%%%%%%%%%%%%%%%%%%%%%%%%
%%%%%%%%%%%%%%%%%%%%%%%%%%%%%%%%%%%%%%%%%%%%%%%%%%%%%%%%%%%%%%%%%%%%%%%%%%%%%%%%%%%%%%%%%%%%%%%%%
%%%%%%%%%%%%%%%%%%%%%%%%%%%%%%%%%%%%%%%%%%%%%%%%%%%%%%%%%%%%%%%%%%%%%%%%%%%%%%%%%%%%%%%%%%%%%%%%%
%%%%%%%%%%%%%%%%%%%%%%%%%%%%%%%%%%%%%%%%%%%%%%%%%%%%%%%%%%%%%%%%%%%%%%%%%%%%%%%%%%%%%%%%%%%%%%%%%
%%%%%%%%%%%%%%%%%%%%%%%%%%%%%%%%%%%%%%%%%%%%%%%%%%%%%%%%%%%%%%%%%%%%%%%%%%%%%%%%%%%%%%%%%%%%%%%%%
%%%%%%%%%%%%%%%%%%%%%%%%%%%%%%%%%%%%%%%%%%%%%%%%%%%%%%%%%%%%%%%%%%%%%%%%%%%%%%%%%%%%%%%%%%%%%%%%%

\chapter{Basic Concepts}\label{chap:dx:diag}

In the following we present (1) the fundamental concepts regarding the diagnosis of KBs and (2) the interactive localization of axioms which must be changed. 

\paragraph{Diagnosis of KBs.} 
Given a knowledge-base $\dxoo$ which is a set of logical sentences (axioms), the user can specify particular requirements during the knowledge-engineering process. The most basic requirement is satisfiability, i.e.\ a logical model exists. A further frequently employed requirement is coherence. Coherence requires that there exists a model s.t.\ the interpretation of every unary predicate is non-empty. In other words, if we add $\exists Y\, a(Y)$ to $\dxoo$ for every unary predicate $a$, then the resulting KB must be satisfiable. 
%Without limiting the generality of our approach, we implicitly assume that coherence is always required. 
%Note, description logic (DL) theorem prover  usually provide a coherence check. Our method assumes a monotonic semantic of the applied logic and therefore is applicable to description logic and to predicate logic such as Datalog +/- \cite{Datalog+-}. In DL unary predicates are called atomic concepts and binary predicates are roles. 
In addition, as it is common practice in software engineering, the knowledge-engineer (user for short) may specify test cases. Test cases are axioms which must (not) be entailed by a valid KB. 

\begin{definition}~\label{def:valid}
Given a set of axioms $\dxTe$ (called positive test cases) and a set of axioms $\dxTne$ (called negative test cases), a knowledge-base $\dxot$ is \emph{valid} iff it fulfills the following requirements:

\begin{enumerate}
\item  $\dxot$ is satisfiable (and coherent if required)
	\item $\dxot \models \dxte\quad \forall \dxte \in \dxTe$
    \item $\dxot \not\models \dxtne\quad \forall \dxtne \in \dxTne$   
\end{enumerate} 
\end{definition}

Let us assume that there is a non-valid KB $\dxoo$, then a set of axioms $\dxmd \subseteq \dxmo$ must be removed and possibly some axioms $EX$ must be added by the user s.t.\ an updated $\dxot$ becomes valid, i.e.\ $\dxot:=(\dxoo \setminus \dxmd) \cup EX$. The goal of diagnosis is to provide information to the users which are the sets of axioms $\dxmd$ (which is called a diagnosis) that must be changed. In order to prevent unnecessary changes, $\dxmd$ is often required to be subset-minimal, i.e.\ the set should be as small as possible. Furthermore, we allow the user to define a set of axioms $\dxmb$ (called the background theory) which must not be changed (i.e.\ the correct axioms). More formally:
  
\begin{definition}\label{def:diag}
Given a diagnosis problem instance (DPI) specified by $\tuple{\dxoo, \dxmb,\dxTe,\dxTne}$ where 
\begin{itemize}
	\item $\dxoo$ is a knowledge-base,
	\item $\dxmb$ a background theory,
	\item $\dxTe$ a set of axioms which must be implied by a valid knowledge-base $\dxot$ and
	\item $\dxTne$ a set of axioms, each of which must \emph{not} be implied by $\dxot$
\end{itemize}
$\dxmd\subseteq\dxoo$ is a \emph{diagnosis} w.r.t.\ $\tuple{\dxoo, \dxmb,\dxTe,\dxTne}$ iff $\dxoo \setminus \dxmd$ can be extended by a set of logical sentences $EX$ such that:
\begin{enumerate}
    \item $(\dxoo \setminus \dxmd) \cup \dxmb \cup EX$ is consistent
	\item $(\dxoo \setminus \dxmd) \cup \dxmb \cup EX \models \dxte$ for all $\dxte \in \dxTe$
    \item $(\dxoo \setminus \dxmd) \cup \dxmb \cup EX \not\models \dxtne$ for all $\dxtne \in \dxTne$
\end{enumerate}
$\dxmd$ is a \emph{minimal diagnosis} iff there is no %proper subset of axioms 
$\dxmd^\prime \subset \dxmd$ such that $\dxmd^\prime$ is a diagnosis. $\dxmd$ is a \emph{minimum cardinality diagnosis} iff there is no diagnosis $\dxmd^\prime$ such that $|\dxmd^\prime|<|\dxmd|$.\footnote{If clear from the context, we will often call $\dxmd$ simply a diagnosis without explicitly stating the DPI w.r.t.\ which it is a diagnosis in the rest of Part~\ref{part:DX}.} 
\end{definition}

The following proposition of~\cite{Shchekotykhin2012} characterizes diagnoses by replacing $EX$ with the positive test cases. 
\begin{corollary}~\label{cor:diag}
Given a DPI $\tuple{\dxoo, \dxmb,\dxTe,\dxTne}$, a set of axioms $\dxmd \subseteq \dxoo$ is a diagnosis w.r.t.\ $\tuple{\dxoo,\dxmb,\dxTe,\dxTne}$ iff 
\begin{equation*}
(\dxoo \setminus \dxmd) \cup \dxmb \cup \{\bigwedge_{\dxte \in \dxTe} \dxte \}
\end{equation*}
is satisfiable (coherent) and 
\begin{equation*}
\forall \dxtne \in \dxTne\;:\; (\dxoo \setminus \dxmd) \cup \dxmb \cup \{ \bigwedge_{\dxte \in \dxTe} \dxte \} \not\models \dxtne
\end{equation*}
\end{corollary}

Hereafter we assume that a diagnosis always exists. 
\begin{proposition}\label{pro:diagnosable}
A diagnosis $\dxmd$ w.r.t.\ a DPI $\tuple{\dxoo,\dxmb,\dxTe,\dxTne}$ exists iff $
\dxmb \cup \{ \bigwedge_{\dxte \in \dxTe}\dxte \}$
is consistent (coherent) and 
$
\forall \dxtne \in \dxTne \;:\;  \dxmb \cup \{ \bigwedge_{\dxte \in \dxTe}\dxte \} \not\models \dxtne
$
\end{proposition}

For the computation of diagnoses \emph{conflict sets} are usually employed to constrain the search space.  A conflict set is the part of the KB that preserves the inconsistency/incoherency.
\begin{definition}
Given a DPI $\tuple{\dxoo,\dxmb,\dxTe,\dxTne}$, a set of axioms $CS \subseteq \dxoo$ is a \emph{conflict set} w.r.t.\ $\tuple{\dxoo,\dxmb,\dxTe,\dxTne}$ iff $CS \cup \dxmb \cup \{\bigwedge_{\dxte \in \dxTe} \dxte\}$ is inconsistent (incoherent) or there is an $\dxtne \in \dxTne$ such that $CS \cup \dxmb \cup \{\bigwedge_{\dxte \in \dxTe} \dxte\} \models \dxtne$.
$CS$ is minimal iff there is no $CS^\prime \subset CS$ such that $CS^\prime$ is a conflict set.\footnote{If clear from the context, we will often call $CS$ simply a conflict set without explicitly stating the DPI w.r.t.\ which it is a conflict set in the rest of Part~\ref{part:DX}.}
\end{definition}

Minimal conflict sets can be used to compute the set of minimal diagnoses as it is shown in~\cite{Reiter87}. The idea is that each diagnosis must include at least one element of each minimal conflict set.\footnote{In the rest of Part~\ref{part:DX}, we consider only minimal conflict sets to avoid the issues concerning the pruning rule~\cite{Reiter87} described in~\cite{greiner1989correction}.}

\begin{proposition}\label{prop:hittingset}
$\dxmd$ is a (minimal) diagnosis w.r.t.\ the DPI $\tuple{\dxoo, \dxmb,\dxTe,\dxTne}$ iff $\dxmd$ is a (minimal) hitting set for the set of all minimal conflict sets w.r.t.\ $\tuple{\dxoo, \dxmb,\dxTe,\dxTne}$.
\end{proposition}

For the generation of a minimal conflict set, diagnosis systems use a divide-and-conquer method (e.g.\ \textsc{QuickXPlain}~\cite{junker04}, for short \textsc{QX}), which we discussed in Sections~\ref{sec:cs_comp} and \ref{sec:cs_comp_correctness}. In the worst case, \textsc{QX} requires $O(|CS|\log(\frac{|\dxoo|}{|CS|}))$ calls to the reasoner, where $CS$ is the returned minimal conflict set. 

The computation of minimal diagnoses in KB debugging systems is implemented using Reiter's Hitting Set \textsc{HS-Tree} algorithm~\cite{Reiter87} (cf.\ Algorithm~\ref{algo:hs} in Chapter~\ref{chap:DiagnosisComputation}). The algorithm constructs a directed  tree from the root to the leaves, where each non-leave node is labeled with a minimal conflict set and leave nodes are labeled by $\checkmark$ (\emph{no conflicts}) or $\times$ (\emph{pruned}). 

Each ($\checkmark$) node corresponds to a minimal diagnosis. The  minimality of the diagnoses is guaranteed by the minimality of conflict sets used for labeling the %non-leave 
nodes, the pruning rule and the breadth-first strategy of the tree generation. Moreover, because of the breadth-first strategy the minimal diagnoses are generated in increasing order of their cardinality. Under the assumption that diagnoses with lower cardinality are more probable than those with higher cardinality, \textsc{HS-Tree} generates most probable minimal diagnoses first.

\paragraph{Diagnoses Discrimination.} For many real-world DPIs, a diagnosis system can return a large number of (minimal) diagnoses.
Each minimal diagnosis corresponds to a different set of axioms in the given KB $\dxmo$. All the axioms of any minimal diagnosis might be deleted from $\dxmo$ or changed accordingly in order to formulate a valid $\dxot$. The user may extend the test cases $P$ and $N$ such that diagnoses are eliminated, thus identifying exactly the correct minimal diagnosis. For discriminating between minimal diagnoses we assume that the user knows some of the sentences a valid $\dxot$ must (not) entail, that is the user serves as an oracle. 

%Ideally we want to query the oracle as less as possible in order to identify the set of axioms to be changed. 

%The user has to decide which minimal diagnosis is exploited (i.e.\ which axioms are changed or removed) in order to formulate an ontology that corresponds to her intention. Note, the ontology which corresponds to the user intention is only observable through the communication with the user.  Interactive ontology debugging discriminates between minimal diagnoses by querying the user where the number of queries should be minimized. We assume that the user (in the role of an oracle) can be asked whether or not a query $Q$ is entailed by $\dxot$. Given a set of diagnoses $\dxmD$ answers to queries are exploited to eliminate diagnoses. 
 
\begin{property}\label{prop:dpiUpdate}
Given a DPI $\tuple{\dxoo, \dxmb,\dxTe,\dxTne}$, a set of diagnoses $\dxmD$ w.r.t.\ $\tuple{\dxoo, \dxmb,\dxTe,\dxTne}$, and a logical sentence $\dxqry$
representing the oracle query $\dxot \models \dxqry\;$. 
If the oracle gives the answer \emph{yes} then $\dxmd_i \in \dxmD$ is a diagnosis w.r.t.\ $\tuple{\dxoo, \dxmb,\dxTe \cup
\{\dxqry\},\dxTne}$ iff both conditions hold:
\begin{align*}
(\dxoo \setminus \dxmd_i)& \cup \dxmb \cup \{\bigwedge_{\dxte \in \dxTe} \dxte\} \cup \{\dxqry\} \; \emph{is consistent}\\
\forall \dxtne \in \dxTne & \;:\;(\dxoo \setminus \dxmd_i)  \cup \dxmb \cup \{\bigwedge_{\dxte \in \dxTe} \dxte\} \cup \{\dxqry\} \not\models \dxtne
\end{align*}

If the oracle gives the answer \emph{no} then $\dxmd_i \in \dxmD$ is a diagnosis w.r.t.\ $\tuple{\dxoo, \dxmb,\dxTe,\dxTne \cup \{\dxqry\}}$ iff both conditions hold:
\begin{align*}
(\dxoo \setminus \dxmd_i)& \cup \dxmb \cup \{\bigwedge_{\dxte \in
\dxTe} \dxte\}\; \emph{is consistent} \\
\forall \dxtne \in (\dxTne \cup
\{\dxqry\}) & \;:\; (\dxoo \setminus \dxmd_i) \cup \dxmb \cup \{\bigwedge_{\dxte \in
\dxTe} \dxte\} \not\models \dxtne
\end{align*}
\end{property}

However, many different queries might exist for some set of diagnoses $|\dxmD| \geq 2$, in the extreme case exponentially many (in $|\dxmD|$). To select the best query, the authors in~\cite{Shchekotykhin2012} suggest two query selection strategies: \textsc{split-in-half} ($\mathsf{SPL}$) and \textsc{entropy} ($\mathsf{ENT}$). The first strategy is a greedy approach preferring queries which allow to remove half of the diagnoses in $\dxmD$, for both answers to the query. The second is an information-theoretic measure, which estimates the information gain for both outcomes of each query and returns the query that maximizes the expected information gain. The \emph{prior fault probabilities} required for evaluating the $\mathsf{ENT}$ measure can be obtained from statistics of previous diagnosis sessions. For instance, if the user has problems to apply ``$\exists$'', then the diagnosis logs are likely to contain more repairs of axioms including this quantifier. Consequently, the prior fault probabilities of axioms including ``$\exists$'' should be higher. Given the fault probabilities of axioms, one can calculate prior fault probabilities of diagnoses as well as evaluate $\mathsf{ENT}$ (see~\cite{Shchekotykhin2012} for more details). The queries for both strategies are constructed by exploiting so called classification and realization services provided by description logic reasoners. Given a KB $\dxoo$ and interpreting unary predicates as classes (rsp.\ concepts), the classification generates the inheritance (subsumption) tree, 
i.e.\ the entailments $\dxoo \models \forall X\, p(X) \rightarrow q(X)$, if $p$ is a subclass of $q$. 
%An implication is direct iff there is no unary predicate $s$ s.t.\ $\dxoo \models \forall X: p(X) \rightarrow s(X) \land \forall X: s(X) \rightarrow q(X).$ 
Realization computes, for each individual
name $t$ occurring in a KB $\dxoo$, a set of most specific classes $p$ s.t.\ $\dxoo \models p(t)$ (see \cite{Baader2007} for details). 

%If $\dxoo \models p(t)$ but there is no unary predicate $q$ s.t.\ $\dxoo \models q(t) \land \forall X : p(X) \rightarrow q(X).$ then $p$ is a most specific atomic concept for $t$. 

Due to the number of diagnoses and the complexity of diagnosis computation, not all diagnoses are exploited for generating queries but 
a set of minimal diagnoses of size less or equal to some (small) predefined number $m$~\cite{Shchekotykhin2012}. We call this set the \emph{leading diagnoses} and denote it by $\dxmD$ from now on. This set comprises the (most probable) minimal diagnoses which represent the set of all diagnoses. 

The sequential KB debugging process can be sketched as follows. As input a DPI and some meta information, such as prior fault estimates $\dxFP$, query selection strategy $\dxqss$ ($\mathsf{SPL}$ or $\mathsf{ENT}$) and stop criterion $\sigma$, are given. As output a minimal diagnosis is returned that has a posterior probability of at least $1-\sigma$. For sufficiently small $\sigma$ this means that the returned diagnosis is highly probable whereas all other minimal diagnoses are highly improbable. 

%Note, in the extreme case for $\sigma = 0$ there is only one minimal diagnosis left after the interactive debugging process stops. 
\begin{enumerate}
		\item Using \textsc{QX} and \textsc{HS-Tree}, compute a set of leading diagnoses $\dxmD$ of cardinality $\min(m,a)$, where $a$ is the number of all minimal diagnoses w.r.t.\ the DPI and $m$ is the number of leading diagnoses predefined by a user.
    \item Use the prior fault probabilities $\dxFP$ and the already specified test cases to compute (posterior) probabilities of diagnoses in $\dxmD$ by the Bayesian Rule (cf.~\cite{Shchekotykhin2012}). 
    \item If some diagnosis $\dxmd \in \dxmD$ has a probability greater than or equal to $1-\sigma$ or the user accepts $\dxmd$ as the axioms to be changed then stop and return $\dxmd$.
		\item Use $\dxmD$ to generate a set of queries and select the best query $\dxqry$ according to $\dxqss$.
		\item Ask the user $\dxot \models \dxqry$ and, depending on the answer, add $\dxqry$ either to $\dxTe$ or to $\dxTne$.
    \item Remove elements from $\dxmD$ violating the newly acquired test case.
    \item Repeat at Step 1. 
\end{enumerate}

%%%%%%%%%%%%%%%%%%%%%%%%%%%%%%%%%%%%%%%%%%%%%%%%%%%%%%%%%%%%%%%%%%%%%%%%%%%%%%%%%%%%%%%%%%%%%%%%%
%%%%%%%%%%%%%%%%%%%%%%%%%%%%%%%%%%%%%%%%%%%%%%%%%%%%%%%%%%%%%%%%%%%%%%%%%%%%%%%%%%%%%%%%%%%%%%%%%
%%%%%%%%%%%%%%%%%%%%%%%%%%%%%%%%%%%%%%%%%%%%%%%%%%%%%%%%%%%%%%%%%%%%%%%%%%%%%%%%%%%%%%%%%%%%%%%%%
%%%%%%%%%%%%%%%%%%%%%%%%%%%%%%%%%%%%%%%%%%%%%%%%%%%%%%%%%%%%%%%%%%%%%%%%%%%%%%%%%%%%%%%%%%%%%%%%%
%%%%%%%%%%%%%%%%%%%%%%%%%%%%%%%%%%%%%%%%%%%%%%%%%%%%%%%%%%%%%%%%%%%%%%%%%%%%%%%%%%%%%%%%%%%%%%%%%
%%%%%%%%%%%%%%%%%%%%%%%%%%%%%%%%%%%%%%%%%%%%%%%%%%%%%%%%%%%%%%%%%%%%%%%%%%%%%%%%%%%%%%%%%%%%%%%%%
%%%%%%%%%%%%%%%%%%%%%%%%%%%%%%%%%%%%%%%%%%%%%%%%%%%%%%%%%%%%%%%%%%%%%%%%%%%%%%%%%%%%%%%%%%%%%%%%%
%%%%%%%%%%%%%%%%%%%%%%%%%%%%%%%%%%%%%%%%%%%%%%%%%%%%%%%%%%%%%%%%%%%%%%%%%%%%%%%%%%%%%%%%%%%%%%%%%
%%%%%%%%%%%%%%%%%%%%%%%%%%%%%%%%%%%%%%%%%%%%%%%%%%%%%%%%%%%%%%%%%%%%%%%%%%%%%%%%%%%%%%%%%%%%%%%%%
%%%%%%%%%%%%%%%%%%%%%%%%%%%%%%%%%%%%%%%%%%%%%%%%%%%%%%%%%%%%%%%%%%%%%%%%%%%%%%%%%%%%%%%%%%%%%%%%%
%%%%%%%%%%%%%%%%%%%%%%%%%%%%%%%%%%%%%%%%%%%%%%%%%%%%%%%%%%%%%%%%%%%%%%%%%%%%%%%%%%%%%%%%%%%%%%%%%

\chapter[Interactive Direct Diagnosis of Knowledge Bases]{Interactive Direct Diagnosis of Knowledge Bases%
\chaptermark{Interactive Direct KB Diagnosis}}
\chaptermark{Interactive Direct KB Diagnosis}
\label{chap:dx:details}
%%%%%%
%\chapter{Interactive Direct Diagnosis of Knowledge Bases}
%\label{chap:dx:details}
%%%%%%
%\fixme{algo zeilennummern anpassen}
%Dual QX
 
The novelty of our approach is the interactivity combined with the direct calculation of diagnoses. To this end we will utilize an ``inverse'' version of the \textsc{QX} algorithm~\cite{junker04} called \textsc{Inv-QX} and an associated ``inverse'' version of \textsc{HS-Tree} termed \textsc{Inv-HS-Tree}. 

This combination of algorithms was first used in~\cite{Felfernig2011}. However, we introduced two modifications: (i) a depth-first search strategy instead of breadth-first and (ii) a new pruning rule which moves axioms from $\dxmo$ to $\dxmb$ instead of just  removing them from $\dxmo$, since not adding them to $\dxmb$ might result in losing some of the minimal diagnoses. 

\paragraph{\textsc{Inv-QX} -- Key Idea.} \textsc{Inv-QX} relies on the monotonic semantics of the used knowledge representation language. The algorithm takes a DPI $\tuple{\dxoo, \dxmb,\dxTe,\dxTne}$ and a ranking heuristic $\prec$ as input and outputs either one minimal diagnosis or 'no diagnosis exists'. The ranking heuristic assigns a fault probability to each axiom in $\dxoo$, if this information is available; otherwise every axiom has the same rank. 

The main idea behind Algorithm~\ref{algo:dx:inv_qx} is to start with the set $\dxmd_0=\emptyset$ and extend it until a subset of axioms $\dxmd \subseteq \dxmo$ is found such that $\dxmd$ is a minimal diagnosis with respect to Definition~\ref{def:diag}. In the first steps (lines~\ref{algoline:inv-qx:set_K'}-\ref{algoline:inv-qx:sort}), Algorithm~\ref{algo:dx:inv_qx} defines a (potentially) faulty set of axioms $\dxmo'$ and a set $\dxmb'$ of axioms assumed to be correct and sorts $\dxmo'$ w.r.t.\ the ranking heuristic (\textsc{sort}).
Next, \textsc{Inv-QX} verifies whether a diagnosis exists for the input data (line~\ref{algoline:inv-qx:verify_existence_of_diag}), i.e.\ if the conditions given by Proposition~\ref{pro:diagnosable} are met. This is accomplished by a call to the \textsc{verify} function (defined in line~\ref{algoline:inv-qx:verify_procedure} ff.) which requires a reasoner that implements consistency checking (\textsc{isConsistent}) and allows to decide whether a set of axioms $\dxmo'$ entails some axiom $n$ or not (\textsc{entails}). Concretely, \textsc{verify} tests for given arguments $\dxmb$ (set of correct axioms), $\dxmd$ (potential minimal diagnosis), $\dxmo$ (potentially faulty set of axioms), $\dxTne$ (negative test cases) whether the set $\dxmd$ is a minimal diagnosis or not according to Corollary~\ref{cor:diag}. In case no diagnosis exists, the algorithm returns 'no diagnosis exists', otherwise it calls the function \textsc{findDiag} in line~\ref{algoline:inv-qx:call_findDiag}.
%
%the input data (lines 4 and 6), i.e.\ if the sets of axioms $\dxmo$, $\dxmb$, $\dxTe$ and $\dxTne$ constitute a valid DPI according to Definition~\ref{def:diag}. Namely, the algorithm requires: a) the background theory together with the positive and negative test cases to be consistent (Proposition~\ref{pro:diagnosable}); and b) the knowledge base to be non-valid (Definition~\ref{def:valid}). In both cases \textsc{Inv-QX} calls the \textsc{verify} function (line~\ref{algoline:inv-qx:verify_procedure}) that tests whether the set $\dxmd$ is a minimal diagnosis or not according to Corollary~\ref{cor:diag}. This function requires a reasoner that implements consistency checking (\textsc{isConsistent}) and allows to decide whether a set of axioms $\dxmo'$ entails some axiom $n$ or not (\textsc{entails}). 

\textsc{findDiag} (line~\ref{algoline:inv-qx:findDiag_procedure}) is the main function of the algorithm which takes six arguments as input. The values of the arguments $\dxmb$, $\dxmo$ and $\dxTne$ remain constant during the recursion and are required only for the verification of requirements, i.e.\ calls to the \textsc{verify} function. The values of $\dxmd$ (potential diagnosis), $\Delta$ (axioms most recently added to $\dxmd$) and $\dxmo_\Delta$ (part of the original knowledge base that is currently analyzed for the inclusion of axioms that are elements of the sought minimal diagnosis) on the other hand change throughout the recursive calls of \textsc{findDiag}. The two latter sets are obtained by recurrently partitioning the set $\dxmo_{\Delta}$ (\textsc{split} and \textsc{getElements} in lines~\ref{algoline:inv-qx:split}-\ref{algoline:inv-qx:getElements2}). 
In most of the implementations \textsc{split} is specified so as to return $k = \lfloor|\dxmo_\Delta|/2\rfloor$ which causes the splitting of $\dxmo_\Delta$ into partitions of equal cardinality (this results in the best worst case time complexity~\cite{junker04}).
%
%In most of the implementations \textsc{split} simply partitions the axioms of $\dxmo$ into two sets of equal cardinality.
%
The algorithm pursues this to divide-and-conquer strategy (lines~\ref{algoline:inv-qx:recursive_call1} and \ref{algoline:inv-qx:recursive_call2}) until it identifies that the set $\dxmd$ is a diagnosis (line~\ref{algoline:inv-qx:delta_verify}). In further iterations the algorithm  minimizes this diagnosis by splitting it into sub-diagnoses of the form $\dxmd = \dxmd' \cup \dxmo_\Delta$, where $\dxmo_\Delta$ contains only one axiom. In case $\dxmd$ is a diagnosis and $\dxmd'$ is not, the algorithm decides that $\dxmo_\Delta$ is a subset of the sought minimal diagnosis. Just as the original \textsc{QX} algorithm, \textsc{Inv-QX} always terminates and it returns a minimal diagnosis for a given DPI (provided there exists one).

\begin{algorithm*}[tb]
\small
\caption{\textsc{Inv-QX$(\dxmo, \dxmb, \dxTe, \dxTne, \prec)$}} \label{algo:dx:inv_qx}
\begin{algorithmic}[1]
\Require 
faulty set of axioms $\dxmo$, set of background axioms $\dxmb$, set of positive test cases $\dxTe$, set of negative test cases $\dxTne$, ranking heuristic $\prec$
\Ensure 
a minimal diagnosis $\dxmd$ or 'no diagnosis exists'
\vspace{10pt}
%%%%%%%%%%%%%%%%%%%%
%%%%%%%%%%%%%%%%%%%%
\State $\dxmo' \gets \dxmo \setminus\dxmb$ \label{algoline:inv-qx:set_K'}
\State $\dxmb' \gets \dxmb \cup \dxTe$
\State $\dxmo' \gets \Call{sort}{\dxmo', \prec}$ \label{algoline:inv-qx:sort}
\If{$\lnot\Call{verify}{\dxmb',\emptyset, \emptyset, \dxTne}$} \label{algoline:inv-qx:verify_existence_of_diag}
	\State \Return 'no diagnosis exists'
\EndIf
%\If{$\Call{verify}{\dxmb',\emptyset, \dxmo', \dxTne}$}
  %\State \Return 'consistent'
%\EndIf
\State \Return $\Call{findDiag}{\dxmb', \emptyset, \dxmo', \dxmo', \dxmo', \dxTne}$ \label{algoline:inv-qx:call_findDiag}

\vspace{10pt}

\Procedure{\textsc{findDiag}}{$\dxmb, \dxmd, \Delta, \dxmo_\Delta, \dxmo, \dxTne$} \textbf{returns} a minimal diagnosis  \label{algoline:inv-qx:findDiag_procedure}
\If{$\Delta \neq \emptyset \land \Call{verify}{\dxmb, \dxmd, \dxmo, \dxTne}$} \label{algoline:inv-qx:delta_verify}              
  \State \Return $\emptyset$
\EndIf
\If{$|\dxmo_\Delta| = 1$}  \label{algoline:inv-qx:is_singleton}             
	\State \Return $\dxmo_\Delta$
\EndIf
\State $k \gets \Call{split}{|\dxmo_\Delta|}$ \label{algoline:inv-qx:split}
\State $\dxmo_1 \gets \Call{getElements}{\dxmo_\Delta, 1, k}$
\State $\dxmo_2 \gets \Call{getElements}{\dxmo_\Delta, k + 1, |\dxmo_\Delta|}$ \label{algoline:inv-qx:getElements2}
\State $\dxmd_2 \gets \Call{findDiag}{\dxmb, \dxmd\cup\dxmo_1, \dxmo_1, \dxmo_2, \dxmo, \dxTne}$ \label{algoline:inv-qx:recursive_call1}
\State $\dxmd_1 \gets \Call{findDiag}{\dxmb, \dxmd\cup\dxmd_2, \dxmd_2, \dxmo_1, \dxmo, \dxTne}$ \label{algoline:inv-qx:recursive_call2}
\State \Return $\dxmd_1 \cup \dxmd_2$
\EndProcedure

\vspace{10pt}

\Procedure{\textsc{verify}}{$\dxmb, \dxmd, \dxmo, \dxTne$} \textbf{returns} $\mathit{true}$ or $\mathit{false}$ \label{algoline:inv-qx:verify_procedure}
\State $\dxmo' \gets (\dxmo \setminus \dxmd)\cup\dxmb$
\If{$\lnot \Call{isConsistent}{\dxmo'}$}
\State \Return $\mathit{false}$
\EndIf
\For{$\dxtne \in \dxTne$}
	\If{$\Call{entails}{\dxmo', \dxtne}$}
		\State \Return $\mathit{false}$
  \EndIf
\EndFor
\State \Return $\mathit{true}$
\EndProcedure
\end{algorithmic}
\normalsize
\end{algorithm*}

\textsc{Inv-QX}  requires $O(|\dxmd|\log(\frac{|\dxoo|}{|\dxmd|}))$ calls to a reasoner to find a minimal diagnosis $\dxmd$.
Moreover, in opposite to SAT or CSP methods, e.g.~\cite{Nica2013}, \textsc{Inv-QX} can be used to compute diagnoses in cases when satisfiability checking is beyond $\textsc{NP}$. For instance, reasoning for most of the KBs used in Chapter~\ref{chap:dx:eval} is \textsc{ExpTime}-complete. 

\textsc{Inv-QX} is a deterministic algorithm and returns one and the same minimal diagnosis if applied twice to one and the same DPI. In order to obtain a different next diagnosis, the DPI used as input for \textsc{Inv-QX} must be modified accordingly. To this end, we employ the \textsc{Inv-HS-Tree} algorithm. 

\paragraph{\textsc{Inv-HS-Tree} -- Construction.} The algorithm is inverse to the \textsc{HS-Tree} algorithm in the sense that nodes are now labeled by minimal diagnoses (instead of minimal conflict sets) and a path from the root to an open node is a partial conflict set (instead of a partial diagnosis). The algorithm constructs a directed  tree from the root to the leaves, where each node $\mathsf{nd}$ is labeled either with a minimal diagnosis $\dxmd$ or $\times$ (\emph{pruned}) which indicates that the node is closed. For each $s \in \dxmd$ there is an outgoing edge labeled by $s$. Let $H(\mathsf{nd})$ be the set of edge labels on the path from the root to the node $\mathsf{nd}$.
Initially the algorithm generates an empty root node and adds it to a LIFO-queue, thereby implementing a \emph{depth-first search} strategy. 
%We can apply depth-first because the minimality of paths (conflict sets) is irrelevant in \textsc{Inv-HS-Tree} whereas in \textsc{HS-Tree} minimality of paths (diagnoses) is essential and therefore breadth-first is employed. 
Until the required number $m$ of minimal diagnoses is reached or the queue is empty, the algorithm removes the first node $\mathsf{nd}$ from the queue and labels $\mathsf{nd}$ 
%with the applicable label with lowest number among the following:
by applying the following steps: 
%from the queue and performs the following checks 1 to 4 in chronological order to label $nd$:
%
\begin{enumerate}
    %\item $\times$ if \textsc{verifyRequirements}$(\dxmb\cup U_\dxTe,\emptyset,H(nd),\dxTne)$ is false, i.e.\	$H(nd)$ is a conflict set 
		%w.r.t. $\tuple{\dxmo,\dxmb,\dxTe,\dxTne}$ (\emph{pruned}), or
		\item (\emph{reuse}): $\dxmd \in \dxmD$ if  $\dxmd \cap H(\mathsf{nd}) = \emptyset$, add for each $s \in \dxmd$ a node to the LIFO-queue, or
    \item (\emph{pruned}): $\times$ if \textsc{Inv-QX}$(\dxmo \setminus H(\mathsf{nd}),\dxmb \cup H(\mathsf{nd}),\dxTe,\dxTne) =$ 'no-diagnosis-exists', (according to Proposition~\ref{pro:diagnosable}), or
		%i.e.\ $H(nd)$ is a conflict set w.r.t.\ $\tuple{\dxmo,\dxmb,\dxTe,\dxTne}$ (\emph{pruned}), or \\
		\item (\emph{compute}): $\dxmd$ if $\textsc{Inv-QX}(\dxmo \setminus H(\mathsf{nd}),\dxmb \cup H(\mathsf{nd}),\dxTe,\dxTne) = \dxmd$; add $\dxmd$ to $\dxmD$ and add for each $s \in \dxmd$ a node to the LIFO-queue. 
\end{enumerate}
Reuse of known diagnoses in Step~1 and the addition of $H(\mathsf{nd})$ to the background theory $\dxmb$ in Steps~2 and 3 
%, i.e.\ one axiom $\dxtax_k$ of each minimal diagnosis $\dxmd_i$ on the path from the root to $nd$,
allows the algorithm to force \textsc{Inv-QX} to search for a minimal diagnosis that is different to all already computed minimal diagnoses in $\dxmD$.
So, if neither Step~1 nor Step~2 are applicable, \textsc{Inv-HS-Tree} calls \textsc{Inv-QX} which is guaranteed to compute a \emph{new} minimal diagnosis $\dxmd$ which is then added to the set $\dxmD$.
%
%The depth-first search strategy maintains only a set of leading diagnoses $\dxmD$ s.t.\ $|\dxmD| \leq m$. No conflicts are stored. This allows a significant reduction of memory usage by \textsc{Inv-HS-Tree} compared to \textsc{HS-Tree}. 
%
%That is, the worst case space complexity of the \textsc{Inv-HS-Tree} computing $m$ leading minimal diagnoses is \emph{linear} and amounts to $O(m)$, whereas the worst case space complexity of \textsc{HS-Tree} is $O(|CS_{\max}|^c)$ where $|CS_{\max}|$ is the maximum cardinality of a minimal conflict set and $c$ is the number of all minimal conflict sets w.r.t.\ a DPI.

\paragraph{\textsc{Inv-HS-Tree} -- Update Procedure for Interactivity.}
Since paths in \textsc{Inv-HS-Tree} are (1)~irrelevant and need not be maintained, and (2)~only a small (linear) number of nodes/paths is in memory due to the application of a depth-first search, the update procedure after a query $Q$ has been answered involves a reconstruction of the tree. In particular, by answering $Q$,
$m-k$ of (maximally) $m$ leading diagnoses are invalidated  and deleted from memory. %, where $1\leq k\leq m-1$. 
The $k$ still valid minimal diagnoses are used to build a new tree. To this end, the root is labeled by any of these $k$ minimal diagnoses and a tree is constructed as described above where the $k$ diagnoses are incorporated for the \emph{reuse} check. Note that the recalculation of a diagnosis that has been invalidated by a query is impossible as in subsequent iterations a new DPI is considered which includes the answered query as a test case.

\paragraph{\textsc{Inv-HS-Tree} -- Comparison to \textsc{HS-Tree}.} 
Since \textsc{Inv-QX}$(\dxmo,\dxmb \cup H(\mathsf{nd}),\dxTe,\dxTne) = $ 'no diagnosis exists' means $H(\mathsf{nd})$ is a conflict set w.r.t.\ the current DPI $\tuple{\dxmo,\dxmb,\dxTe,\dxTne}$, in \textsc{Inv-HS-Tree} any path that is a conflict set is \emph{automatically} closed. This makes a pruning rule similar to the one in \textsc{HS-Tree} which closes a node $\mathsf{nd}$ given an alternative path $H(\mathsf{nd}')$ to a closed node $\mathsf{nd}'$ with $H(\mathsf{nd}') \subseteq H(\mathsf{nd})$ obsolete. So, \textsc{Inv-HS-Tree} benefits from the fact that minimality of diagnoses is independent of path-minimality, and thereby might save time for comparison of exponentially many paths over \textsc{HS-Tree}.

Another great advantage of \textsc{Inv-HS-Tree} over \textsc{HS-Tree} is that it can be constructed using a space-saving depth-first strategy. The reason for this is again that minimality of paths (conflict sets) is irrelevant in \textsc{Inv-HS-Tree} whereas in \textsc{HS-Tree} minimality of paths (diagnoses) is essential. In an implementation where successors of a node are generated one at a time in \textsc{Inv-HS-Tree}, the space complexity of the entire tree construction is \emph{linear} and amounts to $O(2m) = O(m)$ where $m$ is the predefined maximum number of leading diagnoses. 
This holds as $k < m$ still valid diagnoses from the previous iteration are in memory, plus a path in the tree can comprise a maximum of $m$ nodes corresponding to \emph{different} (reused or new) diagnoses before the search is stopped ($|\mD| = m$). No conflict sets are stored.

For \textsc{HS-Tree}, by contrast, the worst-case space complexity is \emph{exponential}, i.e.\ $O(|CS_{\max}|^d)$ where $|CS_{\max}|$ is the size of the minimal conflict set with maximum cardinality (among all minimal conflict sets w.r.t.\ the given DPI) and $d$ is the tree depth were $m$ minimal diagnoses have been generated.

The crucial disadvantage of \textsc{Inv-HS-Tree} compared to \textsc{HS-Tree} is that the former 
cannot guarantee the computation of diagnoses in a special order, e.g.\ minimum cardinality or maximum fault probability first.

\begin{figure*}[tb]
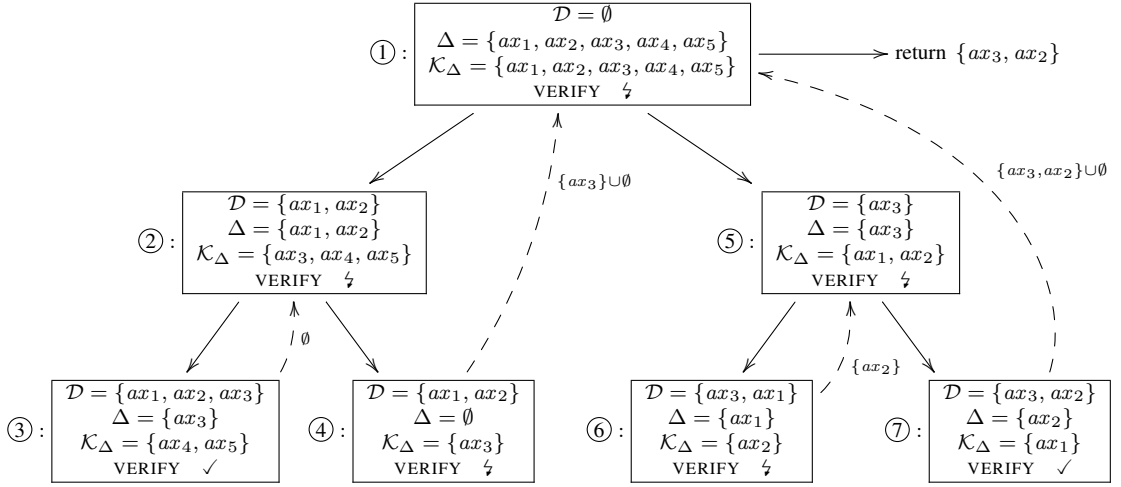

\footnotesize
\centering
\xygraph{
!{<0cm,0cm>;<1.83cm,0cm>:<0cm,2.5cm>::}
!{(1,2)}*+{ \circled{1}:
\begin{array}{|c|} \hline
\dxmd=\emptyset\\
\Delta = \setof{\dxtax_1,\dxtax_2, \dxtax_3,\dxtax_4,\dxtax_5}\\
\dxmo_\Delta = \setof{\dxtax_1,\dxtax_2, \dxtax_3,\dxtax_4,\dxtax_5} \\
\textsc{verify} \quad \text{\lightning} \\ \hline
\end{array}
}="r"
!{(-1,1)}*+{ \circled{2}:
\begin{array}{|c|} \hline
\dxmd=\setof{\dxtax_1,\dxtax_2}\\
\Delta = \setof{\dxtax_1,\dxtax_2}\\
\dxmo_\Delta = \setof{\dxtax_3,\dxtax_4,\dxtax_5} \\
\textsc{verify} \quad \text{\lightning} \\ \hline
\end{array}
}="c1"
!{(-2,0)}*+{ \circled{3}:
\begin{array}{|c|}\hline
\dxmd=\setof{\dxtax_1,\dxtax_2,\dxtax_3}\\
\Delta = \setof{\dxtax_3}\\
\dxmo_\Delta = \setof{\dxtax_4,\dxtax_5} \\
\textsc{verify} \quad \text{\checked} \\ \hline
\end{array}
}="c2"
!{(0,0)}*+{ \circled{4}:
\begin{array}{|c|} \hline
\dxmd=\setof{\dxtax_1,\dxtax_2}\\
\Delta = \emptyset\\
\dxmo_\Delta = \setof{\dxtax_3} \\
\textsc{verify} \quad \text{\lightning} \\ \hline
\end{array}
}="c3"
"r":"c1"_{}
"c1":"c2"_{}
"c2":@{-->}@/_1cm/"c1"_(.7){\emptyset}
"c1":"c3"_{}
"c3":@{-->}@/_.5cm/"r"_(.7){\setof{\dxtax_3}\cup\emptyset}
!{(3,1)}*+{ \circled{5}:
\begin{array}{|c|} \hline
\dxmd=\setof{\dxtax_3}\\
\Delta = \setof{\dxtax_3}\\
\dxmo_\Delta = \setof{\dxtax_1,\dxtax_2} \\
\textsc{verify} \quad \text{\lightning} \\ \hline
\end{array}
}="c4"
!{(2,0)}*+{ \circled{6}:
\begin{array}{|c|} \hline
\dxmd=\setof{\dxtax_3,\dxtax_1}\\
\Delta = \setof{\dxtax_1}\\
\dxmo_\Delta = \setof{\dxtax_2} \\
\textsc{verify} \quad \text{\lightning} \\ \hline
\end{array}
}="c5"
!{(4,0)}*+{ \hspace{15pt} \circled{7}:
\begin{array}{|c|} \hline
\dxmd=\setof{\dxtax_3,\dxtax_2}\\
\Delta = \setof{\dxtax_2}\\
\dxmo_\Delta = \setof{\dxtax_1} \\
\textsc{verify} \quad \text{\checked} \\ \hline
\end{array}
}="c6"
"r":"c4"_{}
"c4":"c5"_{}
"c5":@{-->}@/_1cm/"c4"_(.6){\setof{\dxtax_2}}
"c4":"c6"_{}
"c6":@{-->}@`{c+(1,0),p+(4,0)}"r"_(.6){\setof{\dxtax_3,\dxtax_2}\cup\emptyset}
!{(4,2)}*+{
\text{return } \setof{\dxtax_3,\dxtax_2} }="o"
"r":"o"_{}
}

\caption[\textsc{Inv-QX} Recursion Tree]{\textsc{Inv-QX} recursion tree. Each node shows values of \textsc{findDiag} input variables as well as the result of the \textsc{verify} function called in line~\ref{algoline:inv-qx:delta_verify}.} \label{fig:qx:ex}
\vspace{10pt}
\end{figure*}

\begin{figure*}[tb]
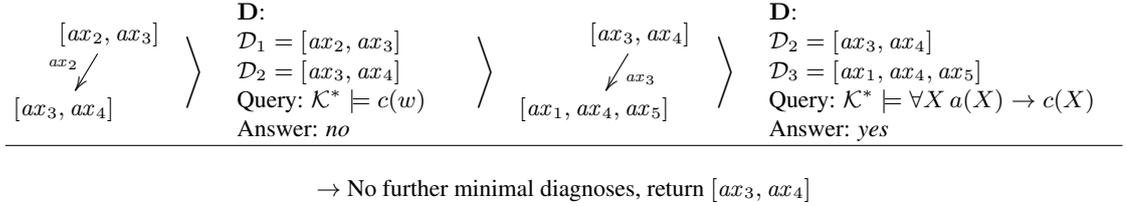

%\hspace{5pt}
\small
\begin{minipage}[c]{0.15\linewidth} 
\xygraph{
!{<0cm,0cm>;<0.6cm,0cm>:<0cm,1cm>::}
%!{(1,0) }*+{[\quad]}="d1"
!{(0,0) }*+{[\dxtax_3, \dxtax_4]}="c3"
!{(1,1)}*+{[\dxtax_2, \dxtax_3]}="c1"
%"c1":"d1"^{\dxtax_3}
"c1":"c3"_{\dxtax_2}
}
\end{minipage}
\begin{minipage}[c]{15pt}
\centering
$\Bigg>$ 
\end{minipage}
\begin{minipage}[c]{0.2\linewidth}
\centering
\begin{tabular}{l}                                          
      $\dxmD$: \\
                $\dxmd_1=[\dxtax_2, \dxtax_3]$ \\
                $\dxmd_2=[\dxtax_3, \dxtax_4]$ \\
         %\\
                Query: $\dxot \models c(w)$\\
                Answer: \emph{no}
		\end{tabular}
\end{minipage}
\begin{minipage}[c]{21pt}
\centering
\hspace{6pt}$\Bigg>$ 
\end{minipage} 
\begin{minipage}[c]{0.17\linewidth}
\xygraph{
!{<0cm,0cm>;<0.6cm,0cm>:<0cm,1cm>::}
!{(0,0) }*+{[\dxtax_1, \dxtax_4, \dxtax_5]}="d1"
%!{(1,0) }*+{[\quad]}="d0"
!{(1,1) }*+{[\dxtax_3, \dxtax_4]}="c3"
%"c3":"d0"^{\dxtax_4}
"c3":"d1"^{\dxtax_3}
}
\end{minipage}
\begin{minipage}[c]{16pt}
\centering
$\Bigg>$ 
\end{minipage}
\begin{minipage}[c]{0.17\linewidth}
\centering
\begin{tabular}{l}                                          
      $\dxmD$: \\
                $\dxmd_2=[\dxtax_3, \dxtax_4]$ \\
                $\dxmd_3=[\dxtax_1, \dxtax_4, \dxtax_5]$ \\
         %\\
                Query: $\dxot \models \forall X\, a(X) \to c(X)$\\
                Answer: \emph{yes} \\
         %\\
         				%No further minimal \\
                %diagnoses, return \\
                %$[\dxtax_3, \dxtax_4]$
								%Return $[\dxtax_3, \dxtax_4]$
		\end{tabular}
\end{minipage}
\hrule
\vspace{2pt}
\begin{center}
$\rightarrow$ No further minimal diagnoses, return $[\dxtax_3, \dxtax_4]$
\end{center}
%\vspace{-5pt}
\caption[Identification of the Target Diagnosis Using \textsc{Inv-HS-Tree} and \textsc{Inv-QX}]{Identification of the target diagnosis $[\dxtax_3,\dxtax_4]$ using \textsc{Inv-HS-Tree}. 
} \label{fig:dual:ex}
%\vspace{-15pt}
\end{figure*}

\begin{figure*}[bt]
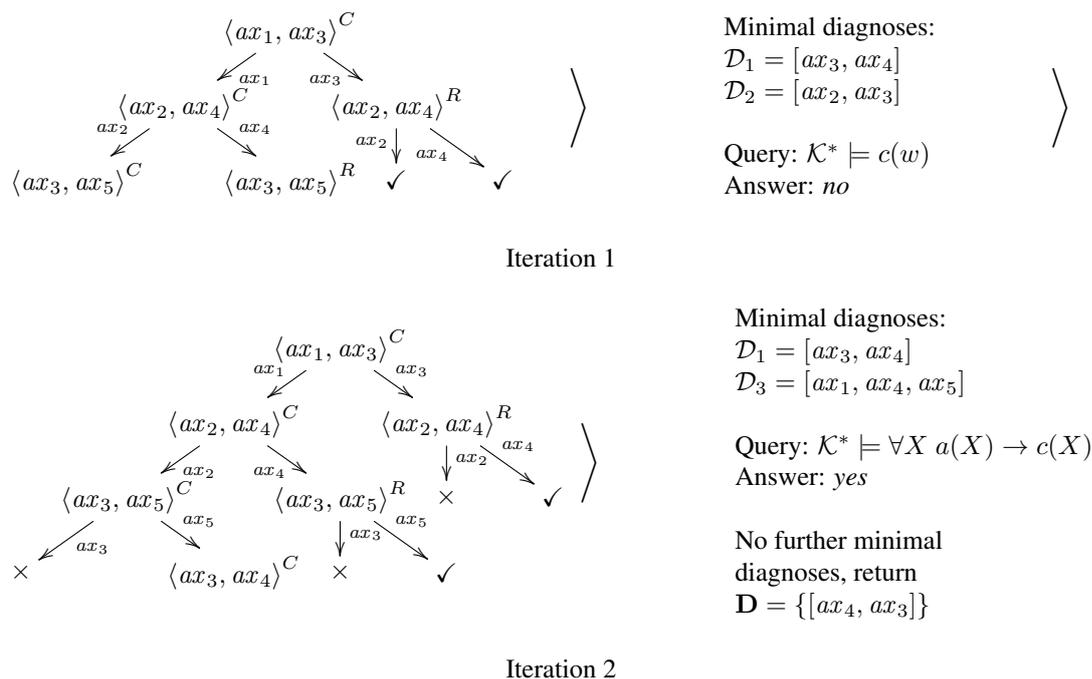

\begin{minipage}[c]{.5\textwidth} 
\xygraph{
!{<0cm,0cm>;<1.4cm,0cm>:<0cm,1cm>::}
!{(4,0) }*+{\checkmark}="d2"
!{(3,0) }*+{\checkmark}="d1"
!{(1,1) }*+{\left<\dxtax_2, \dxtax_4\right>^C}="c2c"
!{(3,1) }*+{\left<\dxtax_2, \dxtax_4\right>^R}="c2r"
!{(0,0) }*+{\left<\dxtax_3, \dxtax_5\right>^C}="c3c"
!{(2,0) }*+{\left<\dxtax_3, \dxtax_5\right>^R}="c3r"
!{(2,2)}*+{\left<\dxtax_1, \dxtax_3\right>^C}="c1c"
"c2c":"c3r"^{\dxtax_4}
"c2c":"c3c"_{\dxtax_2}
"c1c":"c2c"^{\dxtax_1}
"c1c":"c2r"_{\dxtax_3}
"c2r":"d1"_{\dxtax_2}
"c2r":"d2"_{\dxtax_4}
}
\hfill
\end{minipage}
\begin{minipage}[c]{50pt}
$\Bigg>$ \hfill
\end{minipage}
\begin{minipage}[c]{0.3\textwidth}
\begin{tabular}{l}                                          
    Minimal diagnoses: \\
                $\dxmd_1=[\dxtax_3, \dxtax_4]$ \\
                $\dxmd_2=[\dxtax_2, \dxtax_3]$ \\
         \\
                Query: $\dxot \models c(w)$ \\
                Answer: \emph{no}
		\end{tabular}
\end{minipage}
\begin{minipage}[c]{10pt}
$\Bigg>$ 
\end{minipage}
\hfill
\vspace{5pt}
\begin{center}
Iteration 1
\end{center}
\vspace{5pt}
\begin{minipage}[c]{0.5\textwidth}
\vspace{0pt}
\centering
\xygraph{
!{<0cm,0cm>;<1.4cm,0cm>:<0cm,1cm>::}
!{(4,0)}*+{\checkmark}="d2"
!{(4,1)}*+{\times}="o4"
!{(0,0)}*+{\times}="o2"
!{(5,1)}*+{\checkmark}="d1"
!{(3,0)}*+{\times}="o1"
!{(3,1)}*+{\left<\dxtax_3, \dxtax_5\right>^R}="c3r"
!{(4,2)}*+{\left<\dxtax_2, \dxtax_4 \right>^R}="c2r"
!{(2,0)}*+{\left<\dxtax_3, \dxtax_4\right>^C}="c4"
!{(1,1)}*+{\left<\dxtax_3, \dxtax_5\right>^C}="c3"
!{(2,2)}*+{\left<\dxtax_2, \dxtax_4 \right>^C}="c2"
!{(3,3)}*+{\left<\dxtax_1, \dxtax_3\right>^C}="c1"
"c3r":"d2"^{\dxtax_5}
"c3r":"o1"^{\dxtax_3}
"c3":"c4"^{\dxtax_5}
"c3":"o2"^{\dxtax_3}
"c2r":"d1"^{\dxtax_4}
"c2r":"o4"^{\dxtax_2}
"c2":"c3r"_{\dxtax_4}
"c2":"c3"^{\dxtax_2}
"c1":"c2r"^{\dxtax_3}
"c1":"c2"_{\dxtax_1}
}\hfill
\end{minipage}
\begin{minipage}[c]{50pt}
$\Bigg>$ \hfill
\end{minipage} 
\begin{minipage}[c]{0.3\textwidth}
		\begin{tabular}{l}  
                Minimal diagnoses: \\
                $\dxmd_1=[\dxtax_3, \dxtax_4]$ \\
                $\dxmd_3=[\dxtax_1, \dxtax_4, \dxtax_5]$ \\
                \\
                Query: $\dxot \models \forall X\; a(X)\to c(X)$ \\
                Answer: \emph{yes} \\
                \\
                No further minimal \\
                diagnoses, return \\
                $\dxmD = \setof{[\dxtax_4, \dxtax_3]}$
		\end{tabular}
\end{minipage}
\vspace{5pt}
\begin{center}
Iteration 2
\end{center}
\caption[Identification of the Target Diagnosis Using \textsc{HS-Tree} and \textsc{QX}]{
Identification of the target diagnosis $[\dxtax_4,\dxtax_3]$ using \textsc{HS-Tree} and \textsc{QX} computing conflicts on-demand. All computed node labels are denoted with $C$ and all reused with $R$.
} \label{fig:hs:ex}
\end{figure*}

%\paragraph{Example.} 
\begin{example}
Consider a DPI with the following knowledge base $\dxmo$: 
\begin{align*}
&\dxtax_1 : \forall X\,  c(X) \to a(X) \hspace{14pt}  \dxtax_4 : \forall X\,  b(X) \to c(X) \\
&\dxtax_2 : \forall X\,  c(X) \to e(X) \hspace{14pt} \dxtax_5 : \forall X\,  b(X) \to \lnot d(X) \\
&\dxtax_3 : \forall X\, a(X) \to \lnot (c(X) \lor \lnot b(X))
\end{align*}
the background knowledge $\dxmb=\{a(v), b(w), c(s)\}$, one positive $\dxTe=\setof{d(v)}$ and one negative $\dxTne=\setof{e(w)}$ test case. 
%

% Computation of a diagnosis using Inv-QX
Let us first show how a minimal diagnosis is computed by \textsc{Inv-QX} (see Figure~\ref{fig:qx:ex}). The algorithm starts with an empty diagnosis $\dxmd=\emptyset$ and $\dxmo_\Delta$ containing all axioms of $\dxmo$ \circled{1}. \textsc{verify} called in line~\ref{algoline:inv-qx:delta_verify} returns \emph{false} since $(\dxmb \cup \dxTe) \cup (\dxmo \setminus \emptyset)$ is inconsistent. Since moreover $|\dxmo_\Delta| \neq 1$ (line~\ref{algoline:inv-qx:is_singleton}), the algorithm splits $\dxmo_\Delta$ into $\setof{\dxtax_1,\dxtax_2}$ and $\setof{\dxtax_3,\dxtax_4,\dxtax_5}$ (lines~\ref{algoline:inv-qx:split}-\ref{algoline:inv-qx:getElements2}) and passes the sub-problem (line~\ref{algoline:inv-qx:recursive_call1}) to the next level of recursion \circled{2}. Since the set $\dxmd=\setof{\dxtax_1,\dxtax_2}$ is not a diagnosis, i.e.\ the KB $(\dxmb \cup \dxTe) \cup (\dxmo \setminus \dxmd)$ is inconsistent and $|\dxmo_\Delta|=|\setof{\dxtax_3,\dxtax_4,\dxtax_5}| \neq 1$, the problem in $\dxmo_\Delta$ is split one more time (lines~\ref{algoline:inv-qx:split}-\ref{algoline:inv-qx:getElements2}). On the second level of recursion \circled{3} the set $\dxmd$ is a diagnosis, yet not a minimal one. The function \textsc{verify} returns \emph{true} and the algorithm starts to analyze the found diagnosis. Therefore, it verifies whether the last extension of the set $\dxmd$ is a subset of a minimal diagnosis \circled{4}. Since the extension includes only one axiom $\dxtax_3$ and the extended set $\setof{\dxtax_1,\dxtax_2}$ is not a diagnosis, the algorithm concludes that $\dxtax_3$ must be an element of the a minimal diagnosis. The leftmost branch of the recursion tree terminates and returns $\setof{\dxtax_3}$. This axiom is added to the set $\dxmd$ and the algorithm starts investigating whether the two axioms $\setof{\dxtax_1,\dxtax_2}$ also belong to a minimal diagnosis \circled{5}. First, it tests the set $\setof{\dxtax_3,\dxtax_1}$ \circled{6}, which is not a diagnosis, and in the next iteration it identifies $\setof{\dxtax_3,\dxtax_2}$ as a minimal diagnosis in node \circled{7} which is the final output of \textsc{Inv-QX}.

In general, for the sample DPI there are three minimal diagnoses $\{\dxmd_1: [\dxtax_2,\dxtax_3],$ $\dxmd_2: [\dxtax_3,\dxtax_4],$ $\dxmd_3: [\dxtax_1,\dxtax_4,\dxtax_5] \}$ and four minimal conflict sets 
$\{ CS_1: \tuple{\dxtax_1,\dxtax_3},$ $CS_2: \tuple{\dxtax_2,\dxtax_4},$ $CS_3: \tuple{\dxtax_3,\dxtax_5},$ $CS_4: \tuple{\dxtax_3, \dxtax_4}\}$.

Now we show how \textsc{Inv-HS-Tree} can be applied to find the (correct) diagnosis that allows the formulation of a valid KB (with the desired semantics in terms of entailments and non-entailments). Assume that the number of leading diagnoses required for query generation is set to $m=2$. Applied to the sample DPI, \textsc{Inv-HS-Tree} computes a minimal diagnosis $\dxmd_1 := [\dxtax_2, \dxtax_3] = \textsc{Inv-QX}(\dxmo,\dxmb,\dxTe,\dxTne)$ to label the root node, see Figure~\ref{fig:dual:ex}. Next, it generates one successor node that is linked with the root by an edge labeled with $\dxtax_2$.
For this node $\textsc{Inv-QX}(\dxmo \setminus \setof{\dxtax_2},\dxmb \cup \setof{\dxtax_2},\dxTe,\dxTne)$  yields a minimal diagnosis $\dxmd_2:=[\dxtax_3, \dxtax_4]$ disjoint with $\setof{\dxtax_2}$. Now $|\dxmD|=2$ and a query is generated and answered as in Figure~\ref{fig:dual:ex}. 
Adding $c(w)$ to the negative test cases invalidates $\dxmd_1$ since $(\dxmo \setminus \dxmd_1) \cup \dxmb \cup \dxTe \models c(w)$.  
In the course of the update, $\dxmd_1$ is deleted and $\dxmd_2$ used as the root of a new tree.  
An edge labeled with $ax_3$ is created and diagnosis $\dxmd_3:=[\dxtax_1, \dxtax_4, \dxtax_5]$ is generated. 
After the answer to the second query is added to the positive test cases, $\dxmd_3$ is invalidated and all outgoing edge labels $\dxtax_3, \dxtax_4$ of the root $\dxmd_2$ of the new tree are conflict sets for the current DPI 
$\tuple{\dxmo,\dxmb,\setof{d(v), \forall X\, a(X)\to c(X)},\setof{e(w), c(w)}}$, 
i.e.\ all leaf nodes are labeled by $\times$ and the tree construction is complete. So, $\dxmd_2$ is returned as its probability is~1.

% Comparison with HS-Tree
Finally, let us compare the performance of \textsc{HS-Tree}~\cite{Reiter87} with the one of \textsc{Inv-HS-Tree}. Applied to our sample DPI, the standard interactive diagnosis process using \textsc{HS-Tree} first calls \textsc{QX}~\cite{junker04} which returns a minimal conflict set $\tuple{\dxtax_1,\dxtax_3}$ (Figure~\ref{fig:hs:ex}).
This minimal conflict set is used to label the root node of the \textsc{HS-Tree}. By reuse ($R$) of already computed minimal conflict sets or further calls ($C$) to \textsc{QX} (if there is no conflict set to reuse) the algorithm extends the \textsc{HS-Tree} until $m=2$ leading minimal diagnoses $\dxmD := \setof{\dxmd_1, \dxmd_2}$ for the DPI are computed. To discriminate between diagnoses in $\dxmD$, the query $\dxot \models c(w)$ is computed. 
Given the answer \emph{no}, $\dxmd_2$ is invalidated which is reflected by the closing of the corresponding node in the tree (label $\times$). 
The second iteration considers the new DPI $\tuple{\dxmo,\dxmb,\setof{d(v)},\setof{e(w), c(w)}}$ and involves further expansion of (open nodes in) the tree under consideration of the pruning rule until the size of leading diagnoses $\dxmD$ is $2$, i.e.\ $\setof{\dxmd_1,\dxmd_3}$. After the positive answer to the second query and closing of the invalidated diagnosis $\dxmd_3$, the recalculation of $\dxmD$ (not shown in Figure~\ref{fig:hs:ex}) yields no further minimal diagnoses. So, the algorithm terminates and returns $\dxmd_1$.
As we can see, \textsc{HS-Tree} comprises a lot of intermediate nodes in comparison to \textsc{Inv-HS-Tree}. That leads to a dramatic difference in memory consumption between these two approaches.\qed
\end{example}

%%%%%%%%%%%%%%%%%%%%%%%%%%%%%%%%%%%%%%%%%%%%%%%%%%%%%%%%%%%%%%%%%%%%%%%%%%%%%%%%%%%%%%%%%%%%%%%%%
%%%%%%%%%%%%%%%%%%%%%%%%%%%%%%%%%%%%%%%%%%%%%%%%%%%%%%%%%%%%%%%%%%%%%%%%%%%%%%%%%%%%%%%%%%%%%%%%%
%%%%%%%%%%%%%%%%%%%%%%%%%%%%%%%%%%%%%%%%%%%%%%%%%%%%%%%%%%%%%%%%%%%%%%%%%%%%%%%%%%%%%%%%%%%%%%%%%
%%%%%%%%%%%%%%%%%%%%%%%%%%%%%%%%%%%%%%%%%%%%%%%%%%%%%%%%%%%%%%%%%%%%%%%%%%%%%%%%%%%%%%%%%%%%%%%%%
%%%%%%%%%%%%%%%%%%%%%%%%%%%%%%%%%%%%%%%%%%%%%%%%%%%%%%%%%%%%%%%%%%%%%%%%%%%%%%%%%%%%%%%%%%%%%%%%%
%%%%%%%%%%%%%%%%%%%%%%%%%%%%%%%%%%%%%%%%%%%%%%%%%%%%%%%%%%%%%%%%%%%%%%%%%%%%%%%%%%%%%%%%%%%%%%%%%
%%%%%%%%%%%%%%%%%%%%%%%%%%%%%%%%%%%%%%%%%%%%%%%%%%%%%%%%%%%%%%%%%%%%%%%%%%%%%%%%%%%%%%%%%%%%%%%%%
%%%%%%%%%%%%%%%%%%%%%%%%%%%%%%%%%%%%%%%%%%%%%%%%%%%%%%%%%%%%%%%%%%%%%%%%%%%%%%%%%%%%%%%%%%%%%%%%%
%%%%%%%%%%%%%%%%%%%%%%%%%%%%%%%%%%%%%%%%%%%%%%%%%%%%%%%%%%%%%%%%%%%%%%%%%%%%%%%%%%%%%%%%%%%%%%%%%
%%%%%%%%%%%%%%%%%%%%%%%%%%%%%%%%%%%%%%%%%%%%%%%%%%%%%%%%%%%%%%%%%%%%%%%%%%%%%%%%%%%%%%%%%%%%%%%%%
%%%%%%%%%%%%%%%%%%%%%%%%%%%%%%%%%%%%%%%%%%%%%%%%%%%%%%%%%%%%%%%%%%%%%%%%%%%%%%%%%%%%%%%%%%%%%%%%%

\chapter{Evaluation}\label{chap:dx:eval}

%We evaluated our approach $\mathsf{DIR}$ (based on \textsc{Inv-QX} and \textsc{Inv-HS-Tree}) versus the standard technique $\mathsf{STD}$ (based on \textsc{QX} and \textsc{HS-Tree}) using a set of KBs created by automatic matching systems. Given two knowledge bases $\mo_i$ and $\mo_j$ a matching system outputs a set of alignments $M_{ij}$ which express correspondences between semantically related entities of $\mo_i$ and $\mo_j$. Each element of $M_{ij}$ is a tuple $\tuple{x_i,x_j,r,v}$, where $x_i \in Q(\mo_i)$, $x_j \in Q(\mo_j)$, $r\in \setof{\leftarrow, \leftrightarrow, \rightarrow}$ is a semantic relation and $v\in[0,1]$ is a confidence value. The latter expresses the probability of an alignment to be correct and $Q(\mo)$ denotes a set of all elements of $\mo$, e.g.\ names of predicates, for which alignments have to be computed. The result of the matching process is an aligned $\mo_{ij} = \mo_i \cup M_{ij} \cup \mo_j$.

\begin{table*}[bt]
\begin{center}
\scalebox{1}{
\begin{tabular}{|l|l|r|c|r||r|c|r|} \hline

 &  &  \multicolumn{3}{c||}{\textsc{HS-Tree}} & \multicolumn{3}{c|}{\textsc{Inv-HS-Tree}} \\ \cline{3-8}

{System} & {Scoring} &	{Time}	&	 {\#Queries}	&	{Reaction}	&	{Time}	&	{\#Queries}	&	{Reaction}	\\ \hline\hline
AgrMaker	&	ENT	&	19.62	&	1	&	19.10	&	20.83	&	1	&	18.23	\\
AgrMaker	&	SPL	&	36.04	&	4	&	8.76	&	36.03	&	4	&	8.28	\\ \hline
GOMMA-bk	&	ENT	&	18.34	&	1	&	18.07	&	14.47	&	1	&	12.68	\\
GOMMA-bk	&	SPL	&	18.95	&	3	&	6.15	&	19.51	&	3	&	5.91	\\ \hline
GOMMA-nobk	&	ENT	&	18.26	&	1	&	17.98	&	14.26	&	1	&	12.49	\\
GOMMA-nobk	&	SPL	&	18.74	&	3	&	6.08	&	19.47	&	3	&	5.89	\\ \hline
Lily	&	ENT	&	78.54	&	1	&	77.71	&	82.52	&	1	&	72.83	\\
Lily	&	SPL	&	82.94	&	4	&	20.23	&	115.24	&	4	&	26.93	\\ \hline
LogMap	&	ENT	&	6.60	&	1	&	6.30	&	13.41	&	1	&	11.36	\\
LogMap	&	SPL	&	6.61	&	2	&	3.17	&	15.13	&	2	&	6.82	\\ \hline
LogMapLt	&	ENT	&	14.85	&	1	&	14.54	&	12.89	&	1	&	11.34	\\
LogMapLt	&	SPL	&	15.59	&	3	&	5.05	&	17.45	&	3	&	5.29	\\ \hline
MapSSS	&	ENT	&	81.06	&	4	&	19.86	&	56.17	&	3	&	17.32	\\
MapSSS	&	SPL	&	88.32	&	5	&	17.26	&	77.59	&	6	&	12.43	\\
\hline
\end{tabular}
}
\vspace{5pt}
\end{center}
\caption[\textsc{HS-Tree} and \textsc{Inv-HS-Tree} Applied to Anatomy Benchmark]{\textsc{HS-Tree} and \textsc{Inv-HS-Tree} applied to Anatomy benchmark. Time is given in sec, \textbf{Scoring} stands for query selection strategy, \textbf{Reaction} is the average system reaction time between queries.} 
\label{tab:scalability}
\vspace{5pt}
\end{table*}

\begin{table*}[bt]
\small
\begin{center}
\scalebox{1}{
\begin{tabular}{|l@{\extracolsep{-2pt}}|l|l||l|r|c|r|c|c|}
\hline 
 {Ontology (Expressivity)} & {30 Diag} & {min $|\dxmD|$} & {Scoring} & Time & \#Queries & {Reaction} & {\#CC} & {CC} \\
\hline \hline
ldoa-conference-confof	&	 48.06 	&	16	&	ENT	&	11.6	&	6	&	1.5	&	430	&	0.003	\\	
 $\mathcal{SHIN(D)}$	&		&		&	SPL	&	11.3	&	7	&	1.6	&	365	&	0.004	\\	\hline
ldoa-cmt-ekaw	&	 42.28 	&	12	&	ENT	&	48.6	&	21	&	2.2	&	603	&	0.016	\\	
 $\mathcal{SHIN(D)}$	&		&		&	SPL	&	139.1	&	49	&	2.8	&	609	&	0.054	\\	\hline
mappso-confof-ekaw	&	 55.66 	&	10	&	ENT	&	10	&	5	&	1.9	&	341	&	0.007	\\	
 $\mathcal{SHIN(D)}$	&		&		&	SPL	&	31.6	&	13	&	2.3	&	392	&	0.021	\\	\hline
optima-conference-ekaw	&	 62.13 	&	19	&	ENT	&	16.8	&	5	&	2.6	&	553	&	0.008	\\	
 $\mathcal{SHIN(D)}$	&		&		&	SPL	&	16.1	&	8	&	1.9	&	343	&	0.012	\\	\hline
optima-confof-ekaw	&	 44.52 	&	16	&	ENT	&	24	&	20	&	1.1	&	313	&	0.014	\\	
 $\mathcal{SHIN(D)}$	&		&		&	SPL	&	17.6	&	10	&	1.7	&	501	&	0.006	\\	\hline
ldoa-conference-ekaw	&	 56.98 	&	16	&	ENT	&	56.7	&	35	&	1.5	&	253	&	0.053	\\	
 $\mathcal{SHIN(D)}$	&		&		&	SPL	&	25.5	&	9	&	2.7	&	411	&	0.016	\\	\hline
csa-conference-ekaw	&	 62.82 	&	17	&	ENT	&	6.7	&	2	&	2.8	&	499	&	0.003	\\	
 $\mathcal{SHIN(D)}$	&		&		&	SPL	&	22.7	&	8	&	2.7	&	345	&	0.02	\\	\hline
mappso-conference-ekaw	&	 70.46 	&	19	&	ENT	&	27.5	&	13	&	1.9	&	274	&	0.028	\\	
 $\mathcal{SHIN(D)}$	&		&		&	SPL	&	71	&	16	&	4.2	&	519	&	0.041	\\	\hline
ldoa-cmt-edas	&	15.47	&	16	&	ENT	&	24.7	&	22	&	1	&	303	&	0.008	\\	
$\mathcal{ALCOIN(D)}$	&		&		&	SPL	&	11.2	&	7	&	1.4	&	455	&	0.002	\\	\hline
csa-conference-edas	&	39.74	&	26	&	ENT	&	18.4	&	6	&	2.7	&	419	&	0.005	\\	
$\mathcal{ALCHOIN(D)}$	&		&		&	SPL	&	240.8	&	37	&	6.3	&	859	&	0.036	\\	\hline
csa-edas-iasted	&	377.36	&	20	&	ENT	&	1744.6	&	3	&	349.2	&	1021	&	1.3	\\	
$\mathcal{ALCOIN(D)}$	&		&		&	SPL	&	7751.9	&	8	&	795.5	&	577	&	11.5	\\	\hline
ldoa-ekaw-iasted	&	229.72	&	13	&	ENT	&	23871.5	&	9	&	1886	&	287	&	72.6	\\	
$\mathcal{SHIN(D)}$	&		&		&	SPL	&	20449	&	9	&	2100.1	&	517	&	37.2	\\	\hline
mappso-edas-iasted	&	293.74	&	27	&	ENT	&	18400.3	&	5	&	2028.3	&	723	&	17.8	\\	
$\mathcal{ALCOIN(D)}$	&		&		&	SPL	&	159299	&	11	&	13116.6	&	698	&	213.2	\\	\hline
\end{tabular}
}
\vspace{5pt}
\end{center}
\caption[Performance of Sequential Diagnosis Using Direct Computation of Diagnoses]{Sequential diagnosis using direct computation of diagnoses. \textbf{30 Diag} is the time required to find 30 minimal diagnoses, \textbf{min} $|\dxmd|$ is the cardinality of a minimum cardinality diagnosis, \textbf{Scoring} indicates the query selection strategy, \textbf{Reaction} is the average system reaction time between queries, \textbf{\#CC} number of consistency checks, \textbf{CC} gives average time needed for one consistency check. Time is given in sec.}
\label{tab:querysessionsinvhstree}
\vspace{5pt}
\end{table*}

We evaluated our approach $\mathsf{DIR}$ (based on \textsc{Inv-QX} and \textsc{Inv-HS-Tree}) versus the standard technique $\mathsf{STD}$~\cite{Shchekotykhin2012} (based on \textsc{QX} and \textsc{HS-Tree}) using a set of KBs created by automatic matching systems. Given two knowledge bases $\dxmo_i$ and $\dxmo_j$, a matching system outputs an \emph{alignment} $M_{ij}$ which is a set of \emph{correspondences} between semantically related entities of $\dxmo_i$ and $\dxmo_j$. Let $Q(\dxmo)$ denote the set of all elements of $\dxmo$ for which correspondences can be produced, i.e.\ names of predicates. Each correspondence is a tuple $\tuple{x_i,x_j,r,v}$, where $x_i \in Q(\dxmo_i)$, $x_j \in Q(\dxmo_j)$ and $x_i$, $x_j$ have the same arity, $r\in \setof{\leftarrow, \leftrightarrow, \rightarrow}$ is a logical operator and $v\in[0,1]$ is a confidence value. The latter expresses the probability of a correspondence to be correct.  Let $\overline{X}$ be a vector of distinct logical variables with a length equal to the arity of $x_i$, then each $\tuple{x_i,x_j,r,v} \in M_{ij}$ is translated to the axiom $\forall \overline{X}\, x_i(\overline{X})~r~x_j(\overline{X}).$ Let $\dxmo (M_{ij})$ denote the set of axioms resulting from such a translation for the alignment $M_{ij}$. Then the result of the matching process is an aligned KB $\dxmo_{ij} = \dxmo_i \cup \dxmo (M_{ij}) \cup \dxmo_j$.

%In the framework of OAEI 2011 the set of matchable elements of an ontology was limited to ontology classes (unary predicates) and properties (binary predicates).

The KBs considered in this section were created by ontology matching systems participating in the Ontology Alignment Evaluation Initiative (OAEI) 2011~\cite{Ferrara2011}.
Each matching experiment in the framework of OAEI represents a scenario in which a user obtains an  alignment $M_{ij}$ by means of some (semi)automatic tool for two real-world \emph{ontologies} $\dxmo_i$ and $\dxmo_j$. 
The latter are KBs expressed by the Web Ontology Language (OWL)~\cite{Grau2008a}  whose semantics is compatible with the $\mathcal{SROIQ}$ description logic (DL). This DL is a decidable fragment of first-order logic for which a number of effective reasoning methods exist~\cite{Baader2007}. 
%
%This design feature of OWL allows a direct translation of an ontology into a DL KB.
Note that, $\mathcal{SROIQ}$ is a member of a broad family of DL knowledge representation languages. All DL KBs considered in this evaluation are expressible in $\mathcal{SROIQ}$. 

%The first set of aligned ontologies $\dxmo_{ij}$ was drawn from ontologies of the Anatomy benchmark ($\mathsf{Anat}$) and the second set from Conference benchmark ($\mathsf{Conf}$). 

%First, we compared the performance of interactive $\mathsf{STD}$ and interactive $\mathsf{DIR}$ on incoherent aligned $\dxmo_{12}$ generated for Anatomy experiment of OAEI. 

The goal of the first experiment was to compare the performance of $\mathsf{STD}$ and $\mathsf{DIR}$ on a set of large, but diagnostically uncomplicated KBs, generated for the Anatomy experiment of OAEI.\footnote{All KBs and source code of programs used in the evaluation can be downloaded from \url{http://code.google.com/p/rmbd/wiki/DirectDiagnosis}. The tests were performed on Core i7, 64GB RAM running Ubuntu, Java 7 and HermiT as DL reasoner.}
In this experiment the matching systems had to find correspondences between two KBs describing the human and the mouse anatomy. $\dxmo_1$ (\texttt{Human}) and $\dxmo_2$ (\texttt{Mouse}) include 11545 and 4838 axioms, respectively, whereas the size of the alignment $M_{12}$ produced by different matchers varies between 1147 and 1461 correspondences. Seven matching systems produced a classifiable but incoherent output. One system generated a classifiable and coherent aligned KB. However, this system employes a built-in heuristic diagnosis engine which does not guarantee to produce minimal diagnoses. That is, some axioms are removed without reason. Four systems produced KBs which could not be processed by current reasoning systems (e.g.\ HermiT) since these KBs could not be classified within 2 hours. 

For testing the performance of our system we have to define the  correct output of sequential diagnosis which we call the target diagnosis $\dxmd_t$.
We assume that the only available knowledge is $M_{ij}$ together with $\dxmo_i$ and $\dxmo_j$. In order to measure the performance of the matching systems the organizers of OAEI provided a \emph{golden standard} alignment $M_t$ considered as correct. 
Nevertheless, we cannot assume that $M_t$ is explicitly available since the matching system would have used this information. W.r.t.\ the knowledge available, any minimal diagnosis
w.r.t.\ the DPI $\tuple{\dxmo (M_{ij}), \dxmo_i \cup \dxmo_j, \emptyset, \emptyset}$ (i.e.\ $\dxmo (M_{ij})$ is the KB and $\dxmo_i \cup \dxmo_j$ used as background theory)
% which is a subset of $\dxmo (M_{ij})$ with $\dxmo_i \cup \dxmo_j$ used as background theory 
can be selected as $\dxmd_t$. However, for every alignment we selected a minimal diagnosis as target diagnosis $\dxmd_t$ which is outside the golden standard. By this procedure we mimic cases where additional information can be acquired such that no correspondence of the golden standard is removed in order to establish coherence.  We stress that this setting is unfavorable for diagnosis since providing more information by exploiting the golden standard would reduce the number of queries to ask. Consequently, we limit the knowledge to $\dxmo_{ij}$ and use $\dxmo_{ij} \setminus \dxmd_t$ to answer the queries.

In particular, the selection of a target diagnosis $\dxmd_t$ for each $\dxmo_{ij}$ output by a matching system was done in two steps: (i) compute the set of all minimal diagnoses $\mathbf{AD}$ w.r.t.\ the correspondences which are not in the golden standard, i.e.\ $\dxmo (M_{ij}\setminus M_t)$, and use $\dxmo_i \cup \dxmo_j \cup \dxmo (M_{ij} \cap M_t)$ as background theory. The set of test cases are empty. I.e.\ the DPI is   
$\tuple{\dxmo (M_{ij} \setminus M_t), 
\dxmo_i \cup \dxmo_j \cup \dxmo (M_{ij} \cap M_t), \emptyset, \emptyset}$.
(ii) select $\dxmd_t$ randomly from $\mathbf{AD}$. 
The prior fault probabilities of axioms $\dxtax\in \dxmo(M_{ij})$ expressing correspondences were set to $1-v_{\dxtax}$ where $v_{\dxtax}$ is the confidence value provided by the matcher. 

%The knowledge base $\dxmo_{ij} \setminus \dxmd_t$ allowed us to answer the queries automatically.
 
The tests were performed for the mentioned seven incoherent alignments where the input DPI is $\tuple{\dxmo(M_{ij}), \dxmo_i \cup \dxmo_j, \emptyset, \emptyset}$ and the output is a minimal diagnosis. We tested $\mathsf{DIR}$ and $\mathsf{STD}$ with both query selection strategies \textsc{split-in-half} ($\mathsf{SPL}$) and \textsc{entropy} ($\mathsf{ENT}$) in order to evaluate the quality of fault probabilities based on confidence values. Moreover, for generating a query, the number of leading diagnoses was limited to $m=9$.

%
%****Need this note?**** Note that, we excluded the results of \texttt{CODI}, \texttt{CSA}, \texttt{MaasMtch}, \texttt{MapEVO} and \texttt{Aroma}, since the output of these systems was not classifiable within 2 hours, and \texttt{CODI} as it produced coherent alignments using a build-in diagnosis engine that computes a single diagnosis which minimality is not guaranteed. 
%%

The results of the first experiment are presented in Table~\ref{tab:scalability}. $\mathsf{DIR}$ computed $\dxmd_t$ within 36 sec.\ on average and slightly outperformed $\mathsf{STD}$ which required 36.7 sec. The number of asked queries was equal for both methods in all but two cases resulting from KBs produced by the \texttt{MapSSS} system. For these KBs, $\mathsf{DIR}$ required one query more using $\mathsf{ENT}$ and one query less using $\mathsf{SPL}$. In general, the results obtained for the Anatomy case show that $\mathsf{DIR}$ and $\mathsf{STD}$ have similar performance in both runtime and number of queries. Both $\mathsf{DIR}$ and $\mathsf{STD}$ identified the target diagnosis. 
Moreover, the confidence values provided by the matching systems appeared to be a good estimate for fault probabilities. Thus, in many cases $\mathsf{ENT}$ was able to find $\dxmd_t$ using one query only, whereas $\mathsf{SPL}$ used 4 queries on average. 

In the first experiment, the identification of the target diagnosis by sequential $\mathsf{STD}$ required the computation of 19 minimal conflicts on average. Moreover, the average size of a minimum cardinality diagnosis over all KBs in this experiment was 7. In the second experiment (see below), where $\mathsf{STD}$ is not applicable, the cardinality of the target diagnosis is significantly higher. 

%
%The second experiment was performed on KBs of the OAEI Conference benchmark. These KBs are particularly hard from the diagnostic point of view as the quality. In the first experiments the identification of the target diagnosis by $\mathsf{STD}$ required on average the computation of 19 minimal conflicts. Moreover, a minimum cardinality diagnosis comprised on average 7 elements for all KBs in the experiment. In the second experiment, the cardinality of the target diagnosis is significantly higher. 
%
%%
%In 11 of the 13 cases presented in Table~\ref{tab:querysessionsinvhstree} $\mathsf{STD}$ was in all three experiments unable to find any diagnoses. In the other two cases within 2 hours it succeeded in to find one minimal diagnosis for \texttt{csa-conference-ekaw} and nine for \texttt{ldoa-conference-confof}.
%% 
%Whereas $\mathsf{DIR}$ succeeded to find 30 diagnoses for each KB within time acceptable for interactive diagnosis settings.  Moreover, on average $\mathsf{DIR}$ was able to find 1 diagnosis in 8.9 sec.\, 9 diagnoses in 40.83 sec.\ and 30 diagnoses in 107.61 sec. This result shows that $\mathsf{DIR}$ is a stable and practically applicable method even in cases when a knowledge base comprises high-cardinality faults. In Conference case the minimum cardinality diagnoses comprised 18 elements on average.

The second experiment was performed on KBs of the OAEI Conference benchmark which turned out to be problematic for $\mathsf{STD}$. For these KBs we observed that the minimum cardinality diagnoses comprise 18 elements on average. In 11 of the 13 KBs of the second experiment (see Table~\ref{tab:querysessionsinvhstree}), $\mathsf{STD}$ was unable to find any diagnosis within 2 hours. In the other two cases $\mathsf{STD}$ succeeded  to find one minimal diagnosis for \texttt{csa-conference-ekaw} and nine for \texttt{ldoa-conference-confof}. However, $\mathsf{DIR}$ even succeeded to find 30 minimal diagnoses for each KB within time acceptable for interactive diagnosis settings.  Moreover, on average $\mathsf{DIR}$ was able to find 1 minimal diagnosis in 8.9 sec., 9 minimal diagnoses in 40.83 sec.\ and 30 minimal diagnoses in 107.61 sec.\ (see Column 2 of Table~\ref{tab:querysessionsinvhstree}).  This result shows that $\mathsf{DIR}$ is a stable and practically applicable method even in cases where a knowledge base comprises high-cardinality faults.

In the Conference experiment, we first selected the target diagnosis $\dxmd_t$ for each $\dxmo_{ij}$ just as it was done in the described Anatomy case. Next, we evaluated the performance of sequential $\mathsf{DIR}$ using both query selection methods.
The results of the experiment presented in Table~\ref{tab:querysessionsinvhstree} show that $\mathsf{DIR}$ found $\dxmd_t$ for each KB. On average $\mathsf{DIR}$ solved the problems more efficiently using $\mathsf{ENT}$ than $\mathsf{SPL}$ because also in the Conference case the confidence values provided a reasonable estimation of axiom fault probabilities. Only in three cases $\mathsf{ENT}$ required more queries than $\mathsf{SPL}$.

Moreover, the experiments show that the efficiency of debugging methods depends highly on the runtime of the underlying reasoner. For instance, in the hardest case consistency checking took $93.4\%$ of the total time whereas all other operations -- including construction of the search tree, generation and selection of queries -- took only $6.6\%$ of time. Consequently, sequential $\mathsf{DIR}$ requires only a small fraction of computation effort. Runtime improvements can be achieved by advances in reasoning algorithms or the reduction of the number of consistency checks. Currently, in order to generate a query, $\mathsf{DIR}$ requires $O(m * |\dxmd|\log(\frac{|\dxoo|}{|\dxmd|}))$ checks to find $m$ leading diagnoses. %The number of minimal diagnoses to be computed over all iterations of the sequential diagnosis session is, in general, unknown. However, in the worst case, it might require $O(md * |\dxmd|\log(\frac{|\dxoo|}{|\dxmd|}))$ consistency check to find the diagnosis $\dt$, where $md$ is the number of minimal diagnoses existing for the initial DPI.
 
A further source for improvements can be observed for the \texttt{ldoa-ekaw-iasted} ontology where both methods asked the same number of queries. %, but a session using $\mathsf{ENT}$ query selection method took considerably longer than the session using $\mathsf{SPL}$. 
In this case, a sequential diagnosis session using $\mathsf{ENT}$ query selection method required only half of the consistency checks $\mathsf{SPL}$ did. However, an average consistency check made in the session using $\mathsf{ENT}$ took almost twice as long as an average consistency check using $\mathsf{SPL}$.
The analysis of this ontology showed that there is a small subset of axioms (called ``hot spot'' in~\cite{Goncalves2012}) which made reasoning considerably harder. As practice shows, they can be resolved by suitable queries. This can be observed in the \texttt{ldoa-ekaw-iasted} case where $\mathsf{SPL}$ acquired appropriate test cases early and thereby found $\dxmd_t$ faster. Therefore, research and application of methods allowing fast identification of such hot spots might result in a significant improvement of diagnosis runtime. %

\chapter{Summary and Conclusions} \label{chap:dx:conc}

In this part, we presented a sequential diagnosis method for faulty KBs which is based on the direct computation of minimal diagnoses. 
We were able to reduce the number of consistency checks by avoiding the computation of minimized conflict sets and by computing just \emph{some} set of minimal diagnoses instead of a set of most probable diagnoses or a set of minimum cardinality diagnoses. 
%Moreover, we show that the inverse method can also be used with diagnosis discrimination algorithms, thus, allowing interactive ontology debugging. 
The presented evaluation results in Chapter~\ref{chap:dx:eval} indicate that the performance of the suggested sequential diagnosis system is either comparable with or outperforms the existing approach in terms of runtime and required number of queries in case a KB includes a large number of faults. 
%That is, the 
The scalability of the algorithms was demonstrated on a set of large KBs including thousands of axioms.

%%%%%%%%%%%%%%%
%\input{testcases}
%\input{optimal_diagnosis}
%\input{query}
%\input{query_old}
%\input{strategy}
%\input{impl}
%\input{eval}
%%%%%%%%%%%%%%%
\part{Epilog}
\label{part:Epilog}
In this part we provide a discussion of related work in Chapter~\ref{chap:RelatedWork},\footnote{Note that related work specific to topics addressed in Parts~\ref{part:JWS}-\ref{part:DX} is separately treated in these parts.}
%we talk about related work. 
summarize the contributions of this work in Chapter~\ref{chap:summary} and deal with our future work topics in Chapter~\ref{chap:future_work}.
\chapter{Related Work}
\label{chap:RelatedWork}
%kalyanpur, nikitina, meilicke, logmap
To the best of our knowledge no interactive KB debugging methods that ask a user automatically selected queries have been proposed to repair faulty (monotonic) KBs so far (except for our own previous works~\cite{ksgf2010, Shchekotykhin2012, Rodler2013, Shchekotykhin2014}).

Non-interactive debugging methods for KBs (ontologies) are introduced in~\cite{Schlobach2007,Kalyanpur.Just.ISWC07,friedrich2005gdm}. Ranking of diagnoses and proposing a ``best'' diagnosis is presented in~\cite{Kalyanpur2006}. This method uses a number of measures such as (a)~the frequency with which a formula appears in conflict sets, (b)~the impact on the KB in terms of its ``lost'' entailments when some formula is modified or removed, (c)~provenance information about the formula and (d)~syntactic relevance of a formula. All these measures are evaluated for each formula in a conflict set. The scores are then combined in a rank value which is associated with the corresponding formula. These ranks are then used by a modified hitting set tree algorithm that identifies diagnoses with a minimal rank. 
In this work no query generation and selection strategy is proposed if the intended diagnosis cannot be determined reliably with the given a-priori knowledge. In our work additional information is acquired until the minimal diagnosis with the intended semantics can be identified with confidence. 
In general, the work of~\cite{Kalyanpur2006} can be combined with the approaches presented in our work as ranks of logical formulas can be taken into account together with other observations for calculating the prior probabilities of minimal diagnoses (see Section~\ref{sec:prob_space_construction}).

The idea of selecting the next query based on certain query selection measures was exploited in the generation of decisions trees \cite{Quinlan1986} and for selecting measurements in the model-based diagnosis of circuits \cite{dekleer1987} (in both works, the minimal expected entropy measure was used). We extended these methods to query selection in the domain of KB debugging \cite{ksgf2010} and devised further query selection measures \cite{Shchekotykhin2012,Rodler2013}.  

An approach for the debugging of faulty aligned KBs (ontologies) was proposed by \cite{meilicke2011}. An aligned KB is the union of two KBs $\mo_1$ and $\mo_2$ and an alignment $A_{1,2}$ (which is properly formatted as a set of logical formulas, cf.\ Definition~18 in \cite{meilicke2011}). $A_{1,2}$ is
a set of correspondences (each with an associated automatically computed confidence value) produced by an automatic system (an ontology matcher) given $\mo_1$ and $\mo_2$ as inputs where each correspondence represents a (possible) semantic relationship between a term occurring in the first and a term occurring in the second input KB.
% results from the application of an automatic system called (ontology) matcher which gets $\mo_1$ and $\mo_2$ as inputs and generates a set of semantic correspondences where each correspondence represents a possible semantic relationship between a term occurring in the first and a term occurring in the second input KB. 
The goal of a debugging system for faulty aligned KBs is usually the determination of a subset of the alignment $A'_{1,2} \subset A_{1,2}$ such that the aligned KB using $A'_{1,2}$ is not faulty. In terms of our approaches, this corresponds to the setting $\mo := A_{1,2}$ and $\mb := \mo_1 \cup \mo_2$. We have already shown in \cite{Rodler2012OM,Shchekotykhin2012b} that our systems can also be applied for fault localization in aligned KBs. The work of \cite{meilicke2011} describes approximate algorithms for computing a ``local optimal diagnosis'' and complete methods to discover a ``global optimal diagnosis''. Optimality in this context refers to the maximum sum of confidences in the resulting repaired alignment $A'_{1,2}$. In contrast to our framework, diagnoses are determined automatically without support for user interaction. Instead, \cite{meilicke2011} demonstrates techniques for the manual revision of the alignment as a procedure \emph{independent} from debugging. 
Another difference to our approach is the way of detecting sources of faults. We rely on a divide-and-conquer algorithm \cite{junker04} for the identification of a minimal conflict set $C \subseteq A_{1,2}$ (in \cite{meilicke2011} $C$ is called a MIPS, cf.\ \cite{friedrich2005gdm,Schlobach2007}). In the worst case the method we use exhibits only $O(|C|*\log(|A_{1,2}|/|C|))$ calls of some function that performs a check for faults in a KB and internally uses a reasoner (in our case \textsc{isKBValid}, see Algorithm~\ref{algo:qx}). The ``shrink'' strategy applied in \cite{meilicke2011} (which is similar to the ``expand-and-shrink'' method used in \cite{Kalyanpur.Just.ISWC07}), on the other hand, requires a worst case number of $O(|A_{1,2}|)$ calls to such a function. Empirical evaluations and a theoretical analysis of the best and worst case complexity of the ``expand-and-shrink'' method compared to the divide-and-conquer method performed in \cite{Shchekotykhin2008} revealed that the latter is preferable over the former. It should be noted that a similar divide-and-conquer method as used in our work could most probably also be plugged into the system in \cite{meilicke2011} instead of the ``shrink'' method.
%we use and the ``shrink'' method 
%%used in \cite{meilicke2011} 
%performed in \cite{Shchekotykhin2008} show that the 
%\cite{meilicke2011} 

There are some ontology matchers which incorporate alignment repair features: %\cite{Jean-Mary2009,Ngo2012,DBLP:journals/jods/GiunchigliaYS07,ReulP10,conf/semweb/HuberSNM11}. 
CODI \cite{Huber2011}, YAM++ \cite{Ngo2012}, ASMOV \cite{Jean-Mary2009} and KOSIMap \cite{Reul2010}, for instance, 
%employ logic-based methods to refine a set of candidate correspondences, either to avoid inconsistencies or to eliminate unwanted or redundant entailments. This is accomplished means of a set of predefined ``anti-patterns'' w.r.t. pairs or sets of correspondences which must not occur in the aligned ontology.
employ logic-based techniques to search for a set of predefined ``anti-patterns'' which must not occur in the aligned ontology, either to avoid inconsistencies or incoherencies or to eliminate unwanted or redundant entailments. In case such a pattern is revealed, it is resolved by eliminating from the alignment some correspondences responsible for its occurrence.  
All the techniques incorporated in these matchers are distinct from the presented approaches in that they implement incomplete or approximate methods of alignment repair, i.e. not all alternative solutions to the alignment debugging problem are taken into account. As a consequence of this, on the one hand, the final alignment produced by these systems may still trigger faults in the aligned KB. On the other hand, a suboptimal solution may be found, e.g.\ in terms of the user-intended semantics w.r.t. the aligned ontology or other criteria such as alignment confidence or cardinality.

Another ontology matcher, LogMap~2 \cite{Jimenez-Ruiz2012a}, provides integrated debugging features and the opportunity for a user to interact during this process. However, the system is not really comparable with ours since it is very specialized and dedicated to the goal of producing a fault-free alignment. Concretely, there are at least two differences to our approach. First, LogMap 2 uses incomplete reasoning mechanisms in order to speed up the matching process. Hence, the output is not guaranteed to be fault-free. Second, the option for user interaction aims in fact at the revision of a set of correspondences, i.e.\ the sequential assessing of single correspondences as 'faulty' or 'correct'. Our approach, on the contrary, asks the user queries (i.e.\ \emph{entailments} of non-faulty parts of the KB).  

An interactive technique similar to our approaches was presented in~\cite{Nikitina2011}, where a user is successively asked single KB formulas (ontology axioms) in order to 
obtain a partition of a given ontology into a set of desired or correct and a set of undesired or incorrect formulas. 
Whereas our strategies aim at finding a parsimonious solution involving minimal change to the given faulty KB in order to repair it, the method proposed in~\cite{Nikitina2011} pursues a (potentially) more invasive approach to KB quality assurance, namely a (reasoner-supported) exhaustive manual inspection of (parts of) a KB.  
Given an inconsistent/incoherent KB, this technique starts from an empty set of desired formulas aiming at adding to this set only correct formulas of the KB which preserve consistency and coherency. Our approach, on the other hand, works its way forward the other way round in that it starts from the complete KB aiming at finding a minimal set of formulas to be deleted or modified which are responsible for the violation of the pre-specified requirements.
Another difference of our approach compared to the one suggested in \cite{Nikitina2011} is the type of queries asked to the user and the way these are selected. Our method allows for the generation of queries which are not explicit formulas in the KB, but implicit consequences of non-faulty parts of the KB. Besides, the set of selectable queries in our approach differs from one iteration to the next due to the changing set of leading diagnoses whereas queries (i.e.\ KB formulas) in \cite{Nikitina2011} are known in advance and the challenge is to figure out the best ordering of formulas to be assessed by the user. Whereas we apply mostly information theoretic measures (e.g.\ the minimal expected entropy in the set of leading diagnoses after a query has been answered), the authors in \cite{Nikitina2011} employ ``impact measures'' which, roughly speaking, indicate the number of automatically classifiable formulas in case of positive and, respectively, negative classification of a query (i.e.\ a particular formula).  

\chapter{Summary}
\label{chap:summary}
%\label{sec:Summary}
In this work we motivated why appropriate tool assistance is a must when it comes to 
%finding faults and adequately 
repairing faulty KBs. For, KBs that do not satisfy some minimal quality criteria such as logical consistency can make artificial intelligence applications relying on the domain knowledge modeled by this KB completely useless. In such a case, no meaningful reasoning or answering of queries about the domain is possible.

Non-interactive debugging systems published in research literature often cannot localize all possible faults (\emph{incompleteness}), suggest the deletion or modification of unnecessarily large parts of the KB (\emph{non-minimality}), return incorrect solutions which lead to a repaired KB not satisfying the imposed quality requirements (\emph{unsoundness}) or suffer from \emph{poor scalability} due to the inherent complexity of the KB debugging problem \cite{Stuckenschmidt2008}. Even if a system is complete and sound and considers only minimal solutions, there are generally exponentially many solution candidates to select one from. However, any two repaired KBs obtained from these candidates differ in their semantics in terms of entailments and non-entailments. Selection of just any of these repaired KBs might result in unexpected entailments, the loss of desired entailments or unwanted changes to the KB which in turn might cause unexpected new faults during the further development or application of the repaired KB. Also, manual inspection of a large set of solution candidates can be time-consuming (if not practically infeasible), tedious and error-prone since human beings are normally not capable of fully realizing the semantic consequences of deleting a set of formulas from a KB. 

To account for this issue, we evolved a comprehensive theory on which provably complete, sound and optimal (in terms of given probability information) interactive KB debugging systems can be built which suggest only minimal changes to repair a present KB. Interaction with a user is realized by asking the user queries. That is, a conjunction of logical formulas must be classified either as an intended or a non-intended entailment of the correct KB. To construct a query, only a minimal set of two solution candidates must be available. After the answer to a query is known, the search space for solutions is pruned. Iteration of this process until 
%a solution candidate has overwhelming probability or until 
there is only a single solution candidate left yields a (repaired) solution KB which features exactly the semantics desired and expected by the user.

We presented algorithms for the computation of minimal conflict sets, i.e.\ irreducible faulty subsets of the KB, and for the computation of minimal diagnoses, i.e.\ irreducible sets of KB formulas that must be properly modified or deleted in order to repair the KB. We combined these algorithms with methods that derive probabilities of diagnoses from meta information about faults (e.g.\ the outcome of a statistical analysis) to constitute a non-interactive debugging system for monotonic KBs which computes minimal diagnoses in best-first order. Building on the idea of this non-interactive method, we devised a complete and sound best-first algorithm for the interactive debugging of monotonic KBs that allows a user to take part in the debugging process in order to figure out the best solution.  
%find exactly the solution with the intended semantics. 

In order to integrate the new information collected by successive consultations of the user, the diagnoses computation in an interactive system must be regularly stopped. That is, there must be alternating phases, on the one hand for the further exploration of the solution space in order to gain new evidence for query generation and on the other hand for user interaction. To this end, we proposed two new strategies for the iterative computation of minimal diagnoses that exactly serve this purpose.
The first strategy, \textsc{staticHS}, takes advantage of an artificial fixation of the solution set which guarantees the monotonic reduction of the solution space independently of the asked queries, the given answers or other parameters of the algorithm. In this vein, the complexity of this algorithm is initially known and the maximum overhead compared to the non-interactive algorithm is polynomially bound.\footnote{This holds under the reasonable assumption that, in practice, a debugging session will involve only a polynomial number of queries to an interacting user. Recall that a user can abort the debugging session at any time and select the currently most probable diagnosis as their solution to the debugging problem.} On the downside, \textsc{staticHS} cannot optimally exploit the information given by the answered queries and thus cannot employ powerful methods that enable a more efficient pruning of the solution search space. 

Such powerful methods can be incorporated by the second suggested strategy, \textsc{dynamicHS}, the performance of which can be orders of magnitude better than the (initially fixed) performance of \textsc{staticHS} in the best case. That is, the ability to fully incorporate the information gained from user interaction might lead to a modified problem instance for which only a single (best) solution exists with only a small fraction of the time, space and user effort needed by \textsc{staticHS}. Moreover, the (exact) solution located by means of an interactive debugging session applying \textsc{dynamicHS} is generally a better (verified) solution than the (exact) solution found by use of \textsc{staticHS}.  
%the exact solution of the Interactive KB Debugging Problem might be found  
However, the complexity of \textsc{dynamicHS} depends to a great degree on which queries are generated and which input parameters are chosen and the worst case complexity is not initially bound as in case of \textsc{staticHS}. In the design of \textsc{dynamicHS} we put a particular emphasis on memory saving behavior which is manifested, for instance, by the manner how duplicate search tree paths are handled.

For selecting the best subsequent query in interactive debugging we first proposed and exhaustively analyzed two strategies: The ``split-in-half'' strategy prefers queries which allow eliminating a half of the leading diagnoses. The entropy-based strategy employs information theoretic concepts to exploit knowledge about the likelihood of formulas to be faulty. Based on the probability of a formula containing an error we can predict the (expected) information gain produced by a query result, enabling us to select the best subsequent query according to a one-step-lookahead entropy-based scoring function. 

In comprehensive experiments using real-world KBs we compared the entropy-based method with the ``split-in-half'' strategy and witnessed a significant reduction in the number of queries required to identify the correct diagnosis when the entropy-based method is applied. Depending on the quality of the given prior fault probabilities, the required number of queries could be reduced by up to 60\%. 
In order to evaluate the robustness of the entropy-based method we experimented with different prior fault probability distributions as well as different qualities of the prior probabilities. Furthermore, we investigated cases where knowledge about fault probabilities is missing or inaccurate. In case such knowledge is unavailable, the entropy-based methods ranks the diagnoses based on the number of syntax elements contained in a formula and the number of formulas in a diagnosis. Given that this is a reasonable guess (i.e.\ the sought diagnosis is not at the lower end of the diagnoses ranked by their prior probabilities), the entropy-based method outperformed ``split-in-half''. Moreover, even if the initial guess is not reasonable, the entropy-based method improves the accuracy of the probabilities as more questions are asked.
Furthermore, the applicability of the approach to real-world KBs containing thousands of formulas was demonstrated by an extensive set of evaluations.

We showed that unconditional reliance upon the entropy-based method might still be problematic in the presence of fault information that is considerably uncertain. For, the entropy-based strategy fully exploits and gains from the given fault information. In this vein, it proved to speed up the debugging procedure in the normal case. However, we found out in experiments that it might also have a negative impact on the performance in the bad case where the actual solution diagnosis is rated as highly improbable. As an alternative, one might prefer to rely on a tool (e.g.\ ``split-in-half'') which does not consider any fault information at all. In this case, however, possibly well-chosen information cannot be exploited, resulting again in inefficient debugging actions.

Minimal effort for the interacting user can be achieved if both the query selection method is chosen carefully and the provided fault information satisfies some minimum quality requirements. In particular, for deficient fault information and unfavorable strategy for query selection, we reported on cases where the overhead in terms of user effort exceeds 2000\% (!) in comparison to employing a more favorable query selection strategy. Unfortunately, assessment of the fault information is only possible a-poteriori (after the debugging session is finished and the correct solution is known).
To tackle this issue, we proposed a reinforcement learning strategy (RIO) which combines the benefits of the entropy-based and the ``'split-in-half' approaches, i.e.\ high potential (to perform well) and low risk (to perform badly). RIO continuously adapts its behavior depending on the performance achieved and in this vein minimizes the risk of integrating low-quality fault information into the debugging process.

The RIO approach makes interactive debugging practical even in scenarios where reliable fault estimates are difficult to obtain. Tested under various conditions, the RIO algorithm revealed good scalability and reaction time as well as superior average performance to both the entropy-based as well as the ``split-in-half'' strategy in all tested cases w.r.t.\ required amount of user interaction. Highest achieved savings of RIO as against the best other strategy amounted to more than 80\%. Further on, the performed evaluations provided evidence that for 100\% of the cases in the hardest (from the debugging point of view) class of faulty test KBs, RIO performed at least as good as the \emph{best} other strategy and in more than 70\% of these cases it even manifested superior behavior to the \emph{best} other strategy. Choosing RIO over other approaches can involve an improvement by the factor of up to 23, meaning that more than 95\% of user time and effort might be saved per debugging session.

Moreover, we came up with mechanisms for efficiently dealing with KB debugging problems involving high cardinality faults. In the standard interactive debugging approach described in the first parts of this work, the computation of queries is based on the generation of the set of \emph{most probable (or minimum cardinality)} leading diagnoses. By this postulation, certain quality guarantees about the output solution can be given. However, we learned that dropping this requirement can bring about substantial savings in terms of time and especially space complexity of interactive debugging, in particular in debugging scenarios where faulty KBs are (partly) generated as a result of the application of automatic systems, e.g.\ KB (ontology) learning or matching systems.%~\cite{Huber2011,Ngo2012,Jean-Mary2009,Reul2010,Jimenez-Ruiz2012a,meilicke2011}.
	
To cope with such situations, we proposed to base query computation on \emph{any} set of leading diagnoses using a ``direct'' method for diagnosis generation. Contrary to the standard method that exploits minimal conflict sets, this approach takes advantage of the duality between minimal diagnoses and minimal conflict sets and employs ``inverse'' algorithms to those used in the standard approach in order to determine minimal diagnoses directly from the DPI without the indirection via conflict sets.

We studied the application of this direct method to high cardinality faults in KBs and noticed that the number of required queries per debugging session is hardly affected for cases when the standard approach is also applicable. However, the direct method proved applicable and able to locate the correct solution diagnosis also in situations when the standard approach (albeit one that not yet incorporates the powerful search tree pruning techniques introduced in this work) is not due to time or memory issues.

We want to point out that this work is unique in that it provides an in-depth theoretical workup of the topic of interactive KB debugging which (to the best of our knowledge) cannot be found in such a detailed fashion in other works. Furthermore, this is the first work that gives precise definitions of the problems addressed in interactive KB debugging. Additionally, it is unique in that it features (new) algorithms that \emph{provably} solve these interactive KB debugging problems. To account for a tradeoff between solution quality and execution time, these algorithms are equipped with a feature to compute approximate solutions where the goodness of the approximation can be steered by the user. Another unique characteristic of this work is that it deals with an \emph{entire system} of algorithms that are required for the interactive debugging of monotonic KBs, considers and details all algorithms separately, analyzes their complexity, proves their correctness and demonstrates how all these algorithms are orchestrated to make up a full-fledged and provably correct interactive KB debugging system.
%covers and explains \emph{all aspects} of a full-fledged interactive KB debugging system for arbitrary monotonic KBs for which reasoning is decidable (for instance, iterative diagnosis cmputation).  
% for interactive debugging and proofs of the correctness and feasibility of a full-fledged interactive debugging system for arbitrary monotonic KBs for which reasoning is decidable. Correctness of all discussed algorithms is demonstrated and their complexity is analyzed. 
%\vspace{10pt}

\chapter{Future Work Topics}
\label{chap:future_work}
%\fixme{include short discussion on hierarchical reasoning and mention MORe by Horrocks}
%CONCLUSION + SUMMARY
%
%\noindent\textbf{Summary.}
%
%
%\subsubsection*{Topics for Future Work}
	%\label{sec:future_work}
This work has given rise to several questions we will elaborate on in our future work: 

%\noindent\textbf{Query Generation and Selection.}
\paragraph{Query Generation and Selection.}
%\begin{itemize}
	%\item Of course, these assumptions and methods of obtaining fault probabilities of syntactical elements and formulas are only \emph{some} possible ways of doing so. For example, one might argue that the ``authorship'' of a formula is somewhat not clearly defined. What if user $u_1$ has originally written formula $\tax$ and then user $u_2$ alters the formula to become $\tax'$? Who is the author of $\tax'$? $u_1$, $u_2$ or both? For whose fault probability computation should the renewed modification of $\tax'$ to $\tax''$ count? Questions like this one need to be discussed and maybe evaluations using real data need to be accomplished in order to find a practical answer; or perhaps to find out that completely different approaches turn out to be reasonable. This is a topic of our future work.
	%\item 
Our discussions of the presented query generation methods have revealed some drawbacks (cf.\ Chapter~\ref{chap:QueryGeneration}). Albeit being a fixed-parameter tractable problem as argued, the exponential time complexity regarding the number of leading diagnoses $|\mD|$ in case an optimal query 
	%with optimal information theoretic properties such as information gain 
	is desired is clearly an aspect that should be improved. This high complexity arises from the paradigm of computing an optimal query w.r.t.\ some measure $qsm()$ by calculating a (generally exponentially large) pool $\QP$ of queries in a first stage, whereupon the best query in $\QP$ according to $qsm()$ is filtered out in a second stage.
	
	A key to solving this issue is the use of a different paradigm that does not rely on the computation of the pool $\QP$. Instead, qualitative measures can be derived from quantitative measures that have been used in interactive debugging scenarios~\cite{Shchekotykhin2012, Rodler2013, ksgf2010}. These qualitative measures provide a way to estimate the $qsm()$ value of \emph{partial q-partitions}, i.e.\ ones where not all leading diagnoses have been assigned to the respective set in the q-partition yet. In this way a \emph{direct} search for a query with (nearly) optimal properties is possible. A similar strategy called CKK has been employed in~\cite{Shchekotykhin2012} for the information gain measure $qsm() := \mathsf{ENT()}$ (see Section~\ref{sec:query_selection_measures}). From such a technique we can expect to save a high number of reasoner calls. Because usually only a small subset of q-partitions included in a query pool (of exponential cardinality) is required to find a query with desirable properties if the search is implemented by means of a \emph{heuristic} that involves the exploration of seemingly favorable (potential) queries and (partial) q-partitions, respectively, first. 

Another shortcoming of the paradigm of query pool generation and subsequent selection of the best query is the extensive use of reasoning services which may be computationally expensive (depending on the given DPI). Instead of computing a set of common entailments $Q$ of a set of KBs $\mo_i^*$ first and consulting a reasoner to fill up the (q-)partition for $Q$ in order to test whether $Q$ is a query at all (see Chapter~\ref{chap:QueryGeneration}), the idea enabling a significant reduction of reasoner dependence is to compute some kind of \emph{canonical query} without a reasoner and use simple set comparisons to decide whether the associated partition is a q-partition. Guided by qualitative properties mentioned before, a search for such q-partition with desirable properties can be accomplished \emph{without reasoning at all}. Also, a set-minimal version of the optimal canonical query can be computed without reasoning aid. Only for the optional enrichment of the identified optimal canonical query by additional entailments and for the subsequent minimization of the enriched query, the reasoner may be employed. We will present strategies accounting for these ideas in the near future.

Another aspect that can be improved is that \emph{only one} minimized version of each query is computed by Algorithm~\ref{algo:query_gen}. That is, per q-partition $\Pt$, there might be some set-minimal queries which do not occur in the output set $\QP$. From the point of view of how well a query might be understood by an interacting user, of course not all minimized queries can be assumed equally good in general. For instance, consider the minimized queries $Q_4$ and $Q_{10}$ in Table~\ref{tab:queries_partitions} on page~\pageref{tab:queries_partitions}. Both are equally good regarding their q-partitions (just the sets $\dx{}$ and $\dnx{}$ are commuted), but most people will probably agree that $Q_4$ is much easier to comprehend from the logical point of view and thus much easier to answer.

Hence, in order to avoid a situation where a potentially best-understood query w.r.t.\ $\Pt$ is not included in $\QP$, the query minimization process (see Section~\ref{sec:minimization_of_queries}) might be adapted to take into account some information about faults the interacting user is prone to. This could be exploited to estimate how well this user might be able to understand and answer a query. For instance, given that the user frequently has problems to apply $\exists$ in a correct manner to express what they intend to express, but has never made any mistakes in formulating implications $\rightarrow$, then the query $Q_1 = \setof{\forall X\,p(X) \rightarrow q(X), r(a)}$ might be better comprehended than $Q_2 = \setof{\forall X \exists Y s(X,Y)}$. One way to achieve the finding of a well-understood query for some q-partition $\Pt$ is to run the query minimization \textsc{minQ} more than once, each time with a modified input (using a hitting set tree to accomplish this in a systematic manner -- cf.\ Chapter~\ref{chap:DiagnosisComputation}, where an analogue idea is used to compute different minimal conflict sets w.r.t.\ a DPI). In this way, different set-minimal queries for $\Pt$ can be identified and the process can be stopped when a suitable query is found. 

In order to come up with such a strategy, however, one must first gain insight into how well a user might understand certain logical formalisms and what properties make a query easy to comprehend from the logical perspective. It is planned to gather corresponding data about different users in the scope of a user study and to utilize the results to achieve a model of ``query hardness'' (by sticking to a similar overall methodology as used in \cite{Horridge2011b}) in order to come up with strategies for the determination of minimal queries that are easily understood. Note that such a model could also act as a guide how to specify the initial fault probabilities of syntactical elements that are used to obtain diagnoses probabilities (see Section~\ref{sec:DiagnosisProbabilitySpace}).

%\noindent\textbf{Incorporating A-Posteriori Probabilities into Diagnosis Search.} 
\paragraph{Incorporating A-Posteriori Probabilities into Diagnosis Search.}
As we discussed in Remark~\ref{rem:absolute_vs_relative_fault_tolerance_AND_initial_vs_updated_probabilities} on page~\pageref{rem:absolute_vs_relative_fault_tolerance_AND_initial_vs_updated_probabilities}, the a-priori ($p_{\mD,prio}()$) and the a-posteriori ($p_{\mD}()$) diagnoses probabilities might not only differ in terms of the probability values assigned to different diagnoses, but also in terms of the probability order of diagnoses. 
Incorporation of updated probabilities \emph{directly} into the hitting set tree algorithms to be used for the determination of leading diagnoses in the order prescribed by an updated probability measure is only possible if there is an additional update operator (besides Bayes' Theorem for adapting \emph{diagnoses} probabilities) that can be applied to \emph{formula} probabilities. For, the latter are exploited in the hitting set tree to assign probability weights to paths that are \emph{not yet diagnoses} (cf.\ $p_{nodes}()$ specified by Definition~\ref{def:p_node()} and the discussion of Formula~\ref{eq:path_prob_calc}) in order to guide the search for minimal diagnoses in best-first order. Updated \emph{diagnosis} probabilities are not helpful at all for this purpose. Devising a reasonable mechanism of updating formula probabilities seems to be hard mostly due to the lack of suitable data that might be collected during the debugging session to accomplish that. What would be imaginable during the debugging session is to try to learn something about the fault probability of \emph{syntactical elements} by examining the positive (all formulas are \emph{definitely} correct) and singleton negative (the single formula is \emph{definitely} incorrect) test cases. However, a drawback of such a strategy comes into effect when only syntactically very simple queries are used which is, for instance, the case in Example~\ref{example:query_computation} (see the definition of the \textsc{getEntailments} function there). From such queries not many useful insights concerning faulty syntactical elements might be gained. On the other hand, such queries are absolutely desirable from the point of view of how well a user might comprehend the formulas asked by the system. Hence, these two aspects seem to contradict each other. Still, it is a topic for future research to attempt to elaborate a solution for that issue.

%\noindent\textbf{Facilitation of More Informative User Answers.} 
\paragraph{Facilitation of More Informative User Answers.}
The debugging system described in this work is designed to get along with just a ``minimal'' feedback of a user regarding an asked query. That is, we assume the user' answer to a query $Q$ to be merely $\true$, i.e.\ each formula in $Q$ (or the conjunction of formulas in $Q$) must be entailed by the correct KB, or $\false$, i.e.\ at least one formula in $Q$ (or the conjunction of formulas in $Q$) must not be entailed by the correct KB. However, imagine a user being presented $Q$ and think of how they might proceed in order to come up with an answer to $Q$. The first observation is that, in order to respond by $\true$, a user must definitely scrutinize each single formula in $Q$ because otherwise they could never decide for sure whether the conjunction of all formulas in $Q$ is correct. Another observation is that a user might cease to go through the rest of the formulas in case they have already identified one that must not be an entailment of the desired KB. For, in this situation, the overall query $Q$ is already $\false$. This however indicates that at least one formula must be known to be correct or false whatever answer is given to $Q$. Therefore, we can usually expect a user to be able to give exactly this information, namely one formula in $Q$ that must be incorrect, additionally to answering by $\false$. This extra piece of information can be exploited to achieve better space and time efficiency in the context of diagnosis computation since knowing \emph{which} formula must definitely not be entailed gives more information that just a set of formulas of which we know that at least one among those is not entailed. 
Apart from that, there might be other pieces of additional information a user might be \emph{easily} able to give additionally to the ``minimal'' feedback we assume in this work. Proposing more efficient algorithms that exploit such tapes of additional information is on our future work agenda.
	 
%\noindent\textbf{Usage of ``Positive-Impact'' Queries in Combination with \textsc{dynamicHS}.}
\paragraph{Usage of ``Positive-Impact'' Queries in Combination with \textsc{dynamicHS}.}	
	As we discussed in Section~\ref{sec:OverviewAndIntuition} in the context of Algorithm~\ref{algo:inter_onto_debug} in dynamic mode, an added test case might give rise to some pruning steps as well as it might induce the construction of new subtrees (where ``new'' means that these would be no subtress of a hitting set tree w.r.t.\ the DPI not including this test case). The latter situation occurs when ``completely new'' minimal conflict sets (those that are in no subset-relationship with existing ones) are introduced by the addition of a test case. If this is the only impact of a test case, then this test case has only a negative influence on the time and space complexity of Algorithm~\ref{algo:inter_onto_debug} using \textsc{dynamicHS}. In other words, none of the invalidated minimal diagnoses (and no other nodes in the tree) are redundant, but all of them must additionally hit the set of ``completely new'' minimal conflict sets (in order to become diagnoses w.r.t.\ new DPI). Hence, in this case, the transition from one DPI to another including this test case results only in monotonic growth of the tree. If possible, such ``negative-impact test cases'' must be avoided. On the other hand, one must strive for the usage of ``positive-impact test cases'', i.e.\ those that only trigger tree pruning, but no tree expansion.  
Defining and studying properties that constitute such ``positive-impact test cases'' and ``negative-impact test cases'', respectively, and developing specialized algorithms for extracting exactly those types of queries that enable as substantial and effective pruning as possible in the context of \textsc{dynamicHS} is part of our already ongoing research. Note that a rough intuition of which properties make out a ``positive-impact test case'' is illustrated on the basis of an example in Section~\ref{sec:OverviewAndIntuition}.

%\noindent\textbf{Finding the Right Expert to Answer a Query in a Collaborative KB Development Setting.}
\paragraph{Finding the Right Expert to Answer a Query in a Collaborative KB Development Setting.}
As we mentioned in Chapter~\ref{chap:intro}, there are collaborative KB development projects such as the OBO Project\footnote{\url{http://obo.sourceforge.net}} and the NCI Thesaurus\footnote{\url{http://nciterms.nci.nih.gov/ncitbrowser}}, where many different people contribute to the specification of their knowledge in large KBs. In such a setting, it may be hard to decide who is the person that has the highest chance of being able to answer a concrete query correctly. The idea in such a scenario could be to use a combination of different measures such as educational level (e.g.\ professor versus PhD student) or hierarchy of contributors (e.g.\ senior user versus regular user), statistical information about past faults of a contributor (e.g., how many of the formulas originally authored by a person have been corrected by other persons of higher educational level) or provenance information regarding terms occurring in the query (who has authored most of the formulas in which these terms occur?) in order to learn an ``expert model'' and use it to devise some kind of recommender system \cite{jannach2010} that suggests which person to ask a particular query. 

Once established, such an expert model together with provenance information of KB formulas and other types of information discussed in Section~\ref{sec:prob_space_construction} could also be exploited when it comes to the definition of the fault information provided as input to our debugging system. An example of a system which enables the remote collaborative development of KBs (ontologies) and also provides logs of interesting usage data such as formula change logs and provenance information is Web Prot\'{e}g\'{e}~\cite{Tudorache2013}.

\paragraph{Studying the Performance of the Newly Proposed Iterative Diagnosis Computation Mechanisms.} 
We will conduct extensive experiments using faulty real-world KBs in order to assess the impact of the usage of the powerful search tree pruning techniques of the \textsc{dynamicHS} method or the guaranteed ``convergence'' towards the correct solution diagnosis of the \textsc{staticHS} in comparison to interactive debugging algorithms used in our previous works \cite{Shchekotykhin2012,Rodler2013,ksgf2010,Shchekotykhin2014}.    

%\noindent\textbf{Methods for Query Selection without Computation of Diagnoses.}
\paragraph{Methods for Query Selection without Computation of Diagnoses.}
 We are also working on ``conflict-based debugging'' methods that do not rely on the computation of leading diagnoses for query generation. Instead, queries might be generated directly from (minimal) conflict sets. Such methods might be used together with a boolean hitting set search tree (which was originally proposed by \cite{jiang2002} and optimized by \cite{Pill2012}) where the tree is regularly pruned using test cases such that tree branching is mostly or completely suppressed. In this manner, the tree remains small in size and all in all computes only a single diagnosis, i.e.\ the one consistent with all answered queries. Such an approach could be very space saving. Nevertheless, it is unclear whether the number of required queries and/or the computation time might increase. Implementing such an approach and answering these open questions is a topic on our future work agenda.  

%\noindent\textbf{Employing Advanced Reasoning Techniques to Increase Debugging Efficiency.} 
\paragraph{Employing Advanced Reasoning Techniques to Increase Debugging Efficiency.}
To cope with application contexts where reasoning is the main obstacle for efficient debugging, a plan for future work is to integrate advances reasoning techniques into our system. 

For example, a modular combination of reasoners \cite{Romero2012} might be adopted. In such a system there are two sound reasoners are combined where one (R1, e.g.\ HermiT \cite{Shearer2008}) is complete for the full logic $\mathcal{L}$ (e.g.\ OWL~2 \cite{Grau2008a}) and the other one (R2, e.g.\ ELK \cite{kazakov2014}) is complete for only a fragment $\mathcal{L}' \subset \mathcal{L}$ (e.g.\ the OWL~2~EL profile \cite{Grau2008a}), but $\mathcal{L}'$ can be handles much more efficiently by R2. The system in \cite{Romero2012} could be used to assign the bulk of the workload on R2 while relying on R1 only if necessary. 

Another interesting approach might be to employ techniques introduced in \cite{Goncalves2012} for detecting so-called ``hot spots'' in KBs which, when deleted from the KB, lead to much more efficient reasoning. Since reasoning in our approaches is mostly applied to fractions of the faulty KB, we could possibly benefit from such an approach. For instance, queries are entailments of a set of different non-faulty fractions $\mo^*_i = (\mo \setminus \md_i) \cup \mb \cup U_{\Tp}$ of the original KB. Now, given that a hot spot $H$ is included, say in $\mb \cup U_{\Tp}$, then we might well delete $H$ from this subset of $\mo^*_i$ and might still obtain meaningful queries. The reason is that $H$ does not include any formulas in $U_{\mD}$ (where $\mD$ is the set of leading diagnoses) which are essential for query computation from the diagnosis discrimination point of view. Formulas in $\mb \cup U_{\Tp}$, on the other hand, are included in all non-faulty fractions $\mo^*_i$ and thus do not directly serve the discrimination between diagnoses. Since $U_{\mD}$ might be much smaller in size than $\mb \cup U_{\Tp}$ in many scenarios (due to a usually small number of leading diagnoses in $\mD$), there might be a high chance for hot spots to be located in $\mb \cup U_{\Tp}$ rather than in $U_{\mD}$.

%\section{Conclusion}
%\label{sec:conclusion}
%%%%%%%%%%%%%%%%%%%%%%%%%%%%%%%%%%%%%%%%%%%%%%%%%%%%%%%%%%%%%%%%%%%%%%%%%%%%%%%%%%%%%%%%%%%%%% NOT COPIED start
%We have shown problems of state-of-the-art interactive ontology debugging strategies w.r.t.\ the usage of unreliable meta information. To tackle this issue, we proposed a learning strategy which combines the benefits of existing approaches, i.e. high potential and low risk. Depending on the performance of the diagnosis discrimination actions, the trust in the a-priori information is adapted. Tested under various conditions, our algorithm 
%revealed good scalability and reaction time as well as superior average performance to two common approaches in the field in all tested cases w.r.t. required user interaction. Highest achieved savings amounted to more than 80\% and user interaction overheads resulting from the wrong choice of strategy of up to 2300\% could be saved. In the hardest test cases, the new strategy was not only on average, but in 100\% of the test cases at least as good as the best other strategy. 
%%%%%%%%%%%%%%%%%%%%%%%%%%%%%%%%%%%%%%%%%%%%%%%%%%%%%%%%%%%%%%%%%%%%%%%%%%%%%%%%%%%%%%%%%%%%%% NOT COPIED end

%\addcontentsline{toc}{chapter}{References}
\bibliographystyle{alpha}
\bibliography{library}

\newcommand{\etalchar}[1]{$^{#1}$}
\begin{thebibliography}{KPSCG06}

\bibitem[ARW12]{Abreu2012}
Rui Abreu, Andr\'{e} Riboira, and Franz Wotawa.
\newblock {Constraint-based Debugging of Spreadsheets}.
\newblock In {\em CIbSE}, pages 1--14, 2012.

\bibitem[Baa03]{Baader2003c}
Franz Baader.
\newblock {Appendix: Description Logic Terminology}.
\newblock In Franz Baader, Diego Calvanese, Deborah~L. McGuinness, Daniele
  Nardi, and Peter~F. Patel-Schneider, editors, {\em Description Logic
  Handbook}, pages 485--495. Cambridge University Press, 2003.

\bibitem[BATJ91]{Bylander1991}
Tom Bylander, Dean Allemang, Michael Tanner, and John Josephson.
\newblock {The computational complexity of abduction}.
\newblock {\em Artificial Intelligence}, 49:25--60, 1991.

\bibitem[BBL05]{baader2005}
Franz Baader, Sebastian Brandt, and Carsten Lutz.
\newblock {Pushing the EL envelope}.
\newblock In {\em IJCAI}, pages 364--369, 2005.

\bibitem[BCM{\etalchar{+}}07]{Baader2007}
Franz Baader, Diego Calvanese, Deborah~L. McGuinness, Daniele Nardi, and
  Peter~F. Patel-Schneider, editors.
\newblock {\em The Description Logic Handbook: Theory, Implementation, and
  Applications}.
\newblock Cambridge University Press, 2007.

\bibitem[BKP12]{Baader2012}
Franz Baader, Martin Knechtel, and Rafael Penaloza.
\newblock {Context-dependent views to axioms and consequences of Semantic Web
  ontologies}.
\newblock {\em Web Semantics: Science, Services and Agents on the World Wide
  Web}, 12-13:22--40, April 2012.

\bibitem[BLHL{\etalchar{+}}01]{berners2001semantic}
Tim Berners-Lee, James Hendler, Ora Lassila, et~al.
\newblock {The Semantic Web}.
\newblock 2001.
\newblock \url{http://bit.ly/18ZvAXo}.

\bibitem[Bor96]{borgida1996}
Alex Borgida.
\newblock {On the relative expressiveness of description logics and predicate
  logics}.
\newblock {\em Artificial Intelligence}, 82(1-2):353--367, 1996.

\bibitem[BP08]{Baader2008}
Franz Baader and R.~Penaloza.
\newblock {Axiom Pinpointing in General Tableaux}.
\newblock {\em Journal of Logic and Computation}, 20(1):5--34, November 2008.

\bibitem[CFD93]{ConsoleFD93}
Luca Console, Gerhard Friedrich, and Daniele~Theseider Dupre.
\newblock {Model-Based Diagnosis Meets Error Diagnosis in Logic Programs}.
\newblock In {\em IJCAI}, pages 1494--1501, 1993.

\bibitem[CGT89]{Ceri1989a}
Stefano Ceri, Georg Gottlob, and Letizia Tanca.
\newblock {What you always wanted to know about Datalog (and never dared to
  ask)}.
\newblock {\em IEEE Transactions on Knowledge and Data Engineering}, I(1),
  1989.

\bibitem[Chu36]{church1936}
Alonzo Church.
\newblock An unsolvable problem of elementary number theory.
\newblock {\em American Journal of Mathematics}, pages 345--363, 1936.

\bibitem[CL73]{chang1973}
Chin-Liang Chang and Richard Char-Tung Lee.
\newblock {\em {Symbolic Logic and Mechanical Theorem Proving}}.
\newblock Academic Press Inc., 1973.

\bibitem[Coo71]{cook1971}
Stephen~A. Cook.
\newblock The complexity of theorem-proving procedures.
\newblock In {\em Proceedings of the third annual ACM symposium on Theory of
  computing}, pages 151--158. ACM, 1971.

\bibitem[CP71]{Ceraso71}
John Ceraso and Angela Provitera.
\newblock {Sources of error in syllogistic reasoning}.
\newblock {\em Cognitive Psychology}, 2(4):400--410, 1971.

\bibitem[CRV{\etalchar{+}}09]{Corcho09}
Oscar Corcho, Catherine Roussey, {Vilches Bl\'{a}zquez}, Luis Manuel, and Ivan
  P\'{e}rez.
\newblock {Pattern-based OWL Ontology Debugging Guidelines}.
\newblock In Eva Blomqvist, Kurt Sandkuhl, Francois Scharffe, and Vojtech
  Svatek, editors, {\em Workshop on Ontology Patterns (WOP 2009), collocated
  with the 8th International Semantic Web Conference (ISWC 2009).}, CEUR
  Workshop proceedings, pages 68--82, 2009.

\bibitem[DF95]{downey1995}
Rod~G. Downey and Michael~R. Fellows.
\newblock Fixed-parameter tractability and completeness {I}: Basic results.
\newblock {\em SIAM Journal on Computing}, 24(4):873--921, 1995.

\bibitem[dKW87]{dekleer1987}
Johan de~Kleer and Brian~C. Williams.
\newblock {Diagnosing multiple faults}.
\newblock {\em Artificial Intelligence}, 32(1):97--130, April 1987.

\bibitem[DQPS11]{Du2011}
Jianfeng Du, Guilin Qi, Jeff~Z. Pan, and Yi-Dong Shen.
\newblock {A Decomposition-Based Approach to OWL DL Ontology Diagnosis}.
\newblock In {\em Proceedings of 23rd IEEE International Conference on Tools
  with Artificial Intelligence}, pages 659--664. IEEE Press, November 2011.

\bibitem[Dur10]{durrett2010}
Rick Durrett.
\newblock {\em {Probability: Theory and Examples, Fourth Edition}}.
\newblock Cambridge University Press, 2010.

\bibitem[EFvH{\etalchar{+}}11]{Ferrara2011}
J\'{e}r\^{o}me Euzenat, Alfio Ferrara, Willem~Robert van Hage, Laura Hollink,
  Christian Meilicke, Andriy Nikolov, Dominique Ritze, Fran\c{c}ois Scharffe,
  Pavel Shvaiko, Heiner Stuckenschmidt, Ondrej Sv\'{a}b-Zamazal, and
  C\'{a}ssia~Trojahn dos Santos.
\newblock {Final results of the Ontology Alignment Evaluation Initiative 2011}.
\newblock In {\em Proceedings of the 6th International Workshop on Ontology
  Matching}, pages 1--29. CEUR-WS.org, 2011.

\bibitem[FFJS04]{Felfernig2004213}
Alexander Felfernig, Gerhard Friedrich, Dietmar Jannach, and Markus Stumptner.
\newblock Consistency-based diagnosis of configuration knowledge bases.
\newblock {\em Artificial Intelligence}, 152(2):213 -- 234, 2004.

\bibitem[FS05]{friedrich2005gdm}
Gerhard Friedrich and Kostyantyn Shchekotykhin.
\newblock {A General Diagnosis Method for Ontologies}.
\newblock In Yolanda Gil, Enrico Motta, Richard Benjamins, and Mark Musen,
  editors, {\em Proceedings of the 4th International Semantic Web Conference
  (ISWC 2005)}, pages 232--246. Springer, 2005.

\bibitem[FSW99]{Friedrich1999}
Gerhard Friedrich, Markus Stumptner, and Franz Wotawa.
\newblock {Model-based diagnosis of hardware designs}.
\newblock {\em Artif. Intell.}, 111(1-2):3--39, 1999.

\bibitem[FSZ11]{Felfernig2011}
Alexander Felfernig, Monika Schubert, and Christoph Zehentner.
\newblock {An efficient diagnosis algorithm for inconsistent constraint sets}.
\newblock {\em Artificial Intelligence for Engineering Design, Analysis and
  Manufacturing}, 26(1):53--62, June 2011.

\bibitem[GHM{\etalchar{+}}08]{Grau2008a}
Bernardo~Cuenca Grau, Ian Horrocks, Boris Motik, Bijan Parsia, Peter~F.
  Patel-Schneider, and Ulrike Sattler.
\newblock {OWL 2: The next step for OWL}.
\newblock {\em Web Semantics: Science, Services and Agents on the World Wide
  Web}, 6(4):309--322, November 2008.

\bibitem[GPS12]{Goncalves2012}
Rafael Goncalves, Bijan Parsia, and Ulrike Sattler.
\newblock {Performance Heterogeneity and Approximate Reasoning in Description
  Logic Ontologies}.
\newblock In {\em Proceedings of 11th International Semantic Web Conference
  (ISWC 2012)}, pages 82--98, 2012.

\bibitem[GSW89]{greiner1989correction}
Russell Greiner, Barbara~A. Smith, and Ralph~W. Wilkerson.
\newblock {A correction to the algorithm in Reiter's theory of diagnosis}.
\newblock {\em Artificial Intelligence}, 41(1):79--88, 1989.

\bibitem[HBP11]{Horridge2011b}
Matthew Horridge, Samantha Bail, and Bijan Parsia.
\newblock {The cognitive complexity of OWL justifications}.
\newblock In {\em Proceedings of the 10th International Semantic Web Conference
  (ISWC 2011)}. Springer, 2011.

\bibitem[HM01]{Haarslev2001}
Volker Haarslev and Ralf M\"{u}ller.
\newblock {RACER System Description}.
\newblock In Rajeev Gor\'{e}, Alexander Leitsch, and Tobias Nipkow, editors,
  {\em 1st International Joint Conference on Automated Reasoning}, volume 2083
  of {\em Lecture Notes in Computer Science}, pages 701--705, Berlin,
  Heidelberg, 2001. Springer Berlin Heidelberg.

\bibitem[Hor11]{Horridge2011a}
Matthew Horridge.
\newblock {\em Justification based Explanation in Ontologies}.
\newblock PhD thesis, University of Manchester, 2011.

\bibitem[HPS08]{Horridge2008}
Matthew Horridge, Bijan Parsia, and Ulrike Sattler.
\newblock {Laconic and Precise Justifications in OWL}.
\newblock In Amit Shet, Steffen Staab, Mike Dean, Massimo Paolucci, Diana
  Maynard, Timothy Finin, and Krishnaprasad Thirunarayan, editors, {\em
  Proceedings of the 7th International Semantic Web Conference (ISWC 2008)},
  volume 5318 of {\em Lecture Notes in Computer Science}, pages 323--338.
  Springer, 2008.

\bibitem[HPS09]{Horridge2009}
Matthew Horridge, Bijan Parsia, and Ulrike Sattler.
\newblock {Lemmas for Justifications in OWL}.
\newblock In {\em Proceedings of the 22nd Workshop of Description Logics
  DL2009}. CEUR Workshop Proceedings, 2009.

\bibitem[HPS10]{Horridge2010}
Matthew Horridge, Bijan Parsia, and Ulrike Sattler.
\newblock {Justification Oriented Proofs in OWL}.
\newblock In {\em Proceedings of the 9th International Semantic Web Conference
  (ISWC 2010)}. Springer, 2010.

\bibitem[HPS12a]{Horridge2012b}
Matthew Horridge, Bijan Parsia, and Ulrike Sattler.
\newblock {Extracting justifications from BioPortal ontologies}.
\newblock In {\em Proceedings of the 11th International Semantic Web Conference
  (ISWC 2012)}, pages 287--299, 2012.

\bibitem[HPS12b]{Horridge2012}
Matthew Horridge, Bijan Parsia, and Ulrike Sattler.
\newblock {Justification Masking in Ontologies}.
\newblock In {\em Thirteenth International Conference on the Principles of
  Knowledge Representation and Reasoning}, 2012.

\bibitem[HSNM11]{Huber2011}
Jakob Huber, Timo Sztyler, Jan Noessner, and Christian Meilicke.
\newblock {CODI: Combinatorial Optimization for Data Integration - Results for
  OAEI 2011}.
\newblock In {\em Proceedings of the 6th International Workshop on Ontology
  Matching}, 2011.

\bibitem[JL99]{Johnson1999}
Philip~N. Johnson-Laird.
\newblock {Deductive reasoning}.
\newblock {\em Annual review of psychology}, 50:109--135, 1999.

\bibitem[JL02]{jiang2002}
Yun-fei Jiang and Li~Lin.
\newblock {Computing the minimal hitting sets with binary HS-tree}.
\newblock {\em Journal of software}, 13(12):2267--2274, 2002.

\bibitem[JMSK09]{Jean-Mary2009}
Yves~R. Jean-Mary, E.~Patrick Shironoshita, and Mansur~R. Kabuka.
\newblock {Ontology Matching with Semantic Verification.}
\newblock {\em Web Semantics: Science, Services and Agents on the World Wide
  Web}, 7(3):235--251, September 2009.

\bibitem[JRG11]{Jimenez-Ruiz2011}
Ernesto Jim\'{e}nez-Ruiz and Bernardo~Cuenca Grau.
\newblock {Logmap: Logic-based and scalable ontology matching}.
\newblock In {\em Proceedings of the 10th International Semantic Web Conference
  (ISWC 2011)}, pages 273--288. Springer, 2011.

\bibitem[JRGZH12]{Jimenez-Ruiz2012a}
Ernesto Jim\'{e}nez-Ruiz, Bernardo~Cuenca Grau, Yujiao Zhou, and Ian Horrocks.
\newblock {Large-scale interactive ontology matching: Algorithms and
  implementation}.
\newblock In {\em Proceedings of 20th European Conference on Artificial
  Intelligence (ECAI2012)}, pages 444--449, 2012.

\bibitem[Jun04]{junker04}
Ulrich Junker.
\newblock {QUICKXPLAIN: Preferred Explanations and Relaxations for
  Over-Constrained Problems}.
\newblock In Deborah~L. McGuinness and George Ferguson, editors, {\em
  Proceedings of the Nineteenth National Conference on Artificial Intelligence,
  Sixteenth Conference on Innovative Applications of Artificial Intelligence},
  volume~3, pages 167--172. AAAI Press / The MIT Press, 2004.

\bibitem[JZFF10]{jannach2010}
Dietmar Jannach, Markus Zanker, Alexander Felfernig, and Gerhard Friedrich.
\newblock {\em Recommender Systems: An Introduction}.
\newblock Cambridge University Press, New York, NY, USA, 1st edition, 2010.

\bibitem[Kal06]{Kalyanpur2006a}
Aditya Kalyanpur.
\newblock {\em {Debugging and Repair of OWL Ontologies}}.
\newblock PhD thesis, University of Maryland, College Park, 2006.

\bibitem[Kar72]{karp1972}
Richard~M. Karp.
\newblock Reducibility among combinatorial problems.
\newblock {\em Complexity of Computer Computations}, pages 85--103, 1972.

\bibitem[Kaz08]{Kazakov2008}
Yevgeny Kazakov.
\newblock {SRIQ and SROIQ are harder than SHOIQ}.
\newblock In {\em Proceedings of the 21st Workshop of Description Logics
  DL2008}, 2008.

\bibitem[KK06]{kreuzer2006}
Martin Kreuzer and Stefan K\"uhling.
\newblock {\em Logik f\"ur Informatiker}.
\newblock Pearson Studium, M\"unchen, Germany, 2006.

\bibitem[KKLO86]{kk1986}
Narendra Karmarkar, Richard~M. Karp, George~S. Lueker, and Andrew~M. Odlyzko.
\newblock {Probabilistic analysis of optimum partitioning}.
\newblock {\em Journal of Applied Probability}, 23(3):626--645, 1986.

\bibitem[KKS14]{kazakov2014}
Yevgeny Kazakov, Markus Kr{\"o}tzsch, and Franti{\v{s}}ek Siman{\v{c}}{\'\i}k.
\newblock {The incredible ELK}.
\newblock {\em Journal of automated reasoning}, 53(1):1--61, 2014.

\bibitem[Kor98]{Korf1998}
Richard~E. Korf.
\newblock {A complete anytime algorithm for number partitioning}.
\newblock {\em Artificial Intelligence}, 106(2):181--203, December 1998.

\bibitem[KPHS07]{Kalyanpur.Just.ISWC07}
Aditya Kalyanpur, Bijan Parsia, Matthew Horridge, and Evren Sirin.
\newblock {Finding all Justifications of OWL DL Entailments}.
\newblock In Karl Aberer, Key-Sun Choi, Natasha~F. Noy, Dean Allemang, Kyung-Il
  Lee, Lyndon J.~B. Nixon, Jennifer Golbeck, Peter Mika, Diana Maynard,
  Riichiro Mizoguchi, Guus Schreiber, and Philippe Cudr\'{e}-Mauroux, editors,
  {\em The Semantic Web, 6th International Semantic Web Conference, 2nd Asian
  Semantic Web Conference, ISWC 2007 + ASWC 2007}, volume 4825 of {\em LNCS},
  pages 267--280, Berlin, Heidelberg, November 2007. Springer Verlag.

\bibitem[KPS{\etalchar{+}}06]{Kalyanpur2006b}
Aditya Kalyanpur, Bijan Parsia, Evren Sirin, Bernardo~Cuenca Grau, and James
  Hendler.
\newblock {Swoop: A Web Ontology Editing Browser}.
\newblock {\em J. Web Sem.}, 4(2):144--153, 2006.

\bibitem[KPSCG06]{Kalyanpur2006}
Aditya Kalyanpur, Bijan Parsia, Evren Sirin, and Bernardo Cuenca~Grau.
\newblock {Repairing Unsatisfiable Concepts in OWL Ontologies}.
\newblock In York Sure and John Domingue, editors, {\em The Semantic Web:
  Research and Applications, 3rd European Semantic Web Conference, ESWC 2006},
  volume 4011 of {\em Lecture Notes in Computer Science}, pages 170--184,
  Berlin, Heidelberg, 2006. Springer.

\bibitem[KPSH05]{kalyanpur2005}
Aditya Kalyanpur, Bijan Parsia, Evren Sirin, and James Hendler.
\newblock {Debugging Unsatisfiable Classes in OWL Ontologies}.
\newblock {\em Web Semantics: Science, Services and Agents on the World Wide
  Web}, 3(4):268--293, 2005.

\bibitem[MB88]{MuggletonBu88}
Stephen Muggleton and Wray~L. Buntine.
\newblock {Machine Invention of First-order Predicates by Inverting
  Resolution}.
\newblock In J~Laird, editor, {\em Proceedings of the 5th International
  Conference on Machine Learning ({ICML'88})}, pages 339--352. Morgan Kaufmann,
  1988.

\bibitem[Mei11]{meilicke2011}
Christian Meilicke.
\newblock {\em {Alignment Incoherence in Ontology Matching}}.
\newblock PhD thesis, Universit\"at Mannheim, 2011.

\bibitem[Men09]{mendelson2009}
Elliott Mendelson.
\newblock {\em {Introduction to Mathematical Logic, Fifth Edition}}.
\newblock CRC Press, 2009.

\bibitem[MPSP09]{Motik2009a}
Boris Motik, Peter~F. Patel-Schneider, and Bijan Parsia.
\newblock {OWL 2 Web Ontology Language Structural Specification and
  Functional-Style Syntax}.
\newblock {\em W3C recommendation}, pages 1--133, 2009.

\bibitem[MS72]{meyer1972}
Albert~R. Meyer and Larry~J. Stockmeyer.
\newblock The equivalence problem for regular expressions with squaring
  requires exponential space.
\newblock In {\em 13th Annual Symposium on Switching and Automata Theory},
  pages 125--129. IEEE, 1972.

\bibitem[MS09]{MeilickeStuck2009}
Christian Meilicke and Heiner Stuckenschmidt.
\newblock {An Efficient Method for Computing Alignment Diagnoses}.
\newblock In {\em Proceedings of the 3rd International Conference on Web
  Reasoning and Rule Systems}, pages 182--196. Springer-Verlag, 2009.

\bibitem[MSH09]{Motik2009}
Boris Motik, Rob Shearer, and Ian Horrocks.
\newblock {Hypertableau Reasoning for Description Logics}.
\newblock {\em Journal of Artificial Intelligence Research}, 36(1):165--228,
  2009.

\bibitem[MST07]{Meilicke2007}
Christian Meilicke, Heiner Stuckenschmidt, and Andrei Tamilin.
\newblock {Repairing Ontology Mappings}.
\newblock {\em Proceedings of the 22nd National Conference on Artificial
  intelligence - AAAI'07}, pages 1408--1413, 2007.

\bibitem[MST08]{Meilicke2008a}
Christian Meilicke, Heiner Stuckenschmidt, and Andrei Tamilin.
\newblock {Reasoning Support for Mapping Revision}.
\newblock {\em Journal of Logic and Computation}, 19(5):807--829, August 2008.

\bibitem[Mug95]{Muggleton1995}
Stephen Muggleton.
\newblock {Inverse entailment and Progol 1 Introduction}.
\newblock {\em New Generation Computing, Special issue on Inductive Logic
  Programming}, 13(3-4):245--286, 1995.

\bibitem[NB12]{Ngo2012}
Duyhoa Ngo and Zohra Bellahsene.
\newblock {YAM++ - A combination of graph matching and machine learning
  approach to ontology alignment task}.
\newblock {\em Journal of Web Semantics - The Semantic Web Challenge 2011
  Special Issue}, 2012.

\bibitem[NCLM06]{Noy2006a}
Natalya~F. Noy, A.~Chugh, W.~Liu, and Mark~A. Musen.
\newblock {A framework for ontology evolution in collaborative environments}.
\newblock In {\em Proceedings of the 5th International Semantic Web Conference
  (ISWC 2006)}, 2006.

\bibitem[NPQW13]{Nica2013}
Iulia Nica, Ingo Pill, Thomas Quaritsch, and Franz Wotawa.
\newblock {The route to success: A performance comparison of diagnosis
  algorithms}.
\newblock In {\em Proceedings of the Twenty-Third international Joint
  Conference on Artificial Intelligence}, pages 1039--1045, 2013.

\bibitem[NRG12]{Nikitina2011}
Nadeschda Nikitina, Sebastian Rudolph, and Birte Glimm.
\newblock {Interactive Ontology Revision}.
\newblock {\em Web Semantics: Science, Services and Agents on the World Wide
  Web}, 12-13:118--130, 2012.

\bibitem[NSD{\etalchar{+}}00]{Noy2000}
Natalya~F. Noy, Michael Sintek, Stefan Decker, Monica Crub\'{e}zy, Ray~W.
  Fergerson, and Mark~A. Musen.
\newblock {Creating Semantic Web Contents with Prot\'{e}g\'{e}-2000}.
\newblock {\em IEEE Intelligent Systems}, 16(2):60--71, 2000.

\bibitem[PQ12]{Pill2012}
Ingo Pill and Thomas Quaritsch.
\newblock {Optimizations for the Boolean Approach to Computing Minimal Hitting
  Sets}.
\newblock In {\em Proceedings of the 20th European Conference on Artificial
  Intelligence}, pages 648--653, 2012.

\bibitem[PSHH{\etalchar{+}}04]{patel2004owl}
Peter~F. Patel-Schneider, Patrick Hayes, Ian Horrocks, et~al.
\newblock {OWL Web Ontology Language Semantics and Abstract Syntax}.
\newblock {\em W3C recommendation}, 10, 2004.

\bibitem[PSK05]{Parsia2005}
Bijan Parsia, Evren Sirin, and Aditya Kalyanpur.
\newblock {Debugging OWL ontologies}.
\newblock In Allan Ellis and Tatsuya Hagino, editors, {\em Proceedings of the
  14th international conference on World Wide Web}, pages 633--640. ACM Press,
  May 2005.

\bibitem[PW03]{Peischl2003}
Bernhard Peischl and Franz Wotawa.
\newblock {Model-Based Diagnosis or Reasoning from First Principles}.
\newblock {\em IEEE Intelligent Systems}, 18:32--37, 2003.

\bibitem[Qui86]{Quinlan1986}
John~Ross Quinlan.
\newblock {Induction of Decision Trees}.
\newblock {\em Machine Learning}, 1(1):81--106, 1986.

\bibitem[RCVB09]{Roussey2009}
Catherine Roussey, Oscar Corcho, and Luis~Manuel Vilches-Bl\'{a}zquez.
\newblock {A catalogue of OWL ontology antipatterns}.
\newblock In {\em International Conference On Knowledge Capture}, pages
  205--206, Redondo Beach, California, USA, 2009. ACM.

\bibitem[RDH{\etalchar{+}}04]{Rector2004}
Alan Rector, Nick Drummond, Matthew Horridge, Jeremy Rogers, Holger Knublauch,
  Robert Stevens, Hai Wang, and Chris Wroe.
\newblock {OWL Pizzas: Practical Experience of Teaching OWL-DL: Common Errors
  \& Common Patterns}.
\newblock In Enrico Motta, Nigel~R. Shadbolt, Arthur Stutt, and Nick Gibbins,
  editors, {\em Engineering Knowledge in the Age of the SemanticWeb 14th
  International Conference, EKAW 2004}, pages 63--81, Whittenbury Hall, UK,
  2004. Springer.

\bibitem[Rei87]{Reiter87}
Raymond Reiter.
\newblock {A Theory of Diagnosis from First Principles}.
\newblock {\em Artificial Intelligence}, 32(1):57--95, 1987.

\bibitem[RGH12]{Romero2012}
Ana~Armas Romero, Bernardo~Cuenca Grau, and Ian Horrocks.
\newblock {MORe: Modular combination of OWL reasoners for ontology
  classification}.
\newblock In {\em Proceedings of the 11th International Semantic Web Conference
  (ISWC 2012)}, 2012.

\bibitem[RN10]{russellnorvig2003}
Stuart~J. Russell and Peter Norvig.
\newblock {\em Artificial Intelligence: A Modern Approach}.
\newblock Pearson Education, 3rd edition, 2010.

\bibitem[Rod15]{Rodler2015}
Patrick Rodler.
\newblock {A Theory of Interactive Debugging of Knowledge Bases in Monotonic
  Logics}.
\newblock Master's thesis, Alpen-Adria Universit\"at Klagenfurt, 2015.

\bibitem[RP10]{Reul2010}
Quentin Reul and Jeff~Z. Pan.
\newblock {KOSIMap: Use of Description Logic Reasoning to Align Heterogeneous
  Ontologies}.
\newblock In Volker Haarslev, David Toman, and Grant Weddell, editors, {\em
  Proceedings of the 23rd International Workshop on Description Logics DL2010},
  pages 489--500. CEUR Workshop Proceedings, 2010.

\bibitem[RSFF11]{Rodler2011}
Patrick Rodler, Kostyantyn Shchekotykhin, Philipp Fleiss, and Gerhard
  Friedrich.
\newblock {Balancing Brave and Cautious Query Strategies in Ontology
  Debugging}.
\newblock In Tudor~Groza {Vit Novacek, Zhisheng Huang}, editor, {\em
  Proceedings of the Joint Workshop on Knowledge Evolution and Ontology
  Dynamics 2011 (EvoDyn2011)}, Bonn, Germany, 2011. CEUR Workshop Proceedings.

\bibitem[RSFF12]{Rodler2012OM}
Patrick Rodler, Kostyantyn Shchekotykhin, Philipp Fleiss, and Gerhard
  Friedrich.
\newblock {RIO: Minimizing User Interaction in Debugging of Aligned
  Ontologies}.
\newblock In {\em Proceedings of the 7th International Workshop on Ontology
  Matching (OM-2012)}, 2012.

\bibitem[RSFF13]{Rodler2013}
Patrick Rodler, Kostyantyn Shchekotykhin, Philipp Fleiss, and Gerhard
  Friedrich.
\newblock {RIO: Minimizing User Interaction in Ontology Debugging}.
\newblock In Wolfgang Faber and Domenico Lembo, editors, {\em Web Reasoning and
  Rule Systems}, volume 7994 of {\em Lecture Notes in Computer Science}, pages
  153--167. Springer Berlin Heidelberg, 2013.

\bibitem[SE13]{Shvaiko2012}
Pavel Shvaiko and J\'{e}r\^{o}me Euzenat.
\newblock {Ontology matching: State of the art and future challenges}.
\newblock {\em IEEE Transactions on Knowledge and Data Engineering},
  25(1):158--176, 2013.

\bibitem[SEA{\etalchar{+}}02]{Sure2002}
York Sure, Michael Erdmann, Juergen Angele, Steffen Staab, Rudi Studer, and
  Dirk Wenke.
\newblock {OntoEdit: Collaborative Ontology Development for the Semantic Web}.
\newblock In {\em Proceedings of the 1st International Semantic Web Conference
  (ISWC 2002)}, pages 221--235, 2002.

\bibitem[Set12]{settles2012}
Burr Settles.
\newblock {\em {Active Learning}}.
\newblock Morgan and Claypool Publishers, 2012.

\bibitem[SF10]{ksgf2010}
Kostyantyn Shchekotykhin and Gerhard Friedrich.
\newblock {Query strategy for sequential ontology debugging}.
\newblock In Peter~F. Patel-Schneider, Pan Yue, Pascal Hitzler, Peter Mika,
  Zhang Lei, Jeff Pan, Ian Horrocks, and Birte Glimm, editors, {\em Proceedings
  of the 9th International Semantic Web Conference (ISWC 2010)}, pages
  696--712, Shanghai, China, 2010.

\bibitem[SFFR12]{Shchekotykhin2012}
Kostyantyn Shchekotykhin, Gerhard Friedrich, Philipp Fleiss, and Patrick
  Rodler.
\newblock {Interactive Ontology Debugging: Two Query Strategies for Efficient
  Fault Localization}.
\newblock {\em Web Semantics: Science, Services and Agents on the World Wide
  Web}, 12-13:88--103, 2012.

\bibitem[SFJ08]{Shchekotykhin2008}
Kostyantyn Shchekotykhin, Gerhard Friedrich, and Dietmar Jannach.
\newblock {On Computing Minimal Conflicts for Ontology Debugging}.
\newblock In {\em MBS 2008 - Workshop on Model-Based Systems}, 2008.

\bibitem[SFRF12]{Shchekotykhin2012b}
Kostyantyn Shchekotykhin, Philipp Fleiss, Patrick Rodler, and Gerhard
  Friedrich.
\newblock {Direct computation of diagnoses for ontology alignment}.
\newblock In Pavel Shvaiko, J\'{e}r\^{o}me Euzenat, Anastasios Kementsietsidis,
  Ming Mao, Natasha Noy, and Heiner Stuckenschmidt, editors, {\em Proceedings
  of the 7th International Workshop on Ontology Matching (OM2012)}, pages
  244--245, Boston, MA USA, 2012. CEUR Workshop Proceedings.

\bibitem[SFRF14a]{Shchekotykhin2014a}
Kostyantyn Shchekotykhin, Gerhard Friedrich, Patrick Rodler, and Philipp
  Fleiss.
\newblock {A direct approach to sequential diagnosis of high cardinality faults
  in knowledge bases}.
\newblock In {\em DX 2014 - 25th International Workshop on Principles of
  Diagnosis (DX 2014)}, 2014.

\bibitem[SFRF14b]{Shchekotykhin2014b}
Kostyantyn Shchekotykhin, Gerhard Friedrich, Patrick Rodler, and Philipp
  Fleiss.
\newblock {Interactive Ontology Debugging using Direct Diagnosis}.
\newblock In Patrick Lambrix, Guilin Qi, Matthew Horridge, and Bijan Parsia,
  editors, {\em Proceedings of the Third International Workshop on Debugging
  Ontologies and Ontology Mappings (WoDOOM14)}. CEUR Workshop Proceedings,
  2014.

\bibitem[SFRF14c]{Shchekotykhin2014}
Kostyantyn Shchekotykhin, Gerhard Friedrich, Patrick Rodler, and Philipp
  Fleiss.
\newblock {Sequential diagnosis of high cardinality faults in knowledge-bases
  by direct diagnosis generation}.
\newblock In {\em Proceedings of the 21st European Conference on Artificial
  Intelligence (ECAI 2014)}. IOS Press, 2014.

\bibitem[Sha48]{shannon1948}
Claude~Elwood Shannon.
\newblock A mathematical theory of communication.
\newblock {\em Bell System Technical Journal}, 27(3):379--423, 1948.

\bibitem[Sha83]{Shapiro83}
Ehud Shapiro.
\newblock {\em {Algorithmic Program Debugging}}.
\newblock MIT Press, 1983.

\bibitem[SHCH07]{Schlobach2007}
Stefan Schlobach, Zhisheng Huang, Ronald Cornet, and Frank Harmelen.
\newblock {Debugging Incoherent Terminologies}.
\newblock {\em Journal of Automated Reasoning}, 39(3):317--349, 2007.

\bibitem[SKFP12]{Stern2012}
Roni Stern, Meir Kalech, Alexander Feldman, and Gregory Provan.
\newblock {Exploring the Duality in Conflict-Directed Model-Based Diagnosis}.
\newblock In {\em Proceedings of the Twenty-Sixth AAAI Conference on Artificial
  Intelligence Exploring}, pages 828--834, 2012.

\bibitem[SL89]{Selman1989}
Bart Selman and Hector Levesque.
\newblock {Abductive and default reasoning: A computational core}.
\newblock {\em Proceedings of the 8th National Conference on Artificial
  Intelligence}, pages 343--348, 1989.

\bibitem[SMH08]{Shearer2008}
Rob Shearer, Boris Motik, and Ian Horrocks.
\newblock {HermiT : A Highly-Efficient OWL Reasoner}.
\newblock In {\em Proc. of the 5th Int. Workshop on OWL: Experiences and
  Directions (OWLED 2008 EU)}, 2008.

\bibitem[SPG{\etalchar{+}}07]{sirin2007pellet}
Evren Sirin, Bijan Parsia, Bernardo~Cuenca Grau, Aditya Kalyanpur, and Y~Katz.
\newblock {Pellet: A practical OWL-DL reasoner}.
\newblock {\em Web Semantics: Science, Services and Agents on the World Wide
  Web}, 5(2):51--53, 2007.

\bibitem[SQJH08]{Suntisrivaraporn2008}
Boontawee Suntisrivaraporn, Guilin Qi, Qiu Ji, and Peter Haase.
\newblock {A Modularization-Based Approach to Finding All Justifications for
  OWL DL Entailments}.
\newblock In {\em Proceedings of the 7th International Semantic Web Conference
  (ISWC 2008)}, pages 1--15. Springer, 2008.

\bibitem[SRF11]{Shchekotykhin2011}
Kostyantyn Shchekotykhin, Patrick Rodler, and Gerhard Friedrich.
\newblock {Balancing brave and cautious query strategies in ontology
  debugging}.
\newblock In {\em 22nd International Workshop on Principles of Diagnosis (DX
  2011)}, pages 122--129, 2011.

\bibitem[SS89]{schmidt1989}
Manfred Schmidt-Schau{\ss}.
\newblock Subsumption in {KL-ONE} is undecidable.
\newblock In {\em Proceedings of the 1st International Conference on Principles
  of Knowledge Representation and Reasoning}, pages 421--431. Morgan Kaufmann
  Publishers Inc., 1989.

\bibitem[SSZ09]{SattlerSZ09}
Ulrike Sattler, Thomas Schneider, and Michael Zakharyaschev.
\newblock {Which Kind of Module Should I Extract?}
\newblock In Bernardo~Cuenca Grau, Ian Horrocks, Boris Motik, and Ulrike
  Sattler, editors, {\em Proceedings of the 22nd International Workshop on
  Description Logics}, volume 477 of {\em CEUR Workshop Proceedings}.
  CEUR-WS.org, 2009.

\bibitem[Stu08]{Stuckenschmidt2008}
Heiner Stuckenschmidt.
\newblock {Debugging OWL Ontologies - A Reality Check}.
\newblock In Raul Garcia-Castro, Asunci\'{o}n G\'{o}mez-P\'{e}rez, Charles~J.
  Petrie, Emanuele~Della Valle, Ulrich K\"{u}ster, Michal Zaremba, and
  Shafiq~M. Omair, editors, {\em Proceedings of the 6th International Workshop
  on Evaluation of Ontology-based Tools and the Semantic Web Service Challenge
  (EON)}, pages 1--12, Tenerife, Spain, 2008.

\bibitem[SU06]{satoh2006}
Ken Satoh and Takeaki Uno.
\newblock {Enumerating Minimally Revised Specifications Using Dualization}.
\newblock In Takashi Washio, Akito Sakurai, Katsuto Nakajima, Hideaki Takeda,
  Satoshi Tojo, and Makoto Yokoo, editors, {\em New Frontiers in Artificial
  Intelligence}, volume 4012 of {\em Lecture Notes in Computer Science}, pages
  182--189. Springer Berlin Heidelberg, 2006.

\bibitem[SW05]{Steinbauer2005}
Gerald Steinbauer and Franz Wotawa.
\newblock {Detecting and locating faults in the control software of autonomous
  mobile robots}.
\newblock In {\em IJCAI International Joint Conference on Artificial
  Intelligence}, pages 1742--1743, 2005.

\bibitem[SW09]{Steinbauer2009}
Gerald Steinbauer and Franz Wotawa.
\newblock {Robust Plan Execution Using Model-Based Reasoning}.
\newblock {\em Advanced Robotics}, 23(10):1315--1326, 2009.

\bibitem[TH06]{Tsarkov06}
Dmitry Tsarkov and Ian Horrocks.
\newblock {FaCT++ description logic reasoner: System description}.
\newblock In {\em In Proc. of the Int. Joint Conf. on Automated Reasoning
  (IJCAR 2006)}, pages 292--297. Springer, 2006.

\bibitem[TNNM13]{Tudorache2013}
Tania Tudorache, Csongor Nyulas, Natalya~F. Noy, and Mark~A. Musen.
\newblock {WebProt\'{e}g\'{e}: A Collaborative Ontology Editor and Knowledge
  Acquisition Tool for the Web}.
\newblock {\em Semantic Web}, 4(1):89--99, 2013.

\bibitem[Tur37]{turing1937}
Alan~Mathison Turing.
\newblock {On Computable Numbers, with an Application to the
  Entscheidungsproblem}.
\newblock {\em Proceedings of the London Mathematical Society}, 2(1):230--265,
  1937.

\bibitem[WSM02]{wotawa2002}
Franz Wotawa, Markus Stumptner, and Wolfgang Mayer.
\newblock {Model-Based Debugging or How to Diagnose Programs Automatically}.
\newblock In Tim Hendtlass and Moonis Ali, editors, {\em Developments in
  Applied Artificial Intelligence}, volume 2358 of {\em Lecture Notes in
  Computer Science}, pages 746--757. Springer Berlin Heidelberg, 2002.

\end{thebibliography}
\end{document}